\documentclass[11pt]{article}

\usepackage[preprint]{acl}

\usepackage{times}
\usepackage{latexsym}

\usepackage[T1]{fontenc}

\usepackage[utf8]{inputenc}

\usepackage{microtype}

\usepackage{inconsolata}

\usepackage{graphicx}

\usepackage{color,xcolor}
\usepackage{epsfig}
\usepackage{graphicx}
\usepackage{tabularx}
\usepackage{times}
\usepackage{latexsym}

\usepackage{adjustbox}
\usepackage{array}
\usepackage{booktabs}
\usepackage{colortbl}
\usepackage{float,wrapfig}
\usepackage{hhline}
\usepackage{multirow}
\usepackage{subcaption} 
\usepackage[toc,page]{appendix}

\usepackage{amsmath,amsfonts,amsthm,amssymb}
\usepackage{bm}
\usepackage{nicefrac}
\usepackage{microtype}
\usepackage{inconsolata}
\usepackage{pifont}

\usepackage{changepage}
\usepackage{cuted}
\usepackage{extramarks}
\usepackage{fancyhdr}
\usepackage{lastpage}
\usepackage{setspace}
\usepackage{soul}
\usepackage{xspace}
\usepackage{paralist}
\usepackage{booktabs,arydshln}

\usepackage{hyperref}
\hypersetup{
  urlcolor=mydarkblue
}
\usepackage{url}

\usepackage{algorithm, algpseudocode}
\usepackage{enumerate}
\usepackage{lipsum}
\usepackage{physics}
\usepackage{cleveref}

\usepackage{listings}
\usepackage{xcolor}
\usepackage{enumitem}

\definecolor{commentgray}{rgb}{0.5, 0.5, 0.5}

\newcolumntype{L}[1]{>{\raggedright\let\newline\\\arraybackslash\hspace{0pt}}m{#1}}
\newcolumntype{C}[1]{>{\centering\let\newline\\\arraybackslash\hspace{0pt}}m{#1}}
\newcolumntype{R}[1]{>{\raggedleft\let\newline\\\arraybackslash\hspace{0pt}}m{#1}}

\newcommand{\eqn}[1]{Equation~\ref{#1}}

\newcommand{\fig}[1]{Figure~\ref{#1}}

\newcommand{\tab}[1]{Table~\ref{#1}}

\newcommand{\ignorethis}[1]{}

\DeclareMathOperator*{\argmin}{arg\,min}

\makeatletter
\DeclareRobustCommand\onedot{\futurelet\@let@token\@onedot}
\def\@onedot{\ifx\@let@token.\else.\null\fi\xspace}

\makeatother

\makeatletter
\def\adl@drawiv#1#2#3{%
        \hskip.5\tabcolsep
        \xleaders#3{#2.5\@tempdimb #1{1}#2.5\@tempdimb}%
                #2\z@ plus1fil minus1fil\relax
        \hskip.5\tabcolsep}
\newcommand{\cdashlinelr}[1]{%
  \noalign{\vskip\aboverulesep
           \global\let\@dashdrawstore\adl@draw
           \global\let\adl@draw\adl@drawiv}
  \cdashline{#1}
  \noalign{\global\let\adl@draw\@dashdrawstore
           \vskip\belowrulesep}}
\makeatother

\definecolor{tablepink}{RGB}{250, 235, 242}
\definecolor{tableblue}{RGB}{210, 225, 245}
\colorlet{tablelightblue}{tableblue!50}
\definecolor{tablegray}{RGB}{128, 128, 128}
\definecolor{mydarkblue}{rgb}{0,0.08,1}
\definecolor{mydarkgreen}{rgb}{0.02,0.6,0.02}
\definecolor{mydarkred}{rgb}{0.8,0.02,0.02}
\definecolor{mydarkorange}{rgb}{0.40,0.2,0.02}
\definecolor{mypurple}{RGB}{111,0,255}
\definecolor{myred}{rgb}{1.0,0.0,0.0}
\definecolor{mygold}{rgb}{0.75,0.6,0.12}
\definecolor{mydarkgray}{rgb}{0.66, 0.66, 0.66}

\definecolor{darkgreen}{rgb}{0.02,0.6,0.02}
\definecolor{darkred}{rgb}{0.8,0.02,0.02}
\definecolor{darkorange}{rgb}{0.40,0.2,0.02}
\definecolor{darkpurple}{RGB}{111,0,255}

\def\R{\mathbb{R}}

\theoremstyle{plain}
\newtheorem{theorem}{Theorem}[section]
\newtheorem{proposition}[theorem]{Proposition}

\newtheorem{corollary}[theorem]{Corollary}
\theoremstyle{definition}

\theoremstyle{remark}

\providecommand{\one}{\mathbf{1}}

\providecommand{\W}{\mathcal{W}}

\providecommand{\concur}{CONCUR\xspace}

\providecommand{\setval}[2]{\expandafter\gdef\csname pv@#1\endcsname{#2}}
\providecommand{\val}[1]{\ifcsname pv@#1\endcsname\csname pv@#1\endcsname\else--\fi}
\providecommand{\valor}[2]{\ifcsname pv@#1\endcsname\csname pv@#1\endcsname\else#2\fi}
\DeclareRobustCommand{\gentodo}[2]{\IfFileExists{tables/gen_#1.tex}{}{\textcolor{red}{[TODO: #2]}}}

\xspaceaddexceptions{\%}

\providecommand{\qwenfour}{\texttt{Qwen3-4B-\allowbreak Base}\xspace}
\providecommand{\qweneight}{\texttt{Qwen3-8B-\allowbreak Base}\xspace}
\providecommand{\olmo}{\texttt{OLMo-3-7B-\allowbreak Instruct-\allowbreak SFT}\xspace}

\setval{4b.base.bench.aime24}{7.8}
\setval{4b.base.bench.aime25}{7.1}
\setval{4b.base.bench.amc23}{43.9}
\setval{4b.base.bench.creative-writing-v3}{24.8}
\setval{4b.base.bench.guidebench}{24.9}
\setval{4b.base.bench.humaneval-plus}{65.1}
\setval{4b.base.bench.ifbench}{17.0}
\setval{4b.base.bench.ifeval}{47.4}
\setval{4b.base.bench.livecodebench}{13.2}
\setval{4b.base.bench.longbench-write}{44.4}
\setval{4b.base.bench.math500}{67.3}
\setval{4b.base.bench.mbpp-plus}{55.6}
\setval{4b.base.bench.minerva}{31.6}
\setval{4b.base.bench.olympiad-bench}{35.6}
\setval{4b.base.bench.writingbench}{32.4}
\setval{4b.base.eval.avg}{34.5}
\setval{4b.base.eval.code}{44.6}
\setval{4b.base.eval.if}{29.7}
\setval{4b.base.eval.math}{32.2}
\setval{4b.base.eval.writing}{33.9}
\setval{4b.base.judge2.avg}{33.9}
\setval{4b.base.judge2.creative-writing-v3}{20.9}
\setval{4b.base.judge2.longbench-write}{40.9}
\setval{4b.base.judge2.writing}{30.8}
\setval{4b.base.judge2.writingbench}{30.5}
\setval{4b.base.minus.concur-free.aime24}{$-$4.1}
\setval{4b.base.minus.concur-free.aime24-abs}{4.1}
\setval{4b.base.minus.concur-free.aime25}{$-$7.5}
\setval{4b.base.minus.concur-free.aime25-abs}{7.5}
\setval{4b.base.minus.concur-free.amc23}{$-$10.2}
\setval{4b.base.minus.concur-free.amc23-abs}{10.2}
\setval{4b.base.minus.concur-free.avg}{$-$21.0}
\setval{4b.base.minus.concur-free.avg-abs}{21.0}
\setval{4b.base.minus.concur-free.code}{$-$13.7}
\setval{4b.base.minus.concur-free.code-abs}{13.7}
\setval{4b.base.minus.concur-free.creative-writing-v3}{$-$20.7}
\setval{4b.base.minus.concur-free.creative-writing-v3-abs}{20.7}
\setval{4b.base.minus.concur-free.guidebench}{$-$54.1}
\setval{4b.base.minus.concur-free.guidebench-abs}{54.1}
\setval{4b.base.minus.concur-free.humaneval-plus}{$-$16.0}
\setval{4b.base.minus.concur-free.humaneval-plus-abs}{16.0}
\setval{4b.base.minus.concur-free.if}{$-$38.5}
\setval{4b.base.minus.concur-free.if-abs}{38.5}
\setval{4b.base.minus.concur-free.ifbench}{$-$22.0}
\setval{4b.base.minus.concur-free.ifbench-abs}{22.0}
\setval{4b.base.minus.concur-free.ifeval}{$-$39.3}
\setval{4b.base.minus.concur-free.ifeval-abs}{39.3}
\setval{4b.base.minus.concur-free.judge2.avg}{$-$18.9}
\setval{4b.base.minus.concur-free.judge2.avg-abs}{18.9}
\setval{4b.base.minus.concur-free.judge2.creative-writing-v3}{$-$14.4}
\setval{4b.base.minus.concur-free.judge2.creative-writing-v3-abs}{14.4}
\setval{4b.base.minus.concur-free.judge2.longbench-write}{$-$28.4}
\setval{4b.base.minus.concur-free.judge2.longbench-write-abs}{28.4}
\setval{4b.base.minus.concur-free.judge2.writing}{$-$22.3}
\setval{4b.base.minus.concur-free.judge2.writing-abs}{22.3}
\setval{4b.base.minus.concur-free.judge2.writingbench}{$-$24.0}
\setval{4b.base.minus.concur-free.judge2.writingbench-abs}{24.0}
\setval{4b.base.minus.concur-free.livecodebench}{$-$11.0}
\setval{4b.base.minus.concur-free.livecodebench-abs}{11.0}
\setval{4b.base.minus.concur-free.longbench-write}{$-$33.3}
\setval{4b.base.minus.concur-free.longbench-write-abs}{33.3}
\setval{4b.base.minus.concur-free.math}{$-$9.9}
\setval{4b.base.minus.concur-free.math-abs}{9.9}
\setval{4b.base.minus.concur-free.math500}{$-$13.1}
\setval{4b.base.minus.concur-free.math500-abs}{13.1}
\setval{4b.base.minus.concur-free.mbpp-plus}{$-$14.1}
\setval{4b.base.minus.concur-free.mbpp-plus-abs}{14.1}
\setval{4b.base.minus.concur-free.minerva}{$-$14.6}
\setval{4b.base.minus.concur-free.minerva-abs}{14.6}
\setval{4b.base.minus.concur-free.olympiad-bench}{$-$10.1}
\setval{4b.base.minus.concur-free.olympiad-bench-abs}{10.1}
\setval{4b.base.minus.concur-free.writing}{$-$33.0}
\setval{4b.base.minus.concur-free.writing-abs}{33.0}
\setval{4b.base.minus.concur-free.writingbench}{$-$44.9}
\setval{4b.base.minus.concur-free.writingbench-abs}{44.9}
\setval{4b.base.minus.concur-tied.aime24}{$-$11.7}
\setval{4b.base.minus.concur-tied.aime24-abs}{11.7}
\setval{4b.base.minus.concur-tied.aime25}{$-$9.3}
\setval{4b.base.minus.concur-tied.aime25-abs}{9.3}
\setval{4b.base.minus.concur-tied.amc23}{$-$25.3}
\setval{4b.base.minus.concur-tied.amc23-abs}{25.3}
\setval{4b.base.minus.concur-tied.avg}{$-$24.2}
\setval{4b.base.minus.concur-tied.avg-abs}{24.2}
\setval{4b.base.minus.concur-tied.code}{$-$14.5}
\setval{4b.base.minus.concur-tied.code-abs}{14.5}
\setval{4b.base.minus.concur-tied.creative-writing-v3}{$-$23.1}
\setval{4b.base.minus.concur-tied.creative-writing-v3-abs}{23.1}
\setval{4b.base.minus.concur-tied.guidebench}{$-$54.3}
\setval{4b.base.minus.concur-tied.guidebench-abs}{54.3}
\setval{4b.base.minus.concur-tied.humaneval-plus}{$-$19.0}
\setval{4b.base.minus.concur-tied.humaneval-plus-abs}{19.0}
\setval{4b.base.minus.concur-tied.if}{$-$38.5}
\setval{4b.base.minus.concur-tied.if-abs}{38.5}
\setval{4b.base.minus.concur-tied.ifbench}{$-$20.7}
\setval{4b.base.minus.concur-tied.ifbench-abs}{20.7}
\setval{4b.base.minus.concur-tied.ifeval}{$-$40.4}
\setval{4b.base.minus.concur-tied.ifeval-abs}{40.4}
\setval{4b.base.minus.concur-tied.judge2.avg}{$-$21.5}
\setval{4b.base.minus.concur-tied.judge2.avg-abs}{21.5}
\setval{4b.base.minus.concur-tied.judge2.creative-writing-v3}{$-$13.1}
\setval{4b.base.minus.concur-tied.judge2.creative-writing-v3-abs}{13.1}
\setval{4b.base.minus.concur-tied.judge2.longbench-write}{$-$27.0}
\setval{4b.base.minus.concur-tied.judge2.longbench-write-abs}{27.0}
\setval{4b.base.minus.concur-tied.judge2.writing}{$-$21.7}
\setval{4b.base.minus.concur-tied.judge2.writing-abs}{21.7}
\setval{4b.base.minus.concur-tied.judge2.writingbench}{$-$24.9}
\setval{4b.base.minus.concur-tied.judge2.writingbench-abs}{24.9}
\setval{4b.base.minus.concur-tied.livecodebench}{$-$11.5}
\setval{4b.base.minus.concur-tied.livecodebench-abs}{11.5}
\setval{4b.base.minus.concur-tied.longbench-write}{$-$33.3}
\setval{4b.base.minus.concur-tied.longbench-write-abs}{33.3}
\setval{4b.base.minus.concur-tied.math}{$-$16.5}
\setval{4b.base.minus.concur-tied.math-abs}{16.5}
\setval{4b.base.minus.concur-tied.math500}{$-$18.3}
\setval{4b.base.minus.concur-tied.math500-abs}{18.3}
\setval{4b.base.minus.concur-tied.mbpp-plus}{$-$13.0}
\setval{4b.base.minus.concur-tied.mbpp-plus-abs}{13.0}
\setval{4b.base.minus.concur-tied.minerva}{$-$17.5}
\setval{4b.base.minus.concur-tied.minerva-abs}{17.5}
\setval{4b.base.minus.concur-tied.olympiad-bench}{$-$17.2}
\setval{4b.base.minus.concur-tied.olympiad-bench-abs}{17.2}
\setval{4b.base.minus.concur-tied.writing}{$-$34.7}
\setval{4b.base.minus.concur-tied.writing-abs}{34.7}
\setval{4b.base.minus.concur-tied.writingbench}{$-$47.8}
\setval{4b.base.minus.concur-tied.writingbench-abs}{47.8}
\setval{4b.base.minus.concur.aime24}{$-$12.7}
\setval{4b.base.minus.concur.aime24-abs}{12.7}
\setval{4b.base.minus.concur.aime25}{$-$11.7}
\setval{4b.base.minus.concur.aime25-abs}{11.7}
\setval{4b.base.minus.concur.amc23}{$-$25.5}
\setval{4b.base.minus.concur.amc23-abs}{25.5}
\setval{4b.base.minus.concur.avg}{$-$24.3}
\setval{4b.base.minus.concur.avg-abs}{24.3}
\setval{4b.base.minus.concur.code}{$-$15.1}
\setval{4b.base.minus.concur.code-abs}{15.1}
\setval{4b.base.minus.concur.creative-writing-v3}{$-$19.9}
\setval{4b.base.minus.concur.creative-writing-v3-abs}{19.9}
\setval{4b.base.minus.concur.guidebench}{$-$52.5}
\setval{4b.base.minus.concur.guidebench-abs}{52.5}
\setval{4b.base.minus.concur.humaneval-plus}{$-$17.4}
\setval{4b.base.minus.concur.humaneval-plus-abs}{17.4}
\setval{4b.base.minus.concur.if}{$-$38.9}
\setval{4b.base.minus.concur.if-abs}{38.9}
\setval{4b.base.minus.concur.ifbench}{$-$23.3}
\setval{4b.base.minus.concur.ifbench-abs}{23.3}
\setval{4b.base.minus.concur.ifeval}{$-$40.8}
\setval{4b.base.minus.concur.ifeval-abs}{40.8}
\setval{4b.base.minus.concur.judge2.avg}{$-$21.9}
\setval{4b.base.minus.concur.judge2.avg-abs}{21.9}
\setval{4b.base.minus.concur.judge2.creative-writing-v3}{$-$11.5}
\setval{4b.base.minus.concur.judge2.creative-writing-v3-abs}{11.5}
\setval{4b.base.minus.concur.judge2.longbench-write}{$-$25.7}
\setval{4b.base.minus.concur.judge2.longbench-write-abs}{25.7}
\setval{4b.base.minus.concur.judge2.writing}{$-$20.9}
\setval{4b.base.minus.concur.judge2.writing-abs}{20.9}
\setval{4b.base.minus.concur.judge2.writingbench}{$-$25.6}
\setval{4b.base.minus.concur.judge2.writingbench-abs}{25.6}
\setval{4b.base.minus.concur.livecodebench}{$-$11.9}
\setval{4b.base.minus.concur.livecodebench-abs}{11.9}
\setval{4b.base.minus.concur.longbench-write}{$-$31.9}
\setval{4b.base.minus.concur.longbench-write-abs}{31.9}
\setval{4b.base.minus.concur.math}{$-$17.2}
\setval{4b.base.minus.concur.math-abs}{17.2}
\setval{4b.base.minus.concur.math500}{$-$18.0}
\setval{4b.base.minus.concur.math500-abs}{18.0}
\setval{4b.base.minus.concur.mbpp-plus}{$-$15.9}
\setval{4b.base.minus.concur.mbpp-plus-abs}{15.9}
\setval{4b.base.minus.concur.minerva}{$-$17.6}
\setval{4b.base.minus.concur.minerva-abs}{17.6}
\setval{4b.base.minus.concur.olympiad-bench}{$-$18.0}
\setval{4b.base.minus.concur.olympiad-bench-abs}{18.0}
\setval{4b.base.minus.concur.writing}{$-$32.9}
\setval{4b.base.minus.concur.writing-abs}{32.9}
\setval{4b.base.minus.concur.writingbench}{$-$47.0}
\setval{4b.base.minus.concur.writingbench-abs}{47.0}
\setval{4b.base.minus.gd2po.aime24}{$-$4.1}
\setval{4b.base.minus.gd2po.aime24-abs}{4.1}
\setval{4b.base.minus.gd2po.aime25}{+0.3}
\setval{4b.base.minus.gd2po.aime25-abs}{0.3}
\setval{4b.base.minus.gd2po.amc23}{$-$13.8}
\setval{4b.base.minus.gd2po.amc23-abs}{13.8}
\setval{4b.base.minus.gd2po.avg}{$-$19.6}
\setval{4b.base.minus.gd2po.avg-abs}{19.6}
\setval{4b.base.minus.gd2po.code}{$-$14.8}
\setval{4b.base.minus.gd2po.code-abs}{14.8}
\setval{4b.base.minus.gd2po.creative-writing-v3}{$-$19.9}
\setval{4b.base.minus.gd2po.creative-writing-v3-abs}{19.9}
\setval{4b.base.minus.gd2po.guidebench}{$-$54.2}
\setval{4b.base.minus.gd2po.guidebench-abs}{54.2}
\setval{4b.base.minus.gd2po.humaneval-plus}{$-$18.4}
\setval{4b.base.minus.gd2po.humaneval-plus-abs}{18.4}
\setval{4b.base.minus.gd2po.if}{$-$38.4}
\setval{4b.base.minus.gd2po.if-abs}{38.4}
\setval{4b.base.minus.gd2po.ifbench}{$-$20.7}
\setval{4b.base.minus.gd2po.ifbench-abs}{20.7}
\setval{4b.base.minus.gd2po.ifeval}{$-$40.3}
\setval{4b.base.minus.gd2po.ifeval-abs}{40.3}
\setval{4b.base.minus.gd2po.judge2.avg}{$-$17.7}
\setval{4b.base.minus.gd2po.judge2.avg-abs}{17.7}
\setval{4b.base.minus.gd2po.judge2.creative-writing-v3}{$-$12.0}
\setval{4b.base.minus.gd2po.judge2.creative-writing-v3-abs}{12.0}
\setval{4b.base.minus.gd2po.judge2.longbench-write}{$-$20.1}
\setval{4b.base.minus.gd2po.judge2.longbench-write-abs}{20.1}
\setval{4b.base.minus.gd2po.judge2.writing}{$-$18.0}
\setval{4b.base.minus.gd2po.judge2.writing-abs}{18.0}
\setval{4b.base.minus.gd2po.judge2.writingbench}{$-$22.1}
\setval{4b.base.minus.gd2po.judge2.writingbench-abs}{22.1}
\setval{4b.base.minus.gd2po.livecodebench}{$-$10.3}
\setval{4b.base.minus.gd2po.livecodebench-abs}{10.3}
\setval{4b.base.minus.gd2po.longbench-write}{$-$24.5}
\setval{4b.base.minus.gd2po.longbench-write-abs}{24.5}
\setval{4b.base.minus.gd2po.math}{$-$8.6}
\setval{4b.base.minus.gd2po.math-abs}{8.6}
\setval{4b.base.minus.gd2po.math500}{$-$10.4}
\setval{4b.base.minus.gd2po.math500-abs}{10.4}
\setval{4b.base.minus.gd2po.mbpp-plus}{$-$15.9}
\setval{4b.base.minus.gd2po.mbpp-plus-abs}{15.9}
\setval{4b.base.minus.gd2po.minerva}{$-$15.6}
\setval{4b.base.minus.gd2po.minerva-abs}{15.6}
\setval{4b.base.minus.gd2po.olympiad-bench}{$-$7.9}
\setval{4b.base.minus.gd2po.olympiad-bench-abs}{7.9}
\setval{4b.base.minus.gd2po.writing}{$-$27.4}
\setval{4b.base.minus.gd2po.writing-abs}{27.4}
\setval{4b.base.minus.gd2po.writingbench}{$-$37.7}
\setval{4b.base.minus.gd2po.writingbench-abs}{37.7}
\setval{4b.base.minus.gdpo-noaux.aime24}{$-$15.6}
\setval{4b.base.minus.gdpo-noaux.aime24-abs}{15.6}
\setval{4b.base.minus.gdpo-noaux.aime25}{$-$11.7}
\setval{4b.base.minus.gdpo-noaux.aime25-abs}{11.7}
\setval{4b.base.minus.gdpo-noaux.amc23}{$-$21.8}
\setval{4b.base.minus.gdpo-noaux.amc23-abs}{21.8}
\setval{4b.base.minus.gdpo-noaux.avg}{$-$22.1}
\setval{4b.base.minus.gdpo-noaux.avg-abs}{22.1}
\setval{4b.base.minus.gdpo-noaux.code}{$-$16.2}
\setval{4b.base.minus.gdpo-noaux.code-abs}{16.2}
\setval{4b.base.minus.gdpo-noaux.creative-writing-v3}{$-$14.6}
\setval{4b.base.minus.gdpo-noaux.creative-writing-v3-abs}{14.6}
\setval{4b.base.minus.gdpo-noaux.guidebench}{$-$51.1}
\setval{4b.base.minus.gdpo-noaux.guidebench-abs}{51.1}
\setval{4b.base.minus.gdpo-noaux.humaneval-plus}{$-$20.4}
\setval{4b.base.minus.gdpo-noaux.humaneval-plus-abs}{20.4}
\setval{4b.base.minus.gdpo-noaux.if}{$-$37.3}
\setval{4b.base.minus.gdpo-noaux.if-abs}{37.3}
\setval{4b.base.minus.gdpo-noaux.ifbench}{$-$23.3}
\setval{4b.base.minus.gdpo-noaux.ifbench-abs}{23.3}
\setval{4b.base.minus.gdpo-noaux.ifeval}{$-$37.4}
\setval{4b.base.minus.gdpo-noaux.ifeval-abs}{37.4}
\setval{4b.base.minus.gdpo-noaux.judge2.avg}{$-$19.4}
\setval{4b.base.minus.gdpo-noaux.judge2.avg-abs}{19.4}
\setval{4b.base.minus.gdpo-noaux.judge2.creative-writing-v3}{$-$2.5}
\setval{4b.base.minus.gdpo-noaux.judge2.creative-writing-v3-abs}{2.5}
\setval{4b.base.minus.gdpo-noaux.judge2.longbench-write}{$-$9.9}
\setval{4b.base.minus.gdpo-noaux.judge2.longbench-write-abs}{9.9}
\setval{4b.base.minus.gdpo-noaux.judge2.writing}{$-$9.3}
\setval{4b.base.minus.gdpo-noaux.judge2.writing-abs}{9.3}
\setval{4b.base.minus.gdpo-noaux.judge2.writingbench}{$-$15.6}
\setval{4b.base.minus.gdpo-noaux.judge2.writingbench-abs}{15.6}
\setval{4b.base.minus.gdpo-noaux.livecodebench}{$-$11.5}
\setval{4b.base.minus.gdpo-noaux.livecodebench-abs}{11.5}
\setval{4b.base.minus.gdpo-noaux.longbench-write}{$-$19.9}
\setval{4b.base.minus.gdpo-noaux.longbench-write-abs}{19.9}
\setval{4b.base.minus.gdpo-noaux.math}{$-$17.0}
\setval{4b.base.minus.gdpo-noaux.math-abs}{17.0}
\setval{4b.base.minus.gdpo-noaux.math500}{$-$17.5}
\setval{4b.base.minus.gdpo-noaux.math500-abs}{17.5}
\setval{4b.base.minus.gdpo-noaux.mbpp-plus}{$-$16.7}
\setval{4b.base.minus.gdpo-noaux.mbpp-plus-abs}{16.7}
\setval{4b.base.minus.gdpo-noaux.minerva}{$-$17.9}
\setval{4b.base.minus.gdpo-noaux.minerva-abs}{17.9}
\setval{4b.base.minus.gdpo-noaux.olympiad-bench}{$-$17.5}
\setval{4b.base.minus.gdpo-noaux.olympiad-bench-abs}{17.5}
\setval{4b.base.minus.gdpo-noaux.writing}{$-$23.1}
\setval{4b.base.minus.gdpo-noaux.writing-abs}{23.1}
\setval{4b.base.minus.gdpo-noaux.writingbench}{$-$34.9}
\setval{4b.base.minus.gdpo-noaux.writingbench-abs}{34.9}
\setval{4b.base.minus.gdpo-w0.5.aime24}{$-$6.8}
\setval{4b.base.minus.gdpo-w0.5.aime24-abs}{6.8}
\setval{4b.base.minus.gdpo-w0.5.aime25}{$-$3.2}
\setval{4b.base.minus.gdpo-w0.5.aime25-abs}{3.2}
\setval{4b.base.minus.gdpo-w0.5.amc23}{$-$14.5}
\setval{4b.base.minus.gdpo-w0.5.amc23-abs}{14.5}
\setval{4b.base.minus.gdpo-w0.5.avg}{$-$20.1}
\setval{4b.base.minus.gdpo-w0.5.avg-abs}{20.1}
\setval{4b.base.minus.gdpo-w0.5.code}{$-$14.5}
\setval{4b.base.minus.gdpo-w0.5.code-abs}{14.5}
\setval{4b.base.minus.gdpo-w0.5.creative-writing-v3}{$-$13.3}
\setval{4b.base.minus.gdpo-w0.5.creative-writing-v3-abs}{13.3}
\setval{4b.base.minus.gdpo-w0.5.guidebench}{$-$54.7}
\setval{4b.base.minus.gdpo-w0.5.guidebench-abs}{54.7}
\setval{4b.base.minus.gdpo-w0.5.humaneval-plus}{$-$17.0}
\setval{4b.base.minus.gdpo-w0.5.humaneval-plus-abs}{17.0}
\setval{4b.base.minus.gdpo-w0.5.if}{$-$38.8}
\setval{4b.base.minus.gdpo-w0.5.if-abs}{38.8}
\setval{4b.base.minus.gdpo-w0.5.ifbench}{$-$22.3}
\setval{4b.base.minus.gdpo-w0.5.ifbench-abs}{22.3}
\setval{4b.base.minus.gdpo-w0.5.ifeval}{$-$39.4}
\setval{4b.base.minus.gdpo-w0.5.ifeval-abs}{39.4}
\setval{4b.base.minus.gdpo-w0.5.judge2.avg}{$-$18.6}
\setval{4b.base.minus.gdpo-w0.5.judge2.avg-abs}{18.6}
\setval{4b.base.minus.gdpo-w0.5.judge2.creative-writing-v3}{$-$9.2}
\setval{4b.base.minus.gdpo-w0.5.judge2.creative-writing-v3-abs}{9.2}
\setval{4b.base.minus.gdpo-w0.5.judge2.longbench-write}{$-$19.6}
\setval{4b.base.minus.gdpo-w0.5.judge2.longbench-write-abs}{19.6}
\setval{4b.base.minus.gdpo-w0.5.judge2.writing}{$-$17.3}
\setval{4b.base.minus.gdpo-w0.5.judge2.writing-abs}{17.3}
\setval{4b.base.minus.gdpo-w0.5.judge2.writingbench}{$-$23.1}
\setval{4b.base.minus.gdpo-w0.5.judge2.writingbench-abs}{23.1}
\setval{4b.base.minus.gdpo-w0.5.livecodebench}{$-$11.0}
\setval{4b.base.minus.gdpo-w0.5.livecodebench-abs}{11.0}
\setval{4b.base.minus.gdpo-w0.5.longbench-write}{$-$23.0}
\setval{4b.base.minus.gdpo-w0.5.longbench-write-abs}{23.0}
\setval{4b.base.minus.gdpo-w0.5.math}{$-$11.1}
\setval{4b.base.minus.gdpo-w0.5.math-abs}{11.1}
\setval{4b.base.minus.gdpo-w0.5.math500}{$-$14.5}
\setval{4b.base.minus.gdpo-w0.5.math500-abs}{14.5}
\setval{4b.base.minus.gdpo-w0.5.mbpp-plus}{$-$15.6}
\setval{4b.base.minus.gdpo-w0.5.mbpp-plus-abs}{15.6}
\setval{4b.base.minus.gdpo-w0.5.minerva}{$-$14.5}
\setval{4b.base.minus.gdpo-w0.5.minerva-abs}{14.5}
\setval{4b.base.minus.gdpo-w0.5.olympiad-bench}{$-$12.8}
\setval{4b.base.minus.gdpo-w0.5.olympiad-bench-abs}{12.8}
\setval{4b.base.minus.gdpo-w0.5.writing}{$-$24.9}
\setval{4b.base.minus.gdpo-w0.5.writing-abs}{24.9}
\setval{4b.base.minus.gdpo-w0.5.writingbench}{$-$38.3}
\setval{4b.base.minus.gdpo-w0.5.writingbench-abs}{38.3}
\setval{4b.base.minus.gdpo.aime24}{$-$1.2}
\setval{4b.base.minus.gdpo.aime24-abs}{1.2}
\setval{4b.base.minus.gdpo.aime25}{$-$2.8}
\setval{4b.base.minus.gdpo.aime25-abs}{2.8}
\setval{4b.base.minus.gdpo.amc23}{$-$15.7}
\setval{4b.base.minus.gdpo.amc23-abs}{15.7}
\setval{4b.base.minus.gdpo.avg}{$-$19.8}
\setval{4b.base.minus.gdpo.avg-abs}{19.8}
\setval{4b.base.minus.gdpo.code}{$-$14.4}
\setval{4b.base.minus.gdpo.code-abs}{14.4}
\setval{4b.base.minus.gdpo.creative-writing-v3}{$-$21.2}
\setval{4b.base.minus.gdpo.creative-writing-v3-abs}{21.2}
\setval{4b.base.minus.gdpo.guidebench}{$-$55.8}
\setval{4b.base.minus.gdpo.guidebench-abs}{55.8}
\setval{4b.base.minus.gdpo.humaneval-plus}{$-$16.8}
\setval{4b.base.minus.gdpo.humaneval-plus-abs}{16.8}
\setval{4b.base.minus.gdpo.if}{$-$37.8}
\setval{4b.base.minus.gdpo.if-abs}{37.8}
\setval{4b.base.minus.gdpo.ifbench}{$-$19.3}
\setval{4b.base.minus.gdpo.ifbench-abs}{19.3}
\setval{4b.base.minus.gdpo.ifeval}{$-$38.4}
\setval{4b.base.minus.gdpo.ifeval-abs}{38.4}
\setval{4b.base.minus.gdpo.judge2.avg}{$-$18.0}
\setval{4b.base.minus.gdpo.judge2.avg-abs}{18.0}
\setval{4b.base.minus.gdpo.judge2.creative-writing-v3}{$-$13.5}
\setval{4b.base.minus.gdpo.judge2.creative-writing-v3-abs}{13.5}
\setval{4b.base.minus.gdpo.judge2.longbench-write}{$-$21.4}
\setval{4b.base.minus.gdpo.judge2.longbench-write-abs}{21.4}
\setval{4b.base.minus.gdpo.judge2.writing}{$-$19.5}
\setval{4b.base.minus.gdpo.judge2.writing-abs}{19.5}
\setval{4b.base.minus.gdpo.judge2.writingbench}{$-$23.5}
\setval{4b.base.minus.gdpo.judge2.writingbench-abs}{23.5}
\setval{4b.base.minus.gdpo.livecodebench}{$-$11.2}
\setval{4b.base.minus.gdpo.livecodebench-abs}{11.2}
\setval{4b.base.minus.gdpo.longbench-write}{$-$24.6}
\setval{4b.base.minus.gdpo.longbench-write-abs}{24.6}
\setval{4b.base.minus.gdpo.math}{$-$9.1}
\setval{4b.base.minus.gdpo.math-abs}{9.1}
\setval{4b.base.minus.gdpo.math500}{$-$12.0}
\setval{4b.base.minus.gdpo.math500-abs}{12.0}
\setval{4b.base.minus.gdpo.mbpp-plus}{$-$15.3}
\setval{4b.base.minus.gdpo.mbpp-plus-abs}{15.3}
\setval{4b.base.minus.gdpo.minerva}{$-$13.9}
\setval{4b.base.minus.gdpo.minerva-abs}{13.9}
\setval{4b.base.minus.gdpo.olympiad-bench}{$-$9.0}
\setval{4b.base.minus.gdpo.olympiad-bench-abs}{9.0}
\setval{4b.base.minus.gdpo.writing}{$-$28.4}
\setval{4b.base.minus.gdpo.writing-abs}{28.4}
\setval{4b.base.minus.gdpo.writingbench}{$-$39.5}
\setval{4b.base.minus.gdpo.writingbench-abs}{39.5}
\setval{4b.base.minus.geoma.aime24}{$-$3.8}
\setval{4b.base.minus.geoma.aime24-abs}{3.8}
\setval{4b.base.minus.geoma.aime25}{$-$3.5}
\setval{4b.base.minus.geoma.aime25-abs}{3.5}
\setval{4b.base.minus.geoma.amc23}{$-$11.5}
\setval{4b.base.minus.geoma.amc23-abs}{11.5}
\setval{4b.base.minus.geoma.avg}{$-$19.3}
\setval{4b.base.minus.geoma.avg-abs}{19.3}
\setval{4b.base.minus.geoma.code}{$-$14.1}
\setval{4b.base.minus.geoma.code-abs}{14.1}
\setval{4b.base.minus.geoma.creative-writing-v3}{$-$21.1}
\setval{4b.base.minus.geoma.creative-writing-v3-abs}{21.1}
\setval{4b.base.minus.geoma.guidebench}{$-$53.6}
\setval{4b.base.minus.geoma.guidebench-abs}{53.6}
\setval{4b.base.minus.geoma.humaneval-plus}{$-$15.2}
\setval{4b.base.minus.geoma.humaneval-plus-abs}{15.2}
\setval{4b.base.minus.geoma.if}{$-$37.7}
\setval{4b.base.minus.geoma.if-abs}{37.7}
\setval{4b.base.minus.geoma.ifbench}{$-$19.0}
\setval{4b.base.minus.geoma.ifbench-abs}{19.0}
\setval{4b.base.minus.geoma.ifeval}{$-$40.4}
\setval{4b.base.minus.geoma.ifeval-abs}{40.4}
\setval{4b.base.minus.geoma.judge2.avg}{$-$17.2}
\setval{4b.base.minus.geoma.judge2.avg-abs}{17.2}
\setval{4b.base.minus.geoma.judge2.creative-writing-v3}{$-$12.3}
\setval{4b.base.minus.geoma.judge2.creative-writing-v3-abs}{12.3}
\setval{4b.base.minus.geoma.judge2.longbench-write}{$-$16.7}
\setval{4b.base.minus.geoma.judge2.longbench-write-abs}{16.7}
\setval{4b.base.minus.geoma.judge2.writing}{$-$16.4}
\setval{4b.base.minus.geoma.judge2.writing-abs}{16.4}
\setval{4b.base.minus.geoma.judge2.writingbench}{$-$20.2}
\setval{4b.base.minus.geoma.judge2.writingbench-abs}{20.2}
\setval{4b.base.minus.geoma.livecodebench}{$-$12.5}
\setval{4b.base.minus.geoma.livecodebench-abs}{12.5}
\setval{4b.base.minus.geoma.longbench-write}{$-$21.4}
\setval{4b.base.minus.geoma.longbench-write-abs}{21.4}
\setval{4b.base.minus.geoma.math}{$-$9.0}
\setval{4b.base.minus.geoma.math-abs}{9.0}
\setval{4b.base.minus.geoma.math500}{$-$11.6}
\setval{4b.base.minus.geoma.math500-abs}{11.6}
\setval{4b.base.minus.geoma.mbpp-plus}{$-$14.6}
\setval{4b.base.minus.geoma.mbpp-plus-abs}{14.6}
\setval{4b.base.minus.geoma.minerva}{$-$15.3}
\setval{4b.base.minus.geoma.minerva-abs}{15.3}
\setval{4b.base.minus.geoma.olympiad-bench}{$-$8.2}
\setval{4b.base.minus.geoma.olympiad-bench-abs}{8.2}
\setval{4b.base.minus.geoma.writing}{$-$26.5}
\setval{4b.base.minus.geoma.writing-abs}{26.5}
\setval{4b.base.minus.geoma.writingbench}{$-$37.1}
\setval{4b.base.minus.geoma.writingbench-abs}{37.1}
\setval{4b.base.minus.grpo-nodiv.aime24}{$-$4.6}
\setval{4b.base.minus.grpo-nodiv.aime24-abs}{4.6}
\setval{4b.base.minus.grpo-nodiv.aime25}{$-$1.2}
\setval{4b.base.minus.grpo-nodiv.aime25-abs}{1.2}
\setval{4b.base.minus.grpo-nodiv.amc23}{$-$10.6}
\setval{4b.base.minus.grpo-nodiv.amc23-abs}{10.6}
\setval{4b.base.minus.grpo-nodiv.avg}{$-$18.5}
\setval{4b.base.minus.grpo-nodiv.avg-abs}{18.5}
\setval{4b.base.minus.grpo-nodiv.code}{$-$14.1}
\setval{4b.base.minus.grpo-nodiv.code-abs}{14.1}
\setval{4b.base.minus.grpo-nodiv.creative-writing-v3}{$-$17.8}
\setval{4b.base.minus.grpo-nodiv.creative-writing-v3-abs}{17.8}
\setval{4b.base.minus.grpo-nodiv.guidebench}{$-$53.9}
\setval{4b.base.minus.grpo-nodiv.guidebench-abs}{53.9}
\setval{4b.base.minus.grpo-nodiv.humaneval-plus}{$-$15.4}
\setval{4b.base.minus.grpo-nodiv.humaneval-plus-abs}{15.4}
\setval{4b.base.minus.grpo-nodiv.if}{$-$36.8}
\setval{4b.base.minus.grpo-nodiv.if-abs}{36.8}
\setval{4b.base.minus.grpo-nodiv.ifbench}{$-$16.7}
\setval{4b.base.minus.grpo-nodiv.ifbench-abs}{16.7}
\setval{4b.base.minus.grpo-nodiv.ifeval}{$-$39.9}
\setval{4b.base.minus.grpo-nodiv.ifeval-abs}{39.9}
\setval{4b.base.minus.grpo-nodiv.judge2.avg}{$-$16.8}
\setval{4b.base.minus.grpo-nodiv.judge2.avg-abs}{16.8}
\setval{4b.base.minus.grpo-nodiv.judge2.creative-writing-v3}{$-$9.7}
\setval{4b.base.minus.grpo-nodiv.judge2.creative-writing-v3-abs}{9.7}
\setval{4b.base.minus.grpo-nodiv.judge2.longbench-write}{$-$17.7}
\setval{4b.base.minus.grpo-nodiv.judge2.longbench-write-abs}{17.7}
\setval{4b.base.minus.grpo-nodiv.judge2.writing}{$-$15.5}
\setval{4b.base.minus.grpo-nodiv.judge2.writing-abs}{15.5}
\setval{4b.base.minus.grpo-nodiv.judge2.writingbench}{$-$19.1}
\setval{4b.base.minus.grpo-nodiv.judge2.writingbench-abs}{19.1}
\setval{4b.base.minus.grpo-nodiv.livecodebench}{$-$11.3}
\setval{4b.base.minus.grpo-nodiv.livecodebench-abs}{11.3}
\setval{4b.base.minus.grpo-nodiv.longbench-write}{$-$21.6}
\setval{4b.base.minus.grpo-nodiv.longbench-write-abs}{21.6}
\setval{4b.base.minus.grpo-nodiv.math}{$-$8.7}
\setval{4b.base.minus.grpo-nodiv.math-abs}{8.7}
\setval{4b.base.minus.grpo-nodiv.math500}{$-$12.2}
\setval{4b.base.minus.grpo-nodiv.math500-abs}{12.2}
\setval{4b.base.minus.grpo-nodiv.mbpp-plus}{$-$15.6}
\setval{4b.base.minus.grpo-nodiv.mbpp-plus-abs}{15.6}
\setval{4b.base.minus.grpo-nodiv.minerva}{$-$14.2}
\setval{4b.base.minus.grpo-nodiv.minerva-abs}{14.2}
\setval{4b.base.minus.grpo-nodiv.olympiad-bench}{$-$9.2}
\setval{4b.base.minus.grpo-nodiv.olympiad-bench-abs}{9.2}
\setval{4b.base.minus.grpo-nodiv.writing}{$-$24.2}
\setval{4b.base.minus.grpo-nodiv.writing-abs}{24.2}
\setval{4b.base.minus.grpo-nodiv.writingbench}{$-$33.2}
\setval{4b.base.minus.grpo-nodiv.writingbench-abs}{33.2}
\setval{4b.base.minus.grpo.aime24}{$-$4.2}
\setval{4b.base.minus.grpo.aime24-abs}{4.2}
\setval{4b.base.minus.grpo.aime25}{$-$5.5}
\setval{4b.base.minus.grpo.aime25-abs}{5.5}
\setval{4b.base.minus.grpo.amc23}{$-$10.8}
\setval{4b.base.minus.grpo.amc23-abs}{10.8}
\setval{4b.base.minus.grpo.avg}{$-$20.1}
\setval{4b.base.minus.grpo.avg-abs}{20.1}
\setval{4b.base.minus.grpo.code}{$-$14.7}
\setval{4b.base.minus.grpo.code-abs}{14.7}
\setval{4b.base.minus.grpo.creative-writing-v3}{$-$22.6}
\setval{4b.base.minus.grpo.creative-writing-v3-abs}{22.6}
\setval{4b.base.minus.grpo.guidebench}{$-$53.1}
\setval{4b.base.minus.grpo.guidebench-abs}{53.1}
\setval{4b.base.minus.grpo.humaneval-plus}{$-$17.7}
\setval{4b.base.minus.grpo.humaneval-plus-abs}{17.7}
\setval{4b.base.minus.grpo.if}{$-$37.6}
\setval{4b.base.minus.grpo.if-abs}{37.6}
\setval{4b.base.minus.grpo.ifbench}{$-$20.7}
\setval{4b.base.minus.grpo.ifbench-abs}{20.7}
\setval{4b.base.minus.grpo.ifeval}{$-$39.1}
\setval{4b.base.minus.grpo.ifeval-abs}{39.1}
\setval{4b.base.minus.grpo.judge2.avg}{$-$18.0}
\setval{4b.base.minus.grpo.judge2.avg-abs}{18.0}
\setval{4b.base.minus.grpo.judge2.creative-writing-v3}{$-$14.2}
\setval{4b.base.minus.grpo.judge2.creative-writing-v3-abs}{14.2}
\setval{4b.base.minus.grpo.judge2.longbench-write}{$-$19.8}
\setval{4b.base.minus.grpo.judge2.longbench-write-abs}{19.8}
\setval{4b.base.minus.grpo.judge2.writing}{$-$19.0}
\setval{4b.base.minus.grpo.judge2.writing-abs}{19.0}
\setval{4b.base.minus.grpo.judge2.writingbench}{$-$23.0}
\setval{4b.base.minus.grpo.judge2.writingbench-abs}{23.0}
\setval{4b.base.minus.grpo.livecodebench}{$-$10.8}
\setval{4b.base.minus.grpo.livecodebench-abs}{10.8}
\setval{4b.base.minus.grpo.longbench-write}{$-$24.8}
\setval{4b.base.minus.grpo.longbench-write-abs}{24.8}
\setval{4b.base.minus.grpo.math}{$-$9.4}
\setval{4b.base.minus.grpo.math-abs}{9.4}
\setval{4b.base.minus.grpo.math500}{$-$13.0}
\setval{4b.base.minus.grpo.math500-abs}{13.0}
\setval{4b.base.minus.grpo.mbpp-plus}{$-$15.6}
\setval{4b.base.minus.grpo.mbpp-plus-abs}{15.6}
\setval{4b.base.minus.grpo.minerva}{$-$13.7}
\setval{4b.base.minus.grpo.minerva-abs}{13.7}
\setval{4b.base.minus.grpo.olympiad-bench}{$-$9.0}
\setval{4b.base.minus.grpo.olympiad-bench-abs}{9.0}
\setval{4b.base.minus.grpo.writing}{$-$29.7}
\setval{4b.base.minus.grpo.writing-abs}{29.7}
\setval{4b.base.minus.grpo.writingbench}{$-$41.8}
\setval{4b.base.minus.grpo.writingbench-abs}{41.8}
\setval{4b.base.minus.mt-grpo.aime24}{$-$2.9}
\setval{4b.base.minus.mt-grpo.aime24-abs}{2.9}
\setval{4b.base.minus.mt-grpo.aime25}{+0.9}
\setval{4b.base.minus.mt-grpo.aime25-abs}{0.9}
\setval{4b.base.minus.mt-grpo.amc23}{$-$15.9}
\setval{4b.base.minus.mt-grpo.amc23-abs}{15.9}
\setval{4b.base.minus.mt-grpo.avg}{$-$18.5}
\setval{4b.base.minus.mt-grpo.avg-abs}{18.5}
\setval{4b.base.minus.mt-grpo.code}{$-$12.6}
\setval{4b.base.minus.mt-grpo.code-abs}{12.6}
\setval{4b.base.minus.mt-grpo.creative-writing-v3}{$-$19.7}
\setval{4b.base.minus.mt-grpo.creative-writing-v3-abs}{19.7}
\setval{4b.base.minus.mt-grpo.guidebench}{$-$54.8}
\setval{4b.base.minus.mt-grpo.guidebench-abs}{54.8}
\setval{4b.base.minus.mt-grpo.humaneval-plus}{$-$14.0}
\setval{4b.base.minus.mt-grpo.humaneval-plus-abs}{14.0}
\setval{4b.base.minus.mt-grpo.if}{$-$37.2}
\setval{4b.base.minus.mt-grpo.if-abs}{37.2}
\setval{4b.base.minus.mt-grpo.ifbench}{$-$17.3}
\setval{4b.base.minus.mt-grpo.ifbench-abs}{17.3}
\setval{4b.base.minus.mt-grpo.ifeval}{$-$39.3}
\setval{4b.base.minus.mt-grpo.ifeval-abs}{39.3}
\setval{4b.base.minus.mt-grpo.judge2.avg}{$-$16.9}
\setval{4b.base.minus.mt-grpo.judge2.avg-abs}{16.9}
\setval{4b.base.minus.mt-grpo.judge2.creative-writing-v3}{$-$13.2}
\setval{4b.base.minus.mt-grpo.judge2.creative-writing-v3-abs}{13.2}
\setval{4b.base.minus.mt-grpo.judge2.longbench-write}{$-$19.9}
\setval{4b.base.minus.mt-grpo.judge2.longbench-write-abs}{19.9}
\setval{4b.base.minus.mt-grpo.judge2.writing}{$-$18.6}
\setval{4b.base.minus.mt-grpo.judge2.writing-abs}{18.6}
\setval{4b.base.minus.mt-grpo.judge2.writingbench}{$-$22.8}
\setval{4b.base.minus.mt-grpo.judge2.writingbench-abs}{22.8}
\setval{4b.base.minus.mt-grpo.livecodebench}{$-$9.4}
\setval{4b.base.minus.mt-grpo.livecodebench-abs}{9.4}
\setval{4b.base.minus.mt-grpo.longbench-write}{$-$22.4}
\setval{4b.base.minus.mt-grpo.longbench-write-abs}{22.4}
\setval{4b.base.minus.mt-grpo.math}{$-$8.1}
\setval{4b.base.minus.mt-grpo.math-abs}{8.1}
\setval{4b.base.minus.mt-grpo.math500}{$-$9.7}
\setval{4b.base.minus.mt-grpo.math500-abs}{9.7}
\setval{4b.base.minus.mt-grpo.mbpp-plus}{$-$14.5}
\setval{4b.base.minus.mt-grpo.mbpp-plus-abs}{14.5}
\setval{4b.base.minus.mt-grpo.minerva}{$-$14.4}
\setval{4b.base.minus.mt-grpo.minerva-abs}{14.4}
\setval{4b.base.minus.mt-grpo.olympiad-bench}{$-$6.9}
\setval{4b.base.minus.mt-grpo.olympiad-bench-abs}{6.9}
\setval{4b.base.minus.mt-grpo.writing}{$-$26.6}
\setval{4b.base.minus.mt-grpo.writing-abs}{26.6}
\setval{4b.base.minus.mt-grpo.writingbench}{$-$37.7}
\setval{4b.base.minus.mt-grpo.writingbench-abs}{37.7}
\setval{4b.base.minus.papo.aime24}{$-$1.5}
\setval{4b.base.minus.papo.aime24-abs}{1.5}
\setval{4b.base.minus.papo.aime25}{$-$0.4}
\setval{4b.base.minus.papo.aime25-abs}{0.4}
\setval{4b.base.minus.papo.amc23}{$-$9.6}
\setval{4b.base.minus.papo.amc23-abs}{9.6}
\setval{4b.base.minus.papo.avg}{$-$18.1}
\setval{4b.base.minus.papo.avg-abs}{18.1}
\setval{4b.base.minus.papo.code}{$-$14.5}
\setval{4b.base.minus.papo.code-abs}{14.5}
\setval{4b.base.minus.papo.creative-writing-v3}{$-$21.8}
\setval{4b.base.minus.papo.creative-writing-v3-abs}{21.8}
\setval{4b.base.minus.papo.guidebench}{$-$55.9}
\setval{4b.base.minus.papo.guidebench-abs}{55.9}
\setval{4b.base.minus.papo.humaneval-plus}{$-$18.0}
\setval{4b.base.minus.papo.humaneval-plus-abs}{18.0}
\setval{4b.base.minus.papo.if}{$-$36.9}
\setval{4b.base.minus.papo.if-abs}{36.9}
\setval{4b.base.minus.papo.ifbench}{$-$16.3}
\setval{4b.base.minus.papo.ifbench-abs}{16.3}
\setval{4b.base.minus.papo.ifeval}{$-$38.5}
\setval{4b.base.minus.papo.ifeval-abs}{38.5}
\setval{4b.base.minus.papo.judge2.avg}{$-$16.5}
\setval{4b.base.minus.papo.judge2.avg-abs}{16.5}
\setval{4b.base.minus.papo.judge2.creative-writing-v3}{$-$13.4}
\setval{4b.base.minus.papo.judge2.creative-writing-v3-abs}{13.4}
\setval{4b.base.minus.papo.judge2.longbench-write}{$-$16.6}
\setval{4b.base.minus.papo.judge2.longbench-write-abs}{16.6}
\setval{4b.base.minus.papo.judge2.writing}{$-$17.2}
\setval{4b.base.minus.papo.judge2.writing-abs}{17.2}
\setval{4b.base.minus.papo.judge2.writingbench}{$-$21.7}
\setval{4b.base.minus.papo.judge2.writingbench-abs}{21.7}
\setval{4b.base.minus.papo.livecodebench}{$-$10.3}
\setval{4b.base.minus.papo.livecodebench-abs}{10.3}
\setval{4b.base.minus.papo.longbench-write}{$-$19.5}
\setval{4b.base.minus.papo.longbench-write-abs}{19.5}
\setval{4b.base.minus.papo.math}{$-$6.9}
\setval{4b.base.minus.papo.math-abs}{6.9}
\setval{4b.base.minus.papo.math500}{$-$9.3}
\setval{4b.base.minus.papo.math500-abs}{9.3}
\setval{4b.base.minus.papo.mbpp-plus}{$-$15.1}
\setval{4b.base.minus.papo.mbpp-plus-abs}{15.1}
\setval{4b.base.minus.papo.minerva}{$-$14.9}
\setval{4b.base.minus.papo.minerva-abs}{14.9}
\setval{4b.base.minus.papo.olympiad-bench}{$-$5.9}
\setval{4b.base.minus.papo.olympiad-bench-abs}{5.9}
\setval{4b.base.minus.papo.writing}{$-$25.1}
\setval{4b.base.minus.papo.writing-abs}{25.1}
\setval{4b.base.minus.papo.writingbench}{$-$34.1}
\setval{4b.base.minus.papo.writingbench-abs}{34.1}
\setval{4b.base.minus.saw.aime24}{$-$4.7}
\setval{4b.base.minus.saw.aime24-abs}{4.7}
\setval{4b.base.minus.saw.aime25}{$-$4.1}
\setval{4b.base.minus.saw.aime25-abs}{4.1}
\setval{4b.base.minus.saw.amc23}{$-$12.7}
\setval{4b.base.minus.saw.amc23-abs}{12.7}
\setval{4b.base.minus.saw.avg}{$-$19.7}
\setval{4b.base.minus.saw.avg-abs}{19.7}
\setval{4b.base.minus.saw.code}{$-$13.3}
\setval{4b.base.minus.saw.code-abs}{13.3}
\setval{4b.base.minus.saw.creative-writing-v3}{$-$20.3}
\setval{4b.base.minus.saw.creative-writing-v3-abs}{20.3}
\setval{4b.base.minus.saw.guidebench}{$-$56.1}
\setval{4b.base.minus.saw.guidebench-abs}{56.1}
\setval{4b.base.minus.saw.humaneval-plus}{$-$15.7}
\setval{4b.base.minus.saw.humaneval-plus-abs}{15.7}
\setval{4b.base.minus.saw.if}{$-$39.7}
\setval{4b.base.minus.saw.if-abs}{39.7}
\setval{4b.base.minus.saw.ifbench}{$-$23.3}
\setval{4b.base.minus.saw.ifbench-abs}{23.3}
\setval{4b.base.minus.saw.ifeval}{$-$39.6}
\setval{4b.base.minus.saw.ifeval-abs}{39.6}
\setval{4b.base.minus.saw.judge2.avg}{$-$18.2}
\setval{4b.base.minus.saw.judge2.avg-abs}{18.2}
\setval{4b.base.minus.saw.judge2.creative-writing-v3}{$-$14.7}
\setval{4b.base.minus.saw.judge2.creative-writing-v3-abs}{14.7}
\setval{4b.base.minus.saw.judge2.longbench-write}{$-$19.1}
\setval{4b.base.minus.saw.judge2.longbench-write-abs}{19.1}
\setval{4b.base.minus.saw.judge2.writing}{$-$18.9}
\setval{4b.base.minus.saw.judge2.writing-abs}{18.9}
\setval{4b.base.minus.saw.judge2.writingbench}{$-$22.9}
\setval{4b.base.minus.saw.judge2.writingbench-abs}{22.9}
\setval{4b.base.minus.saw.livecodebench}{$-$9.6}
\setval{4b.base.minus.saw.livecodebench-abs}{9.6}
\setval{4b.base.minus.saw.longbench-write}{$-$21.8}
\setval{4b.base.minus.saw.longbench-write-abs}{21.8}
\setval{4b.base.minus.saw.math}{$-$9.5}
\setval{4b.base.minus.saw.math-abs}{9.5}
\setval{4b.base.minus.saw.math500}{$-$10.6}
\setval{4b.base.minus.saw.math500-abs}{10.6}
\setval{4b.base.minus.saw.mbpp-plus}{$-$14.5}
\setval{4b.base.minus.saw.mbpp-plus-abs}{14.5}
\setval{4b.base.minus.saw.minerva}{$-$16.5}
\setval{4b.base.minus.saw.minerva-abs}{16.5}
\setval{4b.base.minus.saw.olympiad-bench}{$-$8.2}
\setval{4b.base.minus.saw.olympiad-bench-abs}{8.2}
\setval{4b.base.minus.saw.writing}{$-$26.5}
\setval{4b.base.minus.saw.writing-abs}{26.5}
\setval{4b.base.minus.saw.writingbench}{$-$37.5}
\setval{4b.base.minus.saw.writingbench-abs}{37.5}
\setval{4b.concur-free.aux-change.code}{+0.20}
\setval{4b.concur-free.aux-change.if}{+0.24}
\setval{4b.concur-free.aux-change.math}{+0.12}
\setval{4b.concur-free.aux-change.writing}{$-$0.12}
\setval{4b.concur-free.aux-change.writing-rm}{+0.35}
\setval{4b.concur-free.aux-full.code.early}{0.74}
\setval{4b.concur-free.aux-full.code.late}{0.98}
\setval{4b.concur-free.aux-full.code.run}{0.94}
\setval{4b.concur-free.aux-full.if.early}{0.65}
\setval{4b.concur-free.aux-full.if.late}{0.90}
\setval{4b.concur-free.aux-full.if.run}{0.84}
\setval{4b.concur-free.aux-full.math.early}{0.65}
\setval{4b.concur-free.aux-full.math.late}{0.88}
\setval{4b.concur-free.aux-full.math.run}{0.84}
\setval{4b.concur-free.aux-full.writing-rm.early}{0.09}
\setval{4b.concur-free.aux-full.writing-rm.late}{0.51}
\setval{4b.concur-free.aux-full.writing-rm.run}{0.41}
\setval{4b.concur-free.aux-full.writing.early}{0.58}
\setval{4b.concur-free.aux-full.writing.late}{0.33}
\setval{4b.concur-free.aux-full.writing.run}{0.31}
\setval{4b.concur-free.aux-level.code.early}{0.93}
\setval{4b.concur-free.aux-level.code.late}{1.00}
\setval{4b.concur-free.aux-level.code.run}{0.99}
\setval{4b.concur-free.aux-level.if.early}{0.82}
\setval{4b.concur-free.aux-level.if.late}{0.94}
\setval{4b.concur-free.aux-level.if.run}{0.91}
\setval{4b.concur-free.aux-level.math.early}{0.93}
\setval{4b.concur-free.aux-level.math.late}{0.97}
\setval{4b.concur-free.aux-level.math.run}{0.97}
\setval{4b.concur-free.aux-level.writing-rm.early}{0.77}
\setval{4b.concur-free.aux-level.writing-rm.late}{0.95}
\setval{4b.concur-free.aux-level.writing-rm.run}{0.92}
\setval{4b.concur-free.aux-level.writing.early}{0.87}
\setval{4b.concur-free.aux-level.writing.late}{0.65}
\setval{4b.concur-free.aux-level.writing.run}{0.63}
\setval{4b.concur-free.bench.aime24}{11.9}
\setval{4b.concur-free.bench.aime25}{14.6}
\setval{4b.concur-free.bench.amc23}{54.1}
\setval{4b.concur-free.bench.creative-writing-v3}{45.4}
\setval{4b.concur-free.bench.guidebench}{79.0}
\setval{4b.concur-free.bench.humaneval-plus}{81.1}
\setval{4b.concur-free.bench.ifbench}{39.0}
\setval{4b.concur-free.bench.ifeval}{86.7}
\setval{4b.concur-free.bench.livecodebench}{24.2}
\setval{4b.concur-free.bench.longbench-write}{77.8}
\setval{4b.concur-free.bench.math500}{80.3}
\setval{4b.concur-free.bench.mbpp-plus}{69.7}
\setval{4b.concur-free.bench.minerva}{46.3}
\setval{4b.concur-free.bench.olympiad-bench}{45.7}
\setval{4b.concur-free.bench.writingbench}{77.3}
\setval{4b.concur-free.eval.avg}{55.5}
\setval{4b.concur-free.eval.code}{58.3}
\setval{4b.concur-free.eval.if}{68.2}
\setval{4b.concur-free.eval.math}{42.2}
\setval{4b.concur-free.eval.writing}{66.8}
\setval{4b.concur-free.gate.code.against.early}{0.03}
\setval{4b.concur-free.gate.code.against.late}{0.01}
\setval{4b.concur-free.gate.code.against.run}{0.02}
\setval{4b.concur-free.gate.code.agreement}{+0.003}
\setval{4b.concur-free.gate.code.ceiling.early}{0.11}
\setval{4b.concur-free.gate.code.ceiling.late}{0.02}
\setval{4b.concur-free.gate.code.ceiling.run}{0.04}
\setval{4b.concur-free.gate.code.cosine.early}{0.84}
\setval{4b.concur-free.gate.code.cosine.late}{0.82}
\setval{4b.concur-free.gate.code.cosine.run}{0.83}
\setval{4b.concur-free.gate.if.against.early}{0.05}
\setval{4b.concur-free.gate.if.against.late}{0.04}
\setval{4b.concur-free.gate.if.against.run}{0.05}
\setval{4b.concur-free.gate.if.agreement}{+0.017}
\setval{4b.concur-free.gate.if.ceiling.early}{0.05}
\setval{4b.concur-free.gate.if.ceiling.late}{0.17}
\setval{4b.concur-free.gate.if.ceiling.run}{0.13}
\setval{4b.concur-free.gate.if.cosine.early}{0.73}
\setval{4b.concur-free.gate.if.cosine.late}{0.51}
\setval{4b.concur-free.gate.if.cosine.run}{0.59}
\setval{4b.concur-free.gate.math.against.early}{0.00}
\setval{4b.concur-free.gate.math.against.late}{0.00}
\setval{4b.concur-free.gate.math.against.run}{0.00}
\setval{4b.concur-free.gate.math.ceiling.early}{0.19}
\setval{4b.concur-free.gate.math.ceiling.late}{0.16}
\setval{4b.concur-free.gate.math.ceiling.run}{0.17}
\setval{4b.concur-free.gate.math.cosine.early}{0.88}
\setval{4b.concur-free.gate.math.cosine.late}{0.90}
\setval{4b.concur-free.gate.math.cosine.run}{0.90}
\setval{4b.concur-free.gate.writing-rm.against.early}{0.21}
\setval{4b.concur-free.gate.writing-rm.against.late}{0.55}
\setval{4b.concur-free.gate.writing-rm.against.run}{0.52}
\setval{4b.concur-free.gate.writing-rm.agreement}{+0.059}
\setval{4b.concur-free.gate.writing-rm.ceiling.early}{0.00}
\setval{4b.concur-free.gate.writing-rm.ceiling.late}{0.00}
\setval{4b.concur-free.gate.writing-rm.ceiling.run}{0.00}
\setval{4b.concur-free.gate.writing-rm.cosine.early}{0.67}
\setval{4b.concur-free.gate.writing-rm.cosine.late}{0.12}
\setval{4b.concur-free.gate.writing-rm.cosine.run}{0.19}
\setval{4b.concur-free.gate.writing.against.early}{0.33}
\setval{4b.concur-free.gate.writing.against.late}{0.68}
\setval{4b.concur-free.gate.writing.against.run}{0.65}
\setval{4b.concur-free.gate.writing.agreement}{$-$0.051}
\setval{4b.concur-free.gate.writing.ceiling.early}{0.00}
\setval{4b.concur-free.gate.writing.ceiling.late}{0.00}
\setval{4b.concur-free.gate.writing.ceiling.run}{0.00}
\setval{4b.concur-free.gate.writing.cosine.early}{0.48}
\setval{4b.concur-free.gate.writing.cosine.late}{-0.04}
\setval{4b.concur-free.gate.writing.cosine.run}{0.02}
\setval{4b.concur-free.judge2.avg}{52.8}
\setval{4b.concur-free.judge2.creative-writing-v3}{35.3}
\setval{4b.concur-free.judge2.longbench-write}{69.3}
\setval{4b.concur-free.judge2.writing}{53.0}
\setval{4b.concur-free.judge2.writingbench}{54.5}
\setval{4b.concur-free.last-step}{300}
\setval{4b.concur-free.minus.base.aime24}{+4.1}
\setval{4b.concur-free.minus.base.aime24-abs}{4.1}
\setval{4b.concur-free.minus.base.aime25}{+7.5}
\setval{4b.concur-free.minus.base.aime25-abs}{7.5}
\setval{4b.concur-free.minus.base.amc23}{+10.2}
\setval{4b.concur-free.minus.base.amc23-abs}{10.2}
\setval{4b.concur-free.minus.base.avg}{+21.0}
\setval{4b.concur-free.minus.base.avg-abs}{21.0}
\setval{4b.concur-free.minus.base.code}{+13.7}
\setval{4b.concur-free.minus.base.code-abs}{13.7}
\setval{4b.concur-free.minus.base.creative-writing-v3}{+20.7}
\setval{4b.concur-free.minus.base.creative-writing-v3-abs}{20.7}
\setval{4b.concur-free.minus.base.guidebench}{+54.1}
\setval{4b.concur-free.minus.base.guidebench-abs}{54.1}
\setval{4b.concur-free.minus.base.humaneval-plus}{+16.0}
\setval{4b.concur-free.minus.base.humaneval-plus-abs}{16.0}
\setval{4b.concur-free.minus.base.if}{+38.5}
\setval{4b.concur-free.minus.base.if-abs}{38.5}
\setval{4b.concur-free.minus.base.ifbench}{+22.0}
\setval{4b.concur-free.minus.base.ifbench-abs}{22.0}
\setval{4b.concur-free.minus.base.ifeval}{+39.3}
\setval{4b.concur-free.minus.base.ifeval-abs}{39.3}
\setval{4b.concur-free.minus.base.judge2.avg}{+18.9}
\setval{4b.concur-free.minus.base.judge2.avg-abs}{18.9}
\setval{4b.concur-free.minus.base.judge2.creative-writing-v3}{+14.4}
\setval{4b.concur-free.minus.base.judge2.creative-writing-v3-abs}{14.4}
\setval{4b.concur-free.minus.base.judge2.longbench-write}{+28.4}
\setval{4b.concur-free.minus.base.judge2.longbench-write-abs}{28.4}
\setval{4b.concur-free.minus.base.judge2.writing}{+22.3}
\setval{4b.concur-free.minus.base.judge2.writing-abs}{22.3}
\setval{4b.concur-free.minus.base.judge2.writingbench}{+24.0}
\setval{4b.concur-free.minus.base.judge2.writingbench-abs}{24.0}
\setval{4b.concur-free.minus.base.livecodebench}{+11.0}
\setval{4b.concur-free.minus.base.livecodebench-abs}{11.0}
\setval{4b.concur-free.minus.base.longbench-write}{+33.3}
\setval{4b.concur-free.minus.base.longbench-write-abs}{33.3}
\setval{4b.concur-free.minus.base.math}{+9.9}
\setval{4b.concur-free.minus.base.math-abs}{9.9}
\setval{4b.concur-free.minus.base.math500}{+13.1}
\setval{4b.concur-free.minus.base.math500-abs}{13.1}
\setval{4b.concur-free.minus.base.mbpp-plus}{+14.1}
\setval{4b.concur-free.minus.base.mbpp-plus-abs}{14.1}
\setval{4b.concur-free.minus.base.minerva}{+14.6}
\setval{4b.concur-free.minus.base.minerva-abs}{14.6}
\setval{4b.concur-free.minus.base.olympiad-bench}{+10.1}
\setval{4b.concur-free.minus.base.olympiad-bench-abs}{10.1}
\setval{4b.concur-free.minus.base.writing}{+33.0}
\setval{4b.concur-free.minus.base.writing-abs}{33.0}
\setval{4b.concur-free.minus.base.writingbench}{+44.9}
\setval{4b.concur-free.minus.base.writingbench-abs}{44.9}
\setval{4b.concur-free.minus.concur-tied.aime24}{$-$7.6}
\setval{4b.concur-free.minus.concur-tied.aime24-abs}{7.6}
\setval{4b.concur-free.minus.concur-tied.aime25}{$-$1.8}
\setval{4b.concur-free.minus.concur-tied.aime25-abs}{1.8}
\setval{4b.concur-free.minus.concur-tied.amc23}{$-$15.1}
\setval{4b.concur-free.minus.concur-tied.amc23-abs}{15.1}
\setval{4b.concur-free.minus.concur-tied.avg}{$-$3.2}
\setval{4b.concur-free.minus.concur-tied.avg-abs}{3.2}
\setval{4b.concur-free.minus.concur-tied.code}{$-$0.8}
\setval{4b.concur-free.minus.concur-tied.code-abs}{0.8}
\setval{4b.concur-free.minus.concur-tied.creative-writing-v3}{$-$2.4}
\setval{4b.concur-free.minus.concur-tied.creative-writing-v3-abs}{2.4}
\setval{4b.concur-free.minus.concur-tied.guidebench}{$-$0.2}
\setval{4b.concur-free.minus.concur-tied.guidebench-abs}{0.2}
\setval{4b.concur-free.minus.concur-tied.humaneval-plus}{$-$2.9}
\setval{4b.concur-free.minus.concur-tied.humaneval-plus-abs}{2.9}
\setval{4b.concur-free.minus.concur-tied.if}{+0.0}
\setval{4b.concur-free.minus.concur-tied.if-abs}{0.0}
\setval{4b.concur-free.minus.concur-tied.ifbench}{+1.3}
\setval{4b.concur-free.minus.concur-tied.ifbench-abs}{1.3}
\setval{4b.concur-free.minus.concur-tied.ifeval}{$-$1.1}
\setval{4b.concur-free.minus.concur-tied.ifeval-abs}{1.1}
\setval{4b.concur-free.minus.concur-tied.judge2.avg}{$-$2.7}
\setval{4b.concur-free.minus.concur-tied.judge2.avg-abs}{2.7}
\setval{4b.concur-free.minus.concur-tied.judge2.creative-writing-v3}{+1.3}
\setval{4b.concur-free.minus.concur-tied.judge2.creative-writing-v3-abs}{1.3}
\setval{4b.concur-free.minus.concur-tied.judge2.longbench-write}{+1.4}
\setval{4b.concur-free.minus.concur-tied.judge2.longbench-write-abs}{1.4}
\setval{4b.concur-free.minus.concur-tied.judge2.writing}{+0.6}
\setval{4b.concur-free.minus.concur-tied.judge2.writing-abs}{0.6}
\setval{4b.concur-free.minus.concur-tied.judge2.writingbench}{$-$0.9}
\setval{4b.concur-free.minus.concur-tied.judge2.writingbench-abs}{0.9}
\setval{4b.concur-free.minus.concur-tied.livecodebench}{$-$0.6}
\setval{4b.concur-free.minus.concur-tied.livecodebench-abs}{0.6}
\setval{4b.concur-free.minus.concur-tied.longbench-write}{+0.0}
\setval{4b.concur-free.minus.concur-tied.longbench-write-abs}{0.0}
\setval{4b.concur-free.minus.concur-tied.math}{$-$6.6}
\setval{4b.concur-free.minus.concur-tied.math-abs}{6.6}
\setval{4b.concur-free.minus.concur-tied.math500}{$-$5.2}
\setval{4b.concur-free.minus.concur-tied.math500-abs}{5.2}
\setval{4b.concur-free.minus.concur-tied.mbpp-plus}{+1.1}
\setval{4b.concur-free.minus.concur-tied.mbpp-plus-abs}{1.1}
\setval{4b.concur-free.minus.concur-tied.minerva}{$-$2.9}
\setval{4b.concur-free.minus.concur-tied.minerva-abs}{2.9}
\setval{4b.concur-free.minus.concur-tied.olympiad-bench}{$-$7.1}
\setval{4b.concur-free.minus.concur-tied.olympiad-bench-abs}{7.1}
\setval{4b.concur-free.minus.concur-tied.writing}{$-$1.8}
\setval{4b.concur-free.minus.concur-tied.writing-abs}{1.8}
\setval{4b.concur-free.minus.concur-tied.writingbench}{$-$2.9}
\setval{4b.concur-free.minus.concur-tied.writingbench-abs}{2.9}
\setval{4b.concur-free.minus.concur.aime24}{$-$8.6}
\setval{4b.concur-free.minus.concur.aime24-abs}{8.6}
\setval{4b.concur-free.minus.concur.aime25}{$-$4.2}
\setval{4b.concur-free.minus.concur.aime25-abs}{4.2}
\setval{4b.concur-free.minus.concur.amc23}{$-$15.2}
\setval{4b.concur-free.minus.concur.amc23-abs}{15.2}
\setval{4b.concur-free.minus.concur.avg}{$-$3.3}
\setval{4b.concur-free.minus.concur.avg-abs}{3.3}
\setval{4b.concur-free.minus.concur.code}{$-$1.4}
\setval{4b.concur-free.minus.concur.code-abs}{1.4}
\setval{4b.concur-free.minus.concur.creative-writing-v3}{+0.7}
\setval{4b.concur-free.minus.concur.creative-writing-v3-abs}{0.7}
\setval{4b.concur-free.minus.concur.guidebench}{+1.6}
\setval{4b.concur-free.minus.concur.guidebench-abs}{1.6}
\setval{4b.concur-free.minus.concur.humaneval-plus}{$-$1.3}
\setval{4b.concur-free.minus.concur.humaneval-plus-abs}{1.3}
\setval{4b.concur-free.minus.concur.if}{$-$0.4}
\setval{4b.concur-free.minus.concur.if-abs}{0.4}
\setval{4b.concur-free.minus.concur.ifbench}{$-$1.3}
\setval{4b.concur-free.minus.concur.ifbench-abs}{1.3}
\setval{4b.concur-free.minus.concur.ifeval}{$-$1.4}
\setval{4b.concur-free.minus.concur.ifeval-abs}{1.4}
\setval{4b.concur-free.minus.concur.judge2.avg}{$-$3.0}
\setval{4b.concur-free.minus.concur.judge2.avg-abs}{3.0}
\setval{4b.concur-free.minus.concur.judge2.creative-writing-v3}{+2.9}
\setval{4b.concur-free.minus.concur.judge2.creative-writing-v3-abs}{2.9}
\setval{4b.concur-free.minus.concur.judge2.longbench-write}{+2.7}
\setval{4b.concur-free.minus.concur.judge2.longbench-write-abs}{2.7}
\setval{4b.concur-free.minus.concur.judge2.writing}{+1.3}
\setval{4b.concur-free.minus.concur.judge2.writing-abs}{1.3}
\setval{4b.concur-free.minus.concur.judge2.writingbench}{$-$1.6}
\setval{4b.concur-free.minus.concur.judge2.writingbench-abs}{1.6}
\setval{4b.concur-free.minus.concur.livecodebench}{$-$1.0}
\setval{4b.concur-free.minus.concur.livecodebench-abs}{1.0}
\setval{4b.concur-free.minus.concur.longbench-write}{+1.5}
\setval{4b.concur-free.minus.concur.longbench-write-abs}{1.5}
\setval{4b.concur-free.minus.concur.math}{$-$7.3}
\setval{4b.concur-free.minus.concur.math-abs}{7.3}
\setval{4b.concur-free.minus.concur.math500}{$-$5.0}
\setval{4b.concur-free.minus.concur.math500-abs}{5.0}
\setval{4b.concur-free.minus.concur.mbpp-plus}{$-$1.9}
\setval{4b.concur-free.minus.concur.mbpp-plus-abs}{1.9}
\setval{4b.concur-free.minus.concur.minerva}{$-$3.0}
\setval{4b.concur-free.minus.concur.minerva-abs}{3.0}
\setval{4b.concur-free.minus.concur.olympiad-bench}{$-$7.9}
\setval{4b.concur-free.minus.concur.olympiad-bench-abs}{7.9}
\setval{4b.concur-free.minus.concur.writing}{+0.0}
\setval{4b.concur-free.minus.concur.writing-abs}{0.0}
\setval{4b.concur-free.minus.concur.writingbench}{$-$2.1}
\setval{4b.concur-free.minus.concur.writingbench-abs}{2.1}
\setval{4b.concur-free.minus.gd2po.aime24}{+0.0}
\setval{4b.concur-free.minus.gd2po.aime24-abs}{0.0}
\setval{4b.concur-free.minus.gd2po.aime25}{+7.8}
\setval{4b.concur-free.minus.gd2po.aime25-abs}{7.8}
\setval{4b.concur-free.minus.gd2po.amc23}{$-$3.6}
\setval{4b.concur-free.minus.gd2po.amc23-abs}{3.6}
\setval{4b.concur-free.minus.gd2po.avg}{+1.4}
\setval{4b.concur-free.minus.gd2po.avg-abs}{1.4}
\setval{4b.concur-free.minus.gd2po.code}{$-$1.1}
\setval{4b.concur-free.minus.gd2po.code-abs}{1.1}
\setval{4b.concur-free.minus.gd2po.creative-writing-v3}{+0.7}
\setval{4b.concur-free.minus.gd2po.creative-writing-v3-abs}{0.7}
\setval{4b.concur-free.minus.gd2po.guidebench}{$-$0.1}
\setval{4b.concur-free.minus.gd2po.guidebench-abs}{0.1}
\setval{4b.concur-free.minus.gd2po.humaneval-plus}{$-$2.3}
\setval{4b.concur-free.minus.gd2po.humaneval-plus-abs}{2.3}
\setval{4b.concur-free.minus.gd2po.if}{+0.1}
\setval{4b.concur-free.minus.gd2po.if-abs}{0.1}
\setval{4b.concur-free.minus.gd2po.ifbench}{+1.3}
\setval{4b.concur-free.minus.gd2po.ifbench-abs}{1.3}
\setval{4b.concur-free.minus.gd2po.ifeval}{$-$1.0}
\setval{4b.concur-free.minus.gd2po.ifeval-abs}{1.0}
\setval{4b.concur-free.minus.gd2po.judge2.avg}{+1.2}
\setval{4b.concur-free.minus.gd2po.judge2.avg-abs}{1.2}
\setval{4b.concur-free.minus.gd2po.judge2.creative-writing-v3}{+2.4}
\setval{4b.concur-free.minus.gd2po.judge2.creative-writing-v3-abs}{2.4}
\setval{4b.concur-free.minus.gd2po.judge2.longbench-write}{+8.4}
\setval{4b.concur-free.minus.gd2po.judge2.longbench-write-abs}{8.4}
\setval{4b.concur-free.minus.gd2po.judge2.writing}{+4.2}
\setval{4b.concur-free.minus.gd2po.judge2.writing-abs}{4.2}
\setval{4b.concur-free.minus.gd2po.judge2.writingbench}{+1.9}
\setval{4b.concur-free.minus.gd2po.judge2.writingbench-abs}{1.9}
\setval{4b.concur-free.minus.gd2po.livecodebench}{+0.7}
\setval{4b.concur-free.minus.gd2po.livecodebench-abs}{0.7}
\setval{4b.concur-free.minus.gd2po.longbench-write}{+8.8}
\setval{4b.concur-free.minus.gd2po.longbench-write-abs}{8.8}
\setval{4b.concur-free.minus.gd2po.math}{+1.4}
\setval{4b.concur-free.minus.gd2po.math-abs}{1.4}
\setval{4b.concur-free.minus.gd2po.math500}{+2.7}
\setval{4b.concur-free.minus.gd2po.math500-abs}{2.7}
\setval{4b.concur-free.minus.gd2po.mbpp-plus}{$-$1.8}
\setval{4b.concur-free.minus.gd2po.mbpp-plus-abs}{1.8}
\setval{4b.concur-free.minus.gd2po.minerva}{$-$0.9}
\setval{4b.concur-free.minus.gd2po.minerva-abs}{0.9}
\setval{4b.concur-free.minus.gd2po.olympiad-bench}{+2.2}
\setval{4b.concur-free.minus.gd2po.olympiad-bench-abs}{2.2}
\setval{4b.concur-free.minus.gd2po.writing}{+5.6}
\setval{4b.concur-free.minus.gd2po.writing-abs}{5.6}
\setval{4b.concur-free.minus.gd2po.writingbench}{+7.1}
\setval{4b.concur-free.minus.gd2po.writingbench-abs}{7.1}
\setval{4b.concur-free.minus.gdpo-noaux.aime24}{$-$11.6}
\setval{4b.concur-free.minus.gdpo-noaux.aime24-abs}{11.6}
\setval{4b.concur-free.minus.gdpo-noaux.aime25}{$-$4.2}
\setval{4b.concur-free.minus.gdpo-noaux.aime25-abs}{4.2}
\setval{4b.concur-free.minus.gdpo-noaux.amc23}{$-$11.6}
\setval{4b.concur-free.minus.gdpo-noaux.amc23-abs}{11.6}
\setval{4b.concur-free.minus.gdpo-noaux.avg}{$-$1.1}
\setval{4b.concur-free.minus.gdpo-noaux.avg-abs}{1.1}
\setval{4b.concur-free.minus.gdpo-noaux.code}{$-$2.5}
\setval{4b.concur-free.minus.gdpo-noaux.code-abs}{2.5}
\setval{4b.concur-free.minus.gdpo-noaux.creative-writing-v3}{+6.1}
\setval{4b.concur-free.minus.gdpo-noaux.creative-writing-v3-abs}{6.1}
\setval{4b.concur-free.minus.gdpo-noaux.guidebench}{+3.1}
\setval{4b.concur-free.minus.gdpo-noaux.guidebench-abs}{3.1}
\setval{4b.concur-free.minus.gdpo-noaux.humaneval-plus}{$-$4.4}
\setval{4b.concur-free.minus.gdpo-noaux.humaneval-plus-abs}{4.4}
\setval{4b.concur-free.minus.gdpo-noaux.if}{+1.2}
\setval{4b.concur-free.minus.gdpo-noaux.if-abs}{1.2}
\setval{4b.concur-free.minus.gdpo-noaux.ifbench}{$-$1.3}
\setval{4b.concur-free.minus.gdpo-noaux.ifbench-abs}{1.3}
\setval{4b.concur-free.minus.gdpo-noaux.ifeval}{+1.9}
\setval{4b.concur-free.minus.gdpo-noaux.ifeval-abs}{1.9}
\setval{4b.concur-free.minus.gdpo-noaux.judge2.avg}{$-$0.5}
\setval{4b.concur-free.minus.gdpo-noaux.judge2.avg-abs}{0.5}
\setval{4b.concur-free.minus.gdpo-noaux.judge2.creative-writing-v3}{+11.9}
\setval{4b.concur-free.minus.gdpo-noaux.judge2.creative-writing-v3-abs}{11.9}
\setval{4b.concur-free.minus.gdpo-noaux.judge2.longbench-write}{+18.6}
\setval{4b.concur-free.minus.gdpo-noaux.judge2.longbench-write-abs}{18.6}
\setval{4b.concur-free.minus.gdpo-noaux.judge2.writing}{+13.0}
\setval{4b.concur-free.minus.gdpo-noaux.judge2.writing-abs}{13.0}
\setval{4b.concur-free.minus.gdpo-noaux.judge2.writingbench}{+8.4}
\setval{4b.concur-free.minus.gdpo-noaux.judge2.writingbench-abs}{8.4}
\setval{4b.concur-free.minus.gdpo-noaux.livecodebench}{$-$0.6}
\setval{4b.concur-free.minus.gdpo-noaux.livecodebench-abs}{0.6}
\setval{4b.concur-free.minus.gdpo-noaux.longbench-write}{+13.4}
\setval{4b.concur-free.minus.gdpo-noaux.longbench-write-abs}{13.4}
\setval{4b.concur-free.minus.gdpo-noaux.math}{$-$7.1}
\setval{4b.concur-free.minus.gdpo-noaux.math-abs}{7.1}
\setval{4b.concur-free.minus.gdpo-noaux.math500}{$-$4.4}
\setval{4b.concur-free.minus.gdpo-noaux.math500-abs}{4.4}
\setval{4b.concur-free.minus.gdpo-noaux.mbpp-plus}{$-$2.7}
\setval{4b.concur-free.minus.gdpo-noaux.mbpp-plus-abs}{2.7}
\setval{4b.concur-free.minus.gdpo-noaux.minerva}{$-$3.3}
\setval{4b.concur-free.minus.gdpo-noaux.minerva-abs}{3.3}
\setval{4b.concur-free.minus.gdpo-noaux.olympiad-bench}{$-$7.4}
\setval{4b.concur-free.minus.gdpo-noaux.olympiad-bench-abs}{7.4}
\setval{4b.concur-free.minus.gdpo-noaux.writing}{+9.8}
\setval{4b.concur-free.minus.gdpo-noaux.writing-abs}{9.8}
\setval{4b.concur-free.minus.gdpo-noaux.writingbench}{+10.0}
\setval{4b.concur-free.minus.gdpo-noaux.writingbench-abs}{10.0}
\setval{4b.concur-free.minus.gdpo-w0.5.aime24}{$-$2.7}
\setval{4b.concur-free.minus.gdpo-w0.5.aime24-abs}{2.7}
\setval{4b.concur-free.minus.gdpo-w0.5.aime25}{+4.3}
\setval{4b.concur-free.minus.gdpo-w0.5.aime25-abs}{4.3}
\setval{4b.concur-free.minus.gdpo-w0.5.amc23}{$-$4.2}
\setval{4b.concur-free.minus.gdpo-w0.5.amc23-abs}{4.2}
\setval{4b.concur-free.minus.gdpo-w0.5.avg}{+0.9}
\setval{4b.concur-free.minus.gdpo-w0.5.avg-abs}{0.9}
\setval{4b.concur-free.minus.gdpo-w0.5.code}{$-$0.8}
\setval{4b.concur-free.minus.gdpo-w0.5.code-abs}{0.8}
\setval{4b.concur-free.minus.gdpo-w0.5.creative-writing-v3}{+7.3}
\setval{4b.concur-free.minus.gdpo-w0.5.creative-writing-v3-abs}{7.3}
\setval{4b.concur-free.minus.gdpo-w0.5.guidebench}{$-$0.6}
\setval{4b.concur-free.minus.gdpo-w0.5.guidebench-abs}{0.6}
\setval{4b.concur-free.minus.gdpo-w0.5.humaneval-plus}{$-$0.9}
\setval{4b.concur-free.minus.gdpo-w0.5.humaneval-plus-abs}{0.9}
\setval{4b.concur-free.minus.gdpo-w0.5.if}{$-$0.3}
\setval{4b.concur-free.minus.gdpo-w0.5.if-abs}{0.3}
\setval{4b.concur-free.minus.gdpo-w0.5.ifbench}{$-$0.3}
\setval{4b.concur-free.minus.gdpo-w0.5.ifbench-abs}{0.3}
\setval{4b.concur-free.minus.gdpo-w0.5.ifeval}{$-$0.1}
\setval{4b.concur-free.minus.gdpo-w0.5.ifeval-abs}{0.1}
\setval{4b.concur-free.minus.gdpo-w0.5.judge2.avg}{+0.3}
\setval{4b.concur-free.minus.gdpo-w0.5.judge2.avg-abs}{0.3}
\setval{4b.concur-free.minus.gdpo-w0.5.judge2.creative-writing-v3}{+5.2}
\setval{4b.concur-free.minus.gdpo-w0.5.judge2.creative-writing-v3-abs}{5.2}
\setval{4b.concur-free.minus.gdpo-w0.5.judge2.longbench-write}{+8.8}
\setval{4b.concur-free.minus.gdpo-w0.5.judge2.longbench-write-abs}{8.8}
\setval{4b.concur-free.minus.gdpo-w0.5.judge2.writing}{+5.0}
\setval{4b.concur-free.minus.gdpo-w0.5.judge2.writing-abs}{5.0}
\setval{4b.concur-free.minus.gdpo-w0.5.judge2.writingbench}{+0.8}
\setval{4b.concur-free.minus.gdpo-w0.5.judge2.writingbench-abs}{0.8}
\setval{4b.concur-free.minus.gdpo-w0.5.livecodebench}{+0.0}
\setval{4b.concur-free.minus.gdpo-w0.5.livecodebench-abs}{0.0}
\setval{4b.concur-free.minus.gdpo-w0.5.longbench-write}{+10.4}
\setval{4b.concur-free.minus.gdpo-w0.5.longbench-write-abs}{10.4}
\setval{4b.concur-free.minus.gdpo-w0.5.math}{$-$1.1}
\setval{4b.concur-free.minus.gdpo-w0.5.math-abs}{1.1}
\setval{4b.concur-free.minus.gdpo-w0.5.math500}{$-$1.5}
\setval{4b.concur-free.minus.gdpo-w0.5.math500-abs}{1.5}
\setval{4b.concur-free.minus.gdpo-w0.5.mbpp-plus}{$-$1.5}
\setval{4b.concur-free.minus.gdpo-w0.5.mbpp-plus-abs}{1.5}
\setval{4b.concur-free.minus.gdpo-w0.5.minerva}{+0.1}
\setval{4b.concur-free.minus.gdpo-w0.5.minerva-abs}{0.1}
\setval{4b.concur-free.minus.gdpo-w0.5.olympiad-bench}{$-$2.7}
\setval{4b.concur-free.minus.gdpo-w0.5.olympiad-bench-abs}{2.7}
\setval{4b.concur-free.minus.gdpo-w0.5.writing}{+8.1}
\setval{4b.concur-free.minus.gdpo-w0.5.writing-abs}{8.1}
\setval{4b.concur-free.minus.gdpo-w0.5.writingbench}{+6.6}
\setval{4b.concur-free.minus.gdpo-w0.5.writingbench-abs}{6.6}
\setval{4b.concur-free.minus.gdpo.aime24}{+2.8}
\setval{4b.concur-free.minus.gdpo.aime24-abs}{2.8}
\setval{4b.concur-free.minus.gdpo.aime25}{+4.7}
\setval{4b.concur-free.minus.gdpo.aime25-abs}{4.7}
\setval{4b.concur-free.minus.gdpo.amc23}{$-$5.5}
\setval{4b.concur-free.minus.gdpo.amc23-abs}{5.5}
\setval{4b.concur-free.minus.gdpo.avg}{+1.2}
\setval{4b.concur-free.minus.gdpo.avg-abs}{1.2}
\setval{4b.concur-free.minus.gdpo.code}{$-$0.7}
\setval{4b.concur-free.minus.gdpo.code-abs}{0.7}
\setval{4b.concur-free.minus.gdpo.creative-writing-v3}{$-$0.5}
\setval{4b.concur-free.minus.gdpo.creative-writing-v3-abs}{0.5}
\setval{4b.concur-free.minus.gdpo.guidebench}{$-$1.6}
\setval{4b.concur-free.minus.gdpo.guidebench-abs}{1.6}
\setval{4b.concur-free.minus.gdpo.humaneval-plus}{$-$0.7}
\setval{4b.concur-free.minus.gdpo.humaneval-plus-abs}{0.7}
\setval{4b.concur-free.minus.gdpo.if}{+0.7}
\setval{4b.concur-free.minus.gdpo.if-abs}{0.7}
\setval{4b.concur-free.minus.gdpo.ifbench}{+2.7}
\setval{4b.concur-free.minus.gdpo.ifbench-abs}{2.7}
\setval{4b.concur-free.minus.gdpo.ifeval}{+1.0}
\setval{4b.concur-free.minus.gdpo.ifeval-abs}{1.0}
\setval{4b.concur-free.minus.gdpo.judge2.avg}{+0.9}
\setval{4b.concur-free.minus.gdpo.judge2.avg-abs}{0.9}
\setval{4b.concur-free.minus.gdpo.judge2.creative-writing-v3}{+0.9}
\setval{4b.concur-free.minus.gdpo.judge2.creative-writing-v3-abs}{0.9}
\setval{4b.concur-free.minus.gdpo.judge2.longbench-write}{+7.0}
\setval{4b.concur-free.minus.gdpo.judge2.longbench-write-abs}{7.0}
\setval{4b.concur-free.minus.gdpo.judge2.writing}{+2.8}
\setval{4b.concur-free.minus.gdpo.judge2.writing-abs}{2.8}
\setval{4b.concur-free.minus.gdpo.judge2.writingbench}{+0.4}
\setval{4b.concur-free.minus.gdpo.judge2.writingbench-abs}{0.4}
\setval{4b.concur-free.minus.gdpo.livecodebench}{$-$0.2}
\setval{4b.concur-free.minus.gdpo.livecodebench-abs}{0.2}
\setval{4b.concur-free.minus.gdpo.longbench-write}{+8.8}
\setval{4b.concur-free.minus.gdpo.longbench-write-abs}{8.8}
\setval{4b.concur-free.minus.gdpo.math}{+0.8}
\setval{4b.concur-free.minus.gdpo.math-abs}{0.8}
\setval{4b.concur-free.minus.gdpo.math500}{+1.0}
\setval{4b.concur-free.minus.gdpo.math500-abs}{1.0}
\setval{4b.concur-free.minus.gdpo.mbpp-plus}{$-$1.2}
\setval{4b.concur-free.minus.gdpo.mbpp-plus-abs}{1.2}
\setval{4b.concur-free.minus.gdpo.minerva}{+0.8}
\setval{4b.concur-free.minus.gdpo.minerva-abs}{0.8}
\setval{4b.concur-free.minus.gdpo.olympiad-bench}{+1.1}
\setval{4b.concur-free.minus.gdpo.olympiad-bench-abs}{1.1}
\setval{4b.concur-free.minus.gdpo.writing}{+4.5}
\setval{4b.concur-free.minus.gdpo.writing-abs}{4.5}
\setval{4b.concur-free.minus.gdpo.writingbench}{+5.4}
\setval{4b.concur-free.minus.gdpo.writingbench-abs}{5.4}
\setval{4b.concur-free.minus.geoma.aime24}{+0.3}
\setval{4b.concur-free.minus.geoma.aime24-abs}{0.3}
\setval{4b.concur-free.minus.geoma.aime25}{+4.0}
\setval{4b.concur-free.minus.geoma.aime25-abs}{4.0}
\setval{4b.concur-free.minus.geoma.amc23}{$-$1.3}
\setval{4b.concur-free.minus.geoma.amc23-abs}{1.3}
\setval{4b.concur-free.minus.geoma.avg}{+1.7}
\setval{4b.concur-free.minus.geoma.avg-abs}{1.7}
\setval{4b.concur-free.minus.geoma.code}{$-$0.4}
\setval{4b.concur-free.minus.geoma.code-abs}{0.4}
\setval{4b.concur-free.minus.geoma.creative-writing-v3}{$-$0.4}
\setval{4b.concur-free.minus.geoma.creative-writing-v3-abs}{0.4}
\setval{4b.concur-free.minus.geoma.guidebench}{+0.5}
\setval{4b.concur-free.minus.geoma.guidebench-abs}{0.5}
\setval{4b.concur-free.minus.geoma.humaneval-plus}{+0.9}
\setval{4b.concur-free.minus.geoma.humaneval-plus-abs}{0.9}
\setval{4b.concur-free.minus.geoma.if}{+0.8}
\setval{4b.concur-free.minus.geoma.if-abs}{0.8}
\setval{4b.concur-free.minus.geoma.ifbench}{+3.0}
\setval{4b.concur-free.minus.geoma.ifbench-abs}{3.0}
\setval{4b.concur-free.minus.geoma.ifeval}{$-$1.1}
\setval{4b.concur-free.minus.geoma.ifeval-abs}{1.1}
\setval{4b.concur-free.minus.geoma.judge2.avg}{+1.6}
\setval{4b.concur-free.minus.geoma.judge2.avg-abs}{1.6}
\setval{4b.concur-free.minus.geoma.judge2.creative-writing-v3}{+2.2}
\setval{4b.concur-free.minus.geoma.judge2.creative-writing-v3-abs}{2.2}
\setval{4b.concur-free.minus.geoma.judge2.longbench-write}{+11.7}
\setval{4b.concur-free.minus.geoma.judge2.longbench-write-abs}{11.7}
\setval{4b.concur-free.minus.geoma.judge2.writing}{+5.9}
\setval{4b.concur-free.minus.geoma.judge2.writing-abs}{5.9}
\setval{4b.concur-free.minus.geoma.judge2.writingbench}{+3.8}
\setval{4b.concur-free.minus.geoma.judge2.writingbench-abs}{3.8}
\setval{4b.concur-free.minus.geoma.livecodebench}{$-$1.5}
\setval{4b.concur-free.minus.geoma.livecodebench-abs}{1.5}
\setval{4b.concur-free.minus.geoma.longbench-write}{+12.0}
\setval{4b.concur-free.minus.geoma.longbench-write-abs}{12.0}
\setval{4b.concur-free.minus.geoma.math}{+0.9}
\setval{4b.concur-free.minus.geoma.math-abs}{0.9}
\setval{4b.concur-free.minus.geoma.math500}{+1.5}
\setval{4b.concur-free.minus.geoma.math500-abs}{1.5}
\setval{4b.concur-free.minus.geoma.mbpp-plus}{$-$0.6}
\setval{4b.concur-free.minus.geoma.mbpp-plus-abs}{0.6}
\setval{4b.concur-free.minus.geoma.minerva}{$-$0.7}
\setval{4b.concur-free.minus.geoma.minerva-abs}{0.7}
\setval{4b.concur-free.minus.geoma.olympiad-bench}{+1.9}
\setval{4b.concur-free.minus.geoma.olympiad-bench-abs}{1.9}
\setval{4b.concur-free.minus.geoma.writing}{+6.4}
\setval{4b.concur-free.minus.geoma.writing-abs}{6.4}
\setval{4b.concur-free.minus.geoma.writingbench}{+7.8}
\setval{4b.concur-free.minus.geoma.writingbench-abs}{7.8}
\setval{4b.concur-free.minus.grpo-nodiv.aime24}{$-$0.5}
\setval{4b.concur-free.minus.grpo-nodiv.aime24-abs}{0.5}
\setval{4b.concur-free.minus.grpo-nodiv.aime25}{+6.3}
\setval{4b.concur-free.minus.grpo-nodiv.aime25-abs}{6.3}
\setval{4b.concur-free.minus.grpo-nodiv.amc23}{$-$0.4}
\setval{4b.concur-free.minus.grpo-nodiv.amc23-abs}{0.4}
\setval{4b.concur-free.minus.grpo-nodiv.avg}{+2.5}
\setval{4b.concur-free.minus.grpo-nodiv.avg-abs}{2.5}
\setval{4b.concur-free.minus.grpo-nodiv.code}{$-$0.4}
\setval{4b.concur-free.minus.grpo-nodiv.code-abs}{0.4}
\setval{4b.concur-free.minus.grpo-nodiv.creative-writing-v3}{+2.9}
\setval{4b.concur-free.minus.grpo-nodiv.creative-writing-v3-abs}{2.9}
\setval{4b.concur-free.minus.grpo-nodiv.guidebench}{+0.2}
\setval{4b.concur-free.minus.grpo-nodiv.guidebench-abs}{0.2}
\setval{4b.concur-free.minus.grpo-nodiv.humaneval-plus}{+0.6}
\setval{4b.concur-free.minus.grpo-nodiv.humaneval-plus-abs}{0.6}
\setval{4b.concur-free.minus.grpo-nodiv.if}{+1.6}
\setval{4b.concur-free.minus.grpo-nodiv.if-abs}{1.6}
\setval{4b.concur-free.minus.grpo-nodiv.ifbench}{+5.3}
\setval{4b.concur-free.minus.grpo-nodiv.ifbench-abs}{5.3}
\setval{4b.concur-free.minus.grpo-nodiv.ifeval}{$-$0.6}
\setval{4b.concur-free.minus.grpo-nodiv.ifeval-abs}{0.6}
\setval{4b.concur-free.minus.grpo-nodiv.judge2.avg}{+2.1}
\setval{4b.concur-free.minus.grpo-nodiv.judge2.avg-abs}{2.1}
\setval{4b.concur-free.minus.grpo-nodiv.judge2.creative-writing-v3}{+4.8}
\setval{4b.concur-free.minus.grpo-nodiv.judge2.creative-writing-v3-abs}{4.8}
\setval{4b.concur-free.minus.grpo-nodiv.judge2.longbench-write}{+10.7}
\setval{4b.concur-free.minus.grpo-nodiv.judge2.longbench-write-abs}{10.7}
\setval{4b.concur-free.minus.grpo-nodiv.judge2.writing}{+6.8}
\setval{4b.concur-free.minus.grpo-nodiv.judge2.writing-abs}{6.8}
\setval{4b.concur-free.minus.grpo-nodiv.judge2.writingbench}{+4.9}
\setval{4b.concur-free.minus.grpo-nodiv.judge2.writingbench-abs}{4.9}
\setval{4b.concur-free.minus.grpo-nodiv.livecodebench}{$-$0.3}
\setval{4b.concur-free.minus.grpo-nodiv.livecodebench-abs}{0.3}
\setval{4b.concur-free.minus.grpo-nodiv.longbench-write}{+11.8}
\setval{4b.concur-free.minus.grpo-nodiv.longbench-write-abs}{11.8}
\setval{4b.concur-free.minus.grpo-nodiv.math}{+1.3}
\setval{4b.concur-free.minus.grpo-nodiv.math-abs}{1.3}
\setval{4b.concur-free.minus.grpo-nodiv.math500}{+0.8}
\setval{4b.concur-free.minus.grpo-nodiv.math500-abs}{0.8}
\setval{4b.concur-free.minus.grpo-nodiv.mbpp-plus}{$-$1.5}
\setval{4b.concur-free.minus.grpo-nodiv.mbpp-plus-abs}{1.5}
\setval{4b.concur-free.minus.grpo-nodiv.minerva}{+0.5}
\setval{4b.concur-free.minus.grpo-nodiv.minerva-abs}{0.5}
\setval{4b.concur-free.minus.grpo-nodiv.olympiad-bench}{+0.9}
\setval{4b.concur-free.minus.grpo-nodiv.olympiad-bench-abs}{0.9}
\setval{4b.concur-free.minus.grpo-nodiv.writing}{+8.8}
\setval{4b.concur-free.minus.grpo-nodiv.writing-abs}{8.8}
\setval{4b.concur-free.minus.grpo-nodiv.writingbench}{+11.7}
\setval{4b.concur-free.minus.grpo-nodiv.writingbench-abs}{11.7}
\setval{4b.concur-free.minus.grpo.aime24}{$-$0.1}
\setval{4b.concur-free.minus.grpo.aime24-abs}{0.1}
\setval{4b.concur-free.minus.grpo.aime25}{+2.0}
\setval{4b.concur-free.minus.grpo.aime25-abs}{2.0}
\setval{4b.concur-free.minus.grpo.amc23}{$-$0.5}
\setval{4b.concur-free.minus.grpo.amc23-abs}{0.5}
\setval{4b.concur-free.minus.grpo.avg}{+0.9}
\setval{4b.concur-free.minus.grpo.avg-abs}{0.9}
\setval{4b.concur-free.minus.grpo.code}{$-$1.0}
\setval{4b.concur-free.minus.grpo.code-abs}{1.0}
\setval{4b.concur-free.minus.grpo.creative-writing-v3}{$-$1.9}
\setval{4b.concur-free.minus.grpo.creative-writing-v3-abs}{1.9}
\setval{4b.concur-free.minus.grpo.guidebench}{+1.1}
\setval{4b.concur-free.minus.grpo.guidebench-abs}{1.1}
\setval{4b.concur-free.minus.grpo.humaneval-plus}{$-$1.6}
\setval{4b.concur-free.minus.grpo.humaneval-plus-abs}{1.6}
\setval{4b.concur-free.minus.grpo.if}{+0.9}
\setval{4b.concur-free.minus.grpo.if-abs}{0.9}
\setval{4b.concur-free.minus.grpo.ifbench}{+1.3}
\setval{4b.concur-free.minus.grpo.ifbench-abs}{1.3}
\setval{4b.concur-free.minus.grpo.ifeval}{+0.2}
\setval{4b.concur-free.minus.grpo.ifeval-abs}{0.2}
\setval{4b.concur-free.minus.grpo.judge2.avg}{+0.9}
\setval{4b.concur-free.minus.grpo.judge2.avg-abs}{0.9}
\setval{4b.concur-free.minus.grpo.judge2.creative-writing-v3}{+0.2}
\setval{4b.concur-free.minus.grpo.judge2.creative-writing-v3-abs}{0.2}
\setval{4b.concur-free.minus.grpo.judge2.longbench-write}{+8.6}
\setval{4b.concur-free.minus.grpo.judge2.longbench-write-abs}{8.6}
\setval{4b.concur-free.minus.grpo.judge2.writing}{+3.3}
\setval{4b.concur-free.minus.grpo.judge2.writing-abs}{3.3}
\setval{4b.concur-free.minus.grpo.judge2.writingbench}{+1.0}
\setval{4b.concur-free.minus.grpo.judge2.writingbench-abs}{1.0}
\setval{4b.concur-free.minus.grpo.livecodebench}{+0.2}
\setval{4b.concur-free.minus.grpo.livecodebench-abs}{0.2}
\setval{4b.concur-free.minus.grpo.longbench-write}{+8.6}
\setval{4b.concur-free.minus.grpo.longbench-write-abs}{8.6}
\setval{4b.concur-free.minus.grpo.math}{+0.6}
\setval{4b.concur-free.minus.grpo.math-abs}{0.6}
\setval{4b.concur-free.minus.grpo.math500}{+0.1}
\setval{4b.concur-free.minus.grpo.math500-abs}{0.1}
\setval{4b.concur-free.minus.grpo.mbpp-plus}{$-$1.5}
\setval{4b.concur-free.minus.grpo.mbpp-plus-abs}{1.5}
\setval{4b.concur-free.minus.grpo.minerva}{+1.0}
\setval{4b.concur-free.minus.grpo.minerva-abs}{1.0}
\setval{4b.concur-free.minus.grpo.olympiad-bench}{+1.1}
\setval{4b.concur-free.minus.grpo.olympiad-bench-abs}{1.1}
\setval{4b.concur-free.minus.grpo.writing}{+3.2}
\setval{4b.concur-free.minus.grpo.writing-abs}{3.2}
\setval{4b.concur-free.minus.grpo.writingbench}{+3.1}
\setval{4b.concur-free.minus.grpo.writingbench-abs}{3.1}
\setval{4b.concur-free.minus.mt-grpo.aime24}{+1.1}
\setval{4b.concur-free.minus.mt-grpo.aime24-abs}{1.1}
\setval{4b.concur-free.minus.mt-grpo.aime25}{+8.4}
\setval{4b.concur-free.minus.mt-grpo.aime25-abs}{8.4}
\setval{4b.concur-free.minus.mt-grpo.amc23}{$-$5.6}
\setval{4b.concur-free.minus.mt-grpo.amc23-abs}{5.6}
\setval{4b.concur-free.minus.mt-grpo.avg}{+2.5}
\setval{4b.concur-free.minus.mt-grpo.avg-abs}{2.5}
\setval{4b.concur-free.minus.mt-grpo.code}{+1.1}
\setval{4b.concur-free.minus.mt-grpo.code-abs}{1.1}
\setval{4b.concur-free.minus.mt-grpo.creative-writing-v3}{+0.9}
\setval{4b.concur-free.minus.mt-grpo.creative-writing-v3-abs}{0.9}
\setval{4b.concur-free.minus.mt-grpo.guidebench}{$-$0.7}
\setval{4b.concur-free.minus.mt-grpo.guidebench-abs}{0.7}
\setval{4b.concur-free.minus.mt-grpo.humaneval-plus}{+2.0}
\setval{4b.concur-free.minus.mt-grpo.humaneval-plus-abs}{2.0}
\setval{4b.concur-free.minus.mt-grpo.if}{+1.3}
\setval{4b.concur-free.minus.mt-grpo.if-abs}{1.3}
\setval{4b.concur-free.minus.mt-grpo.ifbench}{+4.7}
\setval{4b.concur-free.minus.mt-grpo.ifbench-abs}{4.7}
\setval{4b.concur-free.minus.mt-grpo.ifeval}{+0.0}
\setval{4b.concur-free.minus.mt-grpo.ifeval-abs}{0.0}
\setval{4b.concur-free.minus.mt-grpo.judge2.avg}{+1.9}
\setval{4b.concur-free.minus.mt-grpo.judge2.avg-abs}{1.9}
\setval{4b.concur-free.minus.mt-grpo.judge2.creative-writing-v3}{+1.2}
\setval{4b.concur-free.minus.mt-grpo.judge2.creative-writing-v3-abs}{1.2}
\setval{4b.concur-free.minus.mt-grpo.judge2.longbench-write}{+8.6}
\setval{4b.concur-free.minus.mt-grpo.judge2.longbench-write-abs}{8.6}
\setval{4b.concur-free.minus.mt-grpo.judge2.writing}{+3.7}
\setval{4b.concur-free.minus.mt-grpo.judge2.writing-abs}{3.7}
\setval{4b.concur-free.minus.mt-grpo.judge2.writingbench}{+1.2}
\setval{4b.concur-free.minus.mt-grpo.judge2.writingbench-abs}{1.2}
\setval{4b.concur-free.minus.mt-grpo.livecodebench}{+1.5}
\setval{4b.concur-free.minus.mt-grpo.livecodebench-abs}{1.5}
\setval{4b.concur-free.minus.mt-grpo.longbench-write}{+11.0}
\setval{4b.concur-free.minus.mt-grpo.longbench-write-abs}{11.0}
\setval{4b.concur-free.minus.mt-grpo.math}{+1.8}
\setval{4b.concur-free.minus.mt-grpo.math-abs}{1.8}
\setval{4b.concur-free.minus.mt-grpo.math500}{+3.3}
\setval{4b.concur-free.minus.mt-grpo.math500-abs}{3.3}
\setval{4b.concur-free.minus.mt-grpo.mbpp-plus}{$-$0.4}
\setval{4b.concur-free.minus.mt-grpo.mbpp-plus-abs}{0.4}
\setval{4b.concur-free.minus.mt-grpo.minerva}{+0.3}
\setval{4b.concur-free.minus.mt-grpo.minerva-abs}{0.3}
\setval{4b.concur-free.minus.mt-grpo.olympiad-bench}{+3.2}
\setval{4b.concur-free.minus.mt-grpo.olympiad-bench-abs}{3.2}
\setval{4b.concur-free.minus.mt-grpo.writing}{+6.3}
\setval{4b.concur-free.minus.mt-grpo.writing-abs}{6.3}
\setval{4b.concur-free.minus.mt-grpo.writingbench}{+7.2}
\setval{4b.concur-free.minus.mt-grpo.writingbench-abs}{7.2}
\setval{4b.concur-free.minus.papo.aime24}{+2.6}
\setval{4b.concur-free.minus.papo.aime24-abs}{2.6}
\setval{4b.concur-free.minus.papo.aime25}{+7.1}
\setval{4b.concur-free.minus.papo.aime25-abs}{7.1}
\setval{4b.concur-free.minus.papo.amc23}{+0.6}
\setval{4b.concur-free.minus.papo.amc23-abs}{0.6}
\setval{4b.concur-free.minus.papo.avg}{+2.9}
\setval{4b.concur-free.minus.papo.avg-abs}{2.9}
\setval{4b.concur-free.minus.papo.code}{$-$0.8}
\setval{4b.concur-free.minus.papo.code-abs}{0.8}
\setval{4b.concur-free.minus.papo.creative-writing-v3}{$-$1.1}
\setval{4b.concur-free.minus.papo.creative-writing-v3-abs}{1.1}
\setval{4b.concur-free.minus.papo.guidebench}{$-$1.7}
\setval{4b.concur-free.minus.papo.guidebench-abs}{1.7}
\setval{4b.concur-free.minus.papo.humaneval-plus}{$-$2.0}
\setval{4b.concur-free.minus.papo.humaneval-plus-abs}{2.0}
\setval{4b.concur-free.minus.papo.if}{+1.6}
\setval{4b.concur-free.minus.papo.if-abs}{1.6}
\setval{4b.concur-free.minus.papo.ifbench}{+5.7}
\setval{4b.concur-free.minus.papo.ifbench-abs}{5.7}
\setval{4b.concur-free.minus.papo.ifeval}{+0.8}
\setval{4b.concur-free.minus.papo.ifeval-abs}{0.8}
\setval{4b.concur-free.minus.papo.judge2.avg}{+2.4}
\setval{4b.concur-free.minus.papo.judge2.avg-abs}{2.4}
\setval{4b.concur-free.minus.papo.judge2.creative-writing-v3}{+1.1}
\setval{4b.concur-free.minus.papo.judge2.creative-writing-v3-abs}{1.1}
\setval{4b.concur-free.minus.papo.judge2.longbench-write}{+11.8}
\setval{4b.concur-free.minus.papo.judge2.longbench-write-abs}{11.8}
\setval{4b.concur-free.minus.papo.judge2.writing}{+5.1}
\setval{4b.concur-free.minus.papo.judge2.writing-abs}{5.1}
\setval{4b.concur-free.minus.papo.judge2.writingbench}{+2.3}
\setval{4b.concur-free.minus.papo.judge2.writingbench-abs}{2.3}
\setval{4b.concur-free.minus.papo.livecodebench}{+0.6}
\setval{4b.concur-free.minus.papo.livecodebench-abs}{0.6}
\setval{4b.concur-free.minus.papo.longbench-write}{+13.8}
\setval{4b.concur-free.minus.papo.longbench-write-abs}{13.8}
\setval{4b.concur-free.minus.papo.math}{+3.0}
\setval{4b.concur-free.minus.papo.math-abs}{3.0}
\setval{4b.concur-free.minus.papo.math500}{+3.8}
\setval{4b.concur-free.minus.papo.math500-abs}{3.8}
\setval{4b.concur-free.minus.papo.mbpp-plus}{$-$1.1}
\setval{4b.concur-free.minus.papo.mbpp-plus-abs}{1.1}
\setval{4b.concur-free.minus.papo.minerva}{$-$0.3}
\setval{4b.concur-free.minus.papo.minerva-abs}{0.3}
\setval{4b.concur-free.minus.papo.olympiad-bench}{+4.3}
\setval{4b.concur-free.minus.papo.olympiad-bench-abs}{4.3}
\setval{4b.concur-free.minus.papo.writing}{+7.8}
\setval{4b.concur-free.minus.papo.writing-abs}{7.8}
\setval{4b.concur-free.minus.papo.writingbench}{+10.8}
\setval{4b.concur-free.minus.papo.writingbench-abs}{10.8}
\setval{4b.concur-free.minus.saw.aime24}{$-$0.6}
\setval{4b.concur-free.minus.saw.aime24-abs}{0.6}
\setval{4b.concur-free.minus.saw.aime25}{+3.4}
\setval{4b.concur-free.minus.saw.aime25-abs}{3.4}
\setval{4b.concur-free.minus.saw.amc23}{$-$2.5}
\setval{4b.concur-free.minus.saw.amc23-abs}{2.5}
\setval{4b.concur-free.minus.saw.avg}{+1.3}
\setval{4b.concur-free.minus.saw.avg-abs}{1.3}
\setval{4b.concur-free.minus.saw.code}{+0.4}
\setval{4b.concur-free.minus.saw.code-abs}{0.4}
\setval{4b.concur-free.minus.saw.creative-writing-v3}{+0.4}
\setval{4b.concur-free.minus.saw.creative-writing-v3-abs}{0.4}
\setval{4b.concur-free.minus.saw.guidebench}{$-$2.0}
\setval{4b.concur-free.minus.saw.guidebench-abs}{2.0}
\setval{4b.concur-free.minus.saw.humaneval-plus}{+0.4}
\setval{4b.concur-free.minus.saw.humaneval-plus-abs}{0.4}
\setval{4b.concur-free.minus.saw.if}{$-$1.2}
\setval{4b.concur-free.minus.saw.if-abs}{1.2}
\setval{4b.concur-free.minus.saw.ifbench}{$-$1.3}
\setval{4b.concur-free.minus.saw.ifbench-abs}{1.3}
\setval{4b.concur-free.minus.saw.ifeval}{$-$0.2}
\setval{4b.concur-free.minus.saw.ifeval-abs}{0.2}
\setval{4b.concur-free.minus.saw.judge2.avg}{+0.7}
\setval{4b.concur-free.minus.saw.judge2.avg-abs}{0.7}
\setval{4b.concur-free.minus.saw.judge2.creative-writing-v3}{$-$0.3}
\setval{4b.concur-free.minus.saw.judge2.creative-writing-v3-abs}{0.3}
\setval{4b.concur-free.minus.saw.judge2.longbench-write}{+9.3}
\setval{4b.concur-free.minus.saw.judge2.longbench-write-abs}{9.3}
\setval{4b.concur-free.minus.saw.judge2.writing}{+3.4}
\setval{4b.concur-free.minus.saw.judge2.writing-abs}{3.4}
\setval{4b.concur-free.minus.saw.judge2.writingbench}{+1.1}
\setval{4b.concur-free.minus.saw.judge2.writingbench-abs}{1.1}
\setval{4b.concur-free.minus.saw.livecodebench}{+1.4}
\setval{4b.concur-free.minus.saw.livecodebench-abs}{1.4}
\setval{4b.concur-free.minus.saw.longbench-write}{+11.6}
\setval{4b.concur-free.minus.saw.longbench-write-abs}{11.6}
\setval{4b.concur-free.minus.saw.math}{+0.5}
\setval{4b.concur-free.minus.saw.math-abs}{0.5}
\setval{4b.concur-free.minus.saw.math500}{+2.5}
\setval{4b.concur-free.minus.saw.math500-abs}{2.5}
\setval{4b.concur-free.minus.saw.mbpp-plus}{$-$0.4}
\setval{4b.concur-free.minus.saw.mbpp-plus-abs}{0.4}
\setval{4b.concur-free.minus.saw.minerva}{$-$1.9}
\setval{4b.concur-free.minus.saw.minerva-abs}{1.9}
\setval{4b.concur-free.minus.saw.olympiad-bench}{+1.9}
\setval{4b.concur-free.minus.saw.olympiad-bench-abs}{1.9}
\setval{4b.concur-free.minus.saw.writing}{+6.5}
\setval{4b.concur-free.minus.saw.writing-abs}{6.5}
\setval{4b.concur-free.minus.saw.writingbench}{+7.4}
\setval{4b.concur-free.minus.saw.writingbench-abs}{7.4}
\setval{4b.concur-free.passing-at-ceiling.code.early}{0.71}
\setval{4b.concur-free.passing-at-ceiling.code.late}{0.74}
\setval{4b.concur-free.passing-at-ceiling.code.run}{0.73}
\setval{4b.concur-free.passing-at-ceiling.if.early}{0.28}
\setval{4b.concur-free.passing-at-ceiling.if.late}{0.56}
\setval{4b.concur-free.passing-at-ceiling.if.run}{0.48}
\setval{4b.concur-free.passing-at-ceiling.math.early}{1.00}
\setval{4b.concur-free.passing-at-ceiling.math.late}{1.00}
\setval{4b.concur-free.passing-at-ceiling.math.run}{1.00}
\setval{4b.concur-free.passing-at-ceiling.writing.early}{0.00}
\setval{4b.concur-free.passing-at-ceiling.writing.late}{0.01}
\setval{4b.concur-free.passing-at-ceiling.writing.run}{0.01}
\setval{4b.concur-free.primary-cosine.code.early}{0.97}
\setval{4b.concur-free.primary-cosine.code.late}{0.96}
\setval{4b.concur-free.primary-cosine.code.run}{0.96}
\setval{4b.concur-free.primary-cosine.if.early}{0.85}
\setval{4b.concur-free.primary-cosine.if.late}{0.79}
\setval{4b.concur-free.primary-cosine.if.run}{0.82}
\setval{4b.concur-free.primary-cosine.math.early}{0.97}
\setval{4b.concur-free.primary-cosine.math.late}{0.97}
\setval{4b.concur-free.primary-cosine.math.run}{0.97}
\setval{4b.concur-free.primary-cosine.writing.early}{0.94}
\setval{4b.concur-free.primary-cosine.writing.late}{0.92}
\setval{4b.concur-free.primary-cosine.writing.run}{0.92}
\setval{4b.concur-free.primary-level.code.early}{0.53}
\setval{4b.concur-free.primary-level.code.late}{0.73}
\setval{4b.concur-free.primary-level.code.run}{0.69}
\setval{4b.concur-free.primary-level.if.early}{0.32}
\setval{4b.concur-free.primary-level.if.late}{0.75}
\setval{4b.concur-free.primary-level.if.run}{0.63}
\setval{4b.concur-free.primary-level.math.early}{0.48}
\setval{4b.concur-free.primary-level.math.late}{0.68}
\setval{4b.concur-free.primary-level.math.run}{0.62}
\setval{4b.concur-free.primary-level.writing.early}{0.54}
\setval{4b.concur-free.primary-level.writing.late}{0.79}
\setval{4b.concur-free.primary-level.writing.run}{0.75}
\setval{4b.concur-free.solved.code.early}{0.06}
\setval{4b.concur-free.solved.code.late}{0.01}
\setval{4b.concur-free.solved.code.run}{0.03}
\setval{4b.concur-free.solved.if.early}{0.03}
\setval{4b.concur-free.solved.if.late}{0.20}
\setval{4b.concur-free.solved.if.run}{0.13}
\setval{4b.concur-free.solved.math.early}{0.12}
\setval{4b.concur-free.solved.math.late}{0.11}
\setval{4b.concur-free.solved.math.run}{0.13}
\setval{4b.concur-free.solved.writing.early}{0.00}
\setval{4b.concur-free.solved.writing.late}{0.00}
\setval{4b.concur-free.solved.writing.run}{0.00}
\setval{4b.concur-free.trust.code.agreement}{+0.003}
\setval{4b.concur-free.trust.code.errors}{+1.4}
\setval{4b.concur-free.trust.code.first-free}{40}
\setval{4b.concur-free.trust.code.free-share}{0.31}
\setval{4b.concur-free.trust.if.agreement}{+0.017}
\setval{4b.concur-free.trust.if.errors}{+7.0}
\setval{4b.concur-free.trust.if.first-free}{30}
\setval{4b.concur-free.trust.if.free-share}{0.90}
\setval{4b.concur-free.trust.math.first-free}{--}
\setval{4b.concur-free.trust.math.free-share}{0.00}
\setval{4b.concur-free.trust.writing-rm.agreement}{+0.059}
\setval{4b.concur-free.trust.writing-rm.errors}{+36.2}
\setval{4b.concur-free.trust.writing-rm.first-free}{1}
\setval{4b.concur-free.trust.writing-rm.free-share}{1.00}
\setval{4b.concur-free.trust.writing.agreement}{$-$0.051}
\setval{4b.concur-free.trust.writing.errors}{$-$35.1}
\setval{4b.concur-free.trust.writing.first-free}{--}
\setval{4b.concur-free.trust.writing.free-share}{0.00}
\setval{4b.concur-free.update-share.code.early}{0.20}
\setval{4b.concur-free.update-share.code.late}{0.21}
\setval{4b.concur-free.update-share.code.run}{0.22}
\setval{4b.concur-free.update-share.if.early}{0.21}
\setval{4b.concur-free.update-share.if.late}{0.17}
\setval{4b.concur-free.update-share.if.run}{0.20}
\setval{4b.concur-free.update-share.math.early}{0.18}
\setval{4b.concur-free.update-share.math.late}{0.25}
\setval{4b.concur-free.update-share.math.run}{0.24}
\setval{4b.concur-free.update-share.writing.early}{0.41}
\setval{4b.concur-free.update-share.writing.late}{0.37}
\setval{4b.concur-free.update-share.writing.run}{0.34}
\setval{4b.concur-tied.aux-change.code}{+0.18}
\setval{4b.concur-tied.aux-change.if}{+0.22}
\setval{4b.concur-tied.aux-change.math}{$-$0.13}
\setval{4b.concur-tied.aux-change.writing}{$-$0.17}
\setval{4b.concur-tied.aux-change.writing-rm}{+0.35}
\setval{4b.concur-tied.aux-full.code.early}{0.60}
\setval{4b.concur-tied.aux-full.code.late}{0.87}
\setval{4b.concur-tied.aux-full.code.run}{0.83}
\setval{4b.concur-tied.aux-full.if.early}{0.63}
\setval{4b.concur-tied.aux-full.if.late}{0.86}
\setval{4b.concur-tied.aux-full.if.run}{0.80}
\setval{4b.concur-tied.aux-full.math.early}{0.61}
\setval{4b.concur-tied.aux-full.math.late}{0.26}
\setval{4b.concur-tied.aux-full.math.run}{0.44}
\setval{4b.concur-tied.aux-full.writing-rm.early}{0.10}
\setval{4b.concur-tied.aux-full.writing-rm.late}{0.50}
\setval{4b.concur-tied.aux-full.writing-rm.run}{0.42}
\setval{4b.concur-tied.aux-full.writing.early}{0.56}
\setval{4b.concur-tied.aux-full.writing.late}{0.26}
\setval{4b.concur-tied.aux-full.writing.run}{0.26}
\setval{4b.concur-tied.aux-level.code.early}{0.88}
\setval{4b.concur-tied.aux-level.code.late}{0.97}
\setval{4b.concur-tied.aux-level.code.run}{0.96}
\setval{4b.concur-tied.aux-level.if.early}{0.81}
\setval{4b.concur-tied.aux-level.if.late}{0.92}
\setval{4b.concur-tied.aux-level.if.run}{0.89}
\setval{4b.concur-tied.aux-level.math.early}{0.92}
\setval{4b.concur-tied.aux-level.math.late}{0.72}
\setval{4b.concur-tied.aux-level.math.run}{0.83}
\setval{4b.concur-tied.aux-level.writing-rm.early}{0.78}
\setval{4b.concur-tied.aux-level.writing-rm.late}{0.95}
\setval{4b.concur-tied.aux-level.writing-rm.run}{0.92}
\setval{4b.concur-tied.aux-level.writing.early}{0.86}
\setval{4b.concur-tied.aux-level.writing.late}{0.60}
\setval{4b.concur-tied.aux-level.writing.run}{0.56}
\setval{4b.concur-tied.bench.aime24}{19.5}
\setval{4b.concur-tied.bench.aime25}{16.4}
\setval{4b.concur-tied.bench.amc23}{69.2}
\setval{4b.concur-tied.bench.creative-writing-v3}{47.8}
\setval{4b.concur-tied.bench.guidebench}{79.2}
\setval{4b.concur-tied.bench.humaneval-plus}{84.0}
\setval{4b.concur-tied.bench.ifbench}{37.7}
\setval{4b.concur-tied.bench.ifeval}{87.8}
\setval{4b.concur-tied.bench.livecodebench}{24.7}
\setval{4b.concur-tied.bench.longbench-write}{77.7}
\setval{4b.concur-tied.bench.math500}{85.5}
\setval{4b.concur-tied.bench.mbpp-plus}{68.6}
\setval{4b.concur-tied.bench.minerva}{49.1}
\setval{4b.concur-tied.bench.olympiad-bench}{52.9}
\setval{4b.concur-tied.bench.writingbench}{80.2}
\setval{4b.concur-tied.best-count.all}{10}
\setval{4b.concur-tied.best-count.code}{1}
\setval{4b.concur-tied.best-count.if}{0}
\setval{4b.concur-tied.best-count.math}{6}
\setval{4b.concur-tied.best-count.writing}{3}
\setval{4b.concur-tied.eval.avg}{58.7}
\setval{4b.concur-tied.eval.code}{59.1}
\setval{4b.concur-tied.eval.if}{68.2}
\setval{4b.concur-tied.eval.math}{48.8}
\setval{4b.concur-tied.eval.writing}{68.6}
\setval{4b.concur-tied.gate.code.against.early}{0.04}
\setval{4b.concur-tied.gate.code.against.late}{0.05}
\setval{4b.concur-tied.gate.code.against.run}{0.04}
\setval{4b.concur-tied.gate.code.agreement}{+0.007}
\setval{4b.concur-tied.gate.code.ceiling.early}{0.14}
\setval{4b.concur-tied.gate.code.ceiling.late}{0.14}
\setval{4b.concur-tied.gate.code.ceiling.run}{0.15}
\setval{4b.concur-tied.gate.code.cosine.early}{0.81}
\setval{4b.concur-tied.gate.code.cosine.late}{0.70}
\setval{4b.concur-tied.gate.code.cosine.run}{0.74}
\setval{4b.concur-tied.gate.if.against.early}{0.05}
\setval{4b.concur-tied.gate.if.against.late}{0.04}
\setval{4b.concur-tied.gate.if.against.run}{0.05}
\setval{4b.concur-tied.gate.if.agreement}{+0.018}
\setval{4b.concur-tied.gate.if.ceiling.early}{0.06}
\setval{4b.concur-tied.gate.if.ceiling.late}{0.24}
\setval{4b.concur-tied.gate.if.ceiling.run}{0.16}
\setval{4b.concur-tied.gate.if.cosine.early}{0.72}
\setval{4b.concur-tied.gate.if.cosine.late}{0.48}
\setval{4b.concur-tied.gate.if.cosine.run}{0.56}
\setval{4b.concur-tied.gate.math.against.early}{0.00}
\setval{4b.concur-tied.gate.math.against.late}{0.00}
\setval{4b.concur-tied.gate.math.against.run}{0.00}
\setval{4b.concur-tied.gate.math.ceiling.early}{0.21}
\setval{4b.concur-tied.gate.math.ceiling.late}{0.70}
\setval{4b.concur-tied.gate.math.ceiling.run}{0.52}
\setval{4b.concur-tied.gate.math.cosine.early}{0.87}
\setval{4b.concur-tied.gate.math.cosine.late}{0.46}
\setval{4b.concur-tied.gate.math.cosine.run}{0.64}
\setval{4b.concur-tied.gate.writing-rm.against.early}{0.21}
\setval{4b.concur-tied.gate.writing-rm.against.late}{0.55}
\setval{4b.concur-tied.gate.writing-rm.against.run}{0.51}
\setval{4b.concur-tied.gate.writing-rm.agreement}{+0.058}
\setval{4b.concur-tied.gate.writing-rm.ceiling.early}{0.00}
\setval{4b.concur-tied.gate.writing-rm.ceiling.late}{0.00}
\setval{4b.concur-tied.gate.writing-rm.ceiling.run}{0.00}
\setval{4b.concur-tied.gate.writing-rm.cosine.early}{0.67}
\setval{4b.concur-tied.gate.writing-rm.cosine.late}{0.11}
\setval{4b.concur-tied.gate.writing-rm.cosine.run}{0.19}
\setval{4b.concur-tied.gate.writing.against.early}{0.35}
\setval{4b.concur-tied.gate.writing.against.late}{0.67}
\setval{4b.concur-tied.gate.writing.against.run}{0.64}
\setval{4b.concur-tied.gate.writing.agreement}{$-$0.048}
\setval{4b.concur-tied.gate.writing.ceiling.early}{0.00}
\setval{4b.concur-tied.gate.writing.ceiling.late}{0.01}
\setval{4b.concur-tied.gate.writing.ceiling.run}{0.00}
\setval{4b.concur-tied.gate.writing.cosine.early}{0.46}
\setval{4b.concur-tied.gate.writing.cosine.late}{-0.04}
\setval{4b.concur-tied.gate.writing.cosine.run}{0.02}
\setval{4b.concur-tied.judge2.avg}{55.5}
\setval{4b.concur-tied.judge2.creative-writing-v3}{34.0}
\setval{4b.concur-tied.judge2.longbench-write}{68.0}
\setval{4b.concur-tied.judge2.writing}{52.4}
\setval{4b.concur-tied.judge2.writingbench}{55.3}
\setval{4b.concur-tied.last-step}{300}
\setval{4b.concur-tied.minus.base.aime24}{+11.7}
\setval{4b.concur-tied.minus.base.aime24-abs}{11.7}
\setval{4b.concur-tied.minus.base.aime25}{+9.3}
\setval{4b.concur-tied.minus.base.aime25-abs}{9.3}
\setval{4b.concur-tied.minus.base.amc23}{+25.3}
\setval{4b.concur-tied.minus.base.amc23-abs}{25.3}
\setval{4b.concur-tied.minus.base.avg}{+24.2}
\setval{4b.concur-tied.minus.base.avg-abs}{24.2}
\setval{4b.concur-tied.minus.base.code}{+14.5}
\setval{4b.concur-tied.minus.base.code-abs}{14.5}
\setval{4b.concur-tied.minus.base.creative-writing-v3}{+23.1}
\setval{4b.concur-tied.minus.base.creative-writing-v3-abs}{23.1}
\setval{4b.concur-tied.minus.base.guidebench}{+54.3}
\setval{4b.concur-tied.minus.base.guidebench-abs}{54.3}
\setval{4b.concur-tied.minus.base.humaneval-plus}{+19.0}
\setval{4b.concur-tied.minus.base.humaneval-plus-abs}{19.0}
\setval{4b.concur-tied.minus.base.if}{+38.5}
\setval{4b.concur-tied.minus.base.if-abs}{38.5}
\setval{4b.concur-tied.minus.base.ifbench}{+20.7}
\setval{4b.concur-tied.minus.base.ifbench-abs}{20.7}
\setval{4b.concur-tied.minus.base.ifeval}{+40.4}
\setval{4b.concur-tied.minus.base.ifeval-abs}{40.4}
\setval{4b.concur-tied.minus.base.judge2.avg}{+21.5}
\setval{4b.concur-tied.minus.base.judge2.avg-abs}{21.5}
\setval{4b.concur-tied.minus.base.judge2.creative-writing-v3}{+13.1}
\setval{4b.concur-tied.minus.base.judge2.creative-writing-v3-abs}{13.1}
\setval{4b.concur-tied.minus.base.judge2.longbench-write}{+27.0}
\setval{4b.concur-tied.minus.base.judge2.longbench-write-abs}{27.0}
\setval{4b.concur-tied.minus.base.judge2.writing}{+21.7}
\setval{4b.concur-tied.minus.base.judge2.writing-abs}{21.7}
\setval{4b.concur-tied.minus.base.judge2.writingbench}{+24.9}
\setval{4b.concur-tied.minus.base.judge2.writingbench-abs}{24.9}
\setval{4b.concur-tied.minus.base.livecodebench}{+11.5}
\setval{4b.concur-tied.minus.base.livecodebench-abs}{11.5}
\setval{4b.concur-tied.minus.base.longbench-write}{+33.3}
\setval{4b.concur-tied.minus.base.longbench-write-abs}{33.3}
\setval{4b.concur-tied.minus.base.math}{+16.5}
\setval{4b.concur-tied.minus.base.math-abs}{16.5}
\setval{4b.concur-tied.minus.base.math500}{+18.3}
\setval{4b.concur-tied.minus.base.math500-abs}{18.3}
\setval{4b.concur-tied.minus.base.mbpp-plus}{+13.0}
\setval{4b.concur-tied.minus.base.mbpp-plus-abs}{13.0}
\setval{4b.concur-tied.minus.base.minerva}{+17.5}
\setval{4b.concur-tied.minus.base.minerva-abs}{17.5}
\setval{4b.concur-tied.minus.base.olympiad-bench}{+17.2}
\setval{4b.concur-tied.minus.base.olympiad-bench-abs}{17.2}
\setval{4b.concur-tied.minus.base.writing}{+34.7}
\setval{4b.concur-tied.minus.base.writing-abs}{34.7}
\setval{4b.concur-tied.minus.base.writingbench}{+47.8}
\setval{4b.concur-tied.minus.base.writingbench-abs}{47.8}
\setval{4b.concur-tied.minus.best.bench.aime24}{+7.0}
\setval{4b.concur-tied.minus.best.bench.aime25}{+3.8}
\setval{4b.concur-tied.minus.best.bench.amc23}{+9.5}
\setval{4b.concur-tied.minus.best.bench.creative-writing-v3}{+0.5}
\setval{4b.concur-tied.minus.best.bench.guidebench}{$-$1.8}
\setval{4b.concur-tied.minus.best.bench.humaneval-plus}{+0.6}
\setval{4b.concur-tied.minus.best.bench.ifbench}{$-$2.7}
\setval{4b.concur-tied.minus.best.bench.ifeval}{+0.0}
\setval{4b.concur-tied.minus.best.bench.livecodebench}{$-$0.9}
\setval{4b.concur-tied.minus.best.bench.longbench-write}{+8.5}
\setval{4b.concur-tied.minus.best.bench.math500}{+5.3}
\setval{4b.concur-tied.minus.best.bench.mbpp-plus}{$-$2.9}
\setval{4b.concur-tied.minus.best.bench.minerva}{+1.0}
\setval{4b.concur-tied.minus.best.bench.olympiad-bench}{+8.2}
\setval{4b.concur-tied.minus.best.bench.writingbench}{+5.9}
\setval{4b.concur-tied.minus.best.eval.avg}{+4.0}
\setval{4b.concur-tied.minus.best.eval.code}{$-$0.3}
\setval{4b.concur-tied.minus.best.eval.if}{$-$1.2}
\setval{4b.concur-tied.minus.best.eval.math}{+7.1}
\setval{4b.concur-tied.minus.best.eval.writing}{+5.0}
\setval{4b.concur-tied.minus.concur-free.aime24}{+7.6}
\setval{4b.concur-tied.minus.concur-free.aime24-abs}{7.6}
\setval{4b.concur-tied.minus.concur-free.aime25}{+1.8}
\setval{4b.concur-tied.minus.concur-free.aime25-abs}{1.8}
\setval{4b.concur-tied.minus.concur-free.amc23}{+15.1}
\setval{4b.concur-tied.minus.concur-free.amc23-abs}{15.1}
\setval{4b.concur-tied.minus.concur-free.avg}{+3.2}
\setval{4b.concur-tied.minus.concur-free.avg-abs}{3.2}
\setval{4b.concur-tied.minus.concur-free.code}{+0.8}
\setval{4b.concur-tied.minus.concur-free.code-abs}{0.8}
\setval{4b.concur-tied.minus.concur-free.creative-writing-v3}{+2.4}
\setval{4b.concur-tied.minus.concur-free.creative-writing-v3-abs}{2.4}
\setval{4b.concur-tied.minus.concur-free.guidebench}{+0.2}
\setval{4b.concur-tied.minus.concur-free.guidebench-abs}{0.2}
\setval{4b.concur-tied.minus.concur-free.humaneval-plus}{+2.9}
\setval{4b.concur-tied.minus.concur-free.humaneval-plus-abs}{2.9}
\setval{4b.concur-tied.minus.concur-free.if}{+0.0}
\setval{4b.concur-tied.minus.concur-free.if-abs}{0.0}
\setval{4b.concur-tied.minus.concur-free.ifbench}{$-$1.3}
\setval{4b.concur-tied.minus.concur-free.ifbench-abs}{1.3}
\setval{4b.concur-tied.minus.concur-free.ifeval}{+1.1}
\setval{4b.concur-tied.minus.concur-free.ifeval-abs}{1.1}
\setval{4b.concur-tied.minus.concur-free.judge2.avg}{+2.7}
\setval{4b.concur-tied.minus.concur-free.judge2.avg-abs}{2.7}
\setval{4b.concur-tied.minus.concur-free.judge2.creative-writing-v3}{$-$1.3}
\setval{4b.concur-tied.minus.concur-free.judge2.creative-writing-v3-abs}{1.3}
\setval{4b.concur-tied.minus.concur-free.judge2.longbench-write}{$-$1.4}
\setval{4b.concur-tied.minus.concur-free.judge2.longbench-write-abs}{1.4}
\setval{4b.concur-tied.minus.concur-free.judge2.writing}{$-$0.6}
\setval{4b.concur-tied.minus.concur-free.judge2.writing-abs}{0.6}
\setval{4b.concur-tied.minus.concur-free.judge2.writingbench}{+0.9}
\setval{4b.concur-tied.minus.concur-free.judge2.writingbench-abs}{0.9}
\setval{4b.concur-tied.minus.concur-free.livecodebench}{+0.6}
\setval{4b.concur-tied.minus.concur-free.livecodebench-abs}{0.6}
\setval{4b.concur-tied.minus.concur-free.longbench-write}{+0.0}
\setval{4b.concur-tied.minus.concur-free.longbench-write-abs}{0.0}
\setval{4b.concur-tied.minus.concur-free.math}{+6.6}
\setval{4b.concur-tied.minus.concur-free.math-abs}{6.6}
\setval{4b.concur-tied.minus.concur-free.math500}{+5.2}
\setval{4b.concur-tied.minus.concur-free.math500-abs}{5.2}
\setval{4b.concur-tied.minus.concur-free.mbpp-plus}{$-$1.1}
\setval{4b.concur-tied.minus.concur-free.mbpp-plus-abs}{1.1}
\setval{4b.concur-tied.minus.concur-free.minerva}{+2.9}
\setval{4b.concur-tied.minus.concur-free.minerva-abs}{2.9}
\setval{4b.concur-tied.minus.concur-free.olympiad-bench}{+7.1}
\setval{4b.concur-tied.minus.concur-free.olympiad-bench-abs}{7.1}
\setval{4b.concur-tied.minus.concur-free.writing}{+1.8}
\setval{4b.concur-tied.minus.concur-free.writing-abs}{1.8}
\setval{4b.concur-tied.minus.concur-free.writingbench}{+2.9}
\setval{4b.concur-tied.minus.concur-free.writingbench-abs}{2.9}
\setval{4b.concur-tied.minus.concur.aime24}{$-$1.0}
\setval{4b.concur-tied.minus.concur.aime24-abs}{1.0}
\setval{4b.concur-tied.minus.concur.aime25}{$-$2.4}
\setval{4b.concur-tied.minus.concur.aime25-abs}{2.4}
\setval{4b.concur-tied.minus.concur.amc23}{$-$0.2}
\setval{4b.concur-tied.minus.concur.amc23-abs}{0.2}
\setval{4b.concur-tied.minus.concur.avg}{$-$0.1}
\setval{4b.concur-tied.minus.concur.avg-abs}{0.1}
\setval{4b.concur-tied.minus.concur.code}{$-$0.6}
\setval{4b.concur-tied.minus.concur.code-abs}{0.6}
\setval{4b.concur-tied.minus.concur.creative-writing-v3}{+3.1}
\setval{4b.concur-tied.minus.concur.creative-writing-v3-abs}{3.1}
\setval{4b.concur-tied.minus.concur.guidebench}{+1.8}
\setval{4b.concur-tied.minus.concur.guidebench-abs}{1.8}
\setval{4b.concur-tied.minus.concur.humaneval-plus}{+1.6}
\setval{4b.concur-tied.minus.concur.humaneval-plus-abs}{1.6}
\setval{4b.concur-tied.minus.concur.if}{$-$0.4}
\setval{4b.concur-tied.minus.concur.if-abs}{0.4}
\setval{4b.concur-tied.minus.concur.ifbench}{$-$2.7}
\setval{4b.concur-tied.minus.concur.ifbench-abs}{2.7}
\setval{4b.concur-tied.minus.concur.ifeval}{$-$0.4}
\setval{4b.concur-tied.minus.concur.ifeval-abs}{0.4}
\setval{4b.concur-tied.minus.concur.judge2.avg}{$-$0.3}
\setval{4b.concur-tied.minus.concur.judge2.avg-abs}{0.3}
\setval{4b.concur-tied.minus.concur.judge2.creative-writing-v3}{+1.6}
\setval{4b.concur-tied.minus.concur.judge2.creative-writing-v3-abs}{1.6}
\setval{4b.concur-tied.minus.concur.judge2.longbench-write}{+1.3}
\setval{4b.concur-tied.minus.concur.judge2.longbench-write-abs}{1.3}
\setval{4b.concur-tied.minus.concur.judge2.writing}{+0.7}
\setval{4b.concur-tied.minus.concur.judge2.writing-abs}{0.7}
\setval{4b.concur-tied.minus.concur.judge2.writingbench}{$-$0.7}
\setval{4b.concur-tied.minus.concur.judge2.writingbench-abs}{0.7}
\setval{4b.concur-tied.minus.concur.livecodebench}{$-$0.4}
\setval{4b.concur-tied.minus.concur.livecodebench-abs}{0.4}
\setval{4b.concur-tied.minus.concur.longbench-write}{+1.4}
\setval{4b.concur-tied.minus.concur.longbench-write-abs}{1.4}
\setval{4b.concur-tied.minus.concur.math}{$-$0.7}
\setval{4b.concur-tied.minus.concur.math-abs}{0.7}
\setval{4b.concur-tied.minus.concur.math500}{+0.2}
\setval{4b.concur-tied.minus.concur.math500-abs}{0.2}
\setval{4b.concur-tied.minus.concur.mbpp-plus}{$-$2.9}
\setval{4b.concur-tied.minus.concur.mbpp-plus-abs}{2.9}
\setval{4b.concur-tied.minus.concur.minerva}{$-$0.1}
\setval{4b.concur-tied.minus.concur.minerva-abs}{0.1}
\setval{4b.concur-tied.minus.concur.olympiad-bench}{$-$0.7}
\setval{4b.concur-tied.minus.concur.olympiad-bench-abs}{0.7}
\setval{4b.concur-tied.minus.concur.writing}{+1.8}
\setval{4b.concur-tied.minus.concur.writing-abs}{1.8}
\setval{4b.concur-tied.minus.concur.writingbench}{+0.8}
\setval{4b.concur-tied.minus.concur.writingbench-abs}{0.8}
\setval{4b.concur-tied.minus.gd2po.aime24}{+7.6}
\setval{4b.concur-tied.minus.gd2po.aime24-abs}{7.6}
\setval{4b.concur-tied.minus.gd2po.aime25}{+9.6}
\setval{4b.concur-tied.minus.gd2po.aime25-abs}{9.6}
\setval{4b.concur-tied.minus.gd2po.amc23}{+11.5}
\setval{4b.concur-tied.minus.gd2po.amc23-abs}{11.5}
\setval{4b.concur-tied.minus.gd2po.avg}{+4.6}
\setval{4b.concur-tied.minus.gd2po.avg-abs}{4.6}
\setval{4b.concur-tied.minus.gd2po.code}{$-$0.3}
\setval{4b.concur-tied.minus.gd2po.code-abs}{0.3}
\setval{4b.concur-tied.minus.gd2po.creative-writing-v3}{+3.2}
\setval{4b.concur-tied.minus.gd2po.creative-writing-v3-abs}{3.2}
\setval{4b.concur-tied.minus.gd2po.guidebench}{+0.1}
\setval{4b.concur-tied.minus.gd2po.guidebench-abs}{0.1}
\setval{4b.concur-tied.minus.gd2po.humaneval-plus}{+0.6}
\setval{4b.concur-tied.minus.gd2po.humaneval-plus-abs}{0.6}
\setval{4b.concur-tied.minus.gd2po.if}{+0.1}
\setval{4b.concur-tied.minus.gd2po.if-abs}{0.1}
\setval{4b.concur-tied.minus.gd2po.ifbench}{+0.0}
\setval{4b.concur-tied.minus.gd2po.ifbench-abs}{0.0}
\setval{4b.concur-tied.minus.gd2po.ifeval}{+0.1}
\setval{4b.concur-tied.minus.gd2po.ifeval-abs}{0.1}
\setval{4b.concur-tied.minus.gd2po.judge2.avg}{+3.9}
\setval{4b.concur-tied.minus.gd2po.judge2.avg-abs}{3.9}
\setval{4b.concur-tied.minus.gd2po.judge2.creative-writing-v3}{+1.1}
\setval{4b.concur-tied.minus.gd2po.judge2.creative-writing-v3-abs}{1.1}
\setval{4b.concur-tied.minus.gd2po.judge2.longbench-write}{+7.0}
\setval{4b.concur-tied.minus.gd2po.judge2.longbench-write-abs}{7.0}
\setval{4b.concur-tied.minus.gd2po.judge2.writing}{+3.6}
\setval{4b.concur-tied.minus.gd2po.judge2.writing-abs}{3.6}
\setval{4b.concur-tied.minus.gd2po.judge2.writingbench}{+2.8}
\setval{4b.concur-tied.minus.gd2po.judge2.writingbench-abs}{2.8}
\setval{4b.concur-tied.minus.gd2po.livecodebench}{+1.3}
\setval{4b.concur-tied.minus.gd2po.livecodebench-abs}{1.3}
\setval{4b.concur-tied.minus.gd2po.longbench-write}{+8.8}
\setval{4b.concur-tied.minus.gd2po.longbench-write-abs}{8.8}
\setval{4b.concur-tied.minus.gd2po.math}{+8.0}
\setval{4b.concur-tied.minus.gd2po.math-abs}{8.0}
\setval{4b.concur-tied.minus.gd2po.math500}{+7.8}
\setval{4b.concur-tied.minus.gd2po.math500-abs}{7.8}
\setval{4b.concur-tied.minus.gd2po.mbpp-plus}{$-$2.9}
\setval{4b.concur-tied.minus.gd2po.mbpp-plus-abs}{2.9}
\setval{4b.concur-tied.minus.gd2po.minerva}{+1.9}
\setval{4b.concur-tied.minus.gd2po.minerva-abs}{1.9}
\setval{4b.concur-tied.minus.gd2po.olympiad-bench}{+9.3}
\setval{4b.concur-tied.minus.gd2po.olympiad-bench-abs}{9.3}
\setval{4b.concur-tied.minus.gd2po.writing}{+7.3}
\setval{4b.concur-tied.minus.gd2po.writing-abs}{7.3}
\setval{4b.concur-tied.minus.gd2po.writingbench}{+10.0}
\setval{4b.concur-tied.minus.gd2po.writingbench-abs}{10.0}
\setval{4b.concur-tied.minus.gdpo-noaux.aime24}{$-$4.0}
\setval{4b.concur-tied.minus.gdpo-noaux.aime24-abs}{4.0}
\setval{4b.concur-tied.minus.gdpo-noaux.aime25}{$-$2.4}
\setval{4b.concur-tied.minus.gdpo-noaux.aime25-abs}{2.4}
\setval{4b.concur-tied.minus.gdpo-noaux.amc23}{+3.5}
\setval{4b.concur-tied.minus.gdpo-noaux.amc23-abs}{3.5}
\setval{4b.concur-tied.minus.gdpo-noaux.avg}{+2.0}
\setval{4b.concur-tied.minus.gdpo-noaux.avg-abs}{2.0}
\setval{4b.concur-tied.minus.gdpo-noaux.code}{$-$1.7}
\setval{4b.concur-tied.minus.gdpo-noaux.code-abs}{1.7}
\setval{4b.concur-tied.minus.gdpo-noaux.creative-writing-v3}{+8.5}
\setval{4b.concur-tied.minus.gdpo-noaux.creative-writing-v3-abs}{8.5}
\setval{4b.concur-tied.minus.gdpo-noaux.guidebench}{+3.3}
\setval{4b.concur-tied.minus.gdpo-noaux.guidebench-abs}{3.3}
\setval{4b.concur-tied.minus.gdpo-noaux.humaneval-plus}{$-$1.5}
\setval{4b.concur-tied.minus.gdpo-noaux.humaneval-plus-abs}{1.5}
\setval{4b.concur-tied.minus.gdpo-noaux.if}{+1.2}
\setval{4b.concur-tied.minus.gdpo-noaux.if-abs}{1.2}
\setval{4b.concur-tied.minus.gdpo-noaux.ifbench}{$-$2.7}
\setval{4b.concur-tied.minus.gdpo-noaux.ifbench-abs}{2.7}
\setval{4b.concur-tied.minus.gdpo-noaux.ifeval}{+3.0}
\setval{4b.concur-tied.minus.gdpo-noaux.ifeval-abs}{3.0}
\setval{4b.concur-tied.minus.gdpo-noaux.judge2.avg}{+2.2}
\setval{4b.concur-tied.minus.gdpo-noaux.judge2.avg-abs}{2.2}
\setval{4b.concur-tied.minus.gdpo-noaux.judge2.creative-writing-v3}{+10.6}
\setval{4b.concur-tied.minus.gdpo-noaux.judge2.creative-writing-v3-abs}{10.6}
\setval{4b.concur-tied.minus.gdpo-noaux.judge2.longbench-write}{+17.2}
\setval{4b.concur-tied.minus.gdpo-noaux.judge2.longbench-write-abs}{17.2}
\setval{4b.concur-tied.minus.gdpo-noaux.judge2.writing}{+12.4}
\setval{4b.concur-tied.minus.gdpo-noaux.judge2.writing-abs}{12.4}
\setval{4b.concur-tied.minus.gdpo-noaux.judge2.writingbench}{+9.3}
\setval{4b.concur-tied.minus.gdpo-noaux.judge2.writingbench-abs}{9.3}
\setval{4b.concur-tied.minus.gdpo-noaux.livecodebench}{+0.0}
\setval{4b.concur-tied.minus.gdpo-noaux.livecodebench-abs}{0.0}
\setval{4b.concur-tied.minus.gdpo-noaux.longbench-write}{+13.4}
\setval{4b.concur-tied.minus.gdpo-noaux.longbench-write-abs}{13.4}
\setval{4b.concur-tied.minus.gdpo-noaux.math}{$-$0.5}
\setval{4b.concur-tied.minus.gdpo-noaux.math-abs}{0.5}
\setval{4b.concur-tied.minus.gdpo-noaux.math500}{+0.8}
\setval{4b.concur-tied.minus.gdpo-noaux.math500-abs}{0.8}
\setval{4b.concur-tied.minus.gdpo-noaux.mbpp-plus}{$-$3.7}
\setval{4b.concur-tied.minus.gdpo-noaux.mbpp-plus-abs}{3.7}
\setval{4b.concur-tied.minus.gdpo-noaux.minerva}{$-$0.4}
\setval{4b.concur-tied.minus.gdpo-noaux.minerva-abs}{0.4}
\setval{4b.concur-tied.minus.gdpo-noaux.olympiad-bench}{$-$0.3}
\setval{4b.concur-tied.minus.gdpo-noaux.olympiad-bench-abs}{0.3}
\setval{4b.concur-tied.minus.gdpo-noaux.writing}{+11.6}
\setval{4b.concur-tied.minus.gdpo-noaux.writing-abs}{11.6}
\setval{4b.concur-tied.minus.gdpo-noaux.writingbench}{+12.8}
\setval{4b.concur-tied.minus.gdpo-noaux.writingbench-abs}{12.8}
\setval{4b.concur-tied.minus.gdpo-w0.5.aime24}{+4.9}
\setval{4b.concur-tied.minus.gdpo-w0.5.aime24-abs}{4.9}
\setval{4b.concur-tied.minus.gdpo-w0.5.aime25}{+6.0}
\setval{4b.concur-tied.minus.gdpo-w0.5.aime25-abs}{6.0}
\setval{4b.concur-tied.minus.gdpo-w0.5.amc23}{+10.9}
\setval{4b.concur-tied.minus.gdpo-w0.5.amc23-abs}{10.9}
\setval{4b.concur-tied.minus.gdpo-w0.5.avg}{+4.1}
\setval{4b.concur-tied.minus.gdpo-w0.5.avg-abs}{4.1}
\setval{4b.concur-tied.minus.gdpo-w0.5.code}{+0.0}
\setval{4b.concur-tied.minus.gdpo-w0.5.code-abs}{0.0}
\setval{4b.concur-tied.minus.gdpo-w0.5.creative-writing-v3}{+9.7}
\setval{4b.concur-tied.minus.gdpo-w0.5.creative-writing-v3-abs}{9.7}
\setval{4b.concur-tied.minus.gdpo-w0.5.guidebench}{$-$0.4}
\setval{4b.concur-tied.minus.gdpo-w0.5.guidebench-abs}{0.4}
\setval{4b.concur-tied.minus.gdpo-w0.5.humaneval-plus}{+2.0}
\setval{4b.concur-tied.minus.gdpo-w0.5.humaneval-plus-abs}{2.0}
\setval{4b.concur-tied.minus.gdpo-w0.5.if}{$-$0.4}
\setval{4b.concur-tied.minus.gdpo-w0.5.if-abs}{0.4}
\setval{4b.concur-tied.minus.gdpo-w0.5.ifbench}{$-$1.7}
\setval{4b.concur-tied.minus.gdpo-w0.5.ifbench-abs}{1.7}
\setval{4b.concur-tied.minus.gdpo-w0.5.ifeval}{+1.0}
\setval{4b.concur-tied.minus.gdpo-w0.5.ifeval-abs}{1.0}
\setval{4b.concur-tied.minus.gdpo-w0.5.judge2.avg}{+3.0}
\setval{4b.concur-tied.minus.gdpo-w0.5.judge2.avg-abs}{3.0}
\setval{4b.concur-tied.minus.gdpo-w0.5.judge2.creative-writing-v3}{+3.9}
\setval{4b.concur-tied.minus.gdpo-w0.5.judge2.creative-writing-v3-abs}{3.9}
\setval{4b.concur-tied.minus.gdpo-w0.5.judge2.longbench-write}{+7.4}
\setval{4b.concur-tied.minus.gdpo-w0.5.judge2.longbench-write-abs}{7.4}
\setval{4b.concur-tied.minus.gdpo-w0.5.judge2.writing}{+4.4}
\setval{4b.concur-tied.minus.gdpo-w0.5.judge2.writing-abs}{4.4}
\setval{4b.concur-tied.minus.gdpo-w0.5.judge2.writingbench}{+1.7}
\setval{4b.concur-tied.minus.gdpo-w0.5.judge2.writingbench-abs}{1.7}
\setval{4b.concur-tied.minus.gdpo-w0.5.livecodebench}{+0.6}
\setval{4b.concur-tied.minus.gdpo-w0.5.livecodebench-abs}{0.6}
\setval{4b.concur-tied.minus.gdpo-w0.5.longbench-write}{+10.3}
\setval{4b.concur-tied.minus.gdpo-w0.5.longbench-write-abs}{10.3}
\setval{4b.concur-tied.minus.gdpo-w0.5.math}{+5.5}
\setval{4b.concur-tied.minus.gdpo-w0.5.math-abs}{5.5}
\setval{4b.concur-tied.minus.gdpo-w0.5.math500}{+3.7}
\setval{4b.concur-tied.minus.gdpo-w0.5.math500-abs}{3.7}
\setval{4b.concur-tied.minus.gdpo-w0.5.mbpp-plus}{$-$2.6}
\setval{4b.concur-tied.minus.gdpo-w0.5.mbpp-plus-abs}{2.6}
\setval{4b.concur-tied.minus.gdpo-w0.5.minerva}{+3.0}
\setval{4b.concur-tied.minus.gdpo-w0.5.minerva-abs}{3.0}
\setval{4b.concur-tied.minus.gdpo-w0.5.olympiad-bench}{+4.4}
\setval{4b.concur-tied.minus.gdpo-w0.5.olympiad-bench-abs}{4.4}
\setval{4b.concur-tied.minus.gdpo-w0.5.writing}{+9.8}
\setval{4b.concur-tied.minus.gdpo-w0.5.writing-abs}{9.8}
\setval{4b.concur-tied.minus.gdpo-w0.5.writingbench}{+9.4}
\setval{4b.concur-tied.minus.gdpo-w0.5.writingbench-abs}{9.4}
\setval{4b.concur-tied.minus.gdpo.aime24}{+10.4}
\setval{4b.concur-tied.minus.gdpo.aime24-abs}{10.4}
\setval{4b.concur-tied.minus.gdpo.aime25}{+6.5}
\setval{4b.concur-tied.minus.gdpo.aime25-abs}{6.5}
\setval{4b.concur-tied.minus.gdpo.amc23}{+9.6}
\setval{4b.concur-tied.minus.gdpo.amc23-abs}{9.6}
\setval{4b.concur-tied.minus.gdpo.avg}{+4.4}
\setval{4b.concur-tied.minus.gdpo.avg-abs}{4.4}
\setval{4b.concur-tied.minus.gdpo.code}{+0.1}
\setval{4b.concur-tied.minus.gdpo.code-abs}{0.1}
\setval{4b.concur-tied.minus.gdpo.creative-writing-v3}{+1.9}
\setval{4b.concur-tied.minus.gdpo.creative-writing-v3-abs}{1.9}
\setval{4b.concur-tied.minus.gdpo.guidebench}{$-$1.4}
\setval{4b.concur-tied.minus.gdpo.guidebench-abs}{1.4}
\setval{4b.concur-tied.minus.gdpo.humaneval-plus}{+2.2}
\setval{4b.concur-tied.minus.gdpo.humaneval-plus-abs}{2.2}
\setval{4b.concur-tied.minus.gdpo.if}{+0.6}
\setval{4b.concur-tied.minus.gdpo.if-abs}{0.6}
\setval{4b.concur-tied.minus.gdpo.ifbench}{+1.3}
\setval{4b.concur-tied.minus.gdpo.ifbench-abs}{1.3}
\setval{4b.concur-tied.minus.gdpo.ifeval}{+2.0}
\setval{4b.concur-tied.minus.gdpo.ifeval-abs}{2.0}
\setval{4b.concur-tied.minus.gdpo.judge2.avg}{+3.6}
\setval{4b.concur-tied.minus.gdpo.judge2.avg-abs}{3.6}
\setval{4b.concur-tied.minus.gdpo.judge2.creative-writing-v3}{$-$0.4}
\setval{4b.concur-tied.minus.gdpo.judge2.creative-writing-v3-abs}{0.4}
\setval{4b.concur-tied.minus.gdpo.judge2.longbench-write}{+5.7}
\setval{4b.concur-tied.minus.gdpo.judge2.longbench-write-abs}{5.7}
\setval{4b.concur-tied.minus.gdpo.judge2.writing}{+2.2}
\setval{4b.concur-tied.minus.gdpo.judge2.writing-abs}{2.2}
\setval{4b.concur-tied.minus.gdpo.judge2.writingbench}{+1.3}
\setval{4b.concur-tied.minus.gdpo.judge2.writingbench-abs}{1.3}
\setval{4b.concur-tied.minus.gdpo.livecodebench}{+0.3}
\setval{4b.concur-tied.minus.gdpo.livecodebench-abs}{0.3}
\setval{4b.concur-tied.minus.gdpo.longbench-write}{+8.7}
\setval{4b.concur-tied.minus.gdpo.longbench-write-abs}{8.7}
\setval{4b.concur-tied.minus.gdpo.math}{+7.4}
\setval{4b.concur-tied.minus.gdpo.math-abs}{7.4}
\setval{4b.concur-tied.minus.gdpo.math500}{+6.2}
\setval{4b.concur-tied.minus.gdpo.math500-abs}{6.2}
\setval{4b.concur-tied.minus.gdpo.mbpp-plus}{$-$2.3}
\setval{4b.concur-tied.minus.gdpo.mbpp-plus-abs}{2.3}
\setval{4b.concur-tied.minus.gdpo.minerva}{+3.6}
\setval{4b.concur-tied.minus.gdpo.minerva-abs}{3.6}
\setval{4b.concur-tied.minus.gdpo.olympiad-bench}{+8.2}
\setval{4b.concur-tied.minus.gdpo.olympiad-bench-abs}{8.2}
\setval{4b.concur-tied.minus.gdpo.writing}{+6.3}
\setval{4b.concur-tied.minus.gdpo.writing-abs}{6.3}
\setval{4b.concur-tied.minus.gdpo.writingbench}{+8.2}
\setval{4b.concur-tied.minus.gdpo.writingbench-abs}{8.2}
\setval{4b.concur-tied.minus.geoma.aime24}{+7.9}
\setval{4b.concur-tied.minus.geoma.aime24-abs}{7.9}
\setval{4b.concur-tied.minus.geoma.aime25}{+5.7}
\setval{4b.concur-tied.minus.geoma.aime25-abs}{5.7}
\setval{4b.concur-tied.minus.geoma.amc23}{+13.8}
\setval{4b.concur-tied.minus.geoma.amc23-abs}{13.8}
\setval{4b.concur-tied.minus.geoma.avg}{+4.9}
\setval{4b.concur-tied.minus.geoma.avg-abs}{4.9}
\setval{4b.concur-tied.minus.geoma.code}{+0.4}
\setval{4b.concur-tied.minus.geoma.code-abs}{0.4}
\setval{4b.concur-tied.minus.geoma.creative-writing-v3}{+2.0}
\setval{4b.concur-tied.minus.geoma.creative-writing-v3-abs}{2.0}
\setval{4b.concur-tied.minus.geoma.guidebench}{+0.7}
\setval{4b.concur-tied.minus.geoma.guidebench-abs}{0.7}
\setval{4b.concur-tied.minus.geoma.humaneval-plus}{+3.8}
\setval{4b.concur-tied.minus.geoma.humaneval-plus-abs}{3.8}
\setval{4b.concur-tied.minus.geoma.if}{+0.8}
\setval{4b.concur-tied.minus.geoma.if-abs}{0.8}
\setval{4b.concur-tied.minus.geoma.ifbench}{+1.7}
\setval{4b.concur-tied.minus.geoma.ifbench-abs}{1.7}
\setval{4b.concur-tied.minus.geoma.ifeval}{+0.0}
\setval{4b.concur-tied.minus.geoma.ifeval-abs}{0.0}
\setval{4b.concur-tied.minus.geoma.judge2.avg}{+4.3}
\setval{4b.concur-tied.minus.geoma.judge2.avg-abs}{4.3}
\setval{4b.concur-tied.minus.geoma.judge2.creative-writing-v3}{+0.9}
\setval{4b.concur-tied.minus.geoma.judge2.creative-writing-v3-abs}{0.9}
\setval{4b.concur-tied.minus.geoma.judge2.longbench-write}{+10.3}
\setval{4b.concur-tied.minus.geoma.judge2.longbench-write-abs}{10.3}
\setval{4b.concur-tied.minus.geoma.judge2.writing}{+5.3}
\setval{4b.concur-tied.minus.geoma.judge2.writing-abs}{5.3}
\setval{4b.concur-tied.minus.geoma.judge2.writingbench}{+4.6}
\setval{4b.concur-tied.minus.geoma.judge2.writingbench-abs}{4.6}
\setval{4b.concur-tied.minus.geoma.livecodebench}{$-$0.9}
\setval{4b.concur-tied.minus.geoma.livecodebench-abs}{0.9}
\setval{4b.concur-tied.minus.geoma.longbench-write}{+11.9}
\setval{4b.concur-tied.minus.geoma.longbench-write-abs}{11.9}
\setval{4b.concur-tied.minus.geoma.math}{+7.6}
\setval{4b.concur-tied.minus.geoma.math-abs}{7.6}
\setval{4b.concur-tied.minus.geoma.math500}{+6.7}
\setval{4b.concur-tied.minus.geoma.math500-abs}{6.7}
\setval{4b.concur-tied.minus.geoma.mbpp-plus}{$-$1.6}
\setval{4b.concur-tied.minus.geoma.mbpp-plus-abs}{1.6}
\setval{4b.concur-tied.minus.geoma.minerva}{+2.2}
\setval{4b.concur-tied.minus.geoma.minerva-abs}{2.2}
\setval{4b.concur-tied.minus.geoma.olympiad-bench}{+9.0}
\setval{4b.concur-tied.minus.geoma.olympiad-bench-abs}{9.0}
\setval{4b.concur-tied.minus.geoma.writing}{+8.2}
\setval{4b.concur-tied.minus.geoma.writing-abs}{8.2}
\setval{4b.concur-tied.minus.geoma.writingbench}{+10.7}
\setval{4b.concur-tied.minus.geoma.writingbench-abs}{10.7}
\setval{4b.concur-tied.minus.grpo-nodiv.aime24}{+7.1}
\setval{4b.concur-tied.minus.grpo-nodiv.aime24-abs}{7.1}
\setval{4b.concur-tied.minus.grpo-nodiv.aime25}{+8.0}
\setval{4b.concur-tied.minus.grpo-nodiv.aime25-abs}{8.0}
\setval{4b.concur-tied.minus.grpo-nodiv.amc23}{+14.7}
\setval{4b.concur-tied.minus.grpo-nodiv.amc23-abs}{14.7}
\setval{4b.concur-tied.minus.grpo-nodiv.avg}{+5.7}
\setval{4b.concur-tied.minus.grpo-nodiv.avg-abs}{5.7}
\setval{4b.concur-tied.minus.grpo-nodiv.code}{+0.4}
\setval{4b.concur-tied.minus.grpo-nodiv.code-abs}{0.4}
\setval{4b.concur-tied.minus.grpo-nodiv.creative-writing-v3}{+5.3}
\setval{4b.concur-tied.minus.grpo-nodiv.creative-writing-v3-abs}{5.3}
\setval{4b.concur-tied.minus.grpo-nodiv.guidebench}{+0.4}
\setval{4b.concur-tied.minus.grpo-nodiv.guidebench-abs}{0.4}
\setval{4b.concur-tied.minus.grpo-nodiv.humaneval-plus}{+3.5}
\setval{4b.concur-tied.minus.grpo-nodiv.humaneval-plus-abs}{3.5}
\setval{4b.concur-tied.minus.grpo-nodiv.if}{+1.6}
\setval{4b.concur-tied.minus.grpo-nodiv.if-abs}{1.6}
\setval{4b.concur-tied.minus.grpo-nodiv.ifbench}{+4.0}
\setval{4b.concur-tied.minus.grpo-nodiv.ifbench-abs}{4.0}
\setval{4b.concur-tied.minus.grpo-nodiv.ifeval}{+0.5}
\setval{4b.concur-tied.minus.grpo-nodiv.ifeval-abs}{0.5}
\setval{4b.concur-tied.minus.grpo-nodiv.judge2.avg}{+4.8}
\setval{4b.concur-tied.minus.grpo-nodiv.judge2.avg-abs}{4.8}
\setval{4b.concur-tied.minus.grpo-nodiv.judge2.creative-writing-v3}{+3.5}
\setval{4b.concur-tied.minus.grpo-nodiv.judge2.creative-writing-v3-abs}{3.5}
\setval{4b.concur-tied.minus.grpo-nodiv.judge2.longbench-write}{+9.4}
\setval{4b.concur-tied.minus.grpo-nodiv.judge2.longbench-write-abs}{9.4}
\setval{4b.concur-tied.minus.grpo-nodiv.judge2.writing}{+6.2}
\setval{4b.concur-tied.minus.grpo-nodiv.judge2.writing-abs}{6.2}
\setval{4b.concur-tied.minus.grpo-nodiv.judge2.writingbench}{+5.8}
\setval{4b.concur-tied.minus.grpo-nodiv.judge2.writingbench-abs}{5.8}
\setval{4b.concur-tied.minus.grpo-nodiv.livecodebench}{+0.3}
\setval{4b.concur-tied.minus.grpo-nodiv.livecodebench-abs}{0.3}
\setval{4b.concur-tied.minus.grpo-nodiv.longbench-write}{+11.7}
\setval{4b.concur-tied.minus.grpo-nodiv.longbench-write-abs}{11.7}
\setval{4b.concur-tied.minus.grpo-nodiv.math}{+7.9}
\setval{4b.concur-tied.minus.grpo-nodiv.math-abs}{7.9}
\setval{4b.concur-tied.minus.grpo-nodiv.math500}{+6.0}
\setval{4b.concur-tied.minus.grpo-nodiv.math500-abs}{6.0}
\setval{4b.concur-tied.minus.grpo-nodiv.mbpp-plus}{$-$2.6}
\setval{4b.concur-tied.minus.grpo-nodiv.mbpp-plus-abs}{2.6}
\setval{4b.concur-tied.minus.grpo-nodiv.minerva}{+3.3}
\setval{4b.concur-tied.minus.grpo-nodiv.minerva-abs}{3.3}
\setval{4b.concur-tied.minus.grpo-nodiv.olympiad-bench}{+8.0}
\setval{4b.concur-tied.minus.grpo-nodiv.olympiad-bench-abs}{8.0}
\setval{4b.concur-tied.minus.grpo-nodiv.writing}{+10.5}
\setval{4b.concur-tied.minus.grpo-nodiv.writing-abs}{10.5}
\setval{4b.concur-tied.minus.grpo-nodiv.writingbench}{+14.6}
\setval{4b.concur-tied.minus.grpo-nodiv.writingbench-abs}{14.6}
\setval{4b.concur-tied.minus.grpo.aime24}{+7.5}
\setval{4b.concur-tied.minus.grpo.aime24-abs}{7.5}
\setval{4b.concur-tied.minus.grpo.aime25}{+3.8}
\setval{4b.concur-tied.minus.grpo.aime25-abs}{3.8}
\setval{4b.concur-tied.minus.grpo.amc23}{+14.5}
\setval{4b.concur-tied.minus.grpo.amc23-abs}{14.5}
\setval{4b.concur-tied.minus.grpo.avg}{+4.0}
\setval{4b.concur-tied.minus.grpo.avg-abs}{4.0}
\setval{4b.concur-tied.minus.grpo.code}{$-$0.2}
\setval{4b.concur-tied.minus.grpo.code-abs}{0.2}
\setval{4b.concur-tied.minus.grpo.creative-writing-v3}{+0.5}
\setval{4b.concur-tied.minus.grpo.creative-writing-v3-abs}{0.5}
\setval{4b.concur-tied.minus.grpo.guidebench}{+1.2}
\setval{4b.concur-tied.minus.grpo.guidebench-abs}{1.2}
\setval{4b.concur-tied.minus.grpo.humaneval-plus}{+1.3}
\setval{4b.concur-tied.minus.grpo.humaneval-plus-abs}{1.3}
\setval{4b.concur-tied.minus.grpo.if}{+0.9}
\setval{4b.concur-tied.minus.grpo.if-abs}{0.9}
\setval{4b.concur-tied.minus.grpo.ifbench}{+0.0}
\setval{4b.concur-tied.minus.grpo.ifbench-abs}{0.0}
\setval{4b.concur-tied.minus.grpo.ifeval}{+1.3}
\setval{4b.concur-tied.minus.grpo.ifeval-abs}{1.3}
\setval{4b.concur-tied.minus.grpo.judge2.avg}{+3.5}
\setval{4b.concur-tied.minus.grpo.judge2.avg-abs}{3.5}
\setval{4b.concur-tied.minus.grpo.judge2.creative-writing-v3}{$-$1.1}
\setval{4b.concur-tied.minus.grpo.judge2.creative-writing-v3-abs}{1.1}
\setval{4b.concur-tied.minus.grpo.judge2.longbench-write}{+7.2}
\setval{4b.concur-tied.minus.grpo.judge2.longbench-write-abs}{7.2}
\setval{4b.concur-tied.minus.grpo.judge2.writing}{+2.7}
\setval{4b.concur-tied.minus.grpo.judge2.writing-abs}{2.7}
\setval{4b.concur-tied.minus.grpo.judge2.writingbench}{+1.9}
\setval{4b.concur-tied.minus.grpo.judge2.writingbench-abs}{1.9}
\setval{4b.concur-tied.minus.grpo.livecodebench}{+0.7}
\setval{4b.concur-tied.minus.grpo.livecodebench-abs}{0.7}
\setval{4b.concur-tied.minus.grpo.longbench-write}{+8.5}
\setval{4b.concur-tied.minus.grpo.longbench-write-abs}{8.5}
\setval{4b.concur-tied.minus.grpo.math}{+7.2}
\setval{4b.concur-tied.minus.grpo.math-abs}{7.2}
\setval{4b.concur-tied.minus.grpo.math500}{+5.3}
\setval{4b.concur-tied.minus.grpo.math500-abs}{5.3}
\setval{4b.concur-tied.minus.grpo.mbpp-plus}{$-$2.6}
\setval{4b.concur-tied.minus.grpo.mbpp-plus-abs}{2.6}
\setval{4b.concur-tied.minus.grpo.minerva}{+3.8}
\setval{4b.concur-tied.minus.grpo.minerva-abs}{3.8}
\setval{4b.concur-tied.minus.grpo.olympiad-bench}{+8.2}
\setval{4b.concur-tied.minus.grpo.olympiad-bench-abs}{8.2}
\setval{4b.concur-tied.minus.grpo.writing}{+5.0}
\setval{4b.concur-tied.minus.grpo.writing-abs}{5.0}
\setval{4b.concur-tied.minus.grpo.writingbench}{+5.9}
\setval{4b.concur-tied.minus.grpo.writingbench-abs}{5.9}
\setval{4b.concur-tied.minus.mt-grpo.aime24}{+8.8}
\setval{4b.concur-tied.minus.mt-grpo.aime24-abs}{8.8}
\setval{4b.concur-tied.minus.mt-grpo.aime25}{+10.2}
\setval{4b.concur-tied.minus.mt-grpo.aime25-abs}{10.2}
\setval{4b.concur-tied.minus.mt-grpo.amc23}{+9.5}
\setval{4b.concur-tied.minus.mt-grpo.amc23-abs}{9.5}
\setval{4b.concur-tied.minus.mt-grpo.avg}{+5.6}
\setval{4b.concur-tied.minus.mt-grpo.avg-abs}{5.6}
\setval{4b.concur-tied.minus.mt-grpo.code}{+1.9}
\setval{4b.concur-tied.minus.mt-grpo.code-abs}{1.9}
\setval{4b.concur-tied.minus.mt-grpo.creative-writing-v3}{+3.3}
\setval{4b.concur-tied.minus.mt-grpo.creative-writing-v3-abs}{3.3}
\setval{4b.concur-tied.minus.mt-grpo.guidebench}{$-$0.5}
\setval{4b.concur-tied.minus.mt-grpo.guidebench-abs}{0.5}
\setval{4b.concur-tied.minus.mt-grpo.humaneval-plus}{+4.9}
\setval{4b.concur-tied.minus.mt-grpo.humaneval-plus-abs}{4.9}
\setval{4b.concur-tied.minus.mt-grpo.if}{+1.3}
\setval{4b.concur-tied.minus.mt-grpo.if-abs}{1.3}
\setval{4b.concur-tied.minus.mt-grpo.ifbench}{+3.3}
\setval{4b.concur-tied.minus.mt-grpo.ifbench-abs}{3.3}
\setval{4b.concur-tied.minus.mt-grpo.ifeval}{+1.1}
\setval{4b.concur-tied.minus.mt-grpo.ifeval-abs}{1.1}
\setval{4b.concur-tied.minus.mt-grpo.judge2.avg}{+4.6}
\setval{4b.concur-tied.minus.mt-grpo.judge2.avg-abs}{4.6}
\setval{4b.concur-tied.minus.mt-grpo.judge2.creative-writing-v3}{$-$0.1}
\setval{4b.concur-tied.minus.mt-grpo.judge2.creative-writing-v3-abs}{0.1}
\setval{4b.concur-tied.minus.mt-grpo.judge2.longbench-write}{+7.2}
\setval{4b.concur-tied.minus.mt-grpo.judge2.longbench-write-abs}{7.2}
\setval{4b.concur-tied.minus.mt-grpo.judge2.writing}{+3.1}
\setval{4b.concur-tied.minus.mt-grpo.judge2.writing-abs}{3.1}
\setval{4b.concur-tied.minus.mt-grpo.judge2.writingbench}{+2.1}
\setval{4b.concur-tied.minus.mt-grpo.judge2.writingbench-abs}{2.1}
\setval{4b.concur-tied.minus.mt-grpo.livecodebench}{+2.1}
\setval{4b.concur-tied.minus.mt-grpo.livecodebench-abs}{2.1}
\setval{4b.concur-tied.minus.mt-grpo.longbench-write}{+10.9}
\setval{4b.concur-tied.minus.mt-grpo.longbench-write-abs}{10.9}
\setval{4b.concur-tied.minus.mt-grpo.math}{+8.4}
\setval{4b.concur-tied.minus.mt-grpo.math-abs}{8.4}
\setval{4b.concur-tied.minus.mt-grpo.math500}{+8.5}
\setval{4b.concur-tied.minus.mt-grpo.math500-abs}{8.5}
\setval{4b.concur-tied.minus.mt-grpo.mbpp-plus}{$-$1.5}
\setval{4b.concur-tied.minus.mt-grpo.mbpp-plus-abs}{1.5}
\setval{4b.concur-tied.minus.mt-grpo.minerva}{+3.1}
\setval{4b.concur-tied.minus.mt-grpo.minerva-abs}{3.1}
\setval{4b.concur-tied.minus.mt-grpo.olympiad-bench}{+10.3}
\setval{4b.concur-tied.minus.mt-grpo.olympiad-bench-abs}{10.3}
\setval{4b.concur-tied.minus.mt-grpo.writing}{+8.1}
\setval{4b.concur-tied.minus.mt-grpo.writing-abs}{8.1}
\setval{4b.concur-tied.minus.mt-grpo.writingbench}{+10.0}
\setval{4b.concur-tied.minus.mt-grpo.writingbench-abs}{10.0}
\setval{4b.concur-tied.minus.papo.aime24}{+10.2}
\setval{4b.concur-tied.minus.papo.aime24-abs}{10.2}
\setval{4b.concur-tied.minus.papo.aime25}{+8.9}
\setval{4b.concur-tied.minus.papo.aime25-abs}{8.9}
\setval{4b.concur-tied.minus.papo.amc23}{+15.7}
\setval{4b.concur-tied.minus.papo.amc23-abs}{15.7}
\setval{4b.concur-tied.minus.papo.avg}{+6.1}
\setval{4b.concur-tied.minus.papo.avg-abs}{6.1}
\setval{4b.concur-tied.minus.papo.code}{+0.0}
\setval{4b.concur-tied.minus.papo.code-abs}{0.0}
\setval{4b.concur-tied.minus.papo.creative-writing-v3}{+1.3}
\setval{4b.concur-tied.minus.papo.creative-writing-v3-abs}{1.3}
\setval{4b.concur-tied.minus.papo.guidebench}{$-$1.5}
\setval{4b.concur-tied.minus.papo.guidebench-abs}{1.5}
\setval{4b.concur-tied.minus.papo.humaneval-plus}{+1.0}
\setval{4b.concur-tied.minus.papo.humaneval-plus-abs}{1.0}
\setval{4b.concur-tied.minus.papo.if}{+1.6}
\setval{4b.concur-tied.minus.papo.if-abs}{1.6}
\setval{4b.concur-tied.minus.papo.ifbench}{+4.3}
\setval{4b.concur-tied.minus.papo.ifbench-abs}{4.3}
\setval{4b.concur-tied.minus.papo.ifeval}{+1.9}
\setval{4b.concur-tied.minus.papo.ifeval-abs}{1.9}
\setval{4b.concur-tied.minus.papo.judge2.avg}{+5.1}
\setval{4b.concur-tied.minus.papo.judge2.avg-abs}{5.1}
\setval{4b.concur-tied.minus.papo.judge2.creative-writing-v3}{$-$0.2}
\setval{4b.concur-tied.minus.papo.judge2.creative-writing-v3-abs}{0.2}
\setval{4b.concur-tied.minus.papo.judge2.longbench-write}{+10.4}
\setval{4b.concur-tied.minus.papo.judge2.longbench-write-abs}{10.4}
\setval{4b.concur-tied.minus.papo.judge2.writing}{+4.5}
\setval{4b.concur-tied.minus.papo.judge2.writing-abs}{4.5}
\setval{4b.concur-tied.minus.papo.judge2.writingbench}{+3.2}
\setval{4b.concur-tied.minus.papo.judge2.writingbench-abs}{3.2}
\setval{4b.concur-tied.minus.papo.livecodebench}{+1.2}
\setval{4b.concur-tied.minus.papo.livecodebench-abs}{1.2}
\setval{4b.concur-tied.minus.papo.longbench-write}{+13.8}
\setval{4b.concur-tied.minus.papo.longbench-write-abs}{13.8}
\setval{4b.concur-tied.minus.papo.math}{+9.6}
\setval{4b.concur-tied.minus.papo.math-abs}{9.6}
\setval{4b.concur-tied.minus.papo.math500}{+9.0}
\setval{4b.concur-tied.minus.papo.math500-abs}{9.0}
\setval{4b.concur-tied.minus.papo.mbpp-plus}{$-$2.1}
\setval{4b.concur-tied.minus.papo.mbpp-plus-abs}{2.1}
\setval{4b.concur-tied.minus.papo.minerva}{+2.6}
\setval{4b.concur-tied.minus.papo.minerva-abs}{2.6}
\setval{4b.concur-tied.minus.papo.olympiad-bench}{+11.4}
\setval{4b.concur-tied.minus.papo.olympiad-bench-abs}{11.4}
\setval{4b.concur-tied.minus.papo.writing}{+9.6}
\setval{4b.concur-tied.minus.papo.writing-abs}{9.6}
\setval{4b.concur-tied.minus.papo.writingbench}{+13.7}
\setval{4b.concur-tied.minus.papo.writingbench-abs}{13.7}
\setval{4b.concur-tied.minus.saw.aime24}{+7.0}
\setval{4b.concur-tied.minus.saw.aime24-abs}{7.0}
\setval{4b.concur-tied.minus.saw.aime25}{+5.2}
\setval{4b.concur-tied.minus.saw.aime25-abs}{5.2}
\setval{4b.concur-tied.minus.saw.amc23}{+12.6}
\setval{4b.concur-tied.minus.saw.amc23-abs}{12.6}
\setval{4b.concur-tied.minus.saw.avg}{+4.5}
\setval{4b.concur-tied.minus.saw.avg-abs}{4.5}
\setval{4b.concur-tied.minus.saw.code}{+1.2}
\setval{4b.concur-tied.minus.saw.code-abs}{1.2}
\setval{4b.concur-tied.minus.saw.creative-writing-v3}{+2.8}
\setval{4b.concur-tied.minus.saw.creative-writing-v3-abs}{2.8}
\setval{4b.concur-tied.minus.saw.guidebench}{$-$1.8}
\setval{4b.concur-tied.minus.saw.guidebench-abs}{1.8}
\setval{4b.concur-tied.minus.saw.humaneval-plus}{+3.3}
\setval{4b.concur-tied.minus.saw.humaneval-plus-abs}{3.3}
\setval{4b.concur-tied.minus.saw.if}{$-$1.2}
\setval{4b.concur-tied.minus.saw.if-abs}{1.2}
\setval{4b.concur-tied.minus.saw.ifbench}{$-$2.7}
\setval{4b.concur-tied.minus.saw.ifbench-abs}{2.7}
\setval{4b.concur-tied.minus.saw.ifeval}{+0.8}
\setval{4b.concur-tied.minus.saw.ifeval-abs}{0.8}
\setval{4b.concur-tied.minus.saw.judge2.avg}{+3.4}
\setval{4b.concur-tied.minus.saw.judge2.avg-abs}{3.4}
\setval{4b.concur-tied.minus.saw.judge2.creative-writing-v3}{$-$1.6}
\setval{4b.concur-tied.minus.saw.judge2.creative-writing-v3-abs}{1.6}
\setval{4b.concur-tied.minus.saw.judge2.longbench-write}{+7.9}
\setval{4b.concur-tied.minus.saw.judge2.longbench-write-abs}{7.9}
\setval{4b.concur-tied.minus.saw.judge2.writing}{+2.8}
\setval{4b.concur-tied.minus.saw.judge2.writing-abs}{2.8}
\setval{4b.concur-tied.minus.saw.judge2.writingbench}{+2.0}
\setval{4b.concur-tied.minus.saw.judge2.writingbench-abs}{2.0}
\setval{4b.concur-tied.minus.saw.livecodebench}{+1.9}
\setval{4b.concur-tied.minus.saw.livecodebench-abs}{1.9}
\setval{4b.concur-tied.minus.saw.longbench-write}{+11.5}
\setval{4b.concur-tied.minus.saw.longbench-write-abs}{11.5}
\setval{4b.concur-tied.minus.saw.math}{+7.1}
\setval{4b.concur-tied.minus.saw.math-abs}{7.1}
\setval{4b.concur-tied.minus.saw.math500}{+7.7}
\setval{4b.concur-tied.minus.saw.math500-abs}{7.7}
\setval{4b.concur-tied.minus.saw.mbpp-plus}{$-$1.5}
\setval{4b.concur-tied.minus.saw.mbpp-plus-abs}{1.5}
\setval{4b.concur-tied.minus.saw.minerva}{+1.0}
\setval{4b.concur-tied.minus.saw.minerva-abs}{1.0}
\setval{4b.concur-tied.minus.saw.olympiad-bench}{+9.0}
\setval{4b.concur-tied.minus.saw.olympiad-bench-abs}{9.0}
\setval{4b.concur-tied.minus.saw.writing}{+8.2}
\setval{4b.concur-tied.minus.saw.writing-abs}{8.2}
\setval{4b.concur-tied.minus.saw.writingbench}{+10.3}
\setval{4b.concur-tied.minus.saw.writingbench-abs}{10.3}
\setval{4b.concur-tied.passing-at-ceiling.code.early}{0.70}
\setval{4b.concur-tied.passing-at-ceiling.code.late}{0.73}
\setval{4b.concur-tied.passing-at-ceiling.code.run}{0.73}
\setval{4b.concur-tied.passing-at-ceiling.if.early}{0.28}
\setval{4b.concur-tied.passing-at-ceiling.if.late}{0.55}
\setval{4b.concur-tied.passing-at-ceiling.if.run}{0.47}
\setval{4b.concur-tied.passing-at-ceiling.math.early}{1.00}
\setval{4b.concur-tied.passing-at-ceiling.math.late}{1.00}
\setval{4b.concur-tied.passing-at-ceiling.math.run}{1.00}
\setval{4b.concur-tied.passing-at-ceiling.writing.early}{0.00}
\setval{4b.concur-tied.passing-at-ceiling.writing.late}{0.01}
\setval{4b.concur-tied.passing-at-ceiling.writing.run}{0.01}
\setval{4b.concur-tied.primary-cosine.code.early}{0.98}
\setval{4b.concur-tied.primary-cosine.code.late}{0.96}
\setval{4b.concur-tied.primary-cosine.code.run}{0.97}
\setval{4b.concur-tied.primary-cosine.if.early}{0.86}
\setval{4b.concur-tied.primary-cosine.if.late}{0.80}
\setval{4b.concur-tied.primary-cosine.if.run}{0.84}
\setval{4b.concur-tied.primary-cosine.math.early}{1.00}
\setval{4b.concur-tied.primary-cosine.math.late}{0.97}
\setval{4b.concur-tied.primary-cosine.math.run}{0.98}
\setval{4b.concur-tied.primary-cosine.writing.early}{0.94}
\setval{4b.concur-tied.primary-cosine.writing.late}{0.92}
\setval{4b.concur-tied.primary-cosine.writing.run}{0.92}
\setval{4b.concur-tied.primary-level.code.early}{0.53}
\setval{4b.concur-tied.primary-level.code.late}{0.75}
\setval{4b.concur-tied.primary-level.code.run}{0.70}
\setval{4b.concur-tied.primary-level.if.early}{0.32}
\setval{4b.concur-tied.primary-level.if.late}{0.74}
\setval{4b.concur-tied.primary-level.if.run}{0.62}
\setval{4b.concur-tied.primary-level.math.early}{0.47}
\setval{4b.concur-tied.primary-level.math.late}{0.71}
\setval{4b.concur-tied.primary-level.math.run}{0.65}
\setval{4b.concur-tied.primary-level.writing.early}{0.54}
\setval{4b.concur-tied.primary-level.writing.late}{0.80}
\setval{4b.concur-tied.primary-level.writing.run}{0.75}
\setval{4b.concur-tied.solved.code.early}{0.09}
\setval{4b.concur-tied.solved.code.late}{0.11}
\setval{4b.concur-tied.solved.code.run}{0.12}
\setval{4b.concur-tied.solved.if.early}{0.03}
\setval{4b.concur-tied.solved.if.late}{0.21}
\setval{4b.concur-tied.solved.if.run}{0.14}
\setval{4b.concur-tied.solved.math.early}{0.13}
\setval{4b.concur-tied.solved.math.late}{0.48}
\setval{4b.concur-tied.solved.math.run}{0.36}
\setval{4b.concur-tied.solved.writing.early}{0.00}
\setval{4b.concur-tied.solved.writing.late}{0.00}
\setval{4b.concur-tied.solved.writing.run}{0.00}
\setval{4b.concur-tied.trust.code.agreement}{+0.007}
\setval{4b.concur-tied.trust.code.errors}{+1.9}
\setval{4b.concur-tied.trust.code.first-free}{73}
\setval{4b.concur-tied.trust.code.free-share}{0.23}
\setval{4b.concur-tied.trust.if.agreement}{+0.018}
\setval{4b.concur-tied.trust.if.errors}{+6.5}
\setval{4b.concur-tied.trust.if.first-free}{29}
\setval{4b.concur-tied.trust.if.free-share}{0.90}
\setval{4b.concur-tied.trust.math.first-free}{--}
\setval{4b.concur-tied.trust.math.free-share}{0.00}
\setval{4b.concur-tied.trust.writing-rm.agreement}{+0.058}
\setval{4b.concur-tied.trust.writing-rm.errors}{+35.7}
\setval{4b.concur-tied.trust.writing-rm.first-free}{1}
\setval{4b.concur-tied.trust.writing-rm.free-share}{1.00}
\setval{4b.concur-tied.trust.writing.agreement}{$-$0.048}
\setval{4b.concur-tied.trust.writing.errors}{$-$32.7}
\setval{4b.concur-tied.trust.writing.first-free}{--}
\setval{4b.concur-tied.trust.writing.free-share}{0.00}
\setval{4b.concur-tied.update-share.code.early}{0.20}
\setval{4b.concur-tied.update-share.code.late}{0.24}
\setval{4b.concur-tied.update-share.code.run}{0.23}
\setval{4b.concur-tied.update-share.if.early}{0.21}
\setval{4b.concur-tied.update-share.if.late}{0.17}
\setval{4b.concur-tied.update-share.if.run}{0.21}
\setval{4b.concur-tied.update-share.math.early}{0.17}
\setval{4b.concur-tied.update-share.math.late}{0.18}
\setval{4b.concur-tied.update-share.math.run}{0.19}
\setval{4b.concur-tied.update-share.writing.early}{0.42}
\setval{4b.concur-tied.update-share.writing.late}{0.41}
\setval{4b.concur-tied.update-share.writing.run}{0.37}
\setval{4b.concur.aux-change.code}{+0.20}
\setval{4b.concur.aux-change.if}{+0.24}
\setval{4b.concur.aux-change.math}{$-$0.19}
\setval{4b.concur.aux-change.writing}{$-$0.25}
\setval{4b.concur.aux-change.writing-rm}{+0.35}
\setval{4b.concur.aux-full.code.early}{0.62}
\setval{4b.concur.aux-full.code.late}{0.95}
\setval{4b.concur.aux-full.code.run}{0.91}
\setval{4b.concur.aux-full.if.early}{0.63}
\setval{4b.concur.aux-full.if.late}{0.90}
\setval{4b.concur.aux-full.if.run}{0.83}
\setval{4b.concur.aux-full.math.early}{0.59}
\setval{4b.concur.aux-full.math.late}{0.23}
\setval{4b.concur.aux-full.math.run}{0.41}
\setval{4b.concur.aux-full.writing-rm.early}{0.11}
\setval{4b.concur.aux-full.writing-rm.late}{0.48}
\setval{4b.concur.aux-full.writing-rm.run}{0.42}
\setval{4b.concur.aux-full.writing.early}{0.57}
\setval{4b.concur.aux-full.writing.late}{0.23}
\setval{4b.concur.aux-full.writing.run}{0.28}
\setval{4b.concur.aux-level.code.early}{0.90}
\setval{4b.concur.aux-level.code.late}{0.99}
\setval{4b.concur.aux-level.code.run}{0.98}
\setval{4b.concur.aux-level.if.early}{0.82}
\setval{4b.concur.aux-level.if.late}{0.94}
\setval{4b.concur.aux-level.if.run}{0.91}
\setval{4b.concur.aux-level.math.early}{0.91}
\setval{4b.concur.aux-level.math.late}{0.66}
\setval{4b.concur.aux-level.math.run}{0.80}
\setval{4b.concur.aux-level.writing-rm.early}{0.78}
\setval{4b.concur.aux-level.writing-rm.late}{0.95}
\setval{4b.concur.aux-level.writing-rm.run}{0.93}
\setval{4b.concur.aux-level.writing.early}{0.87}
\setval{4b.concur.aux-level.writing.late}{0.52}
\setval{4b.concur.aux-level.writing.run}{0.59}
\setval{4b.concur.bench.aime24}{20.5}
\setval{4b.concur.bench.aime25}{18.8}
\setval{4b.concur.bench.amc23}{69.4}
\setval{4b.concur.bench.creative-writing-v3}{44.7}
\setval{4b.concur.bench.guidebench}{77.4}
\setval{4b.concur.bench.humaneval-plus}{82.4}
\setval{4b.concur.bench.ifbench}{40.3}
\setval{4b.concur.bench.ifeval}{88.1}
\setval{4b.concur.bench.livecodebench}{25.1}
\setval{4b.concur.bench.longbench-write}{76.3}
\setval{4b.concur.bench.math500}{85.3}
\setval{4b.concur.bench.mbpp-plus}{71.5}
\setval{4b.concur.bench.minerva}{49.2}
\setval{4b.concur.bench.olympiad-bench}{53.6}
\setval{4b.concur.bench.writingbench}{79.4}
\setval{4b.concur.best-count.all}{10}
\setval{4b.concur.best-count.code}{1}
\setval{4b.concur.best-count.if}{1}
\setval{4b.concur.best-count.math}{6}
\setval{4b.concur.best-count.writing}{2}
\setval{4b.concur.eval.avg}{58.8}
\setval{4b.concur.eval.code}{59.7}
\setval{4b.concur.eval.if}{68.6}
\setval{4b.concur.eval.math}{49.5}
\setval{4b.concur.eval.writing}{66.8}
\setval{4b.concur.gate.code.against.early}{0.04}
\setval{4b.concur.gate.code.against.late}{0.02}
\setval{4b.concur.gate.code.against.run}{0.02}
\setval{4b.concur.gate.code.agreement}{+0.012}
\setval{4b.concur.gate.code.ceiling.early}{0.14}
\setval{4b.concur.gate.code.ceiling.late}{0.05}
\setval{4b.concur.gate.code.ceiling.run}{0.06}
\setval{4b.concur.gate.code.cosine.early}{0.82}
\setval{4b.concur.gate.code.cosine.late}{0.80}
\setval{4b.concur.gate.code.cosine.run}{0.82}
\setval{4b.concur.gate.if.against.early}{0.05}
\setval{4b.concur.gate.if.against.late}{0.04}
\setval{4b.concur.gate.if.against.run}{0.04}
\setval{4b.concur.gate.if.agreement}{+0.018}
\setval{4b.concur.gate.if.ceiling.early}{0.06}
\setval{4b.concur.gate.if.ceiling.late}{0.18}
\setval{4b.concur.gate.if.ceiling.run}{0.13}
\setval{4b.concur.gate.if.cosine.early}{0.73}
\setval{4b.concur.gate.if.cosine.late}{0.52}
\setval{4b.concur.gate.if.cosine.run}{0.59}
\setval{4b.concur.gate.math.against.early}{0.00}
\setval{4b.concur.gate.math.against.late}{0.00}
\setval{4b.concur.gate.math.against.run}{0.00}
\setval{4b.concur.gate.math.ceiling.early}{0.20}
\setval{4b.concur.gate.math.ceiling.late}{0.76}
\setval{4b.concur.gate.math.ceiling.run}{0.54}
\setval{4b.concur.gate.math.cosine.early}{0.88}
\setval{4b.concur.gate.math.cosine.late}{0.39}
\setval{4b.concur.gate.math.cosine.run}{0.60}
\setval{4b.concur.gate.writing-rm.against.early}{0.20}
\setval{4b.concur.gate.writing-rm.against.late}{0.58}
\setval{4b.concur.gate.writing-rm.against.run}{0.52}
\setval{4b.concur.gate.writing-rm.agreement}{+0.061}
\setval{4b.concur.gate.writing-rm.ceiling.early}{0.00}
\setval{4b.concur.gate.writing-rm.ceiling.late}{0.00}
\setval{4b.concur.gate.writing-rm.ceiling.run}{0.00}
\setval{4b.concur.gate.writing-rm.cosine.early}{0.69}
\setval{4b.concur.gate.writing-rm.cosine.late}{0.10}
\setval{4b.concur.gate.writing-rm.cosine.run}{0.19}
\setval{4b.concur.gate.writing.against.early}{0.33}
\setval{4b.concur.gate.writing.against.late}{0.67}
\setval{4b.concur.gate.writing.against.run}{0.65}
\setval{4b.concur.gate.writing.agreement}{$-$0.054}
\setval{4b.concur.gate.writing.ceiling.early}{0.00}
\setval{4b.concur.gate.writing.ceiling.late}{0.01}
\setval{4b.concur.gate.writing.ceiling.run}{0.00}
\setval{4b.concur.gate.writing.cosine.early}{0.48}
\setval{4b.concur.gate.writing.cosine.late}{-0.05}
\setval{4b.concur.gate.writing.cosine.run}{0.01}
\setval{4b.concur.judge2.avg}{55.8}
\setval{4b.concur.judge2.creative-writing-v3}{32.4}
\setval{4b.concur.judge2.longbench-write}{66.6}
\setval{4b.concur.judge2.writing}{51.7}
\setval{4b.concur.judge2.writingbench}{56.0}
\setval{4b.concur.last-step}{300}
\setval{4b.concur.minus.base.aime24}{+12.7}
\setval{4b.concur.minus.base.aime24-abs}{12.7}
\setval{4b.concur.minus.base.aime25}{+11.7}
\setval{4b.concur.minus.base.aime25-abs}{11.7}
\setval{4b.concur.minus.base.amc23}{+25.5}
\setval{4b.concur.minus.base.amc23-abs}{25.5}
\setval{4b.concur.minus.base.avg}{+24.3}
\setval{4b.concur.minus.base.avg-abs}{24.3}
\setval{4b.concur.minus.base.code}{+15.1}
\setval{4b.concur.minus.base.code-abs}{15.1}
\setval{4b.concur.minus.base.creative-writing-v3}{+19.9}
\setval{4b.concur.minus.base.creative-writing-v3-abs}{19.9}
\setval{4b.concur.minus.base.guidebench}{+52.5}
\setval{4b.concur.minus.base.guidebench-abs}{52.5}
\setval{4b.concur.minus.base.humaneval-plus}{+17.4}
\setval{4b.concur.minus.base.humaneval-plus-abs}{17.4}
\setval{4b.concur.minus.base.if}{+38.9}
\setval{4b.concur.minus.base.if-abs}{38.9}
\setval{4b.concur.minus.base.ifbench}{+23.3}
\setval{4b.concur.minus.base.ifbench-abs}{23.3}
\setval{4b.concur.minus.base.ifeval}{+40.8}
\setval{4b.concur.minus.base.ifeval-abs}{40.8}
\setval{4b.concur.minus.base.judge2.avg}{+21.9}
\setval{4b.concur.minus.base.judge2.avg-abs}{21.9}
\setval{4b.concur.minus.base.judge2.creative-writing-v3}{+11.5}
\setval{4b.concur.minus.base.judge2.creative-writing-v3-abs}{11.5}
\setval{4b.concur.minus.base.judge2.longbench-write}{+25.7}
\setval{4b.concur.minus.base.judge2.longbench-write-abs}{25.7}
\setval{4b.concur.minus.base.judge2.writing}{+20.9}
\setval{4b.concur.minus.base.judge2.writing-abs}{20.9}
\setval{4b.concur.minus.base.judge2.writingbench}{+25.6}
\setval{4b.concur.minus.base.judge2.writingbench-abs}{25.6}
\setval{4b.concur.minus.base.livecodebench}{+11.9}
\setval{4b.concur.minus.base.livecodebench-abs}{11.9}
\setval{4b.concur.minus.base.longbench-write}{+31.9}
\setval{4b.concur.minus.base.longbench-write-abs}{31.9}
\setval{4b.concur.minus.base.math}{+17.2}
\setval{4b.concur.minus.base.math-abs}{17.2}
\setval{4b.concur.minus.base.math500}{+18.0}
\setval{4b.concur.minus.base.math500-abs}{18.0}
\setval{4b.concur.minus.base.mbpp-plus}{+15.9}
\setval{4b.concur.minus.base.mbpp-plus-abs}{15.9}
\setval{4b.concur.minus.base.minerva}{+17.6}
\setval{4b.concur.minus.base.minerva-abs}{17.6}
\setval{4b.concur.minus.base.olympiad-bench}{+18.0}
\setval{4b.concur.minus.base.olympiad-bench-abs}{18.0}
\setval{4b.concur.minus.base.writing}{+32.9}
\setval{4b.concur.minus.base.writing-abs}{32.9}
\setval{4b.concur.minus.base.writingbench}{+47.0}
\setval{4b.concur.minus.base.writingbench-abs}{47.0}
\setval{4b.concur.minus.best.bench.aime24}{+8.0}
\setval{4b.concur.minus.best.bench.aime25}{+6.1}
\setval{4b.concur.minus.best.bench.amc23}{+9.6}
\setval{4b.concur.minus.best.bench.creative-writing-v3}{$-$2.6}
\setval{4b.concur.minus.best.bench.guidebench}{$-$3.6}
\setval{4b.concur.minus.best.bench.humaneval-plus}{$-$1.0}
\setval{4b.concur.minus.best.bench.ifbench}{+0.0}
\setval{4b.concur.minus.best.bench.ifeval}{+0.4}
\setval{4b.concur.minus.best.bench.livecodebench}{$-$0.5}
\setval{4b.concur.minus.best.bench.longbench-write}{+7.1}
\setval{4b.concur.minus.best.bench.math500}{+5.0}
\setval{4b.concur.minus.best.bench.mbpp-plus}{+0.1}
\setval{4b.concur.minus.best.bench.minerva}{+1.1}
\setval{4b.concur.minus.best.bench.olympiad-bench}{+8.9}
\setval{4b.concur.minus.best.bench.writingbench}{+5.2}
\setval{4b.concur.minus.best.eval.avg}{+4.1}
\setval{4b.concur.minus.best.eval.code}{+0.2}
\setval{4b.concur.minus.best.eval.if}{$-$0.8}
\setval{4b.concur.minus.best.eval.math}{+7.8}
\setval{4b.concur.minus.best.eval.writing}{+3.2}
\setval{4b.concur.minus.concur-free.aime24}{+8.6}
\setval{4b.concur.minus.concur-free.aime24-abs}{8.6}
\setval{4b.concur.minus.concur-free.aime25}{+4.2}
\setval{4b.concur.minus.concur-free.aime25-abs}{4.2}
\setval{4b.concur.minus.concur-free.amc23}{+15.2}
\setval{4b.concur.minus.concur-free.amc23-abs}{15.2}
\setval{4b.concur.minus.concur-free.avg}{+3.3}
\setval{4b.concur.minus.concur-free.avg-abs}{3.3}
\setval{4b.concur.minus.concur-free.code}{+1.4}
\setval{4b.concur.minus.concur-free.code-abs}{1.4}
\setval{4b.concur.minus.concur-free.creative-writing-v3}{$-$0.7}
\setval{4b.concur.minus.concur-free.creative-writing-v3-abs}{0.7}
\setval{4b.concur.minus.concur-free.guidebench}{$-$1.6}
\setval{4b.concur.minus.concur-free.guidebench-abs}{1.6}
\setval{4b.concur.minus.concur-free.humaneval-plus}{+1.3}
\setval{4b.concur.minus.concur-free.humaneval-plus-abs}{1.3}
\setval{4b.concur.minus.concur-free.if}{+0.4}
\setval{4b.concur.minus.concur-free.if-abs}{0.4}
\setval{4b.concur.minus.concur-free.ifbench}{+1.3}
\setval{4b.concur.minus.concur-free.ifbench-abs}{1.3}
\setval{4b.concur.minus.concur-free.ifeval}{+1.4}
\setval{4b.concur.minus.concur-free.ifeval-abs}{1.4}
\setval{4b.concur.minus.concur-free.judge2.avg}{+3.0}
\setval{4b.concur.minus.concur-free.judge2.avg-abs}{3.0}
\setval{4b.concur.minus.concur-free.judge2.creative-writing-v3}{$-$2.9}
\setval{4b.concur.minus.concur-free.judge2.creative-writing-v3-abs}{2.9}
\setval{4b.concur.minus.concur-free.judge2.longbench-write}{$-$2.7}
\setval{4b.concur.minus.concur-free.judge2.longbench-write-abs}{2.7}
\setval{4b.concur.minus.concur-free.judge2.writing}{$-$1.3}
\setval{4b.concur.minus.concur-free.judge2.writing-abs}{1.3}
\setval{4b.concur.minus.concur-free.judge2.writingbench}{+1.6}
\setval{4b.concur.minus.concur-free.judge2.writingbench-abs}{1.6}
\setval{4b.concur.minus.concur-free.livecodebench}{+1.0}
\setval{4b.concur.minus.concur-free.livecodebench-abs}{1.0}
\setval{4b.concur.minus.concur-free.longbench-write}{$-$1.5}
\setval{4b.concur.minus.concur-free.longbench-write-abs}{1.5}
\setval{4b.concur.minus.concur-free.math}{+7.3}
\setval{4b.concur.minus.concur-free.math-abs}{7.3}
\setval{4b.concur.minus.concur-free.math500}{+5.0}
\setval{4b.concur.minus.concur-free.math500-abs}{5.0}
\setval{4b.concur.minus.concur-free.mbpp-plus}{+1.9}
\setval{4b.concur.minus.concur-free.mbpp-plus-abs}{1.9}
\setval{4b.concur.minus.concur-free.minerva}{+3.0}
\setval{4b.concur.minus.concur-free.minerva-abs}{3.0}
\setval{4b.concur.minus.concur-free.olympiad-bench}{+7.9}
\setval{4b.concur.minus.concur-free.olympiad-bench-abs}{7.9}
\setval{4b.concur.minus.concur-free.writing}{+0.0}
\setval{4b.concur.minus.concur-free.writing-abs}{0.0}
\setval{4b.concur.minus.concur-free.writingbench}{+2.1}
\setval{4b.concur.minus.concur-free.writingbench-abs}{2.1}
\setval{4b.concur.minus.concur-tied.aime24}{+1.0}
\setval{4b.concur.minus.concur-tied.aime24-abs}{1.0}
\setval{4b.concur.minus.concur-tied.aime25}{+2.4}
\setval{4b.concur.minus.concur-tied.aime25-abs}{2.4}
\setval{4b.concur.minus.concur-tied.amc23}{+0.2}
\setval{4b.concur.minus.concur-tied.amc23-abs}{0.2}
\setval{4b.concur.minus.concur-tied.avg}{+0.1}
\setval{4b.concur.minus.concur-tied.avg-abs}{0.1}
\setval{4b.concur.minus.concur-tied.code}{+0.6}
\setval{4b.concur.minus.concur-tied.code-abs}{0.6}
\setval{4b.concur.minus.concur-tied.creative-writing-v3}{$-$3.1}
\setval{4b.concur.minus.concur-tied.creative-writing-v3-abs}{3.1}
\setval{4b.concur.minus.concur-tied.guidebench}{$-$1.8}
\setval{4b.concur.minus.concur-tied.guidebench-abs}{1.8}
\setval{4b.concur.minus.concur-tied.humaneval-plus}{$-$1.6}
\setval{4b.concur.minus.concur-tied.humaneval-plus-abs}{1.6}
\setval{4b.concur.minus.concur-tied.if}{+0.4}
\setval{4b.concur.minus.concur-tied.if-abs}{0.4}
\setval{4b.concur.minus.concur-tied.ifbench}{+2.7}
\setval{4b.concur.minus.concur-tied.ifbench-abs}{2.7}
\setval{4b.concur.minus.concur-tied.ifeval}{+0.4}
\setval{4b.concur.minus.concur-tied.ifeval-abs}{0.4}
\setval{4b.concur.minus.concur-tied.judge2.avg}{+0.3}
\setval{4b.concur.minus.concur-tied.judge2.avg-abs}{0.3}
\setval{4b.concur.minus.concur-tied.judge2.creative-writing-v3}{$-$1.6}
\setval{4b.concur.minus.concur-tied.judge2.creative-writing-v3-abs}{1.6}
\setval{4b.concur.minus.concur-tied.judge2.longbench-write}{$-$1.3}
\setval{4b.concur.minus.concur-tied.judge2.longbench-write-abs}{1.3}
\setval{4b.concur.minus.concur-tied.judge2.writing}{$-$0.7}
\setval{4b.concur.minus.concur-tied.judge2.writing-abs}{0.7}
\setval{4b.concur.minus.concur-tied.judge2.writingbench}{+0.7}
\setval{4b.concur.minus.concur-tied.judge2.writingbench-abs}{0.7}
\setval{4b.concur.minus.concur-tied.livecodebench}{+0.4}
\setval{4b.concur.minus.concur-tied.livecodebench-abs}{0.4}
\setval{4b.concur.minus.concur-tied.longbench-write}{$-$1.4}
\setval{4b.concur.minus.concur-tied.longbench-write-abs}{1.4}
\setval{4b.concur.minus.concur-tied.math}{+0.7}
\setval{4b.concur.minus.concur-tied.math-abs}{0.7}
\setval{4b.concur.minus.concur-tied.math500}{$-$0.2}
\setval{4b.concur.minus.concur-tied.math500-abs}{0.2}
\setval{4b.concur.minus.concur-tied.mbpp-plus}{+2.9}
\setval{4b.concur.minus.concur-tied.mbpp-plus-abs}{2.9}
\setval{4b.concur.minus.concur-tied.minerva}{+0.1}
\setval{4b.concur.minus.concur-tied.minerva-abs}{0.1}
\setval{4b.concur.minus.concur-tied.olympiad-bench}{+0.7}
\setval{4b.concur.minus.concur-tied.olympiad-bench-abs}{0.7}
\setval{4b.concur.minus.concur-tied.writing}{$-$1.8}
\setval{4b.concur.minus.concur-tied.writing-abs}{1.8}
\setval{4b.concur.minus.concur-tied.writingbench}{$-$0.8}
\setval{4b.concur.minus.concur-tied.writingbench-abs}{0.8}
\setval{4b.concur.minus.gd2po.aime24}{+8.6}
\setval{4b.concur.minus.gd2po.aime24-abs}{8.6}
\setval{4b.concur.minus.gd2po.aime25}{+12.0}
\setval{4b.concur.minus.gd2po.aime25-abs}{12.0}
\setval{4b.concur.minus.gd2po.amc23}{+11.6}
\setval{4b.concur.minus.gd2po.amc23-abs}{11.6}
\setval{4b.concur.minus.gd2po.avg}{+4.7}
\setval{4b.concur.minus.gd2po.avg-abs}{4.7}
\setval{4b.concur.minus.gd2po.code}{+0.2}
\setval{4b.concur.minus.gd2po.code-abs}{0.2}
\setval{4b.concur.minus.gd2po.creative-writing-v3}{+0.0}
\setval{4b.concur.minus.gd2po.creative-writing-v3-abs}{0.0}
\setval{4b.concur.minus.gd2po.guidebench}{$-$1.7}
\setval{4b.concur.minus.gd2po.guidebench-abs}{1.7}
\setval{4b.concur.minus.gd2po.humaneval-plus}{$-$1.0}
\setval{4b.concur.minus.gd2po.humaneval-plus-abs}{1.0}
\setval{4b.concur.minus.gd2po.if}{+0.5}
\setval{4b.concur.minus.gd2po.if-abs}{0.5}
\setval{4b.concur.minus.gd2po.ifbench}{+2.7}
\setval{4b.concur.minus.gd2po.ifbench-abs}{2.7}
\setval{4b.concur.minus.gd2po.ifeval}{+0.5}
\setval{4b.concur.minus.gd2po.ifeval-abs}{0.5}
\setval{4b.concur.minus.gd2po.judge2.avg}{+4.2}
\setval{4b.concur.minus.gd2po.judge2.avg-abs}{4.2}
\setval{4b.concur.minus.gd2po.judge2.creative-writing-v3}{$-$0.5}
\setval{4b.concur.minus.gd2po.judge2.creative-writing-v3-abs}{0.5}
\setval{4b.concur.minus.gd2po.judge2.longbench-write}{+5.7}
\setval{4b.concur.minus.gd2po.judge2.longbench-write-abs}{5.7}
\setval{4b.concur.minus.gd2po.judge2.writing}{+2.9}
\setval{4b.concur.minus.gd2po.judge2.writing-abs}{2.9}
\setval{4b.concur.minus.gd2po.judge2.writingbench}{+3.5}
\setval{4b.concur.minus.gd2po.judge2.writingbench-abs}{3.5}
\setval{4b.concur.minus.gd2po.livecodebench}{+1.7}
\setval{4b.concur.minus.gd2po.livecodebench-abs}{1.7}
\setval{4b.concur.minus.gd2po.longbench-write}{+7.4}
\setval{4b.concur.minus.gd2po.longbench-write-abs}{7.4}
\setval{4b.concur.minus.gd2po.math}{+8.7}
\setval{4b.concur.minus.gd2po.math-abs}{8.7}
\setval{4b.concur.minus.gd2po.math500}{+7.6}
\setval{4b.concur.minus.gd2po.math500-abs}{7.6}
\setval{4b.concur.minus.gd2po.mbpp-plus}{+0.1}
\setval{4b.concur.minus.gd2po.mbpp-plus-abs}{0.1}
\setval{4b.concur.minus.gd2po.minerva}{+2.0}
\setval{4b.concur.minus.gd2po.minerva-abs}{2.0}
\setval{4b.concur.minus.gd2po.olympiad-bench}{+10.0}
\setval{4b.concur.minus.gd2po.olympiad-bench-abs}{10.0}
\setval{4b.concur.minus.gd2po.writing}{+5.6}
\setval{4b.concur.minus.gd2po.writing-abs}{5.6}
\setval{4b.concur.minus.gd2po.writingbench}{+9.3}
\setval{4b.concur.minus.gd2po.writingbench-abs}{9.3}
\setval{4b.concur.minus.gdpo-noaux.aime24}{$-$2.9}
\setval{4b.concur.minus.gdpo-noaux.aime24-abs}{2.9}
\setval{4b.concur.minus.gdpo-noaux.aime25}{+0.0}
\setval{4b.concur.minus.gdpo-noaux.aime25-abs}{0.0}
\setval{4b.concur.minus.gdpo-noaux.amc23}{+3.7}
\setval{4b.concur.minus.gdpo-noaux.amc23-abs}{3.7}
\setval{4b.concur.minus.gdpo-noaux.avg}{+2.1}
\setval{4b.concur.minus.gdpo-noaux.avg-abs}{2.1}
\setval{4b.concur.minus.gdpo-noaux.code}{$-$1.2}
\setval{4b.concur.minus.gdpo-noaux.code-abs}{1.2}
\setval{4b.concur.minus.gdpo-noaux.creative-writing-v3}{+5.4}
\setval{4b.concur.minus.gdpo-noaux.creative-writing-v3-abs}{5.4}
\setval{4b.concur.minus.gdpo-noaux.guidebench}{+1.4}
\setval{4b.concur.minus.gdpo-noaux.guidebench-abs}{1.4}
\setval{4b.concur.minus.gdpo-noaux.humaneval-plus}{$-$3.0}
\setval{4b.concur.minus.gdpo-noaux.humaneval-plus-abs}{3.0}
\setval{4b.concur.minus.gdpo-noaux.if}{+1.6}
\setval{4b.concur.minus.gdpo-noaux.if-abs}{1.6}
\setval{4b.concur.minus.gdpo-noaux.ifbench}{+0.0}
\setval{4b.concur.minus.gdpo-noaux.ifbench-abs}{0.0}
\setval{4b.concur.minus.gdpo-noaux.ifeval}{+3.4}
\setval{4b.concur.minus.gdpo-noaux.ifeval-abs}{3.4}
\setval{4b.concur.minus.gdpo-noaux.judge2.avg}{+2.5}
\setval{4b.concur.minus.gdpo-noaux.judge2.avg-abs}{2.5}
\setval{4b.concur.minus.gdpo-noaux.judge2.creative-writing-v3}{+9.0}
\setval{4b.concur.minus.gdpo-noaux.judge2.creative-writing-v3-abs}{9.0}
\setval{4b.concur.minus.gdpo-noaux.judge2.longbench-write}{+15.9}
\setval{4b.concur.minus.gdpo-noaux.judge2.longbench-write-abs}{15.9}
\setval{4b.concur.minus.gdpo-noaux.judge2.writing}{+11.6}
\setval{4b.concur.minus.gdpo-noaux.judge2.writing-abs}{11.6}
\setval{4b.concur.minus.gdpo-noaux.judge2.writingbench}{+10.0}
\setval{4b.concur.minus.gdpo-noaux.judge2.writingbench-abs}{10.0}
\setval{4b.concur.minus.gdpo-noaux.livecodebench}{+0.4}
\setval{4b.concur.minus.gdpo-noaux.livecodebench-abs}{0.4}
\setval{4b.concur.minus.gdpo-noaux.longbench-write}{+11.9}
\setval{4b.concur.minus.gdpo-noaux.longbench-write-abs}{11.9}
\setval{4b.concur.minus.gdpo-noaux.math}{+0.2}
\setval{4b.concur.minus.gdpo-noaux.math-abs}{0.2}
\setval{4b.concur.minus.gdpo-noaux.math500}{+0.5}
\setval{4b.concur.minus.gdpo-noaux.math500-abs}{0.5}
\setval{4b.concur.minus.gdpo-noaux.mbpp-plus}{$-$0.8}
\setval{4b.concur.minus.gdpo-noaux.mbpp-plus-abs}{0.8}
\setval{4b.concur.minus.gdpo-noaux.minerva}{$-$0.3}
\setval{4b.concur.minus.gdpo-noaux.minerva-abs}{0.3}
\setval{4b.concur.minus.gdpo-noaux.olympiad-bench}{+0.4}
\setval{4b.concur.minus.gdpo-noaux.olympiad-bench-abs}{0.4}
\setval{4b.concur.minus.gdpo-noaux.writing}{+9.8}
\setval{4b.concur.minus.gdpo-noaux.writing-abs}{9.8}
\setval{4b.concur.minus.gdpo-noaux.writingbench}{+12.1}
\setval{4b.concur.minus.gdpo-noaux.writingbench-abs}{12.1}
\setval{4b.concur.minus.gdpo-w0.5.aime24}{+5.9}
\setval{4b.concur.minus.gdpo-w0.5.aime24-abs}{5.9}
\setval{4b.concur.minus.gdpo-w0.5.aime25}{+8.4}
\setval{4b.concur.minus.gdpo-w0.5.aime25-abs}{8.4}
\setval{4b.concur.minus.gdpo-w0.5.amc23}{+11.0}
\setval{4b.concur.minus.gdpo-w0.5.amc23-abs}{11.0}
\setval{4b.concur.minus.gdpo-w0.5.avg}{+4.2}
\setval{4b.concur.minus.gdpo-w0.5.avg-abs}{4.2}
\setval{4b.concur.minus.gdpo-w0.5.code}{+0.6}
\setval{4b.concur.minus.gdpo-w0.5.code-abs}{0.6}
\setval{4b.concur.minus.gdpo-w0.5.creative-writing-v3}{+6.6}
\setval{4b.concur.minus.gdpo-w0.5.creative-writing-v3-abs}{6.6}
\setval{4b.concur.minus.gdpo-w0.5.guidebench}{$-$2.2}
\setval{4b.concur.minus.gdpo-w0.5.guidebench-abs}{2.2}
\setval{4b.concur.minus.gdpo-w0.5.humaneval-plus}{+0.4}
\setval{4b.concur.minus.gdpo-w0.5.humaneval-plus-abs}{0.4}
\setval{4b.concur.minus.gdpo-w0.5.if}{+0.0}
\setval{4b.concur.minus.gdpo-w0.5.if-abs}{0.0}
\setval{4b.concur.minus.gdpo-w0.5.ifbench}{+1.0}
\setval{4b.concur.minus.gdpo-w0.5.ifbench-abs}{1.0}
\setval{4b.concur.minus.gdpo-w0.5.ifeval}{+1.3}
\setval{4b.concur.minus.gdpo-w0.5.ifeval-abs}{1.3}
\setval{4b.concur.minus.gdpo-w0.5.judge2.avg}{+3.3}
\setval{4b.concur.minus.gdpo-w0.5.judge2.avg-abs}{3.3}
\setval{4b.concur.minus.gdpo-w0.5.judge2.creative-writing-v3}{+2.3}
\setval{4b.concur.minus.gdpo-w0.5.judge2.creative-writing-v3-abs}{2.3}
\setval{4b.concur.minus.gdpo-w0.5.judge2.longbench-write}{+6.1}
\setval{4b.concur.minus.gdpo-w0.5.judge2.longbench-write-abs}{6.1}
\setval{4b.concur.minus.gdpo-w0.5.judge2.writing}{+3.6}
\setval{4b.concur.minus.gdpo-w0.5.judge2.writing-abs}{3.6}
\setval{4b.concur.minus.gdpo-w0.5.judge2.writingbench}{+2.4}
\setval{4b.concur.minus.gdpo-w0.5.judge2.writingbench-abs}{2.4}
\setval{4b.concur.minus.gdpo-w0.5.livecodebench}{+1.0}
\setval{4b.concur.minus.gdpo-w0.5.livecodebench-abs}{1.0}
\setval{4b.concur.minus.gdpo-w0.5.longbench-write}{+8.9}
\setval{4b.concur.minus.gdpo-w0.5.longbench-write-abs}{8.9}
\setval{4b.concur.minus.gdpo-w0.5.math}{+6.2}
\setval{4b.concur.minus.gdpo-w0.5.math-abs}{6.2}
\setval{4b.concur.minus.gdpo-w0.5.math500}{+3.5}
\setval{4b.concur.minus.gdpo-w0.5.math500-abs}{3.5}
\setval{4b.concur.minus.gdpo-w0.5.mbpp-plus}{+0.3}
\setval{4b.concur.minus.gdpo-w0.5.mbpp-plus-abs}{0.3}
\setval{4b.concur.minus.gdpo-w0.5.minerva}{+3.1}
\setval{4b.concur.minus.gdpo-w0.5.minerva-abs}{3.1}
\setval{4b.concur.minus.gdpo-w0.5.olympiad-bench}{+5.1}
\setval{4b.concur.minus.gdpo-w0.5.olympiad-bench-abs}{5.1}
\setval{4b.concur.minus.gdpo-w0.5.writing}{+8.1}
\setval{4b.concur.minus.gdpo-w0.5.writing-abs}{8.1}
\setval{4b.concur.minus.gdpo-w0.5.writingbench}{+8.7}
\setval{4b.concur.minus.gdpo-w0.5.writingbench-abs}{8.7}
\setval{4b.concur.minus.gdpo.aime24}{+11.5}
\setval{4b.concur.minus.gdpo.aime24-abs}{11.5}
\setval{4b.concur.minus.gdpo.aime25}{+8.9}
\setval{4b.concur.minus.gdpo.aime25-abs}{8.9}
\setval{4b.concur.minus.gdpo.amc23}{+9.8}
\setval{4b.concur.minus.gdpo.amc23-abs}{9.8}
\setval{4b.concur.minus.gdpo.avg}{+4.5}
\setval{4b.concur.minus.gdpo.avg-abs}{4.5}
\setval{4b.concur.minus.gdpo.code}{+0.7}
\setval{4b.concur.minus.gdpo.code-abs}{0.7}
\setval{4b.concur.minus.gdpo.creative-writing-v3}{$-$1.3}
\setval{4b.concur.minus.gdpo.creative-writing-v3-abs}{1.3}
\setval{4b.concur.minus.gdpo.guidebench}{$-$3.3}
\setval{4b.concur.minus.gdpo.guidebench-abs}{3.3}
\setval{4b.concur.minus.gdpo.humaneval-plus}{+0.6}
\setval{4b.concur.minus.gdpo.humaneval-plus-abs}{0.6}
\setval{4b.concur.minus.gdpo.if}{+1.0}
\setval{4b.concur.minus.gdpo.if-abs}{1.0}
\setval{4b.concur.minus.gdpo.ifbench}{+4.0}
\setval{4b.concur.minus.gdpo.ifbench-abs}{4.0}
\setval{4b.concur.minus.gdpo.ifeval}{+2.4}
\setval{4b.concur.minus.gdpo.ifeval-abs}{2.4}
\setval{4b.concur.minus.gdpo.judge2.avg}{+3.9}
\setval{4b.concur.minus.gdpo.judge2.avg-abs}{3.9}
\setval{4b.concur.minus.gdpo.judge2.creative-writing-v3}{$-$2.0}
\setval{4b.concur.minus.gdpo.judge2.creative-writing-v3-abs}{2.0}
\setval{4b.concur.minus.gdpo.judge2.longbench-write}{+4.3}
\setval{4b.concur.minus.gdpo.judge2.longbench-write-abs}{4.3}
\setval{4b.concur.minus.gdpo.judge2.writing}{+1.4}
\setval{4b.concur.minus.gdpo.judge2.writing-abs}{1.4}
\setval{4b.concur.minus.gdpo.judge2.writingbench}{+2.0}
\setval{4b.concur.minus.gdpo.judge2.writingbench-abs}{2.0}
\setval{4b.concur.minus.gdpo.livecodebench}{+0.7}
\setval{4b.concur.minus.gdpo.livecodebench-abs}{0.7}
\setval{4b.concur.minus.gdpo.longbench-write}{+7.3}
\setval{4b.concur.minus.gdpo.longbench-write-abs}{7.3}
\setval{4b.concur.minus.gdpo.math}{+8.1}
\setval{4b.concur.minus.gdpo.math-abs}{8.1}
\setval{4b.concur.minus.gdpo.math500}{+6.0}
\setval{4b.concur.minus.gdpo.math500-abs}{6.0}
\setval{4b.concur.minus.gdpo.mbpp-plus}{+0.6}
\setval{4b.concur.minus.gdpo.mbpp-plus-abs}{0.6}
\setval{4b.concur.minus.gdpo.minerva}{+3.7}
\setval{4b.concur.minus.gdpo.minerva-abs}{3.7}
\setval{4b.concur.minus.gdpo.olympiad-bench}{+8.9}
\setval{4b.concur.minus.gdpo.olympiad-bench-abs}{8.9}
\setval{4b.concur.minus.gdpo.writing}{+4.5}
\setval{4b.concur.minus.gdpo.writing-abs}{4.5}
\setval{4b.concur.minus.gdpo.writingbench}{+7.5}
\setval{4b.concur.minus.gdpo.writingbench-abs}{7.5}
\setval{4b.concur.minus.geoma.aime24}{+9.0}
\setval{4b.concur.minus.geoma.aime24-abs}{9.0}
\setval{4b.concur.minus.geoma.aime25}{+8.1}
\setval{4b.concur.minus.geoma.aime25-abs}{8.1}
\setval{4b.concur.minus.geoma.amc23}{+14.0}
\setval{4b.concur.minus.geoma.amc23-abs}{14.0}
\setval{4b.concur.minus.geoma.avg}{+5.0}
\setval{4b.concur.minus.geoma.avg-abs}{5.0}
\setval{4b.concur.minus.geoma.code}{+1.0}
\setval{4b.concur.minus.geoma.code-abs}{1.0}
\setval{4b.concur.minus.geoma.creative-writing-v3}{$-$1.1}
\setval{4b.concur.minus.geoma.creative-writing-v3-abs}{1.1}
\setval{4b.concur.minus.geoma.guidebench}{$-$1.2}
\setval{4b.concur.minus.geoma.guidebench-abs}{1.2}
\setval{4b.concur.minus.geoma.humaneval-plus}{+2.2}
\setval{4b.concur.minus.geoma.humaneval-plus-abs}{2.2}
\setval{4b.concur.minus.geoma.if}{+1.2}
\setval{4b.concur.minus.geoma.if-abs}{1.2}
\setval{4b.concur.minus.geoma.ifbench}{+4.3}
\setval{4b.concur.minus.geoma.ifbench-abs}{4.3}
\setval{4b.concur.minus.geoma.ifeval}{+0.4}
\setval{4b.concur.minus.geoma.ifeval-abs}{0.4}
\setval{4b.concur.minus.geoma.judge2.avg}{+4.6}
\setval{4b.concur.minus.geoma.judge2.avg-abs}{4.6}
\setval{4b.concur.minus.geoma.judge2.creative-writing-v3}{$-$0.8}
\setval{4b.concur.minus.geoma.judge2.creative-writing-v3-abs}{0.8}
\setval{4b.concur.minus.geoma.judge2.longbench-write}{+9.0}
\setval{4b.concur.minus.geoma.judge2.longbench-write-abs}{9.0}
\setval{4b.concur.minus.geoma.judge2.writing}{+4.5}
\setval{4b.concur.minus.geoma.judge2.writing-abs}{4.5}
\setval{4b.concur.minus.geoma.judge2.writingbench}{+5.4}
\setval{4b.concur.minus.geoma.judge2.writingbench-abs}{5.4}
\setval{4b.concur.minus.geoma.livecodebench}{$-$0.5}
\setval{4b.concur.minus.geoma.livecodebench-abs}{0.5}
\setval{4b.concur.minus.geoma.longbench-write}{+10.5}
\setval{4b.concur.minus.geoma.longbench-write-abs}{10.5}
\setval{4b.concur.minus.geoma.math}{+8.3}
\setval{4b.concur.minus.geoma.math-abs}{8.3}
\setval{4b.concur.minus.geoma.math500}{+6.4}
\setval{4b.concur.minus.geoma.math500-abs}{6.4}
\setval{4b.concur.minus.geoma.mbpp-plus}{+1.3}
\setval{4b.concur.minus.geoma.mbpp-plus-abs}{1.3}
\setval{4b.concur.minus.geoma.minerva}{+2.3}
\setval{4b.concur.minus.geoma.minerva-abs}{2.3}
\setval{4b.concur.minus.geoma.olympiad-bench}{+9.7}
\setval{4b.concur.minus.geoma.olympiad-bench-abs}{9.7}
\setval{4b.concur.minus.geoma.writing}{+6.4}
\setval{4b.concur.minus.geoma.writing-abs}{6.4}
\setval{4b.concur.minus.geoma.writingbench}{+9.9}
\setval{4b.concur.minus.geoma.writingbench-abs}{9.9}
\setval{4b.concur.minus.grpo-nodiv.aime24}{+8.1}
\setval{4b.concur.minus.grpo-nodiv.aime24-abs}{8.1}
\setval{4b.concur.minus.grpo-nodiv.aime25}{+10.4}
\setval{4b.concur.minus.grpo-nodiv.aime25-abs}{10.4}
\setval{4b.concur.minus.grpo-nodiv.amc23}{+14.8}
\setval{4b.concur.minus.grpo-nodiv.amc23-abs}{14.8}
\setval{4b.concur.minus.grpo-nodiv.avg}{+5.8}
\setval{4b.concur.minus.grpo-nodiv.avg-abs}{5.8}
\setval{4b.concur.minus.grpo-nodiv.code}{+1.0}
\setval{4b.concur.minus.grpo-nodiv.code-abs}{1.0}
\setval{4b.concur.minus.grpo-nodiv.creative-writing-v3}{+2.2}
\setval{4b.concur.minus.grpo-nodiv.creative-writing-v3-abs}{2.2}
\setval{4b.concur.minus.grpo-nodiv.guidebench}{$-$1.4}
\setval{4b.concur.minus.grpo-nodiv.guidebench-abs}{1.4}
\setval{4b.concur.minus.grpo-nodiv.humaneval-plus}{+2.0}
\setval{4b.concur.minus.grpo-nodiv.humaneval-plus-abs}{2.0}
\setval{4b.concur.minus.grpo-nodiv.if}{+2.0}
\setval{4b.concur.minus.grpo-nodiv.if-abs}{2.0}
\setval{4b.concur.minus.grpo-nodiv.ifbench}{+6.7}
\setval{4b.concur.minus.grpo-nodiv.ifbench-abs}{6.7}
\setval{4b.concur.minus.grpo-nodiv.ifeval}{+0.8}
\setval{4b.concur.minus.grpo-nodiv.ifeval-abs}{0.8}
\setval{4b.concur.minus.grpo-nodiv.judge2.avg}{+5.1}
\setval{4b.concur.minus.grpo-nodiv.judge2.avg-abs}{5.1}
\setval{4b.concur.minus.grpo-nodiv.judge2.creative-writing-v3}{+1.8}
\setval{4b.concur.minus.grpo-nodiv.judge2.creative-writing-v3-abs}{1.8}
\setval{4b.concur.minus.grpo-nodiv.judge2.longbench-write}{+8.0}
\setval{4b.concur.minus.grpo-nodiv.judge2.longbench-write-abs}{8.0}
\setval{4b.concur.minus.grpo-nodiv.judge2.writing}{+5.5}
\setval{4b.concur.minus.grpo-nodiv.judge2.writing-abs}{5.5}
\setval{4b.concur.minus.grpo-nodiv.judge2.writingbench}{+6.5}
\setval{4b.concur.minus.grpo-nodiv.judge2.writingbench-abs}{6.5}
\setval{4b.concur.minus.grpo-nodiv.livecodebench}{+0.7}
\setval{4b.concur.minus.grpo-nodiv.livecodebench-abs}{0.7}
\setval{4b.concur.minus.grpo-nodiv.longbench-write}{+10.3}
\setval{4b.concur.minus.grpo-nodiv.longbench-write-abs}{10.3}
\setval{4b.concur.minus.grpo-nodiv.math}{+8.6}
\setval{4b.concur.minus.grpo-nodiv.math-abs}{8.6}
\setval{4b.concur.minus.grpo-nodiv.math500}{+5.8}
\setval{4b.concur.minus.grpo-nodiv.math500-abs}{5.8}
\setval{4b.concur.minus.grpo-nodiv.mbpp-plus}{+0.3}
\setval{4b.concur.minus.grpo-nodiv.mbpp-plus-abs}{0.3}
\setval{4b.concur.minus.grpo-nodiv.minerva}{+3.4}
\setval{4b.concur.minus.grpo-nodiv.minerva-abs}{3.4}
\setval{4b.concur.minus.grpo-nodiv.olympiad-bench}{+8.8}
\setval{4b.concur.minus.grpo-nodiv.olympiad-bench-abs}{8.8}
\setval{4b.concur.minus.grpo-nodiv.writing}{+8.8}
\setval{4b.concur.minus.grpo-nodiv.writing-abs}{8.8}
\setval{4b.concur.minus.grpo-nodiv.writingbench}{+13.8}
\setval{4b.concur.minus.grpo-nodiv.writingbench-abs}{13.8}
\setval{4b.concur.minus.grpo.aime24}{+8.5}
\setval{4b.concur.minus.grpo.aime24-abs}{8.5}
\setval{4b.concur.minus.grpo.aime25}{+6.1}
\setval{4b.concur.minus.grpo.aime25-abs}{6.1}
\setval{4b.concur.minus.grpo.amc23}{+14.7}
\setval{4b.concur.minus.grpo.amc23-abs}{14.7}
\setval{4b.concur.minus.grpo.avg}{+4.1}
\setval{4b.concur.minus.grpo.avg-abs}{4.1}
\setval{4b.concur.minus.grpo.code}{+0.4}
\setval{4b.concur.minus.grpo.code-abs}{0.4}
\setval{4b.concur.minus.grpo.creative-writing-v3}{$-$2.6}
\setval{4b.concur.minus.grpo.creative-writing-v3-abs}{2.6}
\setval{4b.concur.minus.grpo.guidebench}{$-$0.6}
\setval{4b.concur.minus.grpo.guidebench-abs}{0.6}
\setval{4b.concur.minus.grpo.humaneval-plus}{$-$0.3}
\setval{4b.concur.minus.grpo.humaneval-plus-abs}{0.3}
\setval{4b.concur.minus.grpo.if}{+1.3}
\setval{4b.concur.minus.grpo.if-abs}{1.3}
\setval{4b.concur.minus.grpo.ifbench}{+2.7}
\setval{4b.concur.minus.grpo.ifbench-abs}{2.7}
\setval{4b.concur.minus.grpo.ifeval}{+1.7}
\setval{4b.concur.minus.grpo.ifeval-abs}{1.7}
\setval{4b.concur.minus.grpo.judge2.avg}{+3.9}
\setval{4b.concur.minus.grpo.judge2.avg-abs}{3.9}
\setval{4b.concur.minus.grpo.judge2.creative-writing-v3}{$-$2.7}
\setval{4b.concur.minus.grpo.judge2.creative-writing-v3-abs}{2.7}
\setval{4b.concur.minus.grpo.judge2.longbench-write}{+5.9}
\setval{4b.concur.minus.grpo.judge2.longbench-write-abs}{5.9}
\setval{4b.concur.minus.grpo.judge2.writing}{+1.9}
\setval{4b.concur.minus.grpo.judge2.writing-abs}{1.9}
\setval{4b.concur.minus.grpo.judge2.writingbench}{+2.6}
\setval{4b.concur.minus.grpo.judge2.writingbench-abs}{2.6}
\setval{4b.concur.minus.grpo.livecodebench}{+1.1}
\setval{4b.concur.minus.grpo.livecodebench-abs}{1.1}
\setval{4b.concur.minus.grpo.longbench-write}{+7.1}
\setval{4b.concur.minus.grpo.longbench-write-abs}{7.1}
\setval{4b.concur.minus.grpo.math}{+7.9}
\setval{4b.concur.minus.grpo.math-abs}{7.9}
\setval{4b.concur.minus.grpo.math500}{+5.0}
\setval{4b.concur.minus.grpo.math500-abs}{5.0}
\setval{4b.concur.minus.grpo.mbpp-plus}{+0.3}
\setval{4b.concur.minus.grpo.mbpp-plus-abs}{0.3}
\setval{4b.concur.minus.grpo.minerva}{+3.9}
\setval{4b.concur.minus.grpo.minerva-abs}{3.9}
\setval{4b.concur.minus.grpo.olympiad-bench}{+9.0}
\setval{4b.concur.minus.grpo.olympiad-bench-abs}{9.0}
\setval{4b.concur.minus.grpo.writing}{+3.2}
\setval{4b.concur.minus.grpo.writing-abs}{3.2}
\setval{4b.concur.minus.grpo.writingbench}{+5.2}
\setval{4b.concur.minus.grpo.writingbench-abs}{5.2}
\setval{4b.concur.minus.mt-grpo.aime24}{+9.8}
\setval{4b.concur.minus.mt-grpo.aime24-abs}{9.8}
\setval{4b.concur.minus.mt-grpo.aime25}{+12.6}
\setval{4b.concur.minus.mt-grpo.aime25-abs}{12.6}
\setval{4b.concur.minus.mt-grpo.amc23}{+9.6}
\setval{4b.concur.minus.mt-grpo.amc23-abs}{9.6}
\setval{4b.concur.minus.mt-grpo.avg}{+5.7}
\setval{4b.concur.minus.mt-grpo.avg-abs}{5.7}
\setval{4b.concur.minus.mt-grpo.code}{+2.4}
\setval{4b.concur.minus.mt-grpo.code-abs}{2.4}
\setval{4b.concur.minus.mt-grpo.creative-writing-v3}{+0.2}
\setval{4b.concur.minus.mt-grpo.creative-writing-v3-abs}{0.2}
\setval{4b.concur.minus.mt-grpo.guidebench}{$-$2.3}
\setval{4b.concur.minus.mt-grpo.guidebench-abs}{2.3}
\setval{4b.concur.minus.mt-grpo.humaneval-plus}{+3.4}
\setval{4b.concur.minus.mt-grpo.humaneval-plus-abs}{3.4}
\setval{4b.concur.minus.mt-grpo.if}{+1.7}
\setval{4b.concur.minus.mt-grpo.if-abs}{1.7}
\setval{4b.concur.minus.mt-grpo.ifbench}{+6.0}
\setval{4b.concur.minus.mt-grpo.ifbench-abs}{6.0}
\setval{4b.concur.minus.mt-grpo.ifeval}{+1.4}
\setval{4b.concur.minus.mt-grpo.ifeval-abs}{1.4}
\setval{4b.concur.minus.mt-grpo.judge2.avg}{+4.9}
\setval{4b.concur.minus.mt-grpo.judge2.avg-abs}{4.9}
\setval{4b.concur.minus.mt-grpo.judge2.creative-writing-v3}{$-$1.7}
\setval{4b.concur.minus.mt-grpo.judge2.creative-writing-v3-abs}{1.7}
\setval{4b.concur.minus.mt-grpo.judge2.longbench-write}{+5.9}
\setval{4b.concur.minus.mt-grpo.judge2.longbench-write-abs}{5.9}
\setval{4b.concur.minus.mt-grpo.judge2.writing}{+2.3}
\setval{4b.concur.minus.mt-grpo.judge2.writing-abs}{2.3}
\setval{4b.concur.minus.mt-grpo.judge2.writingbench}{+2.8}
\setval{4b.concur.minus.mt-grpo.judge2.writingbench-abs}{2.8}
\setval{4b.concur.minus.mt-grpo.livecodebench}{+2.5}
\setval{4b.concur.minus.mt-grpo.livecodebench-abs}{2.5}
\setval{4b.concur.minus.mt-grpo.longbench-write}{+9.5}
\setval{4b.concur.minus.mt-grpo.longbench-write-abs}{9.5}
\setval{4b.concur.minus.mt-grpo.math}{+9.1}
\setval{4b.concur.minus.mt-grpo.math-abs}{9.1}
\setval{4b.concur.minus.mt-grpo.math500}{+8.3}
\setval{4b.concur.minus.mt-grpo.math500-abs}{8.3}
\setval{4b.concur.minus.mt-grpo.mbpp-plus}{+1.5}
\setval{4b.concur.minus.mt-grpo.mbpp-plus-abs}{1.5}
\setval{4b.concur.minus.mt-grpo.minerva}{+3.2}
\setval{4b.concur.minus.mt-grpo.minerva-abs}{3.2}
\setval{4b.concur.minus.mt-grpo.olympiad-bench}{+11.1}
\setval{4b.concur.minus.mt-grpo.olympiad-bench-abs}{11.1}
\setval{4b.concur.minus.mt-grpo.writing}{+6.3}
\setval{4b.concur.minus.mt-grpo.writing-abs}{6.3}
\setval{4b.concur.minus.mt-grpo.writingbench}{+9.3}
\setval{4b.concur.minus.mt-grpo.writingbench-abs}{9.3}
\setval{4b.concur.minus.papo.aime24}{+11.2}
\setval{4b.concur.minus.papo.aime24-abs}{11.2}
\setval{4b.concur.minus.papo.aime25}{+11.2}
\setval{4b.concur.minus.papo.aime25-abs}{11.2}
\setval{4b.concur.minus.papo.amc23}{+15.9}
\setval{4b.concur.minus.papo.amc23-abs}{15.9}
\setval{4b.concur.minus.papo.avg}{+6.2}
\setval{4b.concur.minus.papo.avg-abs}{6.2}
\setval{4b.concur.minus.papo.code}{+0.6}
\setval{4b.concur.minus.papo.code-abs}{0.6}
\setval{4b.concur.minus.papo.creative-writing-v3}{$-$1.8}
\setval{4b.concur.minus.papo.creative-writing-v3-abs}{1.8}
\setval{4b.concur.minus.papo.guidebench}{$-$3.4}
\setval{4b.concur.minus.papo.guidebench-abs}{3.4}
\setval{4b.concur.minus.papo.humaneval-plus}{$-$0.6}
\setval{4b.concur.minus.papo.humaneval-plus-abs}{0.6}
\setval{4b.concur.minus.papo.if}{+2.0}
\setval{4b.concur.minus.papo.if-abs}{2.0}
\setval{4b.concur.minus.papo.ifbench}{+7.0}
\setval{4b.concur.minus.papo.ifbench-abs}{7.0}
\setval{4b.concur.minus.papo.ifeval}{+2.3}
\setval{4b.concur.minus.papo.ifeval-abs}{2.3}
\setval{4b.concur.minus.papo.judge2.avg}{+5.4}
\setval{4b.concur.minus.papo.judge2.avg-abs}{5.4}
\setval{4b.concur.minus.papo.judge2.creative-writing-v3}{$-$1.9}
\setval{4b.concur.minus.papo.judge2.creative-writing-v3-abs}{1.9}
\setval{4b.concur.minus.papo.judge2.longbench-write}{+9.1}
\setval{4b.concur.minus.papo.judge2.longbench-write-abs}{9.1}
\setval{4b.concur.minus.papo.judge2.writing}{+3.7}
\setval{4b.concur.minus.papo.judge2.writing-abs}{3.7}
\setval{4b.concur.minus.papo.judge2.writingbench}{+3.9}
\setval{4b.concur.minus.papo.judge2.writingbench-abs}{3.9}
\setval{4b.concur.minus.papo.livecodebench}{+1.6}
\setval{4b.concur.minus.papo.livecodebench-abs}{1.6}
\setval{4b.concur.minus.papo.longbench-write}{+12.3}
\setval{4b.concur.minus.papo.longbench-write-abs}{12.3}
\setval{4b.concur.minus.papo.math}{+10.3}
\setval{4b.concur.minus.papo.math-abs}{10.3}
\setval{4b.concur.minus.papo.math500}{+8.8}
\setval{4b.concur.minus.papo.math500-abs}{8.8}
\setval{4b.concur.minus.papo.mbpp-plus}{+0.8}
\setval{4b.concur.minus.papo.mbpp-plus-abs}{0.8}
\setval{4b.concur.minus.papo.minerva}{+2.7}
\setval{4b.concur.minus.papo.minerva-abs}{2.7}
\setval{4b.concur.minus.papo.olympiad-bench}{+12.1}
\setval{4b.concur.minus.papo.olympiad-bench-abs}{12.1}
\setval{4b.concur.minus.papo.writing}{+7.8}
\setval{4b.concur.minus.papo.writing-abs}{7.8}
\setval{4b.concur.minus.papo.writingbench}{+12.9}
\setval{4b.concur.minus.papo.writingbench-abs}{12.9}
\setval{4b.concur.minus.saw.aime24}{+8.0}
\setval{4b.concur.minus.saw.aime24-abs}{8.0}
\setval{4b.concur.minus.saw.aime25}{+7.6}
\setval{4b.concur.minus.saw.aime25-abs}{7.6}
\setval{4b.concur.minus.saw.amc23}{+12.7}
\setval{4b.concur.minus.saw.amc23-abs}{12.7}
\setval{4b.concur.minus.saw.avg}{+4.6}
\setval{4b.concur.minus.saw.avg-abs}{4.6}
\setval{4b.concur.minus.saw.code}{+1.8}
\setval{4b.concur.minus.saw.code-abs}{1.8}
\setval{4b.concur.minus.saw.creative-writing-v3}{$-$0.4}
\setval{4b.concur.minus.saw.creative-writing-v3-abs}{0.4}
\setval{4b.concur.minus.saw.guidebench}{$-$3.6}
\setval{4b.concur.minus.saw.guidebench-abs}{3.6}
\setval{4b.concur.minus.saw.humaneval-plus}{+1.7}
\setval{4b.concur.minus.saw.humaneval-plus-abs}{1.7}
\setval{4b.concur.minus.saw.if}{$-$0.8}
\setval{4b.concur.minus.saw.if-abs}{0.8}
\setval{4b.concur.minus.saw.ifbench}{+0.0}
\setval{4b.concur.minus.saw.ifbench-abs}{0.0}
\setval{4b.concur.minus.saw.ifeval}{+1.2}
\setval{4b.concur.minus.saw.ifeval-abs}{1.2}
\setval{4b.concur.minus.saw.judge2.avg}{+3.7}
\setval{4b.concur.minus.saw.judge2.avg-abs}{3.7}
\setval{4b.concur.minus.saw.judge2.creative-writing-v3}{$-$3.2}
\setval{4b.concur.minus.saw.judge2.creative-writing-v3-abs}{3.2}
\setval{4b.concur.minus.saw.judge2.longbench-write}{+6.6}
\setval{4b.concur.minus.saw.judge2.longbench-write-abs}{6.6}
\setval{4b.concur.minus.saw.judge2.writing}{+2.0}
\setval{4b.concur.minus.saw.judge2.writing-abs}{2.0}
\setval{4b.concur.minus.saw.judge2.writingbench}{+2.7}
\setval{4b.concur.minus.saw.judge2.writingbench-abs}{2.7}
\setval{4b.concur.minus.saw.livecodebench}{+2.3}
\setval{4b.concur.minus.saw.livecodebench-abs}{2.3}
\setval{4b.concur.minus.saw.longbench-write}{+10.1}
\setval{4b.concur.minus.saw.longbench-write-abs}{10.1}
\setval{4b.concur.minus.saw.math}{+7.8}
\setval{4b.concur.minus.saw.math-abs}{7.8}
\setval{4b.concur.minus.saw.math500}{+7.4}
\setval{4b.concur.minus.saw.math500-abs}{7.4}
\setval{4b.concur.minus.saw.mbpp-plus}{+1.4}
\setval{4b.concur.minus.saw.mbpp-plus-abs}{1.4}
\setval{4b.concur.minus.saw.minerva}{+1.1}
\setval{4b.concur.minus.saw.minerva-abs}{1.1}
\setval{4b.concur.minus.saw.olympiad-bench}{+9.7}
\setval{4b.concur.minus.saw.olympiad-bench-abs}{9.7}
\setval{4b.concur.minus.saw.writing}{+6.4}
\setval{4b.concur.minus.saw.writing-abs}{6.4}
\setval{4b.concur.minus.saw.writingbench}{+9.6}
\setval{4b.concur.minus.saw.writingbench-abs}{9.6}
\setval{4b.concur.passing-at-ceiling.code.early}{0.71}
\setval{4b.concur.passing-at-ceiling.code.late}{0.74}
\setval{4b.concur.passing-at-ceiling.code.run}{0.73}
\setval{4b.concur.passing-at-ceiling.if.early}{0.29}
\setval{4b.concur.passing-at-ceiling.if.late}{0.55}
\setval{4b.concur.passing-at-ceiling.if.run}{0.48}
\setval{4b.concur.passing-at-ceiling.math.early}{1.00}
\setval{4b.concur.passing-at-ceiling.math.late}{1.00}
\setval{4b.concur.passing-at-ceiling.math.run}{1.00}
\setval{4b.concur.passing-at-ceiling.writing.early}{0.00}
\setval{4b.concur.passing-at-ceiling.writing.late}{0.02}
\setval{4b.concur.passing-at-ceiling.writing.run}{0.01}
\setval{4b.concur.primary-cosine.code.early}{0.98}
\setval{4b.concur.primary-cosine.code.late}{0.96}
\setval{4b.concur.primary-cosine.code.run}{0.96}
\setval{4b.concur.primary-cosine.if.early}{0.85}
\setval{4b.concur.primary-cosine.if.late}{0.79}
\setval{4b.concur.primary-cosine.if.run}{0.82}
\setval{4b.concur.primary-cosine.math.early}{1.00}
\setval{4b.concur.primary-cosine.math.late}{0.97}
\setval{4b.concur.primary-cosine.math.run}{0.98}
\setval{4b.concur.primary-cosine.writing.early}{0.94}
\setval{4b.concur.primary-cosine.writing.late}{0.92}
\setval{4b.concur.primary-cosine.writing.run}{0.92}
\setval{4b.concur.primary-level.code.early}{0.53}
\setval{4b.concur.primary-level.code.late}{0.75}
\setval{4b.concur.primary-level.code.run}{0.69}
\setval{4b.concur.primary-level.if.early}{0.32}
\setval{4b.concur.primary-level.if.late}{0.74}
\setval{4b.concur.primary-level.if.run}{0.63}
\setval{4b.concur.primary-level.math.early}{0.47}
\setval{4b.concur.primary-level.math.late}{0.71}
\setval{4b.concur.primary-level.math.run}{0.65}
\setval{4b.concur.primary-level.writing.early}{0.53}
\setval{4b.concur.primary-level.writing.late}{0.80}
\setval{4b.concur.primary-level.writing.run}{0.75}
\setval{4b.concur.solved.code.early}{0.09}
\setval{4b.concur.solved.code.late}{0.03}
\setval{4b.concur.solved.code.run}{0.04}
\setval{4b.concur.solved.if.early}{0.03}
\setval{4b.concur.solved.if.late}{0.21}
\setval{4b.concur.solved.if.run}{0.13}
\setval{4b.concur.solved.math.early}{0.12}
\setval{4b.concur.solved.math.late}{0.50}
\setval{4b.concur.solved.math.run}{0.36}
\setval{4b.concur.solved.writing.early}{0.00}
\setval{4b.concur.solved.writing.late}{0.00}
\setval{4b.concur.solved.writing.run}{0.00}
\setval{4b.concur.trust.code.agreement}{+0.012}
\setval{4b.concur.trust.code.errors}{+4.2}
\setval{4b.concur.trust.code.first-free}{27}
\setval{4b.concur.trust.code.free-share}{0.85}
\setval{4b.concur.trust.if.agreement}{+0.018}
\setval{4b.concur.trust.if.errors}{+7.4}
\setval{4b.concur.trust.if.first-free}{6}
\setval{4b.concur.trust.if.free-share}{0.95}
\setval{4b.concur.trust.math.first-free}{--}
\setval{4b.concur.trust.math.free-share}{0.00}
\setval{4b.concur.trust.writing-rm.agreement}{+0.061}
\setval{4b.concur.trust.writing-rm.errors}{+37.4}
\setval{4b.concur.trust.writing-rm.first-free}{2}
\setval{4b.concur.trust.writing-rm.free-share}{1.00}
\setval{4b.concur.trust.writing.agreement}{$-$0.054}
\setval{4b.concur.trust.writing.errors}{$-$36.8}
\setval{4b.concur.trust.writing.first-free}{--}
\setval{4b.concur.trust.writing.free-share}{0.00}
\setval{4b.concur.update-share.code.early}{0.20}
\setval{4b.concur.update-share.code.late}{0.25}
\setval{4b.concur.update-share.code.run}{0.23}
\setval{4b.concur.update-share.if.early}{0.21}
\setval{4b.concur.update-share.if.late}{0.18}
\setval{4b.concur.update-share.if.run}{0.22}
\setval{4b.concur.update-share.math.early}{0.17}
\setval{4b.concur.update-share.math.late}{0.15}
\setval{4b.concur.update-share.math.run}{0.18}
\setval{4b.concur.update-share.writing.early}{0.42}
\setval{4b.concur.update-share.writing.late}{0.41}
\setval{4b.concur.update-share.writing.run}{0.36}
\setval{4b.estimators.aux-full.code.run-max}{0.98}
\setval{4b.estimators.aux-full.code.run-min}{0.91}
\setval{4b.estimators.aux-full.if.run-max}{0.88}
\setval{4b.estimators.aux-full.if.run-min}{0.83}
\setval{4b.estimators.aux-full.math.run-max}{0.94}
\setval{4b.estimators.aux-full.math.run-min}{0.83}
\setval{4b.estimators.aux-full.writing-rm.run-max}{0.53}
\setval{4b.estimators.aux-full.writing-rm.run-min}{0.41}
\setval{4b.estimators.aux-full.writing.run-max}{0.78}
\setval{4b.estimators.aux-full.writing.run-min}{0.63}
\setval{4b.estimators.avg-spread}{1.6}
\setval{4b.estimators.bench.aime24.max}{12.5}
\setval{4b.estimators.bench.aime24.min}{9.1}
\setval{4b.estimators.bench.aime25.max}{12.6}
\setval{4b.estimators.bench.aime25.min}{6.1}
\setval{4b.estimators.bench.amc23.max}{59.8}
\setval{4b.estimators.bench.amc23.min}{53.5}
\setval{4b.estimators.bench.creative-writing-v3.max}{47.3}
\setval{4b.estimators.bench.creative-writing-v3.min}{44.5}
\setval{4b.estimators.bench.guidebench.max}{81.0}
\setval{4b.estimators.bench.guidebench.min}{77.9}
\setval{4b.estimators.bench.humaneval-plus.max}{83.4}
\setval{4b.estimators.bench.humaneval-plus.min}{79.1}
\setval{4b.estimators.bench.ifbench.max}{40.3}
\setval{4b.estimators.bench.ifbench.min}{33.3}
\setval{4b.estimators.bench.ifeval.max}{87.8}
\setval{4b.estimators.bench.ifeval.min}{85.7}
\setval{4b.estimators.bench.livecodebench.max}{25.7}
\setval{4b.estimators.bench.livecodebench.min}{22.6}
\setval{4b.estimators.bench.longbench-write.max}{69.2}
\setval{4b.estimators.bench.longbench-write.min}{64.0}
\setval{4b.estimators.bench.math500.max}{80.2}
\setval{4b.estimators.bench.math500.min}{76.5}
\setval{4b.estimators.bench.mbpp-plus.max}{71.5}
\setval{4b.estimators.bench.mbpp-plus.min}{70.1}
\setval{4b.estimators.bench.minerva.max}{48.1}
\setval{4b.estimators.bench.minerva.min}{45.3}
\setval{4b.estimators.bench.olympiad-bench.max}{44.7}
\setval{4b.estimators.bench.olympiad-bench.min}{41.5}
\setval{4b.estimators.bench.writingbench.max}{74.3}
\setval{4b.estimators.bench.writingbench.min}{66.5}
\setval{4b.estimators.eval.avg.max}{54.7}
\setval{4b.estimators.eval.avg.min}{52.6}
\setval{4b.estimators.eval.code.max}{59.5}
\setval{4b.estimators.eval.code.min}{57.3}
\setval{4b.estimators.eval.if.max}{69.4}
\setval{4b.estimators.eval.if.min}{66.6}
\setval{4b.estimators.eval.math.max}{41.7}
\setval{4b.estimators.eval.math.min}{39.1}
\setval{4b.estimators.eval.writing.max}{63.6}
\setval{4b.estimators.eval.writing.min}{59.0}
\setval{4b.estimators.minus.gdpo-noaux.avg-max}{$-$2.0}
\setval{4b.estimators.minus.gdpo-noaux.avg-min}{$-$4.1}
\setval{4b.gd2po.aux-change.code}{+0.20}
\setval{4b.gd2po.aux-change.if}{+0.24}
\setval{4b.gd2po.aux-change.math}{+0.14}
\setval{4b.gd2po.aux-change.writing}{+0.18}
\setval{4b.gd2po.aux-change.writing-rm}{+0.36}
\setval{4b.gd2po.aux-full.code.early}{0.73}
\setval{4b.gd2po.aux-full.code.late}{0.96}
\setval{4b.gd2po.aux-full.code.run}{0.95}
\setval{4b.gd2po.aux-full.if.early}{0.64}
\setval{4b.gd2po.aux-full.if.late}{0.90}
\setval{4b.gd2po.aux-full.if.run}{0.85}
\setval{4b.gd2po.aux-full.math.early}{0.64}
\setval{4b.gd2po.aux-full.math.late}{0.91}
\setval{4b.gd2po.aux-full.math.run}{0.87}
\setval{4b.gd2po.aux-full.writing-rm.early}{0.09}
\setval{4b.gd2po.aux-full.writing-rm.late}{0.56}
\setval{4b.gd2po.aux-full.writing-rm.run}{0.44}
\setval{4b.gd2po.aux-full.writing.early}{0.61}
\setval{4b.gd2po.aux-full.writing.late}{0.67}
\setval{4b.gd2po.aux-full.writing.run}{0.63}
\setval{4b.gd2po.aux-level.code.early}{0.92}
\setval{4b.gd2po.aux-level.code.late}{0.99}
\setval{4b.gd2po.aux-level.code.run}{0.99}
\setval{4b.gd2po.aux-level.if.early}{0.82}
\setval{4b.gd2po.aux-level.if.late}{0.94}
\setval{4b.gd2po.aux-level.if.run}{0.92}
\setval{4b.gd2po.aux-level.math.early}{0.93}
\setval{4b.gd2po.aux-level.math.late}{0.99}
\setval{4b.gd2po.aux-level.math.run}{0.98}
\setval{4b.gd2po.aux-level.writing-rm.early}{0.77}
\setval{4b.gd2po.aux-level.writing-rm.late}{0.96}
\setval{4b.gd2po.aux-level.writing-rm.run}{0.93}
\setval{4b.gd2po.aux-level.writing.early}{0.89}
\setval{4b.gd2po.aux-level.writing.late}{0.95}
\setval{4b.gd2po.aux-level.writing.run}{0.93}
\setval{4b.gd2po.bench.aime24}{11.9}
\setval{4b.gd2po.bench.aime25}{6.8}
\setval{4b.gd2po.bench.amc23}{57.7}
\setval{4b.gd2po.bench.creative-writing-v3}{44.7}
\setval{4b.gd2po.bench.guidebench}{79.1}
\setval{4b.gd2po.bench.humaneval-plus}{83.4}
\setval{4b.gd2po.bench.ifbench}{37.7}
\setval{4b.gd2po.bench.ifeval}{87.6}
\setval{4b.gd2po.bench.livecodebench}{23.5}
\setval{4b.gd2po.bench.longbench-write}{68.9}
\setval{4b.gd2po.bench.math500}{77.7}
\setval{4b.gd2po.bench.mbpp-plus}{71.5}
\setval{4b.gd2po.bench.minerva}{47.2}
\setval{4b.gd2po.bench.olympiad-bench}{43.5}
\setval{4b.gd2po.bench.writingbench}{70.2}
\setval{4b.gd2po.eval.avg}{54.1}
\setval{4b.gd2po.eval.code}{59.5}
\setval{4b.gd2po.eval.if}{68.1}
\setval{4b.gd2po.eval.math}{40.8}
\setval{4b.gd2po.eval.writing}{61.3}
\setval{4b.gd2po.gate.code.against.early}{0.03}
\setval{4b.gd2po.gate.code.against.late}{0.02}
\setval{4b.gd2po.gate.code.against.run}{0.02}
\setval{4b.gd2po.gate.code.agreement}{+0.004}
\setval{4b.gd2po.gate.code.ceiling.early}{0.11}
\setval{4b.gd2po.gate.code.ceiling.late}{0.03}
\setval{4b.gd2po.gate.code.ceiling.run}{0.04}
\setval{4b.gd2po.gate.code.cosine.early}{0.84}
\setval{4b.gd2po.gate.code.cosine.late}{0.79}
\setval{4b.gd2po.gate.code.cosine.run}{0.82}
\setval{4b.gd2po.gate.if.against.early}{0.05}
\setval{4b.gd2po.gate.if.against.late}{0.04}
\setval{4b.gd2po.gate.if.against.run}{0.04}
\setval{4b.gd2po.gate.if.agreement}{+0.015}
\setval{4b.gd2po.gate.if.ceiling.early}{0.05}
\setval{4b.gd2po.gate.if.ceiling.late}{0.16}
\setval{4b.gd2po.gate.if.ceiling.run}{0.11}
\setval{4b.gd2po.gate.if.cosine.early}{0.74}
\setval{4b.gd2po.gate.if.cosine.late}{0.51}
\setval{4b.gd2po.gate.if.cosine.run}{0.60}
\setval{4b.gd2po.gate.math.against.early}{0.00}
\setval{4b.gd2po.gate.math.against.late}{0.00}
\setval{4b.gd2po.gate.math.against.run}{0.00}
\setval{4b.gd2po.gate.math.ceiling.early}{0.18}
\setval{4b.gd2po.gate.math.ceiling.late}{0.16}
\setval{4b.gd2po.gate.math.ceiling.run}{0.15}
\setval{4b.gd2po.gate.math.cosine.early}{0.89}
\setval{4b.gd2po.gate.math.cosine.late}{0.91}
\setval{4b.gd2po.gate.math.cosine.run}{0.92}
\setval{4b.gd2po.gate.writing-rm.against.early}{0.19}
\setval{4b.gd2po.gate.writing-rm.against.late}{0.53}
\setval{4b.gd2po.gate.writing-rm.against.run}{0.49}
\setval{4b.gd2po.gate.writing-rm.agreement}{+0.063}
\setval{4b.gd2po.gate.writing-rm.ceiling.early}{0.00}
\setval{4b.gd2po.gate.writing-rm.ceiling.late}{0.00}
\setval{4b.gd2po.gate.writing-rm.ceiling.run}{0.00}
\setval{4b.gd2po.gate.writing-rm.cosine.early}{0.70}
\setval{4b.gd2po.gate.writing-rm.cosine.late}{0.15}
\setval{4b.gd2po.gate.writing-rm.cosine.run}{0.23}
\setval{4b.gd2po.gate.writing.against.early}{0.29}
\setval{4b.gd2po.gate.writing.against.late}{0.62}
\setval{4b.gd2po.gate.writing.against.run}{0.59}
\setval{4b.gd2po.gate.writing.agreement}{$-$0.049}
\setval{4b.gd2po.gate.writing.ceiling.early}{0.00}
\setval{4b.gd2po.gate.writing.ceiling.late}{0.00}
\setval{4b.gd2po.gate.writing.ceiling.run}{0.00}
\setval{4b.gd2po.gate.writing.cosine.early}{0.52}
\setval{4b.gd2po.gate.writing.cosine.late}{0.02}
\setval{4b.gd2po.gate.writing.cosine.run}{0.09}
\setval{4b.gd2po.judge2.avg}{51.6}
\setval{4b.gd2po.judge2.creative-writing-v3}{32.9}
\setval{4b.gd2po.judge2.longbench-write}{61.0}
\setval{4b.gd2po.judge2.writing}{48.8}
\setval{4b.gd2po.judge2.writingbench}{52.5}
\setval{4b.gd2po.last-step}{300}
\setval{4b.gd2po.minus.base.aime24}{+4.1}
\setval{4b.gd2po.minus.base.aime24-abs}{4.1}
\setval{4b.gd2po.minus.base.aime25}{$-$0.3}
\setval{4b.gd2po.minus.base.aime25-abs}{0.3}
\setval{4b.gd2po.minus.base.amc23}{+13.8}
\setval{4b.gd2po.minus.base.amc23-abs}{13.8}
\setval{4b.gd2po.minus.base.avg}{+19.6}
\setval{4b.gd2po.minus.base.avg-abs}{19.6}
\setval{4b.gd2po.minus.base.code}{+14.8}
\setval{4b.gd2po.minus.base.code-abs}{14.8}
\setval{4b.gd2po.minus.base.creative-writing-v3}{+19.9}
\setval{4b.gd2po.minus.base.creative-writing-v3-abs}{19.9}
\setval{4b.gd2po.minus.base.guidebench}{+54.2}
\setval{4b.gd2po.minus.base.guidebench-abs}{54.2}
\setval{4b.gd2po.minus.base.humaneval-plus}{+18.4}
\setval{4b.gd2po.minus.base.humaneval-plus-abs}{18.4}
\setval{4b.gd2po.minus.base.if}{+38.4}
\setval{4b.gd2po.minus.base.if-abs}{38.4}
\setval{4b.gd2po.minus.base.ifbench}{+20.7}
\setval{4b.gd2po.minus.base.ifbench-abs}{20.7}
\setval{4b.gd2po.minus.base.ifeval}{+40.3}
\setval{4b.gd2po.minus.base.ifeval-abs}{40.3}
\setval{4b.gd2po.minus.base.judge2.avg}{+17.7}
\setval{4b.gd2po.minus.base.judge2.avg-abs}{17.7}
\setval{4b.gd2po.minus.base.judge2.creative-writing-v3}{+12.0}
\setval{4b.gd2po.minus.base.judge2.creative-writing-v3-abs}{12.0}
\setval{4b.gd2po.minus.base.judge2.longbench-write}{+20.1}
\setval{4b.gd2po.minus.base.judge2.longbench-write-abs}{20.1}
\setval{4b.gd2po.minus.base.judge2.writing}{+18.0}
\setval{4b.gd2po.minus.base.judge2.writing-abs}{18.0}
\setval{4b.gd2po.minus.base.judge2.writingbench}{+22.1}
\setval{4b.gd2po.minus.base.judge2.writingbench-abs}{22.1}
\setval{4b.gd2po.minus.base.livecodebench}{+10.3}
\setval{4b.gd2po.minus.base.livecodebench-abs}{10.3}
\setval{4b.gd2po.minus.base.longbench-write}{+24.5}
\setval{4b.gd2po.minus.base.longbench-write-abs}{24.5}
\setval{4b.gd2po.minus.base.math}{+8.6}
\setval{4b.gd2po.minus.base.math-abs}{8.6}
\setval{4b.gd2po.minus.base.math500}{+10.4}
\setval{4b.gd2po.minus.base.math500-abs}{10.4}
\setval{4b.gd2po.minus.base.mbpp-plus}{+15.9}
\setval{4b.gd2po.minus.base.mbpp-plus-abs}{15.9}
\setval{4b.gd2po.minus.base.minerva}{+15.6}
\setval{4b.gd2po.minus.base.minerva-abs}{15.6}
\setval{4b.gd2po.minus.base.olympiad-bench}{+7.9}
\setval{4b.gd2po.minus.base.olympiad-bench-abs}{7.9}
\setval{4b.gd2po.minus.base.writing}{+27.4}
\setval{4b.gd2po.minus.base.writing-abs}{27.4}
\setval{4b.gd2po.minus.base.writingbench}{+37.7}
\setval{4b.gd2po.minus.base.writingbench-abs}{37.7}
\setval{4b.gd2po.minus.concur-free.aime24}{+0.0}
\setval{4b.gd2po.minus.concur-free.aime24-abs}{0.0}
\setval{4b.gd2po.minus.concur-free.aime25}{$-$7.8}
\setval{4b.gd2po.minus.concur-free.aime25-abs}{7.8}
\setval{4b.gd2po.minus.concur-free.amc23}{+3.6}
\setval{4b.gd2po.minus.concur-free.amc23-abs}{3.6}
\setval{4b.gd2po.minus.concur-free.avg}{$-$1.4}
\setval{4b.gd2po.minus.concur-free.avg-abs}{1.4}
\setval{4b.gd2po.minus.concur-free.code}{+1.1}
\setval{4b.gd2po.minus.concur-free.code-abs}{1.1}
\setval{4b.gd2po.minus.concur-free.creative-writing-v3}{$-$0.7}
\setval{4b.gd2po.minus.concur-free.creative-writing-v3-abs}{0.7}
\setval{4b.gd2po.minus.concur-free.guidebench}{+0.1}
\setval{4b.gd2po.minus.concur-free.guidebench-abs}{0.1}
\setval{4b.gd2po.minus.concur-free.humaneval-plus}{+2.3}
\setval{4b.gd2po.minus.concur-free.humaneval-plus-abs}{2.3}
\setval{4b.gd2po.minus.concur-free.if}{$-$0.1}
\setval{4b.gd2po.minus.concur-free.if-abs}{0.1}
\setval{4b.gd2po.minus.concur-free.ifbench}{$-$1.3}
\setval{4b.gd2po.minus.concur-free.ifbench-abs}{1.3}
\setval{4b.gd2po.minus.concur-free.ifeval}{+1.0}
\setval{4b.gd2po.minus.concur-free.ifeval-abs}{1.0}
\setval{4b.gd2po.minus.concur-free.judge2.avg}{$-$1.2}
\setval{4b.gd2po.minus.concur-free.judge2.avg-abs}{1.2}
\setval{4b.gd2po.minus.concur-free.judge2.creative-writing-v3}{$-$2.4}
\setval{4b.gd2po.minus.concur-free.judge2.creative-writing-v3-abs}{2.4}
\setval{4b.gd2po.minus.concur-free.judge2.longbench-write}{$-$8.4}
\setval{4b.gd2po.minus.concur-free.judge2.longbench-write-abs}{8.4}
\setval{4b.gd2po.minus.concur-free.judge2.writing}{$-$4.2}
\setval{4b.gd2po.minus.concur-free.judge2.writing-abs}{4.2}
\setval{4b.gd2po.minus.concur-free.judge2.writingbench}{$-$1.9}
\setval{4b.gd2po.minus.concur-free.judge2.writingbench-abs}{1.9}
\setval{4b.gd2po.minus.concur-free.livecodebench}{$-$0.7}
\setval{4b.gd2po.minus.concur-free.livecodebench-abs}{0.7}
\setval{4b.gd2po.minus.concur-free.longbench-write}{$-$8.8}
\setval{4b.gd2po.minus.concur-free.longbench-write-abs}{8.8}
\setval{4b.gd2po.minus.concur-free.math}{$-$1.4}
\setval{4b.gd2po.minus.concur-free.math-abs}{1.4}
\setval{4b.gd2po.minus.concur-free.math500}{$-$2.7}
\setval{4b.gd2po.minus.concur-free.math500-abs}{2.7}
\setval{4b.gd2po.minus.concur-free.mbpp-plus}{+1.8}
\setval{4b.gd2po.minus.concur-free.mbpp-plus-abs}{1.8}
\setval{4b.gd2po.minus.concur-free.minerva}{+0.9}
\setval{4b.gd2po.minus.concur-free.minerva-abs}{0.9}
\setval{4b.gd2po.minus.concur-free.olympiad-bench}{$-$2.2}
\setval{4b.gd2po.minus.concur-free.olympiad-bench-abs}{2.2}
\setval{4b.gd2po.minus.concur-free.writing}{$-$5.6}
\setval{4b.gd2po.minus.concur-free.writing-abs}{5.6}
\setval{4b.gd2po.minus.concur-free.writingbench}{$-$7.1}
\setval{4b.gd2po.minus.concur-free.writingbench-abs}{7.1}
\setval{4b.gd2po.minus.concur-tied.aime24}{$-$7.6}
\setval{4b.gd2po.minus.concur-tied.aime24-abs}{7.6}
\setval{4b.gd2po.minus.concur-tied.aime25}{$-$9.6}
\setval{4b.gd2po.minus.concur-tied.aime25-abs}{9.6}
\setval{4b.gd2po.minus.concur-tied.amc23}{$-$11.5}
\setval{4b.gd2po.minus.concur-tied.amc23-abs}{11.5}
\setval{4b.gd2po.minus.concur-tied.avg}{$-$4.6}
\setval{4b.gd2po.minus.concur-tied.avg-abs}{4.6}
\setval{4b.gd2po.minus.concur-tied.code}{+0.3}
\setval{4b.gd2po.minus.concur-tied.code-abs}{0.3}
\setval{4b.gd2po.minus.concur-tied.creative-writing-v3}{$-$3.2}
\setval{4b.gd2po.minus.concur-tied.creative-writing-v3-abs}{3.2}
\setval{4b.gd2po.minus.concur-tied.guidebench}{$-$0.1}
\setval{4b.gd2po.minus.concur-tied.guidebench-abs}{0.1}
\setval{4b.gd2po.minus.concur-tied.humaneval-plus}{$-$0.6}
\setval{4b.gd2po.minus.concur-tied.humaneval-plus-abs}{0.6}
\setval{4b.gd2po.minus.concur-tied.if}{$-$0.1}
\setval{4b.gd2po.minus.concur-tied.if-abs}{0.1}
\setval{4b.gd2po.minus.concur-tied.ifbench}{+0.0}
\setval{4b.gd2po.minus.concur-tied.ifbench-abs}{0.0}
\setval{4b.gd2po.minus.concur-tied.ifeval}{$-$0.1}
\setval{4b.gd2po.minus.concur-tied.ifeval-abs}{0.1}
\setval{4b.gd2po.minus.concur-tied.judge2.avg}{$-$3.9}
\setval{4b.gd2po.minus.concur-tied.judge2.avg-abs}{3.9}
\setval{4b.gd2po.minus.concur-tied.judge2.creative-writing-v3}{$-$1.1}
\setval{4b.gd2po.minus.concur-tied.judge2.creative-writing-v3-abs}{1.1}
\setval{4b.gd2po.minus.concur-tied.judge2.longbench-write}{$-$7.0}
\setval{4b.gd2po.minus.concur-tied.judge2.longbench-write-abs}{7.0}
\setval{4b.gd2po.minus.concur-tied.judge2.writing}{$-$3.6}
\setval{4b.gd2po.minus.concur-tied.judge2.writing-abs}{3.6}
\setval{4b.gd2po.minus.concur-tied.judge2.writingbench}{$-$2.8}
\setval{4b.gd2po.minus.concur-tied.judge2.writingbench-abs}{2.8}
\setval{4b.gd2po.minus.concur-tied.livecodebench}{$-$1.3}
\setval{4b.gd2po.minus.concur-tied.livecodebench-abs}{1.3}
\setval{4b.gd2po.minus.concur-tied.longbench-write}{$-$8.8}
\setval{4b.gd2po.minus.concur-tied.longbench-write-abs}{8.8}
\setval{4b.gd2po.minus.concur-tied.math}{$-$8.0}
\setval{4b.gd2po.minus.concur-tied.math-abs}{8.0}
\setval{4b.gd2po.minus.concur-tied.math500}{$-$7.8}
\setval{4b.gd2po.minus.concur-tied.math500-abs}{7.8}
\setval{4b.gd2po.minus.concur-tied.mbpp-plus}{+2.9}
\setval{4b.gd2po.minus.concur-tied.mbpp-plus-abs}{2.9}
\setval{4b.gd2po.minus.concur-tied.minerva}{$-$1.9}
\setval{4b.gd2po.minus.concur-tied.minerva-abs}{1.9}
\setval{4b.gd2po.minus.concur-tied.olympiad-bench}{$-$9.3}
\setval{4b.gd2po.minus.concur-tied.olympiad-bench-abs}{9.3}
\setval{4b.gd2po.minus.concur-tied.writing}{$-$7.3}
\setval{4b.gd2po.minus.concur-tied.writing-abs}{7.3}
\setval{4b.gd2po.minus.concur-tied.writingbench}{$-$10.0}
\setval{4b.gd2po.minus.concur-tied.writingbench-abs}{10.0}
\setval{4b.gd2po.minus.concur.aime24}{$-$8.6}
\setval{4b.gd2po.minus.concur.aime24-abs}{8.6}
\setval{4b.gd2po.minus.concur.aime25}{$-$12.0}
\setval{4b.gd2po.minus.concur.aime25-abs}{12.0}
\setval{4b.gd2po.minus.concur.amc23}{$-$11.6}
\setval{4b.gd2po.minus.concur.amc23-abs}{11.6}
\setval{4b.gd2po.minus.concur.avg}{$-$4.7}
\setval{4b.gd2po.minus.concur.avg-abs}{4.7}
\setval{4b.gd2po.minus.concur.code}{$-$0.2}
\setval{4b.gd2po.minus.concur.code-abs}{0.2}
\setval{4b.gd2po.minus.concur.creative-writing-v3}{+0.0}
\setval{4b.gd2po.minus.concur.creative-writing-v3-abs}{0.0}
\setval{4b.gd2po.minus.concur.guidebench}{+1.7}
\setval{4b.gd2po.minus.concur.guidebench-abs}{1.7}
\setval{4b.gd2po.minus.concur.humaneval-plus}{+1.0}
\setval{4b.gd2po.minus.concur.humaneval-plus-abs}{1.0}
\setval{4b.gd2po.minus.concur.if}{$-$0.5}
\setval{4b.gd2po.minus.concur.if-abs}{0.5}
\setval{4b.gd2po.minus.concur.ifbench}{$-$2.7}
\setval{4b.gd2po.minus.concur.ifbench-abs}{2.7}
\setval{4b.gd2po.minus.concur.ifeval}{$-$0.5}
\setval{4b.gd2po.minus.concur.ifeval-abs}{0.5}
\setval{4b.gd2po.minus.concur.judge2.avg}{$-$4.2}
\setval{4b.gd2po.minus.concur.judge2.avg-abs}{4.2}
\setval{4b.gd2po.minus.concur.judge2.creative-writing-v3}{+0.5}
\setval{4b.gd2po.minus.concur.judge2.creative-writing-v3-abs}{0.5}
\setval{4b.gd2po.minus.concur.judge2.longbench-write}{$-$5.7}
\setval{4b.gd2po.minus.concur.judge2.longbench-write-abs}{5.7}
\setval{4b.gd2po.minus.concur.judge2.writing}{$-$2.9}
\setval{4b.gd2po.minus.concur.judge2.writing-abs}{2.9}
\setval{4b.gd2po.minus.concur.judge2.writingbench}{$-$3.5}
\setval{4b.gd2po.minus.concur.judge2.writingbench-abs}{3.5}
\setval{4b.gd2po.minus.concur.livecodebench}{$-$1.7}
\setval{4b.gd2po.minus.concur.livecodebench-abs}{1.7}
\setval{4b.gd2po.minus.concur.longbench-write}{$-$7.4}
\setval{4b.gd2po.minus.concur.longbench-write-abs}{7.4}
\setval{4b.gd2po.minus.concur.math}{$-$8.7}
\setval{4b.gd2po.minus.concur.math-abs}{8.7}
\setval{4b.gd2po.minus.concur.math500}{$-$7.6}
\setval{4b.gd2po.minus.concur.math500-abs}{7.6}
\setval{4b.gd2po.minus.concur.mbpp-plus}{$-$0.1}
\setval{4b.gd2po.minus.concur.mbpp-plus-abs}{0.1}
\setval{4b.gd2po.minus.concur.minerva}{$-$2.0}
\setval{4b.gd2po.minus.concur.minerva-abs}{2.0}
\setval{4b.gd2po.minus.concur.olympiad-bench}{$-$10.0}
\setval{4b.gd2po.minus.concur.olympiad-bench-abs}{10.0}
\setval{4b.gd2po.minus.concur.writing}{$-$5.6}
\setval{4b.gd2po.minus.concur.writing-abs}{5.6}
\setval{4b.gd2po.minus.concur.writingbench}{$-$9.3}
\setval{4b.gd2po.minus.concur.writingbench-abs}{9.3}
\setval{4b.gd2po.minus.gdpo-noaux.aime24}{$-$11.6}
\setval{4b.gd2po.minus.gdpo-noaux.aime24-abs}{11.6}
\setval{4b.gd2po.minus.gdpo-noaux.aime25}{$-$12.0}
\setval{4b.gd2po.minus.gdpo-noaux.aime25-abs}{12.0}
\setval{4b.gd2po.minus.gdpo-noaux.amc23}{$-$8.0}
\setval{4b.gd2po.minus.gdpo-noaux.amc23-abs}{8.0}
\setval{4b.gd2po.minus.gdpo-noaux.avg}{$-$2.6}
\setval{4b.gd2po.minus.gdpo-noaux.avg-abs}{2.6}
\setval{4b.gd2po.minus.gdpo-noaux.code}{$-$1.4}
\setval{4b.gd2po.minus.gdpo-noaux.code-abs}{1.4}
\setval{4b.gd2po.minus.gdpo-noaux.creative-writing-v3}{+5.4}
\setval{4b.gd2po.minus.gdpo-noaux.creative-writing-v3-abs}{5.4}
\setval{4b.gd2po.minus.gdpo-noaux.guidebench}{+3.2}
\setval{4b.gd2po.minus.gdpo-noaux.guidebench-abs}{3.2}
\setval{4b.gd2po.minus.gdpo-noaux.humaneval-plus}{$-$2.1}
\setval{4b.gd2po.minus.gdpo-noaux.humaneval-plus-abs}{2.1}
\setval{4b.gd2po.minus.gdpo-noaux.if}{+1.1}
\setval{4b.gd2po.minus.gdpo-noaux.if-abs}{1.1}
\setval{4b.gd2po.minus.gdpo-noaux.ifbench}{$-$2.7}
\setval{4b.gd2po.minus.gdpo-noaux.ifbench-abs}{2.7}
\setval{4b.gd2po.minus.gdpo-noaux.ifeval}{+2.9}
\setval{4b.gd2po.minus.gdpo-noaux.ifeval-abs}{2.9}
\setval{4b.gd2po.minus.gdpo-noaux.judge2.avg}{$-$1.7}
\setval{4b.gd2po.minus.gdpo-noaux.judge2.avg-abs}{1.7}
\setval{4b.gd2po.minus.gdpo-noaux.judge2.creative-writing-v3}{+9.5}
\setval{4b.gd2po.minus.gdpo-noaux.judge2.creative-writing-v3-abs}{9.5}
\setval{4b.gd2po.minus.gdpo-noaux.judge2.longbench-write}{+10.2}
\setval{4b.gd2po.minus.gdpo-noaux.judge2.longbench-write-abs}{10.2}
\setval{4b.gd2po.minus.gdpo-noaux.judge2.writing}{+8.7}
\setval{4b.gd2po.minus.gdpo-noaux.judge2.writing-abs}{8.7}
\setval{4b.gd2po.minus.gdpo-noaux.judge2.writingbench}{+6.5}
\setval{4b.gd2po.minus.gdpo-noaux.judge2.writingbench-abs}{6.5}
\setval{4b.gd2po.minus.gdpo-noaux.livecodebench}{$-$1.3}
\setval{4b.gd2po.minus.gdpo-noaux.livecodebench-abs}{1.3}
\setval{4b.gd2po.minus.gdpo-noaux.longbench-write}{+4.6}
\setval{4b.gd2po.minus.gdpo-noaux.longbench-write-abs}{4.6}
\setval{4b.gd2po.minus.gdpo-noaux.math}{$-$8.4}
\setval{4b.gd2po.minus.gdpo-noaux.math-abs}{8.4}
\setval{4b.gd2po.minus.gdpo-noaux.math500}{$-$7.1}
\setval{4b.gd2po.minus.gdpo-noaux.math500-abs}{7.1}
\setval{4b.gd2po.minus.gdpo-noaux.mbpp-plus}{$-$0.9}
\setval{4b.gd2po.minus.gdpo-noaux.mbpp-plus-abs}{0.9}
\setval{4b.gd2po.minus.gdpo-noaux.minerva}{$-$2.3}
\setval{4b.gd2po.minus.gdpo-noaux.minerva-abs}{2.3}
\setval{4b.gd2po.minus.gdpo-noaux.olympiad-bench}{$-$9.6}
\setval{4b.gd2po.minus.gdpo-noaux.olympiad-bench-abs}{9.6}
\setval{4b.gd2po.minus.gdpo-noaux.writing}{+4.2}
\setval{4b.gd2po.minus.gdpo-noaux.writing-abs}{4.2}
\setval{4b.gd2po.minus.gdpo-noaux.writingbench}{+2.8}
\setval{4b.gd2po.minus.gdpo-noaux.writingbench-abs}{2.8}
\setval{4b.gd2po.minus.gdpo-w0.5.aime24}{$-$2.7}
\setval{4b.gd2po.minus.gdpo-w0.5.aime24-abs}{2.7}
\setval{4b.gd2po.minus.gdpo-w0.5.aime25}{$-$3.5}
\setval{4b.gd2po.minus.gdpo-w0.5.aime25-abs}{3.5}
\setval{4b.gd2po.minus.gdpo-w0.5.amc23}{$-$0.6}
\setval{4b.gd2po.minus.gdpo-w0.5.amc23-abs}{0.6}
\setval{4b.gd2po.minus.gdpo-w0.5.avg}{$-$0.5}
\setval{4b.gd2po.minus.gdpo-w0.5.avg-abs}{0.5}
\setval{4b.gd2po.minus.gdpo-w0.5.code}{+0.3}
\setval{4b.gd2po.minus.gdpo-w0.5.code-abs}{0.3}
\setval{4b.gd2po.minus.gdpo-w0.5.creative-writing-v3}{+6.6}
\setval{4b.gd2po.minus.gdpo-w0.5.creative-writing-v3-abs}{6.6}
\setval{4b.gd2po.minus.gdpo-w0.5.guidebench}{$-$0.5}
\setval{4b.gd2po.minus.gdpo-w0.5.guidebench-abs}{0.5}
\setval{4b.gd2po.minus.gdpo-w0.5.humaneval-plus}{+1.4}
\setval{4b.gd2po.minus.gdpo-w0.5.humaneval-plus-abs}{1.4}
\setval{4b.gd2po.minus.gdpo-w0.5.if}{$-$0.4}
\setval{4b.gd2po.minus.gdpo-w0.5.if-abs}{0.4}
\setval{4b.gd2po.minus.gdpo-w0.5.ifbench}{$-$1.7}
\setval{4b.gd2po.minus.gdpo-w0.5.ifbench-abs}{1.7}
\setval{4b.gd2po.minus.gdpo-w0.5.ifeval}{+0.8}
\setval{4b.gd2po.minus.gdpo-w0.5.ifeval-abs}{0.8}
\setval{4b.gd2po.minus.gdpo-w0.5.judge2.avg}{$-$0.9}
\setval{4b.gd2po.minus.gdpo-w0.5.judge2.avg-abs}{0.9}
\setval{4b.gd2po.minus.gdpo-w0.5.judge2.creative-writing-v3}{+2.8}
\setval{4b.gd2po.minus.gdpo-w0.5.judge2.creative-writing-v3-abs}{2.8}
\setval{4b.gd2po.minus.gdpo-w0.5.judge2.longbench-write}{+0.4}
\setval{4b.gd2po.minus.gdpo-w0.5.judge2.longbench-write-abs}{0.4}
\setval{4b.gd2po.minus.gdpo-w0.5.judge2.writing}{+0.7}
\setval{4b.gd2po.minus.gdpo-w0.5.judge2.writing-abs}{0.7}
\setval{4b.gd2po.minus.gdpo-w0.5.judge2.writingbench}{$-$1.1}
\setval{4b.gd2po.minus.gdpo-w0.5.judge2.writingbench-abs}{1.1}
\setval{4b.gd2po.minus.gdpo-w0.5.livecodebench}{$-$0.7}
\setval{4b.gd2po.minus.gdpo-w0.5.livecodebench-abs}{0.7}
\setval{4b.gd2po.minus.gdpo-w0.5.longbench-write}{+1.5}
\setval{4b.gd2po.minus.gdpo-w0.5.longbench-write-abs}{1.5}
\setval{4b.gd2po.minus.gdpo-w0.5.math}{$-$2.5}
\setval{4b.gd2po.minus.gdpo-w0.5.math-abs}{2.5}
\setval{4b.gd2po.minus.gdpo-w0.5.math500}{$-$4.1}
\setval{4b.gd2po.minus.gdpo-w0.5.math500-abs}{4.1}
\setval{4b.gd2po.minus.gdpo-w0.5.mbpp-plus}{+0.3}
\setval{4b.gd2po.minus.gdpo-w0.5.mbpp-plus-abs}{0.3}
\setval{4b.gd2po.minus.gdpo-w0.5.minerva}{+1.1}
\setval{4b.gd2po.minus.gdpo-w0.5.minerva-abs}{1.1}
\setval{4b.gd2po.minus.gdpo-w0.5.olympiad-bench}{$-$4.9}
\setval{4b.gd2po.minus.gdpo-w0.5.olympiad-bench-abs}{4.9}
\setval{4b.gd2po.minus.gdpo-w0.5.writing}{+2.5}
\setval{4b.gd2po.minus.gdpo-w0.5.writing-abs}{2.5}
\setval{4b.gd2po.minus.gdpo-w0.5.writingbench}{$-$0.6}
\setval{4b.gd2po.minus.gdpo-w0.5.writingbench-abs}{0.6}
\setval{4b.gd2po.minus.gdpo.aime24}{+2.8}
\setval{4b.gd2po.minus.gdpo.aime24-abs}{2.8}
\setval{4b.gd2po.minus.gdpo.aime25}{$-$3.1}
\setval{4b.gd2po.minus.gdpo.aime25-abs}{3.1}
\setval{4b.gd2po.minus.gdpo.amc23}{$-$1.9}
\setval{4b.gd2po.minus.gdpo.amc23-abs}{1.9}
\setval{4b.gd2po.minus.gdpo.avg}{$-$0.2}
\setval{4b.gd2po.minus.gdpo.avg-abs}{0.2}
\setval{4b.gd2po.minus.gdpo.code}{+0.4}
\setval{4b.gd2po.minus.gdpo.code-abs}{0.4}
\setval{4b.gd2po.minus.gdpo.creative-writing-v3}{$-$1.3}
\setval{4b.gd2po.minus.gdpo.creative-writing-v3-abs}{1.3}
\setval{4b.gd2po.minus.gdpo.guidebench}{$-$1.5}
\setval{4b.gd2po.minus.gdpo.guidebench-abs}{1.5}
\setval{4b.gd2po.minus.gdpo.humaneval-plus}{+1.6}
\setval{4b.gd2po.minus.gdpo.humaneval-plus-abs}{1.6}
\setval{4b.gd2po.minus.gdpo.if}{+0.6}
\setval{4b.gd2po.minus.gdpo.if-abs}{0.6}
\setval{4b.gd2po.minus.gdpo.ifbench}{+1.3}
\setval{4b.gd2po.minus.gdpo.ifbench-abs}{1.3}
\setval{4b.gd2po.minus.gdpo.ifeval}{+1.9}
\setval{4b.gd2po.minus.gdpo.ifeval-abs}{1.9}
\setval{4b.gd2po.minus.gdpo.judge2.avg}{$-$0.3}
\setval{4b.gd2po.minus.gdpo.judge2.avg-abs}{0.3}
\setval{4b.gd2po.minus.gdpo.judge2.creative-writing-v3}{$-$1.5}
\setval{4b.gd2po.minus.gdpo.judge2.creative-writing-v3-abs}{1.5}
\setval{4b.gd2po.minus.gdpo.judge2.longbench-write}{$-$1.3}
\setval{4b.gd2po.minus.gdpo.judge2.longbench-write-abs}{1.3}
\setval{4b.gd2po.minus.gdpo.judge2.writing}{$-$1.4}
\setval{4b.gd2po.minus.gdpo.judge2.writing-abs}{1.4}
\setval{4b.gd2po.minus.gdpo.judge2.writingbench}{$-$1.5}
\setval{4b.gd2po.minus.gdpo.judge2.writingbench-abs}{1.5}
\setval{4b.gd2po.minus.gdpo.livecodebench}{$-$0.9}
\setval{4b.gd2po.minus.gdpo.livecodebench-abs}{0.9}
\setval{4b.gd2po.minus.gdpo.longbench-write}{$-$0.1}
\setval{4b.gd2po.minus.gdpo.longbench-write-abs}{0.1}
\setval{4b.gd2po.minus.gdpo.math}{$-$0.5}
\setval{4b.gd2po.minus.gdpo.math-abs}{0.5}
\setval{4b.gd2po.minus.gdpo.math500}{$-$1.6}
\setval{4b.gd2po.minus.gdpo.math500-abs}{1.6}
\setval{4b.gd2po.minus.gdpo.mbpp-plus}{+0.6}
\setval{4b.gd2po.minus.gdpo.mbpp-plus-abs}{0.6}
\setval{4b.gd2po.minus.gdpo.minerva}{+1.7}
\setval{4b.gd2po.minus.gdpo.minerva-abs}{1.7}
\setval{4b.gd2po.minus.gdpo.olympiad-bench}{$-$1.1}
\setval{4b.gd2po.minus.gdpo.olympiad-bench-abs}{1.1}
\setval{4b.gd2po.minus.gdpo.writing}{$-$1.0}
\setval{4b.gd2po.minus.gdpo.writing-abs}{1.0}
\setval{4b.gd2po.minus.gdpo.writingbench}{$-$1.8}
\setval{4b.gd2po.minus.gdpo.writingbench-abs}{1.8}
\setval{4b.gd2po.minus.geoma.aime24}{+0.3}
\setval{4b.gd2po.minus.geoma.aime24-abs}{0.3}
\setval{4b.gd2po.minus.geoma.aime25}{$-$3.9}
\setval{4b.gd2po.minus.geoma.aime25-abs}{3.9}
\setval{4b.gd2po.minus.geoma.amc23}{+2.3}
\setval{4b.gd2po.minus.geoma.amc23-abs}{2.3}
\setval{4b.gd2po.minus.geoma.avg}{+0.3}
\setval{4b.gd2po.minus.geoma.avg-abs}{0.3}
\setval{4b.gd2po.minus.geoma.code}{+0.7}
\setval{4b.gd2po.minus.geoma.code-abs}{0.7}
\setval{4b.gd2po.minus.geoma.creative-writing-v3}{$-$1.2}
\setval{4b.gd2po.minus.geoma.creative-writing-v3-abs}{1.2}
\setval{4b.gd2po.minus.geoma.guidebench}{+0.6}
\setval{4b.gd2po.minus.geoma.guidebench-abs}{0.6}
\setval{4b.gd2po.minus.geoma.humaneval-plus}{+3.2}
\setval{4b.gd2po.minus.geoma.humaneval-plus-abs}{3.2}
\setval{4b.gd2po.minus.geoma.if}{+0.7}
\setval{4b.gd2po.minus.geoma.if-abs}{0.7}
\setval{4b.gd2po.minus.geoma.ifbench}{+1.7}
\setval{4b.gd2po.minus.geoma.ifbench-abs}{1.7}
\setval{4b.gd2po.minus.geoma.ifeval}{$-$0.1}
\setval{4b.gd2po.minus.geoma.ifeval-abs}{0.1}
\setval{4b.gd2po.minus.geoma.judge2.avg}{+0.5}
\setval{4b.gd2po.minus.geoma.judge2.avg-abs}{0.5}
\setval{4b.gd2po.minus.geoma.judge2.creative-writing-v3}{$-$0.2}
\setval{4b.gd2po.minus.geoma.judge2.creative-writing-v3-abs}{0.2}
\setval{4b.gd2po.minus.geoma.judge2.longbench-write}{+3.3}
\setval{4b.gd2po.minus.geoma.judge2.longbench-write-abs}{3.3}
\setval{4b.gd2po.minus.geoma.judge2.writing}{+1.6}
\setval{4b.gd2po.minus.geoma.judge2.writing-abs}{1.6}
\setval{4b.gd2po.minus.geoma.judge2.writingbench}{+1.9}
\setval{4b.gd2po.minus.geoma.judge2.writingbench-abs}{1.9}
\setval{4b.gd2po.minus.geoma.livecodebench}{$-$2.2}
\setval{4b.gd2po.minus.geoma.livecodebench-abs}{2.2}
\setval{4b.gd2po.minus.geoma.longbench-write}{+3.1}
\setval{4b.gd2po.minus.geoma.longbench-write-abs}{3.1}
\setval{4b.gd2po.minus.geoma.math}{$-$0.4}
\setval{4b.gd2po.minus.geoma.math-abs}{0.4}
\setval{4b.gd2po.minus.geoma.math500}{$-$1.2}
\setval{4b.gd2po.minus.geoma.math500-abs}{1.2}
\setval{4b.gd2po.minus.geoma.mbpp-plus}{+1.2}
\setval{4b.gd2po.minus.geoma.mbpp-plus-abs}{1.2}
\setval{4b.gd2po.minus.geoma.minerva}{+0.2}
\setval{4b.gd2po.minus.geoma.minerva-abs}{0.2}
\setval{4b.gd2po.minus.geoma.olympiad-bench}{$-$0.3}
\setval{4b.gd2po.minus.geoma.olympiad-bench-abs}{0.3}
\setval{4b.gd2po.minus.geoma.writing}{+0.9}
\setval{4b.gd2po.minus.geoma.writing-abs}{0.9}
\setval{4b.gd2po.minus.geoma.writingbench}{+0.6}
\setval{4b.gd2po.minus.geoma.writingbench-abs}{0.6}
\setval{4b.gd2po.minus.grpo-nodiv.aime24}{$-$0.5}
\setval{4b.gd2po.minus.grpo-nodiv.aime24-abs}{0.5}
\setval{4b.gd2po.minus.grpo-nodiv.aime25}{$-$1.6}
\setval{4b.gd2po.minus.grpo-nodiv.aime25-abs}{1.6}
\setval{4b.gd2po.minus.grpo-nodiv.amc23}{+3.2}
\setval{4b.gd2po.minus.grpo-nodiv.amc23-abs}{3.2}
\setval{4b.gd2po.minus.grpo-nodiv.avg}{+1.1}
\setval{4b.gd2po.minus.grpo-nodiv.avg-abs}{1.1}
\setval{4b.gd2po.minus.grpo-nodiv.code}{+0.7}
\setval{4b.gd2po.minus.grpo-nodiv.code-abs}{0.7}
\setval{4b.gd2po.minus.grpo-nodiv.creative-writing-v3}{+2.1}
\setval{4b.gd2po.minus.grpo-nodiv.creative-writing-v3-abs}{2.1}
\setval{4b.gd2po.minus.grpo-nodiv.guidebench}{+0.3}
\setval{4b.gd2po.minus.grpo-nodiv.guidebench-abs}{0.3}
\setval{4b.gd2po.minus.grpo-nodiv.humaneval-plus}{+2.9}
\setval{4b.gd2po.minus.grpo-nodiv.humaneval-plus-abs}{2.9}
\setval{4b.gd2po.minus.grpo-nodiv.if}{+1.5}
\setval{4b.gd2po.minus.grpo-nodiv.if-abs}{1.5}
\setval{4b.gd2po.minus.grpo-nodiv.ifbench}{+4.0}
\setval{4b.gd2po.minus.grpo-nodiv.ifbench-abs}{4.0}
\setval{4b.gd2po.minus.grpo-nodiv.ifeval}{+0.4}
\setval{4b.gd2po.minus.grpo-nodiv.ifeval-abs}{0.4}
\setval{4b.gd2po.minus.grpo-nodiv.judge2.avg}{+0.9}
\setval{4b.gd2po.minus.grpo-nodiv.judge2.avg-abs}{0.9}
\setval{4b.gd2po.minus.grpo-nodiv.judge2.creative-writing-v3}{+2.4}
\setval{4b.gd2po.minus.grpo-nodiv.judge2.creative-writing-v3-abs}{2.4}
\setval{4b.gd2po.minus.grpo-nodiv.judge2.longbench-write}{+2.4}
\setval{4b.gd2po.minus.grpo-nodiv.judge2.longbench-write-abs}{2.4}
\setval{4b.gd2po.minus.grpo-nodiv.judge2.writing}{+2.6}
\setval{4b.gd2po.minus.grpo-nodiv.judge2.writing-abs}{2.6}
\setval{4b.gd2po.minus.grpo-nodiv.judge2.writingbench}{+3.0}
\setval{4b.gd2po.minus.grpo-nodiv.judge2.writingbench-abs}{3.0}
\setval{4b.gd2po.minus.grpo-nodiv.livecodebench}{$-$1.0}
\setval{4b.gd2po.minus.grpo-nodiv.livecodebench-abs}{1.0}
\setval{4b.gd2po.minus.grpo-nodiv.longbench-write}{+2.9}
\setval{4b.gd2po.minus.grpo-nodiv.longbench-write-abs}{2.9}
\setval{4b.gd2po.minus.grpo-nodiv.math}{$-$0.1}
\setval{4b.gd2po.minus.grpo-nodiv.math-abs}{0.1}
\setval{4b.gd2po.minus.grpo-nodiv.math500}{$-$1.8}
\setval{4b.gd2po.minus.grpo-nodiv.math500-abs}{1.8}
\setval{4b.gd2po.minus.grpo-nodiv.mbpp-plus}{+0.3}
\setval{4b.gd2po.minus.grpo-nodiv.mbpp-plus-abs}{0.3}
\setval{4b.gd2po.minus.grpo-nodiv.minerva}{+1.4}
\setval{4b.gd2po.minus.grpo-nodiv.minerva-abs}{1.4}
\setval{4b.gd2po.minus.grpo-nodiv.olympiad-bench}{$-$1.3}
\setval{4b.gd2po.minus.grpo-nodiv.olympiad-bench-abs}{1.3}
\setval{4b.gd2po.minus.grpo-nodiv.writing}{+3.2}
\setval{4b.gd2po.minus.grpo-nodiv.writing-abs}{3.2}
\setval{4b.gd2po.minus.grpo-nodiv.writingbench}{+4.5}
\setval{4b.gd2po.minus.grpo-nodiv.writingbench-abs}{4.5}
\setval{4b.gd2po.minus.grpo.aime24}{$-$0.1}
\setval{4b.gd2po.minus.grpo.aime24-abs}{0.1}
\setval{4b.gd2po.minus.grpo.aime25}{$-$5.8}
\setval{4b.gd2po.minus.grpo.aime25-abs}{5.8}
\setval{4b.gd2po.minus.grpo.amc23}{+3.0}
\setval{4b.gd2po.minus.grpo.amc23-abs}{3.0}
\setval{4b.gd2po.minus.grpo.avg}{$-$0.6}
\setval{4b.gd2po.minus.grpo.avg-abs}{0.6}
\setval{4b.gd2po.minus.grpo.code}{+0.1}
\setval{4b.gd2po.minus.grpo.code-abs}{0.1}
\setval{4b.gd2po.minus.grpo.creative-writing-v3}{$-$2.6}
\setval{4b.gd2po.minus.grpo.creative-writing-v3-abs}{2.6}
\setval{4b.gd2po.minus.grpo.guidebench}{+1.2}
\setval{4b.gd2po.minus.grpo.guidebench-abs}{1.2}
\setval{4b.gd2po.minus.grpo.humaneval-plus}{+0.7}
\setval{4b.gd2po.minus.grpo.humaneval-plus-abs}{0.7}
\setval{4b.gd2po.minus.grpo.if}{+0.8}
\setval{4b.gd2po.minus.grpo.if-abs}{0.8}
\setval{4b.gd2po.minus.grpo.ifbench}{+0.0}
\setval{4b.gd2po.minus.grpo.ifbench-abs}{0.0}
\setval{4b.gd2po.minus.grpo.ifeval}{+1.2}
\setval{4b.gd2po.minus.grpo.ifeval-abs}{1.2}
\setval{4b.gd2po.minus.grpo.judge2.avg}{$-$0.3}
\setval{4b.gd2po.minus.grpo.judge2.avg-abs}{0.3}
\setval{4b.gd2po.minus.grpo.judge2.creative-writing-v3}{$-$2.2}
\setval{4b.gd2po.minus.grpo.judge2.creative-writing-v3-abs}{2.2}
\setval{4b.gd2po.minus.grpo.judge2.longbench-write}{+0.3}
\setval{4b.gd2po.minus.grpo.judge2.longbench-write-abs}{0.3}
\setval{4b.gd2po.minus.grpo.judge2.writing}{$-$1.0}
\setval{4b.gd2po.minus.grpo.judge2.writing-abs}{1.0}
\setval{4b.gd2po.minus.grpo.judge2.writingbench}{$-$0.9}
\setval{4b.gd2po.minus.grpo.judge2.writingbench-abs}{0.9}
\setval{4b.gd2po.minus.grpo.livecodebench}{$-$0.5}
\setval{4b.gd2po.minus.grpo.livecodebench-abs}{0.5}
\setval{4b.gd2po.minus.grpo.longbench-write}{$-$0.3}
\setval{4b.gd2po.minus.grpo.longbench-write-abs}{0.3}
\setval{4b.gd2po.minus.grpo.math}{$-$0.8}
\setval{4b.gd2po.minus.grpo.math-abs}{0.8}
\setval{4b.gd2po.minus.grpo.math500}{$-$2.6}
\setval{4b.gd2po.minus.grpo.math500-abs}{2.6}
\setval{4b.gd2po.minus.grpo.mbpp-plus}{+0.3}
\setval{4b.gd2po.minus.grpo.mbpp-plus-abs}{0.3}
\setval{4b.gd2po.minus.grpo.minerva}{+1.9}
\setval{4b.gd2po.minus.grpo.minerva-abs}{1.9}
\setval{4b.gd2po.minus.grpo.olympiad-bench}{$-$1.1}
\setval{4b.gd2po.minus.grpo.olympiad-bench-abs}{1.1}
\setval{4b.gd2po.minus.grpo.writing}{$-$2.3}
\setval{4b.gd2po.minus.grpo.writing-abs}{2.3}
\setval{4b.gd2po.minus.grpo.writingbench}{$-$4.1}
\setval{4b.gd2po.minus.grpo.writingbench-abs}{4.1}
\setval{4b.gd2po.minus.mt-grpo.aime24}{+1.1}
\setval{4b.gd2po.minus.mt-grpo.aime24-abs}{1.1}
\setval{4b.gd2po.minus.mt-grpo.aime25}{+0.6}
\setval{4b.gd2po.minus.mt-grpo.aime25-abs}{0.6}
\setval{4b.gd2po.minus.mt-grpo.amc23}{$-$2.0}
\setval{4b.gd2po.minus.mt-grpo.amc23-abs}{2.0}
\setval{4b.gd2po.minus.mt-grpo.avg}{+1.0}
\setval{4b.gd2po.minus.mt-grpo.avg-abs}{1.0}
\setval{4b.gd2po.minus.mt-grpo.code}{+2.2}
\setval{4b.gd2po.minus.mt-grpo.code-abs}{2.2}
\setval{4b.gd2po.minus.mt-grpo.creative-writing-v3}{+0.2}
\setval{4b.gd2po.minus.mt-grpo.creative-writing-v3-abs}{0.2}
\setval{4b.gd2po.minus.mt-grpo.guidebench}{$-$0.6}
\setval{4b.gd2po.minus.mt-grpo.guidebench-abs}{0.6}
\setval{4b.gd2po.minus.mt-grpo.humaneval-plus}{+4.3}
\setval{4b.gd2po.minus.mt-grpo.humaneval-plus-abs}{4.3}
\setval{4b.gd2po.minus.mt-grpo.if}{+1.2}
\setval{4b.gd2po.minus.mt-grpo.if-abs}{1.2}
\setval{4b.gd2po.minus.mt-grpo.ifbench}{+3.3}
\setval{4b.gd2po.minus.mt-grpo.ifbench-abs}{3.3}
\setval{4b.gd2po.minus.mt-grpo.ifeval}{+1.0}
\setval{4b.gd2po.minus.mt-grpo.ifeval-abs}{1.0}
\setval{4b.gd2po.minus.mt-grpo.judge2.avg}{+0.8}
\setval{4b.gd2po.minus.mt-grpo.judge2.avg-abs}{0.8}
\setval{4b.gd2po.minus.mt-grpo.judge2.creative-writing-v3}{$-$1.2}
\setval{4b.gd2po.minus.mt-grpo.judge2.creative-writing-v3-abs}{1.2}
\setval{4b.gd2po.minus.mt-grpo.judge2.longbench-write}{+0.2}
\setval{4b.gd2po.minus.mt-grpo.judge2.longbench-write-abs}{0.2}
\setval{4b.gd2po.minus.mt-grpo.judge2.writing}{$-$0.6}
\setval{4b.gd2po.minus.mt-grpo.judge2.writing-abs}{0.6}
\setval{4b.gd2po.minus.mt-grpo.judge2.writingbench}{$-$0.7}
\setval{4b.gd2po.minus.mt-grpo.judge2.writingbench-abs}{0.7}
\setval{4b.gd2po.minus.mt-grpo.livecodebench}{+0.9}
\setval{4b.gd2po.minus.mt-grpo.livecodebench-abs}{0.9}
\setval{4b.gd2po.minus.mt-grpo.longbench-write}{+2.1}
\setval{4b.gd2po.minus.mt-grpo.longbench-write-abs}{2.1}
\setval{4b.gd2po.minus.mt-grpo.math}{+0.4}
\setval{4b.gd2po.minus.mt-grpo.math-abs}{0.4}
\setval{4b.gd2po.minus.mt-grpo.math500}{+0.7}
\setval{4b.gd2po.minus.mt-grpo.math500-abs}{0.7}
\setval{4b.gd2po.minus.mt-grpo.mbpp-plus}{+1.4}
\setval{4b.gd2po.minus.mt-grpo.mbpp-plus-abs}{1.4}
\setval{4b.gd2po.minus.mt-grpo.minerva}{+1.2}
\setval{4b.gd2po.minus.mt-grpo.minerva-abs}{1.2}
\setval{4b.gd2po.minus.mt-grpo.olympiad-bench}{+1.0}
\setval{4b.gd2po.minus.mt-grpo.olympiad-bench-abs}{1.0}
\setval{4b.gd2po.minus.mt-grpo.writing}{+0.8}
\setval{4b.gd2po.minus.mt-grpo.writing-abs}{0.8}
\setval{4b.gd2po.minus.mt-grpo.writingbench}{+0.0}
\setval{4b.gd2po.minus.mt-grpo.writingbench-abs}{0.0}
\setval{4b.gd2po.minus.papo.aime24}{+2.6}
\setval{4b.gd2po.minus.papo.aime24-abs}{2.6}
\setval{4b.gd2po.minus.papo.aime25}{$-$0.7}
\setval{4b.gd2po.minus.papo.aime25-abs}{0.7}
\setval{4b.gd2po.minus.papo.amc23}{+4.2}
\setval{4b.gd2po.minus.papo.amc23-abs}{4.2}
\setval{4b.gd2po.minus.papo.avg}{+1.5}
\setval{4b.gd2po.minus.papo.avg-abs}{1.5}
\setval{4b.gd2po.minus.papo.code}{+0.3}
\setval{4b.gd2po.minus.papo.code-abs}{0.3}
\setval{4b.gd2po.minus.papo.creative-writing-v3}{$-$1.9}
\setval{4b.gd2po.minus.papo.creative-writing-v3-abs}{1.9}
\setval{4b.gd2po.minus.papo.guidebench}{$-$1.6}
\setval{4b.gd2po.minus.papo.guidebench-abs}{1.6}
\setval{4b.gd2po.minus.papo.humaneval-plus}{+0.4}
\setval{4b.gd2po.minus.papo.humaneval-plus-abs}{0.4}
\setval{4b.gd2po.minus.papo.if}{+1.5}
\setval{4b.gd2po.minus.papo.if-abs}{1.5}
\setval{4b.gd2po.minus.papo.ifbench}{+4.3}
\setval{4b.gd2po.minus.papo.ifbench-abs}{4.3}
\setval{4b.gd2po.minus.papo.ifeval}{+1.8}
\setval{4b.gd2po.minus.papo.ifeval-abs}{1.8}
\setval{4b.gd2po.minus.papo.judge2.avg}{+1.2}
\setval{4b.gd2po.minus.papo.judge2.avg-abs}{1.2}
\setval{4b.gd2po.minus.papo.judge2.creative-writing-v3}{$-$1.3}
\setval{4b.gd2po.minus.papo.judge2.creative-writing-v3-abs}{1.3}
\setval{4b.gd2po.minus.papo.judge2.longbench-write}{+3.4}
\setval{4b.gd2po.minus.papo.judge2.longbench-write-abs}{3.4}
\setval{4b.gd2po.minus.papo.judge2.writing}{+0.8}
\setval{4b.gd2po.minus.papo.judge2.writing-abs}{0.8}
\setval{4b.gd2po.minus.papo.judge2.writingbench}{+0.4}
\setval{4b.gd2po.minus.papo.judge2.writingbench-abs}{0.4}
\setval{4b.gd2po.minus.papo.livecodebench}{$-$0.1}
\setval{4b.gd2po.minus.papo.livecodebench-abs}{0.1}
\setval{4b.gd2po.minus.papo.longbench-write}{+5.0}
\setval{4b.gd2po.minus.papo.longbench-write-abs}{5.0}
\setval{4b.gd2po.minus.papo.math}{+1.7}
\setval{4b.gd2po.minus.papo.math-abs}{1.7}
\setval{4b.gd2po.minus.papo.math500}{+1.1}
\setval{4b.gd2po.minus.papo.math500-abs}{1.1}
\setval{4b.gd2po.minus.papo.mbpp-plus}{+0.7}
\setval{4b.gd2po.minus.papo.mbpp-plus-abs}{0.7}
\setval{4b.gd2po.minus.papo.minerva}{+0.7}
\setval{4b.gd2po.minus.papo.minerva-abs}{0.7}
\setval{4b.gd2po.minus.papo.olympiad-bench}{+2.1}
\setval{4b.gd2po.minus.papo.olympiad-bench-abs}{2.1}
\setval{4b.gd2po.minus.papo.writing}{+2.3}
\setval{4b.gd2po.minus.papo.writing-abs}{2.3}
\setval{4b.gd2po.minus.papo.writingbench}{+3.7}
\setval{4b.gd2po.minus.papo.writingbench-abs}{3.7}
\setval{4b.gd2po.minus.saw.aime24}{$-$0.6}
\setval{4b.gd2po.minus.saw.aime24-abs}{0.6}
\setval{4b.gd2po.minus.saw.aime25}{$-$4.4}
\setval{4b.gd2po.minus.saw.aime25-abs}{4.4}
\setval{4b.gd2po.minus.saw.amc23}{+1.1}
\setval{4b.gd2po.minus.saw.amc23-abs}{1.1}
\setval{4b.gd2po.minus.saw.avg}{$-$0.1}
\setval{4b.gd2po.minus.saw.avg-abs}{0.1}
\setval{4b.gd2po.minus.saw.code}{+1.6}
\setval{4b.gd2po.minus.saw.code-abs}{1.6}
\setval{4b.gd2po.minus.saw.creative-writing-v3}{$-$0.4}
\setval{4b.gd2po.minus.saw.creative-writing-v3-abs}{0.4}
\setval{4b.gd2po.minus.saw.guidebench}{$-$1.9}
\setval{4b.gd2po.minus.saw.guidebench-abs}{1.9}
\setval{4b.gd2po.minus.saw.humaneval-plus}{+2.7}
\setval{4b.gd2po.minus.saw.humaneval-plus-abs}{2.7}
\setval{4b.gd2po.minus.saw.if}{$-$1.3}
\setval{4b.gd2po.minus.saw.if-abs}{1.3}
\setval{4b.gd2po.minus.saw.ifbench}{$-$2.7}
\setval{4b.gd2po.minus.saw.ifbench-abs}{2.7}
\setval{4b.gd2po.minus.saw.ifeval}{+0.7}
\setval{4b.gd2po.minus.saw.ifeval-abs}{0.7}
\setval{4b.gd2po.minus.saw.judge2.avg}{$-$0.5}
\setval{4b.gd2po.minus.saw.judge2.avg-abs}{0.5}
\setval{4b.gd2po.minus.saw.judge2.creative-writing-v3}{$-$2.7}
\setval{4b.gd2po.minus.saw.judge2.creative-writing-v3-abs}{2.7}
\setval{4b.gd2po.minus.saw.judge2.longbench-write}{+0.9}
\setval{4b.gd2po.minus.saw.judge2.longbench-write-abs}{0.9}
\setval{4b.gd2po.minus.saw.judge2.writing}{$-$0.9}
\setval{4b.gd2po.minus.saw.judge2.writing-abs}{0.9}
\setval{4b.gd2po.minus.saw.judge2.writingbench}{$-$0.8}
\setval{4b.gd2po.minus.saw.judge2.writingbench-abs}{0.8}
\setval{4b.gd2po.minus.saw.livecodebench}{+0.7}
\setval{4b.gd2po.minus.saw.livecodebench-abs}{0.7}
\setval{4b.gd2po.minus.saw.longbench-write}{+2.7}
\setval{4b.gd2po.minus.saw.longbench-write-abs}{2.7}
\setval{4b.gd2po.minus.saw.math}{$-$0.9}
\setval{4b.gd2po.minus.saw.math-abs}{0.9}
\setval{4b.gd2po.minus.saw.math500}{$-$0.2}
\setval{4b.gd2po.minus.saw.math500-abs}{0.2}
\setval{4b.gd2po.minus.saw.mbpp-plus}{+1.3}
\setval{4b.gd2po.minus.saw.mbpp-plus-abs}{1.3}
\setval{4b.gd2po.minus.saw.minerva}{$-$1.0}
\setval{4b.gd2po.minus.saw.minerva-abs}{1.0}
\setval{4b.gd2po.minus.saw.olympiad-bench}{$-$0.3}
\setval{4b.gd2po.minus.saw.olympiad-bench-abs}{0.3}
\setval{4b.gd2po.minus.saw.writing}{+0.9}
\setval{4b.gd2po.minus.saw.writing-abs}{0.9}
\setval{4b.gd2po.minus.saw.writingbench}{+0.3}
\setval{4b.gd2po.minus.saw.writingbench-abs}{0.3}
\setval{4b.gd2po.passing-at-ceiling.code.early}{0.71}
\setval{4b.gd2po.passing-at-ceiling.code.late}{0.73}
\setval{4b.gd2po.passing-at-ceiling.code.run}{0.73}
\setval{4b.gd2po.passing-at-ceiling.if.early}{0.28}
\setval{4b.gd2po.passing-at-ceiling.if.late}{0.56}
\setval{4b.gd2po.passing-at-ceiling.if.run}{0.48}
\setval{4b.gd2po.passing-at-ceiling.math.early}{1.00}
\setval{4b.gd2po.passing-at-ceiling.math.late}{1.00}
\setval{4b.gd2po.passing-at-ceiling.math.run}{1.00}
\setval{4b.gd2po.passing-at-ceiling.writing.early}{0.00}
\setval{4b.gd2po.passing-at-ceiling.writing.late}{0.00}
\setval{4b.gd2po.passing-at-ceiling.writing.run}{0.00}
\setval{4b.gd2po.primary-cosine.code.early}{0.95}
\setval{4b.gd2po.primary-cosine.code.late}{0.95}
\setval{4b.gd2po.primary-cosine.code.run}{0.95}
\setval{4b.gd2po.primary-cosine.if.early}{0.67}
\setval{4b.gd2po.primary-cosine.if.late}{0.71}
\setval{4b.gd2po.primary-cosine.if.run}{0.73}
\setval{4b.gd2po.primary-cosine.math.early}{0.97}
\setval{4b.gd2po.primary-cosine.math.late}{0.97}
\setval{4b.gd2po.primary-cosine.math.run}{0.98}
\setval{4b.gd2po.primary-cosine.writing.early}{0.80}
\setval{4b.gd2po.primary-cosine.writing.late}{0.66}
\setval{4b.gd2po.primary-cosine.writing.run}{0.64}
\setval{4b.gd2po.primary-level.code.early}{0.53}
\setval{4b.gd2po.primary-level.code.late}{0.76}
\setval{4b.gd2po.primary-level.code.run}{0.71}
\setval{4b.gd2po.primary-level.if.early}{0.31}
\setval{4b.gd2po.primary-level.if.late}{0.74}
\setval{4b.gd2po.primary-level.if.run}{0.63}
\setval{4b.gd2po.primary-level.math.early}{0.47}
\setval{4b.gd2po.primary-level.math.late}{0.69}
\setval{4b.gd2po.primary-level.math.run}{0.63}
\setval{4b.gd2po.primary-level.writing.early}{0.53}
\setval{4b.gd2po.primary-level.writing.late}{0.75}
\setval{4b.gd2po.primary-level.writing.run}{0.71}
\setval{4b.gd2po.solved.code.early}{0.06}
\setval{4b.gd2po.solved.code.late}{0.02}
\setval{4b.gd2po.solved.code.run}{0.02}
\setval{4b.gd2po.solved.if.early}{0.02}
\setval{4b.gd2po.solved.if.late}{0.19}
\setval{4b.gd2po.solved.if.run}{0.13}
\setval{4b.gd2po.solved.math.early}{0.12}
\setval{4b.gd2po.solved.math.late}{0.15}
\setval{4b.gd2po.solved.math.run}{0.13}
\setval{4b.gd2po.solved.writing.early}{0.00}
\setval{4b.gd2po.solved.writing.late}{0.00}
\setval{4b.gd2po.solved.writing.run}{0.00}
\setval{4b.gd2po.update-share.code.early}{0.26}
\setval{4b.gd2po.update-share.code.late}{0.30}
\setval{4b.gd2po.update-share.code.run}{0.28}
\setval{4b.gd2po.update-share.if.early}{0.14}
\setval{4b.gd2po.update-share.if.late}{0.16}
\setval{4b.gd2po.update-share.if.run}{0.18}
\setval{4b.gd2po.update-share.math.early}{0.25}
\setval{4b.gd2po.update-share.math.late}{0.34}
\setval{4b.gd2po.update-share.math.run}{0.33}
\setval{4b.gd2po.update-share.writing.early}{0.36}
\setval{4b.gd2po.update-share.writing.late}{0.21}
\setval{4b.gd2po.update-share.writing.run}{0.21}
\setval{4b.gdpo-noaux.aux-change.code}{$-$0.28}
\setval{4b.gdpo-noaux.aux-change.if}{+0.10}
\setval{4b.gdpo-noaux.aux-change.math}{$-$0.31}
\setval{4b.gdpo-noaux.aux-change.writing}{$-$0.61}
\setval{4b.gdpo-noaux.aux-change.writing-rm}{+0.24}
\setval{4b.gdpo-noaux.aux-full.code.early}{0.60}
\setval{4b.gdpo-noaux.aux-full.code.late}{0.14}
\setval{4b.gdpo-noaux.aux-full.code.run}{0.34}
\setval{4b.gdpo-noaux.aux-full.if.early}{0.58}
\setval{4b.gdpo-noaux.aux-full.if.late}{0.67}
\setval{4b.gdpo-noaux.aux-full.if.run}{0.62}
\setval{4b.gdpo-noaux.aux-full.math.early}{0.60}
\setval{4b.gdpo-noaux.aux-full.math.late}{0.14}
\setval{4b.gdpo-noaux.aux-full.math.run}{0.32}
\setval{4b.gdpo-noaux.aux-full.writing-rm.early}{0.10}
\setval{4b.gdpo-noaux.aux-full.writing-rm.late}{0.05}
\setval{4b.gdpo-noaux.aux-full.writing-rm.run}{0.14}
\setval{4b.gdpo-noaux.aux-full.writing.early}{0.54}
\setval{4b.gdpo-noaux.aux-full.writing.late}{0.06}
\setval{4b.gdpo-noaux.aux-full.writing.run}{0.12}
\setval{4b.gdpo-noaux.aux-level.code.early}{0.89}
\setval{4b.gdpo-noaux.aux-level.code.late}{0.51}
\setval{4b.gdpo-noaux.aux-level.code.run}{0.74}
\setval{4b.gdpo-noaux.aux-level.if.early}{0.78}
\setval{4b.gdpo-noaux.aux-level.if.late}{0.81}
\setval{4b.gdpo-noaux.aux-level.if.run}{0.78}
\setval{4b.gdpo-noaux.aux-level.math.early}{0.91}
\setval{4b.gdpo-noaux.aux-level.math.late}{0.54}
\setval{4b.gdpo-noaux.aux-level.math.run}{0.71}
\setval{4b.gdpo-noaux.aux-level.writing-rm.early}{0.77}
\setval{4b.gdpo-noaux.aux-level.writing-rm.late}{0.83}
\setval{4b.gdpo-noaux.aux-level.writing-rm.run}{0.85}
\setval{4b.gdpo-noaux.aux-level.writing.early}{0.85}
\setval{4b.gdpo-noaux.aux-level.writing.late}{0.15}
\setval{4b.gdpo-noaux.aux-level.writing.run}{0.28}
\setval{4b.gdpo-noaux.bench.aime24}{23.4}
\setval{4b.gdpo-noaux.bench.aime25}{18.8}
\setval{4b.gdpo-noaux.bench.amc23}{65.7}
\setval{4b.gdpo-noaux.bench.creative-writing-v3}{39.3}
\setval{4b.gdpo-noaux.bench.guidebench}{75.9}
\setval{4b.gdpo-noaux.bench.humaneval-plus}{85.5}
\setval{4b.gdpo-noaux.bench.ifbench}{40.3}
\setval{4b.gdpo-noaux.bench.ifeval}{84.8}
\setval{4b.gdpo-noaux.bench.livecodebench}{24.7}
\setval{4b.gdpo-noaux.bench.longbench-write}{64.4}
\setval{4b.gdpo-noaux.bench.math500}{84.8}
\setval{4b.gdpo-noaux.bench.mbpp-plus}{72.3}
\setval{4b.gdpo-noaux.bench.minerva}{49.5}
\setval{4b.gdpo-noaux.bench.olympiad-bench}{53.2}
\setval{4b.gdpo-noaux.bench.writingbench}{67.3}
\setval{4b.gdpo-noaux.eval.avg}{56.7}
\setval{4b.gdpo-noaux.eval.code}{60.9}
\setval{4b.gdpo-noaux.eval.if}{67.0}
\setval{4b.gdpo-noaux.eval.math}{49.2}
\setval{4b.gdpo-noaux.eval.writing}{57.0}
\setval{4b.gdpo-noaux.gate.code.against.early}{0.04}
\setval{4b.gdpo-noaux.gate.code.against.late}{0.11}
\setval{4b.gdpo-noaux.gate.code.against.run}{0.09}
\setval{4b.gdpo-noaux.gate.code.agreement}{+0.020}
\setval{4b.gdpo-noaux.gate.code.ceiling.early}{0.14}
\setval{4b.gdpo-noaux.gate.code.ceiling.late}{0.49}
\setval{4b.gdpo-noaux.gate.code.ceiling.run}{0.37}
\setval{4b.gdpo-noaux.gate.code.cosine.early}{0.81}
\setval{4b.gdpo-noaux.gate.code.cosine.late}{0.34}
\setval{4b.gdpo-noaux.gate.code.cosine.run}{0.51}
\setval{4b.gdpo-noaux.gate.if.against.early}{0.05}
\setval{4b.gdpo-noaux.gate.if.against.late}{0.07}
\setval{4b.gdpo-noaux.gate.if.against.run}{0.07}
\setval{4b.gdpo-noaux.gate.if.agreement}{+0.035}
\setval{4b.gdpo-noaux.gate.if.ceiling.early}{0.06}
\setval{4b.gdpo-noaux.gate.if.ceiling.late}{0.29}
\setval{4b.gdpo-noaux.gate.if.ceiling.run}{0.20}
\setval{4b.gdpo-noaux.gate.if.cosine.early}{0.69}
\setval{4b.gdpo-noaux.gate.if.cosine.late}{0.39}
\setval{4b.gdpo-noaux.gate.if.cosine.run}{0.49}
\setval{4b.gdpo-noaux.gate.math.against.early}{0.00}
\setval{4b.gdpo-noaux.gate.math.against.late}{0.00}
\setval{4b.gdpo-noaux.gate.math.against.run}{0.00}
\setval{4b.gdpo-noaux.gate.math.ceiling.early}{0.21}
\setval{4b.gdpo-noaux.gate.math.ceiling.late}{0.79}
\setval{4b.gdpo-noaux.gate.math.ceiling.run}{0.61}
\setval{4b.gdpo-noaux.gate.math.cosine.early}{0.87}
\setval{4b.gdpo-noaux.gate.math.cosine.late}{0.35}
\setval{4b.gdpo-noaux.gate.math.cosine.run}{0.53}
\setval{4b.gdpo-noaux.gate.writing-rm.against.early}{0.22}
\setval{4b.gdpo-noaux.gate.writing-rm.against.late}{0.56}
\setval{4b.gdpo-noaux.gate.writing-rm.against.run}{0.52}
\setval{4b.gdpo-noaux.gate.writing-rm.agreement}{+0.077}
\setval{4b.gdpo-noaux.gate.writing-rm.ceiling.early}{0.00}
\setval{4b.gdpo-noaux.gate.writing-rm.ceiling.late}{0.01}
\setval{4b.gdpo-noaux.gate.writing-rm.ceiling.run}{0.01}
\setval{4b.gdpo-noaux.gate.writing-rm.cosine.early}{0.67}
\setval{4b.gdpo-noaux.gate.writing-rm.cosine.late}{0.12}
\setval{4b.gdpo-noaux.gate.writing-rm.cosine.run}{0.20}
\setval{4b.gdpo-noaux.gate.writing.against.early}{0.34}
\setval{4b.gdpo-noaux.gate.writing.against.late}{0.57}
\setval{4b.gdpo-noaux.gate.writing.against.run}{0.61}
\setval{4b.gdpo-noaux.gate.writing.agreement}{$-$0.016}
\setval{4b.gdpo-noaux.gate.writing.ceiling.early}{0.00}
\setval{4b.gdpo-noaux.gate.writing.ceiling.late}{0.01}
\setval{4b.gdpo-noaux.gate.writing.ceiling.run}{0.00}
\setval{4b.gdpo-noaux.gate.writing.cosine.early}{0.47}
\setval{4b.gdpo-noaux.gate.writing.cosine.late}{0.04}
\setval{4b.gdpo-noaux.gate.writing.cosine.run}{0.05}
\setval{4b.gdpo-noaux.judge2.avg}{53.3}
\setval{4b.gdpo-noaux.judge2.creative-writing-v3}{23.4}
\setval{4b.gdpo-noaux.judge2.longbench-write}{50.8}
\setval{4b.gdpo-noaux.judge2.writing}{40.1}
\setval{4b.gdpo-noaux.judge2.writingbench}{46.0}
\setval{4b.gdpo-noaux.last-step}{300}
\setval{4b.gdpo-noaux.minus.base.aime24}{+15.6}
\setval{4b.gdpo-noaux.minus.base.aime24-abs}{15.6}
\setval{4b.gdpo-noaux.minus.base.aime25}{+11.7}
\setval{4b.gdpo-noaux.minus.base.aime25-abs}{11.7}
\setval{4b.gdpo-noaux.minus.base.amc23}{+21.8}
\setval{4b.gdpo-noaux.minus.base.amc23-abs}{21.8}
\setval{4b.gdpo-noaux.minus.base.avg}{+22.1}
\setval{4b.gdpo-noaux.minus.base.avg-abs}{22.1}
\setval{4b.gdpo-noaux.minus.base.code}{+16.2}
\setval{4b.gdpo-noaux.minus.base.code-abs}{16.2}
\setval{4b.gdpo-noaux.minus.base.creative-writing-v3}{+14.6}
\setval{4b.gdpo-noaux.minus.base.creative-writing-v3-abs}{14.6}
\setval{4b.gdpo-noaux.minus.base.guidebench}{+51.1}
\setval{4b.gdpo-noaux.minus.base.guidebench-abs}{51.1}
\setval{4b.gdpo-noaux.minus.base.humaneval-plus}{+20.4}
\setval{4b.gdpo-noaux.minus.base.humaneval-plus-abs}{20.4}
\setval{4b.gdpo-noaux.minus.base.if}{+37.3}
\setval{4b.gdpo-noaux.minus.base.if-abs}{37.3}
\setval{4b.gdpo-noaux.minus.base.ifbench}{+23.3}
\setval{4b.gdpo-noaux.minus.base.ifbench-abs}{23.3}
\setval{4b.gdpo-noaux.minus.base.ifeval}{+37.4}
\setval{4b.gdpo-noaux.minus.base.ifeval-abs}{37.4}
\setval{4b.gdpo-noaux.minus.base.judge2.avg}{+19.4}
\setval{4b.gdpo-noaux.minus.base.judge2.avg-abs}{19.4}
\setval{4b.gdpo-noaux.minus.base.judge2.creative-writing-v3}{+2.5}
\setval{4b.gdpo-noaux.minus.base.judge2.creative-writing-v3-abs}{2.5}
\setval{4b.gdpo-noaux.minus.base.judge2.longbench-write}{+9.9}
\setval{4b.gdpo-noaux.minus.base.judge2.longbench-write-abs}{9.9}
\setval{4b.gdpo-noaux.minus.base.judge2.writing}{+9.3}
\setval{4b.gdpo-noaux.minus.base.judge2.writing-abs}{9.3}
\setval{4b.gdpo-noaux.minus.base.judge2.writingbench}{+15.6}
\setval{4b.gdpo-noaux.minus.base.judge2.writingbench-abs}{15.6}
\setval{4b.gdpo-noaux.minus.base.livecodebench}{+11.5}
\setval{4b.gdpo-noaux.minus.base.livecodebench-abs}{11.5}
\setval{4b.gdpo-noaux.minus.base.longbench-write}{+19.9}
\setval{4b.gdpo-noaux.minus.base.longbench-write-abs}{19.9}
\setval{4b.gdpo-noaux.minus.base.math}{+17.0}
\setval{4b.gdpo-noaux.minus.base.math-abs}{17.0}
\setval{4b.gdpo-noaux.minus.base.math500}{+17.5}
\setval{4b.gdpo-noaux.minus.base.math500-abs}{17.5}
\setval{4b.gdpo-noaux.minus.base.mbpp-plus}{+16.7}
\setval{4b.gdpo-noaux.minus.base.mbpp-plus-abs}{16.7}
\setval{4b.gdpo-noaux.minus.base.minerva}{+17.9}
\setval{4b.gdpo-noaux.minus.base.minerva-abs}{17.9}
\setval{4b.gdpo-noaux.minus.base.olympiad-bench}{+17.5}
\setval{4b.gdpo-noaux.minus.base.olympiad-bench-abs}{17.5}
\setval{4b.gdpo-noaux.minus.base.writing}{+23.1}
\setval{4b.gdpo-noaux.minus.base.writing-abs}{23.1}
\setval{4b.gdpo-noaux.minus.base.writingbench}{+34.9}
\setval{4b.gdpo-noaux.minus.base.writingbench-abs}{34.9}
\setval{4b.gdpo-noaux.minus.concur-free.aime24}{+11.6}
\setval{4b.gdpo-noaux.minus.concur-free.aime24-abs}{11.6}
\setval{4b.gdpo-noaux.minus.concur-free.aime25}{+4.2}
\setval{4b.gdpo-noaux.minus.concur-free.aime25-abs}{4.2}
\setval{4b.gdpo-noaux.minus.concur-free.amc23}{+11.6}
\setval{4b.gdpo-noaux.minus.concur-free.amc23-abs}{11.6}
\setval{4b.gdpo-noaux.minus.concur-free.avg}{+1.1}
\setval{4b.gdpo-noaux.minus.concur-free.avg-abs}{1.1}
\setval{4b.gdpo-noaux.minus.concur-free.code}{+2.5}
\setval{4b.gdpo-noaux.minus.concur-free.code-abs}{2.5}
\setval{4b.gdpo-noaux.minus.concur-free.creative-writing-v3}{$-$6.1}
\setval{4b.gdpo-noaux.minus.concur-free.creative-writing-v3-abs}{6.1}
\setval{4b.gdpo-noaux.minus.concur-free.guidebench}{$-$3.1}
\setval{4b.gdpo-noaux.minus.concur-free.guidebench-abs}{3.1}
\setval{4b.gdpo-noaux.minus.concur-free.humaneval-plus}{+4.4}
\setval{4b.gdpo-noaux.minus.concur-free.humaneval-plus-abs}{4.4}
\setval{4b.gdpo-noaux.minus.concur-free.if}{$-$1.2}
\setval{4b.gdpo-noaux.minus.concur-free.if-abs}{1.2}
\setval{4b.gdpo-noaux.minus.concur-free.ifbench}{+1.3}
\setval{4b.gdpo-noaux.minus.concur-free.ifbench-abs}{1.3}
\setval{4b.gdpo-noaux.minus.concur-free.ifeval}{$-$1.9}
\setval{4b.gdpo-noaux.minus.concur-free.ifeval-abs}{1.9}
\setval{4b.gdpo-noaux.minus.concur-free.judge2.avg}{+0.5}
\setval{4b.gdpo-noaux.minus.concur-free.judge2.avg-abs}{0.5}
\setval{4b.gdpo-noaux.minus.concur-free.judge2.creative-writing-v3}{$-$11.9}
\setval{4b.gdpo-noaux.minus.concur-free.judge2.creative-writing-v3-abs}{11.9}
\setval{4b.gdpo-noaux.minus.concur-free.judge2.longbench-write}{$-$18.6}
\setval{4b.gdpo-noaux.minus.concur-free.judge2.longbench-write-abs}{18.6}
\setval{4b.gdpo-noaux.minus.concur-free.judge2.writing}{$-$13.0}
\setval{4b.gdpo-noaux.minus.concur-free.judge2.writing-abs}{13.0}
\setval{4b.gdpo-noaux.minus.concur-free.judge2.writingbench}{$-$8.4}
\setval{4b.gdpo-noaux.minus.concur-free.judge2.writingbench-abs}{8.4}
\setval{4b.gdpo-noaux.minus.concur-free.livecodebench}{+0.6}
\setval{4b.gdpo-noaux.minus.concur-free.livecodebench-abs}{0.6}
\setval{4b.gdpo-noaux.minus.concur-free.longbench-write}{$-$13.4}
\setval{4b.gdpo-noaux.minus.concur-free.longbench-write-abs}{13.4}
\setval{4b.gdpo-noaux.minus.concur-free.math}{+7.1}
\setval{4b.gdpo-noaux.minus.concur-free.math-abs}{7.1}
\setval{4b.gdpo-noaux.minus.concur-free.math500}{+4.4}
\setval{4b.gdpo-noaux.minus.concur-free.math500-abs}{4.4}
\setval{4b.gdpo-noaux.minus.concur-free.mbpp-plus}{+2.7}
\setval{4b.gdpo-noaux.minus.concur-free.mbpp-plus-abs}{2.7}
\setval{4b.gdpo-noaux.minus.concur-free.minerva}{+3.3}
\setval{4b.gdpo-noaux.minus.concur-free.minerva-abs}{3.3}
\setval{4b.gdpo-noaux.minus.concur-free.olympiad-bench}{+7.4}
\setval{4b.gdpo-noaux.minus.concur-free.olympiad-bench-abs}{7.4}
\setval{4b.gdpo-noaux.minus.concur-free.writing}{$-$9.8}
\setval{4b.gdpo-noaux.minus.concur-free.writing-abs}{9.8}
\setval{4b.gdpo-noaux.minus.concur-free.writingbench}{$-$10.0}
\setval{4b.gdpo-noaux.minus.concur-free.writingbench-abs}{10.0}
\setval{4b.gdpo-noaux.minus.concur-tied.aime24}{+4.0}
\setval{4b.gdpo-noaux.minus.concur-tied.aime24-abs}{4.0}
\setval{4b.gdpo-noaux.minus.concur-tied.aime25}{+2.4}
\setval{4b.gdpo-noaux.minus.concur-tied.aime25-abs}{2.4}
\setval{4b.gdpo-noaux.minus.concur-tied.amc23}{$-$3.5}
\setval{4b.gdpo-noaux.minus.concur-tied.amc23-abs}{3.5}
\setval{4b.gdpo-noaux.minus.concur-tied.avg}{$-$2.0}
\setval{4b.gdpo-noaux.minus.concur-tied.avg-abs}{2.0}
\setval{4b.gdpo-noaux.minus.concur-tied.code}{+1.7}
\setval{4b.gdpo-noaux.minus.concur-tied.code-abs}{1.7}
\setval{4b.gdpo-noaux.minus.concur-tied.creative-writing-v3}{$-$8.5}
\setval{4b.gdpo-noaux.minus.concur-tied.creative-writing-v3-abs}{8.5}
\setval{4b.gdpo-noaux.minus.concur-tied.guidebench}{$-$3.3}
\setval{4b.gdpo-noaux.minus.concur-tied.guidebench-abs}{3.3}
\setval{4b.gdpo-noaux.minus.concur-tied.humaneval-plus}{+1.5}
\setval{4b.gdpo-noaux.minus.concur-tied.humaneval-plus-abs}{1.5}
\setval{4b.gdpo-noaux.minus.concur-tied.if}{$-$1.2}
\setval{4b.gdpo-noaux.minus.concur-tied.if-abs}{1.2}
\setval{4b.gdpo-noaux.minus.concur-tied.ifbench}{+2.7}
\setval{4b.gdpo-noaux.minus.concur-tied.ifbench-abs}{2.7}
\setval{4b.gdpo-noaux.minus.concur-tied.ifeval}{$-$3.0}
\setval{4b.gdpo-noaux.minus.concur-tied.ifeval-abs}{3.0}
\setval{4b.gdpo-noaux.minus.concur-tied.judge2.avg}{$-$2.2}
\setval{4b.gdpo-noaux.minus.concur-tied.judge2.avg-abs}{2.2}
\setval{4b.gdpo-noaux.minus.concur-tied.judge2.creative-writing-v3}{$-$10.6}
\setval{4b.gdpo-noaux.minus.concur-tied.judge2.creative-writing-v3-abs}{10.6}
\setval{4b.gdpo-noaux.minus.concur-tied.judge2.longbench-write}{$-$17.2}
\setval{4b.gdpo-noaux.minus.concur-tied.judge2.longbench-write-abs}{17.2}
\setval{4b.gdpo-noaux.minus.concur-tied.judge2.writing}{$-$12.4}
\setval{4b.gdpo-noaux.minus.concur-tied.judge2.writing-abs}{12.4}
\setval{4b.gdpo-noaux.minus.concur-tied.judge2.writingbench}{$-$9.3}
\setval{4b.gdpo-noaux.minus.concur-tied.judge2.writingbench-abs}{9.3}
\setval{4b.gdpo-noaux.minus.concur-tied.livecodebench}{+0.0}
\setval{4b.gdpo-noaux.minus.concur-tied.livecodebench-abs}{0.0}
\setval{4b.gdpo-noaux.minus.concur-tied.longbench-write}{$-$13.4}
\setval{4b.gdpo-noaux.minus.concur-tied.longbench-write-abs}{13.4}
\setval{4b.gdpo-noaux.minus.concur-tied.math}{+0.5}
\setval{4b.gdpo-noaux.minus.concur-tied.math-abs}{0.5}
\setval{4b.gdpo-noaux.minus.concur-tied.math500}{$-$0.8}
\setval{4b.gdpo-noaux.minus.concur-tied.math500-abs}{0.8}
\setval{4b.gdpo-noaux.minus.concur-tied.mbpp-plus}{+3.7}
\setval{4b.gdpo-noaux.minus.concur-tied.mbpp-plus-abs}{3.7}
\setval{4b.gdpo-noaux.minus.concur-tied.minerva}{+0.4}
\setval{4b.gdpo-noaux.minus.concur-tied.minerva-abs}{0.4}
\setval{4b.gdpo-noaux.minus.concur-tied.olympiad-bench}{+0.3}
\setval{4b.gdpo-noaux.minus.concur-tied.olympiad-bench-abs}{0.3}
\setval{4b.gdpo-noaux.minus.concur-tied.writing}{$-$11.6}
\setval{4b.gdpo-noaux.minus.concur-tied.writing-abs}{11.6}
\setval{4b.gdpo-noaux.minus.concur-tied.writingbench}{$-$12.8}
\setval{4b.gdpo-noaux.minus.concur-tied.writingbench-abs}{12.8}
\setval{4b.gdpo-noaux.minus.concur.aime24}{+2.9}
\setval{4b.gdpo-noaux.minus.concur.aime24-abs}{2.9}
\setval{4b.gdpo-noaux.minus.concur.aime25}{+0.0}
\setval{4b.gdpo-noaux.minus.concur.aime25-abs}{0.0}
\setval{4b.gdpo-noaux.minus.concur.amc23}{$-$3.7}
\setval{4b.gdpo-noaux.minus.concur.amc23-abs}{3.7}
\setval{4b.gdpo-noaux.minus.concur.avg}{$-$2.1}
\setval{4b.gdpo-noaux.minus.concur.avg-abs}{2.1}
\setval{4b.gdpo-noaux.minus.concur.code}{+1.2}
\setval{4b.gdpo-noaux.minus.concur.code-abs}{1.2}
\setval{4b.gdpo-noaux.minus.concur.creative-writing-v3}{$-$5.4}
\setval{4b.gdpo-noaux.minus.concur.creative-writing-v3-abs}{5.4}
\setval{4b.gdpo-noaux.minus.concur.guidebench}{$-$1.4}
\setval{4b.gdpo-noaux.minus.concur.guidebench-abs}{1.4}
\setval{4b.gdpo-noaux.minus.concur.humaneval-plus}{+3.0}
\setval{4b.gdpo-noaux.minus.concur.humaneval-plus-abs}{3.0}
\setval{4b.gdpo-noaux.minus.concur.if}{$-$1.6}
\setval{4b.gdpo-noaux.minus.concur.if-abs}{1.6}
\setval{4b.gdpo-noaux.minus.concur.ifbench}{+0.0}
\setval{4b.gdpo-noaux.minus.concur.ifbench-abs}{0.0}
\setval{4b.gdpo-noaux.minus.concur.ifeval}{$-$3.4}
\setval{4b.gdpo-noaux.minus.concur.ifeval-abs}{3.4}
\setval{4b.gdpo-noaux.minus.concur.judge2.avg}{$-$2.5}
\setval{4b.gdpo-noaux.minus.concur.judge2.avg-abs}{2.5}
\setval{4b.gdpo-noaux.minus.concur.judge2.creative-writing-v3}{$-$9.0}
\setval{4b.gdpo-noaux.minus.concur.judge2.creative-writing-v3-abs}{9.0}
\setval{4b.gdpo-noaux.minus.concur.judge2.longbench-write}{$-$15.9}
\setval{4b.gdpo-noaux.minus.concur.judge2.longbench-write-abs}{15.9}
\setval{4b.gdpo-noaux.minus.concur.judge2.writing}{$-$11.6}
\setval{4b.gdpo-noaux.minus.concur.judge2.writing-abs}{11.6}
\setval{4b.gdpo-noaux.minus.concur.judge2.writingbench}{$-$10.0}
\setval{4b.gdpo-noaux.minus.concur.judge2.writingbench-abs}{10.0}
\setval{4b.gdpo-noaux.minus.concur.livecodebench}{$-$0.4}
\setval{4b.gdpo-noaux.minus.concur.livecodebench-abs}{0.4}
\setval{4b.gdpo-noaux.minus.concur.longbench-write}{$-$11.9}
\setval{4b.gdpo-noaux.minus.concur.longbench-write-abs}{11.9}
\setval{4b.gdpo-noaux.minus.concur.math}{$-$0.2}
\setval{4b.gdpo-noaux.minus.concur.math-abs}{0.2}
\setval{4b.gdpo-noaux.minus.concur.math500}{$-$0.5}
\setval{4b.gdpo-noaux.minus.concur.math500-abs}{0.5}
\setval{4b.gdpo-noaux.minus.concur.mbpp-plus}{+0.8}
\setval{4b.gdpo-noaux.minus.concur.mbpp-plus-abs}{0.8}
\setval{4b.gdpo-noaux.minus.concur.minerva}{+0.3}
\setval{4b.gdpo-noaux.minus.concur.minerva-abs}{0.3}
\setval{4b.gdpo-noaux.minus.concur.olympiad-bench}{$-$0.4}
\setval{4b.gdpo-noaux.minus.concur.olympiad-bench-abs}{0.4}
\setval{4b.gdpo-noaux.minus.concur.writing}{$-$9.8}
\setval{4b.gdpo-noaux.minus.concur.writing-abs}{9.8}
\setval{4b.gdpo-noaux.minus.concur.writingbench}{$-$12.1}
\setval{4b.gdpo-noaux.minus.concur.writingbench-abs}{12.1}
\setval{4b.gdpo-noaux.minus.gd2po.aime24}{+11.6}
\setval{4b.gdpo-noaux.minus.gd2po.aime24-abs}{11.6}
\setval{4b.gdpo-noaux.minus.gd2po.aime25}{+12.0}
\setval{4b.gdpo-noaux.minus.gd2po.aime25-abs}{12.0}
\setval{4b.gdpo-noaux.minus.gd2po.amc23}{+8.0}
\setval{4b.gdpo-noaux.minus.gd2po.amc23-abs}{8.0}
\setval{4b.gdpo-noaux.minus.gd2po.avg}{+2.6}
\setval{4b.gdpo-noaux.minus.gd2po.avg-abs}{2.6}
\setval{4b.gdpo-noaux.minus.gd2po.code}{+1.4}
\setval{4b.gdpo-noaux.minus.gd2po.code-abs}{1.4}
\setval{4b.gdpo-noaux.minus.gd2po.creative-writing-v3}{$-$5.4}
\setval{4b.gdpo-noaux.minus.gd2po.creative-writing-v3-abs}{5.4}
\setval{4b.gdpo-noaux.minus.gd2po.guidebench}{$-$3.2}
\setval{4b.gdpo-noaux.minus.gd2po.guidebench-abs}{3.2}
\setval{4b.gdpo-noaux.minus.gd2po.humaneval-plus}{+2.1}
\setval{4b.gdpo-noaux.minus.gd2po.humaneval-plus-abs}{2.1}
\setval{4b.gdpo-noaux.minus.gd2po.if}{$-$1.1}
\setval{4b.gdpo-noaux.minus.gd2po.if-abs}{1.1}
\setval{4b.gdpo-noaux.minus.gd2po.ifbench}{+2.7}
\setval{4b.gdpo-noaux.minus.gd2po.ifbench-abs}{2.7}
\setval{4b.gdpo-noaux.minus.gd2po.ifeval}{$-$2.9}
\setval{4b.gdpo-noaux.minus.gd2po.ifeval-abs}{2.9}
\setval{4b.gdpo-noaux.minus.gd2po.judge2.avg}{+1.7}
\setval{4b.gdpo-noaux.minus.gd2po.judge2.avg-abs}{1.7}
\setval{4b.gdpo-noaux.minus.gd2po.judge2.creative-writing-v3}{$-$9.5}
\setval{4b.gdpo-noaux.minus.gd2po.judge2.creative-writing-v3-abs}{9.5}
\setval{4b.gdpo-noaux.minus.gd2po.judge2.longbench-write}{$-$10.2}
\setval{4b.gdpo-noaux.minus.gd2po.judge2.longbench-write-abs}{10.2}
\setval{4b.gdpo-noaux.minus.gd2po.judge2.writing}{$-$8.7}
\setval{4b.gdpo-noaux.minus.gd2po.judge2.writing-abs}{8.7}
\setval{4b.gdpo-noaux.minus.gd2po.judge2.writingbench}{$-$6.5}
\setval{4b.gdpo-noaux.minus.gd2po.judge2.writingbench-abs}{6.5}
\setval{4b.gdpo-noaux.minus.gd2po.livecodebench}{+1.3}
\setval{4b.gdpo-noaux.minus.gd2po.livecodebench-abs}{1.3}
\setval{4b.gdpo-noaux.minus.gd2po.longbench-write}{$-$4.6}
\setval{4b.gdpo-noaux.minus.gd2po.longbench-write-abs}{4.6}
\setval{4b.gdpo-noaux.minus.gd2po.math}{+8.4}
\setval{4b.gdpo-noaux.minus.gd2po.math-abs}{8.4}
\setval{4b.gdpo-noaux.minus.gd2po.math500}{+7.1}
\setval{4b.gdpo-noaux.minus.gd2po.math500-abs}{7.1}
\setval{4b.gdpo-noaux.minus.gd2po.mbpp-plus}{+0.9}
\setval{4b.gdpo-noaux.minus.gd2po.mbpp-plus-abs}{0.9}
\setval{4b.gdpo-noaux.minus.gd2po.minerva}{+2.3}
\setval{4b.gdpo-noaux.minus.gd2po.minerva-abs}{2.3}
\setval{4b.gdpo-noaux.minus.gd2po.olympiad-bench}{+9.6}
\setval{4b.gdpo-noaux.minus.gd2po.olympiad-bench-abs}{9.6}
\setval{4b.gdpo-noaux.minus.gd2po.writing}{$-$4.2}
\setval{4b.gdpo-noaux.minus.gd2po.writing-abs}{4.2}
\setval{4b.gdpo-noaux.minus.gd2po.writingbench}{$-$2.8}
\setval{4b.gdpo-noaux.minus.gd2po.writingbench-abs}{2.8}
\setval{4b.gdpo-noaux.minus.gdpo-w0.5.aime24}{+8.9}
\setval{4b.gdpo-noaux.minus.gdpo-w0.5.aime24-abs}{8.9}
\setval{4b.gdpo-noaux.minus.gdpo-w0.5.aime25}{+8.4}
\setval{4b.gdpo-noaux.minus.gdpo-w0.5.aime25-abs}{8.4}
\setval{4b.gdpo-noaux.minus.gdpo-w0.5.amc23}{+7.3}
\setval{4b.gdpo-noaux.minus.gdpo-w0.5.amc23-abs}{7.3}
\setval{4b.gdpo-noaux.minus.gdpo-w0.5.avg}{+2.1}
\setval{4b.gdpo-noaux.minus.gdpo-w0.5.avg-abs}{2.1}
\setval{4b.gdpo-noaux.minus.gdpo-w0.5.code}{+1.7}
\setval{4b.gdpo-noaux.minus.gdpo-w0.5.code-abs}{1.7}
\setval{4b.gdpo-noaux.minus.gdpo-w0.5.creative-writing-v3}{+1.2}
\setval{4b.gdpo-noaux.minus.gdpo-w0.5.creative-writing-v3-abs}{1.2}
\setval{4b.gdpo-noaux.minus.gdpo-w0.5.guidebench}{$-$3.6}
\setval{4b.gdpo-noaux.minus.gdpo-w0.5.guidebench-abs}{3.6}
\setval{4b.gdpo-noaux.minus.gdpo-w0.5.humaneval-plus}{+3.5}
\setval{4b.gdpo-noaux.minus.gdpo-w0.5.humaneval-plus-abs}{3.5}
\setval{4b.gdpo-noaux.minus.gdpo-w0.5.if}{$-$1.6}
\setval{4b.gdpo-noaux.minus.gdpo-w0.5.if-abs}{1.6}
\setval{4b.gdpo-noaux.minus.gdpo-w0.5.ifbench}{+1.0}
\setval{4b.gdpo-noaux.minus.gdpo-w0.5.ifbench-abs}{1.0}
\setval{4b.gdpo-noaux.minus.gdpo-w0.5.ifeval}{$-$2.0}
\setval{4b.gdpo-noaux.minus.gdpo-w0.5.ifeval-abs}{2.0}
\setval{4b.gdpo-noaux.minus.gdpo-w0.5.judge2.avg}{+0.8}
\setval{4b.gdpo-noaux.minus.gdpo-w0.5.judge2.avg-abs}{0.8}
\setval{4b.gdpo-noaux.minus.gdpo-w0.5.judge2.creative-writing-v3}{$-$6.7}
\setval{4b.gdpo-noaux.minus.gdpo-w0.5.judge2.creative-writing-v3-abs}{6.7}
\setval{4b.gdpo-noaux.minus.gdpo-w0.5.judge2.longbench-write}{$-$9.8}
\setval{4b.gdpo-noaux.minus.gdpo-w0.5.judge2.longbench-write-abs}{9.8}
\setval{4b.gdpo-noaux.minus.gdpo-w0.5.judge2.writing}{$-$8.0}
\setval{4b.gdpo-noaux.minus.gdpo-w0.5.judge2.writing-abs}{8.0}
\setval{4b.gdpo-noaux.minus.gdpo-w0.5.judge2.writingbench}{$-$7.6}
\setval{4b.gdpo-noaux.minus.gdpo-w0.5.judge2.writingbench-abs}{7.6}
\setval{4b.gdpo-noaux.minus.gdpo-w0.5.livecodebench}{+0.6}
\setval{4b.gdpo-noaux.minus.gdpo-w0.5.livecodebench-abs}{0.6}
\setval{4b.gdpo-noaux.minus.gdpo-w0.5.longbench-write}{$-$3.0}
\setval{4b.gdpo-noaux.minus.gdpo-w0.5.longbench-write-abs}{3.0}
\setval{4b.gdpo-noaux.minus.gdpo-w0.5.math}{+6.0}
\setval{4b.gdpo-noaux.minus.gdpo-w0.5.math-abs}{6.0}
\setval{4b.gdpo-noaux.minus.gdpo-w0.5.math500}{+3.0}
\setval{4b.gdpo-noaux.minus.gdpo-w0.5.math500-abs}{3.0}
\setval{4b.gdpo-noaux.minus.gdpo-w0.5.mbpp-plus}{+1.1}
\setval{4b.gdpo-noaux.minus.gdpo-w0.5.mbpp-plus-abs}{1.1}
\setval{4b.gdpo-noaux.minus.gdpo-w0.5.minerva}{+3.4}
\setval{4b.gdpo-noaux.minus.gdpo-w0.5.minerva-abs}{3.4}
\setval{4b.gdpo-noaux.minus.gdpo-w0.5.olympiad-bench}{+4.7}
\setval{4b.gdpo-noaux.minus.gdpo-w0.5.olympiad-bench-abs}{4.7}
\setval{4b.gdpo-noaux.minus.gdpo-w0.5.writing}{$-$1.7}
\setval{4b.gdpo-noaux.minus.gdpo-w0.5.writing-abs}{1.7}
\setval{4b.gdpo-noaux.minus.gdpo-w0.5.writingbench}{$-$3.4}
\setval{4b.gdpo-noaux.minus.gdpo-w0.5.writingbench-abs}{3.4}
\setval{4b.gdpo-noaux.minus.gdpo.aime24}{+14.4}
\setval{4b.gdpo-noaux.minus.gdpo.aime24-abs}{14.4}
\setval{4b.gdpo-noaux.minus.gdpo.aime25}{+8.9}
\setval{4b.gdpo-noaux.minus.gdpo.aime25-abs}{8.9}
\setval{4b.gdpo-noaux.minus.gdpo.amc23}{+6.1}
\setval{4b.gdpo-noaux.minus.gdpo.amc23-abs}{6.1}
\setval{4b.gdpo-noaux.minus.gdpo.avg}{+2.3}
\setval{4b.gdpo-noaux.minus.gdpo.avg-abs}{2.3}
\setval{4b.gdpo-noaux.minus.gdpo.code}{+1.8}
\setval{4b.gdpo-noaux.minus.gdpo.code-abs}{1.8}
\setval{4b.gdpo-noaux.minus.gdpo.creative-writing-v3}{$-$6.6}
\setval{4b.gdpo-noaux.minus.gdpo.creative-writing-v3-abs}{6.6}
\setval{4b.gdpo-noaux.minus.gdpo.guidebench}{$-$4.7}
\setval{4b.gdpo-noaux.minus.gdpo.guidebench-abs}{4.7}
\setval{4b.gdpo-noaux.minus.gdpo.humaneval-plus}{+3.7}
\setval{4b.gdpo-noaux.minus.gdpo.humaneval-plus-abs}{3.7}
\setval{4b.gdpo-noaux.minus.gdpo.if}{$-$0.6}
\setval{4b.gdpo-noaux.minus.gdpo.if-abs}{0.6}
\setval{4b.gdpo-noaux.minus.gdpo.ifbench}{+4.0}
\setval{4b.gdpo-noaux.minus.gdpo.ifbench-abs}{4.0}
\setval{4b.gdpo-noaux.minus.gdpo.ifeval}{$-$1.0}
\setval{4b.gdpo-noaux.minus.gdpo.ifeval-abs}{1.0}
\setval{4b.gdpo-noaux.minus.gdpo.judge2.avg}{+1.4}
\setval{4b.gdpo-noaux.minus.gdpo.judge2.avg-abs}{1.4}
\setval{4b.gdpo-noaux.minus.gdpo.judge2.creative-writing-v3}{$-$11.0}
\setval{4b.gdpo-noaux.minus.gdpo.judge2.creative-writing-v3-abs}{11.0}
\setval{4b.gdpo-noaux.minus.gdpo.judge2.longbench-write}{$-$11.5}
\setval{4b.gdpo-noaux.minus.gdpo.judge2.longbench-write-abs}{11.5}
\setval{4b.gdpo-noaux.minus.gdpo.judge2.writing}{$-$10.2}
\setval{4b.gdpo-noaux.minus.gdpo.judge2.writing-abs}{10.2}
\setval{4b.gdpo-noaux.minus.gdpo.judge2.writingbench}{$-$8.0}
\setval{4b.gdpo-noaux.minus.gdpo.judge2.writingbench-abs}{8.0}
\setval{4b.gdpo-noaux.minus.gdpo.livecodebench}{+0.3}
\setval{4b.gdpo-noaux.minus.gdpo.livecodebench-abs}{0.3}
\setval{4b.gdpo-noaux.minus.gdpo.longbench-write}{$-$4.6}
\setval{4b.gdpo-noaux.minus.gdpo.longbench-write-abs}{4.6}
\setval{4b.gdpo-noaux.minus.gdpo.math}{+7.9}
\setval{4b.gdpo-noaux.minus.gdpo.math-abs}{7.9}
\setval{4b.gdpo-noaux.minus.gdpo.math500}{+5.5}
\setval{4b.gdpo-noaux.minus.gdpo.math500-abs}{5.5}
\setval{4b.gdpo-noaux.minus.gdpo.mbpp-plus}{+1.5}
\setval{4b.gdpo-noaux.minus.gdpo.mbpp-plus-abs}{1.5}
\setval{4b.gdpo-noaux.minus.gdpo.minerva}{+4.0}
\setval{4b.gdpo-noaux.minus.gdpo.minerva-abs}{4.0}
\setval{4b.gdpo-noaux.minus.gdpo.olympiad-bench}{+8.5}
\setval{4b.gdpo-noaux.minus.gdpo.olympiad-bench-abs}{8.5}
\setval{4b.gdpo-noaux.minus.gdpo.writing}{$-$5.3}
\setval{4b.gdpo-noaux.minus.gdpo.writing-abs}{5.3}
\setval{4b.gdpo-noaux.minus.gdpo.writingbench}{$-$4.6}
\setval{4b.gdpo-noaux.minus.gdpo.writingbench-abs}{4.6}
\setval{4b.gdpo-noaux.minus.geoma.aime24}{+11.9}
\setval{4b.gdpo-noaux.minus.geoma.aime24-abs}{11.9}
\setval{4b.gdpo-noaux.minus.geoma.aime25}{+8.1}
\setval{4b.gdpo-noaux.minus.geoma.aime25-abs}{8.1}
\setval{4b.gdpo-noaux.minus.geoma.amc23}{+10.3}
\setval{4b.gdpo-noaux.minus.geoma.amc23-abs}{10.3}
\setval{4b.gdpo-noaux.minus.geoma.avg}{+2.9}
\setval{4b.gdpo-noaux.minus.geoma.avg-abs}{2.9}
\setval{4b.gdpo-noaux.minus.geoma.code}{+2.1}
\setval{4b.gdpo-noaux.minus.geoma.code-abs}{2.1}
\setval{4b.gdpo-noaux.minus.geoma.creative-writing-v3}{$-$6.5}
\setval{4b.gdpo-noaux.minus.geoma.creative-writing-v3-abs}{6.5}
\setval{4b.gdpo-noaux.minus.geoma.guidebench}{$-$2.6}
\setval{4b.gdpo-noaux.minus.geoma.guidebench-abs}{2.6}
\setval{4b.gdpo-noaux.minus.geoma.humaneval-plus}{+5.2}
\setval{4b.gdpo-noaux.minus.geoma.humaneval-plus-abs}{5.2}
\setval{4b.gdpo-noaux.minus.geoma.if}{$-$0.4}
\setval{4b.gdpo-noaux.minus.geoma.if-abs}{0.4}
\setval{4b.gdpo-noaux.minus.geoma.ifbench}{+4.3}
\setval{4b.gdpo-noaux.minus.geoma.ifbench-abs}{4.3}
\setval{4b.gdpo-noaux.minus.geoma.ifeval}{$-$3.0}
\setval{4b.gdpo-noaux.minus.geoma.ifeval-abs}{3.0}
\setval{4b.gdpo-noaux.minus.geoma.judge2.avg}{+2.1}
\setval{4b.gdpo-noaux.minus.geoma.judge2.avg-abs}{2.1}
\setval{4b.gdpo-noaux.minus.geoma.judge2.creative-writing-v3}{$-$9.8}
\setval{4b.gdpo-noaux.minus.geoma.judge2.creative-writing-v3-abs}{9.8}
\setval{4b.gdpo-noaux.minus.geoma.judge2.longbench-write}{$-$6.9}
\setval{4b.gdpo-noaux.minus.geoma.judge2.longbench-write-abs}{6.9}
\setval{4b.gdpo-noaux.minus.geoma.judge2.writing}{$-$7.1}
\setval{4b.gdpo-noaux.minus.geoma.judge2.writing-abs}{7.1}
\setval{4b.gdpo-noaux.minus.geoma.judge2.writingbench}{$-$4.6}
\setval{4b.gdpo-noaux.minus.geoma.judge2.writingbench-abs}{4.6}
\setval{4b.gdpo-noaux.minus.geoma.livecodebench}{$-$0.9}
\setval{4b.gdpo-noaux.minus.geoma.livecodebench-abs}{0.9}
\setval{4b.gdpo-noaux.minus.geoma.longbench-write}{$-$1.5}
\setval{4b.gdpo-noaux.minus.geoma.longbench-write-abs}{1.5}
\setval{4b.gdpo-noaux.minus.geoma.math}{+8.0}
\setval{4b.gdpo-noaux.minus.geoma.math-abs}{8.0}
\setval{4b.gdpo-noaux.minus.geoma.math500}{+5.9}
\setval{4b.gdpo-noaux.minus.geoma.math500-abs}{5.9}
\setval{4b.gdpo-noaux.minus.geoma.mbpp-plus}{+2.1}
\setval{4b.gdpo-noaux.minus.geoma.mbpp-plus-abs}{2.1}
\setval{4b.gdpo-noaux.minus.geoma.minerva}{+2.6}
\setval{4b.gdpo-noaux.minus.geoma.minerva-abs}{2.6}
\setval{4b.gdpo-noaux.minus.geoma.olympiad-bench}{+9.3}
\setval{4b.gdpo-noaux.minus.geoma.olympiad-bench-abs}{9.3}
\setval{4b.gdpo-noaux.minus.geoma.writing}{$-$3.4}
\setval{4b.gdpo-noaux.minus.geoma.writing-abs}{3.4}
\setval{4b.gdpo-noaux.minus.geoma.writingbench}{$-$2.2}
\setval{4b.gdpo-noaux.minus.geoma.writingbench-abs}{2.2}
\setval{4b.gdpo-noaux.minus.grpo-nodiv.aime24}{+11.0}
\setval{4b.gdpo-noaux.minus.grpo-nodiv.aime24-abs}{11.0}
\setval{4b.gdpo-noaux.minus.grpo-nodiv.aime25}{+10.4}
\setval{4b.gdpo-noaux.minus.grpo-nodiv.aime25-abs}{10.4}
\setval{4b.gdpo-noaux.minus.grpo-nodiv.amc23}{+11.2}
\setval{4b.gdpo-noaux.minus.grpo-nodiv.amc23-abs}{11.2}
\setval{4b.gdpo-noaux.minus.grpo-nodiv.avg}{+3.6}
\setval{4b.gdpo-noaux.minus.grpo-nodiv.avg-abs}{3.6}
\setval{4b.gdpo-noaux.minus.grpo-nodiv.code}{+2.1}
\setval{4b.gdpo-noaux.minus.grpo-nodiv.code-abs}{2.1}
\setval{4b.gdpo-noaux.minus.grpo-nodiv.creative-writing-v3}{$-$3.2}
\setval{4b.gdpo-noaux.minus.grpo-nodiv.creative-writing-v3-abs}{3.2}
\setval{4b.gdpo-noaux.minus.grpo-nodiv.guidebench}{$-$2.9}
\setval{4b.gdpo-noaux.minus.grpo-nodiv.guidebench-abs}{2.9}
\setval{4b.gdpo-noaux.minus.grpo-nodiv.humaneval-plus}{+5.0}
\setval{4b.gdpo-noaux.minus.grpo-nodiv.humaneval-plus-abs}{5.0}
\setval{4b.gdpo-noaux.minus.grpo-nodiv.if}{+0.4}
\setval{4b.gdpo-noaux.minus.grpo-nodiv.if-abs}{0.4}
\setval{4b.gdpo-noaux.minus.grpo-nodiv.ifbench}{+6.7}
\setval{4b.gdpo-noaux.minus.grpo-nodiv.ifbench-abs}{6.7}
\setval{4b.gdpo-noaux.minus.grpo-nodiv.ifeval}{$-$2.5}
\setval{4b.gdpo-noaux.minus.grpo-nodiv.ifeval-abs}{2.5}
\setval{4b.gdpo-noaux.minus.grpo-nodiv.judge2.avg}{+2.6}
\setval{4b.gdpo-noaux.minus.grpo-nodiv.judge2.avg-abs}{2.6}
\setval{4b.gdpo-noaux.minus.grpo-nodiv.judge2.creative-writing-v3}{$-$7.2}
\setval{4b.gdpo-noaux.minus.grpo-nodiv.judge2.creative-writing-v3-abs}{7.2}
\setval{4b.gdpo-noaux.minus.grpo-nodiv.judge2.longbench-write}{$-$7.8}
\setval{4b.gdpo-noaux.minus.grpo-nodiv.judge2.longbench-write-abs}{7.8}
\setval{4b.gdpo-noaux.minus.grpo-nodiv.judge2.writing}{$-$6.2}
\setval{4b.gdpo-noaux.minus.grpo-nodiv.judge2.writing-abs}{6.2}
\setval{4b.gdpo-noaux.minus.grpo-nodiv.judge2.writingbench}{$-$3.5}
\setval{4b.gdpo-noaux.minus.grpo-nodiv.judge2.writingbench-abs}{3.5}
\setval{4b.gdpo-noaux.minus.grpo-nodiv.livecodebench}{+0.3}
\setval{4b.gdpo-noaux.minus.grpo-nodiv.livecodebench-abs}{0.3}
\setval{4b.gdpo-noaux.minus.grpo-nodiv.longbench-write}{$-$1.6}
\setval{4b.gdpo-noaux.minus.grpo-nodiv.longbench-write-abs}{1.6}
\setval{4b.gdpo-noaux.minus.grpo-nodiv.math}{+8.3}
\setval{4b.gdpo-noaux.minus.grpo-nodiv.math-abs}{8.3}
\setval{4b.gdpo-noaux.minus.grpo-nodiv.math500}{+5.3}
\setval{4b.gdpo-noaux.minus.grpo-nodiv.math500-abs}{5.3}
\setval{4b.gdpo-noaux.minus.grpo-nodiv.mbpp-plus}{+1.1}
\setval{4b.gdpo-noaux.minus.grpo-nodiv.mbpp-plus-abs}{1.1}
\setval{4b.gdpo-noaux.minus.grpo-nodiv.minerva}{+3.7}
\setval{4b.gdpo-noaux.minus.grpo-nodiv.minerva-abs}{3.7}
\setval{4b.gdpo-noaux.minus.grpo-nodiv.olympiad-bench}{+8.3}
\setval{4b.gdpo-noaux.minus.grpo-nodiv.olympiad-bench-abs}{8.3}
\setval{4b.gdpo-noaux.minus.grpo-nodiv.writing}{$-$1.0}
\setval{4b.gdpo-noaux.minus.grpo-nodiv.writing-abs}{1.0}
\setval{4b.gdpo-noaux.minus.grpo-nodiv.writingbench}{+1.7}
\setval{4b.gdpo-noaux.minus.grpo-nodiv.writingbench-abs}{1.7}
\setval{4b.gdpo-noaux.minus.grpo.aime24}{+11.5}
\setval{4b.gdpo-noaux.minus.grpo.aime24-abs}{11.5}
\setval{4b.gdpo-noaux.minus.grpo.aime25}{+6.1}
\setval{4b.gdpo-noaux.minus.grpo.aime25-abs}{6.1}
\setval{4b.gdpo-noaux.minus.grpo.amc23}{+11.0}
\setval{4b.gdpo-noaux.minus.grpo.amc23-abs}{11.0}
\setval{4b.gdpo-noaux.minus.grpo.avg}{+2.0}
\setval{4b.gdpo-noaux.minus.grpo.avg-abs}{2.0}
\setval{4b.gdpo-noaux.minus.grpo.code}{+1.6}
\setval{4b.gdpo-noaux.minus.grpo.code-abs}{1.6}
\setval{4b.gdpo-noaux.minus.grpo.creative-writing-v3}{$-$8.0}
\setval{4b.gdpo-noaux.minus.grpo.creative-writing-v3-abs}{8.0}
\setval{4b.gdpo-noaux.minus.grpo.guidebench}{$-$2.0}
\setval{4b.gdpo-noaux.minus.grpo.guidebench-abs}{2.0}
\setval{4b.gdpo-noaux.minus.grpo.humaneval-plus}{+2.7}
\setval{4b.gdpo-noaux.minus.grpo.humaneval-plus-abs}{2.7}
\setval{4b.gdpo-noaux.minus.grpo.if}{$-$0.3}
\setval{4b.gdpo-noaux.minus.grpo.if-abs}{0.3}
\setval{4b.gdpo-noaux.minus.grpo.ifbench}{+2.7}
\setval{4b.gdpo-noaux.minus.grpo.ifbench-abs}{2.7}
\setval{4b.gdpo-noaux.minus.grpo.ifeval}{$-$1.7}
\setval{4b.gdpo-noaux.minus.grpo.ifeval-abs}{1.7}
\setval{4b.gdpo-noaux.minus.grpo.judge2.avg}{+1.4}
\setval{4b.gdpo-noaux.minus.grpo.judge2.avg-abs}{1.4}
\setval{4b.gdpo-noaux.minus.grpo.judge2.creative-writing-v3}{$-$11.8}
\setval{4b.gdpo-noaux.minus.grpo.judge2.creative-writing-v3-abs}{11.8}
\setval{4b.gdpo-noaux.minus.grpo.judge2.longbench-write}{$-$9.9}
\setval{4b.gdpo-noaux.minus.grpo.judge2.longbench-write-abs}{9.9}
\setval{4b.gdpo-noaux.minus.grpo.judge2.writing}{$-$9.7}
\setval{4b.gdpo-noaux.minus.grpo.judge2.writing-abs}{9.7}
\setval{4b.gdpo-noaux.minus.grpo.judge2.writingbench}{$-$7.4}
\setval{4b.gdpo-noaux.minus.grpo.judge2.writingbench-abs}{7.4}
\setval{4b.gdpo-noaux.minus.grpo.livecodebench}{+0.7}
\setval{4b.gdpo-noaux.minus.grpo.livecodebench-abs}{0.7}
\setval{4b.gdpo-noaux.minus.grpo.longbench-write}{$-$4.8}
\setval{4b.gdpo-noaux.minus.grpo.longbench-write-abs}{4.8}
\setval{4b.gdpo-noaux.minus.grpo.math}{+7.6}
\setval{4b.gdpo-noaux.minus.grpo.math-abs}{7.6}
\setval{4b.gdpo-noaux.minus.grpo.math500}{+4.5}
\setval{4b.gdpo-noaux.minus.grpo.math500-abs}{4.5}
\setval{4b.gdpo-noaux.minus.grpo.mbpp-plus}{+1.2}
\setval{4b.gdpo-noaux.minus.grpo.mbpp-plus-abs}{1.2}
\setval{4b.gdpo-noaux.minus.grpo.minerva}{+4.2}
\setval{4b.gdpo-noaux.minus.grpo.minerva-abs}{4.2}
\setval{4b.gdpo-noaux.minus.grpo.olympiad-bench}{+8.5}
\setval{4b.gdpo-noaux.minus.grpo.olympiad-bench-abs}{8.5}
\setval{4b.gdpo-noaux.minus.grpo.writing}{$-$6.6}
\setval{4b.gdpo-noaux.minus.grpo.writing-abs}{6.6}
\setval{4b.gdpo-noaux.minus.grpo.writingbench}{$-$6.9}
\setval{4b.gdpo-noaux.minus.grpo.writingbench-abs}{6.9}
\setval{4b.gdpo-noaux.minus.mt-grpo.aime24}{+12.7}
\setval{4b.gdpo-noaux.minus.mt-grpo.aime24-abs}{12.7}
\setval{4b.gdpo-noaux.minus.mt-grpo.aime25}{+12.6}
\setval{4b.gdpo-noaux.minus.mt-grpo.aime25-abs}{12.6}
\setval{4b.gdpo-noaux.minus.mt-grpo.amc23}{+5.9}
\setval{4b.gdpo-noaux.minus.mt-grpo.amc23-abs}{5.9}
\setval{4b.gdpo-noaux.minus.mt-grpo.avg}{+3.6}
\setval{4b.gdpo-noaux.minus.mt-grpo.avg-abs}{3.6}
\setval{4b.gdpo-noaux.minus.mt-grpo.code}{+3.6}
\setval{4b.gdpo-noaux.minus.mt-grpo.code-abs}{3.6}
\setval{4b.gdpo-noaux.minus.mt-grpo.creative-writing-v3}{$-$5.2}
\setval{4b.gdpo-noaux.minus.mt-grpo.creative-writing-v3-abs}{5.2}
\setval{4b.gdpo-noaux.minus.mt-grpo.guidebench}{$-$3.7}
\setval{4b.gdpo-noaux.minus.mt-grpo.guidebench-abs}{3.7}
\setval{4b.gdpo-noaux.minus.mt-grpo.humaneval-plus}{+6.4}
\setval{4b.gdpo-noaux.minus.mt-grpo.humaneval-plus-abs}{6.4}
\setval{4b.gdpo-noaux.minus.mt-grpo.if}{+0.1}
\setval{4b.gdpo-noaux.minus.mt-grpo.if-abs}{0.1}
\setval{4b.gdpo-noaux.minus.mt-grpo.ifbench}{+6.0}
\setval{4b.gdpo-noaux.minus.mt-grpo.ifbench-abs}{6.0}
\setval{4b.gdpo-noaux.minus.mt-grpo.ifeval}{$-$1.9}
\setval{4b.gdpo-noaux.minus.mt-grpo.ifeval-abs}{1.9}
\setval{4b.gdpo-noaux.minus.mt-grpo.judge2.avg}{+2.4}
\setval{4b.gdpo-noaux.minus.mt-grpo.judge2.avg-abs}{2.4}
\setval{4b.gdpo-noaux.minus.mt-grpo.judge2.creative-writing-v3}{$-$10.7}
\setval{4b.gdpo-noaux.minus.mt-grpo.judge2.creative-writing-v3-abs}{10.7}
\setval{4b.gdpo-noaux.minus.mt-grpo.judge2.longbench-write}{$-$10.0}
\setval{4b.gdpo-noaux.minus.mt-grpo.judge2.longbench-write-abs}{10.0}
\setval{4b.gdpo-noaux.minus.mt-grpo.judge2.writing}{$-$9.3}
\setval{4b.gdpo-noaux.minus.mt-grpo.judge2.writing-abs}{9.3}
\setval{4b.gdpo-noaux.minus.mt-grpo.judge2.writingbench}{$-$7.2}
\setval{4b.gdpo-noaux.minus.mt-grpo.judge2.writingbench-abs}{7.2}
\setval{4b.gdpo-noaux.minus.mt-grpo.livecodebench}{+2.1}
\setval{4b.gdpo-noaux.minus.mt-grpo.livecodebench-abs}{2.1}
\setval{4b.gdpo-noaux.minus.mt-grpo.longbench-write}{$-$2.4}
\setval{4b.gdpo-noaux.minus.mt-grpo.longbench-write-abs}{2.4}
\setval{4b.gdpo-noaux.minus.mt-grpo.math}{+8.9}
\setval{4b.gdpo-noaux.minus.mt-grpo.math-abs}{8.9}
\setval{4b.gdpo-noaux.minus.mt-grpo.math500}{+7.8}
\setval{4b.gdpo-noaux.minus.mt-grpo.math500-abs}{7.8}
\setval{4b.gdpo-noaux.minus.mt-grpo.mbpp-plus}{+2.3}
\setval{4b.gdpo-noaux.minus.mt-grpo.mbpp-plus-abs}{2.3}
\setval{4b.gdpo-noaux.minus.mt-grpo.minerva}{+3.5}
\setval{4b.gdpo-noaux.minus.mt-grpo.minerva-abs}{3.5}
\setval{4b.gdpo-noaux.minus.mt-grpo.olympiad-bench}{+10.6}
\setval{4b.gdpo-noaux.minus.mt-grpo.olympiad-bench-abs}{10.6}
\setval{4b.gdpo-noaux.minus.mt-grpo.writing}{$-$3.5}
\setval{4b.gdpo-noaux.minus.mt-grpo.writing-abs}{3.5}
\setval{4b.gdpo-noaux.minus.mt-grpo.writingbench}{$-$2.8}
\setval{4b.gdpo-noaux.minus.mt-grpo.writingbench-abs}{2.8}
\setval{4b.gdpo-noaux.minus.papo.aime24}{+14.2}
\setval{4b.gdpo-noaux.minus.papo.aime24-abs}{14.2}
\setval{4b.gdpo-noaux.minus.papo.aime25}{+11.2}
\setval{4b.gdpo-noaux.minus.papo.aime25-abs}{11.2}
\setval{4b.gdpo-noaux.minus.papo.amc23}{+12.2}
\setval{4b.gdpo-noaux.minus.papo.amc23-abs}{12.2}
\setval{4b.gdpo-noaux.minus.papo.avg}{+4.1}
\setval{4b.gdpo-noaux.minus.papo.avg-abs}{4.1}
\setval{4b.gdpo-noaux.minus.papo.code}{+1.8}
\setval{4b.gdpo-noaux.minus.papo.code-abs}{1.8}
\setval{4b.gdpo-noaux.minus.papo.creative-writing-v3}{$-$7.2}
\setval{4b.gdpo-noaux.minus.papo.creative-writing-v3-abs}{7.2}
\setval{4b.gdpo-noaux.minus.papo.guidebench}{$-$4.8}
\setval{4b.gdpo-noaux.minus.papo.guidebench-abs}{4.8}
\setval{4b.gdpo-noaux.minus.papo.humaneval-plus}{+2.4}
\setval{4b.gdpo-noaux.minus.papo.humaneval-plus-abs}{2.4}
\setval{4b.gdpo-noaux.minus.papo.if}{+0.4}
\setval{4b.gdpo-noaux.minus.papo.if-abs}{0.4}
\setval{4b.gdpo-noaux.minus.papo.ifbench}{+7.0}
\setval{4b.gdpo-noaux.minus.papo.ifbench-abs}{7.0}
\setval{4b.gdpo-noaux.minus.papo.ifeval}{$-$1.1}
\setval{4b.gdpo-noaux.minus.papo.ifeval-abs}{1.1}
\setval{4b.gdpo-noaux.minus.papo.judge2.avg}{+2.9}
\setval{4b.gdpo-noaux.minus.papo.judge2.avg-abs}{2.9}
\setval{4b.gdpo-noaux.minus.papo.judge2.creative-writing-v3}{$-$10.9}
\setval{4b.gdpo-noaux.minus.papo.judge2.creative-writing-v3-abs}{10.9}
\setval{4b.gdpo-noaux.minus.papo.judge2.longbench-write}{$-$6.8}
\setval{4b.gdpo-noaux.minus.papo.judge2.longbench-write-abs}{6.8}
\setval{4b.gdpo-noaux.minus.papo.judge2.writing}{$-$7.9}
\setval{4b.gdpo-noaux.minus.papo.judge2.writing-abs}{7.9}
\setval{4b.gdpo-noaux.minus.papo.judge2.writingbench}{$-$6.1}
\setval{4b.gdpo-noaux.minus.papo.judge2.writingbench-abs}{6.1}
\setval{4b.gdpo-noaux.minus.papo.livecodebench}{+1.2}
\setval{4b.gdpo-noaux.minus.papo.livecodebench-abs}{1.2}
\setval{4b.gdpo-noaux.minus.papo.longbench-write}{+0.4}
\setval{4b.gdpo-noaux.minus.papo.longbench-write-abs}{0.4}
\setval{4b.gdpo-noaux.minus.papo.math}{+10.1}
\setval{4b.gdpo-noaux.minus.papo.math-abs}{10.1}
\setval{4b.gdpo-noaux.minus.papo.math500}{+8.2}
\setval{4b.gdpo-noaux.minus.papo.math500-abs}{8.2}
\setval{4b.gdpo-noaux.minus.papo.mbpp-plus}{+1.6}
\setval{4b.gdpo-noaux.minus.papo.mbpp-plus-abs}{1.6}
\setval{4b.gdpo-noaux.minus.papo.minerva}{+3.0}
\setval{4b.gdpo-noaux.minus.papo.minerva-abs}{3.0}
\setval{4b.gdpo-noaux.minus.papo.olympiad-bench}{+11.7}
\setval{4b.gdpo-noaux.minus.papo.olympiad-bench-abs}{11.7}
\setval{4b.gdpo-noaux.minus.papo.writing}{$-$2.0}
\setval{4b.gdpo-noaux.minus.papo.writing-abs}{2.0}
\setval{4b.gdpo-noaux.minus.papo.writingbench}{+0.8}
\setval{4b.gdpo-noaux.minus.papo.writingbench-abs}{0.8}
\setval{4b.gdpo-noaux.minus.saw.aime24}{+10.9}
\setval{4b.gdpo-noaux.minus.saw.aime24-abs}{10.9}
\setval{4b.gdpo-noaux.minus.saw.aime25}{+7.6}
\setval{4b.gdpo-noaux.minus.saw.aime25-abs}{7.6}
\setval{4b.gdpo-noaux.minus.saw.amc23}{+9.1}
\setval{4b.gdpo-noaux.minus.saw.amc23-abs}{9.1}
\setval{4b.gdpo-noaux.minus.saw.avg}{+2.5}
\setval{4b.gdpo-noaux.minus.saw.avg-abs}{2.5}
\setval{4b.gdpo-noaux.minus.saw.code}{+3.0}
\setval{4b.gdpo-noaux.minus.saw.code-abs}{3.0}
\setval{4b.gdpo-noaux.minus.saw.creative-writing-v3}{$-$5.7}
\setval{4b.gdpo-noaux.minus.saw.creative-writing-v3-abs}{5.7}
\setval{4b.gdpo-noaux.minus.saw.guidebench}{$-$5.1}
\setval{4b.gdpo-noaux.minus.saw.guidebench-abs}{5.1}
\setval{4b.gdpo-noaux.minus.saw.humaneval-plus}{+4.8}
\setval{4b.gdpo-noaux.minus.saw.humaneval-plus-abs}{4.8}
\setval{4b.gdpo-noaux.minus.saw.if}{$-$2.4}
\setval{4b.gdpo-noaux.minus.saw.if-abs}{2.4}
\setval{4b.gdpo-noaux.minus.saw.ifbench}{+0.0}
\setval{4b.gdpo-noaux.minus.saw.ifbench-abs}{0.0}
\setval{4b.gdpo-noaux.minus.saw.ifeval}{$-$2.2}
\setval{4b.gdpo-noaux.minus.saw.ifeval-abs}{2.2}
\setval{4b.gdpo-noaux.minus.saw.judge2.avg}{+1.2}
\setval{4b.gdpo-noaux.minus.saw.judge2.avg-abs}{1.2}
\setval{4b.gdpo-noaux.minus.saw.judge2.creative-writing-v3}{$-$12.2}
\setval{4b.gdpo-noaux.minus.saw.judge2.creative-writing-v3-abs}{12.2}
\setval{4b.gdpo-noaux.minus.saw.judge2.longbench-write}{$-$9.3}
\setval{4b.gdpo-noaux.minus.saw.judge2.longbench-write-abs}{9.3}
\setval{4b.gdpo-noaux.minus.saw.judge2.writing}{$-$9.6}
\setval{4b.gdpo-noaux.minus.saw.judge2.writing-abs}{9.6}
\setval{4b.gdpo-noaux.minus.saw.judge2.writingbench}{$-$7.3}
\setval{4b.gdpo-noaux.minus.saw.judge2.writingbench-abs}{7.3}
\setval{4b.gdpo-noaux.minus.saw.livecodebench}{+1.9}
\setval{4b.gdpo-noaux.minus.saw.livecodebench-abs}{1.9}
\setval{4b.gdpo-noaux.minus.saw.longbench-write}{$-$1.8}
\setval{4b.gdpo-noaux.minus.saw.longbench-write-abs}{1.8}
\setval{4b.gdpo-noaux.minus.saw.math}{+7.5}
\setval{4b.gdpo-noaux.minus.saw.math-abs}{7.5}
\setval{4b.gdpo-noaux.minus.saw.math500}{+6.9}
\setval{4b.gdpo-noaux.minus.saw.math500-abs}{6.9}
\setval{4b.gdpo-noaux.minus.saw.mbpp-plus}{+2.2}
\setval{4b.gdpo-noaux.minus.saw.mbpp-plus-abs}{2.2}
\setval{4b.gdpo-noaux.minus.saw.minerva}{+1.4}
\setval{4b.gdpo-noaux.minus.saw.minerva-abs}{1.4}
\setval{4b.gdpo-noaux.minus.saw.olympiad-bench}{+9.3}
\setval{4b.gdpo-noaux.minus.saw.olympiad-bench-abs}{9.3}
\setval{4b.gdpo-noaux.minus.saw.writing}{$-$3.4}
\setval{4b.gdpo-noaux.minus.saw.writing-abs}{3.4}
\setval{4b.gdpo-noaux.minus.saw.writingbench}{$-$2.5}
\setval{4b.gdpo-noaux.minus.saw.writingbench-abs}{2.5}
\setval{4b.gdpo-noaux.passing-at-ceiling.code.early}{0.71}
\setval{4b.gdpo-noaux.passing-at-ceiling.code.late}{0.74}
\setval{4b.gdpo-noaux.passing-at-ceiling.code.run}{0.73}
\setval{4b.gdpo-noaux.passing-at-ceiling.if.early}{0.28}
\setval{4b.gdpo-noaux.passing-at-ceiling.if.late}{0.52}
\setval{4b.gdpo-noaux.passing-at-ceiling.if.run}{0.44}
\setval{4b.gdpo-noaux.passing-at-ceiling.math.early}{1.00}
\setval{4b.gdpo-noaux.passing-at-ceiling.math.late}{1.00}
\setval{4b.gdpo-noaux.passing-at-ceiling.math.run}{1.00}
\setval{4b.gdpo-noaux.passing-at-ceiling.writing.early}{0.00}
\setval{4b.gdpo-noaux.passing-at-ceiling.writing.late}{0.03}
\setval{4b.gdpo-noaux.passing-at-ceiling.writing.run}{0.01}
\setval{4b.gdpo-noaux.primary-cosine.code.early}{1.00}
\setval{4b.gdpo-noaux.primary-cosine.code.late}{1.00}
\setval{4b.gdpo-noaux.primary-cosine.code.run}{1.00}
\setval{4b.gdpo-noaux.primary-cosine.if.early}{0.75}
\setval{4b.gdpo-noaux.primary-cosine.if.late}{0.83}
\setval{4b.gdpo-noaux.primary-cosine.if.run}{0.82}
\setval{4b.gdpo-noaux.primary-cosine.math.early}{1.00}
\setval{4b.gdpo-noaux.primary-cosine.math.late}{1.00}
\setval{4b.gdpo-noaux.primary-cosine.math.run}{1.00}
\setval{4b.gdpo-noaux.primary-cosine.writing.early}{1.00}
\setval{4b.gdpo-noaux.primary-cosine.writing.late}{1.00}
\setval{4b.gdpo-noaux.primary-cosine.writing.run}{1.00}
\setval{4b.gdpo-noaux.primary-level.code.early}{0.53}
\setval{4b.gdpo-noaux.primary-level.code.late}{0.75}
\setval{4b.gdpo-noaux.primary-level.code.run}{0.70}
\setval{4b.gdpo-noaux.primary-level.if.early}{0.31}
\setval{4b.gdpo-noaux.primary-level.if.late}{0.71}
\setval{4b.gdpo-noaux.primary-level.if.run}{0.59}
\setval{4b.gdpo-noaux.primary-level.math.early}{0.47}
\setval{4b.gdpo-noaux.primary-level.math.late}{0.71}
\setval{4b.gdpo-noaux.primary-level.math.run}{0.65}
\setval{4b.gdpo-noaux.primary-level.writing.early}{0.53}
\setval{4b.gdpo-noaux.primary-level.writing.late}{0.82}
\setval{4b.gdpo-noaux.primary-level.writing.run}{0.76}
\setval{4b.gdpo-noaux.solved.code.early}{0.08}
\setval{4b.gdpo-noaux.solved.code.late}{0.35}
\setval{4b.gdpo-noaux.solved.code.run}{0.28}
\setval{4b.gdpo-noaux.solved.if.early}{0.03}
\setval{4b.gdpo-noaux.solved.if.late}{0.21}
\setval{4b.gdpo-noaux.solved.if.run}{0.14}
\setval{4b.gdpo-noaux.solved.math.early}{0.13}
\setval{4b.gdpo-noaux.solved.math.late}{0.49}
\setval{4b.gdpo-noaux.solved.math.run}{0.37}
\setval{4b.gdpo-noaux.solved.writing.early}{0.00}
\setval{4b.gdpo-noaux.solved.writing.late}{0.00}
\setval{4b.gdpo-noaux.solved.writing.run}{0.00}
\setval{4b.gdpo-noaux.update-share.code.early}{0.22}
\setval{4b.gdpo-noaux.update-share.code.late}{0.21}
\setval{4b.gdpo-noaux.update-share.code.run}{0.21}
\setval{4b.gdpo-noaux.update-share.if.early}{0.34}
\setval{4b.gdpo-noaux.update-share.if.late}{0.24}
\setval{4b.gdpo-noaux.update-share.if.run}{0.28}
\setval{4b.gdpo-noaux.update-share.math.early}{0.18}
\setval{4b.gdpo-noaux.update-share.math.late}{0.16}
\setval{4b.gdpo-noaux.update-share.math.run}{0.17}
\setval{4b.gdpo-noaux.update-share.writing.early}{0.27}
\setval{4b.gdpo-noaux.update-share.writing.late}{0.39}
\setval{4b.gdpo-noaux.update-share.writing.run}{0.35}
\setval{4b.gdpo-w0.5.aux-change.code}{+0.20}
\setval{4b.gdpo-w0.5.aux-change.if}{+0.23}
\setval{4b.gdpo-w0.5.aux-change.math}{+0.10}
\setval{4b.gdpo-w0.5.aux-change.writing}{+0.19}
\setval{4b.gdpo-w0.5.aux-change.writing-rm}{+0.36}
\setval{4b.gdpo-w0.5.aux-full.code.early}{0.72}
\setval{4b.gdpo-w0.5.aux-full.code.late}{0.95}
\setval{4b.gdpo-w0.5.aux-full.code.run}{0.92}
\setval{4b.gdpo-w0.5.aux-full.if.early}{0.64}
\setval{4b.gdpo-w0.5.aux-full.if.late}{0.90}
\setval{4b.gdpo-w0.5.aux-full.if.run}{0.84}
\setval{4b.gdpo-w0.5.aux-full.math.early}{0.62}
\setval{4b.gdpo-w0.5.aux-full.math.late}{0.80}
\setval{4b.gdpo-w0.5.aux-full.math.run}{0.76}
\setval{4b.gdpo-w0.5.aux-full.writing-rm.early}{0.10}
\setval{4b.gdpo-w0.5.aux-full.writing-rm.late}{0.57}
\setval{4b.gdpo-w0.5.aux-full.writing-rm.run}{0.47}
\setval{4b.gdpo-w0.5.aux-full.writing.early}{0.60}
\setval{4b.gdpo-w0.5.aux-full.writing.late}{0.69}
\setval{4b.gdpo-w0.5.aux-full.writing.run}{0.61}
\setval{4b.gdpo-w0.5.aux-level.code.early}{0.92}
\setval{4b.gdpo-w0.5.aux-level.code.late}{0.99}
\setval{4b.gdpo-w0.5.aux-level.code.run}{0.98}
\setval{4b.gdpo-w0.5.aux-level.if.early}{0.83}
\setval{4b.gdpo-w0.5.aux-level.if.late}{0.94}
\setval{4b.gdpo-w0.5.aux-level.if.run}{0.91}
\setval{4b.gdpo-w0.5.aux-level.math.early}{0.92}
\setval{4b.gdpo-w0.5.aux-level.math.late}{0.95}
\setval{4b.gdpo-w0.5.aux-level.math.run}{0.94}
\setval{4b.gdpo-w0.5.aux-level.writing-rm.early}{0.78}
\setval{4b.gdpo-w0.5.aux-level.writing-rm.late}{0.96}
\setval{4b.gdpo-w0.5.aux-level.writing-rm.run}{0.93}
\setval{4b.gdpo-w0.5.aux-level.writing.early}{0.89}
\setval{4b.gdpo-w0.5.aux-level.writing.late}{0.95}
\setval{4b.gdpo-w0.5.aux-level.writing.run}{0.92}
\setval{4b.gdpo-w0.5.bench.aime24}{14.6}
\setval{4b.gdpo-w0.5.bench.aime25}{10.3}
\setval{4b.gdpo-w0.5.bench.amc23}{58.4}
\setval{4b.gdpo-w0.5.bench.creative-writing-v3}{38.1}
\setval{4b.gdpo-w0.5.bench.guidebench}{79.6}
\setval{4b.gdpo-w0.5.bench.humaneval-plus}{82.0}
\setval{4b.gdpo-w0.5.bench.ifbench}{39.3}
\setval{4b.gdpo-w0.5.bench.ifeval}{86.8}
\setval{4b.gdpo-w0.5.bench.livecodebench}{24.2}
\setval{4b.gdpo-w0.5.bench.longbench-write}{67.4}
\setval{4b.gdpo-w0.5.bench.math500}{81.8}
\setval{4b.gdpo-w0.5.bench.mbpp-plus}{71.2}
\setval{4b.gdpo-w0.5.bench.minerva}{46.1}
\setval{4b.gdpo-w0.5.bench.olympiad-bench}{48.5}
\setval{4b.gdpo-w0.5.bench.writingbench}{70.7}
\setval{4b.gdpo-w0.5.eval.avg}{54.6}
\setval{4b.gdpo-w0.5.eval.code}{59.1}
\setval{4b.gdpo-w0.5.eval.if}{68.6}
\setval{4b.gdpo-w0.5.eval.math}{43.3}
\setval{4b.gdpo-w0.5.eval.writing}{58.8}
\setval{4b.gdpo-w0.5.gate.code.against.early}{0.03}
\setval{4b.gdpo-w0.5.gate.code.against.late}{0.02}
\setval{4b.gdpo-w0.5.gate.code.against.run}{0.02}
\setval{4b.gdpo-w0.5.gate.code.agreement}{+0.011}
\setval{4b.gdpo-w0.5.gate.code.ceiling.early}{0.11}
\setval{4b.gdpo-w0.5.gate.code.ceiling.late}{0.04}
\setval{4b.gdpo-w0.5.gate.code.ceiling.run}{0.06}
\setval{4b.gdpo-w0.5.gate.code.cosine.early}{0.84}
\setval{4b.gdpo-w0.5.gate.code.cosine.late}{0.78}
\setval{4b.gdpo-w0.5.gate.code.cosine.run}{0.81}
\setval{4b.gdpo-w0.5.gate.if.against.early}{0.04}
\setval{4b.gdpo-w0.5.gate.if.against.late}{0.04}
\setval{4b.gdpo-w0.5.gate.if.against.run}{0.05}
\setval{4b.gdpo-w0.5.gate.if.agreement}{+0.019}
\setval{4b.gdpo-w0.5.gate.if.ceiling.early}{0.06}
\setval{4b.gdpo-w0.5.gate.if.ceiling.late}{0.19}
\setval{4b.gdpo-w0.5.gate.if.ceiling.run}{0.13}
\setval{4b.gdpo-w0.5.gate.if.cosine.early}{0.73}
\setval{4b.gdpo-w0.5.gate.if.cosine.late}{0.51}
\setval{4b.gdpo-w0.5.gate.if.cosine.run}{0.58}
\setval{4b.gdpo-w0.5.gate.math.against.early}{0.00}
\setval{4b.gdpo-w0.5.gate.math.against.late}{0.00}
\setval{4b.gdpo-w0.5.gate.math.against.run}{0.00}
\setval{4b.gdpo-w0.5.gate.math.ceiling.early}{0.20}
\setval{4b.gdpo-w0.5.gate.math.ceiling.late}{0.30}
\setval{4b.gdpo-w0.5.gate.math.ceiling.run}{0.28}
\setval{4b.gdpo-w0.5.gate.math.cosine.early}{0.88}
\setval{4b.gdpo-w0.5.gate.math.cosine.late}{0.81}
\setval{4b.gdpo-w0.5.gate.math.cosine.run}{0.83}
\setval{4b.gdpo-w0.5.gate.writing-rm.against.early}{0.21}
\setval{4b.gdpo-w0.5.gate.writing-rm.against.late}{0.54}
\setval{4b.gdpo-w0.5.gate.writing-rm.against.run}{0.49}
\setval{4b.gdpo-w0.5.gate.writing-rm.agreement}{+0.060}
\setval{4b.gdpo-w0.5.gate.writing-rm.ceiling.early}{0.00}
\setval{4b.gdpo-w0.5.gate.writing-rm.ceiling.late}{0.00}
\setval{4b.gdpo-w0.5.gate.writing-rm.ceiling.run}{0.00}
\setval{4b.gdpo-w0.5.gate.writing-rm.cosine.early}{0.68}
\setval{4b.gdpo-w0.5.gate.writing-rm.cosine.late}{0.14}
\setval{4b.gdpo-w0.5.gate.writing-rm.cosine.run}{0.22}
\setval{4b.gdpo-w0.5.gate.writing.against.early}{0.31}
\setval{4b.gdpo-w0.5.gate.writing.against.late}{0.61}
\setval{4b.gdpo-w0.5.gate.writing.against.run}{0.59}
\setval{4b.gdpo-w0.5.gate.writing.agreement}{$-$0.046}
\setval{4b.gdpo-w0.5.gate.writing.ceiling.early}{0.00}
\setval{4b.gdpo-w0.5.gate.writing.ceiling.late}{0.00}
\setval{4b.gdpo-w0.5.gate.writing.ceiling.run}{0.00}
\setval{4b.gdpo-w0.5.gate.writing.cosine.early}{0.50}
\setval{4b.gdpo-w0.5.gate.writing.cosine.late}{0.03}
\setval{4b.gdpo-w0.5.gate.writing.cosine.run}{0.07}
\setval{4b.gdpo-w0.5.judge2.avg}{52.5}
\setval{4b.gdpo-w0.5.judge2.creative-writing-v3}{30.1}
\setval{4b.gdpo-w0.5.judge2.longbench-write}{60.5}
\setval{4b.gdpo-w0.5.judge2.writing}{48.1}
\setval{4b.gdpo-w0.5.judge2.writingbench}{53.6}
\setval{4b.gdpo-w0.5.last-step}{300}
\setval{4b.gdpo-w0.5.minus.base.aime24}{+6.8}
\setval{4b.gdpo-w0.5.minus.base.aime24-abs}{6.8}
\setval{4b.gdpo-w0.5.minus.base.aime25}{+3.2}
\setval{4b.gdpo-w0.5.minus.base.aime25-abs}{3.2}
\setval{4b.gdpo-w0.5.minus.base.amc23}{+14.5}
\setval{4b.gdpo-w0.5.minus.base.amc23-abs}{14.5}
\setval{4b.gdpo-w0.5.minus.base.avg}{+20.1}
\setval{4b.gdpo-w0.5.minus.base.avg-abs}{20.1}
\setval{4b.gdpo-w0.5.minus.base.code}{+14.5}
\setval{4b.gdpo-w0.5.minus.base.code-abs}{14.5}
\setval{4b.gdpo-w0.5.minus.base.creative-writing-v3}{+13.3}
\setval{4b.gdpo-w0.5.minus.base.creative-writing-v3-abs}{13.3}
\setval{4b.gdpo-w0.5.minus.base.guidebench}{+54.7}
\setval{4b.gdpo-w0.5.minus.base.guidebench-abs}{54.7}
\setval{4b.gdpo-w0.5.minus.base.humaneval-plus}{+17.0}
\setval{4b.gdpo-w0.5.minus.base.humaneval-plus-abs}{17.0}
\setval{4b.gdpo-w0.5.minus.base.if}{+38.8}
\setval{4b.gdpo-w0.5.minus.base.if-abs}{38.8}
\setval{4b.gdpo-w0.5.minus.base.ifbench}{+22.3}
\setval{4b.gdpo-w0.5.minus.base.ifbench-abs}{22.3}
\setval{4b.gdpo-w0.5.minus.base.ifeval}{+39.4}
\setval{4b.gdpo-w0.5.minus.base.ifeval-abs}{39.4}
\setval{4b.gdpo-w0.5.minus.base.judge2.avg}{+18.6}
\setval{4b.gdpo-w0.5.minus.base.judge2.avg-abs}{18.6}
\setval{4b.gdpo-w0.5.minus.base.judge2.creative-writing-v3}{+9.2}
\setval{4b.gdpo-w0.5.minus.base.judge2.creative-writing-v3-abs}{9.2}
\setval{4b.gdpo-w0.5.minus.base.judge2.longbench-write}{+19.6}
\setval{4b.gdpo-w0.5.minus.base.judge2.longbench-write-abs}{19.6}
\setval{4b.gdpo-w0.5.minus.base.judge2.writing}{+17.3}
\setval{4b.gdpo-w0.5.minus.base.judge2.writing-abs}{17.3}
\setval{4b.gdpo-w0.5.minus.base.judge2.writingbench}{+23.1}
\setval{4b.gdpo-w0.5.minus.base.judge2.writingbench-abs}{23.1}
\setval{4b.gdpo-w0.5.minus.base.livecodebench}{+11.0}
\setval{4b.gdpo-w0.5.minus.base.livecodebench-abs}{11.0}
\setval{4b.gdpo-w0.5.minus.base.longbench-write}{+23.0}
\setval{4b.gdpo-w0.5.minus.base.longbench-write-abs}{23.0}
\setval{4b.gdpo-w0.5.minus.base.math}{+11.1}
\setval{4b.gdpo-w0.5.minus.base.math-abs}{11.1}
\setval{4b.gdpo-w0.5.minus.base.math500}{+14.5}
\setval{4b.gdpo-w0.5.minus.base.math500-abs}{14.5}
\setval{4b.gdpo-w0.5.minus.base.mbpp-plus}{+15.6}
\setval{4b.gdpo-w0.5.minus.base.mbpp-plus-abs}{15.6}
\setval{4b.gdpo-w0.5.minus.base.minerva}{+14.5}
\setval{4b.gdpo-w0.5.minus.base.minerva-abs}{14.5}
\setval{4b.gdpo-w0.5.minus.base.olympiad-bench}{+12.8}
\setval{4b.gdpo-w0.5.minus.base.olympiad-bench-abs}{12.8}
\setval{4b.gdpo-w0.5.minus.base.writing}{+24.9}
\setval{4b.gdpo-w0.5.minus.base.writing-abs}{24.9}
\setval{4b.gdpo-w0.5.minus.base.writingbench}{+38.3}
\setval{4b.gdpo-w0.5.minus.base.writingbench-abs}{38.3}
\setval{4b.gdpo-w0.5.minus.concur-free.aime24}{+2.7}
\setval{4b.gdpo-w0.5.minus.concur-free.aime24-abs}{2.7}
\setval{4b.gdpo-w0.5.minus.concur-free.aime25}{$-$4.3}
\setval{4b.gdpo-w0.5.minus.concur-free.aime25-abs}{4.3}
\setval{4b.gdpo-w0.5.minus.concur-free.amc23}{+4.2}
\setval{4b.gdpo-w0.5.minus.concur-free.amc23-abs}{4.2}
\setval{4b.gdpo-w0.5.minus.concur-free.avg}{$-$0.9}
\setval{4b.gdpo-w0.5.minus.concur-free.avg-abs}{0.9}
\setval{4b.gdpo-w0.5.minus.concur-free.code}{+0.8}
\setval{4b.gdpo-w0.5.minus.concur-free.code-abs}{0.8}
\setval{4b.gdpo-w0.5.minus.concur-free.creative-writing-v3}{$-$7.3}
\setval{4b.gdpo-w0.5.minus.concur-free.creative-writing-v3-abs}{7.3}
\setval{4b.gdpo-w0.5.minus.concur-free.guidebench}{+0.6}
\setval{4b.gdpo-w0.5.minus.concur-free.guidebench-abs}{0.6}
\setval{4b.gdpo-w0.5.minus.concur-free.humaneval-plus}{+0.9}
\setval{4b.gdpo-w0.5.minus.concur-free.humaneval-plus-abs}{0.9}
\setval{4b.gdpo-w0.5.minus.concur-free.if}{+0.3}
\setval{4b.gdpo-w0.5.minus.concur-free.if-abs}{0.3}
\setval{4b.gdpo-w0.5.minus.concur-free.ifbench}{+0.3}
\setval{4b.gdpo-w0.5.minus.concur-free.ifbench-abs}{0.3}
\setval{4b.gdpo-w0.5.minus.concur-free.ifeval}{+0.1}
\setval{4b.gdpo-w0.5.minus.concur-free.ifeval-abs}{0.1}
\setval{4b.gdpo-w0.5.minus.concur-free.judge2.avg}{$-$0.3}
\setval{4b.gdpo-w0.5.minus.concur-free.judge2.avg-abs}{0.3}
\setval{4b.gdpo-w0.5.minus.concur-free.judge2.creative-writing-v3}{$-$5.2}
\setval{4b.gdpo-w0.5.minus.concur-free.judge2.creative-writing-v3-abs}{5.2}
\setval{4b.gdpo-w0.5.minus.concur-free.judge2.longbench-write}{$-$8.8}
\setval{4b.gdpo-w0.5.minus.concur-free.judge2.longbench-write-abs}{8.8}
\setval{4b.gdpo-w0.5.minus.concur-free.judge2.writing}{$-$5.0}
\setval{4b.gdpo-w0.5.minus.concur-free.judge2.writing-abs}{5.0}
\setval{4b.gdpo-w0.5.minus.concur-free.judge2.writingbench}{$-$0.8}
\setval{4b.gdpo-w0.5.minus.concur-free.judge2.writingbench-abs}{0.8}
\setval{4b.gdpo-w0.5.minus.concur-free.livecodebench}{+0.0}
\setval{4b.gdpo-w0.5.minus.concur-free.livecodebench-abs}{0.0}
\setval{4b.gdpo-w0.5.minus.concur-free.longbench-write}{$-$10.4}
\setval{4b.gdpo-w0.5.minus.concur-free.longbench-write-abs}{10.4}
\setval{4b.gdpo-w0.5.minus.concur-free.math}{+1.1}
\setval{4b.gdpo-w0.5.minus.concur-free.math-abs}{1.1}
\setval{4b.gdpo-w0.5.minus.concur-free.math500}{+1.5}
\setval{4b.gdpo-w0.5.minus.concur-free.math500-abs}{1.5}
\setval{4b.gdpo-w0.5.minus.concur-free.mbpp-plus}{+1.5}
\setval{4b.gdpo-w0.5.minus.concur-free.mbpp-plus-abs}{1.5}
\setval{4b.gdpo-w0.5.minus.concur-free.minerva}{$-$0.1}
\setval{4b.gdpo-w0.5.minus.concur-free.minerva-abs}{0.1}
\setval{4b.gdpo-w0.5.minus.concur-free.olympiad-bench}{+2.7}
\setval{4b.gdpo-w0.5.minus.concur-free.olympiad-bench-abs}{2.7}
\setval{4b.gdpo-w0.5.minus.concur-free.writing}{$-$8.1}
\setval{4b.gdpo-w0.5.minus.concur-free.writing-abs}{8.1}
\setval{4b.gdpo-w0.5.minus.concur-free.writingbench}{$-$6.6}
\setval{4b.gdpo-w0.5.minus.concur-free.writingbench-abs}{6.6}
\setval{4b.gdpo-w0.5.minus.concur-tied.aime24}{$-$4.9}
\setval{4b.gdpo-w0.5.minus.concur-tied.aime24-abs}{4.9}
\setval{4b.gdpo-w0.5.minus.concur-tied.aime25}{$-$6.0}
\setval{4b.gdpo-w0.5.minus.concur-tied.aime25-abs}{6.0}
\setval{4b.gdpo-w0.5.minus.concur-tied.amc23}{$-$10.9}
\setval{4b.gdpo-w0.5.minus.concur-tied.amc23-abs}{10.9}
\setval{4b.gdpo-w0.5.minus.concur-tied.avg}{$-$4.1}
\setval{4b.gdpo-w0.5.minus.concur-tied.avg-abs}{4.1}
\setval{4b.gdpo-w0.5.minus.concur-tied.code}{+0.0}
\setval{4b.gdpo-w0.5.minus.concur-tied.code-abs}{0.0}
\setval{4b.gdpo-w0.5.minus.concur-tied.creative-writing-v3}{$-$9.7}
\setval{4b.gdpo-w0.5.minus.concur-tied.creative-writing-v3-abs}{9.7}
\setval{4b.gdpo-w0.5.minus.concur-tied.guidebench}{+0.4}
\setval{4b.gdpo-w0.5.minus.concur-tied.guidebench-abs}{0.4}
\setval{4b.gdpo-w0.5.minus.concur-tied.humaneval-plus}{$-$2.0}
\setval{4b.gdpo-w0.5.minus.concur-tied.humaneval-plus-abs}{2.0}
\setval{4b.gdpo-w0.5.minus.concur-tied.if}{+0.4}
\setval{4b.gdpo-w0.5.minus.concur-tied.if-abs}{0.4}
\setval{4b.gdpo-w0.5.minus.concur-tied.ifbench}{+1.7}
\setval{4b.gdpo-w0.5.minus.concur-tied.ifbench-abs}{1.7}
\setval{4b.gdpo-w0.5.minus.concur-tied.ifeval}{$-$1.0}
\setval{4b.gdpo-w0.5.minus.concur-tied.ifeval-abs}{1.0}
\setval{4b.gdpo-w0.5.minus.concur-tied.judge2.avg}{$-$3.0}
\setval{4b.gdpo-w0.5.minus.concur-tied.judge2.avg-abs}{3.0}
\setval{4b.gdpo-w0.5.minus.concur-tied.judge2.creative-writing-v3}{$-$3.9}
\setval{4b.gdpo-w0.5.minus.concur-tied.judge2.creative-writing-v3-abs}{3.9}
\setval{4b.gdpo-w0.5.minus.concur-tied.judge2.longbench-write}{$-$7.4}
\setval{4b.gdpo-w0.5.minus.concur-tied.judge2.longbench-write-abs}{7.4}
\setval{4b.gdpo-w0.5.minus.concur-tied.judge2.writing}{$-$4.4}
\setval{4b.gdpo-w0.5.minus.concur-tied.judge2.writing-abs}{4.4}
\setval{4b.gdpo-w0.5.minus.concur-tied.judge2.writingbench}{$-$1.7}
\setval{4b.gdpo-w0.5.minus.concur-tied.judge2.writingbench-abs}{1.7}
\setval{4b.gdpo-w0.5.minus.concur-tied.livecodebench}{$-$0.6}
\setval{4b.gdpo-w0.5.minus.concur-tied.livecodebench-abs}{0.6}
\setval{4b.gdpo-w0.5.minus.concur-tied.longbench-write}{$-$10.3}
\setval{4b.gdpo-w0.5.minus.concur-tied.longbench-write-abs}{10.3}
\setval{4b.gdpo-w0.5.minus.concur-tied.math}{$-$5.5}
\setval{4b.gdpo-w0.5.minus.concur-tied.math-abs}{5.5}
\setval{4b.gdpo-w0.5.minus.concur-tied.math500}{$-$3.7}
\setval{4b.gdpo-w0.5.minus.concur-tied.math500-abs}{3.7}
\setval{4b.gdpo-w0.5.minus.concur-tied.mbpp-plus}{+2.6}
\setval{4b.gdpo-w0.5.minus.concur-tied.mbpp-plus-abs}{2.6}
\setval{4b.gdpo-w0.5.minus.concur-tied.minerva}{$-$3.0}
\setval{4b.gdpo-w0.5.minus.concur-tied.minerva-abs}{3.0}
\setval{4b.gdpo-w0.5.minus.concur-tied.olympiad-bench}{$-$4.4}
\setval{4b.gdpo-w0.5.minus.concur-tied.olympiad-bench-abs}{4.4}
\setval{4b.gdpo-w0.5.minus.concur-tied.writing}{$-$9.8}
\setval{4b.gdpo-w0.5.minus.concur-tied.writing-abs}{9.8}
\setval{4b.gdpo-w0.5.minus.concur-tied.writingbench}{$-$9.4}
\setval{4b.gdpo-w0.5.minus.concur-tied.writingbench-abs}{9.4}
\setval{4b.gdpo-w0.5.minus.concur.aime24}{$-$5.9}
\setval{4b.gdpo-w0.5.minus.concur.aime24-abs}{5.9}
\setval{4b.gdpo-w0.5.minus.concur.aime25}{$-$8.4}
\setval{4b.gdpo-w0.5.minus.concur.aime25-abs}{8.4}
\setval{4b.gdpo-w0.5.minus.concur.amc23}{$-$11.0}
\setval{4b.gdpo-w0.5.minus.concur.amc23-abs}{11.0}
\setval{4b.gdpo-w0.5.minus.concur.avg}{$-$4.2}
\setval{4b.gdpo-w0.5.minus.concur.avg-abs}{4.2}
\setval{4b.gdpo-w0.5.minus.concur.code}{$-$0.6}
\setval{4b.gdpo-w0.5.minus.concur.code-abs}{0.6}
\setval{4b.gdpo-w0.5.minus.concur.creative-writing-v3}{$-$6.6}
\setval{4b.gdpo-w0.5.minus.concur.creative-writing-v3-abs}{6.6}
\setval{4b.gdpo-w0.5.minus.concur.guidebench}{+2.2}
\setval{4b.gdpo-w0.5.minus.concur.guidebench-abs}{2.2}
\setval{4b.gdpo-w0.5.minus.concur.humaneval-plus}{$-$0.4}
\setval{4b.gdpo-w0.5.minus.concur.humaneval-plus-abs}{0.4}
\setval{4b.gdpo-w0.5.minus.concur.if}{+0.0}
\setval{4b.gdpo-w0.5.minus.concur.if-abs}{0.0}
\setval{4b.gdpo-w0.5.minus.concur.ifbench}{$-$1.0}
\setval{4b.gdpo-w0.5.minus.concur.ifbench-abs}{1.0}
\setval{4b.gdpo-w0.5.minus.concur.ifeval}{$-$1.3}
\setval{4b.gdpo-w0.5.minus.concur.ifeval-abs}{1.3}
\setval{4b.gdpo-w0.5.minus.concur.judge2.avg}{$-$3.3}
\setval{4b.gdpo-w0.5.minus.concur.judge2.avg-abs}{3.3}
\setval{4b.gdpo-w0.5.minus.concur.judge2.creative-writing-v3}{$-$2.3}
\setval{4b.gdpo-w0.5.minus.concur.judge2.creative-writing-v3-abs}{2.3}
\setval{4b.gdpo-w0.5.minus.concur.judge2.longbench-write}{$-$6.1}
\setval{4b.gdpo-w0.5.minus.concur.judge2.longbench-write-abs}{6.1}
\setval{4b.gdpo-w0.5.minus.concur.judge2.writing}{$-$3.6}
\setval{4b.gdpo-w0.5.minus.concur.judge2.writing-abs}{3.6}
\setval{4b.gdpo-w0.5.minus.concur.judge2.writingbench}{$-$2.4}
\setval{4b.gdpo-w0.5.minus.concur.judge2.writingbench-abs}{2.4}
\setval{4b.gdpo-w0.5.minus.concur.livecodebench}{$-$1.0}
\setval{4b.gdpo-w0.5.minus.concur.livecodebench-abs}{1.0}
\setval{4b.gdpo-w0.5.minus.concur.longbench-write}{$-$8.9}
\setval{4b.gdpo-w0.5.minus.concur.longbench-write-abs}{8.9}
\setval{4b.gdpo-w0.5.minus.concur.math}{$-$6.2}
\setval{4b.gdpo-w0.5.minus.concur.math-abs}{6.2}
\setval{4b.gdpo-w0.5.minus.concur.math500}{$-$3.5}
\setval{4b.gdpo-w0.5.minus.concur.math500-abs}{3.5}
\setval{4b.gdpo-w0.5.minus.concur.mbpp-plus}{$-$0.3}
\setval{4b.gdpo-w0.5.minus.concur.mbpp-plus-abs}{0.3}
\setval{4b.gdpo-w0.5.minus.concur.minerva}{$-$3.1}
\setval{4b.gdpo-w0.5.minus.concur.minerva-abs}{3.1}
\setval{4b.gdpo-w0.5.minus.concur.olympiad-bench}{$-$5.1}
\setval{4b.gdpo-w0.5.minus.concur.olympiad-bench-abs}{5.1}
\setval{4b.gdpo-w0.5.minus.concur.writing}{$-$8.1}
\setval{4b.gdpo-w0.5.minus.concur.writing-abs}{8.1}
\setval{4b.gdpo-w0.5.minus.concur.writingbench}{$-$8.7}
\setval{4b.gdpo-w0.5.minus.concur.writingbench-abs}{8.7}
\setval{4b.gdpo-w0.5.minus.gd2po.aime24}{+2.7}
\setval{4b.gdpo-w0.5.minus.gd2po.aime24-abs}{2.7}
\setval{4b.gdpo-w0.5.minus.gd2po.aime25}{+3.5}
\setval{4b.gdpo-w0.5.minus.gd2po.aime25-abs}{3.5}
\setval{4b.gdpo-w0.5.minus.gd2po.amc23}{+0.6}
\setval{4b.gdpo-w0.5.minus.gd2po.amc23-abs}{0.6}
\setval{4b.gdpo-w0.5.minus.gd2po.avg}{+0.5}
\setval{4b.gdpo-w0.5.minus.gd2po.avg-abs}{0.5}
\setval{4b.gdpo-w0.5.minus.gd2po.code}{$-$0.3}
\setval{4b.gdpo-w0.5.minus.gd2po.code-abs}{0.3}
\setval{4b.gdpo-w0.5.minus.gd2po.creative-writing-v3}{$-$6.6}
\setval{4b.gdpo-w0.5.minus.gd2po.creative-writing-v3-abs}{6.6}
\setval{4b.gdpo-w0.5.minus.gd2po.guidebench}{+0.5}
\setval{4b.gdpo-w0.5.minus.gd2po.guidebench-abs}{0.5}
\setval{4b.gdpo-w0.5.minus.gd2po.humaneval-plus}{$-$1.4}
\setval{4b.gdpo-w0.5.minus.gd2po.humaneval-plus-abs}{1.4}
\setval{4b.gdpo-w0.5.minus.gd2po.if}{+0.4}
\setval{4b.gdpo-w0.5.minus.gd2po.if-abs}{0.4}
\setval{4b.gdpo-w0.5.minus.gd2po.ifbench}{+1.7}
\setval{4b.gdpo-w0.5.minus.gd2po.ifbench-abs}{1.7}
\setval{4b.gdpo-w0.5.minus.gd2po.ifeval}{$-$0.8}
\setval{4b.gdpo-w0.5.minus.gd2po.ifeval-abs}{0.8}
\setval{4b.gdpo-w0.5.minus.gd2po.judge2.avg}{+0.9}
\setval{4b.gdpo-w0.5.minus.gd2po.judge2.avg-abs}{0.9}
\setval{4b.gdpo-w0.5.minus.gd2po.judge2.creative-writing-v3}{$-$2.8}
\setval{4b.gdpo-w0.5.minus.gd2po.judge2.creative-writing-v3-abs}{2.8}
\setval{4b.gdpo-w0.5.minus.gd2po.judge2.longbench-write}{$-$0.4}
\setval{4b.gdpo-w0.5.minus.gd2po.judge2.longbench-write-abs}{0.4}
\setval{4b.gdpo-w0.5.minus.gd2po.judge2.writing}{$-$0.7}
\setval{4b.gdpo-w0.5.minus.gd2po.judge2.writing-abs}{0.7}
\setval{4b.gdpo-w0.5.minus.gd2po.judge2.writingbench}{+1.1}
\setval{4b.gdpo-w0.5.minus.gd2po.judge2.writingbench-abs}{1.1}
\setval{4b.gdpo-w0.5.minus.gd2po.livecodebench}{+0.7}
\setval{4b.gdpo-w0.5.minus.gd2po.livecodebench-abs}{0.7}
\setval{4b.gdpo-w0.5.minus.gd2po.longbench-write}{$-$1.5}
\setval{4b.gdpo-w0.5.minus.gd2po.longbench-write-abs}{1.5}
\setval{4b.gdpo-w0.5.minus.gd2po.math}{+2.5}
\setval{4b.gdpo-w0.5.minus.gd2po.math-abs}{2.5}
\setval{4b.gdpo-w0.5.minus.gd2po.math500}{+4.1}
\setval{4b.gdpo-w0.5.minus.gd2po.math500-abs}{4.1}
\setval{4b.gdpo-w0.5.minus.gd2po.mbpp-plus}{$-$0.3}
\setval{4b.gdpo-w0.5.minus.gd2po.mbpp-plus-abs}{0.3}
\setval{4b.gdpo-w0.5.minus.gd2po.minerva}{$-$1.1}
\setval{4b.gdpo-w0.5.minus.gd2po.minerva-abs}{1.1}
\setval{4b.gdpo-w0.5.minus.gd2po.olympiad-bench}{+4.9}
\setval{4b.gdpo-w0.5.minus.gd2po.olympiad-bench-abs}{4.9}
\setval{4b.gdpo-w0.5.minus.gd2po.writing}{$-$2.5}
\setval{4b.gdpo-w0.5.minus.gd2po.writing-abs}{2.5}
\setval{4b.gdpo-w0.5.minus.gd2po.writingbench}{+0.6}
\setval{4b.gdpo-w0.5.minus.gd2po.writingbench-abs}{0.6}
\setval{4b.gdpo-w0.5.minus.gdpo-noaux.aime24}{$-$8.9}
\setval{4b.gdpo-w0.5.minus.gdpo-noaux.aime24-abs}{8.9}
\setval{4b.gdpo-w0.5.minus.gdpo-noaux.aime25}{$-$8.4}
\setval{4b.gdpo-w0.5.minus.gdpo-noaux.aime25-abs}{8.4}
\setval{4b.gdpo-w0.5.minus.gdpo-noaux.amc23}{$-$7.3}
\setval{4b.gdpo-w0.5.minus.gdpo-noaux.amc23-abs}{7.3}
\setval{4b.gdpo-w0.5.minus.gdpo-noaux.avg}{$-$2.1}
\setval{4b.gdpo-w0.5.minus.gdpo-noaux.avg-abs}{2.1}
\setval{4b.gdpo-w0.5.minus.gdpo-noaux.code}{$-$1.7}
\setval{4b.gdpo-w0.5.minus.gdpo-noaux.code-abs}{1.7}
\setval{4b.gdpo-w0.5.minus.gdpo-noaux.creative-writing-v3}{$-$1.2}
\setval{4b.gdpo-w0.5.minus.gdpo-noaux.creative-writing-v3-abs}{1.2}
\setval{4b.gdpo-w0.5.minus.gdpo-noaux.guidebench}{+3.6}
\setval{4b.gdpo-w0.5.minus.gdpo-noaux.guidebench-abs}{3.6}
\setval{4b.gdpo-w0.5.minus.gdpo-noaux.humaneval-plus}{$-$3.5}
\setval{4b.gdpo-w0.5.minus.gdpo-noaux.humaneval-plus-abs}{3.5}
\setval{4b.gdpo-w0.5.minus.gdpo-noaux.if}{+1.6}
\setval{4b.gdpo-w0.5.minus.gdpo-noaux.if-abs}{1.6}
\setval{4b.gdpo-w0.5.minus.gdpo-noaux.ifbench}{$-$1.0}
\setval{4b.gdpo-w0.5.minus.gdpo-noaux.ifbench-abs}{1.0}
\setval{4b.gdpo-w0.5.minus.gdpo-noaux.ifeval}{+2.0}
\setval{4b.gdpo-w0.5.minus.gdpo-noaux.ifeval-abs}{2.0}
\setval{4b.gdpo-w0.5.minus.gdpo-noaux.judge2.avg}{$-$0.8}
\setval{4b.gdpo-w0.5.minus.gdpo-noaux.judge2.avg-abs}{0.8}
\setval{4b.gdpo-w0.5.minus.gdpo-noaux.judge2.creative-writing-v3}{+6.7}
\setval{4b.gdpo-w0.5.minus.gdpo-noaux.judge2.creative-writing-v3-abs}{6.7}
\setval{4b.gdpo-w0.5.minus.gdpo-noaux.judge2.longbench-write}{+9.8}
\setval{4b.gdpo-w0.5.minus.gdpo-noaux.judge2.longbench-write-abs}{9.8}
\setval{4b.gdpo-w0.5.minus.gdpo-noaux.judge2.writing}{+8.0}
\setval{4b.gdpo-w0.5.minus.gdpo-noaux.judge2.writing-abs}{8.0}
\setval{4b.gdpo-w0.5.minus.gdpo-noaux.judge2.writingbench}{+7.6}
\setval{4b.gdpo-w0.5.minus.gdpo-noaux.judge2.writingbench-abs}{7.6}
\setval{4b.gdpo-w0.5.minus.gdpo-noaux.livecodebench}{$-$0.6}
\setval{4b.gdpo-w0.5.minus.gdpo-noaux.livecodebench-abs}{0.6}
\setval{4b.gdpo-w0.5.minus.gdpo-noaux.longbench-write}{+3.0}
\setval{4b.gdpo-w0.5.minus.gdpo-noaux.longbench-write-abs}{3.0}
\setval{4b.gdpo-w0.5.minus.gdpo-noaux.math}{$-$6.0}
\setval{4b.gdpo-w0.5.minus.gdpo-noaux.math-abs}{6.0}
\setval{4b.gdpo-w0.5.minus.gdpo-noaux.math500}{$-$3.0}
\setval{4b.gdpo-w0.5.minus.gdpo-noaux.math500-abs}{3.0}
\setval{4b.gdpo-w0.5.minus.gdpo-noaux.mbpp-plus}{$-$1.1}
\setval{4b.gdpo-w0.5.minus.gdpo-noaux.mbpp-plus-abs}{1.1}
\setval{4b.gdpo-w0.5.minus.gdpo-noaux.minerva}{$-$3.4}
\setval{4b.gdpo-w0.5.minus.gdpo-noaux.minerva-abs}{3.4}
\setval{4b.gdpo-w0.5.minus.gdpo-noaux.olympiad-bench}{$-$4.7}
\setval{4b.gdpo-w0.5.minus.gdpo-noaux.olympiad-bench-abs}{4.7}
\setval{4b.gdpo-w0.5.minus.gdpo-noaux.writing}{+1.7}
\setval{4b.gdpo-w0.5.minus.gdpo-noaux.writing-abs}{1.7}
\setval{4b.gdpo-w0.5.minus.gdpo-noaux.writingbench}{+3.4}
\setval{4b.gdpo-w0.5.minus.gdpo-noaux.writingbench-abs}{3.4}
\setval{4b.gdpo-w0.5.minus.gdpo.aime24}{+5.5}
\setval{4b.gdpo-w0.5.minus.gdpo.aime24-abs}{5.5}
\setval{4b.gdpo-w0.5.minus.gdpo.aime25}{+0.4}
\setval{4b.gdpo-w0.5.minus.gdpo.aime25-abs}{0.4}
\setval{4b.gdpo-w0.5.minus.gdpo.amc23}{$-$1.2}
\setval{4b.gdpo-w0.5.minus.gdpo.amc23-abs}{1.2}
\setval{4b.gdpo-w0.5.minus.gdpo.avg}{+0.3}
\setval{4b.gdpo-w0.5.minus.gdpo.avg-abs}{0.3}
\setval{4b.gdpo-w0.5.minus.gdpo.code}{+0.1}
\setval{4b.gdpo-w0.5.minus.gdpo.code-abs}{0.1}
\setval{4b.gdpo-w0.5.minus.gdpo.creative-writing-v3}{$-$7.9}
\setval{4b.gdpo-w0.5.minus.gdpo.creative-writing-v3-abs}{7.9}
\setval{4b.gdpo-w0.5.minus.gdpo.guidebench}{$-$1.1}
\setval{4b.gdpo-w0.5.minus.gdpo.guidebench-abs}{1.1}
\setval{4b.gdpo-w0.5.minus.gdpo.humaneval-plus}{+0.2}
\setval{4b.gdpo-w0.5.minus.gdpo.humaneval-plus-abs}{0.2}
\setval{4b.gdpo-w0.5.minus.gdpo.if}{+1.0}
\setval{4b.gdpo-w0.5.minus.gdpo.if-abs}{1.0}
\setval{4b.gdpo-w0.5.minus.gdpo.ifbench}{+3.0}
\setval{4b.gdpo-w0.5.minus.gdpo.ifbench-abs}{3.0}
\setval{4b.gdpo-w0.5.minus.gdpo.ifeval}{+1.1}
\setval{4b.gdpo-w0.5.minus.gdpo.ifeval-abs}{1.1}
\setval{4b.gdpo-w0.5.minus.gdpo.judge2.avg}{+0.6}
\setval{4b.gdpo-w0.5.minus.gdpo.judge2.avg-abs}{0.6}
\setval{4b.gdpo-w0.5.minus.gdpo.judge2.creative-writing-v3}{$-$4.3}
\setval{4b.gdpo-w0.5.minus.gdpo.judge2.creative-writing-v3-abs}{4.3}
\setval{4b.gdpo-w0.5.minus.gdpo.judge2.longbench-write}{$-$1.8}
\setval{4b.gdpo-w0.5.minus.gdpo.judge2.longbench-write-abs}{1.8}
\setval{4b.gdpo-w0.5.minus.gdpo.judge2.writing}{$-$2.2}
\setval{4b.gdpo-w0.5.minus.gdpo.judge2.writing-abs}{2.2}
\setval{4b.gdpo-w0.5.minus.gdpo.judge2.writingbench}{$-$0.4}
\setval{4b.gdpo-w0.5.minus.gdpo.judge2.writingbench-abs}{0.4}
\setval{4b.gdpo-w0.5.minus.gdpo.livecodebench}{$-$0.2}
\setval{4b.gdpo-w0.5.minus.gdpo.livecodebench-abs}{0.2}
\setval{4b.gdpo-w0.5.minus.gdpo.longbench-write}{$-$1.6}
\setval{4b.gdpo-w0.5.minus.gdpo.longbench-write-abs}{1.6}
\setval{4b.gdpo-w0.5.minus.gdpo.math}{+1.9}
\setval{4b.gdpo-w0.5.minus.gdpo.math-abs}{1.9}
\setval{4b.gdpo-w0.5.minus.gdpo.math500}{+2.5}
\setval{4b.gdpo-w0.5.minus.gdpo.math500-abs}{2.5}
\setval{4b.gdpo-w0.5.minus.gdpo.mbpp-plus}{+0.3}
\setval{4b.gdpo-w0.5.minus.gdpo.mbpp-plus-abs}{0.3}
\setval{4b.gdpo-w0.5.minus.gdpo.minerva}{+0.6}
\setval{4b.gdpo-w0.5.minus.gdpo.minerva-abs}{0.6}
\setval{4b.gdpo-w0.5.minus.gdpo.olympiad-bench}{+3.8}
\setval{4b.gdpo-w0.5.minus.gdpo.olympiad-bench-abs}{3.8}
\setval{4b.gdpo-w0.5.minus.gdpo.writing}{$-$3.6}
\setval{4b.gdpo-w0.5.minus.gdpo.writing-abs}{3.6}
\setval{4b.gdpo-w0.5.minus.gdpo.writingbench}{$-$1.2}
\setval{4b.gdpo-w0.5.minus.gdpo.writingbench-abs}{1.2}
\setval{4b.gdpo-w0.5.minus.geoma.aime24}{+3.0}
\setval{4b.gdpo-w0.5.minus.geoma.aime24-abs}{3.0}
\setval{4b.gdpo-w0.5.minus.geoma.aime25}{$-$0.3}
\setval{4b.gdpo-w0.5.minus.geoma.aime25-abs}{0.3}
\setval{4b.gdpo-w0.5.minus.geoma.amc23}{+3.0}
\setval{4b.gdpo-w0.5.minus.geoma.amc23-abs}{3.0}
\setval{4b.gdpo-w0.5.minus.geoma.avg}{+0.8}
\setval{4b.gdpo-w0.5.minus.geoma.avg-abs}{0.8}
\setval{4b.gdpo-w0.5.minus.geoma.code}{+0.4}
\setval{4b.gdpo-w0.5.minus.geoma.code-abs}{0.4}
\setval{4b.gdpo-w0.5.minus.geoma.creative-writing-v3}{$-$7.8}
\setval{4b.gdpo-w0.5.minus.geoma.creative-writing-v3-abs}{7.8}
\setval{4b.gdpo-w0.5.minus.geoma.guidebench}{+1.1}
\setval{4b.gdpo-w0.5.minus.geoma.guidebench-abs}{1.1}
\setval{4b.gdpo-w0.5.minus.geoma.humaneval-plus}{+1.8}
\setval{4b.gdpo-w0.5.minus.geoma.humaneval-plus-abs}{1.8}
\setval{4b.gdpo-w0.5.minus.geoma.if}{+1.1}
\setval{4b.gdpo-w0.5.minus.geoma.if-abs}{1.1}
\setval{4b.gdpo-w0.5.minus.geoma.ifbench}{+3.3}
\setval{4b.gdpo-w0.5.minus.geoma.ifbench-abs}{3.3}
\setval{4b.gdpo-w0.5.minus.geoma.ifeval}{$-$1.0}
\setval{4b.gdpo-w0.5.minus.geoma.ifeval-abs}{1.0}
\setval{4b.gdpo-w0.5.minus.geoma.judge2.avg}{+1.3}
\setval{4b.gdpo-w0.5.minus.geoma.judge2.avg-abs}{1.3}
\setval{4b.gdpo-w0.5.minus.geoma.judge2.creative-writing-v3}{$-$3.1}
\setval{4b.gdpo-w0.5.minus.geoma.judge2.creative-writing-v3-abs}{3.1}
\setval{4b.gdpo-w0.5.minus.geoma.judge2.longbench-write}{+2.9}
\setval{4b.gdpo-w0.5.minus.geoma.judge2.longbench-write-abs}{2.9}
\setval{4b.gdpo-w0.5.minus.geoma.judge2.writing}{+0.9}
\setval{4b.gdpo-w0.5.minus.geoma.judge2.writing-abs}{0.9}
\setval{4b.gdpo-w0.5.minus.geoma.judge2.writingbench}{+2.9}
\setval{4b.gdpo-w0.5.minus.geoma.judge2.writingbench-abs}{2.9}
\setval{4b.gdpo-w0.5.minus.geoma.livecodebench}{$-$1.5}
\setval{4b.gdpo-w0.5.minus.geoma.livecodebench-abs}{1.5}
\setval{4b.gdpo-w0.5.minus.geoma.longbench-write}{+1.6}
\setval{4b.gdpo-w0.5.minus.geoma.longbench-write-abs}{1.6}
\setval{4b.gdpo-w0.5.minus.geoma.math}{+2.1}
\setval{4b.gdpo-w0.5.minus.geoma.math-abs}{2.1}
\setval{4b.gdpo-w0.5.minus.geoma.math500}{+2.9}
\setval{4b.gdpo-w0.5.minus.geoma.math500-abs}{2.9}
\setval{4b.gdpo-w0.5.minus.geoma.mbpp-plus}{+1.0}
\setval{4b.gdpo-w0.5.minus.geoma.mbpp-plus-abs}{1.0}
\setval{4b.gdpo-w0.5.minus.geoma.minerva}{$-$0.8}
\setval{4b.gdpo-w0.5.minus.geoma.minerva-abs}{0.8}
\setval{4b.gdpo-w0.5.minus.geoma.olympiad-bench}{+4.6}
\setval{4b.gdpo-w0.5.minus.geoma.olympiad-bench-abs}{4.6}
\setval{4b.gdpo-w0.5.minus.geoma.writing}{$-$1.7}
\setval{4b.gdpo-w0.5.minus.geoma.writing-abs}{1.7}
\setval{4b.gdpo-w0.5.minus.geoma.writingbench}{+1.2}
\setval{4b.gdpo-w0.5.minus.geoma.writingbench-abs}{1.2}
\setval{4b.gdpo-w0.5.minus.grpo-nodiv.aime24}{+2.2}
\setval{4b.gdpo-w0.5.minus.grpo-nodiv.aime24-abs}{2.2}
\setval{4b.gdpo-w0.5.minus.grpo-nodiv.aime25}{+2.0}
\setval{4b.gdpo-w0.5.minus.grpo-nodiv.aime25-abs}{2.0}
\setval{4b.gdpo-w0.5.minus.grpo-nodiv.amc23}{+3.8}
\setval{4b.gdpo-w0.5.minus.grpo-nodiv.amc23-abs}{3.8}
\setval{4b.gdpo-w0.5.minus.grpo-nodiv.avg}{+1.6}
\setval{4b.gdpo-w0.5.minus.grpo-nodiv.avg-abs}{1.6}
\setval{4b.gdpo-w0.5.minus.grpo-nodiv.code}{+0.4}
\setval{4b.gdpo-w0.5.minus.grpo-nodiv.code-abs}{0.4}
\setval{4b.gdpo-w0.5.minus.grpo-nodiv.creative-writing-v3}{$-$4.4}
\setval{4b.gdpo-w0.5.minus.grpo-nodiv.creative-writing-v3-abs}{4.4}
\setval{4b.gdpo-w0.5.minus.grpo-nodiv.guidebench}{+0.8}
\setval{4b.gdpo-w0.5.minus.grpo-nodiv.guidebench-abs}{0.8}
\setval{4b.gdpo-w0.5.minus.grpo-nodiv.humaneval-plus}{+1.5}
\setval{4b.gdpo-w0.5.minus.grpo-nodiv.humaneval-plus-abs}{1.5}
\setval{4b.gdpo-w0.5.minus.grpo-nodiv.if}{+2.0}
\setval{4b.gdpo-w0.5.minus.grpo-nodiv.if-abs}{2.0}
\setval{4b.gdpo-w0.5.minus.grpo-nodiv.ifbench}{+5.7}
\setval{4b.gdpo-w0.5.minus.grpo-nodiv.ifbench-abs}{5.7}
\setval{4b.gdpo-w0.5.minus.grpo-nodiv.ifeval}{$-$0.5}
\setval{4b.gdpo-w0.5.minus.grpo-nodiv.ifeval-abs}{0.5}
\setval{4b.gdpo-w0.5.minus.grpo-nodiv.judge2.avg}{+1.8}
\setval{4b.gdpo-w0.5.minus.grpo-nodiv.judge2.avg-abs}{1.8}
\setval{4b.gdpo-w0.5.minus.grpo-nodiv.judge2.creative-writing-v3}{$-$0.5}
\setval{4b.gdpo-w0.5.minus.grpo-nodiv.judge2.creative-writing-v3-abs}{0.5}
\setval{4b.gdpo-w0.5.minus.grpo-nodiv.judge2.longbench-write}{+1.9}
\setval{4b.gdpo-w0.5.minus.grpo-nodiv.judge2.longbench-write-abs}{1.9}
\setval{4b.gdpo-w0.5.minus.grpo-nodiv.judge2.writing}{+1.8}
\setval{4b.gdpo-w0.5.minus.grpo-nodiv.judge2.writing-abs}{1.8}
\setval{4b.gdpo-w0.5.minus.grpo-nodiv.judge2.writingbench}{+4.0}
\setval{4b.gdpo-w0.5.minus.grpo-nodiv.judge2.writingbench-abs}{4.0}
\setval{4b.gdpo-w0.5.minus.grpo-nodiv.livecodebench}{$-$0.3}
\setval{4b.gdpo-w0.5.minus.grpo-nodiv.livecodebench-abs}{0.3}
\setval{4b.gdpo-w0.5.minus.grpo-nodiv.longbench-write}{+1.4}
\setval{4b.gdpo-w0.5.minus.grpo-nodiv.longbench-write-abs}{1.4}
\setval{4b.gdpo-w0.5.minus.grpo-nodiv.math}{+2.4}
\setval{4b.gdpo-w0.5.minus.grpo-nodiv.math-abs}{2.4}
\setval{4b.gdpo-w0.5.minus.grpo-nodiv.math500}{+2.3}
\setval{4b.gdpo-w0.5.minus.grpo-nodiv.math500-abs}{2.3}
\setval{4b.gdpo-w0.5.minus.grpo-nodiv.mbpp-plus}{+0.0}
\setval{4b.gdpo-w0.5.minus.grpo-nodiv.mbpp-plus-abs}{0.0}
\setval{4b.gdpo-w0.5.minus.grpo-nodiv.minerva}{+0.3}
\setval{4b.gdpo-w0.5.minus.grpo-nodiv.minerva-abs}{0.3}
\setval{4b.gdpo-w0.5.minus.grpo-nodiv.olympiad-bench}{+3.6}
\setval{4b.gdpo-w0.5.minus.grpo-nodiv.olympiad-bench-abs}{3.6}
\setval{4b.gdpo-w0.5.minus.grpo-nodiv.writing}{+0.7}
\setval{4b.gdpo-w0.5.minus.grpo-nodiv.writing-abs}{0.7}
\setval{4b.gdpo-w0.5.minus.grpo-nodiv.writingbench}{+5.1}
\setval{4b.gdpo-w0.5.minus.grpo-nodiv.writingbench-abs}{5.1}
\setval{4b.gdpo-w0.5.minus.grpo.aime24}{+2.6}
\setval{4b.gdpo-w0.5.minus.grpo.aime24-abs}{2.6}
\setval{4b.gdpo-w0.5.minus.grpo.aime25}{$-$2.3}
\setval{4b.gdpo-w0.5.minus.grpo.aime25-abs}{2.3}
\setval{4b.gdpo-w0.5.minus.grpo.amc23}{+3.7}
\setval{4b.gdpo-w0.5.minus.grpo.amc23-abs}{3.7}
\setval{4b.gdpo-w0.5.minus.grpo.avg}{$-$0.1}
\setval{4b.gdpo-w0.5.minus.grpo.avg-abs}{0.1}
\setval{4b.gdpo-w0.5.minus.grpo.code}{$-$0.2}
\setval{4b.gdpo-w0.5.minus.grpo.code-abs}{0.2}
\setval{4b.gdpo-w0.5.minus.grpo.creative-writing-v3}{$-$9.2}
\setval{4b.gdpo-w0.5.minus.grpo.creative-writing-v3-abs}{9.2}
\setval{4b.gdpo-w0.5.minus.grpo.guidebench}{+1.6}
\setval{4b.gdpo-w0.5.minus.grpo.guidebench-abs}{1.6}
\setval{4b.gdpo-w0.5.minus.grpo.humaneval-plus}{$-$0.7}
\setval{4b.gdpo-w0.5.minus.grpo.humaneval-plus-abs}{0.7}
\setval{4b.gdpo-w0.5.minus.grpo.if}{+1.2}
\setval{4b.gdpo-w0.5.minus.grpo.if-abs}{1.2}
\setval{4b.gdpo-w0.5.minus.grpo.ifbench}{+1.7}
\setval{4b.gdpo-w0.5.minus.grpo.ifbench-abs}{1.7}
\setval{4b.gdpo-w0.5.minus.grpo.ifeval}{+0.4}
\setval{4b.gdpo-w0.5.minus.grpo.ifeval-abs}{0.4}
\setval{4b.gdpo-w0.5.minus.grpo.judge2.avg}{+0.6}
\setval{4b.gdpo-w0.5.minus.grpo.judge2.avg-abs}{0.6}
\setval{4b.gdpo-w0.5.minus.grpo.judge2.creative-writing-v3}{$-$5.0}
\setval{4b.gdpo-w0.5.minus.grpo.judge2.creative-writing-v3-abs}{5.0}
\setval{4b.gdpo-w0.5.minus.grpo.judge2.longbench-write}{$-$0.2}
\setval{4b.gdpo-w0.5.minus.grpo.judge2.longbench-write-abs}{0.2}
\setval{4b.gdpo-w0.5.minus.grpo.judge2.writing}{$-$1.7}
\setval{4b.gdpo-w0.5.minus.grpo.judge2.writing-abs}{1.7}
\setval{4b.gdpo-w0.5.minus.grpo.judge2.writingbench}{+0.2}
\setval{4b.gdpo-w0.5.minus.grpo.judge2.writingbench-abs}{0.2}
\setval{4b.gdpo-w0.5.minus.grpo.livecodebench}{+0.2}
\setval{4b.gdpo-w0.5.minus.grpo.livecodebench-abs}{0.2}
\setval{4b.gdpo-w0.5.minus.grpo.longbench-write}{$-$1.8}
\setval{4b.gdpo-w0.5.minus.grpo.longbench-write-abs}{1.8}
\setval{4b.gdpo-w0.5.minus.grpo.math}{+1.7}
\setval{4b.gdpo-w0.5.minus.grpo.math-abs}{1.7}
\setval{4b.gdpo-w0.5.minus.grpo.math500}{+1.5}
\setval{4b.gdpo-w0.5.minus.grpo.math500-abs}{1.5}
\setval{4b.gdpo-w0.5.minus.grpo.mbpp-plus}{+0.0}
\setval{4b.gdpo-w0.5.minus.grpo.mbpp-plus-abs}{0.0}
\setval{4b.gdpo-w0.5.minus.grpo.minerva}{+0.8}
\setval{4b.gdpo-w0.5.minus.grpo.minerva-abs}{0.8}
\setval{4b.gdpo-w0.5.minus.grpo.olympiad-bench}{+3.8}
\setval{4b.gdpo-w0.5.minus.grpo.olympiad-bench-abs}{3.8}
\setval{4b.gdpo-w0.5.minus.grpo.writing}{$-$4.8}
\setval{4b.gdpo-w0.5.minus.grpo.writing-abs}{4.8}
\setval{4b.gdpo-w0.5.minus.grpo.writingbench}{$-$3.5}
\setval{4b.gdpo-w0.5.minus.grpo.writingbench-abs}{3.5}
\setval{4b.gdpo-w0.5.minus.mt-grpo.aime24}{+3.9}
\setval{4b.gdpo-w0.5.minus.mt-grpo.aime24-abs}{3.9}
\setval{4b.gdpo-w0.5.minus.mt-grpo.aime25}{+4.2}
\setval{4b.gdpo-w0.5.minus.mt-grpo.aime25-abs}{4.2}
\setval{4b.gdpo-w0.5.minus.mt-grpo.amc23}{$-$1.4}
\setval{4b.gdpo-w0.5.minus.mt-grpo.amc23-abs}{1.4}
\setval{4b.gdpo-w0.5.minus.mt-grpo.avg}{+1.5}
\setval{4b.gdpo-w0.5.minus.mt-grpo.avg-abs}{1.5}
\setval{4b.gdpo-w0.5.minus.mt-grpo.code}{+1.9}
\setval{4b.gdpo-w0.5.minus.mt-grpo.code-abs}{1.9}
\setval{4b.gdpo-w0.5.minus.mt-grpo.creative-writing-v3}{$-$6.4}
\setval{4b.gdpo-w0.5.minus.mt-grpo.creative-writing-v3-abs}{6.4}
\setval{4b.gdpo-w0.5.minus.mt-grpo.guidebench}{$-$0.1}
\setval{4b.gdpo-w0.5.minus.mt-grpo.guidebench-abs}{0.1}
\setval{4b.gdpo-w0.5.minus.mt-grpo.humaneval-plus}{+2.9}
\setval{4b.gdpo-w0.5.minus.mt-grpo.humaneval-plus-abs}{2.9}
\setval{4b.gdpo-w0.5.minus.mt-grpo.if}{+1.7}
\setval{4b.gdpo-w0.5.minus.mt-grpo.if-abs}{1.7}
\setval{4b.gdpo-w0.5.minus.mt-grpo.ifbench}{+5.0}
\setval{4b.gdpo-w0.5.minus.mt-grpo.ifbench-abs}{5.0}
\setval{4b.gdpo-w0.5.minus.mt-grpo.ifeval}{+0.1}
\setval{4b.gdpo-w0.5.minus.mt-grpo.ifeval-abs}{0.1}
\setval{4b.gdpo-w0.5.minus.mt-grpo.judge2.avg}{+1.6}
\setval{4b.gdpo-w0.5.minus.mt-grpo.judge2.avg-abs}{1.6}
\setval{4b.gdpo-w0.5.minus.mt-grpo.judge2.creative-writing-v3}{$-$4.0}
\setval{4b.gdpo-w0.5.minus.mt-grpo.judge2.creative-writing-v3-abs}{4.0}
\setval{4b.gdpo-w0.5.minus.mt-grpo.judge2.longbench-write}{$-$0.3}
\setval{4b.gdpo-w0.5.minus.mt-grpo.judge2.longbench-write-abs}{0.3}
\setval{4b.gdpo-w0.5.minus.mt-grpo.judge2.writing}{$-$1.3}
\setval{4b.gdpo-w0.5.minus.mt-grpo.judge2.writing-abs}{1.3}
\setval{4b.gdpo-w0.5.minus.mt-grpo.judge2.writingbench}{+0.3}
\setval{4b.gdpo-w0.5.minus.mt-grpo.judge2.writingbench-abs}{0.3}
\setval{4b.gdpo-w0.5.minus.mt-grpo.livecodebench}{+1.5}
\setval{4b.gdpo-w0.5.minus.mt-grpo.livecodebench-abs}{1.5}
\setval{4b.gdpo-w0.5.minus.mt-grpo.longbench-write}{+0.6}
\setval{4b.gdpo-w0.5.minus.mt-grpo.longbench-write-abs}{0.6}
\setval{4b.gdpo-w0.5.minus.mt-grpo.math}{+2.9}
\setval{4b.gdpo-w0.5.minus.mt-grpo.math-abs}{2.9}
\setval{4b.gdpo-w0.5.minus.mt-grpo.math500}{+4.8}
\setval{4b.gdpo-w0.5.minus.mt-grpo.math500-abs}{4.8}
\setval{4b.gdpo-w0.5.minus.mt-grpo.mbpp-plus}{+1.1}
\setval{4b.gdpo-w0.5.minus.mt-grpo.mbpp-plus-abs}{1.1}
\setval{4b.gdpo-w0.5.minus.mt-grpo.minerva}{+0.1}
\setval{4b.gdpo-w0.5.minus.mt-grpo.minerva-abs}{0.1}
\setval{4b.gdpo-w0.5.minus.mt-grpo.olympiad-bench}{+5.9}
\setval{4b.gdpo-w0.5.minus.mt-grpo.olympiad-bench-abs}{5.9}
\setval{4b.gdpo-w0.5.minus.mt-grpo.writing}{$-$1.7}
\setval{4b.gdpo-w0.5.minus.mt-grpo.writing-abs}{1.7}
\setval{4b.gdpo-w0.5.minus.mt-grpo.writingbench}{+0.6}
\setval{4b.gdpo-w0.5.minus.mt-grpo.writingbench-abs}{0.6}
\setval{4b.gdpo-w0.5.minus.papo.aime24}{+5.3}
\setval{4b.gdpo-w0.5.minus.papo.aime24-abs}{5.3}
\setval{4b.gdpo-w0.5.minus.papo.aime25}{+2.8}
\setval{4b.gdpo-w0.5.minus.papo.aime25-abs}{2.8}
\setval{4b.gdpo-w0.5.minus.papo.amc23}{+4.8}
\setval{4b.gdpo-w0.5.minus.papo.amc23-abs}{4.8}
\setval{4b.gdpo-w0.5.minus.papo.avg}{+2.0}
\setval{4b.gdpo-w0.5.minus.papo.avg-abs}{2.0}
\setval{4b.gdpo-w0.5.minus.papo.code}{+0.0}
\setval{4b.gdpo-w0.5.minus.papo.code-abs}{0.0}
\setval{4b.gdpo-w0.5.minus.papo.creative-writing-v3}{$-$8.4}
\setval{4b.gdpo-w0.5.minus.papo.creative-writing-v3-abs}{8.4}
\setval{4b.gdpo-w0.5.minus.papo.guidebench}{$-$1.2}
\setval{4b.gdpo-w0.5.minus.papo.guidebench-abs}{1.2}
\setval{4b.gdpo-w0.5.minus.papo.humaneval-plus}{$-$1.0}
\setval{4b.gdpo-w0.5.minus.papo.humaneval-plus-abs}{1.0}
\setval{4b.gdpo-w0.5.minus.papo.if}{+1.9}
\setval{4b.gdpo-w0.5.minus.papo.if-abs}{1.9}
\setval{4b.gdpo-w0.5.minus.papo.ifbench}{+6.0}
\setval{4b.gdpo-w0.5.minus.papo.ifbench-abs}{6.0}
\setval{4b.gdpo-w0.5.minus.papo.ifeval}{+1.0}
\setval{4b.gdpo-w0.5.minus.papo.ifeval-abs}{1.0}
\setval{4b.gdpo-w0.5.minus.papo.judge2.avg}{+2.1}
\setval{4b.gdpo-w0.5.minus.papo.judge2.avg-abs}{2.1}
\setval{4b.gdpo-w0.5.minus.papo.judge2.creative-writing-v3}{$-$4.2}
\setval{4b.gdpo-w0.5.minus.papo.judge2.creative-writing-v3-abs}{4.2}
\setval{4b.gdpo-w0.5.minus.papo.judge2.longbench-write}{+3.0}
\setval{4b.gdpo-w0.5.minus.papo.judge2.longbench-write-abs}{3.0}
\setval{4b.gdpo-w0.5.minus.papo.judge2.writing}{+0.1}
\setval{4b.gdpo-w0.5.minus.papo.judge2.writing-abs}{0.1}
\setval{4b.gdpo-w0.5.minus.papo.judge2.writingbench}{+1.5}
\setval{4b.gdpo-w0.5.minus.papo.judge2.writingbench-abs}{1.5}
\setval{4b.gdpo-w0.5.minus.papo.livecodebench}{+0.6}
\setval{4b.gdpo-w0.5.minus.papo.livecodebench-abs}{0.6}
\setval{4b.gdpo-w0.5.minus.papo.longbench-write}{+3.4}
\setval{4b.gdpo-w0.5.minus.papo.longbench-write-abs}{3.4}
\setval{4b.gdpo-w0.5.minus.papo.math}{+4.1}
\setval{4b.gdpo-w0.5.minus.papo.math-abs}{4.1}
\setval{4b.gdpo-w0.5.minus.papo.math500}{+5.3}
\setval{4b.gdpo-w0.5.minus.papo.math500-abs}{5.3}
\setval{4b.gdpo-w0.5.minus.papo.mbpp-plus}{+0.5}
\setval{4b.gdpo-w0.5.minus.papo.mbpp-plus-abs}{0.5}
\setval{4b.gdpo-w0.5.minus.papo.minerva}{$-$0.4}
\setval{4b.gdpo-w0.5.minus.papo.minerva-abs}{0.4}
\setval{4b.gdpo-w0.5.minus.papo.olympiad-bench}{+7.0}
\setval{4b.gdpo-w0.5.minus.papo.olympiad-bench-abs}{7.0}
\setval{4b.gdpo-w0.5.minus.papo.writing}{$-$0.2}
\setval{4b.gdpo-w0.5.minus.papo.writing-abs}{0.2}
\setval{4b.gdpo-w0.5.minus.papo.writingbench}{+4.2}
\setval{4b.gdpo-w0.5.minus.papo.writingbench-abs}{4.2}
\setval{4b.gdpo-w0.5.minus.saw.aime24}{+2.1}
\setval{4b.gdpo-w0.5.minus.saw.aime24-abs}{2.1}
\setval{4b.gdpo-w0.5.minus.saw.aime25}{$-$0.8}
\setval{4b.gdpo-w0.5.minus.saw.aime25-abs}{0.8}
\setval{4b.gdpo-w0.5.minus.saw.amc23}{+1.7}
\setval{4b.gdpo-w0.5.minus.saw.amc23-abs}{1.7}
\setval{4b.gdpo-w0.5.minus.saw.avg}{+0.4}
\setval{4b.gdpo-w0.5.minus.saw.avg-abs}{0.4}
\setval{4b.gdpo-w0.5.minus.saw.code}{+1.2}
\setval{4b.gdpo-w0.5.minus.saw.code-abs}{1.2}
\setval{4b.gdpo-w0.5.minus.saw.creative-writing-v3}{$-$7.0}
\setval{4b.gdpo-w0.5.minus.saw.creative-writing-v3-abs}{7.0}
\setval{4b.gdpo-w0.5.minus.saw.guidebench}{$-$1.4}
\setval{4b.gdpo-w0.5.minus.saw.guidebench-abs}{1.4}
\setval{4b.gdpo-w0.5.minus.saw.humaneval-plus}{+1.3}
\setval{4b.gdpo-w0.5.minus.saw.humaneval-plus-abs}{1.3}
\setval{4b.gdpo-w0.5.minus.saw.if}{$-$0.9}
\setval{4b.gdpo-w0.5.minus.saw.if-abs}{0.9}
\setval{4b.gdpo-w0.5.minus.saw.ifbench}{$-$1.0}
\setval{4b.gdpo-w0.5.minus.saw.ifbench-abs}{1.0}
\setval{4b.gdpo-w0.5.minus.saw.ifeval}{$-$0.1}
\setval{4b.gdpo-w0.5.minus.saw.ifeval-abs}{0.1}
\setval{4b.gdpo-w0.5.minus.saw.judge2.avg}{+0.4}
\setval{4b.gdpo-w0.5.minus.saw.judge2.avg-abs}{0.4}
\setval{4b.gdpo-w0.5.minus.saw.judge2.creative-writing-v3}{$-$5.5}
\setval{4b.gdpo-w0.5.minus.saw.judge2.creative-writing-v3-abs}{5.5}
\setval{4b.gdpo-w0.5.minus.saw.judge2.longbench-write}{+0.5}
\setval{4b.gdpo-w0.5.minus.saw.judge2.longbench-write-abs}{0.5}
\setval{4b.gdpo-w0.5.minus.saw.judge2.writing}{$-$1.6}
\setval{4b.gdpo-w0.5.minus.saw.judge2.writing-abs}{1.6}
\setval{4b.gdpo-w0.5.minus.saw.judge2.writingbench}{+0.2}
\setval{4b.gdpo-w0.5.minus.saw.judge2.writingbench-abs}{0.2}
\setval{4b.gdpo-w0.5.minus.saw.livecodebench}{+1.4}
\setval{4b.gdpo-w0.5.minus.saw.livecodebench-abs}{1.4}
\setval{4b.gdpo-w0.5.minus.saw.longbench-write}{+1.2}
\setval{4b.gdpo-w0.5.minus.saw.longbench-write-abs}{1.2}
\setval{4b.gdpo-w0.5.minus.saw.math}{+1.6}
\setval{4b.gdpo-w0.5.minus.saw.math-abs}{1.6}
\setval{4b.gdpo-w0.5.minus.saw.math500}{+4.0}
\setval{4b.gdpo-w0.5.minus.saw.math500-abs}{4.0}
\setval{4b.gdpo-w0.5.minus.saw.mbpp-plus}{+1.1}
\setval{4b.gdpo-w0.5.minus.saw.mbpp-plus-abs}{1.1}
\setval{4b.gdpo-w0.5.minus.saw.minerva}{$-$2.0}
\setval{4b.gdpo-w0.5.minus.saw.minerva-abs}{2.0}
\setval{4b.gdpo-w0.5.minus.saw.olympiad-bench}{+4.6}
\setval{4b.gdpo-w0.5.minus.saw.olympiad-bench-abs}{4.6}
\setval{4b.gdpo-w0.5.minus.saw.writing}{$-$1.6}
\setval{4b.gdpo-w0.5.minus.saw.writing-abs}{1.6}
\setval{4b.gdpo-w0.5.minus.saw.writingbench}{+0.9}
\setval{4b.gdpo-w0.5.minus.saw.writingbench-abs}{0.9}
\setval{4b.gdpo-w0.5.passing-at-ceiling.code.early}{0.71}
\setval{4b.gdpo-w0.5.passing-at-ceiling.code.late}{0.73}
\setval{4b.gdpo-w0.5.passing-at-ceiling.code.run}{0.73}
\setval{4b.gdpo-w0.5.passing-at-ceiling.if.early}{0.28}
\setval{4b.gdpo-w0.5.passing-at-ceiling.if.late}{0.56}
\setval{4b.gdpo-w0.5.passing-at-ceiling.if.run}{0.48}
\setval{4b.gdpo-w0.5.passing-at-ceiling.math.early}{1.00}
\setval{4b.gdpo-w0.5.passing-at-ceiling.math.late}{1.00}
\setval{4b.gdpo-w0.5.passing-at-ceiling.math.run}{1.00}
\setval{4b.gdpo-w0.5.passing-at-ceiling.writing.early}{0.00}
\setval{4b.gdpo-w0.5.passing-at-ceiling.writing.late}{0.00}
\setval{4b.gdpo-w0.5.passing-at-ceiling.writing.run}{0.00}
\setval{4b.gdpo-w0.5.primary-cosine.code.early}{0.98}
\setval{4b.gdpo-w0.5.primary-cosine.code.late}{0.98}
\setval{4b.gdpo-w0.5.primary-cosine.code.run}{0.98}
\setval{4b.gdpo-w0.5.primary-cosine.if.early}{0.82}
\setval{4b.gdpo-w0.5.primary-cosine.if.late}{0.80}
\setval{4b.gdpo-w0.5.primary-cosine.if.run}{0.83}
\setval{4b.gdpo-w0.5.primary-cosine.math.early}{0.99}
\setval{4b.gdpo-w0.5.primary-cosine.math.late}{0.98}
\setval{4b.gdpo-w0.5.primary-cosine.math.run}{0.98}
\setval{4b.gdpo-w0.5.primary-cosine.writing.early}{0.93}
\setval{4b.gdpo-w0.5.primary-cosine.writing.late}{0.89}
\setval{4b.gdpo-w0.5.primary-cosine.writing.run}{0.89}
\setval{4b.gdpo-w0.5.primary-level.code.early}{0.53}
\setval{4b.gdpo-w0.5.primary-level.code.late}{0.75}
\setval{4b.gdpo-w0.5.primary-level.code.run}{0.70}
\setval{4b.gdpo-w0.5.primary-level.if.early}{0.32}
\setval{4b.gdpo-w0.5.primary-level.if.late}{0.74}
\setval{4b.gdpo-w0.5.primary-level.if.run}{0.63}
\setval{4b.gdpo-w0.5.primary-level.math.early}{0.48}
\setval{4b.gdpo-w0.5.primary-level.math.late}{0.70}
\setval{4b.gdpo-w0.5.primary-level.math.run}{0.64}
\setval{4b.gdpo-w0.5.primary-level.writing.early}{0.53}
\setval{4b.gdpo-w0.5.primary-level.writing.late}{0.76}
\setval{4b.gdpo-w0.5.primary-level.writing.run}{0.72}
\setval{4b.gdpo-w0.5.solved.code.early}{0.06}
\setval{4b.gdpo-w0.5.solved.code.late}{0.03}
\setval{4b.gdpo-w0.5.solved.code.run}{0.04}
\setval{4b.gdpo-w0.5.solved.if.early}{0.03}
\setval{4b.gdpo-w0.5.solved.if.late}{0.19}
\setval{4b.gdpo-w0.5.solved.if.run}{0.13}
\setval{4b.gdpo-w0.5.solved.math.early}{0.12}
\setval{4b.gdpo-w0.5.solved.math.late}{0.22}
\setval{4b.gdpo-w0.5.solved.math.run}{0.20}
\setval{4b.gdpo-w0.5.solved.writing.early}{0.00}
\setval{4b.gdpo-w0.5.solved.writing.late}{0.00}
\setval{4b.gdpo-w0.5.solved.writing.run}{0.00}
\setval{4b.gdpo-w0.5.update-share.code.early}{0.21}
\setval{4b.gdpo-w0.5.update-share.code.late}{0.22}
\setval{4b.gdpo-w0.5.update-share.code.run}{0.22}
\setval{4b.gdpo-w0.5.update-share.if.early}{0.25}
\setval{4b.gdpo-w0.5.update-share.if.late}{0.19}
\setval{4b.gdpo-w0.5.update-share.if.run}{0.23}
\setval{4b.gdpo-w0.5.update-share.math.early}{0.18}
\setval{4b.gdpo-w0.5.update-share.math.late}{0.24}
\setval{4b.gdpo-w0.5.update-share.math.run}{0.23}
\setval{4b.gdpo-w0.5.update-share.writing.early}{0.35}
\setval{4b.gdpo-w0.5.update-share.writing.late}{0.36}
\setval{4b.gdpo-w0.5.update-share.writing.run}{0.32}
\setval{4b.gdpo.aux-change.code}{+0.20}
\setval{4b.gdpo.aux-change.if}{+0.24}
\setval{4b.gdpo.aux-change.math}{+0.12}
\setval{4b.gdpo.aux-change.writing}{+0.20}
\setval{4b.gdpo.aux-change.writing-rm}{+0.37}
\setval{4b.gdpo.aux-full.code.early}{0.77}
\setval{4b.gdpo.aux-full.code.late}{0.96}
\setval{4b.gdpo.aux-full.code.run}{0.95}
\setval{4b.gdpo.aux-full.if.early}{0.67}
\setval{4b.gdpo.aux-full.if.late}{0.91}
\setval{4b.gdpo.aux-full.if.run}{0.86}
\setval{4b.gdpo.aux-full.math.early}{0.64}
\setval{4b.gdpo.aux-full.math.late}{0.87}
\setval{4b.gdpo.aux-full.math.run}{0.84}
\setval{4b.gdpo.aux-full.writing-rm.early}{0.09}
\setval{4b.gdpo.aux-full.writing-rm.late}{0.63}
\setval{4b.gdpo.aux-full.writing-rm.run}{0.50}
\setval{4b.gdpo.aux-full.writing.early}{0.64}
\setval{4b.gdpo.aux-full.writing.late}{0.75}
\setval{4b.gdpo.aux-full.writing.run}{0.73}
\setval{4b.gdpo.aux-level.code.early}{0.93}
\setval{4b.gdpo.aux-level.code.late}{0.99}
\setval{4b.gdpo.aux-level.code.run}{0.99}
\setval{4b.gdpo.aux-level.if.early}{0.84}
\setval{4b.gdpo.aux-level.if.late}{0.95}
\setval{4b.gdpo.aux-level.if.run}{0.93}
\setval{4b.gdpo.aux-level.math.early}{0.93}
\setval{4b.gdpo.aux-level.math.late}{0.97}
\setval{4b.gdpo.aux-level.math.run}{0.97}
\setval{4b.gdpo.aux-level.writing-rm.early}{0.78}
\setval{4b.gdpo.aux-level.writing-rm.late}{0.97}
\setval{4b.gdpo.aux-level.writing-rm.run}{0.94}
\setval{4b.gdpo.aux-level.writing.early}{0.91}
\setval{4b.gdpo.aux-level.writing.late}{0.97}
\setval{4b.gdpo.aux-level.writing.run}{0.96}
\setval{4b.gdpo.bench.aime24}{9.1}
\setval{4b.gdpo.bench.aime25}{9.9}
\setval{4b.gdpo.bench.amc23}{59.6}
\setval{4b.gdpo.bench.creative-writing-v3}{46.0}
\setval{4b.gdpo.bench.guidebench}{80.6}
\setval{4b.gdpo.bench.humaneval-plus}{81.8}
\setval{4b.gdpo.bench.ifbench}{36.3}
\setval{4b.gdpo.bench.ifeval}{85.7}
\setval{4b.gdpo.bench.livecodebench}{24.4}
\setval{4b.gdpo.bench.longbench-write}{69.0}
\setval{4b.gdpo.bench.math500}{79.3}
\setval{4b.gdpo.bench.mbpp-plus}{70.9}
\setval{4b.gdpo.bench.minerva}{45.5}
\setval{4b.gdpo.bench.olympiad-bench}{44.7}
\setval{4b.gdpo.bench.writingbench}{71.9}
\setval{4b.gdpo.eval.avg}{54.3}
\setval{4b.gdpo.eval.code}{59.0}
\setval{4b.gdpo.eval.if}{67.6}
\setval{4b.gdpo.eval.math}{41.3}
\setval{4b.gdpo.eval.writing}{62.3}
\setval{4b.gdpo.gate.code.against.early}{0.03}
\setval{4b.gdpo.gate.code.against.late}{0.03}
\setval{4b.gdpo.gate.code.against.run}{0.02}
\setval{4b.gdpo.gate.code.agreement}{+0.003}
\setval{4b.gdpo.gate.code.ceiling.early}{0.09}
\setval{4b.gdpo.gate.code.ceiling.late}{0.04}
\setval{4b.gdpo.gate.code.ceiling.run}{0.04}
\setval{4b.gdpo.gate.code.cosine.early}{0.85}
\setval{4b.gdpo.gate.code.cosine.late}{0.79}
\setval{4b.gdpo.gate.code.cosine.run}{0.82}
\setval{4b.gdpo.gate.if.against.early}{0.04}
\setval{4b.gdpo.gate.if.against.late}{0.05}
\setval{4b.gdpo.gate.if.against.run}{0.04}
\setval{4b.gdpo.gate.if.agreement}{+0.015}
\setval{4b.gdpo.gate.if.ceiling.early}{0.05}
\setval{4b.gdpo.gate.if.ceiling.late}{0.18}
\setval{4b.gdpo.gate.if.ceiling.run}{0.11}
\setval{4b.gdpo.gate.if.cosine.early}{0.75}
\setval{4b.gdpo.gate.if.cosine.late}{0.48}
\setval{4b.gdpo.gate.if.cosine.run}{0.59}
\setval{4b.gdpo.gate.math.against.early}{0.00}
\setval{4b.gdpo.gate.math.against.late}{0.00}
\setval{4b.gdpo.gate.math.against.run}{0.00}
\setval{4b.gdpo.gate.math.ceiling.early}{0.19}
\setval{4b.gdpo.gate.math.ceiling.late}{0.19}
\setval{4b.gdpo.gate.math.ceiling.run}{0.17}
\setval{4b.gdpo.gate.math.cosine.early}{0.89}
\setval{4b.gdpo.gate.math.cosine.late}{0.89}
\setval{4b.gdpo.gate.math.cosine.run}{0.90}
\setval{4b.gdpo.gate.writing-rm.against.early}{0.22}
\setval{4b.gdpo.gate.writing-rm.against.late}{0.51}
\setval{4b.gdpo.gate.writing-rm.against.run}{0.48}
\setval{4b.gdpo.gate.writing-rm.agreement}{+0.058}
\setval{4b.gdpo.gate.writing-rm.ceiling.early}{0.00}
\setval{4b.gdpo.gate.writing-rm.ceiling.late}{0.00}
\setval{4b.gdpo.gate.writing-rm.ceiling.run}{0.00}
\setval{4b.gdpo.gate.writing-rm.cosine.early}{0.67}
\setval{4b.gdpo.gate.writing-rm.cosine.late}{0.15}
\setval{4b.gdpo.gate.writing-rm.cosine.run}{0.23}
\setval{4b.gdpo.gate.writing.against.early}{0.29}
\setval{4b.gdpo.gate.writing.against.late}{0.60}
\setval{4b.gdpo.gate.writing.against.run}{0.54}
\setval{4b.gdpo.gate.writing.agreement}{$-$0.036}
\setval{4b.gdpo.gate.writing.ceiling.early}{0.00}
\setval{4b.gdpo.gate.writing.ceiling.late}{0.00}
\setval{4b.gdpo.gate.writing.ceiling.run}{0.00}
\setval{4b.gdpo.gate.writing.cosine.early}{0.51}
\setval{4b.gdpo.gate.writing.cosine.late}{0.04}
\setval{4b.gdpo.gate.writing.cosine.run}{0.12}
\setval{4b.gdpo.judge2.avg}{51.9}
\setval{4b.gdpo.judge2.creative-writing-v3}{34.4}
\setval{4b.gdpo.judge2.longbench-write}{62.3}
\setval{4b.gdpo.judge2.writing}{50.2}
\setval{4b.gdpo.judge2.writingbench}{54.0}
\setval{4b.gdpo.last-step}{300}
\setval{4b.gdpo.minus.base.aime24}{+1.2}
\setval{4b.gdpo.minus.base.aime24-abs}{1.2}
\setval{4b.gdpo.minus.base.aime25}{+2.8}
\setval{4b.gdpo.minus.base.aime25-abs}{2.8}
\setval{4b.gdpo.minus.base.amc23}{+15.7}
\setval{4b.gdpo.minus.base.amc23-abs}{15.7}
\setval{4b.gdpo.minus.base.avg}{+19.8}
\setval{4b.gdpo.minus.base.avg-abs}{19.8}
\setval{4b.gdpo.minus.base.code}{+14.4}
\setval{4b.gdpo.minus.base.code-abs}{14.4}
\setval{4b.gdpo.minus.base.creative-writing-v3}{+21.2}
\setval{4b.gdpo.minus.base.creative-writing-v3-abs}{21.2}
\setval{4b.gdpo.minus.base.guidebench}{+55.8}
\setval{4b.gdpo.minus.base.guidebench-abs}{55.8}
\setval{4b.gdpo.minus.base.humaneval-plus}{+16.8}
\setval{4b.gdpo.minus.base.humaneval-plus-abs}{16.8}
\setval{4b.gdpo.minus.base.if}{+37.8}
\setval{4b.gdpo.minus.base.if-abs}{37.8}
\setval{4b.gdpo.minus.base.ifbench}{+19.3}
\setval{4b.gdpo.minus.base.ifbench-abs}{19.3}
\setval{4b.gdpo.minus.base.ifeval}{+38.4}
\setval{4b.gdpo.minus.base.ifeval-abs}{38.4}
\setval{4b.gdpo.minus.base.judge2.avg}{+18.0}
\setval{4b.gdpo.minus.base.judge2.avg-abs}{18.0}
\setval{4b.gdpo.minus.base.judge2.creative-writing-v3}{+13.5}
\setval{4b.gdpo.minus.base.judge2.creative-writing-v3-abs}{13.5}
\setval{4b.gdpo.minus.base.judge2.longbench-write}{+21.4}
\setval{4b.gdpo.minus.base.judge2.longbench-write-abs}{21.4}
\setval{4b.gdpo.minus.base.judge2.writing}{+19.5}
\setval{4b.gdpo.minus.base.judge2.writing-abs}{19.5}
\setval{4b.gdpo.minus.base.judge2.writingbench}{+23.5}
\setval{4b.gdpo.minus.base.judge2.writingbench-abs}{23.5}
\setval{4b.gdpo.minus.base.livecodebench}{+11.2}
\setval{4b.gdpo.minus.base.livecodebench-abs}{11.2}
\setval{4b.gdpo.minus.base.longbench-write}{+24.6}
\setval{4b.gdpo.minus.base.longbench-write-abs}{24.6}
\setval{4b.gdpo.minus.base.math}{+9.1}
\setval{4b.gdpo.minus.base.math-abs}{9.1}
\setval{4b.gdpo.minus.base.math500}{+12.0}
\setval{4b.gdpo.minus.base.math500-abs}{12.0}
\setval{4b.gdpo.minus.base.mbpp-plus}{+15.3}
\setval{4b.gdpo.minus.base.mbpp-plus-abs}{15.3}
\setval{4b.gdpo.minus.base.minerva}{+13.9}
\setval{4b.gdpo.minus.base.minerva-abs}{13.9}
\setval{4b.gdpo.minus.base.olympiad-bench}{+9.0}
\setval{4b.gdpo.minus.base.olympiad-bench-abs}{9.0}
\setval{4b.gdpo.minus.base.writing}{+28.4}
\setval{4b.gdpo.minus.base.writing-abs}{28.4}
\setval{4b.gdpo.minus.base.writingbench}{+39.5}
\setval{4b.gdpo.minus.base.writingbench-abs}{39.5}
\setval{4b.gdpo.minus.concur-free.aime24}{$-$2.8}
\setval{4b.gdpo.minus.concur-free.aime24-abs}{2.8}
\setval{4b.gdpo.minus.concur-free.aime25}{$-$4.7}
\setval{4b.gdpo.minus.concur-free.aime25-abs}{4.7}
\setval{4b.gdpo.minus.concur-free.amc23}{+5.5}
\setval{4b.gdpo.minus.concur-free.amc23-abs}{5.5}
\setval{4b.gdpo.minus.concur-free.avg}{$-$1.2}
\setval{4b.gdpo.minus.concur-free.avg-abs}{1.2}
\setval{4b.gdpo.minus.concur-free.code}{+0.7}
\setval{4b.gdpo.minus.concur-free.code-abs}{0.7}
\setval{4b.gdpo.minus.concur-free.creative-writing-v3}{+0.5}
\setval{4b.gdpo.minus.concur-free.creative-writing-v3-abs}{0.5}
\setval{4b.gdpo.minus.concur-free.guidebench}{+1.6}
\setval{4b.gdpo.minus.concur-free.guidebench-abs}{1.6}
\setval{4b.gdpo.minus.concur-free.humaneval-plus}{+0.7}
\setval{4b.gdpo.minus.concur-free.humaneval-plus-abs}{0.7}
\setval{4b.gdpo.minus.concur-free.if}{$-$0.7}
\setval{4b.gdpo.minus.concur-free.if-abs}{0.7}
\setval{4b.gdpo.minus.concur-free.ifbench}{$-$2.7}
\setval{4b.gdpo.minus.concur-free.ifbench-abs}{2.7}
\setval{4b.gdpo.minus.concur-free.ifeval}{$-$1.0}
\setval{4b.gdpo.minus.concur-free.ifeval-abs}{1.0}
\setval{4b.gdpo.minus.concur-free.judge2.avg}{$-$0.9}
\setval{4b.gdpo.minus.concur-free.judge2.avg-abs}{0.9}
\setval{4b.gdpo.minus.concur-free.judge2.creative-writing-v3}{$-$0.9}
\setval{4b.gdpo.minus.concur-free.judge2.creative-writing-v3-abs}{0.9}
\setval{4b.gdpo.minus.concur-free.judge2.longbench-write}{$-$7.0}
\setval{4b.gdpo.minus.concur-free.judge2.longbench-write-abs}{7.0}
\setval{4b.gdpo.minus.concur-free.judge2.writing}{$-$2.8}
\setval{4b.gdpo.minus.concur-free.judge2.writing-abs}{2.8}
\setval{4b.gdpo.minus.concur-free.judge2.writingbench}{$-$0.4}
\setval{4b.gdpo.minus.concur-free.judge2.writingbench-abs}{0.4}
\setval{4b.gdpo.minus.concur-free.livecodebench}{+0.2}
\setval{4b.gdpo.minus.concur-free.livecodebench-abs}{0.2}
\setval{4b.gdpo.minus.concur-free.longbench-write}{$-$8.8}
\setval{4b.gdpo.minus.concur-free.longbench-write-abs}{8.8}
\setval{4b.gdpo.minus.concur-free.math}{$-$0.8}
\setval{4b.gdpo.minus.concur-free.math-abs}{0.8}
\setval{4b.gdpo.minus.concur-free.math500}{$-$1.0}
\setval{4b.gdpo.minus.concur-free.math500-abs}{1.0}
\setval{4b.gdpo.minus.concur-free.mbpp-plus}{+1.2}
\setval{4b.gdpo.minus.concur-free.mbpp-plus-abs}{1.2}
\setval{4b.gdpo.minus.concur-free.minerva}{$-$0.8}
\setval{4b.gdpo.minus.concur-free.minerva-abs}{0.8}
\setval{4b.gdpo.minus.concur-free.olympiad-bench}{$-$1.1}
\setval{4b.gdpo.minus.concur-free.olympiad-bench-abs}{1.1}
\setval{4b.gdpo.minus.concur-free.writing}{$-$4.5}
\setval{4b.gdpo.minus.concur-free.writing-abs}{4.5}
\setval{4b.gdpo.minus.concur-free.writingbench}{$-$5.4}
\setval{4b.gdpo.minus.concur-free.writingbench-abs}{5.4}
\setval{4b.gdpo.minus.concur-tied.aime24}{$-$10.4}
\setval{4b.gdpo.minus.concur-tied.aime24-abs}{10.4}
\setval{4b.gdpo.minus.concur-tied.aime25}{$-$6.5}
\setval{4b.gdpo.minus.concur-tied.aime25-abs}{6.5}
\setval{4b.gdpo.minus.concur-tied.amc23}{$-$9.6}
\setval{4b.gdpo.minus.concur-tied.amc23-abs}{9.6}
\setval{4b.gdpo.minus.concur-tied.avg}{$-$4.4}
\setval{4b.gdpo.minus.concur-tied.avg-abs}{4.4}
\setval{4b.gdpo.minus.concur-tied.code}{$-$0.1}
\setval{4b.gdpo.minus.concur-tied.code-abs}{0.1}
\setval{4b.gdpo.minus.concur-tied.creative-writing-v3}{$-$1.9}
\setval{4b.gdpo.minus.concur-tied.creative-writing-v3-abs}{1.9}
\setval{4b.gdpo.minus.concur-tied.guidebench}{+1.4}
\setval{4b.gdpo.minus.concur-tied.guidebench-abs}{1.4}
\setval{4b.gdpo.minus.concur-tied.humaneval-plus}{$-$2.2}
\setval{4b.gdpo.minus.concur-tied.humaneval-plus-abs}{2.2}
\setval{4b.gdpo.minus.concur-tied.if}{$-$0.6}
\setval{4b.gdpo.minus.concur-tied.if-abs}{0.6}
\setval{4b.gdpo.minus.concur-tied.ifbench}{$-$1.3}
\setval{4b.gdpo.minus.concur-tied.ifbench-abs}{1.3}
\setval{4b.gdpo.minus.concur-tied.ifeval}{$-$2.0}
\setval{4b.gdpo.minus.concur-tied.ifeval-abs}{2.0}
\setval{4b.gdpo.minus.concur-tied.judge2.avg}{$-$3.6}
\setval{4b.gdpo.minus.concur-tied.judge2.avg-abs}{3.6}
\setval{4b.gdpo.minus.concur-tied.judge2.creative-writing-v3}{+0.4}
\setval{4b.gdpo.minus.concur-tied.judge2.creative-writing-v3-abs}{0.4}
\setval{4b.gdpo.minus.concur-tied.judge2.longbench-write}{$-$5.7}
\setval{4b.gdpo.minus.concur-tied.judge2.longbench-write-abs}{5.7}
\setval{4b.gdpo.minus.concur-tied.judge2.writing}{$-$2.2}
\setval{4b.gdpo.minus.concur-tied.judge2.writing-abs}{2.2}
\setval{4b.gdpo.minus.concur-tied.judge2.writingbench}{$-$1.3}
\setval{4b.gdpo.minus.concur-tied.judge2.writingbench-abs}{1.3}
\setval{4b.gdpo.minus.concur-tied.livecodebench}{$-$0.3}
\setval{4b.gdpo.minus.concur-tied.livecodebench-abs}{0.3}
\setval{4b.gdpo.minus.concur-tied.longbench-write}{$-$8.7}
\setval{4b.gdpo.minus.concur-tied.longbench-write-abs}{8.7}
\setval{4b.gdpo.minus.concur-tied.math}{$-$7.4}
\setval{4b.gdpo.minus.concur-tied.math-abs}{7.4}
\setval{4b.gdpo.minus.concur-tied.math500}{$-$6.2}
\setval{4b.gdpo.minus.concur-tied.math500-abs}{6.2}
\setval{4b.gdpo.minus.concur-tied.mbpp-plus}{+2.3}
\setval{4b.gdpo.minus.concur-tied.mbpp-plus-abs}{2.3}
\setval{4b.gdpo.minus.concur-tied.minerva}{$-$3.6}
\setval{4b.gdpo.minus.concur-tied.minerva-abs}{3.6}
\setval{4b.gdpo.minus.concur-tied.olympiad-bench}{$-$8.2}
\setval{4b.gdpo.minus.concur-tied.olympiad-bench-abs}{8.2}
\setval{4b.gdpo.minus.concur-tied.writing}{$-$6.3}
\setval{4b.gdpo.minus.concur-tied.writing-abs}{6.3}
\setval{4b.gdpo.minus.concur-tied.writingbench}{$-$8.2}
\setval{4b.gdpo.minus.concur-tied.writingbench-abs}{8.2}
\setval{4b.gdpo.minus.concur.aime24}{$-$11.5}
\setval{4b.gdpo.minus.concur.aime24-abs}{11.5}
\setval{4b.gdpo.minus.concur.aime25}{$-$8.9}
\setval{4b.gdpo.minus.concur.aime25-abs}{8.9}
\setval{4b.gdpo.minus.concur.amc23}{$-$9.8}
\setval{4b.gdpo.minus.concur.amc23-abs}{9.8}
\setval{4b.gdpo.minus.concur.avg}{$-$4.5}
\setval{4b.gdpo.minus.concur.avg-abs}{4.5}
\setval{4b.gdpo.minus.concur.code}{$-$0.7}
\setval{4b.gdpo.minus.concur.code-abs}{0.7}
\setval{4b.gdpo.minus.concur.creative-writing-v3}{+1.3}
\setval{4b.gdpo.minus.concur.creative-writing-v3-abs}{1.3}
\setval{4b.gdpo.minus.concur.guidebench}{+3.3}
\setval{4b.gdpo.minus.concur.guidebench-abs}{3.3}
\setval{4b.gdpo.minus.concur.humaneval-plus}{$-$0.6}
\setval{4b.gdpo.minus.concur.humaneval-plus-abs}{0.6}
\setval{4b.gdpo.minus.concur.if}{$-$1.0}
\setval{4b.gdpo.minus.concur.if-abs}{1.0}
\setval{4b.gdpo.minus.concur.ifbench}{$-$4.0}
\setval{4b.gdpo.minus.concur.ifbench-abs}{4.0}
\setval{4b.gdpo.minus.concur.ifeval}{$-$2.4}
\setval{4b.gdpo.minus.concur.ifeval-abs}{2.4}
\setval{4b.gdpo.minus.concur.judge2.avg}{$-$3.9}
\setval{4b.gdpo.minus.concur.judge2.avg-abs}{3.9}
\setval{4b.gdpo.minus.concur.judge2.creative-writing-v3}{+2.0}
\setval{4b.gdpo.minus.concur.judge2.creative-writing-v3-abs}{2.0}
\setval{4b.gdpo.minus.concur.judge2.longbench-write}{$-$4.3}
\setval{4b.gdpo.minus.concur.judge2.longbench-write-abs}{4.3}
\setval{4b.gdpo.minus.concur.judge2.writing}{$-$1.4}
\setval{4b.gdpo.minus.concur.judge2.writing-abs}{1.4}
\setval{4b.gdpo.minus.concur.judge2.writingbench}{$-$2.0}
\setval{4b.gdpo.minus.concur.judge2.writingbench-abs}{2.0}
\setval{4b.gdpo.minus.concur.livecodebench}{$-$0.7}
\setval{4b.gdpo.minus.concur.livecodebench-abs}{0.7}
\setval{4b.gdpo.minus.concur.longbench-write}{$-$7.3}
\setval{4b.gdpo.minus.concur.longbench-write-abs}{7.3}
\setval{4b.gdpo.minus.concur.math}{$-$8.1}
\setval{4b.gdpo.minus.concur.math-abs}{8.1}
\setval{4b.gdpo.minus.concur.math500}{$-$6.0}
\setval{4b.gdpo.minus.concur.math500-abs}{6.0}
\setval{4b.gdpo.minus.concur.mbpp-plus}{$-$0.6}
\setval{4b.gdpo.minus.concur.mbpp-plus-abs}{0.6}
\setval{4b.gdpo.minus.concur.minerva}{$-$3.7}
\setval{4b.gdpo.minus.concur.minerva-abs}{3.7}
\setval{4b.gdpo.minus.concur.olympiad-bench}{$-$8.9}
\setval{4b.gdpo.minus.concur.olympiad-bench-abs}{8.9}
\setval{4b.gdpo.minus.concur.writing}{$-$4.5}
\setval{4b.gdpo.minus.concur.writing-abs}{4.5}
\setval{4b.gdpo.minus.concur.writingbench}{$-$7.5}
\setval{4b.gdpo.minus.concur.writingbench-abs}{7.5}
\setval{4b.gdpo.minus.gd2po.aime24}{$-$2.8}
\setval{4b.gdpo.minus.gd2po.aime24-abs}{2.8}
\setval{4b.gdpo.minus.gd2po.aime25}{+3.1}
\setval{4b.gdpo.minus.gd2po.aime25-abs}{3.1}
\setval{4b.gdpo.minus.gd2po.amc23}{+1.9}
\setval{4b.gdpo.minus.gd2po.amc23-abs}{1.9}
\setval{4b.gdpo.minus.gd2po.avg}{+0.2}
\setval{4b.gdpo.minus.gd2po.avg-abs}{0.2}
\setval{4b.gdpo.minus.gd2po.code}{$-$0.4}
\setval{4b.gdpo.minus.gd2po.code-abs}{0.4}
\setval{4b.gdpo.minus.gd2po.creative-writing-v3}{+1.3}
\setval{4b.gdpo.minus.gd2po.creative-writing-v3-abs}{1.3}
\setval{4b.gdpo.minus.gd2po.guidebench}{+1.5}
\setval{4b.gdpo.minus.gd2po.guidebench-abs}{1.5}
\setval{4b.gdpo.minus.gd2po.humaneval-plus}{$-$1.6}
\setval{4b.gdpo.minus.gd2po.humaneval-plus-abs}{1.6}
\setval{4b.gdpo.minus.gd2po.if}{$-$0.6}
\setval{4b.gdpo.minus.gd2po.if-abs}{0.6}
\setval{4b.gdpo.minus.gd2po.ifbench}{$-$1.3}
\setval{4b.gdpo.minus.gd2po.ifbench-abs}{1.3}
\setval{4b.gdpo.minus.gd2po.ifeval}{$-$1.9}
\setval{4b.gdpo.minus.gd2po.ifeval-abs}{1.9}
\setval{4b.gdpo.minus.gd2po.judge2.avg}{+0.3}
\setval{4b.gdpo.minus.gd2po.judge2.avg-abs}{0.3}
\setval{4b.gdpo.minus.gd2po.judge2.creative-writing-v3}{+1.5}
\setval{4b.gdpo.minus.gd2po.judge2.creative-writing-v3-abs}{1.5}
\setval{4b.gdpo.minus.gd2po.judge2.longbench-write}{+1.3}
\setval{4b.gdpo.minus.gd2po.judge2.longbench-write-abs}{1.3}
\setval{4b.gdpo.minus.gd2po.judge2.writing}{+1.4}
\setval{4b.gdpo.minus.gd2po.judge2.writing-abs}{1.4}
\setval{4b.gdpo.minus.gd2po.judge2.writingbench}{+1.5}
\setval{4b.gdpo.minus.gd2po.judge2.writingbench-abs}{1.5}
\setval{4b.gdpo.minus.gd2po.livecodebench}{+0.9}
\setval{4b.gdpo.minus.gd2po.livecodebench-abs}{0.9}
\setval{4b.gdpo.minus.gd2po.longbench-write}{+0.1}
\setval{4b.gdpo.minus.gd2po.longbench-write-abs}{0.1}
\setval{4b.gdpo.minus.gd2po.math}{+0.5}
\setval{4b.gdpo.minus.gd2po.math-abs}{0.5}
\setval{4b.gdpo.minus.gd2po.math500}{+1.6}
\setval{4b.gdpo.minus.gd2po.math500-abs}{1.6}
\setval{4b.gdpo.minus.gd2po.mbpp-plus}{$-$0.6}
\setval{4b.gdpo.minus.gd2po.mbpp-plus-abs}{0.6}
\setval{4b.gdpo.minus.gd2po.minerva}{$-$1.7}
\setval{4b.gdpo.minus.gd2po.minerva-abs}{1.7}
\setval{4b.gdpo.minus.gd2po.olympiad-bench}{+1.1}
\setval{4b.gdpo.minus.gd2po.olympiad-bench-abs}{1.1}
\setval{4b.gdpo.minus.gd2po.writing}{+1.0}
\setval{4b.gdpo.minus.gd2po.writing-abs}{1.0}
\setval{4b.gdpo.minus.gd2po.writingbench}{+1.8}
\setval{4b.gdpo.minus.gd2po.writingbench-abs}{1.8}
\setval{4b.gdpo.minus.gdpo-noaux.aime24}{$-$14.4}
\setval{4b.gdpo.minus.gdpo-noaux.aime24-abs}{14.4}
\setval{4b.gdpo.minus.gdpo-noaux.aime25}{$-$8.9}
\setval{4b.gdpo.minus.gdpo-noaux.aime25-abs}{8.9}
\setval{4b.gdpo.minus.gdpo-noaux.amc23}{$-$6.1}
\setval{4b.gdpo.minus.gdpo-noaux.amc23-abs}{6.1}
\setval{4b.gdpo.minus.gdpo-noaux.avg}{$-$2.3}
\setval{4b.gdpo.minus.gdpo-noaux.avg-abs}{2.3}
\setval{4b.gdpo.minus.gdpo-noaux.code}{$-$1.8}
\setval{4b.gdpo.minus.gdpo-noaux.code-abs}{1.8}
\setval{4b.gdpo.minus.gdpo-noaux.creative-writing-v3}{+6.6}
\setval{4b.gdpo.minus.gdpo-noaux.creative-writing-v3-abs}{6.6}
\setval{4b.gdpo.minus.gdpo-noaux.guidebench}{+4.7}
\setval{4b.gdpo.minus.gdpo-noaux.guidebench-abs}{4.7}
\setval{4b.gdpo.minus.gdpo-noaux.humaneval-plus}{$-$3.7}
\setval{4b.gdpo.minus.gdpo-noaux.humaneval-plus-abs}{3.7}
\setval{4b.gdpo.minus.gdpo-noaux.if}{+0.6}
\setval{4b.gdpo.minus.gdpo-noaux.if-abs}{0.6}
\setval{4b.gdpo.minus.gdpo-noaux.ifbench}{$-$4.0}
\setval{4b.gdpo.minus.gdpo-noaux.ifbench-abs}{4.0}
\setval{4b.gdpo.minus.gdpo-noaux.ifeval}{+1.0}
\setval{4b.gdpo.minus.gdpo-noaux.ifeval-abs}{1.0}
\setval{4b.gdpo.minus.gdpo-noaux.judge2.avg}{$-$1.4}
\setval{4b.gdpo.minus.gdpo-noaux.judge2.avg-abs}{1.4}
\setval{4b.gdpo.minus.gdpo-noaux.judge2.creative-writing-v3}{+11.0}
\setval{4b.gdpo.minus.gdpo-noaux.judge2.creative-writing-v3-abs}{11.0}
\setval{4b.gdpo.minus.gdpo-noaux.judge2.longbench-write}{+11.5}
\setval{4b.gdpo.minus.gdpo-noaux.judge2.longbench-write-abs}{11.5}
\setval{4b.gdpo.minus.gdpo-noaux.judge2.writing}{+10.2}
\setval{4b.gdpo.minus.gdpo-noaux.judge2.writing-abs}{10.2}
\setval{4b.gdpo.minus.gdpo-noaux.judge2.writingbench}{+8.0}
\setval{4b.gdpo.minus.gdpo-noaux.judge2.writingbench-abs}{8.0}
\setval{4b.gdpo.minus.gdpo-noaux.livecodebench}{$-$0.3}
\setval{4b.gdpo.minus.gdpo-noaux.livecodebench-abs}{0.3}
\setval{4b.gdpo.minus.gdpo-noaux.longbench-write}{+4.6}
\setval{4b.gdpo.minus.gdpo-noaux.longbench-write-abs}{4.6}
\setval{4b.gdpo.minus.gdpo-noaux.math}{$-$7.9}
\setval{4b.gdpo.minus.gdpo-noaux.math-abs}{7.9}
\setval{4b.gdpo.minus.gdpo-noaux.math500}{$-$5.5}
\setval{4b.gdpo.minus.gdpo-noaux.math500-abs}{5.5}
\setval{4b.gdpo.minus.gdpo-noaux.mbpp-plus}{$-$1.5}
\setval{4b.gdpo.minus.gdpo-noaux.mbpp-plus-abs}{1.5}
\setval{4b.gdpo.minus.gdpo-noaux.minerva}{$-$4.0}
\setval{4b.gdpo.minus.gdpo-noaux.minerva-abs}{4.0}
\setval{4b.gdpo.minus.gdpo-noaux.olympiad-bench}{$-$8.5}
\setval{4b.gdpo.minus.gdpo-noaux.olympiad-bench-abs}{8.5}
\setval{4b.gdpo.minus.gdpo-noaux.writing}{+5.3}
\setval{4b.gdpo.minus.gdpo-noaux.writing-abs}{5.3}
\setval{4b.gdpo.minus.gdpo-noaux.writingbench}{+4.6}
\setval{4b.gdpo.minus.gdpo-noaux.writingbench-abs}{4.6}
\setval{4b.gdpo.minus.gdpo-w0.5.aime24}{$-$5.5}
\setval{4b.gdpo.minus.gdpo-w0.5.aime24-abs}{5.5}
\setval{4b.gdpo.minus.gdpo-w0.5.aime25}{$-$0.4}
\setval{4b.gdpo.minus.gdpo-w0.5.aime25-abs}{0.4}
\setval{4b.gdpo.minus.gdpo-w0.5.amc23}{+1.2}
\setval{4b.gdpo.minus.gdpo-w0.5.amc23-abs}{1.2}
\setval{4b.gdpo.minus.gdpo-w0.5.avg}{$-$0.3}
\setval{4b.gdpo.minus.gdpo-w0.5.avg-abs}{0.3}
\setval{4b.gdpo.minus.gdpo-w0.5.code}{$-$0.1}
\setval{4b.gdpo.minus.gdpo-w0.5.code-abs}{0.1}
\setval{4b.gdpo.minus.gdpo-w0.5.creative-writing-v3}{+7.9}
\setval{4b.gdpo.minus.gdpo-w0.5.creative-writing-v3-abs}{7.9}
\setval{4b.gdpo.minus.gdpo-w0.5.guidebench}{+1.1}
\setval{4b.gdpo.minus.gdpo-w0.5.guidebench-abs}{1.1}
\setval{4b.gdpo.minus.gdpo-w0.5.humaneval-plus}{$-$0.2}
\setval{4b.gdpo.minus.gdpo-w0.5.humaneval-plus-abs}{0.2}
\setval{4b.gdpo.minus.gdpo-w0.5.if}{$-$1.0}
\setval{4b.gdpo.minus.gdpo-w0.5.if-abs}{1.0}
\setval{4b.gdpo.minus.gdpo-w0.5.ifbench}{$-$3.0}
\setval{4b.gdpo.minus.gdpo-w0.5.ifbench-abs}{3.0}
\setval{4b.gdpo.minus.gdpo-w0.5.ifeval}{$-$1.1}
\setval{4b.gdpo.minus.gdpo-w0.5.ifeval-abs}{1.1}
\setval{4b.gdpo.minus.gdpo-w0.5.judge2.avg}{$-$0.6}
\setval{4b.gdpo.minus.gdpo-w0.5.judge2.avg-abs}{0.6}
\setval{4b.gdpo.minus.gdpo-w0.5.judge2.creative-writing-v3}{+4.3}
\setval{4b.gdpo.minus.gdpo-w0.5.judge2.creative-writing-v3-abs}{4.3}
\setval{4b.gdpo.minus.gdpo-w0.5.judge2.longbench-write}{+1.8}
\setval{4b.gdpo.minus.gdpo-w0.5.judge2.longbench-write-abs}{1.8}
\setval{4b.gdpo.minus.gdpo-w0.5.judge2.writing}{+2.2}
\setval{4b.gdpo.minus.gdpo-w0.5.judge2.writing-abs}{2.2}
\setval{4b.gdpo.minus.gdpo-w0.5.judge2.writingbench}{+0.4}
\setval{4b.gdpo.minus.gdpo-w0.5.judge2.writingbench-abs}{0.4}
\setval{4b.gdpo.minus.gdpo-w0.5.livecodebench}{+0.2}
\setval{4b.gdpo.minus.gdpo-w0.5.livecodebench-abs}{0.2}
\setval{4b.gdpo.minus.gdpo-w0.5.longbench-write}{+1.6}
\setval{4b.gdpo.minus.gdpo-w0.5.longbench-write-abs}{1.6}
\setval{4b.gdpo.minus.gdpo-w0.5.math}{$-$1.9}
\setval{4b.gdpo.minus.gdpo-w0.5.math-abs}{1.9}
\setval{4b.gdpo.minus.gdpo-w0.5.math500}{$-$2.5}
\setval{4b.gdpo.minus.gdpo-w0.5.math500-abs}{2.5}
\setval{4b.gdpo.minus.gdpo-w0.5.mbpp-plus}{$-$0.3}
\setval{4b.gdpo.minus.gdpo-w0.5.mbpp-plus-abs}{0.3}
\setval{4b.gdpo.minus.gdpo-w0.5.minerva}{$-$0.6}
\setval{4b.gdpo.minus.gdpo-w0.5.minerva-abs}{0.6}
\setval{4b.gdpo.minus.gdpo-w0.5.olympiad-bench}{$-$3.8}
\setval{4b.gdpo.minus.gdpo-w0.5.olympiad-bench-abs}{3.8}
\setval{4b.gdpo.minus.gdpo-w0.5.writing}{+3.6}
\setval{4b.gdpo.minus.gdpo-w0.5.writing-abs}{3.6}
\setval{4b.gdpo.minus.gdpo-w0.5.writingbench}{+1.2}
\setval{4b.gdpo.minus.gdpo-w0.5.writingbench-abs}{1.2}
\setval{4b.gdpo.minus.geoma.aime24}{$-$2.5}
\setval{4b.gdpo.minus.geoma.aime24-abs}{2.5}
\setval{4b.gdpo.minus.geoma.aime25}{$-$0.7}
\setval{4b.gdpo.minus.geoma.aime25-abs}{0.7}
\setval{4b.gdpo.minus.geoma.amc23}{+4.2}
\setval{4b.gdpo.minus.geoma.amc23-abs}{4.2}
\setval{4b.gdpo.minus.geoma.avg}{+0.5}
\setval{4b.gdpo.minus.geoma.avg-abs}{0.5}
\setval{4b.gdpo.minus.geoma.code}{+0.3}
\setval{4b.gdpo.minus.geoma.code-abs}{0.3}
\setval{4b.gdpo.minus.geoma.creative-writing-v3}{+0.1}
\setval{4b.gdpo.minus.geoma.creative-writing-v3-abs}{0.1}
\setval{4b.gdpo.minus.geoma.guidebench}{+2.1}
\setval{4b.gdpo.minus.geoma.guidebench-abs}{2.1}
\setval{4b.gdpo.minus.geoma.humaneval-plus}{+1.6}
\setval{4b.gdpo.minus.geoma.humaneval-plus-abs}{1.6}
\setval{4b.gdpo.minus.geoma.if}{+0.1}
\setval{4b.gdpo.minus.geoma.if-abs}{0.1}
\setval{4b.gdpo.minus.geoma.ifbench}{+0.3}
\setval{4b.gdpo.minus.geoma.ifbench-abs}{0.3}
\setval{4b.gdpo.minus.geoma.ifeval}{$-$2.0}
\setval{4b.gdpo.minus.geoma.ifeval-abs}{2.0}
\setval{4b.gdpo.minus.geoma.judge2.avg}{+0.8}
\setval{4b.gdpo.minus.geoma.judge2.avg-abs}{0.8}
\setval{4b.gdpo.minus.geoma.judge2.creative-writing-v3}{+1.3}
\setval{4b.gdpo.minus.geoma.judge2.creative-writing-v3-abs}{1.3}
\setval{4b.gdpo.minus.geoma.judge2.longbench-write}{+4.6}
\setval{4b.gdpo.minus.geoma.judge2.longbench-write-abs}{4.6}
\setval{4b.gdpo.minus.geoma.judge2.writing}{+3.1}
\setval{4b.gdpo.minus.geoma.judge2.writing-abs}{3.1}
\setval{4b.gdpo.minus.geoma.judge2.writingbench}{+3.3}
\setval{4b.gdpo.minus.geoma.judge2.writingbench-abs}{3.3}
\setval{4b.gdpo.minus.geoma.livecodebench}{$-$1.3}
\setval{4b.gdpo.minus.geoma.livecodebench-abs}{1.3}
\setval{4b.gdpo.minus.geoma.longbench-write}{+3.2}
\setval{4b.gdpo.minus.geoma.longbench-write-abs}{3.2}
\setval{4b.gdpo.minus.geoma.math}{+0.1}
\setval{4b.gdpo.minus.geoma.math-abs}{0.1}
\setval{4b.gdpo.minus.geoma.math500}{+0.4}
\setval{4b.gdpo.minus.geoma.math500-abs}{0.4}
\setval{4b.gdpo.minus.geoma.mbpp-plus}{+0.7}
\setval{4b.gdpo.minus.geoma.mbpp-plus-abs}{0.7}
\setval{4b.gdpo.minus.geoma.minerva}{$-$1.5}
\setval{4b.gdpo.minus.geoma.minerva-abs}{1.5}
\setval{4b.gdpo.minus.geoma.olympiad-bench}{+0.8}
\setval{4b.gdpo.minus.geoma.olympiad-bench-abs}{0.8}
\setval{4b.gdpo.minus.geoma.writing}{+1.9}
\setval{4b.gdpo.minus.geoma.writing-abs}{1.9}
\setval{4b.gdpo.minus.geoma.writingbench}{+2.4}
\setval{4b.gdpo.minus.geoma.writingbench-abs}{2.4}
\setval{4b.gdpo.minus.grpo-nodiv.aime24}{$-$3.3}
\setval{4b.gdpo.minus.grpo-nodiv.aime24-abs}{3.3}
\setval{4b.gdpo.minus.grpo-nodiv.aime25}{+1.6}
\setval{4b.gdpo.minus.grpo-nodiv.aime25-abs}{1.6}
\setval{4b.gdpo.minus.grpo-nodiv.amc23}{+5.1}
\setval{4b.gdpo.minus.grpo-nodiv.amc23-abs}{5.1}
\setval{4b.gdpo.minus.grpo-nodiv.avg}{+1.3}
\setval{4b.gdpo.minus.grpo-nodiv.avg-abs}{1.3}
\setval{4b.gdpo.minus.grpo-nodiv.code}{+0.3}
\setval{4b.gdpo.minus.grpo-nodiv.code-abs}{0.3}
\setval{4b.gdpo.minus.grpo-nodiv.creative-writing-v3}{+3.4}
\setval{4b.gdpo.minus.grpo-nodiv.creative-writing-v3-abs}{3.4}
\setval{4b.gdpo.minus.grpo-nodiv.guidebench}{+1.8}
\setval{4b.gdpo.minus.grpo-nodiv.guidebench-abs}{1.8}
\setval{4b.gdpo.minus.grpo-nodiv.humaneval-plus}{+1.3}
\setval{4b.gdpo.minus.grpo-nodiv.humaneval-plus-abs}{1.3}
\setval{4b.gdpo.minus.grpo-nodiv.if}{+1.0}
\setval{4b.gdpo.minus.grpo-nodiv.if-abs}{1.0}
\setval{4b.gdpo.minus.grpo-nodiv.ifbench}{+2.7}
\setval{4b.gdpo.minus.grpo-nodiv.ifbench-abs}{2.7}
\setval{4b.gdpo.minus.grpo-nodiv.ifeval}{$-$1.6}
\setval{4b.gdpo.minus.grpo-nodiv.ifeval-abs}{1.6}
\setval{4b.gdpo.minus.grpo-nodiv.judge2.avg}{+1.2}
\setval{4b.gdpo.minus.grpo-nodiv.judge2.avg-abs}{1.2}
\setval{4b.gdpo.minus.grpo-nodiv.judge2.creative-writing-v3}{+3.9}
\setval{4b.gdpo.minus.grpo-nodiv.judge2.creative-writing-v3-abs}{3.9}
\setval{4b.gdpo.minus.grpo-nodiv.judge2.longbench-write}{+3.7}
\setval{4b.gdpo.minus.grpo-nodiv.judge2.longbench-write-abs}{3.7}
\setval{4b.gdpo.minus.grpo-nodiv.judge2.writing}{+4.0}
\setval{4b.gdpo.minus.grpo-nodiv.judge2.writing-abs}{4.0}
\setval{4b.gdpo.minus.grpo-nodiv.judge2.writingbench}{+4.5}
\setval{4b.gdpo.minus.grpo-nodiv.judge2.writingbench-abs}{4.5}
\setval{4b.gdpo.minus.grpo-nodiv.livecodebench}{$-$0.1}
\setval{4b.gdpo.minus.grpo-nodiv.livecodebench-abs}{0.1}
\setval{4b.gdpo.minus.grpo-nodiv.longbench-write}{+3.0}
\setval{4b.gdpo.minus.grpo-nodiv.longbench-write-abs}{3.0}
\setval{4b.gdpo.minus.grpo-nodiv.math}{+0.4}
\setval{4b.gdpo.minus.grpo-nodiv.math-abs}{0.4}
\setval{4b.gdpo.minus.grpo-nodiv.math500}{$-$0.2}
\setval{4b.gdpo.minus.grpo-nodiv.math500-abs}{0.2}
\setval{4b.gdpo.minus.grpo-nodiv.mbpp-plus}{$-$0.3}
\setval{4b.gdpo.minus.grpo-nodiv.mbpp-plus-abs}{0.3}
\setval{4b.gdpo.minus.grpo-nodiv.minerva}{$-$0.3}
\setval{4b.gdpo.minus.grpo-nodiv.minerva-abs}{0.3}
\setval{4b.gdpo.minus.grpo-nodiv.olympiad-bench}{$-$0.1}
\setval{4b.gdpo.minus.grpo-nodiv.olympiad-bench-abs}{0.1}
\setval{4b.gdpo.minus.grpo-nodiv.writing}{+4.2}
\setval{4b.gdpo.minus.grpo-nodiv.writing-abs}{4.2}
\setval{4b.gdpo.minus.grpo-nodiv.writingbench}{+6.3}
\setval{4b.gdpo.minus.grpo-nodiv.writingbench-abs}{6.3}
\setval{4b.gdpo.minus.grpo.aime24}{$-$2.9}
\setval{4b.gdpo.minus.grpo.aime24-abs}{2.9}
\setval{4b.gdpo.minus.grpo.aime25}{$-$2.7}
\setval{4b.gdpo.minus.grpo.aime25-abs}{2.7}
\setval{4b.gdpo.minus.grpo.amc23}{+4.9}
\setval{4b.gdpo.minus.grpo.amc23-abs}{4.9}
\setval{4b.gdpo.minus.grpo.avg}{$-$0.4}
\setval{4b.gdpo.minus.grpo.avg-abs}{0.4}
\setval{4b.gdpo.minus.grpo.code}{$-$0.3}
\setval{4b.gdpo.minus.grpo.code-abs}{0.3}
\setval{4b.gdpo.minus.grpo.creative-writing-v3}{$-$1.4}
\setval{4b.gdpo.minus.grpo.creative-writing-v3-abs}{1.4}
\setval{4b.gdpo.minus.grpo.guidebench}{+2.7}
\setval{4b.gdpo.minus.grpo.guidebench-abs}{2.7}
\setval{4b.gdpo.minus.grpo.humaneval-plus}{$-$0.9}
\setval{4b.gdpo.minus.grpo.humaneval-plus-abs}{0.9}
\setval{4b.gdpo.minus.grpo.if}{+0.2}
\setval{4b.gdpo.minus.grpo.if-abs}{0.2}
\setval{4b.gdpo.minus.grpo.ifbench}{$-$1.3}
\setval{4b.gdpo.minus.grpo.ifbench-abs}{1.3}
\setval{4b.gdpo.minus.grpo.ifeval}{$-$0.7}
\setval{4b.gdpo.minus.grpo.ifeval-abs}{0.7}
\setval{4b.gdpo.minus.grpo.judge2.avg}{+0.0}
\setval{4b.gdpo.minus.grpo.judge2.avg-abs}{0.0}
\setval{4b.gdpo.minus.grpo.judge2.creative-writing-v3}{$-$0.7}
\setval{4b.gdpo.minus.grpo.judge2.creative-writing-v3-abs}{0.7}
\setval{4b.gdpo.minus.grpo.judge2.longbench-write}{+1.6}
\setval{4b.gdpo.minus.grpo.judge2.longbench-write-abs}{1.6}
\setval{4b.gdpo.minus.grpo.judge2.writing}{+0.5}
\setval{4b.gdpo.minus.grpo.judge2.writing-abs}{0.5}
\setval{4b.gdpo.minus.grpo.judge2.writingbench}{+0.6}
\setval{4b.gdpo.minus.grpo.judge2.writingbench-abs}{0.6}
\setval{4b.gdpo.minus.grpo.livecodebench}{+0.4}
\setval{4b.gdpo.minus.grpo.livecodebench-abs}{0.4}
\setval{4b.gdpo.minus.grpo.longbench-write}{$-$0.2}
\setval{4b.gdpo.minus.grpo.longbench-write-abs}{0.2}
\setval{4b.gdpo.minus.grpo.math}{$-$0.2}
\setval{4b.gdpo.minus.grpo.math-abs}{0.2}
\setval{4b.gdpo.minus.grpo.math500}{$-$1.0}
\setval{4b.gdpo.minus.grpo.math500-abs}{1.0}
\setval{4b.gdpo.minus.grpo.mbpp-plus}{$-$0.3}
\setval{4b.gdpo.minus.grpo.mbpp-plus-abs}{0.3}
\setval{4b.gdpo.minus.grpo.minerva}{+0.2}
\setval{4b.gdpo.minus.grpo.minerva-abs}{0.2}
\setval{4b.gdpo.minus.grpo.olympiad-bench}{+0.0}
\setval{4b.gdpo.minus.grpo.olympiad-bench-abs}{0.0}
\setval{4b.gdpo.minus.grpo.writing}{$-$1.3}
\setval{4b.gdpo.minus.grpo.writing-abs}{1.3}
\setval{4b.gdpo.minus.grpo.writingbench}{$-$2.3}
\setval{4b.gdpo.minus.grpo.writingbench-abs}{2.3}
\setval{4b.gdpo.minus.mt-grpo.aime24}{$-$1.7}
\setval{4b.gdpo.minus.mt-grpo.aime24-abs}{1.7}
\setval{4b.gdpo.minus.mt-grpo.aime25}{+3.7}
\setval{4b.gdpo.minus.mt-grpo.aime25-abs}{3.7}
\setval{4b.gdpo.minus.mt-grpo.amc23}{$-$0.2}
\setval{4b.gdpo.minus.mt-grpo.amc23-abs}{0.2}
\setval{4b.gdpo.minus.mt-grpo.avg}{+1.2}
\setval{4b.gdpo.minus.mt-grpo.avg-abs}{1.2}
\setval{4b.gdpo.minus.mt-grpo.code}{+1.8}
\setval{4b.gdpo.minus.mt-grpo.code-abs}{1.8}
\setval{4b.gdpo.minus.mt-grpo.creative-writing-v3}{+1.5}
\setval{4b.gdpo.minus.mt-grpo.creative-writing-v3-abs}{1.5}
\setval{4b.gdpo.minus.mt-grpo.guidebench}{+1.0}
\setval{4b.gdpo.minus.mt-grpo.guidebench-abs}{1.0}
\setval{4b.gdpo.minus.mt-grpo.humaneval-plus}{+2.7}
\setval{4b.gdpo.minus.mt-grpo.humaneval-plus-abs}{2.7}
\setval{4b.gdpo.minus.mt-grpo.if}{+0.7}
\setval{4b.gdpo.minus.mt-grpo.if-abs}{0.7}
\setval{4b.gdpo.minus.mt-grpo.ifbench}{+2.0}
\setval{4b.gdpo.minus.mt-grpo.ifbench-abs}{2.0}
\setval{4b.gdpo.minus.mt-grpo.ifeval}{$-$1.0}
\setval{4b.gdpo.minus.mt-grpo.ifeval-abs}{1.0}
\setval{4b.gdpo.minus.mt-grpo.judge2.avg}{+1.1}
\setval{4b.gdpo.minus.mt-grpo.judge2.avg-abs}{1.1}
\setval{4b.gdpo.minus.mt-grpo.judge2.creative-writing-v3}{+0.3}
\setval{4b.gdpo.minus.mt-grpo.judge2.creative-writing-v3-abs}{0.3}
\setval{4b.gdpo.minus.mt-grpo.judge2.longbench-write}{+1.5}
\setval{4b.gdpo.minus.mt-grpo.judge2.longbench-write-abs}{1.5}
\setval{4b.gdpo.minus.mt-grpo.judge2.writing}{+0.9}
\setval{4b.gdpo.minus.mt-grpo.judge2.writing-abs}{0.9}
\setval{4b.gdpo.minus.mt-grpo.judge2.writingbench}{+0.8}
\setval{4b.gdpo.minus.mt-grpo.judge2.writingbench-abs}{0.8}
\setval{4b.gdpo.minus.mt-grpo.livecodebench}{+1.8}
\setval{4b.gdpo.minus.mt-grpo.livecodebench-abs}{1.8}
\setval{4b.gdpo.minus.mt-grpo.longbench-write}{+2.2}
\setval{4b.gdpo.minus.mt-grpo.longbench-write-abs}{2.2}
\setval{4b.gdpo.minus.mt-grpo.math}{+1.0}
\setval{4b.gdpo.minus.mt-grpo.math-abs}{1.0}
\setval{4b.gdpo.minus.mt-grpo.math500}{+2.3}
\setval{4b.gdpo.minus.mt-grpo.math500-abs}{2.3}
\setval{4b.gdpo.minus.mt-grpo.mbpp-plus}{+0.8}
\setval{4b.gdpo.minus.mt-grpo.mbpp-plus-abs}{0.8}
\setval{4b.gdpo.minus.mt-grpo.minerva}{$-$0.5}
\setval{4b.gdpo.minus.mt-grpo.minerva-abs}{0.5}
\setval{4b.gdpo.minus.mt-grpo.olympiad-bench}{+2.2}
\setval{4b.gdpo.minus.mt-grpo.olympiad-bench-abs}{2.2}
\setval{4b.gdpo.minus.mt-grpo.writing}{+1.8}
\setval{4b.gdpo.minus.mt-grpo.writing-abs}{1.8}
\setval{4b.gdpo.minus.mt-grpo.writingbench}{+1.8}
\setval{4b.gdpo.minus.mt-grpo.writingbench-abs}{1.8}
\setval{4b.gdpo.minus.papo.aime24}{$-$0.2}
\setval{4b.gdpo.minus.papo.aime24-abs}{0.2}
\setval{4b.gdpo.minus.papo.aime25}{+2.4}
\setval{4b.gdpo.minus.papo.aime25-abs}{2.4}
\setval{4b.gdpo.minus.papo.amc23}{+6.1}
\setval{4b.gdpo.minus.papo.amc23-abs}{6.1}
\setval{4b.gdpo.minus.papo.avg}{+1.7}
\setval{4b.gdpo.minus.papo.avg-abs}{1.7}
\setval{4b.gdpo.minus.papo.code}{$-$0.1}
\setval{4b.gdpo.minus.papo.code-abs}{0.1}
\setval{4b.gdpo.minus.papo.creative-writing-v3}{$-$0.6}
\setval{4b.gdpo.minus.papo.creative-writing-v3-abs}{0.6}
\setval{4b.gdpo.minus.papo.guidebench}{$-$0.1}
\setval{4b.gdpo.minus.papo.guidebench-abs}{0.1}
\setval{4b.gdpo.minus.papo.humaneval-plus}{$-$1.2}
\setval{4b.gdpo.minus.papo.humaneval-plus-abs}{1.2}
\setval{4b.gdpo.minus.papo.if}{+0.9}
\setval{4b.gdpo.minus.papo.if-abs}{0.9}
\setval{4b.gdpo.minus.papo.ifbench}{+3.0}
\setval{4b.gdpo.minus.papo.ifbench-abs}{3.0}
\setval{4b.gdpo.minus.papo.ifeval}{$-$0.1}
\setval{4b.gdpo.minus.papo.ifeval-abs}{0.1}
\setval{4b.gdpo.minus.papo.judge2.avg}{+1.5}
\setval{4b.gdpo.minus.papo.judge2.avg-abs}{1.5}
\setval{4b.gdpo.minus.papo.judge2.creative-writing-v3}{+0.2}
\setval{4b.gdpo.minus.papo.judge2.creative-writing-v3-abs}{0.2}
\setval{4b.gdpo.minus.papo.judge2.longbench-write}{+4.8}
\setval{4b.gdpo.minus.papo.judge2.longbench-write-abs}{4.8}
\setval{4b.gdpo.minus.papo.judge2.writing}{+2.3}
\setval{4b.gdpo.minus.papo.judge2.writing-abs}{2.3}
\setval{4b.gdpo.minus.papo.judge2.writingbench}{+1.9}
\setval{4b.gdpo.minus.papo.judge2.writingbench-abs}{1.9}
\setval{4b.gdpo.minus.papo.livecodebench}{+0.9}
\setval{4b.gdpo.minus.papo.livecodebench-abs}{0.9}
\setval{4b.gdpo.minus.papo.longbench-write}{+5.0}
\setval{4b.gdpo.minus.papo.longbench-write-abs}{5.0}
\setval{4b.gdpo.minus.papo.math}{+2.2}
\setval{4b.gdpo.minus.papo.math-abs}{2.2}
\setval{4b.gdpo.minus.papo.math500}{+2.8}
\setval{4b.gdpo.minus.papo.math500-abs}{2.8}
\setval{4b.gdpo.minus.papo.mbpp-plus}{+0.2}
\setval{4b.gdpo.minus.papo.mbpp-plus-abs}{0.2}
\setval{4b.gdpo.minus.papo.minerva}{$-$1.0}
\setval{4b.gdpo.minus.papo.minerva-abs}{1.0}
\setval{4b.gdpo.minus.papo.olympiad-bench}{+3.2}
\setval{4b.gdpo.minus.papo.olympiad-bench-abs}{3.2}
\setval{4b.gdpo.minus.papo.writing}{+3.3}
\setval{4b.gdpo.minus.papo.writing-abs}{3.3}
\setval{4b.gdpo.minus.papo.writingbench}{+5.4}
\setval{4b.gdpo.minus.papo.writingbench-abs}{5.4}
\setval{4b.gdpo.minus.saw.aime24}{$-$3.4}
\setval{4b.gdpo.minus.saw.aime24-abs}{3.4}
\setval{4b.gdpo.minus.saw.aime25}{$-$1.3}
\setval{4b.gdpo.minus.saw.aime25-abs}{1.3}
\setval{4b.gdpo.minus.saw.amc23}{+3.0}
\setval{4b.gdpo.minus.saw.amc23-abs}{3.0}
\setval{4b.gdpo.minus.saw.avg}{+0.1}
\setval{4b.gdpo.minus.saw.avg-abs}{0.1}
\setval{4b.gdpo.minus.saw.code}{+1.2}
\setval{4b.gdpo.minus.saw.code-abs}{1.2}
\setval{4b.gdpo.minus.saw.creative-writing-v3}{+0.9}
\setval{4b.gdpo.minus.saw.creative-writing-v3-abs}{0.9}
\setval{4b.gdpo.minus.saw.guidebench}{$-$0.4}
\setval{4b.gdpo.minus.saw.guidebench-abs}{0.4}
\setval{4b.gdpo.minus.saw.humaneval-plus}{+1.1}
\setval{4b.gdpo.minus.saw.humaneval-plus-abs}{1.1}
\setval{4b.gdpo.minus.saw.if}{$-$1.9}
\setval{4b.gdpo.minus.saw.if-abs}{1.9}
\setval{4b.gdpo.minus.saw.ifbench}{$-$4.0}
\setval{4b.gdpo.minus.saw.ifbench-abs}{4.0}
\setval{4b.gdpo.minus.saw.ifeval}{$-$1.2}
\setval{4b.gdpo.minus.saw.ifeval-abs}{1.2}
\setval{4b.gdpo.minus.saw.judge2.avg}{$-$0.2}
\setval{4b.gdpo.minus.saw.judge2.avg-abs}{0.2}
\setval{4b.gdpo.minus.saw.judge2.creative-writing-v3}{$-$1.2}
\setval{4b.gdpo.minus.saw.judge2.creative-writing-v3-abs}{1.2}
\setval{4b.gdpo.minus.saw.judge2.longbench-write}{+2.3}
\setval{4b.gdpo.minus.saw.judge2.longbench-write-abs}{2.3}
\setval{4b.gdpo.minus.saw.judge2.writing}{+0.6}
\setval{4b.gdpo.minus.saw.judge2.writing-abs}{0.6}
\setval{4b.gdpo.minus.saw.judge2.writingbench}{+0.6}
\setval{4b.gdpo.minus.saw.judge2.writingbench-abs}{0.6}
\setval{4b.gdpo.minus.saw.livecodebench}{+1.6}
\setval{4b.gdpo.minus.saw.livecodebench-abs}{1.6}
\setval{4b.gdpo.minus.saw.longbench-write}{+2.8}
\setval{4b.gdpo.minus.saw.longbench-write-abs}{2.8}
\setval{4b.gdpo.minus.saw.math}{$-$0.4}
\setval{4b.gdpo.minus.saw.math-abs}{0.4}
\setval{4b.gdpo.minus.saw.math500}{+1.4}
\setval{4b.gdpo.minus.saw.math500-abs}{1.4}
\setval{4b.gdpo.minus.saw.mbpp-plus}{+0.8}
\setval{4b.gdpo.minus.saw.mbpp-plus-abs}{0.8}
\setval{4b.gdpo.minus.saw.minerva}{$-$2.7}
\setval{4b.gdpo.minus.saw.minerva-abs}{2.7}
\setval{4b.gdpo.minus.saw.olympiad-bench}{+0.8}
\setval{4b.gdpo.minus.saw.olympiad-bench-abs}{0.8}
\setval{4b.gdpo.minus.saw.writing}{+1.9}
\setval{4b.gdpo.minus.saw.writing-abs}{1.9}
\setval{4b.gdpo.minus.saw.writingbench}{+2.1}
\setval{4b.gdpo.minus.saw.writingbench-abs}{2.1}
\setval{4b.gdpo.passing-at-ceiling.code.early}{0.70}
\setval{4b.gdpo.passing-at-ceiling.code.late}{0.74}
\setval{4b.gdpo.passing-at-ceiling.code.run}{0.73}
\setval{4b.gdpo.passing-at-ceiling.if.early}{0.29}
\setval{4b.gdpo.passing-at-ceiling.if.late}{0.57}
\setval{4b.gdpo.passing-at-ceiling.if.run}{0.48}
\setval{4b.gdpo.passing-at-ceiling.math.early}{1.00}
\setval{4b.gdpo.passing-at-ceiling.math.late}{1.00}
\setval{4b.gdpo.passing-at-ceiling.math.run}{1.00}
\setval{4b.gdpo.passing-at-ceiling.writing.early}{0.00}
\setval{4b.gdpo.passing-at-ceiling.writing.late}{0.00}
\setval{4b.gdpo.passing-at-ceiling.writing.run}{0.00}
\setval{4b.gdpo.primary-cosine.code.early}{0.96}
\setval{4b.gdpo.primary-cosine.code.late}{0.95}
\setval{4b.gdpo.primary-cosine.code.run}{0.96}
\setval{4b.gdpo.primary-cosine.if.early}{0.84}
\setval{4b.gdpo.primary-cosine.if.late}{0.78}
\setval{4b.gdpo.primary-cosine.if.run}{0.81}
\setval{4b.gdpo.primary-cosine.math.early}{0.97}
\setval{4b.gdpo.primary-cosine.math.late}{0.97}
\setval{4b.gdpo.primary-cosine.math.run}{0.97}
\setval{4b.gdpo.primary-cosine.writing.early}{0.86}
\setval{4b.gdpo.primary-cosine.writing.late}{0.75}
\setval{4b.gdpo.primary-cosine.writing.run}{0.75}
\setval{4b.gdpo.primary-level.code.early}{0.53}
\setval{4b.gdpo.primary-level.code.late}{0.75}
\setval{4b.gdpo.primary-level.code.run}{0.70}
\setval{4b.gdpo.primary-level.if.early}{0.32}
\setval{4b.gdpo.primary-level.if.late}{0.76}
\setval{4b.gdpo.primary-level.if.run}{0.64}
\setval{4b.gdpo.primary-level.math.early}{0.48}
\setval{4b.gdpo.primary-level.math.late}{0.68}
\setval{4b.gdpo.primary-level.math.run}{0.62}
\setval{4b.gdpo.primary-level.writing.early}{0.53}
\setval{4b.gdpo.primary-level.writing.late}{0.75}
\setval{4b.gdpo.primary-level.writing.run}{0.71}
\setval{4b.gdpo.solved.code.early}{0.05}
\setval{4b.gdpo.solved.code.late}{0.02}
\setval{4b.gdpo.solved.code.run}{0.03}
\setval{4b.gdpo.solved.if.early}{0.03}
\setval{4b.gdpo.solved.if.late}{0.20}
\setval{4b.gdpo.solved.if.run}{0.13}
\setval{4b.gdpo.solved.math.early}{0.13}
\setval{4b.gdpo.solved.math.late}{0.16}
\setval{4b.gdpo.solved.math.run}{0.14}
\setval{4b.gdpo.solved.writing.early}{0.00}
\setval{4b.gdpo.solved.writing.late}{0.00}
\setval{4b.gdpo.solved.writing.run}{0.00}
\setval{4b.gdpo.update-share.code.early}{0.20}
\setval{4b.gdpo.update-share.code.late}{0.23}
\setval{4b.gdpo.update-share.code.run}{0.22}
\setval{4b.gdpo.update-share.if.early}{0.21}
\setval{4b.gdpo.update-share.if.late}{0.16}
\setval{4b.gdpo.update-share.if.run}{0.19}
\setval{4b.gdpo.update-share.math.early}{0.18}
\setval{4b.gdpo.update-share.math.late}{0.27}
\setval{4b.gdpo.update-share.math.run}{0.25}
\setval{4b.gdpo.update-share.writing.early}{0.42}
\setval{4b.gdpo.update-share.writing.late}{0.35}
\setval{4b.gdpo.update-share.writing.run}{0.34}
\setval{4b.geoma.aux-change.code}{+0.20}
\setval{4b.geoma.aux-change.if}{+0.24}
\setval{4b.geoma.aux-change.math}{+0.14}
\setval{4b.geoma.aux-change.writing}{+0.20}
\setval{4b.geoma.aux-change.writing-rm}{+0.37}
\setval{4b.geoma.aux-full.code.early}{0.80}
\setval{4b.geoma.aux-full.code.late}{0.98}
\setval{4b.geoma.aux-full.code.run}{0.96}
\setval{4b.geoma.aux-full.if.early}{0.67}
\setval{4b.geoma.aux-full.if.late}{0.91}
\setval{4b.geoma.aux-full.if.run}{0.87}
\setval{4b.geoma.aux-full.math.early}{0.66}
\setval{4b.geoma.aux-full.math.late}{0.94}
\setval{4b.geoma.aux-full.math.run}{0.91}
\setval{4b.geoma.aux-full.writing-rm.early}{0.09}
\setval{4b.geoma.aux-full.writing-rm.late}{0.61}
\setval{4b.geoma.aux-full.writing-rm.run}{0.49}
\setval{4b.geoma.aux-full.writing.early}{0.66}
\setval{4b.geoma.aux-full.writing.late}{0.77}
\setval{4b.geoma.aux-full.writing.run}{0.75}
\setval{4b.geoma.aux-level.code.early}{0.94}
\setval{4b.geoma.aux-level.code.late}{1.00}
\setval{4b.geoma.aux-level.code.run}{0.99}
\setval{4b.geoma.aux-level.if.early}{0.84}
\setval{4b.geoma.aux-level.if.late}{0.95}
\setval{4b.geoma.aux-level.if.run}{0.93}
\setval{4b.geoma.aux-level.math.early}{0.94}
\setval{4b.geoma.aux-level.math.late}{0.99}
\setval{4b.geoma.aux-level.math.run}{0.98}
\setval{4b.geoma.aux-level.writing-rm.early}{0.77}
\setval{4b.geoma.aux-level.writing-rm.late}{0.96}
\setval{4b.geoma.aux-level.writing-rm.run}{0.94}
\setval{4b.geoma.aux-level.writing.early}{0.91}
\setval{4b.geoma.aux-level.writing.late}{0.97}
\setval{4b.geoma.aux-level.writing.run}{0.96}
\setval{4b.geoma.bench.aime24}{11.6}
\setval{4b.geoma.bench.aime25}{10.6}
\setval{4b.geoma.bench.amc23}{55.4}
\setval{4b.geoma.bench.creative-writing-v3}{45.9}
\setval{4b.geoma.bench.guidebench}{78.5}
\setval{4b.geoma.bench.humaneval-plus}{80.2}
\setval{4b.geoma.bench.ifbench}{36.0}
\setval{4b.geoma.bench.ifeval}{87.8}
\setval{4b.geoma.bench.livecodebench}{25.7}
\setval{4b.geoma.bench.longbench-write}{65.8}
\setval{4b.geoma.bench.math500}{78.8}
\setval{4b.geoma.bench.mbpp-plus}{70.2}
\setval{4b.geoma.bench.minerva}{46.9}
\setval{4b.geoma.bench.olympiad-bench}{43.8}
\setval{4b.geoma.bench.writingbench}{69.5}
\setval{4b.geoma.eval.avg}{53.8}
\setval{4b.geoma.eval.code}{58.7}
\setval{4b.geoma.eval.if}{67.4}
\setval{4b.geoma.eval.math}{41.2}
\setval{4b.geoma.eval.writing}{60.4}
\setval{4b.geoma.gate.code.against.early}{0.03}
\setval{4b.geoma.gate.code.against.late}{0.01}
\setval{4b.geoma.gate.code.against.run}{0.01}
\setval{4b.geoma.gate.code.agreement}{$-$0.001}
\setval{4b.geoma.gate.code.ceiling.early}{0.08}
\setval{4b.geoma.gate.code.ceiling.late}{0.03}
\setval{4b.geoma.gate.code.ceiling.run}{0.03}
\setval{4b.geoma.gate.code.cosine.early}{0.86}
\setval{4b.geoma.gate.code.cosine.late}{0.82}
\setval{4b.geoma.gate.code.cosine.run}{0.83}
\setval{4b.geoma.gate.if.against.early}{0.05}
\setval{4b.geoma.gate.if.against.late}{0.04}
\setval{4b.geoma.gate.if.against.run}{0.04}
\setval{4b.geoma.gate.if.agreement}{+0.011}
\setval{4b.geoma.gate.if.ceiling.early}{0.05}
\setval{4b.geoma.gate.if.ceiling.late}{0.16}
\setval{4b.geoma.gate.if.ceiling.run}{0.10}
\setval{4b.geoma.gate.if.cosine.early}{0.74}
\setval{4b.geoma.gate.if.cosine.late}{0.52}
\setval{4b.geoma.gate.if.cosine.run}{0.61}
\setval{4b.geoma.gate.math.against.early}{0.00}
\setval{4b.geoma.gate.math.against.late}{0.00}
\setval{4b.geoma.gate.math.against.run}{0.00}
\setval{4b.geoma.gate.math.ceiling.early}{0.18}
\setval{4b.geoma.gate.math.ceiling.late}{0.10}
\setval{4b.geoma.gate.math.ceiling.run}{0.10}
\setval{4b.geoma.gate.math.cosine.early}{0.89}
\setval{4b.geoma.gate.math.cosine.late}{0.94}
\setval{4b.geoma.gate.math.cosine.run}{0.94}
\setval{4b.geoma.gate.writing-rm.against.early}{0.20}
\setval{4b.geoma.gate.writing-rm.against.late}{0.51}
\setval{4b.geoma.gate.writing-rm.against.run}{0.47}
\setval{4b.geoma.gate.writing-rm.agreement}{+0.063}
\setval{4b.geoma.gate.writing-rm.ceiling.early}{0.00}
\setval{4b.geoma.gate.writing-rm.ceiling.late}{0.00}
\setval{4b.geoma.gate.writing-rm.ceiling.run}{0.00}
\setval{4b.geoma.gate.writing-rm.cosine.early}{0.69}
\setval{4b.geoma.gate.writing-rm.cosine.late}{0.14}
\setval{4b.geoma.gate.writing-rm.cosine.run}{0.24}
\setval{4b.geoma.gate.writing.against.early}{0.28}
\setval{4b.geoma.gate.writing.against.late}{0.59}
\setval{4b.geoma.gate.writing.against.run}{0.53}
\setval{4b.geoma.gate.writing.agreement}{$-$0.036}
\setval{4b.geoma.gate.writing.ceiling.early}{0.00}
\setval{4b.geoma.gate.writing.ceiling.late}{0.00}
\setval{4b.geoma.gate.writing.ceiling.run}{0.00}
\setval{4b.geoma.gate.writing.cosine.early}{0.52}
\setval{4b.geoma.gate.writing.cosine.late}{0.05}
\setval{4b.geoma.gate.writing.cosine.run}{0.13}
\setval{4b.geoma.judge2.avg}{51.1}
\setval{4b.geoma.judge2.creative-writing-v3}{33.1}
\setval{4b.geoma.judge2.longbench-write}{57.7}
\setval{4b.geoma.judge2.writing}{47.2}
\setval{4b.geoma.judge2.writingbench}{50.7}
\setval{4b.geoma.last-step}{300}
\setval{4b.geoma.minus.base.aime24}{+3.8}
\setval{4b.geoma.minus.base.aime24-abs}{3.8}
\setval{4b.geoma.minus.base.aime25}{+3.5}
\setval{4b.geoma.minus.base.aime25-abs}{3.5}
\setval{4b.geoma.minus.base.amc23}{+11.5}
\setval{4b.geoma.minus.base.amc23-abs}{11.5}
\setval{4b.geoma.minus.base.avg}{+19.3}
\setval{4b.geoma.minus.base.avg-abs}{19.3}
\setval{4b.geoma.minus.base.code}{+14.1}
\setval{4b.geoma.minus.base.code-abs}{14.1}
\setval{4b.geoma.minus.base.creative-writing-v3}{+21.1}
\setval{4b.geoma.minus.base.creative-writing-v3-abs}{21.1}
\setval{4b.geoma.minus.base.guidebench}{+53.6}
\setval{4b.geoma.minus.base.guidebench-abs}{53.6}
\setval{4b.geoma.minus.base.humaneval-plus}{+15.2}
\setval{4b.geoma.minus.base.humaneval-plus-abs}{15.2}
\setval{4b.geoma.minus.base.if}{+37.7}
\setval{4b.geoma.minus.base.if-abs}{37.7}
\setval{4b.geoma.minus.base.ifbench}{+19.0}
\setval{4b.geoma.minus.base.ifbench-abs}{19.0}
\setval{4b.geoma.minus.base.ifeval}{+40.4}
\setval{4b.geoma.minus.base.ifeval-abs}{40.4}
\setval{4b.geoma.minus.base.judge2.avg}{+17.2}
\setval{4b.geoma.minus.base.judge2.avg-abs}{17.2}
\setval{4b.geoma.minus.base.judge2.creative-writing-v3}{+12.3}
\setval{4b.geoma.minus.base.judge2.creative-writing-v3-abs}{12.3}
\setval{4b.geoma.minus.base.judge2.longbench-write}{+16.7}
\setval{4b.geoma.minus.base.judge2.longbench-write-abs}{16.7}
\setval{4b.geoma.minus.base.judge2.writing}{+16.4}
\setval{4b.geoma.minus.base.judge2.writing-abs}{16.4}
\setval{4b.geoma.minus.base.judge2.writingbench}{+20.2}
\setval{4b.geoma.minus.base.judge2.writingbench-abs}{20.2}
\setval{4b.geoma.minus.base.livecodebench}{+12.5}
\setval{4b.geoma.minus.base.livecodebench-abs}{12.5}
\setval{4b.geoma.minus.base.longbench-write}{+21.4}
\setval{4b.geoma.minus.base.longbench-write-abs}{21.4}
\setval{4b.geoma.minus.base.math}{+9.0}
\setval{4b.geoma.minus.base.math-abs}{9.0}
\setval{4b.geoma.minus.base.math500}{+11.6}
\setval{4b.geoma.minus.base.math500-abs}{11.6}
\setval{4b.geoma.minus.base.mbpp-plus}{+14.6}
\setval{4b.geoma.minus.base.mbpp-plus-abs}{14.6}
\setval{4b.geoma.minus.base.minerva}{+15.3}
\setval{4b.geoma.minus.base.minerva-abs}{15.3}
\setval{4b.geoma.minus.base.olympiad-bench}{+8.2}
\setval{4b.geoma.minus.base.olympiad-bench-abs}{8.2}
\setval{4b.geoma.minus.base.writing}{+26.5}
\setval{4b.geoma.minus.base.writing-abs}{26.5}
\setval{4b.geoma.minus.base.writingbench}{+37.1}
\setval{4b.geoma.minus.base.writingbench-abs}{37.1}
\setval{4b.geoma.minus.concur-free.aime24}{$-$0.3}
\setval{4b.geoma.minus.concur-free.aime24-abs}{0.3}
\setval{4b.geoma.minus.concur-free.aime25}{$-$4.0}
\setval{4b.geoma.minus.concur-free.aime25-abs}{4.0}
\setval{4b.geoma.minus.concur-free.amc23}{+1.3}
\setval{4b.geoma.minus.concur-free.amc23-abs}{1.3}
\setval{4b.geoma.minus.concur-free.avg}{$-$1.7}
\setval{4b.geoma.minus.concur-free.avg-abs}{1.7}
\setval{4b.geoma.minus.concur-free.code}{+0.4}
\setval{4b.geoma.minus.concur-free.code-abs}{0.4}
\setval{4b.geoma.minus.concur-free.creative-writing-v3}{+0.4}
\setval{4b.geoma.minus.concur-free.creative-writing-v3-abs}{0.4}
\setval{4b.geoma.minus.concur-free.guidebench}{$-$0.5}
\setval{4b.geoma.minus.concur-free.guidebench-abs}{0.5}
\setval{4b.geoma.minus.concur-free.humaneval-plus}{$-$0.9}
\setval{4b.geoma.minus.concur-free.humaneval-plus-abs}{0.9}
\setval{4b.geoma.minus.concur-free.if}{$-$0.8}
\setval{4b.geoma.minus.concur-free.if-abs}{0.8}
\setval{4b.geoma.minus.concur-free.ifbench}{$-$3.0}
\setval{4b.geoma.minus.concur-free.ifbench-abs}{3.0}
\setval{4b.geoma.minus.concur-free.ifeval}{+1.1}
\setval{4b.geoma.minus.concur-free.ifeval-abs}{1.1}
\setval{4b.geoma.minus.concur-free.judge2.avg}{$-$1.6}
\setval{4b.geoma.minus.concur-free.judge2.avg-abs}{1.6}
\setval{4b.geoma.minus.concur-free.judge2.creative-writing-v3}{$-$2.2}
\setval{4b.geoma.minus.concur-free.judge2.creative-writing-v3-abs}{2.2}
\setval{4b.geoma.minus.concur-free.judge2.longbench-write}{$-$11.7}
\setval{4b.geoma.minus.concur-free.judge2.longbench-write-abs}{11.7}
\setval{4b.geoma.minus.concur-free.judge2.writing}{$-$5.9}
\setval{4b.geoma.minus.concur-free.judge2.writing-abs}{5.9}
\setval{4b.geoma.minus.concur-free.judge2.writingbench}{$-$3.8}
\setval{4b.geoma.minus.concur-free.judge2.writingbench-abs}{3.8}
\setval{4b.geoma.minus.concur-free.livecodebench}{+1.5}
\setval{4b.geoma.minus.concur-free.livecodebench-abs}{1.5}
\setval{4b.geoma.minus.concur-free.longbench-write}{$-$12.0}
\setval{4b.geoma.minus.concur-free.longbench-write-abs}{12.0}
\setval{4b.geoma.minus.concur-free.math}{$-$0.9}
\setval{4b.geoma.minus.concur-free.math-abs}{0.9}
\setval{4b.geoma.minus.concur-free.math500}{$-$1.5}
\setval{4b.geoma.minus.concur-free.math500-abs}{1.5}
\setval{4b.geoma.minus.concur-free.mbpp-plus}{+0.6}
\setval{4b.geoma.minus.concur-free.mbpp-plus-abs}{0.6}
\setval{4b.geoma.minus.concur-free.minerva}{+0.7}
\setval{4b.geoma.minus.concur-free.minerva-abs}{0.7}
\setval{4b.geoma.minus.concur-free.olympiad-bench}{$-$1.9}
\setval{4b.geoma.minus.concur-free.olympiad-bench-abs}{1.9}
\setval{4b.geoma.minus.concur-free.writing}{$-$6.4}
\setval{4b.geoma.minus.concur-free.writing-abs}{6.4}
\setval{4b.geoma.minus.concur-free.writingbench}{$-$7.8}
\setval{4b.geoma.minus.concur-free.writingbench-abs}{7.8}
\setval{4b.geoma.minus.concur-tied.aime24}{$-$7.9}
\setval{4b.geoma.minus.concur-tied.aime24-abs}{7.9}
\setval{4b.geoma.minus.concur-tied.aime25}{$-$5.7}
\setval{4b.geoma.minus.concur-tied.aime25-abs}{5.7}
\setval{4b.geoma.minus.concur-tied.amc23}{$-$13.8}
\setval{4b.geoma.minus.concur-tied.amc23-abs}{13.8}
\setval{4b.geoma.minus.concur-tied.avg}{$-$4.9}
\setval{4b.geoma.minus.concur-tied.avg-abs}{4.9}
\setval{4b.geoma.minus.concur-tied.code}{$-$0.4}
\setval{4b.geoma.minus.concur-tied.code-abs}{0.4}
\setval{4b.geoma.minus.concur-tied.creative-writing-v3}{$-$2.0}
\setval{4b.geoma.minus.concur-tied.creative-writing-v3-abs}{2.0}
\setval{4b.geoma.minus.concur-tied.guidebench}{$-$0.7}
\setval{4b.geoma.minus.concur-tied.guidebench-abs}{0.7}
\setval{4b.geoma.minus.concur-tied.humaneval-plus}{$-$3.8}
\setval{4b.geoma.minus.concur-tied.humaneval-plus-abs}{3.8}
\setval{4b.geoma.minus.concur-tied.if}{$-$0.8}
\setval{4b.geoma.minus.concur-tied.if-abs}{0.8}
\setval{4b.geoma.minus.concur-tied.ifbench}{$-$1.7}
\setval{4b.geoma.minus.concur-tied.ifbench-abs}{1.7}
\setval{4b.geoma.minus.concur-tied.ifeval}{+0.0}
\setval{4b.geoma.minus.concur-tied.ifeval-abs}{0.0}
\setval{4b.geoma.minus.concur-tied.judge2.avg}{$-$4.3}
\setval{4b.geoma.minus.concur-tied.judge2.avg-abs}{4.3}
\setval{4b.geoma.minus.concur-tied.judge2.creative-writing-v3}{$-$0.9}
\setval{4b.geoma.minus.concur-tied.judge2.creative-writing-v3-abs}{0.9}
\setval{4b.geoma.minus.concur-tied.judge2.longbench-write}{$-$10.3}
\setval{4b.geoma.minus.concur-tied.judge2.longbench-write-abs}{10.3}
\setval{4b.geoma.minus.concur-tied.judge2.writing}{$-$5.3}
\setval{4b.geoma.minus.concur-tied.judge2.writing-abs}{5.3}
\setval{4b.geoma.minus.concur-tied.judge2.writingbench}{$-$4.6}
\setval{4b.geoma.minus.concur-tied.judge2.writingbench-abs}{4.6}
\setval{4b.geoma.minus.concur-tied.livecodebench}{+0.9}
\setval{4b.geoma.minus.concur-tied.livecodebench-abs}{0.9}
\setval{4b.geoma.minus.concur-tied.longbench-write}{$-$11.9}
\setval{4b.geoma.minus.concur-tied.longbench-write-abs}{11.9}
\setval{4b.geoma.minus.concur-tied.math}{$-$7.6}
\setval{4b.geoma.minus.concur-tied.math-abs}{7.6}
\setval{4b.geoma.minus.concur-tied.math500}{$-$6.7}
\setval{4b.geoma.minus.concur-tied.math500-abs}{6.7}
\setval{4b.geoma.minus.concur-tied.mbpp-plus}{+1.6}
\setval{4b.geoma.minus.concur-tied.mbpp-plus-abs}{1.6}
\setval{4b.geoma.minus.concur-tied.minerva}{$-$2.2}
\setval{4b.geoma.minus.concur-tied.minerva-abs}{2.2}
\setval{4b.geoma.minus.concur-tied.olympiad-bench}{$-$9.0}
\setval{4b.geoma.minus.concur-tied.olympiad-bench-abs}{9.0}
\setval{4b.geoma.minus.concur-tied.writing}{$-$8.2}
\setval{4b.geoma.minus.concur-tied.writing-abs}{8.2}
\setval{4b.geoma.minus.concur-tied.writingbench}{$-$10.7}
\setval{4b.geoma.minus.concur-tied.writingbench-abs}{10.7}
\setval{4b.geoma.minus.concur.aime24}{$-$9.0}
\setval{4b.geoma.minus.concur.aime24-abs}{9.0}
\setval{4b.geoma.minus.concur.aime25}{$-$8.1}
\setval{4b.geoma.minus.concur.aime25-abs}{8.1}
\setval{4b.geoma.minus.concur.amc23}{$-$14.0}
\setval{4b.geoma.minus.concur.amc23-abs}{14.0}
\setval{4b.geoma.minus.concur.avg}{$-$5.0}
\setval{4b.geoma.minus.concur.avg-abs}{5.0}
\setval{4b.geoma.minus.concur.code}{$-$1.0}
\setval{4b.geoma.minus.concur.code-abs}{1.0}
\setval{4b.geoma.minus.concur.creative-writing-v3}{+1.1}
\setval{4b.geoma.minus.concur.creative-writing-v3-abs}{1.1}
\setval{4b.geoma.minus.concur.guidebench}{+1.2}
\setval{4b.geoma.minus.concur.guidebench-abs}{1.2}
\setval{4b.geoma.minus.concur.humaneval-plus}{$-$2.2}
\setval{4b.geoma.minus.concur.humaneval-plus-abs}{2.2}
\setval{4b.geoma.minus.concur.if}{$-$1.2}
\setval{4b.geoma.minus.concur.if-abs}{1.2}
\setval{4b.geoma.minus.concur.ifbench}{$-$4.3}
\setval{4b.geoma.minus.concur.ifbench-abs}{4.3}
\setval{4b.geoma.minus.concur.ifeval}{$-$0.4}
\setval{4b.geoma.minus.concur.ifeval-abs}{0.4}
\setval{4b.geoma.minus.concur.judge2.avg}{$-$4.6}
\setval{4b.geoma.minus.concur.judge2.avg-abs}{4.6}
\setval{4b.geoma.minus.concur.judge2.creative-writing-v3}{+0.8}
\setval{4b.geoma.minus.concur.judge2.creative-writing-v3-abs}{0.8}
\setval{4b.geoma.minus.concur.judge2.longbench-write}{$-$9.0}
\setval{4b.geoma.minus.concur.judge2.longbench-write-abs}{9.0}
\setval{4b.geoma.minus.concur.judge2.writing}{$-$4.5}
\setval{4b.geoma.minus.concur.judge2.writing-abs}{4.5}
\setval{4b.geoma.minus.concur.judge2.writingbench}{$-$5.4}
\setval{4b.geoma.minus.concur.judge2.writingbench-abs}{5.4}
\setval{4b.geoma.minus.concur.livecodebench}{+0.5}
\setval{4b.geoma.minus.concur.livecodebench-abs}{0.5}
\setval{4b.geoma.minus.concur.longbench-write}{$-$10.5}
\setval{4b.geoma.minus.concur.longbench-write-abs}{10.5}
\setval{4b.geoma.minus.concur.math}{$-$8.3}
\setval{4b.geoma.minus.concur.math-abs}{8.3}
\setval{4b.geoma.minus.concur.math500}{$-$6.4}
\setval{4b.geoma.minus.concur.math500-abs}{6.4}
\setval{4b.geoma.minus.concur.mbpp-plus}{$-$1.3}
\setval{4b.geoma.minus.concur.mbpp-plus-abs}{1.3}
\setval{4b.geoma.minus.concur.minerva}{$-$2.3}
\setval{4b.geoma.minus.concur.minerva-abs}{2.3}
\setval{4b.geoma.minus.concur.olympiad-bench}{$-$9.7}
\setval{4b.geoma.minus.concur.olympiad-bench-abs}{9.7}
\setval{4b.geoma.minus.concur.writing}{$-$6.4}
\setval{4b.geoma.minus.concur.writing-abs}{6.4}
\setval{4b.geoma.minus.concur.writingbench}{$-$9.9}
\setval{4b.geoma.minus.concur.writingbench-abs}{9.9}
\setval{4b.geoma.minus.gd2po.aime24}{$-$0.3}
\setval{4b.geoma.minus.gd2po.aime24-abs}{0.3}
\setval{4b.geoma.minus.gd2po.aime25}{+3.9}
\setval{4b.geoma.minus.gd2po.aime25-abs}{3.9}
\setval{4b.geoma.minus.gd2po.amc23}{$-$2.3}
\setval{4b.geoma.minus.gd2po.amc23-abs}{2.3}
\setval{4b.geoma.minus.gd2po.avg}{$-$0.3}
\setval{4b.geoma.minus.gd2po.avg-abs}{0.3}
\setval{4b.geoma.minus.gd2po.code}{$-$0.7}
\setval{4b.geoma.minus.gd2po.code-abs}{0.7}
\setval{4b.geoma.minus.gd2po.creative-writing-v3}{+1.2}
\setval{4b.geoma.minus.gd2po.creative-writing-v3-abs}{1.2}
\setval{4b.geoma.minus.gd2po.guidebench}{$-$0.6}
\setval{4b.geoma.minus.gd2po.guidebench-abs}{0.6}
\setval{4b.geoma.minus.gd2po.humaneval-plus}{$-$3.2}
\setval{4b.geoma.minus.gd2po.humaneval-plus-abs}{3.2}
\setval{4b.geoma.minus.gd2po.if}{$-$0.7}
\setval{4b.geoma.minus.gd2po.if-abs}{0.7}
\setval{4b.geoma.minus.gd2po.ifbench}{$-$1.7}
\setval{4b.geoma.minus.gd2po.ifbench-abs}{1.7}
\setval{4b.geoma.minus.gd2po.ifeval}{+0.1}
\setval{4b.geoma.minus.gd2po.ifeval-abs}{0.1}
\setval{4b.geoma.minus.gd2po.judge2.avg}{$-$0.5}
\setval{4b.geoma.minus.gd2po.judge2.avg-abs}{0.5}
\setval{4b.geoma.minus.gd2po.judge2.creative-writing-v3}{+0.2}
\setval{4b.geoma.minus.gd2po.judge2.creative-writing-v3-abs}{0.2}
\setval{4b.geoma.minus.gd2po.judge2.longbench-write}{$-$3.3}
\setval{4b.geoma.minus.gd2po.judge2.longbench-write-abs}{3.3}
\setval{4b.geoma.minus.gd2po.judge2.writing}{$-$1.6}
\setval{4b.geoma.minus.gd2po.judge2.writing-abs}{1.6}
\setval{4b.geoma.minus.gd2po.judge2.writingbench}{$-$1.9}
\setval{4b.geoma.minus.gd2po.judge2.writingbench-abs}{1.9}
\setval{4b.geoma.minus.gd2po.livecodebench}{+2.2}
\setval{4b.geoma.minus.gd2po.livecodebench-abs}{2.2}
\setval{4b.geoma.minus.gd2po.longbench-write}{$-$3.1}
\setval{4b.geoma.minus.gd2po.longbench-write-abs}{3.1}
\setval{4b.geoma.minus.gd2po.math}{+0.4}
\setval{4b.geoma.minus.gd2po.math-abs}{0.4}
\setval{4b.geoma.minus.gd2po.math500}{+1.2}
\setval{4b.geoma.minus.gd2po.math500-abs}{1.2}
\setval{4b.geoma.minus.gd2po.mbpp-plus}{$-$1.2}
\setval{4b.geoma.minus.gd2po.mbpp-plus-abs}{1.2}
\setval{4b.geoma.minus.gd2po.minerva}{$-$0.2}
\setval{4b.geoma.minus.gd2po.minerva-abs}{0.2}
\setval{4b.geoma.minus.gd2po.olympiad-bench}{+0.3}
\setval{4b.geoma.minus.gd2po.olympiad-bench-abs}{0.3}
\setval{4b.geoma.minus.gd2po.writing}{$-$0.9}
\setval{4b.geoma.minus.gd2po.writing-abs}{0.9}
\setval{4b.geoma.minus.gd2po.writingbench}{$-$0.6}
\setval{4b.geoma.minus.gd2po.writingbench-abs}{0.6}
\setval{4b.geoma.minus.gdpo-noaux.aime24}{$-$11.9}
\setval{4b.geoma.minus.gdpo-noaux.aime24-abs}{11.9}
\setval{4b.geoma.minus.gdpo-noaux.aime25}{$-$8.1}
\setval{4b.geoma.minus.gdpo-noaux.aime25-abs}{8.1}
\setval{4b.geoma.minus.gdpo-noaux.amc23}{$-$10.3}
\setval{4b.geoma.minus.gdpo-noaux.amc23-abs}{10.3}
\setval{4b.geoma.minus.gdpo-noaux.avg}{$-$2.9}
\setval{4b.geoma.minus.gdpo-noaux.avg-abs}{2.9}
\setval{4b.geoma.minus.gdpo-noaux.code}{$-$2.1}
\setval{4b.geoma.minus.gdpo-noaux.code-abs}{2.1}
\setval{4b.geoma.minus.gdpo-noaux.creative-writing-v3}{+6.5}
\setval{4b.geoma.minus.gdpo-noaux.creative-writing-v3-abs}{6.5}
\setval{4b.geoma.minus.gdpo-noaux.guidebench}{+2.6}
\setval{4b.geoma.minus.gdpo-noaux.guidebench-abs}{2.6}
\setval{4b.geoma.minus.gdpo-noaux.humaneval-plus}{$-$5.2}
\setval{4b.geoma.minus.gdpo-noaux.humaneval-plus-abs}{5.2}
\setval{4b.geoma.minus.gdpo-noaux.if}{+0.4}
\setval{4b.geoma.minus.gdpo-noaux.if-abs}{0.4}
\setval{4b.geoma.minus.gdpo-noaux.ifbench}{$-$4.3}
\setval{4b.geoma.minus.gdpo-noaux.ifbench-abs}{4.3}
\setval{4b.geoma.minus.gdpo-noaux.ifeval}{+3.0}
\setval{4b.geoma.minus.gdpo-noaux.ifeval-abs}{3.0}
\setval{4b.geoma.minus.gdpo-noaux.judge2.avg}{$-$2.1}
\setval{4b.geoma.minus.gdpo-noaux.judge2.avg-abs}{2.1}
\setval{4b.geoma.minus.gdpo-noaux.judge2.creative-writing-v3}{+9.8}
\setval{4b.geoma.minus.gdpo-noaux.judge2.creative-writing-v3-abs}{9.8}
\setval{4b.geoma.minus.gdpo-noaux.judge2.longbench-write}{+6.9}
\setval{4b.geoma.minus.gdpo-noaux.judge2.longbench-write-abs}{6.9}
\setval{4b.geoma.minus.gdpo-noaux.judge2.writing}{+7.1}
\setval{4b.geoma.minus.gdpo-noaux.judge2.writing-abs}{7.1}
\setval{4b.geoma.minus.gdpo-noaux.judge2.writingbench}{+4.6}
\setval{4b.geoma.minus.gdpo-noaux.judge2.writingbench-abs}{4.6}
\setval{4b.geoma.minus.gdpo-noaux.livecodebench}{+0.9}
\setval{4b.geoma.minus.gdpo-noaux.livecodebench-abs}{0.9}
\setval{4b.geoma.minus.gdpo-noaux.longbench-write}{+1.5}
\setval{4b.geoma.minus.gdpo-noaux.longbench-write-abs}{1.5}
\setval{4b.geoma.minus.gdpo-noaux.math}{$-$8.0}
\setval{4b.geoma.minus.gdpo-noaux.math-abs}{8.0}
\setval{4b.geoma.minus.gdpo-noaux.math500}{$-$5.9}
\setval{4b.geoma.minus.gdpo-noaux.math500-abs}{5.9}
\setval{4b.geoma.minus.gdpo-noaux.mbpp-plus}{$-$2.1}
\setval{4b.geoma.minus.gdpo-noaux.mbpp-plus-abs}{2.1}
\setval{4b.geoma.minus.gdpo-noaux.minerva}{$-$2.6}
\setval{4b.geoma.minus.gdpo-noaux.minerva-abs}{2.6}
\setval{4b.geoma.minus.gdpo-noaux.olympiad-bench}{$-$9.3}
\setval{4b.geoma.minus.gdpo-noaux.olympiad-bench-abs}{9.3}
\setval{4b.geoma.minus.gdpo-noaux.writing}{+3.4}
\setval{4b.geoma.minus.gdpo-noaux.writing-abs}{3.4}
\setval{4b.geoma.minus.gdpo-noaux.writingbench}{+2.2}
\setval{4b.geoma.minus.gdpo-noaux.writingbench-abs}{2.2}
\setval{4b.geoma.minus.gdpo-w0.5.aime24}{$-$3.0}
\setval{4b.geoma.minus.gdpo-w0.5.aime24-abs}{3.0}
\setval{4b.geoma.minus.gdpo-w0.5.aime25}{+0.3}
\setval{4b.geoma.minus.gdpo-w0.5.aime25-abs}{0.3}
\setval{4b.geoma.minus.gdpo-w0.5.amc23}{$-$3.0}
\setval{4b.geoma.minus.gdpo-w0.5.amc23-abs}{3.0}
\setval{4b.geoma.minus.gdpo-w0.5.avg}{$-$0.8}
\setval{4b.geoma.minus.gdpo-w0.5.avg-abs}{0.8}
\setval{4b.geoma.minus.gdpo-w0.5.code}{$-$0.4}
\setval{4b.geoma.minus.gdpo-w0.5.code-abs}{0.4}
\setval{4b.geoma.minus.gdpo-w0.5.creative-writing-v3}{+7.8}
\setval{4b.geoma.minus.gdpo-w0.5.creative-writing-v3-abs}{7.8}
\setval{4b.geoma.minus.gdpo-w0.5.guidebench}{$-$1.1}
\setval{4b.geoma.minus.gdpo-w0.5.guidebench-abs}{1.1}
\setval{4b.geoma.minus.gdpo-w0.5.humaneval-plus}{$-$1.8}
\setval{4b.geoma.minus.gdpo-w0.5.humaneval-plus-abs}{1.8}
\setval{4b.geoma.minus.gdpo-w0.5.if}{$-$1.1}
\setval{4b.geoma.minus.gdpo-w0.5.if-abs}{1.1}
\setval{4b.geoma.minus.gdpo-w0.5.ifbench}{$-$3.3}
\setval{4b.geoma.minus.gdpo-w0.5.ifbench-abs}{3.3}
\setval{4b.geoma.minus.gdpo-w0.5.ifeval}{+1.0}
\setval{4b.geoma.minus.gdpo-w0.5.ifeval-abs}{1.0}
\setval{4b.geoma.minus.gdpo-w0.5.judge2.avg}{$-$1.3}
\setval{4b.geoma.minus.gdpo-w0.5.judge2.avg-abs}{1.3}
\setval{4b.geoma.minus.gdpo-w0.5.judge2.creative-writing-v3}{+3.1}
\setval{4b.geoma.minus.gdpo-w0.5.judge2.creative-writing-v3-abs}{3.1}
\setval{4b.geoma.minus.gdpo-w0.5.judge2.longbench-write}{$-$2.9}
\setval{4b.geoma.minus.gdpo-w0.5.judge2.longbench-write-abs}{2.9}
\setval{4b.geoma.minus.gdpo-w0.5.judge2.writing}{$-$0.9}
\setval{4b.geoma.minus.gdpo-w0.5.judge2.writing-abs}{0.9}
\setval{4b.geoma.minus.gdpo-w0.5.judge2.writingbench}{$-$2.9}
\setval{4b.geoma.minus.gdpo-w0.5.judge2.writingbench-abs}{2.9}
\setval{4b.geoma.minus.gdpo-w0.5.livecodebench}{+1.5}
\setval{4b.geoma.minus.gdpo-w0.5.livecodebench-abs}{1.5}
\setval{4b.geoma.minus.gdpo-w0.5.longbench-write}{$-$1.6}
\setval{4b.geoma.minus.gdpo-w0.5.longbench-write-abs}{1.6}
\setval{4b.geoma.minus.gdpo-w0.5.math}{$-$2.1}
\setval{4b.geoma.minus.gdpo-w0.5.math-abs}{2.1}
\setval{4b.geoma.minus.gdpo-w0.5.math500}{$-$2.9}
\setval{4b.geoma.minus.gdpo-w0.5.math500-abs}{2.9}
\setval{4b.geoma.minus.gdpo-w0.5.mbpp-plus}{$-$1.0}
\setval{4b.geoma.minus.gdpo-w0.5.mbpp-plus-abs}{1.0}
\setval{4b.geoma.minus.gdpo-w0.5.minerva}{+0.8}
\setval{4b.geoma.minus.gdpo-w0.5.minerva-abs}{0.8}
\setval{4b.geoma.minus.gdpo-w0.5.olympiad-bench}{$-$4.6}
\setval{4b.geoma.minus.gdpo-w0.5.olympiad-bench-abs}{4.6}
\setval{4b.geoma.minus.gdpo-w0.5.writing}{+1.7}
\setval{4b.geoma.minus.gdpo-w0.5.writing-abs}{1.7}
\setval{4b.geoma.minus.gdpo-w0.5.writingbench}{$-$1.2}
\setval{4b.geoma.minus.gdpo-w0.5.writingbench-abs}{1.2}
\setval{4b.geoma.minus.gdpo.aime24}{+2.5}
\setval{4b.geoma.minus.gdpo.aime24-abs}{2.5}
\setval{4b.geoma.minus.gdpo.aime25}{+0.7}
\setval{4b.geoma.minus.gdpo.aime25-abs}{0.7}
\setval{4b.geoma.minus.gdpo.amc23}{$-$4.2}
\setval{4b.geoma.minus.gdpo.amc23-abs}{4.2}
\setval{4b.geoma.minus.gdpo.avg}{$-$0.5}
\setval{4b.geoma.minus.gdpo.avg-abs}{0.5}
\setval{4b.geoma.minus.gdpo.code}{$-$0.3}
\setval{4b.geoma.minus.gdpo.code-abs}{0.3}
\setval{4b.geoma.minus.gdpo.creative-writing-v3}{$-$0.1}
\setval{4b.geoma.minus.gdpo.creative-writing-v3-abs}{0.1}
\setval{4b.geoma.minus.gdpo.guidebench}{$-$2.1}
\setval{4b.geoma.minus.gdpo.guidebench-abs}{2.1}
\setval{4b.geoma.minus.gdpo.humaneval-plus}{$-$1.6}
\setval{4b.geoma.minus.gdpo.humaneval-plus-abs}{1.6}
\setval{4b.geoma.minus.gdpo.if}{$-$0.1}
\setval{4b.geoma.minus.gdpo.if-abs}{0.1}
\setval{4b.geoma.minus.gdpo.ifbench}{$-$0.3}
\setval{4b.geoma.minus.gdpo.ifbench-abs}{0.3}
\setval{4b.geoma.minus.gdpo.ifeval}{+2.0}
\setval{4b.geoma.minus.gdpo.ifeval-abs}{2.0}
\setval{4b.geoma.minus.gdpo.judge2.avg}{$-$0.8}
\setval{4b.geoma.minus.gdpo.judge2.avg-abs}{0.8}
\setval{4b.geoma.minus.gdpo.judge2.creative-writing-v3}{$-$1.3}
\setval{4b.geoma.minus.gdpo.judge2.creative-writing-v3-abs}{1.3}
\setval{4b.geoma.minus.gdpo.judge2.longbench-write}{$-$4.6}
\setval{4b.geoma.minus.gdpo.judge2.longbench-write-abs}{4.6}
\setval{4b.geoma.minus.gdpo.judge2.writing}{$-$3.1}
\setval{4b.geoma.minus.gdpo.judge2.writing-abs}{3.1}
\setval{4b.geoma.minus.gdpo.judge2.writingbench}{$-$3.3}
\setval{4b.geoma.minus.gdpo.judge2.writingbench-abs}{3.3}
\setval{4b.geoma.minus.gdpo.livecodebench}{+1.3}
\setval{4b.geoma.minus.gdpo.livecodebench-abs}{1.3}
\setval{4b.geoma.minus.gdpo.longbench-write}{$-$3.2}
\setval{4b.geoma.minus.gdpo.longbench-write-abs}{3.2}
\setval{4b.geoma.minus.gdpo.math}{$-$0.1}
\setval{4b.geoma.minus.gdpo.math-abs}{0.1}
\setval{4b.geoma.minus.gdpo.math500}{$-$0.4}
\setval{4b.geoma.minus.gdpo.math500-abs}{0.4}
\setval{4b.geoma.minus.gdpo.mbpp-plus}{$-$0.7}
\setval{4b.geoma.minus.gdpo.mbpp-plus-abs}{0.7}
\setval{4b.geoma.minus.gdpo.minerva}{+1.5}
\setval{4b.geoma.minus.gdpo.minerva-abs}{1.5}
\setval{4b.geoma.minus.gdpo.olympiad-bench}{$-$0.8}
\setval{4b.geoma.minus.gdpo.olympiad-bench-abs}{0.8}
\setval{4b.geoma.minus.gdpo.writing}{$-$1.9}
\setval{4b.geoma.minus.gdpo.writing-abs}{1.9}
\setval{4b.geoma.minus.gdpo.writingbench}{$-$2.4}
\setval{4b.geoma.minus.gdpo.writingbench-abs}{2.4}
\setval{4b.geoma.minus.grpo-nodiv.aime24}{$-$0.8}
\setval{4b.geoma.minus.grpo-nodiv.aime24-abs}{0.8}
\setval{4b.geoma.minus.grpo-nodiv.aime25}{+2.3}
\setval{4b.geoma.minus.grpo-nodiv.aime25-abs}{2.3}
\setval{4b.geoma.minus.grpo-nodiv.amc23}{+0.9}
\setval{4b.geoma.minus.grpo-nodiv.amc23-abs}{0.9}
\setval{4b.geoma.minus.grpo-nodiv.avg}{+0.8}
\setval{4b.geoma.minus.grpo-nodiv.avg-abs}{0.8}
\setval{4b.geoma.minus.grpo-nodiv.code}{+0.0}
\setval{4b.geoma.minus.grpo-nodiv.code-abs}{0.0}
\setval{4b.geoma.minus.grpo-nodiv.creative-writing-v3}{+3.3}
\setval{4b.geoma.minus.grpo-nodiv.creative-writing-v3-abs}{3.3}
\setval{4b.geoma.minus.grpo-nodiv.guidebench}{$-$0.3}
\setval{4b.geoma.minus.grpo-nodiv.guidebench-abs}{0.3}
\setval{4b.geoma.minus.grpo-nodiv.humaneval-plus}{$-$0.2}
\setval{4b.geoma.minus.grpo-nodiv.humaneval-plus-abs}{0.2}
\setval{4b.geoma.minus.grpo-nodiv.if}{+0.8}
\setval{4b.geoma.minus.grpo-nodiv.if-abs}{0.8}
\setval{4b.geoma.minus.grpo-nodiv.ifbench}{+2.3}
\setval{4b.geoma.minus.grpo-nodiv.ifbench-abs}{2.3}
\setval{4b.geoma.minus.grpo-nodiv.ifeval}{+0.5}
\setval{4b.geoma.minus.grpo-nodiv.ifeval-abs}{0.5}
\setval{4b.geoma.minus.grpo-nodiv.judge2.avg}{+0.5}
\setval{4b.geoma.minus.grpo-nodiv.judge2.avg-abs}{0.5}
\setval{4b.geoma.minus.grpo-nodiv.judge2.creative-writing-v3}{+2.6}
\setval{4b.geoma.minus.grpo-nodiv.judge2.creative-writing-v3-abs}{2.6}
\setval{4b.geoma.minus.grpo-nodiv.judge2.longbench-write}{$-$0.9}
\setval{4b.geoma.minus.grpo-nodiv.judge2.longbench-write-abs}{0.9}
\setval{4b.geoma.minus.grpo-nodiv.judge2.writing}{+0.9}
\setval{4b.geoma.minus.grpo-nodiv.judge2.writing-abs}{0.9}
\setval{4b.geoma.minus.grpo-nodiv.judge2.writingbench}{+1.1}
\setval{4b.geoma.minus.grpo-nodiv.judge2.writingbench-abs}{1.1}
\setval{4b.geoma.minus.grpo-nodiv.livecodebench}{+1.2}
\setval{4b.geoma.minus.grpo-nodiv.livecodebench-abs}{1.2}
\setval{4b.geoma.minus.grpo-nodiv.longbench-write}{$-$0.2}
\setval{4b.geoma.minus.grpo-nodiv.longbench-write-abs}{0.2}
\setval{4b.geoma.minus.grpo-nodiv.math}{+0.3}
\setval{4b.geoma.minus.grpo-nodiv.math-abs}{0.3}
\setval{4b.geoma.minus.grpo-nodiv.math500}{$-$0.6}
\setval{4b.geoma.minus.grpo-nodiv.math500-abs}{0.6}
\setval{4b.geoma.minus.grpo-nodiv.mbpp-plus}{$-$1.0}
\setval{4b.geoma.minus.grpo-nodiv.mbpp-plus-abs}{1.0}
\setval{4b.geoma.minus.grpo-nodiv.minerva}{+1.1}
\setval{4b.geoma.minus.grpo-nodiv.minerva-abs}{1.1}
\setval{4b.geoma.minus.grpo-nodiv.olympiad-bench}{$-$1.0}
\setval{4b.geoma.minus.grpo-nodiv.olympiad-bench-abs}{1.0}
\setval{4b.geoma.minus.grpo-nodiv.writing}{+2.3}
\setval{4b.geoma.minus.grpo-nodiv.writing-abs}{2.3}
\setval{4b.geoma.minus.grpo-nodiv.writingbench}{+3.9}
\setval{4b.geoma.minus.grpo-nodiv.writingbench-abs}{3.9}
\setval{4b.geoma.minus.grpo.aime24}{$-$0.4}
\setval{4b.geoma.minus.grpo.aime24-abs}{0.4}
\setval{4b.geoma.minus.grpo.aime25}{$-$2.0}
\setval{4b.geoma.minus.grpo.aime25-abs}{2.0}
\setval{4b.geoma.minus.grpo.amc23}{+0.7}
\setval{4b.geoma.minus.grpo.amc23-abs}{0.7}
\setval{4b.geoma.minus.grpo.avg}{$-$0.9}
\setval{4b.geoma.minus.grpo.avg-abs}{0.9}
\setval{4b.geoma.minus.grpo.code}{$-$0.6}
\setval{4b.geoma.minus.grpo.code-abs}{0.6}
\setval{4b.geoma.minus.grpo.creative-writing-v3}{$-$1.5}
\setval{4b.geoma.minus.grpo.creative-writing-v3-abs}{1.5}
\setval{4b.geoma.minus.grpo.guidebench}{+0.6}
\setval{4b.geoma.minus.grpo.guidebench-abs}{0.6}
\setval{4b.geoma.minus.grpo.humaneval-plus}{$-$2.5}
\setval{4b.geoma.minus.grpo.humaneval-plus-abs}{2.5}
\setval{4b.geoma.minus.grpo.if}{+0.1}
\setval{4b.geoma.minus.grpo.if-abs}{0.1}
\setval{4b.geoma.minus.grpo.ifbench}{$-$1.7}
\setval{4b.geoma.minus.grpo.ifbench-abs}{1.7}
\setval{4b.geoma.minus.grpo.ifeval}{+1.3}
\setval{4b.geoma.minus.grpo.ifeval-abs}{1.3}
\setval{4b.geoma.minus.grpo.judge2.avg}{$-$0.8}
\setval{4b.geoma.minus.grpo.judge2.avg-abs}{0.8}
\setval{4b.geoma.minus.grpo.judge2.creative-writing-v3}{$-$2.0}
\setval{4b.geoma.minus.grpo.judge2.creative-writing-v3-abs}{2.0}
\setval{4b.geoma.minus.grpo.judge2.longbench-write}{$-$3.1}
\setval{4b.geoma.minus.grpo.judge2.longbench-write-abs}{3.1}
\setval{4b.geoma.minus.grpo.judge2.writing}{$-$2.6}
\setval{4b.geoma.minus.grpo.judge2.writing-abs}{2.6}
\setval{4b.geoma.minus.grpo.judge2.writingbench}{$-$2.8}
\setval{4b.geoma.minus.grpo.judge2.writingbench-abs}{2.8}
\setval{4b.geoma.minus.grpo.livecodebench}{+1.7}
\setval{4b.geoma.minus.grpo.livecodebench-abs}{1.7}
\setval{4b.geoma.minus.grpo.longbench-write}{$-$3.4}
\setval{4b.geoma.minus.grpo.longbench-write-abs}{3.4}
\setval{4b.geoma.minus.grpo.math}{$-$0.4}
\setval{4b.geoma.minus.grpo.math-abs}{0.4}
\setval{4b.geoma.minus.grpo.math500}{$-$1.4}
\setval{4b.geoma.minus.grpo.math500-abs}{1.4}
\setval{4b.geoma.minus.grpo.mbpp-plus}{$-$1.0}
\setval{4b.geoma.minus.grpo.mbpp-plus-abs}{1.0}
\setval{4b.geoma.minus.grpo.minerva}{+1.7}
\setval{4b.geoma.minus.grpo.minerva-abs}{1.7}
\setval{4b.geoma.minus.grpo.olympiad-bench}{$-$0.8}
\setval{4b.geoma.minus.grpo.olympiad-bench-abs}{0.8}
\setval{4b.geoma.minus.grpo.writing}{$-$3.2}
\setval{4b.geoma.minus.grpo.writing-abs}{3.2}
\setval{4b.geoma.minus.grpo.writingbench}{$-$4.7}
\setval{4b.geoma.minus.grpo.writingbench-abs}{4.7}
\setval{4b.geoma.minus.mt-grpo.aime24}{+0.8}
\setval{4b.geoma.minus.mt-grpo.aime24-abs}{0.8}
\setval{4b.geoma.minus.mt-grpo.aime25}{+4.5}
\setval{4b.geoma.minus.mt-grpo.aime25-abs}{4.5}
\setval{4b.geoma.minus.mt-grpo.amc23}{$-$4.4}
\setval{4b.geoma.minus.mt-grpo.amc23-abs}{4.4}
\setval{4b.geoma.minus.mt-grpo.avg}{+0.7}
\setval{4b.geoma.minus.mt-grpo.avg-abs}{0.7}
\setval{4b.geoma.minus.mt-grpo.code}{+1.4}
\setval{4b.geoma.minus.mt-grpo.code-abs}{1.4}
\setval{4b.geoma.minus.mt-grpo.creative-writing-v3}{+1.4}
\setval{4b.geoma.minus.mt-grpo.creative-writing-v3-abs}{1.4}
\setval{4b.geoma.minus.mt-grpo.guidebench}{$-$1.2}
\setval{4b.geoma.minus.mt-grpo.guidebench-abs}{1.2}
\setval{4b.geoma.minus.mt-grpo.humaneval-plus}{+1.2}
\setval{4b.geoma.minus.mt-grpo.humaneval-plus-abs}{1.2}
\setval{4b.geoma.minus.mt-grpo.if}{+0.5}
\setval{4b.geoma.minus.mt-grpo.if-abs}{0.5}
\setval{4b.geoma.minus.mt-grpo.ifbench}{+1.7}
\setval{4b.geoma.minus.mt-grpo.ifbench-abs}{1.7}
\setval{4b.geoma.minus.mt-grpo.ifeval}{+1.1}
\setval{4b.geoma.minus.mt-grpo.ifeval-abs}{1.1}
\setval{4b.geoma.minus.mt-grpo.judge2.avg}{+0.3}
\setval{4b.geoma.minus.mt-grpo.judge2.avg-abs}{0.3}
\setval{4b.geoma.minus.mt-grpo.judge2.creative-writing-v3}{$-$0.9}
\setval{4b.geoma.minus.mt-grpo.judge2.creative-writing-v3-abs}{0.9}
\setval{4b.geoma.minus.mt-grpo.judge2.longbench-write}{$-$3.1}
\setval{4b.geoma.minus.mt-grpo.judge2.longbench-write-abs}{3.1}
\setval{4b.geoma.minus.mt-grpo.judge2.writing}{$-$2.2}
\setval{4b.geoma.minus.mt-grpo.judge2.writing-abs}{2.2}
\setval{4b.geoma.minus.mt-grpo.judge2.writingbench}{$-$2.6}
\setval{4b.geoma.minus.mt-grpo.judge2.writingbench-abs}{2.6}
\setval{4b.geoma.minus.mt-grpo.livecodebench}{+3.0}
\setval{4b.geoma.minus.mt-grpo.livecodebench-abs}{3.0}
\setval{4b.geoma.minus.mt-grpo.longbench-write}{$-$1.0}
\setval{4b.geoma.minus.mt-grpo.longbench-write-abs}{1.0}
\setval{4b.geoma.minus.mt-grpo.math}{+0.8}
\setval{4b.geoma.minus.mt-grpo.math-abs}{0.8}
\setval{4b.geoma.minus.mt-grpo.math500}{+1.9}
\setval{4b.geoma.minus.mt-grpo.math500-abs}{1.9}
\setval{4b.geoma.minus.mt-grpo.mbpp-plus}{+0.2}
\setval{4b.geoma.minus.mt-grpo.mbpp-plus-abs}{0.2}
\setval{4b.geoma.minus.mt-grpo.minerva}{+1.0}
\setval{4b.geoma.minus.mt-grpo.minerva-abs}{1.0}
\setval{4b.geoma.minus.mt-grpo.olympiad-bench}{+1.3}
\setval{4b.geoma.minus.mt-grpo.olympiad-bench-abs}{1.3}
\setval{4b.geoma.minus.mt-grpo.writing}{$-$0.1}
\setval{4b.geoma.minus.mt-grpo.writing-abs}{0.1}
\setval{4b.geoma.minus.mt-grpo.writingbench}{$-$0.6}
\setval{4b.geoma.minus.mt-grpo.writingbench-abs}{0.6}
\setval{4b.geoma.minus.papo.aime24}{+2.3}
\setval{4b.geoma.minus.papo.aime24-abs}{2.3}
\setval{4b.geoma.minus.papo.aime25}{+3.1}
\setval{4b.geoma.minus.papo.aime25-abs}{3.1}
\setval{4b.geoma.minus.papo.amc23}{+1.9}
\setval{4b.geoma.minus.papo.amc23-abs}{1.9}
\setval{4b.geoma.minus.papo.avg}{+1.2}
\setval{4b.geoma.minus.papo.avg-abs}{1.2}
\setval{4b.geoma.minus.papo.code}{$-$0.4}
\setval{4b.geoma.minus.papo.code-abs}{0.4}
\setval{4b.geoma.minus.papo.creative-writing-v3}{$-$0.7}
\setval{4b.geoma.minus.papo.creative-writing-v3-abs}{0.7}
\setval{4b.geoma.minus.papo.guidebench}{$-$2.2}
\setval{4b.geoma.minus.papo.guidebench-abs}{2.2}
\setval{4b.geoma.minus.papo.humaneval-plus}{$-$2.8}
\setval{4b.geoma.minus.papo.humaneval-plus-abs}{2.8}
\setval{4b.geoma.minus.papo.if}{+0.8}
\setval{4b.geoma.minus.papo.if-abs}{0.8}
\setval{4b.geoma.minus.papo.ifbench}{+2.7}
\setval{4b.geoma.minus.papo.ifbench-abs}{2.7}
\setval{4b.geoma.minus.papo.ifeval}{+1.9}
\setval{4b.geoma.minus.papo.ifeval-abs}{1.9}
\setval{4b.geoma.minus.papo.judge2.avg}{+0.7}
\setval{4b.geoma.minus.papo.judge2.avg-abs}{0.7}
\setval{4b.geoma.minus.papo.judge2.creative-writing-v3}{$-$1.1}
\setval{4b.geoma.minus.papo.judge2.creative-writing-v3-abs}{1.1}
\setval{4b.geoma.minus.papo.judge2.longbench-write}{+0.1}
\setval{4b.geoma.minus.papo.judge2.longbench-write-abs}{0.1}
\setval{4b.geoma.minus.papo.judge2.writing}{$-$0.8}
\setval{4b.geoma.minus.papo.judge2.writing-abs}{0.8}
\setval{4b.geoma.minus.papo.judge2.writingbench}{$-$1.5}
\setval{4b.geoma.minus.papo.judge2.writingbench-abs}{1.5}
\setval{4b.geoma.minus.papo.livecodebench}{+2.1}
\setval{4b.geoma.minus.papo.livecodebench-abs}{2.1}
\setval{4b.geoma.minus.papo.longbench-write}{+1.9}
\setval{4b.geoma.minus.papo.longbench-write-abs}{1.9}
\setval{4b.geoma.minus.papo.math}{+2.1}
\setval{4b.geoma.minus.papo.math-abs}{2.1}
\setval{4b.geoma.minus.papo.math500}{+2.3}
\setval{4b.geoma.minus.papo.math500-abs}{2.3}
\setval{4b.geoma.minus.papo.mbpp-plus}{$-$0.5}
\setval{4b.geoma.minus.papo.mbpp-plus-abs}{0.5}
\setval{4b.geoma.minus.papo.minerva}{+0.4}
\setval{4b.geoma.minus.papo.minerva-abs}{0.4}
\setval{4b.geoma.minus.papo.olympiad-bench}{+2.4}
\setval{4b.geoma.minus.papo.olympiad-bench-abs}{2.4}
\setval{4b.geoma.minus.papo.writing}{+1.4}
\setval{4b.geoma.minus.papo.writing-abs}{1.4}
\setval{4b.geoma.minus.papo.writingbench}{+3.0}
\setval{4b.geoma.minus.papo.writingbench-abs}{3.0}
\setval{4b.geoma.minus.saw.aime24}{$-$0.9}
\setval{4b.geoma.minus.saw.aime24-abs}{0.9}
\setval{4b.geoma.minus.saw.aime25}{$-$0.5}
\setval{4b.geoma.minus.saw.aime25-abs}{0.5}
\setval{4b.geoma.minus.saw.amc23}{$-$1.2}
\setval{4b.geoma.minus.saw.amc23-abs}{1.2}
\setval{4b.geoma.minus.saw.avg}{$-$0.4}
\setval{4b.geoma.minus.saw.avg-abs}{0.4}
\setval{4b.geoma.minus.saw.code}{+0.8}
\setval{4b.geoma.minus.saw.code-abs}{0.8}
\setval{4b.geoma.minus.saw.creative-writing-v3}{+0.8}
\setval{4b.geoma.minus.saw.creative-writing-v3-abs}{0.8}
\setval{4b.geoma.minus.saw.guidebench}{$-$2.5}
\setval{4b.geoma.minus.saw.guidebench-abs}{2.5}
\setval{4b.geoma.minus.saw.humaneval-plus}{$-$0.5}
\setval{4b.geoma.minus.saw.humaneval-plus-abs}{0.5}
\setval{4b.geoma.minus.saw.if}{$-$2.0}
\setval{4b.geoma.minus.saw.if-abs}{2.0}
\setval{4b.geoma.minus.saw.ifbench}{$-$4.3}
\setval{4b.geoma.minus.saw.ifbench-abs}{4.3}
\setval{4b.geoma.minus.saw.ifeval}{+0.8}
\setval{4b.geoma.minus.saw.ifeval-abs}{0.8}
\setval{4b.geoma.minus.saw.judge2.avg}{$-$0.9}
\setval{4b.geoma.minus.saw.judge2.avg-abs}{0.9}
\setval{4b.geoma.minus.saw.judge2.creative-writing-v3}{$-$2.5}
\setval{4b.geoma.minus.saw.judge2.creative-writing-v3-abs}{2.5}
\setval{4b.geoma.minus.saw.judge2.longbench-write}{$-$2.4}
\setval{4b.geoma.minus.saw.judge2.longbench-write-abs}{2.4}
\setval{4b.geoma.minus.saw.judge2.writing}{$-$2.5}
\setval{4b.geoma.minus.saw.judge2.writing-abs}{2.5}
\setval{4b.geoma.minus.saw.judge2.writingbench}{$-$2.7}
\setval{4b.geoma.minus.saw.judge2.writingbench-abs}{2.7}
\setval{4b.geoma.minus.saw.livecodebench}{+2.9}
\setval{4b.geoma.minus.saw.livecodebench-abs}{2.9}
\setval{4b.geoma.minus.saw.longbench-write}{$-$0.4}
\setval{4b.geoma.minus.saw.longbench-write-abs}{0.4}
\setval{4b.geoma.minus.saw.math}{$-$0.5}
\setval{4b.geoma.minus.saw.math-abs}{0.5}
\setval{4b.geoma.minus.saw.math500}{+1.0}
\setval{4b.geoma.minus.saw.math500-abs}{1.0}
\setval{4b.geoma.minus.saw.mbpp-plus}{+0.1}
\setval{4b.geoma.minus.saw.mbpp-plus-abs}{0.1}
\setval{4b.geoma.minus.saw.minerva}{$-$1.2}
\setval{4b.geoma.minus.saw.minerva-abs}{1.2}
\setval{4b.geoma.minus.saw.olympiad-bench}{+0.0}
\setval{4b.geoma.minus.saw.olympiad-bench-abs}{0.0}
\setval{4b.geoma.minus.saw.writing}{+0.0}
\setval{4b.geoma.minus.saw.writing-abs}{0.0}
\setval{4b.geoma.minus.saw.writingbench}{$-$0.3}
\setval{4b.geoma.minus.saw.writingbench-abs}{0.3}
\setval{4b.geoma.passing-at-ceiling.code.early}{0.70}
\setval{4b.geoma.passing-at-ceiling.code.late}{0.74}
\setval{4b.geoma.passing-at-ceiling.code.run}{0.73}
\setval{4b.geoma.passing-at-ceiling.if.early}{0.28}
\setval{4b.geoma.passing-at-ceiling.if.late}{0.56}
\setval{4b.geoma.passing-at-ceiling.if.run}{0.48}
\setval{4b.geoma.passing-at-ceiling.math.early}{1.00}
\setval{4b.geoma.passing-at-ceiling.math.late}{1.00}
\setval{4b.geoma.passing-at-ceiling.math.run}{1.00}
\setval{4b.geoma.passing-at-ceiling.writing.early}{0.00}
\setval{4b.geoma.passing-at-ceiling.writing.late}{0.00}
\setval{4b.geoma.passing-at-ceiling.writing.run}{0.00}
\setval{4b.geoma.primary-cosine.code.early}{0.94}
\setval{4b.geoma.primary-cosine.code.late}{0.97}
\setval{4b.geoma.primary-cosine.code.run}{0.97}
\setval{4b.geoma.primary-cosine.if.early}{0.69}
\setval{4b.geoma.primary-cosine.if.late}{0.71}
\setval{4b.geoma.primary-cosine.if.run}{0.75}
\setval{4b.geoma.primary-cosine.math.early}{0.91}
\setval{4b.geoma.primary-cosine.math.late}{0.95}
\setval{4b.geoma.primary-cosine.math.run}{0.95}
\setval{4b.geoma.primary-cosine.writing.early}{0.83}
\setval{4b.geoma.primary-cosine.writing.late}{0.75}
\setval{4b.geoma.primary-cosine.writing.run}{0.75}
\setval{4b.geoma.primary-level.code.early}{0.53}
\setval{4b.geoma.primary-level.code.late}{0.75}
\setval{4b.geoma.primary-level.code.run}{0.70}
\setval{4b.geoma.primary-level.if.early}{0.32}
\setval{4b.geoma.primary-level.if.late}{0.74}
\setval{4b.geoma.primary-level.if.run}{0.63}
\setval{4b.geoma.primary-level.math.early}{0.48}
\setval{4b.geoma.primary-level.math.late}{0.68}
\setval{4b.geoma.primary-level.math.run}{0.62}
\setval{4b.geoma.primary-level.writing.early}{0.53}
\setval{4b.geoma.primary-level.writing.late}{0.76}
\setval{4b.geoma.primary-level.writing.run}{0.71}
\setval{4b.geoma.solved.code.early}{0.04}
\setval{4b.geoma.solved.code.late}{0.02}
\setval{4b.geoma.solved.code.run}{0.02}
\setval{4b.geoma.solved.if.early}{0.03}
\setval{4b.geoma.solved.if.late}{0.18}
\setval{4b.geoma.solved.if.run}{0.12}
\setval{4b.geoma.solved.math.early}{0.12}
\setval{4b.geoma.solved.math.late}{0.09}
\setval{4b.geoma.solved.math.run}{0.08}
\setval{4b.geoma.solved.writing.early}{0.00}
\setval{4b.geoma.solved.writing.late}{0.00}
\setval{4b.geoma.solved.writing.run}{0.00}
\setval{4b.geoma.update-share.code.early}{0.24}
\setval{4b.geoma.update-share.code.late}{0.19}
\setval{4b.geoma.update-share.code.run}{0.20}
\setval{4b.geoma.update-share.if.early}{0.26}
\setval{4b.geoma.update-share.if.late}{0.20}
\setval{4b.geoma.update-share.if.run}{0.23}
\setval{4b.geoma.update-share.math.early}{0.22}
\setval{4b.geoma.update-share.math.late}{0.19}
\setval{4b.geoma.update-share.math.run}{0.20}
\setval{4b.geoma.update-share.writing.early}{0.29}
\setval{4b.geoma.update-share.writing.late}{0.42}
\setval{4b.geoma.update-share.writing.run}{0.37}
\setval{4b.grpo-nodiv.aux-change.code}{+0.19}
\setval{4b.grpo-nodiv.aux-change.if}{+0.23}
\setval{4b.grpo-nodiv.aux-change.math}{+0.10}
\setval{4b.grpo-nodiv.aux-change.writing}{+0.19}
\setval{4b.grpo-nodiv.aux-change.writing-rm}{+0.34}
\setval{4b.grpo-nodiv.aux-full.code.early}{0.74}
\setval{4b.grpo-nodiv.aux-full.code.late}{0.89}
\setval{4b.grpo-nodiv.aux-full.code.run}{0.90}
\setval{4b.grpo-nodiv.aux-full.if.early}{0.65}
\setval{4b.grpo-nodiv.aux-full.if.late}{0.88}
\setval{4b.grpo-nodiv.aux-full.if.run}{0.84}
\setval{4b.grpo-nodiv.aux-full.math.early}{0.65}
\setval{4b.grpo-nodiv.aux-full.math.late}{0.77}
\setval{4b.grpo-nodiv.aux-full.math.run}{0.77}
\setval{4b.grpo-nodiv.aux-full.writing-rm.early}{0.08}
\setval{4b.grpo-nodiv.aux-full.writing-rm.late}{0.46}
\setval{4b.grpo-nodiv.aux-full.writing-rm.run}{0.35}
\setval{4b.grpo-nodiv.aux-full.writing.early}{0.65}
\setval{4b.grpo-nodiv.aux-full.writing.late}{0.70}
\setval{4b.grpo-nodiv.aux-full.writing.run}{0.67}
\setval{4b.grpo-nodiv.aux-level.code.early}{0.93}
\setval{4b.grpo-nodiv.aux-level.code.late}{0.98}
\setval{4b.grpo-nodiv.aux-level.code.run}{0.98}
\setval{4b.grpo-nodiv.aux-level.if.early}{0.83}
\setval{4b.grpo-nodiv.aux-level.if.late}{0.93}
\setval{4b.grpo-nodiv.aux-level.if.run}{0.92}
\setval{4b.grpo-nodiv.aux-level.math.early}{0.93}
\setval{4b.grpo-nodiv.aux-level.math.late}{0.95}
\setval{4b.grpo-nodiv.aux-level.math.run}{0.95}
\setval{4b.grpo-nodiv.aux-level.writing-rm.early}{0.77}
\setval{4b.grpo-nodiv.aux-level.writing-rm.late}{0.94}
\setval{4b.grpo-nodiv.aux-level.writing-rm.run}{0.91}
\setval{4b.grpo-nodiv.aux-level.writing.early}{0.91}
\setval{4b.grpo-nodiv.aux-level.writing.late}{0.96}
\setval{4b.grpo-nodiv.aux-level.writing.run}{0.95}
\setval{4b.grpo-nodiv.bench.aime24}{12.4}
\setval{4b.grpo-nodiv.bench.aime25}{8.3}
\setval{4b.grpo-nodiv.bench.amc23}{54.5}
\setval{4b.grpo-nodiv.bench.creative-writing-v3}{42.5}
\setval{4b.grpo-nodiv.bench.guidebench}{78.8}
\setval{4b.grpo-nodiv.bench.humaneval-plus}{80.5}
\setval{4b.grpo-nodiv.bench.ifbench}{33.7}
\setval{4b.grpo-nodiv.bench.ifeval}{87.3}
\setval{4b.grpo-nodiv.bench.livecodebench}{24.5}
\setval{4b.grpo-nodiv.bench.longbench-write}{66.0}
\setval{4b.grpo-nodiv.bench.math500}{79.5}
\setval{4b.grpo-nodiv.bench.mbpp-plus}{71.2}
\setval{4b.grpo-nodiv.bench.minerva}{45.8}
\setval{4b.grpo-nodiv.bench.olympiad-bench}{44.8}
\setval{4b.grpo-nodiv.bench.writingbench}{65.6}
\setval{4b.grpo-nodiv.eval.avg}{53.0}
\setval{4b.grpo-nodiv.eval.code}{58.7}
\setval{4b.grpo-nodiv.eval.if}{66.6}
\setval{4b.grpo-nodiv.eval.math}{40.9}
\setval{4b.grpo-nodiv.eval.writing}{58.1}
\setval{4b.grpo-nodiv.judge2.avg}{50.7}
\setval{4b.grpo-nodiv.judge2.creative-writing-v3}{30.5}
\setval{4b.grpo-nodiv.judge2.longbench-write}{58.6}
\setval{4b.grpo-nodiv.judge2.writing}{46.2}
\setval{4b.grpo-nodiv.judge2.writingbench}{49.6}
\setval{4b.grpo-nodiv.last-step}{300}
\setval{4b.grpo-nodiv.minus.base.aime24}{+4.6}
\setval{4b.grpo-nodiv.minus.base.aime24-abs}{4.6}
\setval{4b.grpo-nodiv.minus.base.aime25}{+1.2}
\setval{4b.grpo-nodiv.minus.base.aime25-abs}{1.2}
\setval{4b.grpo-nodiv.minus.base.amc23}{+10.6}
\setval{4b.grpo-nodiv.minus.base.amc23-abs}{10.6}
\setval{4b.grpo-nodiv.minus.base.avg}{+18.5}
\setval{4b.grpo-nodiv.minus.base.avg-abs}{18.5}
\setval{4b.grpo-nodiv.minus.base.code}{+14.1}
\setval{4b.grpo-nodiv.minus.base.code-abs}{14.1}
\setval{4b.grpo-nodiv.minus.base.creative-writing-v3}{+17.8}
\setval{4b.grpo-nodiv.minus.base.creative-writing-v3-abs}{17.8}
\setval{4b.grpo-nodiv.minus.base.guidebench}{+53.9}
\setval{4b.grpo-nodiv.minus.base.guidebench-abs}{53.9}
\setval{4b.grpo-nodiv.minus.base.humaneval-plus}{+15.4}
\setval{4b.grpo-nodiv.minus.base.humaneval-plus-abs}{15.4}
\setval{4b.grpo-nodiv.minus.base.if}{+36.8}
\setval{4b.grpo-nodiv.minus.base.if-abs}{36.8}
\setval{4b.grpo-nodiv.minus.base.ifbench}{+16.7}
\setval{4b.grpo-nodiv.minus.base.ifbench-abs}{16.7}
\setval{4b.grpo-nodiv.minus.base.ifeval}{+39.9}
\setval{4b.grpo-nodiv.minus.base.ifeval-abs}{39.9}
\setval{4b.grpo-nodiv.minus.base.judge2.avg}{+16.8}
\setval{4b.grpo-nodiv.minus.base.judge2.avg-abs}{16.8}
\setval{4b.grpo-nodiv.minus.base.judge2.creative-writing-v3}{+9.7}
\setval{4b.grpo-nodiv.minus.base.judge2.creative-writing-v3-abs}{9.7}
\setval{4b.grpo-nodiv.minus.base.judge2.longbench-write}{+17.7}
\setval{4b.grpo-nodiv.minus.base.judge2.longbench-write-abs}{17.7}
\setval{4b.grpo-nodiv.minus.base.judge2.writing}{+15.5}
\setval{4b.grpo-nodiv.minus.base.judge2.writing-abs}{15.5}
\setval{4b.grpo-nodiv.minus.base.judge2.writingbench}{+19.1}
\setval{4b.grpo-nodiv.minus.base.judge2.writingbench-abs}{19.1}
\setval{4b.grpo-nodiv.minus.base.livecodebench}{+11.3}
\setval{4b.grpo-nodiv.minus.base.livecodebench-abs}{11.3}
\setval{4b.grpo-nodiv.minus.base.longbench-write}{+21.6}
\setval{4b.grpo-nodiv.minus.base.longbench-write-abs}{21.6}
\setval{4b.grpo-nodiv.minus.base.math}{+8.7}
\setval{4b.grpo-nodiv.minus.base.math-abs}{8.7}
\setval{4b.grpo-nodiv.minus.base.math500}{+12.2}
\setval{4b.grpo-nodiv.minus.base.math500-abs}{12.2}
\setval{4b.grpo-nodiv.minus.base.mbpp-plus}{+15.6}
\setval{4b.grpo-nodiv.minus.base.mbpp-plus-abs}{15.6}
\setval{4b.grpo-nodiv.minus.base.minerva}{+14.2}
\setval{4b.grpo-nodiv.minus.base.minerva-abs}{14.2}
\setval{4b.grpo-nodiv.minus.base.olympiad-bench}{+9.2}
\setval{4b.grpo-nodiv.minus.base.olympiad-bench-abs}{9.2}
\setval{4b.grpo-nodiv.minus.base.writing}{+24.2}
\setval{4b.grpo-nodiv.minus.base.writing-abs}{24.2}
\setval{4b.grpo-nodiv.minus.base.writingbench}{+33.2}
\setval{4b.grpo-nodiv.minus.base.writingbench-abs}{33.2}
\setval{4b.grpo-nodiv.minus.concur-free.aime24}{+0.5}
\setval{4b.grpo-nodiv.minus.concur-free.aime24-abs}{0.5}
\setval{4b.grpo-nodiv.minus.concur-free.aime25}{$-$6.3}
\setval{4b.grpo-nodiv.minus.concur-free.aime25-abs}{6.3}
\setval{4b.grpo-nodiv.minus.concur-free.amc23}{+0.4}
\setval{4b.grpo-nodiv.minus.concur-free.amc23-abs}{0.4}
\setval{4b.grpo-nodiv.minus.concur-free.avg}{$-$2.5}
\setval{4b.grpo-nodiv.minus.concur-free.avg-abs}{2.5}
\setval{4b.grpo-nodiv.minus.concur-free.code}{+0.4}
\setval{4b.grpo-nodiv.minus.concur-free.code-abs}{0.4}
\setval{4b.grpo-nodiv.minus.concur-free.creative-writing-v3}{$-$2.9}
\setval{4b.grpo-nodiv.minus.concur-free.creative-writing-v3-abs}{2.9}
\setval{4b.grpo-nodiv.minus.concur-free.guidebench}{$-$0.2}
\setval{4b.grpo-nodiv.minus.concur-free.guidebench-abs}{0.2}
\setval{4b.grpo-nodiv.minus.concur-free.humaneval-plus}{$-$0.6}
\setval{4b.grpo-nodiv.minus.concur-free.humaneval-plus-abs}{0.6}
\setval{4b.grpo-nodiv.minus.concur-free.if}{$-$1.6}
\setval{4b.grpo-nodiv.minus.concur-free.if-abs}{1.6}
\setval{4b.grpo-nodiv.minus.concur-free.ifbench}{$-$5.3}
\setval{4b.grpo-nodiv.minus.concur-free.ifbench-abs}{5.3}
\setval{4b.grpo-nodiv.minus.concur-free.ifeval}{+0.6}
\setval{4b.grpo-nodiv.minus.concur-free.ifeval-abs}{0.6}
\setval{4b.grpo-nodiv.minus.concur-free.judge2.avg}{$-$2.1}
\setval{4b.grpo-nodiv.minus.concur-free.judge2.avg-abs}{2.1}
\setval{4b.grpo-nodiv.minus.concur-free.judge2.creative-writing-v3}{$-$4.8}
\setval{4b.grpo-nodiv.minus.concur-free.judge2.creative-writing-v3-abs}{4.8}
\setval{4b.grpo-nodiv.minus.concur-free.judge2.longbench-write}{$-$10.7}
\setval{4b.grpo-nodiv.minus.concur-free.judge2.longbench-write-abs}{10.7}
\setval{4b.grpo-nodiv.minus.concur-free.judge2.writing}{$-$6.8}
\setval{4b.grpo-nodiv.minus.concur-free.judge2.writing-abs}{6.8}
\setval{4b.grpo-nodiv.minus.concur-free.judge2.writingbench}{$-$4.9}
\setval{4b.grpo-nodiv.minus.concur-free.judge2.writingbench-abs}{4.9}
\setval{4b.grpo-nodiv.minus.concur-free.livecodebench}{+0.3}
\setval{4b.grpo-nodiv.minus.concur-free.livecodebench-abs}{0.3}
\setval{4b.grpo-nodiv.minus.concur-free.longbench-write}{$-$11.8}
\setval{4b.grpo-nodiv.minus.concur-free.longbench-write-abs}{11.8}
\setval{4b.grpo-nodiv.minus.concur-free.math}{$-$1.3}
\setval{4b.grpo-nodiv.minus.concur-free.math-abs}{1.3}
\setval{4b.grpo-nodiv.minus.concur-free.math500}{$-$0.8}
\setval{4b.grpo-nodiv.minus.concur-free.math500-abs}{0.8}
\setval{4b.grpo-nodiv.minus.concur-free.mbpp-plus}{+1.5}
\setval{4b.grpo-nodiv.minus.concur-free.mbpp-plus-abs}{1.5}
\setval{4b.grpo-nodiv.minus.concur-free.minerva}{$-$0.5}
\setval{4b.grpo-nodiv.minus.concur-free.minerva-abs}{0.5}
\setval{4b.grpo-nodiv.minus.concur-free.olympiad-bench}{$-$0.9}
\setval{4b.grpo-nodiv.minus.concur-free.olympiad-bench-abs}{0.9}
\setval{4b.grpo-nodiv.minus.concur-free.writing}{$-$8.8}
\setval{4b.grpo-nodiv.minus.concur-free.writing-abs}{8.8}
\setval{4b.grpo-nodiv.minus.concur-free.writingbench}{$-$11.7}
\setval{4b.grpo-nodiv.minus.concur-free.writingbench-abs}{11.7}
\setval{4b.grpo-nodiv.minus.concur-tied.aime24}{$-$7.1}
\setval{4b.grpo-nodiv.minus.concur-tied.aime24-abs}{7.1}
\setval{4b.grpo-nodiv.minus.concur-tied.aime25}{$-$8.0}
\setval{4b.grpo-nodiv.minus.concur-tied.aime25-abs}{8.0}
\setval{4b.grpo-nodiv.minus.concur-tied.amc23}{$-$14.7}
\setval{4b.grpo-nodiv.minus.concur-tied.amc23-abs}{14.7}
\setval{4b.grpo-nodiv.minus.concur-tied.avg}{$-$5.7}
\setval{4b.grpo-nodiv.minus.concur-tied.avg-abs}{5.7}
\setval{4b.grpo-nodiv.minus.concur-tied.code}{$-$0.4}
\setval{4b.grpo-nodiv.minus.concur-tied.code-abs}{0.4}
\setval{4b.grpo-nodiv.minus.concur-tied.creative-writing-v3}{$-$5.3}
\setval{4b.grpo-nodiv.minus.concur-tied.creative-writing-v3-abs}{5.3}
\setval{4b.grpo-nodiv.minus.concur-tied.guidebench}{$-$0.4}
\setval{4b.grpo-nodiv.minus.concur-tied.guidebench-abs}{0.4}
\setval{4b.grpo-nodiv.minus.concur-tied.humaneval-plus}{$-$3.5}
\setval{4b.grpo-nodiv.minus.concur-tied.humaneval-plus-abs}{3.5}
\setval{4b.grpo-nodiv.minus.concur-tied.if}{$-$1.6}
\setval{4b.grpo-nodiv.minus.concur-tied.if-abs}{1.6}
\setval{4b.grpo-nodiv.minus.concur-tied.ifbench}{$-$4.0}
\setval{4b.grpo-nodiv.minus.concur-tied.ifbench-abs}{4.0}
\setval{4b.grpo-nodiv.minus.concur-tied.ifeval}{$-$0.5}
\setval{4b.grpo-nodiv.minus.concur-tied.ifeval-abs}{0.5}
\setval{4b.grpo-nodiv.minus.concur-tied.judge2.avg}{$-$4.8}
\setval{4b.grpo-nodiv.minus.concur-tied.judge2.avg-abs}{4.8}
\setval{4b.grpo-nodiv.minus.concur-tied.judge2.creative-writing-v3}{$-$3.5}
\setval{4b.grpo-nodiv.minus.concur-tied.judge2.creative-writing-v3-abs}{3.5}
\setval{4b.grpo-nodiv.minus.concur-tied.judge2.longbench-write}{$-$9.4}
\setval{4b.grpo-nodiv.minus.concur-tied.judge2.longbench-write-abs}{9.4}
\setval{4b.grpo-nodiv.minus.concur-tied.judge2.writing}{$-$6.2}
\setval{4b.grpo-nodiv.minus.concur-tied.judge2.writing-abs}{6.2}
\setval{4b.grpo-nodiv.minus.concur-tied.judge2.writingbench}{$-$5.8}
\setval{4b.grpo-nodiv.minus.concur-tied.judge2.writingbench-abs}{5.8}
\setval{4b.grpo-nodiv.minus.concur-tied.livecodebench}{$-$0.3}
\setval{4b.grpo-nodiv.minus.concur-tied.livecodebench-abs}{0.3}
\setval{4b.grpo-nodiv.minus.concur-tied.longbench-write}{$-$11.7}
\setval{4b.grpo-nodiv.minus.concur-tied.longbench-write-abs}{11.7}
\setval{4b.grpo-nodiv.minus.concur-tied.math}{$-$7.9}
\setval{4b.grpo-nodiv.minus.concur-tied.math-abs}{7.9}
\setval{4b.grpo-nodiv.minus.concur-tied.math500}{$-$6.0}
\setval{4b.grpo-nodiv.minus.concur-tied.math500-abs}{6.0}
\setval{4b.grpo-nodiv.minus.concur-tied.mbpp-plus}{+2.6}
\setval{4b.grpo-nodiv.minus.concur-tied.mbpp-plus-abs}{2.6}
\setval{4b.grpo-nodiv.minus.concur-tied.minerva}{$-$3.3}
\setval{4b.grpo-nodiv.minus.concur-tied.minerva-abs}{3.3}
\setval{4b.grpo-nodiv.minus.concur-tied.olympiad-bench}{$-$8.0}
\setval{4b.grpo-nodiv.minus.concur-tied.olympiad-bench-abs}{8.0}
\setval{4b.grpo-nodiv.minus.concur-tied.writing}{$-$10.5}
\setval{4b.grpo-nodiv.minus.concur-tied.writing-abs}{10.5}
\setval{4b.grpo-nodiv.minus.concur-tied.writingbench}{$-$14.6}
\setval{4b.grpo-nodiv.minus.concur-tied.writingbench-abs}{14.6}
\setval{4b.grpo-nodiv.minus.concur.aime24}{$-$8.1}
\setval{4b.grpo-nodiv.minus.concur.aime24-abs}{8.1}
\setval{4b.grpo-nodiv.minus.concur.aime25}{$-$10.4}
\setval{4b.grpo-nodiv.minus.concur.aime25-abs}{10.4}
\setval{4b.grpo-nodiv.minus.concur.amc23}{$-$14.8}
\setval{4b.grpo-nodiv.minus.concur.amc23-abs}{14.8}
\setval{4b.grpo-nodiv.minus.concur.avg}{$-$5.8}
\setval{4b.grpo-nodiv.minus.concur.avg-abs}{5.8}
\setval{4b.grpo-nodiv.minus.concur.code}{$-$1.0}
\setval{4b.grpo-nodiv.minus.concur.code-abs}{1.0}
\setval{4b.grpo-nodiv.minus.concur.creative-writing-v3}{$-$2.2}
\setval{4b.grpo-nodiv.minus.concur.creative-writing-v3-abs}{2.2}
\setval{4b.grpo-nodiv.minus.concur.guidebench}{+1.4}
\setval{4b.grpo-nodiv.minus.concur.guidebench-abs}{1.4}
\setval{4b.grpo-nodiv.minus.concur.humaneval-plus}{$-$2.0}
\setval{4b.grpo-nodiv.minus.concur.humaneval-plus-abs}{2.0}
\setval{4b.grpo-nodiv.minus.concur.if}{$-$2.0}
\setval{4b.grpo-nodiv.minus.concur.if-abs}{2.0}
\setval{4b.grpo-nodiv.minus.concur.ifbench}{$-$6.7}
\setval{4b.grpo-nodiv.minus.concur.ifbench-abs}{6.7}
\setval{4b.grpo-nodiv.minus.concur.ifeval}{$-$0.8}
\setval{4b.grpo-nodiv.minus.concur.ifeval-abs}{0.8}
\setval{4b.grpo-nodiv.minus.concur.judge2.avg}{$-$5.1}
\setval{4b.grpo-nodiv.minus.concur.judge2.avg-abs}{5.1}
\setval{4b.grpo-nodiv.minus.concur.judge2.creative-writing-v3}{$-$1.8}
\setval{4b.grpo-nodiv.minus.concur.judge2.creative-writing-v3-abs}{1.8}
\setval{4b.grpo-nodiv.minus.concur.judge2.longbench-write}{$-$8.0}
\setval{4b.grpo-nodiv.minus.concur.judge2.longbench-write-abs}{8.0}
\setval{4b.grpo-nodiv.minus.concur.judge2.writing}{$-$5.5}
\setval{4b.grpo-nodiv.minus.concur.judge2.writing-abs}{5.5}
\setval{4b.grpo-nodiv.minus.concur.judge2.writingbench}{$-$6.5}
\setval{4b.grpo-nodiv.minus.concur.judge2.writingbench-abs}{6.5}
\setval{4b.grpo-nodiv.minus.concur.livecodebench}{$-$0.7}
\setval{4b.grpo-nodiv.minus.concur.livecodebench-abs}{0.7}
\setval{4b.grpo-nodiv.minus.concur.longbench-write}{$-$10.3}
\setval{4b.grpo-nodiv.minus.concur.longbench-write-abs}{10.3}
\setval{4b.grpo-nodiv.minus.concur.math}{$-$8.6}
\setval{4b.grpo-nodiv.minus.concur.math-abs}{8.6}
\setval{4b.grpo-nodiv.minus.concur.math500}{$-$5.8}
\setval{4b.grpo-nodiv.minus.concur.math500-abs}{5.8}
\setval{4b.grpo-nodiv.minus.concur.mbpp-plus}{$-$0.3}
\setval{4b.grpo-nodiv.minus.concur.mbpp-plus-abs}{0.3}
\setval{4b.grpo-nodiv.minus.concur.minerva}{$-$3.4}
\setval{4b.grpo-nodiv.minus.concur.minerva-abs}{3.4}
\setval{4b.grpo-nodiv.minus.concur.olympiad-bench}{$-$8.8}
\setval{4b.grpo-nodiv.minus.concur.olympiad-bench-abs}{8.8}
\setval{4b.grpo-nodiv.minus.concur.writing}{$-$8.8}
\setval{4b.grpo-nodiv.minus.concur.writing-abs}{8.8}
\setval{4b.grpo-nodiv.minus.concur.writingbench}{$-$13.8}
\setval{4b.grpo-nodiv.minus.concur.writingbench-abs}{13.8}
\setval{4b.grpo-nodiv.minus.gd2po.aime24}{+0.5}
\setval{4b.grpo-nodiv.minus.gd2po.aime24-abs}{0.5}
\setval{4b.grpo-nodiv.minus.gd2po.aime25}{+1.6}
\setval{4b.grpo-nodiv.minus.gd2po.aime25-abs}{1.6}
\setval{4b.grpo-nodiv.minus.gd2po.amc23}{$-$3.2}
\setval{4b.grpo-nodiv.minus.gd2po.amc23-abs}{3.2}
\setval{4b.grpo-nodiv.minus.gd2po.avg}{$-$1.1}
\setval{4b.grpo-nodiv.minus.gd2po.avg-abs}{1.1}
\setval{4b.grpo-nodiv.minus.gd2po.code}{$-$0.7}
\setval{4b.grpo-nodiv.minus.gd2po.code-abs}{0.7}
\setval{4b.grpo-nodiv.minus.gd2po.creative-writing-v3}{$-$2.1}
\setval{4b.grpo-nodiv.minus.gd2po.creative-writing-v3-abs}{2.1}
\setval{4b.grpo-nodiv.minus.gd2po.guidebench}{$-$0.3}
\setval{4b.grpo-nodiv.minus.gd2po.guidebench-abs}{0.3}
\setval{4b.grpo-nodiv.minus.gd2po.humaneval-plus}{$-$2.9}
\setval{4b.grpo-nodiv.minus.gd2po.humaneval-plus-abs}{2.9}
\setval{4b.grpo-nodiv.minus.gd2po.if}{$-$1.5}
\setval{4b.grpo-nodiv.minus.gd2po.if-abs}{1.5}
\setval{4b.grpo-nodiv.minus.gd2po.ifbench}{$-$4.0}
\setval{4b.grpo-nodiv.minus.gd2po.ifbench-abs}{4.0}
\setval{4b.grpo-nodiv.minus.gd2po.ifeval}{$-$0.4}
\setval{4b.grpo-nodiv.minus.gd2po.ifeval-abs}{0.4}
\setval{4b.grpo-nodiv.minus.gd2po.judge2.avg}{$-$0.9}
\setval{4b.grpo-nodiv.minus.gd2po.judge2.avg-abs}{0.9}
\setval{4b.grpo-nodiv.minus.gd2po.judge2.creative-writing-v3}{$-$2.4}
\setval{4b.grpo-nodiv.minus.gd2po.judge2.creative-writing-v3-abs}{2.4}
\setval{4b.grpo-nodiv.minus.gd2po.judge2.longbench-write}{$-$2.4}
\setval{4b.grpo-nodiv.minus.gd2po.judge2.longbench-write-abs}{2.4}
\setval{4b.grpo-nodiv.minus.gd2po.judge2.writing}{$-$2.6}
\setval{4b.grpo-nodiv.minus.gd2po.judge2.writing-abs}{2.6}
\setval{4b.grpo-nodiv.minus.gd2po.judge2.writingbench}{$-$3.0}
\setval{4b.grpo-nodiv.minus.gd2po.judge2.writingbench-abs}{3.0}
\setval{4b.grpo-nodiv.minus.gd2po.livecodebench}{+1.0}
\setval{4b.grpo-nodiv.minus.gd2po.livecodebench-abs}{1.0}
\setval{4b.grpo-nodiv.minus.gd2po.longbench-write}{$-$2.9}
\setval{4b.grpo-nodiv.minus.gd2po.longbench-write-abs}{2.9}
\setval{4b.grpo-nodiv.minus.gd2po.math}{+0.1}
\setval{4b.grpo-nodiv.minus.gd2po.math-abs}{0.1}
\setval{4b.grpo-nodiv.minus.gd2po.math500}{+1.8}
\setval{4b.grpo-nodiv.minus.gd2po.math500-abs}{1.8}
\setval{4b.grpo-nodiv.minus.gd2po.mbpp-plus}{$-$0.3}
\setval{4b.grpo-nodiv.minus.gd2po.mbpp-plus-abs}{0.3}
\setval{4b.grpo-nodiv.minus.gd2po.minerva}{$-$1.4}
\setval{4b.grpo-nodiv.minus.gd2po.minerva-abs}{1.4}
\setval{4b.grpo-nodiv.minus.gd2po.olympiad-bench}{+1.3}
\setval{4b.grpo-nodiv.minus.gd2po.olympiad-bench-abs}{1.3}
\setval{4b.grpo-nodiv.minus.gd2po.writing}{$-$3.2}
\setval{4b.grpo-nodiv.minus.gd2po.writing-abs}{3.2}
\setval{4b.grpo-nodiv.minus.gd2po.writingbench}{$-$4.5}
\setval{4b.grpo-nodiv.minus.gd2po.writingbench-abs}{4.5}
\setval{4b.grpo-nodiv.minus.gdpo-noaux.aime24}{$-$11.0}
\setval{4b.grpo-nodiv.minus.gdpo-noaux.aime24-abs}{11.0}
\setval{4b.grpo-nodiv.minus.gdpo-noaux.aime25}{$-$10.4}
\setval{4b.grpo-nodiv.minus.gdpo-noaux.aime25-abs}{10.4}
\setval{4b.grpo-nodiv.minus.gdpo-noaux.amc23}{$-$11.2}
\setval{4b.grpo-nodiv.minus.gdpo-noaux.amc23-abs}{11.2}
\setval{4b.grpo-nodiv.minus.gdpo-noaux.avg}{$-$3.6}
\setval{4b.grpo-nodiv.minus.gdpo-noaux.avg-abs}{3.6}
\setval{4b.grpo-nodiv.minus.gdpo-noaux.code}{$-$2.1}
\setval{4b.grpo-nodiv.minus.gdpo-noaux.code-abs}{2.1}
\setval{4b.grpo-nodiv.minus.gdpo-noaux.creative-writing-v3}{+3.2}
\setval{4b.grpo-nodiv.minus.gdpo-noaux.creative-writing-v3-abs}{3.2}
\setval{4b.grpo-nodiv.minus.gdpo-noaux.guidebench}{+2.9}
\setval{4b.grpo-nodiv.minus.gdpo-noaux.guidebench-abs}{2.9}
\setval{4b.grpo-nodiv.minus.gdpo-noaux.humaneval-plus}{$-$5.0}
\setval{4b.grpo-nodiv.minus.gdpo-noaux.humaneval-plus-abs}{5.0}
\setval{4b.grpo-nodiv.minus.gdpo-noaux.if}{$-$0.4}
\setval{4b.grpo-nodiv.minus.gdpo-noaux.if-abs}{0.4}
\setval{4b.grpo-nodiv.minus.gdpo-noaux.ifbench}{$-$6.7}
\setval{4b.grpo-nodiv.minus.gdpo-noaux.ifbench-abs}{6.7}
\setval{4b.grpo-nodiv.minus.gdpo-noaux.ifeval}{+2.5}
\setval{4b.grpo-nodiv.minus.gdpo-noaux.ifeval-abs}{2.5}
\setval{4b.grpo-nodiv.minus.gdpo-noaux.judge2.avg}{$-$2.6}
\setval{4b.grpo-nodiv.minus.gdpo-noaux.judge2.avg-abs}{2.6}
\setval{4b.grpo-nodiv.minus.gdpo-noaux.judge2.creative-writing-v3}{+7.2}
\setval{4b.grpo-nodiv.minus.gdpo-noaux.judge2.creative-writing-v3-abs}{7.2}
\setval{4b.grpo-nodiv.minus.gdpo-noaux.judge2.longbench-write}{+7.8}
\setval{4b.grpo-nodiv.minus.gdpo-noaux.judge2.longbench-write-abs}{7.8}
\setval{4b.grpo-nodiv.minus.gdpo-noaux.judge2.writing}{+6.2}
\setval{4b.grpo-nodiv.minus.gdpo-noaux.judge2.writing-abs}{6.2}
\setval{4b.grpo-nodiv.minus.gdpo-noaux.judge2.writingbench}{+3.5}
\setval{4b.grpo-nodiv.minus.gdpo-noaux.judge2.writingbench-abs}{3.5}
\setval{4b.grpo-nodiv.minus.gdpo-noaux.livecodebench}{$-$0.3}
\setval{4b.grpo-nodiv.minus.gdpo-noaux.livecodebench-abs}{0.3}
\setval{4b.grpo-nodiv.minus.gdpo-noaux.longbench-write}{+1.6}
\setval{4b.grpo-nodiv.minus.gdpo-noaux.longbench-write-abs}{1.6}
\setval{4b.grpo-nodiv.minus.gdpo-noaux.math}{$-$8.3}
\setval{4b.grpo-nodiv.minus.gdpo-noaux.math-abs}{8.3}
\setval{4b.grpo-nodiv.minus.gdpo-noaux.math500}{$-$5.3}
\setval{4b.grpo-nodiv.minus.gdpo-noaux.math500-abs}{5.3}
\setval{4b.grpo-nodiv.minus.gdpo-noaux.mbpp-plus}{$-$1.1}
\setval{4b.grpo-nodiv.minus.gdpo-noaux.mbpp-plus-abs}{1.1}
\setval{4b.grpo-nodiv.minus.gdpo-noaux.minerva}{$-$3.7}
\setval{4b.grpo-nodiv.minus.gdpo-noaux.minerva-abs}{3.7}
\setval{4b.grpo-nodiv.minus.gdpo-noaux.olympiad-bench}{$-$8.3}
\setval{4b.grpo-nodiv.minus.gdpo-noaux.olympiad-bench-abs}{8.3}
\setval{4b.grpo-nodiv.minus.gdpo-noaux.writing}{+1.0}
\setval{4b.grpo-nodiv.minus.gdpo-noaux.writing-abs}{1.0}
\setval{4b.grpo-nodiv.minus.gdpo-noaux.writingbench}{$-$1.7}
\setval{4b.grpo-nodiv.minus.gdpo-noaux.writingbench-abs}{1.7}
\setval{4b.grpo-nodiv.minus.gdpo-w0.5.aime24}{$-$2.2}
\setval{4b.grpo-nodiv.minus.gdpo-w0.5.aime24-abs}{2.2}
\setval{4b.grpo-nodiv.minus.gdpo-w0.5.aime25}{$-$2.0}
\setval{4b.grpo-nodiv.minus.gdpo-w0.5.aime25-abs}{2.0}
\setval{4b.grpo-nodiv.minus.gdpo-w0.5.amc23}{$-$3.8}
\setval{4b.grpo-nodiv.minus.gdpo-w0.5.amc23-abs}{3.8}
\setval{4b.grpo-nodiv.minus.gdpo-w0.5.avg}{$-$1.6}
\setval{4b.grpo-nodiv.minus.gdpo-w0.5.avg-abs}{1.6}
\setval{4b.grpo-nodiv.minus.gdpo-w0.5.code}{$-$0.4}
\setval{4b.grpo-nodiv.minus.gdpo-w0.5.code-abs}{0.4}
\setval{4b.grpo-nodiv.minus.gdpo-w0.5.creative-writing-v3}{+4.4}
\setval{4b.grpo-nodiv.minus.gdpo-w0.5.creative-writing-v3-abs}{4.4}
\setval{4b.grpo-nodiv.minus.gdpo-w0.5.guidebench}{$-$0.8}
\setval{4b.grpo-nodiv.minus.gdpo-w0.5.guidebench-abs}{0.8}
\setval{4b.grpo-nodiv.minus.gdpo-w0.5.humaneval-plus}{$-$1.5}
\setval{4b.grpo-nodiv.minus.gdpo-w0.5.humaneval-plus-abs}{1.5}
\setval{4b.grpo-nodiv.minus.gdpo-w0.5.if}{$-$2.0}
\setval{4b.grpo-nodiv.minus.gdpo-w0.5.if-abs}{2.0}
\setval{4b.grpo-nodiv.minus.gdpo-w0.5.ifbench}{$-$5.7}
\setval{4b.grpo-nodiv.minus.gdpo-w0.5.ifbench-abs}{5.7}
\setval{4b.grpo-nodiv.minus.gdpo-w0.5.ifeval}{+0.5}
\setval{4b.grpo-nodiv.minus.gdpo-w0.5.ifeval-abs}{0.5}
\setval{4b.grpo-nodiv.minus.gdpo-w0.5.judge2.avg}{$-$1.8}
\setval{4b.grpo-nodiv.minus.gdpo-w0.5.judge2.avg-abs}{1.8}
\setval{4b.grpo-nodiv.minus.gdpo-w0.5.judge2.creative-writing-v3}{+0.5}
\setval{4b.grpo-nodiv.minus.gdpo-w0.5.judge2.creative-writing-v3-abs}{0.5}
\setval{4b.grpo-nodiv.minus.gdpo-w0.5.judge2.longbench-write}{$-$1.9}
\setval{4b.grpo-nodiv.minus.gdpo-w0.5.judge2.longbench-write-abs}{1.9}
\setval{4b.grpo-nodiv.minus.gdpo-w0.5.judge2.writing}{$-$1.8}
\setval{4b.grpo-nodiv.minus.gdpo-w0.5.judge2.writing-abs}{1.8}
\setval{4b.grpo-nodiv.minus.gdpo-w0.5.judge2.writingbench}{$-$4.0}
\setval{4b.grpo-nodiv.minus.gdpo-w0.5.judge2.writingbench-abs}{4.0}
\setval{4b.grpo-nodiv.minus.gdpo-w0.5.livecodebench}{+0.3}
\setval{4b.grpo-nodiv.minus.gdpo-w0.5.livecodebench-abs}{0.3}
\setval{4b.grpo-nodiv.minus.gdpo-w0.5.longbench-write}{$-$1.4}
\setval{4b.grpo-nodiv.minus.gdpo-w0.5.longbench-write-abs}{1.4}
\setval{4b.grpo-nodiv.minus.gdpo-w0.5.math}{$-$2.4}
\setval{4b.grpo-nodiv.minus.gdpo-w0.5.math-abs}{2.4}
\setval{4b.grpo-nodiv.minus.gdpo-w0.5.math500}{$-$2.3}
\setval{4b.grpo-nodiv.minus.gdpo-w0.5.math500-abs}{2.3}
\setval{4b.grpo-nodiv.minus.gdpo-w0.5.mbpp-plus}{+0.0}
\setval{4b.grpo-nodiv.minus.gdpo-w0.5.mbpp-plus-abs}{0.0}
\setval{4b.grpo-nodiv.minus.gdpo-w0.5.minerva}{$-$0.3}
\setval{4b.grpo-nodiv.minus.gdpo-w0.5.minerva-abs}{0.3}
\setval{4b.grpo-nodiv.minus.gdpo-w0.5.olympiad-bench}{$-$3.6}
\setval{4b.grpo-nodiv.minus.gdpo-w0.5.olympiad-bench-abs}{3.6}
\setval{4b.grpo-nodiv.minus.gdpo-w0.5.writing}{$-$0.7}
\setval{4b.grpo-nodiv.minus.gdpo-w0.5.writing-abs}{0.7}
\setval{4b.grpo-nodiv.minus.gdpo-w0.5.writingbench}{$-$5.1}
\setval{4b.grpo-nodiv.minus.gdpo-w0.5.writingbench-abs}{5.1}
\setval{4b.grpo-nodiv.minus.gdpo.aime24}{+3.3}
\setval{4b.grpo-nodiv.minus.gdpo.aime24-abs}{3.3}
\setval{4b.grpo-nodiv.minus.gdpo.aime25}{$-$1.6}
\setval{4b.grpo-nodiv.minus.gdpo.aime25-abs}{1.6}
\setval{4b.grpo-nodiv.minus.gdpo.amc23}{$-$5.1}
\setval{4b.grpo-nodiv.minus.gdpo.amc23-abs}{5.1}
\setval{4b.grpo-nodiv.minus.gdpo.avg}{$-$1.3}
\setval{4b.grpo-nodiv.minus.gdpo.avg-abs}{1.3}
\setval{4b.grpo-nodiv.minus.gdpo.code}{$-$0.3}
\setval{4b.grpo-nodiv.minus.gdpo.code-abs}{0.3}
\setval{4b.grpo-nodiv.minus.gdpo.creative-writing-v3}{$-$3.4}
\setval{4b.grpo-nodiv.minus.gdpo.creative-writing-v3-abs}{3.4}
\setval{4b.grpo-nodiv.minus.gdpo.guidebench}{$-$1.8}
\setval{4b.grpo-nodiv.minus.gdpo.guidebench-abs}{1.8}
\setval{4b.grpo-nodiv.minus.gdpo.humaneval-plus}{$-$1.3}
\setval{4b.grpo-nodiv.minus.gdpo.humaneval-plus-abs}{1.3}
\setval{4b.grpo-nodiv.minus.gdpo.if}{$-$1.0}
\setval{4b.grpo-nodiv.minus.gdpo.if-abs}{1.0}
\setval{4b.grpo-nodiv.minus.gdpo.ifbench}{$-$2.7}
\setval{4b.grpo-nodiv.minus.gdpo.ifbench-abs}{2.7}
\setval{4b.grpo-nodiv.minus.gdpo.ifeval}{+1.6}
\setval{4b.grpo-nodiv.minus.gdpo.ifeval-abs}{1.6}
\setval{4b.grpo-nodiv.minus.gdpo.judge2.avg}{$-$1.2}
\setval{4b.grpo-nodiv.minus.gdpo.judge2.avg-abs}{1.2}
\setval{4b.grpo-nodiv.minus.gdpo.judge2.creative-writing-v3}{$-$3.9}
\setval{4b.grpo-nodiv.minus.gdpo.judge2.creative-writing-v3-abs}{3.9}
\setval{4b.grpo-nodiv.minus.gdpo.judge2.longbench-write}{$-$3.7}
\setval{4b.grpo-nodiv.minus.gdpo.judge2.longbench-write-abs}{3.7}
\setval{4b.grpo-nodiv.minus.gdpo.judge2.writing}{$-$4.0}
\setval{4b.grpo-nodiv.minus.gdpo.judge2.writing-abs}{4.0}
\setval{4b.grpo-nodiv.minus.gdpo.judge2.writingbench}{$-$4.5}
\setval{4b.grpo-nodiv.minus.gdpo.judge2.writingbench-abs}{4.5}
\setval{4b.grpo-nodiv.minus.gdpo.livecodebench}{+0.1}
\setval{4b.grpo-nodiv.minus.gdpo.livecodebench-abs}{0.1}
\setval{4b.grpo-nodiv.minus.gdpo.longbench-write}{$-$3.0}
\setval{4b.grpo-nodiv.minus.gdpo.longbench-write-abs}{3.0}
\setval{4b.grpo-nodiv.minus.gdpo.math}{$-$0.4}
\setval{4b.grpo-nodiv.minus.gdpo.math-abs}{0.4}
\setval{4b.grpo-nodiv.minus.gdpo.math500}{+0.2}
\setval{4b.grpo-nodiv.minus.gdpo.math500-abs}{0.2}
\setval{4b.grpo-nodiv.minus.gdpo.mbpp-plus}{+0.3}
\setval{4b.grpo-nodiv.minus.gdpo.mbpp-plus-abs}{0.3}
\setval{4b.grpo-nodiv.minus.gdpo.minerva}{+0.3}
\setval{4b.grpo-nodiv.minus.gdpo.minerva-abs}{0.3}
\setval{4b.grpo-nodiv.minus.gdpo.olympiad-bench}{+0.1}
\setval{4b.grpo-nodiv.minus.gdpo.olympiad-bench-abs}{0.1}
\setval{4b.grpo-nodiv.minus.gdpo.writing}{$-$4.2}
\setval{4b.grpo-nodiv.minus.gdpo.writing-abs}{4.2}
\setval{4b.grpo-nodiv.minus.gdpo.writingbench}{$-$6.3}
\setval{4b.grpo-nodiv.minus.gdpo.writingbench-abs}{6.3}
\setval{4b.grpo-nodiv.minus.geoma.aime24}{+0.8}
\setval{4b.grpo-nodiv.minus.geoma.aime24-abs}{0.8}
\setval{4b.grpo-nodiv.minus.geoma.aime25}{$-$2.3}
\setval{4b.grpo-nodiv.minus.geoma.aime25-abs}{2.3}
\setval{4b.grpo-nodiv.minus.geoma.amc23}{$-$0.9}
\setval{4b.grpo-nodiv.minus.geoma.amc23-abs}{0.9}
\setval{4b.grpo-nodiv.minus.geoma.avg}{$-$0.8}
\setval{4b.grpo-nodiv.minus.geoma.avg-abs}{0.8}
\setval{4b.grpo-nodiv.minus.geoma.code}{+0.0}
\setval{4b.grpo-nodiv.minus.geoma.code-abs}{0.0}
\setval{4b.grpo-nodiv.minus.geoma.creative-writing-v3}{$-$3.3}
\setval{4b.grpo-nodiv.minus.geoma.creative-writing-v3-abs}{3.3}
\setval{4b.grpo-nodiv.minus.geoma.guidebench}{+0.3}
\setval{4b.grpo-nodiv.minus.geoma.guidebench-abs}{0.3}
\setval{4b.grpo-nodiv.minus.geoma.humaneval-plus}{+0.2}
\setval{4b.grpo-nodiv.minus.geoma.humaneval-plus-abs}{0.2}
\setval{4b.grpo-nodiv.minus.geoma.if}{$-$0.8}
\setval{4b.grpo-nodiv.minus.geoma.if-abs}{0.8}
\setval{4b.grpo-nodiv.minus.geoma.ifbench}{$-$2.3}
\setval{4b.grpo-nodiv.minus.geoma.ifbench-abs}{2.3}
\setval{4b.grpo-nodiv.minus.geoma.ifeval}{$-$0.5}
\setval{4b.grpo-nodiv.minus.geoma.ifeval-abs}{0.5}
\setval{4b.grpo-nodiv.minus.geoma.judge2.avg}{$-$0.5}
\setval{4b.grpo-nodiv.minus.geoma.judge2.avg-abs}{0.5}
\setval{4b.grpo-nodiv.minus.geoma.judge2.creative-writing-v3}{$-$2.6}
\setval{4b.grpo-nodiv.minus.geoma.judge2.creative-writing-v3-abs}{2.6}
\setval{4b.grpo-nodiv.minus.geoma.judge2.longbench-write}{+0.9}
\setval{4b.grpo-nodiv.minus.geoma.judge2.longbench-write-abs}{0.9}
\setval{4b.grpo-nodiv.minus.geoma.judge2.writing}{$-$0.9}
\setval{4b.grpo-nodiv.minus.geoma.judge2.writing-abs}{0.9}
\setval{4b.grpo-nodiv.minus.geoma.judge2.writingbench}{$-$1.1}
\setval{4b.grpo-nodiv.minus.geoma.judge2.writingbench-abs}{1.1}
\setval{4b.grpo-nodiv.minus.geoma.livecodebench}{$-$1.2}
\setval{4b.grpo-nodiv.minus.geoma.livecodebench-abs}{1.2}
\setval{4b.grpo-nodiv.minus.geoma.longbench-write}{+0.2}
\setval{4b.grpo-nodiv.minus.geoma.longbench-write-abs}{0.2}
\setval{4b.grpo-nodiv.minus.geoma.math}{$-$0.3}
\setval{4b.grpo-nodiv.minus.geoma.math-abs}{0.3}
\setval{4b.grpo-nodiv.minus.geoma.math500}{+0.6}
\setval{4b.grpo-nodiv.minus.geoma.math500-abs}{0.6}
\setval{4b.grpo-nodiv.minus.geoma.mbpp-plus}{+1.0}
\setval{4b.grpo-nodiv.minus.geoma.mbpp-plus-abs}{1.0}
\setval{4b.grpo-nodiv.minus.geoma.minerva}{$-$1.1}
\setval{4b.grpo-nodiv.minus.geoma.minerva-abs}{1.1}
\setval{4b.grpo-nodiv.minus.geoma.olympiad-bench}{+1.0}
\setval{4b.grpo-nodiv.minus.geoma.olympiad-bench-abs}{1.0}
\setval{4b.grpo-nodiv.minus.geoma.writing}{$-$2.3}
\setval{4b.grpo-nodiv.minus.geoma.writing-abs}{2.3}
\setval{4b.grpo-nodiv.minus.geoma.writingbench}{$-$3.9}
\setval{4b.grpo-nodiv.minus.geoma.writingbench-abs}{3.9}
\setval{4b.grpo-nodiv.minus.grpo.aime24}{+0.4}
\setval{4b.grpo-nodiv.minus.grpo.aime24-abs}{0.4}
\setval{4b.grpo-nodiv.minus.grpo.aime25}{$-$4.3}
\setval{4b.grpo-nodiv.minus.grpo.aime25-abs}{4.3}
\setval{4b.grpo-nodiv.minus.grpo.amc23}{$-$0.2}
\setval{4b.grpo-nodiv.minus.grpo.amc23-abs}{0.2}
\setval{4b.grpo-nodiv.minus.grpo.avg}{$-$1.6}
\setval{4b.grpo-nodiv.minus.grpo.avg-abs}{1.6}
\setval{4b.grpo-nodiv.minus.grpo.code}{$-$0.6}
\setval{4b.grpo-nodiv.minus.grpo.code-abs}{0.6}
\setval{4b.grpo-nodiv.minus.grpo.creative-writing-v3}{$-$4.8}
\setval{4b.grpo-nodiv.minus.grpo.creative-writing-v3-abs}{4.8}
\setval{4b.grpo-nodiv.minus.grpo.guidebench}{+0.9}
\setval{4b.grpo-nodiv.minus.grpo.guidebench-abs}{0.9}
\setval{4b.grpo-nodiv.minus.grpo.humaneval-plus}{$-$2.3}
\setval{4b.grpo-nodiv.minus.grpo.humaneval-plus-abs}{2.3}
\setval{4b.grpo-nodiv.minus.grpo.if}{$-$0.8}
\setval{4b.grpo-nodiv.minus.grpo.if-abs}{0.8}
\setval{4b.grpo-nodiv.minus.grpo.ifbench}{$-$4.0}
\setval{4b.grpo-nodiv.minus.grpo.ifbench-abs}{4.0}
\setval{4b.grpo-nodiv.minus.grpo.ifeval}{+0.8}
\setval{4b.grpo-nodiv.minus.grpo.ifeval-abs}{0.8}
\setval{4b.grpo-nodiv.minus.grpo.judge2.avg}{$-$1.2}
\setval{4b.grpo-nodiv.minus.grpo.judge2.avg-abs}{1.2}
\setval{4b.grpo-nodiv.minus.grpo.judge2.creative-writing-v3}{$-$4.6}
\setval{4b.grpo-nodiv.minus.grpo.judge2.creative-writing-v3-abs}{4.6}
\setval{4b.grpo-nodiv.minus.grpo.judge2.longbench-write}{$-$2.1}
\setval{4b.grpo-nodiv.minus.grpo.judge2.longbench-write-abs}{2.1}
\setval{4b.grpo-nodiv.minus.grpo.judge2.writing}{$-$3.5}
\setval{4b.grpo-nodiv.minus.grpo.judge2.writing-abs}{3.5}
\setval{4b.grpo-nodiv.minus.grpo.judge2.writingbench}{$-$3.9}
\setval{4b.grpo-nodiv.minus.grpo.judge2.writingbench-abs}{3.9}
\setval{4b.grpo-nodiv.minus.grpo.livecodebench}{+0.5}
\setval{4b.grpo-nodiv.minus.grpo.livecodebench-abs}{0.5}
\setval{4b.grpo-nodiv.minus.grpo.longbench-write}{$-$3.2}
\setval{4b.grpo-nodiv.minus.grpo.longbench-write-abs}{3.2}
\setval{4b.grpo-nodiv.minus.grpo.math}{$-$0.7}
\setval{4b.grpo-nodiv.minus.grpo.math-abs}{0.7}
\setval{4b.grpo-nodiv.minus.grpo.math500}{$-$0.8}
\setval{4b.grpo-nodiv.minus.grpo.math500-abs}{0.8}
\setval{4b.grpo-nodiv.minus.grpo.mbpp-plus}{+0.0}
\setval{4b.grpo-nodiv.minus.grpo.mbpp-plus-abs}{0.0}
\setval{4b.grpo-nodiv.minus.grpo.minerva}{+0.5}
\setval{4b.grpo-nodiv.minus.grpo.minerva-abs}{0.5}
\setval{4b.grpo-nodiv.minus.grpo.olympiad-bench}{+0.2}
\setval{4b.grpo-nodiv.minus.grpo.olympiad-bench-abs}{0.2}
\setval{4b.grpo-nodiv.minus.grpo.writing}{$-$5.5}
\setval{4b.grpo-nodiv.minus.grpo.writing-abs}{5.5}
\setval{4b.grpo-nodiv.minus.grpo.writingbench}{$-$8.6}
\setval{4b.grpo-nodiv.minus.grpo.writingbench-abs}{8.6}
\setval{4b.grpo-nodiv.minus.mt-grpo.aime24}{+1.7}
\setval{4b.grpo-nodiv.minus.mt-grpo.aime24-abs}{1.7}
\setval{4b.grpo-nodiv.minus.mt-grpo.aime25}{+2.2}
\setval{4b.grpo-nodiv.minus.mt-grpo.aime25-abs}{2.2}
\setval{4b.grpo-nodiv.minus.mt-grpo.amc23}{$-$5.2}
\setval{4b.grpo-nodiv.minus.mt-grpo.amc23-abs}{5.2}
\setval{4b.grpo-nodiv.minus.mt-grpo.avg}{+0.0}
\setval{4b.grpo-nodiv.minus.mt-grpo.avg-abs}{0.0}
\setval{4b.grpo-nodiv.minus.mt-grpo.code}{+1.5}
\setval{4b.grpo-nodiv.minus.mt-grpo.code-abs}{1.5}
\setval{4b.grpo-nodiv.minus.mt-grpo.creative-writing-v3}{$-$2.0}
\setval{4b.grpo-nodiv.minus.mt-grpo.creative-writing-v3-abs}{2.0}
\setval{4b.grpo-nodiv.minus.mt-grpo.guidebench}{$-$0.9}
\setval{4b.grpo-nodiv.minus.mt-grpo.guidebench-abs}{0.9}
\setval{4b.grpo-nodiv.minus.mt-grpo.humaneval-plus}{+1.4}
\setval{4b.grpo-nodiv.minus.mt-grpo.humaneval-plus-abs}{1.4}
\setval{4b.grpo-nodiv.minus.mt-grpo.if}{$-$0.3}
\setval{4b.grpo-nodiv.minus.mt-grpo.if-abs}{0.3}
\setval{4b.grpo-nodiv.minus.mt-grpo.ifbench}{$-$0.7}
\setval{4b.grpo-nodiv.minus.mt-grpo.ifbench-abs}{0.7}
\setval{4b.grpo-nodiv.minus.mt-grpo.ifeval}{+0.6}
\setval{4b.grpo-nodiv.minus.mt-grpo.ifeval-abs}{0.6}
\setval{4b.grpo-nodiv.minus.mt-grpo.judge2.avg}{$-$0.2}
\setval{4b.grpo-nodiv.minus.mt-grpo.judge2.avg-abs}{0.2}
\setval{4b.grpo-nodiv.minus.mt-grpo.judge2.creative-writing-v3}{$-$3.6}
\setval{4b.grpo-nodiv.minus.mt-grpo.judge2.creative-writing-v3-abs}{3.6}
\setval{4b.grpo-nodiv.minus.mt-grpo.judge2.longbench-write}{$-$2.2}
\setval{4b.grpo-nodiv.minus.mt-grpo.judge2.longbench-write-abs}{2.2}
\setval{4b.grpo-nodiv.minus.mt-grpo.judge2.writing}{$-$3.1}
\setval{4b.grpo-nodiv.minus.mt-grpo.judge2.writing-abs}{3.1}
\setval{4b.grpo-nodiv.minus.mt-grpo.judge2.writingbench}{$-$3.7}
\setval{4b.grpo-nodiv.minus.mt-grpo.judge2.writingbench-abs}{3.7}
\setval{4b.grpo-nodiv.minus.mt-grpo.livecodebench}{+1.8}
\setval{4b.grpo-nodiv.minus.mt-grpo.livecodebench-abs}{1.8}
\setval{4b.grpo-nodiv.minus.mt-grpo.longbench-write}{$-$0.8}
\setval{4b.grpo-nodiv.minus.mt-grpo.longbench-write-abs}{0.8}
\setval{4b.grpo-nodiv.minus.mt-grpo.math}{+0.5}
\setval{4b.grpo-nodiv.minus.mt-grpo.math-abs}{0.5}
\setval{4b.grpo-nodiv.minus.mt-grpo.math500}{+2.5}
\setval{4b.grpo-nodiv.minus.mt-grpo.math500-abs}{2.5}
\setval{4b.grpo-nodiv.minus.mt-grpo.mbpp-plus}{+1.1}
\setval{4b.grpo-nodiv.minus.mt-grpo.mbpp-plus-abs}{1.1}
\setval{4b.grpo-nodiv.minus.mt-grpo.minerva}{$-$0.2}
\setval{4b.grpo-nodiv.minus.mt-grpo.minerva-abs}{0.2}
\setval{4b.grpo-nodiv.minus.mt-grpo.olympiad-bench}{+2.3}
\setval{4b.grpo-nodiv.minus.mt-grpo.olympiad-bench-abs}{2.3}
\setval{4b.grpo-nodiv.minus.mt-grpo.writing}{$-$2.4}
\setval{4b.grpo-nodiv.minus.mt-grpo.writing-abs}{2.4}
\setval{4b.grpo-nodiv.minus.mt-grpo.writingbench}{$-$4.5}
\setval{4b.grpo-nodiv.minus.mt-grpo.writingbench-abs}{4.5}
\setval{4b.grpo-nodiv.minus.papo.aime24}{+3.1}
\setval{4b.grpo-nodiv.minus.papo.aime24-abs}{3.1}
\setval{4b.grpo-nodiv.minus.papo.aime25}{+0.8}
\setval{4b.grpo-nodiv.minus.papo.aime25-abs}{0.8}
\setval{4b.grpo-nodiv.minus.papo.amc23}{+1.0}
\setval{4b.grpo-nodiv.minus.papo.amc23-abs}{1.0}
\setval{4b.grpo-nodiv.minus.papo.avg}{+0.4}
\setval{4b.grpo-nodiv.minus.papo.avg-abs}{0.4}
\setval{4b.grpo-nodiv.minus.papo.code}{$-$0.4}
\setval{4b.grpo-nodiv.minus.papo.code-abs}{0.4}
\setval{4b.grpo-nodiv.minus.papo.creative-writing-v3}{$-$4.0}
\setval{4b.grpo-nodiv.minus.papo.creative-writing-v3-abs}{4.0}
\setval{4b.grpo-nodiv.minus.papo.guidebench}{$-$1.9}
\setval{4b.grpo-nodiv.minus.papo.guidebench-abs}{1.9}
\setval{4b.grpo-nodiv.minus.papo.humaneval-plus}{$-$2.6}
\setval{4b.grpo-nodiv.minus.papo.humaneval-plus-abs}{2.6}
\setval{4b.grpo-nodiv.minus.papo.if}{+0.0}
\setval{4b.grpo-nodiv.minus.papo.if-abs}{0.0}
\setval{4b.grpo-nodiv.minus.papo.ifbench}{+0.3}
\setval{4b.grpo-nodiv.minus.papo.ifbench-abs}{0.3}
\setval{4b.grpo-nodiv.minus.papo.ifeval}{+1.4}
\setval{4b.grpo-nodiv.minus.papo.ifeval-abs}{1.4}
\setval{4b.grpo-nodiv.minus.papo.judge2.avg}{+0.3}
\setval{4b.grpo-nodiv.minus.papo.judge2.avg-abs}{0.3}
\setval{4b.grpo-nodiv.minus.papo.judge2.creative-writing-v3}{$-$3.7}
\setval{4b.grpo-nodiv.minus.papo.judge2.creative-writing-v3-abs}{3.7}
\setval{4b.grpo-nodiv.minus.papo.judge2.longbench-write}{+1.1}
\setval{4b.grpo-nodiv.minus.papo.judge2.longbench-write-abs}{1.1}
\setval{4b.grpo-nodiv.minus.papo.judge2.writing}{$-$1.7}
\setval{4b.grpo-nodiv.minus.papo.judge2.writing-abs}{1.7}
\setval{4b.grpo-nodiv.minus.papo.judge2.writingbench}{$-$2.6}
\setval{4b.grpo-nodiv.minus.papo.judge2.writingbench-abs}{2.6}
\setval{4b.grpo-nodiv.minus.papo.livecodebench}{+0.9}
\setval{4b.grpo-nodiv.minus.papo.livecodebench-abs}{0.9}
\setval{4b.grpo-nodiv.minus.papo.longbench-write}{+2.0}
\setval{4b.grpo-nodiv.minus.papo.longbench-write-abs}{2.0}
\setval{4b.grpo-nodiv.minus.papo.math}{+1.8}
\setval{4b.grpo-nodiv.minus.papo.math-abs}{1.8}
\setval{4b.grpo-nodiv.minus.papo.math500}{+3.0}
\setval{4b.grpo-nodiv.minus.papo.math500-abs}{3.0}
\setval{4b.grpo-nodiv.minus.papo.mbpp-plus}{+0.5}
\setval{4b.grpo-nodiv.minus.papo.mbpp-plus-abs}{0.5}
\setval{4b.grpo-nodiv.minus.papo.minerva}{$-$0.7}
\setval{4b.grpo-nodiv.minus.papo.minerva-abs}{0.7}
\setval{4b.grpo-nodiv.minus.papo.olympiad-bench}{+3.3}
\setval{4b.grpo-nodiv.minus.papo.olympiad-bench-abs}{3.3}
\setval{4b.grpo-nodiv.minus.papo.writing}{$-$0.9}
\setval{4b.grpo-nodiv.minus.papo.writing-abs}{0.9}
\setval{4b.grpo-nodiv.minus.papo.writingbench}{$-$0.9}
\setval{4b.grpo-nodiv.minus.papo.writingbench-abs}{0.9}
\setval{4b.grpo-nodiv.minus.saw.aime24}{$-$0.1}
\setval{4b.grpo-nodiv.minus.saw.aime24-abs}{0.1}
\setval{4b.grpo-nodiv.minus.saw.aime25}{$-$2.8}
\setval{4b.grpo-nodiv.minus.saw.aime25-abs}{2.8}
\setval{4b.grpo-nodiv.minus.saw.amc23}{$-$2.1}
\setval{4b.grpo-nodiv.minus.saw.amc23-abs}{2.1}
\setval{4b.grpo-nodiv.minus.saw.avg}{$-$1.2}
\setval{4b.grpo-nodiv.minus.saw.avg-abs}{1.2}
\setval{4b.grpo-nodiv.minus.saw.code}{+0.8}
\setval{4b.grpo-nodiv.minus.saw.code-abs}{0.8}
\setval{4b.grpo-nodiv.minus.saw.creative-writing-v3}{$-$2.5}
\setval{4b.grpo-nodiv.minus.saw.creative-writing-v3-abs}{2.5}
\setval{4b.grpo-nodiv.minus.saw.guidebench}{$-$2.2}
\setval{4b.grpo-nodiv.minus.saw.guidebench-abs}{2.2}
\setval{4b.grpo-nodiv.minus.saw.humaneval-plus}{$-$0.2}
\setval{4b.grpo-nodiv.minus.saw.humaneval-plus-abs}{0.2}
\setval{4b.grpo-nodiv.minus.saw.if}{$-$2.8}
\setval{4b.grpo-nodiv.minus.saw.if-abs}{2.8}
\setval{4b.grpo-nodiv.minus.saw.ifbench}{$-$6.7}
\setval{4b.grpo-nodiv.minus.saw.ifbench-abs}{6.7}
\setval{4b.grpo-nodiv.minus.saw.ifeval}{+0.4}
\setval{4b.grpo-nodiv.minus.saw.ifeval-abs}{0.4}
\setval{4b.grpo-nodiv.minus.saw.judge2.avg}{$-$1.4}
\setval{4b.grpo-nodiv.minus.saw.judge2.avg-abs}{1.4}
\setval{4b.grpo-nodiv.minus.saw.judge2.creative-writing-v3}{$-$5.1}
\setval{4b.grpo-nodiv.minus.saw.judge2.creative-writing-v3-abs}{5.1}
\setval{4b.grpo-nodiv.minus.saw.judge2.longbench-write}{$-$1.4}
\setval{4b.grpo-nodiv.minus.saw.judge2.longbench-write-abs}{1.4}
\setval{4b.grpo-nodiv.minus.saw.judge2.writing}{$-$3.4}
\setval{4b.grpo-nodiv.minus.saw.judge2.writing-abs}{3.4}
\setval{4b.grpo-nodiv.minus.saw.judge2.writingbench}{$-$3.8}
\setval{4b.grpo-nodiv.minus.saw.judge2.writingbench-abs}{3.8}
\setval{4b.grpo-nodiv.minus.saw.livecodebench}{+1.7}
\setval{4b.grpo-nodiv.minus.saw.livecodebench-abs}{1.7}
\setval{4b.grpo-nodiv.minus.saw.longbench-write}{$-$0.2}
\setval{4b.grpo-nodiv.minus.saw.longbench-write-abs}{0.2}
\setval{4b.grpo-nodiv.minus.saw.math}{$-$0.8}
\setval{4b.grpo-nodiv.minus.saw.math-abs}{0.8}
\setval{4b.grpo-nodiv.minus.saw.math500}{+1.7}
\setval{4b.grpo-nodiv.minus.saw.math500-abs}{1.7}
\setval{4b.grpo-nodiv.minus.saw.mbpp-plus}{+1.1}
\setval{4b.grpo-nodiv.minus.saw.mbpp-plus-abs}{1.1}
\setval{4b.grpo-nodiv.minus.saw.minerva}{$-$2.3}
\setval{4b.grpo-nodiv.minus.saw.minerva-abs}{2.3}
\setval{4b.grpo-nodiv.minus.saw.olympiad-bench}{+1.0}
\setval{4b.grpo-nodiv.minus.saw.olympiad-bench-abs}{1.0}
\setval{4b.grpo-nodiv.minus.saw.writing}{$-$2.3}
\setval{4b.grpo-nodiv.minus.saw.writing-abs}{2.3}
\setval{4b.grpo-nodiv.minus.saw.writingbench}{$-$4.2}
\setval{4b.grpo-nodiv.minus.saw.writingbench-abs}{4.2}
\setval{4b.grpo-nodiv.passing-at-ceiling.code.early}{0.71}
\setval{4b.grpo-nodiv.passing-at-ceiling.code.late}{0.73}
\setval{4b.grpo-nodiv.passing-at-ceiling.code.run}{0.73}
\setval{4b.grpo-nodiv.passing-at-ceiling.if.early}{0.28}
\setval{4b.grpo-nodiv.passing-at-ceiling.if.late}{0.52}
\setval{4b.grpo-nodiv.passing-at-ceiling.if.run}{0.46}
\setval{4b.grpo-nodiv.passing-at-ceiling.math.early}{1.00}
\setval{4b.grpo-nodiv.passing-at-ceiling.math.late}{1.00}
\setval{4b.grpo-nodiv.passing-at-ceiling.math.run}{1.00}
\setval{4b.grpo-nodiv.passing-at-ceiling.writing.early}{0.00}
\setval{4b.grpo-nodiv.passing-at-ceiling.writing.late}{0.00}
\setval{4b.grpo-nodiv.passing-at-ceiling.writing.run}{0.00}
\setval{4b.grpo-nodiv.primary-level.code.early}{0.54}
\setval{4b.grpo-nodiv.primary-level.code.late}{0.76}
\setval{4b.grpo-nodiv.primary-level.code.run}{0.71}
\setval{4b.grpo-nodiv.primary-level.if.early}{0.31}
\setval{4b.grpo-nodiv.primary-level.if.late}{0.72}
\setval{4b.grpo-nodiv.primary-level.if.run}{0.61}
\setval{4b.grpo-nodiv.primary-level.math.early}{0.48}
\setval{4b.grpo-nodiv.primary-level.math.late}{0.70}
\setval{4b.grpo-nodiv.primary-level.math.run}{0.64}
\setval{4b.grpo-nodiv.primary-level.writing.early}{0.52}
\setval{4b.grpo-nodiv.primary-level.writing.late}{0.74}
\setval{4b.grpo-nodiv.primary-level.writing.run}{0.69}
\setval{4b.grpo.aux-change.code}{+0.20}
\setval{4b.grpo.aux-change.if}{+0.23}
\setval{4b.grpo.aux-change.math}{+0.13}
\setval{4b.grpo.aux-change.writing}{+0.20}
\setval{4b.grpo.aux-change.writing-rm}{+0.37}
\setval{4b.grpo.aux-full.code.early}{0.77}
\setval{4b.grpo.aux-full.code.late}{0.96}
\setval{4b.grpo.aux-full.code.run}{0.95}
\setval{4b.grpo.aux-full.if.early}{0.66}
\setval{4b.grpo.aux-full.if.late}{0.89}
\setval{4b.grpo.aux-full.if.run}{0.85}
\setval{4b.grpo.aux-full.math.early}{0.66}
\setval{4b.grpo.aux-full.math.late}{0.91}
\setval{4b.grpo.aux-full.math.run}{0.88}
\setval{4b.grpo.aux-full.writing-rm.early}{0.09}
\setval{4b.grpo.aux-full.writing-rm.late}{0.59}
\setval{4b.grpo.aux-full.writing-rm.run}{0.47}
\setval{4b.grpo.aux-full.writing.early}{0.63}
\setval{4b.grpo.aux-full.writing.late}{0.71}
\setval{4b.grpo.aux-full.writing.run}{0.71}
\setval{4b.grpo.aux-level.code.early}{0.94}
\setval{4b.grpo.aux-level.code.late}{0.99}
\setval{4b.grpo.aux-level.code.run}{0.99}
\setval{4b.grpo.aux-level.if.early}{0.83}
\setval{4b.grpo.aux-level.if.late}{0.94}
\setval{4b.grpo.aux-level.if.run}{0.92}
\setval{4b.grpo.aux-level.math.early}{0.93}
\setval{4b.grpo.aux-level.math.late}{0.98}
\setval{4b.grpo.aux-level.math.run}{0.98}
\setval{4b.grpo.aux-level.writing-rm.early}{0.78}
\setval{4b.grpo.aux-level.writing-rm.late}{0.96}
\setval{4b.grpo.aux-level.writing-rm.run}{0.93}
\setval{4b.grpo.aux-level.writing.early}{0.91}
\setval{4b.grpo.aux-level.writing.late}{0.97}
\setval{4b.grpo.aux-level.writing.run}{0.96}
\setval{4b.grpo.bench.aime24}{12.0}
\setval{4b.grpo.bench.aime25}{12.6}
\setval{4b.grpo.bench.amc23}{54.7}
\setval{4b.grpo.bench.creative-writing-v3}{47.3}
\setval{4b.grpo.bench.guidebench}{77.9}
\setval{4b.grpo.bench.humaneval-plus}{82.7}
\setval{4b.grpo.bench.ifbench}{37.7}
\setval{4b.grpo.bench.ifeval}{86.5}
\setval{4b.grpo.bench.livecodebench}{24.0}
\setval{4b.grpo.bench.longbench-write}{69.2}
\setval{4b.grpo.bench.math500}{80.2}
\setval{4b.grpo.bench.mbpp-plus}{71.2}
\setval{4b.grpo.bench.minerva}{45.3}
\setval{4b.grpo.bench.olympiad-bench}{44.6}
\setval{4b.grpo.bench.writingbench}{74.3}
\setval{4b.grpo.eval.avg}{54.7}
\setval{4b.grpo.eval.code}{59.3}
\setval{4b.grpo.eval.if}{67.3}
\setval{4b.grpo.eval.math}{41.6}
\setval{4b.grpo.eval.writing}{63.6}
\setval{4b.grpo.judge2.avg}{51.9}
\setval{4b.grpo.judge2.creative-writing-v3}{35.1}
\setval{4b.grpo.judge2.longbench-write}{60.7}
\setval{4b.grpo.judge2.writing}{49.8}
\setval{4b.grpo.judge2.writingbench}{53.4}
\setval{4b.grpo.last-step}{300}
\setval{4b.grpo.minus.base.aime24}{+4.2}
\setval{4b.grpo.minus.base.aime24-abs}{4.2}
\setval{4b.grpo.minus.base.aime25}{+5.5}
\setval{4b.grpo.minus.base.aime25-abs}{5.5}
\setval{4b.grpo.minus.base.amc23}{+10.8}
\setval{4b.grpo.minus.base.amc23-abs}{10.8}
\setval{4b.grpo.minus.base.avg}{+20.1}
\setval{4b.grpo.minus.base.avg-abs}{20.1}
\setval{4b.grpo.minus.base.code}{+14.7}
\setval{4b.grpo.minus.base.code-abs}{14.7}
\setval{4b.grpo.minus.base.creative-writing-v3}{+22.6}
\setval{4b.grpo.minus.base.creative-writing-v3-abs}{22.6}
\setval{4b.grpo.minus.base.guidebench}{+53.1}
\setval{4b.grpo.minus.base.guidebench-abs}{53.1}
\setval{4b.grpo.minus.base.humaneval-plus}{+17.7}
\setval{4b.grpo.minus.base.humaneval-plus-abs}{17.7}
\setval{4b.grpo.minus.base.if}{+37.6}
\setval{4b.grpo.minus.base.if-abs}{37.6}
\setval{4b.grpo.minus.base.ifbench}{+20.7}
\setval{4b.grpo.minus.base.ifbench-abs}{20.7}
\setval{4b.grpo.minus.base.ifeval}{+39.1}
\setval{4b.grpo.minus.base.ifeval-abs}{39.1}
\setval{4b.grpo.minus.base.judge2.avg}{+18.0}
\setval{4b.grpo.minus.base.judge2.avg-abs}{18.0}
\setval{4b.grpo.minus.base.judge2.creative-writing-v3}{+14.2}
\setval{4b.grpo.minus.base.judge2.creative-writing-v3-abs}{14.2}
\setval{4b.grpo.minus.base.judge2.longbench-write}{+19.8}
\setval{4b.grpo.minus.base.judge2.longbench-write-abs}{19.8}
\setval{4b.grpo.minus.base.judge2.writing}{+19.0}
\setval{4b.grpo.minus.base.judge2.writing-abs}{19.0}
\setval{4b.grpo.minus.base.judge2.writingbench}{+23.0}
\setval{4b.grpo.minus.base.judge2.writingbench-abs}{23.0}
\setval{4b.grpo.minus.base.livecodebench}{+10.8}
\setval{4b.grpo.minus.base.livecodebench-abs}{10.8}
\setval{4b.grpo.minus.base.longbench-write}{+24.8}
\setval{4b.grpo.minus.base.longbench-write-abs}{24.8}
\setval{4b.grpo.minus.base.math}{+9.4}
\setval{4b.grpo.minus.base.math-abs}{9.4}
\setval{4b.grpo.minus.base.math500}{+13.0}
\setval{4b.grpo.minus.base.math500-abs}{13.0}
\setval{4b.grpo.minus.base.mbpp-plus}{+15.6}
\setval{4b.grpo.minus.base.mbpp-plus-abs}{15.6}
\setval{4b.grpo.minus.base.minerva}{+13.7}
\setval{4b.grpo.minus.base.minerva-abs}{13.7}
\setval{4b.grpo.minus.base.olympiad-bench}{+9.0}
\setval{4b.grpo.minus.base.olympiad-bench-abs}{9.0}
\setval{4b.grpo.minus.base.writing}{+29.7}
\setval{4b.grpo.minus.base.writing-abs}{29.7}
\setval{4b.grpo.minus.base.writingbench}{+41.8}
\setval{4b.grpo.minus.base.writingbench-abs}{41.8}
\setval{4b.grpo.minus.concur-free.aime24}{+0.1}
\setval{4b.grpo.minus.concur-free.aime24-abs}{0.1}
\setval{4b.grpo.minus.concur-free.aime25}{$-$2.0}
\setval{4b.grpo.minus.concur-free.aime25-abs}{2.0}
\setval{4b.grpo.minus.concur-free.amc23}{+0.5}
\setval{4b.grpo.minus.concur-free.amc23-abs}{0.5}
\setval{4b.grpo.minus.concur-free.avg}{$-$0.9}
\setval{4b.grpo.minus.concur-free.avg-abs}{0.9}
\setval{4b.grpo.minus.concur-free.code}{+1.0}
\setval{4b.grpo.minus.concur-free.code-abs}{1.0}
\setval{4b.grpo.minus.concur-free.creative-writing-v3}{+1.9}
\setval{4b.grpo.minus.concur-free.creative-writing-v3-abs}{1.9}
\setval{4b.grpo.minus.concur-free.guidebench}{$-$1.1}
\setval{4b.grpo.minus.concur-free.guidebench-abs}{1.1}
\setval{4b.grpo.minus.concur-free.humaneval-plus}{+1.6}
\setval{4b.grpo.minus.concur-free.humaneval-plus-abs}{1.6}
\setval{4b.grpo.minus.concur-free.if}{$-$0.9}
\setval{4b.grpo.minus.concur-free.if-abs}{0.9}
\setval{4b.grpo.minus.concur-free.ifbench}{$-$1.3}
\setval{4b.grpo.minus.concur-free.ifbench-abs}{1.3}
\setval{4b.grpo.minus.concur-free.ifeval}{$-$0.2}
\setval{4b.grpo.minus.concur-free.ifeval-abs}{0.2}
\setval{4b.grpo.minus.concur-free.judge2.avg}{$-$0.9}
\setval{4b.grpo.minus.concur-free.judge2.avg-abs}{0.9}
\setval{4b.grpo.minus.concur-free.judge2.creative-writing-v3}{$-$0.2}
\setval{4b.grpo.minus.concur-free.judge2.creative-writing-v3-abs}{0.2}
\setval{4b.grpo.minus.concur-free.judge2.longbench-write}{$-$8.6}
\setval{4b.grpo.minus.concur-free.judge2.longbench-write-abs}{8.6}
\setval{4b.grpo.minus.concur-free.judge2.writing}{$-$3.3}
\setval{4b.grpo.minus.concur-free.judge2.writing-abs}{3.3}
\setval{4b.grpo.minus.concur-free.judge2.writingbench}{$-$1.0}
\setval{4b.grpo.minus.concur-free.judge2.writingbench-abs}{1.0}
\setval{4b.grpo.minus.concur-free.livecodebench}{$-$0.2}
\setval{4b.grpo.minus.concur-free.livecodebench-abs}{0.2}
\setval{4b.grpo.minus.concur-free.longbench-write}{$-$8.6}
\setval{4b.grpo.minus.concur-free.longbench-write-abs}{8.6}
\setval{4b.grpo.minus.concur-free.math}{$-$0.6}
\setval{4b.grpo.minus.concur-free.math-abs}{0.6}
\setval{4b.grpo.minus.concur-free.math500}{$-$0.1}
\setval{4b.grpo.minus.concur-free.math500-abs}{0.1}
\setval{4b.grpo.minus.concur-free.mbpp-plus}{+1.5}
\setval{4b.grpo.minus.concur-free.mbpp-plus-abs}{1.5}
\setval{4b.grpo.minus.concur-free.minerva}{$-$1.0}
\setval{4b.grpo.minus.concur-free.minerva-abs}{1.0}
\setval{4b.grpo.minus.concur-free.olympiad-bench}{$-$1.1}
\setval{4b.grpo.minus.concur-free.olympiad-bench-abs}{1.1}
\setval{4b.grpo.minus.concur-free.writing}{$-$3.2}
\setval{4b.grpo.minus.concur-free.writing-abs}{3.2}
\setval{4b.grpo.minus.concur-free.writingbench}{$-$3.1}
\setval{4b.grpo.minus.concur-free.writingbench-abs}{3.1}
\setval{4b.grpo.minus.concur-tied.aime24}{$-$7.5}
\setval{4b.grpo.minus.concur-tied.aime24-abs}{7.5}
\setval{4b.grpo.minus.concur-tied.aime25}{$-$3.8}
\setval{4b.grpo.minus.concur-tied.aime25-abs}{3.8}
\setval{4b.grpo.minus.concur-tied.amc23}{$-$14.5}
\setval{4b.grpo.minus.concur-tied.amc23-abs}{14.5}
\setval{4b.grpo.minus.concur-tied.avg}{$-$4.0}
\setval{4b.grpo.minus.concur-tied.avg-abs}{4.0}
\setval{4b.grpo.minus.concur-tied.code}{+0.2}
\setval{4b.grpo.minus.concur-tied.code-abs}{0.2}
\setval{4b.grpo.minus.concur-tied.creative-writing-v3}{$-$0.5}
\setval{4b.grpo.minus.concur-tied.creative-writing-v3-abs}{0.5}
\setval{4b.grpo.minus.concur-tied.guidebench}{$-$1.2}
\setval{4b.grpo.minus.concur-tied.guidebench-abs}{1.2}
\setval{4b.grpo.minus.concur-tied.humaneval-plus}{$-$1.3}
\setval{4b.grpo.minus.concur-tied.humaneval-plus-abs}{1.3}
\setval{4b.grpo.minus.concur-tied.if}{$-$0.9}
\setval{4b.grpo.minus.concur-tied.if-abs}{0.9}
\setval{4b.grpo.minus.concur-tied.ifbench}{+0.0}
\setval{4b.grpo.minus.concur-tied.ifbench-abs}{0.0}
\setval{4b.grpo.minus.concur-tied.ifeval}{$-$1.3}
\setval{4b.grpo.minus.concur-tied.ifeval-abs}{1.3}
\setval{4b.grpo.minus.concur-tied.judge2.avg}{$-$3.5}
\setval{4b.grpo.minus.concur-tied.judge2.avg-abs}{3.5}
\setval{4b.grpo.minus.concur-tied.judge2.creative-writing-v3}{+1.1}
\setval{4b.grpo.minus.concur-tied.judge2.creative-writing-v3-abs}{1.1}
\setval{4b.grpo.minus.concur-tied.judge2.longbench-write}{$-$7.2}
\setval{4b.grpo.minus.concur-tied.judge2.longbench-write-abs}{7.2}
\setval{4b.grpo.minus.concur-tied.judge2.writing}{$-$2.7}
\setval{4b.grpo.minus.concur-tied.judge2.writing-abs}{2.7}
\setval{4b.grpo.minus.concur-tied.judge2.writingbench}{$-$1.9}
\setval{4b.grpo.minus.concur-tied.judge2.writingbench-abs}{1.9}
\setval{4b.grpo.minus.concur-tied.livecodebench}{$-$0.7}
\setval{4b.grpo.minus.concur-tied.livecodebench-abs}{0.7}
\setval{4b.grpo.minus.concur-tied.longbench-write}{$-$8.5}
\setval{4b.grpo.minus.concur-tied.longbench-write-abs}{8.5}
\setval{4b.grpo.minus.concur-tied.math}{$-$7.2}
\setval{4b.grpo.minus.concur-tied.math-abs}{7.2}
\setval{4b.grpo.minus.concur-tied.math500}{$-$5.3}
\setval{4b.grpo.minus.concur-tied.math500-abs}{5.3}
\setval{4b.grpo.minus.concur-tied.mbpp-plus}{+2.6}
\setval{4b.grpo.minus.concur-tied.mbpp-plus-abs}{2.6}
\setval{4b.grpo.minus.concur-tied.minerva}{$-$3.8}
\setval{4b.grpo.minus.concur-tied.minerva-abs}{3.8}
\setval{4b.grpo.minus.concur-tied.olympiad-bench}{$-$8.2}
\setval{4b.grpo.minus.concur-tied.olympiad-bench-abs}{8.2}
\setval{4b.grpo.minus.concur-tied.writing}{$-$5.0}
\setval{4b.grpo.minus.concur-tied.writing-abs}{5.0}
\setval{4b.grpo.minus.concur-tied.writingbench}{$-$5.9}
\setval{4b.grpo.minus.concur-tied.writingbench-abs}{5.9}
\setval{4b.grpo.minus.concur.aime24}{$-$8.5}
\setval{4b.grpo.minus.concur.aime24-abs}{8.5}
\setval{4b.grpo.minus.concur.aime25}{$-$6.1}
\setval{4b.grpo.minus.concur.aime25-abs}{6.1}
\setval{4b.grpo.minus.concur.amc23}{$-$14.7}
\setval{4b.grpo.minus.concur.amc23-abs}{14.7}
\setval{4b.grpo.minus.concur.avg}{$-$4.1}
\setval{4b.grpo.minus.concur.avg-abs}{4.1}
\setval{4b.grpo.minus.concur.code}{$-$0.4}
\setval{4b.grpo.minus.concur.code-abs}{0.4}
\setval{4b.grpo.minus.concur.creative-writing-v3}{+2.6}
\setval{4b.grpo.minus.concur.creative-writing-v3-abs}{2.6}
\setval{4b.grpo.minus.concur.guidebench}{+0.6}
\setval{4b.grpo.minus.concur.guidebench-abs}{0.6}
\setval{4b.grpo.minus.concur.humaneval-plus}{+0.3}
\setval{4b.grpo.minus.concur.humaneval-plus-abs}{0.3}
\setval{4b.grpo.minus.concur.if}{$-$1.3}
\setval{4b.grpo.minus.concur.if-abs}{1.3}
\setval{4b.grpo.minus.concur.ifbench}{$-$2.7}
\setval{4b.grpo.minus.concur.ifbench-abs}{2.7}
\setval{4b.grpo.minus.concur.ifeval}{$-$1.7}
\setval{4b.grpo.minus.concur.ifeval-abs}{1.7}
\setval{4b.grpo.minus.concur.judge2.avg}{$-$3.9}
\setval{4b.grpo.minus.concur.judge2.avg-abs}{3.9}
\setval{4b.grpo.minus.concur.judge2.creative-writing-v3}{+2.7}
\setval{4b.grpo.minus.concur.judge2.creative-writing-v3-abs}{2.7}
\setval{4b.grpo.minus.concur.judge2.longbench-write}{$-$5.9}
\setval{4b.grpo.minus.concur.judge2.longbench-write-abs}{5.9}
\setval{4b.grpo.minus.concur.judge2.writing}{$-$1.9}
\setval{4b.grpo.minus.concur.judge2.writing-abs}{1.9}
\setval{4b.grpo.minus.concur.judge2.writingbench}{$-$2.6}
\setval{4b.grpo.minus.concur.judge2.writingbench-abs}{2.6}
\setval{4b.grpo.minus.concur.livecodebench}{$-$1.1}
\setval{4b.grpo.minus.concur.livecodebench-abs}{1.1}
\setval{4b.grpo.minus.concur.longbench-write}{$-$7.1}
\setval{4b.grpo.minus.concur.longbench-write-abs}{7.1}
\setval{4b.grpo.minus.concur.math}{$-$7.9}
\setval{4b.grpo.minus.concur.math-abs}{7.9}
\setval{4b.grpo.minus.concur.math500}{$-$5.0}
\setval{4b.grpo.minus.concur.math500-abs}{5.0}
\setval{4b.grpo.minus.concur.mbpp-plus}{$-$0.3}
\setval{4b.grpo.minus.concur.mbpp-plus-abs}{0.3}
\setval{4b.grpo.minus.concur.minerva}{$-$3.9}
\setval{4b.grpo.minus.concur.minerva-abs}{3.9}
\setval{4b.grpo.minus.concur.olympiad-bench}{$-$9.0}
\setval{4b.grpo.minus.concur.olympiad-bench-abs}{9.0}
\setval{4b.grpo.minus.concur.writing}{$-$3.2}
\setval{4b.grpo.minus.concur.writing-abs}{3.2}
\setval{4b.grpo.minus.concur.writingbench}{$-$5.2}
\setval{4b.grpo.minus.concur.writingbench-abs}{5.2}
\setval{4b.grpo.minus.gd2po.aime24}{+0.1}
\setval{4b.grpo.minus.gd2po.aime24-abs}{0.1}
\setval{4b.grpo.minus.gd2po.aime25}{+5.8}
\setval{4b.grpo.minus.gd2po.aime25-abs}{5.8}
\setval{4b.grpo.minus.gd2po.amc23}{$-$3.0}
\setval{4b.grpo.minus.gd2po.amc23-abs}{3.0}
\setval{4b.grpo.minus.gd2po.avg}{+0.6}
\setval{4b.grpo.minus.gd2po.avg-abs}{0.6}
\setval{4b.grpo.minus.gd2po.code}{$-$0.1}
\setval{4b.grpo.minus.gd2po.code-abs}{0.1}
\setval{4b.grpo.minus.gd2po.creative-writing-v3}{+2.6}
\setval{4b.grpo.minus.gd2po.creative-writing-v3-abs}{2.6}
\setval{4b.grpo.minus.gd2po.guidebench}{$-$1.2}
\setval{4b.grpo.minus.gd2po.guidebench-abs}{1.2}
\setval{4b.grpo.minus.gd2po.humaneval-plus}{$-$0.7}
\setval{4b.grpo.minus.gd2po.humaneval-plus-abs}{0.7}
\setval{4b.grpo.minus.gd2po.if}{$-$0.8}
\setval{4b.grpo.minus.gd2po.if-abs}{0.8}
\setval{4b.grpo.minus.gd2po.ifbench}{+0.0}
\setval{4b.grpo.minus.gd2po.ifbench-abs}{0.0}
\setval{4b.grpo.minus.gd2po.ifeval}{$-$1.2}
\setval{4b.grpo.minus.gd2po.ifeval-abs}{1.2}
\setval{4b.grpo.minus.gd2po.judge2.avg}{+0.3}
\setval{4b.grpo.minus.gd2po.judge2.avg-abs}{0.3}
\setval{4b.grpo.minus.gd2po.judge2.creative-writing-v3}{+2.2}
\setval{4b.grpo.minus.gd2po.judge2.creative-writing-v3-abs}{2.2}
\setval{4b.grpo.minus.gd2po.judge2.longbench-write}{$-$0.3}
\setval{4b.grpo.minus.gd2po.judge2.longbench-write-abs}{0.3}
\setval{4b.grpo.minus.gd2po.judge2.writing}{+1.0}
\setval{4b.grpo.minus.gd2po.judge2.writing-abs}{1.0}
\setval{4b.grpo.minus.gd2po.judge2.writingbench}{+0.9}
\setval{4b.grpo.minus.gd2po.judge2.writingbench-abs}{0.9}
\setval{4b.grpo.minus.gd2po.livecodebench}{+0.5}
\setval{4b.grpo.minus.gd2po.livecodebench-abs}{0.5}
\setval{4b.grpo.minus.gd2po.longbench-write}{+0.3}
\setval{4b.grpo.minus.gd2po.longbench-write-abs}{0.3}
\setval{4b.grpo.minus.gd2po.math}{+0.8}
\setval{4b.grpo.minus.gd2po.math-abs}{0.8}
\setval{4b.grpo.minus.gd2po.math500}{+2.6}
\setval{4b.grpo.minus.gd2po.math500-abs}{2.6}
\setval{4b.grpo.minus.gd2po.mbpp-plus}{$-$0.3}
\setval{4b.grpo.minus.gd2po.mbpp-plus-abs}{0.3}
\setval{4b.grpo.minus.gd2po.minerva}{$-$1.9}
\setval{4b.grpo.minus.gd2po.minerva-abs}{1.9}
\setval{4b.grpo.minus.gd2po.olympiad-bench}{+1.1}
\setval{4b.grpo.minus.gd2po.olympiad-bench-abs}{1.1}
\setval{4b.grpo.minus.gd2po.writing}{+2.3}
\setval{4b.grpo.minus.gd2po.writing-abs}{2.3}
\setval{4b.grpo.minus.gd2po.writingbench}{+4.1}
\setval{4b.grpo.minus.gd2po.writingbench-abs}{4.1}
\setval{4b.grpo.minus.gdpo-noaux.aime24}{$-$11.5}
\setval{4b.grpo.minus.gdpo-noaux.aime24-abs}{11.5}
\setval{4b.grpo.minus.gdpo-noaux.aime25}{$-$6.1}
\setval{4b.grpo.minus.gdpo-noaux.aime25-abs}{6.1}
\setval{4b.grpo.minus.gdpo-noaux.amc23}{$-$11.0}
\setval{4b.grpo.minus.gdpo-noaux.amc23-abs}{11.0}
\setval{4b.grpo.minus.gdpo-noaux.avg}{$-$2.0}
\setval{4b.grpo.minus.gdpo-noaux.avg-abs}{2.0}
\setval{4b.grpo.minus.gdpo-noaux.code}{$-$1.6}
\setval{4b.grpo.minus.gdpo-noaux.code-abs}{1.6}
\setval{4b.grpo.minus.gdpo-noaux.creative-writing-v3}{+8.0}
\setval{4b.grpo.minus.gdpo-noaux.creative-writing-v3-abs}{8.0}
\setval{4b.grpo.minus.gdpo-noaux.guidebench}{+2.0}
\setval{4b.grpo.minus.gdpo-noaux.guidebench-abs}{2.0}
\setval{4b.grpo.minus.gdpo-noaux.humaneval-plus}{$-$2.7}
\setval{4b.grpo.minus.gdpo-noaux.humaneval-plus-abs}{2.7}
\setval{4b.grpo.minus.gdpo-noaux.if}{+0.3}
\setval{4b.grpo.minus.gdpo-noaux.if-abs}{0.3}
\setval{4b.grpo.minus.gdpo-noaux.ifbench}{$-$2.7}
\setval{4b.grpo.minus.gdpo-noaux.ifbench-abs}{2.7}
\setval{4b.grpo.minus.gdpo-noaux.ifeval}{+1.7}
\setval{4b.grpo.minus.gdpo-noaux.ifeval-abs}{1.7}
\setval{4b.grpo.minus.gdpo-noaux.judge2.avg}{$-$1.4}
\setval{4b.grpo.minus.gdpo-noaux.judge2.avg-abs}{1.4}
\setval{4b.grpo.minus.gdpo-noaux.judge2.creative-writing-v3}{+11.8}
\setval{4b.grpo.minus.gdpo-noaux.judge2.creative-writing-v3-abs}{11.8}
\setval{4b.grpo.minus.gdpo-noaux.judge2.longbench-write}{+9.9}
\setval{4b.grpo.minus.gdpo-noaux.judge2.longbench-write-abs}{9.9}
\setval{4b.grpo.minus.gdpo-noaux.judge2.writing}{+9.7}
\setval{4b.grpo.minus.gdpo-noaux.judge2.writing-abs}{9.7}
\setval{4b.grpo.minus.gdpo-noaux.judge2.writingbench}{+7.4}
\setval{4b.grpo.minus.gdpo-noaux.judge2.writingbench-abs}{7.4}
\setval{4b.grpo.minus.gdpo-noaux.livecodebench}{$-$0.7}
\setval{4b.grpo.minus.gdpo-noaux.livecodebench-abs}{0.7}
\setval{4b.grpo.minus.gdpo-noaux.longbench-write}{+4.8}
\setval{4b.grpo.minus.gdpo-noaux.longbench-write-abs}{4.8}
\setval{4b.grpo.minus.gdpo-noaux.math}{$-$7.6}
\setval{4b.grpo.minus.gdpo-noaux.math-abs}{7.6}
\setval{4b.grpo.minus.gdpo-noaux.math500}{$-$4.5}
\setval{4b.grpo.minus.gdpo-noaux.math500-abs}{4.5}
\setval{4b.grpo.minus.gdpo-noaux.mbpp-plus}{$-$1.2}
\setval{4b.grpo.minus.gdpo-noaux.mbpp-plus-abs}{1.2}
\setval{4b.grpo.minus.gdpo-noaux.minerva}{$-$4.2}
\setval{4b.grpo.minus.gdpo-noaux.minerva-abs}{4.2}
\setval{4b.grpo.minus.gdpo-noaux.olympiad-bench}{$-$8.5}
\setval{4b.grpo.minus.gdpo-noaux.olympiad-bench-abs}{8.5}
\setval{4b.grpo.minus.gdpo-noaux.writing}{+6.6}
\setval{4b.grpo.minus.gdpo-noaux.writing-abs}{6.6}
\setval{4b.grpo.minus.gdpo-noaux.writingbench}{+6.9}
\setval{4b.grpo.minus.gdpo-noaux.writingbench-abs}{6.9}
\setval{4b.grpo.minus.gdpo-w0.5.aime24}{$-$2.6}
\setval{4b.grpo.minus.gdpo-w0.5.aime24-abs}{2.6}
\setval{4b.grpo.minus.gdpo-w0.5.aime25}{+2.3}
\setval{4b.grpo.minus.gdpo-w0.5.aime25-abs}{2.3}
\setval{4b.grpo.minus.gdpo-w0.5.amc23}{$-$3.7}
\setval{4b.grpo.minus.gdpo-w0.5.amc23-abs}{3.7}
\setval{4b.grpo.minus.gdpo-w0.5.avg}{+0.1}
\setval{4b.grpo.minus.gdpo-w0.5.avg-abs}{0.1}
\setval{4b.grpo.minus.gdpo-w0.5.code}{+0.2}
\setval{4b.grpo.minus.gdpo-w0.5.code-abs}{0.2}
\setval{4b.grpo.minus.gdpo-w0.5.creative-writing-v3}{+9.2}
\setval{4b.grpo.minus.gdpo-w0.5.creative-writing-v3-abs}{9.2}
\setval{4b.grpo.minus.gdpo-w0.5.guidebench}{$-$1.6}
\setval{4b.grpo.minus.gdpo-w0.5.guidebench-abs}{1.6}
\setval{4b.grpo.minus.gdpo-w0.5.humaneval-plus}{+0.7}
\setval{4b.grpo.minus.gdpo-w0.5.humaneval-plus-abs}{0.7}
\setval{4b.grpo.minus.gdpo-w0.5.if}{$-$1.2}
\setval{4b.grpo.minus.gdpo-w0.5.if-abs}{1.2}
\setval{4b.grpo.minus.gdpo-w0.5.ifbench}{$-$1.7}
\setval{4b.grpo.minus.gdpo-w0.5.ifbench-abs}{1.7}
\setval{4b.grpo.minus.gdpo-w0.5.ifeval}{$-$0.4}
\setval{4b.grpo.minus.gdpo-w0.5.ifeval-abs}{0.4}
\setval{4b.grpo.minus.gdpo-w0.5.judge2.avg}{$-$0.6}
\setval{4b.grpo.minus.gdpo-w0.5.judge2.avg-abs}{0.6}
\setval{4b.grpo.minus.gdpo-w0.5.judge2.creative-writing-v3}{+5.0}
\setval{4b.grpo.minus.gdpo-w0.5.judge2.creative-writing-v3-abs}{5.0}
\setval{4b.grpo.minus.gdpo-w0.5.judge2.longbench-write}{+0.2}
\setval{4b.grpo.minus.gdpo-w0.5.judge2.longbench-write-abs}{0.2}
\setval{4b.grpo.minus.gdpo-w0.5.judge2.writing}{+1.7}
\setval{4b.grpo.minus.gdpo-w0.5.judge2.writing-abs}{1.7}
\setval{4b.grpo.minus.gdpo-w0.5.judge2.writingbench}{$-$0.2}
\setval{4b.grpo.minus.gdpo-w0.5.judge2.writingbench-abs}{0.2}
\setval{4b.grpo.minus.gdpo-w0.5.livecodebench}{$-$0.2}
\setval{4b.grpo.minus.gdpo-w0.5.livecodebench-abs}{0.2}
\setval{4b.grpo.minus.gdpo-w0.5.longbench-write}{+1.8}
\setval{4b.grpo.minus.gdpo-w0.5.longbench-write-abs}{1.8}
\setval{4b.grpo.minus.gdpo-w0.5.math}{$-$1.7}
\setval{4b.grpo.minus.gdpo-w0.5.math-abs}{1.7}
\setval{4b.grpo.minus.gdpo-w0.5.math500}{$-$1.5}
\setval{4b.grpo.minus.gdpo-w0.5.math500-abs}{1.5}
\setval{4b.grpo.minus.gdpo-w0.5.mbpp-plus}{+0.0}
\setval{4b.grpo.minus.gdpo-w0.5.mbpp-plus-abs}{0.0}
\setval{4b.grpo.minus.gdpo-w0.5.minerva}{$-$0.8}
\setval{4b.grpo.minus.gdpo-w0.5.minerva-abs}{0.8}
\setval{4b.grpo.minus.gdpo-w0.5.olympiad-bench}{$-$3.8}
\setval{4b.grpo.minus.gdpo-w0.5.olympiad-bench-abs}{3.8}
\setval{4b.grpo.minus.gdpo-w0.5.writing}{+4.8}
\setval{4b.grpo.minus.gdpo-w0.5.writing-abs}{4.8}
\setval{4b.grpo.minus.gdpo-w0.5.writingbench}{+3.5}
\setval{4b.grpo.minus.gdpo-w0.5.writingbench-abs}{3.5}
\setval{4b.grpo.minus.gdpo.aime24}{+2.9}
\setval{4b.grpo.minus.gdpo.aime24-abs}{2.9}
\setval{4b.grpo.minus.gdpo.aime25}{+2.7}
\setval{4b.grpo.minus.gdpo.aime25-abs}{2.7}
\setval{4b.grpo.minus.gdpo.amc23}{$-$4.9}
\setval{4b.grpo.minus.gdpo.amc23-abs}{4.9}
\setval{4b.grpo.minus.gdpo.avg}{+0.4}
\setval{4b.grpo.minus.gdpo.avg-abs}{0.4}
\setval{4b.grpo.minus.gdpo.code}{+0.3}
\setval{4b.grpo.minus.gdpo.code-abs}{0.3}
\setval{4b.grpo.minus.gdpo.creative-writing-v3}{+1.4}
\setval{4b.grpo.minus.gdpo.creative-writing-v3-abs}{1.4}
\setval{4b.grpo.minus.gdpo.guidebench}{$-$2.7}
\setval{4b.grpo.minus.gdpo.guidebench-abs}{2.7}
\setval{4b.grpo.minus.gdpo.humaneval-plus}{+0.9}
\setval{4b.grpo.minus.gdpo.humaneval-plus-abs}{0.9}
\setval{4b.grpo.minus.gdpo.if}{$-$0.2}
\setval{4b.grpo.minus.gdpo.if-abs}{0.2}
\setval{4b.grpo.minus.gdpo.ifbench}{+1.3}
\setval{4b.grpo.minus.gdpo.ifbench-abs}{1.3}
\setval{4b.grpo.minus.gdpo.ifeval}{+0.7}
\setval{4b.grpo.minus.gdpo.ifeval-abs}{0.7}
\setval{4b.grpo.minus.gdpo.judge2.avg}{+0.0}
\setval{4b.grpo.minus.gdpo.judge2.avg-abs}{0.0}
\setval{4b.grpo.minus.gdpo.judge2.creative-writing-v3}{+0.7}
\setval{4b.grpo.minus.gdpo.judge2.creative-writing-v3-abs}{0.7}
\setval{4b.grpo.minus.gdpo.judge2.longbench-write}{$-$1.6}
\setval{4b.grpo.minus.gdpo.judge2.longbench-write-abs}{1.6}
\setval{4b.grpo.minus.gdpo.judge2.writing}{$-$0.5}
\setval{4b.grpo.minus.gdpo.judge2.writing-abs}{0.5}
\setval{4b.grpo.minus.gdpo.judge2.writingbench}{$-$0.6}
\setval{4b.grpo.minus.gdpo.judge2.writingbench-abs}{0.6}
\setval{4b.grpo.minus.gdpo.livecodebench}{$-$0.4}
\setval{4b.grpo.minus.gdpo.livecodebench-abs}{0.4}
\setval{4b.grpo.minus.gdpo.longbench-write}{+0.2}
\setval{4b.grpo.minus.gdpo.longbench-write-abs}{0.2}
\setval{4b.grpo.minus.gdpo.math}{+0.2}
\setval{4b.grpo.minus.gdpo.math-abs}{0.2}
\setval{4b.grpo.minus.gdpo.math500}{+1.0}
\setval{4b.grpo.minus.gdpo.math500-abs}{1.0}
\setval{4b.grpo.minus.gdpo.mbpp-plus}{+0.3}
\setval{4b.grpo.minus.gdpo.mbpp-plus-abs}{0.3}
\setval{4b.grpo.minus.gdpo.minerva}{$-$0.2}
\setval{4b.grpo.minus.gdpo.minerva-abs}{0.2}
\setval{4b.grpo.minus.gdpo.olympiad-bench}{+0.0}
\setval{4b.grpo.minus.gdpo.olympiad-bench-abs}{0.0}
\setval{4b.grpo.minus.gdpo.writing}{+1.3}
\setval{4b.grpo.minus.gdpo.writing-abs}{1.3}
\setval{4b.grpo.minus.gdpo.writingbench}{+2.3}
\setval{4b.grpo.minus.gdpo.writingbench-abs}{2.3}
\setval{4b.grpo.minus.geoma.aime24}{+0.4}
\setval{4b.grpo.minus.geoma.aime24-abs}{0.4}
\setval{4b.grpo.minus.geoma.aime25}{+2.0}
\setval{4b.grpo.minus.geoma.aime25-abs}{2.0}
\setval{4b.grpo.minus.geoma.amc23}{$-$0.7}
\setval{4b.grpo.minus.geoma.amc23-abs}{0.7}
\setval{4b.grpo.minus.geoma.avg}{+0.9}
\setval{4b.grpo.minus.geoma.avg-abs}{0.9}
\setval{4b.grpo.minus.geoma.code}{+0.6}
\setval{4b.grpo.minus.geoma.code-abs}{0.6}
\setval{4b.grpo.minus.geoma.creative-writing-v3}{+1.5}
\setval{4b.grpo.minus.geoma.creative-writing-v3-abs}{1.5}
\setval{4b.grpo.minus.geoma.guidebench}{$-$0.6}
\setval{4b.grpo.minus.geoma.guidebench-abs}{0.6}
\setval{4b.grpo.minus.geoma.humaneval-plus}{+2.5}
\setval{4b.grpo.minus.geoma.humaneval-plus-abs}{2.5}
\setval{4b.grpo.minus.geoma.if}{$-$0.1}
\setval{4b.grpo.minus.geoma.if-abs}{0.1}
\setval{4b.grpo.minus.geoma.ifbench}{+1.7}
\setval{4b.grpo.minus.geoma.ifbench-abs}{1.7}
\setval{4b.grpo.minus.geoma.ifeval}{$-$1.3}
\setval{4b.grpo.minus.geoma.ifeval-abs}{1.3}
\setval{4b.grpo.minus.geoma.judge2.avg}{+0.8}
\setval{4b.grpo.minus.geoma.judge2.avg-abs}{0.8}
\setval{4b.grpo.minus.geoma.judge2.creative-writing-v3}{+2.0}
\setval{4b.grpo.minus.geoma.judge2.creative-writing-v3-abs}{2.0}
\setval{4b.grpo.minus.geoma.judge2.longbench-write}{+3.1}
\setval{4b.grpo.minus.geoma.judge2.longbench-write-abs}{3.1}
\setval{4b.grpo.minus.geoma.judge2.writing}{+2.6}
\setval{4b.grpo.minus.geoma.judge2.writing-abs}{2.6}
\setval{4b.grpo.minus.geoma.judge2.writingbench}{+2.8}
\setval{4b.grpo.minus.geoma.judge2.writingbench-abs}{2.8}
\setval{4b.grpo.minus.geoma.livecodebench}{$-$1.7}
\setval{4b.grpo.minus.geoma.livecodebench-abs}{1.7}
\setval{4b.grpo.minus.geoma.longbench-write}{+3.4}
\setval{4b.grpo.minus.geoma.longbench-write-abs}{3.4}
\setval{4b.grpo.minus.geoma.math}{+0.4}
\setval{4b.grpo.minus.geoma.math-abs}{0.4}
\setval{4b.grpo.minus.geoma.math500}{+1.4}
\setval{4b.grpo.minus.geoma.math500-abs}{1.4}
\setval{4b.grpo.minus.geoma.mbpp-plus}{+1.0}
\setval{4b.grpo.minus.geoma.mbpp-plus-abs}{1.0}
\setval{4b.grpo.minus.geoma.minerva}{$-$1.7}
\setval{4b.grpo.minus.geoma.minerva-abs}{1.7}
\setval{4b.grpo.minus.geoma.olympiad-bench}{+0.8}
\setval{4b.grpo.minus.geoma.olympiad-bench-abs}{0.8}
\setval{4b.grpo.minus.geoma.writing}{+3.2}
\setval{4b.grpo.minus.geoma.writing-abs}{3.2}
\setval{4b.grpo.minus.geoma.writingbench}{+4.7}
\setval{4b.grpo.minus.geoma.writingbench-abs}{4.7}
\setval{4b.grpo.minus.grpo-nodiv.aime24}{$-$0.4}
\setval{4b.grpo.minus.grpo-nodiv.aime24-abs}{0.4}
\setval{4b.grpo.minus.grpo-nodiv.aime25}{+4.3}
\setval{4b.grpo.minus.grpo-nodiv.aime25-abs}{4.3}
\setval{4b.grpo.minus.grpo-nodiv.amc23}{+0.2}
\setval{4b.grpo.minus.grpo-nodiv.amc23-abs}{0.2}
\setval{4b.grpo.minus.grpo-nodiv.avg}{+1.6}
\setval{4b.grpo.minus.grpo-nodiv.avg-abs}{1.6}
\setval{4b.grpo.minus.grpo-nodiv.code}{+0.6}
\setval{4b.grpo.minus.grpo-nodiv.code-abs}{0.6}
\setval{4b.grpo.minus.grpo-nodiv.creative-writing-v3}{+4.8}
\setval{4b.grpo.minus.grpo-nodiv.creative-writing-v3-abs}{4.8}
\setval{4b.grpo.minus.grpo-nodiv.guidebench}{$-$0.9}
\setval{4b.grpo.minus.grpo-nodiv.guidebench-abs}{0.9}
\setval{4b.grpo.minus.grpo-nodiv.humaneval-plus}{+2.3}
\setval{4b.grpo.minus.grpo-nodiv.humaneval-plus-abs}{2.3}
\setval{4b.grpo.minus.grpo-nodiv.if}{+0.8}
\setval{4b.grpo.minus.grpo-nodiv.if-abs}{0.8}
\setval{4b.grpo.minus.grpo-nodiv.ifbench}{+4.0}
\setval{4b.grpo.minus.grpo-nodiv.ifbench-abs}{4.0}
\setval{4b.grpo.minus.grpo-nodiv.ifeval}{$-$0.8}
\setval{4b.grpo.minus.grpo-nodiv.ifeval-abs}{0.8}
\setval{4b.grpo.minus.grpo-nodiv.judge2.avg}{+1.2}
\setval{4b.grpo.minus.grpo-nodiv.judge2.avg-abs}{1.2}
\setval{4b.grpo.minus.grpo-nodiv.judge2.creative-writing-v3}{+4.6}
\setval{4b.grpo.minus.grpo-nodiv.judge2.creative-writing-v3-abs}{4.6}
\setval{4b.grpo.minus.grpo-nodiv.judge2.longbench-write}{+2.1}
\setval{4b.grpo.minus.grpo-nodiv.judge2.longbench-write-abs}{2.1}
\setval{4b.grpo.minus.grpo-nodiv.judge2.writing}{+3.5}
\setval{4b.grpo.minus.grpo-nodiv.judge2.writing-abs}{3.5}
\setval{4b.grpo.minus.grpo-nodiv.judge2.writingbench}{+3.9}
\setval{4b.grpo.minus.grpo-nodiv.judge2.writingbench-abs}{3.9}
\setval{4b.grpo.minus.grpo-nodiv.livecodebench}{$-$0.5}
\setval{4b.grpo.minus.grpo-nodiv.livecodebench-abs}{0.5}
\setval{4b.grpo.minus.grpo-nodiv.longbench-write}{+3.2}
\setval{4b.grpo.minus.grpo-nodiv.longbench-write-abs}{3.2}
\setval{4b.grpo.minus.grpo-nodiv.math}{+0.7}
\setval{4b.grpo.minus.grpo-nodiv.math-abs}{0.7}
\setval{4b.grpo.minus.grpo-nodiv.math500}{+0.8}
\setval{4b.grpo.minus.grpo-nodiv.math500-abs}{0.8}
\setval{4b.grpo.minus.grpo-nodiv.mbpp-plus}{+0.0}
\setval{4b.grpo.minus.grpo-nodiv.mbpp-plus-abs}{0.0}
\setval{4b.grpo.minus.grpo-nodiv.minerva}{$-$0.5}
\setval{4b.grpo.minus.grpo-nodiv.minerva-abs}{0.5}
\setval{4b.grpo.minus.grpo-nodiv.olympiad-bench}{$-$0.2}
\setval{4b.grpo.minus.grpo-nodiv.olympiad-bench-abs}{0.2}
\setval{4b.grpo.minus.grpo-nodiv.writing}{+5.5}
\setval{4b.grpo.minus.grpo-nodiv.writing-abs}{5.5}
\setval{4b.grpo.minus.grpo-nodiv.writingbench}{+8.6}
\setval{4b.grpo.minus.grpo-nodiv.writingbench-abs}{8.6}
\setval{4b.grpo.minus.mt-grpo.aime24}{+1.3}
\setval{4b.grpo.minus.mt-grpo.aime24-abs}{1.3}
\setval{4b.grpo.minus.mt-grpo.aime25}{+6.5}
\setval{4b.grpo.minus.mt-grpo.aime25-abs}{6.5}
\setval{4b.grpo.minus.mt-grpo.amc23}{$-$5.1}
\setval{4b.grpo.minus.mt-grpo.amc23-abs}{5.1}
\setval{4b.grpo.minus.mt-grpo.avg}{+1.6}
\setval{4b.grpo.minus.mt-grpo.avg-abs}{1.6}
\setval{4b.grpo.minus.mt-grpo.code}{+2.0}
\setval{4b.grpo.minus.mt-grpo.code-abs}{2.0}
\setval{4b.grpo.minus.mt-grpo.creative-writing-v3}{+2.8}
\setval{4b.grpo.minus.mt-grpo.creative-writing-v3-abs}{2.8}
\setval{4b.grpo.minus.mt-grpo.guidebench}{$-$1.7}
\setval{4b.grpo.minus.mt-grpo.guidebench-abs}{1.7}
\setval{4b.grpo.minus.mt-grpo.humaneval-plus}{+3.7}
\setval{4b.grpo.minus.mt-grpo.humaneval-plus-abs}{3.7}
\setval{4b.grpo.minus.mt-grpo.if}{+0.5}
\setval{4b.grpo.minus.mt-grpo.if-abs}{0.5}
\setval{4b.grpo.minus.mt-grpo.ifbench}{+3.3}
\setval{4b.grpo.minus.mt-grpo.ifbench-abs}{3.3}
\setval{4b.grpo.minus.mt-grpo.ifeval}{$-$0.2}
\setval{4b.grpo.minus.mt-grpo.ifeval-abs}{0.2}
\setval{4b.grpo.minus.mt-grpo.judge2.avg}{+1.1}
\setval{4b.grpo.minus.mt-grpo.judge2.avg-abs}{1.1}
\setval{4b.grpo.minus.mt-grpo.judge2.creative-writing-v3}{+1.0}
\setval{4b.grpo.minus.mt-grpo.judge2.creative-writing-v3-abs}{1.0}
\setval{4b.grpo.minus.mt-grpo.judge2.longbench-write}{$-$0.1}
\setval{4b.grpo.minus.mt-grpo.judge2.longbench-write-abs}{0.1}
\setval{4b.grpo.minus.mt-grpo.judge2.writing}{+0.4}
\setval{4b.grpo.minus.mt-grpo.judge2.writing-abs}{0.4}
\setval{4b.grpo.minus.mt-grpo.judge2.writingbench}{+0.2}
\setval{4b.grpo.minus.mt-grpo.judge2.writingbench-abs}{0.2}
\setval{4b.grpo.minus.mt-grpo.livecodebench}{+1.4}
\setval{4b.grpo.minus.mt-grpo.livecodebench-abs}{1.4}
\setval{4b.grpo.minus.mt-grpo.longbench-write}{+2.4}
\setval{4b.grpo.minus.mt-grpo.longbench-write-abs}{2.4}
\setval{4b.grpo.minus.mt-grpo.math}{+1.2}
\setval{4b.grpo.minus.mt-grpo.math-abs}{1.2}
\setval{4b.grpo.minus.mt-grpo.math500}{+3.3}
\setval{4b.grpo.minus.mt-grpo.math500-abs}{3.3}
\setval{4b.grpo.minus.mt-grpo.mbpp-plus}{+1.1}
\setval{4b.grpo.minus.mt-grpo.mbpp-plus-abs}{1.1}
\setval{4b.grpo.minus.mt-grpo.minerva}{$-$0.7}
\setval{4b.grpo.minus.mt-grpo.minerva-abs}{0.7}
\setval{4b.grpo.minus.mt-grpo.olympiad-bench}{+2.1}
\setval{4b.grpo.minus.mt-grpo.olympiad-bench-abs}{2.1}
\setval{4b.grpo.minus.mt-grpo.writing}{+3.1}
\setval{4b.grpo.minus.mt-grpo.writing-abs}{3.1}
\setval{4b.grpo.minus.mt-grpo.writingbench}{+4.1}
\setval{4b.grpo.minus.mt-grpo.writingbench-abs}{4.1}
\setval{4b.grpo.minus.papo.aime24}{+2.7}
\setval{4b.grpo.minus.papo.aime24-abs}{2.7}
\setval{4b.grpo.minus.papo.aime25}{+5.1}
\setval{4b.grpo.minus.papo.aime25-abs}{5.1}
\setval{4b.grpo.minus.papo.amc23}{+1.2}
\setval{4b.grpo.minus.papo.amc23-abs}{1.2}
\setval{4b.grpo.minus.papo.avg}{+2.1}
\setval{4b.grpo.minus.papo.avg-abs}{2.1}
\setval{4b.grpo.minus.papo.code}{+0.2}
\setval{4b.grpo.minus.papo.code-abs}{0.2}
\setval{4b.grpo.minus.papo.creative-writing-v3}{+0.8}
\setval{4b.grpo.minus.papo.creative-writing-v3-abs}{0.8}
\setval{4b.grpo.minus.papo.guidebench}{$-$2.8}
\setval{4b.grpo.minus.papo.guidebench-abs}{2.8}
\setval{4b.grpo.minus.papo.humaneval-plus}{$-$0.3}
\setval{4b.grpo.minus.papo.humaneval-plus-abs}{0.3}
\setval{4b.grpo.minus.papo.if}{+0.7}
\setval{4b.grpo.minus.papo.if-abs}{0.7}
\setval{4b.grpo.minus.papo.ifbench}{+4.3}
\setval{4b.grpo.minus.papo.ifbench-abs}{4.3}
\setval{4b.grpo.minus.papo.ifeval}{+0.6}
\setval{4b.grpo.minus.papo.ifeval-abs}{0.6}
\setval{4b.grpo.minus.papo.judge2.avg}{+1.5}
\setval{4b.grpo.minus.papo.judge2.avg-abs}{1.5}
\setval{4b.grpo.minus.papo.judge2.creative-writing-v3}{+0.9}
\setval{4b.grpo.minus.papo.judge2.creative-writing-v3-abs}{0.9}
\setval{4b.grpo.minus.papo.judge2.longbench-write}{+3.2}
\setval{4b.grpo.minus.papo.judge2.longbench-write-abs}{3.2}
\setval{4b.grpo.minus.papo.judge2.writing}{+1.8}
\setval{4b.grpo.minus.papo.judge2.writing-abs}{1.8}
\setval{4b.grpo.minus.papo.judge2.writingbench}{+1.3}
\setval{4b.grpo.minus.papo.judge2.writingbench-abs}{1.3}
\setval{4b.grpo.minus.papo.livecodebench}{+0.5}
\setval{4b.grpo.minus.papo.livecodebench-abs}{0.5}
\setval{4b.grpo.minus.papo.longbench-write}{+5.2}
\setval{4b.grpo.minus.papo.longbench-write-abs}{5.2}
\setval{4b.grpo.minus.papo.math}{+2.4}
\setval{4b.grpo.minus.papo.math-abs}{2.4}
\setval{4b.grpo.minus.papo.math500}{+3.7}
\setval{4b.grpo.minus.papo.math500-abs}{3.7}
\setval{4b.grpo.minus.papo.mbpp-plus}{+0.4}
\setval{4b.grpo.minus.papo.mbpp-plus-abs}{0.4}
\setval{4b.grpo.minus.papo.minerva}{$-$1.3}
\setval{4b.grpo.minus.papo.minerva-abs}{1.3}
\setval{4b.grpo.minus.papo.olympiad-bench}{+3.1}
\setval{4b.grpo.minus.papo.olympiad-bench-abs}{3.1}
\setval{4b.grpo.minus.papo.writing}{+4.6}
\setval{4b.grpo.minus.papo.writing-abs}{4.6}
\setval{4b.grpo.minus.papo.writingbench}{+7.7}
\setval{4b.grpo.minus.papo.writingbench-abs}{7.7}
\setval{4b.grpo.minus.saw.aime24}{$-$0.5}
\setval{4b.grpo.minus.saw.aime24-abs}{0.5}
\setval{4b.grpo.minus.saw.aime25}{+1.5}
\setval{4b.grpo.minus.saw.aime25-abs}{1.5}
\setval{4b.grpo.minus.saw.amc23}{$-$2.0}
\setval{4b.grpo.minus.saw.amc23-abs}{2.0}
\setval{4b.grpo.minus.saw.avg}{+0.5}
\setval{4b.grpo.minus.saw.avg-abs}{0.5}
\setval{4b.grpo.minus.saw.code}{+1.4}
\setval{4b.grpo.minus.saw.code-abs}{1.4}
\setval{4b.grpo.minus.saw.creative-writing-v3}{+2.3}
\setval{4b.grpo.minus.saw.creative-writing-v3-abs}{2.3}
\setval{4b.grpo.minus.saw.guidebench}{$-$3.1}
\setval{4b.grpo.minus.saw.guidebench-abs}{3.1}
\setval{4b.grpo.minus.saw.humaneval-plus}{+2.0}
\setval{4b.grpo.minus.saw.humaneval-plus-abs}{2.0}
\setval{4b.grpo.minus.saw.if}{$-$2.1}
\setval{4b.grpo.minus.saw.if-abs}{2.1}
\setval{4b.grpo.minus.saw.ifbench}{$-$2.7}
\setval{4b.grpo.minus.saw.ifbench-abs}{2.7}
\setval{4b.grpo.minus.saw.ifeval}{$-$0.5}
\setval{4b.grpo.minus.saw.ifeval-abs}{0.5}
\setval{4b.grpo.minus.saw.judge2.avg}{$-$0.2}
\setval{4b.grpo.minus.saw.judge2.avg-abs}{0.2}
\setval{4b.grpo.minus.saw.judge2.creative-writing-v3}{$-$0.5}
\setval{4b.grpo.minus.saw.judge2.creative-writing-v3-abs}{0.5}
\setval{4b.grpo.minus.saw.judge2.longbench-write}{+0.7}
\setval{4b.grpo.minus.saw.judge2.longbench-write-abs}{0.7}
\setval{4b.grpo.minus.saw.judge2.writing}{+0.1}
\setval{4b.grpo.minus.saw.judge2.writing-abs}{0.1}
\setval{4b.grpo.minus.saw.judge2.writingbench}{+0.1}
\setval{4b.grpo.minus.saw.judge2.writingbench-abs}{0.1}
\setval{4b.grpo.minus.saw.livecodebench}{+1.2}
\setval{4b.grpo.minus.saw.livecodebench-abs}{1.2}
\setval{4b.grpo.minus.saw.longbench-write}{+3.0}
\setval{4b.grpo.minus.saw.longbench-write-abs}{3.0}
\setval{4b.grpo.minus.saw.math}{$-$0.1}
\setval{4b.grpo.minus.saw.math-abs}{0.1}
\setval{4b.grpo.minus.saw.math500}{+2.4}
\setval{4b.grpo.minus.saw.math500-abs}{2.4}
\setval{4b.grpo.minus.saw.mbpp-plus}{+1.1}
\setval{4b.grpo.minus.saw.mbpp-plus-abs}{1.1}
\setval{4b.grpo.minus.saw.minerva}{$-$2.9}
\setval{4b.grpo.minus.saw.minerva-abs}{2.9}
\setval{4b.grpo.minus.saw.olympiad-bench}{+0.8}
\setval{4b.grpo.minus.saw.olympiad-bench-abs}{0.8}
\setval{4b.grpo.minus.saw.writing}{+3.2}
\setval{4b.grpo.minus.saw.writing-abs}{3.2}
\setval{4b.grpo.minus.saw.writingbench}{+4.4}
\setval{4b.grpo.minus.saw.writingbench-abs}{4.4}
\setval{4b.grpo.passing-at-ceiling.code.early}{0.71}
\setval{4b.grpo.passing-at-ceiling.code.late}{0.74}
\setval{4b.grpo.passing-at-ceiling.code.run}{0.73}
\setval{4b.grpo.passing-at-ceiling.if.early}{0.29}
\setval{4b.grpo.passing-at-ceiling.if.late}{0.57}
\setval{4b.grpo.passing-at-ceiling.if.run}{0.48}
\setval{4b.grpo.passing-at-ceiling.math.early}{1.00}
\setval{4b.grpo.passing-at-ceiling.math.late}{1.00}
\setval{4b.grpo.passing-at-ceiling.math.run}{1.00}
\setval{4b.grpo.passing-at-ceiling.writing.early}{0.00}
\setval{4b.grpo.passing-at-ceiling.writing.late}{0.01}
\setval{4b.grpo.passing-at-ceiling.writing.run}{0.00}
\setval{4b.grpo.primary-level.code.early}{0.54}
\setval{4b.grpo.primary-level.code.late}{0.76}
\setval{4b.grpo.primary-level.code.run}{0.70}
\setval{4b.grpo.primary-level.if.early}{0.31}
\setval{4b.grpo.primary-level.if.late}{0.75}
\setval{4b.grpo.primary-level.if.run}{0.62}
\setval{4b.grpo.primary-level.math.early}{0.48}
\setval{4b.grpo.primary-level.math.late}{0.68}
\setval{4b.grpo.primary-level.math.run}{0.62}
\setval{4b.grpo.primary-level.writing.early}{0.53}
\setval{4b.grpo.primary-level.writing.late}{0.76}
\setval{4b.grpo.primary-level.writing.run}{0.72}
\setval{4b.mt-grpo.aux-change.code}{+0.19}
\setval{4b.mt-grpo.aux-change.if}{+0.22}
\setval{4b.mt-grpo.aux-change.math}{+0.13}
\setval{4b.mt-grpo.aux-change.writing}{+0.20}
\setval{4b.mt-grpo.aux-change.writing-rm}{+0.36}
\setval{4b.mt-grpo.aux-full.code.early}{0.74}
\setval{4b.mt-grpo.aux-full.code.late}{0.92}
\setval{4b.mt-grpo.aux-full.code.run}{0.91}
\setval{4b.mt-grpo.aux-full.if.early}{0.67}
\setval{4b.mt-grpo.aux-full.if.late}{0.85}
\setval{4b.mt-grpo.aux-full.if.run}{0.83}
\setval{4b.mt-grpo.aux-full.math.early}{0.65}
\setval{4b.mt-grpo.aux-full.math.late}{0.85}
\setval{4b.mt-grpo.aux-full.math.run}{0.83}
\setval{4b.mt-grpo.aux-full.writing-rm.early}{0.08}
\setval{4b.mt-grpo.aux-full.writing-rm.late}{0.55}
\setval{4b.mt-grpo.aux-full.writing-rm.run}{0.41}
\setval{4b.mt-grpo.aux-full.writing.early}{0.66}
\setval{4b.mt-grpo.aux-full.writing.late}{0.72}
\setval{4b.mt-grpo.aux-full.writing.run}{0.69}
\setval{4b.mt-grpo.aux-level.code.early}{0.92}
\setval{4b.mt-grpo.aux-level.code.late}{0.99}
\setval{4b.mt-grpo.aux-level.code.run}{0.98}
\setval{4b.mt-grpo.aux-level.if.early}{0.84}
\setval{4b.mt-grpo.aux-level.if.late}{0.92}
\setval{4b.mt-grpo.aux-level.if.run}{0.91}
\setval{4b.mt-grpo.aux-level.math.early}{0.93}
\setval{4b.mt-grpo.aux-level.math.late}{0.98}
\setval{4b.mt-grpo.aux-level.math.run}{0.97}
\setval{4b.mt-grpo.aux-level.writing-rm.early}{0.77}
\setval{4b.mt-grpo.aux-level.writing-rm.late}{0.96}
\setval{4b.mt-grpo.aux-level.writing-rm.run}{0.92}
\setval{4b.mt-grpo.aux-level.writing.early}{0.92}
\setval{4b.mt-grpo.aux-level.writing.late}{0.97}
\setval{4b.mt-grpo.aux-level.writing.run}{0.95}
\setval{4b.mt-grpo.bench.aime24}{10.7}
\setval{4b.mt-grpo.bench.aime25}{6.1}
\setval{4b.mt-grpo.bench.amc23}{59.8}
\setval{4b.mt-grpo.bench.creative-writing-v3}{44.5}
\setval{4b.mt-grpo.bench.guidebench}{79.7}
\setval{4b.mt-grpo.bench.humaneval-plus}{79.1}
\setval{4b.mt-grpo.bench.ifbench}{34.3}
\setval{4b.mt-grpo.bench.ifeval}{86.7}
\setval{4b.mt-grpo.bench.livecodebench}{22.6}
\setval{4b.mt-grpo.bench.longbench-write}{66.8}
\setval{4b.mt-grpo.bench.math500}{77.0}
\setval{4b.mt-grpo.bench.mbpp-plus}{70.1}
\setval{4b.mt-grpo.bench.minerva}{46.0}
\setval{4b.mt-grpo.bench.olympiad-bench}{42.5}
\setval{4b.mt-grpo.bench.writingbench}{70.1}
\setval{4b.mt-grpo.eval.avg}{53.1}
\setval{4b.mt-grpo.eval.code}{57.3}
\setval{4b.mt-grpo.eval.if}{66.9}
\setval{4b.mt-grpo.eval.math}{40.4}
\setval{4b.mt-grpo.eval.writing}{60.5}
\setval{4b.mt-grpo.judge2.avg}{50.8}
\setval{4b.mt-grpo.judge2.creative-writing-v3}{34.1}
\setval{4b.mt-grpo.judge2.longbench-write}{60.8}
\setval{4b.mt-grpo.judge2.writing}{49.4}
\setval{4b.mt-grpo.judge2.writingbench}{53.3}
\setval{4b.mt-grpo.last-step}{300}
\setval{4b.mt-grpo.minus.base.aime24}{+2.9}
\setval{4b.mt-grpo.minus.base.aime24-abs}{2.9}
\setval{4b.mt-grpo.minus.base.aime25}{$-$0.9}
\setval{4b.mt-grpo.minus.base.aime25-abs}{0.9}
\setval{4b.mt-grpo.minus.base.amc23}{+15.9}
\setval{4b.mt-grpo.minus.base.amc23-abs}{15.9}
\setval{4b.mt-grpo.minus.base.avg}{+18.5}
\setval{4b.mt-grpo.minus.base.avg-abs}{18.5}
\setval{4b.mt-grpo.minus.base.code}{+12.6}
\setval{4b.mt-grpo.minus.base.code-abs}{12.6}
\setval{4b.mt-grpo.minus.base.creative-writing-v3}{+19.7}
\setval{4b.mt-grpo.minus.base.creative-writing-v3-abs}{19.7}
\setval{4b.mt-grpo.minus.base.guidebench}{+54.8}
\setval{4b.mt-grpo.minus.base.guidebench-abs}{54.8}
\setval{4b.mt-grpo.minus.base.humaneval-plus}{+14.0}
\setval{4b.mt-grpo.minus.base.humaneval-plus-abs}{14.0}
\setval{4b.mt-grpo.minus.base.if}{+37.2}
\setval{4b.mt-grpo.minus.base.if-abs}{37.2}
\setval{4b.mt-grpo.minus.base.ifbench}{+17.3}
\setval{4b.mt-grpo.minus.base.ifbench-abs}{17.3}
\setval{4b.mt-grpo.minus.base.ifeval}{+39.3}
\setval{4b.mt-grpo.minus.base.ifeval-abs}{39.3}
\setval{4b.mt-grpo.minus.base.judge2.avg}{+16.9}
\setval{4b.mt-grpo.minus.base.judge2.avg-abs}{16.9}
\setval{4b.mt-grpo.minus.base.judge2.creative-writing-v3}{+13.2}
\setval{4b.mt-grpo.minus.base.judge2.creative-writing-v3-abs}{13.2}
\setval{4b.mt-grpo.minus.base.judge2.longbench-write}{+19.9}
\setval{4b.mt-grpo.minus.base.judge2.longbench-write-abs}{19.9}
\setval{4b.mt-grpo.minus.base.judge2.writing}{+18.6}
\setval{4b.mt-grpo.minus.base.judge2.writing-abs}{18.6}
\setval{4b.mt-grpo.minus.base.judge2.writingbench}{+22.8}
\setval{4b.mt-grpo.minus.base.judge2.writingbench-abs}{22.8}
\setval{4b.mt-grpo.minus.base.livecodebench}{+9.4}
\setval{4b.mt-grpo.minus.base.livecodebench-abs}{9.4}
\setval{4b.mt-grpo.minus.base.longbench-write}{+22.4}
\setval{4b.mt-grpo.minus.base.longbench-write-abs}{22.4}
\setval{4b.mt-grpo.minus.base.math}{+8.1}
\setval{4b.mt-grpo.minus.base.math-abs}{8.1}
\setval{4b.mt-grpo.minus.base.math500}{+9.7}
\setval{4b.mt-grpo.minus.base.math500-abs}{9.7}
\setval{4b.mt-grpo.minus.base.mbpp-plus}{+14.5}
\setval{4b.mt-grpo.minus.base.mbpp-plus-abs}{14.5}
\setval{4b.mt-grpo.minus.base.minerva}{+14.4}
\setval{4b.mt-grpo.minus.base.minerva-abs}{14.4}
\setval{4b.mt-grpo.minus.base.olympiad-bench}{+6.9}
\setval{4b.mt-grpo.minus.base.olympiad-bench-abs}{6.9}
\setval{4b.mt-grpo.minus.base.writing}{+26.6}
\setval{4b.mt-grpo.minus.base.writing-abs}{26.6}
\setval{4b.mt-grpo.minus.base.writingbench}{+37.7}
\setval{4b.mt-grpo.minus.base.writingbench-abs}{37.7}
\setval{4b.mt-grpo.minus.concur-free.aime24}{$-$1.1}
\setval{4b.mt-grpo.minus.concur-free.aime24-abs}{1.1}
\setval{4b.mt-grpo.minus.concur-free.aime25}{$-$8.4}
\setval{4b.mt-grpo.minus.concur-free.aime25-abs}{8.4}
\setval{4b.mt-grpo.minus.concur-free.amc23}{+5.6}
\setval{4b.mt-grpo.minus.concur-free.amc23-abs}{5.6}
\setval{4b.mt-grpo.minus.concur-free.avg}{$-$2.5}
\setval{4b.mt-grpo.minus.concur-free.avg-abs}{2.5}
\setval{4b.mt-grpo.minus.concur-free.code}{$-$1.1}
\setval{4b.mt-grpo.minus.concur-free.code-abs}{1.1}
\setval{4b.mt-grpo.minus.concur-free.creative-writing-v3}{$-$0.9}
\setval{4b.mt-grpo.minus.concur-free.creative-writing-v3-abs}{0.9}
\setval{4b.mt-grpo.minus.concur-free.guidebench}{+0.7}
\setval{4b.mt-grpo.minus.concur-free.guidebench-abs}{0.7}
\setval{4b.mt-grpo.minus.concur-free.humaneval-plus}{$-$2.0}
\setval{4b.mt-grpo.minus.concur-free.humaneval-plus-abs}{2.0}
\setval{4b.mt-grpo.minus.concur-free.if}{$-$1.3}
\setval{4b.mt-grpo.minus.concur-free.if-abs}{1.3}
\setval{4b.mt-grpo.minus.concur-free.ifbench}{$-$4.7}
\setval{4b.mt-grpo.minus.concur-free.ifbench-abs}{4.7}
\setval{4b.mt-grpo.minus.concur-free.ifeval}{+0.0}
\setval{4b.mt-grpo.minus.concur-free.ifeval-abs}{0.0}
\setval{4b.mt-grpo.minus.concur-free.judge2.avg}{$-$1.9}
\setval{4b.mt-grpo.minus.concur-free.judge2.avg-abs}{1.9}
\setval{4b.mt-grpo.minus.concur-free.judge2.creative-writing-v3}{$-$1.2}
\setval{4b.mt-grpo.minus.concur-free.judge2.creative-writing-v3-abs}{1.2}
\setval{4b.mt-grpo.minus.concur-free.judge2.longbench-write}{$-$8.6}
\setval{4b.mt-grpo.minus.concur-free.judge2.longbench-write-abs}{8.6}
\setval{4b.mt-grpo.minus.concur-free.judge2.writing}{$-$3.7}
\setval{4b.mt-grpo.minus.concur-free.judge2.writing-abs}{3.7}
\setval{4b.mt-grpo.minus.concur-free.judge2.writingbench}{$-$1.2}
\setval{4b.mt-grpo.minus.concur-free.judge2.writingbench-abs}{1.2}
\setval{4b.mt-grpo.minus.concur-free.livecodebench}{$-$1.5}
\setval{4b.mt-grpo.minus.concur-free.livecodebench-abs}{1.5}
\setval{4b.mt-grpo.minus.concur-free.longbench-write}{$-$11.0}
\setval{4b.mt-grpo.minus.concur-free.longbench-write-abs}{11.0}
\setval{4b.mt-grpo.minus.concur-free.math}{$-$1.8}
\setval{4b.mt-grpo.minus.concur-free.math-abs}{1.8}
\setval{4b.mt-grpo.minus.concur-free.math500}{$-$3.3}
\setval{4b.mt-grpo.minus.concur-free.math500-abs}{3.3}
\setval{4b.mt-grpo.minus.concur-free.mbpp-plus}{+0.4}
\setval{4b.mt-grpo.minus.concur-free.mbpp-plus-abs}{0.4}
\setval{4b.mt-grpo.minus.concur-free.minerva}{$-$0.3}
\setval{4b.mt-grpo.minus.concur-free.minerva-abs}{0.3}
\setval{4b.mt-grpo.minus.concur-free.olympiad-bench}{$-$3.2}
\setval{4b.mt-grpo.minus.concur-free.olympiad-bench-abs}{3.2}
\setval{4b.mt-grpo.minus.concur-free.writing}{$-$6.3}
\setval{4b.mt-grpo.minus.concur-free.writing-abs}{6.3}
\setval{4b.mt-grpo.minus.concur-free.writingbench}{$-$7.2}
\setval{4b.mt-grpo.minus.concur-free.writingbench-abs}{7.2}
\setval{4b.mt-grpo.minus.concur-tied.aime24}{$-$8.8}
\setval{4b.mt-grpo.minus.concur-tied.aime24-abs}{8.8}
\setval{4b.mt-grpo.minus.concur-tied.aime25}{$-$10.2}
\setval{4b.mt-grpo.minus.concur-tied.aime25-abs}{10.2}
\setval{4b.mt-grpo.minus.concur-tied.amc23}{$-$9.5}
\setval{4b.mt-grpo.minus.concur-tied.amc23-abs}{9.5}
\setval{4b.mt-grpo.minus.concur-tied.avg}{$-$5.6}
\setval{4b.mt-grpo.minus.concur-tied.avg-abs}{5.6}
\setval{4b.mt-grpo.minus.concur-tied.code}{$-$1.9}
\setval{4b.mt-grpo.minus.concur-tied.code-abs}{1.9}
\setval{4b.mt-grpo.minus.concur-tied.creative-writing-v3}{$-$3.3}
\setval{4b.mt-grpo.minus.concur-tied.creative-writing-v3-abs}{3.3}
\setval{4b.mt-grpo.minus.concur-tied.guidebench}{+0.5}
\setval{4b.mt-grpo.minus.concur-tied.guidebench-abs}{0.5}
\setval{4b.mt-grpo.minus.concur-tied.humaneval-plus}{$-$4.9}
\setval{4b.mt-grpo.minus.concur-tied.humaneval-plus-abs}{4.9}
\setval{4b.mt-grpo.minus.concur-tied.if}{$-$1.3}
\setval{4b.mt-grpo.minus.concur-tied.if-abs}{1.3}
\setval{4b.mt-grpo.minus.concur-tied.ifbench}{$-$3.3}
\setval{4b.mt-grpo.minus.concur-tied.ifbench-abs}{3.3}
\setval{4b.mt-grpo.minus.concur-tied.ifeval}{$-$1.1}
\setval{4b.mt-grpo.minus.concur-tied.ifeval-abs}{1.1}
\setval{4b.mt-grpo.minus.concur-tied.judge2.avg}{$-$4.6}
\setval{4b.mt-grpo.minus.concur-tied.judge2.avg-abs}{4.6}
\setval{4b.mt-grpo.minus.concur-tied.judge2.creative-writing-v3}{+0.1}
\setval{4b.mt-grpo.minus.concur-tied.judge2.creative-writing-v3-abs}{0.1}
\setval{4b.mt-grpo.minus.concur-tied.judge2.longbench-write}{$-$7.2}
\setval{4b.mt-grpo.minus.concur-tied.judge2.longbench-write-abs}{7.2}
\setval{4b.mt-grpo.minus.concur-tied.judge2.writing}{$-$3.1}
\setval{4b.mt-grpo.minus.concur-tied.judge2.writing-abs}{3.1}
\setval{4b.mt-grpo.minus.concur-tied.judge2.writingbench}{$-$2.1}
\setval{4b.mt-grpo.minus.concur-tied.judge2.writingbench-abs}{2.1}
\setval{4b.mt-grpo.minus.concur-tied.livecodebench}{$-$2.1}
\setval{4b.mt-grpo.minus.concur-tied.livecodebench-abs}{2.1}
\setval{4b.mt-grpo.minus.concur-tied.longbench-write}{$-$10.9}
\setval{4b.mt-grpo.minus.concur-tied.longbench-write-abs}{10.9}
\setval{4b.mt-grpo.minus.concur-tied.math}{$-$8.4}
\setval{4b.mt-grpo.minus.concur-tied.math-abs}{8.4}
\setval{4b.mt-grpo.minus.concur-tied.math500}{$-$8.5}
\setval{4b.mt-grpo.minus.concur-tied.math500-abs}{8.5}
\setval{4b.mt-grpo.minus.concur-tied.mbpp-plus}{+1.5}
\setval{4b.mt-grpo.minus.concur-tied.mbpp-plus-abs}{1.5}
\setval{4b.mt-grpo.minus.concur-tied.minerva}{$-$3.1}
\setval{4b.mt-grpo.minus.concur-tied.minerva-abs}{3.1}
\setval{4b.mt-grpo.minus.concur-tied.olympiad-bench}{$-$10.3}
\setval{4b.mt-grpo.minus.concur-tied.olympiad-bench-abs}{10.3}
\setval{4b.mt-grpo.minus.concur-tied.writing}{$-$8.1}
\setval{4b.mt-grpo.minus.concur-tied.writing-abs}{8.1}
\setval{4b.mt-grpo.minus.concur-tied.writingbench}{$-$10.0}
\setval{4b.mt-grpo.minus.concur-tied.writingbench-abs}{10.0}
\setval{4b.mt-grpo.minus.concur.aime24}{$-$9.8}
\setval{4b.mt-grpo.minus.concur.aime24-abs}{9.8}
\setval{4b.mt-grpo.minus.concur.aime25}{$-$12.6}
\setval{4b.mt-grpo.minus.concur.aime25-abs}{12.6}
\setval{4b.mt-grpo.minus.concur.amc23}{$-$9.6}
\setval{4b.mt-grpo.minus.concur.amc23-abs}{9.6}
\setval{4b.mt-grpo.minus.concur.avg}{$-$5.7}
\setval{4b.mt-grpo.minus.concur.avg-abs}{5.7}
\setval{4b.mt-grpo.minus.concur.code}{$-$2.4}
\setval{4b.mt-grpo.minus.concur.code-abs}{2.4}
\setval{4b.mt-grpo.minus.concur.creative-writing-v3}{$-$0.2}
\setval{4b.mt-grpo.minus.concur.creative-writing-v3-abs}{0.2}
\setval{4b.mt-grpo.minus.concur.guidebench}{+2.3}
\setval{4b.mt-grpo.minus.concur.guidebench-abs}{2.3}
\setval{4b.mt-grpo.minus.concur.humaneval-plus}{$-$3.4}
\setval{4b.mt-grpo.minus.concur.humaneval-plus-abs}{3.4}
\setval{4b.mt-grpo.minus.concur.if}{$-$1.7}
\setval{4b.mt-grpo.minus.concur.if-abs}{1.7}
\setval{4b.mt-grpo.minus.concur.ifbench}{$-$6.0}
\setval{4b.mt-grpo.minus.concur.ifbench-abs}{6.0}
\setval{4b.mt-grpo.minus.concur.ifeval}{$-$1.4}
\setval{4b.mt-grpo.minus.concur.ifeval-abs}{1.4}
\setval{4b.mt-grpo.minus.concur.judge2.avg}{$-$4.9}
\setval{4b.mt-grpo.minus.concur.judge2.avg-abs}{4.9}
\setval{4b.mt-grpo.minus.concur.judge2.creative-writing-v3}{+1.7}
\setval{4b.mt-grpo.minus.concur.judge2.creative-writing-v3-abs}{1.7}
\setval{4b.mt-grpo.minus.concur.judge2.longbench-write}{$-$5.9}
\setval{4b.mt-grpo.minus.concur.judge2.longbench-write-abs}{5.9}
\setval{4b.mt-grpo.minus.concur.judge2.writing}{$-$2.3}
\setval{4b.mt-grpo.minus.concur.judge2.writing-abs}{2.3}
\setval{4b.mt-grpo.minus.concur.judge2.writingbench}{$-$2.8}
\setval{4b.mt-grpo.minus.concur.judge2.writingbench-abs}{2.8}
\setval{4b.mt-grpo.minus.concur.livecodebench}{$-$2.5}
\setval{4b.mt-grpo.minus.concur.livecodebench-abs}{2.5}
\setval{4b.mt-grpo.minus.concur.longbench-write}{$-$9.5}
\setval{4b.mt-grpo.minus.concur.longbench-write-abs}{9.5}
\setval{4b.mt-grpo.minus.concur.math}{$-$9.1}
\setval{4b.mt-grpo.minus.concur.math-abs}{9.1}
\setval{4b.mt-grpo.minus.concur.math500}{$-$8.3}
\setval{4b.mt-grpo.minus.concur.math500-abs}{8.3}
\setval{4b.mt-grpo.minus.concur.mbpp-plus}{$-$1.5}
\setval{4b.mt-grpo.minus.concur.mbpp-plus-abs}{1.5}
\setval{4b.mt-grpo.minus.concur.minerva}{$-$3.2}
\setval{4b.mt-grpo.minus.concur.minerva-abs}{3.2}
\setval{4b.mt-grpo.minus.concur.olympiad-bench}{$-$11.1}
\setval{4b.mt-grpo.minus.concur.olympiad-bench-abs}{11.1}
\setval{4b.mt-grpo.minus.concur.writing}{$-$6.3}
\setval{4b.mt-grpo.minus.concur.writing-abs}{6.3}
\setval{4b.mt-grpo.minus.concur.writingbench}{$-$9.3}
\setval{4b.mt-grpo.minus.concur.writingbench-abs}{9.3}
\setval{4b.mt-grpo.minus.gd2po.aime24}{$-$1.1}
\setval{4b.mt-grpo.minus.gd2po.aime24-abs}{1.1}
\setval{4b.mt-grpo.minus.gd2po.aime25}{$-$0.6}
\setval{4b.mt-grpo.minus.gd2po.aime25-abs}{0.6}
\setval{4b.mt-grpo.minus.gd2po.amc23}{+2.0}
\setval{4b.mt-grpo.minus.gd2po.amc23-abs}{2.0}
\setval{4b.mt-grpo.minus.gd2po.avg}{$-$1.0}
\setval{4b.mt-grpo.minus.gd2po.avg-abs}{1.0}
\setval{4b.mt-grpo.minus.gd2po.code}{$-$2.2}
\setval{4b.mt-grpo.minus.gd2po.code-abs}{2.2}
\setval{4b.mt-grpo.minus.gd2po.creative-writing-v3}{$-$0.2}
\setval{4b.mt-grpo.minus.gd2po.creative-writing-v3-abs}{0.2}
\setval{4b.mt-grpo.minus.gd2po.guidebench}{+0.6}
\setval{4b.mt-grpo.minus.gd2po.guidebench-abs}{0.6}
\setval{4b.mt-grpo.minus.gd2po.humaneval-plus}{$-$4.3}
\setval{4b.mt-grpo.minus.gd2po.humaneval-plus-abs}{4.3}
\setval{4b.mt-grpo.minus.gd2po.if}{$-$1.2}
\setval{4b.mt-grpo.minus.gd2po.if-abs}{1.2}
\setval{4b.mt-grpo.minus.gd2po.ifbench}{$-$3.3}
\setval{4b.mt-grpo.minus.gd2po.ifbench-abs}{3.3}
\setval{4b.mt-grpo.minus.gd2po.ifeval}{$-$1.0}
\setval{4b.mt-grpo.minus.gd2po.ifeval-abs}{1.0}
\setval{4b.mt-grpo.minus.gd2po.judge2.avg}{$-$0.8}
\setval{4b.mt-grpo.minus.gd2po.judge2.avg-abs}{0.8}
\setval{4b.mt-grpo.minus.gd2po.judge2.creative-writing-v3}{+1.2}
\setval{4b.mt-grpo.minus.gd2po.judge2.creative-writing-v3-abs}{1.2}
\setval{4b.mt-grpo.minus.gd2po.judge2.longbench-write}{$-$0.2}
\setval{4b.mt-grpo.minus.gd2po.judge2.longbench-write-abs}{0.2}
\setval{4b.mt-grpo.minus.gd2po.judge2.writing}{+0.6}
\setval{4b.mt-grpo.minus.gd2po.judge2.writing-abs}{0.6}
\setval{4b.mt-grpo.minus.gd2po.judge2.writingbench}{+0.7}
\setval{4b.mt-grpo.minus.gd2po.judge2.writingbench-abs}{0.7}
\setval{4b.mt-grpo.minus.gd2po.livecodebench}{$-$0.9}
\setval{4b.mt-grpo.minus.gd2po.livecodebench-abs}{0.9}
\setval{4b.mt-grpo.minus.gd2po.longbench-write}{$-$2.1}
\setval{4b.mt-grpo.minus.gd2po.longbench-write-abs}{2.1}
\setval{4b.mt-grpo.minus.gd2po.math}{$-$0.4}
\setval{4b.mt-grpo.minus.gd2po.math-abs}{0.4}
\setval{4b.mt-grpo.minus.gd2po.math500}{$-$0.7}
\setval{4b.mt-grpo.minus.gd2po.math500-abs}{0.7}
\setval{4b.mt-grpo.minus.gd2po.mbpp-plus}{$-$1.4}
\setval{4b.mt-grpo.minus.gd2po.mbpp-plus-abs}{1.4}
\setval{4b.mt-grpo.minus.gd2po.minerva}{$-$1.2}
\setval{4b.mt-grpo.minus.gd2po.minerva-abs}{1.2}
\setval{4b.mt-grpo.minus.gd2po.olympiad-bench}{$-$1.0}
\setval{4b.mt-grpo.minus.gd2po.olympiad-bench-abs}{1.0}
\setval{4b.mt-grpo.minus.gd2po.writing}{$-$0.8}
\setval{4b.mt-grpo.minus.gd2po.writing-abs}{0.8}
\setval{4b.mt-grpo.minus.gd2po.writingbench}{+0.0}
\setval{4b.mt-grpo.minus.gd2po.writingbench-abs}{0.0}
\setval{4b.mt-grpo.minus.gdpo-noaux.aime24}{$-$12.7}
\setval{4b.mt-grpo.minus.gdpo-noaux.aime24-abs}{12.7}
\setval{4b.mt-grpo.minus.gdpo-noaux.aime25}{$-$12.6}
\setval{4b.mt-grpo.minus.gdpo-noaux.aime25-abs}{12.6}
\setval{4b.mt-grpo.minus.gdpo-noaux.amc23}{$-$5.9}
\setval{4b.mt-grpo.minus.gdpo-noaux.amc23-abs}{5.9}
\setval{4b.mt-grpo.minus.gdpo-noaux.avg}{$-$3.6}
\setval{4b.mt-grpo.minus.gdpo-noaux.avg-abs}{3.6}
\setval{4b.mt-grpo.minus.gdpo-noaux.code}{$-$3.6}
\setval{4b.mt-grpo.minus.gdpo-noaux.code-abs}{3.6}
\setval{4b.mt-grpo.minus.gdpo-noaux.creative-writing-v3}{+5.2}
\setval{4b.mt-grpo.minus.gdpo-noaux.creative-writing-v3-abs}{5.2}
\setval{4b.mt-grpo.minus.gdpo-noaux.guidebench}{+3.7}
\setval{4b.mt-grpo.minus.gdpo-noaux.guidebench-abs}{3.7}
\setval{4b.mt-grpo.minus.gdpo-noaux.humaneval-plus}{$-$6.4}
\setval{4b.mt-grpo.minus.gdpo-noaux.humaneval-plus-abs}{6.4}
\setval{4b.mt-grpo.minus.gdpo-noaux.if}{$-$0.1}
\setval{4b.mt-grpo.minus.gdpo-noaux.if-abs}{0.1}
\setval{4b.mt-grpo.minus.gdpo-noaux.ifbench}{$-$6.0}
\setval{4b.mt-grpo.minus.gdpo-noaux.ifbench-abs}{6.0}
\setval{4b.mt-grpo.minus.gdpo-noaux.ifeval}{+1.9}
\setval{4b.mt-grpo.minus.gdpo-noaux.ifeval-abs}{1.9}
\setval{4b.mt-grpo.minus.gdpo-noaux.judge2.avg}{$-$2.4}
\setval{4b.mt-grpo.minus.gdpo-noaux.judge2.avg-abs}{2.4}
\setval{4b.mt-grpo.minus.gdpo-noaux.judge2.creative-writing-v3}{+10.7}
\setval{4b.mt-grpo.minus.gdpo-noaux.judge2.creative-writing-v3-abs}{10.7}
\setval{4b.mt-grpo.minus.gdpo-noaux.judge2.longbench-write}{+10.0}
\setval{4b.mt-grpo.minus.gdpo-noaux.judge2.longbench-write-abs}{10.0}
\setval{4b.mt-grpo.minus.gdpo-noaux.judge2.writing}{+9.3}
\setval{4b.mt-grpo.minus.gdpo-noaux.judge2.writing-abs}{9.3}
\setval{4b.mt-grpo.minus.gdpo-noaux.judge2.writingbench}{+7.2}
\setval{4b.mt-grpo.minus.gdpo-noaux.judge2.writingbench-abs}{7.2}
\setval{4b.mt-grpo.minus.gdpo-noaux.livecodebench}{$-$2.1}
\setval{4b.mt-grpo.minus.gdpo-noaux.livecodebench-abs}{2.1}
\setval{4b.mt-grpo.minus.gdpo-noaux.longbench-write}{+2.4}
\setval{4b.mt-grpo.minus.gdpo-noaux.longbench-write-abs}{2.4}
\setval{4b.mt-grpo.minus.gdpo-noaux.math}{$-$8.9}
\setval{4b.mt-grpo.minus.gdpo-noaux.math-abs}{8.9}
\setval{4b.mt-grpo.minus.gdpo-noaux.math500}{$-$7.8}
\setval{4b.mt-grpo.minus.gdpo-noaux.math500-abs}{7.8}
\setval{4b.mt-grpo.minus.gdpo-noaux.mbpp-plus}{$-$2.3}
\setval{4b.mt-grpo.minus.gdpo-noaux.mbpp-plus-abs}{2.3}
\setval{4b.mt-grpo.minus.gdpo-noaux.minerva}{$-$3.5}
\setval{4b.mt-grpo.minus.gdpo-noaux.minerva-abs}{3.5}
\setval{4b.mt-grpo.minus.gdpo-noaux.olympiad-bench}{$-$10.6}
\setval{4b.mt-grpo.minus.gdpo-noaux.olympiad-bench-abs}{10.6}
\setval{4b.mt-grpo.minus.gdpo-noaux.writing}{+3.5}
\setval{4b.mt-grpo.minus.gdpo-noaux.writing-abs}{3.5}
\setval{4b.mt-grpo.minus.gdpo-noaux.writingbench}{+2.8}
\setval{4b.mt-grpo.minus.gdpo-noaux.writingbench-abs}{2.8}
\setval{4b.mt-grpo.minus.gdpo-w0.5.aime24}{$-$3.9}
\setval{4b.mt-grpo.minus.gdpo-w0.5.aime24-abs}{3.9}
\setval{4b.mt-grpo.minus.gdpo-w0.5.aime25}{$-$4.2}
\setval{4b.mt-grpo.minus.gdpo-w0.5.aime25-abs}{4.2}
\setval{4b.mt-grpo.minus.gdpo-w0.5.amc23}{+1.4}
\setval{4b.mt-grpo.minus.gdpo-w0.5.amc23-abs}{1.4}
\setval{4b.mt-grpo.minus.gdpo-w0.5.avg}{$-$1.5}
\setval{4b.mt-grpo.minus.gdpo-w0.5.avg-abs}{1.5}
\setval{4b.mt-grpo.minus.gdpo-w0.5.code}{$-$1.9}
\setval{4b.mt-grpo.minus.gdpo-w0.5.code-abs}{1.9}
\setval{4b.mt-grpo.minus.gdpo-w0.5.creative-writing-v3}{+6.4}
\setval{4b.mt-grpo.minus.gdpo-w0.5.creative-writing-v3-abs}{6.4}
\setval{4b.mt-grpo.minus.gdpo-w0.5.guidebench}{+0.1}
\setval{4b.mt-grpo.minus.gdpo-w0.5.guidebench-abs}{0.1}
\setval{4b.mt-grpo.minus.gdpo-w0.5.humaneval-plus}{$-$2.9}
\setval{4b.mt-grpo.minus.gdpo-w0.5.humaneval-plus-abs}{2.9}
\setval{4b.mt-grpo.minus.gdpo-w0.5.if}{$-$1.7}
\setval{4b.mt-grpo.minus.gdpo-w0.5.if-abs}{1.7}
\setval{4b.mt-grpo.minus.gdpo-w0.5.ifbench}{$-$5.0}
\setval{4b.mt-grpo.minus.gdpo-w0.5.ifbench-abs}{5.0}
\setval{4b.mt-grpo.minus.gdpo-w0.5.ifeval}{$-$0.1}
\setval{4b.mt-grpo.minus.gdpo-w0.5.ifeval-abs}{0.1}
\setval{4b.mt-grpo.minus.gdpo-w0.5.judge2.avg}{$-$1.6}
\setval{4b.mt-grpo.minus.gdpo-w0.5.judge2.avg-abs}{1.6}
\setval{4b.mt-grpo.minus.gdpo-w0.5.judge2.creative-writing-v3}{+4.0}
\setval{4b.mt-grpo.minus.gdpo-w0.5.judge2.creative-writing-v3-abs}{4.0}
\setval{4b.mt-grpo.minus.gdpo-w0.5.judge2.longbench-write}{+0.3}
\setval{4b.mt-grpo.minus.gdpo-w0.5.judge2.longbench-write-abs}{0.3}
\setval{4b.mt-grpo.minus.gdpo-w0.5.judge2.writing}{+1.3}
\setval{4b.mt-grpo.minus.gdpo-w0.5.judge2.writing-abs}{1.3}
\setval{4b.mt-grpo.minus.gdpo-w0.5.judge2.writingbench}{$-$0.3}
\setval{4b.mt-grpo.minus.gdpo-w0.5.judge2.writingbench-abs}{0.3}
\setval{4b.mt-grpo.minus.gdpo-w0.5.livecodebench}{$-$1.5}
\setval{4b.mt-grpo.minus.gdpo-w0.5.livecodebench-abs}{1.5}
\setval{4b.mt-grpo.minus.gdpo-w0.5.longbench-write}{$-$0.6}
\setval{4b.mt-grpo.minus.gdpo-w0.5.longbench-write-abs}{0.6}
\setval{4b.mt-grpo.minus.gdpo-w0.5.math}{$-$2.9}
\setval{4b.mt-grpo.minus.gdpo-w0.5.math-abs}{2.9}
\setval{4b.mt-grpo.minus.gdpo-w0.5.math500}{$-$4.8}
\setval{4b.mt-grpo.minus.gdpo-w0.5.math500-abs}{4.8}
\setval{4b.mt-grpo.minus.gdpo-w0.5.mbpp-plus}{$-$1.1}
\setval{4b.mt-grpo.minus.gdpo-w0.5.mbpp-plus-abs}{1.1}
\setval{4b.mt-grpo.minus.gdpo-w0.5.minerva}{$-$0.1}
\setval{4b.mt-grpo.minus.gdpo-w0.5.minerva-abs}{0.1}
\setval{4b.mt-grpo.minus.gdpo-w0.5.olympiad-bench}{$-$5.9}
\setval{4b.mt-grpo.minus.gdpo-w0.5.olympiad-bench-abs}{5.9}
\setval{4b.mt-grpo.minus.gdpo-w0.5.writing}{+1.7}
\setval{4b.mt-grpo.minus.gdpo-w0.5.writing-abs}{1.7}
\setval{4b.mt-grpo.minus.gdpo-w0.5.writingbench}{$-$0.6}
\setval{4b.mt-grpo.minus.gdpo-w0.5.writingbench-abs}{0.6}
\setval{4b.mt-grpo.minus.gdpo.aime24}{+1.7}
\setval{4b.mt-grpo.minus.gdpo.aime24-abs}{1.7}
\setval{4b.mt-grpo.minus.gdpo.aime25}{$-$3.7}
\setval{4b.mt-grpo.minus.gdpo.aime25-abs}{3.7}
\setval{4b.mt-grpo.minus.gdpo.amc23}{+0.2}
\setval{4b.mt-grpo.minus.gdpo.amc23-abs}{0.2}
\setval{4b.mt-grpo.minus.gdpo.avg}{$-$1.2}
\setval{4b.mt-grpo.minus.gdpo.avg-abs}{1.2}
\setval{4b.mt-grpo.minus.gdpo.code}{$-$1.8}
\setval{4b.mt-grpo.minus.gdpo.code-abs}{1.8}
\setval{4b.mt-grpo.minus.gdpo.creative-writing-v3}{$-$1.5}
\setval{4b.mt-grpo.minus.gdpo.creative-writing-v3-abs}{1.5}
\setval{4b.mt-grpo.minus.gdpo.guidebench}{$-$1.0}
\setval{4b.mt-grpo.minus.gdpo.guidebench-abs}{1.0}
\setval{4b.mt-grpo.minus.gdpo.humaneval-plus}{$-$2.7}
\setval{4b.mt-grpo.minus.gdpo.humaneval-plus-abs}{2.7}
\setval{4b.mt-grpo.minus.gdpo.if}{$-$0.7}
\setval{4b.mt-grpo.minus.gdpo.if-abs}{0.7}
\setval{4b.mt-grpo.minus.gdpo.ifbench}{$-$2.0}
\setval{4b.mt-grpo.minus.gdpo.ifbench-abs}{2.0}
\setval{4b.mt-grpo.minus.gdpo.ifeval}{+1.0}
\setval{4b.mt-grpo.minus.gdpo.ifeval-abs}{1.0}
\setval{4b.mt-grpo.minus.gdpo.judge2.avg}{$-$1.1}
\setval{4b.mt-grpo.minus.gdpo.judge2.avg-abs}{1.1}
\setval{4b.mt-grpo.minus.gdpo.judge2.creative-writing-v3}{$-$0.3}
\setval{4b.mt-grpo.minus.gdpo.judge2.creative-writing-v3-abs}{0.3}
\setval{4b.mt-grpo.minus.gdpo.judge2.longbench-write}{$-$1.5}
\setval{4b.mt-grpo.minus.gdpo.judge2.longbench-write-abs}{1.5}
\setval{4b.mt-grpo.minus.gdpo.judge2.writing}{$-$0.9}
\setval{4b.mt-grpo.minus.gdpo.judge2.writing-abs}{0.9}
\setval{4b.mt-grpo.minus.gdpo.judge2.writingbench}{$-$0.8}
\setval{4b.mt-grpo.minus.gdpo.judge2.writingbench-abs}{0.8}
\setval{4b.mt-grpo.minus.gdpo.livecodebench}{$-$1.8}
\setval{4b.mt-grpo.minus.gdpo.livecodebench-abs}{1.8}
\setval{4b.mt-grpo.minus.gdpo.longbench-write}{$-$2.2}
\setval{4b.mt-grpo.minus.gdpo.longbench-write-abs}{2.2}
\setval{4b.mt-grpo.minus.gdpo.math}{$-$1.0}
\setval{4b.mt-grpo.minus.gdpo.math-abs}{1.0}
\setval{4b.mt-grpo.minus.gdpo.math500}{$-$2.3}
\setval{4b.mt-grpo.minus.gdpo.math500-abs}{2.3}
\setval{4b.mt-grpo.minus.gdpo.mbpp-plus}{$-$0.8}
\setval{4b.mt-grpo.minus.gdpo.mbpp-plus-abs}{0.8}
\setval{4b.mt-grpo.minus.gdpo.minerva}{+0.5}
\setval{4b.mt-grpo.minus.gdpo.minerva-abs}{0.5}
\setval{4b.mt-grpo.minus.gdpo.olympiad-bench}{$-$2.2}
\setval{4b.mt-grpo.minus.gdpo.olympiad-bench-abs}{2.2}
\setval{4b.mt-grpo.minus.gdpo.writing}{$-$1.8}
\setval{4b.mt-grpo.minus.gdpo.writing-abs}{1.8}
\setval{4b.mt-grpo.minus.gdpo.writingbench}{$-$1.8}
\setval{4b.mt-grpo.minus.gdpo.writingbench-abs}{1.8}
\setval{4b.mt-grpo.minus.geoma.aime24}{$-$0.8}
\setval{4b.mt-grpo.minus.geoma.aime24-abs}{0.8}
\setval{4b.mt-grpo.minus.geoma.aime25}{$-$4.5}
\setval{4b.mt-grpo.minus.geoma.aime25-abs}{4.5}
\setval{4b.mt-grpo.minus.geoma.amc23}{+4.4}
\setval{4b.mt-grpo.minus.geoma.amc23-abs}{4.4}
\setval{4b.mt-grpo.minus.geoma.avg}{$-$0.7}
\setval{4b.mt-grpo.minus.geoma.avg-abs}{0.7}
\setval{4b.mt-grpo.minus.geoma.code}{$-$1.4}
\setval{4b.mt-grpo.minus.geoma.code-abs}{1.4}
\setval{4b.mt-grpo.minus.geoma.creative-writing-v3}{$-$1.4}
\setval{4b.mt-grpo.minus.geoma.creative-writing-v3-abs}{1.4}
\setval{4b.mt-grpo.minus.geoma.guidebench}{+1.2}
\setval{4b.mt-grpo.minus.geoma.guidebench-abs}{1.2}
\setval{4b.mt-grpo.minus.geoma.humaneval-plus}{$-$1.2}
\setval{4b.mt-grpo.minus.geoma.humaneval-plus-abs}{1.2}
\setval{4b.mt-grpo.minus.geoma.if}{$-$0.5}
\setval{4b.mt-grpo.minus.geoma.if-abs}{0.5}
\setval{4b.mt-grpo.minus.geoma.ifbench}{$-$1.7}
\setval{4b.mt-grpo.minus.geoma.ifbench-abs}{1.7}
\setval{4b.mt-grpo.minus.geoma.ifeval}{$-$1.1}
\setval{4b.mt-grpo.minus.geoma.ifeval-abs}{1.1}
\setval{4b.mt-grpo.minus.geoma.judge2.avg}{$-$0.3}
\setval{4b.mt-grpo.minus.geoma.judge2.avg-abs}{0.3}
\setval{4b.mt-grpo.minus.geoma.judge2.creative-writing-v3}{+0.9}
\setval{4b.mt-grpo.minus.geoma.judge2.creative-writing-v3-abs}{0.9}
\setval{4b.mt-grpo.minus.geoma.judge2.longbench-write}{+3.1}
\setval{4b.mt-grpo.minus.geoma.judge2.longbench-write-abs}{3.1}
\setval{4b.mt-grpo.minus.geoma.judge2.writing}{+2.2}
\setval{4b.mt-grpo.minus.geoma.judge2.writing-abs}{2.2}
\setval{4b.mt-grpo.minus.geoma.judge2.writingbench}{+2.6}
\setval{4b.mt-grpo.minus.geoma.judge2.writingbench-abs}{2.6}
\setval{4b.mt-grpo.minus.geoma.livecodebench}{$-$3.0}
\setval{4b.mt-grpo.minus.geoma.livecodebench-abs}{3.0}
\setval{4b.mt-grpo.minus.geoma.longbench-write}{+1.0}
\setval{4b.mt-grpo.minus.geoma.longbench-write-abs}{1.0}
\setval{4b.mt-grpo.minus.geoma.math}{$-$0.8}
\setval{4b.mt-grpo.minus.geoma.math-abs}{0.8}
\setval{4b.mt-grpo.minus.geoma.math500}{$-$1.9}
\setval{4b.mt-grpo.minus.geoma.math500-abs}{1.9}
\setval{4b.mt-grpo.minus.geoma.mbpp-plus}{$-$0.2}
\setval{4b.mt-grpo.minus.geoma.mbpp-plus-abs}{0.2}
\setval{4b.mt-grpo.minus.geoma.minerva}{$-$1.0}
\setval{4b.mt-grpo.minus.geoma.minerva-abs}{1.0}
\setval{4b.mt-grpo.minus.geoma.olympiad-bench}{$-$1.3}
\setval{4b.mt-grpo.minus.geoma.olympiad-bench-abs}{1.3}
\setval{4b.mt-grpo.minus.geoma.writing}{+0.1}
\setval{4b.mt-grpo.minus.geoma.writing-abs}{0.1}
\setval{4b.mt-grpo.minus.geoma.writingbench}{+0.6}
\setval{4b.mt-grpo.minus.geoma.writingbench-abs}{0.6}
\setval{4b.mt-grpo.minus.grpo-nodiv.aime24}{$-$1.7}
\setval{4b.mt-grpo.minus.grpo-nodiv.aime24-abs}{1.7}
\setval{4b.mt-grpo.minus.grpo-nodiv.aime25}{$-$2.2}
\setval{4b.mt-grpo.minus.grpo-nodiv.aime25-abs}{2.2}
\setval{4b.mt-grpo.minus.grpo-nodiv.amc23}{+5.2}
\setval{4b.mt-grpo.minus.grpo-nodiv.amc23-abs}{5.2}
\setval{4b.mt-grpo.minus.grpo-nodiv.avg}{+0.0}
\setval{4b.mt-grpo.minus.grpo-nodiv.avg-abs}{0.0}
\setval{4b.mt-grpo.minus.grpo-nodiv.code}{$-$1.5}
\setval{4b.mt-grpo.minus.grpo-nodiv.code-abs}{1.5}
\setval{4b.mt-grpo.minus.grpo-nodiv.creative-writing-v3}{+2.0}
\setval{4b.mt-grpo.minus.grpo-nodiv.creative-writing-v3-abs}{2.0}
\setval{4b.mt-grpo.minus.grpo-nodiv.guidebench}{+0.9}
\setval{4b.mt-grpo.minus.grpo-nodiv.guidebench-abs}{0.9}
\setval{4b.mt-grpo.minus.grpo-nodiv.humaneval-plus}{$-$1.4}
\setval{4b.mt-grpo.minus.grpo-nodiv.humaneval-plus-abs}{1.4}
\setval{4b.mt-grpo.minus.grpo-nodiv.if}{+0.3}
\setval{4b.mt-grpo.minus.grpo-nodiv.if-abs}{0.3}
\setval{4b.mt-grpo.minus.grpo-nodiv.ifbench}{+0.7}
\setval{4b.mt-grpo.minus.grpo-nodiv.ifbench-abs}{0.7}
\setval{4b.mt-grpo.minus.grpo-nodiv.ifeval}{$-$0.6}
\setval{4b.mt-grpo.minus.grpo-nodiv.ifeval-abs}{0.6}
\setval{4b.mt-grpo.minus.grpo-nodiv.judge2.avg}{+0.2}
\setval{4b.mt-grpo.minus.grpo-nodiv.judge2.avg-abs}{0.2}
\setval{4b.mt-grpo.minus.grpo-nodiv.judge2.creative-writing-v3}{+3.6}
\setval{4b.mt-grpo.minus.grpo-nodiv.judge2.creative-writing-v3-abs}{3.6}
\setval{4b.mt-grpo.minus.grpo-nodiv.judge2.longbench-write}{+2.2}
\setval{4b.mt-grpo.minus.grpo-nodiv.judge2.longbench-write-abs}{2.2}
\setval{4b.mt-grpo.minus.grpo-nodiv.judge2.writing}{+3.1}
\setval{4b.mt-grpo.minus.grpo-nodiv.judge2.writing-abs}{3.1}
\setval{4b.mt-grpo.minus.grpo-nodiv.judge2.writingbench}{+3.7}
\setval{4b.mt-grpo.minus.grpo-nodiv.judge2.writingbench-abs}{3.7}
\setval{4b.mt-grpo.minus.grpo-nodiv.livecodebench}{$-$1.8}
\setval{4b.mt-grpo.minus.grpo-nodiv.livecodebench-abs}{1.8}
\setval{4b.mt-grpo.minus.grpo-nodiv.longbench-write}{+0.8}
\setval{4b.mt-grpo.minus.grpo-nodiv.longbench-write-abs}{0.8}
\setval{4b.mt-grpo.minus.grpo-nodiv.math}{$-$0.5}
\setval{4b.mt-grpo.minus.grpo-nodiv.math-abs}{0.5}
\setval{4b.mt-grpo.minus.grpo-nodiv.math500}{$-$2.5}
\setval{4b.mt-grpo.minus.grpo-nodiv.math500-abs}{2.5}
\setval{4b.mt-grpo.minus.grpo-nodiv.mbpp-plus}{$-$1.1}
\setval{4b.mt-grpo.minus.grpo-nodiv.mbpp-plus-abs}{1.1}
\setval{4b.mt-grpo.minus.grpo-nodiv.minerva}{+0.2}
\setval{4b.mt-grpo.minus.grpo-nodiv.minerva-abs}{0.2}
\setval{4b.mt-grpo.minus.grpo-nodiv.olympiad-bench}{$-$2.3}
\setval{4b.mt-grpo.minus.grpo-nodiv.olympiad-bench-abs}{2.3}
\setval{4b.mt-grpo.minus.grpo-nodiv.writing}{+2.4}
\setval{4b.mt-grpo.minus.grpo-nodiv.writing-abs}{2.4}
\setval{4b.mt-grpo.minus.grpo-nodiv.writingbench}{+4.5}
\setval{4b.mt-grpo.minus.grpo-nodiv.writingbench-abs}{4.5}
\setval{4b.mt-grpo.minus.grpo.aime24}{$-$1.3}
\setval{4b.mt-grpo.minus.grpo.aime24-abs}{1.3}
\setval{4b.mt-grpo.minus.grpo.aime25}{$-$6.5}
\setval{4b.mt-grpo.minus.grpo.aime25-abs}{6.5}
\setval{4b.mt-grpo.minus.grpo.amc23}{+5.1}
\setval{4b.mt-grpo.minus.grpo.amc23-abs}{5.1}
\setval{4b.mt-grpo.minus.grpo.avg}{$-$1.6}
\setval{4b.mt-grpo.minus.grpo.avg-abs}{1.6}
\setval{4b.mt-grpo.minus.grpo.code}{$-$2.0}
\setval{4b.mt-grpo.minus.grpo.code-abs}{2.0}
\setval{4b.mt-grpo.minus.grpo.creative-writing-v3}{$-$2.8}
\setval{4b.mt-grpo.minus.grpo.creative-writing-v3-abs}{2.8}
\setval{4b.mt-grpo.minus.grpo.guidebench}{+1.7}
\setval{4b.mt-grpo.minus.grpo.guidebench-abs}{1.7}
\setval{4b.mt-grpo.minus.grpo.humaneval-plus}{$-$3.7}
\setval{4b.mt-grpo.minus.grpo.humaneval-plus-abs}{3.7}
\setval{4b.mt-grpo.minus.grpo.if}{$-$0.5}
\setval{4b.mt-grpo.minus.grpo.if-abs}{0.5}
\setval{4b.mt-grpo.minus.grpo.ifbench}{$-$3.3}
\setval{4b.mt-grpo.minus.grpo.ifbench-abs}{3.3}
\setval{4b.mt-grpo.minus.grpo.ifeval}{+0.2}
\setval{4b.mt-grpo.minus.grpo.ifeval-abs}{0.2}
\setval{4b.mt-grpo.minus.grpo.judge2.avg}{$-$1.1}
\setval{4b.mt-grpo.minus.grpo.judge2.avg-abs}{1.1}
\setval{4b.mt-grpo.minus.grpo.judge2.creative-writing-v3}{$-$1.0}
\setval{4b.mt-grpo.minus.grpo.judge2.creative-writing-v3-abs}{1.0}
\setval{4b.mt-grpo.minus.grpo.judge2.longbench-write}{+0.1}
\setval{4b.mt-grpo.minus.grpo.judge2.longbench-write-abs}{0.1}
\setval{4b.mt-grpo.minus.grpo.judge2.writing}{$-$0.4}
\setval{4b.mt-grpo.minus.grpo.judge2.writing-abs}{0.4}
\setval{4b.mt-grpo.minus.grpo.judge2.writingbench}{$-$0.2}
\setval{4b.mt-grpo.minus.grpo.judge2.writingbench-abs}{0.2}
\setval{4b.mt-grpo.minus.grpo.livecodebench}{$-$1.4}
\setval{4b.mt-grpo.minus.grpo.livecodebench-abs}{1.4}
\setval{4b.mt-grpo.minus.grpo.longbench-write}{$-$2.4}
\setval{4b.mt-grpo.minus.grpo.longbench-write-abs}{2.4}
\setval{4b.mt-grpo.minus.grpo.math}{$-$1.2}
\setval{4b.mt-grpo.minus.grpo.math-abs}{1.2}
\setval{4b.mt-grpo.minus.grpo.math500}{$-$3.3}
\setval{4b.mt-grpo.minus.grpo.math500-abs}{3.3}
\setval{4b.mt-grpo.minus.grpo.mbpp-plus}{$-$1.1}
\setval{4b.mt-grpo.minus.grpo.mbpp-plus-abs}{1.1}
\setval{4b.mt-grpo.minus.grpo.minerva}{+0.7}
\setval{4b.mt-grpo.minus.grpo.minerva-abs}{0.7}
\setval{4b.mt-grpo.minus.grpo.olympiad-bench}{$-$2.1}
\setval{4b.mt-grpo.minus.grpo.olympiad-bench-abs}{2.1}
\setval{4b.mt-grpo.minus.grpo.writing}{$-$3.1}
\setval{4b.mt-grpo.minus.grpo.writing-abs}{3.1}
\setval{4b.mt-grpo.minus.grpo.writingbench}{$-$4.1}
\setval{4b.mt-grpo.minus.grpo.writingbench-abs}{4.1}
\setval{4b.mt-grpo.minus.papo.aime24}{+1.5}
\setval{4b.mt-grpo.minus.papo.aime24-abs}{1.5}
\setval{4b.mt-grpo.minus.papo.aime25}{$-$1.4}
\setval{4b.mt-grpo.minus.papo.aime25-abs}{1.4}
\setval{4b.mt-grpo.minus.papo.amc23}{+6.2}
\setval{4b.mt-grpo.minus.papo.amc23-abs}{6.2}
\setval{4b.mt-grpo.minus.papo.avg}{+0.5}
\setval{4b.mt-grpo.minus.papo.avg-abs}{0.5}
\setval{4b.mt-grpo.minus.papo.code}{$-$1.8}
\setval{4b.mt-grpo.minus.papo.code-abs}{1.8}
\setval{4b.mt-grpo.minus.papo.creative-writing-v3}{$-$2.0}
\setval{4b.mt-grpo.minus.papo.creative-writing-v3-abs}{2.0}
\setval{4b.mt-grpo.minus.papo.guidebench}{$-$1.1}
\setval{4b.mt-grpo.minus.papo.guidebench-abs}{1.1}
\setval{4b.mt-grpo.minus.papo.humaneval-plus}{$-$4.0}
\setval{4b.mt-grpo.minus.papo.humaneval-plus-abs}{4.0}
\setval{4b.mt-grpo.minus.papo.if}{+0.3}
\setval{4b.mt-grpo.minus.papo.if-abs}{0.3}
\setval{4b.mt-grpo.minus.papo.ifbench}{+1.0}
\setval{4b.mt-grpo.minus.papo.ifbench-abs}{1.0}
\setval{4b.mt-grpo.minus.papo.ifeval}{+0.8}
\setval{4b.mt-grpo.minus.papo.ifeval-abs}{0.8}
\setval{4b.mt-grpo.minus.papo.judge2.avg}{+0.4}
\setval{4b.mt-grpo.minus.papo.judge2.avg-abs}{0.4}
\setval{4b.mt-grpo.minus.papo.judge2.creative-writing-v3}{$-$0.2}
\setval{4b.mt-grpo.minus.papo.judge2.creative-writing-v3-abs}{0.2}
\setval{4b.mt-grpo.minus.papo.judge2.longbench-write}{+3.2}
\setval{4b.mt-grpo.minus.papo.judge2.longbench-write-abs}{3.2}
\setval{4b.mt-grpo.minus.papo.judge2.writing}{+1.4}
\setval{4b.mt-grpo.minus.papo.judge2.writing-abs}{1.4}
\setval{4b.mt-grpo.minus.papo.judge2.writingbench}{+1.1}
\setval{4b.mt-grpo.minus.papo.judge2.writingbench-abs}{1.1}
\setval{4b.mt-grpo.minus.papo.livecodebench}{$-$0.9}
\setval{4b.mt-grpo.minus.papo.livecodebench-abs}{0.9}
\setval{4b.mt-grpo.minus.papo.longbench-write}{+2.8}
\setval{4b.mt-grpo.minus.papo.longbench-write-abs}{2.8}
\setval{4b.mt-grpo.minus.papo.math}{+1.2}
\setval{4b.mt-grpo.minus.papo.math-abs}{1.2}
\setval{4b.mt-grpo.minus.papo.math500}{+0.5}
\setval{4b.mt-grpo.minus.papo.math500-abs}{0.5}
\setval{4b.mt-grpo.minus.papo.mbpp-plus}{$-$0.7}
\setval{4b.mt-grpo.minus.papo.mbpp-plus-abs}{0.7}
\setval{4b.mt-grpo.minus.papo.minerva}{$-$0.6}
\setval{4b.mt-grpo.minus.papo.minerva-abs}{0.6}
\setval{4b.mt-grpo.minus.papo.olympiad-bench}{+1.0}
\setval{4b.mt-grpo.minus.papo.olympiad-bench-abs}{1.0}
\setval{4b.mt-grpo.minus.papo.writing}{+1.5}
\setval{4b.mt-grpo.minus.papo.writing-abs}{1.5}
\setval{4b.mt-grpo.minus.papo.writingbench}{+3.6}
\setval{4b.mt-grpo.minus.papo.writingbench-abs}{3.6}
\setval{4b.mt-grpo.minus.saw.aime24}{$-$1.8}
\setval{4b.mt-grpo.minus.saw.aime24-abs}{1.8}
\setval{4b.mt-grpo.minus.saw.aime25}{$-$5.0}
\setval{4b.mt-grpo.minus.saw.aime25-abs}{5.0}
\setval{4b.mt-grpo.minus.saw.amc23}{+3.1}
\setval{4b.mt-grpo.minus.saw.amc23-abs}{3.1}
\setval{4b.mt-grpo.minus.saw.avg}{$-$1.1}
\setval{4b.mt-grpo.minus.saw.avg-abs}{1.1}
\setval{4b.mt-grpo.minus.saw.code}{$-$0.6}
\setval{4b.mt-grpo.minus.saw.code-abs}{0.6}
\setval{4b.mt-grpo.minus.saw.creative-writing-v3}{$-$0.6}
\setval{4b.mt-grpo.minus.saw.creative-writing-v3-abs}{0.6}
\setval{4b.mt-grpo.minus.saw.guidebench}{$-$1.3}
\setval{4b.mt-grpo.minus.saw.guidebench-abs}{1.3}
\setval{4b.mt-grpo.minus.saw.humaneval-plus}{$-$1.6}
\setval{4b.mt-grpo.minus.saw.humaneval-plus-abs}{1.6}
\setval{4b.mt-grpo.minus.saw.if}{$-$2.5}
\setval{4b.mt-grpo.minus.saw.if-abs}{2.5}
\setval{4b.mt-grpo.minus.saw.ifbench}{$-$6.0}
\setval{4b.mt-grpo.minus.saw.ifbench-abs}{6.0}
\setval{4b.mt-grpo.minus.saw.ifeval}{$-$0.2}
\setval{4b.mt-grpo.minus.saw.ifeval-abs}{0.2}
\setval{4b.mt-grpo.minus.saw.judge2.avg}{$-$1.2}
\setval{4b.mt-grpo.minus.saw.judge2.avg-abs}{1.2}
\setval{4b.mt-grpo.minus.saw.judge2.creative-writing-v3}{$-$1.5}
\setval{4b.mt-grpo.minus.saw.judge2.creative-writing-v3-abs}{1.5}
\setval{4b.mt-grpo.minus.saw.judge2.longbench-write}{+0.7}
\setval{4b.mt-grpo.minus.saw.judge2.longbench-write-abs}{0.7}
\setval{4b.mt-grpo.minus.saw.judge2.writing}{$-$0.3}
\setval{4b.mt-grpo.minus.saw.judge2.writing-abs}{0.3}
\setval{4b.mt-grpo.minus.saw.judge2.writingbench}{$-$0.1}
\setval{4b.mt-grpo.minus.saw.judge2.writingbench-abs}{0.1}
\setval{4b.mt-grpo.minus.saw.livecodebench}{$-$0.2}
\setval{4b.mt-grpo.minus.saw.livecodebench-abs}{0.2}
\setval{4b.mt-grpo.minus.saw.longbench-write}{+0.6}
\setval{4b.mt-grpo.minus.saw.longbench-write-abs}{0.6}
\setval{4b.mt-grpo.minus.saw.math}{$-$1.3}
\setval{4b.mt-grpo.minus.saw.math-abs}{1.3}
\setval{4b.mt-grpo.minus.saw.math500}{$-$0.8}
\setval{4b.mt-grpo.minus.saw.math500-abs}{0.8}
\setval{4b.mt-grpo.minus.saw.mbpp-plus}{$-$0.1}
\setval{4b.mt-grpo.minus.saw.mbpp-plus-abs}{0.1}
\setval{4b.mt-grpo.minus.saw.minerva}{$-$2.2}
\setval{4b.mt-grpo.minus.saw.minerva-abs}{2.2}
\setval{4b.mt-grpo.minus.saw.olympiad-bench}{$-$1.3}
\setval{4b.mt-grpo.minus.saw.olympiad-bench-abs}{1.3}
\setval{4b.mt-grpo.minus.saw.writing}{+0.1}
\setval{4b.mt-grpo.minus.saw.writing-abs}{0.1}
\setval{4b.mt-grpo.minus.saw.writingbench}{+0.3}
\setval{4b.mt-grpo.minus.saw.writingbench-abs}{0.3}
\setval{4b.mt-grpo.passing-at-ceiling.code.early}{0.65}
\setval{4b.mt-grpo.passing-at-ceiling.code.late}{0.44}
\setval{4b.mt-grpo.passing-at-ceiling.code.run}{0.50}
\setval{4b.mt-grpo.passing-at-ceiling.if.early}{0.29}
\setval{4b.mt-grpo.passing-at-ceiling.if.late}{0.46}
\setval{4b.mt-grpo.passing-at-ceiling.if.run}{0.42}
\setval{4b.mt-grpo.passing-at-ceiling.math.early}{1.00}
\setval{4b.mt-grpo.passing-at-ceiling.math.late}{1.00}
\setval{4b.mt-grpo.passing-at-ceiling.math.run}{1.00}
\setval{4b.mt-grpo.passing-at-ceiling.writing.early}{0.00}
\setval{4b.mt-grpo.passing-at-ceiling.writing.late}{0.00}
\setval{4b.mt-grpo.passing-at-ceiling.writing.run}{0.00}
\setval{4b.mt-grpo.primary-level.code.early}{0.51}
\setval{4b.mt-grpo.primary-level.code.late}{0.61}
\setval{4b.mt-grpo.primary-level.code.run}{0.59}
\setval{4b.mt-grpo.primary-level.if.early}{0.35}
\setval{4b.mt-grpo.primary-level.if.late}{0.70}
\setval{4b.mt-grpo.primary-level.if.run}{0.64}
\setval{4b.mt-grpo.primary-level.math.early}{0.54}
\setval{4b.mt-grpo.primary-level.math.late}{0.70}
\setval{4b.mt-grpo.primary-level.math.run}{0.63}
\setval{4b.mt-grpo.primary-level.writing.early}{0.52}
\setval{4b.mt-grpo.primary-level.writing.late}{0.75}
\setval{4b.mt-grpo.primary-level.writing.run}{0.69}
\setval{4b.papo.aux-change.code}{+0.21}
\setval{4b.papo.aux-change.if}{+0.25}
\setval{4b.papo.aux-change.math}{+0.15}
\setval{4b.papo.aux-change.writing}{+0.21}
\setval{4b.papo.aux-change.writing-rm}{+0.37}
\setval{4b.papo.aux-full.code.early}{0.84}
\setval{4b.papo.aux-full.code.late}{0.99}
\setval{4b.papo.aux-full.code.run}{0.98}
\setval{4b.papo.aux-full.if.early}{0.69}
\setval{4b.papo.aux-full.if.late}{0.92}
\setval{4b.papo.aux-full.if.run}{0.88}
\setval{4b.papo.aux-full.math.early}{0.71}
\setval{4b.papo.aux-full.math.late}{0.97}
\setval{4b.papo.aux-full.math.run}{0.94}
\setval{4b.papo.aux-full.writing-rm.early}{0.09}
\setval{4b.papo.aux-full.writing-rm.late}{0.65}
\setval{4b.papo.aux-full.writing-rm.run}{0.53}
\setval{4b.papo.aux-full.writing.early}{0.69}
\setval{4b.papo.aux-full.writing.late}{0.81}
\setval{4b.papo.aux-full.writing.run}{0.78}
\setval{4b.papo.aux-level.code.early}{0.95}
\setval{4b.papo.aux-level.code.late}{1.00}
\setval{4b.papo.aux-level.code.run}{1.00}
\setval{4b.papo.aux-level.if.early}{0.85}
\setval{4b.papo.aux-level.if.late}{0.95}
\setval{4b.papo.aux-level.if.run}{0.93}
\setval{4b.papo.aux-level.math.early}{0.94}
\setval{4b.papo.aux-level.math.late}{1.00}
\setval{4b.papo.aux-level.math.run}{0.99}
\setval{4b.papo.aux-level.writing-rm.early}{0.78}
\setval{4b.papo.aux-level.writing-rm.late}{0.97}
\setval{4b.papo.aux-level.writing-rm.run}{0.94}
\setval{4b.papo.aux-level.writing.early}{0.92}
\setval{4b.papo.aux-level.writing.late}{0.98}
\setval{4b.papo.aux-level.writing.run}{0.97}
\setval{4b.papo.bench.aime24}{9.3}
\setval{4b.papo.bench.aime25}{7.5}
\setval{4b.papo.bench.amc23}{53.5}
\setval{4b.papo.bench.creative-writing-v3}{46.5}
\setval{4b.papo.bench.guidebench}{80.7}
\setval{4b.papo.bench.humaneval-plus}{83.0}
\setval{4b.papo.bench.ifbench}{33.3}
\setval{4b.papo.bench.ifeval}{85.9}
\setval{4b.papo.bench.livecodebench}{23.5}
\setval{4b.papo.bench.longbench-write}{64.0}
\setval{4b.papo.bench.math500}{76.5}
\setval{4b.papo.bench.mbpp-plus}{70.7}
\setval{4b.papo.bench.minerva}{46.5}
\setval{4b.papo.bench.olympiad-bench}{41.5}
\setval{4b.papo.bench.writingbench}{66.5}
\setval{4b.papo.eval.avg}{52.6}
\setval{4b.papo.eval.code}{59.1}
\setval{4b.papo.eval.if}{66.6}
\setval{4b.papo.eval.math}{39.1}
\setval{4b.papo.eval.writing}{59.0}
\setval{4b.papo.gate.code.against.early}{0.02}
\setval{4b.papo.gate.code.against.late}{0.00}
\setval{4b.papo.gate.code.against.run}{0.00}
\setval{4b.papo.gate.code.agreement}{+0.001}
\setval{4b.papo.gate.code.ceiling.early}{0.06}
\setval{4b.papo.gate.code.ceiling.late}{0.01}
\setval{4b.papo.gate.code.ceiling.run}{0.01}
\setval{4b.papo.gate.code.cosine.early}{0.89}
\setval{4b.papo.gate.code.cosine.late}{0.84}
\setval{4b.papo.gate.code.cosine.run}{0.86}
\setval{4b.papo.gate.if.against.early}{0.04}
\setval{4b.papo.gate.if.against.late}{0.04}
\setval{4b.papo.gate.if.against.run}{0.04}
\setval{4b.papo.gate.if.agreement}{+0.003}
\setval{4b.papo.gate.if.ceiling.early}{0.05}
\setval{4b.papo.gate.if.ceiling.late}{0.15}
\setval{4b.papo.gate.if.ceiling.run}{0.09}
\setval{4b.papo.gate.if.cosine.early}{0.76}
\setval{4b.papo.gate.if.cosine.late}{0.54}
\setval{4b.papo.gate.if.cosine.run}{0.63}
\setval{4b.papo.gate.math.against.early}{0.00}
\setval{4b.papo.gate.math.against.late}{0.00}
\setval{4b.papo.gate.math.against.run}{0.00}
\setval{4b.papo.gate.math.ceiling.early}{0.17}
\setval{4b.papo.gate.math.ceiling.late}{0.04}
\setval{4b.papo.gate.math.ceiling.run}{0.06}
\setval{4b.papo.gate.math.cosine.early}{0.90}
\setval{4b.papo.gate.math.cosine.late}{0.98}
\setval{4b.papo.gate.math.cosine.run}{0.97}
\setval{4b.papo.gate.writing-rm.against.early}{0.20}
\setval{4b.papo.gate.writing-rm.against.late}{0.50}
\setval{4b.papo.gate.writing-rm.against.run}{0.46}
\setval{4b.papo.gate.writing-rm.agreement}{+0.057}
\setval{4b.papo.gate.writing-rm.ceiling.early}{0.00}
\setval{4b.papo.gate.writing-rm.ceiling.late}{0.00}
\setval{4b.papo.gate.writing-rm.ceiling.run}{0.00}
\setval{4b.papo.gate.writing-rm.cosine.early}{0.69}
\setval{4b.papo.gate.writing-rm.cosine.late}{0.16}
\setval{4b.papo.gate.writing-rm.cosine.run}{0.25}
\setval{4b.papo.gate.writing.against.early}{0.25}
\setval{4b.papo.gate.writing.against.late}{0.52}
\setval{4b.papo.gate.writing.against.run}{0.48}
\setval{4b.papo.gate.writing.agreement}{$-$0.028}
\setval{4b.papo.gate.writing.ceiling.early}{0.00}
\setval{4b.papo.gate.writing.ceiling.late}{0.00}
\setval{4b.papo.gate.writing.ceiling.run}{0.00}
\setval{4b.papo.gate.writing.cosine.early}{0.55}
\setval{4b.papo.gate.writing.cosine.late}{0.10}
\setval{4b.papo.gate.writing.cosine.run}{0.16}
\setval{4b.papo.judge2.avg}{50.4}
\setval{4b.papo.judge2.creative-writing-v3}{34.2}
\setval{4b.papo.judge2.longbench-write}{57.5}
\setval{4b.papo.judge2.writing}{48.0}
\setval{4b.papo.judge2.writingbench}{52.1}
\setval{4b.papo.last-step}{300}
\setval{4b.papo.minus.base.aime24}{+1.5}
\setval{4b.papo.minus.base.aime24-abs}{1.5}
\setval{4b.papo.minus.base.aime25}{+0.4}
\setval{4b.papo.minus.base.aime25-abs}{0.4}
\setval{4b.papo.minus.base.amc23}{+9.6}
\setval{4b.papo.minus.base.amc23-abs}{9.6}
\setval{4b.papo.minus.base.avg}{+18.1}
\setval{4b.papo.minus.base.avg-abs}{18.1}
\setval{4b.papo.minus.base.code}{+14.5}
\setval{4b.papo.minus.base.code-abs}{14.5}
\setval{4b.papo.minus.base.creative-writing-v3}{+21.8}
\setval{4b.papo.minus.base.creative-writing-v3-abs}{21.8}
\setval{4b.papo.minus.base.guidebench}{+55.9}
\setval{4b.papo.minus.base.guidebench-abs}{55.9}
\setval{4b.papo.minus.base.humaneval-plus}{+18.0}
\setval{4b.papo.minus.base.humaneval-plus-abs}{18.0}
\setval{4b.papo.minus.base.if}{+36.9}
\setval{4b.papo.minus.base.if-abs}{36.9}
\setval{4b.papo.minus.base.ifbench}{+16.3}
\setval{4b.papo.minus.base.ifbench-abs}{16.3}
\setval{4b.papo.minus.base.ifeval}{+38.5}
\setval{4b.papo.minus.base.ifeval-abs}{38.5}
\setval{4b.papo.minus.base.judge2.avg}{+16.5}
\setval{4b.papo.minus.base.judge2.avg-abs}{16.5}
\setval{4b.papo.minus.base.judge2.creative-writing-v3}{+13.4}
\setval{4b.papo.minus.base.judge2.creative-writing-v3-abs}{13.4}
\setval{4b.papo.minus.base.judge2.longbench-write}{+16.6}
\setval{4b.papo.minus.base.judge2.longbench-write-abs}{16.6}
\setval{4b.papo.minus.base.judge2.writing}{+17.2}
\setval{4b.papo.minus.base.judge2.writing-abs}{17.2}
\setval{4b.papo.minus.base.judge2.writingbench}{+21.7}
\setval{4b.papo.minus.base.judge2.writingbench-abs}{21.7}
\setval{4b.papo.minus.base.livecodebench}{+10.3}
\setval{4b.papo.minus.base.livecodebench-abs}{10.3}
\setval{4b.papo.minus.base.longbench-write}{+19.5}
\setval{4b.papo.minus.base.longbench-write-abs}{19.5}
\setval{4b.papo.minus.base.math}{+6.9}
\setval{4b.papo.minus.base.math-abs}{6.9}
\setval{4b.papo.minus.base.math500}{+9.3}
\setval{4b.papo.minus.base.math500-abs}{9.3}
\setval{4b.papo.minus.base.mbpp-plus}{+15.1}
\setval{4b.papo.minus.base.mbpp-plus-abs}{15.1}
\setval{4b.papo.minus.base.minerva}{+14.9}
\setval{4b.papo.minus.base.minerva-abs}{14.9}
\setval{4b.papo.minus.base.olympiad-bench}{+5.9}
\setval{4b.papo.minus.base.olympiad-bench-abs}{5.9}
\setval{4b.papo.minus.base.writing}{+25.1}
\setval{4b.papo.minus.base.writing-abs}{25.1}
\setval{4b.papo.minus.base.writingbench}{+34.1}
\setval{4b.papo.minus.base.writingbench-abs}{34.1}
\setval{4b.papo.minus.concur-free.aime24}{$-$2.6}
\setval{4b.papo.minus.concur-free.aime24-abs}{2.6}
\setval{4b.papo.minus.concur-free.aime25}{$-$7.1}
\setval{4b.papo.minus.concur-free.aime25-abs}{7.1}
\setval{4b.papo.minus.concur-free.amc23}{$-$0.6}
\setval{4b.papo.minus.concur-free.amc23-abs}{0.6}
\setval{4b.papo.minus.concur-free.avg}{$-$2.9}
\setval{4b.papo.minus.concur-free.avg-abs}{2.9}
\setval{4b.papo.minus.concur-free.code}{+0.8}
\setval{4b.papo.minus.concur-free.code-abs}{0.8}
\setval{4b.papo.minus.concur-free.creative-writing-v3}{+1.1}
\setval{4b.papo.minus.concur-free.creative-writing-v3-abs}{1.1}
\setval{4b.papo.minus.concur-free.guidebench}{+1.7}
\setval{4b.papo.minus.concur-free.guidebench-abs}{1.7}
\setval{4b.papo.minus.concur-free.humaneval-plus}{+2.0}
\setval{4b.papo.minus.concur-free.humaneval-plus-abs}{2.0}
\setval{4b.papo.minus.concur-free.if}{$-$1.6}
\setval{4b.papo.minus.concur-free.if-abs}{1.6}
\setval{4b.papo.minus.concur-free.ifbench}{$-$5.7}
\setval{4b.papo.minus.concur-free.ifbench-abs}{5.7}
\setval{4b.papo.minus.concur-free.ifeval}{$-$0.8}
\setval{4b.papo.minus.concur-free.ifeval-abs}{0.8}
\setval{4b.papo.minus.concur-free.judge2.avg}{$-$2.4}
\setval{4b.papo.minus.concur-free.judge2.avg-abs}{2.4}
\setval{4b.papo.minus.concur-free.judge2.creative-writing-v3}{$-$1.1}
\setval{4b.papo.minus.concur-free.judge2.creative-writing-v3-abs}{1.1}
\setval{4b.papo.minus.concur-free.judge2.longbench-write}{$-$11.8}
\setval{4b.papo.minus.concur-free.judge2.longbench-write-abs}{11.8}
\setval{4b.papo.minus.concur-free.judge2.writing}{$-$5.1}
\setval{4b.papo.minus.concur-free.judge2.writing-abs}{5.1}
\setval{4b.papo.minus.concur-free.judge2.writingbench}{$-$2.3}
\setval{4b.papo.minus.concur-free.judge2.writingbench-abs}{2.3}
\setval{4b.papo.minus.concur-free.livecodebench}{$-$0.6}
\setval{4b.papo.minus.concur-free.livecodebench-abs}{0.6}
\setval{4b.papo.minus.concur-free.longbench-write}{$-$13.8}
\setval{4b.papo.minus.concur-free.longbench-write-abs}{13.8}
\setval{4b.papo.minus.concur-free.math}{$-$3.0}
\setval{4b.papo.minus.concur-free.math-abs}{3.0}
\setval{4b.papo.minus.concur-free.math500}{$-$3.8}
\setval{4b.papo.minus.concur-free.math500-abs}{3.8}
\setval{4b.papo.minus.concur-free.mbpp-plus}{+1.1}
\setval{4b.papo.minus.concur-free.mbpp-plus-abs}{1.1}
\setval{4b.papo.minus.concur-free.minerva}{+0.3}
\setval{4b.papo.minus.concur-free.minerva-abs}{0.3}
\setval{4b.papo.minus.concur-free.olympiad-bench}{$-$4.3}
\setval{4b.papo.minus.concur-free.olympiad-bench-abs}{4.3}
\setval{4b.papo.minus.concur-free.writing}{$-$7.8}
\setval{4b.papo.minus.concur-free.writing-abs}{7.8}
\setval{4b.papo.minus.concur-free.writingbench}{$-$10.8}
\setval{4b.papo.minus.concur-free.writingbench-abs}{10.8}
\setval{4b.papo.minus.concur-tied.aime24}{$-$10.2}
\setval{4b.papo.minus.concur-tied.aime24-abs}{10.2}
\setval{4b.papo.minus.concur-tied.aime25}{$-$8.9}
\setval{4b.papo.minus.concur-tied.aime25-abs}{8.9}
\setval{4b.papo.minus.concur-tied.amc23}{$-$15.7}
\setval{4b.papo.minus.concur-tied.amc23-abs}{15.7}
\setval{4b.papo.minus.concur-tied.avg}{$-$6.1}
\setval{4b.papo.minus.concur-tied.avg-abs}{6.1}
\setval{4b.papo.minus.concur-tied.code}{+0.0}
\setval{4b.papo.minus.concur-tied.code-abs}{0.0}
\setval{4b.papo.minus.concur-tied.creative-writing-v3}{$-$1.3}
\setval{4b.papo.minus.concur-tied.creative-writing-v3-abs}{1.3}
\setval{4b.papo.minus.concur-tied.guidebench}{+1.5}
\setval{4b.papo.minus.concur-tied.guidebench-abs}{1.5}
\setval{4b.papo.minus.concur-tied.humaneval-plus}{$-$1.0}
\setval{4b.papo.minus.concur-tied.humaneval-plus-abs}{1.0}
\setval{4b.papo.minus.concur-tied.if}{$-$1.6}
\setval{4b.papo.minus.concur-tied.if-abs}{1.6}
\setval{4b.papo.minus.concur-tied.ifbench}{$-$4.3}
\setval{4b.papo.minus.concur-tied.ifbench-abs}{4.3}
\setval{4b.papo.minus.concur-tied.ifeval}{$-$1.9}
\setval{4b.papo.minus.concur-tied.ifeval-abs}{1.9}
\setval{4b.papo.minus.concur-tied.judge2.avg}{$-$5.1}
\setval{4b.papo.minus.concur-tied.judge2.avg-abs}{5.1}
\setval{4b.papo.minus.concur-tied.judge2.creative-writing-v3}{+0.2}
\setval{4b.papo.minus.concur-tied.judge2.creative-writing-v3-abs}{0.2}
\setval{4b.papo.minus.concur-tied.judge2.longbench-write}{$-$10.4}
\setval{4b.papo.minus.concur-tied.judge2.longbench-write-abs}{10.4}
\setval{4b.papo.minus.concur-tied.judge2.writing}{$-$4.5}
\setval{4b.papo.minus.concur-tied.judge2.writing-abs}{4.5}
\setval{4b.papo.minus.concur-tied.judge2.writingbench}{$-$3.2}
\setval{4b.papo.minus.concur-tied.judge2.writingbench-abs}{3.2}
\setval{4b.papo.minus.concur-tied.livecodebench}{$-$1.2}
\setval{4b.papo.minus.concur-tied.livecodebench-abs}{1.2}
\setval{4b.papo.minus.concur-tied.longbench-write}{$-$13.8}
\setval{4b.papo.minus.concur-tied.longbench-write-abs}{13.8}
\setval{4b.papo.minus.concur-tied.math}{$-$9.6}
\setval{4b.papo.minus.concur-tied.math-abs}{9.6}
\setval{4b.papo.minus.concur-tied.math500}{$-$9.0}
\setval{4b.papo.minus.concur-tied.math500-abs}{9.0}
\setval{4b.papo.minus.concur-tied.mbpp-plus}{+2.1}
\setval{4b.papo.minus.concur-tied.mbpp-plus-abs}{2.1}
\setval{4b.papo.minus.concur-tied.minerva}{$-$2.6}
\setval{4b.papo.minus.concur-tied.minerva-abs}{2.6}
\setval{4b.papo.minus.concur-tied.olympiad-bench}{$-$11.4}
\setval{4b.papo.minus.concur-tied.olympiad-bench-abs}{11.4}
\setval{4b.papo.minus.concur-tied.writing}{$-$9.6}
\setval{4b.papo.minus.concur-tied.writing-abs}{9.6}
\setval{4b.papo.minus.concur-tied.writingbench}{$-$13.7}
\setval{4b.papo.minus.concur-tied.writingbench-abs}{13.7}
\setval{4b.papo.minus.concur.aime24}{$-$11.2}
\setval{4b.papo.minus.concur.aime24-abs}{11.2}
\setval{4b.papo.minus.concur.aime25}{$-$11.2}
\setval{4b.papo.minus.concur.aime25-abs}{11.2}
\setval{4b.papo.minus.concur.amc23}{$-$15.9}
\setval{4b.papo.minus.concur.amc23-abs}{15.9}
\setval{4b.papo.minus.concur.avg}{$-$6.2}
\setval{4b.papo.minus.concur.avg-abs}{6.2}
\setval{4b.papo.minus.concur.code}{$-$0.6}
\setval{4b.papo.minus.concur.code-abs}{0.6}
\setval{4b.papo.minus.concur.creative-writing-v3}{+1.8}
\setval{4b.papo.minus.concur.creative-writing-v3-abs}{1.8}
\setval{4b.papo.minus.concur.guidebench}{+3.4}
\setval{4b.papo.minus.concur.guidebench-abs}{3.4}
\setval{4b.papo.minus.concur.humaneval-plus}{+0.6}
\setval{4b.papo.minus.concur.humaneval-plus-abs}{0.6}
\setval{4b.papo.minus.concur.if}{$-$2.0}
\setval{4b.papo.minus.concur.if-abs}{2.0}
\setval{4b.papo.minus.concur.ifbench}{$-$7.0}
\setval{4b.papo.minus.concur.ifbench-abs}{7.0}
\setval{4b.papo.minus.concur.ifeval}{$-$2.3}
\setval{4b.papo.minus.concur.ifeval-abs}{2.3}
\setval{4b.papo.minus.concur.judge2.avg}{$-$5.4}
\setval{4b.papo.minus.concur.judge2.avg-abs}{5.4}
\setval{4b.papo.minus.concur.judge2.creative-writing-v3}{+1.9}
\setval{4b.papo.minus.concur.judge2.creative-writing-v3-abs}{1.9}
\setval{4b.papo.minus.concur.judge2.longbench-write}{$-$9.1}
\setval{4b.papo.minus.concur.judge2.longbench-write-abs}{9.1}
\setval{4b.papo.minus.concur.judge2.writing}{$-$3.7}
\setval{4b.papo.minus.concur.judge2.writing-abs}{3.7}
\setval{4b.papo.minus.concur.judge2.writingbench}{$-$3.9}
\setval{4b.papo.minus.concur.judge2.writingbench-abs}{3.9}
\setval{4b.papo.minus.concur.livecodebench}{$-$1.6}
\setval{4b.papo.minus.concur.livecodebench-abs}{1.6}
\setval{4b.papo.minus.concur.longbench-write}{$-$12.3}
\setval{4b.papo.minus.concur.longbench-write-abs}{12.3}
\setval{4b.papo.minus.concur.math}{$-$10.3}
\setval{4b.papo.minus.concur.math-abs}{10.3}
\setval{4b.papo.minus.concur.math500}{$-$8.8}
\setval{4b.papo.minus.concur.math500-abs}{8.8}
\setval{4b.papo.minus.concur.mbpp-plus}{$-$0.8}
\setval{4b.papo.minus.concur.mbpp-plus-abs}{0.8}
\setval{4b.papo.minus.concur.minerva}{$-$2.7}
\setval{4b.papo.minus.concur.minerva-abs}{2.7}
\setval{4b.papo.minus.concur.olympiad-bench}{$-$12.1}
\setval{4b.papo.minus.concur.olympiad-bench-abs}{12.1}
\setval{4b.papo.minus.concur.writing}{$-$7.8}
\setval{4b.papo.minus.concur.writing-abs}{7.8}
\setval{4b.papo.minus.concur.writingbench}{$-$12.9}
\setval{4b.papo.minus.concur.writingbench-abs}{12.9}
\setval{4b.papo.minus.gd2po.aime24}{$-$2.6}
\setval{4b.papo.minus.gd2po.aime24-abs}{2.6}
\setval{4b.papo.minus.gd2po.aime25}{+0.7}
\setval{4b.papo.minus.gd2po.aime25-abs}{0.7}
\setval{4b.papo.minus.gd2po.amc23}{$-$4.2}
\setval{4b.papo.minus.gd2po.amc23-abs}{4.2}
\setval{4b.papo.minus.gd2po.avg}{$-$1.5}
\setval{4b.papo.minus.gd2po.avg-abs}{1.5}
\setval{4b.papo.minus.gd2po.code}{$-$0.3}
\setval{4b.papo.minus.gd2po.code-abs}{0.3}
\setval{4b.papo.minus.gd2po.creative-writing-v3}{+1.9}
\setval{4b.papo.minus.gd2po.creative-writing-v3-abs}{1.9}
\setval{4b.papo.minus.gd2po.guidebench}{+1.6}
\setval{4b.papo.minus.gd2po.guidebench-abs}{1.6}
\setval{4b.papo.minus.gd2po.humaneval-plus}{$-$0.4}
\setval{4b.papo.minus.gd2po.humaneval-plus-abs}{0.4}
\setval{4b.papo.minus.gd2po.if}{$-$1.5}
\setval{4b.papo.minus.gd2po.if-abs}{1.5}
\setval{4b.papo.minus.gd2po.ifbench}{$-$4.3}
\setval{4b.papo.minus.gd2po.ifbench-abs}{4.3}
\setval{4b.papo.minus.gd2po.ifeval}{$-$1.8}
\setval{4b.papo.minus.gd2po.ifeval-abs}{1.8}
\setval{4b.papo.minus.gd2po.judge2.avg}{$-$1.2}
\setval{4b.papo.minus.gd2po.judge2.avg-abs}{1.2}
\setval{4b.papo.minus.gd2po.judge2.creative-writing-v3}{+1.3}
\setval{4b.papo.minus.gd2po.judge2.creative-writing-v3-abs}{1.3}
\setval{4b.papo.minus.gd2po.judge2.longbench-write}{$-$3.4}
\setval{4b.papo.minus.gd2po.judge2.longbench-write-abs}{3.4}
\setval{4b.papo.minus.gd2po.judge2.writing}{$-$0.8}
\setval{4b.papo.minus.gd2po.judge2.writing-abs}{0.8}
\setval{4b.papo.minus.gd2po.judge2.writingbench}{$-$0.4}
\setval{4b.papo.minus.gd2po.judge2.writingbench-abs}{0.4}
\setval{4b.papo.minus.gd2po.livecodebench}{+0.1}
\setval{4b.papo.minus.gd2po.livecodebench-abs}{0.1}
\setval{4b.papo.minus.gd2po.longbench-write}{$-$5.0}
\setval{4b.papo.minus.gd2po.longbench-write-abs}{5.0}
\setval{4b.papo.minus.gd2po.math}{$-$1.7}
\setval{4b.papo.minus.gd2po.math-abs}{1.7}
\setval{4b.papo.minus.gd2po.math500}{$-$1.1}
\setval{4b.papo.minus.gd2po.math500-abs}{1.1}
\setval{4b.papo.minus.gd2po.mbpp-plus}{$-$0.7}
\setval{4b.papo.minus.gd2po.mbpp-plus-abs}{0.7}
\setval{4b.papo.minus.gd2po.minerva}{$-$0.7}
\setval{4b.papo.minus.gd2po.minerva-abs}{0.7}
\setval{4b.papo.minus.gd2po.olympiad-bench}{$-$2.1}
\setval{4b.papo.minus.gd2po.olympiad-bench-abs}{2.1}
\setval{4b.papo.minus.gd2po.writing}{$-$2.3}
\setval{4b.papo.minus.gd2po.writing-abs}{2.3}
\setval{4b.papo.minus.gd2po.writingbench}{$-$3.7}
\setval{4b.papo.minus.gd2po.writingbench-abs}{3.7}
\setval{4b.papo.minus.gdpo-noaux.aime24}{$-$14.2}
\setval{4b.papo.minus.gdpo-noaux.aime24-abs}{14.2}
\setval{4b.papo.minus.gdpo-noaux.aime25}{$-$11.2}
\setval{4b.papo.minus.gdpo-noaux.aime25-abs}{11.2}
\setval{4b.papo.minus.gdpo-noaux.amc23}{$-$12.2}
\setval{4b.papo.minus.gdpo-noaux.amc23-abs}{12.2}
\setval{4b.papo.minus.gdpo-noaux.avg}{$-$4.1}
\setval{4b.papo.minus.gdpo-noaux.avg-abs}{4.1}
\setval{4b.papo.minus.gdpo-noaux.code}{$-$1.8}
\setval{4b.papo.minus.gdpo-noaux.code-abs}{1.8}
\setval{4b.papo.minus.gdpo-noaux.creative-writing-v3}{+7.2}
\setval{4b.papo.minus.gdpo-noaux.creative-writing-v3-abs}{7.2}
\setval{4b.papo.minus.gdpo-noaux.guidebench}{+4.8}
\setval{4b.papo.minus.gdpo-noaux.guidebench-abs}{4.8}
\setval{4b.papo.minus.gdpo-noaux.humaneval-plus}{$-$2.4}
\setval{4b.papo.minus.gdpo-noaux.humaneval-plus-abs}{2.4}
\setval{4b.papo.minus.gdpo-noaux.if}{$-$0.4}
\setval{4b.papo.minus.gdpo-noaux.if-abs}{0.4}
\setval{4b.papo.minus.gdpo-noaux.ifbench}{$-$7.0}
\setval{4b.papo.minus.gdpo-noaux.ifbench-abs}{7.0}
\setval{4b.papo.minus.gdpo-noaux.ifeval}{+1.1}
\setval{4b.papo.minus.gdpo-noaux.ifeval-abs}{1.1}
\setval{4b.papo.minus.gdpo-noaux.judge2.avg}{$-$2.9}
\setval{4b.papo.minus.gdpo-noaux.judge2.avg-abs}{2.9}
\setval{4b.papo.minus.gdpo-noaux.judge2.creative-writing-v3}{+10.9}
\setval{4b.papo.minus.gdpo-noaux.judge2.creative-writing-v3-abs}{10.9}
\setval{4b.papo.minus.gdpo-noaux.judge2.longbench-write}{+6.8}
\setval{4b.papo.minus.gdpo-noaux.judge2.longbench-write-abs}{6.8}
\setval{4b.papo.minus.gdpo-noaux.judge2.writing}{+7.9}
\setval{4b.papo.minus.gdpo-noaux.judge2.writing-abs}{7.9}
\setval{4b.papo.minus.gdpo-noaux.judge2.writingbench}{+6.1}
\setval{4b.papo.minus.gdpo-noaux.judge2.writingbench-abs}{6.1}
\setval{4b.papo.minus.gdpo-noaux.livecodebench}{$-$1.2}
\setval{4b.papo.minus.gdpo-noaux.livecodebench-abs}{1.2}
\setval{4b.papo.minus.gdpo-noaux.longbench-write}{$-$0.4}
\setval{4b.papo.minus.gdpo-noaux.longbench-write-abs}{0.4}
\setval{4b.papo.minus.gdpo-noaux.math}{$-$10.1}
\setval{4b.papo.minus.gdpo-noaux.math-abs}{10.1}
\setval{4b.papo.minus.gdpo-noaux.math500}{$-$8.2}
\setval{4b.papo.minus.gdpo-noaux.math500-abs}{8.2}
\setval{4b.papo.minus.gdpo-noaux.mbpp-plus}{$-$1.6}
\setval{4b.papo.minus.gdpo-noaux.mbpp-plus-abs}{1.6}
\setval{4b.papo.minus.gdpo-noaux.minerva}{$-$3.0}
\setval{4b.papo.minus.gdpo-noaux.minerva-abs}{3.0}
\setval{4b.papo.minus.gdpo-noaux.olympiad-bench}{$-$11.7}
\setval{4b.papo.minus.gdpo-noaux.olympiad-bench-abs}{11.7}
\setval{4b.papo.minus.gdpo-noaux.writing}{+2.0}
\setval{4b.papo.minus.gdpo-noaux.writing-abs}{2.0}
\setval{4b.papo.minus.gdpo-noaux.writingbench}{$-$0.8}
\setval{4b.papo.minus.gdpo-noaux.writingbench-abs}{0.8}
\setval{4b.papo.minus.gdpo-w0.5.aime24}{$-$5.3}
\setval{4b.papo.minus.gdpo-w0.5.aime24-abs}{5.3}
\setval{4b.papo.minus.gdpo-w0.5.aime25}{$-$2.8}
\setval{4b.papo.minus.gdpo-w0.5.aime25-abs}{2.8}
\setval{4b.papo.minus.gdpo-w0.5.amc23}{$-$4.8}
\setval{4b.papo.minus.gdpo-w0.5.amc23-abs}{4.8}
\setval{4b.papo.minus.gdpo-w0.5.avg}{$-$2.0}
\setval{4b.papo.minus.gdpo-w0.5.avg-abs}{2.0}
\setval{4b.papo.minus.gdpo-w0.5.code}{+0.0}
\setval{4b.papo.minus.gdpo-w0.5.code-abs}{0.0}
\setval{4b.papo.minus.gdpo-w0.5.creative-writing-v3}{+8.4}
\setval{4b.papo.minus.gdpo-w0.5.creative-writing-v3-abs}{8.4}
\setval{4b.papo.minus.gdpo-w0.5.guidebench}{+1.2}
\setval{4b.papo.minus.gdpo-w0.5.guidebench-abs}{1.2}
\setval{4b.papo.minus.gdpo-w0.5.humaneval-plus}{+1.0}
\setval{4b.papo.minus.gdpo-w0.5.humaneval-plus-abs}{1.0}
\setval{4b.papo.minus.gdpo-w0.5.if}{$-$1.9}
\setval{4b.papo.minus.gdpo-w0.5.if-abs}{1.9}
\setval{4b.papo.minus.gdpo-w0.5.ifbench}{$-$6.0}
\setval{4b.papo.minus.gdpo-w0.5.ifbench-abs}{6.0}
\setval{4b.papo.minus.gdpo-w0.5.ifeval}{$-$1.0}
\setval{4b.papo.minus.gdpo-w0.5.ifeval-abs}{1.0}
\setval{4b.papo.minus.gdpo-w0.5.judge2.avg}{$-$2.1}
\setval{4b.papo.minus.gdpo-w0.5.judge2.avg-abs}{2.1}
\setval{4b.papo.minus.gdpo-w0.5.judge2.creative-writing-v3}{+4.2}
\setval{4b.papo.minus.gdpo-w0.5.judge2.creative-writing-v3-abs}{4.2}
\setval{4b.papo.minus.gdpo-w0.5.judge2.longbench-write}{$-$3.0}
\setval{4b.papo.minus.gdpo-w0.5.judge2.longbench-write-abs}{3.0}
\setval{4b.papo.minus.gdpo-w0.5.judge2.writing}{$-$0.1}
\setval{4b.papo.minus.gdpo-w0.5.judge2.writing-abs}{0.1}
\setval{4b.papo.minus.gdpo-w0.5.judge2.writingbench}{$-$1.5}
\setval{4b.papo.minus.gdpo-w0.5.judge2.writingbench-abs}{1.5}
\setval{4b.papo.minus.gdpo-w0.5.livecodebench}{$-$0.6}
\setval{4b.papo.minus.gdpo-w0.5.livecodebench-abs}{0.6}
\setval{4b.papo.minus.gdpo-w0.5.longbench-write}{$-$3.4}
\setval{4b.papo.minus.gdpo-w0.5.longbench-write-abs}{3.4}
\setval{4b.papo.minus.gdpo-w0.5.math}{$-$4.1}
\setval{4b.papo.minus.gdpo-w0.5.math-abs}{4.1}
\setval{4b.papo.minus.gdpo-w0.5.math500}{$-$5.3}
\setval{4b.papo.minus.gdpo-w0.5.math500-abs}{5.3}
\setval{4b.papo.minus.gdpo-w0.5.mbpp-plus}{$-$0.5}
\setval{4b.papo.minus.gdpo-w0.5.mbpp-plus-abs}{0.5}
\setval{4b.papo.minus.gdpo-w0.5.minerva}{+0.4}
\setval{4b.papo.minus.gdpo-w0.5.minerva-abs}{0.4}
\setval{4b.papo.minus.gdpo-w0.5.olympiad-bench}{$-$7.0}
\setval{4b.papo.minus.gdpo-w0.5.olympiad-bench-abs}{7.0}
\setval{4b.papo.minus.gdpo-w0.5.writing}{+0.2}
\setval{4b.papo.minus.gdpo-w0.5.writing-abs}{0.2}
\setval{4b.papo.minus.gdpo-w0.5.writingbench}{$-$4.2}
\setval{4b.papo.minus.gdpo-w0.5.writingbench-abs}{4.2}
\setval{4b.papo.minus.gdpo.aime24}{+0.2}
\setval{4b.papo.minus.gdpo.aime24-abs}{0.2}
\setval{4b.papo.minus.gdpo.aime25}{$-$2.4}
\setval{4b.papo.minus.gdpo.aime25-abs}{2.4}
\setval{4b.papo.minus.gdpo.amc23}{$-$6.1}
\setval{4b.papo.minus.gdpo.amc23-abs}{6.1}
\setval{4b.papo.minus.gdpo.avg}{$-$1.7}
\setval{4b.papo.minus.gdpo.avg-abs}{1.7}
\setval{4b.papo.minus.gdpo.code}{+0.1}
\setval{4b.papo.minus.gdpo.code-abs}{0.1}
\setval{4b.papo.minus.gdpo.creative-writing-v3}{+0.6}
\setval{4b.papo.minus.gdpo.creative-writing-v3-abs}{0.6}
\setval{4b.papo.minus.gdpo.guidebench}{+0.1}
\setval{4b.papo.minus.gdpo.guidebench-abs}{0.1}
\setval{4b.papo.minus.gdpo.humaneval-plus}{+1.2}
\setval{4b.papo.minus.gdpo.humaneval-plus-abs}{1.2}
\setval{4b.papo.minus.gdpo.if}{$-$0.9}
\setval{4b.papo.minus.gdpo.if-abs}{0.9}
\setval{4b.papo.minus.gdpo.ifbench}{$-$3.0}
\setval{4b.papo.minus.gdpo.ifbench-abs}{3.0}
\setval{4b.papo.minus.gdpo.ifeval}{+0.1}
\setval{4b.papo.minus.gdpo.ifeval-abs}{0.1}
\setval{4b.papo.minus.gdpo.judge2.avg}{$-$1.5}
\setval{4b.papo.minus.gdpo.judge2.avg-abs}{1.5}
\setval{4b.papo.minus.gdpo.judge2.creative-writing-v3}{$-$0.2}
\setval{4b.papo.minus.gdpo.judge2.creative-writing-v3-abs}{0.2}
\setval{4b.papo.minus.gdpo.judge2.longbench-write}{$-$4.8}
\setval{4b.papo.minus.gdpo.judge2.longbench-write-abs}{4.8}
\setval{4b.papo.minus.gdpo.judge2.writing}{$-$2.3}
\setval{4b.papo.minus.gdpo.judge2.writing-abs}{2.3}
\setval{4b.papo.minus.gdpo.judge2.writingbench}{$-$1.9}
\setval{4b.papo.minus.gdpo.judge2.writingbench-abs}{1.9}
\setval{4b.papo.minus.gdpo.livecodebench}{$-$0.9}
\setval{4b.papo.minus.gdpo.livecodebench-abs}{0.9}
\setval{4b.papo.minus.gdpo.longbench-write}{$-$5.0}
\setval{4b.papo.minus.gdpo.longbench-write-abs}{5.0}
\setval{4b.papo.minus.gdpo.math}{$-$2.2}
\setval{4b.papo.minus.gdpo.math-abs}{2.2}
\setval{4b.papo.minus.gdpo.math500}{$-$2.8}
\setval{4b.papo.minus.gdpo.math500-abs}{2.8}
\setval{4b.papo.minus.gdpo.mbpp-plus}{$-$0.2}
\setval{4b.papo.minus.gdpo.mbpp-plus-abs}{0.2}
\setval{4b.papo.minus.gdpo.minerva}{+1.0}
\setval{4b.papo.minus.gdpo.minerva-abs}{1.0}
\setval{4b.papo.minus.gdpo.olympiad-bench}{$-$3.2}
\setval{4b.papo.minus.gdpo.olympiad-bench-abs}{3.2}
\setval{4b.papo.minus.gdpo.writing}{$-$3.3}
\setval{4b.papo.minus.gdpo.writing-abs}{3.3}
\setval{4b.papo.minus.gdpo.writingbench}{$-$5.4}
\setval{4b.papo.minus.gdpo.writingbench-abs}{5.4}
\setval{4b.papo.minus.geoma.aime24}{$-$2.3}
\setval{4b.papo.minus.geoma.aime24-abs}{2.3}
\setval{4b.papo.minus.geoma.aime25}{$-$3.1}
\setval{4b.papo.minus.geoma.aime25-abs}{3.1}
\setval{4b.papo.minus.geoma.amc23}{$-$1.9}
\setval{4b.papo.minus.geoma.amc23-abs}{1.9}
\setval{4b.papo.minus.geoma.avg}{$-$1.2}
\setval{4b.papo.minus.geoma.avg-abs}{1.2}
\setval{4b.papo.minus.geoma.code}{+0.4}
\setval{4b.papo.minus.geoma.code-abs}{0.4}
\setval{4b.papo.minus.geoma.creative-writing-v3}{+0.7}
\setval{4b.papo.minus.geoma.creative-writing-v3-abs}{0.7}
\setval{4b.papo.minus.geoma.guidebench}{+2.2}
\setval{4b.papo.minus.geoma.guidebench-abs}{2.2}
\setval{4b.papo.minus.geoma.humaneval-plus}{+2.8}
\setval{4b.papo.minus.geoma.humaneval-plus-abs}{2.8}
\setval{4b.papo.minus.geoma.if}{$-$0.8}
\setval{4b.papo.minus.geoma.if-abs}{0.8}
\setval{4b.papo.minus.geoma.ifbench}{$-$2.7}
\setval{4b.papo.minus.geoma.ifbench-abs}{2.7}
\setval{4b.papo.minus.geoma.ifeval}{$-$1.9}
\setval{4b.papo.minus.geoma.ifeval-abs}{1.9}
\setval{4b.papo.minus.geoma.judge2.avg}{$-$0.7}
\setval{4b.papo.minus.geoma.judge2.avg-abs}{0.7}
\setval{4b.papo.minus.geoma.judge2.creative-writing-v3}{+1.1}
\setval{4b.papo.minus.geoma.judge2.creative-writing-v3-abs}{1.1}
\setval{4b.papo.minus.geoma.judge2.longbench-write}{$-$0.1}
\setval{4b.papo.minus.geoma.judge2.longbench-write-abs}{0.1}
\setval{4b.papo.minus.geoma.judge2.writing}{+0.8}
\setval{4b.papo.minus.geoma.judge2.writing-abs}{0.8}
\setval{4b.papo.minus.geoma.judge2.writingbench}{+1.5}
\setval{4b.papo.minus.geoma.judge2.writingbench-abs}{1.5}
\setval{4b.papo.minus.geoma.livecodebench}{$-$2.1}
\setval{4b.papo.minus.geoma.livecodebench-abs}{2.1}
\setval{4b.papo.minus.geoma.longbench-write}{$-$1.9}
\setval{4b.papo.minus.geoma.longbench-write-abs}{1.9}
\setval{4b.papo.minus.geoma.math}{$-$2.1}
\setval{4b.papo.minus.geoma.math-abs}{2.1}
\setval{4b.papo.minus.geoma.math500}{$-$2.3}
\setval{4b.papo.minus.geoma.math500-abs}{2.3}
\setval{4b.papo.minus.geoma.mbpp-plus}{+0.5}
\setval{4b.papo.minus.geoma.mbpp-plus-abs}{0.5}
\setval{4b.papo.minus.geoma.minerva}{$-$0.4}
\setval{4b.papo.minus.geoma.minerva-abs}{0.4}
\setval{4b.papo.minus.geoma.olympiad-bench}{$-$2.4}
\setval{4b.papo.minus.geoma.olympiad-bench-abs}{2.4}
\setval{4b.papo.minus.geoma.writing}{$-$1.4}
\setval{4b.papo.minus.geoma.writing-abs}{1.4}
\setval{4b.papo.minus.geoma.writingbench}{$-$3.0}
\setval{4b.papo.minus.geoma.writingbench-abs}{3.0}
\setval{4b.papo.minus.grpo-nodiv.aime24}{$-$3.1}
\setval{4b.papo.minus.grpo-nodiv.aime24-abs}{3.1}
\setval{4b.papo.minus.grpo-nodiv.aime25}{$-$0.8}
\setval{4b.papo.minus.grpo-nodiv.aime25-abs}{0.8}
\setval{4b.papo.minus.grpo-nodiv.amc23}{$-$1.0}
\setval{4b.papo.minus.grpo-nodiv.amc23-abs}{1.0}
\setval{4b.papo.minus.grpo-nodiv.avg}{$-$0.4}
\setval{4b.papo.minus.grpo-nodiv.avg-abs}{0.4}
\setval{4b.papo.minus.grpo-nodiv.code}{+0.4}
\setval{4b.papo.minus.grpo-nodiv.code-abs}{0.4}
\setval{4b.papo.minus.grpo-nodiv.creative-writing-v3}{+4.0}
\setval{4b.papo.minus.grpo-nodiv.creative-writing-v3-abs}{4.0}
\setval{4b.papo.minus.grpo-nodiv.guidebench}{+1.9}
\setval{4b.papo.minus.grpo-nodiv.guidebench-abs}{1.9}
\setval{4b.papo.minus.grpo-nodiv.humaneval-plus}{+2.6}
\setval{4b.papo.minus.grpo-nodiv.humaneval-plus-abs}{2.6}
\setval{4b.papo.minus.grpo-nodiv.if}{+0.0}
\setval{4b.papo.minus.grpo-nodiv.if-abs}{0.0}
\setval{4b.papo.minus.grpo-nodiv.ifbench}{$-$0.3}
\setval{4b.papo.minus.grpo-nodiv.ifbench-abs}{0.3}
\setval{4b.papo.minus.grpo-nodiv.ifeval}{$-$1.4}
\setval{4b.papo.minus.grpo-nodiv.ifeval-abs}{1.4}
\setval{4b.papo.minus.grpo-nodiv.judge2.avg}{$-$0.3}
\setval{4b.papo.minus.grpo-nodiv.judge2.avg-abs}{0.3}
\setval{4b.papo.minus.grpo-nodiv.judge2.creative-writing-v3}{+3.7}
\setval{4b.papo.minus.grpo-nodiv.judge2.creative-writing-v3-abs}{3.7}
\setval{4b.papo.minus.grpo-nodiv.judge2.longbench-write}{$-$1.1}
\setval{4b.papo.minus.grpo-nodiv.judge2.longbench-write-abs}{1.1}
\setval{4b.papo.minus.grpo-nodiv.judge2.writing}{+1.7}
\setval{4b.papo.minus.grpo-nodiv.judge2.writing-abs}{1.7}
\setval{4b.papo.minus.grpo-nodiv.judge2.writingbench}{+2.6}
\setval{4b.papo.minus.grpo-nodiv.judge2.writingbench-abs}{2.6}
\setval{4b.papo.minus.grpo-nodiv.livecodebench}{$-$0.9}
\setval{4b.papo.minus.grpo-nodiv.livecodebench-abs}{0.9}
\setval{4b.papo.minus.grpo-nodiv.longbench-write}{$-$2.0}
\setval{4b.papo.minus.grpo-nodiv.longbench-write-abs}{2.0}
\setval{4b.papo.minus.grpo-nodiv.math}{$-$1.8}
\setval{4b.papo.minus.grpo-nodiv.math-abs}{1.8}
\setval{4b.papo.minus.grpo-nodiv.math500}{$-$3.0}
\setval{4b.papo.minus.grpo-nodiv.math500-abs}{3.0}
\setval{4b.papo.minus.grpo-nodiv.mbpp-plus}{$-$0.5}
\setval{4b.papo.minus.grpo-nodiv.mbpp-plus-abs}{0.5}
\setval{4b.papo.minus.grpo-nodiv.minerva}{+0.7}
\setval{4b.papo.minus.grpo-nodiv.minerva-abs}{0.7}
\setval{4b.papo.minus.grpo-nodiv.olympiad-bench}{$-$3.3}
\setval{4b.papo.minus.grpo-nodiv.olympiad-bench-abs}{3.3}
\setval{4b.papo.minus.grpo-nodiv.writing}{+0.9}
\setval{4b.papo.minus.grpo-nodiv.writing-abs}{0.9}
\setval{4b.papo.minus.grpo-nodiv.writingbench}{+0.9}
\setval{4b.papo.minus.grpo-nodiv.writingbench-abs}{0.9}
\setval{4b.papo.minus.grpo.aime24}{$-$2.7}
\setval{4b.papo.minus.grpo.aime24-abs}{2.7}
\setval{4b.papo.minus.grpo.aime25}{$-$5.1}
\setval{4b.papo.minus.grpo.aime25-abs}{5.1}
\setval{4b.papo.minus.grpo.amc23}{$-$1.2}
\setval{4b.papo.minus.grpo.amc23-abs}{1.2}
\setval{4b.papo.minus.grpo.avg}{$-$2.1}
\setval{4b.papo.minus.grpo.avg-abs}{2.1}
\setval{4b.papo.minus.grpo.code}{$-$0.2}
\setval{4b.papo.minus.grpo.code-abs}{0.2}
\setval{4b.papo.minus.grpo.creative-writing-v3}{$-$0.8}
\setval{4b.papo.minus.grpo.creative-writing-v3-abs}{0.8}
\setval{4b.papo.minus.grpo.guidebench}{+2.8}
\setval{4b.papo.minus.grpo.guidebench-abs}{2.8}
\setval{4b.papo.minus.grpo.humaneval-plus}{+0.3}
\setval{4b.papo.minus.grpo.humaneval-plus-abs}{0.3}
\setval{4b.papo.minus.grpo.if}{$-$0.7}
\setval{4b.papo.minus.grpo.if-abs}{0.7}
\setval{4b.papo.minus.grpo.ifbench}{$-$4.3}
\setval{4b.papo.minus.grpo.ifbench-abs}{4.3}
\setval{4b.papo.minus.grpo.ifeval}{$-$0.6}
\setval{4b.papo.minus.grpo.ifeval-abs}{0.6}
\setval{4b.papo.minus.grpo.judge2.avg}{$-$1.5}
\setval{4b.papo.minus.grpo.judge2.avg-abs}{1.5}
\setval{4b.papo.minus.grpo.judge2.creative-writing-v3}{$-$0.9}
\setval{4b.papo.minus.grpo.judge2.creative-writing-v3-abs}{0.9}
\setval{4b.papo.minus.grpo.judge2.longbench-write}{$-$3.2}
\setval{4b.papo.minus.grpo.judge2.longbench-write-abs}{3.2}
\setval{4b.papo.minus.grpo.judge2.writing}{$-$1.8}
\setval{4b.papo.minus.grpo.judge2.writing-abs}{1.8}
\setval{4b.papo.minus.grpo.judge2.writingbench}{$-$1.3}
\setval{4b.papo.minus.grpo.judge2.writingbench-abs}{1.3}
\setval{4b.papo.minus.grpo.livecodebench}{$-$0.5}
\setval{4b.papo.minus.grpo.livecodebench-abs}{0.5}
\setval{4b.papo.minus.grpo.longbench-write}{$-$5.2}
\setval{4b.papo.minus.grpo.longbench-write-abs}{5.2}
\setval{4b.papo.minus.grpo.math}{$-$2.4}
\setval{4b.papo.minus.grpo.math-abs}{2.4}
\setval{4b.papo.minus.grpo.math500}{$-$3.7}
\setval{4b.papo.minus.grpo.math500-abs}{3.7}
\setval{4b.papo.minus.grpo.mbpp-plus}{$-$0.4}
\setval{4b.papo.minus.grpo.mbpp-plus-abs}{0.4}
\setval{4b.papo.minus.grpo.minerva}{+1.3}
\setval{4b.papo.minus.grpo.minerva-abs}{1.3}
\setval{4b.papo.minus.grpo.olympiad-bench}{$-$3.1}
\setval{4b.papo.minus.grpo.olympiad-bench-abs}{3.1}
\setval{4b.papo.minus.grpo.writing}{$-$4.6}
\setval{4b.papo.minus.grpo.writing-abs}{4.6}
\setval{4b.papo.minus.grpo.writingbench}{$-$7.7}
\setval{4b.papo.minus.grpo.writingbench-abs}{7.7}
\setval{4b.papo.minus.mt-grpo.aime24}{$-$1.5}
\setval{4b.papo.minus.mt-grpo.aime24-abs}{1.5}
\setval{4b.papo.minus.mt-grpo.aime25}{+1.4}
\setval{4b.papo.minus.mt-grpo.aime25-abs}{1.4}
\setval{4b.papo.minus.mt-grpo.amc23}{$-$6.2}
\setval{4b.papo.minus.mt-grpo.amc23-abs}{6.2}
\setval{4b.papo.minus.mt-grpo.avg}{$-$0.5}
\setval{4b.papo.minus.mt-grpo.avg-abs}{0.5}
\setval{4b.papo.minus.mt-grpo.code}{+1.8}
\setval{4b.papo.minus.mt-grpo.code-abs}{1.8}
\setval{4b.papo.minus.mt-grpo.creative-writing-v3}{+2.0}
\setval{4b.papo.minus.mt-grpo.creative-writing-v3-abs}{2.0}
\setval{4b.papo.minus.mt-grpo.guidebench}{+1.1}
\setval{4b.papo.minus.mt-grpo.guidebench-abs}{1.1}
\setval{4b.papo.minus.mt-grpo.humaneval-plus}{+4.0}
\setval{4b.papo.minus.mt-grpo.humaneval-plus-abs}{4.0}
\setval{4b.papo.minus.mt-grpo.if}{$-$0.3}
\setval{4b.papo.minus.mt-grpo.if-abs}{0.3}
\setval{4b.papo.minus.mt-grpo.ifbench}{$-$1.0}
\setval{4b.papo.minus.mt-grpo.ifbench-abs}{1.0}
\setval{4b.papo.minus.mt-grpo.ifeval}{$-$0.8}
\setval{4b.papo.minus.mt-grpo.ifeval-abs}{0.8}
\setval{4b.papo.minus.mt-grpo.judge2.avg}{$-$0.4}
\setval{4b.papo.minus.mt-grpo.judge2.avg-abs}{0.4}
\setval{4b.papo.minus.mt-grpo.judge2.creative-writing-v3}{+0.2}
\setval{4b.papo.minus.mt-grpo.judge2.creative-writing-v3-abs}{0.2}
\setval{4b.papo.minus.mt-grpo.judge2.longbench-write}{$-$3.2}
\setval{4b.papo.minus.mt-grpo.judge2.longbench-write-abs}{3.2}
\setval{4b.papo.minus.mt-grpo.judge2.writing}{$-$1.4}
\setval{4b.papo.minus.mt-grpo.judge2.writing-abs}{1.4}
\setval{4b.papo.minus.mt-grpo.judge2.writingbench}{$-$1.1}
\setval{4b.papo.minus.mt-grpo.judge2.writingbench-abs}{1.1}
\setval{4b.papo.minus.mt-grpo.livecodebench}{+0.9}
\setval{4b.papo.minus.mt-grpo.livecodebench-abs}{0.9}
\setval{4b.papo.minus.mt-grpo.longbench-write}{$-$2.8}
\setval{4b.papo.minus.mt-grpo.longbench-write-abs}{2.8}
\setval{4b.papo.minus.mt-grpo.math}{$-$1.2}
\setval{4b.papo.minus.mt-grpo.math-abs}{1.2}
\setval{4b.papo.minus.mt-grpo.math500}{$-$0.5}
\setval{4b.papo.minus.mt-grpo.math500-abs}{0.5}
\setval{4b.papo.minus.mt-grpo.mbpp-plus}{+0.7}
\setval{4b.papo.minus.mt-grpo.mbpp-plus-abs}{0.7}
\setval{4b.papo.minus.mt-grpo.minerva}{+0.6}
\setval{4b.papo.minus.mt-grpo.minerva-abs}{0.6}
\setval{4b.papo.minus.mt-grpo.olympiad-bench}{$-$1.0}
\setval{4b.papo.minus.mt-grpo.olympiad-bench-abs}{1.0}
\setval{4b.papo.minus.mt-grpo.writing}{$-$1.5}
\setval{4b.papo.minus.mt-grpo.writing-abs}{1.5}
\setval{4b.papo.minus.mt-grpo.writingbench}{$-$3.6}
\setval{4b.papo.minus.mt-grpo.writingbench-abs}{3.6}
\setval{4b.papo.minus.saw.aime24}{$-$3.2}
\setval{4b.papo.minus.saw.aime24-abs}{3.2}
\setval{4b.papo.minus.saw.aime25}{$-$3.6}
\setval{4b.papo.minus.saw.aime25-abs}{3.6}
\setval{4b.papo.minus.saw.amc23}{$-$3.1}
\setval{4b.papo.minus.saw.amc23-abs}{3.1}
\setval{4b.papo.minus.saw.avg}{$-$1.6}
\setval{4b.papo.minus.saw.avg-abs}{1.6}
\setval{4b.papo.minus.saw.code}{+1.2}
\setval{4b.papo.minus.saw.code-abs}{1.2}
\setval{4b.papo.minus.saw.creative-writing-v3}{+1.5}
\setval{4b.papo.minus.saw.creative-writing-v3-abs}{1.5}
\setval{4b.papo.minus.saw.guidebench}{$-$0.3}
\setval{4b.papo.minus.saw.guidebench-abs}{0.3}
\setval{4b.papo.minus.saw.humaneval-plus}{+2.3}
\setval{4b.papo.minus.saw.humaneval-plus-abs}{2.3}
\setval{4b.papo.minus.saw.if}{$-$2.8}
\setval{4b.papo.minus.saw.if-abs}{2.8}
\setval{4b.papo.minus.saw.ifbench}{$-$7.0}
\setval{4b.papo.minus.saw.ifbench-abs}{7.0}
\setval{4b.papo.minus.saw.ifeval}{$-$1.1}
\setval{4b.papo.minus.saw.ifeval-abs}{1.1}
\setval{4b.papo.minus.saw.judge2.avg}{$-$1.7}
\setval{4b.papo.minus.saw.judge2.avg-abs}{1.7}
\setval{4b.papo.minus.saw.judge2.creative-writing-v3}{$-$1.4}
\setval{4b.papo.minus.saw.judge2.creative-writing-v3-abs}{1.4}
\setval{4b.papo.minus.saw.judge2.longbench-write}{$-$2.5}
\setval{4b.papo.minus.saw.judge2.longbench-write-abs}{2.5}
\setval{4b.papo.minus.saw.judge2.writing}{$-$1.7}
\setval{4b.papo.minus.saw.judge2.writing-abs}{1.7}
\setval{4b.papo.minus.saw.judge2.writingbench}{$-$1.2}
\setval{4b.papo.minus.saw.judge2.writingbench-abs}{1.2}
\setval{4b.papo.minus.saw.livecodebench}{+0.7}
\setval{4b.papo.minus.saw.livecodebench-abs}{0.7}
\setval{4b.papo.minus.saw.longbench-write}{$-$2.2}
\setval{4b.papo.minus.saw.longbench-write-abs}{2.2}
\setval{4b.papo.minus.saw.math}{$-$2.5}
\setval{4b.papo.minus.saw.math-abs}{2.5}
\setval{4b.papo.minus.saw.math500}{$-$1.3}
\setval{4b.papo.minus.saw.math500-abs}{1.3}
\setval{4b.papo.minus.saw.mbpp-plus}{+0.6}
\setval{4b.papo.minus.saw.mbpp-plus-abs}{0.6}
\setval{4b.papo.minus.saw.minerva}{$-$1.6}
\setval{4b.papo.minus.saw.minerva-abs}{1.6}
\setval{4b.papo.minus.saw.olympiad-bench}{$-$2.4}
\setval{4b.papo.minus.saw.olympiad-bench-abs}{2.4}
\setval{4b.papo.minus.saw.writing}{$-$1.4}
\setval{4b.papo.minus.saw.writing-abs}{1.4}
\setval{4b.papo.minus.saw.writingbench}{$-$3.4}
\setval{4b.papo.minus.saw.writingbench-abs}{3.4}
\setval{4b.papo.passing-at-ceiling.code.early}{0.71}
\setval{4b.papo.passing-at-ceiling.code.late}{0.74}
\setval{4b.papo.passing-at-ceiling.code.run}{0.73}
\setval{4b.papo.passing-at-ceiling.if.early}{0.29}
\setval{4b.papo.passing-at-ceiling.if.late}{0.57}
\setval{4b.papo.passing-at-ceiling.if.run}{0.48}
\setval{4b.papo.passing-at-ceiling.math.early}{1.00}
\setval{4b.papo.passing-at-ceiling.math.late}{1.00}
\setval{4b.papo.passing-at-ceiling.math.run}{1.00}
\setval{4b.papo.passing-at-ceiling.writing.early}{0.00}
\setval{4b.papo.passing-at-ceiling.writing.late}{0.00}
\setval{4b.papo.passing-at-ceiling.writing.run}{0.00}
\setval{4b.papo.primary-cosine.code.early}{0.93}
\setval{4b.papo.primary-cosine.code.late}{0.99}
\setval{4b.papo.primary-cosine.code.run}{0.99}
\setval{4b.papo.primary-cosine.if.early}{0.70}
\setval{4b.papo.primary-cosine.if.late}{0.72}
\setval{4b.papo.primary-cosine.if.run}{0.75}
\setval{4b.papo.primary-cosine.math.early}{0.86}
\setval{4b.papo.primary-cosine.math.late}{0.96}
\setval{4b.papo.primary-cosine.math.run}{0.95}
\setval{4b.papo.primary-cosine.writing.early}{0.75}
\setval{4b.papo.primary-cosine.writing.late}{0.77}
\setval{4b.papo.primary-cosine.writing.run}{0.76}
\setval{4b.papo.primary-level.code.early}{0.52}
\setval{4b.papo.primary-level.code.late}{0.74}
\setval{4b.papo.primary-level.code.run}{0.69}
\setval{4b.papo.primary-level.if.early}{0.33}
\setval{4b.papo.primary-level.if.late}{0.75}
\setval{4b.papo.primary-level.if.run}{0.63}
\setval{4b.papo.primary-level.math.early}{0.47}
\setval{4b.papo.primary-level.math.late}{0.65}
\setval{4b.papo.primary-level.math.run}{0.59}
\setval{4b.papo.primary-level.writing.early}{0.53}
\setval{4b.papo.primary-level.writing.late}{0.74}
\setval{4b.papo.primary-level.writing.run}{0.70}
\setval{4b.papo.solved.code.early}{0.03}
\setval{4b.papo.solved.code.late}{0.01}
\setval{4b.papo.solved.code.run}{0.01}
\setval{4b.papo.solved.if.early}{0.03}
\setval{4b.papo.solved.if.late}{0.19}
\setval{4b.papo.solved.if.run}{0.12}
\setval{4b.papo.solved.math.early}{0.11}
\setval{4b.papo.solved.math.late}{0.03}
\setval{4b.papo.solved.math.run}{0.05}
\setval{4b.papo.solved.writing.early}{0.00}
\setval{4b.papo.solved.writing.late}{0.00}
\setval{4b.papo.solved.writing.run}{0.00}
\setval{4b.papo.update-share.code.early}{0.17}
\setval{4b.papo.update-share.code.late}{0.13}
\setval{4b.papo.update-share.code.run}{0.14}
\setval{4b.papo.update-share.if.early}{0.28}
\setval{4b.papo.update-share.if.late}{0.20}
\setval{4b.papo.update-share.if.run}{0.23}
\setval{4b.papo.update-share.math.early}{0.17}
\setval{4b.papo.update-share.math.late}{0.14}
\setval{4b.papo.update-share.math.run}{0.14}
\setval{4b.papo.update-share.writing.early}{0.38}
\setval{4b.papo.update-share.writing.late}{0.53}
\setval{4b.papo.update-share.writing.run}{0.49}
\setval{4b.saw.aux-change.code}{+0.20}
\setval{4b.saw.aux-change.if}{+0.24}
\setval{4b.saw.aux-change.math}{+0.14}
\setval{4b.saw.aux-change.writing}{+0.21}
\setval{4b.saw.aux-change.writing-rm}{+0.37}
\setval{4b.saw.aux-full.code.early}{0.75}
\setval{4b.saw.aux-full.code.late}{0.95}
\setval{4b.saw.aux-full.code.run}{0.94}
\setval{4b.saw.aux-full.if.early}{0.67}
\setval{4b.saw.aux-full.if.late}{0.91}
\setval{4b.saw.aux-full.if.run}{0.87}
\setval{4b.saw.aux-full.math.early}{0.66}
\setval{4b.saw.aux-full.math.late}{0.91}
\setval{4b.saw.aux-full.math.run}{0.87}
\setval{4b.saw.aux-full.writing-rm.early}{0.10}
\setval{4b.saw.aux-full.writing-rm.late}{0.62}
\setval{4b.saw.aux-full.writing-rm.run}{0.49}
\setval{4b.saw.aux-full.writing.early}{0.64}
\setval{4b.saw.aux-full.writing.late}{0.77}
\setval{4b.saw.aux-full.writing.run}{0.76}
\setval{4b.saw.aux-level.code.early}{0.93}
\setval{4b.saw.aux-level.code.late}{0.99}
\setval{4b.saw.aux-level.code.run}{0.99}
\setval{4b.saw.aux-level.if.early}{0.84}
\setval{4b.saw.aux-level.if.late}{0.94}
\setval{4b.saw.aux-level.if.run}{0.93}
\setval{4b.saw.aux-level.math.early}{0.93}
\setval{4b.saw.aux-level.math.late}{0.99}
\setval{4b.saw.aux-level.math.run}{0.98}
\setval{4b.saw.aux-level.writing-rm.early}{0.78}
\setval{4b.saw.aux-level.writing-rm.late}{0.96}
\setval{4b.saw.aux-level.writing-rm.run}{0.94}
\setval{4b.saw.aux-level.writing.early}{0.91}
\setval{4b.saw.aux-level.writing.late}{0.97}
\setval{4b.saw.aux-level.writing.run}{0.97}
\setval{4b.saw.bench.aime24}{12.5}
\setval{4b.saw.bench.aime25}{11.1}
\setval{4b.saw.bench.amc23}{56.6}
\setval{4b.saw.bench.creative-writing-v3}{45.1}
\setval{4b.saw.bench.guidebench}{81.0}
\setval{4b.saw.bench.humaneval-plus}{80.7}
\setval{4b.saw.bench.ifbench}{40.3}
\setval{4b.saw.bench.ifeval}{86.9}
\setval{4b.saw.bench.livecodebench}{22.8}
\setval{4b.saw.bench.longbench-write}{66.2}
\setval{4b.saw.bench.math500}{77.8}
\setval{4b.saw.bench.mbpp-plus}{70.1}
\setval{4b.saw.bench.minerva}{48.1}
\setval{4b.saw.bench.olympiad-bench}{43.9}
\setval{4b.saw.bench.writingbench}{69.9}
\setval{4b.saw.eval.avg}{54.2}
\setval{4b.saw.eval.code}{57.9}
\setval{4b.saw.eval.if}{69.4}
\setval{4b.saw.eval.math}{41.7}
\setval{4b.saw.eval.writing}{60.4}
\setval{4b.saw.gate.code.against.early}{0.03}
\setval{4b.saw.gate.code.against.late}{0.02}
\setval{4b.saw.gate.code.against.run}{0.02}
\setval{4b.saw.gate.code.agreement}{+0.001}
\setval{4b.saw.gate.code.ceiling.early}{0.11}
\setval{4b.saw.gate.code.ceiling.late}{0.04}
\setval{4b.saw.gate.code.ceiling.run}{0.04}
\setval{4b.saw.gate.code.cosine.early}{0.85}
\setval{4b.saw.gate.code.cosine.late}{0.78}
\setval{4b.saw.gate.code.cosine.run}{0.82}
\setval{4b.saw.gate.if.against.early}{0.04}
\setval{4b.saw.gate.if.against.late}{0.03}
\setval{4b.saw.gate.if.against.run}{0.04}
\setval{4b.saw.gate.if.agreement}{+0.008}
\setval{4b.saw.gate.if.ceiling.early}{0.05}
\setval{4b.saw.gate.if.ceiling.late}{0.18}
\setval{4b.saw.gate.if.ceiling.run}{0.12}
\setval{4b.saw.gate.if.cosine.early}{0.76}
\setval{4b.saw.gate.if.cosine.late}{0.51}
\setval{4b.saw.gate.if.cosine.run}{0.60}
\setval{4b.saw.gate.math.against.early}{0.00}
\setval{4b.saw.gate.math.against.late}{0.00}
\setval{4b.saw.gate.math.against.run}{0.00}
\setval{4b.saw.gate.math.ceiling.early}{0.19}
\setval{4b.saw.gate.math.ceiling.late}{0.15}
\setval{4b.saw.gate.math.ceiling.run}{0.14}
\setval{4b.saw.gate.math.cosine.early}{0.89}
\setval{4b.saw.gate.math.cosine.late}{0.92}
\setval{4b.saw.gate.math.cosine.run}{0.92}
\setval{4b.saw.gate.writing-rm.against.early}{0.20}
\setval{4b.saw.gate.writing-rm.against.late}{0.52}
\setval{4b.saw.gate.writing-rm.against.run}{0.48}
\setval{4b.saw.gate.writing-rm.agreement}{+0.058}
\setval{4b.saw.gate.writing-rm.ceiling.early}{0.00}
\setval{4b.saw.gate.writing-rm.ceiling.late}{0.00}
\setval{4b.saw.gate.writing-rm.ceiling.run}{0.00}
\setval{4b.saw.gate.writing-rm.cosine.early}{0.69}
\setval{4b.saw.gate.writing-rm.cosine.late}{0.14}
\setval{4b.saw.gate.writing-rm.cosine.run}{0.23}
\setval{4b.saw.gate.writing.against.early}{0.28}
\setval{4b.saw.gate.writing.against.late}{0.58}
\setval{4b.saw.gate.writing.against.run}{0.52}
\setval{4b.saw.gate.writing.agreement}{$-$0.034}
\setval{4b.saw.gate.writing.ceiling.early}{0.00}
\setval{4b.saw.gate.writing.ceiling.late}{0.00}
\setval{4b.saw.gate.writing.ceiling.run}{0.00}
\setval{4b.saw.gate.writing.cosine.early}{0.53}
\setval{4b.saw.gate.writing.cosine.late}{0.06}
\setval{4b.saw.gate.writing.cosine.run}{0.13}
\setval{4b.saw.judge2.avg}{52.1}
\setval{4b.saw.judge2.creative-writing-v3}{35.6}
\setval{4b.saw.judge2.longbench-write}{60.0}
\setval{4b.saw.judge2.writing}{49.7}
\setval{4b.saw.judge2.writingbench}{53.4}
\setval{4b.saw.last-step}{300}
\setval{4b.saw.minus.base.aime24}{+4.7}
\setval{4b.saw.minus.base.aime24-abs}{4.7}
\setval{4b.saw.minus.base.aime25}{+4.1}
\setval{4b.saw.minus.base.aime25-abs}{4.1}
\setval{4b.saw.minus.base.amc23}{+12.7}
\setval{4b.saw.minus.base.amc23-abs}{12.7}
\setval{4b.saw.minus.base.avg}{+19.7}
\setval{4b.saw.minus.base.avg-abs}{19.7}
\setval{4b.saw.minus.base.code}{+13.3}
\setval{4b.saw.minus.base.code-abs}{13.3}
\setval{4b.saw.minus.base.creative-writing-v3}{+20.3}
\setval{4b.saw.minus.base.creative-writing-v3-abs}{20.3}
\setval{4b.saw.minus.base.guidebench}{+56.1}
\setval{4b.saw.minus.base.guidebench-abs}{56.1}
\setval{4b.saw.minus.base.humaneval-plus}{+15.7}
\setval{4b.saw.minus.base.humaneval-plus-abs}{15.7}
\setval{4b.saw.minus.base.if}{+39.7}
\setval{4b.saw.minus.base.if-abs}{39.7}
\setval{4b.saw.minus.base.ifbench}{+23.3}
\setval{4b.saw.minus.base.ifbench-abs}{23.3}
\setval{4b.saw.minus.base.ifeval}{+39.6}
\setval{4b.saw.minus.base.ifeval-abs}{39.6}
\setval{4b.saw.minus.base.judge2.avg}{+18.2}
\setval{4b.saw.minus.base.judge2.avg-abs}{18.2}
\setval{4b.saw.minus.base.judge2.creative-writing-v3}{+14.7}
\setval{4b.saw.minus.base.judge2.creative-writing-v3-abs}{14.7}
\setval{4b.saw.minus.base.judge2.longbench-write}{+19.1}
\setval{4b.saw.minus.base.judge2.longbench-write-abs}{19.1}
\setval{4b.saw.minus.base.judge2.writing}{+18.9}
\setval{4b.saw.minus.base.judge2.writing-abs}{18.9}
\setval{4b.saw.minus.base.judge2.writingbench}{+22.9}
\setval{4b.saw.minus.base.judge2.writingbench-abs}{22.9}
\setval{4b.saw.minus.base.livecodebench}{+9.6}
\setval{4b.saw.minus.base.livecodebench-abs}{9.6}
\setval{4b.saw.minus.base.longbench-write}{+21.8}
\setval{4b.saw.minus.base.longbench-write-abs}{21.8}
\setval{4b.saw.minus.base.math}{+9.5}
\setval{4b.saw.minus.base.math-abs}{9.5}
\setval{4b.saw.minus.base.math500}{+10.6}
\setval{4b.saw.minus.base.math500-abs}{10.6}
\setval{4b.saw.minus.base.mbpp-plus}{+14.5}
\setval{4b.saw.minus.base.mbpp-plus-abs}{14.5}
\setval{4b.saw.minus.base.minerva}{+16.5}
\setval{4b.saw.minus.base.minerva-abs}{16.5}
\setval{4b.saw.minus.base.olympiad-bench}{+8.2}
\setval{4b.saw.minus.base.olympiad-bench-abs}{8.2}
\setval{4b.saw.minus.base.writing}{+26.5}
\setval{4b.saw.minus.base.writing-abs}{26.5}
\setval{4b.saw.minus.base.writingbench}{+37.5}
\setval{4b.saw.minus.base.writingbench-abs}{37.5}
\setval{4b.saw.minus.concur-free.aime24}{+0.6}
\setval{4b.saw.minus.concur-free.aime24-abs}{0.6}
\setval{4b.saw.minus.concur-free.aime25}{$-$3.4}
\setval{4b.saw.minus.concur-free.aime25-abs}{3.4}
\setval{4b.saw.minus.concur-free.amc23}{+2.5}
\setval{4b.saw.minus.concur-free.amc23-abs}{2.5}
\setval{4b.saw.minus.concur-free.avg}{$-$1.3}
\setval{4b.saw.minus.concur-free.avg-abs}{1.3}
\setval{4b.saw.minus.concur-free.code}{$-$0.4}
\setval{4b.saw.minus.concur-free.code-abs}{0.4}
\setval{4b.saw.minus.concur-free.creative-writing-v3}{$-$0.4}
\setval{4b.saw.minus.concur-free.creative-writing-v3-abs}{0.4}
\setval{4b.saw.minus.concur-free.guidebench}{+2.0}
\setval{4b.saw.minus.concur-free.guidebench-abs}{2.0}
\setval{4b.saw.minus.concur-free.humaneval-plus}{$-$0.4}
\setval{4b.saw.minus.concur-free.humaneval-plus-abs}{0.4}
\setval{4b.saw.minus.concur-free.if}{+1.2}
\setval{4b.saw.minus.concur-free.if-abs}{1.2}
\setval{4b.saw.minus.concur-free.ifbench}{+1.3}
\setval{4b.saw.minus.concur-free.ifbench-abs}{1.3}
\setval{4b.saw.minus.concur-free.ifeval}{+0.2}
\setval{4b.saw.minus.concur-free.ifeval-abs}{0.2}
\setval{4b.saw.minus.concur-free.judge2.avg}{$-$0.7}
\setval{4b.saw.minus.concur-free.judge2.avg-abs}{0.7}
\setval{4b.saw.minus.concur-free.judge2.creative-writing-v3}{+0.3}
\setval{4b.saw.minus.concur-free.judge2.creative-writing-v3-abs}{0.3}
\setval{4b.saw.minus.concur-free.judge2.longbench-write}{$-$9.3}
\setval{4b.saw.minus.concur-free.judge2.longbench-write-abs}{9.3}
\setval{4b.saw.minus.concur-free.judge2.writing}{$-$3.4}
\setval{4b.saw.minus.concur-free.judge2.writing-abs}{3.4}
\setval{4b.saw.minus.concur-free.judge2.writingbench}{$-$1.1}
\setval{4b.saw.minus.concur-free.judge2.writingbench-abs}{1.1}
\setval{4b.saw.minus.concur-free.livecodebench}{$-$1.4}
\setval{4b.saw.minus.concur-free.livecodebench-abs}{1.4}
\setval{4b.saw.minus.concur-free.longbench-write}{$-$11.6}
\setval{4b.saw.minus.concur-free.longbench-write-abs}{11.6}
\setval{4b.saw.minus.concur-free.math}{$-$0.5}
\setval{4b.saw.minus.concur-free.math-abs}{0.5}
\setval{4b.saw.minus.concur-free.math500}{$-$2.5}
\setval{4b.saw.minus.concur-free.math500-abs}{2.5}
\setval{4b.saw.minus.concur-free.mbpp-plus}{+0.4}
\setval{4b.saw.minus.concur-free.mbpp-plus-abs}{0.4}
\setval{4b.saw.minus.concur-free.minerva}{+1.9}
\setval{4b.saw.minus.concur-free.minerva-abs}{1.9}
\setval{4b.saw.minus.concur-free.olympiad-bench}{$-$1.9}
\setval{4b.saw.minus.concur-free.olympiad-bench-abs}{1.9}
\setval{4b.saw.minus.concur-free.writing}{$-$6.5}
\setval{4b.saw.minus.concur-free.writing-abs}{6.5}
\setval{4b.saw.minus.concur-free.writingbench}{$-$7.4}
\setval{4b.saw.minus.concur-free.writingbench-abs}{7.4}
\setval{4b.saw.minus.concur-tied.aime24}{$-$7.0}
\setval{4b.saw.minus.concur-tied.aime24-abs}{7.0}
\setval{4b.saw.minus.concur-tied.aime25}{$-$5.2}
\setval{4b.saw.minus.concur-tied.aime25-abs}{5.2}
\setval{4b.saw.minus.concur-tied.amc23}{$-$12.6}
\setval{4b.saw.minus.concur-tied.amc23-abs}{12.6}
\setval{4b.saw.minus.concur-tied.avg}{$-$4.5}
\setval{4b.saw.minus.concur-tied.avg-abs}{4.5}
\setval{4b.saw.minus.concur-tied.code}{$-$1.2}
\setval{4b.saw.minus.concur-tied.code-abs}{1.2}
\setval{4b.saw.minus.concur-tied.creative-writing-v3}{$-$2.8}
\setval{4b.saw.minus.concur-tied.creative-writing-v3-abs}{2.8}
\setval{4b.saw.minus.concur-tied.guidebench}{+1.8}
\setval{4b.saw.minus.concur-tied.guidebench-abs}{1.8}
\setval{4b.saw.minus.concur-tied.humaneval-plus}{$-$3.3}
\setval{4b.saw.minus.concur-tied.humaneval-plus-abs}{3.3}
\setval{4b.saw.minus.concur-tied.if}{+1.2}
\setval{4b.saw.minus.concur-tied.if-abs}{1.2}
\setval{4b.saw.minus.concur-tied.ifbench}{+2.7}
\setval{4b.saw.minus.concur-tied.ifbench-abs}{2.7}
\setval{4b.saw.minus.concur-tied.ifeval}{$-$0.8}
\setval{4b.saw.minus.concur-tied.ifeval-abs}{0.8}
\setval{4b.saw.minus.concur-tied.judge2.avg}{$-$3.4}
\setval{4b.saw.minus.concur-tied.judge2.avg-abs}{3.4}
\setval{4b.saw.minus.concur-tied.judge2.creative-writing-v3}{+1.6}
\setval{4b.saw.minus.concur-tied.judge2.creative-writing-v3-abs}{1.6}
\setval{4b.saw.minus.concur-tied.judge2.longbench-write}{$-$7.9}
\setval{4b.saw.minus.concur-tied.judge2.longbench-write-abs}{7.9}
\setval{4b.saw.minus.concur-tied.judge2.writing}{$-$2.8}
\setval{4b.saw.minus.concur-tied.judge2.writing-abs}{2.8}
\setval{4b.saw.minus.concur-tied.judge2.writingbench}{$-$2.0}
\setval{4b.saw.minus.concur-tied.judge2.writingbench-abs}{2.0}
\setval{4b.saw.minus.concur-tied.livecodebench}{$-$1.9}
\setval{4b.saw.minus.concur-tied.livecodebench-abs}{1.9}
\setval{4b.saw.minus.concur-tied.longbench-write}{$-$11.5}
\setval{4b.saw.minus.concur-tied.longbench-write-abs}{11.5}
\setval{4b.saw.minus.concur-tied.math}{$-$7.1}
\setval{4b.saw.minus.concur-tied.math-abs}{7.1}
\setval{4b.saw.minus.concur-tied.math500}{$-$7.7}
\setval{4b.saw.minus.concur-tied.math500-abs}{7.7}
\setval{4b.saw.minus.concur-tied.mbpp-plus}{+1.5}
\setval{4b.saw.minus.concur-tied.mbpp-plus-abs}{1.5}
\setval{4b.saw.minus.concur-tied.minerva}{$-$1.0}
\setval{4b.saw.minus.concur-tied.minerva-abs}{1.0}
\setval{4b.saw.minus.concur-tied.olympiad-bench}{$-$9.0}
\setval{4b.saw.minus.concur-tied.olympiad-bench-abs}{9.0}
\setval{4b.saw.minus.concur-tied.writing}{$-$8.2}
\setval{4b.saw.minus.concur-tied.writing-abs}{8.2}
\setval{4b.saw.minus.concur-tied.writingbench}{$-$10.3}
\setval{4b.saw.minus.concur-tied.writingbench-abs}{10.3}
\setval{4b.saw.minus.concur.aime24}{$-$8.0}
\setval{4b.saw.minus.concur.aime24-abs}{8.0}
\setval{4b.saw.minus.concur.aime25}{$-$7.6}
\setval{4b.saw.minus.concur.aime25-abs}{7.6}
\setval{4b.saw.minus.concur.amc23}{$-$12.7}
\setval{4b.saw.minus.concur.amc23-abs}{12.7}
\setval{4b.saw.minus.concur.avg}{$-$4.6}
\setval{4b.saw.minus.concur.avg-abs}{4.6}
\setval{4b.saw.minus.concur.code}{$-$1.8}
\setval{4b.saw.minus.concur.code-abs}{1.8}
\setval{4b.saw.minus.concur.creative-writing-v3}{+0.4}
\setval{4b.saw.minus.concur.creative-writing-v3-abs}{0.4}
\setval{4b.saw.minus.concur.guidebench}{+3.6}
\setval{4b.saw.minus.concur.guidebench-abs}{3.6}
\setval{4b.saw.minus.concur.humaneval-plus}{$-$1.7}
\setval{4b.saw.minus.concur.humaneval-plus-abs}{1.7}
\setval{4b.saw.minus.concur.if}{+0.8}
\setval{4b.saw.minus.concur.if-abs}{0.8}
\setval{4b.saw.minus.concur.ifbench}{+0.0}
\setval{4b.saw.minus.concur.ifbench-abs}{0.0}
\setval{4b.saw.minus.concur.ifeval}{$-$1.2}
\setval{4b.saw.minus.concur.ifeval-abs}{1.2}
\setval{4b.saw.minus.concur.judge2.avg}{$-$3.7}
\setval{4b.saw.minus.concur.judge2.avg-abs}{3.7}
\setval{4b.saw.minus.concur.judge2.creative-writing-v3}{+3.2}
\setval{4b.saw.minus.concur.judge2.creative-writing-v3-abs}{3.2}
\setval{4b.saw.minus.concur.judge2.longbench-write}{$-$6.6}
\setval{4b.saw.minus.concur.judge2.longbench-write-abs}{6.6}
\setval{4b.saw.minus.concur.judge2.writing}{$-$2.0}
\setval{4b.saw.minus.concur.judge2.writing-abs}{2.0}
\setval{4b.saw.minus.concur.judge2.writingbench}{$-$2.7}
\setval{4b.saw.minus.concur.judge2.writingbench-abs}{2.7}
\setval{4b.saw.minus.concur.livecodebench}{$-$2.3}
\setval{4b.saw.minus.concur.livecodebench-abs}{2.3}
\setval{4b.saw.minus.concur.longbench-write}{$-$10.1}
\setval{4b.saw.minus.concur.longbench-write-abs}{10.1}
\setval{4b.saw.minus.concur.math}{$-$7.8}
\setval{4b.saw.minus.concur.math-abs}{7.8}
\setval{4b.saw.minus.concur.math500}{$-$7.4}
\setval{4b.saw.minus.concur.math500-abs}{7.4}
\setval{4b.saw.minus.concur.mbpp-plus}{$-$1.4}
\setval{4b.saw.minus.concur.mbpp-plus-abs}{1.4}
\setval{4b.saw.minus.concur.minerva}{$-$1.1}
\setval{4b.saw.minus.concur.minerva-abs}{1.1}
\setval{4b.saw.minus.concur.olympiad-bench}{$-$9.7}
\setval{4b.saw.minus.concur.olympiad-bench-abs}{9.7}
\setval{4b.saw.minus.concur.writing}{$-$6.4}
\setval{4b.saw.minus.concur.writing-abs}{6.4}
\setval{4b.saw.minus.concur.writingbench}{$-$9.6}
\setval{4b.saw.minus.concur.writingbench-abs}{9.6}
\setval{4b.saw.minus.gd2po.aime24}{+0.6}
\setval{4b.saw.minus.gd2po.aime24-abs}{0.6}
\setval{4b.saw.minus.gd2po.aime25}{+4.4}
\setval{4b.saw.minus.gd2po.aime25-abs}{4.4}
\setval{4b.saw.minus.gd2po.amc23}{$-$1.1}
\setval{4b.saw.minus.gd2po.amc23-abs}{1.1}
\setval{4b.saw.minus.gd2po.avg}{+0.1}
\setval{4b.saw.minus.gd2po.avg-abs}{0.1}
\setval{4b.saw.minus.gd2po.code}{$-$1.6}
\setval{4b.saw.minus.gd2po.code-abs}{1.6}
\setval{4b.saw.minus.gd2po.creative-writing-v3}{+0.4}
\setval{4b.saw.minus.gd2po.creative-writing-v3-abs}{0.4}
\setval{4b.saw.minus.gd2po.guidebench}{+1.9}
\setval{4b.saw.minus.gd2po.guidebench-abs}{1.9}
\setval{4b.saw.minus.gd2po.humaneval-plus}{$-$2.7}
\setval{4b.saw.minus.gd2po.humaneval-plus-abs}{2.7}
\setval{4b.saw.minus.gd2po.if}{+1.3}
\setval{4b.saw.minus.gd2po.if-abs}{1.3}
\setval{4b.saw.minus.gd2po.ifbench}{+2.7}
\setval{4b.saw.minus.gd2po.ifbench-abs}{2.7}
\setval{4b.saw.minus.gd2po.ifeval}{$-$0.7}
\setval{4b.saw.minus.gd2po.ifeval-abs}{0.7}
\setval{4b.saw.minus.gd2po.judge2.avg}{+0.5}
\setval{4b.saw.minus.gd2po.judge2.avg-abs}{0.5}
\setval{4b.saw.minus.gd2po.judge2.creative-writing-v3}{+2.7}
\setval{4b.saw.minus.gd2po.judge2.creative-writing-v3-abs}{2.7}
\setval{4b.saw.minus.gd2po.judge2.longbench-write}{$-$0.9}
\setval{4b.saw.minus.gd2po.judge2.longbench-write-abs}{0.9}
\setval{4b.saw.minus.gd2po.judge2.writing}{+0.9}
\setval{4b.saw.minus.gd2po.judge2.writing-abs}{0.9}
\setval{4b.saw.minus.gd2po.judge2.writingbench}{+0.8}
\setval{4b.saw.minus.gd2po.judge2.writingbench-abs}{0.8}
\setval{4b.saw.minus.gd2po.livecodebench}{$-$0.7}
\setval{4b.saw.minus.gd2po.livecodebench-abs}{0.7}
\setval{4b.saw.minus.gd2po.longbench-write}{$-$2.7}
\setval{4b.saw.minus.gd2po.longbench-write-abs}{2.7}
\setval{4b.saw.minus.gd2po.math}{+0.9}
\setval{4b.saw.minus.gd2po.math-abs}{0.9}
\setval{4b.saw.minus.gd2po.math500}{+0.2}
\setval{4b.saw.minus.gd2po.math500-abs}{0.2}
\setval{4b.saw.minus.gd2po.mbpp-plus}{$-$1.3}
\setval{4b.saw.minus.gd2po.mbpp-plus-abs}{1.3}
\setval{4b.saw.minus.gd2po.minerva}{+1.0}
\setval{4b.saw.minus.gd2po.minerva-abs}{1.0}
\setval{4b.saw.minus.gd2po.olympiad-bench}{+0.3}
\setval{4b.saw.minus.gd2po.olympiad-bench-abs}{0.3}
\setval{4b.saw.minus.gd2po.writing}{$-$0.9}
\setval{4b.saw.minus.gd2po.writing-abs}{0.9}
\setval{4b.saw.minus.gd2po.writingbench}{$-$0.3}
\setval{4b.saw.minus.gd2po.writingbench-abs}{0.3}
\setval{4b.saw.minus.gdpo-noaux.aime24}{$-$10.9}
\setval{4b.saw.minus.gdpo-noaux.aime24-abs}{10.9}
\setval{4b.saw.minus.gdpo-noaux.aime25}{$-$7.6}
\setval{4b.saw.minus.gdpo-noaux.aime25-abs}{7.6}
\setval{4b.saw.minus.gdpo-noaux.amc23}{$-$9.1}
\setval{4b.saw.minus.gdpo-noaux.amc23-abs}{9.1}
\setval{4b.saw.minus.gdpo-noaux.avg}{$-$2.5}
\setval{4b.saw.minus.gdpo-noaux.avg-abs}{2.5}
\setval{4b.saw.minus.gdpo-noaux.code}{$-$3.0}
\setval{4b.saw.minus.gdpo-noaux.code-abs}{3.0}
\setval{4b.saw.minus.gdpo-noaux.creative-writing-v3}{+5.7}
\setval{4b.saw.minus.gdpo-noaux.creative-writing-v3-abs}{5.7}
\setval{4b.saw.minus.gdpo-noaux.guidebench}{+5.1}
\setval{4b.saw.minus.gdpo-noaux.guidebench-abs}{5.1}
\setval{4b.saw.minus.gdpo-noaux.humaneval-plus}{$-$4.8}
\setval{4b.saw.minus.gdpo-noaux.humaneval-plus-abs}{4.8}
\setval{4b.saw.minus.gdpo-noaux.if}{+2.4}
\setval{4b.saw.minus.gdpo-noaux.if-abs}{2.4}
\setval{4b.saw.minus.gdpo-noaux.ifbench}{+0.0}
\setval{4b.saw.minus.gdpo-noaux.ifbench-abs}{0.0}
\setval{4b.saw.minus.gdpo-noaux.ifeval}{+2.2}
\setval{4b.saw.minus.gdpo-noaux.ifeval-abs}{2.2}
\setval{4b.saw.minus.gdpo-noaux.judge2.avg}{$-$1.2}
\setval{4b.saw.minus.gdpo-noaux.judge2.avg-abs}{1.2}
\setval{4b.saw.minus.gdpo-noaux.judge2.creative-writing-v3}{+12.2}
\setval{4b.saw.minus.gdpo-noaux.judge2.creative-writing-v3-abs}{12.2}
\setval{4b.saw.minus.gdpo-noaux.judge2.longbench-write}{+9.3}
\setval{4b.saw.minus.gdpo-noaux.judge2.longbench-write-abs}{9.3}
\setval{4b.saw.minus.gdpo-noaux.judge2.writing}{+9.6}
\setval{4b.saw.minus.gdpo-noaux.judge2.writing-abs}{9.6}
\setval{4b.saw.minus.gdpo-noaux.judge2.writingbench}{+7.3}
\setval{4b.saw.minus.gdpo-noaux.judge2.writingbench-abs}{7.3}
\setval{4b.saw.minus.gdpo-noaux.livecodebench}{$-$1.9}
\setval{4b.saw.minus.gdpo-noaux.livecodebench-abs}{1.9}
\setval{4b.saw.minus.gdpo-noaux.longbench-write}{+1.8}
\setval{4b.saw.minus.gdpo-noaux.longbench-write-abs}{1.8}
\setval{4b.saw.minus.gdpo-noaux.math}{$-$7.5}
\setval{4b.saw.minus.gdpo-noaux.math-abs}{7.5}
\setval{4b.saw.minus.gdpo-noaux.math500}{$-$6.9}
\setval{4b.saw.minus.gdpo-noaux.math500-abs}{6.9}
\setval{4b.saw.minus.gdpo-noaux.mbpp-plus}{$-$2.2}
\setval{4b.saw.minus.gdpo-noaux.mbpp-plus-abs}{2.2}
\setval{4b.saw.minus.gdpo-noaux.minerva}{$-$1.4}
\setval{4b.saw.minus.gdpo-noaux.minerva-abs}{1.4}
\setval{4b.saw.minus.gdpo-noaux.olympiad-bench}{$-$9.3}
\setval{4b.saw.minus.gdpo-noaux.olympiad-bench-abs}{9.3}
\setval{4b.saw.minus.gdpo-noaux.writing}{+3.4}
\setval{4b.saw.minus.gdpo-noaux.writing-abs}{3.4}
\setval{4b.saw.minus.gdpo-noaux.writingbench}{+2.5}
\setval{4b.saw.minus.gdpo-noaux.writingbench-abs}{2.5}
\setval{4b.saw.minus.gdpo-w0.5.aime24}{$-$2.1}
\setval{4b.saw.minus.gdpo-w0.5.aime24-abs}{2.1}
\setval{4b.saw.minus.gdpo-w0.5.aime25}{+0.8}
\setval{4b.saw.minus.gdpo-w0.5.aime25-abs}{0.8}
\setval{4b.saw.minus.gdpo-w0.5.amc23}{$-$1.7}
\setval{4b.saw.minus.gdpo-w0.5.amc23-abs}{1.7}
\setval{4b.saw.minus.gdpo-w0.5.avg}{$-$0.4}
\setval{4b.saw.minus.gdpo-w0.5.avg-abs}{0.4}
\setval{4b.saw.minus.gdpo-w0.5.code}{$-$1.2}
\setval{4b.saw.minus.gdpo-w0.5.code-abs}{1.2}
\setval{4b.saw.minus.gdpo-w0.5.creative-writing-v3}{+7.0}
\setval{4b.saw.minus.gdpo-w0.5.creative-writing-v3-abs}{7.0}
\setval{4b.saw.minus.gdpo-w0.5.guidebench}{+1.4}
\setval{4b.saw.minus.gdpo-w0.5.guidebench-abs}{1.4}
\setval{4b.saw.minus.gdpo-w0.5.humaneval-plus}{$-$1.3}
\setval{4b.saw.minus.gdpo-w0.5.humaneval-plus-abs}{1.3}
\setval{4b.saw.minus.gdpo-w0.5.if}{+0.9}
\setval{4b.saw.minus.gdpo-w0.5.if-abs}{0.9}
\setval{4b.saw.minus.gdpo-w0.5.ifbench}{+1.0}
\setval{4b.saw.minus.gdpo-w0.5.ifbench-abs}{1.0}
\setval{4b.saw.minus.gdpo-w0.5.ifeval}{+0.1}
\setval{4b.saw.minus.gdpo-w0.5.ifeval-abs}{0.1}
\setval{4b.saw.minus.gdpo-w0.5.judge2.avg}{$-$0.4}
\setval{4b.saw.minus.gdpo-w0.5.judge2.avg-abs}{0.4}
\setval{4b.saw.minus.gdpo-w0.5.judge2.creative-writing-v3}{+5.5}
\setval{4b.saw.minus.gdpo-w0.5.judge2.creative-writing-v3-abs}{5.5}
\setval{4b.saw.minus.gdpo-w0.5.judge2.longbench-write}{$-$0.5}
\setval{4b.saw.minus.gdpo-w0.5.judge2.longbench-write-abs}{0.5}
\setval{4b.saw.minus.gdpo-w0.5.judge2.writing}{+1.6}
\setval{4b.saw.minus.gdpo-w0.5.judge2.writing-abs}{1.6}
\setval{4b.saw.minus.gdpo-w0.5.judge2.writingbench}{$-$0.2}
\setval{4b.saw.minus.gdpo-w0.5.judge2.writingbench-abs}{0.2}
\setval{4b.saw.minus.gdpo-w0.5.livecodebench}{$-$1.4}
\setval{4b.saw.minus.gdpo-w0.5.livecodebench-abs}{1.4}
\setval{4b.saw.minus.gdpo-w0.5.longbench-write}{$-$1.2}
\setval{4b.saw.minus.gdpo-w0.5.longbench-write-abs}{1.2}
\setval{4b.saw.minus.gdpo-w0.5.math}{$-$1.6}
\setval{4b.saw.minus.gdpo-w0.5.math-abs}{1.6}
\setval{4b.saw.minus.gdpo-w0.5.math500}{$-$4.0}
\setval{4b.saw.minus.gdpo-w0.5.math500-abs}{4.0}
\setval{4b.saw.minus.gdpo-w0.5.mbpp-plus}{$-$1.1}
\setval{4b.saw.minus.gdpo-w0.5.mbpp-plus-abs}{1.1}
\setval{4b.saw.minus.gdpo-w0.5.minerva}{+2.0}
\setval{4b.saw.minus.gdpo-w0.5.minerva-abs}{2.0}
\setval{4b.saw.minus.gdpo-w0.5.olympiad-bench}{$-$4.6}
\setval{4b.saw.minus.gdpo-w0.5.olympiad-bench-abs}{4.6}
\setval{4b.saw.minus.gdpo-w0.5.writing}{+1.6}
\setval{4b.saw.minus.gdpo-w0.5.writing-abs}{1.6}
\setval{4b.saw.minus.gdpo-w0.5.writingbench}{$-$0.9}
\setval{4b.saw.minus.gdpo-w0.5.writingbench-abs}{0.9}
\setval{4b.saw.minus.gdpo.aime24}{+3.4}
\setval{4b.saw.minus.gdpo.aime24-abs}{3.4}
\setval{4b.saw.minus.gdpo.aime25}{+1.3}
\setval{4b.saw.minus.gdpo.aime25-abs}{1.3}
\setval{4b.saw.minus.gdpo.amc23}{$-$3.0}
\setval{4b.saw.minus.gdpo.amc23-abs}{3.0}
\setval{4b.saw.minus.gdpo.avg}{$-$0.1}
\setval{4b.saw.minus.gdpo.avg-abs}{0.1}
\setval{4b.saw.minus.gdpo.code}{$-$1.2}
\setval{4b.saw.minus.gdpo.code-abs}{1.2}
\setval{4b.saw.minus.gdpo.creative-writing-v3}{$-$0.9}
\setval{4b.saw.minus.gdpo.creative-writing-v3-abs}{0.9}
\setval{4b.saw.minus.gdpo.guidebench}{+0.4}
\setval{4b.saw.minus.gdpo.guidebench-abs}{0.4}
\setval{4b.saw.minus.gdpo.humaneval-plus}{$-$1.1}
\setval{4b.saw.minus.gdpo.humaneval-plus-abs}{1.1}
\setval{4b.saw.minus.gdpo.if}{+1.9}
\setval{4b.saw.minus.gdpo.if-abs}{1.9}
\setval{4b.saw.minus.gdpo.ifbench}{+4.0}
\setval{4b.saw.minus.gdpo.ifbench-abs}{4.0}
\setval{4b.saw.minus.gdpo.ifeval}{+1.2}
\setval{4b.saw.minus.gdpo.ifeval-abs}{1.2}
\setval{4b.saw.minus.gdpo.judge2.avg}{+0.2}
\setval{4b.saw.minus.gdpo.judge2.avg-abs}{0.2}
\setval{4b.saw.minus.gdpo.judge2.creative-writing-v3}{+1.2}
\setval{4b.saw.minus.gdpo.judge2.creative-writing-v3-abs}{1.2}
\setval{4b.saw.minus.gdpo.judge2.longbench-write}{$-$2.3}
\setval{4b.saw.minus.gdpo.judge2.longbench-write-abs}{2.3}
\setval{4b.saw.minus.gdpo.judge2.writing}{$-$0.6}
\setval{4b.saw.minus.gdpo.judge2.writing-abs}{0.6}
\setval{4b.saw.minus.gdpo.judge2.writingbench}{$-$0.6}
\setval{4b.saw.minus.gdpo.judge2.writingbench-abs}{0.6}
\setval{4b.saw.minus.gdpo.livecodebench}{$-$1.6}
\setval{4b.saw.minus.gdpo.livecodebench-abs}{1.6}
\setval{4b.saw.minus.gdpo.longbench-write}{$-$2.8}
\setval{4b.saw.minus.gdpo.longbench-write-abs}{2.8}
\setval{4b.saw.minus.gdpo.math}{+0.4}
\setval{4b.saw.minus.gdpo.math-abs}{0.4}
\setval{4b.saw.minus.gdpo.math500}{$-$1.4}
\setval{4b.saw.minus.gdpo.math500-abs}{1.4}
\setval{4b.saw.minus.gdpo.mbpp-plus}{$-$0.8}
\setval{4b.saw.minus.gdpo.mbpp-plus-abs}{0.8}
\setval{4b.saw.minus.gdpo.minerva}{+2.7}
\setval{4b.saw.minus.gdpo.minerva-abs}{2.7}
\setval{4b.saw.minus.gdpo.olympiad-bench}{$-$0.8}
\setval{4b.saw.minus.gdpo.olympiad-bench-abs}{0.8}
\setval{4b.saw.minus.gdpo.writing}{$-$1.9}
\setval{4b.saw.minus.gdpo.writing-abs}{1.9}
\setval{4b.saw.minus.gdpo.writingbench}{$-$2.1}
\setval{4b.saw.minus.gdpo.writingbench-abs}{2.1}
\setval{4b.saw.minus.geoma.aime24}{+0.9}
\setval{4b.saw.minus.geoma.aime24-abs}{0.9}
\setval{4b.saw.minus.geoma.aime25}{+0.5}
\setval{4b.saw.minus.geoma.aime25-abs}{0.5}
\setval{4b.saw.minus.geoma.amc23}{+1.2}
\setval{4b.saw.minus.geoma.amc23-abs}{1.2}
\setval{4b.saw.minus.geoma.avg}{+0.4}
\setval{4b.saw.minus.geoma.avg-abs}{0.4}
\setval{4b.saw.minus.geoma.code}{$-$0.8}
\setval{4b.saw.minus.geoma.code-abs}{0.8}
\setval{4b.saw.minus.geoma.creative-writing-v3}{$-$0.8}
\setval{4b.saw.minus.geoma.creative-writing-v3-abs}{0.8}
\setval{4b.saw.minus.geoma.guidebench}{+2.5}
\setval{4b.saw.minus.geoma.guidebench-abs}{2.5}
\setval{4b.saw.minus.geoma.humaneval-plus}{+0.5}
\setval{4b.saw.minus.geoma.humaneval-plus-abs}{0.5}
\setval{4b.saw.minus.geoma.if}{+2.0}
\setval{4b.saw.minus.geoma.if-abs}{2.0}
\setval{4b.saw.minus.geoma.ifbench}{+4.3}
\setval{4b.saw.minus.geoma.ifbench-abs}{4.3}
\setval{4b.saw.minus.geoma.ifeval}{$-$0.8}
\setval{4b.saw.minus.geoma.ifeval-abs}{0.8}
\setval{4b.saw.minus.geoma.judge2.avg}{+0.9}
\setval{4b.saw.minus.geoma.judge2.avg-abs}{0.9}
\setval{4b.saw.minus.geoma.judge2.creative-writing-v3}{+2.5}
\setval{4b.saw.minus.geoma.judge2.creative-writing-v3-abs}{2.5}
\setval{4b.saw.minus.geoma.judge2.longbench-write}{+2.4}
\setval{4b.saw.minus.geoma.judge2.longbench-write-abs}{2.4}
\setval{4b.saw.minus.geoma.judge2.writing}{+2.5}
\setval{4b.saw.minus.geoma.judge2.writing-abs}{2.5}
\setval{4b.saw.minus.geoma.judge2.writingbench}{+2.7}
\setval{4b.saw.minus.geoma.judge2.writingbench-abs}{2.7}
\setval{4b.saw.minus.geoma.livecodebench}{$-$2.9}
\setval{4b.saw.minus.geoma.livecodebench-abs}{2.9}
\setval{4b.saw.minus.geoma.longbench-write}{+0.4}
\setval{4b.saw.minus.geoma.longbench-write-abs}{0.4}
\setval{4b.saw.minus.geoma.math}{+0.5}
\setval{4b.saw.minus.geoma.math-abs}{0.5}
\setval{4b.saw.minus.geoma.math500}{$-$1.0}
\setval{4b.saw.minus.geoma.math500-abs}{1.0}
\setval{4b.saw.minus.geoma.mbpp-plus}{$-$0.1}
\setval{4b.saw.minus.geoma.mbpp-plus-abs}{0.1}
\setval{4b.saw.minus.geoma.minerva}{+1.2}
\setval{4b.saw.minus.geoma.minerva-abs}{1.2}
\setval{4b.saw.minus.geoma.olympiad-bench}{+0.0}
\setval{4b.saw.minus.geoma.olympiad-bench-abs}{0.0}
\setval{4b.saw.minus.geoma.writing}{+0.0}
\setval{4b.saw.minus.geoma.writing-abs}{0.0}
\setval{4b.saw.minus.geoma.writingbench}{+0.3}
\setval{4b.saw.minus.geoma.writingbench-abs}{0.3}
\setval{4b.saw.minus.grpo-nodiv.aime24}{+0.1}
\setval{4b.saw.minus.grpo-nodiv.aime24-abs}{0.1}
\setval{4b.saw.minus.grpo-nodiv.aime25}{+2.8}
\setval{4b.saw.minus.grpo-nodiv.aime25-abs}{2.8}
\setval{4b.saw.minus.grpo-nodiv.amc23}{+2.1}
\setval{4b.saw.minus.grpo-nodiv.amc23-abs}{2.1}
\setval{4b.saw.minus.grpo-nodiv.avg}{+1.2}
\setval{4b.saw.minus.grpo-nodiv.avg-abs}{1.2}
\setval{4b.saw.minus.grpo-nodiv.code}{$-$0.8}
\setval{4b.saw.minus.grpo-nodiv.code-abs}{0.8}
\setval{4b.saw.minus.grpo-nodiv.creative-writing-v3}{+2.5}
\setval{4b.saw.minus.grpo-nodiv.creative-writing-v3-abs}{2.5}
\setval{4b.saw.minus.grpo-nodiv.guidebench}{+2.2}
\setval{4b.saw.minus.grpo-nodiv.guidebench-abs}{2.2}
\setval{4b.saw.minus.grpo-nodiv.humaneval-plus}{+0.2}
\setval{4b.saw.minus.grpo-nodiv.humaneval-plus-abs}{0.2}
\setval{4b.saw.minus.grpo-nodiv.if}{+2.8}
\setval{4b.saw.minus.grpo-nodiv.if-abs}{2.8}
\setval{4b.saw.minus.grpo-nodiv.ifbench}{+6.7}
\setval{4b.saw.minus.grpo-nodiv.ifbench-abs}{6.7}
\setval{4b.saw.minus.grpo-nodiv.ifeval}{$-$0.4}
\setval{4b.saw.minus.grpo-nodiv.ifeval-abs}{0.4}
\setval{4b.saw.minus.grpo-nodiv.judge2.avg}{+1.4}
\setval{4b.saw.minus.grpo-nodiv.judge2.avg-abs}{1.4}
\setval{4b.saw.minus.grpo-nodiv.judge2.creative-writing-v3}{+5.1}
\setval{4b.saw.minus.grpo-nodiv.judge2.creative-writing-v3-abs}{5.1}
\setval{4b.saw.minus.grpo-nodiv.judge2.longbench-write}{+1.4}
\setval{4b.saw.minus.grpo-nodiv.judge2.longbench-write-abs}{1.4}
\setval{4b.saw.minus.grpo-nodiv.judge2.writing}{+3.4}
\setval{4b.saw.minus.grpo-nodiv.judge2.writing-abs}{3.4}
\setval{4b.saw.minus.grpo-nodiv.judge2.writingbench}{+3.8}
\setval{4b.saw.minus.grpo-nodiv.judge2.writingbench-abs}{3.8}
\setval{4b.saw.minus.grpo-nodiv.livecodebench}{$-$1.7}
\setval{4b.saw.minus.grpo-nodiv.livecodebench-abs}{1.7}
\setval{4b.saw.minus.grpo-nodiv.longbench-write}{+0.2}
\setval{4b.saw.minus.grpo-nodiv.longbench-write-abs}{0.2}
\setval{4b.saw.minus.grpo-nodiv.math}{+0.8}
\setval{4b.saw.minus.grpo-nodiv.math-abs}{0.8}
\setval{4b.saw.minus.grpo-nodiv.math500}{$-$1.7}
\setval{4b.saw.minus.grpo-nodiv.math500-abs}{1.7}
\setval{4b.saw.minus.grpo-nodiv.mbpp-plus}{$-$1.1}
\setval{4b.saw.minus.grpo-nodiv.mbpp-plus-abs}{1.1}
\setval{4b.saw.minus.grpo-nodiv.minerva}{+2.3}
\setval{4b.saw.minus.grpo-nodiv.minerva-abs}{2.3}
\setval{4b.saw.minus.grpo-nodiv.olympiad-bench}{$-$1.0}
\setval{4b.saw.minus.grpo-nodiv.olympiad-bench-abs}{1.0}
\setval{4b.saw.minus.grpo-nodiv.writing}{+2.3}
\setval{4b.saw.minus.grpo-nodiv.writing-abs}{2.3}
\setval{4b.saw.minus.grpo-nodiv.writingbench}{+4.2}
\setval{4b.saw.minus.grpo-nodiv.writingbench-abs}{4.2}
\setval{4b.saw.minus.grpo.aime24}{+0.5}
\setval{4b.saw.minus.grpo.aime24-abs}{0.5}
\setval{4b.saw.minus.grpo.aime25}{$-$1.5}
\setval{4b.saw.minus.grpo.aime25-abs}{1.5}
\setval{4b.saw.minus.grpo.amc23}{+2.0}
\setval{4b.saw.minus.grpo.amc23-abs}{2.0}
\setval{4b.saw.minus.grpo.avg}{$-$0.5}
\setval{4b.saw.minus.grpo.avg-abs}{0.5}
\setval{4b.saw.minus.grpo.code}{$-$1.4}
\setval{4b.saw.minus.grpo.code-abs}{1.4}
\setval{4b.saw.minus.grpo.creative-writing-v3}{$-$2.3}
\setval{4b.saw.minus.grpo.creative-writing-v3-abs}{2.3}
\setval{4b.saw.minus.grpo.guidebench}{+3.1}
\setval{4b.saw.minus.grpo.guidebench-abs}{3.1}
\setval{4b.saw.minus.grpo.humaneval-plus}{$-$2.0}
\setval{4b.saw.minus.grpo.humaneval-plus-abs}{2.0}
\setval{4b.saw.minus.grpo.if}{+2.1}
\setval{4b.saw.minus.grpo.if-abs}{2.1}
\setval{4b.saw.minus.grpo.ifbench}{+2.7}
\setval{4b.saw.minus.grpo.ifbench-abs}{2.7}
\setval{4b.saw.minus.grpo.ifeval}{+0.5}
\setval{4b.saw.minus.grpo.ifeval-abs}{0.5}
\setval{4b.saw.minus.grpo.judge2.avg}{+0.2}
\setval{4b.saw.minus.grpo.judge2.avg-abs}{0.2}
\setval{4b.saw.minus.grpo.judge2.creative-writing-v3}{+0.5}
\setval{4b.saw.minus.grpo.judge2.creative-writing-v3-abs}{0.5}
\setval{4b.saw.minus.grpo.judge2.longbench-write}{$-$0.7}
\setval{4b.saw.minus.grpo.judge2.longbench-write-abs}{0.7}
\setval{4b.saw.minus.grpo.judge2.writing}{$-$0.1}
\setval{4b.saw.minus.grpo.judge2.writing-abs}{0.1}
\setval{4b.saw.minus.grpo.judge2.writingbench}{$-$0.1}
\setval{4b.saw.minus.grpo.judge2.writingbench-abs}{0.1}
\setval{4b.saw.minus.grpo.livecodebench}{$-$1.2}
\setval{4b.saw.minus.grpo.livecodebench-abs}{1.2}
\setval{4b.saw.minus.grpo.longbench-write}{$-$3.0}
\setval{4b.saw.minus.grpo.longbench-write-abs}{3.0}
\setval{4b.saw.minus.grpo.math}{+0.1}
\setval{4b.saw.minus.grpo.math-abs}{0.1}
\setval{4b.saw.minus.grpo.math500}{$-$2.4}
\setval{4b.saw.minus.grpo.math500-abs}{2.4}
\setval{4b.saw.minus.grpo.mbpp-plus}{$-$1.1}
\setval{4b.saw.minus.grpo.mbpp-plus-abs}{1.1}
\setval{4b.saw.minus.grpo.minerva}{+2.9}
\setval{4b.saw.minus.grpo.minerva-abs}{2.9}
\setval{4b.saw.minus.grpo.olympiad-bench}{$-$0.8}
\setval{4b.saw.minus.grpo.olympiad-bench-abs}{0.8}
\setval{4b.saw.minus.grpo.writing}{$-$3.2}
\setval{4b.saw.minus.grpo.writing-abs}{3.2}
\setval{4b.saw.minus.grpo.writingbench}{$-$4.4}
\setval{4b.saw.minus.grpo.writingbench-abs}{4.4}
\setval{4b.saw.minus.mt-grpo.aime24}{+1.8}
\setval{4b.saw.minus.mt-grpo.aime24-abs}{1.8}
\setval{4b.saw.minus.mt-grpo.aime25}{+5.0}
\setval{4b.saw.minus.mt-grpo.aime25-abs}{5.0}
\setval{4b.saw.minus.mt-grpo.amc23}{$-$3.1}
\setval{4b.saw.minus.mt-grpo.amc23-abs}{3.1}
\setval{4b.saw.minus.mt-grpo.avg}{+1.1}
\setval{4b.saw.minus.mt-grpo.avg-abs}{1.1}
\setval{4b.saw.minus.mt-grpo.code}{+0.6}
\setval{4b.saw.minus.mt-grpo.code-abs}{0.6}
\setval{4b.saw.minus.mt-grpo.creative-writing-v3}{+0.6}
\setval{4b.saw.minus.mt-grpo.creative-writing-v3-abs}{0.6}
\setval{4b.saw.minus.mt-grpo.guidebench}{+1.3}
\setval{4b.saw.minus.mt-grpo.guidebench-abs}{1.3}
\setval{4b.saw.minus.mt-grpo.humaneval-plus}{+1.6}
\setval{4b.saw.minus.mt-grpo.humaneval-plus-abs}{1.6}
\setval{4b.saw.minus.mt-grpo.if}{+2.5}
\setval{4b.saw.minus.mt-grpo.if-abs}{2.5}
\setval{4b.saw.minus.mt-grpo.ifbench}{+6.0}
\setval{4b.saw.minus.mt-grpo.ifbench-abs}{6.0}
\setval{4b.saw.minus.mt-grpo.ifeval}{+0.2}
\setval{4b.saw.minus.mt-grpo.ifeval-abs}{0.2}
\setval{4b.saw.minus.mt-grpo.judge2.avg}{+1.2}
\setval{4b.saw.minus.mt-grpo.judge2.avg-abs}{1.2}
\setval{4b.saw.minus.mt-grpo.judge2.creative-writing-v3}{+1.5}
\setval{4b.saw.minus.mt-grpo.judge2.creative-writing-v3-abs}{1.5}
\setval{4b.saw.minus.mt-grpo.judge2.longbench-write}{$-$0.7}
\setval{4b.saw.minus.mt-grpo.judge2.longbench-write-abs}{0.7}
\setval{4b.saw.minus.mt-grpo.judge2.writing}{+0.3}
\setval{4b.saw.minus.mt-grpo.judge2.writing-abs}{0.3}
\setval{4b.saw.minus.mt-grpo.judge2.writingbench}{+0.1}
\setval{4b.saw.minus.mt-grpo.judge2.writingbench-abs}{0.1}
\setval{4b.saw.minus.mt-grpo.livecodebench}{+0.2}
\setval{4b.saw.minus.mt-grpo.livecodebench-abs}{0.2}
\setval{4b.saw.minus.mt-grpo.longbench-write}{$-$0.6}
\setval{4b.saw.minus.mt-grpo.longbench-write-abs}{0.6}
\setval{4b.saw.minus.mt-grpo.math}{+1.3}
\setval{4b.saw.minus.mt-grpo.math-abs}{1.3}
\setval{4b.saw.minus.mt-grpo.math500}{+0.8}
\setval{4b.saw.minus.mt-grpo.math500-abs}{0.8}
\setval{4b.saw.minus.mt-grpo.mbpp-plus}{+0.1}
\setval{4b.saw.minus.mt-grpo.mbpp-plus-abs}{0.1}
\setval{4b.saw.minus.mt-grpo.minerva}{+2.2}
\setval{4b.saw.minus.mt-grpo.minerva-abs}{2.2}
\setval{4b.saw.minus.mt-grpo.olympiad-bench}{+1.3}
\setval{4b.saw.minus.mt-grpo.olympiad-bench-abs}{1.3}
\setval{4b.saw.minus.mt-grpo.writing}{$-$0.1}
\setval{4b.saw.minus.mt-grpo.writing-abs}{0.1}
\setval{4b.saw.minus.mt-grpo.writingbench}{$-$0.3}
\setval{4b.saw.minus.mt-grpo.writingbench-abs}{0.3}
\setval{4b.saw.minus.papo.aime24}{+3.2}
\setval{4b.saw.minus.papo.aime24-abs}{3.2}
\setval{4b.saw.minus.papo.aime25}{+3.6}
\setval{4b.saw.minus.papo.aime25-abs}{3.6}
\setval{4b.saw.minus.papo.amc23}{+3.1}
\setval{4b.saw.minus.papo.amc23-abs}{3.1}
\setval{4b.saw.minus.papo.avg}{+1.6}
\setval{4b.saw.minus.papo.avg-abs}{1.6}
\setval{4b.saw.minus.papo.code}{$-$1.2}
\setval{4b.saw.minus.papo.code-abs}{1.2}
\setval{4b.saw.minus.papo.creative-writing-v3}{$-$1.5}
\setval{4b.saw.minus.papo.creative-writing-v3-abs}{1.5}
\setval{4b.saw.minus.papo.guidebench}{+0.3}
\setval{4b.saw.minus.papo.guidebench-abs}{0.3}
\setval{4b.saw.minus.papo.humaneval-plus}{$-$2.3}
\setval{4b.saw.minus.papo.humaneval-plus-abs}{2.3}
\setval{4b.saw.minus.papo.if}{+2.8}
\setval{4b.saw.minus.papo.if-abs}{2.8}
\setval{4b.saw.minus.papo.ifbench}{+7.0}
\setval{4b.saw.minus.papo.ifbench-abs}{7.0}
\setval{4b.saw.minus.papo.ifeval}{+1.1}
\setval{4b.saw.minus.papo.ifeval-abs}{1.1}
\setval{4b.saw.minus.papo.judge2.avg}{+1.7}
\setval{4b.saw.minus.papo.judge2.avg-abs}{1.7}
\setval{4b.saw.minus.papo.judge2.creative-writing-v3}{+1.4}
\setval{4b.saw.minus.papo.judge2.creative-writing-v3-abs}{1.4}
\setval{4b.saw.minus.papo.judge2.longbench-write}{+2.5}
\setval{4b.saw.minus.papo.judge2.longbench-write-abs}{2.5}
\setval{4b.saw.minus.papo.judge2.writing}{+1.7}
\setval{4b.saw.minus.papo.judge2.writing-abs}{1.7}
\setval{4b.saw.minus.papo.judge2.writingbench}{+1.2}
\setval{4b.saw.minus.papo.judge2.writingbench-abs}{1.2}
\setval{4b.saw.minus.papo.livecodebench}{$-$0.7}
\setval{4b.saw.minus.papo.livecodebench-abs}{0.7}
\setval{4b.saw.minus.papo.longbench-write}{+2.2}
\setval{4b.saw.minus.papo.longbench-write-abs}{2.2}
\setval{4b.saw.minus.papo.math}{+2.5}
\setval{4b.saw.minus.papo.math-abs}{2.5}
\setval{4b.saw.minus.papo.math500}{+1.3}
\setval{4b.saw.minus.papo.math500-abs}{1.3}
\setval{4b.saw.minus.papo.mbpp-plus}{$-$0.6}
\setval{4b.saw.minus.papo.mbpp-plus-abs}{0.6}
\setval{4b.saw.minus.papo.minerva}{+1.6}
\setval{4b.saw.minus.papo.minerva-abs}{1.6}
\setval{4b.saw.minus.papo.olympiad-bench}{+2.4}
\setval{4b.saw.minus.papo.olympiad-bench-abs}{2.4}
\setval{4b.saw.minus.papo.writing}{+1.4}
\setval{4b.saw.minus.papo.writing-abs}{1.4}
\setval{4b.saw.minus.papo.writingbench}{+3.4}
\setval{4b.saw.minus.papo.writingbench-abs}{3.4}
\setval{4b.saw.passing-at-ceiling.code.early}{0.71}
\setval{4b.saw.passing-at-ceiling.code.late}{0.73}
\setval{4b.saw.passing-at-ceiling.code.run}{0.73}
\setval{4b.saw.passing-at-ceiling.if.early}{0.28}
\setval{4b.saw.passing-at-ceiling.if.late}{0.57}
\setval{4b.saw.passing-at-ceiling.if.run}{0.49}
\setval{4b.saw.passing-at-ceiling.math.early}{1.00}
\setval{4b.saw.passing-at-ceiling.math.late}{1.00}
\setval{4b.saw.passing-at-ceiling.math.run}{1.00}
\setval{4b.saw.passing-at-ceiling.writing.early}{0.00}
\setval{4b.saw.passing-at-ceiling.writing.late}{0.00}
\setval{4b.saw.passing-at-ceiling.writing.run}{0.00}
\setval{4b.saw.primary-cosine.code.early}{0.96}
\setval{4b.saw.primary-cosine.code.late}{0.95}
\setval{4b.saw.primary-cosine.code.run}{0.96}
\setval{4b.saw.primary-cosine.if.early}{0.87}
\setval{4b.saw.primary-cosine.if.late}{0.76}
\setval{4b.saw.primary-cosine.if.run}{0.80}
\setval{4b.saw.primary-cosine.math.early}{0.97}
\setval{4b.saw.primary-cosine.math.late}{0.98}
\setval{4b.saw.primary-cosine.math.run}{0.98}
\setval{4b.saw.primary-cosine.writing.early}{0.83}
\setval{4b.saw.primary-cosine.writing.late}{0.76}
\setval{4b.saw.primary-cosine.writing.run}{0.75}
\setval{4b.saw.primary-level.code.early}{0.53}
\setval{4b.saw.primary-level.code.late}{0.75}
\setval{4b.saw.primary-level.code.run}{0.70}
\setval{4b.saw.primary-level.if.early}{0.32}
\setval{4b.saw.primary-level.if.late}{0.76}
\setval{4b.saw.primary-level.if.run}{0.65}
\setval{4b.saw.primary-level.math.early}{0.48}
\setval{4b.saw.primary-level.math.late}{0.68}
\setval{4b.saw.primary-level.math.run}{0.62}
\setval{4b.saw.primary-level.writing.early}{0.53}
\setval{4b.saw.primary-level.writing.late}{0.75}
\setval{4b.saw.primary-level.writing.run}{0.71}
\setval{4b.saw.solved.code.early}{0.07}
\setval{4b.saw.solved.code.late}{0.02}
\setval{4b.saw.solved.code.run}{0.03}
\setval{4b.saw.solved.if.early}{0.03}
\setval{4b.saw.solved.if.late}{0.21}
\setval{4b.saw.solved.if.run}{0.14}
\setval{4b.saw.solved.math.early}{0.13}
\setval{4b.saw.solved.math.late}{0.13}
\setval{4b.saw.solved.math.run}{0.12}
\setval{4b.saw.solved.writing.early}{0.00}
\setval{4b.saw.solved.writing.late}{0.00}
\setval{4b.saw.solved.writing.run}{0.00}
\setval{4b.saw.update-share.code.early}{0.19}
\setval{4b.saw.update-share.code.late}{0.23}
\setval{4b.saw.update-share.code.run}{0.22}
\setval{4b.saw.update-share.if.early}{0.23}
\setval{4b.saw.update-share.if.late}{0.16}
\setval{4b.saw.update-share.if.run}{0.20}
\setval{4b.saw.update-share.math.early}{0.17}
\setval{4b.saw.update-share.math.late}{0.26}
\setval{4b.saw.update-share.math.run}{0.24}
\setval{4b.saw.update-share.writing.early}{0.41}
\setval{4b.saw.update-share.writing.late}{0.36}
\setval{4b.saw.update-share.writing.run}{0.34}
\setval{4b.start.aux-level.code}{0.79}
\setval{4b.start.aux-level.if}{0.70}
\setval{4b.start.aux-level.math}{0.85}
\setval{4b.start.aux-level.writing}{0.77}
\setval{4b.start.aux-level.writing-rm}{0.59}
\setval{4b.train.step}{300}
\setval{8b.base.bench.aime24}{11.4}
\setval{8b.base.bench.aime25}{8.5}
\setval{8b.base.bench.amc23}{50.2}
\setval{8b.base.bench.creative-writing-v3}{26.7}
\setval{8b.base.bench.guidebench}{45.6}
\setval{8b.base.bench.humaneval-plus}{80.1}
\setval{8b.base.bench.ifbench}{17.3}
\setval{8b.base.bench.ifeval}{53.0}
\setval{8b.base.bench.livecodebench}{20.9}
\setval{8b.base.bench.longbench-write}{51.4}
\setval{8b.base.bench.math500}{72.5}
\setval{8b.base.bench.mbpp-plus}{67.3}
\setval{8b.base.bench.minerva}{40.9}
\setval{8b.base.bench.olympiad-bench}{39.4}
\setval{8b.base.bench.writingbench}{36.8}
\setval{8b.base.eval.avg}{41.5}
\setval{8b.base.eval.code}{56.1}
\setval{8b.base.eval.if}{38.6}
\setval{8b.base.eval.math}{37.2}
\setval{8b.base.eval.writing}{38.3}
\setval{8b.base.judge2.avg}{41.0}
\setval{8b.base.judge2.creative-writing-v3}{26.4}
\setval{8b.base.judge2.longbench-write}{48.1}
\setval{8b.base.judge2.writing}{36.2}
\setval{8b.base.judge2.writingbench}{34.0}
\setval{8b.base.minus.concur-free.aime24}{$-$8.3}
\setval{8b.base.minus.concur-free.aime24-abs}{8.3}
\setval{8b.base.minus.concur-free.aime25}{$-$8.0}
\setval{8b.base.minus.concur-free.aime25-abs}{8.0}
\setval{8b.base.minus.concur-free.amc23}{$-$14.5}
\setval{8b.base.minus.concur-free.amc23-abs}{14.5}
\setval{8b.base.minus.concur-free.avg}{$-$18.8}
\setval{8b.base.minus.concur-free.avg-abs}{18.8}
\setval{8b.base.minus.concur-free.code}{$-$6.4}
\setval{8b.base.minus.concur-free.code-abs}{6.4}
\setval{8b.base.minus.concur-free.creative-writing-v3}{$-$24.9}
\setval{8b.base.minus.concur-free.creative-writing-v3-abs}{24.9}
\setval{8b.base.minus.concur-free.guidebench}{$-$37.3}
\setval{8b.base.minus.concur-free.guidebench-abs}{37.3}
\setval{8b.base.minus.concur-free.humaneval-plus}{$-$5.2}
\setval{8b.base.minus.concur-free.humaneval-plus-abs}{5.2}
\setval{8b.base.minus.concur-free.if}{$-$31.6}
\setval{8b.base.minus.concur-free.if-abs}{31.6}
\setval{8b.base.minus.concur-free.ifbench}{$-$21.3}
\setval{8b.base.minus.concur-free.ifbench-abs}{21.3}
\setval{8b.base.minus.concur-free.ifeval}{$-$36.2}
\setval{8b.base.minus.concur-free.ifeval-abs}{36.2}
\setval{8b.base.minus.concur-free.judge2.avg}{$-$16.3}
\setval{8b.base.minus.concur-free.judge2.avg-abs}{16.3}
\setval{8b.base.minus.concur-free.judge2.creative-writing-v3}{$-$16.1}
\setval{8b.base.minus.concur-free.judge2.creative-writing-v3-abs}{16.1}
\setval{8b.base.minus.concur-free.judge2.longbench-write}{$-$26.0}
\setval{8b.base.minus.concur-free.judge2.longbench-write-abs}{26.0}
\setval{8b.base.minus.concur-free.judge2.writing}{$-$22.9}
\setval{8b.base.minus.concur-free.judge2.writing-abs}{22.9}
\setval{8b.base.minus.concur-free.judge2.writingbench}{$-$26.7}
\setval{8b.base.minus.concur-free.judge2.writingbench-abs}{26.7}
\setval{8b.base.minus.concur-free.livecodebench}{$-$6.2}
\setval{8b.base.minus.concur-free.livecodebench-abs}{6.2}
\setval{8b.base.minus.concur-free.longbench-write}{$-$31.5}
\setval{8b.base.minus.concur-free.longbench-write-abs}{31.5}
\setval{8b.base.minus.concur-free.math}{$-$10.2}
\setval{8b.base.minus.concur-free.math-abs}{10.2}
\setval{8b.base.minus.concur-free.math500}{$-$11.0}
\setval{8b.base.minus.concur-free.math500-abs}{11.0}
\setval{8b.base.minus.concur-free.mbpp-plus}{$-$7.8}
\setval{8b.base.minus.concur-free.mbpp-plus-abs}{7.8}
\setval{8b.base.minus.concur-free.minerva}{$-$8.9}
\setval{8b.base.minus.concur-free.minerva-abs}{8.9}
\setval{8b.base.minus.concur-free.olympiad-bench}{$-$10.6}
\setval{8b.base.minus.concur-free.olympiad-bench-abs}{10.6}
\setval{8b.base.minus.concur-free.writing}{$-$35.5}
\setval{8b.base.minus.concur-free.writing-abs}{35.5}
\setval{8b.base.minus.concur-free.writingbench}{$-$50.0}
\setval{8b.base.minus.concur-free.writingbench-abs}{50.0}
\setval{8b.base.minus.concur-tied.aime24}{$-$19.9}
\setval{8b.base.minus.concur-tied.aime24-abs}{19.9}
\setval{8b.base.minus.concur-tied.aime25}{$-$15.4}
\setval{8b.base.minus.concur-tied.aime25-abs}{15.4}
\setval{8b.base.minus.concur-tied.amc23}{$-$28.3}
\setval{8b.base.minus.concur-tied.amc23-abs}{28.3}
\setval{8b.base.minus.concur-tied.avg}{$-$22.5}
\setval{8b.base.minus.concur-tied.avg-abs}{22.5}
\setval{8b.base.minus.concur-tied.code}{$-$7.1}
\setval{8b.base.minus.concur-tied.code-abs}{7.1}
\setval{8b.base.minus.concur-tied.creative-writing-v3}{$-$25.7}
\setval{8b.base.minus.concur-tied.creative-writing-v3-abs}{25.7}
\setval{8b.base.minus.concur-tied.guidebench}{$-$37.8}
\setval{8b.base.minus.concur-tied.guidebench-abs}{37.8}
\setval{8b.base.minus.concur-tied.humaneval-plus}{$-$7.3}
\setval{8b.base.minus.concur-tied.humaneval-plus-abs}{7.3}
\setval{8b.base.minus.concur-tied.if}{$-$31.9}
\setval{8b.base.minus.concur-tied.if-abs}{31.9}
\setval{8b.base.minus.concur-tied.ifbench}{$-$22.0}
\setval{8b.base.minus.concur-tied.ifbench-abs}{22.0}
\setval{8b.base.minus.concur-tied.ifeval}{$-$35.9}
\setval{8b.base.minus.concur-tied.ifeval-abs}{35.9}
\setval{8b.base.minus.concur-tied.judge2.avg}{$-$20.3}
\setval{8b.base.minus.concur-tied.judge2.avg-abs}{20.3}
\setval{8b.base.minus.concur-tied.judge2.creative-writing-v3}{$-$15.3}
\setval{8b.base.minus.concur-tied.judge2.creative-writing-v3-abs}{15.3}
\setval{8b.base.minus.concur-tied.judge2.longbench-write}{$-$26.7}
\setval{8b.base.minus.concur-tied.judge2.longbench-write-abs}{26.7}
\setval{8b.base.minus.concur-tied.judge2.writing}{$-$23.9}
\setval{8b.base.minus.concur-tied.judge2.writing-abs}{23.9}
\setval{8b.base.minus.concur-tied.judge2.writingbench}{$-$29.8}
\setval{8b.base.minus.concur-tied.judge2.writingbench-abs}{29.8}
\setval{8b.base.minus.concur-tied.livecodebench}{$-$6.3}
\setval{8b.base.minus.concur-tied.livecodebench-abs}{6.3}
\setval{8b.base.minus.concur-tied.longbench-write}{$-$29.8}
\setval{8b.base.minus.concur-tied.longbench-write-abs}{29.8}
\setval{8b.base.minus.concur-tied.math}{$-$19.2}
\setval{8b.base.minus.concur-tied.math-abs}{19.2}
\setval{8b.base.minus.concur-tied.math500}{$-$17.1}
\setval{8b.base.minus.concur-tied.math500-abs}{17.1}
\setval{8b.base.minus.concur-tied.mbpp-plus}{$-$7.6}
\setval{8b.base.minus.concur-tied.mbpp-plus-abs}{7.6}
\setval{8b.base.minus.concur-tied.minerva}{$-$14.5}
\setval{8b.base.minus.concur-tied.minerva-abs}{14.5}
\setval{8b.base.minus.concur-tied.olympiad-bench}{$-$20.1}
\setval{8b.base.minus.concur-tied.olympiad-bench-abs}{20.1}
\setval{8b.base.minus.concur-tied.writing}{$-$35.3}
\setval{8b.base.minus.concur-tied.writing-abs}{35.3}
\setval{8b.base.minus.concur-tied.writingbench}{$-$50.4}
\setval{8b.base.minus.concur-tied.writingbench-abs}{50.4}
\setval{8b.base.minus.concur.aime24}{$-$17.4}
\setval{8b.base.minus.concur.aime24-abs}{17.4}
\setval{8b.base.minus.concur.aime25}{$-$14.6}
\setval{8b.base.minus.concur.aime25-abs}{14.6}
\setval{8b.base.minus.concur.amc23}{$-$28.7}
\setval{8b.base.minus.concur.amc23-abs}{28.7}
\setval{8b.base.minus.concur.avg}{$-$21.7}
\setval{8b.base.minus.concur.avg-abs}{21.7}
\setval{8b.base.minus.concur.code}{$-$4.7}
\setval{8b.base.minus.concur.code-abs}{4.7}
\setval{8b.base.minus.concur.creative-writing-v3}{$-$23.2}
\setval{8b.base.minus.concur.creative-writing-v3-abs}{23.2}
\setval{8b.base.minus.concur.guidebench}{$-$37.8}
\setval{8b.base.minus.concur.guidebench-abs}{37.8}
\setval{8b.base.minus.concur.humaneval-plus}{$-$2.7}
\setval{8b.base.minus.concur.humaneval-plus-abs}{2.7}
\setval{8b.base.minus.concur.if}{$-$31.4}
\setval{8b.base.minus.concur.if-abs}{31.4}
\setval{8b.base.minus.concur.ifbench}{$-$21.3}
\setval{8b.base.minus.concur.ifbench-abs}{21.3}
\setval{8b.base.minus.concur.ifeval}{$-$35.1}
\setval{8b.base.minus.concur.ifeval-abs}{35.1}
\setval{8b.base.minus.concur.judge2.avg}{$-$19.3}
\setval{8b.base.minus.concur.judge2.avg-abs}{19.3}
\setval{8b.base.minus.concur.judge2.creative-writing-v3}{$-$11.3}
\setval{8b.base.minus.concur.judge2.creative-writing-v3-abs}{11.3}
\setval{8b.base.minus.concur.judge2.longbench-write}{$-$27.4}
\setval{8b.base.minus.concur.judge2.longbench-write-abs}{27.4}
\setval{8b.base.minus.concur.judge2.writing}{$-$22.9}
\setval{8b.base.minus.concur.judge2.writing-abs}{22.9}
\setval{8b.base.minus.concur.judge2.writingbench}{$-$29.9}
\setval{8b.base.minus.concur.judge2.writingbench-abs}{29.9}
\setval{8b.base.minus.concur.livecodebench}{$-$5.4}
\setval{8b.base.minus.concur.livecodebench-abs}{5.4}
\setval{8b.base.minus.concur.longbench-write}{$-$30.2}
\setval{8b.base.minus.concur.longbench-write-abs}{30.2}
\setval{8b.base.minus.concur.math}{$-$18.8}
\setval{8b.base.minus.concur.math-abs}{18.8}
\setval{8b.base.minus.concur.math500}{$-$17.7}
\setval{8b.base.minus.concur.math500-abs}{17.7}
\setval{8b.base.minus.concur.mbpp-plus}{$-$5.9}
\setval{8b.base.minus.concur.mbpp-plus-abs}{5.9}
\setval{8b.base.minus.concur.minerva}{$-$13.3}
\setval{8b.base.minus.concur.minerva-abs}{13.3}
\setval{8b.base.minus.concur.olympiad-bench}{$-$20.9}
\setval{8b.base.minus.concur.olympiad-bench-abs}{20.9}
\setval{8b.base.minus.concur.writing}{$-$34.9}
\setval{8b.base.minus.concur.writing-abs}{34.9}
\setval{8b.base.minus.concur.writingbench}{$-$51.3}
\setval{8b.base.minus.concur.writingbench-abs}{51.3}
\setval{8b.base.minus.gd2po.aime24}{$-$4.9}
\setval{8b.base.minus.gd2po.aime24-abs}{4.9}
\setval{8b.base.minus.gd2po.aime25}{$-$3.6}
\setval{8b.base.minus.gd2po.aime25-abs}{3.6}
\setval{8b.base.minus.gd2po.amc23}{$-$9.4}
\setval{8b.base.minus.gd2po.amc23-abs}{9.4}
\setval{8b.base.minus.gd2po.avg}{$-$15.9}
\setval{8b.base.minus.gd2po.avg-abs}{15.9}
\setval{8b.base.minus.gd2po.code}{$-$5.7}
\setval{8b.base.minus.gd2po.code-abs}{5.7}
\setval{8b.base.minus.gd2po.creative-writing-v3}{$-$21.4}
\setval{8b.base.minus.gd2po.creative-writing-v3-abs}{21.4}
\setval{8b.base.minus.gd2po.guidebench}{$-$38.1}
\setval{8b.base.minus.gd2po.guidebench-abs}{38.1}
\setval{8b.base.minus.gd2po.humaneval-plus}{$-$3.9}
\setval{8b.base.minus.gd2po.humaneval-plus-abs}{3.9}
\setval{8b.base.minus.gd2po.if}{$-$33.4}
\setval{8b.base.minus.gd2po.if-abs}{33.4}
\setval{8b.base.minus.gd2po.ifbench}{$-$25.7}
\setval{8b.base.minus.gd2po.ifbench-abs}{25.7}
\setval{8b.base.minus.gd2po.ifeval}{$-$36.3}
\setval{8b.base.minus.gd2po.ifeval-abs}{36.3}
\setval{8b.base.minus.gd2po.judge2.avg}{$-$13.8}
\setval{8b.base.minus.gd2po.judge2.avg-abs}{13.8}
\setval{8b.base.minus.gd2po.judge2.creative-writing-v3}{$-$10.3}
\setval{8b.base.minus.gd2po.judge2.creative-writing-v3-abs}{10.3}
\setval{8b.base.minus.gd2po.judge2.longbench-write}{$-$14.1}
\setval{8b.base.minus.gd2po.judge2.longbench-write-abs}{14.1}
\setval{8b.base.minus.gd2po.judge2.writing}{$-$15.2}
\setval{8b.base.minus.gd2po.judge2.writing-abs}{15.2}
\setval{8b.base.minus.gd2po.judge2.writingbench}{$-$21.2}
\setval{8b.base.minus.gd2po.judge2.writingbench-abs}{21.2}
\setval{8b.base.minus.gd2po.livecodebench}{$-$5.4}
\setval{8b.base.minus.gd2po.livecodebench-abs}{5.4}
\setval{8b.base.minus.gd2po.longbench-write}{$-$18.5}
\setval{8b.base.minus.gd2po.longbench-write-abs}{18.5}
\setval{8b.base.minus.gd2po.math}{$-$7.5}
\setval{8b.base.minus.gd2po.math-abs}{7.5}
\setval{8b.base.minus.gd2po.math500}{$-$8.8}
\setval{8b.base.minus.gd2po.math500-abs}{8.8}
\setval{8b.base.minus.gd2po.mbpp-plus}{$-$7.8}
\setval{8b.base.minus.gd2po.mbpp-plus-abs}{7.8}
\setval{8b.base.minus.gd2po.minerva}{$-$9.0}
\setval{8b.base.minus.gd2po.minerva-abs}{9.0}
\setval{8b.base.minus.gd2po.olympiad-bench}{$-$9.1}
\setval{8b.base.minus.gd2po.olympiad-bench-abs}{9.1}
\setval{8b.base.minus.gd2po.writing}{$-$25.7}
\setval{8b.base.minus.gd2po.writing-abs}{25.7}
\setval{8b.base.minus.gd2po.writingbench}{$-$37.1}
\setval{8b.base.minus.gd2po.writingbench-abs}{37.1}
\setval{8b.base.minus.gdpo-noaux.aime24}{$-$16.6}
\setval{8b.base.minus.gdpo-noaux.aime24-abs}{16.6}
\setval{8b.base.minus.gdpo-noaux.aime25}{$-$15.4}
\setval{8b.base.minus.gdpo-noaux.aime25-abs}{15.4}
\setval{8b.base.minus.gdpo-noaux.amc23}{$-$26.9}
\setval{8b.base.minus.gdpo-noaux.amc23-abs}{26.9}
\setval{8b.base.minus.gdpo-noaux.avg}{$-$19.1}
\setval{8b.base.minus.gdpo-noaux.avg-abs}{19.1}
\setval{8b.base.minus.gdpo-noaux.code}{$-$6.7}
\setval{8b.base.minus.gdpo-noaux.code-abs}{6.7}
\setval{8b.base.minus.gdpo-noaux.creative-writing-v3}{$-$17.2}
\setval{8b.base.minus.gdpo-noaux.creative-writing-v3-abs}{17.2}
\setval{8b.base.minus.gdpo-noaux.guidebench}{$-$35.8}
\setval{8b.base.minus.gdpo-noaux.guidebench-abs}{35.8}
\setval{8b.base.minus.gdpo-noaux.humaneval-plus}{$-$5.9}
\setval{8b.base.minus.gdpo-noaux.humaneval-plus-abs}{5.9}
\setval{8b.base.minus.gdpo-noaux.if}{$-$28.6}
\setval{8b.base.minus.gdpo-noaux.if-abs}{28.6}
\setval{8b.base.minus.gdpo-noaux.ifbench}{$-$16.3}
\setval{8b.base.minus.gdpo-noaux.ifbench-abs}{16.3}
\setval{8b.base.minus.gdpo-noaux.ifeval}{$-$33.6}
\setval{8b.base.minus.gdpo-noaux.ifeval-abs}{33.6}
\setval{8b.base.minus.gdpo-noaux.judge2.avg}{$-$16.0}
\setval{8b.base.minus.gdpo-noaux.judge2.avg-abs}{16.0}
\setval{8b.base.minus.gdpo-noaux.judge2.creative-writing-v3}{$-$2.4}
\setval{8b.base.minus.gdpo-noaux.judge2.creative-writing-v3-abs}{2.4}
\setval{8b.base.minus.gdpo-noaux.judge2.longbench-write}{$-$6.4}
\setval{8b.base.minus.gdpo-noaux.judge2.longbench-write-abs}{6.4}
\setval{8b.base.minus.gdpo-noaux.judge2.writing}{$-$8.3}
\setval{8b.base.minus.gdpo-noaux.judge2.writing-abs}{8.3}
\setval{8b.base.minus.gdpo-noaux.judge2.writingbench}{$-$16.0}
\setval{8b.base.minus.gdpo-noaux.judge2.writingbench-abs}{16.0}
\setval{8b.base.minus.gdpo-noaux.livecodebench}{$-$6.5}
\setval{8b.base.minus.gdpo-noaux.livecodebench-abs}{6.5}
\setval{8b.base.minus.gdpo-noaux.longbench-write}{$-$14.5}
\setval{8b.base.minus.gdpo-noaux.longbench-write-abs}{14.5}
\setval{8b.base.minus.gdpo-noaux.math}{$-$18.4}
\setval{8b.base.minus.gdpo-noaux.math-abs}{18.4}
\setval{8b.base.minus.gdpo-noaux.math500}{$-$17.8}
\setval{8b.base.minus.gdpo-noaux.math500-abs}{17.8}
\setval{8b.base.minus.gdpo-noaux.mbpp-plus}{$-$7.6}
\setval{8b.base.minus.gdpo-noaux.mbpp-plus-abs}{7.6}
\setval{8b.base.minus.gdpo-noaux.minerva}{$-$13.1}
\setval{8b.base.minus.gdpo-noaux.minerva-abs}{13.1}
\setval{8b.base.minus.gdpo-noaux.olympiad-bench}{$-$20.5}
\setval{8b.base.minus.gdpo-noaux.olympiad-bench-abs}{20.5}
\setval{8b.base.minus.gdpo-noaux.writing}{$-$23.3}
\setval{8b.base.minus.gdpo-noaux.writing-abs}{23.3}
\setval{8b.base.minus.gdpo-noaux.writingbench}{$-$38.2}
\setval{8b.base.minus.gdpo-noaux.writingbench-abs}{38.2}
\setval{8b.base.minus.gdpo-w0.5.aime24}{$-$7.9}
\setval{8b.base.minus.gdpo-w0.5.aime24-abs}{7.9}
\setval{8b.base.minus.gdpo-w0.5.aime25}{$-$6.9}
\setval{8b.base.minus.gdpo-w0.5.aime25-abs}{6.9}
\setval{8b.base.minus.gdpo-w0.5.amc23}{$-$13.2}
\setval{8b.base.minus.gdpo-w0.5.amc23-abs}{13.2}
\setval{8b.base.minus.gdpo-w0.5.avg}{$-$16.9}
\setval{8b.base.minus.gdpo-w0.5.avg-abs}{16.9}
\setval{8b.base.minus.gdpo-w0.5.code}{$-$5.1}
\setval{8b.base.minus.gdpo-w0.5.code-abs}{5.1}
\setval{8b.base.minus.gdpo-w0.5.creative-writing-v3}{$-$24.8}
\setval{8b.base.minus.gdpo-w0.5.creative-writing-v3-abs}{24.8}
\setval{8b.base.minus.gdpo-w0.5.guidebench}{$-$38.4}
\setval{8b.base.minus.gdpo-w0.5.guidebench-abs}{38.4}
\setval{8b.base.minus.gdpo-w0.5.humaneval-plus}{$-$3.7}
\setval{8b.base.minus.gdpo-w0.5.humaneval-plus-abs}{3.7}
\setval{8b.base.minus.gdpo-w0.5.if}{$-$31.2}
\setval{8b.base.minus.gdpo-w0.5.if-abs}{31.2}
\setval{8b.base.minus.gdpo-w0.5.ifbench}{$-$20.3}
\setval{8b.base.minus.gdpo-w0.5.ifbench-abs}{20.3}
\setval{8b.base.minus.gdpo-w0.5.ifeval}{$-$34.8}
\setval{8b.base.minus.gdpo-w0.5.ifeval-abs}{34.8}
\setval{8b.base.minus.gdpo-w0.5.judge2.avg}{$-$14.8}
\setval{8b.base.minus.gdpo-w0.5.judge2.avg-abs}{14.8}
\setval{8b.base.minus.gdpo-w0.5.judge2.creative-writing-v3}{$-$13.9}
\setval{8b.base.minus.gdpo-w0.5.judge2.creative-writing-v3-abs}{13.9}
\setval{8b.base.minus.gdpo-w0.5.judge2.longbench-write}{$-$15.6}
\setval{8b.base.minus.gdpo-w0.5.judge2.longbench-write-abs}{15.6}
\setval{8b.base.minus.gdpo-w0.5.judge2.writing}{$-$17.2}
\setval{8b.base.minus.gdpo-w0.5.judge2.writing-abs}{17.2}
\setval{8b.base.minus.gdpo-w0.5.judge2.writingbench}{$-$22.2}
\setval{8b.base.minus.gdpo-w0.5.judge2.writingbench-abs}{22.2}
\setval{8b.base.minus.gdpo-w0.5.livecodebench}{$-$5.9}
\setval{8b.base.minus.gdpo-w0.5.livecodebench-abs}{5.9}
\setval{8b.base.minus.gdpo-w0.5.longbench-write}{$-$19.8}
\setval{8b.base.minus.gdpo-w0.5.longbench-write-abs}{19.8}
\setval{8b.base.minus.gdpo-w0.5.math}{$-$10.3}
\setval{8b.base.minus.gdpo-w0.5.math-abs}{10.3}
\setval{8b.base.minus.gdpo-w0.5.math500}{$-$11.6}
\setval{8b.base.minus.gdpo-w0.5.math500-abs}{11.6}
\setval{8b.base.minus.gdpo-w0.5.mbpp-plus}{$-$5.6}
\setval{8b.base.minus.gdpo-w0.5.mbpp-plus-abs}{5.6}
\setval{8b.base.minus.gdpo-w0.5.minerva}{$-$10.3}
\setval{8b.base.minus.gdpo-w0.5.minerva-abs}{10.3}
\setval{8b.base.minus.gdpo-w0.5.olympiad-bench}{$-$11.7}
\setval{8b.base.minus.gdpo-w0.5.olympiad-bench-abs}{11.7}
\setval{8b.base.minus.gdpo-w0.5.writing}{$-$27.6}
\setval{8b.base.minus.gdpo-w0.5.writing-abs}{27.6}
\setval{8b.base.minus.gdpo-w0.5.writingbench}{$-$38.3}
\setval{8b.base.minus.gdpo-w0.5.writingbench-abs}{38.3}
\setval{8b.base.minus.gdpo.aime24}{$-$10.0}
\setval{8b.base.minus.gdpo.aime24-abs}{10.0}
\setval{8b.base.minus.gdpo.aime25}{$-$5.8}
\setval{8b.base.minus.gdpo.aime25-abs}{5.8}
\setval{8b.base.minus.gdpo.amc23}{$-$12.0}
\setval{8b.base.minus.gdpo.amc23-abs}{12.0}
\setval{8b.base.minus.gdpo.avg}{$-$16.9}
\setval{8b.base.minus.gdpo.avg-abs}{16.9}
\setval{8b.base.minus.gdpo.code}{$-$5.5}
\setval{8b.base.minus.gdpo.code-abs}{5.5}
\setval{8b.base.minus.gdpo.creative-writing-v3}{$-$26.2}
\setval{8b.base.minus.gdpo.creative-writing-v3-abs}{26.2}
\setval{8b.base.minus.gdpo.guidebench}{$-$39.0}
\setval{8b.base.minus.gdpo.guidebench-abs}{39.0}
\setval{8b.base.minus.gdpo.humaneval-plus}{$-$4.0}
\setval{8b.base.minus.gdpo.humaneval-plus-abs}{4.0}
\setval{8b.base.minus.gdpo.if}{$-$32.0}
\setval{8b.base.minus.gdpo.if-abs}{32.0}
\setval{8b.base.minus.gdpo.ifbench}{$-$22.0}
\setval{8b.base.minus.gdpo.ifbench-abs}{22.0}
\setval{8b.base.minus.gdpo.ifeval}{$-$35.1}
\setval{8b.base.minus.gdpo.ifeval-abs}{35.1}
\setval{8b.base.minus.gdpo.judge2.avg}{$-$15.0}
\setval{8b.base.minus.gdpo.judge2.avg-abs}{15.0}
\setval{8b.base.minus.gdpo.judge2.creative-writing-v3}{$-$14.6}
\setval{8b.base.minus.gdpo.judge2.creative-writing-v3-abs}{14.6}
\setval{8b.base.minus.gdpo.judge2.longbench-write}{$-$16.1}
\setval{8b.base.minus.gdpo.judge2.longbench-write-abs}{16.1}
\setval{8b.base.minus.gdpo.judge2.writing}{$-$18.4}
\setval{8b.base.minus.gdpo.judge2.writing-abs}{18.4}
\setval{8b.base.minus.gdpo.judge2.writingbench}{$-$24.4}
\setval{8b.base.minus.gdpo.judge2.writingbench-abs}{24.4}
\setval{8b.base.minus.gdpo.livecodebench}{$-$5.3}
\setval{8b.base.minus.gdpo.livecodebench-abs}{5.3}
\setval{8b.base.minus.gdpo.longbench-write}{$-$18.2}
\setval{8b.base.minus.gdpo.longbench-write-abs}{18.2}
\setval{8b.base.minus.gdpo.math}{$-$9.5}
\setval{8b.base.minus.gdpo.math-abs}{9.5}
\setval{8b.base.minus.gdpo.math500}{$-$10.5}
\setval{8b.base.minus.gdpo.math500-abs}{10.5}
\setval{8b.base.minus.gdpo.mbpp-plus}{$-$7.2}
\setval{8b.base.minus.gdpo.mbpp-plus-abs}{7.2}
\setval{8b.base.minus.gdpo.minerva}{$-$8.7}
\setval{8b.base.minus.gdpo.minerva-abs}{8.7}
\setval{8b.base.minus.gdpo.olympiad-bench}{$-$9.9}
\setval{8b.base.minus.gdpo.olympiad-bench-abs}{9.9}
\setval{8b.base.minus.gdpo.writing}{$-$27.9}
\setval{8b.base.minus.gdpo.writing-abs}{27.9}
\setval{8b.base.minus.gdpo.writingbench}{$-$39.3}
\setval{8b.base.minus.gdpo.writingbench-abs}{39.3}
\setval{8b.base.minus.geoma.aime24}{$-$2.7}
\setval{8b.base.minus.geoma.aime24-abs}{2.7}
\setval{8b.base.minus.geoma.aime25}{$-$4.0}
\setval{8b.base.minus.geoma.aime25-abs}{4.0}
\setval{8b.base.minus.geoma.amc23}{$-$13.3}
\setval{8b.base.minus.geoma.amc23-abs}{13.3}
\setval{8b.base.minus.geoma.avg}{$-$15.2}
\setval{8b.base.minus.geoma.avg-abs}{15.2}
\setval{8b.base.minus.geoma.code}{$-$5.3}
\setval{8b.base.minus.geoma.code-abs}{5.3}
\setval{8b.base.minus.geoma.creative-writing-v3}{$-$23.7}
\setval{8b.base.minus.geoma.creative-writing-v3-abs}{23.7}
\setval{8b.base.minus.geoma.guidebench}{$-$38.8}
\setval{8b.base.minus.geoma.guidebench-abs}{38.8}
\setval{8b.base.minus.geoma.humaneval-plus}{$-$2.7}
\setval{8b.base.minus.geoma.humaneval-plus-abs}{2.7}
\setval{8b.base.minus.geoma.if}{$-$29.9}
\setval{8b.base.minus.geoma.if-abs}{29.9}
\setval{8b.base.minus.geoma.ifbench}{$-$16.7}
\setval{8b.base.minus.geoma.ifbench-abs}{16.7}
\setval{8b.base.minus.geoma.ifeval}{$-$34.3}
\setval{8b.base.minus.geoma.ifeval-abs}{34.3}
\setval{8b.base.minus.geoma.judge2.avg}{$-$13.4}
\setval{8b.base.minus.geoma.judge2.avg-abs}{13.4}
\setval{8b.base.minus.geoma.judge2.creative-writing-v3}{$-$11.1}
\setval{8b.base.minus.geoma.judge2.creative-writing-v3-abs}{11.1}
\setval{8b.base.minus.geoma.judge2.longbench-write}{$-$16.8}
\setval{8b.base.minus.geoma.judge2.longbench-write-abs}{16.8}
\setval{8b.base.minus.geoma.judge2.writing}{$-$17.4}
\setval{8b.base.minus.geoma.judge2.writing-abs}{17.4}
\setval{8b.base.minus.geoma.judge2.writingbench}{$-$24.3}
\setval{8b.base.minus.geoma.judge2.writingbench-abs}{24.3}
\setval{8b.base.minus.geoma.livecodebench}{$-$4.8}
\setval{8b.base.minus.geoma.livecodebench-abs}{4.8}
\setval{8b.base.minus.geoma.longbench-write}{$-$17.5}
\setval{8b.base.minus.geoma.longbench-write-abs}{17.5}
\setval{8b.base.minus.geoma.math}{$-$7.1}
\setval{8b.base.minus.geoma.math-abs}{7.1}
\setval{8b.base.minus.geoma.math500}{$-$7.2}
\setval{8b.base.minus.geoma.math500-abs}{7.2}
\setval{8b.base.minus.geoma.mbpp-plus}{$-$8.3}
\setval{8b.base.minus.geoma.mbpp-plus-abs}{8.3}
\setval{8b.base.minus.geoma.minerva}{$-$8.9}
\setval{8b.base.minus.geoma.minerva-abs}{8.9}
\setval{8b.base.minus.geoma.olympiad-bench}{$-$6.6}
\setval{8b.base.minus.geoma.olympiad-bench-abs}{6.6}
\setval{8b.base.minus.geoma.writing}{$-$26.8}
\setval{8b.base.minus.geoma.writing-abs}{26.8}
\setval{8b.base.minus.geoma.writingbench}{$-$39.2}
\setval{8b.base.minus.geoma.writingbench-abs}{39.2}
\setval{8b.base.minus.grpo-nodiv.aime24}{$-$9.3}
\setval{8b.base.minus.grpo-nodiv.aime24-abs}{9.3}
\setval{8b.base.minus.grpo-nodiv.aime25}{$-$6.2}
\setval{8b.base.minus.grpo-nodiv.aime25-abs}{6.2}
\setval{8b.base.minus.grpo-nodiv.amc23}{$-$13.4}
\setval{8b.base.minus.grpo-nodiv.amc23-abs}{13.4}
\setval{8b.base.minus.grpo-nodiv.avg}{$-$16.6}
\setval{8b.base.minus.grpo-nodiv.avg-abs}{16.6}
\setval{8b.base.minus.grpo-nodiv.code}{$-$6.4}
\setval{8b.base.minus.grpo-nodiv.code-abs}{6.4}
\setval{8b.base.minus.grpo-nodiv.creative-writing-v3}{$-$19.9}
\setval{8b.base.minus.grpo-nodiv.creative-writing-v3-abs}{19.9}
\setval{8b.base.minus.grpo-nodiv.guidebench}{$-$39.3}
\setval{8b.base.minus.grpo-nodiv.guidebench-abs}{39.3}
\setval{8b.base.minus.grpo-nodiv.humaneval-plus}{$-$6.3}
\setval{8b.base.minus.grpo-nodiv.humaneval-plus-abs}{6.3}
\setval{8b.base.minus.grpo-nodiv.if}{$-$31.6}
\setval{8b.base.minus.grpo-nodiv.if-abs}{31.6}
\setval{8b.base.minus.grpo-nodiv.ifbench}{$-$19.0}
\setval{8b.base.minus.grpo-nodiv.ifbench-abs}{19.0}
\setval{8b.base.minus.grpo-nodiv.ifeval}{$-$36.5}
\setval{8b.base.minus.grpo-nodiv.ifeval-abs}{36.5}
\setval{8b.base.minus.grpo-nodiv.judge2.avg}{$-$15.0}
\setval{8b.base.minus.grpo-nodiv.judge2.avg-abs}{15.0}
\setval{8b.base.minus.grpo-nodiv.judge2.creative-writing-v3}{$-$11.0}
\setval{8b.base.minus.grpo-nodiv.judge2.creative-writing-v3-abs}{11.0}
\setval{8b.base.minus.grpo-nodiv.judge2.longbench-write}{$-$18.5}
\setval{8b.base.minus.grpo-nodiv.judge2.longbench-write-abs}{18.5}
\setval{8b.base.minus.grpo-nodiv.judge2.writing}{$-$16.5}
\setval{8b.base.minus.grpo-nodiv.judge2.writing-abs}{16.5}
\setval{8b.base.minus.grpo-nodiv.judge2.writingbench}{$-$19.9}
\setval{8b.base.minus.grpo-nodiv.judge2.writingbench-abs}{19.9}
\setval{8b.base.minus.grpo-nodiv.livecodebench}{$-$4.1}
\setval{8b.base.minus.grpo-nodiv.livecodebench-abs}{4.1}
\setval{8b.base.minus.grpo-nodiv.longbench-write}{$-$19.6}
\setval{8b.base.minus.grpo-nodiv.longbench-write-abs}{19.6}
\setval{8b.base.minus.grpo-nodiv.math}{$-$10.4}
\setval{8b.base.minus.grpo-nodiv.math-abs}{10.4}
\setval{8b.base.minus.grpo-nodiv.math500}{$-$11.7}
\setval{8b.base.minus.grpo-nodiv.math500-abs}{11.7}
\setval{8b.base.minus.grpo-nodiv.mbpp-plus}{$-$8.8}
\setval{8b.base.minus.grpo-nodiv.mbpp-plus-abs}{8.8}
\setval{8b.base.minus.grpo-nodiv.minerva}{$-$9.9}
\setval{8b.base.minus.grpo-nodiv.minerva-abs}{9.9}
\setval{8b.base.minus.grpo-nodiv.olympiad-bench}{$-$12.1}
\setval{8b.base.minus.grpo-nodiv.olympiad-bench-abs}{12.1}
\setval{8b.base.minus.grpo-nodiv.writing}{$-$24.4}
\setval{8b.base.minus.grpo-nodiv.writing-abs}{24.4}
\setval{8b.base.minus.grpo-nodiv.writingbench}{$-$33.6}
\setval{8b.base.minus.grpo-nodiv.writingbench-abs}{33.6}
\setval{8b.base.minus.grpo.aime24}{$-$4.5}
\setval{8b.base.minus.grpo.aime24-abs}{4.5}
\setval{8b.base.minus.grpo.aime25}{$-$7.0}
\setval{8b.base.minus.grpo.aime25-abs}{7.0}
\setval{8b.base.minus.grpo.amc23}{$-$14.1}
\setval{8b.base.minus.grpo.amc23-abs}{14.1}
\setval{8b.base.minus.grpo.avg}{$-$16.4}
\setval{8b.base.minus.grpo.avg-abs}{16.4}
\setval{8b.base.minus.grpo.code}{$-$5.5}
\setval{8b.base.minus.grpo.code-abs}{5.5}
\setval{8b.base.minus.grpo.creative-writing-v3}{$-$25.8}
\setval{8b.base.minus.grpo.creative-writing-v3-abs}{25.8}
\setval{8b.base.minus.grpo.guidebench}{$-$36.3}
\setval{8b.base.minus.grpo.guidebench-abs}{36.3}
\setval{8b.base.minus.grpo.humaneval-plus}{$-$3.7}
\setval{8b.base.minus.grpo.humaneval-plus-abs}{3.7}
\setval{8b.base.minus.grpo.if}{$-$31.6}
\setval{8b.base.minus.grpo.if-abs}{31.6}
\setval{8b.base.minus.grpo.ifbench}{$-$24.7}
\setval{8b.base.minus.grpo.ifbench-abs}{24.7}
\setval{8b.base.minus.grpo.ifeval}{$-$33.8}
\setval{8b.base.minus.grpo.ifeval-abs}{33.8}
\setval{8b.base.minus.grpo.judge2.avg}{$-$14.2}
\setval{8b.base.minus.grpo.judge2.avg-abs}{14.2}
\setval{8b.base.minus.grpo.judge2.creative-writing-v3}{$-$12.8}
\setval{8b.base.minus.grpo.judge2.creative-writing-v3-abs}{12.8}
\setval{8b.base.minus.grpo.judge2.longbench-write}{$-$14.4}
\setval{8b.base.minus.grpo.judge2.longbench-write-abs}{14.4}
\setval{8b.base.minus.grpo.judge2.writing}{$-$16.7}
\setval{8b.base.minus.grpo.judge2.writing-abs}{16.7}
\setval{8b.base.minus.grpo.judge2.writingbench}{$-$22.9}
\setval{8b.base.minus.grpo.judge2.writingbench-abs}{22.9}
\setval{8b.base.minus.grpo.livecodebench}{$-$5.6}
\setval{8b.base.minus.grpo.livecodebench-abs}{5.6}
\setval{8b.base.minus.grpo.longbench-write}{$-$18.4}
\setval{8b.base.minus.grpo.longbench-write-abs}{18.4}
\setval{8b.base.minus.grpo.math}{$-$8.5}
\setval{8b.base.minus.grpo.math-abs}{8.5}
\setval{8b.base.minus.grpo.math500}{$-$8.2}
\setval{8b.base.minus.grpo.math500-abs}{8.2}
\setval{8b.base.minus.grpo.mbpp-plus}{$-$7.4}
\setval{8b.base.minus.grpo.mbpp-plus-abs}{7.4}
\setval{8b.base.minus.grpo.minerva}{$-$8.9}
\setval{8b.base.minus.grpo.minerva-abs}{8.9}
\setval{8b.base.minus.grpo.olympiad-bench}{$-$8.5}
\setval{8b.base.minus.grpo.olympiad-bench-abs}{8.5}
\setval{8b.base.minus.grpo.writing}{$-$27.9}
\setval{8b.base.minus.grpo.writing-abs}{27.9}
\setval{8b.base.minus.grpo.writingbench}{$-$39.7}
\setval{8b.base.minus.grpo.writingbench-abs}{39.7}
\setval{8b.base.minus.mt-grpo.aime24}{$-$3.1}
\setval{8b.base.minus.mt-grpo.aime24-abs}{3.1}
\setval{8b.base.minus.mt-grpo.aime25}{$-$7.2}
\setval{8b.base.minus.mt-grpo.aime25-abs}{7.2}
\setval{8b.base.minus.mt-grpo.amc23}{$-$15.4}
\setval{8b.base.minus.mt-grpo.amc23-abs}{15.4}
\setval{8b.base.minus.mt-grpo.avg}{$-$15.7}
\setval{8b.base.minus.mt-grpo.avg-abs}{15.7}
\setval{8b.base.minus.mt-grpo.code}{$-$6.6}
\setval{8b.base.minus.mt-grpo.code-abs}{6.6}
\setval{8b.base.minus.mt-grpo.creative-writing-v3}{$-$23.6}
\setval{8b.base.minus.mt-grpo.creative-writing-v3-abs}{23.6}
\setval{8b.base.minus.mt-grpo.guidebench}{$-$37.5}
\setval{8b.base.minus.mt-grpo.guidebench-abs}{37.5}
\setval{8b.base.minus.mt-grpo.humaneval-plus}{$-$5.0}
\setval{8b.base.minus.mt-grpo.humaneval-plus-abs}{5.0}
\setval{8b.base.minus.mt-grpo.if}{$-$30.5}
\setval{8b.base.minus.mt-grpo.if-abs}{30.5}
\setval{8b.base.minus.mt-grpo.ifbench}{$-$20.0}
\setval{8b.base.minus.mt-grpo.ifbench-abs}{20.0}
\setval{8b.base.minus.mt-grpo.ifeval}{$-$33.9}
\setval{8b.base.minus.mt-grpo.ifeval-abs}{33.9}
\setval{8b.base.minus.mt-grpo.judge2.avg}{$-$14.2}
\setval{8b.base.minus.mt-grpo.judge2.avg-abs}{14.2}
\setval{8b.base.minus.mt-grpo.judge2.creative-writing-v3}{$-$15.1}
\setval{8b.base.minus.mt-grpo.judge2.creative-writing-v3-abs}{15.1}
\setval{8b.base.minus.mt-grpo.judge2.longbench-write}{$-$14.4}
\setval{8b.base.minus.mt-grpo.judge2.longbench-write-abs}{14.4}
\setval{8b.base.minus.mt-grpo.judge2.writing}{$-$17.0}
\setval{8b.base.minus.mt-grpo.judge2.writing-abs}{17.0}
\setval{8b.base.minus.mt-grpo.judge2.writingbench}{$-$21.5}
\setval{8b.base.minus.mt-grpo.judge2.writingbench-abs}{21.5}
\setval{8b.base.minus.mt-grpo.livecodebench}{$-$6.7}
\setval{8b.base.minus.mt-grpo.livecodebench-abs}{6.7}
\setval{8b.base.minus.mt-grpo.longbench-write}{$-$16.0}
\setval{8b.base.minus.mt-grpo.longbench-write-abs}{16.0}
\setval{8b.base.minus.mt-grpo.math}{$-$8.4}
\setval{8b.base.minus.mt-grpo.math-abs}{8.4}
\setval{8b.base.minus.mt-grpo.math500}{$-$8.1}
\setval{8b.base.minus.mt-grpo.math500-abs}{8.1}
\setval{8b.base.minus.mt-grpo.mbpp-plus}{$-$8.0}
\setval{8b.base.minus.mt-grpo.mbpp-plus-abs}{8.0}
\setval{8b.base.minus.mt-grpo.minerva}{$-$10.1}
\setval{8b.base.minus.mt-grpo.minerva-abs}{10.1}
\setval{8b.base.minus.mt-grpo.olympiad-bench}{$-$6.3}
\setval{8b.base.minus.mt-grpo.olympiad-bench-abs}{6.3}
\setval{8b.base.minus.mt-grpo.writing}{$-$24.9}
\setval{8b.base.minus.mt-grpo.writing-abs}{24.9}
\setval{8b.base.minus.mt-grpo.writingbench}{$-$35.0}
\setval{8b.base.minus.mt-grpo.writingbench-abs}{35.0}
\setval{8b.base.minus.papo.aime24}{$-$0.8}
\setval{8b.base.minus.papo.aime24-abs}{0.8}
\setval{8b.base.minus.papo.aime25}{$-$1.6}
\setval{8b.base.minus.papo.aime25-abs}{1.6}
\setval{8b.base.minus.papo.amc23}{$-$11.8}
\setval{8b.base.minus.papo.amc23-abs}{11.8}
\setval{8b.base.minus.papo.avg}{$-$14.4}
\setval{8b.base.minus.papo.avg-abs}{14.4}
\setval{8b.base.minus.papo.code}{$-$4.8}
\setval{8b.base.minus.papo.code-abs}{4.8}
\setval{8b.base.minus.papo.creative-writing-v3}{$-$24.4}
\setval{8b.base.minus.papo.creative-writing-v3-abs}{24.4}
\setval{8b.base.minus.papo.guidebench}{$-$38.8}
\setval{8b.base.minus.papo.guidebench-abs}{38.8}
\setval{8b.base.minus.papo.humaneval-plus}{$-$2.4}
\setval{8b.base.minus.papo.humaneval-plus-abs}{2.4}
\setval{8b.base.minus.papo.if}{$-$30.7}
\setval{8b.base.minus.papo.if-abs}{30.7}
\setval{8b.base.minus.papo.ifbench}{$-$20.0}
\setval{8b.base.minus.papo.ifbench-abs}{20.0}
\setval{8b.base.minus.papo.ifeval}{$-$33.3}
\setval{8b.base.minus.papo.ifeval-abs}{33.3}
\setval{8b.base.minus.papo.judge2.avg}{$-$12.6}
\setval{8b.base.minus.papo.judge2.avg-abs}{12.6}
\setval{8b.base.minus.papo.judge2.creative-writing-v3}{$-$15.4}
\setval{8b.base.minus.papo.judge2.creative-writing-v3-abs}{15.4}
\setval{8b.base.minus.papo.judge2.longbench-write}{$-$11.5}
\setval{8b.base.minus.papo.judge2.longbench-write-abs}{11.5}
\setval{8b.base.minus.papo.judge2.writing}{$-$16.4}
\setval{8b.base.minus.papo.judge2.writing-abs}{16.4}
\setval{8b.base.minus.papo.judge2.writingbench}{$-$22.3}
\setval{8b.base.minus.papo.judge2.writingbench-abs}{22.3}
\setval{8b.base.minus.papo.livecodebench}{$-$5.7}
\setval{8b.base.minus.papo.livecodebench-abs}{5.7}
\setval{8b.base.minus.papo.longbench-write}{$-$14.1}
\setval{8b.base.minus.papo.longbench-write-abs}{14.1}
\setval{8b.base.minus.papo.math}{$-$5.7}
\setval{8b.base.minus.papo.math-abs}{5.7}
\setval{8b.base.minus.papo.math500}{$-$6.0}
\setval{8b.base.minus.papo.math500-abs}{6.0}
\setval{8b.base.minus.papo.mbpp-plus}{$-$6.1}
\setval{8b.base.minus.papo.mbpp-plus-abs}{6.1}
\setval{8b.base.minus.papo.minerva}{$-$8.4}
\setval{8b.base.minus.papo.minerva-abs}{8.4}
\setval{8b.base.minus.papo.olympiad-bench}{$-$5.4}
\setval{8b.base.minus.papo.olympiad-bench-abs}{5.4}
\setval{8b.base.minus.papo.writing}{$-$25.0}
\setval{8b.base.minus.papo.writing-abs}{25.0}
\setval{8b.base.minus.papo.writingbench}{$-$36.5}
\setval{8b.base.minus.papo.writingbench-abs}{36.5}
\setval{8b.base.minus.saw.aime24}{$-$4.0}
\setval{8b.base.minus.saw.aime24-abs}{4.0}
\setval{8b.base.minus.saw.aime25}{$-$7.0}
\setval{8b.base.minus.saw.aime25-abs}{7.0}
\setval{8b.base.minus.saw.amc23}{$-$12.0}
\setval{8b.base.minus.saw.amc23-abs}{12.0}
\setval{8b.base.minus.saw.avg}{$-$16.5}
\setval{8b.base.minus.saw.avg-abs}{16.5}
\setval{8b.base.minus.saw.code}{$-$6.2}
\setval{8b.base.minus.saw.code-abs}{6.2}
\setval{8b.base.minus.saw.creative-writing-v3}{$-$23.0}
\setval{8b.base.minus.saw.creative-writing-v3-abs}{23.0}
\setval{8b.base.minus.saw.guidebench}{$-$38.7}
\setval{8b.base.minus.saw.guidebench-abs}{38.7}
\setval{8b.base.minus.saw.humaneval-plus}{$-$5.8}
\setval{8b.base.minus.saw.humaneval-plus-abs}{5.8}
\setval{8b.base.minus.saw.if}{$-$32.0}
\setval{8b.base.minus.saw.if-abs}{32.0}
\setval{8b.base.minus.saw.ifbench}{$-$22.7}
\setval{8b.base.minus.saw.ifbench-abs}{22.7}
\setval{8b.base.minus.saw.ifeval}{$-$34.7}
\setval{8b.base.minus.saw.ifeval-abs}{34.7}
\setval{8b.base.minus.saw.judge2.avg}{$-$14.4}
\setval{8b.base.minus.saw.judge2.avg-abs}{14.4}
\setval{8b.base.minus.saw.judge2.creative-writing-v3}{$-$10.1}
\setval{8b.base.minus.saw.judge2.creative-writing-v3-abs}{10.1}
\setval{8b.base.minus.saw.judge2.longbench-write}{$-$16.8}
\setval{8b.base.minus.saw.judge2.longbench-write-abs}{16.8}
\setval{8b.base.minus.saw.judge2.writing}{$-$16.6}
\setval{8b.base.minus.saw.judge2.writing-abs}{16.6}
\setval{8b.base.minus.saw.judge2.writingbench}{$-$22.9}
\setval{8b.base.minus.saw.judge2.writingbench-abs}{22.9}
\setval{8b.base.minus.saw.livecodebench}{$-$5.8}
\setval{8b.base.minus.saw.livecodebench-abs}{5.8}
\setval{8b.base.minus.saw.longbench-write}{$-$19.6}
\setval{8b.base.minus.saw.longbench-write-abs}{19.6}
\setval{8b.base.minus.saw.math}{$-$8.7}
\setval{8b.base.minus.saw.math-abs}{8.7}
\setval{8b.base.minus.saw.math500}{$-$9.5}
\setval{8b.base.minus.saw.math500-abs}{9.5}
\setval{8b.base.minus.saw.mbpp-plus}{$-$7.1}
\setval{8b.base.minus.saw.mbpp-plus-abs}{7.1}
\setval{8b.base.minus.saw.minerva}{$-$10.9}
\setval{8b.base.minus.saw.minerva-abs}{10.9}
\setval{8b.base.minus.saw.olympiad-bench}{$-$8.7}
\setval{8b.base.minus.saw.olympiad-bench-abs}{8.7}
\setval{8b.base.minus.saw.writing}{$-$27.1}
\setval{8b.base.minus.saw.writing-abs}{27.1}
\setval{8b.base.minus.saw.writingbench}{$-$38.6}
\setval{8b.base.minus.saw.writingbench-abs}{38.6}
\setval{8b.concur-free.aux-change.code}{+0.18}
\setval{8b.concur-free.aux-change.if}{+0.18}
\setval{8b.concur-free.aux-change.math}{+0.13}
\setval{8b.concur-free.aux-change.writing}{$-$0.42}
\setval{8b.concur-free.aux-change.writing-rm}{+0.31}
\setval{8b.concur-free.aux-full.code.early}{0.68}
\setval{8b.concur-free.aux-full.code.late}{0.96}
\setval{8b.concur-free.aux-full.code.run}{0.93}
\setval{8b.concur-free.aux-full.if.early}{0.69}
\setval{8b.concur-free.aux-full.if.late}{0.93}
\setval{8b.concur-free.aux-full.if.run}{0.88}
\setval{8b.concur-free.aux-full.math.early}{0.55}
\setval{8b.concur-free.aux-full.math.late}{0.87}
\setval{8b.concur-free.aux-full.math.run}{0.81}
\setval{8b.concur-free.aux-full.writing-rm.early}{0.07}
\setval{8b.concur-free.aux-full.writing-rm.late}{0.48}
\setval{8b.concur-free.aux-full.writing-rm.run}{0.41}
\setval{8b.concur-free.aux-full.writing.early}{0.61}
\setval{8b.concur-free.aux-full.writing.late}{0.21}
\setval{8b.concur-free.aux-full.writing.run}{0.24}
\setval{8b.concur-free.aux-level.code.early}{0.92}
\setval{8b.concur-free.aux-level.code.late}{0.99}
\setval{8b.concur-free.aux-level.code.run}{0.98}
\setval{8b.concur-free.aux-level.if.early}{0.86}
\setval{8b.concur-free.aux-level.if.late}{0.97}
\setval{8b.concur-free.aux-level.if.run}{0.94}
\setval{8b.concur-free.aux-level.math.early}{0.90}
\setval{8b.concur-free.aux-level.math.late}{0.97}
\setval{8b.concur-free.aux-level.math.run}{0.96}
\setval{8b.concur-free.aux-level.writing-rm.early}{0.77}
\setval{8b.concur-free.aux-level.writing-rm.late}{0.95}
\setval{8b.concur-free.aux-level.writing-rm.run}{0.92}
\setval{8b.concur-free.aux-level.writing.early}{0.88}
\setval{8b.concur-free.aux-level.writing.late}{0.42}
\setval{8b.concur-free.aux-level.writing.run}{0.47}
\setval{8b.concur-free.bench.aime24}{19.7}
\setval{8b.concur-free.bench.aime25}{16.6}
\setval{8b.concur-free.bench.amc23}{64.7}
\setval{8b.concur-free.bench.creative-writing-v3}{51.6}
\setval{8b.concur-free.bench.guidebench}{82.9}
\setval{8b.concur-free.bench.humaneval-plus}{85.3}
\setval{8b.concur-free.bench.ifbench}{38.7}
\setval{8b.concur-free.bench.ifeval}{89.2}
\setval{8b.concur-free.bench.livecodebench}{27.1}
\setval{8b.concur-free.bench.longbench-write}{83.0}
\setval{8b.concur-free.bench.math500}{83.4}
\setval{8b.concur-free.bench.mbpp-plus}{75.1}
\setval{8b.concur-free.bench.minerva}{49.8}
\setval{8b.concur-free.bench.olympiad-bench}{50.0}
\setval{8b.concur-free.bench.writingbench}{86.8}
\setval{8b.concur-free.eval.avg}{60.3}
\setval{8b.concur-free.eval.code}{62.5}
\setval{8b.concur-free.eval.if}{70.3}
\setval{8b.concur-free.eval.math}{47.4}
\setval{8b.concur-free.eval.writing}{73.8}
\setval{8b.concur-free.gate.code.against.early}{0.04}
\setval{8b.concur-free.gate.code.against.late}{0.02}
\setval{8b.concur-free.gate.code.against.run}{0.03}
\setval{8b.concur-free.gate.code.agreement}{+0.006}
\setval{8b.concur-free.gate.code.ceiling.early}{0.13}
\setval{8b.concur-free.gate.code.ceiling.late}{0.03}
\setval{8b.concur-free.gate.code.ceiling.run}{0.05}
\setval{8b.concur-free.gate.code.cosine.early}{0.82}
\setval{8b.concur-free.gate.code.cosine.late}{0.78}
\setval{8b.concur-free.gate.code.cosine.run}{0.81}
\setval{8b.concur-free.gate.if.against.early}{0.05}
\setval{8b.concur-free.gate.if.against.late}{0.04}
\setval{8b.concur-free.gate.if.against.run}{0.04}
\setval{8b.concur-free.gate.if.agreement}{+0.016}
\setval{8b.concur-free.gate.if.ceiling.early}{0.06}
\setval{8b.concur-free.gate.if.ceiling.late}{0.14}
\setval{8b.concur-free.gate.if.ceiling.run}{0.12}
\setval{8b.concur-free.gate.if.cosine.early}{0.74}
\setval{8b.concur-free.gate.if.cosine.late}{0.54}
\setval{8b.concur-free.gate.if.cosine.run}{0.61}
\setval{8b.concur-free.gate.math.against.early}{0.00}
\setval{8b.concur-free.gate.math.against.late}{0.00}
\setval{8b.concur-free.gate.math.against.run}{0.00}
\setval{8b.concur-free.gate.math.ceiling.early}{0.24}
\setval{8b.concur-free.gate.math.ceiling.late}{0.20}
\setval{8b.concur-free.gate.math.ceiling.run}{0.20}
\setval{8b.concur-free.gate.math.cosine.early}{0.84}
\setval{8b.concur-free.gate.math.cosine.late}{0.88}
\setval{8b.concur-free.gate.math.cosine.run}{0.87}
\setval{8b.concur-free.gate.writing-rm.against.early}{0.20}
\setval{8b.concur-free.gate.writing-rm.against.late}{0.57}
\setval{8b.concur-free.gate.writing-rm.against.run}{0.53}
\setval{8b.concur-free.gate.writing-rm.agreement}{+0.049}
\setval{8b.concur-free.gate.writing-rm.ceiling.early}{0.00}
\setval{8b.concur-free.gate.writing-rm.ceiling.late}{0.01}
\setval{8b.concur-free.gate.writing-rm.ceiling.run}{0.00}
\setval{8b.concur-free.gate.writing-rm.cosine.early}{0.69}
\setval{8b.concur-free.gate.writing-rm.cosine.late}{0.08}
\setval{8b.concur-free.gate.writing-rm.cosine.run}{0.17}
\setval{8b.concur-free.gate.writing.against.early}{0.32}
\setval{8b.concur-free.gate.writing.against.late}{0.65}
\setval{8b.concur-free.gate.writing.against.run}{0.64}
\setval{8b.concur-free.gate.writing.agreement}{$-$0.045}
\setval{8b.concur-free.gate.writing.ceiling.early}{0.00}
\setval{8b.concur-free.gate.writing.ceiling.late}{0.01}
\setval{8b.concur-free.gate.writing.ceiling.run}{0.00}
\setval{8b.concur-free.gate.writing.cosine.early}{0.49}
\setval{8b.concur-free.gate.writing.cosine.late}{-0.03}
\setval{8b.concur-free.gate.writing.cosine.run}{0.01}
\setval{8b.concur-free.judge2.avg}{57.3}
\setval{8b.concur-free.judge2.creative-writing-v3}{42.5}
\setval{8b.concur-free.judge2.longbench-write}{74.1}
\setval{8b.concur-free.judge2.writing}{59.1}
\setval{8b.concur-free.judge2.writingbench}{60.7}
\setval{8b.concur-free.last-step}{300}
\setval{8b.concur-free.minus.base.aime24}{+8.3}
\setval{8b.concur-free.minus.base.aime24-abs}{8.3}
\setval{8b.concur-free.minus.base.aime25}{+8.0}
\setval{8b.concur-free.minus.base.aime25-abs}{8.0}
\setval{8b.concur-free.minus.base.amc23}{+14.5}
\setval{8b.concur-free.minus.base.amc23-abs}{14.5}
\setval{8b.concur-free.minus.base.avg}{+18.8}
\setval{8b.concur-free.minus.base.avg-abs}{18.8}
\setval{8b.concur-free.minus.base.code}{+6.4}
\setval{8b.concur-free.minus.base.code-abs}{6.4}
\setval{8b.concur-free.minus.base.creative-writing-v3}{+24.9}
\setval{8b.concur-free.minus.base.creative-writing-v3-abs}{24.9}
\setval{8b.concur-free.minus.base.guidebench}{+37.3}
\setval{8b.concur-free.minus.base.guidebench-abs}{37.3}
\setval{8b.concur-free.minus.base.humaneval-plus}{+5.2}
\setval{8b.concur-free.minus.base.humaneval-plus-abs}{5.2}
\setval{8b.concur-free.minus.base.if}{+31.6}
\setval{8b.concur-free.minus.base.if-abs}{31.6}
\setval{8b.concur-free.minus.base.ifbench}{+21.3}
\setval{8b.concur-free.minus.base.ifbench-abs}{21.3}
\setval{8b.concur-free.minus.base.ifeval}{+36.2}
\setval{8b.concur-free.minus.base.ifeval-abs}{36.2}
\setval{8b.concur-free.minus.base.judge2.avg}{+16.3}
\setval{8b.concur-free.minus.base.judge2.avg-abs}{16.3}
\setval{8b.concur-free.minus.base.judge2.creative-writing-v3}{+16.1}
\setval{8b.concur-free.minus.base.judge2.creative-writing-v3-abs}{16.1}
\setval{8b.concur-free.minus.base.judge2.longbench-write}{+26.0}
\setval{8b.concur-free.minus.base.judge2.longbench-write-abs}{26.0}
\setval{8b.concur-free.minus.base.judge2.writing}{+22.9}
\setval{8b.concur-free.minus.base.judge2.writing-abs}{22.9}
\setval{8b.concur-free.minus.base.judge2.writingbench}{+26.7}
\setval{8b.concur-free.minus.base.judge2.writingbench-abs}{26.7}
\setval{8b.concur-free.minus.base.livecodebench}{+6.2}
\setval{8b.concur-free.minus.base.livecodebench-abs}{6.2}
\setval{8b.concur-free.minus.base.longbench-write}{+31.5}
\setval{8b.concur-free.minus.base.longbench-write-abs}{31.5}
\setval{8b.concur-free.minus.base.math}{+10.2}
\setval{8b.concur-free.minus.base.math-abs}{10.2}
\setval{8b.concur-free.minus.base.math500}{+11.0}
\setval{8b.concur-free.minus.base.math500-abs}{11.0}
\setval{8b.concur-free.minus.base.mbpp-plus}{+7.8}
\setval{8b.concur-free.minus.base.mbpp-plus-abs}{7.8}
\setval{8b.concur-free.minus.base.minerva}{+8.9}
\setval{8b.concur-free.minus.base.minerva-abs}{8.9}
\setval{8b.concur-free.minus.base.olympiad-bench}{+10.6}
\setval{8b.concur-free.minus.base.olympiad-bench-abs}{10.6}
\setval{8b.concur-free.minus.base.writing}{+35.5}
\setval{8b.concur-free.minus.base.writing-abs}{35.5}
\setval{8b.concur-free.minus.base.writingbench}{+50.0}
\setval{8b.concur-free.minus.base.writingbench-abs}{50.0}
\setval{8b.concur-free.minus.concur-tied.aime24}{$-$11.6}
\setval{8b.concur-free.minus.concur-tied.aime24-abs}{11.6}
\setval{8b.concur-free.minus.concur-tied.aime25}{$-$7.4}
\setval{8b.concur-free.minus.concur-tied.aime25-abs}{7.4}
\setval{8b.concur-free.minus.concur-tied.amc23}{$-$13.8}
\setval{8b.concur-free.minus.concur-tied.amc23-abs}{13.8}
\setval{8b.concur-free.minus.concur-tied.avg}{$-$3.8}
\setval{8b.concur-free.minus.concur-tied.avg-abs}{3.8}
\setval{8b.concur-free.minus.concur-tied.code}{$-$0.6}
\setval{8b.concur-free.minus.concur-tied.code-abs}{0.6}
\setval{8b.concur-free.minus.concur-tied.creative-writing-v3}{$-$0.8}
\setval{8b.concur-free.minus.concur-tied.creative-writing-v3-abs}{0.8}
\setval{8b.concur-free.minus.concur-tied.guidebench}{$-$0.5}
\setval{8b.concur-free.minus.concur-tied.guidebench-abs}{0.5}
\setval{8b.concur-free.minus.concur-tied.humaneval-plus}{$-$2.1}
\setval{8b.concur-free.minus.concur-tied.humaneval-plus-abs}{2.1}
\setval{8b.concur-free.minus.concur-tied.if}{$-$0.3}
\setval{8b.concur-free.minus.concur-tied.if-abs}{0.3}
\setval{8b.concur-free.minus.concur-tied.ifbench}{$-$0.7}
\setval{8b.concur-free.minus.concur-tied.ifbench-abs}{0.7}
\setval{8b.concur-free.minus.concur-tied.ifeval}{+0.4}
\setval{8b.concur-free.minus.concur-tied.ifeval-abs}{0.4}
\setval{8b.concur-free.minus.concur-tied.judge2.avg}{$-$4.0}
\setval{8b.concur-free.minus.concur-tied.judge2.avg-abs}{4.0}
\setval{8b.concur-free.minus.concur-tied.judge2.creative-writing-v3}{+0.7}
\setval{8b.concur-free.minus.concur-tied.judge2.creative-writing-v3-abs}{0.7}
\setval{8b.concur-free.minus.concur-tied.judge2.longbench-write}{$-$0.7}
\setval{8b.concur-free.minus.concur-tied.judge2.longbench-write-abs}{0.7}
\setval{8b.concur-free.minus.concur-tied.judge2.writing}{$-$1.0}
\setval{8b.concur-free.minus.concur-tied.judge2.writing-abs}{1.0}
\setval{8b.concur-free.minus.concur-tied.judge2.writingbench}{$-$3.1}
\setval{8b.concur-free.minus.concur-tied.judge2.writingbench-abs}{3.1}
\setval{8b.concur-free.minus.concur-tied.livecodebench}{$-$0.1}
\setval{8b.concur-free.minus.concur-tied.livecodebench-abs}{0.1}
\setval{8b.concur-free.minus.concur-tied.longbench-write}{+1.8}
\setval{8b.concur-free.minus.concur-tied.longbench-write-abs}{1.8}
\setval{8b.concur-free.minus.concur-tied.math}{$-$9.0}
\setval{8b.concur-free.minus.concur-tied.math-abs}{9.0}
\setval{8b.concur-free.minus.concur-tied.math500}{$-$6.2}
\setval{8b.concur-free.minus.concur-tied.math500-abs}{6.2}
\setval{8b.concur-free.minus.concur-tied.mbpp-plus}{+0.2}
\setval{8b.concur-free.minus.concur-tied.mbpp-plus-abs}{0.2}
\setval{8b.concur-free.minus.concur-tied.minerva}{$-$5.6}
\setval{8b.concur-free.minus.concur-tied.minerva-abs}{5.6}
\setval{8b.concur-free.minus.concur-tied.olympiad-bench}{$-$9.5}
\setval{8b.concur-free.minus.concur-tied.olympiad-bench-abs}{9.5}
\setval{8b.concur-free.minus.concur-tied.writing}{+0.2}
\setval{8b.concur-free.minus.concur-tied.writing-abs}{0.2}
\setval{8b.concur-free.minus.concur-tied.writingbench}{$-$0.4}
\setval{8b.concur-free.minus.concur-tied.writingbench-abs}{0.4}
\setval{8b.concur-free.minus.concur.aime24}{$-$9.1}
\setval{8b.concur-free.minus.concur.aime24-abs}{9.1}
\setval{8b.concur-free.minus.concur.aime25}{$-$6.6}
\setval{8b.concur-free.minus.concur.aime25-abs}{6.6}
\setval{8b.concur-free.minus.concur.amc23}{$-$14.2}
\setval{8b.concur-free.minus.concur.amc23-abs}{14.2}
\setval{8b.concur-free.minus.concur.avg}{$-$2.9}
\setval{8b.concur-free.minus.concur.avg-abs}{2.9}
\setval{8b.concur-free.minus.concur.code}{+1.7}
\setval{8b.concur-free.minus.concur.code-abs}{1.7}
\setval{8b.concur-free.minus.concur.creative-writing-v3}{+1.7}
\setval{8b.concur-free.minus.concur.creative-writing-v3-abs}{1.7}
\setval{8b.concur-free.minus.concur.guidebench}{$-$0.5}
\setval{8b.concur-free.minus.concur.guidebench-abs}{0.5}
\setval{8b.concur-free.minus.concur.humaneval-plus}{+2.6}
\setval{8b.concur-free.minus.concur.humaneval-plus-abs}{2.6}
\setval{8b.concur-free.minus.concur.if}{+0.2}
\setval{8b.concur-free.minus.concur.if-abs}{0.2}
\setval{8b.concur-free.minus.concur.ifbench}{+0.0}
\setval{8b.concur-free.minus.concur.ifbench-abs}{0.0}
\setval{8b.concur-free.minus.concur.ifeval}{+1.1}
\setval{8b.concur-free.minus.concur.ifeval-abs}{1.1}
\setval{8b.concur-free.minus.concur.judge2.avg}{$-$3.0}
\setval{8b.concur-free.minus.concur.judge2.avg-abs}{3.0}
\setval{8b.concur-free.minus.concur.judge2.creative-writing-v3}{+4.8}
\setval{8b.concur-free.minus.concur.judge2.creative-writing-v3-abs}{4.8}
\setval{8b.concur-free.minus.concur.judge2.longbench-write}{$-$1.4}
\setval{8b.concur-free.minus.concur.judge2.longbench-write-abs}{1.4}
\setval{8b.concur-free.minus.concur.judge2.writing}{+0.0}
\setval{8b.concur-free.minus.concur.judge2.writing-abs}{0.0}
\setval{8b.concur-free.minus.concur.judge2.writingbench}{$-$3.3}
\setval{8b.concur-free.minus.concur.judge2.writingbench-abs}{3.3}
\setval{8b.concur-free.minus.concur.livecodebench}{+0.7}
\setval{8b.concur-free.minus.concur.livecodebench-abs}{0.7}
\setval{8b.concur-free.minus.concur.longbench-write}{+1.4}
\setval{8b.concur-free.minus.concur.longbench-write-abs}{1.4}
\setval{8b.concur-free.minus.concur.math}{$-$8.5}
\setval{8b.concur-free.minus.concur.math-abs}{8.5}
\setval{8b.concur-free.minus.concur.math500}{$-$6.7}
\setval{8b.concur-free.minus.concur.math500-abs}{6.7}
\setval{8b.concur-free.minus.concur.mbpp-plus}{+1.9}
\setval{8b.concur-free.minus.concur.mbpp-plus-abs}{1.9}
\setval{8b.concur-free.minus.concur.minerva}{$-$4.4}
\setval{8b.concur-free.minus.concur.minerva-abs}{4.4}
\setval{8b.concur-free.minus.concur.olympiad-bench}{$-$10.3}
\setval{8b.concur-free.minus.concur.olympiad-bench-abs}{10.3}
\setval{8b.concur-free.minus.concur.writing}{+0.6}
\setval{8b.concur-free.minus.concur.writing-abs}{0.6}
\setval{8b.concur-free.minus.concur.writingbench}{$-$1.3}
\setval{8b.concur-free.minus.concur.writingbench-abs}{1.3}
\setval{8b.concur-free.minus.gd2po.aime24}{+3.4}
\setval{8b.concur-free.minus.gd2po.aime24-abs}{3.4}
\setval{8b.concur-free.minus.gd2po.aime25}{+4.4}
\setval{8b.concur-free.minus.gd2po.aime25-abs}{4.4}
\setval{8b.concur-free.minus.gd2po.amc23}{+5.1}
\setval{8b.concur-free.minus.gd2po.amc23-abs}{5.1}
\setval{8b.concur-free.minus.gd2po.avg}{+2.9}
\setval{8b.concur-free.minus.gd2po.avg-abs}{2.9}
\setval{8b.concur-free.minus.gd2po.code}{+0.7}
\setval{8b.concur-free.minus.gd2po.code-abs}{0.7}
\setval{8b.concur-free.minus.gd2po.creative-writing-v3}{+3.5}
\setval{8b.concur-free.minus.gd2po.creative-writing-v3-abs}{3.5}
\setval{8b.concur-free.minus.gd2po.guidebench}{$-$0.8}
\setval{8b.concur-free.minus.gd2po.guidebench-abs}{0.8}
\setval{8b.concur-free.minus.gd2po.humaneval-plus}{+1.3}
\setval{8b.concur-free.minus.gd2po.humaneval-plus-abs}{1.3}
\setval{8b.concur-free.minus.gd2po.if}{$-$1.7}
\setval{8b.concur-free.minus.gd2po.if-abs}{1.7}
\setval{8b.concur-free.minus.gd2po.ifbench}{$-$4.3}
\setval{8b.concur-free.minus.gd2po.ifbench-abs}{4.3}
\setval{8b.concur-free.minus.gd2po.ifeval}{$-$0.1}
\setval{8b.concur-free.minus.gd2po.ifeval-abs}{0.1}
\setval{8b.concur-free.minus.gd2po.judge2.avg}{+2.4}
\setval{8b.concur-free.minus.gd2po.judge2.avg-abs}{2.4}
\setval{8b.concur-free.minus.gd2po.judge2.creative-writing-v3}{+5.8}
\setval{8b.concur-free.minus.gd2po.judge2.creative-writing-v3-abs}{5.8}
\setval{8b.concur-free.minus.gd2po.judge2.longbench-write}{+11.9}
\setval{8b.concur-free.minus.gd2po.judge2.longbench-write-abs}{11.9}
\setval{8b.concur-free.minus.gd2po.judge2.writing}{+7.7}
\setval{8b.concur-free.minus.gd2po.judge2.writing-abs}{7.7}
\setval{8b.concur-free.minus.gd2po.judge2.writingbench}{+5.4}
\setval{8b.concur-free.minus.gd2po.judge2.writingbench-abs}{5.4}
\setval{8b.concur-free.minus.gd2po.livecodebench}{+0.8}
\setval{8b.concur-free.minus.gd2po.livecodebench-abs}{0.8}
\setval{8b.concur-free.minus.gd2po.longbench-write}{+13.1}
\setval{8b.concur-free.minus.gd2po.longbench-write-abs}{13.1}
\setval{8b.concur-free.minus.gd2po.math}{+2.7}
\setval{8b.concur-free.minus.gd2po.math-abs}{2.7}
\setval{8b.concur-free.minus.gd2po.math500}{+2.2}
\setval{8b.concur-free.minus.gd2po.math500-abs}{2.2}
\setval{8b.concur-free.minus.gd2po.mbpp-plus}{+0.0}
\setval{8b.concur-free.minus.gd2po.mbpp-plus-abs}{0.0}
\setval{8b.concur-free.minus.gd2po.minerva}{$-$0.1}
\setval{8b.concur-free.minus.gd2po.minerva-abs}{0.1}
\setval{8b.concur-free.minus.gd2po.olympiad-bench}{+1.5}
\setval{8b.concur-free.minus.gd2po.olympiad-bench-abs}{1.5}
\setval{8b.concur-free.minus.gd2po.writing}{+9.8}
\setval{8b.concur-free.minus.gd2po.writing-abs}{9.8}
\setval{8b.concur-free.minus.gd2po.writingbench}{+12.9}
\setval{8b.concur-free.minus.gd2po.writingbench-abs}{12.9}
\setval{8b.concur-free.minus.gdpo-noaux.aime24}{$-$8.2}
\setval{8b.concur-free.minus.gdpo-noaux.aime24-abs}{8.2}
\setval{8b.concur-free.minus.gdpo-noaux.aime25}{$-$7.4}
\setval{8b.concur-free.minus.gdpo-noaux.aime25-abs}{7.4}
\setval{8b.concur-free.minus.gdpo-noaux.amc23}{$-$12.4}
\setval{8b.concur-free.minus.gdpo-noaux.amc23-abs}{12.4}
\setval{8b.concur-free.minus.gdpo-noaux.avg}{$-$0.3}
\setval{8b.concur-free.minus.gdpo-noaux.avg-abs}{0.3}
\setval{8b.concur-free.minus.gdpo-noaux.code}{$-$0.2}
\setval{8b.concur-free.minus.gdpo-noaux.code-abs}{0.2}
\setval{8b.concur-free.minus.gdpo-noaux.creative-writing-v3}{+7.8}
\setval{8b.concur-free.minus.gdpo-noaux.creative-writing-v3-abs}{7.8}
\setval{8b.concur-free.minus.gdpo-noaux.guidebench}{+1.5}
\setval{8b.concur-free.minus.gdpo-noaux.guidebench-abs}{1.5}
\setval{8b.concur-free.minus.gdpo-noaux.humaneval-plus}{$-$0.7}
\setval{8b.concur-free.minus.gdpo-noaux.humaneval-plus-abs}{0.7}
\setval{8b.concur-free.minus.gdpo-noaux.if}{+3.1}
\setval{8b.concur-free.minus.gdpo-noaux.if-abs}{3.1}
\setval{8b.concur-free.minus.gdpo-noaux.ifbench}{+5.0}
\setval{8b.concur-free.minus.gdpo-noaux.ifbench-abs}{5.0}
\setval{8b.concur-free.minus.gdpo-noaux.ifeval}{+2.6}
\setval{8b.concur-free.minus.gdpo-noaux.ifeval-abs}{2.6}
\setval{8b.concur-free.minus.gdpo-noaux.judge2.avg}{+0.2}
\setval{8b.concur-free.minus.gdpo-noaux.judge2.avg-abs}{0.2}
\setval{8b.concur-free.minus.gdpo-noaux.judge2.creative-writing-v3}{+13.6}
\setval{8b.concur-free.minus.gdpo-noaux.judge2.creative-writing-v3-abs}{13.6}
\setval{8b.concur-free.minus.gdpo-noaux.judge2.longbench-write}{+19.6}
\setval{8b.concur-free.minus.gdpo-noaux.judge2.longbench-write-abs}{19.6}
\setval{8b.concur-free.minus.gdpo-noaux.judge2.writing}{+14.6}
\setval{8b.concur-free.minus.gdpo-noaux.judge2.writing-abs}{14.6}
\setval{8b.concur-free.minus.gdpo-noaux.judge2.writingbench}{+10.6}
\setval{8b.concur-free.minus.gdpo-noaux.judge2.writingbench-abs}{10.6}
\setval{8b.concur-free.minus.gdpo-noaux.livecodebench}{$-$0.3}
\setval{8b.concur-free.minus.gdpo-noaux.livecodebench-abs}{0.3}
\setval{8b.concur-free.minus.gdpo-noaux.longbench-write}{+17.0}
\setval{8b.concur-free.minus.gdpo-noaux.longbench-write-abs}{17.0}
\setval{8b.concur-free.minus.gdpo-noaux.math}{$-$8.2}
\setval{8b.concur-free.minus.gdpo-noaux.math-abs}{8.2}
\setval{8b.concur-free.minus.gdpo-noaux.math500}{$-$6.8}
\setval{8b.concur-free.minus.gdpo-noaux.math500-abs}{6.8}
\setval{8b.concur-free.minus.gdpo-noaux.mbpp-plus}{+0.2}
\setval{8b.concur-free.minus.gdpo-noaux.mbpp-plus-abs}{0.2}
\setval{8b.concur-free.minus.gdpo-noaux.minerva}{$-$4.2}
\setval{8b.concur-free.minus.gdpo-noaux.minerva-abs}{4.2}
\setval{8b.concur-free.minus.gdpo-noaux.olympiad-bench}{$-$9.9}
\setval{8b.concur-free.minus.gdpo-noaux.olympiad-bench-abs}{9.9}
\setval{8b.concur-free.minus.gdpo-noaux.writing}{+12.2}
\setval{8b.concur-free.minus.gdpo-noaux.writing-abs}{12.2}
\setval{8b.concur-free.minus.gdpo-noaux.writingbench}{+11.8}
\setval{8b.concur-free.minus.gdpo-noaux.writingbench-abs}{11.8}
\setval{8b.concur-free.minus.gdpo-w0.5.aime24}{+0.4}
\setval{8b.concur-free.minus.gdpo-w0.5.aime24-abs}{0.4}
\setval{8b.concur-free.minus.gdpo-w0.5.aime25}{+1.1}
\setval{8b.concur-free.minus.gdpo-w0.5.aime25-abs}{1.1}
\setval{8b.concur-free.minus.gdpo-w0.5.amc23}{+1.2}
\setval{8b.concur-free.minus.gdpo-w0.5.amc23-abs}{1.2}
\setval{8b.concur-free.minus.gdpo-w0.5.avg}{+1.9}
\setval{8b.concur-free.minus.gdpo-w0.5.avg-abs}{1.9}
\setval{8b.concur-free.minus.gdpo-w0.5.code}{+1.4}
\setval{8b.concur-free.minus.gdpo-w0.5.code-abs}{1.4}
\setval{8b.concur-free.minus.gdpo-w0.5.creative-writing-v3}{+0.1}
\setval{8b.concur-free.minus.gdpo-w0.5.creative-writing-v3-abs}{0.1}
\setval{8b.concur-free.minus.gdpo-w0.5.guidebench}{$-$1.1}
\setval{8b.concur-free.minus.gdpo-w0.5.guidebench-abs}{1.1}
\setval{8b.concur-free.minus.gdpo-w0.5.humaneval-plus}{+1.5}
\setval{8b.concur-free.minus.gdpo-w0.5.humaneval-plus-abs}{1.5}
\setval{8b.concur-free.minus.gdpo-w0.5.if}{+0.5}
\setval{8b.concur-free.minus.gdpo-w0.5.if-abs}{0.5}
\setval{8b.concur-free.minus.gdpo-w0.5.ifbench}{+1.0}
\setval{8b.concur-free.minus.gdpo-w0.5.ifbench-abs}{1.0}
\setval{8b.concur-free.minus.gdpo-w0.5.ifeval}{+1.4}
\setval{8b.concur-free.minus.gdpo-w0.5.ifeval-abs}{1.4}
\setval{8b.concur-free.minus.gdpo-w0.5.judge2.avg}{+1.5}
\setval{8b.concur-free.minus.gdpo-w0.5.judge2.avg-abs}{1.5}
\setval{8b.concur-free.minus.gdpo-w0.5.judge2.creative-writing-v3}{+2.1}
\setval{8b.concur-free.minus.gdpo-w0.5.judge2.creative-writing-v3-abs}{2.1}
\setval{8b.concur-free.minus.gdpo-w0.5.judge2.longbench-write}{+10.4}
\setval{8b.concur-free.minus.gdpo-w0.5.judge2.longbench-write-abs}{10.4}
\setval{8b.concur-free.minus.gdpo-w0.5.judge2.writing}{+5.7}
\setval{8b.concur-free.minus.gdpo-w0.5.judge2.writing-abs}{5.7}
\setval{8b.concur-free.minus.gdpo-w0.5.judge2.writingbench}{+4.4}
\setval{8b.concur-free.minus.gdpo-w0.5.judge2.writingbench-abs}{4.4}
\setval{8b.concur-free.minus.gdpo-w0.5.livecodebench}{+0.3}
\setval{8b.concur-free.minus.gdpo-w0.5.livecodebench-abs}{0.3}
\setval{8b.concur-free.minus.gdpo-w0.5.longbench-write}{+11.8}
\setval{8b.concur-free.minus.gdpo-w0.5.longbench-write-abs}{11.8}
\setval{8b.concur-free.minus.gdpo-w0.5.math}{$-$0.1}
\setval{8b.concur-free.minus.gdpo-w0.5.math-abs}{0.1}
\setval{8b.concur-free.minus.gdpo-w0.5.math500}{$-$0.6}
\setval{8b.concur-free.minus.gdpo-w0.5.math500-abs}{0.6}
\setval{8b.concur-free.minus.gdpo-w0.5.mbpp-plus}{+2.3}
\setval{8b.concur-free.minus.gdpo-w0.5.mbpp-plus-abs}{2.3}
\setval{8b.concur-free.minus.gdpo-w0.5.minerva}{$-$1.4}
\setval{8b.concur-free.minus.gdpo-w0.5.minerva-abs}{1.4}
\setval{8b.concur-free.minus.gdpo-w0.5.olympiad-bench}{$-$1.1}
\setval{8b.concur-free.minus.gdpo-w0.5.olympiad-bench-abs}{1.1}
\setval{8b.concur-free.minus.gdpo-w0.5.writing}{+7.9}
\setval{8b.concur-free.minus.gdpo-w0.5.writing-abs}{7.9}
\setval{8b.concur-free.minus.gdpo-w0.5.writingbench}{+11.7}
\setval{8b.concur-free.minus.gdpo-w0.5.writingbench-abs}{11.7}
\setval{8b.concur-free.minus.gdpo.aime24}{$-$1.7}
\setval{8b.concur-free.minus.gdpo.aime24-abs}{1.7}
\setval{8b.concur-free.minus.gdpo.aime25}{+2.2}
\setval{8b.concur-free.minus.gdpo.aime25-abs}{2.2}
\setval{8b.concur-free.minus.gdpo.amc23}{+2.5}
\setval{8b.concur-free.minus.gdpo.amc23-abs}{2.5}
\setval{8b.concur-free.minus.gdpo.avg}{+1.9}
\setval{8b.concur-free.minus.gdpo.avg-abs}{1.9}
\setval{8b.concur-free.minus.gdpo.code}{+0.9}
\setval{8b.concur-free.minus.gdpo.code-abs}{0.9}
\setval{8b.concur-free.minus.gdpo.creative-writing-v3}{$-$1.3}
\setval{8b.concur-free.minus.gdpo.creative-writing-v3-abs}{1.3}
\setval{8b.concur-free.minus.gdpo.guidebench}{$-$1.6}
\setval{8b.concur-free.minus.gdpo.guidebench-abs}{1.6}
\setval{8b.concur-free.minus.gdpo.humaneval-plus}{+1.2}
\setval{8b.concur-free.minus.gdpo.humaneval-plus-abs}{1.2}
\setval{8b.concur-free.minus.gdpo.if}{$-$0.4}
\setval{8b.concur-free.minus.gdpo.if-abs}{0.4}
\setval{8b.concur-free.minus.gdpo.ifbench}{$-$0.7}
\setval{8b.concur-free.minus.gdpo.ifbench-abs}{0.7}
\setval{8b.concur-free.minus.gdpo.ifeval}{+1.1}
\setval{8b.concur-free.minus.gdpo.ifeval-abs}{1.1}
\setval{8b.concur-free.minus.gdpo.judge2.avg}{+1.3}
\setval{8b.concur-free.minus.gdpo.judge2.avg-abs}{1.3}
\setval{8b.concur-free.minus.gdpo.judge2.creative-writing-v3}{+1.5}
\setval{8b.concur-free.minus.gdpo.judge2.creative-writing-v3-abs}{1.5}
\setval{8b.concur-free.minus.gdpo.judge2.longbench-write}{+9.9}
\setval{8b.concur-free.minus.gdpo.judge2.longbench-write-abs}{9.9}
\setval{8b.concur-free.minus.gdpo.judge2.writing}{+4.6}
\setval{8b.concur-free.minus.gdpo.judge2.writing-abs}{4.6}
\setval{8b.concur-free.minus.gdpo.judge2.writingbench}{+2.3}
\setval{8b.concur-free.minus.gdpo.judge2.writingbench-abs}{2.3}
\setval{8b.concur-free.minus.gdpo.livecodebench}{+0.9}
\setval{8b.concur-free.minus.gdpo.livecodebench-abs}{0.9}
\setval{8b.concur-free.minus.gdpo.longbench-write}{+13.4}
\setval{8b.concur-free.minus.gdpo.longbench-write-abs}{13.4}
\setval{8b.concur-free.minus.gdpo.math}{+0.7}
\setval{8b.concur-free.minus.gdpo.math-abs}{0.7}
\setval{8b.concur-free.minus.gdpo.math500}{+0.4}
\setval{8b.concur-free.minus.gdpo.math500-abs}{0.4}
\setval{8b.concur-free.minus.gdpo.mbpp-plus}{+0.6}
\setval{8b.concur-free.minus.gdpo.mbpp-plus-abs}{0.6}
\setval{8b.concur-free.minus.gdpo.minerva}{+0.2}
\setval{8b.concur-free.minus.gdpo.minerva-abs}{0.2}
\setval{8b.concur-free.minus.gdpo.olympiad-bench}{+0.7}
\setval{8b.concur-free.minus.gdpo.olympiad-bench-abs}{0.7}
\setval{8b.concur-free.minus.gdpo.writing}{+7.6}
\setval{8b.concur-free.minus.gdpo.writing-abs}{7.6}
\setval{8b.concur-free.minus.gdpo.writingbench}{+10.7}
\setval{8b.concur-free.minus.gdpo.writingbench-abs}{10.7}
\setval{8b.concur-free.minus.geoma.aime24}{+5.6}
\setval{8b.concur-free.minus.geoma.aime24-abs}{5.6}
\setval{8b.concur-free.minus.geoma.aime25}{+4.1}
\setval{8b.concur-free.minus.geoma.aime25-abs}{4.1}
\setval{8b.concur-free.minus.geoma.amc23}{+1.2}
\setval{8b.concur-free.minus.geoma.amc23-abs}{1.2}
\setval{8b.concur-free.minus.geoma.avg}{+3.6}
\setval{8b.concur-free.minus.geoma.avg-abs}{3.6}
\setval{8b.concur-free.minus.geoma.code}{+1.1}
\setval{8b.concur-free.minus.geoma.code-abs}{1.1}
\setval{8b.concur-free.minus.geoma.creative-writing-v3}{+1.2}
\setval{8b.concur-free.minus.geoma.creative-writing-v3-abs}{1.2}
\setval{8b.concur-free.minus.geoma.guidebench}{$-$1.4}
\setval{8b.concur-free.minus.geoma.guidebench-abs}{1.4}
\setval{8b.concur-free.minus.geoma.humaneval-plus}{+2.6}
\setval{8b.concur-free.minus.geoma.humaneval-plus-abs}{2.6}
\setval{8b.concur-free.minus.geoma.if}{+1.7}
\setval{8b.concur-free.minus.geoma.if-abs}{1.7}
\setval{8b.concur-free.minus.geoma.ifbench}{+4.7}
\setval{8b.concur-free.minus.geoma.ifbench-abs}{4.7}
\setval{8b.concur-free.minus.geoma.ifeval}{+1.9}
\setval{8b.concur-free.minus.geoma.ifeval-abs}{1.9}
\setval{8b.concur-free.minus.geoma.judge2.avg}{+2.9}
\setval{8b.concur-free.minus.geoma.judge2.avg-abs}{2.9}
\setval{8b.concur-free.minus.geoma.judge2.creative-writing-v3}{+5.0}
\setval{8b.concur-free.minus.geoma.judge2.creative-writing-v3-abs}{5.0}
\setval{8b.concur-free.minus.geoma.judge2.longbench-write}{+9.2}
\setval{8b.concur-free.minus.geoma.judge2.longbench-write-abs}{9.2}
\setval{8b.concur-free.minus.geoma.judge2.writing}{+5.5}
\setval{8b.concur-free.minus.geoma.judge2.writing-abs}{5.5}
\setval{8b.concur-free.minus.geoma.judge2.writingbench}{+2.4}
\setval{8b.concur-free.minus.geoma.judge2.writingbench-abs}{2.4}
\setval{8b.concur-free.minus.geoma.livecodebench}{+1.4}
\setval{8b.concur-free.minus.geoma.livecodebench-abs}{1.4}
\setval{8b.concur-free.minus.geoma.longbench-write}{+14.0}
\setval{8b.concur-free.minus.geoma.longbench-write-abs}{14.0}
\setval{8b.concur-free.minus.geoma.math}{+3.1}
\setval{8b.concur-free.minus.geoma.math-abs}{3.1}
\setval{8b.concur-free.minus.geoma.math500}{+3.8}
\setval{8b.concur-free.minus.geoma.math500-abs}{3.8}
\setval{8b.concur-free.minus.geoma.mbpp-plus}{$-$0.5}
\setval{8b.concur-free.minus.geoma.mbpp-plus-abs}{0.5}
\setval{8b.concur-free.minus.geoma.minerva}{+0.0}
\setval{8b.concur-free.minus.geoma.minerva-abs}{0.0}
\setval{8b.concur-free.minus.geoma.olympiad-bench}{+4.1}
\setval{8b.concur-free.minus.geoma.olympiad-bench-abs}{4.1}
\setval{8b.concur-free.minus.geoma.writing}{+8.7}
\setval{8b.concur-free.minus.geoma.writing-abs}{8.7}
\setval{8b.concur-free.minus.geoma.writingbench}{+10.8}
\setval{8b.concur-free.minus.geoma.writingbench-abs}{10.8}
\setval{8b.concur-free.minus.grpo-nodiv.aime24}{$-$0.9}
\setval{8b.concur-free.minus.grpo-nodiv.aime24-abs}{0.9}
\setval{8b.concur-free.minus.grpo-nodiv.aime25}{+1.8}
\setval{8b.concur-free.minus.grpo-nodiv.aime25-abs}{1.8}
\setval{8b.concur-free.minus.grpo-nodiv.amc23}{+1.1}
\setval{8b.concur-free.minus.grpo-nodiv.amc23-abs}{1.1}
\setval{8b.concur-free.minus.grpo-nodiv.avg}{+2.2}
\setval{8b.concur-free.minus.grpo-nodiv.avg-abs}{2.2}
\setval{8b.concur-free.minus.grpo-nodiv.code}{+0.1}
\setval{8b.concur-free.minus.grpo-nodiv.code-abs}{0.1}
\setval{8b.concur-free.minus.grpo-nodiv.creative-writing-v3}{+5.0}
\setval{8b.concur-free.minus.grpo-nodiv.creative-writing-v3-abs}{5.0}
\setval{8b.concur-free.minus.grpo-nodiv.guidebench}{$-$1.9}
\setval{8b.concur-free.minus.grpo-nodiv.guidebench-abs}{1.9}
\setval{8b.concur-free.minus.grpo-nodiv.humaneval-plus}{$-$1.0}
\setval{8b.concur-free.minus.grpo-nodiv.humaneval-plus-abs}{1.0}
\setval{8b.concur-free.minus.grpo-nodiv.if}{+0.1}
\setval{8b.concur-free.minus.grpo-nodiv.if-abs}{0.1}
\setval{8b.concur-free.minus.grpo-nodiv.ifbench}{+2.3}
\setval{8b.concur-free.minus.grpo-nodiv.ifbench-abs}{2.3}
\setval{8b.concur-free.minus.grpo-nodiv.ifeval}{$-$0.2}
\setval{8b.concur-free.minus.grpo-nodiv.ifeval-abs}{0.2}
\setval{8b.concur-free.minus.grpo-nodiv.judge2.avg}{+1.2}
\setval{8b.concur-free.minus.grpo-nodiv.judge2.avg-abs}{1.2}
\setval{8b.concur-free.minus.grpo-nodiv.judge2.creative-writing-v3}{+5.0}
\setval{8b.concur-free.minus.grpo-nodiv.judge2.creative-writing-v3-abs}{5.0}
\setval{8b.concur-free.minus.grpo-nodiv.judge2.longbench-write}{+7.6}
\setval{8b.concur-free.minus.grpo-nodiv.judge2.longbench-write-abs}{7.6}
\setval{8b.concur-free.minus.grpo-nodiv.judge2.writing}{+6.5}
\setval{8b.concur-free.minus.grpo-nodiv.judge2.writing-abs}{6.5}
\setval{8b.concur-free.minus.grpo-nodiv.judge2.writingbench}{+6.8}
\setval{8b.concur-free.minus.grpo-nodiv.judge2.writingbench-abs}{6.8}
\setval{8b.concur-free.minus.grpo-nodiv.livecodebench}{+2.1}
\setval{8b.concur-free.minus.grpo-nodiv.livecodebench-abs}{2.1}
\setval{8b.concur-free.minus.grpo-nodiv.longbench-write}{+11.9}
\setval{8b.concur-free.minus.grpo-nodiv.longbench-write-abs}{11.9}
\setval{8b.concur-free.minus.grpo-nodiv.math}{$-$0.2}
\setval{8b.concur-free.minus.grpo-nodiv.math-abs}{0.2}
\setval{8b.concur-free.minus.grpo-nodiv.math500}{$-$0.7}
\setval{8b.concur-free.minus.grpo-nodiv.math500-abs}{0.7}
\setval{8b.concur-free.minus.grpo-nodiv.mbpp-plus}{$-$0.9}
\setval{8b.concur-free.minus.grpo-nodiv.mbpp-plus-abs}{0.9}
\setval{8b.concur-free.minus.grpo-nodiv.minerva}{$-$1.0}
\setval{8b.concur-free.minus.grpo-nodiv.minerva-abs}{1.0}
\setval{8b.concur-free.minus.grpo-nodiv.olympiad-bench}{$-$1.5}
\setval{8b.concur-free.minus.grpo-nodiv.olympiad-bench-abs}{1.5}
\setval{8b.concur-free.minus.grpo-nodiv.writing}{+11.1}
\setval{8b.concur-free.minus.grpo-nodiv.writing-abs}{11.1}
\setval{8b.concur-free.minus.grpo-nodiv.writingbench}{+16.4}
\setval{8b.concur-free.minus.grpo-nodiv.writingbench-abs}{16.4}
\setval{8b.concur-free.minus.grpo.aime24}{+3.9}
\setval{8b.concur-free.minus.grpo.aime24-abs}{3.9}
\setval{8b.concur-free.minus.grpo.aime25}{+1.0}
\setval{8b.concur-free.minus.grpo.aime25-abs}{1.0}
\setval{8b.concur-free.minus.grpo.amc23}{+0.3}
\setval{8b.concur-free.minus.grpo.amc23-abs}{0.3}
\setval{8b.concur-free.minus.grpo.avg}{+2.4}
\setval{8b.concur-free.minus.grpo.avg-abs}{2.4}
\setval{8b.concur-free.minus.grpo.code}{+0.9}
\setval{8b.concur-free.minus.grpo.code-abs}{0.9}
\setval{8b.concur-free.minus.grpo.creative-writing-v3}{$-$0.8}
\setval{8b.concur-free.minus.grpo.creative-writing-v3-abs}{0.8}
\setval{8b.concur-free.minus.grpo.guidebench}{+1.1}
\setval{8b.concur-free.minus.grpo.guidebench-abs}{1.1}
\setval{8b.concur-free.minus.grpo.humaneval-plus}{+1.6}
\setval{8b.concur-free.minus.grpo.humaneval-plus-abs}{1.6}
\setval{8b.concur-free.minus.grpo.if}{+0.0}
\setval{8b.concur-free.minus.grpo.if-abs}{0.0}
\setval{8b.concur-free.minus.grpo.ifbench}{$-$3.3}
\setval{8b.concur-free.minus.grpo.ifbench-abs}{3.3}
\setval{8b.concur-free.minus.grpo.ifeval}{+2.4}
\setval{8b.concur-free.minus.grpo.ifeval-abs}{2.4}
\setval{8b.concur-free.minus.grpo.judge2.avg}{+2.1}
\setval{8b.concur-free.minus.grpo.judge2.avg-abs}{2.1}
\setval{8b.concur-free.minus.grpo.judge2.creative-writing-v3}{+3.2}
\setval{8b.concur-free.minus.grpo.judge2.creative-writing-v3-abs}{3.2}
\setval{8b.concur-free.minus.grpo.judge2.longbench-write}{+11.7}
\setval{8b.concur-free.minus.grpo.judge2.longbench-write-abs}{11.7}
\setval{8b.concur-free.minus.grpo.judge2.writing}{+6.2}
\setval{8b.concur-free.minus.grpo.judge2.writing-abs}{6.2}
\setval{8b.concur-free.minus.grpo.judge2.writingbench}{+3.8}
\setval{8b.concur-free.minus.grpo.judge2.writingbench-abs}{3.8}
\setval{8b.concur-free.minus.grpo.livecodebench}{+0.6}
\setval{8b.concur-free.minus.grpo.livecodebench-abs}{0.6}
\setval{8b.concur-free.minus.grpo.longbench-write}{+13.2}
\setval{8b.concur-free.minus.grpo.longbench-write-abs}{13.2}
\setval{8b.concur-free.minus.grpo.math}{+1.7}
\setval{8b.concur-free.minus.grpo.math-abs}{1.7}
\setval{8b.concur-free.minus.grpo.math500}{+2.8}
\setval{8b.concur-free.minus.grpo.math500-abs}{2.8}
\setval{8b.concur-free.minus.grpo.mbpp-plus}{+0.4}
\setval{8b.concur-free.minus.grpo.mbpp-plus-abs}{0.4}
\setval{8b.concur-free.minus.grpo.minerva}{$-$0.1}
\setval{8b.concur-free.minus.grpo.minerva-abs}{0.1}
\setval{8b.concur-free.minus.grpo.olympiad-bench}{+2.1}
\setval{8b.concur-free.minus.grpo.olympiad-bench-abs}{2.1}
\setval{8b.concur-free.minus.grpo.writing}{+7.6}
\setval{8b.concur-free.minus.grpo.writing-abs}{7.6}
\setval{8b.concur-free.minus.grpo.writingbench}{+10.3}
\setval{8b.concur-free.minus.grpo.writingbench-abs}{10.3}
\setval{8b.concur-free.minus.mt-grpo.aime24}{+5.2}
\setval{8b.concur-free.minus.mt-grpo.aime24-abs}{5.2}
\setval{8b.concur-free.minus.mt-grpo.aime25}{+0.8}
\setval{8b.concur-free.minus.mt-grpo.aime25-abs}{0.8}
\setval{8b.concur-free.minus.mt-grpo.amc23}{$-$0.9}
\setval{8b.concur-free.minus.mt-grpo.amc23-abs}{0.9}
\setval{8b.concur-free.minus.mt-grpo.avg}{+3.1}
\setval{8b.concur-free.minus.mt-grpo.avg-abs}{3.1}
\setval{8b.concur-free.minus.mt-grpo.code}{$-$0.2}
\setval{8b.concur-free.minus.mt-grpo.code-abs}{0.2}
\setval{8b.concur-free.minus.mt-grpo.creative-writing-v3}{+1.3}
\setval{8b.concur-free.minus.mt-grpo.creative-writing-v3-abs}{1.3}
\setval{8b.concur-free.minus.mt-grpo.guidebench}{$-$0.2}
\setval{8b.concur-free.minus.mt-grpo.guidebench-abs}{0.2}
\setval{8b.concur-free.minus.mt-grpo.humaneval-plus}{+0.2}
\setval{8b.concur-free.minus.mt-grpo.humaneval-plus-abs}{0.2}
\setval{8b.concur-free.minus.mt-grpo.if}{+1.1}
\setval{8b.concur-free.minus.mt-grpo.if-abs}{1.1}
\setval{8b.concur-free.minus.mt-grpo.ifbench}{+1.3}
\setval{8b.concur-free.minus.mt-grpo.ifbench-abs}{1.3}
\setval{8b.concur-free.minus.mt-grpo.ifeval}{+2.3}
\setval{8b.concur-free.minus.mt-grpo.ifeval-abs}{2.3}
\setval{8b.concur-free.minus.mt-grpo.judge2.avg}{+2.1}
\setval{8b.concur-free.minus.mt-grpo.judge2.avg-abs}{2.1}
\setval{8b.concur-free.minus.mt-grpo.judge2.creative-writing-v3}{+1.0}
\setval{8b.concur-free.minus.mt-grpo.judge2.creative-writing-v3-abs}{1.0}
\setval{8b.concur-free.minus.mt-grpo.judge2.longbench-write}{+11.6}
\setval{8b.concur-free.minus.mt-grpo.judge2.longbench-write-abs}{11.6}
\setval{8b.concur-free.minus.mt-grpo.judge2.writing}{+5.9}
\setval{8b.concur-free.minus.mt-grpo.judge2.writing-abs}{5.9}
\setval{8b.concur-free.minus.mt-grpo.judge2.writingbench}{+5.2}
\setval{8b.concur-free.minus.mt-grpo.judge2.writingbench-abs}{5.2}
\setval{8b.concur-free.minus.mt-grpo.livecodebench}{$-$0.6}
\setval{8b.concur-free.minus.mt-grpo.livecodebench-abs}{0.6}
\setval{8b.concur-free.minus.mt-grpo.longbench-write}{+15.5}
\setval{8b.concur-free.minus.mt-grpo.longbench-write-abs}{15.5}
\setval{8b.concur-free.minus.mt-grpo.math}{+1.8}
\setval{8b.concur-free.minus.mt-grpo.math-abs}{1.8}
\setval{8b.concur-free.minus.mt-grpo.math500}{+2.8}
\setval{8b.concur-free.minus.mt-grpo.math500-abs}{2.8}
\setval{8b.concur-free.minus.mt-grpo.mbpp-plus}{$-$0.2}
\setval{8b.concur-free.minus.mt-grpo.mbpp-plus-abs}{0.2}
\setval{8b.concur-free.minus.mt-grpo.minerva}{$-$1.3}
\setval{8b.concur-free.minus.mt-grpo.minerva-abs}{1.3}
\setval{8b.concur-free.minus.mt-grpo.olympiad-bench}{+4.3}
\setval{8b.concur-free.minus.mt-grpo.olympiad-bench-abs}{4.3}
\setval{8b.concur-free.minus.mt-grpo.writing}{+10.6}
\setval{8b.concur-free.minus.mt-grpo.writing-abs}{10.6}
\setval{8b.concur-free.minus.mt-grpo.writingbench}{+15.1}
\setval{8b.concur-free.minus.mt-grpo.writingbench-abs}{15.1}
\setval{8b.concur-free.minus.papo.aime24}{+7.5}
\setval{8b.concur-free.minus.papo.aime24-abs}{7.5}
\setval{8b.concur-free.minus.papo.aime25}{+6.5}
\setval{8b.concur-free.minus.papo.aime25-abs}{6.5}
\setval{8b.concur-free.minus.papo.amc23}{+2.7}
\setval{8b.concur-free.minus.papo.amc23-abs}{2.7}
\setval{8b.concur-free.minus.papo.avg}{+4.4}
\setval{8b.concur-free.minus.papo.avg-abs}{4.4}
\setval{8b.concur-free.minus.papo.code}{+1.7}
\setval{8b.concur-free.minus.papo.code-abs}{1.7}
\setval{8b.concur-free.minus.papo.creative-writing-v3}{+0.5}
\setval{8b.concur-free.minus.papo.creative-writing-v3-abs}{0.5}
\setval{8b.concur-free.minus.papo.guidebench}{$-$1.4}
\setval{8b.concur-free.minus.papo.guidebench-abs}{1.4}
\setval{8b.concur-free.minus.papo.humaneval-plus}{+2.8}
\setval{8b.concur-free.minus.papo.humaneval-plus-abs}{2.8}
\setval{8b.concur-free.minus.papo.if}{+0.9}
\setval{8b.concur-free.minus.papo.if-abs}{0.9}
\setval{8b.concur-free.minus.papo.ifbench}{+1.3}
\setval{8b.concur-free.minus.papo.ifbench-abs}{1.3}
\setval{8b.concur-free.minus.papo.ifeval}{+2.9}
\setval{8b.concur-free.minus.papo.ifeval-abs}{2.9}
\setval{8b.concur-free.minus.papo.judge2.avg}{+3.6}
\setval{8b.concur-free.minus.papo.judge2.avg-abs}{3.6}
\setval{8b.concur-free.minus.papo.judge2.creative-writing-v3}{+0.6}
\setval{8b.concur-free.minus.papo.judge2.creative-writing-v3-abs}{0.6}
\setval{8b.concur-free.minus.papo.judge2.longbench-write}{+14.6}
\setval{8b.concur-free.minus.papo.judge2.longbench-write-abs}{14.6}
\setval{8b.concur-free.minus.papo.judge2.writing}{+6.5}
\setval{8b.concur-free.minus.papo.judge2.writing-abs}{6.5}
\setval{8b.concur-free.minus.papo.judge2.writingbench}{+4.4}
\setval{8b.concur-free.minus.papo.judge2.writingbench-abs}{4.4}
\setval{8b.concur-free.minus.papo.livecodebench}{+0.5}
\setval{8b.concur-free.minus.papo.livecodebench-abs}{0.5}
\setval{8b.concur-free.minus.papo.longbench-write}{+17.5}
\setval{8b.concur-free.minus.papo.longbench-write-abs}{17.5}
\setval{8b.concur-free.minus.papo.math}{+4.5}
\setval{8b.concur-free.minus.papo.math-abs}{4.5}
\setval{8b.concur-free.minus.papo.math500}{+5.0}
\setval{8b.concur-free.minus.papo.math500-abs}{5.0}
\setval{8b.concur-free.minus.papo.mbpp-plus}{+1.7}
\setval{8b.concur-free.minus.papo.mbpp-plus-abs}{1.7}
\setval{8b.concur-free.minus.papo.minerva}{+0.4}
\setval{8b.concur-free.minus.papo.minerva-abs}{0.4}
\setval{8b.concur-free.minus.papo.olympiad-bench}{+5.2}
\setval{8b.concur-free.minus.papo.olympiad-bench-abs}{5.2}
\setval{8b.concur-free.minus.papo.writing}{+10.5}
\setval{8b.concur-free.minus.papo.writing-abs}{10.5}
\setval{8b.concur-free.minus.papo.writingbench}{+13.6}
\setval{8b.concur-free.minus.papo.writingbench-abs}{13.6}
\setval{8b.concur-free.minus.saw.aime24}{+4.4}
\setval{8b.concur-free.minus.saw.aime24-abs}{4.4}
\setval{8b.concur-free.minus.saw.aime25}{+1.0}
\setval{8b.concur-free.minus.saw.aime25-abs}{1.0}
\setval{8b.concur-free.minus.saw.amc23}{+2.4}
\setval{8b.concur-free.minus.saw.amc23-abs}{2.4}
\setval{8b.concur-free.minus.saw.avg}{+2.3}
\setval{8b.concur-free.minus.saw.avg-abs}{2.3}
\setval{8b.concur-free.minus.saw.code}{+0.2}
\setval{8b.concur-free.minus.saw.code-abs}{0.2}
\setval{8b.concur-free.minus.saw.creative-writing-v3}{+2.0}
\setval{8b.concur-free.minus.saw.creative-writing-v3-abs}{2.0}
\setval{8b.concur-free.minus.saw.guidebench}{$-$1.3}
\setval{8b.concur-free.minus.saw.guidebench-abs}{1.3}
\setval{8b.concur-free.minus.saw.humaneval-plus}{$-$0.5}
\setval{8b.concur-free.minus.saw.humaneval-plus-abs}{0.5}
\setval{8b.concur-free.minus.saw.if}{$-$0.4}
\setval{8b.concur-free.minus.saw.if-abs}{0.4}
\setval{8b.concur-free.minus.saw.ifbench}{$-$1.3}
\setval{8b.concur-free.minus.saw.ifbench-abs}{1.3}
\setval{8b.concur-free.minus.saw.ifeval}{+1.6}
\setval{8b.concur-free.minus.saw.ifeval-abs}{1.6}
\setval{8b.concur-free.minus.saw.judge2.avg}{+1.9}
\setval{8b.concur-free.minus.saw.judge2.avg-abs}{1.9}
\setval{8b.concur-free.minus.saw.judge2.creative-writing-v3}{+5.9}
\setval{8b.concur-free.minus.saw.judge2.creative-writing-v3-abs}{5.9}
\setval{8b.concur-free.minus.saw.judge2.longbench-write}{+9.3}
\setval{8b.concur-free.minus.saw.judge2.longbench-write-abs}{9.3}
\setval{8b.concur-free.minus.saw.judge2.writing}{+6.3}
\setval{8b.concur-free.minus.saw.judge2.writing-abs}{6.3}
\setval{8b.concur-free.minus.saw.judge2.writingbench}{+3.7}
\setval{8b.concur-free.minus.saw.judge2.writingbench-abs}{3.7}
\setval{8b.concur-free.minus.saw.livecodebench}{+0.4}
\setval{8b.concur-free.minus.saw.livecodebench-abs}{0.4}
\setval{8b.concur-free.minus.saw.longbench-write}{+11.9}
\setval{8b.concur-free.minus.saw.longbench-write-abs}{11.9}
\setval{8b.concur-free.minus.saw.math}{+1.5}
\setval{8b.concur-free.minus.saw.math-abs}{1.5}
\setval{8b.concur-free.minus.saw.math500}{+1.5}
\setval{8b.concur-free.minus.saw.math500-abs}{1.5}
\setval{8b.concur-free.minus.saw.mbpp-plus}{+0.8}
\setval{8b.concur-free.minus.saw.mbpp-plus-abs}{0.8}
\setval{8b.concur-free.minus.saw.minerva}{$-$2.0}
\setval{8b.concur-free.minus.saw.minerva-abs}{2.0}
\setval{8b.concur-free.minus.saw.olympiad-bench}{+1.9}
\setval{8b.concur-free.minus.saw.olympiad-bench-abs}{1.9}
\setval{8b.concur-free.minus.saw.writing}{+8.4}
\setval{8b.concur-free.minus.saw.writing-abs}{8.4}
\setval{8b.concur-free.minus.saw.writingbench}{+11.4}
\setval{8b.concur-free.minus.saw.writingbench-abs}{11.4}
\setval{8b.concur-free.passing-at-ceiling.code.early}{0.71}
\setval{8b.concur-free.passing-at-ceiling.code.late}{0.75}
\setval{8b.concur-free.passing-at-ceiling.code.run}{0.74}
\setval{8b.concur-free.passing-at-ceiling.if.early}{0.32}
\setval{8b.concur-free.passing-at-ceiling.if.late}{0.55}
\setval{8b.concur-free.passing-at-ceiling.if.run}{0.48}
\setval{8b.concur-free.passing-at-ceiling.math.early}{1.00}
\setval{8b.concur-free.passing-at-ceiling.math.late}{1.00}
\setval{8b.concur-free.passing-at-ceiling.math.run}{1.00}
\setval{8b.concur-free.passing-at-ceiling.writing.early}{0.00}
\setval{8b.concur-free.passing-at-ceiling.writing.late}{0.03}
\setval{8b.concur-free.passing-at-ceiling.writing.run}{0.01}
\setval{8b.concur-free.primary-cosine.code.early}{0.97}
\setval{8b.concur-free.primary-cosine.code.late}{0.96}
\setval{8b.concur-free.primary-cosine.code.run}{0.96}
\setval{8b.concur-free.primary-cosine.if.early}{0.85}
\setval{8b.concur-free.primary-cosine.if.late}{0.80}
\setval{8b.concur-free.primary-cosine.if.run}{0.83}
\setval{8b.concur-free.primary-cosine.math.early}{0.96}
\setval{8b.concur-free.primary-cosine.math.late}{0.97}
\setval{8b.concur-free.primary-cosine.math.run}{0.97}
\setval{8b.concur-free.primary-cosine.writing.early}{0.94}
\setval{8b.concur-free.primary-cosine.writing.late}{0.91}
\setval{8b.concur-free.primary-cosine.writing.run}{0.92}
\setval{8b.concur-free.primary-level.code.early}{0.54}
\setval{8b.concur-free.primary-level.code.late}{0.76}
\setval{8b.concur-free.primary-level.code.run}{0.71}
\setval{8b.concur-free.primary-level.if.early}{0.36}
\setval{8b.concur-free.primary-level.if.late}{0.75}
\setval{8b.concur-free.primary-level.if.run}{0.64}
\setval{8b.concur-free.primary-level.math.early}{0.49}
\setval{8b.concur-free.primary-level.math.late}{0.70}
\setval{8b.concur-free.primary-level.math.run}{0.64}
\setval{8b.concur-free.primary-level.writing.early}{0.54}
\setval{8b.concur-free.primary-level.writing.late}{0.82}
\setval{8b.concur-free.primary-level.writing.run}{0.77}
\setval{8b.concur-free.solved.code.early}{0.08}
\setval{8b.concur-free.solved.code.late}{0.02}
\setval{8b.concur-free.solved.code.run}{0.04}
\setval{8b.concur-free.solved.if.early}{0.03}
\setval{8b.concur-free.solved.if.late}{0.17}
\setval{8b.concur-free.solved.if.run}{0.13}
\setval{8b.concur-free.solved.math.early}{0.15}
\setval{8b.concur-free.solved.math.late}{0.16}
\setval{8b.concur-free.solved.math.run}{0.15}
\setval{8b.concur-free.solved.writing.early}{0.00}
\setval{8b.concur-free.solved.writing.late}{0.00}
\setval{8b.concur-free.solved.writing.run}{0.00}
\setval{8b.concur-free.trust.code.agreement}{+0.006}
\setval{8b.concur-free.trust.code.errors}{+2.3}
\setval{8b.concur-free.trust.code.first-free}{229}
\setval{8b.concur-free.trust.code.free-share}{0.20}
\setval{8b.concur-free.trust.if.agreement}{+0.016}
\setval{8b.concur-free.trust.if.errors}{+7.2}
\setval{8b.concur-free.trust.if.first-free}{6}
\setval{8b.concur-free.trust.if.free-share}{0.97}
\setval{8b.concur-free.trust.math.first-free}{--}
\setval{8b.concur-free.trust.math.free-share}{0.00}
\setval{8b.concur-free.trust.writing-rm.agreement}{+0.049}
\setval{8b.concur-free.trust.writing-rm.errors}{+29.8}
\setval{8b.concur-free.trust.writing-rm.first-free}{1}
\setval{8b.concur-free.trust.writing-rm.free-share}{1.00}
\setval{8b.concur-free.trust.writing.agreement}{$-$0.045}
\setval{8b.concur-free.trust.writing.errors}{$-$33.7}
\setval{8b.concur-free.trust.writing.first-free}{--}
\setval{8b.concur-free.trust.writing.free-share}{0.00}
\setval{8b.concur-free.update-share.code.early}{0.20}
\setval{8b.concur-free.update-share.code.late}{0.22}
\setval{8b.concur-free.update-share.code.run}{0.22}
\setval{8b.concur-free.update-share.if.early}{0.22}
\setval{8b.concur-free.update-share.if.late}{0.18}
\setval{8b.concur-free.update-share.if.run}{0.21}
\setval{8b.concur-free.update-share.math.early}{0.18}
\setval{8b.concur-free.update-share.math.late}{0.25}
\setval{8b.concur-free.update-share.math.run}{0.24}
\setval{8b.concur-free.update-share.writing.early}{0.41}
\setval{8b.concur-free.update-share.writing.late}{0.35}
\setval{8b.concur-free.update-share.writing.run}{0.33}
\setval{8b.concur-tied.aux-change.code}{+0.16}
\setval{8b.concur-tied.aux-change.if}{+0.17}
\setval{8b.concur-tied.aux-change.math}{$-$0.55}
\setval{8b.concur-tied.aux-change.writing}{$-$0.39}
\setval{8b.concur-tied.aux-change.writing-rm}{+0.32}
\setval{8b.concur-tied.aux-full.code.early}{0.56}
\setval{8b.concur-tied.aux-full.code.late}{0.82}
\setval{8b.concur-tied.aux-full.code.run}{0.74}
\setval{8b.concur-tied.aux-full.if.early}{0.69}
\setval{8b.concur-tied.aux-full.if.late}{0.91}
\setval{8b.concur-tied.aux-full.if.run}{0.83}
\setval{8b.concur-tied.aux-full.math.early}{0.52}
\setval{8b.concur-tied.aux-full.math.late}{0.10}
\setval{8b.concur-tied.aux-full.math.run}{0.18}
\setval{8b.concur-tied.aux-full.writing-rm.early}{0.08}
\setval{8b.concur-tied.aux-full.writing-rm.late}{0.51}
\setval{8b.concur-tied.aux-full.writing-rm.run}{0.42}
\setval{8b.concur-tied.aux-full.writing.early}{0.61}
\setval{8b.concur-tied.aux-full.writing.late}{0.23}
\setval{8b.concur-tied.aux-full.writing.run}{0.26}
\setval{8b.concur-tied.aux-level.code.early}{0.88}
\setval{8b.concur-tied.aux-level.code.late}{0.97}
\setval{8b.concur-tied.aux-level.code.run}{0.95}
\setval{8b.concur-tied.aux-level.if.early}{0.86}
\setval{8b.concur-tied.aux-level.if.late}{0.96}
\setval{8b.concur-tied.aux-level.if.run}{0.92}
\setval{8b.concur-tied.aux-level.math.early}{0.89}
\setval{8b.concur-tied.aux-level.math.late}{0.29}
\setval{8b.concur-tied.aux-level.math.run}{0.45}
\setval{8b.concur-tied.aux-level.writing-rm.early}{0.77}
\setval{8b.concur-tied.aux-level.writing-rm.late}{0.95}
\setval{8b.concur-tied.aux-level.writing-rm.run}{0.93}
\setval{8b.concur-tied.aux-level.writing.early}{0.89}
\setval{8b.concur-tied.aux-level.writing.late}{0.45}
\setval{8b.concur-tied.aux-level.writing.run}{0.53}
\setval{8b.concur-tied.bench.aime24}{31.2}
\setval{8b.concur-tied.bench.aime25}{24.0}
\setval{8b.concur-tied.bench.amc23}{78.5}
\setval{8b.concur-tied.bench.creative-writing-v3}{52.4}
\setval{8b.concur-tied.bench.guidebench}{83.4}
\setval{8b.concur-tied.bench.humaneval-plus}{87.4}
\setval{8b.concur-tied.bench.ifbench}{39.3}
\setval{8b.concur-tied.bench.ifeval}{88.8}
\setval{8b.concur-tied.bench.livecodebench}{27.2}
\setval{8b.concur-tied.bench.longbench-write}{81.2}
\setval{8b.concur-tied.bench.math500}{89.6}
\setval{8b.concur-tied.bench.mbpp-plus}{74.9}
\setval{8b.concur-tied.bench.minerva}{55.4}
\setval{8b.concur-tied.bench.olympiad-bench}{59.5}
\setval{8b.concur-tied.bench.writingbench}{87.2}
\setval{8b.concur-tied.best-count.all}{9}
\setval{8b.concur-tied.best-count.code}{1}
\setval{8b.concur-tied.best-count.if}{0}
\setval{8b.concur-tied.best-count.math}{6}
\setval{8b.concur-tied.best-count.writing}{2}
\setval{8b.concur-tied.eval.avg}{64.0}
\setval{8b.concur-tied.eval.code}{63.1}
\setval{8b.concur-tied.eval.if}{70.5}
\setval{8b.concur-tied.eval.math}{56.4}
\setval{8b.concur-tied.eval.writing}{73.6}
\setval{8b.concur-tied.gate.code.against.early}{0.05}
\setval{8b.concur-tied.gate.code.against.late}{0.06}
\setval{8b.concur-tied.gate.code.against.run}{0.05}
\setval{8b.concur-tied.gate.code.agreement}{+0.005}
\setval{8b.concur-tied.gate.code.ceiling.early}{0.17}
\setval{8b.concur-tied.gate.code.ceiling.late}{0.23}
\setval{8b.concur-tied.gate.code.ceiling.run}{0.24}
\setval{8b.concur-tied.gate.code.cosine.early}{0.79}
\setval{8b.concur-tied.gate.code.cosine.late}{0.63}
\setval{8b.concur-tied.gate.code.cosine.run}{0.68}
\setval{8b.concur-tied.gate.if.against.early}{0.05}
\setval{8b.concur-tied.gate.if.against.late}{0.04}
\setval{8b.concur-tied.gate.if.against.run}{0.05}
\setval{8b.concur-tied.gate.if.agreement}{+0.022}
\setval{8b.concur-tied.gate.if.ceiling.early}{0.06}
\setval{8b.concur-tied.gate.if.ceiling.late}{0.18}
\setval{8b.concur-tied.gate.if.ceiling.run}{0.15}
\setval{8b.concur-tied.gate.if.cosine.early}{0.74}
\setval{8b.concur-tied.gate.if.cosine.late}{0.53}
\setval{8b.concur-tied.gate.if.cosine.run}{0.58}
\setval{8b.concur-tied.gate.math.against.early}{0.00}
\setval{8b.concur-tied.gate.math.against.late}{0.00}
\setval{8b.concur-tied.gate.math.against.run}{0.00}
\setval{8b.concur-tied.gate.math.ceiling.early}{0.25}
\setval{8b.concur-tied.gate.math.ceiling.late}{0.89}
\setval{8b.concur-tied.gate.math.ceiling.run}{0.75}
\setval{8b.concur-tied.gate.math.cosine.early}{0.84}
\setval{8b.concur-tied.gate.math.cosine.late}{0.19}
\setval{8b.concur-tied.gate.math.cosine.run}{0.34}
\setval{8b.concur-tied.gate.writing-rm.against.early}{0.20}
\setval{8b.concur-tied.gate.writing-rm.against.late}{0.56}
\setval{8b.concur-tied.gate.writing-rm.against.run}{0.53}
\setval{8b.concur-tied.gate.writing-rm.agreement}{+0.050}
\setval{8b.concur-tied.gate.writing-rm.ceiling.early}{0.00}
\setval{8b.concur-tied.gate.writing-rm.ceiling.late}{0.01}
\setval{8b.concur-tied.gate.writing-rm.ceiling.run}{0.00}
\setval{8b.concur-tied.gate.writing-rm.cosine.early}{0.69}
\setval{8b.concur-tied.gate.writing-rm.cosine.late}{0.09}
\setval{8b.concur-tied.gate.writing-rm.cosine.run}{0.17}
\setval{8b.concur-tied.gate.writing.against.early}{0.31}
\setval{8b.concur-tied.gate.writing.against.late}{0.65}
\setval{8b.concur-tied.gate.writing.against.run}{0.64}
\setval{8b.concur-tied.gate.writing.agreement}{$-$0.047}
\setval{8b.concur-tied.gate.writing.ceiling.early}{0.00}
\setval{8b.concur-tied.gate.writing.ceiling.late}{0.01}
\setval{8b.concur-tied.gate.writing.ceiling.run}{0.00}
\setval{8b.concur-tied.gate.writing.cosine.early}{0.49}
\setval{8b.concur-tied.gate.writing.cosine.late}{-0.02}
\setval{8b.concur-tied.gate.writing.cosine.run}{0.01}
\setval{8b.concur-tied.judge2.avg}{61.3}
\setval{8b.concur-tied.judge2.creative-writing-v3}{41.8}
\setval{8b.concur-tied.judge2.longbench-write}{74.8}
\setval{8b.concur-tied.judge2.writing}{60.1}
\setval{8b.concur-tied.judge2.writingbench}{63.8}
\setval{8b.concur-tied.last-step}{300}
\setval{8b.concur-tied.minus.base.aime24}{+19.9}
\setval{8b.concur-tied.minus.base.aime24-abs}{19.9}
\setval{8b.concur-tied.minus.base.aime25}{+15.4}
\setval{8b.concur-tied.minus.base.aime25-abs}{15.4}
\setval{8b.concur-tied.minus.base.amc23}{+28.3}
\setval{8b.concur-tied.minus.base.amc23-abs}{28.3}
\setval{8b.concur-tied.minus.base.avg}{+22.5}
\setval{8b.concur-tied.minus.base.avg-abs}{22.5}
\setval{8b.concur-tied.minus.base.code}{+7.1}
\setval{8b.concur-tied.minus.base.code-abs}{7.1}
\setval{8b.concur-tied.minus.base.creative-writing-v3}{+25.7}
\setval{8b.concur-tied.minus.base.creative-writing-v3-abs}{25.7}
\setval{8b.concur-tied.minus.base.guidebench}{+37.8}
\setval{8b.concur-tied.minus.base.guidebench-abs}{37.8}
\setval{8b.concur-tied.minus.base.humaneval-plus}{+7.3}
\setval{8b.concur-tied.minus.base.humaneval-plus-abs}{7.3}
\setval{8b.concur-tied.minus.base.if}{+31.9}
\setval{8b.concur-tied.minus.base.if-abs}{31.9}
\setval{8b.concur-tied.minus.base.ifbench}{+22.0}
\setval{8b.concur-tied.minus.base.ifbench-abs}{22.0}
\setval{8b.concur-tied.minus.base.ifeval}{+35.9}
\setval{8b.concur-tied.minus.base.ifeval-abs}{35.9}
\setval{8b.concur-tied.minus.base.judge2.avg}{+20.3}
\setval{8b.concur-tied.minus.base.judge2.avg-abs}{20.3}
\setval{8b.concur-tied.minus.base.judge2.creative-writing-v3}{+15.3}
\setval{8b.concur-tied.minus.base.judge2.creative-writing-v3-abs}{15.3}
\setval{8b.concur-tied.minus.base.judge2.longbench-write}{+26.7}
\setval{8b.concur-tied.minus.base.judge2.longbench-write-abs}{26.7}
\setval{8b.concur-tied.minus.base.judge2.writing}{+23.9}
\setval{8b.concur-tied.minus.base.judge2.writing-abs}{23.9}
\setval{8b.concur-tied.minus.base.judge2.writingbench}{+29.8}
\setval{8b.concur-tied.minus.base.judge2.writingbench-abs}{29.8}
\setval{8b.concur-tied.minus.base.livecodebench}{+6.3}
\setval{8b.concur-tied.minus.base.livecodebench-abs}{6.3}
\setval{8b.concur-tied.minus.base.longbench-write}{+29.8}
\setval{8b.concur-tied.minus.base.longbench-write-abs}{29.8}
\setval{8b.concur-tied.minus.base.math}{+19.2}
\setval{8b.concur-tied.minus.base.math-abs}{19.2}
\setval{8b.concur-tied.minus.base.math500}{+17.1}
\setval{8b.concur-tied.minus.base.math500-abs}{17.1}
\setval{8b.concur-tied.minus.base.mbpp-plus}{+7.6}
\setval{8b.concur-tied.minus.base.mbpp-plus-abs}{7.6}
\setval{8b.concur-tied.minus.base.minerva}{+14.5}
\setval{8b.concur-tied.minus.base.minerva-abs}{14.5}
\setval{8b.concur-tied.minus.base.olympiad-bench}{+20.1}
\setval{8b.concur-tied.minus.base.olympiad-bench-abs}{20.1}
\setval{8b.concur-tied.minus.base.writing}{+35.3}
\setval{8b.concur-tied.minus.base.writing-abs}{35.3}
\setval{8b.concur-tied.minus.base.writingbench}{+50.4}
\setval{8b.concur-tied.minus.base.writingbench-abs}{50.4}
\setval{8b.concur-tied.minus.best.bench.aime24}{+9.9}
\setval{8b.concur-tied.minus.best.bench.aime25}{+8.2}
\setval{8b.concur-tied.minus.best.bench.amc23}{+12.9}
\setval{8b.concur-tied.minus.best.bench.creative-writing-v3}{$-$0.5}
\setval{8b.concur-tied.minus.best.bench.guidebench}{$-$1.2}
\setval{8b.concur-tied.minus.best.bench.humaneval-plus}{+1.5}
\setval{8b.concur-tied.minus.best.bench.ifbench}{$-$3.7}
\setval{8b.concur-tied.minus.best.bench.ifeval}{$-$0.5}
\setval{8b.concur-tied.minus.best.bench.livecodebench}{$-$0.5}
\setval{8b.concur-tied.minus.best.bench.longbench-write}{+10.2}
\setval{8b.concur-tied.minus.best.bench.math500}{+6.6}
\setval{8b.concur-tied.minus.best.bench.mbpp-plus}{$-$0.7}
\setval{8b.concur-tied.minus.best.bench.minerva}{+3.6}
\setval{8b.concur-tied.minus.best.bench.olympiad-bench}{+10.2}
\setval{8b.concur-tied.minus.best.bench.writingbench}{+10.7}
\setval{8b.concur-tied.minus.best.eval.avg}{+5.7}
\setval{8b.concur-tied.minus.best.eval.code}{+0.5}
\setval{8b.concur-tied.minus.best.eval.if}{$-$1.5}
\setval{8b.concur-tied.minus.best.eval.math}{+9.7}
\setval{8b.concur-tied.minus.best.eval.writing}{+7.4}
\setval{8b.concur-tied.minus.concur-free.aime24}{+11.6}
\setval{8b.concur-tied.minus.concur-free.aime24-abs}{11.6}
\setval{8b.concur-tied.minus.concur-free.aime25}{+7.4}
\setval{8b.concur-tied.minus.concur-free.aime25-abs}{7.4}
\setval{8b.concur-tied.minus.concur-free.amc23}{+13.8}
\setval{8b.concur-tied.minus.concur-free.amc23-abs}{13.8}
\setval{8b.concur-tied.minus.concur-free.avg}{+3.8}
\setval{8b.concur-tied.minus.concur-free.avg-abs}{3.8}
\setval{8b.concur-tied.minus.concur-free.code}{+0.6}
\setval{8b.concur-tied.minus.concur-free.code-abs}{0.6}
\setval{8b.concur-tied.minus.concur-free.creative-writing-v3}{+0.8}
\setval{8b.concur-tied.minus.concur-free.creative-writing-v3-abs}{0.8}
\setval{8b.concur-tied.minus.concur-free.guidebench}{+0.5}
\setval{8b.concur-tied.minus.concur-free.guidebench-abs}{0.5}
\setval{8b.concur-tied.minus.concur-free.humaneval-plus}{+2.1}
\setval{8b.concur-tied.minus.concur-free.humaneval-plus-abs}{2.1}
\setval{8b.concur-tied.minus.concur-free.if}{+0.3}
\setval{8b.concur-tied.minus.concur-free.if-abs}{0.3}
\setval{8b.concur-tied.minus.concur-free.ifbench}{+0.7}
\setval{8b.concur-tied.minus.concur-free.ifbench-abs}{0.7}
\setval{8b.concur-tied.minus.concur-free.ifeval}{$-$0.4}
\setval{8b.concur-tied.minus.concur-free.ifeval-abs}{0.4}
\setval{8b.concur-tied.minus.concur-free.judge2.avg}{+4.0}
\setval{8b.concur-tied.minus.concur-free.judge2.avg-abs}{4.0}
\setval{8b.concur-tied.minus.concur-free.judge2.creative-writing-v3}{$-$0.7}
\setval{8b.concur-tied.minus.concur-free.judge2.creative-writing-v3-abs}{0.7}
\setval{8b.concur-tied.minus.concur-free.judge2.longbench-write}{+0.7}
\setval{8b.concur-tied.minus.concur-free.judge2.longbench-write-abs}{0.7}
\setval{8b.concur-tied.minus.concur-free.judge2.writing}{+1.0}
\setval{8b.concur-tied.minus.concur-free.judge2.writing-abs}{1.0}
\setval{8b.concur-tied.minus.concur-free.judge2.writingbench}{+3.1}
\setval{8b.concur-tied.minus.concur-free.judge2.writingbench-abs}{3.1}
\setval{8b.concur-tied.minus.concur-free.livecodebench}{+0.1}
\setval{8b.concur-tied.minus.concur-free.livecodebench-abs}{0.1}
\setval{8b.concur-tied.minus.concur-free.longbench-write}{$-$1.8}
\setval{8b.concur-tied.minus.concur-free.longbench-write-abs}{1.8}
\setval{8b.concur-tied.minus.concur-free.math}{+9.0}
\setval{8b.concur-tied.minus.concur-free.math-abs}{9.0}
\setval{8b.concur-tied.minus.concur-free.math500}{+6.2}
\setval{8b.concur-tied.minus.concur-free.math500-abs}{6.2}
\setval{8b.concur-tied.minus.concur-free.mbpp-plus}{$-$0.2}
\setval{8b.concur-tied.minus.concur-free.mbpp-plus-abs}{0.2}
\setval{8b.concur-tied.minus.concur-free.minerva}{+5.6}
\setval{8b.concur-tied.minus.concur-free.minerva-abs}{5.6}
\setval{8b.concur-tied.minus.concur-free.olympiad-bench}{+9.5}
\setval{8b.concur-tied.minus.concur-free.olympiad-bench-abs}{9.5}
\setval{8b.concur-tied.minus.concur-free.writing}{$-$0.2}
\setval{8b.concur-tied.minus.concur-free.writing-abs}{0.2}
\setval{8b.concur-tied.minus.concur-free.writingbench}{+0.4}
\setval{8b.concur-tied.minus.concur-free.writingbench-abs}{0.4}
\setval{8b.concur-tied.minus.concur.aime24}{+2.5}
\setval{8b.concur-tied.minus.concur.aime24-abs}{2.5}
\setval{8b.concur-tied.minus.concur.aime25}{+0.8}
\setval{8b.concur-tied.minus.concur.aime25-abs}{0.8}
\setval{8b.concur-tied.minus.concur.amc23}{$-$0.4}
\setval{8b.concur-tied.minus.concur.amc23-abs}{0.4}
\setval{8b.concur-tied.minus.concur.avg}{+0.8}
\setval{8b.concur-tied.minus.concur.avg-abs}{0.8}
\setval{8b.concur-tied.minus.concur.code}{+2.4}
\setval{8b.concur-tied.minus.concur.code-abs}{2.4}
\setval{8b.concur-tied.minus.concur.creative-writing-v3}{+2.5}
\setval{8b.concur-tied.minus.concur.creative-writing-v3-abs}{2.5}
\setval{8b.concur-tied.minus.concur.guidebench}{+0.0}
\setval{8b.concur-tied.minus.concur.guidebench-abs}{0.0}
\setval{8b.concur-tied.minus.concur.humaneval-plus}{+4.6}
\setval{8b.concur-tied.minus.concur.humaneval-plus-abs}{4.6}
\setval{8b.concur-tied.minus.concur.if}{+0.5}
\setval{8b.concur-tied.minus.concur.if-abs}{0.5}
\setval{8b.concur-tied.minus.concur.ifbench}{+0.7}
\setval{8b.concur-tied.minus.concur.ifbench-abs}{0.7}
\setval{8b.concur-tied.minus.concur.ifeval}{+0.7}
\setval{8b.concur-tied.minus.concur.ifeval-abs}{0.7}
\setval{8b.concur-tied.minus.concur.judge2.avg}{+1.0}
\setval{8b.concur-tied.minus.concur.judge2.avg-abs}{1.0}
\setval{8b.concur-tied.minus.concur.judge2.creative-writing-v3}{+4.0}
\setval{8b.concur-tied.minus.concur.judge2.creative-writing-v3-abs}{4.0}
\setval{8b.concur-tied.minus.concur.judge2.longbench-write}{$-$0.7}
\setval{8b.concur-tied.minus.concur.judge2.longbench-write-abs}{0.7}
\setval{8b.concur-tied.minus.concur.judge2.writing}{+1.0}
\setval{8b.concur-tied.minus.concur.judge2.writing-abs}{1.0}
\setval{8b.concur-tied.minus.concur.judge2.writingbench}{$-$0.2}
\setval{8b.concur-tied.minus.concur.judge2.writingbench-abs}{0.2}
\setval{8b.concur-tied.minus.concur.livecodebench}{+0.9}
\setval{8b.concur-tied.minus.concur.livecodebench-abs}{0.9}
\setval{8b.concur-tied.minus.concur.longbench-write}{$-$0.4}
\setval{8b.concur-tied.minus.concur.longbench-write-abs}{0.4}
\setval{8b.concur-tied.minus.concur.math}{+0.5}
\setval{8b.concur-tied.minus.concur.math-abs}{0.5}
\setval{8b.concur-tied.minus.concur.math500}{$-$0.5}
\setval{8b.concur-tied.minus.concur.math500-abs}{0.5}
\setval{8b.concur-tied.minus.concur.mbpp-plus}{+1.7}
\setval{8b.concur-tied.minus.concur.mbpp-plus-abs}{1.7}
\setval{8b.concur-tied.minus.concur.minerva}{+1.2}
\setval{8b.concur-tied.minus.concur.minerva-abs}{1.2}
\setval{8b.concur-tied.minus.concur.olympiad-bench}{$-$0.8}
\setval{8b.concur-tied.minus.concur.olympiad-bench-abs}{0.8}
\setval{8b.concur-tied.minus.concur.writing}{+0.4}
\setval{8b.concur-tied.minus.concur.writing-abs}{0.4}
\setval{8b.concur-tied.minus.concur.writingbench}{$-$0.9}
\setval{8b.concur-tied.minus.concur.writingbench-abs}{0.9}
\setval{8b.concur-tied.minus.gd2po.aime24}{+15.0}
\setval{8b.concur-tied.minus.gd2po.aime24-abs}{15.0}
\setval{8b.concur-tied.minus.gd2po.aime25}{+11.8}
\setval{8b.concur-tied.minus.gd2po.aime25-abs}{11.8}
\setval{8b.concur-tied.minus.gd2po.amc23}{+18.9}
\setval{8b.concur-tied.minus.gd2po.amc23-abs}{18.9}
\setval{8b.concur-tied.minus.gd2po.avg}{+6.6}
\setval{8b.concur-tied.minus.gd2po.avg-abs}{6.6}
\setval{8b.concur-tied.minus.gd2po.code}{+1.4}
\setval{8b.concur-tied.minus.gd2po.code-abs}{1.4}
\setval{8b.concur-tied.minus.gd2po.creative-writing-v3}{+4.3}
\setval{8b.concur-tied.minus.gd2po.creative-writing-v3-abs}{4.3}
\setval{8b.concur-tied.minus.gd2po.guidebench}{$-$0.3}
\setval{8b.concur-tied.minus.gd2po.guidebench-abs}{0.3}
\setval{8b.concur-tied.minus.gd2po.humaneval-plus}{+3.4}
\setval{8b.concur-tied.minus.gd2po.humaneval-plus-abs}{3.4}
\setval{8b.concur-tied.minus.gd2po.if}{$-$1.5}
\setval{8b.concur-tied.minus.gd2po.if-abs}{1.5}
\setval{8b.concur-tied.minus.gd2po.ifbench}{$-$3.7}
\setval{8b.concur-tied.minus.gd2po.ifbench-abs}{3.7}
\setval{8b.concur-tied.minus.gd2po.ifeval}{$-$0.5}
\setval{8b.concur-tied.minus.gd2po.ifeval-abs}{0.5}
\setval{8b.concur-tied.minus.gd2po.judge2.avg}{+6.4}
\setval{8b.concur-tied.minus.gd2po.judge2.avg-abs}{6.4}
\setval{8b.concur-tied.minus.gd2po.judge2.creative-writing-v3}{+5.0}
\setval{8b.concur-tied.minus.gd2po.judge2.creative-writing-v3-abs}{5.0}
\setval{8b.concur-tied.minus.gd2po.judge2.longbench-write}{+12.6}
\setval{8b.concur-tied.minus.gd2po.judge2.longbench-write-abs}{12.6}
\setval{8b.concur-tied.minus.gd2po.judge2.writing}{+8.7}
\setval{8b.concur-tied.minus.gd2po.judge2.writing-abs}{8.7}
\setval{8b.concur-tied.minus.gd2po.judge2.writingbench}{+8.5}
\setval{8b.concur-tied.minus.gd2po.judge2.writingbench-abs}{8.5}
\setval{8b.concur-tied.minus.gd2po.livecodebench}{+0.9}
\setval{8b.concur-tied.minus.gd2po.livecodebench-abs}{0.9}
\setval{8b.concur-tied.minus.gd2po.longbench-write}{+11.3}
\setval{8b.concur-tied.minus.gd2po.longbench-write-abs}{11.3}
\setval{8b.concur-tied.minus.gd2po.math}{+11.8}
\setval{8b.concur-tied.minus.gd2po.math-abs}{11.8}
\setval{8b.concur-tied.minus.gd2po.math500}{+8.4}
\setval{8b.concur-tied.minus.gd2po.math500-abs}{8.4}
\setval{8b.concur-tied.minus.gd2po.mbpp-plus}{$-$0.2}
\setval{8b.concur-tied.minus.gd2po.mbpp-plus-abs}{0.2}
\setval{8b.concur-tied.minus.gd2po.minerva}{+5.5}
\setval{8b.concur-tied.minus.gd2po.minerva-abs}{5.5}
\setval{8b.concur-tied.minus.gd2po.olympiad-bench}{+11.0}
\setval{8b.concur-tied.minus.gd2po.olympiad-bench-abs}{11.0}
\setval{8b.concur-tied.minus.gd2po.writing}{+9.6}
\setval{8b.concur-tied.minus.gd2po.writing-abs}{9.6}
\setval{8b.concur-tied.minus.gd2po.writingbench}{+13.3}
\setval{8b.concur-tied.minus.gd2po.writingbench-abs}{13.3}
\setval{8b.concur-tied.minus.gdpo-noaux.aime24}{+3.3}
\setval{8b.concur-tied.minus.gdpo-noaux.aime24-abs}{3.3}
\setval{8b.concur-tied.minus.gdpo-noaux.aime25}{+0.0}
\setval{8b.concur-tied.minus.gdpo-noaux.aime25-abs}{0.0}
\setval{8b.concur-tied.minus.gdpo-noaux.amc23}{+1.4}
\setval{8b.concur-tied.minus.gdpo-noaux.amc23-abs}{1.4}
\setval{8b.concur-tied.minus.gdpo-noaux.avg}{+3.5}
\setval{8b.concur-tied.minus.gdpo-noaux.avg-abs}{3.5}
\setval{8b.concur-tied.minus.gdpo-noaux.code}{+0.4}
\setval{8b.concur-tied.minus.gdpo-noaux.code-abs}{0.4}
\setval{8b.concur-tied.minus.gdpo-noaux.creative-writing-v3}{+8.5}
\setval{8b.concur-tied.minus.gdpo-noaux.creative-writing-v3-abs}{8.5}
\setval{8b.concur-tied.minus.gdpo-noaux.guidebench}{+2.0}
\setval{8b.concur-tied.minus.gdpo-noaux.guidebench-abs}{2.0}
\setval{8b.concur-tied.minus.gdpo-noaux.humaneval-plus}{+1.4}
\setval{8b.concur-tied.minus.gdpo-noaux.humaneval-plus-abs}{1.4}
\setval{8b.concur-tied.minus.gdpo-noaux.if}{+3.3}
\setval{8b.concur-tied.minus.gdpo-noaux.if-abs}{3.3}
\setval{8b.concur-tied.minus.gdpo-noaux.ifbench}{+5.7}
\setval{8b.concur-tied.minus.gdpo-noaux.ifbench-abs}{5.7}
\setval{8b.concur-tied.minus.gdpo-noaux.ifeval}{+2.3}
\setval{8b.concur-tied.minus.gdpo-noaux.ifeval-abs}{2.3}
\setval{8b.concur-tied.minus.gdpo-noaux.judge2.avg}{+4.2}
\setval{8b.concur-tied.minus.gdpo-noaux.judge2.avg-abs}{4.2}
\setval{8b.concur-tied.minus.gdpo-noaux.judge2.creative-writing-v3}{+12.9}
\setval{8b.concur-tied.minus.gdpo-noaux.judge2.creative-writing-v3-abs}{12.9}
\setval{8b.concur-tied.minus.gdpo-noaux.judge2.longbench-write}{+20.3}
\setval{8b.concur-tied.minus.gdpo-noaux.judge2.longbench-write-abs}{20.3}
\setval{8b.concur-tied.minus.gdpo-noaux.judge2.writing}{+15.6}
\setval{8b.concur-tied.minus.gdpo-noaux.judge2.writing-abs}{15.6}
\setval{8b.concur-tied.minus.gdpo-noaux.judge2.writingbench}{+13.7}
\setval{8b.concur-tied.minus.gdpo-noaux.judge2.writingbench-abs}{13.7}
\setval{8b.concur-tied.minus.gdpo-noaux.livecodebench}{$-$0.2}
\setval{8b.concur-tied.minus.gdpo-noaux.livecodebench-abs}{0.2}
\setval{8b.concur-tied.minus.gdpo-noaux.longbench-write}{+15.3}
\setval{8b.concur-tied.minus.gdpo-noaux.longbench-write-abs}{15.3}
\setval{8b.concur-tied.minus.gdpo-noaux.math}{+0.9}
\setval{8b.concur-tied.minus.gdpo-noaux.math-abs}{0.9}
\setval{8b.concur-tied.minus.gdpo-noaux.math500}{$-$0.6}
\setval{8b.concur-tied.minus.gdpo-noaux.math500-abs}{0.6}
\setval{8b.concur-tied.minus.gdpo-noaux.mbpp-plus}{+0.0}
\setval{8b.concur-tied.minus.gdpo-noaux.mbpp-plus-abs}{0.0}
\setval{8b.concur-tied.minus.gdpo-noaux.minerva}{+1.4}
\setval{8b.concur-tied.minus.gdpo-noaux.minerva-abs}{1.4}
\setval{8b.concur-tied.minus.gdpo-noaux.olympiad-bench}{$-$0.4}
\setval{8b.concur-tied.minus.gdpo-noaux.olympiad-bench-abs}{0.4}
\setval{8b.concur-tied.minus.gdpo-noaux.writing}{+12.0}
\setval{8b.concur-tied.minus.gdpo-noaux.writing-abs}{12.0}
\setval{8b.concur-tied.minus.gdpo-noaux.writingbench}{+12.2}
\setval{8b.concur-tied.minus.gdpo-noaux.writingbench-abs}{12.2}
\setval{8b.concur-tied.minus.gdpo-w0.5.aime24}{+12.0}
\setval{8b.concur-tied.minus.gdpo-w0.5.aime24-abs}{12.0}
\setval{8b.concur-tied.minus.gdpo-w0.5.aime25}{+8.5}
\setval{8b.concur-tied.minus.gdpo-w0.5.aime25-abs}{8.5}
\setval{8b.concur-tied.minus.gdpo-w0.5.amc23}{+15.1}
\setval{8b.concur-tied.minus.gdpo-w0.5.amc23-abs}{15.1}
\setval{8b.concur-tied.minus.gdpo-w0.5.avg}{+5.7}
\setval{8b.concur-tied.minus.gdpo-w0.5.avg-abs}{5.7}
\setval{8b.concur-tied.minus.gdpo-w0.5.code}{+2.0}
\setval{8b.concur-tied.minus.gdpo-w0.5.code-abs}{2.0}
\setval{8b.concur-tied.minus.gdpo-w0.5.creative-writing-v3}{+0.9}
\setval{8b.concur-tied.minus.gdpo-w0.5.creative-writing-v3-abs}{0.9}
\setval{8b.concur-tied.minus.gdpo-w0.5.guidebench}{$-$0.6}
\setval{8b.concur-tied.minus.gdpo-w0.5.guidebench-abs}{0.6}
\setval{8b.concur-tied.minus.gdpo-w0.5.humaneval-plus}{+3.6}
\setval{8b.concur-tied.minus.gdpo-w0.5.humaneval-plus-abs}{3.6}
\setval{8b.concur-tied.minus.gdpo-w0.5.if}{+0.7}
\setval{8b.concur-tied.minus.gdpo-w0.5.if-abs}{0.7}
\setval{8b.concur-tied.minus.gdpo-w0.5.ifbench}{+1.7}
\setval{8b.concur-tied.minus.gdpo-w0.5.ifbench-abs}{1.7}
\setval{8b.concur-tied.minus.gdpo-w0.5.ifeval}{+1.1}
\setval{8b.concur-tied.minus.gdpo-w0.5.ifeval-abs}{1.1}
\setval{8b.concur-tied.minus.gdpo-w0.5.judge2.avg}{+5.5}
\setval{8b.concur-tied.minus.gdpo-w0.5.judge2.avg-abs}{5.5}
\setval{8b.concur-tied.minus.gdpo-w0.5.judge2.creative-writing-v3}{+1.4}
\setval{8b.concur-tied.minus.gdpo-w0.5.judge2.creative-writing-v3-abs}{1.4}
\setval{8b.concur-tied.minus.gdpo-w0.5.judge2.longbench-write}{+11.1}
\setval{8b.concur-tied.minus.gdpo-w0.5.judge2.longbench-write-abs}{11.1}
\setval{8b.concur-tied.minus.gdpo-w0.5.judge2.writing}{+6.7}
\setval{8b.concur-tied.minus.gdpo-w0.5.judge2.writing-abs}{6.7}
\setval{8b.concur-tied.minus.gdpo-w0.5.judge2.writingbench}{+7.5}
\setval{8b.concur-tied.minus.gdpo-w0.5.judge2.writingbench-abs}{7.5}
\setval{8b.concur-tied.minus.gdpo-w0.5.livecodebench}{+0.4}
\setval{8b.concur-tied.minus.gdpo-w0.5.livecodebench-abs}{0.4}
\setval{8b.concur-tied.minus.gdpo-w0.5.longbench-write}{+10.0}
\setval{8b.concur-tied.minus.gdpo-w0.5.longbench-write-abs}{10.0}
\setval{8b.concur-tied.minus.gdpo-w0.5.math}{+9.0}
\setval{8b.concur-tied.minus.gdpo-w0.5.math-abs}{9.0}
\setval{8b.concur-tied.minus.gdpo-w0.5.math500}{+5.5}
\setval{8b.concur-tied.minus.gdpo-w0.5.math500-abs}{5.5}
\setval{8b.concur-tied.minus.gdpo-w0.5.mbpp-plus}{+2.0}
\setval{8b.concur-tied.minus.gdpo-w0.5.mbpp-plus-abs}{2.0}
\setval{8b.concur-tied.minus.gdpo-w0.5.minerva}{+4.2}
\setval{8b.concur-tied.minus.gdpo-w0.5.minerva-abs}{4.2}
\setval{8b.concur-tied.minus.gdpo-w0.5.olympiad-bench}{+8.4}
\setval{8b.concur-tied.minus.gdpo-w0.5.olympiad-bench-abs}{8.4}
\setval{8b.concur-tied.minus.gdpo-w0.5.writing}{+7.7}
\setval{8b.concur-tied.minus.gdpo-w0.5.writing-abs}{7.7}
\setval{8b.concur-tied.minus.gdpo-w0.5.writingbench}{+12.2}
\setval{8b.concur-tied.minus.gdpo-w0.5.writingbench-abs}{12.2}
\setval{8b.concur-tied.minus.gdpo.aime24}{+9.9}
\setval{8b.concur-tied.minus.gdpo.aime24-abs}{9.9}
\setval{8b.concur-tied.minus.gdpo.aime25}{+9.6}
\setval{8b.concur-tied.minus.gdpo.aime25-abs}{9.6}
\setval{8b.concur-tied.minus.gdpo.amc23}{+16.3}
\setval{8b.concur-tied.minus.gdpo.amc23-abs}{16.3}
\setval{8b.concur-tied.minus.gdpo.avg}{+5.7}
\setval{8b.concur-tied.minus.gdpo.avg-abs}{5.7}
\setval{8b.concur-tied.minus.gdpo.code}{+1.5}
\setval{8b.concur-tied.minus.gdpo.code-abs}{1.5}
\setval{8b.concur-tied.minus.gdpo.creative-writing-v3}{$-$0.5}
\setval{8b.concur-tied.minus.gdpo.creative-writing-v3-abs}{0.5}
\setval{8b.concur-tied.minus.gdpo.guidebench}{$-$1.2}
\setval{8b.concur-tied.minus.gdpo.guidebench-abs}{1.2}
\setval{8b.concur-tied.minus.gdpo.humaneval-plus}{+3.3}
\setval{8b.concur-tied.minus.gdpo.humaneval-plus-abs}{3.3}
\setval{8b.concur-tied.minus.gdpo.if}{$-$0.1}
\setval{8b.concur-tied.minus.gdpo.if-abs}{0.1}
\setval{8b.concur-tied.minus.gdpo.ifbench}{+0.0}
\setval{8b.concur-tied.minus.gdpo.ifbench-abs}{0.0}
\setval{8b.concur-tied.minus.gdpo.ifeval}{+0.7}
\setval{8b.concur-tied.minus.gdpo.ifeval-abs}{0.7}
\setval{8b.concur-tied.minus.gdpo.judge2.avg}{+5.3}
\setval{8b.concur-tied.minus.gdpo.judge2.avg-abs}{5.3}
\setval{8b.concur-tied.minus.gdpo.judge2.creative-writing-v3}{+0.8}
\setval{8b.concur-tied.minus.gdpo.judge2.creative-writing-v3-abs}{0.8}
\setval{8b.concur-tied.minus.gdpo.judge2.longbench-write}{+10.6}
\setval{8b.concur-tied.minus.gdpo.judge2.longbench-write-abs}{10.6}
\setval{8b.concur-tied.minus.gdpo.judge2.writing}{+5.6}
\setval{8b.concur-tied.minus.gdpo.judge2.writing-abs}{5.6}
\setval{8b.concur-tied.minus.gdpo.judge2.writingbench}{+5.4}
\setval{8b.concur-tied.minus.gdpo.judge2.writingbench-abs}{5.4}
\setval{8b.concur-tied.minus.gdpo.livecodebench}{+1.0}
\setval{8b.concur-tied.minus.gdpo.livecodebench-abs}{1.0}
\setval{8b.concur-tied.minus.gdpo.longbench-write}{+11.6}
\setval{8b.concur-tied.minus.gdpo.longbench-write-abs}{11.6}
\setval{8b.concur-tied.minus.gdpo.math}{+9.7}
\setval{8b.concur-tied.minus.gdpo.math-abs}{9.7}
\setval{8b.concur-tied.minus.gdpo.math500}{+6.6}
\setval{8b.concur-tied.minus.gdpo.math500-abs}{6.6}
\setval{8b.concur-tied.minus.gdpo.mbpp-plus}{+0.3}
\setval{8b.concur-tied.minus.gdpo.mbpp-plus-abs}{0.3}
\setval{8b.concur-tied.minus.gdpo.minerva}{+5.8}
\setval{8b.concur-tied.minus.gdpo.minerva-abs}{5.8}
\setval{8b.concur-tied.minus.gdpo.olympiad-bench}{+10.2}
\setval{8b.concur-tied.minus.gdpo.olympiad-bench-abs}{10.2}
\setval{8b.concur-tied.minus.gdpo.writing}{+7.4}
\setval{8b.concur-tied.minus.gdpo.writing-abs}{7.4}
\setval{8b.concur-tied.minus.gdpo.writingbench}{+11.1}
\setval{8b.concur-tied.minus.gdpo.writingbench-abs}{11.1}
\setval{8b.concur-tied.minus.geoma.aime24}{+17.2}
\setval{8b.concur-tied.minus.geoma.aime24-abs}{17.2}
\setval{8b.concur-tied.minus.geoma.aime25}{+11.5}
\setval{8b.concur-tied.minus.geoma.aime25-abs}{11.5}
\setval{8b.concur-tied.minus.geoma.amc23}{+15.0}
\setval{8b.concur-tied.minus.geoma.amc23-abs}{15.0}
\setval{8b.concur-tied.minus.geoma.avg}{+7.3}
\setval{8b.concur-tied.minus.geoma.avg-abs}{7.3}
\setval{8b.concur-tied.minus.geoma.code}{+1.8}
\setval{8b.concur-tied.minus.geoma.code-abs}{1.8}
\setval{8b.concur-tied.minus.geoma.creative-writing-v3}{+2.0}
\setval{8b.concur-tied.minus.geoma.creative-writing-v3-abs}{2.0}
\setval{8b.concur-tied.minus.geoma.guidebench}{$-$1.0}
\setval{8b.concur-tied.minus.geoma.guidebench-abs}{1.0}
\setval{8b.concur-tied.minus.geoma.humaneval-plus}{+4.6}
\setval{8b.concur-tied.minus.geoma.humaneval-plus-abs}{4.6}
\setval{8b.concur-tied.minus.geoma.if}{+2.0}
\setval{8b.concur-tied.minus.geoma.if-abs}{2.0}
\setval{8b.concur-tied.minus.geoma.ifbench}{+5.3}
\setval{8b.concur-tied.minus.geoma.ifbench-abs}{5.3}
\setval{8b.concur-tied.minus.geoma.ifeval}{+1.6}
\setval{8b.concur-tied.minus.geoma.ifeval-abs}{1.6}
\setval{8b.concur-tied.minus.geoma.judge2.avg}{+6.9}
\setval{8b.concur-tied.minus.geoma.judge2.avg-abs}{6.9}
\setval{8b.concur-tied.minus.geoma.judge2.creative-writing-v3}{+4.2}
\setval{8b.concur-tied.minus.geoma.judge2.creative-writing-v3-abs}{4.2}
\setval{8b.concur-tied.minus.geoma.judge2.longbench-write}{+9.9}
\setval{8b.concur-tied.minus.geoma.judge2.longbench-write-abs}{9.9}
\setval{8b.concur-tied.minus.geoma.judge2.writing}{+6.5}
\setval{8b.concur-tied.minus.geoma.judge2.writing-abs}{6.5}
\setval{8b.concur-tied.minus.geoma.judge2.writingbench}{+5.5}
\setval{8b.concur-tied.minus.geoma.judge2.writingbench-abs}{5.5}
\setval{8b.concur-tied.minus.geoma.livecodebench}{+1.5}
\setval{8b.concur-tied.minus.geoma.livecodebench-abs}{1.5}
\setval{8b.concur-tied.minus.geoma.longbench-write}{+12.3}
\setval{8b.concur-tied.minus.geoma.longbench-write-abs}{12.3}
\setval{8b.concur-tied.minus.geoma.math}{+12.1}
\setval{8b.concur-tied.minus.geoma.math-abs}{12.1}
\setval{8b.concur-tied.minus.geoma.math500}{+10.0}
\setval{8b.concur-tied.minus.geoma.math500-abs}{10.0}
\setval{8b.concur-tied.minus.geoma.mbpp-plus}{$-$0.7}
\setval{8b.concur-tied.minus.geoma.mbpp-plus-abs}{0.7}
\setval{8b.concur-tied.minus.geoma.minerva}{+5.6}
\setval{8b.concur-tied.minus.geoma.minerva-abs}{5.6}
\setval{8b.concur-tied.minus.geoma.olympiad-bench}{+13.6}
\setval{8b.concur-tied.minus.geoma.olympiad-bench-abs}{13.6}
\setval{8b.concur-tied.minus.geoma.writing}{+8.5}
\setval{8b.concur-tied.minus.geoma.writing-abs}{8.5}
\setval{8b.concur-tied.minus.geoma.writingbench}{+11.3}
\setval{8b.concur-tied.minus.geoma.writingbench-abs}{11.3}
\setval{8b.concur-tied.minus.grpo-nodiv.aime24}{+10.6}
\setval{8b.concur-tied.minus.grpo-nodiv.aime24-abs}{10.6}
\setval{8b.concur-tied.minus.grpo-nodiv.aime25}{+9.2}
\setval{8b.concur-tied.minus.grpo-nodiv.aime25-abs}{9.2}
\setval{8b.concur-tied.minus.grpo-nodiv.amc23}{+14.9}
\setval{8b.concur-tied.minus.grpo-nodiv.amc23-abs}{14.9}
\setval{8b.concur-tied.minus.grpo-nodiv.avg}{+5.9}
\setval{8b.concur-tied.minus.grpo-nodiv.avg-abs}{5.9}
\setval{8b.concur-tied.minus.grpo-nodiv.code}{+0.7}
\setval{8b.concur-tied.minus.grpo-nodiv.code-abs}{0.7}
\setval{8b.concur-tied.minus.grpo-nodiv.creative-writing-v3}{+5.8}
\setval{8b.concur-tied.minus.grpo-nodiv.creative-writing-v3-abs}{5.8}
\setval{8b.concur-tied.minus.grpo-nodiv.guidebench}{$-$1.4}
\setval{8b.concur-tied.minus.grpo-nodiv.guidebench-abs}{1.4}
\setval{8b.concur-tied.minus.grpo-nodiv.humaneval-plus}{+1.0}
\setval{8b.concur-tied.minus.grpo-nodiv.humaneval-plus-abs}{1.0}
\setval{8b.concur-tied.minus.grpo-nodiv.if}{+0.3}
\setval{8b.concur-tied.minus.grpo-nodiv.if-abs}{0.3}
\setval{8b.concur-tied.minus.grpo-nodiv.ifbench}{+3.0}
\setval{8b.concur-tied.minus.grpo-nodiv.ifbench-abs}{3.0}
\setval{8b.concur-tied.minus.grpo-nodiv.ifeval}{$-$0.6}
\setval{8b.concur-tied.minus.grpo-nodiv.ifeval-abs}{0.6}
\setval{8b.concur-tied.minus.grpo-nodiv.judge2.avg}{+5.2}
\setval{8b.concur-tied.minus.grpo-nodiv.judge2.avg-abs}{5.2}
\setval{8b.concur-tied.minus.grpo-nodiv.judge2.creative-writing-v3}{+4.3}
\setval{8b.concur-tied.minus.grpo-nodiv.judge2.creative-writing-v3-abs}{4.3}
\setval{8b.concur-tied.minus.grpo-nodiv.judge2.longbench-write}{+8.2}
\setval{8b.concur-tied.minus.grpo-nodiv.judge2.longbench-write-abs}{8.2}
\setval{8b.concur-tied.minus.grpo-nodiv.judge2.writing}{+7.5}
\setval{8b.concur-tied.minus.grpo-nodiv.judge2.writing-abs}{7.5}
\setval{8b.concur-tied.minus.grpo-nodiv.judge2.writingbench}{+9.9}
\setval{8b.concur-tied.minus.grpo-nodiv.judge2.writingbench-abs}{9.9}
\setval{8b.concur-tied.minus.grpo-nodiv.livecodebench}{+2.2}
\setval{8b.concur-tied.minus.grpo-nodiv.livecodebench-abs}{2.2}
\setval{8b.concur-tied.minus.grpo-nodiv.longbench-write}{+10.2}
\setval{8b.concur-tied.minus.grpo-nodiv.longbench-write-abs}{10.2}
\setval{8b.concur-tied.minus.grpo-nodiv.math}{+8.8}
\setval{8b.concur-tied.minus.grpo-nodiv.math-abs}{8.8}
\setval{8b.concur-tied.minus.grpo-nodiv.math500}{+5.5}
\setval{8b.concur-tied.minus.grpo-nodiv.math500-abs}{5.5}
\setval{8b.concur-tied.minus.grpo-nodiv.mbpp-plus}{$-$1.2}
\setval{8b.concur-tied.minus.grpo-nodiv.mbpp-plus-abs}{1.2}
\setval{8b.concur-tied.minus.grpo-nodiv.minerva}{+4.6}
\setval{8b.concur-tied.minus.grpo-nodiv.minerva-abs}{4.6}
\setval{8b.concur-tied.minus.grpo-nodiv.olympiad-bench}{+8.0}
\setval{8b.concur-tied.minus.grpo-nodiv.olympiad-bench-abs}{8.0}
\setval{8b.concur-tied.minus.grpo-nodiv.writing}{+10.9}
\setval{8b.concur-tied.minus.grpo-nodiv.writing-abs}{10.9}
\setval{8b.concur-tied.minus.grpo-nodiv.writingbench}{+16.9}
\setval{8b.concur-tied.minus.grpo-nodiv.writingbench-abs}{16.9}
\setval{8b.concur-tied.minus.grpo.aime24}{+15.4}
\setval{8b.concur-tied.minus.grpo.aime24-abs}{15.4}
\setval{8b.concur-tied.minus.grpo.aime25}{+8.4}
\setval{8b.concur-tied.minus.grpo.aime25-abs}{8.4}
\setval{8b.concur-tied.minus.grpo.amc23}{+14.1}
\setval{8b.concur-tied.minus.grpo.amc23-abs}{14.1}
\setval{8b.concur-tied.minus.grpo.avg}{+6.1}
\setval{8b.concur-tied.minus.grpo.avg-abs}{6.1}
\setval{8b.concur-tied.minus.grpo.code}{+1.5}
\setval{8b.concur-tied.minus.grpo.code-abs}{1.5}
\setval{8b.concur-tied.minus.grpo.creative-writing-v3}{$-$0.1}
\setval{8b.concur-tied.minus.grpo.creative-writing-v3-abs}{0.1}
\setval{8b.concur-tied.minus.grpo.guidebench}{+1.5}
\setval{8b.concur-tied.minus.grpo.guidebench-abs}{1.5}
\setval{8b.concur-tied.minus.grpo.humaneval-plus}{+3.7}
\setval{8b.concur-tied.minus.grpo.humaneval-plus-abs}{3.7}
\setval{8b.concur-tied.minus.grpo.if}{+0.3}
\setval{8b.concur-tied.minus.grpo.if-abs}{0.3}
\setval{8b.concur-tied.minus.grpo.ifbench}{$-$2.7}
\setval{8b.concur-tied.minus.grpo.ifbench-abs}{2.7}
\setval{8b.concur-tied.minus.grpo.ifeval}{+2.0}
\setval{8b.concur-tied.minus.grpo.ifeval-abs}{2.0}
\setval{8b.concur-tied.minus.grpo.judge2.avg}{+6.1}
\setval{8b.concur-tied.minus.grpo.judge2.avg-abs}{6.1}
\setval{8b.concur-tied.minus.grpo.judge2.creative-writing-v3}{+2.5}
\setval{8b.concur-tied.minus.grpo.judge2.creative-writing-v3-abs}{2.5}
\setval{8b.concur-tied.minus.grpo.judge2.longbench-write}{+12.3}
\setval{8b.concur-tied.minus.grpo.judge2.longbench-write-abs}{12.3}
\setval{8b.concur-tied.minus.grpo.judge2.writing}{+7.2}
\setval{8b.concur-tied.minus.grpo.judge2.writing-abs}{7.2}
\setval{8b.concur-tied.minus.grpo.judge2.writingbench}{+6.9}
\setval{8b.concur-tied.minus.grpo.judge2.writingbench-abs}{6.9}
\setval{8b.concur-tied.minus.grpo.livecodebench}{+0.7}
\setval{8b.concur-tied.minus.grpo.livecodebench-abs}{0.7}
\setval{8b.concur-tied.minus.grpo.longbench-write}{+11.4}
\setval{8b.concur-tied.minus.grpo.longbench-write-abs}{11.4}
\setval{8b.concur-tied.minus.grpo.math}{+10.7}
\setval{8b.concur-tied.minus.grpo.math-abs}{10.7}
\setval{8b.concur-tied.minus.grpo.math500}{+9.0}
\setval{8b.concur-tied.minus.grpo.math500-abs}{9.0}
\setval{8b.concur-tied.minus.grpo.mbpp-plus}{+0.2}
\setval{8b.concur-tied.minus.grpo.mbpp-plus-abs}{0.2}
\setval{8b.concur-tied.minus.grpo.minerva}{+5.5}
\setval{8b.concur-tied.minus.grpo.minerva-abs}{5.5}
\setval{8b.concur-tied.minus.grpo.olympiad-bench}{+11.6}
\setval{8b.concur-tied.minus.grpo.olympiad-bench-abs}{11.6}
\setval{8b.concur-tied.minus.grpo.writing}{+7.4}
\setval{8b.concur-tied.minus.grpo.writing-abs}{7.4}
\setval{8b.concur-tied.minus.grpo.writingbench}{+10.7}
\setval{8b.concur-tied.minus.grpo.writingbench-abs}{10.7}
\setval{8b.concur-tied.minus.mt-grpo.aime24}{+16.8}
\setval{8b.concur-tied.minus.mt-grpo.aime24-abs}{16.8}
\setval{8b.concur-tied.minus.mt-grpo.aime25}{+8.2}
\setval{8b.concur-tied.minus.mt-grpo.aime25-abs}{8.2}
\setval{8b.concur-tied.minus.mt-grpo.amc23}{+12.9}
\setval{8b.concur-tied.minus.mt-grpo.amc23-abs}{12.9}
\setval{8b.concur-tied.minus.mt-grpo.avg}{+6.8}
\setval{8b.concur-tied.minus.mt-grpo.avg-abs}{6.8}
\setval{8b.concur-tied.minus.mt-grpo.code}{+0.5}
\setval{8b.concur-tied.minus.mt-grpo.code-abs}{0.5}
\setval{8b.concur-tied.minus.mt-grpo.creative-writing-v3}{+2.1}
\setval{8b.concur-tied.minus.mt-grpo.creative-writing-v3-abs}{2.1}
\setval{8b.concur-tied.minus.mt-grpo.guidebench}{+0.3}
\setval{8b.concur-tied.minus.mt-grpo.guidebench-abs}{0.3}
\setval{8b.concur-tied.minus.mt-grpo.humaneval-plus}{+2.3}
\setval{8b.concur-tied.minus.mt-grpo.humaneval-plus-abs}{2.3}
\setval{8b.concur-tied.minus.mt-grpo.if}{+1.4}
\setval{8b.concur-tied.minus.mt-grpo.if-abs}{1.4}
\setval{8b.concur-tied.minus.mt-grpo.ifbench}{+2.0}
\setval{8b.concur-tied.minus.mt-grpo.ifbench-abs}{2.0}
\setval{8b.concur-tied.minus.mt-grpo.ifeval}{+1.9}
\setval{8b.concur-tied.minus.mt-grpo.ifeval-abs}{1.9}
\setval{8b.concur-tied.minus.mt-grpo.judge2.avg}{+6.1}
\setval{8b.concur-tied.minus.mt-grpo.judge2.avg-abs}{6.1}
\setval{8b.concur-tied.minus.mt-grpo.judge2.creative-writing-v3}{+0.2}
\setval{8b.concur-tied.minus.mt-grpo.judge2.creative-writing-v3-abs}{0.2}
\setval{8b.concur-tied.minus.mt-grpo.judge2.longbench-write}{+12.3}
\setval{8b.concur-tied.minus.mt-grpo.judge2.longbench-write-abs}{12.3}
\setval{8b.concur-tied.minus.mt-grpo.judge2.writing}{+6.9}
\setval{8b.concur-tied.minus.mt-grpo.judge2.writing-abs}{6.9}
\setval{8b.concur-tied.minus.mt-grpo.judge2.writingbench}{+8.3}
\setval{8b.concur-tied.minus.mt-grpo.judge2.writingbench-abs}{8.3}
\setval{8b.concur-tied.minus.mt-grpo.livecodebench}{$-$0.5}
\setval{8b.concur-tied.minus.mt-grpo.livecodebench-abs}{0.5}
\setval{8b.concur-tied.minus.mt-grpo.longbench-write}{+13.8}
\setval{8b.concur-tied.minus.mt-grpo.longbench-write-abs}{13.8}
\setval{8b.concur-tied.minus.mt-grpo.math}{+10.8}
\setval{8b.concur-tied.minus.mt-grpo.math-abs}{10.8}
\setval{8b.concur-tied.minus.mt-grpo.math500}{+9.0}
\setval{8b.concur-tied.minus.mt-grpo.math500-abs}{9.0}
\setval{8b.concur-tied.minus.mt-grpo.mbpp-plus}{$-$0.4}
\setval{8b.concur-tied.minus.mt-grpo.mbpp-plus-abs}{0.4}
\setval{8b.concur-tied.minus.mt-grpo.minerva}{+4.4}
\setval{8b.concur-tied.minus.mt-grpo.minerva-abs}{4.4}
\setval{8b.concur-tied.minus.mt-grpo.olympiad-bench}{+13.8}
\setval{8b.concur-tied.minus.mt-grpo.olympiad-bench-abs}{13.8}
\setval{8b.concur-tied.minus.mt-grpo.writing}{+10.5}
\setval{8b.concur-tied.minus.mt-grpo.writing-abs}{10.5}
\setval{8b.concur-tied.minus.mt-grpo.writingbench}{+15.5}
\setval{8b.concur-tied.minus.mt-grpo.writingbench-abs}{15.5}
\setval{8b.concur-tied.minus.papo.aime24}{+19.1}
\setval{8b.concur-tied.minus.papo.aime24-abs}{19.1}
\setval{8b.concur-tied.minus.papo.aime25}{+13.9}
\setval{8b.concur-tied.minus.papo.aime25-abs}{13.9}
\setval{8b.concur-tied.minus.papo.amc23}{+16.5}
\setval{8b.concur-tied.minus.papo.amc23-abs}{16.5}
\setval{8b.concur-tied.minus.papo.avg}{+8.2}
\setval{8b.concur-tied.minus.papo.avg-abs}{8.2}
\setval{8b.concur-tied.minus.papo.code}{+2.3}
\setval{8b.concur-tied.minus.papo.code-abs}{2.3}
\setval{8b.concur-tied.minus.papo.creative-writing-v3}{+1.3}
\setval{8b.concur-tied.minus.papo.creative-writing-v3-abs}{1.3}
\setval{8b.concur-tied.minus.papo.guidebench}{$-$1.0}
\setval{8b.concur-tied.minus.papo.guidebench-abs}{1.0}
\setval{8b.concur-tied.minus.papo.humaneval-plus}{+4.9}
\setval{8b.concur-tied.minus.papo.humaneval-plus-abs}{4.9}
\setval{8b.concur-tied.minus.papo.if}{+1.2}
\setval{8b.concur-tied.minus.papo.if-abs}{1.2}
\setval{8b.concur-tied.minus.papo.ifbench}{+2.0}
\setval{8b.concur-tied.minus.papo.ifbench-abs}{2.0}
\setval{8b.concur-tied.minus.papo.ifeval}{+2.5}
\setval{8b.concur-tied.minus.papo.ifeval-abs}{2.5}
\setval{8b.concur-tied.minus.papo.judge2.avg}{+7.6}
\setval{8b.concur-tied.minus.papo.judge2.avg-abs}{7.6}
\setval{8b.concur-tied.minus.papo.judge2.creative-writing-v3}{$-$0.1}
\setval{8b.concur-tied.minus.papo.judge2.creative-writing-v3-abs}{0.1}
\setval{8b.concur-tied.minus.papo.judge2.longbench-write}{+15.2}
\setval{8b.concur-tied.minus.papo.judge2.longbench-write-abs}{15.2}
\setval{8b.concur-tied.minus.papo.judge2.writing}{+7.5}
\setval{8b.concur-tied.minus.papo.judge2.writing-abs}{7.5}
\setval{8b.concur-tied.minus.papo.judge2.writingbench}{+7.5}
\setval{8b.concur-tied.minus.papo.judge2.writingbench-abs}{7.5}
\setval{8b.concur-tied.minus.papo.livecodebench}{+0.6}
\setval{8b.concur-tied.minus.papo.livecodebench-abs}{0.6}
\setval{8b.concur-tied.minus.papo.longbench-write}{+15.7}
\setval{8b.concur-tied.minus.papo.longbench-write-abs}{15.7}
\setval{8b.concur-tied.minus.papo.math}{+13.6}
\setval{8b.concur-tied.minus.papo.math-abs}{13.6}
\setval{8b.concur-tied.minus.papo.math500}{+11.1}
\setval{8b.concur-tied.minus.papo.math500-abs}{11.1}
\setval{8b.concur-tied.minus.papo.mbpp-plus}{+1.5}
\setval{8b.concur-tied.minus.papo.mbpp-plus-abs}{1.5}
\setval{8b.concur-tied.minus.papo.minerva}{+6.1}
\setval{8b.concur-tied.minus.papo.minerva-abs}{6.1}
\setval{8b.concur-tied.minus.papo.olympiad-bench}{+14.7}
\setval{8b.concur-tied.minus.papo.olympiad-bench-abs}{14.7}
\setval{8b.concur-tied.minus.papo.writing}{+10.3}
\setval{8b.concur-tied.minus.papo.writing-abs}{10.3}
\setval{8b.concur-tied.minus.papo.writingbench}{+14.0}
\setval{8b.concur-tied.minus.papo.writingbench-abs}{14.0}
\setval{8b.concur-tied.minus.saw.aime24}{+15.9}
\setval{8b.concur-tied.minus.saw.aime24-abs}{15.9}
\setval{8b.concur-tied.minus.saw.aime25}{+8.4}
\setval{8b.concur-tied.minus.saw.aime25-abs}{8.4}
\setval{8b.concur-tied.minus.saw.amc23}{+16.2}
\setval{8b.concur-tied.minus.saw.amc23-abs}{16.2}
\setval{8b.concur-tied.minus.saw.avg}{+6.0}
\setval{8b.concur-tied.minus.saw.avg-abs}{6.0}
\setval{8b.concur-tied.minus.saw.code}{+0.9}
\setval{8b.concur-tied.minus.saw.code-abs}{0.9}
\setval{8b.concur-tied.minus.saw.creative-writing-v3}{+2.8}
\setval{8b.concur-tied.minus.saw.creative-writing-v3-abs}{2.8}
\setval{8b.concur-tied.minus.saw.guidebench}{$-$0.9}
\setval{8b.concur-tied.minus.saw.guidebench-abs}{0.9}
\setval{8b.concur-tied.minus.saw.humaneval-plus}{+1.5}
\setval{8b.concur-tied.minus.saw.humaneval-plus-abs}{1.5}
\setval{8b.concur-tied.minus.saw.if}{$-$0.1}
\setval{8b.concur-tied.minus.saw.if-abs}{0.1}
\setval{8b.concur-tied.minus.saw.ifbench}{$-$0.7}
\setval{8b.concur-tied.minus.saw.ifbench-abs}{0.7}
\setval{8b.concur-tied.minus.saw.ifeval}{+1.2}
\setval{8b.concur-tied.minus.saw.ifeval-abs}{1.2}
\setval{8b.concur-tied.minus.saw.judge2.avg}{+5.8}
\setval{8b.concur-tied.minus.saw.judge2.avg-abs}{5.8}
\setval{8b.concur-tied.minus.saw.judge2.creative-writing-v3}{+5.2}
\setval{8b.concur-tied.minus.saw.judge2.creative-writing-v3-abs}{5.2}
\setval{8b.concur-tied.minus.saw.judge2.longbench-write}{+10.0}
\setval{8b.concur-tied.minus.saw.judge2.longbench-write-abs}{10.0}
\setval{8b.concur-tied.minus.saw.judge2.writing}{+7.3}
\setval{8b.concur-tied.minus.saw.judge2.writing-abs}{7.3}
\setval{8b.concur-tied.minus.saw.judge2.writingbench}{+6.8}
\setval{8b.concur-tied.minus.saw.judge2.writingbench-abs}{6.8}
\setval{8b.concur-tied.minus.saw.livecodebench}{+0.5}
\setval{8b.concur-tied.minus.saw.livecodebench-abs}{0.5}
\setval{8b.concur-tied.minus.saw.longbench-write}{+10.2}
\setval{8b.concur-tied.minus.saw.longbench-write-abs}{10.2}
\setval{8b.concur-tied.minus.saw.math}{+10.6}
\setval{8b.concur-tied.minus.saw.math-abs}{10.6}
\setval{8b.concur-tied.minus.saw.math500}{+7.7}
\setval{8b.concur-tied.minus.saw.math500-abs}{7.7}
\setval{8b.concur-tied.minus.saw.mbpp-plus}{+0.5}
\setval{8b.concur-tied.minus.saw.mbpp-plus-abs}{0.5}
\setval{8b.concur-tied.minus.saw.minerva}{+3.6}
\setval{8b.concur-tied.minus.saw.minerva-abs}{3.6}
\setval{8b.concur-tied.minus.saw.olympiad-bench}{+11.4}
\setval{8b.concur-tied.minus.saw.olympiad-bench-abs}{11.4}
\setval{8b.concur-tied.minus.saw.writing}{+8.2}
\setval{8b.concur-tied.minus.saw.writing-abs}{8.2}
\setval{8b.concur-tied.minus.saw.writingbench}{+11.8}
\setval{8b.concur-tied.minus.saw.writingbench-abs}{11.8}
\setval{8b.concur-tied.passing-at-ceiling.code.early}{0.71}
\setval{8b.concur-tied.passing-at-ceiling.code.late}{0.75}
\setval{8b.concur-tied.passing-at-ceiling.code.run}{0.74}
\setval{8b.concur-tied.passing-at-ceiling.if.early}{0.32}
\setval{8b.concur-tied.passing-at-ceiling.if.late}{0.55}
\setval{8b.concur-tied.passing-at-ceiling.if.run}{0.47}
\setval{8b.concur-tied.passing-at-ceiling.math.early}{1.00}
\setval{8b.concur-tied.passing-at-ceiling.math.late}{1.00}
\setval{8b.concur-tied.passing-at-ceiling.math.run}{1.00}
\setval{8b.concur-tied.passing-at-ceiling.writing.early}{0.00}
\setval{8b.concur-tied.passing-at-ceiling.writing.late}{0.02}
\setval{8b.concur-tied.passing-at-ceiling.writing.run}{0.01}
\setval{8b.concur-tied.primary-cosine.code.early}{0.98}
\setval{8b.concur-tied.primary-cosine.code.late}{0.95}
\setval{8b.concur-tied.primary-cosine.code.run}{0.96}
\setval{8b.concur-tied.primary-cosine.if.early}{0.86}
\setval{8b.concur-tied.primary-cosine.if.late}{0.81}
\setval{8b.concur-tied.primary-cosine.if.run}{0.85}
\setval{8b.concur-tied.primary-cosine.math.early}{0.99}
\setval{8b.concur-tied.primary-cosine.math.late}{0.97}
\setval{8b.concur-tied.primary-cosine.math.run}{0.97}
\setval{8b.concur-tied.primary-cosine.writing.early}{0.94}
\setval{8b.concur-tied.primary-cosine.writing.late}{0.92}
\setval{8b.concur-tied.primary-cosine.writing.run}{0.92}
\setval{8b.concur-tied.primary-level.code.early}{0.54}
\setval{8b.concur-tied.primary-level.code.late}{0.76}
\setval{8b.concur-tied.primary-level.code.run}{0.71}
\setval{8b.concur-tied.primary-level.if.early}{0.35}
\setval{8b.concur-tied.primary-level.if.late}{0.74}
\setval{8b.concur-tied.primary-level.if.run}{0.63}
\setval{8b.concur-tied.primary-level.math.early}{0.49}
\setval{8b.concur-tied.primary-level.math.late}{0.74}
\setval{8b.concur-tied.primary-level.math.run}{0.69}
\setval{8b.concur-tied.primary-level.writing.early}{0.54}
\setval{8b.concur-tied.primary-level.writing.late}{0.81}
\setval{8b.concur-tied.primary-level.writing.run}{0.76}
\setval{8b.concur-tied.solved.code.early}{0.11}
\setval{8b.concur-tied.solved.code.late}{0.18}
\setval{8b.concur-tied.solved.code.run}{0.19}
\setval{8b.concur-tied.solved.if.early}{0.03}
\setval{8b.concur-tied.solved.if.late}{0.20}
\setval{8b.concur-tied.solved.if.run}{0.13}
\setval{8b.concur-tied.solved.math.early}{0.15}
\setval{8b.concur-tied.solved.math.late}{0.46}
\setval{8b.concur-tied.solved.math.run}{0.39}
\setval{8b.concur-tied.solved.writing.early}{0.00}
\setval{8b.concur-tied.solved.writing.late}{0.00}
\setval{8b.concur-tied.solved.writing.run}{0.00}
\setval{8b.concur-tied.trust.code.agreement}{+0.005}
\setval{8b.concur-tied.trust.code.errors}{+1.2}
\setval{8b.concur-tied.trust.code.first-free}{--}
\setval{8b.concur-tied.trust.code.free-share}{0.00}
\setval{8b.concur-tied.trust.if.agreement}{+0.022}
\setval{8b.concur-tied.trust.if.errors}{+8.4}
\setval{8b.concur-tied.trust.if.first-free}{13}
\setval{8b.concur-tied.trust.if.free-share}{0.96}
\setval{8b.concur-tied.trust.math.first-free}{--}
\setval{8b.concur-tied.trust.math.free-share}{0.00}
\setval{8b.concur-tied.trust.writing-rm.agreement}{+0.050}
\setval{8b.concur-tied.trust.writing-rm.errors}{+30.5}
\setval{8b.concur-tied.trust.writing-rm.first-free}{1}
\setval{8b.concur-tied.trust.writing-rm.free-share}{1.00}
\setval{8b.concur-tied.trust.writing.agreement}{$-$0.047}
\setval{8b.concur-tied.trust.writing.errors}{$-$34.0}
\setval{8b.concur-tied.trust.writing.first-free}{--}
\setval{8b.concur-tied.trust.writing.free-share}{0.00}
\setval{8b.concur-tied.update-share.code.early}{0.19}
\setval{8b.concur-tied.update-share.code.late}{0.26}
\setval{8b.concur-tied.update-share.code.run}{0.25}
\setval{8b.concur-tied.update-share.if.early}{0.22}
\setval{8b.concur-tied.update-share.if.late}{0.20}
\setval{8b.concur-tied.update-share.if.run}{0.24}
\setval{8b.concur-tied.update-share.math.early}{0.17}
\setval{8b.concur-tied.update-share.math.late}{0.12}
\setval{8b.concur-tied.update-share.math.run}{0.13}
\setval{8b.concur-tied.update-share.writing.early}{0.42}
\setval{8b.concur-tied.update-share.writing.late}{0.42}
\setval{8b.concur-tied.update-share.writing.run}{0.38}
\setval{8b.concur.aux-change.code}{+0.18}
\setval{8b.concur.aux-change.if}{+0.18}
\setval{8b.concur.aux-change.math}{$-$0.59}
\setval{8b.concur.aux-change.writing}{$-$0.29}
\setval{8b.concur.aux-change.writing-rm}{+0.32}
\setval{8b.concur.aux-full.code.early}{0.64}
\setval{8b.concur.aux-full.code.late}{0.96}
\setval{8b.concur.aux-full.code.run}{0.92}
\setval{8b.concur.aux-full.if.early}{0.69}
\setval{8b.concur.aux-full.if.late}{0.93}
\setval{8b.concur.aux-full.if.run}{0.87}
\setval{8b.concur.aux-full.math.early}{0.53}
\setval{8b.concur.aux-full.math.late}{0.05}
\setval{8b.concur.aux-full.math.run}{0.21}
\setval{8b.concur.aux-full.writing-rm.early}{0.07}
\setval{8b.concur.aux-full.writing-rm.late}{0.53}
\setval{8b.concur.aux-full.writing-rm.run}{0.44}
\setval{8b.concur.aux-full.writing.early}{0.61}
\setval{8b.concur.aux-full.writing.late}{0.30}
\setval{8b.concur.aux-full.writing.run}{0.27}
\setval{8b.concur.aux-level.code.early}{0.91}
\setval{8b.concur.aux-level.code.late}{0.99}
\setval{8b.concur.aux-level.code.run}{0.98}
\setval{8b.concur.aux-level.if.early}{0.85}
\setval{8b.concur.aux-level.if.late}{0.97}
\setval{8b.concur.aux-level.if.run}{0.94}
\setval{8b.concur.aux-level.math.early}{0.89}
\setval{8b.concur.aux-level.math.late}{0.25}
\setval{8b.concur.aux-level.math.run}{0.50}
\setval{8b.concur.aux-level.writing-rm.early}{0.77}
\setval{8b.concur.aux-level.writing-rm.late}{0.95}
\setval{8b.concur.aux-level.writing-rm.run}{0.93}
\setval{8b.concur.aux-level.writing.early}{0.89}
\setval{8b.concur.aux-level.writing.late}{0.55}
\setval{8b.concur.aux-level.writing.run}{0.54}
\setval{8b.concur.bench.aime24}{28.7}
\setval{8b.concur.bench.aime25}{23.1}
\setval{8b.concur.bench.amc23}{78.9}
\setval{8b.concur.bench.creative-writing-v3}{49.9}
\setval{8b.concur.bench.guidebench}{83.4}
\setval{8b.concur.bench.humaneval-plus}{82.7}
\setval{8b.concur.bench.ifbench}{38.7}
\setval{8b.concur.bench.ifeval}{88.1}
\setval{8b.concur.bench.livecodebench}{26.3}
\setval{8b.concur.bench.longbench-write}{81.6}
\setval{8b.concur.bench.math500}{90.1}
\setval{8b.concur.bench.mbpp-plus}{73.2}
\setval{8b.concur.bench.minerva}{54.2}
\setval{8b.concur.bench.olympiad-bench}{60.4}
\setval{8b.concur.bench.writingbench}{88.1}
\setval{8b.concur.best-count.all}{8}
\setval{8b.concur.best-count.code}{0}
\setval{8b.concur.best-count.if}{0}
\setval{8b.concur.best-count.math}{6}
\setval{8b.concur.best-count.writing}{2}
\setval{8b.concur.eval.avg}{63.2}
\setval{8b.concur.eval.code}{60.8}
\setval{8b.concur.eval.if}{70.1}
\setval{8b.concur.eval.math}{55.9}
\setval{8b.concur.eval.writing}{73.2}
\setval{8b.concur.gate.code.against.early}{0.04}
\setval{8b.concur.gate.code.against.late}{0.03}
\setval{8b.concur.gate.code.against.run}{0.03}
\setval{8b.concur.gate.code.agreement}{+0.011}
\setval{8b.concur.gate.code.ceiling.early}{0.14}
\setval{8b.concur.gate.code.ceiling.late}{0.03}
\setval{8b.concur.gate.code.ceiling.run}{0.05}
\setval{8b.concur.gate.code.cosine.early}{0.81}
\setval{8b.concur.gate.code.cosine.late}{0.78}
\setval{8b.concur.gate.code.cosine.run}{0.80}
\setval{8b.concur.gate.if.against.early}{0.05}
\setval{8b.concur.gate.if.against.late}{0.05}
\setval{8b.concur.gate.if.against.run}{0.04}
\setval{8b.concur.gate.if.agreement}{+0.018}
\setval{8b.concur.gate.if.ceiling.early}{0.06}
\setval{8b.concur.gate.if.ceiling.late}{0.14}
\setval{8b.concur.gate.if.ceiling.run}{0.12}
\setval{8b.concur.gate.if.cosine.early}{0.74}
\setval{8b.concur.gate.if.cosine.late}{0.53}
\setval{8b.concur.gate.if.cosine.run}{0.60}
\setval{8b.concur.gate.math.against.early}{0.00}
\setval{8b.concur.gate.math.against.late}{0.00}
\setval{8b.concur.gate.math.against.run}{0.00}
\setval{8b.concur.gate.math.ceiling.early}{0.26}
\setval{8b.concur.gate.math.ceiling.late}{0.91}
\setval{8b.concur.gate.math.ceiling.run}{0.73}
\setval{8b.concur.gate.math.cosine.early}{0.84}
\setval{8b.concur.gate.math.cosine.late}{0.16}
\setval{8b.concur.gate.math.cosine.run}{0.37}
\setval{8b.concur.gate.writing-rm.against.early}{0.20}
\setval{8b.concur.gate.writing-rm.against.late}{0.55}
\setval{8b.concur.gate.writing-rm.against.run}{0.52}
\setval{8b.concur.gate.writing-rm.agreement}{+0.051}
\setval{8b.concur.gate.writing-rm.ceiling.early}{0.00}
\setval{8b.concur.gate.writing-rm.ceiling.late}{0.01}
\setval{8b.concur.gate.writing-rm.ceiling.run}{0.00}
\setval{8b.concur.gate.writing-rm.cosine.early}{0.69}
\setval{8b.concur.gate.writing-rm.cosine.late}{0.10}
\setval{8b.concur.gate.writing-rm.cosine.run}{0.17}
\setval{8b.concur.gate.writing.against.early}{0.32}
\setval{8b.concur.gate.writing.against.late}{0.67}
\setval{8b.concur.gate.writing.against.run}{0.64}
\setval{8b.concur.gate.writing.agreement}{$-$0.047}
\setval{8b.concur.gate.writing.ceiling.early}{0.00}
\setval{8b.concur.gate.writing.ceiling.late}{0.01}
\setval{8b.concur.gate.writing.ceiling.run}{0.00}
\setval{8b.concur.gate.writing.cosine.early}{0.49}
\setval{8b.concur.gate.writing.cosine.late}{-0.04}
\setval{8b.concur.gate.writing.cosine.run}{0.01}
\setval{8b.concur.judge2.avg}{60.3}
\setval{8b.concur.judge2.creative-writing-v3}{37.8}
\setval{8b.concur.judge2.longbench-write}{75.5}
\setval{8b.concur.judge2.writing}{59.1}
\setval{8b.concur.judge2.writingbench}{63.9}
\setval{8b.concur.last-step}{300}
\setval{8b.concur.minus.base.aime24}{+17.4}
\setval{8b.concur.minus.base.aime24-abs}{17.4}
\setval{8b.concur.minus.base.aime25}{+14.6}
\setval{8b.concur.minus.base.aime25-abs}{14.6}
\setval{8b.concur.minus.base.amc23}{+28.7}
\setval{8b.concur.minus.base.amc23-abs}{28.7}
\setval{8b.concur.minus.base.avg}{+21.7}
\setval{8b.concur.minus.base.avg-abs}{21.7}
\setval{8b.concur.minus.base.code}{+4.7}
\setval{8b.concur.minus.base.code-abs}{4.7}
\setval{8b.concur.minus.base.creative-writing-v3}{+23.2}
\setval{8b.concur.minus.base.creative-writing-v3-abs}{23.2}
\setval{8b.concur.minus.base.guidebench}{+37.8}
\setval{8b.concur.minus.base.guidebench-abs}{37.8}
\setval{8b.concur.minus.base.humaneval-plus}{+2.7}
\setval{8b.concur.minus.base.humaneval-plus-abs}{2.7}
\setval{8b.concur.minus.base.if}{+31.4}
\setval{8b.concur.minus.base.if-abs}{31.4}
\setval{8b.concur.minus.base.ifbench}{+21.3}
\setval{8b.concur.minus.base.ifbench-abs}{21.3}
\setval{8b.concur.minus.base.ifeval}{+35.1}
\setval{8b.concur.minus.base.ifeval-abs}{35.1}
\setval{8b.concur.minus.base.judge2.avg}{+19.3}
\setval{8b.concur.minus.base.judge2.avg-abs}{19.3}
\setval{8b.concur.minus.base.judge2.creative-writing-v3}{+11.3}
\setval{8b.concur.minus.base.judge2.creative-writing-v3-abs}{11.3}
\setval{8b.concur.minus.base.judge2.longbench-write}{+27.4}
\setval{8b.concur.minus.base.judge2.longbench-write-abs}{27.4}
\setval{8b.concur.minus.base.judge2.writing}{+22.9}
\setval{8b.concur.minus.base.judge2.writing-abs}{22.9}
\setval{8b.concur.minus.base.judge2.writingbench}{+29.9}
\setval{8b.concur.minus.base.judge2.writingbench-abs}{29.9}
\setval{8b.concur.minus.base.livecodebench}{+5.4}
\setval{8b.concur.minus.base.livecodebench-abs}{5.4}
\setval{8b.concur.minus.base.longbench-write}{+30.2}
\setval{8b.concur.minus.base.longbench-write-abs}{30.2}
\setval{8b.concur.minus.base.math}{+18.8}
\setval{8b.concur.minus.base.math-abs}{18.8}
\setval{8b.concur.minus.base.math500}{+17.7}
\setval{8b.concur.minus.base.math500-abs}{17.7}
\setval{8b.concur.minus.base.mbpp-plus}{+5.9}
\setval{8b.concur.minus.base.mbpp-plus-abs}{5.9}
\setval{8b.concur.minus.base.minerva}{+13.3}
\setval{8b.concur.minus.base.minerva-abs}{13.3}
\setval{8b.concur.minus.base.olympiad-bench}{+20.9}
\setval{8b.concur.minus.base.olympiad-bench-abs}{20.9}
\setval{8b.concur.minus.base.writing}{+34.9}
\setval{8b.concur.minus.base.writing-abs}{34.9}
\setval{8b.concur.minus.base.writingbench}{+51.3}
\setval{8b.concur.minus.base.writingbench-abs}{51.3}
\setval{8b.concur.minus.best.bench.aime24}{+7.4}
\setval{8b.concur.minus.best.bench.aime25}{+7.4}
\setval{8b.concur.minus.best.bench.amc23}{+13.3}
\setval{8b.concur.minus.best.bench.creative-writing-v3}{$-$3.0}
\setval{8b.concur.minus.best.bench.guidebench}{$-$1.2}
\setval{8b.concur.minus.best.bench.humaneval-plus}{$-$3.1}
\setval{8b.concur.minus.best.bench.ifbench}{$-$4.3}
\setval{8b.concur.minus.best.bench.ifeval}{$-$1.2}
\setval{8b.concur.minus.best.bench.livecodebench}{$-$1.3}
\setval{8b.concur.minus.best.bench.longbench-write}{+10.6}
\setval{8b.concur.minus.best.bench.math500}{+7.1}
\setval{8b.concur.minus.best.bench.mbpp-plus}{$-$2.4}
\setval{8b.concur.minus.best.bench.minerva}{+2.4}
\setval{8b.concur.minus.best.bench.olympiad-bench}{+11.1}
\setval{8b.concur.minus.best.bench.writingbench}{+11.6}
\setval{8b.concur.minus.best.eval.avg}{+4.8}
\setval{8b.concur.minus.best.eval.code}{$-$1.9}
\setval{8b.concur.minus.best.eval.if}{$-$1.9}
\setval{8b.concur.minus.best.eval.math}{+9.3}
\setval{8b.concur.minus.best.eval.writing}{+6.9}
\setval{8b.concur.minus.concur-free.aime24}{+9.1}
\setval{8b.concur.minus.concur-free.aime24-abs}{9.1}
\setval{8b.concur.minus.concur-free.aime25}{+6.6}
\setval{8b.concur.minus.concur-free.aime25-abs}{6.6}
\setval{8b.concur.minus.concur-free.amc23}{+14.2}
\setval{8b.concur.minus.concur-free.amc23-abs}{14.2}
\setval{8b.concur.minus.concur-free.avg}{+2.9}
\setval{8b.concur.minus.concur-free.avg-abs}{2.9}
\setval{8b.concur.minus.concur-free.code}{$-$1.7}
\setval{8b.concur.minus.concur-free.code-abs}{1.7}
\setval{8b.concur.minus.concur-free.creative-writing-v3}{$-$1.7}
\setval{8b.concur.minus.concur-free.creative-writing-v3-abs}{1.7}
\setval{8b.concur.minus.concur-free.guidebench}{+0.5}
\setval{8b.concur.minus.concur-free.guidebench-abs}{0.5}
\setval{8b.concur.minus.concur-free.humaneval-plus}{$-$2.6}
\setval{8b.concur.minus.concur-free.humaneval-plus-abs}{2.6}
\setval{8b.concur.minus.concur-free.if}{$-$0.2}
\setval{8b.concur.minus.concur-free.if-abs}{0.2}
\setval{8b.concur.minus.concur-free.ifbench}{+0.0}
\setval{8b.concur.minus.concur-free.ifbench-abs}{0.0}
\setval{8b.concur.minus.concur-free.ifeval}{$-$1.1}
\setval{8b.concur.minus.concur-free.ifeval-abs}{1.1}
\setval{8b.concur.minus.concur-free.judge2.avg}{+3.0}
\setval{8b.concur.minus.concur-free.judge2.avg-abs}{3.0}
\setval{8b.concur.minus.concur-free.judge2.creative-writing-v3}{$-$4.8}
\setval{8b.concur.minus.concur-free.judge2.creative-writing-v3-abs}{4.8}
\setval{8b.concur.minus.concur-free.judge2.longbench-write}{+1.4}
\setval{8b.concur.minus.concur-free.judge2.longbench-write-abs}{1.4}
\setval{8b.concur.minus.concur-free.judge2.writing}{+0.0}
\setval{8b.concur.minus.concur-free.judge2.writing-abs}{0.0}
\setval{8b.concur.minus.concur-free.judge2.writingbench}{+3.3}
\setval{8b.concur.minus.concur-free.judge2.writingbench-abs}{3.3}
\setval{8b.concur.minus.concur-free.livecodebench}{$-$0.7}
\setval{8b.concur.minus.concur-free.livecodebench-abs}{0.7}
\setval{8b.concur.minus.concur-free.longbench-write}{$-$1.4}
\setval{8b.concur.minus.concur-free.longbench-write-abs}{1.4}
\setval{8b.concur.minus.concur-free.math}{+8.5}
\setval{8b.concur.minus.concur-free.math-abs}{8.5}
\setval{8b.concur.minus.concur-free.math500}{+6.7}
\setval{8b.concur.minus.concur-free.math500-abs}{6.7}
\setval{8b.concur.minus.concur-free.mbpp-plus}{$-$1.9}
\setval{8b.concur.minus.concur-free.mbpp-plus-abs}{1.9}
\setval{8b.concur.minus.concur-free.minerva}{+4.4}
\setval{8b.concur.minus.concur-free.minerva-abs}{4.4}
\setval{8b.concur.minus.concur-free.olympiad-bench}{+10.3}
\setval{8b.concur.minus.concur-free.olympiad-bench-abs}{10.3}
\setval{8b.concur.minus.concur-free.writing}{$-$0.6}
\setval{8b.concur.minus.concur-free.writing-abs}{0.6}
\setval{8b.concur.minus.concur-free.writingbench}{+1.3}
\setval{8b.concur.minus.concur-free.writingbench-abs}{1.3}
\setval{8b.concur.minus.concur-tied.aime24}{$-$2.5}
\setval{8b.concur.minus.concur-tied.aime24-abs}{2.5}
\setval{8b.concur.minus.concur-tied.aime25}{$-$0.8}
\setval{8b.concur.minus.concur-tied.aime25-abs}{0.8}
\setval{8b.concur.minus.concur-tied.amc23}{+0.4}
\setval{8b.concur.minus.concur-tied.amc23-abs}{0.4}
\setval{8b.concur.minus.concur-tied.avg}{$-$0.8}
\setval{8b.concur.minus.concur-tied.avg-abs}{0.8}
\setval{8b.concur.minus.concur-tied.code}{$-$2.4}
\setval{8b.concur.minus.concur-tied.code-abs}{2.4}
\setval{8b.concur.minus.concur-tied.creative-writing-v3}{$-$2.5}
\setval{8b.concur.minus.concur-tied.creative-writing-v3-abs}{2.5}
\setval{8b.concur.minus.concur-tied.guidebench}{+0.0}
\setval{8b.concur.minus.concur-tied.guidebench-abs}{0.0}
\setval{8b.concur.minus.concur-tied.humaneval-plus}{$-$4.6}
\setval{8b.concur.minus.concur-tied.humaneval-plus-abs}{4.6}
\setval{8b.concur.minus.concur-tied.if}{$-$0.5}
\setval{8b.concur.minus.concur-tied.if-abs}{0.5}
\setval{8b.concur.minus.concur-tied.ifbench}{$-$0.7}
\setval{8b.concur.minus.concur-tied.ifbench-abs}{0.7}
\setval{8b.concur.minus.concur-tied.ifeval}{$-$0.7}
\setval{8b.concur.minus.concur-tied.ifeval-abs}{0.7}
\setval{8b.concur.minus.concur-tied.judge2.avg}{$-$1.0}
\setval{8b.concur.minus.concur-tied.judge2.avg-abs}{1.0}
\setval{8b.concur.minus.concur-tied.judge2.creative-writing-v3}{$-$4.0}
\setval{8b.concur.minus.concur-tied.judge2.creative-writing-v3-abs}{4.0}
\setval{8b.concur.minus.concur-tied.judge2.longbench-write}{+0.7}
\setval{8b.concur.minus.concur-tied.judge2.longbench-write-abs}{0.7}
\setval{8b.concur.minus.concur-tied.judge2.writing}{$-$1.0}
\setval{8b.concur.minus.concur-tied.judge2.writing-abs}{1.0}
\setval{8b.concur.minus.concur-tied.judge2.writingbench}{+0.2}
\setval{8b.concur.minus.concur-tied.judge2.writingbench-abs}{0.2}
\setval{8b.concur.minus.concur-tied.livecodebench}{$-$0.9}
\setval{8b.concur.minus.concur-tied.livecodebench-abs}{0.9}
\setval{8b.concur.minus.concur-tied.longbench-write}{+0.4}
\setval{8b.concur.minus.concur-tied.longbench-write-abs}{0.4}
\setval{8b.concur.minus.concur-tied.math}{$-$0.5}
\setval{8b.concur.minus.concur-tied.math-abs}{0.5}
\setval{8b.concur.minus.concur-tied.math500}{+0.5}
\setval{8b.concur.minus.concur-tied.math500-abs}{0.5}
\setval{8b.concur.minus.concur-tied.mbpp-plus}{$-$1.7}
\setval{8b.concur.minus.concur-tied.mbpp-plus-abs}{1.7}
\setval{8b.concur.minus.concur-tied.minerva}{$-$1.2}
\setval{8b.concur.minus.concur-tied.minerva-abs}{1.2}
\setval{8b.concur.minus.concur-tied.olympiad-bench}{+0.8}
\setval{8b.concur.minus.concur-tied.olympiad-bench-abs}{0.8}
\setval{8b.concur.minus.concur-tied.writing}{$-$0.4}
\setval{8b.concur.minus.concur-tied.writing-abs}{0.4}
\setval{8b.concur.minus.concur-tied.writingbench}{+0.9}
\setval{8b.concur.minus.concur-tied.writingbench-abs}{0.9}
\setval{8b.concur.minus.gd2po.aime24}{+12.5}
\setval{8b.concur.minus.gd2po.aime24-abs}{12.5}
\setval{8b.concur.minus.gd2po.aime25}{+10.9}
\setval{8b.concur.minus.gd2po.aime25-abs}{10.9}
\setval{8b.concur.minus.gd2po.amc23}{+19.3}
\setval{8b.concur.minus.gd2po.amc23-abs}{19.3}
\setval{8b.concur.minus.gd2po.avg}{+5.8}
\setval{8b.concur.minus.gd2po.avg-abs}{5.8}
\setval{8b.concur.minus.gd2po.code}{$-$1.0}
\setval{8b.concur.minus.gd2po.code-abs}{1.0}
\setval{8b.concur.minus.gd2po.creative-writing-v3}{+1.8}
\setval{8b.concur.minus.gd2po.creative-writing-v3-abs}{1.8}
\setval{8b.concur.minus.gd2po.guidebench}{$-$0.3}
\setval{8b.concur.minus.gd2po.guidebench-abs}{0.3}
\setval{8b.concur.minus.gd2po.humaneval-plus}{$-$1.2}
\setval{8b.concur.minus.gd2po.humaneval-plus-abs}{1.2}
\setval{8b.concur.minus.gd2po.if}{$-$1.9}
\setval{8b.concur.minus.gd2po.if-abs}{1.9}
\setval{8b.concur.minus.gd2po.ifbench}{$-$4.3}
\setval{8b.concur.minus.gd2po.ifbench-abs}{4.3}
\setval{8b.concur.minus.gd2po.ifeval}{$-$1.2}
\setval{8b.concur.minus.gd2po.ifeval-abs}{1.2}
\setval{8b.concur.minus.gd2po.judge2.avg}{+5.5}
\setval{8b.concur.minus.gd2po.judge2.avg-abs}{5.5}
\setval{8b.concur.minus.gd2po.judge2.creative-writing-v3}{+1.0}
\setval{8b.concur.minus.gd2po.judge2.creative-writing-v3-abs}{1.0}
\setval{8b.concur.minus.gd2po.judge2.longbench-write}{+13.3}
\setval{8b.concur.minus.gd2po.judge2.longbench-write-abs}{13.3}
\setval{8b.concur.minus.gd2po.judge2.writing}{+7.7}
\setval{8b.concur.minus.gd2po.judge2.writing-abs}{7.7}
\setval{8b.concur.minus.gd2po.judge2.writingbench}{+8.7}
\setval{8b.concur.minus.gd2po.judge2.writingbench-abs}{8.7}
\setval{8b.concur.minus.gd2po.livecodebench}{+0.1}
\setval{8b.concur.minus.gd2po.livecodebench-abs}{0.1}
\setval{8b.concur.minus.gd2po.longbench-write}{+11.7}
\setval{8b.concur.minus.gd2po.longbench-write-abs}{11.7}
\setval{8b.concur.minus.gd2po.math}{+11.3}
\setval{8b.concur.minus.gd2po.math-abs}{11.3}
\setval{8b.concur.minus.gd2po.math500}{+8.9}
\setval{8b.concur.minus.gd2po.math500-abs}{8.9}
\setval{8b.concur.minus.gd2po.mbpp-plus}{$-$1.9}
\setval{8b.concur.minus.gd2po.mbpp-plus-abs}{1.9}
\setval{8b.concur.minus.gd2po.minerva}{+4.3}
\setval{8b.concur.minus.gd2po.minerva-abs}{4.3}
\setval{8b.concur.minus.gd2po.olympiad-bench}{+11.8}
\setval{8b.concur.minus.gd2po.olympiad-bench-abs}{11.8}
\setval{8b.concur.minus.gd2po.writing}{+9.2}
\setval{8b.concur.minus.gd2po.writing-abs}{9.2}
\setval{8b.concur.minus.gd2po.writingbench}{+14.2}
\setval{8b.concur.minus.gd2po.writingbench-abs}{14.2}
\setval{8b.concur.minus.gdpo-noaux.aime24}{+0.8}
\setval{8b.concur.minus.gdpo-noaux.aime24-abs}{0.8}
\setval{8b.concur.minus.gdpo-noaux.aime25}{$-$0.8}
\setval{8b.concur.minus.gdpo-noaux.aime25-abs}{0.8}
\setval{8b.concur.minus.gdpo-noaux.amc23}{+1.8}
\setval{8b.concur.minus.gdpo-noaux.amc23-abs}{1.8}
\setval{8b.concur.minus.gdpo-noaux.avg}{+2.6}
\setval{8b.concur.minus.gdpo-noaux.avg-abs}{2.6}
\setval{8b.concur.minus.gdpo-noaux.code}{$-$2.0}
\setval{8b.concur.minus.gdpo-noaux.code-abs}{2.0}
\setval{8b.concur.minus.gdpo-noaux.creative-writing-v3}{+6.0}
\setval{8b.concur.minus.gdpo-noaux.creative-writing-v3-abs}{6.0}
\setval{8b.concur.minus.gdpo-noaux.guidebench}{+2.0}
\setval{8b.concur.minus.gdpo-noaux.guidebench-abs}{2.0}
\setval{8b.concur.minus.gdpo-noaux.humaneval-plus}{$-$3.2}
\setval{8b.concur.minus.gdpo-noaux.humaneval-plus-abs}{3.2}
\setval{8b.concur.minus.gdpo-noaux.if}{+2.9}
\setval{8b.concur.minus.gdpo-noaux.if-abs}{2.9}
\setval{8b.concur.minus.gdpo-noaux.ifbench}{+5.0}
\setval{8b.concur.minus.gdpo-noaux.ifbench-abs}{5.0}
\setval{8b.concur.minus.gdpo-noaux.ifeval}{+1.6}
\setval{8b.concur.minus.gdpo-noaux.ifeval-abs}{1.6}
\setval{8b.concur.minus.gdpo-noaux.judge2.avg}{+3.3}
\setval{8b.concur.minus.gdpo-noaux.judge2.avg-abs}{3.3}
\setval{8b.concur.minus.gdpo-noaux.judge2.creative-writing-v3}{+8.9}
\setval{8b.concur.minus.gdpo-noaux.judge2.creative-writing-v3-abs}{8.9}
\setval{8b.concur.minus.gdpo-noaux.judge2.longbench-write}{+21.0}
\setval{8b.concur.minus.gdpo-noaux.judge2.longbench-write-abs}{21.0}
\setval{8b.concur.minus.gdpo-noaux.judge2.writing}{+14.6}
\setval{8b.concur.minus.gdpo-noaux.judge2.writing-abs}{14.6}
\setval{8b.concur.minus.gdpo-noaux.judge2.writingbench}{+13.9}
\setval{8b.concur.minus.gdpo-noaux.judge2.writingbench-abs}{13.9}
\setval{8b.concur.minus.gdpo-noaux.livecodebench}{$-$1.0}
\setval{8b.concur.minus.gdpo-noaux.livecodebench-abs}{1.0}
\setval{8b.concur.minus.gdpo-noaux.longbench-write}{+15.6}
\setval{8b.concur.minus.gdpo-noaux.longbench-write-abs}{15.6}
\setval{8b.concur.minus.gdpo-noaux.math}{+0.4}
\setval{8b.concur.minus.gdpo-noaux.math-abs}{0.4}
\setval{8b.concur.minus.gdpo-noaux.math500}{$-$0.1}
\setval{8b.concur.minus.gdpo-noaux.math500-abs}{0.1}
\setval{8b.concur.minus.gdpo-noaux.mbpp-plus}{$-$1.7}
\setval{8b.concur.minus.gdpo-noaux.mbpp-plus-abs}{1.7}
\setval{8b.concur.minus.gdpo-noaux.minerva}{+0.2}
\setval{8b.concur.minus.gdpo-noaux.minerva-abs}{0.2}
\setval{8b.concur.minus.gdpo-noaux.olympiad-bench}{+0.4}
\setval{8b.concur.minus.gdpo-noaux.olympiad-bench-abs}{0.4}
\setval{8b.concur.minus.gdpo-noaux.writing}{+11.6}
\setval{8b.concur.minus.gdpo-noaux.writing-abs}{11.6}
\setval{8b.concur.minus.gdpo-noaux.writingbench}{+13.1}
\setval{8b.concur.minus.gdpo-noaux.writingbench-abs}{13.1}
\setval{8b.concur.minus.gdpo-w0.5.aime24}{+9.5}
\setval{8b.concur.minus.gdpo-w0.5.aime24-abs}{9.5}
\setval{8b.concur.minus.gdpo-w0.5.aime25}{+7.7}
\setval{8b.concur.minus.gdpo-w0.5.aime25-abs}{7.7}
\setval{8b.concur.minus.gdpo-w0.5.amc23}{+15.5}
\setval{8b.concur.minus.gdpo-w0.5.amc23-abs}{15.5}
\setval{8b.concur.minus.gdpo-w0.5.avg}{+4.8}
\setval{8b.concur.minus.gdpo-w0.5.avg-abs}{4.8}
\setval{8b.concur.minus.gdpo-w0.5.code}{$-$0.4}
\setval{8b.concur.minus.gdpo-w0.5.code-abs}{0.4}
\setval{8b.concur.minus.gdpo-w0.5.creative-writing-v3}{$-$1.6}
\setval{8b.concur.minus.gdpo-w0.5.creative-writing-v3-abs}{1.6}
\setval{8b.concur.minus.gdpo-w0.5.guidebench}{$-$0.6}
\setval{8b.concur.minus.gdpo-w0.5.guidebench-abs}{0.6}
\setval{8b.concur.minus.gdpo-w0.5.humaneval-plus}{$-$1.0}
\setval{8b.concur.minus.gdpo-w0.5.humaneval-plus-abs}{1.0}
\setval{8b.concur.minus.gdpo-w0.5.if}{+0.3}
\setval{8b.concur.minus.gdpo-w0.5.if-abs}{0.3}
\setval{8b.concur.minus.gdpo-w0.5.ifbench}{+1.0}
\setval{8b.concur.minus.gdpo-w0.5.ifbench-abs}{1.0}
\setval{8b.concur.minus.gdpo-w0.5.ifeval}{+0.4}
\setval{8b.concur.minus.gdpo-w0.5.ifeval-abs}{0.4}
\setval{8b.concur.minus.gdpo-w0.5.judge2.avg}{+4.5}
\setval{8b.concur.minus.gdpo-w0.5.judge2.avg-abs}{4.5}
\setval{8b.concur.minus.gdpo-w0.5.judge2.creative-writing-v3}{$-$2.6}
\setval{8b.concur.minus.gdpo-w0.5.judge2.creative-writing-v3-abs}{2.6}
\setval{8b.concur.minus.gdpo-w0.5.judge2.longbench-write}{+11.8}
\setval{8b.concur.minus.gdpo-w0.5.judge2.longbench-write-abs}{11.8}
\setval{8b.concur.minus.gdpo-w0.5.judge2.writing}{+5.6}
\setval{8b.concur.minus.gdpo-w0.5.judge2.writing-abs}{5.6}
\setval{8b.concur.minus.gdpo-w0.5.judge2.writingbench}{+7.7}
\setval{8b.concur.minus.gdpo-w0.5.judge2.writingbench-abs}{7.7}
\setval{8b.concur.minus.gdpo-w0.5.livecodebench}{$-$0.5}
\setval{8b.concur.minus.gdpo-w0.5.livecodebench-abs}{0.5}
\setval{8b.concur.minus.gdpo-w0.5.longbench-write}{+10.4}
\setval{8b.concur.minus.gdpo-w0.5.longbench-write-abs}{10.4}
\setval{8b.concur.minus.gdpo-w0.5.math}{+8.5}
\setval{8b.concur.minus.gdpo-w0.5.math-abs}{8.5}
\setval{8b.concur.minus.gdpo-w0.5.math500}{+6.1}
\setval{8b.concur.minus.gdpo-w0.5.math500-abs}{6.1}
\setval{8b.concur.minus.gdpo-w0.5.mbpp-plus}{+0.4}
\setval{8b.concur.minus.gdpo-w0.5.mbpp-plus-abs}{0.4}
\setval{8b.concur.minus.gdpo-w0.5.minerva}{+3.0}
\setval{8b.concur.minus.gdpo-w0.5.minerva-abs}{3.0}
\setval{8b.concur.minus.gdpo-w0.5.olympiad-bench}{+9.2}
\setval{8b.concur.minus.gdpo-w0.5.olympiad-bench-abs}{9.2}
\setval{8b.concur.minus.gdpo-w0.5.writing}{+7.3}
\setval{8b.concur.minus.gdpo-w0.5.writing-abs}{7.3}
\setval{8b.concur.minus.gdpo-w0.5.writingbench}{+13.0}
\setval{8b.concur.minus.gdpo-w0.5.writingbench-abs}{13.0}
\setval{8b.concur.minus.gdpo.aime24}{+7.4}
\setval{8b.concur.minus.gdpo.aime24-abs}{7.4}
\setval{8b.concur.minus.gdpo.aime25}{+8.8}
\setval{8b.concur.minus.gdpo.aime25-abs}{8.8}
\setval{8b.concur.minus.gdpo.amc23}{+16.7}
\setval{8b.concur.minus.gdpo.amc23-abs}{16.7}
\setval{8b.concur.minus.gdpo.avg}{+4.8}
\setval{8b.concur.minus.gdpo.avg-abs}{4.8}
\setval{8b.concur.minus.gdpo.code}{$-$0.8}
\setval{8b.concur.minus.gdpo.code-abs}{0.8}
\setval{8b.concur.minus.gdpo.creative-writing-v3}{$-$3.0}
\setval{8b.concur.minus.gdpo.creative-writing-v3-abs}{3.0}
\setval{8b.concur.minus.gdpo.guidebench}{$-$1.2}
\setval{8b.concur.minus.gdpo.guidebench-abs}{1.2}
\setval{8b.concur.minus.gdpo.humaneval-plus}{$-$1.3}
\setval{8b.concur.minus.gdpo.humaneval-plus-abs}{1.3}
\setval{8b.concur.minus.gdpo.if}{$-$0.6}
\setval{8b.concur.minus.gdpo.if-abs}{0.6}
\setval{8b.concur.minus.gdpo.ifbench}{$-$0.7}
\setval{8b.concur.minus.gdpo.ifbench-abs}{0.7}
\setval{8b.concur.minus.gdpo.ifeval}{+0.0}
\setval{8b.concur.minus.gdpo.ifeval-abs}{0.0}
\setval{8b.concur.minus.gdpo.judge2.avg}{+4.3}
\setval{8b.concur.minus.gdpo.judge2.avg-abs}{4.3}
\setval{8b.concur.minus.gdpo.judge2.creative-writing-v3}{$-$3.3}
\setval{8b.concur.minus.gdpo.judge2.creative-writing-v3-abs}{3.3}
\setval{8b.concur.minus.gdpo.judge2.longbench-write}{+11.3}
\setval{8b.concur.minus.gdpo.judge2.longbench-write-abs}{11.3}
\setval{8b.concur.minus.gdpo.judge2.writing}{+4.5}
\setval{8b.concur.minus.gdpo.judge2.writing-abs}{4.5}
\setval{8b.concur.minus.gdpo.judge2.writingbench}{+5.6}
\setval{8b.concur.minus.gdpo.judge2.writingbench-abs}{5.6}
\setval{8b.concur.minus.gdpo.livecodebench}{+0.1}
\setval{8b.concur.minus.gdpo.livecodebench-abs}{0.1}
\setval{8b.concur.minus.gdpo.longbench-write}{+12.0}
\setval{8b.concur.minus.gdpo.longbench-write-abs}{12.0}
\setval{8b.concur.minus.gdpo.math}{+9.3}
\setval{8b.concur.minus.gdpo.math-abs}{9.3}
\setval{8b.concur.minus.gdpo.math500}{+7.1}
\setval{8b.concur.minus.gdpo.math500-abs}{7.1}
\setval{8b.concur.minus.gdpo.mbpp-plus}{$-$1.3}
\setval{8b.concur.minus.gdpo.mbpp-plus-abs}{1.3}
\setval{8b.concur.minus.gdpo.minerva}{+4.6}
\setval{8b.concur.minus.gdpo.minerva-abs}{4.6}
\setval{8b.concur.minus.gdpo.olympiad-bench}{+11.1}
\setval{8b.concur.minus.gdpo.olympiad-bench-abs}{11.1}
\setval{8b.concur.minus.gdpo.writing}{+7.0}
\setval{8b.concur.minus.gdpo.writing-abs}{7.0}
\setval{8b.concur.minus.gdpo.writingbench}{+12.0}
\setval{8b.concur.minus.gdpo.writingbench-abs}{12.0}
\setval{8b.concur.minus.geoma.aime24}{+14.7}
\setval{8b.concur.minus.geoma.aime24-abs}{14.7}
\setval{8b.concur.minus.geoma.aime25}{+10.6}
\setval{8b.concur.minus.geoma.aime25-abs}{10.6}
\setval{8b.concur.minus.geoma.amc23}{+15.4}
\setval{8b.concur.minus.geoma.amc23-abs}{15.4}
\setval{8b.concur.minus.geoma.avg}{+6.5}
\setval{8b.concur.minus.geoma.avg-abs}{6.5}
\setval{8b.concur.minus.geoma.code}{$-$0.6}
\setval{8b.concur.minus.geoma.code-abs}{0.6}
\setval{8b.concur.minus.geoma.creative-writing-v3}{$-$0.5}
\setval{8b.concur.minus.geoma.creative-writing-v3-abs}{0.5}
\setval{8b.concur.minus.geoma.guidebench}{$-$1.0}
\setval{8b.concur.minus.geoma.guidebench-abs}{1.0}
\setval{8b.concur.minus.geoma.humaneval-plus}{+0.0}
\setval{8b.concur.minus.geoma.humaneval-plus-abs}{0.0}
\setval{8b.concur.minus.geoma.if}{+1.5}
\setval{8b.concur.minus.geoma.if-abs}{1.5}
\setval{8b.concur.minus.geoma.ifbench}{+4.7}
\setval{8b.concur.minus.geoma.ifbench-abs}{4.7}
\setval{8b.concur.minus.geoma.ifeval}{+0.8}
\setval{8b.concur.minus.geoma.ifeval-abs}{0.8}
\setval{8b.concur.minus.geoma.judge2.avg}{+6.0}
\setval{8b.concur.minus.geoma.judge2.avg-abs}{6.0}
\setval{8b.concur.minus.geoma.judge2.creative-writing-v3}{+0.2}
\setval{8b.concur.minus.geoma.judge2.creative-writing-v3-abs}{0.2}
\setval{8b.concur.minus.geoma.judge2.longbench-write}{+10.6}
\setval{8b.concur.minus.geoma.judge2.longbench-write-abs}{10.6}
\setval{8b.concur.minus.geoma.judge2.writing}{+5.5}
\setval{8b.concur.minus.geoma.judge2.writing-abs}{5.5}
\setval{8b.concur.minus.geoma.judge2.writingbench}{+5.6}
\setval{8b.concur.minus.geoma.judge2.writingbench-abs}{5.6}
\setval{8b.concur.minus.geoma.livecodebench}{+0.6}
\setval{8b.concur.minus.geoma.livecodebench-abs}{0.6}
\setval{8b.concur.minus.geoma.longbench-write}{+12.6}
\setval{8b.concur.minus.geoma.longbench-write-abs}{12.6}
\setval{8b.concur.minus.geoma.math}{+11.7}
\setval{8b.concur.minus.geoma.math-abs}{11.7}
\setval{8b.concur.minus.geoma.math500}{+10.5}
\setval{8b.concur.minus.geoma.math500-abs}{10.5}
\setval{8b.concur.minus.geoma.mbpp-plus}{$-$2.4}
\setval{8b.concur.minus.geoma.mbpp-plus-abs}{2.4}
\setval{8b.concur.minus.geoma.minerva}{+4.4}
\setval{8b.concur.minus.geoma.minerva-abs}{4.4}
\setval{8b.concur.minus.geoma.olympiad-bench}{+14.4}
\setval{8b.concur.minus.geoma.olympiad-bench-abs}{14.4}
\setval{8b.concur.minus.geoma.writing}{+8.1}
\setval{8b.concur.minus.geoma.writing-abs}{8.1}
\setval{8b.concur.minus.geoma.writingbench}{+12.1}
\setval{8b.concur.minus.geoma.writingbench-abs}{12.1}
\setval{8b.concur.minus.grpo-nodiv.aime24}{+8.1}
\setval{8b.concur.minus.grpo-nodiv.aime24-abs}{8.1}
\setval{8b.concur.minus.grpo-nodiv.aime25}{+8.3}
\setval{8b.concur.minus.grpo-nodiv.aime25-abs}{8.3}
\setval{8b.concur.minus.grpo-nodiv.amc23}{+15.3}
\setval{8b.concur.minus.grpo-nodiv.amc23-abs}{15.3}
\setval{8b.concur.minus.grpo-nodiv.avg}{+5.1}
\setval{8b.concur.minus.grpo-nodiv.avg-abs}{5.1}
\setval{8b.concur.minus.grpo-nodiv.code}{$-$1.7}
\setval{8b.concur.minus.grpo-nodiv.code-abs}{1.7}
\setval{8b.concur.minus.grpo-nodiv.creative-writing-v3}{+3.3}
\setval{8b.concur.minus.grpo-nodiv.creative-writing-v3-abs}{3.3}
\setval{8b.concur.minus.grpo-nodiv.guidebench}{$-$1.4}
\setval{8b.concur.minus.grpo-nodiv.guidebench-abs}{1.4}
\setval{8b.concur.minus.grpo-nodiv.humaneval-plus}{$-$3.6}
\setval{8b.concur.minus.grpo-nodiv.humaneval-plus-abs}{3.6}
\setval{8b.concur.minus.grpo-nodiv.if}{$-$0.1}
\setval{8b.concur.minus.grpo-nodiv.if-abs}{0.1}
\setval{8b.concur.minus.grpo-nodiv.ifbench}{+2.3}
\setval{8b.concur.minus.grpo-nodiv.ifbench-abs}{2.3}
\setval{8b.concur.minus.grpo-nodiv.ifeval}{$-$1.3}
\setval{8b.concur.minus.grpo-nodiv.ifeval-abs}{1.3}
\setval{8b.concur.minus.grpo-nodiv.judge2.avg}{+4.3}
\setval{8b.concur.minus.grpo-nodiv.judge2.avg-abs}{4.3}
\setval{8b.concur.minus.grpo-nodiv.judge2.creative-writing-v3}{+0.3}
\setval{8b.concur.minus.grpo-nodiv.judge2.creative-writing-v3-abs}{0.3}
\setval{8b.concur.minus.grpo-nodiv.judge2.longbench-write}{+8.9}
\setval{8b.concur.minus.grpo-nodiv.judge2.longbench-write-abs}{8.9}
\setval{8b.concur.minus.grpo-nodiv.judge2.writing}{+6.4}
\setval{8b.concur.minus.grpo-nodiv.judge2.writing-abs}{6.4}
\setval{8b.concur.minus.grpo-nodiv.judge2.writingbench}{+10.0}
\setval{8b.concur.minus.grpo-nodiv.judge2.writingbench-abs}{10.0}
\setval{8b.concur.minus.grpo-nodiv.livecodebench}{+1.4}
\setval{8b.concur.minus.grpo-nodiv.livecodebench-abs}{1.4}
\setval{8b.concur.minus.grpo-nodiv.longbench-write}{+10.5}
\setval{8b.concur.minus.grpo-nodiv.longbench-write-abs}{10.5}
\setval{8b.concur.minus.grpo-nodiv.math}{+8.3}
\setval{8b.concur.minus.grpo-nodiv.math-abs}{8.3}
\setval{8b.concur.minus.grpo-nodiv.math500}{+6.0}
\setval{8b.concur.minus.grpo-nodiv.math500-abs}{6.0}
\setval{8b.concur.minus.grpo-nodiv.mbpp-plus}{$-$2.8}
\setval{8b.concur.minus.grpo-nodiv.mbpp-plus-abs}{2.8}
\setval{8b.concur.minus.grpo-nodiv.minerva}{+3.4}
\setval{8b.concur.minus.grpo-nodiv.minerva-abs}{3.4}
\setval{8b.concur.minus.grpo-nodiv.olympiad-bench}{+8.9}
\setval{8b.concur.minus.grpo-nodiv.olympiad-bench-abs}{8.9}
\setval{8b.concur.minus.grpo-nodiv.writing}{+10.5}
\setval{8b.concur.minus.grpo-nodiv.writing-abs}{10.5}
\setval{8b.concur.minus.grpo-nodiv.writingbench}{+17.7}
\setval{8b.concur.minus.grpo-nodiv.writingbench-abs}{17.7}
\setval{8b.concur.minus.grpo.aime24}{+12.9}
\setval{8b.concur.minus.grpo.aime24-abs}{12.9}
\setval{8b.concur.minus.grpo.aime25}{+7.6}
\setval{8b.concur.minus.grpo.aime25-abs}{7.6}
\setval{8b.concur.minus.grpo.amc23}{+14.5}
\setval{8b.concur.minus.grpo.amc23-abs}{14.5}
\setval{8b.concur.minus.grpo.avg}{+5.3}
\setval{8b.concur.minus.grpo.avg-abs}{5.3}
\setval{8b.concur.minus.grpo.code}{$-$0.9}
\setval{8b.concur.minus.grpo.code-abs}{0.9}
\setval{8b.concur.minus.grpo.creative-writing-v3}{$-$2.6}
\setval{8b.concur.minus.grpo.creative-writing-v3-abs}{2.6}
\setval{8b.concur.minus.grpo.guidebench}{+1.5}
\setval{8b.concur.minus.grpo.guidebench-abs}{1.5}
\setval{8b.concur.minus.grpo.humaneval-plus}{$-$1.0}
\setval{8b.concur.minus.grpo.humaneval-plus-abs}{1.0}
\setval{8b.concur.minus.grpo.if}{$-$0.2}
\setval{8b.concur.minus.grpo.if-abs}{0.2}
\setval{8b.concur.minus.grpo.ifbench}{$-$3.3}
\setval{8b.concur.minus.grpo.ifbench-abs}{3.3}
\setval{8b.concur.minus.grpo.ifeval}{+1.3}
\setval{8b.concur.minus.grpo.ifeval-abs}{1.3}
\setval{8b.concur.minus.grpo.judge2.avg}{+5.1}
\setval{8b.concur.minus.grpo.judge2.avg-abs}{5.1}
\setval{8b.concur.minus.grpo.judge2.creative-writing-v3}{$-$1.5}
\setval{8b.concur.minus.grpo.judge2.creative-writing-v3-abs}{1.5}
\setval{8b.concur.minus.grpo.judge2.longbench-write}{+13.1}
\setval{8b.concur.minus.grpo.judge2.longbench-write-abs}{13.1}
\setval{8b.concur.minus.grpo.judge2.writing}{+6.2}
\setval{8b.concur.minus.grpo.judge2.writing-abs}{6.2}
\setval{8b.concur.minus.grpo.judge2.writingbench}{+7.0}
\setval{8b.concur.minus.grpo.judge2.writingbench-abs}{7.0}
\setval{8b.concur.minus.grpo.livecodebench}{$-$0.2}
\setval{8b.concur.minus.grpo.livecodebench-abs}{0.2}
\setval{8b.concur.minus.grpo.longbench-write}{+11.8}
\setval{8b.concur.minus.grpo.longbench-write-abs}{11.8}
\setval{8b.concur.minus.grpo.math}{+10.2}
\setval{8b.concur.minus.grpo.math-abs}{10.2}
\setval{8b.concur.minus.grpo.math500}{+9.5}
\setval{8b.concur.minus.grpo.math500-abs}{9.5}
\setval{8b.concur.minus.grpo.mbpp-plus}{$-$1.5}
\setval{8b.concur.minus.grpo.mbpp-plus-abs}{1.5}
\setval{8b.concur.minus.grpo.minerva}{+4.3}
\setval{8b.concur.minus.grpo.minerva-abs}{4.3}
\setval{8b.concur.minus.grpo.olympiad-bench}{+12.4}
\setval{8b.concur.minus.grpo.olympiad-bench-abs}{12.4}
\setval{8b.concur.minus.grpo.writing}{+6.9}
\setval{8b.concur.minus.grpo.writing-abs}{6.9}
\setval{8b.concur.minus.grpo.writingbench}{+11.6}
\setval{8b.concur.minus.grpo.writingbench-abs}{11.6}
\setval{8b.concur.minus.mt-grpo.aime24}{+14.3}
\setval{8b.concur.minus.mt-grpo.aime24-abs}{14.3}
\setval{8b.concur.minus.mt-grpo.aime25}{+7.4}
\setval{8b.concur.minus.mt-grpo.aime25-abs}{7.4}
\setval{8b.concur.minus.mt-grpo.amc23}{+13.3}
\setval{8b.concur.minus.mt-grpo.amc23-abs}{13.3}
\setval{8b.concur.minus.mt-grpo.avg}{+6.0}
\setval{8b.concur.minus.mt-grpo.avg-abs}{6.0}
\setval{8b.concur.minus.mt-grpo.code}{$-$1.9}
\setval{8b.concur.minus.mt-grpo.code-abs}{1.9}
\setval{8b.concur.minus.mt-grpo.creative-writing-v3}{$-$0.4}
\setval{8b.concur.minus.mt-grpo.creative-writing-v3-abs}{0.4}
\setval{8b.concur.minus.mt-grpo.guidebench}{+0.3}
\setval{8b.concur.minus.mt-grpo.guidebench-abs}{0.3}
\setval{8b.concur.minus.mt-grpo.humaneval-plus}{$-$2.3}
\setval{8b.concur.minus.mt-grpo.humaneval-plus-abs}{2.3}
\setval{8b.concur.minus.mt-grpo.if}{+0.9}
\setval{8b.concur.minus.mt-grpo.if-abs}{0.9}
\setval{8b.concur.minus.mt-grpo.ifbench}{+1.3}
\setval{8b.concur.minus.mt-grpo.ifbench-abs}{1.3}
\setval{8b.concur.minus.mt-grpo.ifeval}{+1.2}
\setval{8b.concur.minus.mt-grpo.ifeval-abs}{1.2}
\setval{8b.concur.minus.mt-grpo.judge2.avg}{+5.1}
\setval{8b.concur.minus.mt-grpo.judge2.avg-abs}{5.1}
\setval{8b.concur.minus.mt-grpo.judge2.creative-writing-v3}{$-$3.8}
\setval{8b.concur.minus.mt-grpo.judge2.creative-writing-v3-abs}{3.8}
\setval{8b.concur.minus.mt-grpo.judge2.longbench-write}{+13.0}
\setval{8b.concur.minus.mt-grpo.judge2.longbench-write-abs}{13.0}
\setval{8b.concur.minus.mt-grpo.judge2.writing}{+5.9}
\setval{8b.concur.minus.mt-grpo.judge2.writing-abs}{5.9}
\setval{8b.concur.minus.mt-grpo.judge2.writingbench}{+8.4}
\setval{8b.concur.minus.mt-grpo.judge2.writingbench-abs}{8.4}
\setval{8b.concur.minus.mt-grpo.livecodebench}{$-$1.3}
\setval{8b.concur.minus.mt-grpo.livecodebench-abs}{1.3}
\setval{8b.concur.minus.mt-grpo.longbench-write}{+14.2}
\setval{8b.concur.minus.mt-grpo.longbench-write-abs}{14.2}
\setval{8b.concur.minus.mt-grpo.math}{+10.4}
\setval{8b.concur.minus.mt-grpo.math-abs}{10.4}
\setval{8b.concur.minus.mt-grpo.math500}{+9.5}
\setval{8b.concur.minus.mt-grpo.math500-abs}{9.5}
\setval{8b.concur.minus.mt-grpo.mbpp-plus}{$-$2.1}
\setval{8b.concur.minus.mt-grpo.mbpp-plus-abs}{2.1}
\setval{8b.concur.minus.mt-grpo.minerva}{+3.1}
\setval{8b.concur.minus.mt-grpo.minerva-abs}{3.1}
\setval{8b.concur.minus.mt-grpo.olympiad-bench}{+14.6}
\setval{8b.concur.minus.mt-grpo.olympiad-bench-abs}{14.6}
\setval{8b.concur.minus.mt-grpo.writing}{+10.0}
\setval{8b.concur.minus.mt-grpo.writing-abs}{10.0}
\setval{8b.concur.minus.mt-grpo.writingbench}{+16.4}
\setval{8b.concur.minus.mt-grpo.writingbench-abs}{16.4}
\setval{8b.concur.minus.papo.aime24}{+16.6}
\setval{8b.concur.minus.papo.aime24-abs}{16.6}
\setval{8b.concur.minus.papo.aime25}{+13.0}
\setval{8b.concur.minus.papo.aime25-abs}{13.0}
\setval{8b.concur.minus.papo.amc23}{+16.9}
\setval{8b.concur.minus.papo.amc23-abs}{16.9}
\setval{8b.concur.minus.papo.avg}{+7.3}
\setval{8b.concur.minus.papo.avg-abs}{7.3}
\setval{8b.concur.minus.papo.code}{$-$0.1}
\setval{8b.concur.minus.papo.code-abs}{0.1}
\setval{8b.concur.minus.papo.creative-writing-v3}{$-$1.2}
\setval{8b.concur.minus.papo.creative-writing-v3-abs}{1.2}
\setval{8b.concur.minus.papo.guidebench}{$-$1.0}
\setval{8b.concur.minus.papo.guidebench-abs}{1.0}
\setval{8b.concur.minus.papo.humaneval-plus}{+0.2}
\setval{8b.concur.minus.papo.humaneval-plus-abs}{0.2}
\setval{8b.concur.minus.papo.if}{+0.7}
\setval{8b.concur.minus.papo.if-abs}{0.7}
\setval{8b.concur.minus.papo.ifbench}{+1.3}
\setval{8b.concur.minus.papo.ifbench-abs}{1.3}
\setval{8b.concur.minus.papo.ifeval}{+1.8}
\setval{8b.concur.minus.papo.ifeval-abs}{1.8}
\setval{8b.concur.minus.papo.judge2.avg}{+6.7}
\setval{8b.concur.minus.papo.judge2.avg-abs}{6.7}
\setval{8b.concur.minus.papo.judge2.creative-writing-v3}{$-$4.1}
\setval{8b.concur.minus.papo.judge2.creative-writing-v3-abs}{4.1}
\setval{8b.concur.minus.papo.judge2.longbench-write}{+15.9}
\setval{8b.concur.minus.papo.judge2.longbench-write-abs}{15.9}
\setval{8b.concur.minus.papo.judge2.writing}{+6.5}
\setval{8b.concur.minus.papo.judge2.writing-abs}{6.5}
\setval{8b.concur.minus.papo.judge2.writingbench}{+7.6}
\setval{8b.concur.minus.papo.judge2.writingbench-abs}{7.6}
\setval{8b.concur.minus.papo.livecodebench}{$-$0.3}
\setval{8b.concur.minus.papo.livecodebench-abs}{0.3}
\setval{8b.concur.minus.papo.longbench-write}{+16.1}
\setval{8b.concur.minus.papo.longbench-write-abs}{16.1}
\setval{8b.concur.minus.papo.math}{+13.1}
\setval{8b.concur.minus.papo.math-abs}{13.1}
\setval{8b.concur.minus.papo.math500}{+11.7}
\setval{8b.concur.minus.papo.math500-abs}{11.7}
\setval{8b.concur.minus.papo.mbpp-plus}{$-$0.2}
\setval{8b.concur.minus.papo.mbpp-plus-abs}{0.2}
\setval{8b.concur.minus.papo.minerva}{+4.9}
\setval{8b.concur.minus.papo.minerva-abs}{4.9}
\setval{8b.concur.minus.papo.olympiad-bench}{+15.6}
\setval{8b.concur.minus.papo.olympiad-bench-abs}{15.6}
\setval{8b.concur.minus.papo.writing}{+9.9}
\setval{8b.concur.minus.papo.writing-abs}{9.9}
\setval{8b.concur.minus.papo.writingbench}{+14.9}
\setval{8b.concur.minus.papo.writingbench-abs}{14.9}
\setval{8b.concur.minus.saw.aime24}{+13.4}
\setval{8b.concur.minus.saw.aime24-abs}{13.4}
\setval{8b.concur.minus.saw.aime25}{+7.6}
\setval{8b.concur.minus.saw.aime25-abs}{7.6}
\setval{8b.concur.minus.saw.amc23}{+16.6}
\setval{8b.concur.minus.saw.amc23-abs}{16.6}
\setval{8b.concur.minus.saw.avg}{+5.2}
\setval{8b.concur.minus.saw.avg-abs}{5.2}
\setval{8b.concur.minus.saw.code}{$-$1.5}
\setval{8b.concur.minus.saw.code-abs}{1.5}
\setval{8b.concur.minus.saw.creative-writing-v3}{+0.2}
\setval{8b.concur.minus.saw.creative-writing-v3-abs}{0.2}
\setval{8b.concur.minus.saw.guidebench}{$-$0.9}
\setval{8b.concur.minus.saw.guidebench-abs}{0.9}
\setval{8b.concur.minus.saw.humaneval-plus}{$-$3.1}
\setval{8b.concur.minus.saw.humaneval-plus-abs}{3.1}
\setval{8b.concur.minus.saw.if}{$-$0.6}
\setval{8b.concur.minus.saw.if-abs}{0.6}
\setval{8b.concur.minus.saw.ifbench}{$-$1.3}
\setval{8b.concur.minus.saw.ifbench-abs}{1.3}
\setval{8b.concur.minus.saw.ifeval}{+0.5}
\setval{8b.concur.minus.saw.ifeval-abs}{0.5}
\setval{8b.concur.minus.saw.judge2.avg}{+4.9}
\setval{8b.concur.minus.saw.judge2.avg-abs}{4.9}
\setval{8b.concur.minus.saw.judge2.creative-writing-v3}{+1.2}
\setval{8b.concur.minus.saw.judge2.creative-writing-v3-abs}{1.2}
\setval{8b.concur.minus.saw.judge2.longbench-write}{+10.7}
\setval{8b.concur.minus.saw.judge2.longbench-write-abs}{10.7}
\setval{8b.concur.minus.saw.judge2.writing}{+6.3}
\setval{8b.concur.minus.saw.judge2.writing-abs}{6.3}
\setval{8b.concur.minus.saw.judge2.writingbench}{+7.0}
\setval{8b.concur.minus.saw.judge2.writingbench-abs}{7.0}
\setval{8b.concur.minus.saw.livecodebench}{$-$0.3}
\setval{8b.concur.minus.saw.livecodebench-abs}{0.3}
\setval{8b.concur.minus.saw.longbench-write}{+10.6}
\setval{8b.concur.minus.saw.longbench-write-abs}{10.6}
\setval{8b.concur.minus.saw.math}{+10.1}
\setval{8b.concur.minus.saw.math-abs}{10.1}
\setval{8b.concur.minus.saw.math500}{+8.2}
\setval{8b.concur.minus.saw.math500-abs}{8.2}
\setval{8b.concur.minus.saw.mbpp-plus}{$-$1.1}
\setval{8b.concur.minus.saw.mbpp-plus-abs}{1.1}
\setval{8b.concur.minus.saw.minerva}{+2.4}
\setval{8b.concur.minus.saw.minerva-abs}{2.4}
\setval{8b.concur.minus.saw.olympiad-bench}{+12.2}
\setval{8b.concur.minus.saw.olympiad-bench-abs}{12.2}
\setval{8b.concur.minus.saw.writing}{+7.8}
\setval{8b.concur.minus.saw.writing-abs}{7.8}
\setval{8b.concur.minus.saw.writingbench}{+12.7}
\setval{8b.concur.minus.saw.writingbench-abs}{12.7}
\setval{8b.concur.passing-at-ceiling.code.early}{0.72}
\setval{8b.concur.passing-at-ceiling.code.late}{0.76}
\setval{8b.concur.passing-at-ceiling.code.run}{0.74}
\setval{8b.concur.passing-at-ceiling.if.early}{0.32}
\setval{8b.concur.passing-at-ceiling.if.late}{0.56}
\setval{8b.concur.passing-at-ceiling.if.run}{0.48}
\setval{8b.concur.passing-at-ceiling.math.early}{1.00}
\setval{8b.concur.passing-at-ceiling.math.late}{1.00}
\setval{8b.concur.passing-at-ceiling.math.run}{1.00}
\setval{8b.concur.passing-at-ceiling.writing.early}{0.00}
\setval{8b.concur.passing-at-ceiling.writing.late}{0.02}
\setval{8b.concur.passing-at-ceiling.writing.run}{0.01}
\setval{8b.concur.primary-cosine.code.early}{0.97}
\setval{8b.concur.primary-cosine.code.late}{0.96}
\setval{8b.concur.primary-cosine.code.run}{0.96}
\setval{8b.concur.primary-cosine.if.early}{0.86}
\setval{8b.concur.primary-cosine.if.late}{0.80}
\setval{8b.concur.primary-cosine.if.run}{0.83}
\setval{8b.concur.primary-cosine.math.early}{0.99}
\setval{8b.concur.primary-cosine.math.late}{0.97}
\setval{8b.concur.primary-cosine.math.run}{0.97}
\setval{8b.concur.primary-cosine.writing.early}{0.94}
\setval{8b.concur.primary-cosine.writing.late}{0.92}
\setval{8b.concur.primary-cosine.writing.run}{0.92}
\setval{8b.concur.primary-level.code.early}{0.54}
\setval{8b.concur.primary-level.code.late}{0.76}
\setval{8b.concur.primary-level.code.run}{0.71}
\setval{8b.concur.primary-level.if.early}{0.35}
\setval{8b.concur.primary-level.if.late}{0.75}
\setval{8b.concur.primary-level.if.run}{0.64}
\setval{8b.concur.primary-level.math.early}{0.49}
\setval{8b.concur.primary-level.math.late}{0.75}
\setval{8b.concur.primary-level.math.run}{0.68}
\setval{8b.concur.primary-level.writing.early}{0.54}
\setval{8b.concur.primary-level.writing.late}{0.81}
\setval{8b.concur.primary-level.writing.run}{0.76}
\setval{8b.concur.solved.code.early}{0.09}
\setval{8b.concur.solved.code.late}{0.02}
\setval{8b.concur.solved.code.run}{0.04}
\setval{8b.concur.solved.if.early}{0.03}
\setval{8b.concur.solved.if.late}{0.19}
\setval{8b.concur.solved.if.run}{0.13}
\setval{8b.concur.solved.math.early}{0.16}
\setval{8b.concur.solved.math.late}{0.48}
\setval{8b.concur.solved.math.run}{0.40}
\setval{8b.concur.solved.writing.early}{0.00}
\setval{8b.concur.solved.writing.late}{0.00}
\setval{8b.concur.solved.writing.run}{0.00}
\setval{8b.concur.trust.code.agreement}{+0.011}
\setval{8b.concur.trust.code.errors}{+3.8}
\setval{8b.concur.trust.code.first-free}{10}
\setval{8b.concur.trust.code.free-share}{0.97}
\setval{8b.concur.trust.if.agreement}{+0.018}
\setval{8b.concur.trust.if.errors}{+7.3}
\setval{8b.concur.trust.if.first-free}{8}
\setval{8b.concur.trust.if.free-share}{0.95}
\setval{8b.concur.trust.math.first-free}{--}
\setval{8b.concur.trust.math.free-share}{0.00}
\setval{8b.concur.trust.writing-rm.agreement}{+0.051}
\setval{8b.concur.trust.writing-rm.errors}{+31.6}
\setval{8b.concur.trust.writing-rm.first-free}{2}
\setval{8b.concur.trust.writing-rm.free-share}{1.00}
\setval{8b.concur.trust.writing.agreement}{$-$0.047}
\setval{8b.concur.trust.writing.errors}{$-$34.1}
\setval{8b.concur.trust.writing.first-free}{--}
\setval{8b.concur.trust.writing.free-share}{0.00}
\setval{8b.concur.update-share.code.early}{0.20}
\setval{8b.concur.update-share.code.late}{0.27}
\setval{8b.concur.update-share.code.run}{0.25}
\setval{8b.concur.update-share.if.early}{0.22}
\setval{8b.concur.update-share.if.late}{0.21}
\setval{8b.concur.update-share.if.run}{0.24}
\setval{8b.concur.update-share.math.early}{0.17}
\setval{8b.concur.update-share.math.late}{0.10}
\setval{8b.concur.update-share.math.run}{0.14}
\setval{8b.concur.update-share.writing.early}{0.42}
\setval{8b.concur.update-share.writing.late}{0.42}
\setval{8b.concur.update-share.writing.run}{0.38}
\setval{8b.estimators.aux-full.code.run-max}{0.97}
\setval{8b.estimators.aux-full.code.run-min}{0.88}
\setval{8b.estimators.aux-full.if.run-max}{0.91}
\setval{8b.estimators.aux-full.if.run-min}{0.86}
\setval{8b.estimators.aux-full.math.run-max}{0.93}
\setval{8b.estimators.aux-full.math.run-min}{0.81}
\setval{8b.estimators.aux-full.writing-rm.run-max}{0.54}
\setval{8b.estimators.aux-full.writing-rm.run-min}{0.43}
\setval{8b.estimators.aux-full.writing.run-max}{0.79}
\setval{8b.estimators.aux-full.writing.run-min}{0.62}
\setval{8b.estimators.avg-spread}{1.7}
\setval{8b.estimators.bench.aime24.max}{21.4}
\setval{8b.estimators.bench.aime24.min}{12.2}
\setval{8b.estimators.bench.aime25.max}{15.7}
\setval{8b.estimators.bench.aime25.min}{10.1}
\setval{8b.estimators.bench.amc23.max}{65.6}
\setval{8b.estimators.bench.amc23.min}{59.6}
\setval{8b.estimators.bench.creative-writing-v3.max}{52.9}
\setval{8b.estimators.bench.creative-writing-v3.min}{48.2}
\setval{8b.estimators.bench.guidebench.max}{84.5}
\setval{8b.estimators.bench.guidebench.min}{81.9}
\setval{8b.estimators.bench.humaneval-plus.max}{85.9}
\setval{8b.estimators.bench.humaneval-plus.min}{82.5}
\setval{8b.estimators.bench.ifbench.max}{43.0}
\setval{8b.estimators.bench.ifbench.min}{34.0}
\setval{8b.estimators.bench.ifeval.max}{89.3}
\setval{8b.estimators.bench.ifeval.min}{86.3}
\setval{8b.estimators.bench.livecodebench.max}{27.7}
\setval{8b.estimators.bench.livecodebench.min}{25.7}
\setval{8b.estimators.bench.longbench-write.max}{71.0}
\setval{8b.estimators.bench.longbench-write.min}{65.5}
\setval{8b.estimators.bench.math500.max}{83.0}
\setval{8b.estimators.bench.math500.min}{78.5}
\setval{8b.estimators.bench.mbpp-plus.max}{75.6}
\setval{8b.estimators.bench.mbpp-plus.min}{73.4}
\setval{8b.estimators.bench.minerva.max}{51.8}
\setval{8b.estimators.bench.minerva.min}{49.3}
\setval{8b.estimators.bench.olympiad-bench.max}{49.3}
\setval{8b.estimators.bench.olympiad-bench.min}{44.8}
\setval{8b.estimators.bench.writingbench.max}{76.5}
\setval{8b.estimators.bench.writingbench.min}{71.7}
\setval{8b.estimators.eval.avg.max}{58.3}
\setval{8b.estimators.eval.avg.min}{55.8}
\setval{8b.estimators.eval.code.max}{62.7}
\setval{8b.estimators.eval.code.min}{60.8}
\setval{8b.estimators.eval.if.max}{72.0}
\setval{8b.estimators.eval.if.min}{68.5}
\setval{8b.estimators.eval.math.max}{46.6}
\setval{8b.estimators.eval.math.min}{42.8}
\setval{8b.estimators.eval.writing.max}{66.2}
\setval{8b.estimators.eval.writing.min}{63.2}
\setval{8b.estimators.minus.gdpo-noaux.avg-max}{$-$2.2}
\setval{8b.estimators.minus.gdpo-noaux.avg-min}{$-$4.7}
\setval{8b.gd2po.aux-change.code}{+0.18}
\setval{8b.gd2po.aux-change.if}{+0.18}
\setval{8b.gd2po.aux-change.math}{+0.13}
\setval{8b.gd2po.aux-change.writing}{+0.11}
\setval{8b.gd2po.aux-change.writing-rm}{+0.32}
\setval{8b.gd2po.aux-full.code.early}{0.70}
\setval{8b.gd2po.aux-full.code.late}{0.97}
\setval{8b.gd2po.aux-full.code.run}{0.94}
\setval{8b.gd2po.aux-full.if.early}{0.69}
\setval{8b.gd2po.aux-full.if.late}{0.93}
\setval{8b.gd2po.aux-full.if.run}{0.88}
\setval{8b.gd2po.aux-full.math.early}{0.56}
\setval{8b.gd2po.aux-full.math.late}{0.90}
\setval{8b.gd2po.aux-full.math.run}{0.84}
\setval{8b.gd2po.aux-full.writing-rm.early}{0.07}
\setval{8b.gd2po.aux-full.writing-rm.late}{0.61}
\setval{8b.gd2po.aux-full.writing-rm.run}{0.47}
\setval{8b.gd2po.aux-full.writing.early}{0.64}
\setval{8b.gd2po.aux-full.writing.late}{0.70}
\setval{8b.gd2po.aux-full.writing.run}{0.62}
\setval{8b.gd2po.aux-level.code.early}{0.93}
\setval{8b.gd2po.aux-level.code.late}{0.99}
\setval{8b.gd2po.aux-level.code.run}{0.99}
\setval{8b.gd2po.aux-level.if.early}{0.86}
\setval{8b.gd2po.aux-level.if.late}{0.97}
\setval{8b.gd2po.aux-level.if.run}{0.94}
\setval{8b.gd2po.aux-level.math.early}{0.90}
\setval{8b.gd2po.aux-level.math.late}{0.98}
\setval{8b.gd2po.aux-level.math.run}{0.97}
\setval{8b.gd2po.aux-level.writing-rm.early}{0.77}
\setval{8b.gd2po.aux-level.writing-rm.late}{0.96}
\setval{8b.gd2po.aux-level.writing-rm.run}{0.93}
\setval{8b.gd2po.aux-level.writing.early}{0.90}
\setval{8b.gd2po.aux-level.writing.late}{0.95}
\setval{8b.gd2po.aux-level.writing.run}{0.92}
\setval{8b.gd2po.bench.aime24}{16.2}
\setval{8b.gd2po.bench.aime25}{12.2}
\setval{8b.gd2po.bench.amc23}{59.6}
\setval{8b.gd2po.bench.creative-writing-v3}{48.2}
\setval{8b.gd2po.bench.guidebench}{83.7}
\setval{8b.gd2po.bench.humaneval-plus}{84.0}
\setval{8b.gd2po.bench.ifbench}{43.0}
\setval{8b.gd2po.bench.ifeval}{89.3}
\setval{8b.gd2po.bench.livecodebench}{26.3}
\setval{8b.gd2po.bench.longbench-write}{69.9}
\setval{8b.gd2po.bench.math500}{81.2}
\setval{8b.gd2po.bench.mbpp-plus}{75.1}
\setval{8b.gd2po.bench.minerva}{49.9}
\setval{8b.gd2po.bench.olympiad-bench}{48.6}
\setval{8b.gd2po.bench.writingbench}{73.9}
\setval{8b.gd2po.eval.avg}{57.4}
\setval{8b.gd2po.eval.code}{61.8}
\setval{8b.gd2po.eval.if}{72.0}
\setval{8b.gd2po.eval.math}{44.6}
\setval{8b.gd2po.eval.writing}{64.0}
\setval{8b.gd2po.gate.code.against.early}{0.04}
\setval{8b.gd2po.gate.code.against.late}{0.03}
\setval{8b.gd2po.gate.code.against.run}{0.03}
\setval{8b.gd2po.gate.code.agreement}{+0.002}
\setval{8b.gd2po.gate.code.ceiling.early}{0.12}
\setval{8b.gd2po.gate.code.ceiling.late}{0.04}
\setval{8b.gd2po.gate.code.ceiling.run}{0.05}
\setval{8b.gd2po.gate.code.cosine.early}{0.83}
\setval{8b.gd2po.gate.code.cosine.late}{0.76}
\setval{8b.gd2po.gate.code.cosine.run}{0.79}
\setval{8b.gd2po.gate.if.against.early}{0.05}
\setval{8b.gd2po.gate.if.against.late}{0.04}
\setval{8b.gd2po.gate.if.against.run}{0.05}
\setval{8b.gd2po.gate.if.agreement}{+0.013}
\setval{8b.gd2po.gate.if.ceiling.early}{0.06}
\setval{8b.gd2po.gate.if.ceiling.late}{0.18}
\setval{8b.gd2po.gate.if.ceiling.run}{0.12}
\setval{8b.gd2po.gate.if.cosine.early}{0.75}
\setval{8b.gd2po.gate.if.cosine.late}{0.49}
\setval{8b.gd2po.gate.if.cosine.run}{0.59}
\setval{8b.gd2po.gate.math.against.early}{0.00}
\setval{8b.gd2po.gate.math.against.late}{0.00}
\setval{8b.gd2po.gate.math.against.run}{0.00}
\setval{8b.gd2po.gate.math.ceiling.early}{0.23}
\setval{8b.gd2po.gate.math.ceiling.late}{0.19}
\setval{8b.gd2po.gate.math.ceiling.run}{0.19}
\setval{8b.gd2po.gate.math.cosine.early}{0.86}
\setval{8b.gd2po.gate.math.cosine.late}{0.89}
\setval{8b.gd2po.gate.math.cosine.run}{0.89}
\setval{8b.gd2po.gate.writing-rm.against.early}{0.19}
\setval{8b.gd2po.gate.writing-rm.against.late}{0.52}
\setval{8b.gd2po.gate.writing-rm.against.run}{0.48}
\setval{8b.gd2po.gate.writing-rm.agreement}{+0.055}
\setval{8b.gd2po.gate.writing-rm.ceiling.early}{0.00}
\setval{8b.gd2po.gate.writing-rm.ceiling.late}{0.00}
\setval{8b.gd2po.gate.writing-rm.ceiling.run}{0.00}
\setval{8b.gd2po.gate.writing-rm.cosine.early}{0.70}
\setval{8b.gd2po.gate.writing-rm.cosine.late}{0.13}
\setval{8b.gd2po.gate.writing-rm.cosine.run}{0.22}
\setval{8b.gd2po.gate.writing.against.early}{0.29}
\setval{8b.gd2po.gate.writing.against.late}{0.61}
\setval{8b.gd2po.gate.writing.against.run}{0.59}
\setval{8b.gd2po.gate.writing.agreement}{$-$0.049}
\setval{8b.gd2po.gate.writing.ceiling.early}{0.00}
\setval{8b.gd2po.gate.writing.ceiling.late}{0.00}
\setval{8b.gd2po.gate.writing.ceiling.run}{0.00}
\setval{8b.gd2po.gate.writing.cosine.early}{0.51}
\setval{8b.gd2po.gate.writing.cosine.late}{0.03}
\setval{8b.gd2po.gate.writing.cosine.run}{0.08}
\setval{8b.gd2po.judge2.avg}{54.9}
\setval{8b.gd2po.judge2.creative-writing-v3}{36.7}
\setval{8b.gd2po.judge2.longbench-write}{62.2}
\setval{8b.gd2po.judge2.writing}{51.4}
\setval{8b.gd2po.judge2.writingbench}{55.2}
\setval{8b.gd2po.last-step}{300}
\setval{8b.gd2po.minus.base.aime24}{+4.9}
\setval{8b.gd2po.minus.base.aime24-abs}{4.9}
\setval{8b.gd2po.minus.base.aime25}{+3.6}
\setval{8b.gd2po.minus.base.aime25-abs}{3.6}
\setval{8b.gd2po.minus.base.amc23}{+9.4}
\setval{8b.gd2po.minus.base.amc23-abs}{9.4}
\setval{8b.gd2po.minus.base.avg}{+15.9}
\setval{8b.gd2po.minus.base.avg-abs}{15.9}
\setval{8b.gd2po.minus.base.code}{+5.7}
\setval{8b.gd2po.minus.base.code-abs}{5.7}
\setval{8b.gd2po.minus.base.creative-writing-v3}{+21.4}
\setval{8b.gd2po.minus.base.creative-writing-v3-abs}{21.4}
\setval{8b.gd2po.minus.base.guidebench}{+38.1}
\setval{8b.gd2po.minus.base.guidebench-abs}{38.1}
\setval{8b.gd2po.minus.base.humaneval-plus}{+3.9}
\setval{8b.gd2po.minus.base.humaneval-plus-abs}{3.9}
\setval{8b.gd2po.minus.base.if}{+33.4}
\setval{8b.gd2po.minus.base.if-abs}{33.4}
\setval{8b.gd2po.minus.base.ifbench}{+25.7}
\setval{8b.gd2po.minus.base.ifbench-abs}{25.7}
\setval{8b.gd2po.minus.base.ifeval}{+36.3}
\setval{8b.gd2po.minus.base.ifeval-abs}{36.3}
\setval{8b.gd2po.minus.base.judge2.avg}{+13.8}
\setval{8b.gd2po.minus.base.judge2.avg-abs}{13.8}
\setval{8b.gd2po.minus.base.judge2.creative-writing-v3}{+10.3}
\setval{8b.gd2po.minus.base.judge2.creative-writing-v3-abs}{10.3}
\setval{8b.gd2po.minus.base.judge2.longbench-write}{+14.1}
\setval{8b.gd2po.minus.base.judge2.longbench-write-abs}{14.1}
\setval{8b.gd2po.minus.base.judge2.writing}{+15.2}
\setval{8b.gd2po.minus.base.judge2.writing-abs}{15.2}
\setval{8b.gd2po.minus.base.judge2.writingbench}{+21.2}
\setval{8b.gd2po.minus.base.judge2.writingbench-abs}{21.2}
\setval{8b.gd2po.minus.base.livecodebench}{+5.4}
\setval{8b.gd2po.minus.base.livecodebench-abs}{5.4}
\setval{8b.gd2po.minus.base.longbench-write}{+18.5}
\setval{8b.gd2po.minus.base.longbench-write-abs}{18.5}
\setval{8b.gd2po.minus.base.math}{+7.5}
\setval{8b.gd2po.minus.base.math-abs}{7.5}
\setval{8b.gd2po.minus.base.math500}{+8.8}
\setval{8b.gd2po.minus.base.math500-abs}{8.8}
\setval{8b.gd2po.minus.base.mbpp-plus}{+7.8}
\setval{8b.gd2po.minus.base.mbpp-plus-abs}{7.8}
\setval{8b.gd2po.minus.base.minerva}{+9.0}
\setval{8b.gd2po.minus.base.minerva-abs}{9.0}
\setval{8b.gd2po.minus.base.olympiad-bench}{+9.1}
\setval{8b.gd2po.minus.base.olympiad-bench-abs}{9.1}
\setval{8b.gd2po.minus.base.writing}{+25.7}
\setval{8b.gd2po.minus.base.writing-abs}{25.7}
\setval{8b.gd2po.minus.base.writingbench}{+37.1}
\setval{8b.gd2po.minus.base.writingbench-abs}{37.1}
\setval{8b.gd2po.minus.concur-free.aime24}{$-$3.4}
\setval{8b.gd2po.minus.concur-free.aime24-abs}{3.4}
\setval{8b.gd2po.minus.concur-free.aime25}{$-$4.4}
\setval{8b.gd2po.minus.concur-free.aime25-abs}{4.4}
\setval{8b.gd2po.minus.concur-free.amc23}{$-$5.1}
\setval{8b.gd2po.minus.concur-free.amc23-abs}{5.1}
\setval{8b.gd2po.minus.concur-free.avg}{$-$2.9}
\setval{8b.gd2po.minus.concur-free.avg-abs}{2.9}
\setval{8b.gd2po.minus.concur-free.code}{$-$0.7}
\setval{8b.gd2po.minus.concur-free.code-abs}{0.7}
\setval{8b.gd2po.minus.concur-free.creative-writing-v3}{$-$3.5}
\setval{8b.gd2po.minus.concur-free.creative-writing-v3-abs}{3.5}
\setval{8b.gd2po.minus.concur-free.guidebench}{+0.8}
\setval{8b.gd2po.minus.concur-free.guidebench-abs}{0.8}
\setval{8b.gd2po.minus.concur-free.humaneval-plus}{$-$1.3}
\setval{8b.gd2po.minus.concur-free.humaneval-plus-abs}{1.3}
\setval{8b.gd2po.minus.concur-free.if}{+1.7}
\setval{8b.gd2po.minus.concur-free.if-abs}{1.7}
\setval{8b.gd2po.minus.concur-free.ifbench}{+4.3}
\setval{8b.gd2po.minus.concur-free.ifbench-abs}{4.3}
\setval{8b.gd2po.minus.concur-free.ifeval}{+0.1}
\setval{8b.gd2po.minus.concur-free.ifeval-abs}{0.1}
\setval{8b.gd2po.minus.concur-free.judge2.avg}{$-$2.4}
\setval{8b.gd2po.minus.concur-free.judge2.avg-abs}{2.4}
\setval{8b.gd2po.minus.concur-free.judge2.creative-writing-v3}{$-$5.8}
\setval{8b.gd2po.minus.concur-free.judge2.creative-writing-v3-abs}{5.8}
\setval{8b.gd2po.minus.concur-free.judge2.longbench-write}{$-$11.9}
\setval{8b.gd2po.minus.concur-free.judge2.longbench-write-abs}{11.9}
\setval{8b.gd2po.minus.concur-free.judge2.writing}{$-$7.7}
\setval{8b.gd2po.minus.concur-free.judge2.writing-abs}{7.7}
\setval{8b.gd2po.minus.concur-free.judge2.writingbench}{$-$5.4}
\setval{8b.gd2po.minus.concur-free.judge2.writingbench-abs}{5.4}
\setval{8b.gd2po.minus.concur-free.livecodebench}{$-$0.8}
\setval{8b.gd2po.minus.concur-free.livecodebench-abs}{0.8}
\setval{8b.gd2po.minus.concur-free.longbench-write}{$-$13.1}
\setval{8b.gd2po.minus.concur-free.longbench-write-abs}{13.1}
\setval{8b.gd2po.minus.concur-free.math}{$-$2.7}
\setval{8b.gd2po.minus.concur-free.math-abs}{2.7}
\setval{8b.gd2po.minus.concur-free.math500}{$-$2.2}
\setval{8b.gd2po.minus.concur-free.math500-abs}{2.2}
\setval{8b.gd2po.minus.concur-free.mbpp-plus}{+0.0}
\setval{8b.gd2po.minus.concur-free.mbpp-plus-abs}{0.0}
\setval{8b.gd2po.minus.concur-free.minerva}{+0.1}
\setval{8b.gd2po.minus.concur-free.minerva-abs}{0.1}
\setval{8b.gd2po.minus.concur-free.olympiad-bench}{$-$1.5}
\setval{8b.gd2po.minus.concur-free.olympiad-bench-abs}{1.5}
\setval{8b.gd2po.minus.concur-free.writing}{$-$9.8}
\setval{8b.gd2po.minus.concur-free.writing-abs}{9.8}
\setval{8b.gd2po.minus.concur-free.writingbench}{$-$12.9}
\setval{8b.gd2po.minus.concur-free.writingbench-abs}{12.9}
\setval{8b.gd2po.minus.concur-tied.aime24}{$-$15.0}
\setval{8b.gd2po.minus.concur-tied.aime24-abs}{15.0}
\setval{8b.gd2po.minus.concur-tied.aime25}{$-$11.8}
\setval{8b.gd2po.minus.concur-tied.aime25-abs}{11.8}
\setval{8b.gd2po.minus.concur-tied.amc23}{$-$18.9}
\setval{8b.gd2po.minus.concur-tied.amc23-abs}{18.9}
\setval{8b.gd2po.minus.concur-tied.avg}{$-$6.6}
\setval{8b.gd2po.minus.concur-tied.avg-abs}{6.6}
\setval{8b.gd2po.minus.concur-tied.code}{$-$1.4}
\setval{8b.gd2po.minus.concur-tied.code-abs}{1.4}
\setval{8b.gd2po.minus.concur-tied.creative-writing-v3}{$-$4.3}
\setval{8b.gd2po.minus.concur-tied.creative-writing-v3-abs}{4.3}
\setval{8b.gd2po.minus.concur-tied.guidebench}{+0.3}
\setval{8b.gd2po.minus.concur-tied.guidebench-abs}{0.3}
\setval{8b.gd2po.minus.concur-tied.humaneval-plus}{$-$3.4}
\setval{8b.gd2po.minus.concur-tied.humaneval-plus-abs}{3.4}
\setval{8b.gd2po.minus.concur-tied.if}{+1.5}
\setval{8b.gd2po.minus.concur-tied.if-abs}{1.5}
\setval{8b.gd2po.minus.concur-tied.ifbench}{+3.7}
\setval{8b.gd2po.minus.concur-tied.ifbench-abs}{3.7}
\setval{8b.gd2po.minus.concur-tied.ifeval}{+0.5}
\setval{8b.gd2po.minus.concur-tied.ifeval-abs}{0.5}
\setval{8b.gd2po.minus.concur-tied.judge2.avg}{$-$6.4}
\setval{8b.gd2po.minus.concur-tied.judge2.avg-abs}{6.4}
\setval{8b.gd2po.minus.concur-tied.judge2.creative-writing-v3}{$-$5.0}
\setval{8b.gd2po.minus.concur-tied.judge2.creative-writing-v3-abs}{5.0}
\setval{8b.gd2po.minus.concur-tied.judge2.longbench-write}{$-$12.6}
\setval{8b.gd2po.minus.concur-tied.judge2.longbench-write-abs}{12.6}
\setval{8b.gd2po.minus.concur-tied.judge2.writing}{$-$8.7}
\setval{8b.gd2po.minus.concur-tied.judge2.writing-abs}{8.7}
\setval{8b.gd2po.minus.concur-tied.judge2.writingbench}{$-$8.5}
\setval{8b.gd2po.minus.concur-tied.judge2.writingbench-abs}{8.5}
\setval{8b.gd2po.minus.concur-tied.livecodebench}{$-$0.9}
\setval{8b.gd2po.minus.concur-tied.livecodebench-abs}{0.9}
\setval{8b.gd2po.minus.concur-tied.longbench-write}{$-$11.3}
\setval{8b.gd2po.minus.concur-tied.longbench-write-abs}{11.3}
\setval{8b.gd2po.minus.concur-tied.math}{$-$11.8}
\setval{8b.gd2po.minus.concur-tied.math-abs}{11.8}
\setval{8b.gd2po.minus.concur-tied.math500}{$-$8.4}
\setval{8b.gd2po.minus.concur-tied.math500-abs}{8.4}
\setval{8b.gd2po.minus.concur-tied.mbpp-plus}{+0.2}
\setval{8b.gd2po.minus.concur-tied.mbpp-plus-abs}{0.2}
\setval{8b.gd2po.minus.concur-tied.minerva}{$-$5.5}
\setval{8b.gd2po.minus.concur-tied.minerva-abs}{5.5}
\setval{8b.gd2po.minus.concur-tied.olympiad-bench}{$-$11.0}
\setval{8b.gd2po.minus.concur-tied.olympiad-bench-abs}{11.0}
\setval{8b.gd2po.minus.concur-tied.writing}{$-$9.6}
\setval{8b.gd2po.minus.concur-tied.writing-abs}{9.6}
\setval{8b.gd2po.minus.concur-tied.writingbench}{$-$13.3}
\setval{8b.gd2po.minus.concur-tied.writingbench-abs}{13.3}
\setval{8b.gd2po.minus.concur.aime24}{$-$12.5}
\setval{8b.gd2po.minus.concur.aime24-abs}{12.5}
\setval{8b.gd2po.minus.concur.aime25}{$-$10.9}
\setval{8b.gd2po.minus.concur.aime25-abs}{10.9}
\setval{8b.gd2po.minus.concur.amc23}{$-$19.3}
\setval{8b.gd2po.minus.concur.amc23-abs}{19.3}
\setval{8b.gd2po.minus.concur.avg}{$-$5.8}
\setval{8b.gd2po.minus.concur.avg-abs}{5.8}
\setval{8b.gd2po.minus.concur.code}{+1.0}
\setval{8b.gd2po.minus.concur.code-abs}{1.0}
\setval{8b.gd2po.minus.concur.creative-writing-v3}{$-$1.8}
\setval{8b.gd2po.minus.concur.creative-writing-v3-abs}{1.8}
\setval{8b.gd2po.minus.concur.guidebench}{+0.3}
\setval{8b.gd2po.minus.concur.guidebench-abs}{0.3}
\setval{8b.gd2po.minus.concur.humaneval-plus}{+1.2}
\setval{8b.gd2po.minus.concur.humaneval-plus-abs}{1.2}
\setval{8b.gd2po.minus.concur.if}{+1.9}
\setval{8b.gd2po.minus.concur.if-abs}{1.9}
\setval{8b.gd2po.minus.concur.ifbench}{+4.3}
\setval{8b.gd2po.minus.concur.ifbench-abs}{4.3}
\setval{8b.gd2po.minus.concur.ifeval}{+1.2}
\setval{8b.gd2po.minus.concur.ifeval-abs}{1.2}
\setval{8b.gd2po.minus.concur.judge2.avg}{$-$5.5}
\setval{8b.gd2po.minus.concur.judge2.avg-abs}{5.5}
\setval{8b.gd2po.minus.concur.judge2.creative-writing-v3}{$-$1.0}
\setval{8b.gd2po.minus.concur.judge2.creative-writing-v3-abs}{1.0}
\setval{8b.gd2po.minus.concur.judge2.longbench-write}{$-$13.3}
\setval{8b.gd2po.minus.concur.judge2.longbench-write-abs}{13.3}
\setval{8b.gd2po.minus.concur.judge2.writing}{$-$7.7}
\setval{8b.gd2po.minus.concur.judge2.writing-abs}{7.7}
\setval{8b.gd2po.minus.concur.judge2.writingbench}{$-$8.7}
\setval{8b.gd2po.minus.concur.judge2.writingbench-abs}{8.7}
\setval{8b.gd2po.minus.concur.livecodebench}{$-$0.1}
\setval{8b.gd2po.minus.concur.livecodebench-abs}{0.1}
\setval{8b.gd2po.minus.concur.longbench-write}{$-$11.7}
\setval{8b.gd2po.minus.concur.longbench-write-abs}{11.7}
\setval{8b.gd2po.minus.concur.math}{$-$11.3}
\setval{8b.gd2po.minus.concur.math-abs}{11.3}
\setval{8b.gd2po.minus.concur.math500}{$-$8.9}
\setval{8b.gd2po.minus.concur.math500-abs}{8.9}
\setval{8b.gd2po.minus.concur.mbpp-plus}{+1.9}
\setval{8b.gd2po.minus.concur.mbpp-plus-abs}{1.9}
\setval{8b.gd2po.minus.concur.minerva}{$-$4.3}
\setval{8b.gd2po.minus.concur.minerva-abs}{4.3}
\setval{8b.gd2po.minus.concur.olympiad-bench}{$-$11.8}
\setval{8b.gd2po.minus.concur.olympiad-bench-abs}{11.8}
\setval{8b.gd2po.minus.concur.writing}{$-$9.2}
\setval{8b.gd2po.minus.concur.writing-abs}{9.2}
\setval{8b.gd2po.minus.concur.writingbench}{$-$14.2}
\setval{8b.gd2po.minus.concur.writingbench-abs}{14.2}
\setval{8b.gd2po.minus.gdpo-noaux.aime24}{$-$11.7}
\setval{8b.gd2po.minus.gdpo-noaux.aime24-abs}{11.7}
\setval{8b.gd2po.minus.gdpo-noaux.aime25}{$-$11.8}
\setval{8b.gd2po.minus.gdpo-noaux.aime25-abs}{11.8}
\setval{8b.gd2po.minus.gdpo-noaux.amc23}{$-$17.5}
\setval{8b.gd2po.minus.gdpo-noaux.amc23-abs}{17.5}
\setval{8b.gd2po.minus.gdpo-noaux.avg}{$-$3.1}
\setval{8b.gd2po.minus.gdpo-noaux.avg-abs}{3.1}
\setval{8b.gd2po.minus.gdpo-noaux.code}{$-$1.0}
\setval{8b.gd2po.minus.gdpo-noaux.code-abs}{1.0}
\setval{8b.gd2po.minus.gdpo-noaux.creative-writing-v3}{+4.3}
\setval{8b.gd2po.minus.gdpo-noaux.creative-writing-v3-abs}{4.3}
\setval{8b.gd2po.minus.gdpo-noaux.guidebench}{+2.3}
\setval{8b.gd2po.minus.gdpo-noaux.guidebench-abs}{2.3}
\setval{8b.gd2po.minus.gdpo-noaux.humaneval-plus}{$-$2.0}
\setval{8b.gd2po.minus.gdpo-noaux.humaneval-plus-abs}{2.0}
\setval{8b.gd2po.minus.gdpo-noaux.if}{+4.8}
\setval{8b.gd2po.minus.gdpo-noaux.if-abs}{4.8}
\setval{8b.gd2po.minus.gdpo-noaux.ifbench}{+9.3}
\setval{8b.gd2po.minus.gdpo-noaux.ifbench-abs}{9.3}
\setval{8b.gd2po.minus.gdpo-noaux.ifeval}{+2.8}
\setval{8b.gd2po.minus.gdpo-noaux.ifeval-abs}{2.8}
\setval{8b.gd2po.minus.gdpo-noaux.judge2.avg}{$-$2.2}
\setval{8b.gd2po.minus.gdpo-noaux.judge2.avg-abs}{2.2}
\setval{8b.gd2po.minus.gdpo-noaux.judge2.creative-writing-v3}{+7.9}
\setval{8b.gd2po.minus.gdpo-noaux.judge2.creative-writing-v3-abs}{7.9}
\setval{8b.gd2po.minus.gdpo-noaux.judge2.longbench-write}{+7.7}
\setval{8b.gd2po.minus.gdpo-noaux.judge2.longbench-write-abs}{7.7}
\setval{8b.gd2po.minus.gdpo-noaux.judge2.writing}{+6.9}
\setval{8b.gd2po.minus.gdpo-noaux.judge2.writing-abs}{6.9}
\setval{8b.gd2po.minus.gdpo-noaux.judge2.writingbench}{+5.2}
\setval{8b.gd2po.minus.gdpo-noaux.judge2.writingbench-abs}{5.2}
\setval{8b.gd2po.minus.gdpo-noaux.livecodebench}{$-$1.1}
\setval{8b.gd2po.minus.gdpo-noaux.livecodebench-abs}{1.1}
\setval{8b.gd2po.minus.gdpo-noaux.longbench-write}{+3.9}
\setval{8b.gd2po.minus.gdpo-noaux.longbench-write-abs}{3.9}
\setval{8b.gd2po.minus.gdpo-noaux.math}{$-$10.9}
\setval{8b.gd2po.minus.gdpo-noaux.math-abs}{10.9}
\setval{8b.gd2po.minus.gdpo-noaux.math500}{$-$9.0}
\setval{8b.gd2po.minus.gdpo-noaux.math500-abs}{9.0}
\setval{8b.gd2po.minus.gdpo-noaux.mbpp-plus}{+0.2}
\setval{8b.gd2po.minus.gdpo-noaux.mbpp-plus-abs}{0.2}
\setval{8b.gd2po.minus.gdpo-noaux.minerva}{$-$4.1}
\setval{8b.gd2po.minus.gdpo-noaux.minerva-abs}{4.1}
\setval{8b.gd2po.minus.gdpo-noaux.olympiad-bench}{$-$11.4}
\setval{8b.gd2po.minus.gdpo-noaux.olympiad-bench-abs}{11.4}
\setval{8b.gd2po.minus.gdpo-noaux.writing}{+2.4}
\setval{8b.gd2po.minus.gdpo-noaux.writing-abs}{2.4}
\setval{8b.gd2po.minus.gdpo-noaux.writingbench}{$-$1.1}
\setval{8b.gd2po.minus.gdpo-noaux.writingbench-abs}{1.1}
\setval{8b.gd2po.minus.gdpo-w0.5.aime24}{$-$3.0}
\setval{8b.gd2po.minus.gdpo-w0.5.aime24-abs}{3.0}
\setval{8b.gd2po.minus.gdpo-w0.5.aime25}{$-$3.2}
\setval{8b.gd2po.minus.gdpo-w0.5.aime25-abs}{3.2}
\setval{8b.gd2po.minus.gdpo-w0.5.amc23}{$-$3.8}
\setval{8b.gd2po.minus.gdpo-w0.5.amc23-abs}{3.8}
\setval{8b.gd2po.minus.gdpo-w0.5.avg}{$-$0.9}
\setval{8b.gd2po.minus.gdpo-w0.5.avg-abs}{0.9}
\setval{8b.gd2po.minus.gdpo-w0.5.code}{+0.6}
\setval{8b.gd2po.minus.gdpo-w0.5.code-abs}{0.6}
\setval{8b.gd2po.minus.gdpo-w0.5.creative-writing-v3}{$-$3.4}
\setval{8b.gd2po.minus.gdpo-w0.5.creative-writing-v3-abs}{3.4}
\setval{8b.gd2po.minus.gdpo-w0.5.guidebench}{$-$0.3}
\setval{8b.gd2po.minus.gdpo-w0.5.guidebench-abs}{0.3}
\setval{8b.gd2po.minus.gdpo-w0.5.humaneval-plus}{+0.2}
\setval{8b.gd2po.minus.gdpo-w0.5.humaneval-plus-abs}{0.2}
\setval{8b.gd2po.minus.gdpo-w0.5.if}{+2.2}
\setval{8b.gd2po.minus.gdpo-w0.5.if-abs}{2.2}
\setval{8b.gd2po.minus.gdpo-w0.5.ifbench}{+5.3}
\setval{8b.gd2po.minus.gdpo-w0.5.ifbench-abs}{5.3}
\setval{8b.gd2po.minus.gdpo-w0.5.ifeval}{+1.6}
\setval{8b.gd2po.minus.gdpo-w0.5.ifeval-abs}{1.6}
\setval{8b.gd2po.minus.gdpo-w0.5.judge2.avg}{$-$1.0}
\setval{8b.gd2po.minus.gdpo-w0.5.judge2.avg-abs}{1.0}
\setval{8b.gd2po.minus.gdpo-w0.5.judge2.creative-writing-v3}{$-$3.6}
\setval{8b.gd2po.minus.gdpo-w0.5.judge2.creative-writing-v3-abs}{3.6}
\setval{8b.gd2po.minus.gdpo-w0.5.judge2.longbench-write}{$-$1.5}
\setval{8b.gd2po.minus.gdpo-w0.5.judge2.longbench-write-abs}{1.5}
\setval{8b.gd2po.minus.gdpo-w0.5.judge2.writing}{$-$2.0}
\setval{8b.gd2po.minus.gdpo-w0.5.judge2.writing-abs}{2.0}
\setval{8b.gd2po.minus.gdpo-w0.5.judge2.writingbench}{$-$1.0}
\setval{8b.gd2po.minus.gdpo-w0.5.judge2.writingbench-abs}{1.0}
\setval{8b.gd2po.minus.gdpo-w0.5.livecodebench}{$-$0.5}
\setval{8b.gd2po.minus.gdpo-w0.5.livecodebench-abs}{0.5}
\setval{8b.gd2po.minus.gdpo-w0.5.longbench-write}{$-$1.3}
\setval{8b.gd2po.minus.gdpo-w0.5.longbench-write-abs}{1.3}
\setval{8b.gd2po.minus.gdpo-w0.5.math}{$-$2.8}
\setval{8b.gd2po.minus.gdpo-w0.5.math-abs}{2.8}
\setval{8b.gd2po.minus.gdpo-w0.5.math500}{$-$2.8}
\setval{8b.gd2po.minus.gdpo-w0.5.math500-abs}{2.8}
\setval{8b.gd2po.minus.gdpo-w0.5.mbpp-plus}{+2.2}
\setval{8b.gd2po.minus.gdpo-w0.5.mbpp-plus-abs}{2.2}
\setval{8b.gd2po.minus.gdpo-w0.5.minerva}{$-$1.3}
\setval{8b.gd2po.minus.gdpo-w0.5.minerva-abs}{1.3}
\setval{8b.gd2po.minus.gdpo-w0.5.olympiad-bench}{$-$2.6}
\setval{8b.gd2po.minus.gdpo-w0.5.olympiad-bench-abs}{2.6}
\setval{8b.gd2po.minus.gdpo-w0.5.writing}{$-$2.0}
\setval{8b.gd2po.minus.gdpo-w0.5.writing-abs}{2.0}
\setval{8b.gd2po.minus.gdpo-w0.5.writingbench}{$-$1.1}
\setval{8b.gd2po.minus.gdpo-w0.5.writingbench-abs}{1.1}
\setval{8b.gd2po.minus.gdpo.aime24}{$-$5.1}
\setval{8b.gd2po.minus.gdpo.aime24-abs}{5.1}
\setval{8b.gd2po.minus.gdpo.aime25}{$-$2.2}
\setval{8b.gd2po.minus.gdpo.aime25-abs}{2.2}
\setval{8b.gd2po.minus.gdpo.amc23}{$-$2.6}
\setval{8b.gd2po.minus.gdpo.amc23-abs}{2.6}
\setval{8b.gd2po.minus.gdpo.avg}{$-$0.9}
\setval{8b.gd2po.minus.gdpo.avg-abs}{0.9}
\setval{8b.gd2po.minus.gdpo.code}{+0.2}
\setval{8b.gd2po.minus.gdpo.code-abs}{0.2}
\setval{8b.gd2po.minus.gdpo.creative-writing-v3}{$-$4.7}
\setval{8b.gd2po.minus.gdpo.creative-writing-v3-abs}{4.7}
\setval{8b.gd2po.minus.gdpo.guidebench}{$-$0.9}
\setval{8b.gd2po.minus.gdpo.guidebench-abs}{0.9}
\setval{8b.gd2po.minus.gdpo.humaneval-plus}{$-$0.1}
\setval{8b.gd2po.minus.gdpo.humaneval-plus-abs}{0.1}
\setval{8b.gd2po.minus.gdpo.if}{+1.3}
\setval{8b.gd2po.minus.gdpo.if-abs}{1.3}
\setval{8b.gd2po.minus.gdpo.ifbench}{+3.7}
\setval{8b.gd2po.minus.gdpo.ifbench-abs}{3.7}
\setval{8b.gd2po.minus.gdpo.ifeval}{+1.2}
\setval{8b.gd2po.minus.gdpo.ifeval-abs}{1.2}
\setval{8b.gd2po.minus.gdpo.judge2.avg}{$-$1.1}
\setval{8b.gd2po.minus.gdpo.judge2.avg-abs}{1.1}
\setval{8b.gd2po.minus.gdpo.judge2.creative-writing-v3}{$-$4.3}
\setval{8b.gd2po.minus.gdpo.judge2.creative-writing-v3-abs}{4.3}
\setval{8b.gd2po.minus.gdpo.judge2.longbench-write}{$-$2.0}
\setval{8b.gd2po.minus.gdpo.judge2.longbench-write-abs}{2.0}
\setval{8b.gd2po.minus.gdpo.judge2.writing}{$-$3.1}
\setval{8b.gd2po.minus.gdpo.judge2.writing-abs}{3.1}
\setval{8b.gd2po.minus.gdpo.judge2.writingbench}{$-$3.1}
\setval{8b.gd2po.minus.gdpo.judge2.writingbench-abs}{3.1}
\setval{8b.gd2po.minus.gdpo.livecodebench}{+0.1}
\setval{8b.gd2po.minus.gdpo.livecodebench-abs}{0.1}
\setval{8b.gd2po.minus.gdpo.longbench-write}{+0.3}
\setval{8b.gd2po.minus.gdpo.longbench-write-abs}{0.3}
\setval{8b.gd2po.minus.gdpo.math}{$-$2.0}
\setval{8b.gd2po.minus.gdpo.math-abs}{2.0}
\setval{8b.gd2po.minus.gdpo.math500}{$-$1.8}
\setval{8b.gd2po.minus.gdpo.math500-abs}{1.8}
\setval{8b.gd2po.minus.gdpo.mbpp-plus}{+0.6}
\setval{8b.gd2po.minus.gdpo.mbpp-plus-abs}{0.6}
\setval{8b.gd2po.minus.gdpo.minerva}{+0.3}
\setval{8b.gd2po.minus.gdpo.minerva-abs}{0.3}
\setval{8b.gd2po.minus.gdpo.olympiad-bench}{$-$0.8}
\setval{8b.gd2po.minus.gdpo.olympiad-bench-abs}{0.8}
\setval{8b.gd2po.minus.gdpo.writing}{$-$2.2}
\setval{8b.gd2po.minus.gdpo.writing-abs}{2.2}
\setval{8b.gd2po.minus.gdpo.writingbench}{$-$2.2}
\setval{8b.gd2po.minus.gdpo.writingbench-abs}{2.2}
\setval{8b.gd2po.minus.geoma.aime24}{+2.2}
\setval{8b.gd2po.minus.geoma.aime24-abs}{2.2}
\setval{8b.gd2po.minus.geoma.aime25}{$-$0.3}
\setval{8b.gd2po.minus.geoma.aime25-abs}{0.3}
\setval{8b.gd2po.minus.geoma.amc23}{$-$3.9}
\setval{8b.gd2po.minus.geoma.amc23-abs}{3.9}
\setval{8b.gd2po.minus.geoma.avg}{+0.7}
\setval{8b.gd2po.minus.geoma.avg-abs}{0.7}
\setval{8b.gd2po.minus.geoma.code}{+0.4}
\setval{8b.gd2po.minus.geoma.code-abs}{0.4}
\setval{8b.gd2po.minus.geoma.creative-writing-v3}{$-$2.2}
\setval{8b.gd2po.minus.geoma.creative-writing-v3-abs}{2.2}
\setval{8b.gd2po.minus.geoma.guidebench}{$-$0.7}
\setval{8b.gd2po.minus.geoma.guidebench-abs}{0.7}
\setval{8b.gd2po.minus.geoma.humaneval-plus}{+1.2}
\setval{8b.gd2po.minus.geoma.humaneval-plus-abs}{1.2}
\setval{8b.gd2po.minus.geoma.if}{+3.5}
\setval{8b.gd2po.minus.geoma.if-abs}{3.5}
\setval{8b.gd2po.minus.geoma.ifbench}{+9.0}
\setval{8b.gd2po.minus.geoma.ifbench-abs}{9.0}
\setval{8b.gd2po.minus.geoma.ifeval}{+2.0}
\setval{8b.gd2po.minus.geoma.ifeval-abs}{2.0}
\setval{8b.gd2po.minus.geoma.judge2.avg}{+0.5}
\setval{8b.gd2po.minus.geoma.judge2.avg-abs}{0.5}
\setval{8b.gd2po.minus.geoma.judge2.creative-writing-v3}{$-$0.8}
\setval{8b.gd2po.minus.geoma.judge2.creative-writing-v3-abs}{0.8}
\setval{8b.gd2po.minus.geoma.judge2.longbench-write}{$-$2.7}
\setval{8b.gd2po.minus.geoma.judge2.longbench-write-abs}{2.7}
\setval{8b.gd2po.minus.geoma.judge2.writing}{$-$2.2}
\setval{8b.gd2po.minus.geoma.judge2.writing-abs}{2.2}
\setval{8b.gd2po.minus.geoma.judge2.writingbench}{$-$3.1}
\setval{8b.gd2po.minus.geoma.judge2.writingbench-abs}{3.1}
\setval{8b.gd2po.minus.geoma.livecodebench}{+0.6}
\setval{8b.gd2po.minus.geoma.livecodebench-abs}{0.6}
\setval{8b.gd2po.minus.geoma.longbench-write}{+0.9}
\setval{8b.gd2po.minus.geoma.longbench-write-abs}{0.9}
\setval{8b.gd2po.minus.geoma.math}{+0.4}
\setval{8b.gd2po.minus.geoma.math-abs}{0.4}
\setval{8b.gd2po.minus.geoma.math500}{+1.6}
\setval{8b.gd2po.minus.geoma.math500-abs}{1.6}
\setval{8b.gd2po.minus.geoma.mbpp-plus}{$-$0.5}
\setval{8b.gd2po.minus.geoma.mbpp-plus-abs}{0.5}
\setval{8b.gd2po.minus.geoma.minerva}{+0.1}
\setval{8b.gd2po.minus.geoma.minerva-abs}{0.1}
\setval{8b.gd2po.minus.geoma.olympiad-bench}{+2.6}
\setval{8b.gd2po.minus.geoma.olympiad-bench-abs}{2.6}
\setval{8b.gd2po.minus.geoma.writing}{$-$1.1}
\setval{8b.gd2po.minus.geoma.writing-abs}{1.1}
\setval{8b.gd2po.minus.geoma.writingbench}{$-$2.0}
\setval{8b.gd2po.minus.geoma.writingbench-abs}{2.0}
\setval{8b.gd2po.minus.grpo-nodiv.aime24}{$-$4.4}
\setval{8b.gd2po.minus.grpo-nodiv.aime24-abs}{4.4}
\setval{8b.gd2po.minus.grpo-nodiv.aime25}{$-$2.6}
\setval{8b.gd2po.minus.grpo-nodiv.aime25-abs}{2.6}
\setval{8b.gd2po.minus.grpo-nodiv.amc23}{$-$4.0}
\setval{8b.gd2po.minus.grpo-nodiv.amc23-abs}{4.0}
\setval{8b.gd2po.minus.grpo-nodiv.avg}{$-$0.7}
\setval{8b.gd2po.minus.grpo-nodiv.avg-abs}{0.7}
\setval{8b.gd2po.minus.grpo-nodiv.code}{$-$0.7}
\setval{8b.gd2po.minus.grpo-nodiv.code-abs}{0.7}
\setval{8b.gd2po.minus.grpo-nodiv.creative-writing-v3}{+1.5}
\setval{8b.gd2po.minus.grpo-nodiv.creative-writing-v3-abs}{1.5}
\setval{8b.gd2po.minus.grpo-nodiv.guidebench}{$-$1.2}
\setval{8b.gd2po.minus.grpo-nodiv.guidebench-abs}{1.2}
\setval{8b.gd2po.minus.grpo-nodiv.humaneval-plus}{$-$2.4}
\setval{8b.gd2po.minus.grpo-nodiv.humaneval-plus-abs}{2.4}
\setval{8b.gd2po.minus.grpo-nodiv.if}{+1.8}
\setval{8b.gd2po.minus.grpo-nodiv.if-abs}{1.8}
\setval{8b.gd2po.minus.grpo-nodiv.ifbench}{+6.7}
\setval{8b.gd2po.minus.grpo-nodiv.ifbench-abs}{6.7}
\setval{8b.gd2po.minus.grpo-nodiv.ifeval}{$-$0.1}
\setval{8b.gd2po.minus.grpo-nodiv.ifeval-abs}{0.1}
\setval{8b.gd2po.minus.grpo-nodiv.judge2.avg}{$-$1.2}
\setval{8b.gd2po.minus.grpo-nodiv.judge2.avg-abs}{1.2}
\setval{8b.gd2po.minus.grpo-nodiv.judge2.creative-writing-v3}{$-$0.8}
\setval{8b.gd2po.minus.grpo-nodiv.judge2.creative-writing-v3-abs}{0.8}
\setval{8b.gd2po.minus.grpo-nodiv.judge2.longbench-write}{$-$4.4}
\setval{8b.gd2po.minus.grpo-nodiv.judge2.longbench-write-abs}{4.4}
\setval{8b.gd2po.minus.grpo-nodiv.judge2.writing}{$-$1.3}
\setval{8b.gd2po.minus.grpo-nodiv.judge2.writing-abs}{1.3}
\setval{8b.gd2po.minus.grpo-nodiv.judge2.writingbench}{+1.3}
\setval{8b.gd2po.minus.grpo-nodiv.judge2.writingbench-abs}{1.3}
\setval{8b.gd2po.minus.grpo-nodiv.livecodebench}{+1.3}
\setval{8b.gd2po.minus.grpo-nodiv.livecodebench-abs}{1.3}
\setval{8b.gd2po.minus.grpo-nodiv.longbench-write}{$-$1.2}
\setval{8b.gd2po.minus.grpo-nodiv.longbench-write-abs}{1.2}
\setval{8b.gd2po.minus.grpo-nodiv.math}{$-$2.9}
\setval{8b.gd2po.minus.grpo-nodiv.math-abs}{2.9}
\setval{8b.gd2po.minus.grpo-nodiv.math500}{$-$2.9}
\setval{8b.gd2po.minus.grpo-nodiv.math500-abs}{2.9}
\setval{8b.gd2po.minus.grpo-nodiv.mbpp-plus}{$-$1.0}
\setval{8b.gd2po.minus.grpo-nodiv.mbpp-plus-abs}{1.0}
\setval{8b.gd2po.minus.grpo-nodiv.minerva}{$-$0.9}
\setval{8b.gd2po.minus.grpo-nodiv.minerva-abs}{0.9}
\setval{8b.gd2po.minus.grpo-nodiv.olympiad-bench}{$-$2.9}
\setval{8b.gd2po.minus.grpo-nodiv.olympiad-bench-abs}{2.9}
\setval{8b.gd2po.minus.grpo-nodiv.writing}{+1.3}
\setval{8b.gd2po.minus.grpo-nodiv.writing-abs}{1.3}
\setval{8b.gd2po.minus.grpo-nodiv.writingbench}{+3.6}
\setval{8b.gd2po.minus.grpo-nodiv.writingbench-abs}{3.6}
\setval{8b.gd2po.minus.grpo.aime24}{+0.4}
\setval{8b.gd2po.minus.grpo.aime24-abs}{0.4}
\setval{8b.gd2po.minus.grpo.aime25}{$-$3.3}
\setval{8b.gd2po.minus.grpo.aime25-abs}{3.3}
\setval{8b.gd2po.minus.grpo.amc23}{$-$4.8}
\setval{8b.gd2po.minus.grpo.amc23-abs}{4.8}
\setval{8b.gd2po.minus.grpo.avg}{$-$0.5}
\setval{8b.gd2po.minus.grpo.avg-abs}{0.5}
\setval{8b.gd2po.minus.grpo.code}{+0.1}
\setval{8b.gd2po.minus.grpo.code-abs}{0.1}
\setval{8b.gd2po.minus.grpo.creative-writing-v3}{$-$4.3}
\setval{8b.gd2po.minus.grpo.creative-writing-v3-abs}{4.3}
\setval{8b.gd2po.minus.grpo.guidebench}{+1.8}
\setval{8b.gd2po.minus.grpo.guidebench-abs}{1.8}
\setval{8b.gd2po.minus.grpo.humaneval-plus}{+0.2}
\setval{8b.gd2po.minus.grpo.humaneval-plus-abs}{0.2}
\setval{8b.gd2po.minus.grpo.if}{+1.8}
\setval{8b.gd2po.minus.grpo.if-abs}{1.8}
\setval{8b.gd2po.minus.grpo.ifbench}{+1.0}
\setval{8b.gd2po.minus.grpo.ifbench-abs}{1.0}
\setval{8b.gd2po.minus.grpo.ifeval}{+2.5}
\setval{8b.gd2po.minus.grpo.ifeval-abs}{2.5}
\setval{8b.gd2po.minus.grpo.judge2.avg}{$-$0.3}
\setval{8b.gd2po.minus.grpo.judge2.avg-abs}{0.3}
\setval{8b.gd2po.minus.grpo.judge2.creative-writing-v3}{$-$2.5}
\setval{8b.gd2po.minus.grpo.judge2.creative-writing-v3-abs}{2.5}
\setval{8b.gd2po.minus.grpo.judge2.longbench-write}{$-$0.3}
\setval{8b.gd2po.minus.grpo.judge2.longbench-write-abs}{0.3}
\setval{8b.gd2po.minus.grpo.judge2.writing}{$-$1.5}
\setval{8b.gd2po.minus.grpo.judge2.writing-abs}{1.5}
\setval{8b.gd2po.minus.grpo.judge2.writingbench}{$-$1.7}
\setval{8b.gd2po.minus.grpo.judge2.writingbench-abs}{1.7}
\setval{8b.gd2po.minus.grpo.livecodebench}{$-$0.2}
\setval{8b.gd2po.minus.grpo.livecodebench-abs}{0.2}
\setval{8b.gd2po.minus.grpo.longbench-write}{+0.1}
\setval{8b.gd2po.minus.grpo.longbench-write-abs}{0.1}
\setval{8b.gd2po.minus.grpo.math}{$-$1.1}
\setval{8b.gd2po.minus.grpo.math-abs}{1.1}
\setval{8b.gd2po.minus.grpo.math500}{+0.6}
\setval{8b.gd2po.minus.grpo.math500-abs}{0.6}
\setval{8b.gd2po.minus.grpo.mbpp-plus}{+0.4}
\setval{8b.gd2po.minus.grpo.mbpp-plus-abs}{0.4}
\setval{8b.gd2po.minus.grpo.minerva}{+0.1}
\setval{8b.gd2po.minus.grpo.minerva-abs}{0.1}
\setval{8b.gd2po.minus.grpo.olympiad-bench}{+0.6}
\setval{8b.gd2po.minus.grpo.olympiad-bench-abs}{0.6}
\setval{8b.gd2po.minus.grpo.writing}{$-$2.3}
\setval{8b.gd2po.minus.grpo.writing-abs}{2.3}
\setval{8b.gd2po.minus.grpo.writingbench}{$-$2.6}
\setval{8b.gd2po.minus.grpo.writingbench-abs}{2.6}
\setval{8b.gd2po.minus.mt-grpo.aime24}{+1.8}
\setval{8b.gd2po.minus.mt-grpo.aime24-abs}{1.8}
\setval{8b.gd2po.minus.mt-grpo.aime25}{$-$3.5}
\setval{8b.gd2po.minus.mt-grpo.aime25-abs}{3.5}
\setval{8b.gd2po.minus.mt-grpo.amc23}{$-$6.0}
\setval{8b.gd2po.minus.mt-grpo.amc23-abs}{6.0}
\setval{8b.gd2po.minus.mt-grpo.avg}{+0.2}
\setval{8b.gd2po.minus.mt-grpo.avg-abs}{0.2}
\setval{8b.gd2po.minus.mt-grpo.code}{$-$0.9}
\setval{8b.gd2po.minus.mt-grpo.code-abs}{0.9}
\setval{8b.gd2po.minus.mt-grpo.creative-writing-v3}{$-$2.2}
\setval{8b.gd2po.minus.mt-grpo.creative-writing-v3-abs}{2.2}
\setval{8b.gd2po.minus.mt-grpo.guidebench}{+0.6}
\setval{8b.gd2po.minus.mt-grpo.guidebench-abs}{0.6}
\setval{8b.gd2po.minus.mt-grpo.humaneval-plus}{$-$1.1}
\setval{8b.gd2po.minus.mt-grpo.humaneval-plus-abs}{1.1}
\setval{8b.gd2po.minus.mt-grpo.if}{+2.9}
\setval{8b.gd2po.minus.mt-grpo.if-abs}{2.9}
\setval{8b.gd2po.minus.mt-grpo.ifbench}{+5.7}
\setval{8b.gd2po.minus.mt-grpo.ifbench-abs}{5.7}
\setval{8b.gd2po.minus.mt-grpo.ifeval}{+2.4}
\setval{8b.gd2po.minus.mt-grpo.ifeval-abs}{2.4}
\setval{8b.gd2po.minus.mt-grpo.judge2.avg}{$-$0.3}
\setval{8b.gd2po.minus.mt-grpo.judge2.avg-abs}{0.3}
\setval{8b.gd2po.minus.mt-grpo.judge2.creative-writing-v3}{$-$4.8}
\setval{8b.gd2po.minus.mt-grpo.judge2.creative-writing-v3-abs}{4.8}
\setval{8b.gd2po.minus.mt-grpo.judge2.longbench-write}{$-$0.3}
\setval{8b.gd2po.minus.mt-grpo.judge2.longbench-write-abs}{0.3}
\setval{8b.gd2po.minus.mt-grpo.judge2.writing}{$-$1.8}
\setval{8b.gd2po.minus.mt-grpo.judge2.writing-abs}{1.8}
\setval{8b.gd2po.minus.mt-grpo.judge2.writingbench}{$-$0.3}
\setval{8b.gd2po.minus.mt-grpo.judge2.writingbench-abs}{0.3}
\setval{8b.gd2po.minus.mt-grpo.livecodebench}{$-$1.4}
\setval{8b.gd2po.minus.mt-grpo.livecodebench-abs}{1.4}
\setval{8b.gd2po.minus.mt-grpo.longbench-write}{+2.4}
\setval{8b.gd2po.minus.mt-grpo.longbench-write-abs}{2.4}
\setval{8b.gd2po.minus.mt-grpo.math}{$-$0.9}
\setval{8b.gd2po.minus.mt-grpo.math-abs}{0.9}
\setval{8b.gd2po.minus.mt-grpo.math500}{+0.6}
\setval{8b.gd2po.minus.mt-grpo.math500-abs}{0.6}
\setval{8b.gd2po.minus.mt-grpo.mbpp-plus}{$-$0.2}
\setval{8b.gd2po.minus.mt-grpo.mbpp-plus-abs}{0.2}
\setval{8b.gd2po.minus.mt-grpo.minerva}{$-$1.1}
\setval{8b.gd2po.minus.mt-grpo.minerva-abs}{1.1}
\setval{8b.gd2po.minus.mt-grpo.olympiad-bench}{+2.8}
\setval{8b.gd2po.minus.mt-grpo.olympiad-bench-abs}{2.8}
\setval{8b.gd2po.minus.mt-grpo.writing}{+0.8}
\setval{8b.gd2po.minus.mt-grpo.writing-abs}{0.8}
\setval{8b.gd2po.minus.mt-grpo.writingbench}{+2.2}
\setval{8b.gd2po.minus.mt-grpo.writingbench-abs}{2.2}
\setval{8b.gd2po.minus.papo.aime24}{+4.1}
\setval{8b.gd2po.minus.papo.aime24-abs}{4.1}
\setval{8b.gd2po.minus.papo.aime25}{+2.1}
\setval{8b.gd2po.minus.papo.aime25-abs}{2.1}
\setval{8b.gd2po.minus.papo.amc23}{$-$2.4}
\setval{8b.gd2po.minus.papo.amc23-abs}{2.4}
\setval{8b.gd2po.minus.papo.avg}{+1.6}
\setval{8b.gd2po.minus.papo.avg-abs}{1.6}
\setval{8b.gd2po.minus.papo.code}{+0.9}
\setval{8b.gd2po.minus.papo.code-abs}{0.9}
\setval{8b.gd2po.minus.papo.creative-writing-v3}{$-$3.0}
\setval{8b.gd2po.minus.papo.creative-writing-v3-abs}{3.0}
\setval{8b.gd2po.minus.papo.guidebench}{$-$0.7}
\setval{8b.gd2po.minus.papo.guidebench-abs}{0.7}
\setval{8b.gd2po.minus.papo.humaneval-plus}{+1.5}
\setval{8b.gd2po.minus.papo.humaneval-plus-abs}{1.5}
\setval{8b.gd2po.minus.papo.if}{+2.7}
\setval{8b.gd2po.minus.papo.if-abs}{2.7}
\setval{8b.gd2po.minus.papo.ifbench}{+5.7}
\setval{8b.gd2po.minus.papo.ifbench-abs}{5.7}
\setval{8b.gd2po.minus.papo.ifeval}{+3.0}
\setval{8b.gd2po.minus.papo.ifeval-abs}{3.0}
\setval{8b.gd2po.minus.papo.judge2.avg}{+1.2}
\setval{8b.gd2po.minus.papo.judge2.avg-abs}{1.2}
\setval{8b.gd2po.minus.papo.judge2.creative-writing-v3}{$-$5.1}
\setval{8b.gd2po.minus.papo.judge2.creative-writing-v3-abs}{5.1}
\setval{8b.gd2po.minus.papo.judge2.longbench-write}{+2.6}
\setval{8b.gd2po.minus.papo.judge2.longbench-write-abs}{2.6}
\setval{8b.gd2po.minus.papo.judge2.writing}{$-$1.2}
\setval{8b.gd2po.minus.papo.judge2.writing-abs}{1.2}
\setval{8b.gd2po.minus.papo.judge2.writingbench}{$-$1.1}
\setval{8b.gd2po.minus.papo.judge2.writingbench-abs}{1.1}
\setval{8b.gd2po.minus.papo.livecodebench}{$-$0.3}
\setval{8b.gd2po.minus.papo.livecodebench-abs}{0.3}
\setval{8b.gd2po.minus.papo.longbench-write}{+4.4}
\setval{8b.gd2po.minus.papo.longbench-write-abs}{4.4}
\setval{8b.gd2po.minus.papo.math}{+1.8}
\setval{8b.gd2po.minus.papo.math-abs}{1.8}
\setval{8b.gd2po.minus.papo.math500}{+2.7}
\setval{8b.gd2po.minus.papo.math500-abs}{2.7}
\setval{8b.gd2po.minus.papo.mbpp-plus}{+1.7}
\setval{8b.gd2po.minus.papo.mbpp-plus-abs}{1.7}
\setval{8b.gd2po.minus.papo.minerva}{+0.6}
\setval{8b.gd2po.minus.papo.minerva-abs}{0.6}
\setval{8b.gd2po.minus.papo.olympiad-bench}{+3.8}
\setval{8b.gd2po.minus.papo.olympiad-bench-abs}{3.8}
\setval{8b.gd2po.minus.papo.writing}{+0.7}
\setval{8b.gd2po.minus.papo.writing-abs}{0.7}
\setval{8b.gd2po.minus.papo.writingbench}{+0.7}
\setval{8b.gd2po.minus.papo.writingbench-abs}{0.7}
\setval{8b.gd2po.minus.saw.aime24}{+0.9}
\setval{8b.gd2po.minus.saw.aime24-abs}{0.9}
\setval{8b.gd2po.minus.saw.aime25}{$-$3.3}
\setval{8b.gd2po.minus.saw.aime25-abs}{3.3}
\setval{8b.gd2po.minus.saw.amc23}{$-$2.7}
\setval{8b.gd2po.minus.saw.amc23-abs}{2.7}
\setval{8b.gd2po.minus.saw.avg}{$-$0.6}
\setval{8b.gd2po.minus.saw.avg-abs}{0.6}
\setval{8b.gd2po.minus.saw.code}{$-$0.5}
\setval{8b.gd2po.minus.saw.code-abs}{0.5}
\setval{8b.gd2po.minus.saw.creative-writing-v3}{$-$1.5}
\setval{8b.gd2po.minus.saw.creative-writing-v3-abs}{1.5}
\setval{8b.gd2po.minus.saw.guidebench}{$-$0.6}
\setval{8b.gd2po.minus.saw.guidebench-abs}{0.6}
\setval{8b.gd2po.minus.saw.humaneval-plus}{$-$1.9}
\setval{8b.gd2po.minus.saw.humaneval-plus-abs}{1.9}
\setval{8b.gd2po.minus.saw.if}{+1.4}
\setval{8b.gd2po.minus.saw.if-abs}{1.4}
\setval{8b.gd2po.minus.saw.ifbench}{+3.0}
\setval{8b.gd2po.minus.saw.ifbench-abs}{3.0}
\setval{8b.gd2po.minus.saw.ifeval}{+1.7}
\setval{8b.gd2po.minus.saw.ifeval-abs}{1.7}
\setval{8b.gd2po.minus.saw.judge2.avg}{$-$0.6}
\setval{8b.gd2po.minus.saw.judge2.avg-abs}{0.6}
\setval{8b.gd2po.minus.saw.judge2.creative-writing-v3}{+0.2}
\setval{8b.gd2po.minus.saw.judge2.creative-writing-v3-abs}{0.2}
\setval{8b.gd2po.minus.saw.judge2.longbench-write}{$-$2.6}
\setval{8b.gd2po.minus.saw.judge2.longbench-write-abs}{2.6}
\setval{8b.gd2po.minus.saw.judge2.writing}{$-$1.4}
\setval{8b.gd2po.minus.saw.judge2.writing-abs}{1.4}
\setval{8b.gd2po.minus.saw.judge2.writingbench}{$-$1.7}
\setval{8b.gd2po.minus.saw.judge2.writingbench-abs}{1.7}
\setval{8b.gd2po.minus.saw.livecodebench}{$-$0.4}
\setval{8b.gd2po.minus.saw.livecodebench-abs}{0.4}
\setval{8b.gd2po.minus.saw.longbench-write}{$-$1.1}
\setval{8b.gd2po.minus.saw.longbench-write-abs}{1.1}
\setval{8b.gd2po.minus.saw.math}{$-$1.2}
\setval{8b.gd2po.minus.saw.math-abs}{1.2}
\setval{8b.gd2po.minus.saw.math500}{$-$0.7}
\setval{8b.gd2po.minus.saw.math500-abs}{0.7}
\setval{8b.gd2po.minus.saw.mbpp-plus}{+0.7}
\setval{8b.gd2po.minus.saw.mbpp-plus-abs}{0.7}
\setval{8b.gd2po.minus.saw.minerva}{$-$1.8}
\setval{8b.gd2po.minus.saw.minerva-abs}{1.8}
\setval{8b.gd2po.minus.saw.olympiad-bench}{+0.4}
\setval{8b.gd2po.minus.saw.olympiad-bench-abs}{0.4}
\setval{8b.gd2po.minus.saw.writing}{$-$1.4}
\setval{8b.gd2po.minus.saw.writing-abs}{1.4}
\setval{8b.gd2po.minus.saw.writingbench}{$-$1.5}
\setval{8b.gd2po.minus.saw.writingbench-abs}{1.5}
\setval{8b.gd2po.passing-at-ceiling.code.early}{0.72}
\setval{8b.gd2po.passing-at-ceiling.code.late}{0.76}
\setval{8b.gd2po.passing-at-ceiling.code.run}{0.75}
\setval{8b.gd2po.passing-at-ceiling.if.early}{0.32}
\setval{8b.gd2po.passing-at-ceiling.if.late}{0.59}
\setval{8b.gd2po.passing-at-ceiling.if.run}{0.49}
\setval{8b.gd2po.passing-at-ceiling.math.early}{1.00}
\setval{8b.gd2po.passing-at-ceiling.math.late}{1.00}
\setval{8b.gd2po.passing-at-ceiling.math.run}{1.00}
\setval{8b.gd2po.passing-at-ceiling.writing.early}{0.00}
\setval{8b.gd2po.passing-at-ceiling.writing.late}{0.01}
\setval{8b.gd2po.passing-at-ceiling.writing.run}{0.00}
\setval{8b.gd2po.primary-cosine.code.early}{0.94}
\setval{8b.gd2po.primary-cosine.code.late}{0.94}
\setval{8b.gd2po.primary-cosine.code.run}{0.94}
\setval{8b.gd2po.primary-cosine.if.early}{0.68}
\setval{8b.gd2po.primary-cosine.if.late}{0.72}
\setval{8b.gd2po.primary-cosine.if.run}{0.75}
\setval{8b.gd2po.primary-cosine.math.early}{0.95}
\setval{8b.gd2po.primary-cosine.math.late}{0.97}
\setval{8b.gd2po.primary-cosine.math.run}{0.97}
\setval{8b.gd2po.primary-cosine.writing.early}{0.80}
\setval{8b.gd2po.primary-cosine.writing.late}{0.68}
\setval{8b.gd2po.primary-cosine.writing.run}{0.65}
\setval{8b.gd2po.primary-level.code.early}{0.55}
\setval{8b.gd2po.primary-level.code.late}{0.77}
\setval{8b.gd2po.primary-level.code.run}{0.72}
\setval{8b.gd2po.primary-level.if.early}{0.35}
\setval{8b.gd2po.primary-level.if.late}{0.77}
\setval{8b.gd2po.primary-level.if.run}{0.65}
\setval{8b.gd2po.primary-level.math.early}{0.49}
\setval{8b.gd2po.primary-level.math.late}{0.71}
\setval{8b.gd2po.primary-level.math.run}{0.64}
\setval{8b.gd2po.primary-level.writing.early}{0.53}
\setval{8b.gd2po.primary-level.writing.late}{0.76}
\setval{8b.gd2po.primary-level.writing.run}{0.72}
\setval{8b.gd2po.solved.code.early}{0.07}
\setval{8b.gd2po.solved.code.late}{0.02}
\setval{8b.gd2po.solved.code.run}{0.04}
\setval{8b.gd2po.solved.if.early}{0.03}
\setval{8b.gd2po.solved.if.late}{0.21}
\setval{8b.gd2po.solved.if.run}{0.13}
\setval{8b.gd2po.solved.math.early}{0.15}
\setval{8b.gd2po.solved.math.late}{0.15}
\setval{8b.gd2po.solved.math.run}{0.15}
\setval{8b.gd2po.solved.writing.early}{0.00}
\setval{8b.gd2po.solved.writing.late}{0.00}
\setval{8b.gd2po.solved.writing.run}{0.00}
\setval{8b.gd2po.update-share.code.early}{0.25}
\setval{8b.gd2po.update-share.code.late}{0.27}
\setval{8b.gd2po.update-share.code.run}{0.27}
\setval{8b.gd2po.update-share.if.early}{0.14}
\setval{8b.gd2po.update-share.if.late}{0.16}
\setval{8b.gd2po.update-share.if.run}{0.18}
\setval{8b.gd2po.update-share.math.early}{0.25}
\setval{8b.gd2po.update-share.math.late}{0.36}
\setval{8b.gd2po.update-share.math.run}{0.34}
\setval{8b.gd2po.update-share.writing.early}{0.36}
\setval{8b.gd2po.update-share.writing.late}{0.21}
\setval{8b.gd2po.update-share.writing.run}{0.21}
\setval{8b.gdpo-noaux.aux-change.code}{$-$0.25}
\setval{8b.gdpo-noaux.aux-change.if}{+0.05}
\setval{8b.gdpo-noaux.aux-change.math}{$-$0.70}
\setval{8b.gdpo-noaux.aux-change.writing}{$-$0.72}
\setval{8b.gdpo-noaux.aux-change.writing-rm}{+0.20}
\setval{8b.gdpo-noaux.aux-full.code.early}{0.49}
\setval{8b.gdpo-noaux.aux-full.code.late}{0.16}
\setval{8b.gdpo-noaux.aux-full.code.run}{0.24}
\setval{8b.gdpo-noaux.aux-full.if.early}{0.65}
\setval{8b.gdpo-noaux.aux-full.if.late}{0.69}
\setval{8b.gdpo-noaux.aux-full.if.run}{0.65}
\setval{8b.gdpo-noaux.aux-full.math.early}{0.51}
\setval{8b.gdpo-noaux.aux-full.math.late}{0.02}
\setval{8b.gdpo-noaux.aux-full.math.run}{0.09}
\setval{8b.gdpo-noaux.aux-full.writing-rm.early}{0.06}
\setval{8b.gdpo-noaux.aux-full.writing-rm.late}{0.03}
\setval{8b.gdpo-noaux.aux-full.writing-rm.run}{0.12}
\setval{8b.gdpo-noaux.aux-full.writing.early}{0.59}
\setval{8b.gdpo-noaux.aux-full.writing.late}{0.06}
\setval{8b.gdpo-noaux.aux-full.writing.run}{0.12}
\setval{8b.gdpo-noaux.aux-level.code.early}{0.86}
\setval{8b.gdpo-noaux.aux-level.code.late}{0.57}
\setval{8b.gdpo-noaux.aux-level.code.run}{0.69}
\setval{8b.gdpo-noaux.aux-level.if.early}{0.83}
\setval{8b.gdpo-noaux.aux-level.if.late}{0.83}
\setval{8b.gdpo-noaux.aux-level.if.run}{0.81}
\setval{8b.gdpo-noaux.aux-level.math.early}{0.88}
\setval{8b.gdpo-noaux.aux-level.math.late}{0.15}
\setval{8b.gdpo-noaux.aux-level.math.run}{0.29}
\setval{8b.gdpo-noaux.aux-level.writing-rm.early}{0.76}
\setval{8b.gdpo-noaux.aux-level.writing-rm.late}{0.84}
\setval{8b.gdpo-noaux.aux-level.writing-rm.run}{0.86}
\setval{8b.gdpo-noaux.aux-level.writing.early}{0.87}
\setval{8b.gdpo-noaux.aux-level.writing.late}{0.11}
\setval{8b.gdpo-noaux.aux-level.writing.run}{0.23}
\setval{8b.gdpo-noaux.bench.aime24}{27.9}
\setval{8b.gdpo-noaux.bench.aime25}{24.0}
\setval{8b.gdpo-noaux.bench.amc23}{77.1}
\setval{8b.gdpo-noaux.bench.creative-writing-v3}{43.9}
\setval{8b.gdpo-noaux.bench.guidebench}{81.4}
\setval{8b.gdpo-noaux.bench.humaneval-plus}{86.0}
\setval{8b.gdpo-noaux.bench.ifbench}{33.7}
\setval{8b.gdpo-noaux.bench.ifeval}{86.6}
\setval{8b.gdpo-noaux.bench.livecodebench}{27.4}
\setval{8b.gdpo-noaux.bench.longbench-write}{66.0}
\setval{8b.gdpo-noaux.bench.math500}{90.2}
\setval{8b.gdpo-noaux.bench.mbpp-plus}{74.9}
\setval{8b.gdpo-noaux.bench.minerva}{54.0}
\setval{8b.gdpo-noaux.bench.olympiad-bench}{59.9}
\setval{8b.gdpo-noaux.bench.writingbench}{75.0}
\setval{8b.gdpo-noaux.eval.avg}{60.5}
\setval{8b.gdpo-noaux.eval.code}{62.7}
\setval{8b.gdpo-noaux.eval.if}{67.2}
\setval{8b.gdpo-noaux.eval.math}{55.5}
\setval{8b.gdpo-noaux.eval.writing}{61.6}
\setval{8b.gdpo-noaux.gate.code.against.early}{0.05}
\setval{8b.gdpo-noaux.gate.code.against.late}{0.12}
\setval{8b.gdpo-noaux.gate.code.against.run}{0.10}
\setval{8b.gdpo-noaux.gate.code.agreement}{+0.040}
\setval{8b.gdpo-noaux.gate.code.ceiling.early}{0.17}
\setval{8b.gdpo-noaux.gate.code.ceiling.late}{0.54}
\setval{8b.gdpo-noaux.gate.code.ceiling.run}{0.42}
\setval{8b.gdpo-noaux.gate.code.cosine.early}{0.78}
\setval{8b.gdpo-noaux.gate.code.cosine.late}{0.27}
\setval{8b.gdpo-noaux.gate.code.cosine.run}{0.46}
\setval{8b.gdpo-noaux.gate.if.against.early}{0.05}
\setval{8b.gdpo-noaux.gate.if.against.late}{0.07}
\setval{8b.gdpo-noaux.gate.if.against.run}{0.08}
\setval{8b.gdpo-noaux.gate.if.agreement}{+0.034}
\setval{8b.gdpo-noaux.gate.if.ceiling.early}{0.07}
\setval{8b.gdpo-noaux.gate.if.ceiling.late}{0.29}
\setval{8b.gdpo-noaux.gate.if.ceiling.run}{0.19}
\setval{8b.gdpo-noaux.gate.if.cosine.early}{0.72}
\setval{8b.gdpo-noaux.gate.if.cosine.late}{0.40}
\setval{8b.gdpo-noaux.gate.if.cosine.run}{0.50}
\setval{8b.gdpo-noaux.gate.math.against.early}{0.00}
\setval{8b.gdpo-noaux.gate.math.against.late}{0.00}
\setval{8b.gdpo-noaux.gate.math.against.run}{0.00}
\setval{8b.gdpo-noaux.gate.math.ceiling.early}{0.27}
\setval{8b.gdpo-noaux.gate.math.ceiling.late}{0.91}
\setval{8b.gdpo-noaux.gate.math.ceiling.run}{0.79}
\setval{8b.gdpo-noaux.gate.math.cosine.early}{0.82}
\setval{8b.gdpo-noaux.gate.math.cosine.late}{0.14}
\setval{8b.gdpo-noaux.gate.math.cosine.run}{0.27}
\setval{8b.gdpo-noaux.gate.writing-rm.against.early}{0.20}
\setval{8b.gdpo-noaux.gate.writing-rm.against.late}{0.57}
\setval{8b.gdpo-noaux.gate.writing-rm.against.run}{0.52}
\setval{8b.gdpo-noaux.gate.writing-rm.agreement}{+0.076}
\setval{8b.gdpo-noaux.gate.writing-rm.ceiling.early}{0.00}
\setval{8b.gdpo-noaux.gate.writing-rm.ceiling.late}{0.01}
\setval{8b.gdpo-noaux.gate.writing-rm.ceiling.run}{0.01}
\setval{8b.gdpo-noaux.gate.writing-rm.cosine.early}{0.70}
\setval{8b.gdpo-noaux.gate.writing-rm.cosine.late}{0.09}
\setval{8b.gdpo-noaux.gate.writing-rm.cosine.run}{0.19}
\setval{8b.gdpo-noaux.gate.writing.against.early}{0.32}
\setval{8b.gdpo-noaux.gate.writing.against.late}{0.61}
\setval{8b.gdpo-noaux.gate.writing.against.run}{0.62}
\setval{8b.gdpo-noaux.gate.writing.agreement}{$-$0.016}
\setval{8b.gdpo-noaux.gate.writing.ceiling.early}{0.00}
\setval{8b.gdpo-noaux.gate.writing.ceiling.late}{0.01}
\setval{8b.gdpo-noaux.gate.writing.ceiling.run}{0.01}
\setval{8b.gdpo-noaux.gate.writing.cosine.early}{0.50}
\setval{8b.gdpo-noaux.gate.writing.cosine.late}{0.01}
\setval{8b.gdpo-noaux.gate.writing.cosine.run}{0.05}
\setval{8b.gdpo-noaux.judge2.avg}{57.1}
\setval{8b.gdpo-noaux.judge2.creative-writing-v3}{28.9}
\setval{8b.gdpo-noaux.judge2.longbench-write}{54.5}
\setval{8b.gdpo-noaux.judge2.writing}{44.5}
\setval{8b.gdpo-noaux.judge2.writingbench}{50.0}
\setval{8b.gdpo-noaux.last-step}{300}
\setval{8b.gdpo-noaux.minus.base.aime24}{+16.6}
\setval{8b.gdpo-noaux.minus.base.aime24-abs}{16.6}
\setval{8b.gdpo-noaux.minus.base.aime25}{+15.4}
\setval{8b.gdpo-noaux.minus.base.aime25-abs}{15.4}
\setval{8b.gdpo-noaux.minus.base.amc23}{+26.9}
\setval{8b.gdpo-noaux.minus.base.amc23-abs}{26.9}
\setval{8b.gdpo-noaux.minus.base.avg}{+19.1}
\setval{8b.gdpo-noaux.minus.base.avg-abs}{19.1}
\setval{8b.gdpo-noaux.minus.base.code}{+6.7}
\setval{8b.gdpo-noaux.minus.base.code-abs}{6.7}
\setval{8b.gdpo-noaux.minus.base.creative-writing-v3}{+17.2}
\setval{8b.gdpo-noaux.minus.base.creative-writing-v3-abs}{17.2}
\setval{8b.gdpo-noaux.minus.base.guidebench}{+35.8}
\setval{8b.gdpo-noaux.minus.base.guidebench-abs}{35.8}
\setval{8b.gdpo-noaux.minus.base.humaneval-plus}{+5.9}
\setval{8b.gdpo-noaux.minus.base.humaneval-plus-abs}{5.9}
\setval{8b.gdpo-noaux.minus.base.if}{+28.6}
\setval{8b.gdpo-noaux.minus.base.if-abs}{28.6}
\setval{8b.gdpo-noaux.minus.base.ifbench}{+16.3}
\setval{8b.gdpo-noaux.minus.base.ifbench-abs}{16.3}
\setval{8b.gdpo-noaux.minus.base.ifeval}{+33.6}
\setval{8b.gdpo-noaux.minus.base.ifeval-abs}{33.6}
\setval{8b.gdpo-noaux.minus.base.judge2.avg}{+16.0}
\setval{8b.gdpo-noaux.minus.base.judge2.avg-abs}{16.0}
\setval{8b.gdpo-noaux.minus.base.judge2.creative-writing-v3}{+2.4}
\setval{8b.gdpo-noaux.minus.base.judge2.creative-writing-v3-abs}{2.4}
\setval{8b.gdpo-noaux.minus.base.judge2.longbench-write}{+6.4}
\setval{8b.gdpo-noaux.minus.base.judge2.longbench-write-abs}{6.4}
\setval{8b.gdpo-noaux.minus.base.judge2.writing}{+8.3}
\setval{8b.gdpo-noaux.minus.base.judge2.writing-abs}{8.3}
\setval{8b.gdpo-noaux.minus.base.judge2.writingbench}{+16.0}
\setval{8b.gdpo-noaux.minus.base.judge2.writingbench-abs}{16.0}
\setval{8b.gdpo-noaux.minus.base.livecodebench}{+6.5}
\setval{8b.gdpo-noaux.minus.base.livecodebench-abs}{6.5}
\setval{8b.gdpo-noaux.minus.base.longbench-write}{+14.5}
\setval{8b.gdpo-noaux.minus.base.longbench-write-abs}{14.5}
\setval{8b.gdpo-noaux.minus.base.math}{+18.4}
\setval{8b.gdpo-noaux.minus.base.math-abs}{18.4}
\setval{8b.gdpo-noaux.minus.base.math500}{+17.8}
\setval{8b.gdpo-noaux.minus.base.math500-abs}{17.8}
\setval{8b.gdpo-noaux.minus.base.mbpp-plus}{+7.6}
\setval{8b.gdpo-noaux.minus.base.mbpp-plus-abs}{7.6}
\setval{8b.gdpo-noaux.minus.base.minerva}{+13.1}
\setval{8b.gdpo-noaux.minus.base.minerva-abs}{13.1}
\setval{8b.gdpo-noaux.minus.base.olympiad-bench}{+20.5}
\setval{8b.gdpo-noaux.minus.base.olympiad-bench-abs}{20.5}
\setval{8b.gdpo-noaux.minus.base.writing}{+23.3}
\setval{8b.gdpo-noaux.minus.base.writing-abs}{23.3}
\setval{8b.gdpo-noaux.minus.base.writingbench}{+38.2}
\setval{8b.gdpo-noaux.minus.base.writingbench-abs}{38.2}
\setval{8b.gdpo-noaux.minus.concur-free.aime24}{+8.2}
\setval{8b.gdpo-noaux.minus.concur-free.aime24-abs}{8.2}
\setval{8b.gdpo-noaux.minus.concur-free.aime25}{+7.4}
\setval{8b.gdpo-noaux.minus.concur-free.aime25-abs}{7.4}
\setval{8b.gdpo-noaux.minus.concur-free.amc23}{+12.4}
\setval{8b.gdpo-noaux.minus.concur-free.amc23-abs}{12.4}
\setval{8b.gdpo-noaux.minus.concur-free.avg}{+0.3}
\setval{8b.gdpo-noaux.minus.concur-free.avg-abs}{0.3}
\setval{8b.gdpo-noaux.minus.concur-free.code}{+0.2}
\setval{8b.gdpo-noaux.minus.concur-free.code-abs}{0.2}
\setval{8b.gdpo-noaux.minus.concur-free.creative-writing-v3}{$-$7.8}
\setval{8b.gdpo-noaux.minus.concur-free.creative-writing-v3-abs}{7.8}
\setval{8b.gdpo-noaux.minus.concur-free.guidebench}{$-$1.5}
\setval{8b.gdpo-noaux.minus.concur-free.guidebench-abs}{1.5}
\setval{8b.gdpo-noaux.minus.concur-free.humaneval-plus}{+0.7}
\setval{8b.gdpo-noaux.minus.concur-free.humaneval-plus-abs}{0.7}
\setval{8b.gdpo-noaux.minus.concur-free.if}{$-$3.1}
\setval{8b.gdpo-noaux.minus.concur-free.if-abs}{3.1}
\setval{8b.gdpo-noaux.minus.concur-free.ifbench}{$-$5.0}
\setval{8b.gdpo-noaux.minus.concur-free.ifbench-abs}{5.0}
\setval{8b.gdpo-noaux.minus.concur-free.ifeval}{$-$2.6}
\setval{8b.gdpo-noaux.minus.concur-free.ifeval-abs}{2.6}
\setval{8b.gdpo-noaux.minus.concur-free.judge2.avg}{$-$0.2}
\setval{8b.gdpo-noaux.minus.concur-free.judge2.avg-abs}{0.2}
\setval{8b.gdpo-noaux.minus.concur-free.judge2.creative-writing-v3}{$-$13.6}
\setval{8b.gdpo-noaux.minus.concur-free.judge2.creative-writing-v3-abs}{13.6}
\setval{8b.gdpo-noaux.minus.concur-free.judge2.longbench-write}{$-$19.6}
\setval{8b.gdpo-noaux.minus.concur-free.judge2.longbench-write-abs}{19.6}
\setval{8b.gdpo-noaux.minus.concur-free.judge2.writing}{$-$14.6}
\setval{8b.gdpo-noaux.minus.concur-free.judge2.writing-abs}{14.6}
\setval{8b.gdpo-noaux.minus.concur-free.judge2.writingbench}{$-$10.6}
\setval{8b.gdpo-noaux.minus.concur-free.judge2.writingbench-abs}{10.6}
\setval{8b.gdpo-noaux.minus.concur-free.livecodebench}{+0.3}
\setval{8b.gdpo-noaux.minus.concur-free.livecodebench-abs}{0.3}
\setval{8b.gdpo-noaux.minus.concur-free.longbench-write}{$-$17.0}
\setval{8b.gdpo-noaux.minus.concur-free.longbench-write-abs}{17.0}
\setval{8b.gdpo-noaux.minus.concur-free.math}{+8.2}
\setval{8b.gdpo-noaux.minus.concur-free.math-abs}{8.2}
\setval{8b.gdpo-noaux.minus.concur-free.math500}{+6.8}
\setval{8b.gdpo-noaux.minus.concur-free.math500-abs}{6.8}
\setval{8b.gdpo-noaux.minus.concur-free.mbpp-plus}{$-$0.2}
\setval{8b.gdpo-noaux.minus.concur-free.mbpp-plus-abs}{0.2}
\setval{8b.gdpo-noaux.minus.concur-free.minerva}{+4.2}
\setval{8b.gdpo-noaux.minus.concur-free.minerva-abs}{4.2}
\setval{8b.gdpo-noaux.minus.concur-free.olympiad-bench}{+9.9}
\setval{8b.gdpo-noaux.minus.concur-free.olympiad-bench-abs}{9.9}
\setval{8b.gdpo-noaux.minus.concur-free.writing}{$-$12.2}
\setval{8b.gdpo-noaux.minus.concur-free.writing-abs}{12.2}
\setval{8b.gdpo-noaux.minus.concur-free.writingbench}{$-$11.8}
\setval{8b.gdpo-noaux.minus.concur-free.writingbench-abs}{11.8}
\setval{8b.gdpo-noaux.minus.concur-tied.aime24}{$-$3.3}
\setval{8b.gdpo-noaux.minus.concur-tied.aime24-abs}{3.3}
\setval{8b.gdpo-noaux.minus.concur-tied.aime25}{+0.0}
\setval{8b.gdpo-noaux.minus.concur-tied.aime25-abs}{0.0}
\setval{8b.gdpo-noaux.minus.concur-tied.amc23}{$-$1.4}
\setval{8b.gdpo-noaux.minus.concur-tied.amc23-abs}{1.4}
\setval{8b.gdpo-noaux.minus.concur-tied.avg}{$-$3.5}
\setval{8b.gdpo-noaux.minus.concur-tied.avg-abs}{3.5}
\setval{8b.gdpo-noaux.minus.concur-tied.code}{$-$0.4}
\setval{8b.gdpo-noaux.minus.concur-tied.code-abs}{0.4}
\setval{8b.gdpo-noaux.minus.concur-tied.creative-writing-v3}{$-$8.5}
\setval{8b.gdpo-noaux.minus.concur-tied.creative-writing-v3-abs}{8.5}
\setval{8b.gdpo-noaux.minus.concur-tied.guidebench}{$-$2.0}
\setval{8b.gdpo-noaux.minus.concur-tied.guidebench-abs}{2.0}
\setval{8b.gdpo-noaux.minus.concur-tied.humaneval-plus}{$-$1.4}
\setval{8b.gdpo-noaux.minus.concur-tied.humaneval-plus-abs}{1.4}
\setval{8b.gdpo-noaux.minus.concur-tied.if}{$-$3.3}
\setval{8b.gdpo-noaux.minus.concur-tied.if-abs}{3.3}
\setval{8b.gdpo-noaux.minus.concur-tied.ifbench}{$-$5.7}
\setval{8b.gdpo-noaux.minus.concur-tied.ifbench-abs}{5.7}
\setval{8b.gdpo-noaux.minus.concur-tied.ifeval}{$-$2.3}
\setval{8b.gdpo-noaux.minus.concur-tied.ifeval-abs}{2.3}
\setval{8b.gdpo-noaux.minus.concur-tied.judge2.avg}{$-$4.2}
\setval{8b.gdpo-noaux.minus.concur-tied.judge2.avg-abs}{4.2}
\setval{8b.gdpo-noaux.minus.concur-tied.judge2.creative-writing-v3}{$-$12.9}
\setval{8b.gdpo-noaux.minus.concur-tied.judge2.creative-writing-v3-abs}{12.9}
\setval{8b.gdpo-noaux.minus.concur-tied.judge2.longbench-write}{$-$20.3}
\setval{8b.gdpo-noaux.minus.concur-tied.judge2.longbench-write-abs}{20.3}
\setval{8b.gdpo-noaux.minus.concur-tied.judge2.writing}{$-$15.6}
\setval{8b.gdpo-noaux.minus.concur-tied.judge2.writing-abs}{15.6}
\setval{8b.gdpo-noaux.minus.concur-tied.judge2.writingbench}{$-$13.7}
\setval{8b.gdpo-noaux.minus.concur-tied.judge2.writingbench-abs}{13.7}
\setval{8b.gdpo-noaux.minus.concur-tied.livecodebench}{+0.2}
\setval{8b.gdpo-noaux.minus.concur-tied.livecodebench-abs}{0.2}
\setval{8b.gdpo-noaux.minus.concur-tied.longbench-write}{$-$15.3}
\setval{8b.gdpo-noaux.minus.concur-tied.longbench-write-abs}{15.3}
\setval{8b.gdpo-noaux.minus.concur-tied.math}{$-$0.9}
\setval{8b.gdpo-noaux.minus.concur-tied.math-abs}{0.9}
\setval{8b.gdpo-noaux.minus.concur-tied.math500}{+0.6}
\setval{8b.gdpo-noaux.minus.concur-tied.math500-abs}{0.6}
\setval{8b.gdpo-noaux.minus.concur-tied.mbpp-plus}{+0.0}
\setval{8b.gdpo-noaux.minus.concur-tied.mbpp-plus-abs}{0.0}
\setval{8b.gdpo-noaux.minus.concur-tied.minerva}{$-$1.4}
\setval{8b.gdpo-noaux.minus.concur-tied.minerva-abs}{1.4}
\setval{8b.gdpo-noaux.minus.concur-tied.olympiad-bench}{+0.4}
\setval{8b.gdpo-noaux.minus.concur-tied.olympiad-bench-abs}{0.4}
\setval{8b.gdpo-noaux.minus.concur-tied.writing}{$-$12.0}
\setval{8b.gdpo-noaux.minus.concur-tied.writing-abs}{12.0}
\setval{8b.gdpo-noaux.minus.concur-tied.writingbench}{$-$12.2}
\setval{8b.gdpo-noaux.minus.concur-tied.writingbench-abs}{12.2}
\setval{8b.gdpo-noaux.minus.concur.aime24}{$-$0.8}
\setval{8b.gdpo-noaux.minus.concur.aime24-abs}{0.8}
\setval{8b.gdpo-noaux.minus.concur.aime25}{+0.8}
\setval{8b.gdpo-noaux.minus.concur.aime25-abs}{0.8}
\setval{8b.gdpo-noaux.minus.concur.amc23}{$-$1.8}
\setval{8b.gdpo-noaux.minus.concur.amc23-abs}{1.8}
\setval{8b.gdpo-noaux.minus.concur.avg}{$-$2.6}
\setval{8b.gdpo-noaux.minus.concur.avg-abs}{2.6}
\setval{8b.gdpo-noaux.minus.concur.code}{+2.0}
\setval{8b.gdpo-noaux.minus.concur.code-abs}{2.0}
\setval{8b.gdpo-noaux.minus.concur.creative-writing-v3}{$-$6.0}
\setval{8b.gdpo-noaux.minus.concur.creative-writing-v3-abs}{6.0}
\setval{8b.gdpo-noaux.minus.concur.guidebench}{$-$2.0}
\setval{8b.gdpo-noaux.minus.concur.guidebench-abs}{2.0}
\setval{8b.gdpo-noaux.minus.concur.humaneval-plus}{+3.2}
\setval{8b.gdpo-noaux.minus.concur.humaneval-plus-abs}{3.2}
\setval{8b.gdpo-noaux.minus.concur.if}{$-$2.9}
\setval{8b.gdpo-noaux.minus.concur.if-abs}{2.9}
\setval{8b.gdpo-noaux.minus.concur.ifbench}{$-$5.0}
\setval{8b.gdpo-noaux.minus.concur.ifbench-abs}{5.0}
\setval{8b.gdpo-noaux.minus.concur.ifeval}{$-$1.6}
\setval{8b.gdpo-noaux.minus.concur.ifeval-abs}{1.6}
\setval{8b.gdpo-noaux.minus.concur.judge2.avg}{$-$3.3}
\setval{8b.gdpo-noaux.minus.concur.judge2.avg-abs}{3.3}
\setval{8b.gdpo-noaux.minus.concur.judge2.creative-writing-v3}{$-$8.9}
\setval{8b.gdpo-noaux.minus.concur.judge2.creative-writing-v3-abs}{8.9}
\setval{8b.gdpo-noaux.minus.concur.judge2.longbench-write}{$-$21.0}
\setval{8b.gdpo-noaux.minus.concur.judge2.longbench-write-abs}{21.0}
\setval{8b.gdpo-noaux.minus.concur.judge2.writing}{$-$14.6}
\setval{8b.gdpo-noaux.minus.concur.judge2.writing-abs}{14.6}
\setval{8b.gdpo-noaux.minus.concur.judge2.writingbench}{$-$13.9}
\setval{8b.gdpo-noaux.minus.concur.judge2.writingbench-abs}{13.9}
\setval{8b.gdpo-noaux.minus.concur.livecodebench}{+1.0}
\setval{8b.gdpo-noaux.minus.concur.livecodebench-abs}{1.0}
\setval{8b.gdpo-noaux.minus.concur.longbench-write}{$-$15.6}
\setval{8b.gdpo-noaux.minus.concur.longbench-write-abs}{15.6}
\setval{8b.gdpo-noaux.minus.concur.math}{$-$0.4}
\setval{8b.gdpo-noaux.minus.concur.math-abs}{0.4}
\setval{8b.gdpo-noaux.minus.concur.math500}{+0.1}
\setval{8b.gdpo-noaux.minus.concur.math500-abs}{0.1}
\setval{8b.gdpo-noaux.minus.concur.mbpp-plus}{+1.7}
\setval{8b.gdpo-noaux.minus.concur.mbpp-plus-abs}{1.7}
\setval{8b.gdpo-noaux.minus.concur.minerva}{$-$0.2}
\setval{8b.gdpo-noaux.minus.concur.minerva-abs}{0.2}
\setval{8b.gdpo-noaux.minus.concur.olympiad-bench}{$-$0.4}
\setval{8b.gdpo-noaux.minus.concur.olympiad-bench-abs}{0.4}
\setval{8b.gdpo-noaux.minus.concur.writing}{$-$11.6}
\setval{8b.gdpo-noaux.minus.concur.writing-abs}{11.6}
\setval{8b.gdpo-noaux.minus.concur.writingbench}{$-$13.1}
\setval{8b.gdpo-noaux.minus.concur.writingbench-abs}{13.1}
\setval{8b.gdpo-noaux.minus.gd2po.aime24}{+11.7}
\setval{8b.gdpo-noaux.minus.gd2po.aime24-abs}{11.7}
\setval{8b.gdpo-noaux.minus.gd2po.aime25}{+11.8}
\setval{8b.gdpo-noaux.minus.gd2po.aime25-abs}{11.8}
\setval{8b.gdpo-noaux.minus.gd2po.amc23}{+17.5}
\setval{8b.gdpo-noaux.minus.gd2po.amc23-abs}{17.5}
\setval{8b.gdpo-noaux.minus.gd2po.avg}{+3.1}
\setval{8b.gdpo-noaux.minus.gd2po.avg-abs}{3.1}
\setval{8b.gdpo-noaux.minus.gd2po.code}{+1.0}
\setval{8b.gdpo-noaux.minus.gd2po.code-abs}{1.0}
\setval{8b.gdpo-noaux.minus.gd2po.creative-writing-v3}{$-$4.3}
\setval{8b.gdpo-noaux.minus.gd2po.creative-writing-v3-abs}{4.3}
\setval{8b.gdpo-noaux.minus.gd2po.guidebench}{$-$2.3}
\setval{8b.gdpo-noaux.minus.gd2po.guidebench-abs}{2.3}
\setval{8b.gdpo-noaux.minus.gd2po.humaneval-plus}{+2.0}
\setval{8b.gdpo-noaux.minus.gd2po.humaneval-plus-abs}{2.0}
\setval{8b.gdpo-noaux.minus.gd2po.if}{$-$4.8}
\setval{8b.gdpo-noaux.minus.gd2po.if-abs}{4.8}
\setval{8b.gdpo-noaux.minus.gd2po.ifbench}{$-$9.3}
\setval{8b.gdpo-noaux.minus.gd2po.ifbench-abs}{9.3}
\setval{8b.gdpo-noaux.minus.gd2po.ifeval}{$-$2.8}
\setval{8b.gdpo-noaux.minus.gd2po.ifeval-abs}{2.8}
\setval{8b.gdpo-noaux.minus.gd2po.judge2.avg}{+2.2}
\setval{8b.gdpo-noaux.minus.gd2po.judge2.avg-abs}{2.2}
\setval{8b.gdpo-noaux.minus.gd2po.judge2.creative-writing-v3}{$-$7.9}
\setval{8b.gdpo-noaux.minus.gd2po.judge2.creative-writing-v3-abs}{7.9}
\setval{8b.gdpo-noaux.minus.gd2po.judge2.longbench-write}{$-$7.7}
\setval{8b.gdpo-noaux.minus.gd2po.judge2.longbench-write-abs}{7.7}
\setval{8b.gdpo-noaux.minus.gd2po.judge2.writing}{$-$6.9}
\setval{8b.gdpo-noaux.minus.gd2po.judge2.writing-abs}{6.9}
\setval{8b.gdpo-noaux.minus.gd2po.judge2.writingbench}{$-$5.2}
\setval{8b.gdpo-noaux.minus.gd2po.judge2.writingbench-abs}{5.2}
\setval{8b.gdpo-noaux.minus.gd2po.livecodebench}{+1.1}
\setval{8b.gdpo-noaux.minus.gd2po.livecodebench-abs}{1.1}
\setval{8b.gdpo-noaux.minus.gd2po.longbench-write}{$-$3.9}
\setval{8b.gdpo-noaux.minus.gd2po.longbench-write-abs}{3.9}
\setval{8b.gdpo-noaux.minus.gd2po.math}{+10.9}
\setval{8b.gdpo-noaux.minus.gd2po.math-abs}{10.9}
\setval{8b.gdpo-noaux.minus.gd2po.math500}{+9.0}
\setval{8b.gdpo-noaux.minus.gd2po.math500-abs}{9.0}
\setval{8b.gdpo-noaux.minus.gd2po.mbpp-plus}{$-$0.2}
\setval{8b.gdpo-noaux.minus.gd2po.mbpp-plus-abs}{0.2}
\setval{8b.gdpo-noaux.minus.gd2po.minerva}{+4.1}
\setval{8b.gdpo-noaux.minus.gd2po.minerva-abs}{4.1}
\setval{8b.gdpo-noaux.minus.gd2po.olympiad-bench}{+11.4}
\setval{8b.gdpo-noaux.minus.gd2po.olympiad-bench-abs}{11.4}
\setval{8b.gdpo-noaux.minus.gd2po.writing}{$-$2.4}
\setval{8b.gdpo-noaux.minus.gd2po.writing-abs}{2.4}
\setval{8b.gdpo-noaux.minus.gd2po.writingbench}{+1.1}
\setval{8b.gdpo-noaux.minus.gd2po.writingbench-abs}{1.1}
\setval{8b.gdpo-noaux.minus.gdpo-w0.5.aime24}{+8.6}
\setval{8b.gdpo-noaux.minus.gdpo-w0.5.aime24-abs}{8.6}
\setval{8b.gdpo-noaux.minus.gdpo-w0.5.aime25}{+8.5}
\setval{8b.gdpo-noaux.minus.gdpo-w0.5.aime25-abs}{8.5}
\setval{8b.gdpo-noaux.minus.gdpo-w0.5.amc23}{+13.7}
\setval{8b.gdpo-noaux.minus.gdpo-w0.5.amc23-abs}{13.7}
\setval{8b.gdpo-noaux.minus.gdpo-w0.5.avg}{+2.2}
\setval{8b.gdpo-noaux.minus.gdpo-w0.5.avg-abs}{2.2}
\setval{8b.gdpo-noaux.minus.gdpo-w0.5.code}{+1.6}
\setval{8b.gdpo-noaux.minus.gdpo-w0.5.code-abs}{1.6}
\setval{8b.gdpo-noaux.minus.gdpo-w0.5.creative-writing-v3}{$-$7.7}
\setval{8b.gdpo-noaux.minus.gdpo-w0.5.creative-writing-v3-abs}{7.7}
\setval{8b.gdpo-noaux.minus.gdpo-w0.5.guidebench}{$-$2.6}
\setval{8b.gdpo-noaux.minus.gdpo-w0.5.guidebench-abs}{2.6}
\setval{8b.gdpo-noaux.minus.gdpo-w0.5.humaneval-plus}{+2.2}
\setval{8b.gdpo-noaux.minus.gdpo-w0.5.humaneval-plus-abs}{2.2}
\setval{8b.gdpo-noaux.minus.gdpo-w0.5.if}{$-$2.6}
\setval{8b.gdpo-noaux.minus.gdpo-w0.5.if-abs}{2.6}
\setval{8b.gdpo-noaux.minus.gdpo-w0.5.ifbench}{$-$4.0}
\setval{8b.gdpo-noaux.minus.gdpo-w0.5.ifbench-abs}{4.0}
\setval{8b.gdpo-noaux.minus.gdpo-w0.5.ifeval}{$-$1.2}
\setval{8b.gdpo-noaux.minus.gdpo-w0.5.ifeval-abs}{1.2}
\setval{8b.gdpo-noaux.minus.gdpo-w0.5.judge2.avg}{+1.3}
\setval{8b.gdpo-noaux.minus.gdpo-w0.5.judge2.avg-abs}{1.3}
\setval{8b.gdpo-noaux.minus.gdpo-w0.5.judge2.creative-writing-v3}{$-$11.5}
\setval{8b.gdpo-noaux.minus.gdpo-w0.5.judge2.creative-writing-v3-abs}{11.5}
\setval{8b.gdpo-noaux.minus.gdpo-w0.5.judge2.longbench-write}{$-$9.2}
\setval{8b.gdpo-noaux.minus.gdpo-w0.5.judge2.longbench-write-abs}{9.2}
\setval{8b.gdpo-noaux.minus.gdpo-w0.5.judge2.writing}{$-$8.9}
\setval{8b.gdpo-noaux.minus.gdpo-w0.5.judge2.writing-abs}{8.9}
\setval{8b.gdpo-noaux.minus.gdpo-w0.5.judge2.writingbench}{$-$6.2}
\setval{8b.gdpo-noaux.minus.gdpo-w0.5.judge2.writingbench-abs}{6.2}
\setval{8b.gdpo-noaux.minus.gdpo-w0.5.livecodebench}{+0.6}
\setval{8b.gdpo-noaux.minus.gdpo-w0.5.livecodebench-abs}{0.6}
\setval{8b.gdpo-noaux.minus.gdpo-w0.5.longbench-write}{$-$5.2}
\setval{8b.gdpo-noaux.minus.gdpo-w0.5.longbench-write-abs}{5.2}
\setval{8b.gdpo-noaux.minus.gdpo-w0.5.math}{+8.1}
\setval{8b.gdpo-noaux.minus.gdpo-w0.5.math-abs}{8.1}
\setval{8b.gdpo-noaux.minus.gdpo-w0.5.math500}{+6.2}
\setval{8b.gdpo-noaux.minus.gdpo-w0.5.math500-abs}{6.2}
\setval{8b.gdpo-noaux.minus.gdpo-w0.5.mbpp-plus}{+2.0}
\setval{8b.gdpo-noaux.minus.gdpo-w0.5.mbpp-plus-abs}{2.0}
\setval{8b.gdpo-noaux.minus.gdpo-w0.5.minerva}{+2.8}
\setval{8b.gdpo-noaux.minus.gdpo-w0.5.minerva-abs}{2.8}
\setval{8b.gdpo-noaux.minus.gdpo-w0.5.olympiad-bench}{+8.8}
\setval{8b.gdpo-noaux.minus.gdpo-w0.5.olympiad-bench-abs}{8.8}
\setval{8b.gdpo-noaux.minus.gdpo-w0.5.writing}{$-$4.3}
\setval{8b.gdpo-noaux.minus.gdpo-w0.5.writing-abs}{4.3}
\setval{8b.gdpo-noaux.minus.gdpo-w0.5.writingbench}{+0.0}
\setval{8b.gdpo-noaux.minus.gdpo-w0.5.writingbench-abs}{0.0}
\setval{8b.gdpo-noaux.minus.gdpo.aime24}{+6.6}
\setval{8b.gdpo-noaux.minus.gdpo.aime24-abs}{6.6}
\setval{8b.gdpo-noaux.minus.gdpo.aime25}{+9.6}
\setval{8b.gdpo-noaux.minus.gdpo.aime25-abs}{9.6}
\setval{8b.gdpo-noaux.minus.gdpo.amc23}{+14.9}
\setval{8b.gdpo-noaux.minus.gdpo.amc23-abs}{14.9}
\setval{8b.gdpo-noaux.minus.gdpo.avg}{+2.2}
\setval{8b.gdpo-noaux.minus.gdpo.avg-abs}{2.2}
\setval{8b.gdpo-noaux.minus.gdpo.code}{+1.1}
\setval{8b.gdpo-noaux.minus.gdpo.code-abs}{1.1}
\setval{8b.gdpo-noaux.minus.gdpo.creative-writing-v3}{$-$9.0}
\setval{8b.gdpo-noaux.minus.gdpo.creative-writing-v3-abs}{9.0}
\setval{8b.gdpo-noaux.minus.gdpo.guidebench}{$-$3.2}
\setval{8b.gdpo-noaux.minus.gdpo.guidebench-abs}{3.2}
\setval{8b.gdpo-noaux.minus.gdpo.humaneval-plus}{+1.9}
\setval{8b.gdpo-noaux.minus.gdpo.humaneval-plus-abs}{1.9}
\setval{8b.gdpo-noaux.minus.gdpo.if}{$-$3.5}
\setval{8b.gdpo-noaux.minus.gdpo.if-abs}{3.5}
\setval{8b.gdpo-noaux.minus.gdpo.ifbench}{$-$5.7}
\setval{8b.gdpo-noaux.minus.gdpo.ifbench-abs}{5.7}
\setval{8b.gdpo-noaux.minus.gdpo.ifeval}{$-$1.6}
\setval{8b.gdpo-noaux.minus.gdpo.ifeval-abs}{1.6}
\setval{8b.gdpo-noaux.minus.gdpo.judge2.avg}{+1.1}
\setval{8b.gdpo-noaux.minus.gdpo.judge2.avg-abs}{1.1}
\setval{8b.gdpo-noaux.minus.gdpo.judge2.creative-writing-v3}{$-$12.2}
\setval{8b.gdpo-noaux.minus.gdpo.judge2.creative-writing-v3-abs}{12.2}
\setval{8b.gdpo-noaux.minus.gdpo.judge2.longbench-write}{$-$9.7}
\setval{8b.gdpo-noaux.minus.gdpo.judge2.longbench-write-abs}{9.7}
\setval{8b.gdpo-noaux.minus.gdpo.judge2.writing}{$-$10.1}
\setval{8b.gdpo-noaux.minus.gdpo.judge2.writing-abs}{10.1}
\setval{8b.gdpo-noaux.minus.gdpo.judge2.writingbench}{$-$8.3}
\setval{8b.gdpo-noaux.minus.gdpo.judge2.writingbench-abs}{8.3}
\setval{8b.gdpo-noaux.minus.gdpo.livecodebench}{+1.1}
\setval{8b.gdpo-noaux.minus.gdpo.livecodebench-abs}{1.1}
\setval{8b.gdpo-noaux.minus.gdpo.longbench-write}{$-$3.6}
\setval{8b.gdpo-noaux.minus.gdpo.longbench-write-abs}{3.6}
\setval{8b.gdpo-noaux.minus.gdpo.math}{+8.9}
\setval{8b.gdpo-noaux.minus.gdpo.math-abs}{8.9}
\setval{8b.gdpo-noaux.minus.gdpo.math500}{+7.2}
\setval{8b.gdpo-noaux.minus.gdpo.math500-abs}{7.2}
\setval{8b.gdpo-noaux.minus.gdpo.mbpp-plus}{+0.3}
\setval{8b.gdpo-noaux.minus.gdpo.mbpp-plus-abs}{0.3}
\setval{8b.gdpo-noaux.minus.gdpo.minerva}{+4.4}
\setval{8b.gdpo-noaux.minus.gdpo.minerva-abs}{4.4}
\setval{8b.gdpo-noaux.minus.gdpo.olympiad-bench}{+10.6}
\setval{8b.gdpo-noaux.minus.gdpo.olympiad-bench-abs}{10.6}
\setval{8b.gdpo-noaux.minus.gdpo.writing}{$-$4.6}
\setval{8b.gdpo-noaux.minus.gdpo.writing-abs}{4.6}
\setval{8b.gdpo-noaux.minus.gdpo.writingbench}{$-$1.1}
\setval{8b.gdpo-noaux.minus.gdpo.writingbench-abs}{1.1}
\setval{8b.gdpo-noaux.minus.geoma.aime24}{+13.9}
\setval{8b.gdpo-noaux.minus.geoma.aime24-abs}{13.9}
\setval{8b.gdpo-noaux.minus.geoma.aime25}{+11.5}
\setval{8b.gdpo-noaux.minus.geoma.aime25-abs}{11.5}
\setval{8b.gdpo-noaux.minus.geoma.amc23}{+13.6}
\setval{8b.gdpo-noaux.minus.geoma.amc23-abs}{13.6}
\setval{8b.gdpo-noaux.minus.geoma.avg}{+3.8}
\setval{8b.gdpo-noaux.minus.geoma.avg-abs}{3.8}
\setval{8b.gdpo-noaux.minus.geoma.code}{+1.4}
\setval{8b.gdpo-noaux.minus.geoma.code-abs}{1.4}
\setval{8b.gdpo-noaux.minus.geoma.creative-writing-v3}{$-$6.5}
\setval{8b.gdpo-noaux.minus.geoma.creative-writing-v3-abs}{6.5}
\setval{8b.gdpo-noaux.minus.geoma.guidebench}{$-$3.0}
\setval{8b.gdpo-noaux.minus.geoma.guidebench-abs}{3.0}
\setval{8b.gdpo-noaux.minus.geoma.humaneval-plus}{+3.2}
\setval{8b.gdpo-noaux.minus.geoma.humaneval-plus-abs}{3.2}
\setval{8b.gdpo-noaux.minus.geoma.if}{$-$1.3}
\setval{8b.gdpo-noaux.minus.geoma.if-abs}{1.3}
\setval{8b.gdpo-noaux.minus.geoma.ifbench}{$-$0.3}
\setval{8b.gdpo-noaux.minus.geoma.ifbench-abs}{0.3}
\setval{8b.gdpo-noaux.minus.geoma.ifeval}{$-$0.7}
\setval{8b.gdpo-noaux.minus.geoma.ifeval-abs}{0.7}
\setval{8b.gdpo-noaux.minus.geoma.judge2.avg}{+2.7}
\setval{8b.gdpo-noaux.minus.geoma.judge2.avg-abs}{2.7}
\setval{8b.gdpo-noaux.minus.geoma.judge2.creative-writing-v3}{$-$8.7}
\setval{8b.gdpo-noaux.minus.geoma.judge2.creative-writing-v3-abs}{8.7}
\setval{8b.gdpo-noaux.minus.geoma.judge2.longbench-write}{$-$10.4}
\setval{8b.gdpo-noaux.minus.geoma.judge2.longbench-write-abs}{10.4}
\setval{8b.gdpo-noaux.minus.geoma.judge2.writing}{$-$9.1}
\setval{8b.gdpo-noaux.minus.geoma.judge2.writing-abs}{9.1}
\setval{8b.gdpo-noaux.minus.geoma.judge2.writingbench}{$-$8.3}
\setval{8b.gdpo-noaux.minus.geoma.judge2.writingbench-abs}{8.3}
\setval{8b.gdpo-noaux.minus.geoma.livecodebench}{+1.7}
\setval{8b.gdpo-noaux.minus.geoma.livecodebench-abs}{1.7}
\setval{8b.gdpo-noaux.minus.geoma.longbench-write}{$-$3.0}
\setval{8b.gdpo-noaux.minus.geoma.longbench-write-abs}{3.0}
\setval{8b.gdpo-noaux.minus.geoma.math}{+11.3}
\setval{8b.gdpo-noaux.minus.geoma.math-abs}{11.3}
\setval{8b.gdpo-noaux.minus.geoma.math500}{+10.6}
\setval{8b.gdpo-noaux.minus.geoma.math500-abs}{10.6}
\setval{8b.gdpo-noaux.minus.geoma.mbpp-plus}{$-$0.7}
\setval{8b.gdpo-noaux.minus.geoma.mbpp-plus-abs}{0.7}
\setval{8b.gdpo-noaux.minus.geoma.minerva}{+4.2}
\setval{8b.gdpo-noaux.minus.geoma.minerva-abs}{4.2}
\setval{8b.gdpo-noaux.minus.geoma.olympiad-bench}{+13.9}
\setval{8b.gdpo-noaux.minus.geoma.olympiad-bench-abs}{13.9}
\setval{8b.gdpo-noaux.minus.geoma.writing}{$-$3.5}
\setval{8b.gdpo-noaux.minus.geoma.writing-abs}{3.5}
\setval{8b.gdpo-noaux.minus.geoma.writingbench}{$-$0.9}
\setval{8b.gdpo-noaux.minus.geoma.writingbench-abs}{0.9}
\setval{8b.gdpo-noaux.minus.grpo-nodiv.aime24}{+7.3}
\setval{8b.gdpo-noaux.minus.grpo-nodiv.aime24-abs}{7.3}
\setval{8b.gdpo-noaux.minus.grpo-nodiv.aime25}{+9.2}
\setval{8b.gdpo-noaux.minus.grpo-nodiv.aime25-abs}{9.2}
\setval{8b.gdpo-noaux.minus.grpo-nodiv.amc23}{+13.5}
\setval{8b.gdpo-noaux.minus.grpo-nodiv.amc23-abs}{13.5}
\setval{8b.gdpo-noaux.minus.grpo-nodiv.avg}{+2.4}
\setval{8b.gdpo-noaux.minus.grpo-nodiv.avg-abs}{2.4}
\setval{8b.gdpo-noaux.minus.grpo-nodiv.code}{+0.3}
\setval{8b.gdpo-noaux.minus.grpo-nodiv.code-abs}{0.3}
\setval{8b.gdpo-noaux.minus.grpo-nodiv.creative-writing-v3}{$-$2.8}
\setval{8b.gdpo-noaux.minus.grpo-nodiv.creative-writing-v3-abs}{2.8}
\setval{8b.gdpo-noaux.minus.grpo-nodiv.guidebench}{$-$3.5}
\setval{8b.gdpo-noaux.minus.grpo-nodiv.guidebench-abs}{3.5}
\setval{8b.gdpo-noaux.minus.grpo-nodiv.humaneval-plus}{$-$0.4}
\setval{8b.gdpo-noaux.minus.grpo-nodiv.humaneval-plus-abs}{0.4}
\setval{8b.gdpo-noaux.minus.grpo-nodiv.if}{$-$3.0}
\setval{8b.gdpo-noaux.minus.grpo-nodiv.if-abs}{3.0}
\setval{8b.gdpo-noaux.minus.grpo-nodiv.ifbench}{$-$2.7}
\setval{8b.gdpo-noaux.minus.grpo-nodiv.ifbench-abs}{2.7}
\setval{8b.gdpo-noaux.minus.grpo-nodiv.ifeval}{$-$2.9}
\setval{8b.gdpo-noaux.minus.grpo-nodiv.ifeval-abs}{2.9}
\setval{8b.gdpo-noaux.minus.grpo-nodiv.judge2.avg}{+1.0}
\setval{8b.gdpo-noaux.minus.grpo-nodiv.judge2.avg-abs}{1.0}
\setval{8b.gdpo-noaux.minus.grpo-nodiv.judge2.creative-writing-v3}{$-$8.6}
\setval{8b.gdpo-noaux.minus.grpo-nodiv.judge2.creative-writing-v3-abs}{8.6}
\setval{8b.gdpo-noaux.minus.grpo-nodiv.judge2.longbench-write}{$-$12.1}
\setval{8b.gdpo-noaux.minus.grpo-nodiv.judge2.longbench-write-abs}{12.1}
\setval{8b.gdpo-noaux.minus.grpo-nodiv.judge2.writing}{$-$8.2}
\setval{8b.gdpo-noaux.minus.grpo-nodiv.judge2.writing-abs}{8.2}
\setval{8b.gdpo-noaux.minus.grpo-nodiv.judge2.writingbench}{$-$3.8}
\setval{8b.gdpo-noaux.minus.grpo-nodiv.judge2.writingbench-abs}{3.8}
\setval{8b.gdpo-noaux.minus.grpo-nodiv.livecodebench}{+2.4}
\setval{8b.gdpo-noaux.minus.grpo-nodiv.livecodebench-abs}{2.4}
\setval{8b.gdpo-noaux.minus.grpo-nodiv.longbench-write}{$-$5.1}
\setval{8b.gdpo-noaux.minus.grpo-nodiv.longbench-write-abs}{5.1}
\setval{8b.gdpo-noaux.minus.grpo-nodiv.math}{+7.9}
\setval{8b.gdpo-noaux.minus.grpo-nodiv.math-abs}{7.9}
\setval{8b.gdpo-noaux.minus.grpo-nodiv.math500}{+6.1}
\setval{8b.gdpo-noaux.minus.grpo-nodiv.math500-abs}{6.1}
\setval{8b.gdpo-noaux.minus.grpo-nodiv.mbpp-plus}{$-$1.2}
\setval{8b.gdpo-noaux.minus.grpo-nodiv.mbpp-plus-abs}{1.2}
\setval{8b.gdpo-noaux.minus.grpo-nodiv.minerva}{+3.2}
\setval{8b.gdpo-noaux.minus.grpo-nodiv.minerva-abs}{3.2}
\setval{8b.gdpo-noaux.minus.grpo-nodiv.olympiad-bench}{+8.4}
\setval{8b.gdpo-noaux.minus.grpo-nodiv.olympiad-bench-abs}{8.4}
\setval{8b.gdpo-noaux.minus.grpo-nodiv.writing}{$-$1.1}
\setval{8b.gdpo-noaux.minus.grpo-nodiv.writing-abs}{1.1}
\setval{8b.gdpo-noaux.minus.grpo-nodiv.writingbench}{+4.7}
\setval{8b.gdpo-noaux.minus.grpo-nodiv.writingbench-abs}{4.7}
\setval{8b.gdpo-noaux.minus.grpo.aime24}{+12.1}
\setval{8b.gdpo-noaux.minus.grpo.aime24-abs}{12.1}
\setval{8b.gdpo-noaux.minus.grpo.aime25}{+8.4}
\setval{8b.gdpo-noaux.minus.grpo.aime25-abs}{8.4}
\setval{8b.gdpo-noaux.minus.grpo.amc23}{+12.7}
\setval{8b.gdpo-noaux.minus.grpo.amc23-abs}{12.7}
\setval{8b.gdpo-noaux.minus.grpo.avg}{+2.6}
\setval{8b.gdpo-noaux.minus.grpo.avg-abs}{2.6}
\setval{8b.gdpo-noaux.minus.grpo.code}{+1.1}
\setval{8b.gdpo-noaux.minus.grpo.code-abs}{1.1}
\setval{8b.gdpo-noaux.minus.grpo.creative-writing-v3}{$-$8.6}
\setval{8b.gdpo-noaux.minus.grpo.creative-writing-v3-abs}{8.6}
\setval{8b.gdpo-noaux.minus.grpo.guidebench}{$-$0.5}
\setval{8b.gdpo-noaux.minus.grpo.guidebench-abs}{0.5}
\setval{8b.gdpo-noaux.minus.grpo.humaneval-plus}{+2.3}
\setval{8b.gdpo-noaux.minus.grpo.humaneval-plus-abs}{2.3}
\setval{8b.gdpo-noaux.minus.grpo.if}{$-$3.0}
\setval{8b.gdpo-noaux.minus.grpo.if-abs}{3.0}
\setval{8b.gdpo-noaux.minus.grpo.ifbench}{$-$8.3}
\setval{8b.gdpo-noaux.minus.grpo.ifbench-abs}{8.3}
\setval{8b.gdpo-noaux.minus.grpo.ifeval}{$-$0.2}
\setval{8b.gdpo-noaux.minus.grpo.ifeval-abs}{0.2}
\setval{8b.gdpo-noaux.minus.grpo.judge2.avg}{+1.9}
\setval{8b.gdpo-noaux.minus.grpo.judge2.avg-abs}{1.9}
\setval{8b.gdpo-noaux.minus.grpo.judge2.creative-writing-v3}{$-$10.4}
\setval{8b.gdpo-noaux.minus.grpo.judge2.creative-writing-v3-abs}{10.4}
\setval{8b.gdpo-noaux.minus.grpo.judge2.longbench-write}{$-$7.9}
\setval{8b.gdpo-noaux.minus.grpo.judge2.longbench-write-abs}{7.9}
\setval{8b.gdpo-noaux.minus.grpo.judge2.writing}{$-$8.4}
\setval{8b.gdpo-noaux.minus.grpo.judge2.writing-abs}{8.4}
\setval{8b.gdpo-noaux.minus.grpo.judge2.writingbench}{$-$6.9}
\setval{8b.gdpo-noaux.minus.grpo.judge2.writingbench-abs}{6.9}
\setval{8b.gdpo-noaux.minus.grpo.livecodebench}{+0.9}
\setval{8b.gdpo-noaux.minus.grpo.livecodebench-abs}{0.9}
\setval{8b.gdpo-noaux.minus.grpo.longbench-write}{$-$3.8}
\setval{8b.gdpo-noaux.minus.grpo.longbench-write-abs}{3.8}
\setval{8b.gdpo-noaux.minus.grpo.math}{+9.8}
\setval{8b.gdpo-noaux.minus.grpo.math-abs}{9.8}
\setval{8b.gdpo-noaux.minus.grpo.math500}{+9.6}
\setval{8b.gdpo-noaux.minus.grpo.math500-abs}{9.6}
\setval{8b.gdpo-noaux.minus.grpo.mbpp-plus}{+0.2}
\setval{8b.gdpo-noaux.minus.grpo.mbpp-plus-abs}{0.2}
\setval{8b.gdpo-noaux.minus.grpo.minerva}{+4.1}
\setval{8b.gdpo-noaux.minus.grpo.minerva-abs}{4.1}
\setval{8b.gdpo-noaux.minus.grpo.olympiad-bench}{+12.0}
\setval{8b.gdpo-noaux.minus.grpo.olympiad-bench-abs}{12.0}
\setval{8b.gdpo-noaux.minus.grpo.writing}{$-$4.6}
\setval{8b.gdpo-noaux.minus.grpo.writing-abs}{4.6}
\setval{8b.gdpo-noaux.minus.grpo.writingbench}{$-$1.5}
\setval{8b.gdpo-noaux.minus.grpo.writingbench-abs}{1.5}
\setval{8b.gdpo-noaux.minus.mt-grpo.aime24}{+13.4}
\setval{8b.gdpo-noaux.minus.mt-grpo.aime24-abs}{13.4}
\setval{8b.gdpo-noaux.minus.mt-grpo.aime25}{+8.2}
\setval{8b.gdpo-noaux.minus.mt-grpo.aime25-abs}{8.2}
\setval{8b.gdpo-noaux.minus.mt-grpo.amc23}{+11.5}
\setval{8b.gdpo-noaux.minus.mt-grpo.amc23-abs}{11.5}
\setval{8b.gdpo-noaux.minus.mt-grpo.avg}{+3.3}
\setval{8b.gdpo-noaux.minus.mt-grpo.avg-abs}{3.3}
\setval{8b.gdpo-noaux.minus.mt-grpo.code}{+0.1}
\setval{8b.gdpo-noaux.minus.mt-grpo.code-abs}{0.1}
\setval{8b.gdpo-noaux.minus.mt-grpo.creative-writing-v3}{$-$6.4}
\setval{8b.gdpo-noaux.minus.mt-grpo.creative-writing-v3-abs}{6.4}
\setval{8b.gdpo-noaux.minus.mt-grpo.guidebench}{$-$1.7}
\setval{8b.gdpo-noaux.minus.mt-grpo.guidebench-abs}{1.7}
\setval{8b.gdpo-noaux.minus.mt-grpo.humaneval-plus}{+0.9}
\setval{8b.gdpo-noaux.minus.mt-grpo.humaneval-plus-abs}{0.9}
\setval{8b.gdpo-noaux.minus.mt-grpo.if}{$-$1.9}
\setval{8b.gdpo-noaux.minus.mt-grpo.if-abs}{1.9}
\setval{8b.gdpo-noaux.minus.mt-grpo.ifbench}{$-$3.7}
\setval{8b.gdpo-noaux.minus.mt-grpo.ifbench-abs}{3.7}
\setval{8b.gdpo-noaux.minus.mt-grpo.ifeval}{$-$0.4}
\setval{8b.gdpo-noaux.minus.mt-grpo.ifeval-abs}{0.4}
\setval{8b.gdpo-noaux.minus.mt-grpo.judge2.avg}{+1.9}
\setval{8b.gdpo-noaux.minus.mt-grpo.judge2.avg-abs}{1.9}
\setval{8b.gdpo-noaux.minus.mt-grpo.judge2.creative-writing-v3}{$-$12.7}
\setval{8b.gdpo-noaux.minus.mt-grpo.judge2.creative-writing-v3-abs}{12.7}
\setval{8b.gdpo-noaux.minus.mt-grpo.judge2.longbench-write}{$-$8.0}
\setval{8b.gdpo-noaux.minus.mt-grpo.judge2.longbench-write-abs}{8.0}
\setval{8b.gdpo-noaux.minus.mt-grpo.judge2.writing}{$-$8.7}
\setval{8b.gdpo-noaux.minus.mt-grpo.judge2.writing-abs}{8.7}
\setval{8b.gdpo-noaux.minus.mt-grpo.judge2.writingbench}{$-$5.5}
\setval{8b.gdpo-noaux.minus.mt-grpo.judge2.writingbench-abs}{5.5}
\setval{8b.gdpo-noaux.minus.mt-grpo.livecodebench}{$-$0.3}
\setval{8b.gdpo-noaux.minus.mt-grpo.livecodebench-abs}{0.3}
\setval{8b.gdpo-noaux.minus.mt-grpo.longbench-write}{$-$1.5}
\setval{8b.gdpo-noaux.minus.mt-grpo.longbench-write-abs}{1.5}
\setval{8b.gdpo-noaux.minus.mt-grpo.math}{+10.0}
\setval{8b.gdpo-noaux.minus.mt-grpo.math-abs}{10.0}
\setval{8b.gdpo-noaux.minus.mt-grpo.math500}{+9.7}
\setval{8b.gdpo-noaux.minus.mt-grpo.math500-abs}{9.7}
\setval{8b.gdpo-noaux.minus.mt-grpo.mbpp-plus}{$-$0.4}
\setval{8b.gdpo-noaux.minus.mt-grpo.mbpp-plus-abs}{0.4}
\setval{8b.gdpo-noaux.minus.mt-grpo.minerva}{+2.9}
\setval{8b.gdpo-noaux.minus.mt-grpo.minerva-abs}{2.9}
\setval{8b.gdpo-noaux.minus.mt-grpo.olympiad-bench}{+14.2}
\setval{8b.gdpo-noaux.minus.mt-grpo.olympiad-bench-abs}{14.2}
\setval{8b.gdpo-noaux.minus.mt-grpo.writing}{$-$1.5}
\setval{8b.gdpo-noaux.minus.mt-grpo.writing-abs}{1.5}
\setval{8b.gdpo-noaux.minus.mt-grpo.writingbench}{+3.3}
\setval{8b.gdpo-noaux.minus.mt-grpo.writingbench-abs}{3.3}
\setval{8b.gdpo-noaux.minus.papo.aime24}{+15.7}
\setval{8b.gdpo-noaux.minus.papo.aime24-abs}{15.7}
\setval{8b.gdpo-noaux.minus.papo.aime25}{+13.9}
\setval{8b.gdpo-noaux.minus.papo.aime25-abs}{13.9}
\setval{8b.gdpo-noaux.minus.papo.amc23}{+15.1}
\setval{8b.gdpo-noaux.minus.papo.amc23-abs}{15.1}
\setval{8b.gdpo-noaux.minus.papo.avg}{+4.7}
\setval{8b.gdpo-noaux.minus.papo.avg-abs}{4.7}
\setval{8b.gdpo-noaux.minus.papo.code}{+1.9}
\setval{8b.gdpo-noaux.minus.papo.code-abs}{1.9}
\setval{8b.gdpo-noaux.minus.papo.creative-writing-v3}{$-$7.3}
\setval{8b.gdpo-noaux.minus.papo.creative-writing-v3-abs}{7.3}
\setval{8b.gdpo-noaux.minus.papo.guidebench}{$-$3.0}
\setval{8b.gdpo-noaux.minus.papo.guidebench-abs}{3.0}
\setval{8b.gdpo-noaux.minus.papo.humaneval-plus}{+3.5}
\setval{8b.gdpo-noaux.minus.papo.humaneval-plus-abs}{3.5}
\setval{8b.gdpo-noaux.minus.papo.if}{$-$2.1}
\setval{8b.gdpo-noaux.minus.papo.if-abs}{2.1}
\setval{8b.gdpo-noaux.minus.papo.ifbench}{$-$3.7}
\setval{8b.gdpo-noaux.minus.papo.ifbench-abs}{3.7}
\setval{8b.gdpo-noaux.minus.papo.ifeval}{+0.2}
\setval{8b.gdpo-noaux.minus.papo.ifeval-abs}{0.2}
\setval{8b.gdpo-noaux.minus.papo.judge2.avg}{+3.4}
\setval{8b.gdpo-noaux.minus.papo.judge2.avg-abs}{3.4}
\setval{8b.gdpo-noaux.minus.papo.judge2.creative-writing-v3}{$-$13.0}
\setval{8b.gdpo-noaux.minus.papo.judge2.creative-writing-v3-abs}{13.0}
\setval{8b.gdpo-noaux.minus.papo.judge2.longbench-write}{$-$5.1}
\setval{8b.gdpo-noaux.minus.papo.judge2.longbench-write-abs}{5.1}
\setval{8b.gdpo-noaux.minus.papo.judge2.writing}{$-$8.1}
\setval{8b.gdpo-noaux.minus.papo.judge2.writing-abs}{8.1}
\setval{8b.gdpo-noaux.minus.papo.judge2.writingbench}{$-$6.3}
\setval{8b.gdpo-noaux.minus.papo.judge2.writingbench-abs}{6.3}
\setval{8b.gdpo-noaux.minus.papo.livecodebench}{+0.7}
\setval{8b.gdpo-noaux.minus.papo.livecodebench-abs}{0.7}
\setval{8b.gdpo-noaux.minus.papo.longbench-write}{+0.5}
\setval{8b.gdpo-noaux.minus.papo.longbench-write-abs}{0.5}
\setval{8b.gdpo-noaux.minus.papo.math}{+12.7}
\setval{8b.gdpo-noaux.minus.papo.math-abs}{12.7}
\setval{8b.gdpo-noaux.minus.papo.math500}{+11.8}
\setval{8b.gdpo-noaux.minus.papo.math500-abs}{11.8}
\setval{8b.gdpo-noaux.minus.papo.mbpp-plus}{+1.5}
\setval{8b.gdpo-noaux.minus.papo.mbpp-plus-abs}{1.5}
\setval{8b.gdpo-noaux.minus.papo.minerva}{+4.6}
\setval{8b.gdpo-noaux.minus.papo.minerva-abs}{4.6}
\setval{8b.gdpo-noaux.minus.papo.olympiad-bench}{+15.1}
\setval{8b.gdpo-noaux.minus.papo.olympiad-bench-abs}{15.1}
\setval{8b.gdpo-noaux.minus.papo.writing}{$-$1.7}
\setval{8b.gdpo-noaux.minus.papo.writing-abs}{1.7}
\setval{8b.gdpo-noaux.minus.papo.writingbench}{+1.8}
\setval{8b.gdpo-noaux.minus.papo.writingbench-abs}{1.8}
\setval{8b.gdpo-noaux.minus.saw.aime24}{+12.6}
\setval{8b.gdpo-noaux.minus.saw.aime24-abs}{12.6}
\setval{8b.gdpo-noaux.minus.saw.aime25}{+8.4}
\setval{8b.gdpo-noaux.minus.saw.aime25-abs}{8.4}
\setval{8b.gdpo-noaux.minus.saw.amc23}{+14.8}
\setval{8b.gdpo-noaux.minus.saw.amc23-abs}{14.8}
\setval{8b.gdpo-noaux.minus.saw.avg}{+2.5}
\setval{8b.gdpo-noaux.minus.saw.avg-abs}{2.5}
\setval{8b.gdpo-noaux.minus.saw.code}{+0.4}
\setval{8b.gdpo-noaux.minus.saw.code-abs}{0.4}
\setval{8b.gdpo-noaux.minus.saw.creative-writing-v3}{$-$5.8}
\setval{8b.gdpo-noaux.minus.saw.creative-writing-v3-abs}{5.8}
\setval{8b.gdpo-noaux.minus.saw.guidebench}{$-$2.9}
\setval{8b.gdpo-noaux.minus.saw.guidebench-abs}{2.9}
\setval{8b.gdpo-noaux.minus.saw.humaneval-plus}{+0.1}
\setval{8b.gdpo-noaux.minus.saw.humaneval-plus-abs}{0.1}
\setval{8b.gdpo-noaux.minus.saw.if}{$-$3.4}
\setval{8b.gdpo-noaux.minus.saw.if-abs}{3.4}
\setval{8b.gdpo-noaux.minus.saw.ifbench}{$-$6.3}
\setval{8b.gdpo-noaux.minus.saw.ifbench-abs}{6.3}
\setval{8b.gdpo-noaux.minus.saw.ifeval}{$-$1.1}
\setval{8b.gdpo-noaux.minus.saw.ifeval-abs}{1.1}
\setval{8b.gdpo-noaux.minus.saw.judge2.avg}{+1.6}
\setval{8b.gdpo-noaux.minus.saw.judge2.avg-abs}{1.6}
\setval{8b.gdpo-noaux.minus.saw.judge2.creative-writing-v3}{$-$7.7}
\setval{8b.gdpo-noaux.minus.saw.judge2.creative-writing-v3-abs}{7.7}
\setval{8b.gdpo-noaux.minus.saw.judge2.longbench-write}{$-$10.3}
\setval{8b.gdpo-noaux.minus.saw.judge2.longbench-write-abs}{10.3}
\setval{8b.gdpo-noaux.minus.saw.judge2.writing}{$-$8.3}
\setval{8b.gdpo-noaux.minus.saw.judge2.writing-abs}{8.3}
\setval{8b.gdpo-noaux.minus.saw.judge2.writingbench}{$-$6.9}
\setval{8b.gdpo-noaux.minus.saw.judge2.writingbench-abs}{6.9}
\setval{8b.gdpo-noaux.minus.saw.livecodebench}{+0.7}
\setval{8b.gdpo-noaux.minus.saw.livecodebench-abs}{0.7}
\setval{8b.gdpo-noaux.minus.saw.longbench-write}{$-$5.1}
\setval{8b.gdpo-noaux.minus.saw.longbench-write-abs}{5.1}
\setval{8b.gdpo-noaux.minus.saw.math}{+9.7}
\setval{8b.gdpo-noaux.minus.saw.math-abs}{9.7}
\setval{8b.gdpo-noaux.minus.saw.math500}{+8.3}
\setval{8b.gdpo-noaux.minus.saw.math500-abs}{8.3}
\setval{8b.gdpo-noaux.minus.saw.mbpp-plus}{+0.5}
\setval{8b.gdpo-noaux.minus.saw.mbpp-plus-abs}{0.5}
\setval{8b.gdpo-noaux.minus.saw.minerva}{+2.2}
\setval{8b.gdpo-noaux.minus.saw.minerva-abs}{2.2}
\setval{8b.gdpo-noaux.minus.saw.olympiad-bench}{+11.8}
\setval{8b.gdpo-noaux.minus.saw.olympiad-bench-abs}{11.8}
\setval{8b.gdpo-noaux.minus.saw.writing}{$-$3.8}
\setval{8b.gdpo-noaux.minus.saw.writing-abs}{3.8}
\setval{8b.gdpo-noaux.minus.saw.writingbench}{$-$0.4}
\setval{8b.gdpo-noaux.minus.saw.writingbench-abs}{0.4}
\setval{8b.gdpo-noaux.passing-at-ceiling.code.early}{0.70}
\setval{8b.gdpo-noaux.passing-at-ceiling.code.late}{0.76}
\setval{8b.gdpo-noaux.passing-at-ceiling.code.run}{0.75}
\setval{8b.gdpo-noaux.passing-at-ceiling.if.early}{0.32}
\setval{8b.gdpo-noaux.passing-at-ceiling.if.late}{0.52}
\setval{8b.gdpo-noaux.passing-at-ceiling.if.run}{0.44}
\setval{8b.gdpo-noaux.passing-at-ceiling.math.early}{1.00}
\setval{8b.gdpo-noaux.passing-at-ceiling.math.late}{1.00}
\setval{8b.gdpo-noaux.passing-at-ceiling.math.run}{1.00}
\setval{8b.gdpo-noaux.passing-at-ceiling.writing.early}{0.00}
\setval{8b.gdpo-noaux.passing-at-ceiling.writing.late}{0.03}
\setval{8b.gdpo-noaux.passing-at-ceiling.writing.run}{0.02}
\setval{8b.gdpo-noaux.primary-cosine.code.early}{1.00}
\setval{8b.gdpo-noaux.primary-cosine.code.late}{1.00}
\setval{8b.gdpo-noaux.primary-cosine.code.run}{1.00}
\setval{8b.gdpo-noaux.primary-cosine.if.early}{0.74}
\setval{8b.gdpo-noaux.primary-cosine.if.late}{0.85}
\setval{8b.gdpo-noaux.primary-cosine.if.run}{0.84}
\setval{8b.gdpo-noaux.primary-cosine.math.early}{1.00}
\setval{8b.gdpo-noaux.primary-cosine.math.late}{1.00}
\setval{8b.gdpo-noaux.primary-cosine.math.run}{1.00}
\setval{8b.gdpo-noaux.primary-cosine.writing.early}{1.00}
\setval{8b.gdpo-noaux.primary-cosine.writing.late}{1.00}
\setval{8b.gdpo-noaux.primary-cosine.writing.run}{1.00}
\setval{8b.gdpo-noaux.primary-level.code.early}{0.54}
\setval{8b.gdpo-noaux.primary-level.code.late}{0.76}
\setval{8b.gdpo-noaux.primary-level.code.run}{0.71}
\setval{8b.gdpo-noaux.primary-level.if.early}{0.35}
\setval{8b.gdpo-noaux.primary-level.if.late}{0.70}
\setval{8b.gdpo-noaux.primary-level.if.run}{0.59}
\setval{8b.gdpo-noaux.primary-level.math.early}{0.49}
\setval{8b.gdpo-noaux.primary-level.math.late}{0.74}
\setval{8b.gdpo-noaux.primary-level.math.run}{0.69}
\setval{8b.gdpo-noaux.primary-level.writing.early}{0.53}
\setval{8b.gdpo-noaux.primary-level.writing.late}{0.83}
\setval{8b.gdpo-noaux.primary-level.writing.run}{0.77}
\setval{8b.gdpo-noaux.solved.code.early}{0.11}
\setval{8b.gdpo-noaux.solved.code.late}{0.38}
\setval{8b.gdpo-noaux.solved.code.run}{0.31}
\setval{8b.gdpo-noaux.solved.if.early}{0.03}
\setval{8b.gdpo-noaux.solved.if.late}{0.20}
\setval{8b.gdpo-noaux.solved.if.run}{0.13}
\setval{8b.gdpo-noaux.solved.math.early}{0.16}
\setval{8b.gdpo-noaux.solved.math.late}{0.40}
\setval{8b.gdpo-noaux.solved.math.run}{0.34}
\setval{8b.gdpo-noaux.solved.writing.early}{0.00}
\setval{8b.gdpo-noaux.solved.writing.late}{0.01}
\setval{8b.gdpo-noaux.solved.writing.run}{0.00}
\setval{8b.gdpo-noaux.update-share.code.early}{0.21}
\setval{8b.gdpo-noaux.update-share.code.late}{0.20}
\setval{8b.gdpo-noaux.update-share.code.run}{0.21}
\setval{8b.gdpo-noaux.update-share.if.early}{0.34}
\setval{8b.gdpo-noaux.update-share.if.late}{0.25}
\setval{8b.gdpo-noaux.update-share.if.run}{0.28}
\setval{8b.gdpo-noaux.update-share.math.early}{0.18}
\setval{8b.gdpo-noaux.update-share.math.late}{0.16}
\setval{8b.gdpo-noaux.update-share.math.run}{0.17}
\setval{8b.gdpo-noaux.update-share.writing.early}{0.26}
\setval{8b.gdpo-noaux.update-share.writing.late}{0.39}
\setval{8b.gdpo-noaux.update-share.writing.run}{0.34}
\setval{8b.gdpo-w0.5.aux-change.code}{+0.18}
\setval{8b.gdpo-w0.5.aux-change.if}{+0.18}
\setval{8b.gdpo-w0.5.aux-change.math}{+0.10}
\setval{8b.gdpo-w0.5.aux-change.writing}{+0.12}
\setval{8b.gdpo-w0.5.aux-change.writing-rm}{+0.32}
\setval{8b.gdpo-w0.5.aux-full.code.early}{0.67}
\setval{8b.gdpo-w0.5.aux-full.code.late}{0.94}
\setval{8b.gdpo-w0.5.aux-full.code.run}{0.92}
\setval{8b.gdpo-w0.5.aux-full.if.early}{0.69}
\setval{8b.gdpo-w0.5.aux-full.if.late}{0.92}
\setval{8b.gdpo-w0.5.aux-full.if.run}{0.87}
\setval{8b.gdpo-w0.5.aux-full.math.early}{0.55}
\setval{8b.gdpo-w0.5.aux-full.math.late}{0.83}
\setval{8b.gdpo-w0.5.aux-full.math.run}{0.76}
\setval{8b.gdpo-w0.5.aux-full.writing-rm.early}{0.07}
\setval{8b.gdpo-w0.5.aux-full.writing-rm.late}{0.59}
\setval{8b.gdpo-w0.5.aux-full.writing-rm.run}{0.50}
\setval{8b.gdpo-w0.5.aux-full.writing.early}{0.63}
\setval{8b.gdpo-w0.5.aux-full.writing.late}{0.73}
\setval{8b.gdpo-w0.5.aux-full.writing.run}{0.61}
\setval{8b.gdpo-w0.5.aux-level.code.early}{0.92}
\setval{8b.gdpo-w0.5.aux-level.code.late}{0.99}
\setval{8b.gdpo-w0.5.aux-level.code.run}{0.98}
\setval{8b.gdpo-w0.5.aux-level.if.early}{0.85}
\setval{8b.gdpo-w0.5.aux-level.if.late}{0.96}
\setval{8b.gdpo-w0.5.aux-level.if.run}{0.94}
\setval{8b.gdpo-w0.5.aux-level.math.early}{0.90}
\setval{8b.gdpo-w0.5.aux-level.math.late}{0.94}
\setval{8b.gdpo-w0.5.aux-level.math.run}{0.93}
\setval{8b.gdpo-w0.5.aux-level.writing-rm.early}{0.77}
\setval{8b.gdpo-w0.5.aux-level.writing-rm.late}{0.96}
\setval{8b.gdpo-w0.5.aux-level.writing-rm.run}{0.93}
\setval{8b.gdpo-w0.5.aux-level.writing.early}{0.90}
\setval{8b.gdpo-w0.5.aux-level.writing.late}{0.96}
\setval{8b.gdpo-w0.5.aux-level.writing.run}{0.91}
\setval{8b.gdpo-w0.5.bench.aime24}{19.3}
\setval{8b.gdpo-w0.5.bench.aime25}{15.4}
\setval{8b.gdpo-w0.5.bench.amc23}{63.4}
\setval{8b.gdpo-w0.5.bench.creative-writing-v3}{51.6}
\setval{8b.gdpo-w0.5.bench.guidebench}{84.0}
\setval{8b.gdpo-w0.5.bench.humaneval-plus}{83.8}
\setval{8b.gdpo-w0.5.bench.ifbench}{37.7}
\setval{8b.gdpo-w0.5.bench.ifeval}{87.8}
\setval{8b.gdpo-w0.5.bench.livecodebench}{26.8}
\setval{8b.gdpo-w0.5.bench.longbench-write}{71.2}
\setval{8b.gdpo-w0.5.bench.math500}{84.1}
\setval{8b.gdpo-w0.5.bench.mbpp-plus}{72.8}
\setval{8b.gdpo-w0.5.bench.minerva}{51.2}
\setval{8b.gdpo-w0.5.bench.olympiad-bench}{51.1}
\setval{8b.gdpo-w0.5.bench.writingbench}{75.0}
\setval{8b.gdpo-w0.5.eval.avg}{58.3}
\setval{8b.gdpo-w0.5.eval.code}{61.1}
\setval{8b.gdpo-w0.5.eval.if}{69.8}
\setval{8b.gdpo-w0.5.eval.math}{47.4}
\setval{8b.gdpo-w0.5.eval.writing}{65.9}
\setval{8b.gdpo-w0.5.gate.code.against.early}{0.04}
\setval{8b.gdpo-w0.5.gate.code.against.late}{0.05}
\setval{8b.gdpo-w0.5.gate.code.against.run}{0.03}
\setval{8b.gdpo-w0.5.gate.code.agreement}{+0.007}
\setval{8b.gdpo-w0.5.gate.code.ceiling.early}{0.13}
\setval{8b.gdpo-w0.5.gate.code.ceiling.late}{0.07}
\setval{8b.gdpo-w0.5.gate.code.ceiling.run}{0.06}
\setval{8b.gdpo-w0.5.gate.code.cosine.early}{0.82}
\setval{8b.gdpo-w0.5.gate.code.cosine.late}{0.73}
\setval{8b.gdpo-w0.5.gate.code.cosine.run}{0.79}
\setval{8b.gdpo-w0.5.gate.if.against.early}{0.05}
\setval{8b.gdpo-w0.5.gate.if.against.late}{0.04}
\setval{8b.gdpo-w0.5.gate.if.against.run}{0.04}
\setval{8b.gdpo-w0.5.gate.if.agreement}{+0.015}
\setval{8b.gdpo-w0.5.gate.if.ceiling.early}{0.06}
\setval{8b.gdpo-w0.5.gate.if.ceiling.late}{0.17}
\setval{8b.gdpo-w0.5.gate.if.ceiling.run}{0.13}
\setval{8b.gdpo-w0.5.gate.if.cosine.early}{0.74}
\setval{8b.gdpo-w0.5.gate.if.cosine.late}{0.51}
\setval{8b.gdpo-w0.5.gate.if.cosine.run}{0.59}
\setval{8b.gdpo-w0.5.gate.math.against.early}{0.00}
\setval{8b.gdpo-w0.5.gate.math.against.late}{0.00}
\setval{8b.gdpo-w0.5.gate.math.against.run}{0.00}
\setval{8b.gdpo-w0.5.gate.math.ceiling.early}{0.24}
\setval{8b.gdpo-w0.5.gate.math.ceiling.late}{0.29}
\setval{8b.gdpo-w0.5.gate.math.ceiling.run}{0.30}
\setval{8b.gdpo-w0.5.gate.math.cosine.early}{0.85}
\setval{8b.gdpo-w0.5.gate.math.cosine.late}{0.81}
\setval{8b.gdpo-w0.5.gate.math.cosine.run}{0.81}
\setval{8b.gdpo-w0.5.gate.writing-rm.against.early}{0.19}
\setval{8b.gdpo-w0.5.gate.writing-rm.against.late}{0.55}
\setval{8b.gdpo-w0.5.gate.writing-rm.against.run}{0.50}
\setval{8b.gdpo-w0.5.gate.writing-rm.agreement}{+0.049}
\setval{8b.gdpo-w0.5.gate.writing-rm.ceiling.early}{0.00}
\setval{8b.gdpo-w0.5.gate.writing-rm.ceiling.late}{0.01}
\setval{8b.gdpo-w0.5.gate.writing-rm.ceiling.run}{0.00}
\setval{8b.gdpo-w0.5.gate.writing-rm.cosine.early}{0.69}
\setval{8b.gdpo-w0.5.gate.writing-rm.cosine.late}{0.11}
\setval{8b.gdpo-w0.5.gate.writing-rm.cosine.run}{0.20}
\setval{8b.gdpo-w0.5.gate.writing.against.early}{0.29}
\setval{8b.gdpo-w0.5.gate.writing.against.late}{0.63}
\setval{8b.gdpo-w0.5.gate.writing.against.run}{0.61}
\setval{8b.gdpo-w0.5.gate.writing.agreement}{$-$0.047}
\setval{8b.gdpo-w0.5.gate.writing.ceiling.early}{-0.00}
\setval{8b.gdpo-w0.5.gate.writing.ceiling.late}{0.00}
\setval{8b.gdpo-w0.5.gate.writing.ceiling.run}{0.00}
\setval{8b.gdpo-w0.5.gate.writing.cosine.early}{0.53}
\setval{8b.gdpo-w0.5.gate.writing.cosine.late}{0.02}
\setval{8b.gdpo-w0.5.gate.writing.cosine.run}{0.06}
\setval{8b.gdpo-w0.5.judge2.avg}{55.8}
\setval{8b.gdpo-w0.5.judge2.creative-writing-v3}{40.4}
\setval{8b.gdpo-w0.5.judge2.longbench-write}{63.7}
\setval{8b.gdpo-w0.5.judge2.writing}{53.4}
\setval{8b.gdpo-w0.5.judge2.writingbench}{56.2}
\setval{8b.gdpo-w0.5.last-step}{300}
\setval{8b.gdpo-w0.5.minus.base.aime24}{+7.9}
\setval{8b.gdpo-w0.5.minus.base.aime24-abs}{7.9}
\setval{8b.gdpo-w0.5.minus.base.aime25}{+6.9}
\setval{8b.gdpo-w0.5.minus.base.aime25-abs}{6.9}
\setval{8b.gdpo-w0.5.minus.base.amc23}{+13.2}
\setval{8b.gdpo-w0.5.minus.base.amc23-abs}{13.2}
\setval{8b.gdpo-w0.5.minus.base.avg}{+16.9}
\setval{8b.gdpo-w0.5.minus.base.avg-abs}{16.9}
\setval{8b.gdpo-w0.5.minus.base.code}{+5.1}
\setval{8b.gdpo-w0.5.minus.base.code-abs}{5.1}
\setval{8b.gdpo-w0.5.minus.base.creative-writing-v3}{+24.8}
\setval{8b.gdpo-w0.5.minus.base.creative-writing-v3-abs}{24.8}
\setval{8b.gdpo-w0.5.minus.base.guidebench}{+38.4}
\setval{8b.gdpo-w0.5.minus.base.guidebench-abs}{38.4}
\setval{8b.gdpo-w0.5.minus.base.humaneval-plus}{+3.7}
\setval{8b.gdpo-w0.5.minus.base.humaneval-plus-abs}{3.7}
\setval{8b.gdpo-w0.5.minus.base.if}{+31.2}
\setval{8b.gdpo-w0.5.minus.base.if-abs}{31.2}
\setval{8b.gdpo-w0.5.minus.base.ifbench}{+20.3}
\setval{8b.gdpo-w0.5.minus.base.ifbench-abs}{20.3}
\setval{8b.gdpo-w0.5.minus.base.ifeval}{+34.8}
\setval{8b.gdpo-w0.5.minus.base.ifeval-abs}{34.8}
\setval{8b.gdpo-w0.5.minus.base.judge2.avg}{+14.8}
\setval{8b.gdpo-w0.5.minus.base.judge2.avg-abs}{14.8}
\setval{8b.gdpo-w0.5.minus.base.judge2.creative-writing-v3}{+13.9}
\setval{8b.gdpo-w0.5.minus.base.judge2.creative-writing-v3-abs}{13.9}
\setval{8b.gdpo-w0.5.minus.base.judge2.longbench-write}{+15.6}
\setval{8b.gdpo-w0.5.minus.base.judge2.longbench-write-abs}{15.6}
\setval{8b.gdpo-w0.5.minus.base.judge2.writing}{+17.2}
\setval{8b.gdpo-w0.5.minus.base.judge2.writing-abs}{17.2}
\setval{8b.gdpo-w0.5.minus.base.judge2.writingbench}{+22.2}
\setval{8b.gdpo-w0.5.minus.base.judge2.writingbench-abs}{22.2}
\setval{8b.gdpo-w0.5.minus.base.livecodebench}{+5.9}
\setval{8b.gdpo-w0.5.minus.base.livecodebench-abs}{5.9}
\setval{8b.gdpo-w0.5.minus.base.longbench-write}{+19.8}
\setval{8b.gdpo-w0.5.minus.base.longbench-write-abs}{19.8}
\setval{8b.gdpo-w0.5.minus.base.math}{+10.3}
\setval{8b.gdpo-w0.5.minus.base.math-abs}{10.3}
\setval{8b.gdpo-w0.5.minus.base.math500}{+11.6}
\setval{8b.gdpo-w0.5.minus.base.math500-abs}{11.6}
\setval{8b.gdpo-w0.5.minus.base.mbpp-plus}{+5.6}
\setval{8b.gdpo-w0.5.minus.base.mbpp-plus-abs}{5.6}
\setval{8b.gdpo-w0.5.minus.base.minerva}{+10.3}
\setval{8b.gdpo-w0.5.minus.base.minerva-abs}{10.3}
\setval{8b.gdpo-w0.5.minus.base.olympiad-bench}{+11.7}
\setval{8b.gdpo-w0.5.minus.base.olympiad-bench-abs}{11.7}
\setval{8b.gdpo-w0.5.minus.base.writing}{+27.6}
\setval{8b.gdpo-w0.5.minus.base.writing-abs}{27.6}
\setval{8b.gdpo-w0.5.minus.base.writingbench}{+38.3}
\setval{8b.gdpo-w0.5.minus.base.writingbench-abs}{38.3}
\setval{8b.gdpo-w0.5.minus.concur-free.aime24}{$-$0.4}
\setval{8b.gdpo-w0.5.minus.concur-free.aime24-abs}{0.4}
\setval{8b.gdpo-w0.5.minus.concur-free.aime25}{$-$1.1}
\setval{8b.gdpo-w0.5.minus.concur-free.aime25-abs}{1.1}
\setval{8b.gdpo-w0.5.minus.concur-free.amc23}{$-$1.2}
\setval{8b.gdpo-w0.5.minus.concur-free.amc23-abs}{1.2}
\setval{8b.gdpo-w0.5.minus.concur-free.avg}{$-$1.9}
\setval{8b.gdpo-w0.5.minus.concur-free.avg-abs}{1.9}
\setval{8b.gdpo-w0.5.minus.concur-free.code}{$-$1.4}
\setval{8b.gdpo-w0.5.minus.concur-free.code-abs}{1.4}
\setval{8b.gdpo-w0.5.minus.concur-free.creative-writing-v3}{$-$0.1}
\setval{8b.gdpo-w0.5.minus.concur-free.creative-writing-v3-abs}{0.1}
\setval{8b.gdpo-w0.5.minus.concur-free.guidebench}{+1.1}
\setval{8b.gdpo-w0.5.minus.concur-free.guidebench-abs}{1.1}
\setval{8b.gdpo-w0.5.minus.concur-free.humaneval-plus}{$-$1.5}
\setval{8b.gdpo-w0.5.minus.concur-free.humaneval-plus-abs}{1.5}
\setval{8b.gdpo-w0.5.minus.concur-free.if}{$-$0.5}
\setval{8b.gdpo-w0.5.minus.concur-free.if-abs}{0.5}
\setval{8b.gdpo-w0.5.minus.concur-free.ifbench}{$-$1.0}
\setval{8b.gdpo-w0.5.minus.concur-free.ifbench-abs}{1.0}
\setval{8b.gdpo-w0.5.minus.concur-free.ifeval}{$-$1.4}
\setval{8b.gdpo-w0.5.minus.concur-free.ifeval-abs}{1.4}
\setval{8b.gdpo-w0.5.minus.concur-free.judge2.avg}{$-$1.5}
\setval{8b.gdpo-w0.5.minus.concur-free.judge2.avg-abs}{1.5}
\setval{8b.gdpo-w0.5.minus.concur-free.judge2.creative-writing-v3}{$-$2.1}
\setval{8b.gdpo-w0.5.minus.concur-free.judge2.creative-writing-v3-abs}{2.1}
\setval{8b.gdpo-w0.5.minus.concur-free.judge2.longbench-write}{$-$10.4}
\setval{8b.gdpo-w0.5.minus.concur-free.judge2.longbench-write-abs}{10.4}
\setval{8b.gdpo-w0.5.minus.concur-free.judge2.writing}{$-$5.7}
\setval{8b.gdpo-w0.5.minus.concur-free.judge2.writing-abs}{5.7}
\setval{8b.gdpo-w0.5.minus.concur-free.judge2.writingbench}{$-$4.4}
\setval{8b.gdpo-w0.5.minus.concur-free.judge2.writingbench-abs}{4.4}
\setval{8b.gdpo-w0.5.minus.concur-free.livecodebench}{$-$0.3}
\setval{8b.gdpo-w0.5.minus.concur-free.livecodebench-abs}{0.3}
\setval{8b.gdpo-w0.5.minus.concur-free.longbench-write}{$-$11.8}
\setval{8b.gdpo-w0.5.minus.concur-free.longbench-write-abs}{11.8}
\setval{8b.gdpo-w0.5.minus.concur-free.math}{+0.1}
\setval{8b.gdpo-w0.5.minus.concur-free.math-abs}{0.1}
\setval{8b.gdpo-w0.5.minus.concur-free.math500}{+0.6}
\setval{8b.gdpo-w0.5.minus.concur-free.math500-abs}{0.6}
\setval{8b.gdpo-w0.5.minus.concur-free.mbpp-plus}{$-$2.3}
\setval{8b.gdpo-w0.5.minus.concur-free.mbpp-plus-abs}{2.3}
\setval{8b.gdpo-w0.5.minus.concur-free.minerva}{+1.4}
\setval{8b.gdpo-w0.5.minus.concur-free.minerva-abs}{1.4}
\setval{8b.gdpo-w0.5.minus.concur-free.olympiad-bench}{+1.1}
\setval{8b.gdpo-w0.5.minus.concur-free.olympiad-bench-abs}{1.1}
\setval{8b.gdpo-w0.5.minus.concur-free.writing}{$-$7.9}
\setval{8b.gdpo-w0.5.minus.concur-free.writing-abs}{7.9}
\setval{8b.gdpo-w0.5.minus.concur-free.writingbench}{$-$11.7}
\setval{8b.gdpo-w0.5.minus.concur-free.writingbench-abs}{11.7}
\setval{8b.gdpo-w0.5.minus.concur-tied.aime24}{$-$12.0}
\setval{8b.gdpo-w0.5.minus.concur-tied.aime24-abs}{12.0}
\setval{8b.gdpo-w0.5.minus.concur-tied.aime25}{$-$8.5}
\setval{8b.gdpo-w0.5.minus.concur-tied.aime25-abs}{8.5}
\setval{8b.gdpo-w0.5.minus.concur-tied.amc23}{$-$15.1}
\setval{8b.gdpo-w0.5.minus.concur-tied.amc23-abs}{15.1}
\setval{8b.gdpo-w0.5.minus.concur-tied.avg}{$-$5.7}
\setval{8b.gdpo-w0.5.minus.concur-tied.avg-abs}{5.7}
\setval{8b.gdpo-w0.5.minus.concur-tied.code}{$-$2.0}
\setval{8b.gdpo-w0.5.minus.concur-tied.code-abs}{2.0}
\setval{8b.gdpo-w0.5.minus.concur-tied.creative-writing-v3}{$-$0.9}
\setval{8b.gdpo-w0.5.minus.concur-tied.creative-writing-v3-abs}{0.9}
\setval{8b.gdpo-w0.5.minus.concur-tied.guidebench}{+0.6}
\setval{8b.gdpo-w0.5.minus.concur-tied.guidebench-abs}{0.6}
\setval{8b.gdpo-w0.5.minus.concur-tied.humaneval-plus}{$-$3.6}
\setval{8b.gdpo-w0.5.minus.concur-tied.humaneval-plus-abs}{3.6}
\setval{8b.gdpo-w0.5.minus.concur-tied.if}{$-$0.7}
\setval{8b.gdpo-w0.5.minus.concur-tied.if-abs}{0.7}
\setval{8b.gdpo-w0.5.minus.concur-tied.ifbench}{$-$1.7}
\setval{8b.gdpo-w0.5.minus.concur-tied.ifbench-abs}{1.7}
\setval{8b.gdpo-w0.5.minus.concur-tied.ifeval}{$-$1.1}
\setval{8b.gdpo-w0.5.minus.concur-tied.ifeval-abs}{1.1}
\setval{8b.gdpo-w0.5.minus.concur-tied.judge2.avg}{$-$5.5}
\setval{8b.gdpo-w0.5.minus.concur-tied.judge2.avg-abs}{5.5}
\setval{8b.gdpo-w0.5.minus.concur-tied.judge2.creative-writing-v3}{$-$1.4}
\setval{8b.gdpo-w0.5.minus.concur-tied.judge2.creative-writing-v3-abs}{1.4}
\setval{8b.gdpo-w0.5.minus.concur-tied.judge2.longbench-write}{$-$11.1}
\setval{8b.gdpo-w0.5.minus.concur-tied.judge2.longbench-write-abs}{11.1}
\setval{8b.gdpo-w0.5.minus.concur-tied.judge2.writing}{$-$6.7}
\setval{8b.gdpo-w0.5.minus.concur-tied.judge2.writing-abs}{6.7}
\setval{8b.gdpo-w0.5.minus.concur-tied.judge2.writingbench}{$-$7.5}
\setval{8b.gdpo-w0.5.minus.concur-tied.judge2.writingbench-abs}{7.5}
\setval{8b.gdpo-w0.5.minus.concur-tied.livecodebench}{$-$0.4}
\setval{8b.gdpo-w0.5.minus.concur-tied.livecodebench-abs}{0.4}
\setval{8b.gdpo-w0.5.minus.concur-tied.longbench-write}{$-$10.0}
\setval{8b.gdpo-w0.5.minus.concur-tied.longbench-write-abs}{10.0}
\setval{8b.gdpo-w0.5.minus.concur-tied.math}{$-$9.0}
\setval{8b.gdpo-w0.5.minus.concur-tied.math-abs}{9.0}
\setval{8b.gdpo-w0.5.minus.concur-tied.math500}{$-$5.5}
\setval{8b.gdpo-w0.5.minus.concur-tied.math500-abs}{5.5}
\setval{8b.gdpo-w0.5.minus.concur-tied.mbpp-plus}{$-$2.0}
\setval{8b.gdpo-w0.5.minus.concur-tied.mbpp-plus-abs}{2.0}
\setval{8b.gdpo-w0.5.minus.concur-tied.minerva}{$-$4.2}
\setval{8b.gdpo-w0.5.minus.concur-tied.minerva-abs}{4.2}
\setval{8b.gdpo-w0.5.minus.concur-tied.olympiad-bench}{$-$8.4}
\setval{8b.gdpo-w0.5.minus.concur-tied.olympiad-bench-abs}{8.4}
\setval{8b.gdpo-w0.5.minus.concur-tied.writing}{$-$7.7}
\setval{8b.gdpo-w0.5.minus.concur-tied.writing-abs}{7.7}
\setval{8b.gdpo-w0.5.minus.concur-tied.writingbench}{$-$12.2}
\setval{8b.gdpo-w0.5.minus.concur-tied.writingbench-abs}{12.2}
\setval{8b.gdpo-w0.5.minus.concur.aime24}{$-$9.5}
\setval{8b.gdpo-w0.5.minus.concur.aime24-abs}{9.5}
\setval{8b.gdpo-w0.5.minus.concur.aime25}{$-$7.7}
\setval{8b.gdpo-w0.5.minus.concur.aime25-abs}{7.7}
\setval{8b.gdpo-w0.5.minus.concur.amc23}{$-$15.5}
\setval{8b.gdpo-w0.5.minus.concur.amc23-abs}{15.5}
\setval{8b.gdpo-w0.5.minus.concur.avg}{$-$4.8}
\setval{8b.gdpo-w0.5.minus.concur.avg-abs}{4.8}
\setval{8b.gdpo-w0.5.minus.concur.code}{+0.4}
\setval{8b.gdpo-w0.5.minus.concur.code-abs}{0.4}
\setval{8b.gdpo-w0.5.minus.concur.creative-writing-v3}{+1.6}
\setval{8b.gdpo-w0.5.minus.concur.creative-writing-v3-abs}{1.6}
\setval{8b.gdpo-w0.5.minus.concur.guidebench}{+0.6}
\setval{8b.gdpo-w0.5.minus.concur.guidebench-abs}{0.6}
\setval{8b.gdpo-w0.5.minus.concur.humaneval-plus}{+1.0}
\setval{8b.gdpo-w0.5.minus.concur.humaneval-plus-abs}{1.0}
\setval{8b.gdpo-w0.5.minus.concur.if}{$-$0.3}
\setval{8b.gdpo-w0.5.minus.concur.if-abs}{0.3}
\setval{8b.gdpo-w0.5.minus.concur.ifbench}{$-$1.0}
\setval{8b.gdpo-w0.5.minus.concur.ifbench-abs}{1.0}
\setval{8b.gdpo-w0.5.minus.concur.ifeval}{$-$0.4}
\setval{8b.gdpo-w0.5.minus.concur.ifeval-abs}{0.4}
\setval{8b.gdpo-w0.5.minus.concur.judge2.avg}{$-$4.5}
\setval{8b.gdpo-w0.5.minus.concur.judge2.avg-abs}{4.5}
\setval{8b.gdpo-w0.5.minus.concur.judge2.creative-writing-v3}{+2.6}
\setval{8b.gdpo-w0.5.minus.concur.judge2.creative-writing-v3-abs}{2.6}
\setval{8b.gdpo-w0.5.minus.concur.judge2.longbench-write}{$-$11.8}
\setval{8b.gdpo-w0.5.minus.concur.judge2.longbench-write-abs}{11.8}
\setval{8b.gdpo-w0.5.minus.concur.judge2.writing}{$-$5.6}
\setval{8b.gdpo-w0.5.minus.concur.judge2.writing-abs}{5.6}
\setval{8b.gdpo-w0.5.minus.concur.judge2.writingbench}{$-$7.7}
\setval{8b.gdpo-w0.5.minus.concur.judge2.writingbench-abs}{7.7}
\setval{8b.gdpo-w0.5.minus.concur.livecodebench}{+0.5}
\setval{8b.gdpo-w0.5.minus.concur.livecodebench-abs}{0.5}
\setval{8b.gdpo-w0.5.minus.concur.longbench-write}{$-$10.4}
\setval{8b.gdpo-w0.5.minus.concur.longbench-write-abs}{10.4}
\setval{8b.gdpo-w0.5.minus.concur.math}{$-$8.5}
\setval{8b.gdpo-w0.5.minus.concur.math-abs}{8.5}
\setval{8b.gdpo-w0.5.minus.concur.math500}{$-$6.1}
\setval{8b.gdpo-w0.5.minus.concur.math500-abs}{6.1}
\setval{8b.gdpo-w0.5.minus.concur.mbpp-plus}{$-$0.4}
\setval{8b.gdpo-w0.5.minus.concur.mbpp-plus-abs}{0.4}
\setval{8b.gdpo-w0.5.minus.concur.minerva}{$-$3.0}
\setval{8b.gdpo-w0.5.minus.concur.minerva-abs}{3.0}
\setval{8b.gdpo-w0.5.minus.concur.olympiad-bench}{$-$9.2}
\setval{8b.gdpo-w0.5.minus.concur.olympiad-bench-abs}{9.2}
\setval{8b.gdpo-w0.5.minus.concur.writing}{$-$7.3}
\setval{8b.gdpo-w0.5.minus.concur.writing-abs}{7.3}
\setval{8b.gdpo-w0.5.minus.concur.writingbench}{$-$13.0}
\setval{8b.gdpo-w0.5.minus.concur.writingbench-abs}{13.0}
\setval{8b.gdpo-w0.5.minus.gd2po.aime24}{+3.0}
\setval{8b.gdpo-w0.5.minus.gd2po.aime24-abs}{3.0}
\setval{8b.gdpo-w0.5.minus.gd2po.aime25}{+3.2}
\setval{8b.gdpo-w0.5.minus.gd2po.aime25-abs}{3.2}
\setval{8b.gdpo-w0.5.minus.gd2po.amc23}{+3.8}
\setval{8b.gdpo-w0.5.minus.gd2po.amc23-abs}{3.8}
\setval{8b.gdpo-w0.5.minus.gd2po.avg}{+0.9}
\setval{8b.gdpo-w0.5.minus.gd2po.avg-abs}{0.9}
\setval{8b.gdpo-w0.5.minus.gd2po.code}{$-$0.6}
\setval{8b.gdpo-w0.5.minus.gd2po.code-abs}{0.6}
\setval{8b.gdpo-w0.5.minus.gd2po.creative-writing-v3}{+3.4}
\setval{8b.gdpo-w0.5.minus.gd2po.creative-writing-v3-abs}{3.4}
\setval{8b.gdpo-w0.5.minus.gd2po.guidebench}{+0.3}
\setval{8b.gdpo-w0.5.minus.gd2po.guidebench-abs}{0.3}
\setval{8b.gdpo-w0.5.minus.gd2po.humaneval-plus}{$-$0.2}
\setval{8b.gdpo-w0.5.minus.gd2po.humaneval-plus-abs}{0.2}
\setval{8b.gdpo-w0.5.minus.gd2po.if}{$-$2.2}
\setval{8b.gdpo-w0.5.minus.gd2po.if-abs}{2.2}
\setval{8b.gdpo-w0.5.minus.gd2po.ifbench}{$-$5.3}
\setval{8b.gdpo-w0.5.minus.gd2po.ifbench-abs}{5.3}
\setval{8b.gdpo-w0.5.minus.gd2po.ifeval}{$-$1.6}
\setval{8b.gdpo-w0.5.minus.gd2po.ifeval-abs}{1.6}
\setval{8b.gdpo-w0.5.minus.gd2po.judge2.avg}{+1.0}
\setval{8b.gdpo-w0.5.minus.gd2po.judge2.avg-abs}{1.0}
\setval{8b.gdpo-w0.5.minus.gd2po.judge2.creative-writing-v3}{+3.6}
\setval{8b.gdpo-w0.5.minus.gd2po.judge2.creative-writing-v3-abs}{3.6}
\setval{8b.gdpo-w0.5.minus.gd2po.judge2.longbench-write}{+1.5}
\setval{8b.gdpo-w0.5.minus.gd2po.judge2.longbench-write-abs}{1.5}
\setval{8b.gdpo-w0.5.minus.gd2po.judge2.writing}{+2.0}
\setval{8b.gdpo-w0.5.minus.gd2po.judge2.writing-abs}{2.0}
\setval{8b.gdpo-w0.5.minus.gd2po.judge2.writingbench}{+1.0}
\setval{8b.gdpo-w0.5.minus.gd2po.judge2.writingbench-abs}{1.0}
\setval{8b.gdpo-w0.5.minus.gd2po.livecodebench}{+0.5}
\setval{8b.gdpo-w0.5.minus.gd2po.livecodebench-abs}{0.5}
\setval{8b.gdpo-w0.5.minus.gd2po.longbench-write}{+1.3}
\setval{8b.gdpo-w0.5.minus.gd2po.longbench-write-abs}{1.3}
\setval{8b.gdpo-w0.5.minus.gd2po.math}{+2.8}
\setval{8b.gdpo-w0.5.minus.gd2po.math-abs}{2.8}
\setval{8b.gdpo-w0.5.minus.gd2po.math500}{+2.8}
\setval{8b.gdpo-w0.5.minus.gd2po.math500-abs}{2.8}
\setval{8b.gdpo-w0.5.minus.gd2po.mbpp-plus}{$-$2.2}
\setval{8b.gdpo-w0.5.minus.gd2po.mbpp-plus-abs}{2.2}
\setval{8b.gdpo-w0.5.minus.gd2po.minerva}{+1.3}
\setval{8b.gdpo-w0.5.minus.gd2po.minerva-abs}{1.3}
\setval{8b.gdpo-w0.5.minus.gd2po.olympiad-bench}{+2.6}
\setval{8b.gdpo-w0.5.minus.gd2po.olympiad-bench-abs}{2.6}
\setval{8b.gdpo-w0.5.minus.gd2po.writing}{+2.0}
\setval{8b.gdpo-w0.5.minus.gd2po.writing-abs}{2.0}
\setval{8b.gdpo-w0.5.minus.gd2po.writingbench}{+1.1}
\setval{8b.gdpo-w0.5.minus.gd2po.writingbench-abs}{1.1}
\setval{8b.gdpo-w0.5.minus.gdpo-noaux.aime24}{$-$8.6}
\setval{8b.gdpo-w0.5.minus.gdpo-noaux.aime24-abs}{8.6}
\setval{8b.gdpo-w0.5.minus.gdpo-noaux.aime25}{$-$8.5}
\setval{8b.gdpo-w0.5.minus.gdpo-noaux.aime25-abs}{8.5}
\setval{8b.gdpo-w0.5.minus.gdpo-noaux.amc23}{$-$13.7}
\setval{8b.gdpo-w0.5.minus.gdpo-noaux.amc23-abs}{13.7}
\setval{8b.gdpo-w0.5.minus.gdpo-noaux.avg}{$-$2.2}
\setval{8b.gdpo-w0.5.minus.gdpo-noaux.avg-abs}{2.2}
\setval{8b.gdpo-w0.5.minus.gdpo-noaux.code}{$-$1.6}
\setval{8b.gdpo-w0.5.minus.gdpo-noaux.code-abs}{1.6}
\setval{8b.gdpo-w0.5.minus.gdpo-noaux.creative-writing-v3}{+7.7}
\setval{8b.gdpo-w0.5.minus.gdpo-noaux.creative-writing-v3-abs}{7.7}
\setval{8b.gdpo-w0.5.minus.gdpo-noaux.guidebench}{+2.6}
\setval{8b.gdpo-w0.5.minus.gdpo-noaux.guidebench-abs}{2.6}
\setval{8b.gdpo-w0.5.minus.gdpo-noaux.humaneval-plus}{$-$2.2}
\setval{8b.gdpo-w0.5.minus.gdpo-noaux.humaneval-plus-abs}{2.2}
\setval{8b.gdpo-w0.5.minus.gdpo-noaux.if}{+2.6}
\setval{8b.gdpo-w0.5.minus.gdpo-noaux.if-abs}{2.6}
\setval{8b.gdpo-w0.5.minus.gdpo-noaux.ifbench}{+4.0}
\setval{8b.gdpo-w0.5.minus.gdpo-noaux.ifbench-abs}{4.0}
\setval{8b.gdpo-w0.5.minus.gdpo-noaux.ifeval}{+1.2}
\setval{8b.gdpo-w0.5.minus.gdpo-noaux.ifeval-abs}{1.2}
\setval{8b.gdpo-w0.5.minus.gdpo-noaux.judge2.avg}{$-$1.3}
\setval{8b.gdpo-w0.5.minus.gdpo-noaux.judge2.avg-abs}{1.3}
\setval{8b.gdpo-w0.5.minus.gdpo-noaux.judge2.creative-writing-v3}{+11.5}
\setval{8b.gdpo-w0.5.minus.gdpo-noaux.judge2.creative-writing-v3-abs}{11.5}
\setval{8b.gdpo-w0.5.minus.gdpo-noaux.judge2.longbench-write}{+9.2}
\setval{8b.gdpo-w0.5.minus.gdpo-noaux.judge2.longbench-write-abs}{9.2}
\setval{8b.gdpo-w0.5.minus.gdpo-noaux.judge2.writing}{+8.9}
\setval{8b.gdpo-w0.5.minus.gdpo-noaux.judge2.writing-abs}{8.9}
\setval{8b.gdpo-w0.5.minus.gdpo-noaux.judge2.writingbench}{+6.2}
\setval{8b.gdpo-w0.5.minus.gdpo-noaux.judge2.writingbench-abs}{6.2}
\setval{8b.gdpo-w0.5.minus.gdpo-noaux.livecodebench}{$-$0.6}
\setval{8b.gdpo-w0.5.minus.gdpo-noaux.livecodebench-abs}{0.6}
\setval{8b.gdpo-w0.5.minus.gdpo-noaux.longbench-write}{+5.2}
\setval{8b.gdpo-w0.5.minus.gdpo-noaux.longbench-write-abs}{5.2}
\setval{8b.gdpo-w0.5.minus.gdpo-noaux.math}{$-$8.1}
\setval{8b.gdpo-w0.5.minus.gdpo-noaux.math-abs}{8.1}
\setval{8b.gdpo-w0.5.minus.gdpo-noaux.math500}{$-$6.2}
\setval{8b.gdpo-w0.5.minus.gdpo-noaux.math500-abs}{6.2}
\setval{8b.gdpo-w0.5.minus.gdpo-noaux.mbpp-plus}{$-$2.0}
\setval{8b.gdpo-w0.5.minus.gdpo-noaux.mbpp-plus-abs}{2.0}
\setval{8b.gdpo-w0.5.minus.gdpo-noaux.minerva}{$-$2.8}
\setval{8b.gdpo-w0.5.minus.gdpo-noaux.minerva-abs}{2.8}
\setval{8b.gdpo-w0.5.minus.gdpo-noaux.olympiad-bench}{$-$8.8}
\setval{8b.gdpo-w0.5.minus.gdpo-noaux.olympiad-bench-abs}{8.8}
\setval{8b.gdpo-w0.5.minus.gdpo-noaux.writing}{+4.3}
\setval{8b.gdpo-w0.5.minus.gdpo-noaux.writing-abs}{4.3}
\setval{8b.gdpo-w0.5.minus.gdpo-noaux.writingbench}{+0.0}
\setval{8b.gdpo-w0.5.minus.gdpo-noaux.writingbench-abs}{0.0}
\setval{8b.gdpo-w0.5.minus.gdpo.aime24}{$-$2.1}
\setval{8b.gdpo-w0.5.minus.gdpo.aime24-abs}{2.1}
\setval{8b.gdpo-w0.5.minus.gdpo.aime25}{+1.0}
\setval{8b.gdpo-w0.5.minus.gdpo.aime25-abs}{1.0}
\setval{8b.gdpo-w0.5.minus.gdpo.amc23}{+1.3}
\setval{8b.gdpo-w0.5.minus.gdpo.amc23-abs}{1.3}
\setval{8b.gdpo-w0.5.minus.gdpo.avg}{+0.0}
\setval{8b.gdpo-w0.5.minus.gdpo.avg-abs}{0.0}
\setval{8b.gdpo-w0.5.minus.gdpo.code}{$-$0.5}
\setval{8b.gdpo-w0.5.minus.gdpo.code-abs}{0.5}
\setval{8b.gdpo-w0.5.minus.gdpo.creative-writing-v3}{$-$1.3}
\setval{8b.gdpo-w0.5.minus.gdpo.creative-writing-v3-abs}{1.3}
\setval{8b.gdpo-w0.5.minus.gdpo.guidebench}{$-$0.6}
\setval{8b.gdpo-w0.5.minus.gdpo.guidebench-abs}{0.6}
\setval{8b.gdpo-w0.5.minus.gdpo.humaneval-plus}{$-$0.3}
\setval{8b.gdpo-w0.5.minus.gdpo.humaneval-plus-abs}{0.3}
\setval{8b.gdpo-w0.5.minus.gdpo.if}{$-$0.9}
\setval{8b.gdpo-w0.5.minus.gdpo.if-abs}{0.9}
\setval{8b.gdpo-w0.5.minus.gdpo.ifbench}{$-$1.7}
\setval{8b.gdpo-w0.5.minus.gdpo.ifbench-abs}{1.7}
\setval{8b.gdpo-w0.5.minus.gdpo.ifeval}{$-$0.4}
\setval{8b.gdpo-w0.5.minus.gdpo.ifeval-abs}{0.4}
\setval{8b.gdpo-w0.5.minus.gdpo.judge2.avg}{$-$0.2}
\setval{8b.gdpo-w0.5.minus.gdpo.judge2.avg-abs}{0.2}
\setval{8b.gdpo-w0.5.minus.gdpo.judge2.creative-writing-v3}{$-$0.7}
\setval{8b.gdpo-w0.5.minus.gdpo.judge2.creative-writing-v3-abs}{0.7}
\setval{8b.gdpo-w0.5.minus.gdpo.judge2.longbench-write}{$-$0.5}
\setval{8b.gdpo-w0.5.minus.gdpo.judge2.longbench-write-abs}{0.5}
\setval{8b.gdpo-w0.5.minus.gdpo.judge2.writing}{$-$1.1}
\setval{8b.gdpo-w0.5.minus.gdpo.judge2.writing-abs}{1.1}
\setval{8b.gdpo-w0.5.minus.gdpo.judge2.writingbench}{$-$2.1}
\setval{8b.gdpo-w0.5.minus.gdpo.judge2.writingbench-abs}{2.1}
\setval{8b.gdpo-w0.5.minus.gdpo.livecodebench}{+0.6}
\setval{8b.gdpo-w0.5.minus.gdpo.livecodebench-abs}{0.6}
\setval{8b.gdpo-w0.5.minus.gdpo.longbench-write}{+1.6}
\setval{8b.gdpo-w0.5.minus.gdpo.longbench-write-abs}{1.6}
\setval{8b.gdpo-w0.5.minus.gdpo.math}{+0.8}
\setval{8b.gdpo-w0.5.minus.gdpo.math-abs}{0.8}
\setval{8b.gdpo-w0.5.minus.gdpo.math500}{+1.1}
\setval{8b.gdpo-w0.5.minus.gdpo.math500-abs}{1.1}
\setval{8b.gdpo-w0.5.minus.gdpo.mbpp-plus}{$-$1.7}
\setval{8b.gdpo-w0.5.minus.gdpo.mbpp-plus-abs}{1.7}
\setval{8b.gdpo-w0.5.minus.gdpo.minerva}{+1.6}
\setval{8b.gdpo-w0.5.minus.gdpo.minerva-abs}{1.6}
\setval{8b.gdpo-w0.5.minus.gdpo.olympiad-bench}{+1.8}
\setval{8b.gdpo-w0.5.minus.gdpo.olympiad-bench-abs}{1.8}
\setval{8b.gdpo-w0.5.minus.gdpo.writing}{$-$0.3}
\setval{8b.gdpo-w0.5.minus.gdpo.writing-abs}{0.3}
\setval{8b.gdpo-w0.5.minus.gdpo.writingbench}{$-$1.1}
\setval{8b.gdpo-w0.5.minus.gdpo.writingbench-abs}{1.1}
\setval{8b.gdpo-w0.5.minus.geoma.aime24}{+5.2}
\setval{8b.gdpo-w0.5.minus.geoma.aime24-abs}{5.2}
\setval{8b.gdpo-w0.5.minus.geoma.aime25}{+2.9}
\setval{8b.gdpo-w0.5.minus.geoma.aime25-abs}{2.9}
\setval{8b.gdpo-w0.5.minus.geoma.amc23}{$-$0.1}
\setval{8b.gdpo-w0.5.minus.geoma.amc23-abs}{0.1}
\setval{8b.gdpo-w0.5.minus.geoma.avg}{+1.6}
\setval{8b.gdpo-w0.5.minus.geoma.avg-abs}{1.6}
\setval{8b.gdpo-w0.5.minus.geoma.code}{$-$0.2}
\setval{8b.gdpo-w0.5.minus.geoma.code-abs}{0.2}
\setval{8b.gdpo-w0.5.minus.geoma.creative-writing-v3}{+1.2}
\setval{8b.gdpo-w0.5.minus.geoma.creative-writing-v3-abs}{1.2}
\setval{8b.gdpo-w0.5.minus.geoma.guidebench}{$-$0.4}
\setval{8b.gdpo-w0.5.minus.geoma.guidebench-abs}{0.4}
\setval{8b.gdpo-w0.5.minus.geoma.humaneval-plus}{+1.0}
\setval{8b.gdpo-w0.5.minus.geoma.humaneval-plus-abs}{1.0}
\setval{8b.gdpo-w0.5.minus.geoma.if}{+1.3}
\setval{8b.gdpo-w0.5.minus.geoma.if-abs}{1.3}
\setval{8b.gdpo-w0.5.minus.geoma.ifbench}{+3.7}
\setval{8b.gdpo-w0.5.minus.geoma.ifbench-abs}{3.7}
\setval{8b.gdpo-w0.5.minus.geoma.ifeval}{+0.5}
\setval{8b.gdpo-w0.5.minus.geoma.ifeval-abs}{0.5}
\setval{8b.gdpo-w0.5.minus.geoma.judge2.avg}{+1.4}
\setval{8b.gdpo-w0.5.minus.geoma.judge2.avg-abs}{1.4}
\setval{8b.gdpo-w0.5.minus.geoma.judge2.creative-writing-v3}{+2.8}
\setval{8b.gdpo-w0.5.minus.geoma.judge2.creative-writing-v3-abs}{2.8}
\setval{8b.gdpo-w0.5.minus.geoma.judge2.longbench-write}{$-$1.2}
\setval{8b.gdpo-w0.5.minus.geoma.judge2.longbench-write-abs}{1.2}
\setval{8b.gdpo-w0.5.minus.geoma.judge2.writing}{$-$0.2}
\setval{8b.gdpo-w0.5.minus.geoma.judge2.writing-abs}{0.2}
\setval{8b.gdpo-w0.5.minus.geoma.judge2.writingbench}{$-$2.1}
\setval{8b.gdpo-w0.5.minus.geoma.judge2.writingbench-abs}{2.1}
\setval{8b.gdpo-w0.5.minus.geoma.livecodebench}{+1.1}
\setval{8b.gdpo-w0.5.minus.geoma.livecodebench-abs}{1.1}
\setval{8b.gdpo-w0.5.minus.geoma.longbench-write}{+2.2}
\setval{8b.gdpo-w0.5.minus.geoma.longbench-write-abs}{2.2}
\setval{8b.gdpo-w0.5.minus.geoma.math}{+3.2}
\setval{8b.gdpo-w0.5.minus.geoma.math-abs}{3.2}
\setval{8b.gdpo-w0.5.minus.geoma.math500}{+4.4}
\setval{8b.gdpo-w0.5.minus.geoma.math500-abs}{4.4}
\setval{8b.gdpo-w0.5.minus.geoma.mbpp-plus}{$-$2.8}
\setval{8b.gdpo-w0.5.minus.geoma.mbpp-plus-abs}{2.8}
\setval{8b.gdpo-w0.5.minus.geoma.minerva}{+1.4}
\setval{8b.gdpo-w0.5.minus.geoma.minerva-abs}{1.4}
\setval{8b.gdpo-w0.5.minus.geoma.olympiad-bench}{+5.1}
\setval{8b.gdpo-w0.5.minus.geoma.olympiad-bench-abs}{5.1}
\setval{8b.gdpo-w0.5.minus.geoma.writing}{+0.8}
\setval{8b.gdpo-w0.5.minus.geoma.writing-abs}{0.8}
\setval{8b.gdpo-w0.5.minus.geoma.writingbench}{$-$0.9}
\setval{8b.gdpo-w0.5.minus.geoma.writingbench-abs}{0.9}
\setval{8b.gdpo-w0.5.minus.grpo-nodiv.aime24}{$-$1.4}
\setval{8b.gdpo-w0.5.minus.grpo-nodiv.aime24-abs}{1.4}
\setval{8b.gdpo-w0.5.minus.grpo-nodiv.aime25}{+0.6}
\setval{8b.gdpo-w0.5.minus.grpo-nodiv.aime25-abs}{0.6}
\setval{8b.gdpo-w0.5.minus.grpo-nodiv.amc23}{$-$0.2}
\setval{8b.gdpo-w0.5.minus.grpo-nodiv.amc23-abs}{0.2}
\setval{8b.gdpo-w0.5.minus.grpo-nodiv.avg}{+0.2}
\setval{8b.gdpo-w0.5.minus.grpo-nodiv.avg-abs}{0.2}
\setval{8b.gdpo-w0.5.minus.grpo-nodiv.code}{$-$1.3}
\setval{8b.gdpo-w0.5.minus.grpo-nodiv.code-abs}{1.3}
\setval{8b.gdpo-w0.5.minus.grpo-nodiv.creative-writing-v3}{+4.9}
\setval{8b.gdpo-w0.5.minus.grpo-nodiv.creative-writing-v3-abs}{4.9}
\setval{8b.gdpo-w0.5.minus.grpo-nodiv.guidebench}{$-$0.9}
\setval{8b.gdpo-w0.5.minus.grpo-nodiv.guidebench-abs}{0.9}
\setval{8b.gdpo-w0.5.minus.grpo-nodiv.humaneval-plus}{$-$2.6}
\setval{8b.gdpo-w0.5.minus.grpo-nodiv.humaneval-plus-abs}{2.6}
\setval{8b.gdpo-w0.5.minus.grpo-nodiv.if}{$-$0.4}
\setval{8b.gdpo-w0.5.minus.grpo-nodiv.if-abs}{0.4}
\setval{8b.gdpo-w0.5.minus.grpo-nodiv.ifbench}{+1.3}
\setval{8b.gdpo-w0.5.minus.grpo-nodiv.ifbench-abs}{1.3}
\setval{8b.gdpo-w0.5.minus.grpo-nodiv.ifeval}{$-$1.7}
\setval{8b.gdpo-w0.5.minus.grpo-nodiv.ifeval-abs}{1.7}
\setval{8b.gdpo-w0.5.minus.grpo-nodiv.judge2.avg}{$-$0.2}
\setval{8b.gdpo-w0.5.minus.grpo-nodiv.judge2.avg-abs}{0.2}
\setval{8b.gdpo-w0.5.minus.grpo-nodiv.judge2.creative-writing-v3}{+2.9}
\setval{8b.gdpo-w0.5.minus.grpo-nodiv.judge2.creative-writing-v3-abs}{2.9}
\setval{8b.gdpo-w0.5.minus.grpo-nodiv.judge2.longbench-write}{$-$2.9}
\setval{8b.gdpo-w0.5.minus.grpo-nodiv.judge2.longbench-write-abs}{2.9}
\setval{8b.gdpo-w0.5.minus.grpo-nodiv.judge2.writing}{+0.8}
\setval{8b.gdpo-w0.5.minus.grpo-nodiv.judge2.writing-abs}{0.8}
\setval{8b.gdpo-w0.5.minus.grpo-nodiv.judge2.writingbench}{+2.3}
\setval{8b.gdpo-w0.5.minus.grpo-nodiv.judge2.writingbench-abs}{2.3}
\setval{8b.gdpo-w0.5.minus.grpo-nodiv.livecodebench}{+1.8}
\setval{8b.gdpo-w0.5.minus.grpo-nodiv.livecodebench-abs}{1.8}
\setval{8b.gdpo-w0.5.minus.grpo-nodiv.longbench-write}{+0.1}
\setval{8b.gdpo-w0.5.minus.grpo-nodiv.longbench-write-abs}{0.1}
\setval{8b.gdpo-w0.5.minus.grpo-nodiv.math}{$-$0.2}
\setval{8b.gdpo-w0.5.minus.grpo-nodiv.math-abs}{0.2}
\setval{8b.gdpo-w0.5.minus.grpo-nodiv.math500}{$-$0.1}
\setval{8b.gdpo-w0.5.minus.grpo-nodiv.math500-abs}{0.1}
\setval{8b.gdpo-w0.5.minus.grpo-nodiv.mbpp-plus}{$-$3.2}
\setval{8b.gdpo-w0.5.minus.grpo-nodiv.mbpp-plus-abs}{3.2}
\setval{8b.gdpo-w0.5.minus.grpo-nodiv.minerva}{+0.4}
\setval{8b.gdpo-w0.5.minus.grpo-nodiv.minerva-abs}{0.4}
\setval{8b.gdpo-w0.5.minus.grpo-nodiv.olympiad-bench}{$-$0.4}
\setval{8b.gdpo-w0.5.minus.grpo-nodiv.olympiad-bench-abs}{0.4}
\setval{8b.gdpo-w0.5.minus.grpo-nodiv.writing}{+3.2}
\setval{8b.gdpo-w0.5.minus.grpo-nodiv.writing-abs}{3.2}
\setval{8b.gdpo-w0.5.minus.grpo-nodiv.writingbench}{+4.7}
\setval{8b.gdpo-w0.5.minus.grpo-nodiv.writingbench-abs}{4.7}
\setval{8b.gdpo-w0.5.minus.grpo.aime24}{+3.4}
\setval{8b.gdpo-w0.5.minus.grpo.aime24-abs}{3.4}
\setval{8b.gdpo-w0.5.minus.grpo.aime25}{$-$0.1}
\setval{8b.gdpo-w0.5.minus.grpo.aime25-abs}{0.1}
\setval{8b.gdpo-w0.5.minus.grpo.amc23}{$-$0.9}
\setval{8b.gdpo-w0.5.minus.grpo.amc23-abs}{0.9}
\setval{8b.gdpo-w0.5.minus.grpo.avg}{+0.4}
\setval{8b.gdpo-w0.5.minus.grpo.avg-abs}{0.4}
\setval{8b.gdpo-w0.5.minus.grpo.code}{$-$0.5}
\setval{8b.gdpo-w0.5.minus.grpo.code-abs}{0.5}
\setval{8b.gdpo-w0.5.minus.grpo.creative-writing-v3}{$-$0.9}
\setval{8b.gdpo-w0.5.minus.grpo.creative-writing-v3-abs}{0.9}
\setval{8b.gdpo-w0.5.minus.grpo.guidebench}{+2.1}
\setval{8b.gdpo-w0.5.minus.grpo.guidebench-abs}{2.1}
\setval{8b.gdpo-w0.5.minus.grpo.humaneval-plus}{+0.1}
\setval{8b.gdpo-w0.5.minus.grpo.humaneval-plus-abs}{0.1}
\setval{8b.gdpo-w0.5.minus.grpo.if}{$-$0.4}
\setval{8b.gdpo-w0.5.minus.grpo.if-abs}{0.4}
\setval{8b.gdpo-w0.5.minus.grpo.ifbench}{$-$4.3}
\setval{8b.gdpo-w0.5.minus.grpo.ifbench-abs}{4.3}
\setval{8b.gdpo-w0.5.minus.grpo.ifeval}{+1.0}
\setval{8b.gdpo-w0.5.minus.grpo.ifeval-abs}{1.0}
\setval{8b.gdpo-w0.5.minus.grpo.judge2.avg}{+0.6}
\setval{8b.gdpo-w0.5.minus.grpo.judge2.avg-abs}{0.6}
\setval{8b.gdpo-w0.5.minus.grpo.judge2.creative-writing-v3}{+1.1}
\setval{8b.gdpo-w0.5.minus.grpo.judge2.creative-writing-v3-abs}{1.1}
\setval{8b.gdpo-w0.5.minus.grpo.judge2.longbench-write}{+1.2}
\setval{8b.gdpo-w0.5.minus.grpo.judge2.longbench-write-abs}{1.2}
\setval{8b.gdpo-w0.5.minus.grpo.judge2.writing}{+0.6}
\setval{8b.gdpo-w0.5.minus.grpo.judge2.writing-abs}{0.6}
\setval{8b.gdpo-w0.5.minus.grpo.judge2.writingbench}{$-$0.7}
\setval{8b.gdpo-w0.5.minus.grpo.judge2.writingbench-abs}{0.7}
\setval{8b.gdpo-w0.5.minus.grpo.livecodebench}{+0.3}
\setval{8b.gdpo-w0.5.minus.grpo.livecodebench-abs}{0.3}
\setval{8b.gdpo-w0.5.minus.grpo.longbench-write}{+1.4}
\setval{8b.gdpo-w0.5.minus.grpo.longbench-write-abs}{1.4}
\setval{8b.gdpo-w0.5.minus.grpo.math}{+1.7}
\setval{8b.gdpo-w0.5.minus.grpo.math-abs}{1.7}
\setval{8b.gdpo-w0.5.minus.grpo.math500}{+3.4}
\setval{8b.gdpo-w0.5.minus.grpo.math500-abs}{3.4}
\setval{8b.gdpo-w0.5.minus.grpo.mbpp-plus}{$-$1.8}
\setval{8b.gdpo-w0.5.minus.grpo.mbpp-plus-abs}{1.8}
\setval{8b.gdpo-w0.5.minus.grpo.minerva}{+1.4}
\setval{8b.gdpo-w0.5.minus.grpo.minerva-abs}{1.4}
\setval{8b.gdpo-w0.5.minus.grpo.olympiad-bench}{+3.2}
\setval{8b.gdpo-w0.5.minus.grpo.olympiad-bench-abs}{3.2}
\setval{8b.gdpo-w0.5.minus.grpo.writing}{$-$0.3}
\setval{8b.gdpo-w0.5.minus.grpo.writing-abs}{0.3}
\setval{8b.gdpo-w0.5.minus.grpo.writingbench}{$-$1.4}
\setval{8b.gdpo-w0.5.minus.grpo.writingbench-abs}{1.4}
\setval{8b.gdpo-w0.5.minus.mt-grpo.aime24}{+4.8}
\setval{8b.gdpo-w0.5.minus.mt-grpo.aime24-abs}{4.8}
\setval{8b.gdpo-w0.5.minus.mt-grpo.aime25}{$-$0.3}
\setval{8b.gdpo-w0.5.minus.mt-grpo.aime25-abs}{0.3}
\setval{8b.gdpo-w0.5.minus.mt-grpo.amc23}{$-$2.2}
\setval{8b.gdpo-w0.5.minus.mt-grpo.amc23-abs}{2.2}
\setval{8b.gdpo-w0.5.minus.mt-grpo.avg}{+1.1}
\setval{8b.gdpo-w0.5.minus.mt-grpo.avg-abs}{1.1}
\setval{8b.gdpo-w0.5.minus.mt-grpo.code}{$-$1.5}
\setval{8b.gdpo-w0.5.minus.mt-grpo.code-abs}{1.5}
\setval{8b.gdpo-w0.5.minus.mt-grpo.creative-writing-v3}{+1.2}
\setval{8b.gdpo-w0.5.minus.mt-grpo.creative-writing-v3-abs}{1.2}
\setval{8b.gdpo-w0.5.minus.mt-grpo.guidebench}{+0.9}
\setval{8b.gdpo-w0.5.minus.mt-grpo.guidebench-abs}{0.9}
\setval{8b.gdpo-w0.5.minus.mt-grpo.humaneval-plus}{$-$1.3}
\setval{8b.gdpo-w0.5.minus.mt-grpo.humaneval-plus-abs}{1.3}
\setval{8b.gdpo-w0.5.minus.mt-grpo.if}{+0.7}
\setval{8b.gdpo-w0.5.minus.mt-grpo.if-abs}{0.7}
\setval{8b.gdpo-w0.5.minus.mt-grpo.ifbench}{+0.3}
\setval{8b.gdpo-w0.5.minus.mt-grpo.ifbench-abs}{0.3}
\setval{8b.gdpo-w0.5.minus.mt-grpo.ifeval}{+0.8}
\setval{8b.gdpo-w0.5.minus.mt-grpo.ifeval-abs}{0.8}
\setval{8b.gdpo-w0.5.minus.mt-grpo.judge2.avg}{+0.6}
\setval{8b.gdpo-w0.5.minus.mt-grpo.judge2.avg-abs}{0.6}
\setval{8b.gdpo-w0.5.minus.mt-grpo.judge2.creative-writing-v3}{$-$1.2}
\setval{8b.gdpo-w0.5.minus.mt-grpo.judge2.creative-writing-v3-abs}{1.2}
\setval{8b.gdpo-w0.5.minus.mt-grpo.judge2.longbench-write}{+1.2}
\setval{8b.gdpo-w0.5.minus.mt-grpo.judge2.longbench-write-abs}{1.2}
\setval{8b.gdpo-w0.5.minus.mt-grpo.judge2.writing}{+0.2}
\setval{8b.gdpo-w0.5.minus.mt-grpo.judge2.writing-abs}{0.2}
\setval{8b.gdpo-w0.5.minus.mt-grpo.judge2.writingbench}{+0.7}
\setval{8b.gdpo-w0.5.minus.mt-grpo.judge2.writingbench-abs}{0.7}
\setval{8b.gdpo-w0.5.minus.mt-grpo.livecodebench}{$-$0.9}
\setval{8b.gdpo-w0.5.minus.mt-grpo.livecodebench-abs}{0.9}
\setval{8b.gdpo-w0.5.minus.mt-grpo.longbench-write}{+3.7}
\setval{8b.gdpo-w0.5.minus.mt-grpo.longbench-write-abs}{3.7}
\setval{8b.gdpo-w0.5.minus.mt-grpo.math}{+1.9}
\setval{8b.gdpo-w0.5.minus.mt-grpo.math-abs}{1.9}
\setval{8b.gdpo-w0.5.minus.mt-grpo.math500}{+3.5}
\setval{8b.gdpo-w0.5.minus.mt-grpo.math500-abs}{3.5}
\setval{8b.gdpo-w0.5.minus.mt-grpo.mbpp-plus}{$-$2.4}
\setval{8b.gdpo-w0.5.minus.mt-grpo.mbpp-plus-abs}{2.4}
\setval{8b.gdpo-w0.5.minus.mt-grpo.minerva}{+0.2}
\setval{8b.gdpo-w0.5.minus.mt-grpo.minerva-abs}{0.2}
\setval{8b.gdpo-w0.5.minus.mt-grpo.olympiad-bench}{+5.4}
\setval{8b.gdpo-w0.5.minus.mt-grpo.olympiad-bench-abs}{5.4}
\setval{8b.gdpo-w0.5.minus.mt-grpo.writing}{+2.8}
\setval{8b.gdpo-w0.5.minus.mt-grpo.writing-abs}{2.8}
\setval{8b.gdpo-w0.5.minus.mt-grpo.writingbench}{+3.3}
\setval{8b.gdpo-w0.5.minus.mt-grpo.writingbench-abs}{3.3}
\setval{8b.gdpo-w0.5.minus.papo.aime24}{+7.1}
\setval{8b.gdpo-w0.5.minus.papo.aime24-abs}{7.1}
\setval{8b.gdpo-w0.5.minus.papo.aime25}{+5.3}
\setval{8b.gdpo-w0.5.minus.papo.aime25-abs}{5.3}
\setval{8b.gdpo-w0.5.minus.papo.amc23}{+1.4}
\setval{8b.gdpo-w0.5.minus.papo.amc23-abs}{1.4}
\setval{8b.gdpo-w0.5.minus.papo.avg}{+2.5}
\setval{8b.gdpo-w0.5.minus.papo.avg-abs}{2.5}
\setval{8b.gdpo-w0.5.minus.papo.code}{+0.3}
\setval{8b.gdpo-w0.5.minus.papo.code-abs}{0.3}
\setval{8b.gdpo-w0.5.minus.papo.creative-writing-v3}{+0.4}
\setval{8b.gdpo-w0.5.minus.papo.creative-writing-v3-abs}{0.4}
\setval{8b.gdpo-w0.5.minus.papo.guidebench}{$-$0.4}
\setval{8b.gdpo-w0.5.minus.papo.guidebench-abs}{0.4}
\setval{8b.gdpo-w0.5.minus.papo.humaneval-plus}{+1.3}
\setval{8b.gdpo-w0.5.minus.papo.humaneval-plus-abs}{1.3}
\setval{8b.gdpo-w0.5.minus.papo.if}{+0.5}
\setval{8b.gdpo-w0.5.minus.papo.if-abs}{0.5}
\setval{8b.gdpo-w0.5.minus.papo.ifbench}{+0.3}
\setval{8b.gdpo-w0.5.minus.papo.ifbench-abs}{0.3}
\setval{8b.gdpo-w0.5.minus.papo.ifeval}{+1.4}
\setval{8b.gdpo-w0.5.minus.papo.ifeval-abs}{1.4}
\setval{8b.gdpo-w0.5.minus.papo.judge2.avg}{+2.2}
\setval{8b.gdpo-w0.5.minus.papo.judge2.avg-abs}{2.2}
\setval{8b.gdpo-w0.5.minus.papo.judge2.creative-writing-v3}{$-$1.5}
\setval{8b.gdpo-w0.5.minus.papo.judge2.creative-writing-v3-abs}{1.5}
\setval{8b.gdpo-w0.5.minus.papo.judge2.longbench-write}{+4.1}
\setval{8b.gdpo-w0.5.minus.papo.judge2.longbench-write-abs}{4.1}
\setval{8b.gdpo-w0.5.minus.papo.judge2.writing}{+0.8}
\setval{8b.gdpo-w0.5.minus.papo.judge2.writing-abs}{0.8}
\setval{8b.gdpo-w0.5.minus.papo.judge2.writingbench}{$-$0.1}
\setval{8b.gdpo-w0.5.minus.papo.judge2.writingbench-abs}{0.1}
\setval{8b.gdpo-w0.5.minus.papo.livecodebench}{+0.2}
\setval{8b.gdpo-w0.5.minus.papo.livecodebench-abs}{0.2}
\setval{8b.gdpo-w0.5.minus.papo.longbench-write}{+5.7}
\setval{8b.gdpo-w0.5.minus.papo.longbench-write-abs}{5.7}
\setval{8b.gdpo-w0.5.minus.papo.math}{+4.6}
\setval{8b.gdpo-w0.5.minus.papo.math-abs}{4.6}
\setval{8b.gdpo-w0.5.minus.papo.math500}{+5.6}
\setval{8b.gdpo-w0.5.minus.papo.math500-abs}{5.6}
\setval{8b.gdpo-w0.5.minus.papo.mbpp-plus}{$-$0.6}
\setval{8b.gdpo-w0.5.minus.papo.mbpp-plus-abs}{0.6}
\setval{8b.gdpo-w0.5.minus.papo.minerva}{+1.9}
\setval{8b.gdpo-w0.5.minus.papo.minerva-abs}{1.9}
\setval{8b.gdpo-w0.5.minus.papo.olympiad-bench}{+6.3}
\setval{8b.gdpo-w0.5.minus.papo.olympiad-bench-abs}{6.3}
\setval{8b.gdpo-w0.5.minus.papo.writing}{+2.6}
\setval{8b.gdpo-w0.5.minus.papo.writing-abs}{2.6}
\setval{8b.gdpo-w0.5.minus.papo.writingbench}{+1.8}
\setval{8b.gdpo-w0.5.minus.papo.writingbench-abs}{1.8}
\setval{8b.gdpo-w0.5.minus.saw.aime24}{+4.0}
\setval{8b.gdpo-w0.5.minus.saw.aime24-abs}{4.0}
\setval{8b.gdpo-w0.5.minus.saw.aime25}{$-$0.1}
\setval{8b.gdpo-w0.5.minus.saw.aime25-abs}{0.1}
\setval{8b.gdpo-w0.5.minus.saw.amc23}{+1.2}
\setval{8b.gdpo-w0.5.minus.saw.amc23-abs}{1.2}
\setval{8b.gdpo-w0.5.minus.saw.avg}{+0.4}
\setval{8b.gdpo-w0.5.minus.saw.avg-abs}{0.4}
\setval{8b.gdpo-w0.5.minus.saw.code}{$-$1.2}
\setval{8b.gdpo-w0.5.minus.saw.code-abs}{1.2}
\setval{8b.gdpo-w0.5.minus.saw.creative-writing-v3}{+1.9}
\setval{8b.gdpo-w0.5.minus.saw.creative-writing-v3-abs}{1.9}
\setval{8b.gdpo-w0.5.minus.saw.guidebench}{$-$0.3}
\setval{8b.gdpo-w0.5.minus.saw.guidebench-abs}{0.3}
\setval{8b.gdpo-w0.5.minus.saw.humaneval-plus}{$-$2.1}
\setval{8b.gdpo-w0.5.minus.saw.humaneval-plus-abs}{2.1}
\setval{8b.gdpo-w0.5.minus.saw.if}{$-$0.8}
\setval{8b.gdpo-w0.5.minus.saw.if-abs}{0.8}
\setval{8b.gdpo-w0.5.minus.saw.ifbench}{$-$2.3}
\setval{8b.gdpo-w0.5.minus.saw.ifbench-abs}{2.3}
\setval{8b.gdpo-w0.5.minus.saw.ifeval}{+0.1}
\setval{8b.gdpo-w0.5.minus.saw.ifeval-abs}{0.1}
\setval{8b.gdpo-w0.5.minus.saw.judge2.avg}{+0.4}
\setval{8b.gdpo-w0.5.minus.saw.judge2.avg-abs}{0.4}
\setval{8b.gdpo-w0.5.minus.saw.judge2.creative-writing-v3}{+3.8}
\setval{8b.gdpo-w0.5.minus.saw.judge2.creative-writing-v3-abs}{3.8}
\setval{8b.gdpo-w0.5.minus.saw.judge2.longbench-write}{$-$1.1}
\setval{8b.gdpo-w0.5.minus.saw.judge2.longbench-write-abs}{1.1}
\setval{8b.gdpo-w0.5.minus.saw.judge2.writing}{+0.6}
\setval{8b.gdpo-w0.5.minus.saw.judge2.writing-abs}{0.6}
\setval{8b.gdpo-w0.5.minus.saw.judge2.writingbench}{$-$0.7}
\setval{8b.gdpo-w0.5.minus.saw.judge2.writingbench-abs}{0.7}
\setval{8b.gdpo-w0.5.minus.saw.livecodebench}{+0.1}
\setval{8b.gdpo-w0.5.minus.saw.livecodebench-abs}{0.1}
\setval{8b.gdpo-w0.5.minus.saw.longbench-write}{+0.2}
\setval{8b.gdpo-w0.5.minus.saw.longbench-write-abs}{0.2}
\setval{8b.gdpo-w0.5.minus.saw.math}{+1.6}
\setval{8b.gdpo-w0.5.minus.saw.math-abs}{1.6}
\setval{8b.gdpo-w0.5.minus.saw.math500}{+2.1}
\setval{8b.gdpo-w0.5.minus.saw.math500-abs}{2.1}
\setval{8b.gdpo-w0.5.minus.saw.mbpp-plus}{$-$1.5}
\setval{8b.gdpo-w0.5.minus.saw.mbpp-plus-abs}{1.5}
\setval{8b.gdpo-w0.5.minus.saw.minerva}{$-$0.6}
\setval{8b.gdpo-w0.5.minus.saw.minerva-abs}{0.6}
\setval{8b.gdpo-w0.5.minus.saw.olympiad-bench}{+3.0}
\setval{8b.gdpo-w0.5.minus.saw.olympiad-bench-abs}{3.0}
\setval{8b.gdpo-w0.5.minus.saw.writing}{+0.6}
\setval{8b.gdpo-w0.5.minus.saw.writing-abs}{0.6}
\setval{8b.gdpo-w0.5.minus.saw.writingbench}{$-$0.4}
\setval{8b.gdpo-w0.5.minus.saw.writingbench-abs}{0.4}
\setval{8b.gdpo-w0.5.passing-at-ceiling.code.early}{0.71}
\setval{8b.gdpo-w0.5.passing-at-ceiling.code.late}{0.75}
\setval{8b.gdpo-w0.5.passing-at-ceiling.code.run}{0.74}
\setval{8b.gdpo-w0.5.passing-at-ceiling.if.early}{0.32}
\setval{8b.gdpo-w0.5.passing-at-ceiling.if.late}{0.57}
\setval{8b.gdpo-w0.5.passing-at-ceiling.if.run}{0.49}
\setval{8b.gdpo-w0.5.passing-at-ceiling.math.early}{1.00}
\setval{8b.gdpo-w0.5.passing-at-ceiling.math.late}{1.00}
\setval{8b.gdpo-w0.5.passing-at-ceiling.math.run}{1.00}
\setval{8b.gdpo-w0.5.passing-at-ceiling.writing.early}{0.00}
\setval{8b.gdpo-w0.5.passing-at-ceiling.writing.late}{0.01}
\setval{8b.gdpo-w0.5.passing-at-ceiling.writing.run}{0.00}
\setval{8b.gdpo-w0.5.primary-cosine.code.early}{0.98}
\setval{8b.gdpo-w0.5.primary-cosine.code.late}{0.97}
\setval{8b.gdpo-w0.5.primary-cosine.code.run}{0.98}
\setval{8b.gdpo-w0.5.primary-cosine.if.early}{0.82}
\setval{8b.gdpo-w0.5.primary-cosine.if.late}{0.81}
\setval{8b.gdpo-w0.5.primary-cosine.if.run}{0.83}
\setval{8b.gdpo-w0.5.primary-cosine.math.early}{0.98}
\setval{8b.gdpo-w0.5.primary-cosine.math.late}{0.98}
\setval{8b.gdpo-w0.5.primary-cosine.math.run}{0.98}
\setval{8b.gdpo-w0.5.primary-cosine.writing.early}{0.93}
\setval{8b.gdpo-w0.5.primary-cosine.writing.late}{0.90}
\setval{8b.gdpo-w0.5.primary-cosine.writing.run}{0.89}
\setval{8b.gdpo-w0.5.primary-level.code.early}{0.54}
\setval{8b.gdpo-w0.5.primary-level.code.late}{0.77}
\setval{8b.gdpo-w0.5.primary-level.code.run}{0.72}
\setval{8b.gdpo-w0.5.primary-level.if.early}{0.35}
\setval{8b.gdpo-w0.5.primary-level.if.late}{0.76}
\setval{8b.gdpo-w0.5.primary-level.if.run}{0.65}
\setval{8b.gdpo-w0.5.primary-level.math.early}{0.49}
\setval{8b.gdpo-w0.5.primary-level.math.late}{0.71}
\setval{8b.gdpo-w0.5.primary-level.math.run}{0.65}
\setval{8b.gdpo-w0.5.primary-level.writing.early}{0.53}
\setval{8b.gdpo-w0.5.primary-level.writing.late}{0.77}
\setval{8b.gdpo-w0.5.primary-level.writing.run}{0.73}
\setval{8b.gdpo-w0.5.solved.code.early}{0.08}
\setval{8b.gdpo-w0.5.solved.code.late}{0.05}
\setval{8b.gdpo-w0.5.solved.code.run}{0.04}
\setval{8b.gdpo-w0.5.solved.if.early}{0.03}
\setval{8b.gdpo-w0.5.solved.if.late}{0.19}
\setval{8b.gdpo-w0.5.solved.if.run}{0.13}
\setval{8b.gdpo-w0.5.solved.math.early}{0.15}
\setval{8b.gdpo-w0.5.solved.math.late}{0.19}
\setval{8b.gdpo-w0.5.solved.math.run}{0.20}
\setval{8b.gdpo-w0.5.solved.writing.early}{0.00}
\setval{8b.gdpo-w0.5.solved.writing.late}{0.00}
\setval{8b.gdpo-w0.5.solved.writing.run}{0.00}
\setval{8b.gdpo-w0.5.update-share.code.early}{0.21}
\setval{8b.gdpo-w0.5.update-share.code.late}{0.22}
\setval{8b.gdpo-w0.5.update-share.code.run}{0.22}
\setval{8b.gdpo-w0.5.update-share.if.early}{0.25}
\setval{8b.gdpo-w0.5.update-share.if.late}{0.18}
\setval{8b.gdpo-w0.5.update-share.if.run}{0.22}
\setval{8b.gdpo-w0.5.update-share.math.early}{0.18}
\setval{8b.gdpo-w0.5.update-share.math.late}{0.25}
\setval{8b.gdpo-w0.5.update-share.math.run}{0.23}
\setval{8b.gdpo-w0.5.update-share.writing.early}{0.36}
\setval{8b.gdpo-w0.5.update-share.writing.late}{0.35}
\setval{8b.gdpo-w0.5.update-share.writing.run}{0.32}
\setval{8b.gdpo.aux-change.code}{+0.18}
\setval{8b.gdpo.aux-change.if}{+0.18}
\setval{8b.gdpo.aux-change.math}{+0.13}
\setval{8b.gdpo.aux-change.writing}{+0.14}
\setval{8b.gdpo.aux-change.writing-rm}{+0.33}
\setval{8b.gdpo.aux-full.code.early}{0.71}
\setval{8b.gdpo.aux-full.code.late}{0.96}
\setval{8b.gdpo.aux-full.code.run}{0.93}
\setval{8b.gdpo.aux-full.if.early}{0.72}
\setval{8b.gdpo.aux-full.if.late}{0.94}
\setval{8b.gdpo.aux-full.if.run}{0.90}
\setval{8b.gdpo.aux-full.math.early}{0.57}
\setval{8b.gdpo.aux-full.math.late}{0.88}
\setval{8b.gdpo.aux-full.math.run}{0.82}
\setval{8b.gdpo.aux-full.writing-rm.early}{0.07}
\setval{8b.gdpo.aux-full.writing-rm.late}{0.66}
\setval{8b.gdpo.aux-full.writing-rm.run}{0.53}
\setval{8b.gdpo.aux-full.writing.early}{0.66}
\setval{8b.gdpo.aux-full.writing.late}{0.81}
\setval{8b.gdpo.aux-full.writing.run}{0.73}
\setval{8b.gdpo.aux-level.code.early}{0.93}
\setval{8b.gdpo.aux-level.code.late}{0.99}
\setval{8b.gdpo.aux-level.code.run}{0.99}
\setval{8b.gdpo.aux-level.if.early}{0.88}
\setval{8b.gdpo.aux-level.if.late}{0.97}
\setval{8b.gdpo.aux-level.if.run}{0.95}
\setval{8b.gdpo.aux-level.math.early}{0.91}
\setval{8b.gdpo.aux-level.math.late}{0.97}
\setval{8b.gdpo.aux-level.math.run}{0.96}
\setval{8b.gdpo.aux-level.writing-rm.early}{0.77}
\setval{8b.gdpo.aux-level.writing-rm.late}{0.97}
\setval{8b.gdpo.aux-level.writing-rm.run}{0.94}
\setval{8b.gdpo.aux-level.writing.early}{0.91}
\setval{8b.gdpo.aux-level.writing.late}{0.98}
\setval{8b.gdpo.aux-level.writing.run}{0.96}
\setval{8b.gdpo.bench.aime24}{21.4}
\setval{8b.gdpo.bench.aime25}{14.4}
\setval{8b.gdpo.bench.amc23}{62.2}
\setval{8b.gdpo.bench.creative-writing-v3}{52.9}
\setval{8b.gdpo.bench.guidebench}{84.5}
\setval{8b.gdpo.bench.humaneval-plus}{84.1}
\setval{8b.gdpo.bench.ifbench}{39.3}
\setval{8b.gdpo.bench.ifeval}{88.1}
\setval{8b.gdpo.bench.livecodebench}{26.2}
\setval{8b.gdpo.bench.longbench-write}{69.6}
\setval{8b.gdpo.bench.math500}{83.0}
\setval{8b.gdpo.bench.mbpp-plus}{74.5}
\setval{8b.gdpo.bench.minerva}{49.6}
\setval{8b.gdpo.bench.olympiad-bench}{49.3}
\setval{8b.gdpo.bench.writingbench}{76.1}
\setval{8b.gdpo.eval.avg}{58.3}
\setval{8b.gdpo.eval.code}{61.6}
\setval{8b.gdpo.eval.if}{70.7}
\setval{8b.gdpo.eval.math}{46.6}
\setval{8b.gdpo.eval.writing}{66.2}
\setval{8b.gdpo.gate.code.against.early}{0.04}
\setval{8b.gdpo.gate.code.against.late}{0.02}
\setval{8b.gdpo.gate.code.against.run}{0.02}
\setval{8b.gdpo.gate.code.agreement}{+0.003}
\setval{8b.gdpo.gate.code.ceiling.early}{0.11}
\setval{8b.gdpo.gate.code.ceiling.late}{0.04}
\setval{8b.gdpo.gate.code.ceiling.run}{0.05}
\setval{8b.gdpo.gate.code.cosine.early}{0.84}
\setval{8b.gdpo.gate.code.cosine.late}{0.77}
\setval{8b.gdpo.gate.code.cosine.run}{0.80}
\setval{8b.gdpo.gate.if.against.early}{0.04}
\setval{8b.gdpo.gate.if.against.late}{0.03}
\setval{8b.gdpo.gate.if.against.run}{0.04}
\setval{8b.gdpo.gate.if.agreement}{+0.011}
\setval{8b.gdpo.gate.if.ceiling.early}{0.05}
\setval{8b.gdpo.gate.if.ceiling.late}{0.15}
\setval{8b.gdpo.gate.if.ceiling.run}{0.11}
\setval{8b.gdpo.gate.if.cosine.early}{0.76}
\setval{8b.gdpo.gate.if.cosine.late}{0.53}
\setval{8b.gdpo.gate.if.cosine.run}{0.60}
\setval{8b.gdpo.gate.math.against.early}{0.00}
\setval{8b.gdpo.gate.math.against.late}{0.00}
\setval{8b.gdpo.gate.math.against.run}{0.00}
\setval{8b.gdpo.gate.math.ceiling.early}{0.23}
\setval{8b.gdpo.gate.math.ceiling.late}{0.21}
\setval{8b.gdpo.gate.math.ceiling.run}{0.21}
\setval{8b.gdpo.gate.math.cosine.early}{0.86}
\setval{8b.gdpo.gate.math.cosine.late}{0.87}
\setval{8b.gdpo.gate.math.cosine.run}{0.87}
\setval{8b.gdpo.gate.writing-rm.against.early}{0.19}
\setval{8b.gdpo.gate.writing-rm.against.late}{0.53}
\setval{8b.gdpo.gate.writing-rm.against.run}{0.48}
\setval{8b.gdpo.gate.writing-rm.agreement}{+0.048}
\setval{8b.gdpo.gate.writing-rm.ceiling.early}{0.00}
\setval{8b.gdpo.gate.writing-rm.ceiling.late}{0.00}
\setval{8b.gdpo.gate.writing-rm.ceiling.run}{0.00}
\setval{8b.gdpo.gate.writing-rm.cosine.early}{0.71}
\setval{8b.gdpo.gate.writing-rm.cosine.late}{0.11}
\setval{8b.gdpo.gate.writing-rm.cosine.run}{0.21}
\setval{8b.gdpo.gate.writing.against.early}{0.27}
\setval{8b.gdpo.gate.writing.against.late}{0.59}
\setval{8b.gdpo.gate.writing.against.run}{0.55}
\setval{8b.gdpo.gate.writing.agreement}{$-$0.036}
\setval{8b.gdpo.gate.writing.ceiling.early}{0.00}
\setval{8b.gdpo.gate.writing.ceiling.late}{0.00}
\setval{8b.gdpo.gate.writing.ceiling.run}{0.00}
\setval{8b.gdpo.gate.writing.cosine.early}{0.54}
\setval{8b.gdpo.gate.writing.cosine.late}{0.04}
\setval{8b.gdpo.gate.writing.cosine.run}{0.10}
\setval{8b.gdpo.judge2.avg}{56.0}
\setval{8b.gdpo.judge2.creative-writing-v3}{41.0}
\setval{8b.gdpo.judge2.longbench-write}{64.2}
\setval{8b.gdpo.judge2.writing}{54.5}
\setval{8b.gdpo.judge2.writingbench}{58.4}
\setval{8b.gdpo.last-step}{300}
\setval{8b.gdpo.minus.base.aime24}{+10.0}
\setval{8b.gdpo.minus.base.aime24-abs}{10.0}
\setval{8b.gdpo.minus.base.aime25}{+5.8}
\setval{8b.gdpo.minus.base.aime25-abs}{5.8}
\setval{8b.gdpo.minus.base.amc23}{+12.0}
\setval{8b.gdpo.minus.base.amc23-abs}{12.0}
\setval{8b.gdpo.minus.base.avg}{+16.9}
\setval{8b.gdpo.minus.base.avg-abs}{16.9}
\setval{8b.gdpo.minus.base.code}{+5.5}
\setval{8b.gdpo.minus.base.code-abs}{5.5}
\setval{8b.gdpo.minus.base.creative-writing-v3}{+26.2}
\setval{8b.gdpo.minus.base.creative-writing-v3-abs}{26.2}
\setval{8b.gdpo.minus.base.guidebench}{+39.0}
\setval{8b.gdpo.minus.base.guidebench-abs}{39.0}
\setval{8b.gdpo.minus.base.humaneval-plus}{+4.0}
\setval{8b.gdpo.minus.base.humaneval-plus-abs}{4.0}
\setval{8b.gdpo.minus.base.if}{+32.0}
\setval{8b.gdpo.minus.base.if-abs}{32.0}
\setval{8b.gdpo.minus.base.ifbench}{+22.0}
\setval{8b.gdpo.minus.base.ifbench-abs}{22.0}
\setval{8b.gdpo.minus.base.ifeval}{+35.1}
\setval{8b.gdpo.minus.base.ifeval-abs}{35.1}
\setval{8b.gdpo.minus.base.judge2.avg}{+15.0}
\setval{8b.gdpo.minus.base.judge2.avg-abs}{15.0}
\setval{8b.gdpo.minus.base.judge2.creative-writing-v3}{+14.6}
\setval{8b.gdpo.minus.base.judge2.creative-writing-v3-abs}{14.6}
\setval{8b.gdpo.minus.base.judge2.longbench-write}{+16.1}
\setval{8b.gdpo.minus.base.judge2.longbench-write-abs}{16.1}
\setval{8b.gdpo.minus.base.judge2.writing}{+18.4}
\setval{8b.gdpo.minus.base.judge2.writing-abs}{18.4}
\setval{8b.gdpo.minus.base.judge2.writingbench}{+24.4}
\setval{8b.gdpo.minus.base.judge2.writingbench-abs}{24.4}
\setval{8b.gdpo.minus.base.livecodebench}{+5.3}
\setval{8b.gdpo.minus.base.livecodebench-abs}{5.3}
\setval{8b.gdpo.minus.base.longbench-write}{+18.2}
\setval{8b.gdpo.minus.base.longbench-write-abs}{18.2}
\setval{8b.gdpo.minus.base.math}{+9.5}
\setval{8b.gdpo.minus.base.math-abs}{9.5}
\setval{8b.gdpo.minus.base.math500}{+10.5}
\setval{8b.gdpo.minus.base.math500-abs}{10.5}
\setval{8b.gdpo.minus.base.mbpp-plus}{+7.2}
\setval{8b.gdpo.minus.base.mbpp-plus-abs}{7.2}
\setval{8b.gdpo.minus.base.minerva}{+8.7}
\setval{8b.gdpo.minus.base.minerva-abs}{8.7}
\setval{8b.gdpo.minus.base.olympiad-bench}{+9.9}
\setval{8b.gdpo.minus.base.olympiad-bench-abs}{9.9}
\setval{8b.gdpo.minus.base.writing}{+27.9}
\setval{8b.gdpo.minus.base.writing-abs}{27.9}
\setval{8b.gdpo.minus.base.writingbench}{+39.3}
\setval{8b.gdpo.minus.base.writingbench-abs}{39.3}
\setval{8b.gdpo.minus.concur-free.aime24}{+1.7}
\setval{8b.gdpo.minus.concur-free.aime24-abs}{1.7}
\setval{8b.gdpo.minus.concur-free.aime25}{$-$2.2}
\setval{8b.gdpo.minus.concur-free.aime25-abs}{2.2}
\setval{8b.gdpo.minus.concur-free.amc23}{$-$2.5}
\setval{8b.gdpo.minus.concur-free.amc23-abs}{2.5}
\setval{8b.gdpo.minus.concur-free.avg}{$-$1.9}
\setval{8b.gdpo.minus.concur-free.avg-abs}{1.9}
\setval{8b.gdpo.minus.concur-free.code}{$-$0.9}
\setval{8b.gdpo.minus.concur-free.code-abs}{0.9}
\setval{8b.gdpo.minus.concur-free.creative-writing-v3}{+1.3}
\setval{8b.gdpo.minus.concur-free.creative-writing-v3-abs}{1.3}
\setval{8b.gdpo.minus.concur-free.guidebench}{+1.6}
\setval{8b.gdpo.minus.concur-free.guidebench-abs}{1.6}
\setval{8b.gdpo.minus.concur-free.humaneval-plus}{$-$1.2}
\setval{8b.gdpo.minus.concur-free.humaneval-plus-abs}{1.2}
\setval{8b.gdpo.minus.concur-free.if}{+0.4}
\setval{8b.gdpo.minus.concur-free.if-abs}{0.4}
\setval{8b.gdpo.minus.concur-free.ifbench}{+0.7}
\setval{8b.gdpo.minus.concur-free.ifbench-abs}{0.7}
\setval{8b.gdpo.minus.concur-free.ifeval}{$-$1.1}
\setval{8b.gdpo.minus.concur-free.ifeval-abs}{1.1}
\setval{8b.gdpo.minus.concur-free.judge2.avg}{$-$1.3}
\setval{8b.gdpo.minus.concur-free.judge2.avg-abs}{1.3}
\setval{8b.gdpo.minus.concur-free.judge2.creative-writing-v3}{$-$1.5}
\setval{8b.gdpo.minus.concur-free.judge2.creative-writing-v3-abs}{1.5}
\setval{8b.gdpo.minus.concur-free.judge2.longbench-write}{$-$9.9}
\setval{8b.gdpo.minus.concur-free.judge2.longbench-write-abs}{9.9}
\setval{8b.gdpo.minus.concur-free.judge2.writing}{$-$4.6}
\setval{8b.gdpo.minus.concur-free.judge2.writing-abs}{4.6}
\setval{8b.gdpo.minus.concur-free.judge2.writingbench}{$-$2.3}
\setval{8b.gdpo.minus.concur-free.judge2.writingbench-abs}{2.3}
\setval{8b.gdpo.minus.concur-free.livecodebench}{$-$0.9}
\setval{8b.gdpo.minus.concur-free.livecodebench-abs}{0.9}
\setval{8b.gdpo.minus.concur-free.longbench-write}{$-$13.4}
\setval{8b.gdpo.minus.concur-free.longbench-write-abs}{13.4}
\setval{8b.gdpo.minus.concur-free.math}{$-$0.7}
\setval{8b.gdpo.minus.concur-free.math-abs}{0.7}
\setval{8b.gdpo.minus.concur-free.math500}{$-$0.4}
\setval{8b.gdpo.minus.concur-free.math500-abs}{0.4}
\setval{8b.gdpo.minus.concur-free.mbpp-plus}{$-$0.6}
\setval{8b.gdpo.minus.concur-free.mbpp-plus-abs}{0.6}
\setval{8b.gdpo.minus.concur-free.minerva}{$-$0.2}
\setval{8b.gdpo.minus.concur-free.minerva-abs}{0.2}
\setval{8b.gdpo.minus.concur-free.olympiad-bench}{$-$0.7}
\setval{8b.gdpo.minus.concur-free.olympiad-bench-abs}{0.7}
\setval{8b.gdpo.minus.concur-free.writing}{$-$7.6}
\setval{8b.gdpo.minus.concur-free.writing-abs}{7.6}
\setval{8b.gdpo.minus.concur-free.writingbench}{$-$10.7}
\setval{8b.gdpo.minus.concur-free.writingbench-abs}{10.7}
\setval{8b.gdpo.minus.concur-tied.aime24}{$-$9.9}
\setval{8b.gdpo.minus.concur-tied.aime24-abs}{9.9}
\setval{8b.gdpo.minus.concur-tied.aime25}{$-$9.6}
\setval{8b.gdpo.minus.concur-tied.aime25-abs}{9.6}
\setval{8b.gdpo.minus.concur-tied.amc23}{$-$16.3}
\setval{8b.gdpo.minus.concur-tied.amc23-abs}{16.3}
\setval{8b.gdpo.minus.concur-tied.avg}{$-$5.7}
\setval{8b.gdpo.minus.concur-tied.avg-abs}{5.7}
\setval{8b.gdpo.minus.concur-tied.code}{$-$1.5}
\setval{8b.gdpo.minus.concur-tied.code-abs}{1.5}
\setval{8b.gdpo.minus.concur-tied.creative-writing-v3}{+0.5}
\setval{8b.gdpo.minus.concur-tied.creative-writing-v3-abs}{0.5}
\setval{8b.gdpo.minus.concur-tied.guidebench}{+1.2}
\setval{8b.gdpo.minus.concur-tied.guidebench-abs}{1.2}
\setval{8b.gdpo.minus.concur-tied.humaneval-plus}{$-$3.3}
\setval{8b.gdpo.minus.concur-tied.humaneval-plus-abs}{3.3}
\setval{8b.gdpo.minus.concur-tied.if}{+0.1}
\setval{8b.gdpo.minus.concur-tied.if-abs}{0.1}
\setval{8b.gdpo.minus.concur-tied.ifbench}{+0.0}
\setval{8b.gdpo.minus.concur-tied.ifbench-abs}{0.0}
\setval{8b.gdpo.minus.concur-tied.ifeval}{$-$0.7}
\setval{8b.gdpo.minus.concur-tied.ifeval-abs}{0.7}
\setval{8b.gdpo.minus.concur-tied.judge2.avg}{$-$5.3}
\setval{8b.gdpo.minus.concur-tied.judge2.avg-abs}{5.3}
\setval{8b.gdpo.minus.concur-tied.judge2.creative-writing-v3}{$-$0.8}
\setval{8b.gdpo.minus.concur-tied.judge2.creative-writing-v3-abs}{0.8}
\setval{8b.gdpo.minus.concur-tied.judge2.longbench-write}{$-$10.6}
\setval{8b.gdpo.minus.concur-tied.judge2.longbench-write-abs}{10.6}
\setval{8b.gdpo.minus.concur-tied.judge2.writing}{$-$5.6}
\setval{8b.gdpo.minus.concur-tied.judge2.writing-abs}{5.6}
\setval{8b.gdpo.minus.concur-tied.judge2.writingbench}{$-$5.4}
\setval{8b.gdpo.minus.concur-tied.judge2.writingbench-abs}{5.4}
\setval{8b.gdpo.minus.concur-tied.livecodebench}{$-$1.0}
\setval{8b.gdpo.minus.concur-tied.livecodebench-abs}{1.0}
\setval{8b.gdpo.minus.concur-tied.longbench-write}{$-$11.6}
\setval{8b.gdpo.minus.concur-tied.longbench-write-abs}{11.6}
\setval{8b.gdpo.minus.concur-tied.math}{$-$9.7}
\setval{8b.gdpo.minus.concur-tied.math-abs}{9.7}
\setval{8b.gdpo.minus.concur-tied.math500}{$-$6.6}
\setval{8b.gdpo.minus.concur-tied.math500-abs}{6.6}
\setval{8b.gdpo.minus.concur-tied.mbpp-plus}{$-$0.3}
\setval{8b.gdpo.minus.concur-tied.mbpp-plus-abs}{0.3}
\setval{8b.gdpo.minus.concur-tied.minerva}{$-$5.8}
\setval{8b.gdpo.minus.concur-tied.minerva-abs}{5.8}
\setval{8b.gdpo.minus.concur-tied.olympiad-bench}{$-$10.2}
\setval{8b.gdpo.minus.concur-tied.olympiad-bench-abs}{10.2}
\setval{8b.gdpo.minus.concur-tied.writing}{$-$7.4}
\setval{8b.gdpo.minus.concur-tied.writing-abs}{7.4}
\setval{8b.gdpo.minus.concur-tied.writingbench}{$-$11.1}
\setval{8b.gdpo.minus.concur-tied.writingbench-abs}{11.1}
\setval{8b.gdpo.minus.concur.aime24}{$-$7.4}
\setval{8b.gdpo.minus.concur.aime24-abs}{7.4}
\setval{8b.gdpo.minus.concur.aime25}{$-$8.8}
\setval{8b.gdpo.minus.concur.aime25-abs}{8.8}
\setval{8b.gdpo.minus.concur.amc23}{$-$16.7}
\setval{8b.gdpo.minus.concur.amc23-abs}{16.7}
\setval{8b.gdpo.minus.concur.avg}{$-$4.8}
\setval{8b.gdpo.minus.concur.avg-abs}{4.8}
\setval{8b.gdpo.minus.concur.code}{+0.8}
\setval{8b.gdpo.minus.concur.code-abs}{0.8}
\setval{8b.gdpo.minus.concur.creative-writing-v3}{+3.0}
\setval{8b.gdpo.minus.concur.creative-writing-v3-abs}{3.0}
\setval{8b.gdpo.minus.concur.guidebench}{+1.2}
\setval{8b.gdpo.minus.concur.guidebench-abs}{1.2}
\setval{8b.gdpo.minus.concur.humaneval-plus}{+1.3}
\setval{8b.gdpo.minus.concur.humaneval-plus-abs}{1.3}
\setval{8b.gdpo.minus.concur.if}{+0.6}
\setval{8b.gdpo.minus.concur.if-abs}{0.6}
\setval{8b.gdpo.minus.concur.ifbench}{+0.7}
\setval{8b.gdpo.minus.concur.ifbench-abs}{0.7}
\setval{8b.gdpo.minus.concur.ifeval}{+0.0}
\setval{8b.gdpo.minus.concur.ifeval-abs}{0.0}
\setval{8b.gdpo.minus.concur.judge2.avg}{$-$4.3}
\setval{8b.gdpo.minus.concur.judge2.avg-abs}{4.3}
\setval{8b.gdpo.minus.concur.judge2.creative-writing-v3}{+3.3}
\setval{8b.gdpo.minus.concur.judge2.creative-writing-v3-abs}{3.3}
\setval{8b.gdpo.minus.concur.judge2.longbench-write}{$-$11.3}
\setval{8b.gdpo.minus.concur.judge2.longbench-write-abs}{11.3}
\setval{8b.gdpo.minus.concur.judge2.writing}{$-$4.5}
\setval{8b.gdpo.minus.concur.judge2.writing-abs}{4.5}
\setval{8b.gdpo.minus.concur.judge2.writingbench}{$-$5.6}
\setval{8b.gdpo.minus.concur.judge2.writingbench-abs}{5.6}
\setval{8b.gdpo.minus.concur.livecodebench}{$-$0.1}
\setval{8b.gdpo.minus.concur.livecodebench-abs}{0.1}
\setval{8b.gdpo.minus.concur.longbench-write}{$-$12.0}
\setval{8b.gdpo.minus.concur.longbench-write-abs}{12.0}
\setval{8b.gdpo.minus.concur.math}{$-$9.3}
\setval{8b.gdpo.minus.concur.math-abs}{9.3}
\setval{8b.gdpo.minus.concur.math500}{$-$7.1}
\setval{8b.gdpo.minus.concur.math500-abs}{7.1}
\setval{8b.gdpo.minus.concur.mbpp-plus}{+1.3}
\setval{8b.gdpo.minus.concur.mbpp-plus-abs}{1.3}
\setval{8b.gdpo.minus.concur.minerva}{$-$4.6}
\setval{8b.gdpo.minus.concur.minerva-abs}{4.6}
\setval{8b.gdpo.minus.concur.olympiad-bench}{$-$11.1}
\setval{8b.gdpo.minus.concur.olympiad-bench-abs}{11.1}
\setval{8b.gdpo.minus.concur.writing}{$-$7.0}
\setval{8b.gdpo.minus.concur.writing-abs}{7.0}
\setval{8b.gdpo.minus.concur.writingbench}{$-$12.0}
\setval{8b.gdpo.minus.concur.writingbench-abs}{12.0}
\setval{8b.gdpo.minus.gd2po.aime24}{+5.1}
\setval{8b.gdpo.minus.gd2po.aime24-abs}{5.1}
\setval{8b.gdpo.minus.gd2po.aime25}{+2.2}
\setval{8b.gdpo.minus.gd2po.aime25-abs}{2.2}
\setval{8b.gdpo.minus.gd2po.amc23}{+2.6}
\setval{8b.gdpo.minus.gd2po.amc23-abs}{2.6}
\setval{8b.gdpo.minus.gd2po.avg}{+0.9}
\setval{8b.gdpo.minus.gd2po.avg-abs}{0.9}
\setval{8b.gdpo.minus.gd2po.code}{$-$0.2}
\setval{8b.gdpo.minus.gd2po.code-abs}{0.2}
\setval{8b.gdpo.minus.gd2po.creative-writing-v3}{+4.7}
\setval{8b.gdpo.minus.gd2po.creative-writing-v3-abs}{4.7}
\setval{8b.gdpo.minus.gd2po.guidebench}{+0.9}
\setval{8b.gdpo.minus.gd2po.guidebench-abs}{0.9}
\setval{8b.gdpo.minus.gd2po.humaneval-plus}{+0.1}
\setval{8b.gdpo.minus.gd2po.humaneval-plus-abs}{0.1}
\setval{8b.gdpo.minus.gd2po.if}{$-$1.3}
\setval{8b.gdpo.minus.gd2po.if-abs}{1.3}
\setval{8b.gdpo.minus.gd2po.ifbench}{$-$3.7}
\setval{8b.gdpo.minus.gd2po.ifbench-abs}{3.7}
\setval{8b.gdpo.minus.gd2po.ifeval}{$-$1.2}
\setval{8b.gdpo.minus.gd2po.ifeval-abs}{1.2}
\setval{8b.gdpo.minus.gd2po.judge2.avg}{+1.1}
\setval{8b.gdpo.minus.gd2po.judge2.avg-abs}{1.1}
\setval{8b.gdpo.minus.gd2po.judge2.creative-writing-v3}{+4.3}
\setval{8b.gdpo.minus.gd2po.judge2.creative-writing-v3-abs}{4.3}
\setval{8b.gdpo.minus.gd2po.judge2.longbench-write}{+2.0}
\setval{8b.gdpo.minus.gd2po.judge2.longbench-write-abs}{2.0}
\setval{8b.gdpo.minus.gd2po.judge2.writing}{+3.1}
\setval{8b.gdpo.minus.gd2po.judge2.writing-abs}{3.1}
\setval{8b.gdpo.minus.gd2po.judge2.writingbench}{+3.1}
\setval{8b.gdpo.minus.gd2po.judge2.writingbench-abs}{3.1}
\setval{8b.gdpo.minus.gd2po.livecodebench}{$-$0.1}
\setval{8b.gdpo.minus.gd2po.livecodebench-abs}{0.1}
\setval{8b.gdpo.minus.gd2po.longbench-write}{$-$0.3}
\setval{8b.gdpo.minus.gd2po.longbench-write-abs}{0.3}
\setval{8b.gdpo.minus.gd2po.math}{+2.0}
\setval{8b.gdpo.minus.gd2po.math-abs}{2.0}
\setval{8b.gdpo.minus.gd2po.math500}{+1.8}
\setval{8b.gdpo.minus.gd2po.math500-abs}{1.8}
\setval{8b.gdpo.minus.gd2po.mbpp-plus}{$-$0.6}
\setval{8b.gdpo.minus.gd2po.mbpp-plus-abs}{0.6}
\setval{8b.gdpo.minus.gd2po.minerva}{$-$0.3}
\setval{8b.gdpo.minus.gd2po.minerva-abs}{0.3}
\setval{8b.gdpo.minus.gd2po.olympiad-bench}{+0.8}
\setval{8b.gdpo.minus.gd2po.olympiad-bench-abs}{0.8}
\setval{8b.gdpo.minus.gd2po.writing}{+2.2}
\setval{8b.gdpo.minus.gd2po.writing-abs}{2.2}
\setval{8b.gdpo.minus.gd2po.writingbench}{+2.2}
\setval{8b.gdpo.minus.gd2po.writingbench-abs}{2.2}
\setval{8b.gdpo.minus.gdpo-noaux.aime24}{$-$6.6}
\setval{8b.gdpo.minus.gdpo-noaux.aime24-abs}{6.6}
\setval{8b.gdpo.minus.gdpo-noaux.aime25}{$-$9.6}
\setval{8b.gdpo.minus.gdpo-noaux.aime25-abs}{9.6}
\setval{8b.gdpo.minus.gdpo-noaux.amc23}{$-$14.9}
\setval{8b.gdpo.minus.gdpo-noaux.amc23-abs}{14.9}
\setval{8b.gdpo.minus.gdpo-noaux.avg}{$-$2.2}
\setval{8b.gdpo.minus.gdpo-noaux.avg-abs}{2.2}
\setval{8b.gdpo.minus.gdpo-noaux.code}{$-$1.1}
\setval{8b.gdpo.minus.gdpo-noaux.code-abs}{1.1}
\setval{8b.gdpo.minus.gdpo-noaux.creative-writing-v3}{+9.0}
\setval{8b.gdpo.minus.gdpo-noaux.creative-writing-v3-abs}{9.0}
\setval{8b.gdpo.minus.gdpo-noaux.guidebench}{+3.2}
\setval{8b.gdpo.minus.gdpo-noaux.guidebench-abs}{3.2}
\setval{8b.gdpo.minus.gdpo-noaux.humaneval-plus}{$-$1.9}
\setval{8b.gdpo.minus.gdpo-noaux.humaneval-plus-abs}{1.9}
\setval{8b.gdpo.minus.gdpo-noaux.if}{+3.5}
\setval{8b.gdpo.minus.gdpo-noaux.if-abs}{3.5}
\setval{8b.gdpo.minus.gdpo-noaux.ifbench}{+5.7}
\setval{8b.gdpo.minus.gdpo-noaux.ifbench-abs}{5.7}
\setval{8b.gdpo.minus.gdpo-noaux.ifeval}{+1.6}
\setval{8b.gdpo.minus.gdpo-noaux.ifeval-abs}{1.6}
\setval{8b.gdpo.minus.gdpo-noaux.judge2.avg}{$-$1.1}
\setval{8b.gdpo.minus.gdpo-noaux.judge2.avg-abs}{1.1}
\setval{8b.gdpo.minus.gdpo-noaux.judge2.creative-writing-v3}{+12.2}
\setval{8b.gdpo.minus.gdpo-noaux.judge2.creative-writing-v3-abs}{12.2}
\setval{8b.gdpo.minus.gdpo-noaux.judge2.longbench-write}{+9.7}
\setval{8b.gdpo.minus.gdpo-noaux.judge2.longbench-write-abs}{9.7}
\setval{8b.gdpo.minus.gdpo-noaux.judge2.writing}{+10.1}
\setval{8b.gdpo.minus.gdpo-noaux.judge2.writing-abs}{10.1}
\setval{8b.gdpo.minus.gdpo-noaux.judge2.writingbench}{+8.3}
\setval{8b.gdpo.minus.gdpo-noaux.judge2.writingbench-abs}{8.3}
\setval{8b.gdpo.minus.gdpo-noaux.livecodebench}{$-$1.1}
\setval{8b.gdpo.minus.gdpo-noaux.livecodebench-abs}{1.1}
\setval{8b.gdpo.minus.gdpo-noaux.longbench-write}{+3.6}
\setval{8b.gdpo.minus.gdpo-noaux.longbench-write-abs}{3.6}
\setval{8b.gdpo.minus.gdpo-noaux.math}{$-$8.9}
\setval{8b.gdpo.minus.gdpo-noaux.math-abs}{8.9}
\setval{8b.gdpo.minus.gdpo-noaux.math500}{$-$7.2}
\setval{8b.gdpo.minus.gdpo-noaux.math500-abs}{7.2}
\setval{8b.gdpo.minus.gdpo-noaux.mbpp-plus}{$-$0.3}
\setval{8b.gdpo.minus.gdpo-noaux.mbpp-plus-abs}{0.3}
\setval{8b.gdpo.minus.gdpo-noaux.minerva}{$-$4.4}
\setval{8b.gdpo.minus.gdpo-noaux.minerva-abs}{4.4}
\setval{8b.gdpo.minus.gdpo-noaux.olympiad-bench}{$-$10.6}
\setval{8b.gdpo.minus.gdpo-noaux.olympiad-bench-abs}{10.6}
\setval{8b.gdpo.minus.gdpo-noaux.writing}{+4.6}
\setval{8b.gdpo.minus.gdpo-noaux.writing-abs}{4.6}
\setval{8b.gdpo.minus.gdpo-noaux.writingbench}{+1.1}
\setval{8b.gdpo.minus.gdpo-noaux.writingbench-abs}{1.1}
\setval{8b.gdpo.minus.gdpo-w0.5.aime24}{+2.1}
\setval{8b.gdpo.minus.gdpo-w0.5.aime24-abs}{2.1}
\setval{8b.gdpo.minus.gdpo-w0.5.aime25}{$-$1.0}
\setval{8b.gdpo.minus.gdpo-w0.5.aime25-abs}{1.0}
\setval{8b.gdpo.minus.gdpo-w0.5.amc23}{$-$1.3}
\setval{8b.gdpo.minus.gdpo-w0.5.amc23-abs}{1.3}
\setval{8b.gdpo.minus.gdpo-w0.5.avg}{+0.0}
\setval{8b.gdpo.minus.gdpo-w0.5.avg-abs}{0.0}
\setval{8b.gdpo.minus.gdpo-w0.5.code}{+0.5}
\setval{8b.gdpo.minus.gdpo-w0.5.code-abs}{0.5}
\setval{8b.gdpo.minus.gdpo-w0.5.creative-writing-v3}{+1.3}
\setval{8b.gdpo.minus.gdpo-w0.5.creative-writing-v3-abs}{1.3}
\setval{8b.gdpo.minus.gdpo-w0.5.guidebench}{+0.6}
\setval{8b.gdpo.minus.gdpo-w0.5.guidebench-abs}{0.6}
\setval{8b.gdpo.minus.gdpo-w0.5.humaneval-plus}{+0.3}
\setval{8b.gdpo.minus.gdpo-w0.5.humaneval-plus-abs}{0.3}
\setval{8b.gdpo.minus.gdpo-w0.5.if}{+0.9}
\setval{8b.gdpo.minus.gdpo-w0.5.if-abs}{0.9}
\setval{8b.gdpo.minus.gdpo-w0.5.ifbench}{+1.7}
\setval{8b.gdpo.minus.gdpo-w0.5.ifbench-abs}{1.7}
\setval{8b.gdpo.minus.gdpo-w0.5.ifeval}{+0.4}
\setval{8b.gdpo.minus.gdpo-w0.5.ifeval-abs}{0.4}
\setval{8b.gdpo.minus.gdpo-w0.5.judge2.avg}{+0.2}
\setval{8b.gdpo.minus.gdpo-w0.5.judge2.avg-abs}{0.2}
\setval{8b.gdpo.minus.gdpo-w0.5.judge2.creative-writing-v3}{+0.7}
\setval{8b.gdpo.minus.gdpo-w0.5.judge2.creative-writing-v3-abs}{0.7}
\setval{8b.gdpo.minus.gdpo-w0.5.judge2.longbench-write}{+0.5}
\setval{8b.gdpo.minus.gdpo-w0.5.judge2.longbench-write-abs}{0.5}
\setval{8b.gdpo.minus.gdpo-w0.5.judge2.writing}{+1.1}
\setval{8b.gdpo.minus.gdpo-w0.5.judge2.writing-abs}{1.1}
\setval{8b.gdpo.minus.gdpo-w0.5.judge2.writingbench}{+2.1}
\setval{8b.gdpo.minus.gdpo-w0.5.judge2.writingbench-abs}{2.1}
\setval{8b.gdpo.minus.gdpo-w0.5.livecodebench}{$-$0.6}
\setval{8b.gdpo.minus.gdpo-w0.5.livecodebench-abs}{0.6}
\setval{8b.gdpo.minus.gdpo-w0.5.longbench-write}{$-$1.6}
\setval{8b.gdpo.minus.gdpo-w0.5.longbench-write-abs}{1.6}
\setval{8b.gdpo.minus.gdpo-w0.5.math}{$-$0.8}
\setval{8b.gdpo.minus.gdpo-w0.5.math-abs}{0.8}
\setval{8b.gdpo.minus.gdpo-w0.5.math500}{$-$1.1}
\setval{8b.gdpo.minus.gdpo-w0.5.math500-abs}{1.1}
\setval{8b.gdpo.minus.gdpo-w0.5.mbpp-plus}{+1.7}
\setval{8b.gdpo.minus.gdpo-w0.5.mbpp-plus-abs}{1.7}
\setval{8b.gdpo.minus.gdpo-w0.5.minerva}{$-$1.6}
\setval{8b.gdpo.minus.gdpo-w0.5.minerva-abs}{1.6}
\setval{8b.gdpo.minus.gdpo-w0.5.olympiad-bench}{$-$1.8}
\setval{8b.gdpo.minus.gdpo-w0.5.olympiad-bench-abs}{1.8}
\setval{8b.gdpo.minus.gdpo-w0.5.writing}{+0.3}
\setval{8b.gdpo.minus.gdpo-w0.5.writing-abs}{0.3}
\setval{8b.gdpo.minus.gdpo-w0.5.writingbench}{+1.1}
\setval{8b.gdpo.minus.gdpo-w0.5.writingbench-abs}{1.1}
\setval{8b.gdpo.minus.geoma.aime24}{+7.3}
\setval{8b.gdpo.minus.geoma.aime24-abs}{7.3}
\setval{8b.gdpo.minus.geoma.aime25}{+1.9}
\setval{8b.gdpo.minus.geoma.aime25-abs}{1.9}
\setval{8b.gdpo.minus.geoma.amc23}{$-$1.3}
\setval{8b.gdpo.minus.geoma.amc23-abs}{1.3}
\setval{8b.gdpo.minus.geoma.avg}{+1.7}
\setval{8b.gdpo.minus.geoma.avg-abs}{1.7}
\setval{8b.gdpo.minus.geoma.code}{+0.3}
\setval{8b.gdpo.minus.geoma.code-abs}{0.3}
\setval{8b.gdpo.minus.geoma.creative-writing-v3}{+2.5}
\setval{8b.gdpo.minus.geoma.creative-writing-v3-abs}{2.5}
\setval{8b.gdpo.minus.geoma.guidebench}{+0.2}
\setval{8b.gdpo.minus.geoma.guidebench-abs}{0.2}
\setval{8b.gdpo.minus.geoma.humaneval-plus}{+1.3}
\setval{8b.gdpo.minus.geoma.humaneval-plus-abs}{1.3}
\setval{8b.gdpo.minus.geoma.if}{+2.1}
\setval{8b.gdpo.minus.geoma.if-abs}{2.1}
\setval{8b.gdpo.minus.geoma.ifbench}{+5.3}
\setval{8b.gdpo.minus.geoma.ifbench-abs}{5.3}
\setval{8b.gdpo.minus.geoma.ifeval}{+0.8}
\setval{8b.gdpo.minus.geoma.ifeval-abs}{0.8}
\setval{8b.gdpo.minus.geoma.judge2.avg}{+1.6}
\setval{8b.gdpo.minus.geoma.judge2.avg-abs}{1.6}
\setval{8b.gdpo.minus.geoma.judge2.creative-writing-v3}{+3.5}
\setval{8b.gdpo.minus.geoma.judge2.creative-writing-v3-abs}{3.5}
\setval{8b.gdpo.minus.geoma.judge2.longbench-write}{$-$0.7}
\setval{8b.gdpo.minus.geoma.judge2.longbench-write-abs}{0.7}
\setval{8b.gdpo.minus.geoma.judge2.writing}{+1.0}
\setval{8b.gdpo.minus.geoma.judge2.writing-abs}{1.0}
\setval{8b.gdpo.minus.geoma.judge2.writingbench}{+0.1}
\setval{8b.gdpo.minus.geoma.judge2.writingbench-abs}{0.1}
\setval{8b.gdpo.minus.geoma.livecodebench}{+0.5}
\setval{8b.gdpo.minus.geoma.livecodebench-abs}{0.5}
\setval{8b.gdpo.minus.geoma.longbench-write}{+0.6}
\setval{8b.gdpo.minus.geoma.longbench-write-abs}{0.6}
\setval{8b.gdpo.minus.geoma.math}{+2.4}
\setval{8b.gdpo.minus.geoma.math-abs}{2.4}
\setval{8b.gdpo.minus.geoma.math500}{+3.4}
\setval{8b.gdpo.minus.geoma.math500-abs}{3.4}
\setval{8b.gdpo.minus.geoma.mbpp-plus}{$-$1.1}
\setval{8b.gdpo.minus.geoma.mbpp-plus-abs}{1.1}
\setval{8b.gdpo.minus.geoma.minerva}{$-$0.2}
\setval{8b.gdpo.minus.geoma.minerva-abs}{0.2}
\setval{8b.gdpo.minus.geoma.olympiad-bench}{+3.3}
\setval{8b.gdpo.minus.geoma.olympiad-bench-abs}{3.3}
\setval{8b.gdpo.minus.geoma.writing}{+1.1}
\setval{8b.gdpo.minus.geoma.writing-abs}{1.1}
\setval{8b.gdpo.minus.geoma.writingbench}{+0.2}
\setval{8b.gdpo.minus.geoma.writingbench-abs}{0.2}
\setval{8b.gdpo.minus.grpo-nodiv.aime24}{+0.7}
\setval{8b.gdpo.minus.grpo-nodiv.aime24-abs}{0.7}
\setval{8b.gdpo.minus.grpo-nodiv.aime25}{$-$0.4}
\setval{8b.gdpo.minus.grpo-nodiv.aime25-abs}{0.4}
\setval{8b.gdpo.minus.grpo-nodiv.amc23}{$-$1.4}
\setval{8b.gdpo.minus.grpo-nodiv.amc23-abs}{1.4}
\setval{8b.gdpo.minus.grpo-nodiv.avg}{+0.3}
\setval{8b.gdpo.minus.grpo-nodiv.avg-abs}{0.3}
\setval{8b.gdpo.minus.grpo-nodiv.code}{$-$0.8}
\setval{8b.gdpo.minus.grpo-nodiv.code-abs}{0.8}
\setval{8b.gdpo.minus.grpo-nodiv.creative-writing-v3}{+6.2}
\setval{8b.gdpo.minus.grpo-nodiv.creative-writing-v3-abs}{6.2}
\setval{8b.gdpo.minus.grpo-nodiv.guidebench}{$-$0.3}
\setval{8b.gdpo.minus.grpo-nodiv.guidebench-abs}{0.3}
\setval{8b.gdpo.minus.grpo-nodiv.humaneval-plus}{$-$2.3}
\setval{8b.gdpo.minus.grpo-nodiv.humaneval-plus-abs}{2.3}
\setval{8b.gdpo.minus.grpo-nodiv.if}{+0.5}
\setval{8b.gdpo.minus.grpo-nodiv.if-abs}{0.5}
\setval{8b.gdpo.minus.grpo-nodiv.ifbench}{+3.0}
\setval{8b.gdpo.minus.grpo-nodiv.ifbench-abs}{3.0}
\setval{8b.gdpo.minus.grpo-nodiv.ifeval}{$-$1.3}
\setval{8b.gdpo.minus.grpo-nodiv.ifeval-abs}{1.3}
\setval{8b.gdpo.minus.grpo-nodiv.judge2.avg}{$-$0.1}
\setval{8b.gdpo.minus.grpo-nodiv.judge2.avg-abs}{0.1}
\setval{8b.gdpo.minus.grpo-nodiv.judge2.creative-writing-v3}{+3.5}
\setval{8b.gdpo.minus.grpo-nodiv.judge2.creative-writing-v3-abs}{3.5}
\setval{8b.gdpo.minus.grpo-nodiv.judge2.longbench-write}{$-$2.4}
\setval{8b.gdpo.minus.grpo-nodiv.judge2.longbench-write-abs}{2.4}
\setval{8b.gdpo.minus.grpo-nodiv.judge2.writing}{+1.9}
\setval{8b.gdpo.minus.grpo-nodiv.judge2.writing-abs}{1.9}
\setval{8b.gdpo.minus.grpo-nodiv.judge2.writingbench}{+4.5}
\setval{8b.gdpo.minus.grpo-nodiv.judge2.writingbench-abs}{4.5}
\setval{8b.gdpo.minus.grpo-nodiv.livecodebench}{+1.3}
\setval{8b.gdpo.minus.grpo-nodiv.livecodebench-abs}{1.3}
\setval{8b.gdpo.minus.grpo-nodiv.longbench-write}{$-$1.5}
\setval{8b.gdpo.minus.grpo-nodiv.longbench-write-abs}{1.5}
\setval{8b.gdpo.minus.grpo-nodiv.math}{$-$0.9}
\setval{8b.gdpo.minus.grpo-nodiv.math-abs}{0.9}
\setval{8b.gdpo.minus.grpo-nodiv.math500}{$-$1.1}
\setval{8b.gdpo.minus.grpo-nodiv.math500-abs}{1.1}
\setval{8b.gdpo.minus.grpo-nodiv.mbpp-plus}{$-$1.5}
\setval{8b.gdpo.minus.grpo-nodiv.mbpp-plus-abs}{1.5}
\setval{8b.gdpo.minus.grpo-nodiv.minerva}{$-$1.2}
\setval{8b.gdpo.minus.grpo-nodiv.minerva-abs}{1.2}
\setval{8b.gdpo.minus.grpo-nodiv.olympiad-bench}{$-$2.2}
\setval{8b.gdpo.minus.grpo-nodiv.olympiad-bench-abs}{2.2}
\setval{8b.gdpo.minus.grpo-nodiv.writing}{+3.5}
\setval{8b.gdpo.minus.grpo-nodiv.writing-abs}{3.5}
\setval{8b.gdpo.minus.grpo-nodiv.writingbench}{+5.8}
\setval{8b.gdpo.minus.grpo-nodiv.writingbench-abs}{5.8}
\setval{8b.gdpo.minus.grpo.aime24}{+5.5}
\setval{8b.gdpo.minus.grpo.aime24-abs}{5.5}
\setval{8b.gdpo.minus.grpo.aime25}{$-$1.1}
\setval{8b.gdpo.minus.grpo.aime25-abs}{1.1}
\setval{8b.gdpo.minus.grpo.amc23}{$-$2.2}
\setval{8b.gdpo.minus.grpo.amc23-abs}{2.2}
\setval{8b.gdpo.minus.grpo.avg}{+0.5}
\setval{8b.gdpo.minus.grpo.avg-abs}{0.5}
\setval{8b.gdpo.minus.grpo.code}{+0.0}
\setval{8b.gdpo.minus.grpo.code-abs}{0.0}
\setval{8b.gdpo.minus.grpo.creative-writing-v3}{+0.4}
\setval{8b.gdpo.minus.grpo.creative-writing-v3-abs}{0.4}
\setval{8b.gdpo.minus.grpo.guidebench}{+2.7}
\setval{8b.gdpo.minus.grpo.guidebench-abs}{2.7}
\setval{8b.gdpo.minus.grpo.humaneval-plus}{+0.4}
\setval{8b.gdpo.minus.grpo.humaneval-plus-abs}{0.4}
\setval{8b.gdpo.minus.grpo.if}{+0.4}
\setval{8b.gdpo.minus.grpo.if-abs}{0.4}
\setval{8b.gdpo.minus.grpo.ifbench}{$-$2.7}
\setval{8b.gdpo.minus.grpo.ifbench-abs}{2.7}
\setval{8b.gdpo.minus.grpo.ifeval}{+1.3}
\setval{8b.gdpo.minus.grpo.ifeval-abs}{1.3}
\setval{8b.gdpo.minus.grpo.judge2.avg}{+0.8}
\setval{8b.gdpo.minus.grpo.judge2.avg-abs}{0.8}
\setval{8b.gdpo.minus.grpo.judge2.creative-writing-v3}{+1.8}
\setval{8b.gdpo.minus.grpo.judge2.creative-writing-v3-abs}{1.8}
\setval{8b.gdpo.minus.grpo.judge2.longbench-write}{+1.7}
\setval{8b.gdpo.minus.grpo.judge2.longbench-write-abs}{1.7}
\setval{8b.gdpo.minus.grpo.judge2.writing}{+1.7}
\setval{8b.gdpo.minus.grpo.judge2.writing-abs}{1.7}
\setval{8b.gdpo.minus.grpo.judge2.writingbench}{+1.5}
\setval{8b.gdpo.minus.grpo.judge2.writingbench-abs}{1.5}
\setval{8b.gdpo.minus.grpo.livecodebench}{$-$0.3}
\setval{8b.gdpo.minus.grpo.livecodebench-abs}{0.3}
\setval{8b.gdpo.minus.grpo.longbench-write}{$-$0.2}
\setval{8b.gdpo.minus.grpo.longbench-write-abs}{0.2}
\setval{8b.gdpo.minus.grpo.math}{+0.9}
\setval{8b.gdpo.minus.grpo.math-abs}{0.9}
\setval{8b.gdpo.minus.grpo.math500}{+2.4}
\setval{8b.gdpo.minus.grpo.math500-abs}{2.4}
\setval{8b.gdpo.minus.grpo.mbpp-plus}{$-$0.1}
\setval{8b.gdpo.minus.grpo.mbpp-plus-abs}{0.1}
\setval{8b.gdpo.minus.grpo.minerva}{$-$0.3}
\setval{8b.gdpo.minus.grpo.minerva-abs}{0.3}
\setval{8b.gdpo.minus.grpo.olympiad-bench}{+1.4}
\setval{8b.gdpo.minus.grpo.olympiad-bench-abs}{1.4}
\setval{8b.gdpo.minus.grpo.writing}{$-$0.1}
\setval{8b.gdpo.minus.grpo.writing-abs}{0.1}
\setval{8b.gdpo.minus.grpo.writingbench}{$-$0.4}
\setval{8b.gdpo.minus.grpo.writingbench-abs}{0.4}
\setval{8b.gdpo.minus.mt-grpo.aime24}{+6.9}
\setval{8b.gdpo.minus.mt-grpo.aime24-abs}{6.9}
\setval{8b.gdpo.minus.mt-grpo.aime25}{$-$1.4}
\setval{8b.gdpo.minus.mt-grpo.aime25-abs}{1.4}
\setval{8b.gdpo.minus.mt-grpo.amc23}{$-$3.4}
\setval{8b.gdpo.minus.mt-grpo.amc23-abs}{3.4}
\setval{8b.gdpo.minus.mt-grpo.avg}{+1.1}
\setval{8b.gdpo.minus.mt-grpo.avg-abs}{1.1}
\setval{8b.gdpo.minus.mt-grpo.code}{$-$1.0}
\setval{8b.gdpo.minus.mt-grpo.code-abs}{1.0}
\setval{8b.gdpo.minus.mt-grpo.creative-writing-v3}{+2.6}
\setval{8b.gdpo.minus.mt-grpo.creative-writing-v3-abs}{2.6}
\setval{8b.gdpo.minus.mt-grpo.guidebench}{+1.4}
\setval{8b.gdpo.minus.mt-grpo.guidebench-abs}{1.4}
\setval{8b.gdpo.minus.mt-grpo.humaneval-plus}{$-$1.0}
\setval{8b.gdpo.minus.mt-grpo.humaneval-plus-abs}{1.0}
\setval{8b.gdpo.minus.mt-grpo.if}{+1.5}
\setval{8b.gdpo.minus.mt-grpo.if-abs}{1.5}
\setval{8b.gdpo.minus.mt-grpo.ifbench}{+2.0}
\setval{8b.gdpo.minus.mt-grpo.ifbench-abs}{2.0}
\setval{8b.gdpo.minus.mt-grpo.ifeval}{+1.2}
\setval{8b.gdpo.minus.mt-grpo.ifeval-abs}{1.2}
\setval{8b.gdpo.minus.mt-grpo.judge2.avg}{+0.8}
\setval{8b.gdpo.minus.mt-grpo.judge2.avg-abs}{0.8}
\setval{8b.gdpo.minus.mt-grpo.judge2.creative-writing-v3}{$-$0.5}
\setval{8b.gdpo.minus.mt-grpo.judge2.creative-writing-v3-abs}{0.5}
\setval{8b.gdpo.minus.mt-grpo.judge2.longbench-write}{+1.7}
\setval{8b.gdpo.minus.mt-grpo.judge2.longbench-write-abs}{1.7}
\setval{8b.gdpo.minus.mt-grpo.judge2.writing}{+1.4}
\setval{8b.gdpo.minus.mt-grpo.judge2.writing-abs}{1.4}
\setval{8b.gdpo.minus.mt-grpo.judge2.writingbench}{+2.9}
\setval{8b.gdpo.minus.mt-grpo.judge2.writingbench-abs}{2.9}
\setval{8b.gdpo.minus.mt-grpo.livecodebench}{$-$1.4}
\setval{8b.gdpo.minus.mt-grpo.livecodebench-abs}{1.4}
\setval{8b.gdpo.minus.mt-grpo.longbench-write}{+2.1}
\setval{8b.gdpo.minus.mt-grpo.longbench-write-abs}{2.1}
\setval{8b.gdpo.minus.mt-grpo.math}{+1.1}
\setval{8b.gdpo.minus.mt-grpo.math-abs}{1.1}
\setval{8b.gdpo.minus.mt-grpo.math500}{+2.4}
\setval{8b.gdpo.minus.mt-grpo.math500-abs}{2.4}
\setval{8b.gdpo.minus.mt-grpo.mbpp-plus}{$-$0.7}
\setval{8b.gdpo.minus.mt-grpo.mbpp-plus-abs}{0.7}
\setval{8b.gdpo.minus.mt-grpo.minerva}{$-$1.4}
\setval{8b.gdpo.minus.mt-grpo.minerva-abs}{1.4}
\setval{8b.gdpo.minus.mt-grpo.olympiad-bench}{+3.6}
\setval{8b.gdpo.minus.mt-grpo.olympiad-bench-abs}{3.6}
\setval{8b.gdpo.minus.mt-grpo.writing}{+3.0}
\setval{8b.gdpo.minus.mt-grpo.writing-abs}{3.0}
\setval{8b.gdpo.minus.mt-grpo.writingbench}{+4.4}
\setval{8b.gdpo.minus.mt-grpo.writingbench-abs}{4.4}
\setval{8b.gdpo.minus.papo.aime24}{+9.2}
\setval{8b.gdpo.minus.papo.aime24-abs}{9.2}
\setval{8b.gdpo.minus.papo.aime25}{+4.3}
\setval{8b.gdpo.minus.papo.aime25-abs}{4.3}
\setval{8b.gdpo.minus.papo.amc23}{+0.2}
\setval{8b.gdpo.minus.papo.amc23-abs}{0.2}
\setval{8b.gdpo.minus.papo.avg}{+2.5}
\setval{8b.gdpo.minus.papo.avg-abs}{2.5}
\setval{8b.gdpo.minus.papo.code}{+0.8}
\setval{8b.gdpo.minus.papo.code-abs}{0.8}
\setval{8b.gdpo.minus.papo.creative-writing-v3}{+1.8}
\setval{8b.gdpo.minus.papo.creative-writing-v3-abs}{1.8}
\setval{8b.gdpo.minus.papo.guidebench}{+0.2}
\setval{8b.gdpo.minus.papo.guidebench-abs}{0.2}
\setval{8b.gdpo.minus.papo.humaneval-plus}{+1.6}
\setval{8b.gdpo.minus.papo.humaneval-plus-abs}{1.6}
\setval{8b.gdpo.minus.papo.if}{+1.3}
\setval{8b.gdpo.minus.papo.if-abs}{1.3}
\setval{8b.gdpo.minus.papo.ifbench}{+2.0}
\setval{8b.gdpo.minus.papo.ifbench-abs}{2.0}
\setval{8b.gdpo.minus.papo.ifeval}{+1.8}
\setval{8b.gdpo.minus.papo.ifeval-abs}{1.8}
\setval{8b.gdpo.minus.papo.judge2.avg}{+2.3}
\setval{8b.gdpo.minus.papo.judge2.avg-abs}{2.3}
\setval{8b.gdpo.minus.papo.judge2.creative-writing-v3}{$-$0.8}
\setval{8b.gdpo.minus.papo.judge2.creative-writing-v3-abs}{0.8}
\setval{8b.gdpo.minus.papo.judge2.longbench-write}{+4.6}
\setval{8b.gdpo.minus.papo.judge2.longbench-write-abs}{4.6}
\setval{8b.gdpo.minus.papo.judge2.writing}{+2.0}
\setval{8b.gdpo.minus.papo.judge2.writing-abs}{2.0}
\setval{8b.gdpo.minus.papo.judge2.writingbench}{+2.1}
\setval{8b.gdpo.minus.papo.judge2.writingbench-abs}{2.1}
\setval{8b.gdpo.minus.papo.livecodebench}{$-$0.4}
\setval{8b.gdpo.minus.papo.livecodebench-abs}{0.4}
\setval{8b.gdpo.minus.papo.longbench-write}{+4.1}
\setval{8b.gdpo.minus.papo.longbench-write-abs}{4.1}
\setval{8b.gdpo.minus.papo.math}{+3.8}
\setval{8b.gdpo.minus.papo.math-abs}{3.8}
\setval{8b.gdpo.minus.papo.math500}{+4.5}
\setval{8b.gdpo.minus.papo.math500-abs}{4.5}
\setval{8b.gdpo.minus.papo.mbpp-plus}{+1.1}
\setval{8b.gdpo.minus.papo.mbpp-plus-abs}{1.1}
\setval{8b.gdpo.minus.papo.minerva}{+0.3}
\setval{8b.gdpo.minus.papo.minerva-abs}{0.3}
\setval{8b.gdpo.minus.papo.olympiad-bench}{+4.5}
\setval{8b.gdpo.minus.papo.olympiad-bench-abs}{4.5}
\setval{8b.gdpo.minus.papo.writing}{+2.9}
\setval{8b.gdpo.minus.papo.writing-abs}{2.9}
\setval{8b.gdpo.minus.papo.writingbench}{+2.9}
\setval{8b.gdpo.minus.papo.writingbench-abs}{2.9}
\setval{8b.gdpo.minus.saw.aime24}{+6.0}
\setval{8b.gdpo.minus.saw.aime24-abs}{6.0}
\setval{8b.gdpo.minus.saw.aime25}{$-$1.1}
\setval{8b.gdpo.minus.saw.aime25-abs}{1.1}
\setval{8b.gdpo.minus.saw.amc23}{$-$0.1}
\setval{8b.gdpo.minus.saw.amc23-abs}{0.1}
\setval{8b.gdpo.minus.saw.avg}{+0.4}
\setval{8b.gdpo.minus.saw.avg-abs}{0.4}
\setval{8b.gdpo.minus.saw.code}{$-$0.7}
\setval{8b.gdpo.minus.saw.code-abs}{0.7}
\setval{8b.gdpo.minus.saw.creative-writing-v3}{+3.2}
\setval{8b.gdpo.minus.saw.creative-writing-v3-abs}{3.2}
\setval{8b.gdpo.minus.saw.guidebench}{+0.3}
\setval{8b.gdpo.minus.saw.guidebench-abs}{0.3}
\setval{8b.gdpo.minus.saw.humaneval-plus}{$-$1.8}
\setval{8b.gdpo.minus.saw.humaneval-plus-abs}{1.8}
\setval{8b.gdpo.minus.saw.if}{+0.0}
\setval{8b.gdpo.minus.saw.if-abs}{0.0}
\setval{8b.gdpo.minus.saw.ifbench}{$-$0.7}
\setval{8b.gdpo.minus.saw.ifbench-abs}{0.7}
\setval{8b.gdpo.minus.saw.ifeval}{+0.5}
\setval{8b.gdpo.minus.saw.ifeval-abs}{0.5}
\setval{8b.gdpo.minus.saw.judge2.avg}{+0.6}
\setval{8b.gdpo.minus.saw.judge2.avg-abs}{0.6}
\setval{8b.gdpo.minus.saw.judge2.creative-writing-v3}{+4.5}
\setval{8b.gdpo.minus.saw.judge2.creative-writing-v3-abs}{4.5}
\setval{8b.gdpo.minus.saw.judge2.longbench-write}{$-$0.6}
\setval{8b.gdpo.minus.saw.judge2.longbench-write-abs}{0.6}
\setval{8b.gdpo.minus.saw.judge2.writing}{+1.8}
\setval{8b.gdpo.minus.saw.judge2.writing-abs}{1.8}
\setval{8b.gdpo.minus.saw.judge2.writingbench}{+1.4}
\setval{8b.gdpo.minus.saw.judge2.writingbench-abs}{1.4}
\setval{8b.gdpo.minus.saw.livecodebench}{$-$0.5}
\setval{8b.gdpo.minus.saw.livecodebench-abs}{0.5}
\setval{8b.gdpo.minus.saw.longbench-write}{$-$1.5}
\setval{8b.gdpo.minus.saw.longbench-write-abs}{1.5}
\setval{8b.gdpo.minus.saw.math}{+0.8}
\setval{8b.gdpo.minus.saw.math-abs}{0.8}
\setval{8b.gdpo.minus.saw.math500}{+1.1}
\setval{8b.gdpo.minus.saw.math500-abs}{1.1}
\setval{8b.gdpo.minus.saw.mbpp-plus}{+0.2}
\setval{8b.gdpo.minus.saw.mbpp-plus-abs}{0.2}
\setval{8b.gdpo.minus.saw.minerva}{$-$2.2}
\setval{8b.gdpo.minus.saw.minerva-abs}{2.2}
\setval{8b.gdpo.minus.saw.olympiad-bench}{+1.2}
\setval{8b.gdpo.minus.saw.olympiad-bench-abs}{1.2}
\setval{8b.gdpo.minus.saw.writing}{+0.8}
\setval{8b.gdpo.minus.saw.writing-abs}{0.8}
\setval{8b.gdpo.minus.saw.writingbench}{+0.7}
\setval{8b.gdpo.minus.saw.writingbench-abs}{0.7}
\setval{8b.gdpo.passing-at-ceiling.code.early}{0.71}
\setval{8b.gdpo.passing-at-ceiling.code.late}{0.75}
\setval{8b.gdpo.passing-at-ceiling.code.run}{0.74}
\setval{8b.gdpo.passing-at-ceiling.if.early}{0.32}
\setval{8b.gdpo.passing-at-ceiling.if.late}{0.57}
\setval{8b.gdpo.passing-at-ceiling.if.run}{0.49}
\setval{8b.gdpo.passing-at-ceiling.math.early}{1.00}
\setval{8b.gdpo.passing-at-ceiling.math.late}{1.00}
\setval{8b.gdpo.passing-at-ceiling.math.run}{1.00}
\setval{8b.gdpo.passing-at-ceiling.writing.early}{0.00}
\setval{8b.gdpo.passing-at-ceiling.writing.late}{0.01}
\setval{8b.gdpo.passing-at-ceiling.writing.run}{0.00}
\setval{8b.gdpo.primary-cosine.code.early}{0.96}
\setval{8b.gdpo.primary-cosine.code.late}{0.95}
\setval{8b.gdpo.primary-cosine.code.run}{0.95}
\setval{8b.gdpo.primary-cosine.if.early}{0.84}
\setval{8b.gdpo.primary-cosine.if.late}{0.79}
\setval{8b.gdpo.primary-cosine.if.run}{0.82}
\setval{8b.gdpo.primary-cosine.math.early}{0.96}
\setval{8b.gdpo.primary-cosine.math.late}{0.97}
\setval{8b.gdpo.primary-cosine.math.run}{0.97}
\setval{8b.gdpo.primary-cosine.writing.early}{0.87}
\setval{8b.gdpo.primary-cosine.writing.late}{0.76}
\setval{8b.gdpo.primary-cosine.writing.run}{0.75}
\setval{8b.gdpo.primary-level.code.early}{0.54}
\setval{8b.gdpo.primary-level.code.late}{0.77}
\setval{8b.gdpo.primary-level.code.run}{0.72}
\setval{8b.gdpo.primary-level.if.early}{0.35}
\setval{8b.gdpo.primary-level.if.late}{0.77}
\setval{8b.gdpo.primary-level.if.run}{0.66}
\setval{8b.gdpo.primary-level.math.early}{0.49}
\setval{8b.gdpo.primary-level.math.late}{0.71}
\setval{8b.gdpo.primary-level.math.run}{0.64}
\setval{8b.gdpo.primary-level.writing.early}{0.53}
\setval{8b.gdpo.primary-level.writing.late}{0.76}
\setval{8b.gdpo.primary-level.writing.run}{0.72}
\setval{8b.gdpo.solved.code.early}{0.07}
\setval{8b.gdpo.solved.code.late}{0.03}
\setval{8b.gdpo.solved.code.run}{0.04}
\setval{8b.gdpo.solved.if.early}{0.03}
\setval{8b.gdpo.solved.if.late}{0.20}
\setval{8b.gdpo.solved.if.run}{0.13}
\setval{8b.gdpo.solved.math.early}{0.15}
\setval{8b.gdpo.solved.math.late}{0.15}
\setval{8b.gdpo.solved.math.run}{0.15}
\setval{8b.gdpo.solved.writing.early}{0.00}
\setval{8b.gdpo.solved.writing.late}{0.00}
\setval{8b.gdpo.solved.writing.run}{0.00}
\setval{8b.gdpo.update-share.code.early}{0.20}
\setval{8b.gdpo.update-share.code.late}{0.22}
\setval{8b.gdpo.update-share.code.run}{0.22}
\setval{8b.gdpo.update-share.if.early}{0.21}
\setval{8b.gdpo.update-share.if.late}{0.16}
\setval{8b.gdpo.update-share.if.run}{0.19}
\setval{8b.gdpo.update-share.math.early}{0.18}
\setval{8b.gdpo.update-share.math.late}{0.28}
\setval{8b.gdpo.update-share.math.run}{0.25}
\setval{8b.gdpo.update-share.writing.early}{0.41}
\setval{8b.gdpo.update-share.writing.late}{0.34}
\setval{8b.gdpo.update-share.writing.run}{0.33}
\setval{8b.geoma.aux-change.code}{+0.18}
\setval{8b.geoma.aux-change.if}{+0.18}
\setval{8b.geoma.aux-change.math}{+0.15}
\setval{8b.geoma.aux-change.writing}{+0.14}
\setval{8b.geoma.aux-change.writing-rm}{+0.33}
\setval{8b.geoma.aux-full.code.early}{0.75}
\setval{8b.geoma.aux-full.code.late}{0.97}
\setval{8b.geoma.aux-full.code.run}{0.95}
\setval{8b.geoma.aux-full.if.early}{0.73}
\setval{8b.geoma.aux-full.if.late}{0.94}
\setval{8b.geoma.aux-full.if.run}{0.91}
\setval{8b.geoma.aux-full.math.early}{0.60}
\setval{8b.geoma.aux-full.math.late}{0.96}
\setval{8b.geoma.aux-full.math.run}{0.90}
\setval{8b.geoma.aux-full.writing-rm.early}{0.07}
\setval{8b.geoma.aux-full.writing-rm.late}{0.64}
\setval{8b.geoma.aux-full.writing-rm.run}{0.52}
\setval{8b.geoma.aux-full.writing.early}{0.68}
\setval{8b.geoma.aux-full.writing.late}{0.83}
\setval{8b.geoma.aux-full.writing.run}{0.75}
\setval{8b.geoma.aux-level.code.early}{0.94}
\setval{8b.geoma.aux-level.code.late}{0.99}
\setval{8b.geoma.aux-level.code.run}{0.99}
\setval{8b.geoma.aux-level.if.early}{0.88}
\setval{8b.geoma.aux-level.if.late}{0.97}
\setval{8b.geoma.aux-level.if.run}{0.95}
\setval{8b.geoma.aux-level.math.early}{0.92}
\setval{8b.geoma.aux-level.math.late}{0.99}
\setval{8b.geoma.aux-level.math.run}{0.98}
\setval{8b.geoma.aux-level.writing-rm.early}{0.77}
\setval{8b.geoma.aux-level.writing-rm.late}{0.97}
\setval{8b.geoma.aux-level.writing-rm.run}{0.94}
\setval{8b.geoma.aux-level.writing.early}{0.92}
\setval{8b.geoma.aux-level.writing.late}{0.98}
\setval{8b.geoma.aux-level.writing.run}{0.96}
\setval{8b.geoma.bench.aime24}{14.1}
\setval{8b.geoma.bench.aime25}{12.5}
\setval{8b.geoma.bench.amc23}{63.5}
\setval{8b.geoma.bench.creative-writing-v3}{50.4}
\setval{8b.geoma.bench.guidebench}{84.4}
\setval{8b.geoma.bench.humaneval-plus}{82.7}
\setval{8b.geoma.bench.ifbench}{34.0}
\setval{8b.geoma.bench.ifeval}{87.3}
\setval{8b.geoma.bench.livecodebench}{25.7}
\setval{8b.geoma.bench.longbench-write}{69.0}
\setval{8b.geoma.bench.math500}{79.6}
\setval{8b.geoma.bench.mbpp-plus}{75.6}
\setval{8b.geoma.bench.minerva}{49.8}
\setval{8b.geoma.bench.olympiad-bench}{46.0}
\setval{8b.geoma.bench.writingbench}{75.9}
\setval{8b.geoma.eval.avg}{56.7}
\setval{8b.geoma.eval.code}{61.4}
\setval{8b.geoma.eval.if}{68.5}
\setval{8b.geoma.eval.math}{44.2}
\setval{8b.geoma.eval.writing}{65.1}
\setval{8b.geoma.gate.code.against.early}{0.03}
\setval{8b.geoma.gate.code.against.late}{0.02}
\setval{8b.geoma.gate.code.against.run}{0.02}
\setval{8b.geoma.gate.code.agreement}{+0.004}
\setval{8b.geoma.gate.code.ceiling.early}{0.10}
\setval{8b.geoma.gate.code.ceiling.late}{0.03}
\setval{8b.geoma.gate.code.ceiling.run}{0.03}
\setval{8b.geoma.gate.code.cosine.early}{0.85}
\setval{8b.geoma.gate.code.cosine.late}{0.78}
\setval{8b.geoma.gate.code.cosine.run}{0.82}
\setval{8b.geoma.gate.if.against.early}{0.04}
\setval{8b.geoma.gate.if.against.late}{0.04}
\setval{8b.geoma.gate.if.against.run}{0.03}
\setval{8b.geoma.gate.if.agreement}{+0.011}
\setval{8b.geoma.gate.if.ceiling.early}{0.05}
\setval{8b.geoma.gate.if.ceiling.late}{0.12}
\setval{8b.geoma.gate.if.ceiling.run}{0.08}
\setval{8b.geoma.gate.if.cosine.early}{0.76}
\setval{8b.geoma.gate.if.cosine.late}{0.55}
\setval{8b.geoma.gate.if.cosine.run}{0.63}
\setval{8b.geoma.gate.math.against.early}{0.00}
\setval{8b.geoma.gate.math.against.late}{0.00}
\setval{8b.geoma.gate.math.against.run}{0.00}
\setval{8b.geoma.gate.math.ceiling.early}{0.21}
\setval{8b.geoma.gate.math.ceiling.late}{0.08}
\setval{8b.geoma.gate.math.ceiling.run}{0.11}
\setval{8b.geoma.gate.math.cosine.early}{0.87}
\setval{8b.geoma.gate.math.cosine.late}{0.95}
\setval{8b.geoma.gate.math.cosine.run}{0.94}
\setval{8b.geoma.gate.writing-rm.against.early}{0.19}
\setval{8b.geoma.gate.writing-rm.against.late}{0.54}
\setval{8b.geoma.gate.writing-rm.against.run}{0.47}
\setval{8b.geoma.gate.writing-rm.agreement}{+0.051}
\setval{8b.geoma.gate.writing-rm.ceiling.early}{0.00}
\setval{8b.geoma.gate.writing-rm.ceiling.late}{0.00}
\setval{8b.geoma.gate.writing-rm.ceiling.run}{0.00}
\setval{8b.geoma.gate.writing-rm.cosine.early}{0.71}
\setval{8b.geoma.gate.writing-rm.cosine.late}{0.11}
\setval{8b.geoma.gate.writing-rm.cosine.run}{0.22}
\setval{8b.geoma.gate.writing.against.early}{0.25}
\setval{8b.geoma.gate.writing.against.late}{0.55}
\setval{8b.geoma.gate.writing.against.run}{0.53}
\setval{8b.geoma.gate.writing.agreement}{$-$0.031}
\setval{8b.geoma.gate.writing.ceiling.early}{0.00}
\setval{8b.geoma.gate.writing.ceiling.late}{0.00}
\setval{8b.geoma.gate.writing.ceiling.run}{0.00}
\setval{8b.geoma.gate.writing.cosine.early}{0.56}
\setval{8b.geoma.gate.writing.cosine.late}{0.07}
\setval{8b.geoma.gate.writing.cosine.run}{0.13}
\setval{8b.geoma.judge2.avg}{54.4}
\setval{8b.geoma.judge2.creative-writing-v3}{37.5}
\setval{8b.geoma.judge2.longbench-write}{64.9}
\setval{8b.geoma.judge2.writing}{53.6}
\setval{8b.geoma.judge2.writingbench}{58.3}
\setval{8b.geoma.last-step}{300}
\setval{8b.geoma.minus.base.aime24}{+2.7}
\setval{8b.geoma.minus.base.aime24-abs}{2.7}
\setval{8b.geoma.minus.base.aime25}{+4.0}
\setval{8b.geoma.minus.base.aime25-abs}{4.0}
\setval{8b.geoma.minus.base.amc23}{+13.3}
\setval{8b.geoma.minus.base.amc23-abs}{13.3}
\setval{8b.geoma.minus.base.avg}{+15.2}
\setval{8b.geoma.minus.base.avg-abs}{15.2}
\setval{8b.geoma.minus.base.code}{+5.3}
\setval{8b.geoma.minus.base.code-abs}{5.3}
\setval{8b.geoma.minus.base.creative-writing-v3}{+23.7}
\setval{8b.geoma.minus.base.creative-writing-v3-abs}{23.7}
\setval{8b.geoma.minus.base.guidebench}{+38.8}
\setval{8b.geoma.minus.base.guidebench-abs}{38.8}
\setval{8b.geoma.minus.base.humaneval-plus}{+2.7}
\setval{8b.geoma.minus.base.humaneval-plus-abs}{2.7}
\setval{8b.geoma.minus.base.if}{+29.9}
\setval{8b.geoma.minus.base.if-abs}{29.9}
\setval{8b.geoma.minus.base.ifbench}{+16.7}
\setval{8b.geoma.minus.base.ifbench-abs}{16.7}
\setval{8b.geoma.minus.base.ifeval}{+34.3}
\setval{8b.geoma.minus.base.ifeval-abs}{34.3}
\setval{8b.geoma.minus.base.judge2.avg}{+13.4}
\setval{8b.geoma.minus.base.judge2.avg-abs}{13.4}
\setval{8b.geoma.minus.base.judge2.creative-writing-v3}{+11.1}
\setval{8b.geoma.minus.base.judge2.creative-writing-v3-abs}{11.1}
\setval{8b.geoma.minus.base.judge2.longbench-write}{+16.8}
\setval{8b.geoma.minus.base.judge2.longbench-write-abs}{16.8}
\setval{8b.geoma.minus.base.judge2.writing}{+17.4}
\setval{8b.geoma.minus.base.judge2.writing-abs}{17.4}
\setval{8b.geoma.minus.base.judge2.writingbench}{+24.3}
\setval{8b.geoma.minus.base.judge2.writingbench-abs}{24.3}
\setval{8b.geoma.minus.base.livecodebench}{+4.8}
\setval{8b.geoma.minus.base.livecodebench-abs}{4.8}
\setval{8b.geoma.minus.base.longbench-write}{+17.5}
\setval{8b.geoma.minus.base.longbench-write-abs}{17.5}
\setval{8b.geoma.minus.base.math}{+7.1}
\setval{8b.geoma.minus.base.math-abs}{7.1}
\setval{8b.geoma.minus.base.math500}{+7.2}
\setval{8b.geoma.minus.base.math500-abs}{7.2}
\setval{8b.geoma.minus.base.mbpp-plus}{+8.3}
\setval{8b.geoma.minus.base.mbpp-plus-abs}{8.3}
\setval{8b.geoma.minus.base.minerva}{+8.9}
\setval{8b.geoma.minus.base.minerva-abs}{8.9}
\setval{8b.geoma.minus.base.olympiad-bench}{+6.6}
\setval{8b.geoma.minus.base.olympiad-bench-abs}{6.6}
\setval{8b.geoma.minus.base.writing}{+26.8}
\setval{8b.geoma.minus.base.writing-abs}{26.8}
\setval{8b.geoma.minus.base.writingbench}{+39.2}
\setval{8b.geoma.minus.base.writingbench-abs}{39.2}
\setval{8b.geoma.minus.concur-free.aime24}{$-$5.6}
\setval{8b.geoma.minus.concur-free.aime24-abs}{5.6}
\setval{8b.geoma.minus.concur-free.aime25}{$-$4.1}
\setval{8b.geoma.minus.concur-free.aime25-abs}{4.1}
\setval{8b.geoma.minus.concur-free.amc23}{$-$1.2}
\setval{8b.geoma.minus.concur-free.amc23-abs}{1.2}
\setval{8b.geoma.minus.concur-free.avg}{$-$3.6}
\setval{8b.geoma.minus.concur-free.avg-abs}{3.6}
\setval{8b.geoma.minus.concur-free.code}{$-$1.1}
\setval{8b.geoma.minus.concur-free.code-abs}{1.1}
\setval{8b.geoma.minus.concur-free.creative-writing-v3}{$-$1.2}
\setval{8b.geoma.minus.concur-free.creative-writing-v3-abs}{1.2}
\setval{8b.geoma.minus.concur-free.guidebench}{+1.4}
\setval{8b.geoma.minus.concur-free.guidebench-abs}{1.4}
\setval{8b.geoma.minus.concur-free.humaneval-plus}{$-$2.6}
\setval{8b.geoma.minus.concur-free.humaneval-plus-abs}{2.6}
\setval{8b.geoma.minus.concur-free.if}{$-$1.7}
\setval{8b.geoma.minus.concur-free.if-abs}{1.7}
\setval{8b.geoma.minus.concur-free.ifbench}{$-$4.7}
\setval{8b.geoma.minus.concur-free.ifbench-abs}{4.7}
\setval{8b.geoma.minus.concur-free.ifeval}{$-$1.9}
\setval{8b.geoma.minus.concur-free.ifeval-abs}{1.9}
\setval{8b.geoma.minus.concur-free.judge2.avg}{$-$2.9}
\setval{8b.geoma.minus.concur-free.judge2.avg-abs}{2.9}
\setval{8b.geoma.minus.concur-free.judge2.creative-writing-v3}{$-$5.0}
\setval{8b.geoma.minus.concur-free.judge2.creative-writing-v3-abs}{5.0}
\setval{8b.geoma.minus.concur-free.judge2.longbench-write}{$-$9.2}
\setval{8b.geoma.minus.concur-free.judge2.longbench-write-abs}{9.2}
\setval{8b.geoma.minus.concur-free.judge2.writing}{$-$5.5}
\setval{8b.geoma.minus.concur-free.judge2.writing-abs}{5.5}
\setval{8b.geoma.minus.concur-free.judge2.writingbench}{$-$2.4}
\setval{8b.geoma.minus.concur-free.judge2.writingbench-abs}{2.4}
\setval{8b.geoma.minus.concur-free.livecodebench}{$-$1.4}
\setval{8b.geoma.minus.concur-free.livecodebench-abs}{1.4}
\setval{8b.geoma.minus.concur-free.longbench-write}{$-$14.0}
\setval{8b.geoma.minus.concur-free.longbench-write-abs}{14.0}
\setval{8b.geoma.minus.concur-free.math}{$-$3.1}
\setval{8b.geoma.minus.concur-free.math-abs}{3.1}
\setval{8b.geoma.minus.concur-free.math500}{$-$3.8}
\setval{8b.geoma.minus.concur-free.math500-abs}{3.8}
\setval{8b.geoma.minus.concur-free.mbpp-plus}{+0.5}
\setval{8b.geoma.minus.concur-free.mbpp-plus-abs}{0.5}
\setval{8b.geoma.minus.concur-free.minerva}{+0.0}
\setval{8b.geoma.minus.concur-free.minerva-abs}{0.0}
\setval{8b.geoma.minus.concur-free.olympiad-bench}{$-$4.1}
\setval{8b.geoma.minus.concur-free.olympiad-bench-abs}{4.1}
\setval{8b.geoma.minus.concur-free.writing}{$-$8.7}
\setval{8b.geoma.minus.concur-free.writing-abs}{8.7}
\setval{8b.geoma.minus.concur-free.writingbench}{$-$10.8}
\setval{8b.geoma.minus.concur-free.writingbench-abs}{10.8}
\setval{8b.geoma.minus.concur-tied.aime24}{$-$17.2}
\setval{8b.geoma.minus.concur-tied.aime24-abs}{17.2}
\setval{8b.geoma.minus.concur-tied.aime25}{$-$11.5}
\setval{8b.geoma.minus.concur-tied.aime25-abs}{11.5}
\setval{8b.geoma.minus.concur-tied.amc23}{$-$15.0}
\setval{8b.geoma.minus.concur-tied.amc23-abs}{15.0}
\setval{8b.geoma.minus.concur-tied.avg}{$-$7.3}
\setval{8b.geoma.minus.concur-tied.avg-abs}{7.3}
\setval{8b.geoma.minus.concur-tied.code}{$-$1.8}
\setval{8b.geoma.minus.concur-tied.code-abs}{1.8}
\setval{8b.geoma.minus.concur-tied.creative-writing-v3}{$-$2.0}
\setval{8b.geoma.minus.concur-tied.creative-writing-v3-abs}{2.0}
\setval{8b.geoma.minus.concur-tied.guidebench}{+1.0}
\setval{8b.geoma.minus.concur-tied.guidebench-abs}{1.0}
\setval{8b.geoma.minus.concur-tied.humaneval-plus}{$-$4.6}
\setval{8b.geoma.minus.concur-tied.humaneval-plus-abs}{4.6}
\setval{8b.geoma.minus.concur-tied.if}{$-$2.0}
\setval{8b.geoma.minus.concur-tied.if-abs}{2.0}
\setval{8b.geoma.minus.concur-tied.ifbench}{$-$5.3}
\setval{8b.geoma.minus.concur-tied.ifbench-abs}{5.3}
\setval{8b.geoma.minus.concur-tied.ifeval}{$-$1.6}
\setval{8b.geoma.minus.concur-tied.ifeval-abs}{1.6}
\setval{8b.geoma.minus.concur-tied.judge2.avg}{$-$6.9}
\setval{8b.geoma.minus.concur-tied.judge2.avg-abs}{6.9}
\setval{8b.geoma.minus.concur-tied.judge2.creative-writing-v3}{$-$4.2}
\setval{8b.geoma.minus.concur-tied.judge2.creative-writing-v3-abs}{4.2}
\setval{8b.geoma.minus.concur-tied.judge2.longbench-write}{$-$9.9}
\setval{8b.geoma.minus.concur-tied.judge2.longbench-write-abs}{9.9}
\setval{8b.geoma.minus.concur-tied.judge2.writing}{$-$6.5}
\setval{8b.geoma.minus.concur-tied.judge2.writing-abs}{6.5}
\setval{8b.geoma.minus.concur-tied.judge2.writingbench}{$-$5.5}
\setval{8b.geoma.minus.concur-tied.judge2.writingbench-abs}{5.5}
\setval{8b.geoma.minus.concur-tied.livecodebench}{$-$1.5}
\setval{8b.geoma.minus.concur-tied.livecodebench-abs}{1.5}
\setval{8b.geoma.minus.concur-tied.longbench-write}{$-$12.3}
\setval{8b.geoma.minus.concur-tied.longbench-write-abs}{12.3}
\setval{8b.geoma.minus.concur-tied.math}{$-$12.1}
\setval{8b.geoma.minus.concur-tied.math-abs}{12.1}
\setval{8b.geoma.minus.concur-tied.math500}{$-$10.0}
\setval{8b.geoma.minus.concur-tied.math500-abs}{10.0}
\setval{8b.geoma.minus.concur-tied.mbpp-plus}{+0.7}
\setval{8b.geoma.minus.concur-tied.mbpp-plus-abs}{0.7}
\setval{8b.geoma.minus.concur-tied.minerva}{$-$5.6}
\setval{8b.geoma.minus.concur-tied.minerva-abs}{5.6}
\setval{8b.geoma.minus.concur-tied.olympiad-bench}{$-$13.6}
\setval{8b.geoma.minus.concur-tied.olympiad-bench-abs}{13.6}
\setval{8b.geoma.minus.concur-tied.writing}{$-$8.5}
\setval{8b.geoma.minus.concur-tied.writing-abs}{8.5}
\setval{8b.geoma.minus.concur-tied.writingbench}{$-$11.3}
\setval{8b.geoma.minus.concur-tied.writingbench-abs}{11.3}
\setval{8b.geoma.minus.concur.aime24}{$-$14.7}
\setval{8b.geoma.minus.concur.aime24-abs}{14.7}
\setval{8b.geoma.minus.concur.aime25}{$-$10.6}
\setval{8b.geoma.minus.concur.aime25-abs}{10.6}
\setval{8b.geoma.minus.concur.amc23}{$-$15.4}
\setval{8b.geoma.minus.concur.amc23-abs}{15.4}
\setval{8b.geoma.minus.concur.avg}{$-$6.5}
\setval{8b.geoma.minus.concur.avg-abs}{6.5}
\setval{8b.geoma.minus.concur.code}{+0.6}
\setval{8b.geoma.minus.concur.code-abs}{0.6}
\setval{8b.geoma.minus.concur.creative-writing-v3}{+0.5}
\setval{8b.geoma.minus.concur.creative-writing-v3-abs}{0.5}
\setval{8b.geoma.minus.concur.guidebench}{+1.0}
\setval{8b.geoma.minus.concur.guidebench-abs}{1.0}
\setval{8b.geoma.minus.concur.humaneval-plus}{+0.0}
\setval{8b.geoma.minus.concur.humaneval-plus-abs}{0.0}
\setval{8b.geoma.minus.concur.if}{$-$1.5}
\setval{8b.geoma.minus.concur.if-abs}{1.5}
\setval{8b.geoma.minus.concur.ifbench}{$-$4.7}
\setval{8b.geoma.minus.concur.ifbench-abs}{4.7}
\setval{8b.geoma.minus.concur.ifeval}{$-$0.8}
\setval{8b.geoma.minus.concur.ifeval-abs}{0.8}
\setval{8b.geoma.minus.concur.judge2.avg}{$-$6.0}
\setval{8b.geoma.minus.concur.judge2.avg-abs}{6.0}
\setval{8b.geoma.minus.concur.judge2.creative-writing-v3}{$-$0.2}
\setval{8b.geoma.minus.concur.judge2.creative-writing-v3-abs}{0.2}
\setval{8b.geoma.minus.concur.judge2.longbench-write}{$-$10.6}
\setval{8b.geoma.minus.concur.judge2.longbench-write-abs}{10.6}
\setval{8b.geoma.minus.concur.judge2.writing}{$-$5.5}
\setval{8b.geoma.minus.concur.judge2.writing-abs}{5.5}
\setval{8b.geoma.minus.concur.judge2.writingbench}{$-$5.6}
\setval{8b.geoma.minus.concur.judge2.writingbench-abs}{5.6}
\setval{8b.geoma.minus.concur.livecodebench}{$-$0.6}
\setval{8b.geoma.minus.concur.livecodebench-abs}{0.6}
\setval{8b.geoma.minus.concur.longbench-write}{$-$12.6}
\setval{8b.geoma.minus.concur.longbench-write-abs}{12.6}
\setval{8b.geoma.minus.concur.math}{$-$11.7}
\setval{8b.geoma.minus.concur.math-abs}{11.7}
\setval{8b.geoma.minus.concur.math500}{$-$10.5}
\setval{8b.geoma.minus.concur.math500-abs}{10.5}
\setval{8b.geoma.minus.concur.mbpp-plus}{+2.4}
\setval{8b.geoma.minus.concur.mbpp-plus-abs}{2.4}
\setval{8b.geoma.minus.concur.minerva}{$-$4.4}
\setval{8b.geoma.minus.concur.minerva-abs}{4.4}
\setval{8b.geoma.minus.concur.olympiad-bench}{$-$14.4}
\setval{8b.geoma.minus.concur.olympiad-bench-abs}{14.4}
\setval{8b.geoma.minus.concur.writing}{$-$8.1}
\setval{8b.geoma.minus.concur.writing-abs}{8.1}
\setval{8b.geoma.minus.concur.writingbench}{$-$12.1}
\setval{8b.geoma.minus.concur.writingbench-abs}{12.1}
\setval{8b.geoma.minus.gd2po.aime24}{$-$2.2}
\setval{8b.geoma.minus.gd2po.aime24-abs}{2.2}
\setval{8b.geoma.minus.gd2po.aime25}{+0.3}
\setval{8b.geoma.minus.gd2po.aime25-abs}{0.3}
\setval{8b.geoma.minus.gd2po.amc23}{+3.9}
\setval{8b.geoma.minus.gd2po.amc23-abs}{3.9}
\setval{8b.geoma.minus.gd2po.avg}{$-$0.7}
\setval{8b.geoma.minus.gd2po.avg-abs}{0.7}
\setval{8b.geoma.minus.gd2po.code}{$-$0.4}
\setval{8b.geoma.minus.gd2po.code-abs}{0.4}
\setval{8b.geoma.minus.gd2po.creative-writing-v3}{+2.2}
\setval{8b.geoma.minus.gd2po.creative-writing-v3-abs}{2.2}
\setval{8b.geoma.minus.gd2po.guidebench}{+0.7}
\setval{8b.geoma.minus.gd2po.guidebench-abs}{0.7}
\setval{8b.geoma.minus.gd2po.humaneval-plus}{$-$1.2}
\setval{8b.geoma.minus.gd2po.humaneval-plus-abs}{1.2}
\setval{8b.geoma.minus.gd2po.if}{$-$3.5}
\setval{8b.geoma.minus.gd2po.if-abs}{3.5}
\setval{8b.geoma.minus.gd2po.ifbench}{$-$9.0}
\setval{8b.geoma.minus.gd2po.ifbench-abs}{9.0}
\setval{8b.geoma.minus.gd2po.ifeval}{$-$2.0}
\setval{8b.geoma.minus.gd2po.ifeval-abs}{2.0}
\setval{8b.geoma.minus.gd2po.judge2.avg}{$-$0.5}
\setval{8b.geoma.minus.gd2po.judge2.avg-abs}{0.5}
\setval{8b.geoma.minus.gd2po.judge2.creative-writing-v3}{+0.8}
\setval{8b.geoma.minus.gd2po.judge2.creative-writing-v3-abs}{0.8}
\setval{8b.geoma.minus.gd2po.judge2.longbench-write}{+2.7}
\setval{8b.geoma.minus.gd2po.judge2.longbench-write-abs}{2.7}
\setval{8b.geoma.minus.gd2po.judge2.writing}{+2.2}
\setval{8b.geoma.minus.gd2po.judge2.writing-abs}{2.2}
\setval{8b.geoma.minus.gd2po.judge2.writingbench}{+3.1}
\setval{8b.geoma.minus.gd2po.judge2.writingbench-abs}{3.1}
\setval{8b.geoma.minus.gd2po.livecodebench}{$-$0.6}
\setval{8b.geoma.minus.gd2po.livecodebench-abs}{0.6}
\setval{8b.geoma.minus.gd2po.longbench-write}{$-$0.9}
\setval{8b.geoma.minus.gd2po.longbench-write-abs}{0.9}
\setval{8b.geoma.minus.gd2po.math}{$-$0.4}
\setval{8b.geoma.minus.gd2po.math-abs}{0.4}
\setval{8b.geoma.minus.gd2po.math500}{$-$1.6}
\setval{8b.geoma.minus.gd2po.math500-abs}{1.6}
\setval{8b.geoma.minus.gd2po.mbpp-plus}{+0.5}
\setval{8b.geoma.minus.gd2po.mbpp-plus-abs}{0.5}
\setval{8b.geoma.minus.gd2po.minerva}{$-$0.1}
\setval{8b.geoma.minus.gd2po.minerva-abs}{0.1}
\setval{8b.geoma.minus.gd2po.olympiad-bench}{$-$2.6}
\setval{8b.geoma.minus.gd2po.olympiad-bench-abs}{2.6}
\setval{8b.geoma.minus.gd2po.writing}{+1.1}
\setval{8b.geoma.minus.gd2po.writing-abs}{1.1}
\setval{8b.geoma.minus.gd2po.writingbench}{+2.0}
\setval{8b.geoma.minus.gd2po.writingbench-abs}{2.0}
\setval{8b.geoma.minus.gdpo-noaux.aime24}{$-$13.9}
\setval{8b.geoma.minus.gdpo-noaux.aime24-abs}{13.9}
\setval{8b.geoma.minus.gdpo-noaux.aime25}{$-$11.5}
\setval{8b.geoma.minus.gdpo-noaux.aime25-abs}{11.5}
\setval{8b.geoma.minus.gdpo-noaux.amc23}{$-$13.6}
\setval{8b.geoma.minus.gdpo-noaux.amc23-abs}{13.6}
\setval{8b.geoma.minus.gdpo-noaux.avg}{$-$3.8}
\setval{8b.geoma.minus.gdpo-noaux.avg-abs}{3.8}
\setval{8b.geoma.minus.gdpo-noaux.code}{$-$1.4}
\setval{8b.geoma.minus.gdpo-noaux.code-abs}{1.4}
\setval{8b.geoma.minus.gdpo-noaux.creative-writing-v3}{+6.5}
\setval{8b.geoma.minus.gdpo-noaux.creative-writing-v3-abs}{6.5}
\setval{8b.geoma.minus.gdpo-noaux.guidebench}{+3.0}
\setval{8b.geoma.minus.gdpo-noaux.guidebench-abs}{3.0}
\setval{8b.geoma.minus.gdpo-noaux.humaneval-plus}{$-$3.2}
\setval{8b.geoma.minus.gdpo-noaux.humaneval-plus-abs}{3.2}
\setval{8b.geoma.minus.gdpo-noaux.if}{+1.3}
\setval{8b.geoma.minus.gdpo-noaux.if-abs}{1.3}
\setval{8b.geoma.minus.gdpo-noaux.ifbench}{+0.3}
\setval{8b.geoma.minus.gdpo-noaux.ifbench-abs}{0.3}
\setval{8b.geoma.minus.gdpo-noaux.ifeval}{+0.7}
\setval{8b.geoma.minus.gdpo-noaux.ifeval-abs}{0.7}
\setval{8b.geoma.minus.gdpo-noaux.judge2.avg}{$-$2.7}
\setval{8b.geoma.minus.gdpo-noaux.judge2.avg-abs}{2.7}
\setval{8b.geoma.minus.gdpo-noaux.judge2.creative-writing-v3}{+8.7}
\setval{8b.geoma.minus.gdpo-noaux.judge2.creative-writing-v3-abs}{8.7}
\setval{8b.geoma.minus.gdpo-noaux.judge2.longbench-write}{+10.4}
\setval{8b.geoma.minus.gdpo-noaux.judge2.longbench-write-abs}{10.4}
\setval{8b.geoma.minus.gdpo-noaux.judge2.writing}{+9.1}
\setval{8b.geoma.minus.gdpo-noaux.judge2.writing-abs}{9.1}
\setval{8b.geoma.minus.gdpo-noaux.judge2.writingbench}{+8.3}
\setval{8b.geoma.minus.gdpo-noaux.judge2.writingbench-abs}{8.3}
\setval{8b.geoma.minus.gdpo-noaux.livecodebench}{$-$1.7}
\setval{8b.geoma.minus.gdpo-noaux.livecodebench-abs}{1.7}
\setval{8b.geoma.minus.gdpo-noaux.longbench-write}{+3.0}
\setval{8b.geoma.minus.gdpo-noaux.longbench-write-abs}{3.0}
\setval{8b.geoma.minus.gdpo-noaux.math}{$-$11.3}
\setval{8b.geoma.minus.gdpo-noaux.math-abs}{11.3}
\setval{8b.geoma.minus.gdpo-noaux.math500}{$-$10.6}
\setval{8b.geoma.minus.gdpo-noaux.math500-abs}{10.6}
\setval{8b.geoma.minus.gdpo-noaux.mbpp-plus}{+0.7}
\setval{8b.geoma.minus.gdpo-noaux.mbpp-plus-abs}{0.7}
\setval{8b.geoma.minus.gdpo-noaux.minerva}{$-$4.2}
\setval{8b.geoma.minus.gdpo-noaux.minerva-abs}{4.2}
\setval{8b.geoma.minus.gdpo-noaux.olympiad-bench}{$-$13.9}
\setval{8b.geoma.minus.gdpo-noaux.olympiad-bench-abs}{13.9}
\setval{8b.geoma.minus.gdpo-noaux.writing}{+3.5}
\setval{8b.geoma.minus.gdpo-noaux.writing-abs}{3.5}
\setval{8b.geoma.minus.gdpo-noaux.writingbench}{+0.9}
\setval{8b.geoma.minus.gdpo-noaux.writingbench-abs}{0.9}
\setval{8b.geoma.minus.gdpo-w0.5.aime24}{$-$5.2}
\setval{8b.geoma.minus.gdpo-w0.5.aime24-abs}{5.2}
\setval{8b.geoma.minus.gdpo-w0.5.aime25}{$-$2.9}
\setval{8b.geoma.minus.gdpo-w0.5.aime25-abs}{2.9}
\setval{8b.geoma.minus.gdpo-w0.5.amc23}{+0.1}
\setval{8b.geoma.minus.gdpo-w0.5.amc23-abs}{0.1}
\setval{8b.geoma.minus.gdpo-w0.5.avg}{$-$1.6}
\setval{8b.geoma.minus.gdpo-w0.5.avg-abs}{1.6}
\setval{8b.geoma.minus.gdpo-w0.5.code}{+0.2}
\setval{8b.geoma.minus.gdpo-w0.5.code-abs}{0.2}
\setval{8b.geoma.minus.gdpo-w0.5.creative-writing-v3}{$-$1.2}
\setval{8b.geoma.minus.gdpo-w0.5.creative-writing-v3-abs}{1.2}
\setval{8b.geoma.minus.gdpo-w0.5.guidebench}{+0.4}
\setval{8b.geoma.minus.gdpo-w0.5.guidebench-abs}{0.4}
\setval{8b.geoma.minus.gdpo-w0.5.humaneval-plus}{$-$1.0}
\setval{8b.geoma.minus.gdpo-w0.5.humaneval-plus-abs}{1.0}
\setval{8b.geoma.minus.gdpo-w0.5.if}{$-$1.3}
\setval{8b.geoma.minus.gdpo-w0.5.if-abs}{1.3}
\setval{8b.geoma.minus.gdpo-w0.5.ifbench}{$-$3.7}
\setval{8b.geoma.minus.gdpo-w0.5.ifbench-abs}{3.7}
\setval{8b.geoma.minus.gdpo-w0.5.ifeval}{$-$0.5}
\setval{8b.geoma.minus.gdpo-w0.5.ifeval-abs}{0.5}
\setval{8b.geoma.minus.gdpo-w0.5.judge2.avg}{$-$1.4}
\setval{8b.geoma.minus.gdpo-w0.5.judge2.avg-abs}{1.4}
\setval{8b.geoma.minus.gdpo-w0.5.judge2.creative-writing-v3}{$-$2.8}
\setval{8b.geoma.minus.gdpo-w0.5.judge2.creative-writing-v3-abs}{2.8}
\setval{8b.geoma.minus.gdpo-w0.5.judge2.longbench-write}{+1.2}
\setval{8b.geoma.minus.gdpo-w0.5.judge2.longbench-write-abs}{1.2}
\setval{8b.geoma.minus.gdpo-w0.5.judge2.writing}{+0.2}
\setval{8b.geoma.minus.gdpo-w0.5.judge2.writing-abs}{0.2}
\setval{8b.geoma.minus.gdpo-w0.5.judge2.writingbench}{+2.1}
\setval{8b.geoma.minus.gdpo-w0.5.judge2.writingbench-abs}{2.1}
\setval{8b.geoma.minus.gdpo-w0.5.livecodebench}{$-$1.1}
\setval{8b.geoma.minus.gdpo-w0.5.livecodebench-abs}{1.1}
\setval{8b.geoma.minus.gdpo-w0.5.longbench-write}{$-$2.2}
\setval{8b.geoma.minus.gdpo-w0.5.longbench-write-abs}{2.2}
\setval{8b.geoma.minus.gdpo-w0.5.math}{$-$3.2}
\setval{8b.geoma.minus.gdpo-w0.5.math-abs}{3.2}
\setval{8b.geoma.minus.gdpo-w0.5.math500}{$-$4.4}
\setval{8b.geoma.minus.gdpo-w0.5.math500-abs}{4.4}
\setval{8b.geoma.minus.gdpo-w0.5.mbpp-plus}{+2.8}
\setval{8b.geoma.minus.gdpo-w0.5.mbpp-plus-abs}{2.8}
\setval{8b.geoma.minus.gdpo-w0.5.minerva}{$-$1.4}
\setval{8b.geoma.minus.gdpo-w0.5.minerva-abs}{1.4}
\setval{8b.geoma.minus.gdpo-w0.5.olympiad-bench}{$-$5.1}
\setval{8b.geoma.minus.gdpo-w0.5.olympiad-bench-abs}{5.1}
\setval{8b.geoma.minus.gdpo-w0.5.writing}{$-$0.8}
\setval{8b.geoma.minus.gdpo-w0.5.writing-abs}{0.8}
\setval{8b.geoma.minus.gdpo-w0.5.writingbench}{+0.9}
\setval{8b.geoma.minus.gdpo-w0.5.writingbench-abs}{0.9}
\setval{8b.geoma.minus.gdpo.aime24}{$-$7.3}
\setval{8b.geoma.minus.gdpo.aime24-abs}{7.3}
\setval{8b.geoma.minus.gdpo.aime25}{$-$1.9}
\setval{8b.geoma.minus.gdpo.aime25-abs}{1.9}
\setval{8b.geoma.minus.gdpo.amc23}{+1.3}
\setval{8b.geoma.minus.gdpo.amc23-abs}{1.3}
\setval{8b.geoma.minus.gdpo.avg}{$-$1.7}
\setval{8b.geoma.minus.gdpo.avg-abs}{1.7}
\setval{8b.geoma.minus.gdpo.code}{$-$0.3}
\setval{8b.geoma.minus.gdpo.code-abs}{0.3}
\setval{8b.geoma.minus.gdpo.creative-writing-v3}{$-$2.5}
\setval{8b.geoma.minus.gdpo.creative-writing-v3-abs}{2.5}
\setval{8b.geoma.minus.gdpo.guidebench}{$-$0.2}
\setval{8b.geoma.minus.gdpo.guidebench-abs}{0.2}
\setval{8b.geoma.minus.gdpo.humaneval-plus}{$-$1.3}
\setval{8b.geoma.minus.gdpo.humaneval-plus-abs}{1.3}
\setval{8b.geoma.minus.gdpo.if}{$-$2.1}
\setval{8b.geoma.minus.gdpo.if-abs}{2.1}
\setval{8b.geoma.minus.gdpo.ifbench}{$-$5.3}
\setval{8b.geoma.minus.gdpo.ifbench-abs}{5.3}
\setval{8b.geoma.minus.gdpo.ifeval}{$-$0.8}
\setval{8b.geoma.minus.gdpo.ifeval-abs}{0.8}
\setval{8b.geoma.minus.gdpo.judge2.avg}{$-$1.6}
\setval{8b.geoma.minus.gdpo.judge2.avg-abs}{1.6}
\setval{8b.geoma.minus.gdpo.judge2.creative-writing-v3}{$-$3.5}
\setval{8b.geoma.minus.gdpo.judge2.creative-writing-v3-abs}{3.5}
\setval{8b.geoma.minus.gdpo.judge2.longbench-write}{+0.7}
\setval{8b.geoma.minus.gdpo.judge2.longbench-write-abs}{0.7}
\setval{8b.geoma.minus.gdpo.judge2.writing}{$-$1.0}
\setval{8b.geoma.minus.gdpo.judge2.writing-abs}{1.0}
\setval{8b.geoma.minus.gdpo.judge2.writingbench}{$-$0.1}
\setval{8b.geoma.minus.gdpo.judge2.writingbench-abs}{0.1}
\setval{8b.geoma.minus.gdpo.livecodebench}{$-$0.5}
\setval{8b.geoma.minus.gdpo.livecodebench-abs}{0.5}
\setval{8b.geoma.minus.gdpo.longbench-write}{$-$0.6}
\setval{8b.geoma.minus.gdpo.longbench-write-abs}{0.6}
\setval{8b.geoma.minus.gdpo.math}{$-$2.4}
\setval{8b.geoma.minus.gdpo.math-abs}{2.4}
\setval{8b.geoma.minus.gdpo.math500}{$-$3.4}
\setval{8b.geoma.minus.gdpo.math500-abs}{3.4}
\setval{8b.geoma.minus.gdpo.mbpp-plus}{+1.1}
\setval{8b.geoma.minus.gdpo.mbpp-plus-abs}{1.1}
\setval{8b.geoma.minus.gdpo.minerva}{+0.2}
\setval{8b.geoma.minus.gdpo.minerva-abs}{0.2}
\setval{8b.geoma.minus.gdpo.olympiad-bench}{$-$3.3}
\setval{8b.geoma.minus.gdpo.olympiad-bench-abs}{3.3}
\setval{8b.geoma.minus.gdpo.writing}{$-$1.1}
\setval{8b.geoma.minus.gdpo.writing-abs}{1.1}
\setval{8b.geoma.minus.gdpo.writingbench}{$-$0.2}
\setval{8b.geoma.minus.gdpo.writingbench-abs}{0.2}
\setval{8b.geoma.minus.grpo-nodiv.aime24}{$-$6.6}
\setval{8b.geoma.minus.grpo-nodiv.aime24-abs}{6.6}
\setval{8b.geoma.minus.grpo-nodiv.aime25}{$-$2.3}
\setval{8b.geoma.minus.grpo-nodiv.aime25-abs}{2.3}
\setval{8b.geoma.minus.grpo-nodiv.amc23}{$-$0.1}
\setval{8b.geoma.minus.grpo-nodiv.amc23-abs}{0.1}
\setval{8b.geoma.minus.grpo-nodiv.avg}{$-$1.4}
\setval{8b.geoma.minus.grpo-nodiv.avg-abs}{1.4}
\setval{8b.geoma.minus.grpo-nodiv.code}{$-$1.1}
\setval{8b.geoma.minus.grpo-nodiv.code-abs}{1.1}
\setval{8b.geoma.minus.grpo-nodiv.creative-writing-v3}{+3.7}
\setval{8b.geoma.minus.grpo-nodiv.creative-writing-v3-abs}{3.7}
\setval{8b.geoma.minus.grpo-nodiv.guidebench}{$-$0.5}
\setval{8b.geoma.minus.grpo-nodiv.guidebench-abs}{0.5}
\setval{8b.geoma.minus.grpo-nodiv.humaneval-plus}{$-$3.6}
\setval{8b.geoma.minus.grpo-nodiv.humaneval-plus-abs}{3.6}
\setval{8b.geoma.minus.grpo-nodiv.if}{$-$1.7}
\setval{8b.geoma.minus.grpo-nodiv.if-abs}{1.7}
\setval{8b.geoma.minus.grpo-nodiv.ifbench}{$-$2.3}
\setval{8b.geoma.minus.grpo-nodiv.ifbench-abs}{2.3}
\setval{8b.geoma.minus.grpo-nodiv.ifeval}{$-$2.2}
\setval{8b.geoma.minus.grpo-nodiv.ifeval-abs}{2.2}
\setval{8b.geoma.minus.grpo-nodiv.judge2.avg}{$-$1.7}
\setval{8b.geoma.minus.grpo-nodiv.judge2.avg-abs}{1.7}
\setval{8b.geoma.minus.grpo-nodiv.judge2.creative-writing-v3}{+0.1}
\setval{8b.geoma.minus.grpo-nodiv.judge2.creative-writing-v3-abs}{0.1}
\setval{8b.geoma.minus.grpo-nodiv.judge2.longbench-write}{$-$1.7}
\setval{8b.geoma.minus.grpo-nodiv.judge2.longbench-write-abs}{1.7}
\setval{8b.geoma.minus.grpo-nodiv.judge2.writing}{+0.9}
\setval{8b.geoma.minus.grpo-nodiv.judge2.writing-abs}{0.9}
\setval{8b.geoma.minus.grpo-nodiv.judge2.writingbench}{+4.4}
\setval{8b.geoma.minus.grpo-nodiv.judge2.writingbench-abs}{4.4}
\setval{8b.geoma.minus.grpo-nodiv.livecodebench}{+0.7}
\setval{8b.geoma.minus.grpo-nodiv.livecodebench-abs}{0.7}
\setval{8b.geoma.minus.grpo-nodiv.longbench-write}{$-$2.1}
\setval{8b.geoma.minus.grpo-nodiv.longbench-write-abs}{2.1}
\setval{8b.geoma.minus.grpo-nodiv.math}{$-$3.3}
\setval{8b.geoma.minus.grpo-nodiv.math-abs}{3.3}
\setval{8b.geoma.minus.grpo-nodiv.math500}{$-$4.5}
\setval{8b.geoma.minus.grpo-nodiv.math500-abs}{4.5}
\setval{8b.geoma.minus.grpo-nodiv.mbpp-plus}{$-$0.4}
\setval{8b.geoma.minus.grpo-nodiv.mbpp-plus-abs}{0.4}
\setval{8b.geoma.minus.grpo-nodiv.minerva}{$-$1.0}
\setval{8b.geoma.minus.grpo-nodiv.minerva-abs}{1.0}
\setval{8b.geoma.minus.grpo-nodiv.olympiad-bench}{$-$5.5}
\setval{8b.geoma.minus.grpo-nodiv.olympiad-bench-abs}{5.5}
\setval{8b.geoma.minus.grpo-nodiv.writing}{+2.4}
\setval{8b.geoma.minus.grpo-nodiv.writing-abs}{2.4}
\setval{8b.geoma.minus.grpo-nodiv.writingbench}{+5.6}
\setval{8b.geoma.minus.grpo-nodiv.writingbench-abs}{5.6}
\setval{8b.geoma.minus.grpo.aime24}{$-$1.8}
\setval{8b.geoma.minus.grpo.aime24-abs}{1.8}
\setval{8b.geoma.minus.grpo.aime25}{$-$3.0}
\setval{8b.geoma.minus.grpo.aime25-abs}{3.0}
\setval{8b.geoma.minus.grpo.amc23}{$-$0.9}
\setval{8b.geoma.minus.grpo.amc23-abs}{0.9}
\setval{8b.geoma.minus.grpo.avg}{$-$1.2}
\setval{8b.geoma.minus.grpo.avg-abs}{1.2}
\setval{8b.geoma.minus.grpo.code}{$-$0.3}
\setval{8b.geoma.minus.grpo.code-abs}{0.3}
\setval{8b.geoma.minus.grpo.creative-writing-v3}{$-$2.1}
\setval{8b.geoma.minus.grpo.creative-writing-v3-abs}{2.1}
\setval{8b.geoma.minus.grpo.guidebench}{+2.5}
\setval{8b.geoma.minus.grpo.guidebench-abs}{2.5}
\setval{8b.geoma.minus.grpo.humaneval-plus}{$-$1.0}
\setval{8b.geoma.minus.grpo.humaneval-plus-abs}{1.0}
\setval{8b.geoma.minus.grpo.if}{$-$1.7}
\setval{8b.geoma.minus.grpo.if-abs}{1.7}
\setval{8b.geoma.minus.grpo.ifbench}{$-$8.0}
\setval{8b.geoma.minus.grpo.ifbench-abs}{8.0}
\setval{8b.geoma.minus.grpo.ifeval}{+0.5}
\setval{8b.geoma.minus.grpo.ifeval-abs}{0.5}
\setval{8b.geoma.minus.grpo.judge2.avg}{$-$0.8}
\setval{8b.geoma.minus.grpo.judge2.avg-abs}{0.8}
\setval{8b.geoma.minus.grpo.judge2.creative-writing-v3}{$-$1.7}
\setval{8b.geoma.minus.grpo.judge2.creative-writing-v3-abs}{1.7}
\setval{8b.geoma.minus.grpo.judge2.longbench-write}{+2.4}
\setval{8b.geoma.minus.grpo.judge2.longbench-write-abs}{2.4}
\setval{8b.geoma.minus.grpo.judge2.writing}{+0.7}
\setval{8b.geoma.minus.grpo.judge2.writing-abs}{0.7}
\setval{8b.geoma.minus.grpo.judge2.writingbench}{+1.4}
\setval{8b.geoma.minus.grpo.judge2.writingbench-abs}{1.4}
\setval{8b.geoma.minus.grpo.livecodebench}{$-$0.8}
\setval{8b.geoma.minus.grpo.livecodebench-abs}{0.8}
\setval{8b.geoma.minus.grpo.longbench-write}{$-$0.8}
\setval{8b.geoma.minus.grpo.longbench-write-abs}{0.8}
\setval{8b.geoma.minus.grpo.math}{$-$1.5}
\setval{8b.geoma.minus.grpo.math-abs}{1.5}
\setval{8b.geoma.minus.grpo.math500}{$-$1.0}
\setval{8b.geoma.minus.grpo.math500-abs}{1.0}
\setval{8b.geoma.minus.grpo.mbpp-plus}{+1.0}
\setval{8b.geoma.minus.grpo.mbpp-plus-abs}{1.0}
\setval{8b.geoma.minus.grpo.minerva}{$-$0.1}
\setval{8b.geoma.minus.grpo.minerva-abs}{0.1}
\setval{8b.geoma.minus.grpo.olympiad-bench}{$-$1.9}
\setval{8b.geoma.minus.grpo.olympiad-bench-abs}{1.9}
\setval{8b.geoma.minus.grpo.writing}{$-$1.1}
\setval{8b.geoma.minus.grpo.writing-abs}{1.1}
\setval{8b.geoma.minus.grpo.writingbench}{$-$0.5}
\setval{8b.geoma.minus.grpo.writingbench-abs}{0.5}
\setval{8b.geoma.minus.mt-grpo.aime24}{$-$0.4}
\setval{8b.geoma.minus.mt-grpo.aime24-abs}{0.4}
\setval{8b.geoma.minus.mt-grpo.aime25}{$-$3.2}
\setval{8b.geoma.minus.mt-grpo.aime25-abs}{3.2}
\setval{8b.geoma.minus.mt-grpo.amc23}{$-$2.1}
\setval{8b.geoma.minus.mt-grpo.amc23-abs}{2.1}
\setval{8b.geoma.minus.mt-grpo.avg}{$-$0.5}
\setval{8b.geoma.minus.mt-grpo.avg-abs}{0.5}
\setval{8b.geoma.minus.mt-grpo.code}{$-$1.3}
\setval{8b.geoma.minus.mt-grpo.code-abs}{1.3}
\setval{8b.geoma.minus.mt-grpo.creative-writing-v3}{+0.1}
\setval{8b.geoma.minus.mt-grpo.creative-writing-v3-abs}{0.1}
\setval{8b.geoma.minus.mt-grpo.guidebench}{+1.2}
\setval{8b.geoma.minus.mt-grpo.guidebench-abs}{1.2}
\setval{8b.geoma.minus.mt-grpo.humaneval-plus}{$-$2.3}
\setval{8b.geoma.minus.mt-grpo.humaneval-plus-abs}{2.3}
\setval{8b.geoma.minus.mt-grpo.if}{$-$0.6}
\setval{8b.geoma.minus.mt-grpo.if-abs}{0.6}
\setval{8b.geoma.minus.mt-grpo.ifbench}{$-$3.3}
\setval{8b.geoma.minus.mt-grpo.ifbench-abs}{3.3}
\setval{8b.geoma.minus.mt-grpo.ifeval}{+0.4}
\setval{8b.geoma.minus.mt-grpo.ifeval-abs}{0.4}
\setval{8b.geoma.minus.mt-grpo.judge2.avg}{$-$0.8}
\setval{8b.geoma.minus.mt-grpo.judge2.avg-abs}{0.8}
\setval{8b.geoma.minus.mt-grpo.judge2.creative-writing-v3}{$-$4.0}
\setval{8b.geoma.minus.mt-grpo.judge2.creative-writing-v3-abs}{4.0}
\setval{8b.geoma.minus.mt-grpo.judge2.longbench-write}{+2.4}
\setval{8b.geoma.minus.mt-grpo.judge2.longbench-write-abs}{2.4}
\setval{8b.geoma.minus.mt-grpo.judge2.writing}{+0.4}
\setval{8b.geoma.minus.mt-grpo.judge2.writing-abs}{0.4}
\setval{8b.geoma.minus.mt-grpo.judge2.writingbench}{+2.8}
\setval{8b.geoma.minus.mt-grpo.judge2.writingbench-abs}{2.8}
\setval{8b.geoma.minus.mt-grpo.livecodebench}{$-$1.9}
\setval{8b.geoma.minus.mt-grpo.livecodebench-abs}{1.9}
\setval{8b.geoma.minus.mt-grpo.longbench-write}{+1.5}
\setval{8b.geoma.minus.mt-grpo.longbench-write-abs}{1.5}
\setval{8b.geoma.minus.mt-grpo.math}{$-$1.3}
\setval{8b.geoma.minus.mt-grpo.math-abs}{1.3}
\setval{8b.geoma.minus.mt-grpo.math500}{$-$1.0}
\setval{8b.geoma.minus.mt-grpo.math500-abs}{1.0}
\setval{8b.geoma.minus.mt-grpo.mbpp-plus}{+0.3}
\setval{8b.geoma.minus.mt-grpo.mbpp-plus-abs}{0.3}
\setval{8b.geoma.minus.mt-grpo.minerva}{$-$1.3}
\setval{8b.geoma.minus.mt-grpo.minerva-abs}{1.3}
\setval{8b.geoma.minus.mt-grpo.olympiad-bench}{+0.2}
\setval{8b.geoma.minus.mt-grpo.olympiad-bench-abs}{0.2}
\setval{8b.geoma.minus.mt-grpo.writing}{+1.9}
\setval{8b.geoma.minus.mt-grpo.writing-abs}{1.9}
\setval{8b.geoma.minus.mt-grpo.writingbench}{+4.2}
\setval{8b.geoma.minus.mt-grpo.writingbench-abs}{4.2}
\setval{8b.geoma.minus.papo.aime24}{+1.9}
\setval{8b.geoma.minus.papo.aime24-abs}{1.9}
\setval{8b.geoma.minus.papo.aime25}{+2.4}
\setval{8b.geoma.minus.papo.aime25-abs}{2.4}
\setval{8b.geoma.minus.papo.amc23}{+1.5}
\setval{8b.geoma.minus.papo.amc23-abs}{1.5}
\setval{8b.geoma.minus.papo.avg}{+0.9}
\setval{8b.geoma.minus.papo.avg-abs}{0.9}
\setval{8b.geoma.minus.papo.code}{+0.5}
\setval{8b.geoma.minus.papo.code-abs}{0.5}
\setval{8b.geoma.minus.papo.creative-writing-v3}{$-$0.8}
\setval{8b.geoma.minus.papo.creative-writing-v3-abs}{0.8}
\setval{8b.geoma.minus.papo.guidebench}{+0.0}
\setval{8b.geoma.minus.papo.guidebench-abs}{0.0}
\setval{8b.geoma.minus.papo.humaneval-plus}{+0.2}
\setval{8b.geoma.minus.papo.humaneval-plus-abs}{0.2}
\setval{8b.geoma.minus.papo.if}{$-$0.8}
\setval{8b.geoma.minus.papo.if-abs}{0.8}
\setval{8b.geoma.minus.papo.ifbench}{$-$3.3}
\setval{8b.geoma.minus.papo.ifbench-abs}{3.3}
\setval{8b.geoma.minus.papo.ifeval}{+1.0}
\setval{8b.geoma.minus.papo.ifeval-abs}{1.0}
\setval{8b.geoma.minus.papo.judge2.avg}{+0.7}
\setval{8b.geoma.minus.papo.judge2.avg-abs}{0.7}
\setval{8b.geoma.minus.papo.judge2.creative-writing-v3}{$-$4.3}
\setval{8b.geoma.minus.papo.judge2.creative-writing-v3-abs}{4.3}
\setval{8b.geoma.minus.papo.judge2.longbench-write}{+5.3}
\setval{8b.geoma.minus.papo.judge2.longbench-write-abs}{5.3}
\setval{8b.geoma.minus.papo.judge2.writing}{+1.0}
\setval{8b.geoma.minus.papo.judge2.writing-abs}{1.0}
\setval{8b.geoma.minus.papo.judge2.writingbench}{+2.0}
\setval{8b.geoma.minus.papo.judge2.writingbench-abs}{2.0}
\setval{8b.geoma.minus.papo.livecodebench}{$-$0.9}
\setval{8b.geoma.minus.papo.livecodebench-abs}{0.9}
\setval{8b.geoma.minus.papo.longbench-write}{+3.5}
\setval{8b.geoma.minus.papo.longbench-write-abs}{3.5}
\setval{8b.geoma.minus.papo.math}{+1.4}
\setval{8b.geoma.minus.papo.math-abs}{1.4}
\setval{8b.geoma.minus.papo.math500}{+1.2}
\setval{8b.geoma.minus.papo.math500-abs}{1.2}
\setval{8b.geoma.minus.papo.mbpp-plus}{+2.2}
\setval{8b.geoma.minus.papo.mbpp-plus-abs}{2.2}
\setval{8b.geoma.minus.papo.minerva}{+0.4}
\setval{8b.geoma.minus.papo.minerva-abs}{0.4}
\setval{8b.geoma.minus.papo.olympiad-bench}{+1.2}
\setval{8b.geoma.minus.papo.olympiad-bench-abs}{1.2}
\setval{8b.geoma.minus.papo.writing}{+1.8}
\setval{8b.geoma.minus.papo.writing-abs}{1.8}
\setval{8b.geoma.minus.papo.writingbench}{+2.7}
\setval{8b.geoma.minus.papo.writingbench-abs}{2.7}
\setval{8b.geoma.minus.saw.aime24}{$-$1.3}
\setval{8b.geoma.minus.saw.aime24-abs}{1.3}
\setval{8b.geoma.minus.saw.aime25}{$-$3.0}
\setval{8b.geoma.minus.saw.aime25-abs}{3.0}
\setval{8b.geoma.minus.saw.amc23}{+1.2}
\setval{8b.geoma.minus.saw.amc23-abs}{1.2}
\setval{8b.geoma.minus.saw.avg}{$-$1.3}
\setval{8b.geoma.minus.saw.avg-abs}{1.3}
\setval{8b.geoma.minus.saw.code}{$-$0.9}
\setval{8b.geoma.minus.saw.code-abs}{0.9}
\setval{8b.geoma.minus.saw.creative-writing-v3}{+0.7}
\setval{8b.geoma.minus.saw.creative-writing-v3-abs}{0.7}
\setval{8b.geoma.minus.saw.guidebench}{+0.1}
\setval{8b.geoma.minus.saw.guidebench-abs}{0.1}
\setval{8b.geoma.minus.saw.humaneval-plus}{$-$3.1}
\setval{8b.geoma.minus.saw.humaneval-plus-abs}{3.1}
\setval{8b.geoma.minus.saw.if}{$-$2.1}
\setval{8b.geoma.minus.saw.if-abs}{2.1}
\setval{8b.geoma.minus.saw.ifbench}{$-$6.0}
\setval{8b.geoma.minus.saw.ifbench-abs}{6.0}
\setval{8b.geoma.minus.saw.ifeval}{$-$0.4}
\setval{8b.geoma.minus.saw.ifeval-abs}{0.4}
\setval{8b.geoma.minus.saw.judge2.avg}{$-$1.1}
\setval{8b.geoma.minus.saw.judge2.avg-abs}{1.1}
\setval{8b.geoma.minus.saw.judge2.creative-writing-v3}{+1.0}
\setval{8b.geoma.minus.saw.judge2.creative-writing-v3-abs}{1.0}
\setval{8b.geoma.minus.saw.judge2.longbench-write}{+0.1}
\setval{8b.geoma.minus.saw.judge2.longbench-write-abs}{0.1}
\setval{8b.geoma.minus.saw.judge2.writing}{+0.8}
\setval{8b.geoma.minus.saw.judge2.writing-abs}{0.8}
\setval{8b.geoma.minus.saw.judge2.writingbench}{+1.4}
\setval{8b.geoma.minus.saw.judge2.writingbench-abs}{1.4}
\setval{8b.geoma.minus.saw.livecodebench}{$-$1.0}
\setval{8b.geoma.minus.saw.livecodebench-abs}{1.0}
\setval{8b.geoma.minus.saw.longbench-write}{$-$2.1}
\setval{8b.geoma.minus.saw.longbench-write-abs}{2.1}
\setval{8b.geoma.minus.saw.math}{$-$1.6}
\setval{8b.geoma.minus.saw.math-abs}{1.6}
\setval{8b.geoma.minus.saw.math500}{$-$2.3}
\setval{8b.geoma.minus.saw.math500-abs}{2.3}
\setval{8b.geoma.minus.saw.mbpp-plus}{+1.3}
\setval{8b.geoma.minus.saw.mbpp-plus-abs}{1.3}
\setval{8b.geoma.minus.saw.minerva}{$-$2.0}
\setval{8b.geoma.minus.saw.minerva-abs}{2.0}
\setval{8b.geoma.minus.saw.olympiad-bench}{$-$2.1}
\setval{8b.geoma.minus.saw.olympiad-bench-abs}{2.1}
\setval{8b.geoma.minus.saw.writing}{$-$0.3}
\setval{8b.geoma.minus.saw.writing-abs}{0.3}
\setval{8b.geoma.minus.saw.writingbench}{+0.5}
\setval{8b.geoma.minus.saw.writingbench-abs}{0.5}
\setval{8b.geoma.passing-at-ceiling.code.early}{0.71}
\setval{8b.geoma.passing-at-ceiling.code.late}{0.76}
\setval{8b.geoma.passing-at-ceiling.code.run}{0.75}
\setval{8b.geoma.passing-at-ceiling.if.early}{0.32}
\setval{8b.geoma.passing-at-ceiling.if.late}{0.56}
\setval{8b.geoma.passing-at-ceiling.if.run}{0.49}
\setval{8b.geoma.passing-at-ceiling.math.early}{1.00}
\setval{8b.geoma.passing-at-ceiling.math.late}{1.00}
\setval{8b.geoma.passing-at-ceiling.math.run}{1.00}
\setval{8b.geoma.passing-at-ceiling.writing.early}{0.00}
\setval{8b.geoma.passing-at-ceiling.writing.late}{0.01}
\setval{8b.geoma.passing-at-ceiling.writing.run}{0.00}
\setval{8b.geoma.primary-cosine.code.early}{0.93}
\setval{8b.geoma.primary-cosine.code.late}{0.97}
\setval{8b.geoma.primary-cosine.code.run}{0.96}
\setval{8b.geoma.primary-cosine.if.early}{0.70}
\setval{8b.geoma.primary-cosine.if.late}{0.75}
\setval{8b.geoma.primary-cosine.if.run}{0.77}
\setval{8b.geoma.primary-cosine.math.early}{0.89}
\setval{8b.geoma.primary-cosine.math.late}{0.96}
\setval{8b.geoma.primary-cosine.math.run}{0.94}
\setval{8b.geoma.primary-cosine.writing.early}{0.84}
\setval{8b.geoma.primary-cosine.writing.late}{0.79}
\setval{8b.geoma.primary-cosine.writing.run}{0.76}
\setval{8b.geoma.primary-level.code.early}{0.55}
\setval{8b.geoma.primary-level.code.late}{0.77}
\setval{8b.geoma.primary-level.code.run}{0.72}
\setval{8b.geoma.primary-level.if.early}{0.35}
\setval{8b.geoma.primary-level.if.late}{0.75}
\setval{8b.geoma.primary-level.if.run}{0.65}
\setval{8b.geoma.primary-level.math.early}{0.48}
\setval{8b.geoma.primary-level.math.late}{0.67}
\setval{8b.geoma.primary-level.math.run}{0.62}
\setval{8b.geoma.primary-level.writing.early}{0.53}
\setval{8b.geoma.primary-level.writing.late}{0.76}
\setval{8b.geoma.primary-level.writing.run}{0.72}
\setval{8b.geoma.solved.code.early}{0.06}
\setval{8b.geoma.solved.code.late}{0.02}
\setval{8b.geoma.solved.code.run}{0.02}
\setval{8b.geoma.solved.if.early}{0.03}
\setval{8b.geoma.solved.if.late}{0.17}
\setval{8b.geoma.solved.if.run}{0.11}
\setval{8b.geoma.solved.math.early}{0.14}
\setval{8b.geoma.solved.math.late}{0.08}
\setval{8b.geoma.solved.math.run}{0.09}
\setval{8b.geoma.solved.writing.early}{0.00}
\setval{8b.geoma.solved.writing.late}{0.00}
\setval{8b.geoma.solved.writing.run}{0.00}
\setval{8b.geoma.update-share.code.early}{0.24}
\setval{8b.geoma.update-share.code.late}{0.19}
\setval{8b.geoma.update-share.code.run}{0.20}
\setval{8b.geoma.update-share.if.early}{0.26}
\setval{8b.geoma.update-share.if.late}{0.21}
\setval{8b.geoma.update-share.if.run}{0.23}
\setval{8b.geoma.update-share.math.early}{0.23}
\setval{8b.geoma.update-share.math.late}{0.19}
\setval{8b.geoma.update-share.math.run}{0.20}
\setval{8b.geoma.update-share.writing.early}{0.28}
\setval{8b.geoma.update-share.writing.late}{0.41}
\setval{8b.geoma.update-share.writing.run}{0.37}
\setval{8b.grpo-nodiv.aux-change.code}{+0.18}
\setval{8b.grpo-nodiv.aux-change.if}{+0.18}
\setval{8b.grpo-nodiv.aux-change.math}{+0.10}
\setval{8b.grpo-nodiv.aux-change.writing}{+0.13}
\setval{8b.grpo-nodiv.aux-change.writing-rm}{+0.31}
\setval{8b.grpo-nodiv.aux-full.code.early}{0.68}
\setval{8b.grpo-nodiv.aux-full.code.late}{0.94}
\setval{8b.grpo-nodiv.aux-full.code.run}{0.88}
\setval{8b.grpo-nodiv.aux-full.if.early}{0.71}
\setval{8b.grpo-nodiv.aux-full.if.late}{0.92}
\setval{8b.grpo-nodiv.aux-full.if.run}{0.87}
\setval{8b.grpo-nodiv.aux-full.math.early}{0.56}
\setval{8b.grpo-nodiv.aux-full.math.late}{0.72}
\setval{8b.grpo-nodiv.aux-full.math.run}{0.68}
\setval{8b.grpo-nodiv.aux-full.writing-rm.early}{0.05}
\setval{8b.grpo-nodiv.aux-full.writing-rm.late}{0.50}
\setval{8b.grpo-nodiv.aux-full.writing-rm.run}{0.36}
\setval{8b.grpo-nodiv.aux-full.writing.early}{0.69}
\setval{8b.grpo-nodiv.aux-full.writing.late}{0.74}
\setval{8b.grpo-nodiv.aux-full.writing.run}{0.67}
\setval{8b.grpo-nodiv.aux-level.code.early}{0.92}
\setval{8b.grpo-nodiv.aux-level.code.late}{0.99}
\setval{8b.grpo-nodiv.aux-level.code.run}{0.98}
\setval{8b.grpo-nodiv.aux-level.if.early}{0.87}
\setval{8b.grpo-nodiv.aux-level.if.late}{0.96}
\setval{8b.grpo-nodiv.aux-level.if.run}{0.94}
\setval{8b.grpo-nodiv.aux-level.math.early}{0.90}
\setval{8b.grpo-nodiv.aux-level.math.late}{0.94}
\setval{8b.grpo-nodiv.aux-level.math.run}{0.93}
\setval{8b.grpo-nodiv.aux-level.writing-rm.early}{0.76}
\setval{8b.grpo-nodiv.aux-level.writing-rm.late}{0.94}
\setval{8b.grpo-nodiv.aux-level.writing-rm.run}{0.91}
\setval{8b.grpo-nodiv.aux-level.writing.early}{0.92}
\setval{8b.grpo-nodiv.aux-level.writing.late}{0.97}
\setval{8b.grpo-nodiv.aux-level.writing.run}{0.94}
\setval{8b.grpo-nodiv.bench.aime24}{20.6}
\setval{8b.grpo-nodiv.bench.aime25}{14.8}
\setval{8b.grpo-nodiv.bench.amc23}{63.6}
\setval{8b.grpo-nodiv.bench.creative-writing-v3}{46.7}
\setval{8b.grpo-nodiv.bench.guidebench}{84.8}
\setval{8b.grpo-nodiv.bench.humaneval-plus}{86.3}
\setval{8b.grpo-nodiv.bench.ifbench}{36.3}
\setval{8b.grpo-nodiv.bench.ifeval}{89.4}
\setval{8b.grpo-nodiv.bench.livecodebench}{25.0}
\setval{8b.grpo-nodiv.bench.longbench-write}{71.1}
\setval{8b.grpo-nodiv.bench.math500}{84.1}
\setval{8b.grpo-nodiv.bench.mbpp-plus}{76.0}
\setval{8b.grpo-nodiv.bench.minerva}{50.8}
\setval{8b.grpo-nodiv.bench.olympiad-bench}{51.5}
\setval{8b.grpo-nodiv.bench.writingbench}{70.3}
\setval{8b.grpo-nodiv.eval.avg}{58.1}
\setval{8b.grpo-nodiv.eval.code}{62.4}
\setval{8b.grpo-nodiv.eval.if}{70.2}
\setval{8b.grpo-nodiv.eval.math}{47.6}
\setval{8b.grpo-nodiv.eval.writing}{62.7}
\setval{8b.grpo-nodiv.judge2.avg}{56.1}
\setval{8b.grpo-nodiv.judge2.creative-writing-v3}{37.5}
\setval{8b.grpo-nodiv.judge2.longbench-write}{66.5}
\setval{8b.grpo-nodiv.judge2.writing}{52.6}
\setval{8b.grpo-nodiv.judge2.writingbench}{53.9}
\setval{8b.grpo-nodiv.last-step}{300}
\setval{8b.grpo-nodiv.minus.base.aime24}{+9.3}
\setval{8b.grpo-nodiv.minus.base.aime24-abs}{9.3}
\setval{8b.grpo-nodiv.minus.base.aime25}{+6.2}
\setval{8b.grpo-nodiv.minus.base.aime25-abs}{6.2}
\setval{8b.grpo-nodiv.minus.base.amc23}{+13.4}
\setval{8b.grpo-nodiv.minus.base.amc23-abs}{13.4}
\setval{8b.grpo-nodiv.minus.base.avg}{+16.6}
\setval{8b.grpo-nodiv.minus.base.avg-abs}{16.6}
\setval{8b.grpo-nodiv.minus.base.code}{+6.4}
\setval{8b.grpo-nodiv.minus.base.code-abs}{6.4}
\setval{8b.grpo-nodiv.minus.base.creative-writing-v3}{+19.9}
\setval{8b.grpo-nodiv.minus.base.creative-writing-v3-abs}{19.9}
\setval{8b.grpo-nodiv.minus.base.guidebench}{+39.3}
\setval{8b.grpo-nodiv.minus.base.guidebench-abs}{39.3}
\setval{8b.grpo-nodiv.minus.base.humaneval-plus}{+6.3}
\setval{8b.grpo-nodiv.minus.base.humaneval-plus-abs}{6.3}
\setval{8b.grpo-nodiv.minus.base.if}{+31.6}
\setval{8b.grpo-nodiv.minus.base.if-abs}{31.6}
\setval{8b.grpo-nodiv.minus.base.ifbench}{+19.0}
\setval{8b.grpo-nodiv.minus.base.ifbench-abs}{19.0}
\setval{8b.grpo-nodiv.minus.base.ifeval}{+36.5}
\setval{8b.grpo-nodiv.minus.base.ifeval-abs}{36.5}
\setval{8b.grpo-nodiv.minus.base.judge2.avg}{+15.0}
\setval{8b.grpo-nodiv.minus.base.judge2.avg-abs}{15.0}
\setval{8b.grpo-nodiv.minus.base.judge2.creative-writing-v3}{+11.0}
\setval{8b.grpo-nodiv.minus.base.judge2.creative-writing-v3-abs}{11.0}
\setval{8b.grpo-nodiv.minus.base.judge2.longbench-write}{+18.5}
\setval{8b.grpo-nodiv.minus.base.judge2.longbench-write-abs}{18.5}
\setval{8b.grpo-nodiv.minus.base.judge2.writing}{+16.5}
\setval{8b.grpo-nodiv.minus.base.judge2.writing-abs}{16.5}
\setval{8b.grpo-nodiv.minus.base.judge2.writingbench}{+19.9}
\setval{8b.grpo-nodiv.minus.base.judge2.writingbench-abs}{19.9}
\setval{8b.grpo-nodiv.minus.base.livecodebench}{+4.1}
\setval{8b.grpo-nodiv.minus.base.livecodebench-abs}{4.1}
\setval{8b.grpo-nodiv.minus.base.longbench-write}{+19.6}
\setval{8b.grpo-nodiv.minus.base.longbench-write-abs}{19.6}
\setval{8b.grpo-nodiv.minus.base.math}{+10.4}
\setval{8b.grpo-nodiv.minus.base.math-abs}{10.4}
\setval{8b.grpo-nodiv.minus.base.math500}{+11.7}
\setval{8b.grpo-nodiv.minus.base.math500-abs}{11.7}
\setval{8b.grpo-nodiv.minus.base.mbpp-plus}{+8.8}
\setval{8b.grpo-nodiv.minus.base.mbpp-plus-abs}{8.8}
\setval{8b.grpo-nodiv.minus.base.minerva}{+9.9}
\setval{8b.grpo-nodiv.minus.base.minerva-abs}{9.9}
\setval{8b.grpo-nodiv.minus.base.olympiad-bench}{+12.1}
\setval{8b.grpo-nodiv.minus.base.olympiad-bench-abs}{12.1}
\setval{8b.grpo-nodiv.minus.base.writing}{+24.4}
\setval{8b.grpo-nodiv.minus.base.writing-abs}{24.4}
\setval{8b.grpo-nodiv.minus.base.writingbench}{+33.6}
\setval{8b.grpo-nodiv.minus.base.writingbench-abs}{33.6}
\setval{8b.grpo-nodiv.minus.concur-free.aime24}{+0.9}
\setval{8b.grpo-nodiv.minus.concur-free.aime24-abs}{0.9}
\setval{8b.grpo-nodiv.minus.concur-free.aime25}{$-$1.8}
\setval{8b.grpo-nodiv.minus.concur-free.aime25-abs}{1.8}
\setval{8b.grpo-nodiv.minus.concur-free.amc23}{$-$1.1}
\setval{8b.grpo-nodiv.minus.concur-free.amc23-abs}{1.1}
\setval{8b.grpo-nodiv.minus.concur-free.avg}{$-$2.2}
\setval{8b.grpo-nodiv.minus.concur-free.avg-abs}{2.2}
\setval{8b.grpo-nodiv.minus.concur-free.code}{$-$0.1}
\setval{8b.grpo-nodiv.minus.concur-free.code-abs}{0.1}
\setval{8b.grpo-nodiv.minus.concur-free.creative-writing-v3}{$-$5.0}
\setval{8b.grpo-nodiv.minus.concur-free.creative-writing-v3-abs}{5.0}
\setval{8b.grpo-nodiv.minus.concur-free.guidebench}{+1.9}
\setval{8b.grpo-nodiv.minus.concur-free.guidebench-abs}{1.9}
\setval{8b.grpo-nodiv.minus.concur-free.humaneval-plus}{+1.0}
\setval{8b.grpo-nodiv.minus.concur-free.humaneval-plus-abs}{1.0}
\setval{8b.grpo-nodiv.minus.concur-free.if}{$-$0.1}
\setval{8b.grpo-nodiv.minus.concur-free.if-abs}{0.1}
\setval{8b.grpo-nodiv.minus.concur-free.ifbench}{$-$2.3}
\setval{8b.grpo-nodiv.minus.concur-free.ifbench-abs}{2.3}
\setval{8b.grpo-nodiv.minus.concur-free.ifeval}{+0.2}
\setval{8b.grpo-nodiv.minus.concur-free.ifeval-abs}{0.2}
\setval{8b.grpo-nodiv.minus.concur-free.judge2.avg}{$-$1.2}
\setval{8b.grpo-nodiv.minus.concur-free.judge2.avg-abs}{1.2}
\setval{8b.grpo-nodiv.minus.concur-free.judge2.creative-writing-v3}{$-$5.0}
\setval{8b.grpo-nodiv.minus.concur-free.judge2.creative-writing-v3-abs}{5.0}
\setval{8b.grpo-nodiv.minus.concur-free.judge2.longbench-write}{$-$7.6}
\setval{8b.grpo-nodiv.minus.concur-free.judge2.longbench-write-abs}{7.6}
\setval{8b.grpo-nodiv.minus.concur-free.judge2.writing}{$-$6.5}
\setval{8b.grpo-nodiv.minus.concur-free.judge2.writing-abs}{6.5}
\setval{8b.grpo-nodiv.minus.concur-free.judge2.writingbench}{$-$6.8}
\setval{8b.grpo-nodiv.minus.concur-free.judge2.writingbench-abs}{6.8}
\setval{8b.grpo-nodiv.minus.concur-free.livecodebench}{$-$2.1}
\setval{8b.grpo-nodiv.minus.concur-free.livecodebench-abs}{2.1}
\setval{8b.grpo-nodiv.minus.concur-free.longbench-write}{$-$11.9}
\setval{8b.grpo-nodiv.minus.concur-free.longbench-write-abs}{11.9}
\setval{8b.grpo-nodiv.minus.concur-free.math}{+0.2}
\setval{8b.grpo-nodiv.minus.concur-free.math-abs}{0.2}
\setval{8b.grpo-nodiv.minus.concur-free.math500}{+0.7}
\setval{8b.grpo-nodiv.minus.concur-free.math500-abs}{0.7}
\setval{8b.grpo-nodiv.minus.concur-free.mbpp-plus}{+0.9}
\setval{8b.grpo-nodiv.minus.concur-free.mbpp-plus-abs}{0.9}
\setval{8b.grpo-nodiv.minus.concur-free.minerva}{+1.0}
\setval{8b.grpo-nodiv.minus.concur-free.minerva-abs}{1.0}
\setval{8b.grpo-nodiv.minus.concur-free.olympiad-bench}{+1.5}
\setval{8b.grpo-nodiv.minus.concur-free.olympiad-bench-abs}{1.5}
\setval{8b.grpo-nodiv.minus.concur-free.writing}{$-$11.1}
\setval{8b.grpo-nodiv.minus.concur-free.writing-abs}{11.1}
\setval{8b.grpo-nodiv.minus.concur-free.writingbench}{$-$16.4}
\setval{8b.grpo-nodiv.minus.concur-free.writingbench-abs}{16.4}
\setval{8b.grpo-nodiv.minus.concur-tied.aime24}{$-$10.6}
\setval{8b.grpo-nodiv.minus.concur-tied.aime24-abs}{10.6}
\setval{8b.grpo-nodiv.minus.concur-tied.aime25}{$-$9.2}
\setval{8b.grpo-nodiv.minus.concur-tied.aime25-abs}{9.2}
\setval{8b.grpo-nodiv.minus.concur-tied.amc23}{$-$14.9}
\setval{8b.grpo-nodiv.minus.concur-tied.amc23-abs}{14.9}
\setval{8b.grpo-nodiv.minus.concur-tied.avg}{$-$5.9}
\setval{8b.grpo-nodiv.minus.concur-tied.avg-abs}{5.9}
\setval{8b.grpo-nodiv.minus.concur-tied.code}{$-$0.7}
\setval{8b.grpo-nodiv.minus.concur-tied.code-abs}{0.7}
\setval{8b.grpo-nodiv.minus.concur-tied.creative-writing-v3}{$-$5.8}
\setval{8b.grpo-nodiv.minus.concur-tied.creative-writing-v3-abs}{5.8}
\setval{8b.grpo-nodiv.minus.concur-tied.guidebench}{+1.4}
\setval{8b.grpo-nodiv.minus.concur-tied.guidebench-abs}{1.4}
\setval{8b.grpo-nodiv.minus.concur-tied.humaneval-plus}{$-$1.0}
\setval{8b.grpo-nodiv.minus.concur-tied.humaneval-plus-abs}{1.0}
\setval{8b.grpo-nodiv.minus.concur-tied.if}{$-$0.3}
\setval{8b.grpo-nodiv.minus.concur-tied.if-abs}{0.3}
\setval{8b.grpo-nodiv.minus.concur-tied.ifbench}{$-$3.0}
\setval{8b.grpo-nodiv.minus.concur-tied.ifbench-abs}{3.0}
\setval{8b.grpo-nodiv.minus.concur-tied.ifeval}{+0.6}
\setval{8b.grpo-nodiv.minus.concur-tied.ifeval-abs}{0.6}
\setval{8b.grpo-nodiv.minus.concur-tied.judge2.avg}{$-$5.2}
\setval{8b.grpo-nodiv.minus.concur-tied.judge2.avg-abs}{5.2}
\setval{8b.grpo-nodiv.minus.concur-tied.judge2.creative-writing-v3}{$-$4.3}
\setval{8b.grpo-nodiv.minus.concur-tied.judge2.creative-writing-v3-abs}{4.3}
\setval{8b.grpo-nodiv.minus.concur-tied.judge2.longbench-write}{$-$8.2}
\setval{8b.grpo-nodiv.minus.concur-tied.judge2.longbench-write-abs}{8.2}
\setval{8b.grpo-nodiv.minus.concur-tied.judge2.writing}{$-$7.5}
\setval{8b.grpo-nodiv.minus.concur-tied.judge2.writing-abs}{7.5}
\setval{8b.grpo-nodiv.minus.concur-tied.judge2.writingbench}{$-$9.9}
\setval{8b.grpo-nodiv.minus.concur-tied.judge2.writingbench-abs}{9.9}
\setval{8b.grpo-nodiv.minus.concur-tied.livecodebench}{$-$2.2}
\setval{8b.grpo-nodiv.minus.concur-tied.livecodebench-abs}{2.2}
\setval{8b.grpo-nodiv.minus.concur-tied.longbench-write}{$-$10.2}
\setval{8b.grpo-nodiv.minus.concur-tied.longbench-write-abs}{10.2}
\setval{8b.grpo-nodiv.minus.concur-tied.math}{$-$8.8}
\setval{8b.grpo-nodiv.minus.concur-tied.math-abs}{8.8}
\setval{8b.grpo-nodiv.minus.concur-tied.math500}{$-$5.5}
\setval{8b.grpo-nodiv.minus.concur-tied.math500-abs}{5.5}
\setval{8b.grpo-nodiv.minus.concur-tied.mbpp-plus}{+1.2}
\setval{8b.grpo-nodiv.minus.concur-tied.mbpp-plus-abs}{1.2}
\setval{8b.grpo-nodiv.minus.concur-tied.minerva}{$-$4.6}
\setval{8b.grpo-nodiv.minus.concur-tied.minerva-abs}{4.6}
\setval{8b.grpo-nodiv.minus.concur-tied.olympiad-bench}{$-$8.0}
\setval{8b.grpo-nodiv.minus.concur-tied.olympiad-bench-abs}{8.0}
\setval{8b.grpo-nodiv.minus.concur-tied.writing}{$-$10.9}
\setval{8b.grpo-nodiv.minus.concur-tied.writing-abs}{10.9}
\setval{8b.grpo-nodiv.minus.concur-tied.writingbench}{$-$16.9}
\setval{8b.grpo-nodiv.minus.concur-tied.writingbench-abs}{16.9}
\setval{8b.grpo-nodiv.minus.concur.aime24}{$-$8.1}
\setval{8b.grpo-nodiv.minus.concur.aime24-abs}{8.1}
\setval{8b.grpo-nodiv.minus.concur.aime25}{$-$8.3}
\setval{8b.grpo-nodiv.minus.concur.aime25-abs}{8.3}
\setval{8b.grpo-nodiv.minus.concur.amc23}{$-$15.3}
\setval{8b.grpo-nodiv.minus.concur.amc23-abs}{15.3}
\setval{8b.grpo-nodiv.minus.concur.avg}{$-$5.1}
\setval{8b.grpo-nodiv.minus.concur.avg-abs}{5.1}
\setval{8b.grpo-nodiv.minus.concur.code}{+1.7}
\setval{8b.grpo-nodiv.minus.concur.code-abs}{1.7}
\setval{8b.grpo-nodiv.minus.concur.creative-writing-v3}{$-$3.3}
\setval{8b.grpo-nodiv.minus.concur.creative-writing-v3-abs}{3.3}
\setval{8b.grpo-nodiv.minus.concur.guidebench}{+1.4}
\setval{8b.grpo-nodiv.minus.concur.guidebench-abs}{1.4}
\setval{8b.grpo-nodiv.minus.concur.humaneval-plus}{+3.6}
\setval{8b.grpo-nodiv.minus.concur.humaneval-plus-abs}{3.6}
\setval{8b.grpo-nodiv.minus.concur.if}{+0.1}
\setval{8b.grpo-nodiv.minus.concur.if-abs}{0.1}
\setval{8b.grpo-nodiv.minus.concur.ifbench}{$-$2.3}
\setval{8b.grpo-nodiv.minus.concur.ifbench-abs}{2.3}
\setval{8b.grpo-nodiv.minus.concur.ifeval}{+1.3}
\setval{8b.grpo-nodiv.minus.concur.ifeval-abs}{1.3}
\setval{8b.grpo-nodiv.minus.concur.judge2.avg}{$-$4.3}
\setval{8b.grpo-nodiv.minus.concur.judge2.avg-abs}{4.3}
\setval{8b.grpo-nodiv.minus.concur.judge2.creative-writing-v3}{$-$0.3}
\setval{8b.grpo-nodiv.minus.concur.judge2.creative-writing-v3-abs}{0.3}
\setval{8b.grpo-nodiv.minus.concur.judge2.longbench-write}{$-$8.9}
\setval{8b.grpo-nodiv.minus.concur.judge2.longbench-write-abs}{8.9}
\setval{8b.grpo-nodiv.minus.concur.judge2.writing}{$-$6.4}
\setval{8b.grpo-nodiv.minus.concur.judge2.writing-abs}{6.4}
\setval{8b.grpo-nodiv.minus.concur.judge2.writingbench}{$-$10.0}
\setval{8b.grpo-nodiv.minus.concur.judge2.writingbench-abs}{10.0}
\setval{8b.grpo-nodiv.minus.concur.livecodebench}{$-$1.4}
\setval{8b.grpo-nodiv.minus.concur.livecodebench-abs}{1.4}
\setval{8b.grpo-nodiv.minus.concur.longbench-write}{$-$10.5}
\setval{8b.grpo-nodiv.minus.concur.longbench-write-abs}{10.5}
\setval{8b.grpo-nodiv.minus.concur.math}{$-$8.3}
\setval{8b.grpo-nodiv.minus.concur.math-abs}{8.3}
\setval{8b.grpo-nodiv.minus.concur.math500}{$-$6.0}
\setval{8b.grpo-nodiv.minus.concur.math500-abs}{6.0}
\setval{8b.grpo-nodiv.minus.concur.mbpp-plus}{+2.8}
\setval{8b.grpo-nodiv.minus.concur.mbpp-plus-abs}{2.8}
\setval{8b.grpo-nodiv.minus.concur.minerva}{$-$3.4}
\setval{8b.grpo-nodiv.minus.concur.minerva-abs}{3.4}
\setval{8b.grpo-nodiv.minus.concur.olympiad-bench}{$-$8.9}
\setval{8b.grpo-nodiv.minus.concur.olympiad-bench-abs}{8.9}
\setval{8b.grpo-nodiv.minus.concur.writing}{$-$10.5}
\setval{8b.grpo-nodiv.minus.concur.writing-abs}{10.5}
\setval{8b.grpo-nodiv.minus.concur.writingbench}{$-$17.7}
\setval{8b.grpo-nodiv.minus.concur.writingbench-abs}{17.7}
\setval{8b.grpo-nodiv.minus.gd2po.aime24}{+4.4}
\setval{8b.grpo-nodiv.minus.gd2po.aime24-abs}{4.4}
\setval{8b.grpo-nodiv.minus.gd2po.aime25}{+2.6}
\setval{8b.grpo-nodiv.minus.gd2po.aime25-abs}{2.6}
\setval{8b.grpo-nodiv.minus.gd2po.amc23}{+4.0}
\setval{8b.grpo-nodiv.minus.gd2po.amc23-abs}{4.0}
\setval{8b.grpo-nodiv.minus.gd2po.avg}{+0.7}
\setval{8b.grpo-nodiv.minus.gd2po.avg-abs}{0.7}
\setval{8b.grpo-nodiv.minus.gd2po.code}{+0.7}
\setval{8b.grpo-nodiv.minus.gd2po.code-abs}{0.7}
\setval{8b.grpo-nodiv.minus.gd2po.creative-writing-v3}{$-$1.5}
\setval{8b.grpo-nodiv.minus.gd2po.creative-writing-v3-abs}{1.5}
\setval{8b.grpo-nodiv.minus.gd2po.guidebench}{+1.2}
\setval{8b.grpo-nodiv.minus.gd2po.guidebench-abs}{1.2}
\setval{8b.grpo-nodiv.minus.gd2po.humaneval-plus}{+2.4}
\setval{8b.grpo-nodiv.minus.gd2po.humaneval-plus-abs}{2.4}
\setval{8b.grpo-nodiv.minus.gd2po.if}{$-$1.8}
\setval{8b.grpo-nodiv.minus.gd2po.if-abs}{1.8}
\setval{8b.grpo-nodiv.minus.gd2po.ifbench}{$-$6.7}
\setval{8b.grpo-nodiv.minus.gd2po.ifbench-abs}{6.7}
\setval{8b.grpo-nodiv.minus.gd2po.ifeval}{+0.1}
\setval{8b.grpo-nodiv.minus.gd2po.ifeval-abs}{0.1}
\setval{8b.grpo-nodiv.minus.gd2po.judge2.avg}{+1.2}
\setval{8b.grpo-nodiv.minus.gd2po.judge2.avg-abs}{1.2}
\setval{8b.grpo-nodiv.minus.gd2po.judge2.creative-writing-v3}{+0.8}
\setval{8b.grpo-nodiv.minus.gd2po.judge2.creative-writing-v3-abs}{0.8}
\setval{8b.grpo-nodiv.minus.gd2po.judge2.longbench-write}{+4.4}
\setval{8b.grpo-nodiv.minus.gd2po.judge2.longbench-write-abs}{4.4}
\setval{8b.grpo-nodiv.minus.gd2po.judge2.writing}{+1.3}
\setval{8b.grpo-nodiv.minus.gd2po.judge2.writing-abs}{1.3}
\setval{8b.grpo-nodiv.minus.gd2po.judge2.writingbench}{$-$1.3}
\setval{8b.grpo-nodiv.minus.gd2po.judge2.writingbench-abs}{1.3}
\setval{8b.grpo-nodiv.minus.gd2po.livecodebench}{$-$1.3}
\setval{8b.grpo-nodiv.minus.gd2po.livecodebench-abs}{1.3}
\setval{8b.grpo-nodiv.minus.gd2po.longbench-write}{+1.2}
\setval{8b.grpo-nodiv.minus.gd2po.longbench-write-abs}{1.2}
\setval{8b.grpo-nodiv.minus.gd2po.math}{+2.9}
\setval{8b.grpo-nodiv.minus.gd2po.math-abs}{2.9}
\setval{8b.grpo-nodiv.minus.gd2po.math500}{+2.9}
\setval{8b.grpo-nodiv.minus.gd2po.math500-abs}{2.9}
\setval{8b.grpo-nodiv.minus.gd2po.mbpp-plus}{+1.0}
\setval{8b.grpo-nodiv.minus.gd2po.mbpp-plus-abs}{1.0}
\setval{8b.grpo-nodiv.minus.gd2po.minerva}{+0.9}
\setval{8b.grpo-nodiv.minus.gd2po.minerva-abs}{0.9}
\setval{8b.grpo-nodiv.minus.gd2po.olympiad-bench}{+2.9}
\setval{8b.grpo-nodiv.minus.gd2po.olympiad-bench-abs}{2.9}
\setval{8b.grpo-nodiv.minus.gd2po.writing}{$-$1.3}
\setval{8b.grpo-nodiv.minus.gd2po.writing-abs}{1.3}
\setval{8b.grpo-nodiv.minus.gd2po.writingbench}{$-$3.6}
\setval{8b.grpo-nodiv.minus.gd2po.writingbench-abs}{3.6}
\setval{8b.grpo-nodiv.minus.gdpo-noaux.aime24}{$-$7.3}
\setval{8b.grpo-nodiv.minus.gdpo-noaux.aime24-abs}{7.3}
\setval{8b.grpo-nodiv.minus.gdpo-noaux.aime25}{$-$9.2}
\setval{8b.grpo-nodiv.minus.gdpo-noaux.aime25-abs}{9.2}
\setval{8b.grpo-nodiv.minus.gdpo-noaux.amc23}{$-$13.5}
\setval{8b.grpo-nodiv.minus.gdpo-noaux.amc23-abs}{13.5}
\setval{8b.grpo-nodiv.minus.gdpo-noaux.avg}{$-$2.4}
\setval{8b.grpo-nodiv.minus.gdpo-noaux.avg-abs}{2.4}
\setval{8b.grpo-nodiv.minus.gdpo-noaux.code}{$-$0.3}
\setval{8b.grpo-nodiv.minus.gdpo-noaux.code-abs}{0.3}
\setval{8b.grpo-nodiv.minus.gdpo-noaux.creative-writing-v3}{+2.8}
\setval{8b.grpo-nodiv.minus.gdpo-noaux.creative-writing-v3-abs}{2.8}
\setval{8b.grpo-nodiv.minus.gdpo-noaux.guidebench}{+3.5}
\setval{8b.grpo-nodiv.minus.gdpo-noaux.guidebench-abs}{3.5}
\setval{8b.grpo-nodiv.minus.gdpo-noaux.humaneval-plus}{+0.4}
\setval{8b.grpo-nodiv.minus.gdpo-noaux.humaneval-plus-abs}{0.4}
\setval{8b.grpo-nodiv.minus.gdpo-noaux.if}{+3.0}
\setval{8b.grpo-nodiv.minus.gdpo-noaux.if-abs}{3.0}
\setval{8b.grpo-nodiv.minus.gdpo-noaux.ifbench}{+2.7}
\setval{8b.grpo-nodiv.minus.gdpo-noaux.ifbench-abs}{2.7}
\setval{8b.grpo-nodiv.minus.gdpo-noaux.ifeval}{+2.9}
\setval{8b.grpo-nodiv.minus.gdpo-noaux.ifeval-abs}{2.9}
\setval{8b.grpo-nodiv.minus.gdpo-noaux.judge2.avg}{$-$1.0}
\setval{8b.grpo-nodiv.minus.gdpo-noaux.judge2.avg-abs}{1.0}
\setval{8b.grpo-nodiv.minus.gdpo-noaux.judge2.creative-writing-v3}{+8.6}
\setval{8b.grpo-nodiv.minus.gdpo-noaux.judge2.creative-writing-v3-abs}{8.6}
\setval{8b.grpo-nodiv.minus.gdpo-noaux.judge2.longbench-write}{+12.1}
\setval{8b.grpo-nodiv.minus.gdpo-noaux.judge2.longbench-write-abs}{12.1}
\setval{8b.grpo-nodiv.minus.gdpo-noaux.judge2.writing}{+8.2}
\setval{8b.grpo-nodiv.minus.gdpo-noaux.judge2.writing-abs}{8.2}
\setval{8b.grpo-nodiv.minus.gdpo-noaux.judge2.writingbench}{+3.8}
\setval{8b.grpo-nodiv.minus.gdpo-noaux.judge2.writingbench-abs}{3.8}
\setval{8b.grpo-nodiv.minus.gdpo-noaux.livecodebench}{$-$2.4}
\setval{8b.grpo-nodiv.minus.gdpo-noaux.livecodebench-abs}{2.4}
\setval{8b.grpo-nodiv.minus.gdpo-noaux.longbench-write}{+5.1}
\setval{8b.grpo-nodiv.minus.gdpo-noaux.longbench-write-abs}{5.1}
\setval{8b.grpo-nodiv.minus.gdpo-noaux.math}{$-$7.9}
\setval{8b.grpo-nodiv.minus.gdpo-noaux.math-abs}{7.9}
\setval{8b.grpo-nodiv.minus.gdpo-noaux.math500}{$-$6.1}
\setval{8b.grpo-nodiv.minus.gdpo-noaux.math500-abs}{6.1}
\setval{8b.grpo-nodiv.minus.gdpo-noaux.mbpp-plus}{+1.2}
\setval{8b.grpo-nodiv.minus.gdpo-noaux.mbpp-plus-abs}{1.2}
\setval{8b.grpo-nodiv.minus.gdpo-noaux.minerva}{$-$3.2}
\setval{8b.grpo-nodiv.minus.gdpo-noaux.minerva-abs}{3.2}
\setval{8b.grpo-nodiv.minus.gdpo-noaux.olympiad-bench}{$-$8.4}
\setval{8b.grpo-nodiv.minus.gdpo-noaux.olympiad-bench-abs}{8.4}
\setval{8b.grpo-nodiv.minus.gdpo-noaux.writing}{+1.1}
\setval{8b.grpo-nodiv.minus.gdpo-noaux.writing-abs}{1.1}
\setval{8b.grpo-nodiv.minus.gdpo-noaux.writingbench}{$-$4.7}
\setval{8b.grpo-nodiv.minus.gdpo-noaux.writingbench-abs}{4.7}
\setval{8b.grpo-nodiv.minus.gdpo-w0.5.aime24}{+1.4}
\setval{8b.grpo-nodiv.minus.gdpo-w0.5.aime24-abs}{1.4}
\setval{8b.grpo-nodiv.minus.gdpo-w0.5.aime25}{$-$0.6}
\setval{8b.grpo-nodiv.minus.gdpo-w0.5.aime25-abs}{0.6}
\setval{8b.grpo-nodiv.minus.gdpo-w0.5.amc23}{+0.2}
\setval{8b.grpo-nodiv.minus.gdpo-w0.5.amc23-abs}{0.2}
\setval{8b.grpo-nodiv.minus.gdpo-w0.5.avg}{$-$0.2}
\setval{8b.grpo-nodiv.minus.gdpo-w0.5.avg-abs}{0.2}
\setval{8b.grpo-nodiv.minus.gdpo-w0.5.code}{+1.3}
\setval{8b.grpo-nodiv.minus.gdpo-w0.5.code-abs}{1.3}
\setval{8b.grpo-nodiv.minus.gdpo-w0.5.creative-writing-v3}{$-$4.9}
\setval{8b.grpo-nodiv.minus.gdpo-w0.5.creative-writing-v3-abs}{4.9}
\setval{8b.grpo-nodiv.minus.gdpo-w0.5.guidebench}{+0.9}
\setval{8b.grpo-nodiv.minus.gdpo-w0.5.guidebench-abs}{0.9}
\setval{8b.grpo-nodiv.minus.gdpo-w0.5.humaneval-plus}{+2.6}
\setval{8b.grpo-nodiv.minus.gdpo-w0.5.humaneval-plus-abs}{2.6}
\setval{8b.grpo-nodiv.minus.gdpo-w0.5.if}{+0.4}
\setval{8b.grpo-nodiv.minus.gdpo-w0.5.if-abs}{0.4}
\setval{8b.grpo-nodiv.minus.gdpo-w0.5.ifbench}{$-$1.3}
\setval{8b.grpo-nodiv.minus.gdpo-w0.5.ifbench-abs}{1.3}
\setval{8b.grpo-nodiv.minus.gdpo-w0.5.ifeval}{+1.7}
\setval{8b.grpo-nodiv.minus.gdpo-w0.5.ifeval-abs}{1.7}
\setval{8b.grpo-nodiv.minus.gdpo-w0.5.judge2.avg}{+0.2}
\setval{8b.grpo-nodiv.minus.gdpo-w0.5.judge2.avg-abs}{0.2}
\setval{8b.grpo-nodiv.minus.gdpo-w0.5.judge2.creative-writing-v3}{$-$2.9}
\setval{8b.grpo-nodiv.minus.gdpo-w0.5.judge2.creative-writing-v3-abs}{2.9}
\setval{8b.grpo-nodiv.minus.gdpo-w0.5.judge2.longbench-write}{+2.9}
\setval{8b.grpo-nodiv.minus.gdpo-w0.5.judge2.longbench-write-abs}{2.9}
\setval{8b.grpo-nodiv.minus.gdpo-w0.5.judge2.writing}{$-$0.8}
\setval{8b.grpo-nodiv.minus.gdpo-w0.5.judge2.writing-abs}{0.8}
\setval{8b.grpo-nodiv.minus.gdpo-w0.5.judge2.writingbench}{$-$2.3}
\setval{8b.grpo-nodiv.minus.gdpo-w0.5.judge2.writingbench-abs}{2.3}
\setval{8b.grpo-nodiv.minus.gdpo-w0.5.livecodebench}{$-$1.8}
\setval{8b.grpo-nodiv.minus.gdpo-w0.5.livecodebench-abs}{1.8}
\setval{8b.grpo-nodiv.minus.gdpo-w0.5.longbench-write}{$-$0.1}
\setval{8b.grpo-nodiv.minus.gdpo-w0.5.longbench-write-abs}{0.1}
\setval{8b.grpo-nodiv.minus.gdpo-w0.5.math}{+0.2}
\setval{8b.grpo-nodiv.minus.gdpo-w0.5.math-abs}{0.2}
\setval{8b.grpo-nodiv.minus.gdpo-w0.5.math500}{+0.1}
\setval{8b.grpo-nodiv.minus.gdpo-w0.5.math500-abs}{0.1}
\setval{8b.grpo-nodiv.minus.gdpo-w0.5.mbpp-plus}{+3.2}
\setval{8b.grpo-nodiv.minus.gdpo-w0.5.mbpp-plus-abs}{3.2}
\setval{8b.grpo-nodiv.minus.gdpo-w0.5.minerva}{$-$0.4}
\setval{8b.grpo-nodiv.minus.gdpo-w0.5.minerva-abs}{0.4}
\setval{8b.grpo-nodiv.minus.gdpo-w0.5.olympiad-bench}{+0.4}
\setval{8b.grpo-nodiv.minus.gdpo-w0.5.olympiad-bench-abs}{0.4}
\setval{8b.grpo-nodiv.minus.gdpo-w0.5.writing}{$-$3.2}
\setval{8b.grpo-nodiv.minus.gdpo-w0.5.writing-abs}{3.2}
\setval{8b.grpo-nodiv.minus.gdpo-w0.5.writingbench}{$-$4.7}
\setval{8b.grpo-nodiv.minus.gdpo-w0.5.writingbench-abs}{4.7}
\setval{8b.grpo-nodiv.minus.gdpo.aime24}{$-$0.7}
\setval{8b.grpo-nodiv.minus.gdpo.aime24-abs}{0.7}
\setval{8b.grpo-nodiv.minus.gdpo.aime25}{+0.4}
\setval{8b.grpo-nodiv.minus.gdpo.aime25-abs}{0.4}
\setval{8b.grpo-nodiv.minus.gdpo.amc23}{+1.4}
\setval{8b.grpo-nodiv.minus.gdpo.amc23-abs}{1.4}
\setval{8b.grpo-nodiv.minus.gdpo.avg}{$-$0.3}
\setval{8b.grpo-nodiv.minus.gdpo.avg-abs}{0.3}
\setval{8b.grpo-nodiv.minus.gdpo.code}{+0.8}
\setval{8b.grpo-nodiv.minus.gdpo.code-abs}{0.8}
\setval{8b.grpo-nodiv.minus.gdpo.creative-writing-v3}{$-$6.2}
\setval{8b.grpo-nodiv.minus.gdpo.creative-writing-v3-abs}{6.2}
\setval{8b.grpo-nodiv.minus.gdpo.guidebench}{+0.3}
\setval{8b.grpo-nodiv.minus.gdpo.guidebench-abs}{0.3}
\setval{8b.grpo-nodiv.minus.gdpo.humaneval-plus}{+2.3}
\setval{8b.grpo-nodiv.minus.gdpo.humaneval-plus-abs}{2.3}
\setval{8b.grpo-nodiv.minus.gdpo.if}{$-$0.5}
\setval{8b.grpo-nodiv.minus.gdpo.if-abs}{0.5}
\setval{8b.grpo-nodiv.minus.gdpo.ifbench}{$-$3.0}
\setval{8b.grpo-nodiv.minus.gdpo.ifbench-abs}{3.0}
\setval{8b.grpo-nodiv.minus.gdpo.ifeval}{+1.3}
\setval{8b.grpo-nodiv.minus.gdpo.ifeval-abs}{1.3}
\setval{8b.grpo-nodiv.minus.gdpo.judge2.avg}{+0.1}
\setval{8b.grpo-nodiv.minus.gdpo.judge2.avg-abs}{0.1}
\setval{8b.grpo-nodiv.minus.gdpo.judge2.creative-writing-v3}{$-$3.5}
\setval{8b.grpo-nodiv.minus.gdpo.judge2.creative-writing-v3-abs}{3.5}
\setval{8b.grpo-nodiv.minus.gdpo.judge2.longbench-write}{+2.4}
\setval{8b.grpo-nodiv.minus.gdpo.judge2.longbench-write-abs}{2.4}
\setval{8b.grpo-nodiv.minus.gdpo.judge2.writing}{$-$1.9}
\setval{8b.grpo-nodiv.minus.gdpo.judge2.writing-abs}{1.9}
\setval{8b.grpo-nodiv.minus.gdpo.judge2.writingbench}{$-$4.5}
\setval{8b.grpo-nodiv.minus.gdpo.judge2.writingbench-abs}{4.5}
\setval{8b.grpo-nodiv.minus.gdpo.livecodebench}{$-$1.3}
\setval{8b.grpo-nodiv.minus.gdpo.livecodebench-abs}{1.3}
\setval{8b.grpo-nodiv.minus.gdpo.longbench-write}{+1.5}
\setval{8b.grpo-nodiv.minus.gdpo.longbench-write-abs}{1.5}
\setval{8b.grpo-nodiv.minus.gdpo.math}{+0.9}
\setval{8b.grpo-nodiv.minus.gdpo.math-abs}{0.9}
\setval{8b.grpo-nodiv.minus.gdpo.math500}{+1.1}
\setval{8b.grpo-nodiv.minus.gdpo.math500-abs}{1.1}
\setval{8b.grpo-nodiv.minus.gdpo.mbpp-plus}{+1.5}
\setval{8b.grpo-nodiv.minus.gdpo.mbpp-plus-abs}{1.5}
\setval{8b.grpo-nodiv.minus.gdpo.minerva}{+1.2}
\setval{8b.grpo-nodiv.minus.gdpo.minerva-abs}{1.2}
\setval{8b.grpo-nodiv.minus.gdpo.olympiad-bench}{+2.2}
\setval{8b.grpo-nodiv.minus.gdpo.olympiad-bench-abs}{2.2}
\setval{8b.grpo-nodiv.minus.gdpo.writing}{$-$3.5}
\setval{8b.grpo-nodiv.minus.gdpo.writing-abs}{3.5}
\setval{8b.grpo-nodiv.minus.gdpo.writingbench}{$-$5.8}
\setval{8b.grpo-nodiv.minus.gdpo.writingbench-abs}{5.8}
\setval{8b.grpo-nodiv.minus.geoma.aime24}{+6.6}
\setval{8b.grpo-nodiv.minus.geoma.aime24-abs}{6.6}
\setval{8b.grpo-nodiv.minus.geoma.aime25}{+2.3}
\setval{8b.grpo-nodiv.minus.geoma.aime25-abs}{2.3}
\setval{8b.grpo-nodiv.minus.geoma.amc23}{+0.1}
\setval{8b.grpo-nodiv.minus.geoma.amc23-abs}{0.1}
\setval{8b.grpo-nodiv.minus.geoma.avg}{+1.4}
\setval{8b.grpo-nodiv.minus.geoma.avg-abs}{1.4}
\setval{8b.grpo-nodiv.minus.geoma.code}{+1.1}
\setval{8b.grpo-nodiv.minus.geoma.code-abs}{1.1}
\setval{8b.grpo-nodiv.minus.geoma.creative-writing-v3}{$-$3.7}
\setval{8b.grpo-nodiv.minus.geoma.creative-writing-v3-abs}{3.7}
\setval{8b.grpo-nodiv.minus.geoma.guidebench}{+0.5}
\setval{8b.grpo-nodiv.minus.geoma.guidebench-abs}{0.5}
\setval{8b.grpo-nodiv.minus.geoma.humaneval-plus}{+3.6}
\setval{8b.grpo-nodiv.minus.geoma.humaneval-plus-abs}{3.6}
\setval{8b.grpo-nodiv.minus.geoma.if}{+1.7}
\setval{8b.grpo-nodiv.minus.geoma.if-abs}{1.7}
\setval{8b.grpo-nodiv.minus.geoma.ifbench}{+2.3}
\setval{8b.grpo-nodiv.minus.geoma.ifbench-abs}{2.3}
\setval{8b.grpo-nodiv.minus.geoma.ifeval}{+2.2}
\setval{8b.grpo-nodiv.minus.geoma.ifeval-abs}{2.2}
\setval{8b.grpo-nodiv.minus.geoma.judge2.avg}{+1.7}
\setval{8b.grpo-nodiv.minus.geoma.judge2.avg-abs}{1.7}
\setval{8b.grpo-nodiv.minus.geoma.judge2.creative-writing-v3}{$-$0.1}
\setval{8b.grpo-nodiv.minus.geoma.judge2.creative-writing-v3-abs}{0.1}
\setval{8b.grpo-nodiv.minus.geoma.judge2.longbench-write}{+1.7}
\setval{8b.grpo-nodiv.minus.geoma.judge2.longbench-write-abs}{1.7}
\setval{8b.grpo-nodiv.minus.geoma.judge2.writing}{$-$0.9}
\setval{8b.grpo-nodiv.minus.geoma.judge2.writing-abs}{0.9}
\setval{8b.grpo-nodiv.minus.geoma.judge2.writingbench}{$-$4.4}
\setval{8b.grpo-nodiv.minus.geoma.judge2.writingbench-abs}{4.4}
\setval{8b.grpo-nodiv.minus.geoma.livecodebench}{$-$0.7}
\setval{8b.grpo-nodiv.minus.geoma.livecodebench-abs}{0.7}
\setval{8b.grpo-nodiv.minus.geoma.longbench-write}{+2.1}
\setval{8b.grpo-nodiv.minus.geoma.longbench-write-abs}{2.1}
\setval{8b.grpo-nodiv.minus.geoma.math}{+3.3}
\setval{8b.grpo-nodiv.minus.geoma.math-abs}{3.3}
\setval{8b.grpo-nodiv.minus.geoma.math500}{+4.5}
\setval{8b.grpo-nodiv.minus.geoma.math500-abs}{4.5}
\setval{8b.grpo-nodiv.minus.geoma.mbpp-plus}{+0.4}
\setval{8b.grpo-nodiv.minus.geoma.mbpp-plus-abs}{0.4}
\setval{8b.grpo-nodiv.minus.geoma.minerva}{+1.0}
\setval{8b.grpo-nodiv.minus.geoma.minerva-abs}{1.0}
\setval{8b.grpo-nodiv.minus.geoma.olympiad-bench}{+5.5}
\setval{8b.grpo-nodiv.minus.geoma.olympiad-bench-abs}{5.5}
\setval{8b.grpo-nodiv.minus.geoma.writing}{$-$2.4}
\setval{8b.grpo-nodiv.minus.geoma.writing-abs}{2.4}
\setval{8b.grpo-nodiv.minus.geoma.writingbench}{$-$5.6}
\setval{8b.grpo-nodiv.minus.geoma.writingbench-abs}{5.6}
\setval{8b.grpo-nodiv.minus.grpo.aime24}{+4.8}
\setval{8b.grpo-nodiv.minus.grpo.aime24-abs}{4.8}
\setval{8b.grpo-nodiv.minus.grpo.aime25}{$-$0.7}
\setval{8b.grpo-nodiv.minus.grpo.aime25-abs}{0.7}
\setval{8b.grpo-nodiv.minus.grpo.amc23}{$-$0.8}
\setval{8b.grpo-nodiv.minus.grpo.amc23-abs}{0.8}
\setval{8b.grpo-nodiv.minus.grpo.avg}{+0.2}
\setval{8b.grpo-nodiv.minus.grpo.avg-abs}{0.2}
\setval{8b.grpo-nodiv.minus.grpo.code}{+0.8}
\setval{8b.grpo-nodiv.minus.grpo.code-abs}{0.8}
\setval{8b.grpo-nodiv.minus.grpo.creative-writing-v3}{$-$5.8}
\setval{8b.grpo-nodiv.minus.grpo.creative-writing-v3-abs}{5.8}
\setval{8b.grpo-nodiv.minus.grpo.guidebench}{+3.0}
\setval{8b.grpo-nodiv.minus.grpo.guidebench-abs}{3.0}
\setval{8b.grpo-nodiv.minus.grpo.humaneval-plus}{+2.6}
\setval{8b.grpo-nodiv.minus.grpo.humaneval-plus-abs}{2.6}
\setval{8b.grpo-nodiv.minus.grpo.if}{+0.0}
\setval{8b.grpo-nodiv.minus.grpo.if-abs}{0.0}
\setval{8b.grpo-nodiv.minus.grpo.ifbench}{$-$5.7}
\setval{8b.grpo-nodiv.minus.grpo.ifbench-abs}{5.7}
\setval{8b.grpo-nodiv.minus.grpo.ifeval}{+2.6}
\setval{8b.grpo-nodiv.minus.grpo.ifeval-abs}{2.6}
\setval{8b.grpo-nodiv.minus.grpo.judge2.avg}{+0.9}
\setval{8b.grpo-nodiv.minus.grpo.judge2.avg-abs}{0.9}
\setval{8b.grpo-nodiv.minus.grpo.judge2.creative-writing-v3}{$-$1.8}
\setval{8b.grpo-nodiv.minus.grpo.judge2.creative-writing-v3-abs}{1.8}
\setval{8b.grpo-nodiv.minus.grpo.judge2.longbench-write}{+4.1}
\setval{8b.grpo-nodiv.minus.grpo.judge2.longbench-write-abs}{4.1}
\setval{8b.grpo-nodiv.minus.grpo.judge2.writing}{$-$0.2}
\setval{8b.grpo-nodiv.minus.grpo.judge2.writing-abs}{0.2}
\setval{8b.grpo-nodiv.minus.grpo.judge2.writingbench}{$-$3.0}
\setval{8b.grpo-nodiv.minus.grpo.judge2.writingbench-abs}{3.0}
\setval{8b.grpo-nodiv.minus.grpo.livecodebench}{$-$1.5}
\setval{8b.grpo-nodiv.minus.grpo.livecodebench-abs}{1.5}
\setval{8b.grpo-nodiv.minus.grpo.longbench-write}{+1.3}
\setval{8b.grpo-nodiv.minus.grpo.longbench-write-abs}{1.3}
\setval{8b.grpo-nodiv.minus.grpo.math}{+1.9}
\setval{8b.grpo-nodiv.minus.grpo.math-abs}{1.9}
\setval{8b.grpo-nodiv.minus.grpo.math500}{+3.5}
\setval{8b.grpo-nodiv.minus.grpo.math500-abs}{3.5}
\setval{8b.grpo-nodiv.minus.grpo.mbpp-plus}{+1.4}
\setval{8b.grpo-nodiv.minus.grpo.mbpp-plus-abs}{1.4}
\setval{8b.grpo-nodiv.minus.grpo.minerva}{+0.9}
\setval{8b.grpo-nodiv.minus.grpo.minerva-abs}{0.9}
\setval{8b.grpo-nodiv.minus.grpo.olympiad-bench}{+3.6}
\setval{8b.grpo-nodiv.minus.grpo.olympiad-bench-abs}{3.6}
\setval{8b.grpo-nodiv.minus.grpo.writing}{$-$3.6}
\setval{8b.grpo-nodiv.minus.grpo.writing-abs}{3.6}
\setval{8b.grpo-nodiv.minus.grpo.writingbench}{$-$6.1}
\setval{8b.grpo-nodiv.minus.grpo.writingbench-abs}{6.1}
\setval{8b.grpo-nodiv.minus.mt-grpo.aime24}{+6.1}
\setval{8b.grpo-nodiv.minus.mt-grpo.aime24-abs}{6.1}
\setval{8b.grpo-nodiv.minus.mt-grpo.aime25}{$-$0.9}
\setval{8b.grpo-nodiv.minus.mt-grpo.aime25-abs}{0.9}
\setval{8b.grpo-nodiv.minus.mt-grpo.amc23}{$-$2.0}
\setval{8b.grpo-nodiv.minus.mt-grpo.amc23-abs}{2.0}
\setval{8b.grpo-nodiv.minus.mt-grpo.avg}{+0.9}
\setval{8b.grpo-nodiv.minus.mt-grpo.avg-abs}{0.9}
\setval{8b.grpo-nodiv.minus.mt-grpo.code}{$-$0.2}
\setval{8b.grpo-nodiv.minus.mt-grpo.code-abs}{0.2}
\setval{8b.grpo-nodiv.minus.mt-grpo.creative-writing-v3}{$-$3.7}
\setval{8b.grpo-nodiv.minus.mt-grpo.creative-writing-v3-abs}{3.7}
\setval{8b.grpo-nodiv.minus.mt-grpo.guidebench}{+1.7}
\setval{8b.grpo-nodiv.minus.mt-grpo.guidebench-abs}{1.7}
\setval{8b.grpo-nodiv.minus.mt-grpo.humaneval-plus}{+1.3}
\setval{8b.grpo-nodiv.minus.mt-grpo.humaneval-plus-abs}{1.3}
\setval{8b.grpo-nodiv.minus.mt-grpo.if}{+1.1}
\setval{8b.grpo-nodiv.minus.mt-grpo.if-abs}{1.1}
\setval{8b.grpo-nodiv.minus.mt-grpo.ifbench}{$-$1.0}
\setval{8b.grpo-nodiv.minus.mt-grpo.ifbench-abs}{1.0}
\setval{8b.grpo-nodiv.minus.mt-grpo.ifeval}{+2.5}
\setval{8b.grpo-nodiv.minus.mt-grpo.ifeval-abs}{2.5}
\setval{8b.grpo-nodiv.minus.mt-grpo.judge2.avg}{+0.9}
\setval{8b.grpo-nodiv.minus.mt-grpo.judge2.avg-abs}{0.9}
\setval{8b.grpo-nodiv.minus.mt-grpo.judge2.creative-writing-v3}{$-$4.0}
\setval{8b.grpo-nodiv.minus.mt-grpo.judge2.creative-writing-v3-abs}{4.0}
\setval{8b.grpo-nodiv.minus.mt-grpo.judge2.longbench-write}{+4.1}
\setval{8b.grpo-nodiv.minus.mt-grpo.judge2.longbench-write-abs}{4.1}
\setval{8b.grpo-nodiv.minus.mt-grpo.judge2.writing}{$-$0.5}
\setval{8b.grpo-nodiv.minus.mt-grpo.judge2.writing-abs}{0.5}
\setval{8b.grpo-nodiv.minus.mt-grpo.judge2.writingbench}{$-$1.6}
\setval{8b.grpo-nodiv.minus.mt-grpo.judge2.writingbench-abs}{1.6}
\setval{8b.grpo-nodiv.minus.mt-grpo.livecodebench}{$-$2.7}
\setval{8b.grpo-nodiv.minus.mt-grpo.livecodebench-abs}{2.7}
\setval{8b.grpo-nodiv.minus.mt-grpo.longbench-write}{+3.6}
\setval{8b.grpo-nodiv.minus.mt-grpo.longbench-write-abs}{3.6}
\setval{8b.grpo-nodiv.minus.mt-grpo.math}{+2.0}
\setval{8b.grpo-nodiv.minus.mt-grpo.math-abs}{2.0}
\setval{8b.grpo-nodiv.minus.mt-grpo.math500}{+3.5}
\setval{8b.grpo-nodiv.minus.mt-grpo.math500-abs}{3.5}
\setval{8b.grpo-nodiv.minus.mt-grpo.mbpp-plus}{+0.8}
\setval{8b.grpo-nodiv.minus.mt-grpo.mbpp-plus-abs}{0.8}
\setval{8b.grpo-nodiv.minus.mt-grpo.minerva}{$-$0.2}
\setval{8b.grpo-nodiv.minus.mt-grpo.minerva-abs}{0.2}
\setval{8b.grpo-nodiv.minus.mt-grpo.olympiad-bench}{+5.7}
\setval{8b.grpo-nodiv.minus.mt-grpo.olympiad-bench-abs}{5.7}
\setval{8b.grpo-nodiv.minus.mt-grpo.writing}{$-$0.5}
\setval{8b.grpo-nodiv.minus.mt-grpo.writing-abs}{0.5}
\setval{8b.grpo-nodiv.minus.mt-grpo.writingbench}{$-$1.4}
\setval{8b.grpo-nodiv.minus.mt-grpo.writingbench-abs}{1.4}
\setval{8b.grpo-nodiv.minus.papo.aime24}{+8.4}
\setval{8b.grpo-nodiv.minus.papo.aime24-abs}{8.4}
\setval{8b.grpo-nodiv.minus.papo.aime25}{+4.7}
\setval{8b.grpo-nodiv.minus.papo.aime25-abs}{4.7}
\setval{8b.grpo-nodiv.minus.papo.amc23}{+1.6}
\setval{8b.grpo-nodiv.minus.papo.amc23-abs}{1.6}
\setval{8b.grpo-nodiv.minus.papo.avg}{+2.3}
\setval{8b.grpo-nodiv.minus.papo.avg-abs}{2.3}
\setval{8b.grpo-nodiv.minus.papo.code}{+1.6}
\setval{8b.grpo-nodiv.minus.papo.code-abs}{1.6}
\setval{8b.grpo-nodiv.minus.papo.creative-writing-v3}{$-$4.5}
\setval{8b.grpo-nodiv.minus.papo.creative-writing-v3-abs}{4.5}
\setval{8b.grpo-nodiv.minus.papo.guidebench}{+0.5}
\setval{8b.grpo-nodiv.minus.papo.guidebench-abs}{0.5}
\setval{8b.grpo-nodiv.minus.papo.humaneval-plus}{+3.8}
\setval{8b.grpo-nodiv.minus.papo.humaneval-plus-abs}{3.8}
\setval{8b.grpo-nodiv.minus.papo.if}{+0.9}
\setval{8b.grpo-nodiv.minus.papo.if-abs}{0.9}
\setval{8b.grpo-nodiv.minus.papo.ifbench}{$-$1.0}
\setval{8b.grpo-nodiv.minus.papo.ifbench-abs}{1.0}
\setval{8b.grpo-nodiv.minus.papo.ifeval}{+3.1}
\setval{8b.grpo-nodiv.minus.papo.ifeval-abs}{3.1}
\setval{8b.grpo-nodiv.minus.papo.judge2.avg}{+2.4}
\setval{8b.grpo-nodiv.minus.papo.judge2.avg-abs}{2.4}
\setval{8b.grpo-nodiv.minus.papo.judge2.creative-writing-v3}{$-$4.4}
\setval{8b.grpo-nodiv.minus.papo.judge2.creative-writing-v3-abs}{4.4}
\setval{8b.grpo-nodiv.minus.papo.judge2.longbench-write}{+7.0}
\setval{8b.grpo-nodiv.minus.papo.judge2.longbench-write-abs}{7.0}
\setval{8b.grpo-nodiv.minus.papo.judge2.writing}{+0.1}
\setval{8b.grpo-nodiv.minus.papo.judge2.writing-abs}{0.1}
\setval{8b.grpo-nodiv.minus.papo.judge2.writingbench}{$-$2.4}
\setval{8b.grpo-nodiv.minus.papo.judge2.writingbench-abs}{2.4}
\setval{8b.grpo-nodiv.minus.papo.livecodebench}{$-$1.7}
\setval{8b.grpo-nodiv.minus.papo.livecodebench-abs}{1.7}
\setval{8b.grpo-nodiv.minus.papo.longbench-write}{+5.6}
\setval{8b.grpo-nodiv.minus.papo.longbench-write-abs}{5.6}
\setval{8b.grpo-nodiv.minus.papo.math}{+4.7}
\setval{8b.grpo-nodiv.minus.papo.math-abs}{4.7}
\setval{8b.grpo-nodiv.minus.papo.math500}{+5.7}
\setval{8b.grpo-nodiv.minus.papo.math500-abs}{5.7}
\setval{8b.grpo-nodiv.minus.papo.mbpp-plus}{+2.6}
\setval{8b.grpo-nodiv.minus.papo.mbpp-plus-abs}{2.6}
\setval{8b.grpo-nodiv.minus.papo.minerva}{+1.5}
\setval{8b.grpo-nodiv.minus.papo.minerva-abs}{1.5}
\setval{8b.grpo-nodiv.minus.papo.olympiad-bench}{+6.7}
\setval{8b.grpo-nodiv.minus.papo.olympiad-bench-abs}{6.7}
\setval{8b.grpo-nodiv.minus.papo.writing}{$-$0.6}
\setval{8b.grpo-nodiv.minus.papo.writing-abs}{0.6}
\setval{8b.grpo-nodiv.minus.papo.writingbench}{$-$2.9}
\setval{8b.grpo-nodiv.minus.papo.writingbench-abs}{2.9}
\setval{8b.grpo-nodiv.minus.saw.aime24}{+5.3}
\setval{8b.grpo-nodiv.minus.saw.aime24-abs}{5.3}
\setval{8b.grpo-nodiv.minus.saw.aime25}{$-$0.7}
\setval{8b.grpo-nodiv.minus.saw.aime25-abs}{0.7}
\setval{8b.grpo-nodiv.minus.saw.amc23}{+1.3}
\setval{8b.grpo-nodiv.minus.saw.amc23-abs}{1.3}
\setval{8b.grpo-nodiv.minus.saw.avg}{+0.1}
\setval{8b.grpo-nodiv.minus.saw.avg-abs}{0.1}
\setval{8b.grpo-nodiv.minus.saw.code}{+0.2}
\setval{8b.grpo-nodiv.minus.saw.code-abs}{0.2}
\setval{8b.grpo-nodiv.minus.saw.creative-writing-v3}{$-$3.0}
\setval{8b.grpo-nodiv.minus.saw.creative-writing-v3-abs}{3.0}
\setval{8b.grpo-nodiv.minus.saw.guidebench}{+0.6}
\setval{8b.grpo-nodiv.minus.saw.guidebench-abs}{0.6}
\setval{8b.grpo-nodiv.minus.saw.humaneval-plus}{+0.5}
\setval{8b.grpo-nodiv.minus.saw.humaneval-plus-abs}{0.5}
\setval{8b.grpo-nodiv.minus.saw.if}{$-$0.4}
\setval{8b.grpo-nodiv.minus.saw.if-abs}{0.4}
\setval{8b.grpo-nodiv.minus.saw.ifbench}{$-$3.7}
\setval{8b.grpo-nodiv.minus.saw.ifbench-abs}{3.7}
\setval{8b.grpo-nodiv.minus.saw.ifeval}{+1.8}
\setval{8b.grpo-nodiv.minus.saw.ifeval-abs}{1.8}
\setval{8b.grpo-nodiv.minus.saw.judge2.avg}{+0.6}
\setval{8b.grpo-nodiv.minus.saw.judge2.avg-abs}{0.6}
\setval{8b.grpo-nodiv.minus.saw.judge2.creative-writing-v3}{+0.9}
\setval{8b.grpo-nodiv.minus.saw.judge2.creative-writing-v3-abs}{0.9}
\setval{8b.grpo-nodiv.minus.saw.judge2.longbench-write}{+1.7}
\setval{8b.grpo-nodiv.minus.saw.judge2.longbench-write-abs}{1.7}
\setval{8b.grpo-nodiv.minus.saw.judge2.writing}{$-$0.1}
\setval{8b.grpo-nodiv.minus.saw.judge2.writing-abs}{0.1}
\setval{8b.grpo-nodiv.minus.saw.judge2.writingbench}{$-$3.0}
\setval{8b.grpo-nodiv.minus.saw.judge2.writingbench-abs}{3.0}
\setval{8b.grpo-nodiv.minus.saw.livecodebench}{$-$1.7}
\setval{8b.grpo-nodiv.minus.saw.livecodebench-abs}{1.7}
\setval{8b.grpo-nodiv.minus.saw.longbench-write}{+0.0}
\setval{8b.grpo-nodiv.minus.saw.longbench-write-abs}{0.0}
\setval{8b.grpo-nodiv.minus.saw.math}{+1.8}
\setval{8b.grpo-nodiv.minus.saw.math-abs}{1.8}
\setval{8b.grpo-nodiv.minus.saw.math500}{+2.2}
\setval{8b.grpo-nodiv.minus.saw.math500-abs}{2.2}
\setval{8b.grpo-nodiv.minus.saw.mbpp-plus}{+1.7}
\setval{8b.grpo-nodiv.minus.saw.mbpp-plus-abs}{1.7}
\setval{8b.grpo-nodiv.minus.saw.minerva}{$-$1.0}
\setval{8b.grpo-nodiv.minus.saw.minerva-abs}{1.0}
\setval{8b.grpo-nodiv.minus.saw.olympiad-bench}{+3.4}
\setval{8b.grpo-nodiv.minus.saw.olympiad-bench-abs}{3.4}
\setval{8b.grpo-nodiv.minus.saw.writing}{$-$2.7}
\setval{8b.grpo-nodiv.minus.saw.writing-abs}{2.7}
\setval{8b.grpo-nodiv.minus.saw.writingbench}{$-$5.1}
\setval{8b.grpo-nodiv.minus.saw.writingbench-abs}{5.1}
\setval{8b.grpo-nodiv.passing-at-ceiling.code.early}{0.72}
\setval{8b.grpo-nodiv.passing-at-ceiling.code.late}{0.75}
\setval{8b.grpo-nodiv.passing-at-ceiling.code.run}{0.74}
\setval{8b.grpo-nodiv.passing-at-ceiling.if.early}{0.31}
\setval{8b.grpo-nodiv.passing-at-ceiling.if.late}{0.55}
\setval{8b.grpo-nodiv.passing-at-ceiling.if.run}{0.47}
\setval{8b.grpo-nodiv.passing-at-ceiling.math.early}{1.00}
\setval{8b.grpo-nodiv.passing-at-ceiling.math.late}{1.00}
\setval{8b.grpo-nodiv.passing-at-ceiling.math.run}{1.00}
\setval{8b.grpo-nodiv.passing-at-ceiling.writing.early}{0.00}
\setval{8b.grpo-nodiv.passing-at-ceiling.writing.late}{0.00}
\setval{8b.grpo-nodiv.passing-at-ceiling.writing.run}{0.00}
\setval{8b.grpo-nodiv.primary-level.code.early}{0.55}
\setval{8b.grpo-nodiv.primary-level.code.late}{0.77}
\setval{8b.grpo-nodiv.primary-level.code.run}{0.72}
\setval{8b.grpo-nodiv.primary-level.if.early}{0.35}
\setval{8b.grpo-nodiv.primary-level.if.late}{0.75}
\setval{8b.grpo-nodiv.primary-level.if.run}{0.64}
\setval{8b.grpo-nodiv.primary-level.math.early}{0.49}
\setval{8b.grpo-nodiv.primary-level.math.late}{0.73}
\setval{8b.grpo-nodiv.primary-level.math.run}{0.66}
\setval{8b.grpo-nodiv.primary-level.writing.early}{0.52}
\setval{8b.grpo-nodiv.primary-level.writing.late}{0.74}
\setval{8b.grpo-nodiv.primary-level.writing.run}{0.69}
\setval{8b.grpo.aux-change.code}{+0.18}
\setval{8b.grpo.aux-change.if}{+0.18}
\setval{8b.grpo.aux-change.math}{+0.14}
\setval{8b.grpo.aux-change.writing}{+0.14}
\setval{8b.grpo.aux-change.writing-rm}{+0.32}
\setval{8b.grpo.aux-full.code.early}{0.72}
\setval{8b.grpo.aux-full.code.late}{0.97}
\setval{8b.grpo.aux-full.code.run}{0.95}
\setval{8b.grpo.aux-full.if.early}{0.73}
\setval{8b.grpo.aux-full.if.late}{0.94}
\setval{8b.grpo.aux-full.if.run}{0.90}
\setval{8b.grpo.aux-full.math.early}{0.59}
\setval{8b.grpo.aux-full.math.late}{0.93}
\setval{8b.grpo.aux-full.math.run}{0.88}
\setval{8b.grpo.aux-full.writing-rm.early}{0.07}
\setval{8b.grpo.aux-full.writing-rm.late}{0.59}
\setval{8b.grpo.aux-full.writing-rm.run}{0.48}
\setval{8b.grpo.aux-full.writing.early}{0.67}
\setval{8b.grpo.aux-full.writing.late}{0.78}
\setval{8b.grpo.aux-full.writing.run}{0.74}
\setval{8b.grpo.aux-level.code.early}{0.93}
\setval{8b.grpo.aux-level.code.late}{0.99}
\setval{8b.grpo.aux-level.code.run}{0.99}
\setval{8b.grpo.aux-level.if.early}{0.88}
\setval{8b.grpo.aux-level.if.late}{0.97}
\setval{8b.grpo.aux-level.if.run}{0.95}
\setval{8b.grpo.aux-level.math.early}{0.91}
\setval{8b.grpo.aux-level.math.late}{0.99}
\setval{8b.grpo.aux-level.math.run}{0.98}
\setval{8b.grpo.aux-level.writing-rm.early}{0.77}
\setval{8b.grpo.aux-level.writing-rm.late}{0.96}
\setval{8b.grpo.aux-level.writing-rm.run}{0.93}
\setval{8b.grpo.aux-level.writing.early}{0.91}
\setval{8b.grpo.aux-level.writing.late}{0.98}
\setval{8b.grpo.aux-level.writing.run}{0.96}
\setval{8b.grpo.bench.aime24}{15.8}
\setval{8b.grpo.bench.aime25}{15.5}
\setval{8b.grpo.bench.amc23}{64.4}
\setval{8b.grpo.bench.creative-writing-v3}{52.5}
\setval{8b.grpo.bench.guidebench}{81.9}
\setval{8b.grpo.bench.humaneval-plus}{83.7}
\setval{8b.grpo.bench.ifbench}{42.0}
\setval{8b.grpo.bench.ifeval}{86.8}
\setval{8b.grpo.bench.livecodebench}{26.5}
\setval{8b.grpo.bench.longbench-write}{69.8}
\setval{8b.grpo.bench.math500}{80.6}
\setval{8b.grpo.bench.mbpp-plus}{74.7}
\setval{8b.grpo.bench.minerva}{49.9}
\setval{8b.grpo.bench.olympiad-bench}{47.9}
\setval{8b.grpo.bench.writingbench}{76.5}
\setval{8b.grpo.eval.avg}{57.9}
\setval{8b.grpo.eval.code}{61.6}
\setval{8b.grpo.eval.if}{70.2}
\setval{8b.grpo.eval.math}{45.7}
\setval{8b.grpo.eval.writing}{66.2}
\setval{8b.grpo.judge2.avg}{55.2}
\setval{8b.grpo.judge2.creative-writing-v3}{39.3}
\setval{8b.grpo.judge2.longbench-write}{62.4}
\setval{8b.grpo.judge2.writing}{52.9}
\setval{8b.grpo.judge2.writingbench}{56.9}
\setval{8b.grpo.last-step}{300}
\setval{8b.grpo.minus.base.aime24}{+4.5}
\setval{8b.grpo.minus.base.aime24-abs}{4.5}
\setval{8b.grpo.minus.base.aime25}{+7.0}
\setval{8b.grpo.minus.base.aime25-abs}{7.0}
\setval{8b.grpo.minus.base.amc23}{+14.1}
\setval{8b.grpo.minus.base.amc23-abs}{14.1}
\setval{8b.grpo.minus.base.avg}{+16.4}
\setval{8b.grpo.minus.base.avg-abs}{16.4}
\setval{8b.grpo.minus.base.code}{+5.5}
\setval{8b.grpo.minus.base.code-abs}{5.5}
\setval{8b.grpo.minus.base.creative-writing-v3}{+25.8}
\setval{8b.grpo.minus.base.creative-writing-v3-abs}{25.8}
\setval{8b.grpo.minus.base.guidebench}{+36.3}
\setval{8b.grpo.minus.base.guidebench-abs}{36.3}
\setval{8b.grpo.minus.base.humaneval-plus}{+3.7}
\setval{8b.grpo.minus.base.humaneval-plus-abs}{3.7}
\setval{8b.grpo.minus.base.if}{+31.6}
\setval{8b.grpo.minus.base.if-abs}{31.6}
\setval{8b.grpo.minus.base.ifbench}{+24.7}
\setval{8b.grpo.minus.base.ifbench-abs}{24.7}
\setval{8b.grpo.minus.base.ifeval}{+33.8}
\setval{8b.grpo.minus.base.ifeval-abs}{33.8}
\setval{8b.grpo.minus.base.judge2.avg}{+14.2}
\setval{8b.grpo.minus.base.judge2.avg-abs}{14.2}
\setval{8b.grpo.minus.base.judge2.creative-writing-v3}{+12.8}
\setval{8b.grpo.minus.base.judge2.creative-writing-v3-abs}{12.8}
\setval{8b.grpo.minus.base.judge2.longbench-write}{+14.4}
\setval{8b.grpo.minus.base.judge2.longbench-write-abs}{14.4}
\setval{8b.grpo.minus.base.judge2.writing}{+16.7}
\setval{8b.grpo.minus.base.judge2.writing-abs}{16.7}
\setval{8b.grpo.minus.base.judge2.writingbench}{+22.9}
\setval{8b.grpo.minus.base.judge2.writingbench-abs}{22.9}
\setval{8b.grpo.minus.base.livecodebench}{+5.6}
\setval{8b.grpo.minus.base.livecodebench-abs}{5.6}
\setval{8b.grpo.minus.base.longbench-write}{+18.4}
\setval{8b.grpo.minus.base.longbench-write-abs}{18.4}
\setval{8b.grpo.minus.base.math}{+8.5}
\setval{8b.grpo.minus.base.math-abs}{8.5}
\setval{8b.grpo.minus.base.math500}{+8.2}
\setval{8b.grpo.minus.base.math500-abs}{8.2}
\setval{8b.grpo.minus.base.mbpp-plus}{+7.4}
\setval{8b.grpo.minus.base.mbpp-plus-abs}{7.4}
\setval{8b.grpo.minus.base.minerva}{+8.9}
\setval{8b.grpo.minus.base.minerva-abs}{8.9}
\setval{8b.grpo.minus.base.olympiad-bench}{+8.5}
\setval{8b.grpo.minus.base.olympiad-bench-abs}{8.5}
\setval{8b.grpo.minus.base.writing}{+27.9}
\setval{8b.grpo.minus.base.writing-abs}{27.9}
\setval{8b.grpo.minus.base.writingbench}{+39.7}
\setval{8b.grpo.minus.base.writingbench-abs}{39.7}
\setval{8b.grpo.minus.concur-free.aime24}{$-$3.9}
\setval{8b.grpo.minus.concur-free.aime24-abs}{3.9}
\setval{8b.grpo.minus.concur-free.aime25}{$-$1.0}
\setval{8b.grpo.minus.concur-free.aime25-abs}{1.0}
\setval{8b.grpo.minus.concur-free.amc23}{$-$0.3}
\setval{8b.grpo.minus.concur-free.amc23-abs}{0.3}
\setval{8b.grpo.minus.concur-free.avg}{$-$2.4}
\setval{8b.grpo.minus.concur-free.avg-abs}{2.4}
\setval{8b.grpo.minus.concur-free.code}{$-$0.9}
\setval{8b.grpo.minus.concur-free.code-abs}{0.9}
\setval{8b.grpo.minus.concur-free.creative-writing-v3}{+0.8}
\setval{8b.grpo.minus.concur-free.creative-writing-v3-abs}{0.8}
\setval{8b.grpo.minus.concur-free.guidebench}{$-$1.1}
\setval{8b.grpo.minus.concur-free.guidebench-abs}{1.1}
\setval{8b.grpo.minus.concur-free.humaneval-plus}{$-$1.6}
\setval{8b.grpo.minus.concur-free.humaneval-plus-abs}{1.6}
\setval{8b.grpo.minus.concur-free.if}{+0.0}
\setval{8b.grpo.minus.concur-free.if-abs}{0.0}
\setval{8b.grpo.minus.concur-free.ifbench}{+3.3}
\setval{8b.grpo.minus.concur-free.ifbench-abs}{3.3}
\setval{8b.grpo.minus.concur-free.ifeval}{$-$2.4}
\setval{8b.grpo.minus.concur-free.ifeval-abs}{2.4}
\setval{8b.grpo.minus.concur-free.judge2.avg}{$-$2.1}
\setval{8b.grpo.minus.concur-free.judge2.avg-abs}{2.1}
\setval{8b.grpo.minus.concur-free.judge2.creative-writing-v3}{$-$3.2}
\setval{8b.grpo.minus.concur-free.judge2.creative-writing-v3-abs}{3.2}
\setval{8b.grpo.minus.concur-free.judge2.longbench-write}{$-$11.7}
\setval{8b.grpo.minus.concur-free.judge2.longbench-write-abs}{11.7}
\setval{8b.grpo.minus.concur-free.judge2.writing}{$-$6.2}
\setval{8b.grpo.minus.concur-free.judge2.writing-abs}{6.2}
\setval{8b.grpo.minus.concur-free.judge2.writingbench}{$-$3.8}
\setval{8b.grpo.minus.concur-free.judge2.writingbench-abs}{3.8}
\setval{8b.grpo.minus.concur-free.livecodebench}{$-$0.6}
\setval{8b.grpo.minus.concur-free.livecodebench-abs}{0.6}
\setval{8b.grpo.minus.concur-free.longbench-write}{$-$13.2}
\setval{8b.grpo.minus.concur-free.longbench-write-abs}{13.2}
\setval{8b.grpo.minus.concur-free.math}{$-$1.7}
\setval{8b.grpo.minus.concur-free.math-abs}{1.7}
\setval{8b.grpo.minus.concur-free.math500}{$-$2.8}
\setval{8b.grpo.minus.concur-free.math500-abs}{2.8}
\setval{8b.grpo.minus.concur-free.mbpp-plus}{$-$0.4}
\setval{8b.grpo.minus.concur-free.mbpp-plus-abs}{0.4}
\setval{8b.grpo.minus.concur-free.minerva}{+0.1}
\setval{8b.grpo.minus.concur-free.minerva-abs}{0.1}
\setval{8b.grpo.minus.concur-free.olympiad-bench}{$-$2.1}
\setval{8b.grpo.minus.concur-free.olympiad-bench-abs}{2.1}
\setval{8b.grpo.minus.concur-free.writing}{$-$7.6}
\setval{8b.grpo.minus.concur-free.writing-abs}{7.6}
\setval{8b.grpo.minus.concur-free.writingbench}{$-$10.3}
\setval{8b.grpo.minus.concur-free.writingbench-abs}{10.3}
\setval{8b.grpo.minus.concur-tied.aime24}{$-$15.4}
\setval{8b.grpo.minus.concur-tied.aime24-abs}{15.4}
\setval{8b.grpo.minus.concur-tied.aime25}{$-$8.4}
\setval{8b.grpo.minus.concur-tied.aime25-abs}{8.4}
\setval{8b.grpo.minus.concur-tied.amc23}{$-$14.1}
\setval{8b.grpo.minus.concur-tied.amc23-abs}{14.1}
\setval{8b.grpo.minus.concur-tied.avg}{$-$6.1}
\setval{8b.grpo.minus.concur-tied.avg-abs}{6.1}
\setval{8b.grpo.minus.concur-tied.code}{$-$1.5}
\setval{8b.grpo.minus.concur-tied.code-abs}{1.5}
\setval{8b.grpo.minus.concur-tied.creative-writing-v3}{+0.1}
\setval{8b.grpo.minus.concur-tied.creative-writing-v3-abs}{0.1}
\setval{8b.grpo.minus.concur-tied.guidebench}{$-$1.5}
\setval{8b.grpo.minus.concur-tied.guidebench-abs}{1.5}
\setval{8b.grpo.minus.concur-tied.humaneval-plus}{$-$3.7}
\setval{8b.grpo.minus.concur-tied.humaneval-plus-abs}{3.7}
\setval{8b.grpo.minus.concur-tied.if}{$-$0.3}
\setval{8b.grpo.minus.concur-tied.if-abs}{0.3}
\setval{8b.grpo.minus.concur-tied.ifbench}{+2.7}
\setval{8b.grpo.minus.concur-tied.ifbench-abs}{2.7}
\setval{8b.grpo.minus.concur-tied.ifeval}{$-$2.0}
\setval{8b.grpo.minus.concur-tied.ifeval-abs}{2.0}
\setval{8b.grpo.minus.concur-tied.judge2.avg}{$-$6.1}
\setval{8b.grpo.minus.concur-tied.judge2.avg-abs}{6.1}
\setval{8b.grpo.minus.concur-tied.judge2.creative-writing-v3}{$-$2.5}
\setval{8b.grpo.minus.concur-tied.judge2.creative-writing-v3-abs}{2.5}
\setval{8b.grpo.minus.concur-tied.judge2.longbench-write}{$-$12.3}
\setval{8b.grpo.minus.concur-tied.judge2.longbench-write-abs}{12.3}
\setval{8b.grpo.minus.concur-tied.judge2.writing}{$-$7.2}
\setval{8b.grpo.minus.concur-tied.judge2.writing-abs}{7.2}
\setval{8b.grpo.minus.concur-tied.judge2.writingbench}{$-$6.9}
\setval{8b.grpo.minus.concur-tied.judge2.writingbench-abs}{6.9}
\setval{8b.grpo.minus.concur-tied.livecodebench}{$-$0.7}
\setval{8b.grpo.minus.concur-tied.livecodebench-abs}{0.7}
\setval{8b.grpo.minus.concur-tied.longbench-write}{$-$11.4}
\setval{8b.grpo.minus.concur-tied.longbench-write-abs}{11.4}
\setval{8b.grpo.minus.concur-tied.math}{$-$10.7}
\setval{8b.grpo.minus.concur-tied.math-abs}{10.7}
\setval{8b.grpo.minus.concur-tied.math500}{$-$9.0}
\setval{8b.grpo.minus.concur-tied.math500-abs}{9.0}
\setval{8b.grpo.minus.concur-tied.mbpp-plus}{$-$0.2}
\setval{8b.grpo.minus.concur-tied.mbpp-plus-abs}{0.2}
\setval{8b.grpo.minus.concur-tied.minerva}{$-$5.5}
\setval{8b.grpo.minus.concur-tied.minerva-abs}{5.5}
\setval{8b.grpo.minus.concur-tied.olympiad-bench}{$-$11.6}
\setval{8b.grpo.minus.concur-tied.olympiad-bench-abs}{11.6}
\setval{8b.grpo.minus.concur-tied.writing}{$-$7.4}
\setval{8b.grpo.minus.concur-tied.writing-abs}{7.4}
\setval{8b.grpo.minus.concur-tied.writingbench}{$-$10.7}
\setval{8b.grpo.minus.concur-tied.writingbench-abs}{10.7}
\setval{8b.grpo.minus.concur.aime24}{$-$12.9}
\setval{8b.grpo.minus.concur.aime24-abs}{12.9}
\setval{8b.grpo.minus.concur.aime25}{$-$7.6}
\setval{8b.grpo.minus.concur.aime25-abs}{7.6}
\setval{8b.grpo.minus.concur.amc23}{$-$14.5}
\setval{8b.grpo.minus.concur.amc23-abs}{14.5}
\setval{8b.grpo.minus.concur.avg}{$-$5.3}
\setval{8b.grpo.minus.concur.avg-abs}{5.3}
\setval{8b.grpo.minus.concur.code}{+0.9}
\setval{8b.grpo.minus.concur.code-abs}{0.9}
\setval{8b.grpo.minus.concur.creative-writing-v3}{+2.6}
\setval{8b.grpo.minus.concur.creative-writing-v3-abs}{2.6}
\setval{8b.grpo.minus.concur.guidebench}{$-$1.5}
\setval{8b.grpo.minus.concur.guidebench-abs}{1.5}
\setval{8b.grpo.minus.concur.humaneval-plus}{+1.0}
\setval{8b.grpo.minus.concur.humaneval-plus-abs}{1.0}
\setval{8b.grpo.minus.concur.if}{+0.2}
\setval{8b.grpo.minus.concur.if-abs}{0.2}
\setval{8b.grpo.minus.concur.ifbench}{+3.3}
\setval{8b.grpo.minus.concur.ifbench-abs}{3.3}
\setval{8b.grpo.minus.concur.ifeval}{$-$1.3}
\setval{8b.grpo.minus.concur.ifeval-abs}{1.3}
\setval{8b.grpo.minus.concur.judge2.avg}{$-$5.1}
\setval{8b.grpo.minus.concur.judge2.avg-abs}{5.1}
\setval{8b.grpo.minus.concur.judge2.creative-writing-v3}{+1.5}
\setval{8b.grpo.minus.concur.judge2.creative-writing-v3-abs}{1.5}
\setval{8b.grpo.minus.concur.judge2.longbench-write}{$-$13.1}
\setval{8b.grpo.minus.concur.judge2.longbench-write-abs}{13.1}
\setval{8b.grpo.minus.concur.judge2.writing}{$-$6.2}
\setval{8b.grpo.minus.concur.judge2.writing-abs}{6.2}
\setval{8b.grpo.minus.concur.judge2.writingbench}{$-$7.0}
\setval{8b.grpo.minus.concur.judge2.writingbench-abs}{7.0}
\setval{8b.grpo.minus.concur.livecodebench}{+0.2}
\setval{8b.grpo.minus.concur.livecodebench-abs}{0.2}
\setval{8b.grpo.minus.concur.longbench-write}{$-$11.8}
\setval{8b.grpo.minus.concur.longbench-write-abs}{11.8}
\setval{8b.grpo.minus.concur.math}{$-$10.2}
\setval{8b.grpo.minus.concur.math-abs}{10.2}
\setval{8b.grpo.minus.concur.math500}{$-$9.5}
\setval{8b.grpo.minus.concur.math500-abs}{9.5}
\setval{8b.grpo.minus.concur.mbpp-plus}{+1.5}
\setval{8b.grpo.minus.concur.mbpp-plus-abs}{1.5}
\setval{8b.grpo.minus.concur.minerva}{$-$4.3}
\setval{8b.grpo.minus.concur.minerva-abs}{4.3}
\setval{8b.grpo.minus.concur.olympiad-bench}{$-$12.4}
\setval{8b.grpo.minus.concur.olympiad-bench-abs}{12.4}
\setval{8b.grpo.minus.concur.writing}{$-$6.9}
\setval{8b.grpo.minus.concur.writing-abs}{6.9}
\setval{8b.grpo.minus.concur.writingbench}{$-$11.6}
\setval{8b.grpo.minus.concur.writingbench-abs}{11.6}
\setval{8b.grpo.minus.gd2po.aime24}{$-$0.4}
\setval{8b.grpo.minus.gd2po.aime24-abs}{0.4}
\setval{8b.grpo.minus.gd2po.aime25}{+3.3}
\setval{8b.grpo.minus.gd2po.aime25-abs}{3.3}
\setval{8b.grpo.minus.gd2po.amc23}{+4.8}
\setval{8b.grpo.minus.gd2po.amc23-abs}{4.8}
\setval{8b.grpo.minus.gd2po.avg}{+0.5}
\setval{8b.grpo.minus.gd2po.avg-abs}{0.5}
\setval{8b.grpo.minus.gd2po.code}{$-$0.1}
\setval{8b.grpo.minus.gd2po.code-abs}{0.1}
\setval{8b.grpo.minus.gd2po.creative-writing-v3}{+4.3}
\setval{8b.grpo.minus.gd2po.creative-writing-v3-abs}{4.3}
\setval{8b.grpo.minus.gd2po.guidebench}{$-$1.8}
\setval{8b.grpo.minus.gd2po.guidebench-abs}{1.8}
\setval{8b.grpo.minus.gd2po.humaneval-plus}{$-$0.2}
\setval{8b.grpo.minus.gd2po.humaneval-plus-abs}{0.2}
\setval{8b.grpo.minus.gd2po.if}{$-$1.8}
\setval{8b.grpo.minus.gd2po.if-abs}{1.8}
\setval{8b.grpo.minus.gd2po.ifbench}{$-$1.0}
\setval{8b.grpo.minus.gd2po.ifbench-abs}{1.0}
\setval{8b.grpo.minus.gd2po.ifeval}{$-$2.5}
\setval{8b.grpo.minus.gd2po.ifeval-abs}{2.5}
\setval{8b.grpo.minus.gd2po.judge2.avg}{+0.3}
\setval{8b.grpo.minus.gd2po.judge2.avg-abs}{0.3}
\setval{8b.grpo.minus.gd2po.judge2.creative-writing-v3}{+2.5}
\setval{8b.grpo.minus.gd2po.judge2.creative-writing-v3-abs}{2.5}
\setval{8b.grpo.minus.gd2po.judge2.longbench-write}{+0.3}
\setval{8b.grpo.minus.gd2po.judge2.longbench-write-abs}{0.3}
\setval{8b.grpo.minus.gd2po.judge2.writing}{+1.5}
\setval{8b.grpo.minus.gd2po.judge2.writing-abs}{1.5}
\setval{8b.grpo.minus.gd2po.judge2.writingbench}{+1.7}
\setval{8b.grpo.minus.gd2po.judge2.writingbench-abs}{1.7}
\setval{8b.grpo.minus.gd2po.livecodebench}{+0.2}
\setval{8b.grpo.minus.gd2po.livecodebench-abs}{0.2}
\setval{8b.grpo.minus.gd2po.longbench-write}{$-$0.1}
\setval{8b.grpo.minus.gd2po.longbench-write-abs}{0.1}
\setval{8b.grpo.minus.gd2po.math}{+1.1}
\setval{8b.grpo.minus.gd2po.math-abs}{1.1}
\setval{8b.grpo.minus.gd2po.math500}{$-$0.6}
\setval{8b.grpo.minus.gd2po.math500-abs}{0.6}
\setval{8b.grpo.minus.gd2po.mbpp-plus}{$-$0.4}
\setval{8b.grpo.minus.gd2po.mbpp-plus-abs}{0.4}
\setval{8b.grpo.minus.gd2po.minerva}{$-$0.1}
\setval{8b.grpo.minus.gd2po.minerva-abs}{0.1}
\setval{8b.grpo.minus.gd2po.olympiad-bench}{$-$0.6}
\setval{8b.grpo.minus.gd2po.olympiad-bench-abs}{0.6}
\setval{8b.grpo.minus.gd2po.writing}{+2.3}
\setval{8b.grpo.minus.gd2po.writing-abs}{2.3}
\setval{8b.grpo.minus.gd2po.writingbench}{+2.6}
\setval{8b.grpo.minus.gd2po.writingbench-abs}{2.6}
\setval{8b.grpo.minus.gdpo-noaux.aime24}{$-$12.1}
\setval{8b.grpo.minus.gdpo-noaux.aime24-abs}{12.1}
\setval{8b.grpo.minus.gdpo-noaux.aime25}{$-$8.4}
\setval{8b.grpo.minus.gdpo-noaux.aime25-abs}{8.4}
\setval{8b.grpo.minus.gdpo-noaux.amc23}{$-$12.7}
\setval{8b.grpo.minus.gdpo-noaux.amc23-abs}{12.7}
\setval{8b.grpo.minus.gdpo-noaux.avg}{$-$2.6}
\setval{8b.grpo.minus.gdpo-noaux.avg-abs}{2.6}
\setval{8b.grpo.minus.gdpo-noaux.code}{$-$1.1}
\setval{8b.grpo.minus.gdpo-noaux.code-abs}{1.1}
\setval{8b.grpo.minus.gdpo-noaux.creative-writing-v3}{+8.6}
\setval{8b.grpo.minus.gdpo-noaux.creative-writing-v3-abs}{8.6}
\setval{8b.grpo.minus.gdpo-noaux.guidebench}{+0.5}
\setval{8b.grpo.minus.gdpo-noaux.guidebench-abs}{0.5}
\setval{8b.grpo.minus.gdpo-noaux.humaneval-plus}{$-$2.3}
\setval{8b.grpo.minus.gdpo-noaux.humaneval-plus-abs}{2.3}
\setval{8b.grpo.minus.gdpo-noaux.if}{+3.0}
\setval{8b.grpo.minus.gdpo-noaux.if-abs}{3.0}
\setval{8b.grpo.minus.gdpo-noaux.ifbench}{+8.3}
\setval{8b.grpo.minus.gdpo-noaux.ifbench-abs}{8.3}
\setval{8b.grpo.minus.gdpo-noaux.ifeval}{+0.2}
\setval{8b.grpo.minus.gdpo-noaux.ifeval-abs}{0.2}
\setval{8b.grpo.minus.gdpo-noaux.judge2.avg}{$-$1.9}
\setval{8b.grpo.minus.gdpo-noaux.judge2.avg-abs}{1.9}
\setval{8b.grpo.minus.gdpo-noaux.judge2.creative-writing-v3}{+10.4}
\setval{8b.grpo.minus.gdpo-noaux.judge2.creative-writing-v3-abs}{10.4}
\setval{8b.grpo.minus.gdpo-noaux.judge2.longbench-write}{+7.9}
\setval{8b.grpo.minus.gdpo-noaux.judge2.longbench-write-abs}{7.9}
\setval{8b.grpo.minus.gdpo-noaux.judge2.writing}{+8.4}
\setval{8b.grpo.minus.gdpo-noaux.judge2.writing-abs}{8.4}
\setval{8b.grpo.minus.gdpo-noaux.judge2.writingbench}{+6.9}
\setval{8b.grpo.minus.gdpo-noaux.judge2.writingbench-abs}{6.9}
\setval{8b.grpo.minus.gdpo-noaux.livecodebench}{$-$0.9}
\setval{8b.grpo.minus.gdpo-noaux.livecodebench-abs}{0.9}
\setval{8b.grpo.minus.gdpo-noaux.longbench-write}{+3.8}
\setval{8b.grpo.minus.gdpo-noaux.longbench-write-abs}{3.8}
\setval{8b.grpo.minus.gdpo-noaux.math}{$-$9.8}
\setval{8b.grpo.minus.gdpo-noaux.math-abs}{9.8}
\setval{8b.grpo.minus.gdpo-noaux.math500}{$-$9.6}
\setval{8b.grpo.minus.gdpo-noaux.math500-abs}{9.6}
\setval{8b.grpo.minus.gdpo-noaux.mbpp-plus}{$-$0.2}
\setval{8b.grpo.minus.gdpo-noaux.mbpp-plus-abs}{0.2}
\setval{8b.grpo.minus.gdpo-noaux.minerva}{$-$4.1}
\setval{8b.grpo.minus.gdpo-noaux.minerva-abs}{4.1}
\setval{8b.grpo.minus.gdpo-noaux.olympiad-bench}{$-$12.0}
\setval{8b.grpo.minus.gdpo-noaux.olympiad-bench-abs}{12.0}
\setval{8b.grpo.minus.gdpo-noaux.writing}{+4.6}
\setval{8b.grpo.minus.gdpo-noaux.writing-abs}{4.6}
\setval{8b.grpo.minus.gdpo-noaux.writingbench}{+1.5}
\setval{8b.grpo.minus.gdpo-noaux.writingbench-abs}{1.5}
\setval{8b.grpo.minus.gdpo-w0.5.aime24}{$-$3.4}
\setval{8b.grpo.minus.gdpo-w0.5.aime24-abs}{3.4}
\setval{8b.grpo.minus.gdpo-w0.5.aime25}{+0.1}
\setval{8b.grpo.minus.gdpo-w0.5.aime25-abs}{0.1}
\setval{8b.grpo.minus.gdpo-w0.5.amc23}{+0.9}
\setval{8b.grpo.minus.gdpo-w0.5.amc23-abs}{0.9}
\setval{8b.grpo.minus.gdpo-w0.5.avg}{$-$0.4}
\setval{8b.grpo.minus.gdpo-w0.5.avg-abs}{0.4}
\setval{8b.grpo.minus.gdpo-w0.5.code}{+0.5}
\setval{8b.grpo.minus.gdpo-w0.5.code-abs}{0.5}
\setval{8b.grpo.minus.gdpo-w0.5.creative-writing-v3}{+0.9}
\setval{8b.grpo.minus.gdpo-w0.5.creative-writing-v3-abs}{0.9}
\setval{8b.grpo.minus.gdpo-w0.5.guidebench}{$-$2.1}
\setval{8b.grpo.minus.gdpo-w0.5.guidebench-abs}{2.1}
\setval{8b.grpo.minus.gdpo-w0.5.humaneval-plus}{$-$0.1}
\setval{8b.grpo.minus.gdpo-w0.5.humaneval-plus-abs}{0.1}
\setval{8b.grpo.minus.gdpo-w0.5.if}{+0.4}
\setval{8b.grpo.minus.gdpo-w0.5.if-abs}{0.4}
\setval{8b.grpo.minus.gdpo-w0.5.ifbench}{+4.3}
\setval{8b.grpo.minus.gdpo-w0.5.ifbench-abs}{4.3}
\setval{8b.grpo.minus.gdpo-w0.5.ifeval}{$-$1.0}
\setval{8b.grpo.minus.gdpo-w0.5.ifeval-abs}{1.0}
\setval{8b.grpo.minus.gdpo-w0.5.judge2.avg}{$-$0.6}
\setval{8b.grpo.minus.gdpo-w0.5.judge2.avg-abs}{0.6}
\setval{8b.grpo.minus.gdpo-w0.5.judge2.creative-writing-v3}{$-$1.1}
\setval{8b.grpo.minus.gdpo-w0.5.judge2.creative-writing-v3-abs}{1.1}
\setval{8b.grpo.minus.gdpo-w0.5.judge2.longbench-write}{$-$1.2}
\setval{8b.grpo.minus.gdpo-w0.5.judge2.longbench-write-abs}{1.2}
\setval{8b.grpo.minus.gdpo-w0.5.judge2.writing}{$-$0.6}
\setval{8b.grpo.minus.gdpo-w0.5.judge2.writing-abs}{0.6}
\setval{8b.grpo.minus.gdpo-w0.5.judge2.writingbench}{+0.7}
\setval{8b.grpo.minus.gdpo-w0.5.judge2.writingbench-abs}{0.7}
\setval{8b.grpo.minus.gdpo-w0.5.livecodebench}{$-$0.3}
\setval{8b.grpo.minus.gdpo-w0.5.livecodebench-abs}{0.3}
\setval{8b.grpo.minus.gdpo-w0.5.longbench-write}{$-$1.4}
\setval{8b.grpo.minus.gdpo-w0.5.longbench-write-abs}{1.4}
\setval{8b.grpo.minus.gdpo-w0.5.math}{$-$1.7}
\setval{8b.grpo.minus.gdpo-w0.5.math-abs}{1.7}
\setval{8b.grpo.minus.gdpo-w0.5.math500}{$-$3.4}
\setval{8b.grpo.minus.gdpo-w0.5.math500-abs}{3.4}
\setval{8b.grpo.minus.gdpo-w0.5.mbpp-plus}{+1.8}
\setval{8b.grpo.minus.gdpo-w0.5.mbpp-plus-abs}{1.8}
\setval{8b.grpo.minus.gdpo-w0.5.minerva}{$-$1.4}
\setval{8b.grpo.minus.gdpo-w0.5.minerva-abs}{1.4}
\setval{8b.grpo.minus.gdpo-w0.5.olympiad-bench}{$-$3.2}
\setval{8b.grpo.minus.gdpo-w0.5.olympiad-bench-abs}{3.2}
\setval{8b.grpo.minus.gdpo-w0.5.writing}{+0.3}
\setval{8b.grpo.minus.gdpo-w0.5.writing-abs}{0.3}
\setval{8b.grpo.minus.gdpo-w0.5.writingbench}{+1.4}
\setval{8b.grpo.minus.gdpo-w0.5.writingbench-abs}{1.4}
\setval{8b.grpo.minus.gdpo.aime24}{$-$5.5}
\setval{8b.grpo.minus.gdpo.aime24-abs}{5.5}
\setval{8b.grpo.minus.gdpo.aime25}{+1.1}
\setval{8b.grpo.minus.gdpo.aime25-abs}{1.1}
\setval{8b.grpo.minus.gdpo.amc23}{+2.2}
\setval{8b.grpo.minus.gdpo.amc23-abs}{2.2}
\setval{8b.grpo.minus.gdpo.avg}{$-$0.5}
\setval{8b.grpo.minus.gdpo.avg-abs}{0.5}
\setval{8b.grpo.minus.gdpo.code}{+0.0}
\setval{8b.grpo.minus.gdpo.code-abs}{0.0}
\setval{8b.grpo.minus.gdpo.creative-writing-v3}{$-$0.4}
\setval{8b.grpo.minus.gdpo.creative-writing-v3-abs}{0.4}
\setval{8b.grpo.minus.gdpo.guidebench}{$-$2.7}
\setval{8b.grpo.minus.gdpo.guidebench-abs}{2.7}
\setval{8b.grpo.minus.gdpo.humaneval-plus}{$-$0.4}
\setval{8b.grpo.minus.gdpo.humaneval-plus-abs}{0.4}
\setval{8b.grpo.minus.gdpo.if}{$-$0.4}
\setval{8b.grpo.minus.gdpo.if-abs}{0.4}
\setval{8b.grpo.minus.gdpo.ifbench}{+2.7}
\setval{8b.grpo.minus.gdpo.ifbench-abs}{2.7}
\setval{8b.grpo.minus.gdpo.ifeval}{$-$1.3}
\setval{8b.grpo.minus.gdpo.ifeval-abs}{1.3}
\setval{8b.grpo.minus.gdpo.judge2.avg}{$-$0.8}
\setval{8b.grpo.minus.gdpo.judge2.avg-abs}{0.8}
\setval{8b.grpo.minus.gdpo.judge2.creative-writing-v3}{$-$1.8}
\setval{8b.grpo.minus.gdpo.judge2.creative-writing-v3-abs}{1.8}
\setval{8b.grpo.minus.gdpo.judge2.longbench-write}{$-$1.7}
\setval{8b.grpo.minus.gdpo.judge2.longbench-write-abs}{1.7}
\setval{8b.grpo.minus.gdpo.judge2.writing}{$-$1.7}
\setval{8b.grpo.minus.gdpo.judge2.writing-abs}{1.7}
\setval{8b.grpo.minus.gdpo.judge2.writingbench}{$-$1.5}
\setval{8b.grpo.minus.gdpo.judge2.writingbench-abs}{1.5}
\setval{8b.grpo.minus.gdpo.livecodebench}{+0.3}
\setval{8b.grpo.minus.gdpo.livecodebench-abs}{0.3}
\setval{8b.grpo.minus.gdpo.longbench-write}{+0.2}
\setval{8b.grpo.minus.gdpo.longbench-write-abs}{0.2}
\setval{8b.grpo.minus.gdpo.math}{$-$0.9}
\setval{8b.grpo.minus.gdpo.math-abs}{0.9}
\setval{8b.grpo.minus.gdpo.math500}{$-$2.4}
\setval{8b.grpo.minus.gdpo.math500-abs}{2.4}
\setval{8b.grpo.minus.gdpo.mbpp-plus}{+0.1}
\setval{8b.grpo.minus.gdpo.mbpp-plus-abs}{0.1}
\setval{8b.grpo.minus.gdpo.minerva}{+0.3}
\setval{8b.grpo.minus.gdpo.minerva-abs}{0.3}
\setval{8b.grpo.minus.gdpo.olympiad-bench}{$-$1.4}
\setval{8b.grpo.minus.gdpo.olympiad-bench-abs}{1.4}
\setval{8b.grpo.minus.gdpo.writing}{+0.1}
\setval{8b.grpo.minus.gdpo.writing-abs}{0.1}
\setval{8b.grpo.minus.gdpo.writingbench}{+0.4}
\setval{8b.grpo.minus.gdpo.writingbench-abs}{0.4}
\setval{8b.grpo.minus.geoma.aime24}{+1.8}
\setval{8b.grpo.minus.geoma.aime24-abs}{1.8}
\setval{8b.grpo.minus.geoma.aime25}{+3.0}
\setval{8b.grpo.minus.geoma.aime25-abs}{3.0}
\setval{8b.grpo.minus.geoma.amc23}{+0.9}
\setval{8b.grpo.minus.geoma.amc23-abs}{0.9}
\setval{8b.grpo.minus.geoma.avg}{+1.2}
\setval{8b.grpo.minus.geoma.avg-abs}{1.2}
\setval{8b.grpo.minus.geoma.code}{+0.3}
\setval{8b.grpo.minus.geoma.code-abs}{0.3}
\setval{8b.grpo.minus.geoma.creative-writing-v3}{+2.1}
\setval{8b.grpo.minus.geoma.creative-writing-v3-abs}{2.1}
\setval{8b.grpo.minus.geoma.guidebench}{$-$2.5}
\setval{8b.grpo.minus.geoma.guidebench-abs}{2.5}
\setval{8b.grpo.minus.geoma.humaneval-plus}{+1.0}
\setval{8b.grpo.minus.geoma.humaneval-plus-abs}{1.0}
\setval{8b.grpo.minus.geoma.if}{+1.7}
\setval{8b.grpo.minus.geoma.if-abs}{1.7}
\setval{8b.grpo.minus.geoma.ifbench}{+8.0}
\setval{8b.grpo.minus.geoma.ifbench-abs}{8.0}
\setval{8b.grpo.minus.geoma.ifeval}{$-$0.5}
\setval{8b.grpo.minus.geoma.ifeval-abs}{0.5}
\setval{8b.grpo.minus.geoma.judge2.avg}{+0.8}
\setval{8b.grpo.minus.geoma.judge2.avg-abs}{0.8}
\setval{8b.grpo.minus.geoma.judge2.creative-writing-v3}{+1.7}
\setval{8b.grpo.minus.geoma.judge2.creative-writing-v3-abs}{1.7}
\setval{8b.grpo.minus.geoma.judge2.longbench-write}{$-$2.4}
\setval{8b.grpo.minus.geoma.judge2.longbench-write-abs}{2.4}
\setval{8b.grpo.minus.geoma.judge2.writing}{$-$0.7}
\setval{8b.grpo.minus.geoma.judge2.writing-abs}{0.7}
\setval{8b.grpo.minus.geoma.judge2.writingbench}{$-$1.4}
\setval{8b.grpo.minus.geoma.judge2.writingbench-abs}{1.4}
\setval{8b.grpo.minus.geoma.livecodebench}{+0.8}
\setval{8b.grpo.minus.geoma.livecodebench-abs}{0.8}
\setval{8b.grpo.minus.geoma.longbench-write}{+0.8}
\setval{8b.grpo.minus.geoma.longbench-write-abs}{0.8}
\setval{8b.grpo.minus.geoma.math}{+1.5}
\setval{8b.grpo.minus.geoma.math-abs}{1.5}
\setval{8b.grpo.minus.geoma.math500}{+1.0}
\setval{8b.grpo.minus.geoma.math500-abs}{1.0}
\setval{8b.grpo.minus.geoma.mbpp-plus}{$-$1.0}
\setval{8b.grpo.minus.geoma.mbpp-plus-abs}{1.0}
\setval{8b.grpo.minus.geoma.minerva}{+0.1}
\setval{8b.grpo.minus.geoma.minerva-abs}{0.1}
\setval{8b.grpo.minus.geoma.olympiad-bench}{+1.9}
\setval{8b.grpo.minus.geoma.olympiad-bench-abs}{1.9}
\setval{8b.grpo.minus.geoma.writing}{+1.1}
\setval{8b.grpo.minus.geoma.writing-abs}{1.1}
\setval{8b.grpo.minus.geoma.writingbench}{+0.5}
\setval{8b.grpo.minus.geoma.writingbench-abs}{0.5}
\setval{8b.grpo.minus.grpo-nodiv.aime24}{$-$4.8}
\setval{8b.grpo.minus.grpo-nodiv.aime24-abs}{4.8}
\setval{8b.grpo.minus.grpo-nodiv.aime25}{+0.7}
\setval{8b.grpo.minus.grpo-nodiv.aime25-abs}{0.7}
\setval{8b.grpo.minus.grpo-nodiv.amc23}{+0.8}
\setval{8b.grpo.minus.grpo-nodiv.amc23-abs}{0.8}
\setval{8b.grpo.minus.grpo-nodiv.avg}{$-$0.2}
\setval{8b.grpo.minus.grpo-nodiv.avg-abs}{0.2}
\setval{8b.grpo.minus.grpo-nodiv.code}{$-$0.8}
\setval{8b.grpo.minus.grpo-nodiv.code-abs}{0.8}
\setval{8b.grpo.minus.grpo-nodiv.creative-writing-v3}{+5.8}
\setval{8b.grpo.minus.grpo-nodiv.creative-writing-v3-abs}{5.8}
\setval{8b.grpo.minus.grpo-nodiv.guidebench}{$-$3.0}
\setval{8b.grpo.minus.grpo-nodiv.guidebench-abs}{3.0}
\setval{8b.grpo.minus.grpo-nodiv.humaneval-plus}{$-$2.6}
\setval{8b.grpo.minus.grpo-nodiv.humaneval-plus-abs}{2.6}
\setval{8b.grpo.minus.grpo-nodiv.if}{+0.0}
\setval{8b.grpo.minus.grpo-nodiv.if-abs}{0.0}
\setval{8b.grpo.minus.grpo-nodiv.ifbench}{+5.7}
\setval{8b.grpo.minus.grpo-nodiv.ifbench-abs}{5.7}
\setval{8b.grpo.minus.grpo-nodiv.ifeval}{$-$2.6}
\setval{8b.grpo.minus.grpo-nodiv.ifeval-abs}{2.6}
\setval{8b.grpo.minus.grpo-nodiv.judge2.avg}{$-$0.9}
\setval{8b.grpo.minus.grpo-nodiv.judge2.avg-abs}{0.9}
\setval{8b.grpo.minus.grpo-nodiv.judge2.creative-writing-v3}{+1.8}
\setval{8b.grpo.minus.grpo-nodiv.judge2.creative-writing-v3-abs}{1.8}
\setval{8b.grpo.minus.grpo-nodiv.judge2.longbench-write}{$-$4.1}
\setval{8b.grpo.minus.grpo-nodiv.judge2.longbench-write-abs}{4.1}
\setval{8b.grpo.minus.grpo-nodiv.judge2.writing}{+0.2}
\setval{8b.grpo.minus.grpo-nodiv.judge2.writing-abs}{0.2}
\setval{8b.grpo.minus.grpo-nodiv.judge2.writingbench}{+3.0}
\setval{8b.grpo.minus.grpo-nodiv.judge2.writingbench-abs}{3.0}
\setval{8b.grpo.minus.grpo-nodiv.livecodebench}{+1.5}
\setval{8b.grpo.minus.grpo-nodiv.livecodebench-abs}{1.5}
\setval{8b.grpo.minus.grpo-nodiv.longbench-write}{$-$1.3}
\setval{8b.grpo.minus.grpo-nodiv.longbench-write-abs}{1.3}
\setval{8b.grpo.minus.grpo-nodiv.math}{$-$1.9}
\setval{8b.grpo.minus.grpo-nodiv.math-abs}{1.9}
\setval{8b.grpo.minus.grpo-nodiv.math500}{$-$3.5}
\setval{8b.grpo.minus.grpo-nodiv.math500-abs}{3.5}
\setval{8b.grpo.minus.grpo-nodiv.mbpp-plus}{$-$1.4}
\setval{8b.grpo.minus.grpo-nodiv.mbpp-plus-abs}{1.4}
\setval{8b.grpo.minus.grpo-nodiv.minerva}{$-$0.9}
\setval{8b.grpo.minus.grpo-nodiv.minerva-abs}{0.9}
\setval{8b.grpo.minus.grpo-nodiv.olympiad-bench}{$-$3.6}
\setval{8b.grpo.minus.grpo-nodiv.olympiad-bench-abs}{3.6}
\setval{8b.grpo.minus.grpo-nodiv.writing}{+3.6}
\setval{8b.grpo.minus.grpo-nodiv.writing-abs}{3.6}
\setval{8b.grpo.minus.grpo-nodiv.writingbench}{+6.1}
\setval{8b.grpo.minus.grpo-nodiv.writingbench-abs}{6.1}
\setval{8b.grpo.minus.mt-grpo.aime24}{+1.4}
\setval{8b.grpo.minus.mt-grpo.aime24-abs}{1.4}
\setval{8b.grpo.minus.mt-grpo.aime25}{$-$0.2}
\setval{8b.grpo.minus.mt-grpo.aime25-abs}{0.2}
\setval{8b.grpo.minus.mt-grpo.amc23}{$-$1.2}
\setval{8b.grpo.minus.mt-grpo.amc23-abs}{1.2}
\setval{8b.grpo.minus.mt-grpo.avg}{+0.7}
\setval{8b.grpo.minus.mt-grpo.avg-abs}{0.7}
\setval{8b.grpo.minus.mt-grpo.code}{$-$1.0}
\setval{8b.grpo.minus.mt-grpo.code-abs}{1.0}
\setval{8b.grpo.minus.mt-grpo.creative-writing-v3}{+2.1}
\setval{8b.grpo.minus.mt-grpo.creative-writing-v3-abs}{2.1}
\setval{8b.grpo.minus.mt-grpo.guidebench}{$-$1.2}
\setval{8b.grpo.minus.mt-grpo.guidebench-abs}{1.2}
\setval{8b.grpo.minus.mt-grpo.humaneval-plus}{$-$1.3}
\setval{8b.grpo.minus.mt-grpo.humaneval-plus-abs}{1.3}
\setval{8b.grpo.minus.mt-grpo.if}{+1.1}
\setval{8b.grpo.minus.mt-grpo.if-abs}{1.1}
\setval{8b.grpo.minus.mt-grpo.ifbench}{+4.7}
\setval{8b.grpo.minus.mt-grpo.ifbench-abs}{4.7}
\setval{8b.grpo.minus.mt-grpo.ifeval}{$-$0.1}
\setval{8b.grpo.minus.mt-grpo.ifeval-abs}{0.1}
\setval{8b.grpo.minus.mt-grpo.judge2.avg}{+0.0}
\setval{8b.grpo.minus.mt-grpo.judge2.avg-abs}{0.0}
\setval{8b.grpo.minus.mt-grpo.judge2.creative-writing-v3}{$-$2.3}
\setval{8b.grpo.minus.mt-grpo.judge2.creative-writing-v3-abs}{2.3}
\setval{8b.grpo.minus.mt-grpo.judge2.longbench-write}{+0.0}
\setval{8b.grpo.minus.mt-grpo.judge2.longbench-write-abs}{0.0}
\setval{8b.grpo.minus.mt-grpo.judge2.writing}{$-$0.3}
\setval{8b.grpo.minus.mt-grpo.judge2.writing-abs}{0.3}
\setval{8b.grpo.minus.mt-grpo.judge2.writingbench}{+1.4}
\setval{8b.grpo.minus.mt-grpo.judge2.writingbench-abs}{1.4}
\setval{8b.grpo.minus.mt-grpo.livecodebench}{$-$1.1}
\setval{8b.grpo.minus.mt-grpo.livecodebench-abs}{1.1}
\setval{8b.grpo.minus.mt-grpo.longbench-write}{+2.3}
\setval{8b.grpo.minus.mt-grpo.longbench-write-abs}{2.3}
\setval{8b.grpo.minus.mt-grpo.math}{+0.2}
\setval{8b.grpo.minus.mt-grpo.math-abs}{0.2}
\setval{8b.grpo.minus.mt-grpo.math500}{+0.1}
\setval{8b.grpo.minus.mt-grpo.math500-abs}{0.1}
\setval{8b.grpo.minus.mt-grpo.mbpp-plus}{$-$0.6}
\setval{8b.grpo.minus.mt-grpo.mbpp-plus-abs}{0.6}
\setval{8b.grpo.minus.mt-grpo.minerva}{$-$1.2}
\setval{8b.grpo.minus.mt-grpo.minerva-abs}{1.2}
\setval{8b.grpo.minus.mt-grpo.olympiad-bench}{+2.2}
\setval{8b.grpo.minus.mt-grpo.olympiad-bench-abs}{2.2}
\setval{8b.grpo.minus.mt-grpo.writing}{+3.1}
\setval{8b.grpo.minus.mt-grpo.writing-abs}{3.1}
\setval{8b.grpo.minus.mt-grpo.writingbench}{+4.8}
\setval{8b.grpo.minus.mt-grpo.writingbench-abs}{4.8}
\setval{8b.grpo.minus.papo.aime24}{+3.6}
\setval{8b.grpo.minus.papo.aime24-abs}{3.6}
\setval{8b.grpo.minus.papo.aime25}{+5.4}
\setval{8b.grpo.minus.papo.aime25-abs}{5.4}
\setval{8b.grpo.minus.papo.amc23}{+2.3}
\setval{8b.grpo.minus.papo.amc23-abs}{2.3}
\setval{8b.grpo.minus.papo.avg}{+2.1}
\setval{8b.grpo.minus.papo.avg-abs}{2.1}
\setval{8b.grpo.minus.papo.code}{+0.8}
\setval{8b.grpo.minus.papo.code-abs}{0.8}
\setval{8b.grpo.minus.papo.creative-writing-v3}{+1.3}
\setval{8b.grpo.minus.papo.creative-writing-v3-abs}{1.3}
\setval{8b.grpo.minus.papo.guidebench}{$-$2.5}
\setval{8b.grpo.minus.papo.guidebench-abs}{2.5}
\setval{8b.grpo.minus.papo.humaneval-plus}{+1.2}
\setval{8b.grpo.minus.papo.humaneval-plus-abs}{1.2}
\setval{8b.grpo.minus.papo.if}{+0.9}
\setval{8b.grpo.minus.papo.if-abs}{0.9}
\setval{8b.grpo.minus.papo.ifbench}{+4.7}
\setval{8b.grpo.minus.papo.ifbench-abs}{4.7}
\setval{8b.grpo.minus.papo.ifeval}{+0.5}
\setval{8b.grpo.minus.papo.ifeval-abs}{0.5}
\setval{8b.grpo.minus.papo.judge2.avg}{+1.5}
\setval{8b.grpo.minus.papo.judge2.avg-abs}{1.5}
\setval{8b.grpo.minus.papo.judge2.creative-writing-v3}{$-$2.6}
\setval{8b.grpo.minus.papo.judge2.creative-writing-v3-abs}{2.6}
\setval{8b.grpo.minus.papo.judge2.longbench-write}{+2.9}
\setval{8b.grpo.minus.papo.judge2.longbench-write-abs}{2.9}
\setval{8b.grpo.minus.papo.judge2.writing}{+0.3}
\setval{8b.grpo.minus.papo.judge2.writing-abs}{0.3}
\setval{8b.grpo.minus.papo.judge2.writingbench}{+0.6}
\setval{8b.grpo.minus.papo.judge2.writingbench-abs}{0.6}
\setval{8b.grpo.minus.papo.livecodebench}{$-$0.1}
\setval{8b.grpo.minus.papo.livecodebench-abs}{0.1}
\setval{8b.grpo.minus.papo.longbench-write}{+4.3}
\setval{8b.grpo.minus.papo.longbench-write-abs}{4.3}
\setval{8b.grpo.minus.papo.math}{+2.9}
\setval{8b.grpo.minus.papo.math-abs}{2.9}
\setval{8b.grpo.minus.papo.math500}{+2.2}
\setval{8b.grpo.minus.papo.math500-abs}{2.2}
\setval{8b.grpo.minus.papo.mbpp-plus}{+1.2}
\setval{8b.grpo.minus.papo.mbpp-plus-abs}{1.2}
\setval{8b.grpo.minus.papo.minerva}{+0.5}
\setval{8b.grpo.minus.papo.minerva-abs}{0.5}
\setval{8b.grpo.minus.papo.olympiad-bench}{+3.1}
\setval{8b.grpo.minus.papo.olympiad-bench-abs}{3.1}
\setval{8b.grpo.minus.papo.writing}{+3.0}
\setval{8b.grpo.minus.papo.writing-abs}{3.0}
\setval{8b.grpo.minus.papo.writingbench}{+3.3}
\setval{8b.grpo.minus.papo.writingbench-abs}{3.3}
\setval{8b.grpo.minus.saw.aime24}{+0.5}
\setval{8b.grpo.minus.saw.aime24-abs}{0.5}
\setval{8b.grpo.minus.saw.aime25}{+0.0}
\setval{8b.grpo.minus.saw.aime25-abs}{0.0}
\setval{8b.grpo.minus.saw.amc23}{+2.1}
\setval{8b.grpo.minus.saw.amc23-abs}{2.1}
\setval{8b.grpo.minus.saw.avg}{$-$0.1}
\setval{8b.grpo.minus.saw.avg-abs}{0.1}
\setval{8b.grpo.minus.saw.code}{$-$0.7}
\setval{8b.grpo.minus.saw.code-abs}{0.7}
\setval{8b.grpo.minus.saw.creative-writing-v3}{+2.8}
\setval{8b.grpo.minus.saw.creative-writing-v3-abs}{2.8}
\setval{8b.grpo.minus.saw.guidebench}{$-$2.4}
\setval{8b.grpo.minus.saw.guidebench-abs}{2.4}
\setval{8b.grpo.minus.saw.humaneval-plus}{$-$2.1}
\setval{8b.grpo.minus.saw.humaneval-plus-abs}{2.1}
\setval{8b.grpo.minus.saw.if}{$-$0.4}
\setval{8b.grpo.minus.saw.if-abs}{0.4}
\setval{8b.grpo.minus.saw.ifbench}{+2.0}
\setval{8b.grpo.minus.saw.ifbench-abs}{2.0}
\setval{8b.grpo.minus.saw.ifeval}{$-$0.8}
\setval{8b.grpo.minus.saw.ifeval-abs}{0.8}
\setval{8b.grpo.minus.saw.judge2.avg}{$-$0.2}
\setval{8b.grpo.minus.saw.judge2.avg-abs}{0.2}
\setval{8b.grpo.minus.saw.judge2.creative-writing-v3}{+2.7}
\setval{8b.grpo.minus.saw.judge2.creative-writing-v3-abs}{2.7}
\setval{8b.grpo.minus.saw.judge2.longbench-write}{$-$2.4}
\setval{8b.grpo.minus.saw.judge2.longbench-write-abs}{2.4}
\setval{8b.grpo.minus.saw.judge2.writing}{+0.1}
\setval{8b.grpo.minus.saw.judge2.writing-abs}{0.1}
\setval{8b.grpo.minus.saw.judge2.writingbench}{+0.0}
\setval{8b.grpo.minus.saw.judge2.writingbench-abs}{0.0}
\setval{8b.grpo.minus.saw.livecodebench}{$-$0.2}
\setval{8b.grpo.minus.saw.livecodebench-abs}{0.2}
\setval{8b.grpo.minus.saw.longbench-write}{$-$1.2}
\setval{8b.grpo.minus.saw.longbench-write-abs}{1.2}
\setval{8b.grpo.minus.saw.math}{$-$0.1}
\setval{8b.grpo.minus.saw.math-abs}{0.1}
\setval{8b.grpo.minus.saw.math500}{$-$1.3}
\setval{8b.grpo.minus.saw.math500-abs}{1.3}
\setval{8b.grpo.minus.saw.mbpp-plus}{+0.3}
\setval{8b.grpo.minus.saw.mbpp-plus-abs}{0.3}
\setval{8b.grpo.minus.saw.minerva}{$-$1.9}
\setval{8b.grpo.minus.saw.minerva-abs}{1.9}
\setval{8b.grpo.minus.saw.olympiad-bench}{$-$0.2}
\setval{8b.grpo.minus.saw.olympiad-bench-abs}{0.2}
\setval{8b.grpo.minus.saw.writing}{+0.9}
\setval{8b.grpo.minus.saw.writing-abs}{0.9}
\setval{8b.grpo.minus.saw.writingbench}{+1.1}
\setval{8b.grpo.minus.saw.writingbench-abs}{1.1}
\setval{8b.grpo.passing-at-ceiling.code.early}{0.71}
\setval{8b.grpo.passing-at-ceiling.code.late}{0.77}
\setval{8b.grpo.passing-at-ceiling.code.run}{0.75}
\setval{8b.grpo.passing-at-ceiling.if.early}{0.32}
\setval{8b.grpo.passing-at-ceiling.if.late}{0.56}
\setval{8b.grpo.passing-at-ceiling.if.run}{0.48}
\setval{8b.grpo.passing-at-ceiling.math.early}{1.00}
\setval{8b.grpo.passing-at-ceiling.math.late}{1.00}
\setval{8b.grpo.passing-at-ceiling.math.run}{1.00}
\setval{8b.grpo.passing-at-ceiling.writing.early}{0.00}
\setval{8b.grpo.passing-at-ceiling.writing.late}{0.01}
\setval{8b.grpo.passing-at-ceiling.writing.run}{0.00}
\setval{8b.grpo.primary-level.code.early}{0.54}
\setval{8b.grpo.primary-level.code.late}{0.77}
\setval{8b.grpo.primary-level.code.run}{0.72}
\setval{8b.grpo.primary-level.if.early}{0.35}
\setval{8b.grpo.primary-level.if.late}{0.74}
\setval{8b.grpo.primary-level.if.run}{0.64}
\setval{8b.grpo.primary-level.math.early}{0.49}
\setval{8b.grpo.primary-level.math.late}{0.69}
\setval{8b.grpo.primary-level.math.run}{0.63}
\setval{8b.grpo.primary-level.writing.early}{0.53}
\setval{8b.grpo.primary-level.writing.late}{0.76}
\setval{8b.grpo.primary-level.writing.run}{0.72}
\setval{8b.mt-grpo.aux-change.code}{+0.17}
\setval{8b.mt-grpo.aux-change.if}{+0.16}
\setval{8b.mt-grpo.aux-change.math}{+0.13}
\setval{8b.mt-grpo.aux-change.writing}{+0.13}
\setval{8b.mt-grpo.aux-change.writing-rm}{+0.32}
\setval{8b.mt-grpo.aux-full.code.early}{0.70}
\setval{8b.mt-grpo.aux-full.code.late}{0.89}
\setval{8b.mt-grpo.aux-full.code.run}{0.88}
\setval{8b.mt-grpo.aux-full.if.early}{0.72}
\setval{8b.mt-grpo.aux-full.if.late}{0.88}
\setval{8b.mt-grpo.aux-full.if.run}{0.86}
\setval{8b.mt-grpo.aux-full.math.early}{0.58}
\setval{8b.mt-grpo.aux-full.math.late}{0.86}
\setval{8b.mt-grpo.aux-full.math.run}{0.81}
\setval{8b.mt-grpo.aux-full.writing-rm.early}{0.04}
\setval{8b.mt-grpo.aux-full.writing-rm.late}{0.57}
\setval{8b.mt-grpo.aux-full.writing-rm.run}{0.43}
\setval{8b.mt-grpo.aux-full.writing.early}{0.73}
\setval{8b.mt-grpo.aux-full.writing.late}{0.75}
\setval{8b.mt-grpo.aux-full.writing.run}{0.71}
\setval{8b.mt-grpo.aux-level.code.early}{0.93}
\setval{8b.mt-grpo.aux-level.code.late}{0.98}
\setval{8b.mt-grpo.aux-level.code.run}{0.98}
\setval{8b.mt-grpo.aux-level.if.early}{0.87}
\setval{8b.mt-grpo.aux-level.if.late}{0.95}
\setval{8b.mt-grpo.aux-level.if.run}{0.94}
\setval{8b.mt-grpo.aux-level.math.early}{0.91}
\setval{8b.mt-grpo.aux-level.math.late}{0.98}
\setval{8b.mt-grpo.aux-level.math.run}{0.96}
\setval{8b.mt-grpo.aux-level.writing-rm.early}{0.75}
\setval{8b.mt-grpo.aux-level.writing-rm.late}{0.96}
\setval{8b.mt-grpo.aux-level.writing-rm.run}{0.92}
\setval{8b.mt-grpo.aux-level.writing.early}{0.94}
\setval{8b.mt-grpo.aux-level.writing.late}{0.97}
\setval{8b.mt-grpo.aux-level.writing.run}{0.96}
\setval{8b.mt-grpo.bench.aime24}{14.5}
\setval{8b.mt-grpo.bench.aime25}{15.7}
\setval{8b.mt-grpo.bench.amc23}{65.6}
\setval{8b.mt-grpo.bench.creative-writing-v3}{50.3}
\setval{8b.mt-grpo.bench.guidebench}{83.1}
\setval{8b.mt-grpo.bench.humaneval-plus}{85.1}
\setval{8b.mt-grpo.bench.ifbench}{37.3}
\setval{8b.mt-grpo.bench.ifeval}{86.9}
\setval{8b.mt-grpo.bench.livecodebench}{27.7}
\setval{8b.mt-grpo.bench.longbench-write}{67.4}
\setval{8b.mt-grpo.bench.math500}{80.6}
\setval{8b.mt-grpo.bench.mbpp-plus}{75.3}
\setval{8b.mt-grpo.bench.minerva}{51.0}
\setval{8b.mt-grpo.bench.olympiad-bench}{45.7}
\setval{8b.mt-grpo.bench.writingbench}{71.7}
\setval{8b.mt-grpo.eval.avg}{57.2}
\setval{8b.mt-grpo.eval.code}{62.7}
\setval{8b.mt-grpo.eval.if}{69.1}
\setval{8b.mt-grpo.eval.math}{45.5}
\setval{8b.mt-grpo.eval.writing}{63.2}
\setval{8b.mt-grpo.judge2.avg}{55.2}
\setval{8b.mt-grpo.judge2.creative-writing-v3}{41.5}
\setval{8b.mt-grpo.judge2.longbench-write}{62.5}
\setval{8b.mt-grpo.judge2.writing}{53.2}
\setval{8b.mt-grpo.judge2.writingbench}{55.5}
\setval{8b.mt-grpo.last-step}{300}
\setval{8b.mt-grpo.minus.base.aime24}{+3.1}
\setval{8b.mt-grpo.minus.base.aime24-abs}{3.1}
\setval{8b.mt-grpo.minus.base.aime25}{+7.2}
\setval{8b.mt-grpo.minus.base.aime25-abs}{7.2}
\setval{8b.mt-grpo.minus.base.amc23}{+15.4}
\setval{8b.mt-grpo.minus.base.amc23-abs}{15.4}
\setval{8b.mt-grpo.minus.base.avg}{+15.7}
\setval{8b.mt-grpo.minus.base.avg-abs}{15.7}
\setval{8b.mt-grpo.minus.base.code}{+6.6}
\setval{8b.mt-grpo.minus.base.code-abs}{6.6}
\setval{8b.mt-grpo.minus.base.creative-writing-v3}{+23.6}
\setval{8b.mt-grpo.minus.base.creative-writing-v3-abs}{23.6}
\setval{8b.mt-grpo.minus.base.guidebench}{+37.5}
\setval{8b.mt-grpo.minus.base.guidebench-abs}{37.5}
\setval{8b.mt-grpo.minus.base.humaneval-plus}{+5.0}
\setval{8b.mt-grpo.minus.base.humaneval-plus-abs}{5.0}
\setval{8b.mt-grpo.minus.base.if}{+30.5}
\setval{8b.mt-grpo.minus.base.if-abs}{30.5}
\setval{8b.mt-grpo.minus.base.ifbench}{+20.0}
\setval{8b.mt-grpo.minus.base.ifbench-abs}{20.0}
\setval{8b.mt-grpo.minus.base.ifeval}{+33.9}
\setval{8b.mt-grpo.minus.base.ifeval-abs}{33.9}
\setval{8b.mt-grpo.minus.base.judge2.avg}{+14.2}
\setval{8b.mt-grpo.minus.base.judge2.avg-abs}{14.2}
\setval{8b.mt-grpo.minus.base.judge2.creative-writing-v3}{+15.1}
\setval{8b.mt-grpo.minus.base.judge2.creative-writing-v3-abs}{15.1}
\setval{8b.mt-grpo.minus.base.judge2.longbench-write}{+14.4}
\setval{8b.mt-grpo.minus.base.judge2.longbench-write-abs}{14.4}
\setval{8b.mt-grpo.minus.base.judge2.writing}{+17.0}
\setval{8b.mt-grpo.minus.base.judge2.writing-abs}{17.0}
\setval{8b.mt-grpo.minus.base.judge2.writingbench}{+21.5}
\setval{8b.mt-grpo.minus.base.judge2.writingbench-abs}{21.5}
\setval{8b.mt-grpo.minus.base.livecodebench}{+6.7}
\setval{8b.mt-grpo.minus.base.livecodebench-abs}{6.7}
\setval{8b.mt-grpo.minus.base.longbench-write}{+16.0}
\setval{8b.mt-grpo.minus.base.longbench-write-abs}{16.0}
\setval{8b.mt-grpo.minus.base.math}{+8.4}
\setval{8b.mt-grpo.minus.base.math-abs}{8.4}
\setval{8b.mt-grpo.minus.base.math500}{+8.1}
\setval{8b.mt-grpo.minus.base.math500-abs}{8.1}
\setval{8b.mt-grpo.minus.base.mbpp-plus}{+8.0}
\setval{8b.mt-grpo.minus.base.mbpp-plus-abs}{8.0}
\setval{8b.mt-grpo.minus.base.minerva}{+10.1}
\setval{8b.mt-grpo.minus.base.minerva-abs}{10.1}
\setval{8b.mt-grpo.minus.base.olympiad-bench}{+6.3}
\setval{8b.mt-grpo.minus.base.olympiad-bench-abs}{6.3}
\setval{8b.mt-grpo.minus.base.writing}{+24.9}
\setval{8b.mt-grpo.minus.base.writing-abs}{24.9}
\setval{8b.mt-grpo.minus.base.writingbench}{+35.0}
\setval{8b.mt-grpo.minus.base.writingbench-abs}{35.0}
\setval{8b.mt-grpo.minus.concur-free.aime24}{$-$5.2}
\setval{8b.mt-grpo.minus.concur-free.aime24-abs}{5.2}
\setval{8b.mt-grpo.minus.concur-free.aime25}{$-$0.8}
\setval{8b.mt-grpo.minus.concur-free.aime25-abs}{0.8}
\setval{8b.mt-grpo.minus.concur-free.amc23}{+0.9}
\setval{8b.mt-grpo.minus.concur-free.amc23-abs}{0.9}
\setval{8b.mt-grpo.minus.concur-free.avg}{$-$3.1}
\setval{8b.mt-grpo.minus.concur-free.avg-abs}{3.1}
\setval{8b.mt-grpo.minus.concur-free.code}{+0.2}
\setval{8b.mt-grpo.minus.concur-free.code-abs}{0.2}
\setval{8b.mt-grpo.minus.concur-free.creative-writing-v3}{$-$1.3}
\setval{8b.mt-grpo.minus.concur-free.creative-writing-v3-abs}{1.3}
\setval{8b.mt-grpo.minus.concur-free.guidebench}{+0.2}
\setval{8b.mt-grpo.minus.concur-free.guidebench-abs}{0.2}
\setval{8b.mt-grpo.minus.concur-free.humaneval-plus}{$-$0.2}
\setval{8b.mt-grpo.minus.concur-free.humaneval-plus-abs}{0.2}
\setval{8b.mt-grpo.minus.concur-free.if}{$-$1.1}
\setval{8b.mt-grpo.minus.concur-free.if-abs}{1.1}
\setval{8b.mt-grpo.minus.concur-free.ifbench}{$-$1.3}
\setval{8b.mt-grpo.minus.concur-free.ifbench-abs}{1.3}
\setval{8b.mt-grpo.minus.concur-free.ifeval}{$-$2.3}
\setval{8b.mt-grpo.minus.concur-free.ifeval-abs}{2.3}
\setval{8b.mt-grpo.minus.concur-free.judge2.avg}{$-$2.1}
\setval{8b.mt-grpo.minus.concur-free.judge2.avg-abs}{2.1}
\setval{8b.mt-grpo.minus.concur-free.judge2.creative-writing-v3}{$-$1.0}
\setval{8b.mt-grpo.minus.concur-free.judge2.creative-writing-v3-abs}{1.0}
\setval{8b.mt-grpo.minus.concur-free.judge2.longbench-write}{$-$11.6}
\setval{8b.mt-grpo.minus.concur-free.judge2.longbench-write-abs}{11.6}
\setval{8b.mt-grpo.minus.concur-free.judge2.writing}{$-$5.9}
\setval{8b.mt-grpo.minus.concur-free.judge2.writing-abs}{5.9}
\setval{8b.mt-grpo.minus.concur-free.judge2.writingbench}{$-$5.2}
\setval{8b.mt-grpo.minus.concur-free.judge2.writingbench-abs}{5.2}
\setval{8b.mt-grpo.minus.concur-free.livecodebench}{+0.6}
\setval{8b.mt-grpo.minus.concur-free.livecodebench-abs}{0.6}
\setval{8b.mt-grpo.minus.concur-free.longbench-write}{$-$15.5}
\setval{8b.mt-grpo.minus.concur-free.longbench-write-abs}{15.5}
\setval{8b.mt-grpo.minus.concur-free.math}{$-$1.8}
\setval{8b.mt-grpo.minus.concur-free.math-abs}{1.8}
\setval{8b.mt-grpo.minus.concur-free.math500}{$-$2.8}
\setval{8b.mt-grpo.minus.concur-free.math500-abs}{2.8}
\setval{8b.mt-grpo.minus.concur-free.mbpp-plus}{+0.2}
\setval{8b.mt-grpo.minus.concur-free.mbpp-plus-abs}{0.2}
\setval{8b.mt-grpo.minus.concur-free.minerva}{+1.3}
\setval{8b.mt-grpo.minus.concur-free.minerva-abs}{1.3}
\setval{8b.mt-grpo.minus.concur-free.olympiad-bench}{$-$4.3}
\setval{8b.mt-grpo.minus.concur-free.olympiad-bench-abs}{4.3}
\setval{8b.mt-grpo.minus.concur-free.writing}{$-$10.6}
\setval{8b.mt-grpo.minus.concur-free.writing-abs}{10.6}
\setval{8b.mt-grpo.minus.concur-free.writingbench}{$-$15.1}
\setval{8b.mt-grpo.minus.concur-free.writingbench-abs}{15.1}
\setval{8b.mt-grpo.minus.concur-tied.aime24}{$-$16.8}
\setval{8b.mt-grpo.minus.concur-tied.aime24-abs}{16.8}
\setval{8b.mt-grpo.minus.concur-tied.aime25}{$-$8.2}
\setval{8b.mt-grpo.minus.concur-tied.aime25-abs}{8.2}
\setval{8b.mt-grpo.minus.concur-tied.amc23}{$-$12.9}
\setval{8b.mt-grpo.minus.concur-tied.amc23-abs}{12.9}
\setval{8b.mt-grpo.minus.concur-tied.avg}{$-$6.8}
\setval{8b.mt-grpo.minus.concur-tied.avg-abs}{6.8}
\setval{8b.mt-grpo.minus.concur-tied.code}{$-$0.5}
\setval{8b.mt-grpo.minus.concur-tied.code-abs}{0.5}
\setval{8b.mt-grpo.minus.concur-tied.creative-writing-v3}{$-$2.1}
\setval{8b.mt-grpo.minus.concur-tied.creative-writing-v3-abs}{2.1}
\setval{8b.mt-grpo.minus.concur-tied.guidebench}{$-$0.3}
\setval{8b.mt-grpo.minus.concur-tied.guidebench-abs}{0.3}
\setval{8b.mt-grpo.minus.concur-tied.humaneval-plus}{$-$2.3}
\setval{8b.mt-grpo.minus.concur-tied.humaneval-plus-abs}{2.3}
\setval{8b.mt-grpo.minus.concur-tied.if}{$-$1.4}
\setval{8b.mt-grpo.minus.concur-tied.if-abs}{1.4}
\setval{8b.mt-grpo.minus.concur-tied.ifbench}{$-$2.0}
\setval{8b.mt-grpo.minus.concur-tied.ifbench-abs}{2.0}
\setval{8b.mt-grpo.minus.concur-tied.ifeval}{$-$1.9}
\setval{8b.mt-grpo.minus.concur-tied.ifeval-abs}{1.9}
\setval{8b.mt-grpo.minus.concur-tied.judge2.avg}{$-$6.1}
\setval{8b.mt-grpo.minus.concur-tied.judge2.avg-abs}{6.1}
\setval{8b.mt-grpo.minus.concur-tied.judge2.creative-writing-v3}{$-$0.2}
\setval{8b.mt-grpo.minus.concur-tied.judge2.creative-writing-v3-abs}{0.2}
\setval{8b.mt-grpo.minus.concur-tied.judge2.longbench-write}{$-$12.3}
\setval{8b.mt-grpo.minus.concur-tied.judge2.longbench-write-abs}{12.3}
\setval{8b.mt-grpo.minus.concur-tied.judge2.writing}{$-$6.9}
\setval{8b.mt-grpo.minus.concur-tied.judge2.writing-abs}{6.9}
\setval{8b.mt-grpo.minus.concur-tied.judge2.writingbench}{$-$8.3}
\setval{8b.mt-grpo.minus.concur-tied.judge2.writingbench-abs}{8.3}
\setval{8b.mt-grpo.minus.concur-tied.livecodebench}{+0.5}
\setval{8b.mt-grpo.minus.concur-tied.livecodebench-abs}{0.5}
\setval{8b.mt-grpo.minus.concur-tied.longbench-write}{$-$13.8}
\setval{8b.mt-grpo.minus.concur-tied.longbench-write-abs}{13.8}
\setval{8b.mt-grpo.minus.concur-tied.math}{$-$10.8}
\setval{8b.mt-grpo.minus.concur-tied.math-abs}{10.8}
\setval{8b.mt-grpo.minus.concur-tied.math500}{$-$9.0}
\setval{8b.mt-grpo.minus.concur-tied.math500-abs}{9.0}
\setval{8b.mt-grpo.minus.concur-tied.mbpp-plus}{+0.4}
\setval{8b.mt-grpo.minus.concur-tied.mbpp-plus-abs}{0.4}
\setval{8b.mt-grpo.minus.concur-tied.minerva}{$-$4.4}
\setval{8b.mt-grpo.minus.concur-tied.minerva-abs}{4.4}
\setval{8b.mt-grpo.minus.concur-tied.olympiad-bench}{$-$13.8}
\setval{8b.mt-grpo.minus.concur-tied.olympiad-bench-abs}{13.8}
\setval{8b.mt-grpo.minus.concur-tied.writing}{$-$10.5}
\setval{8b.mt-grpo.minus.concur-tied.writing-abs}{10.5}
\setval{8b.mt-grpo.minus.concur-tied.writingbench}{$-$15.5}
\setval{8b.mt-grpo.minus.concur-tied.writingbench-abs}{15.5}
\setval{8b.mt-grpo.minus.concur.aime24}{$-$14.3}
\setval{8b.mt-grpo.minus.concur.aime24-abs}{14.3}
\setval{8b.mt-grpo.minus.concur.aime25}{$-$7.4}
\setval{8b.mt-grpo.minus.concur.aime25-abs}{7.4}
\setval{8b.mt-grpo.minus.concur.amc23}{$-$13.3}
\setval{8b.mt-grpo.minus.concur.amc23-abs}{13.3}
\setval{8b.mt-grpo.minus.concur.avg}{$-$6.0}
\setval{8b.mt-grpo.minus.concur.avg-abs}{6.0}
\setval{8b.mt-grpo.minus.concur.code}{+1.9}
\setval{8b.mt-grpo.minus.concur.code-abs}{1.9}
\setval{8b.mt-grpo.minus.concur.creative-writing-v3}{+0.4}
\setval{8b.mt-grpo.minus.concur.creative-writing-v3-abs}{0.4}
\setval{8b.mt-grpo.minus.concur.guidebench}{$-$0.3}
\setval{8b.mt-grpo.minus.concur.guidebench-abs}{0.3}
\setval{8b.mt-grpo.minus.concur.humaneval-plus}{+2.3}
\setval{8b.mt-grpo.minus.concur.humaneval-plus-abs}{2.3}
\setval{8b.mt-grpo.minus.concur.if}{$-$0.9}
\setval{8b.mt-grpo.minus.concur.if-abs}{0.9}
\setval{8b.mt-grpo.minus.concur.ifbench}{$-$1.3}
\setval{8b.mt-grpo.minus.concur.ifbench-abs}{1.3}
\setval{8b.mt-grpo.minus.concur.ifeval}{$-$1.2}
\setval{8b.mt-grpo.minus.concur.ifeval-abs}{1.2}
\setval{8b.mt-grpo.minus.concur.judge2.avg}{$-$5.1}
\setval{8b.mt-grpo.minus.concur.judge2.avg-abs}{5.1}
\setval{8b.mt-grpo.minus.concur.judge2.creative-writing-v3}{+3.8}
\setval{8b.mt-grpo.minus.concur.judge2.creative-writing-v3-abs}{3.8}
\setval{8b.mt-grpo.minus.concur.judge2.longbench-write}{$-$13.0}
\setval{8b.mt-grpo.minus.concur.judge2.longbench-write-abs}{13.0}
\setval{8b.mt-grpo.minus.concur.judge2.writing}{$-$5.9}
\setval{8b.mt-grpo.minus.concur.judge2.writing-abs}{5.9}
\setval{8b.mt-grpo.minus.concur.judge2.writingbench}{$-$8.4}
\setval{8b.mt-grpo.minus.concur.judge2.writingbench-abs}{8.4}
\setval{8b.mt-grpo.minus.concur.livecodebench}{+1.3}
\setval{8b.mt-grpo.minus.concur.livecodebench-abs}{1.3}
\setval{8b.mt-grpo.minus.concur.longbench-write}{$-$14.2}
\setval{8b.mt-grpo.minus.concur.longbench-write-abs}{14.2}
\setval{8b.mt-grpo.minus.concur.math}{$-$10.4}
\setval{8b.mt-grpo.minus.concur.math-abs}{10.4}
\setval{8b.mt-grpo.minus.concur.math500}{$-$9.5}
\setval{8b.mt-grpo.minus.concur.math500-abs}{9.5}
\setval{8b.mt-grpo.minus.concur.mbpp-plus}{+2.1}
\setval{8b.mt-grpo.minus.concur.mbpp-plus-abs}{2.1}
\setval{8b.mt-grpo.minus.concur.minerva}{$-$3.1}
\setval{8b.mt-grpo.minus.concur.minerva-abs}{3.1}
\setval{8b.mt-grpo.minus.concur.olympiad-bench}{$-$14.6}
\setval{8b.mt-grpo.minus.concur.olympiad-bench-abs}{14.6}
\setval{8b.mt-grpo.minus.concur.writing}{$-$10.0}
\setval{8b.mt-grpo.minus.concur.writing-abs}{10.0}
\setval{8b.mt-grpo.minus.concur.writingbench}{$-$16.4}
\setval{8b.mt-grpo.minus.concur.writingbench-abs}{16.4}
\setval{8b.mt-grpo.minus.gd2po.aime24}{$-$1.8}
\setval{8b.mt-grpo.minus.gd2po.aime24-abs}{1.8}
\setval{8b.mt-grpo.minus.gd2po.aime25}{+3.5}
\setval{8b.mt-grpo.minus.gd2po.aime25-abs}{3.5}
\setval{8b.mt-grpo.minus.gd2po.amc23}{+6.0}
\setval{8b.mt-grpo.minus.gd2po.amc23-abs}{6.0}
\setval{8b.mt-grpo.minus.gd2po.avg}{$-$0.2}
\setval{8b.mt-grpo.minus.gd2po.avg-abs}{0.2}
\setval{8b.mt-grpo.minus.gd2po.code}{+0.9}
\setval{8b.mt-grpo.minus.gd2po.code-abs}{0.9}
\setval{8b.mt-grpo.minus.gd2po.creative-writing-v3}{+2.2}
\setval{8b.mt-grpo.minus.gd2po.creative-writing-v3-abs}{2.2}
\setval{8b.mt-grpo.minus.gd2po.guidebench}{$-$0.6}
\setval{8b.mt-grpo.minus.gd2po.guidebench-abs}{0.6}
\setval{8b.mt-grpo.minus.gd2po.humaneval-plus}{+1.1}
\setval{8b.mt-grpo.minus.gd2po.humaneval-plus-abs}{1.1}
\setval{8b.mt-grpo.minus.gd2po.if}{$-$2.9}
\setval{8b.mt-grpo.minus.gd2po.if-abs}{2.9}
\setval{8b.mt-grpo.minus.gd2po.ifbench}{$-$5.7}
\setval{8b.mt-grpo.minus.gd2po.ifbench-abs}{5.7}
\setval{8b.mt-grpo.minus.gd2po.ifeval}{$-$2.4}
\setval{8b.mt-grpo.minus.gd2po.ifeval-abs}{2.4}
\setval{8b.mt-grpo.minus.gd2po.judge2.avg}{+0.3}
\setval{8b.mt-grpo.minus.gd2po.judge2.avg-abs}{0.3}
\setval{8b.mt-grpo.minus.gd2po.judge2.creative-writing-v3}{+4.8}
\setval{8b.mt-grpo.minus.gd2po.judge2.creative-writing-v3-abs}{4.8}
\setval{8b.mt-grpo.minus.gd2po.judge2.longbench-write}{+0.3}
\setval{8b.mt-grpo.minus.gd2po.judge2.longbench-write-abs}{0.3}
\setval{8b.mt-grpo.minus.gd2po.judge2.writing}{+1.8}
\setval{8b.mt-grpo.minus.gd2po.judge2.writing-abs}{1.8}
\setval{8b.mt-grpo.minus.gd2po.judge2.writingbench}{+0.3}
\setval{8b.mt-grpo.minus.gd2po.judge2.writingbench-abs}{0.3}
\setval{8b.mt-grpo.minus.gd2po.livecodebench}{+1.4}
\setval{8b.mt-grpo.minus.gd2po.livecodebench-abs}{1.4}
\setval{8b.mt-grpo.minus.gd2po.longbench-write}{$-$2.4}
\setval{8b.mt-grpo.minus.gd2po.longbench-write-abs}{2.4}
\setval{8b.mt-grpo.minus.gd2po.math}{+0.9}
\setval{8b.mt-grpo.minus.gd2po.math-abs}{0.9}
\setval{8b.mt-grpo.minus.gd2po.math500}{$-$0.6}
\setval{8b.mt-grpo.minus.gd2po.math500-abs}{0.6}
\setval{8b.mt-grpo.minus.gd2po.mbpp-plus}{+0.2}
\setval{8b.mt-grpo.minus.gd2po.mbpp-plus-abs}{0.2}
\setval{8b.mt-grpo.minus.gd2po.minerva}{+1.1}
\setval{8b.mt-grpo.minus.gd2po.minerva-abs}{1.1}
\setval{8b.mt-grpo.minus.gd2po.olympiad-bench}{$-$2.8}
\setval{8b.mt-grpo.minus.gd2po.olympiad-bench-abs}{2.8}
\setval{8b.mt-grpo.minus.gd2po.writing}{$-$0.8}
\setval{8b.mt-grpo.minus.gd2po.writing-abs}{0.8}
\setval{8b.mt-grpo.minus.gd2po.writingbench}{$-$2.2}
\setval{8b.mt-grpo.minus.gd2po.writingbench-abs}{2.2}
\setval{8b.mt-grpo.minus.gdpo-noaux.aime24}{$-$13.4}
\setval{8b.mt-grpo.minus.gdpo-noaux.aime24-abs}{13.4}
\setval{8b.mt-grpo.minus.gdpo-noaux.aime25}{$-$8.2}
\setval{8b.mt-grpo.minus.gdpo-noaux.aime25-abs}{8.2}
\setval{8b.mt-grpo.minus.gdpo-noaux.amc23}{$-$11.5}
\setval{8b.mt-grpo.minus.gdpo-noaux.amc23-abs}{11.5}
\setval{8b.mt-grpo.minus.gdpo-noaux.avg}{$-$3.3}
\setval{8b.mt-grpo.minus.gdpo-noaux.avg-abs}{3.3}
\setval{8b.mt-grpo.minus.gdpo-noaux.code}{$-$0.1}
\setval{8b.mt-grpo.minus.gdpo-noaux.code-abs}{0.1}
\setval{8b.mt-grpo.minus.gdpo-noaux.creative-writing-v3}{+6.4}
\setval{8b.mt-grpo.minus.gdpo-noaux.creative-writing-v3-abs}{6.4}
\setval{8b.mt-grpo.minus.gdpo-noaux.guidebench}{+1.7}
\setval{8b.mt-grpo.minus.gdpo-noaux.guidebench-abs}{1.7}
\setval{8b.mt-grpo.minus.gdpo-noaux.humaneval-plus}{$-$0.9}
\setval{8b.mt-grpo.minus.gdpo-noaux.humaneval-plus-abs}{0.9}
\setval{8b.mt-grpo.minus.gdpo-noaux.if}{+1.9}
\setval{8b.mt-grpo.minus.gdpo-noaux.if-abs}{1.9}
\setval{8b.mt-grpo.minus.gdpo-noaux.ifbench}{+3.7}
\setval{8b.mt-grpo.minus.gdpo-noaux.ifbench-abs}{3.7}
\setval{8b.mt-grpo.minus.gdpo-noaux.ifeval}{+0.4}
\setval{8b.mt-grpo.minus.gdpo-noaux.ifeval-abs}{0.4}
\setval{8b.mt-grpo.minus.gdpo-noaux.judge2.avg}{$-$1.9}
\setval{8b.mt-grpo.minus.gdpo-noaux.judge2.avg-abs}{1.9}
\setval{8b.mt-grpo.minus.gdpo-noaux.judge2.creative-writing-v3}{+12.7}
\setval{8b.mt-grpo.minus.gdpo-noaux.judge2.creative-writing-v3-abs}{12.7}
\setval{8b.mt-grpo.minus.gdpo-noaux.judge2.longbench-write}{+8.0}
\setval{8b.mt-grpo.minus.gdpo-noaux.judge2.longbench-write-abs}{8.0}
\setval{8b.mt-grpo.minus.gdpo-noaux.judge2.writing}{+8.7}
\setval{8b.mt-grpo.minus.gdpo-noaux.judge2.writing-abs}{8.7}
\setval{8b.mt-grpo.minus.gdpo-noaux.judge2.writingbench}{+5.5}
\setval{8b.mt-grpo.minus.gdpo-noaux.judge2.writingbench-abs}{5.5}
\setval{8b.mt-grpo.minus.gdpo-noaux.livecodebench}{+0.3}
\setval{8b.mt-grpo.minus.gdpo-noaux.livecodebench-abs}{0.3}
\setval{8b.mt-grpo.minus.gdpo-noaux.longbench-write}{+1.5}
\setval{8b.mt-grpo.minus.gdpo-noaux.longbench-write-abs}{1.5}
\setval{8b.mt-grpo.minus.gdpo-noaux.math}{$-$10.0}
\setval{8b.mt-grpo.minus.gdpo-noaux.math-abs}{10.0}
\setval{8b.mt-grpo.minus.gdpo-noaux.math500}{$-$9.7}
\setval{8b.mt-grpo.minus.gdpo-noaux.math500-abs}{9.7}
\setval{8b.mt-grpo.minus.gdpo-noaux.mbpp-plus}{+0.4}
\setval{8b.mt-grpo.minus.gdpo-noaux.mbpp-plus-abs}{0.4}
\setval{8b.mt-grpo.minus.gdpo-noaux.minerva}{$-$2.9}
\setval{8b.mt-grpo.minus.gdpo-noaux.minerva-abs}{2.9}
\setval{8b.mt-grpo.minus.gdpo-noaux.olympiad-bench}{$-$14.2}
\setval{8b.mt-grpo.minus.gdpo-noaux.olympiad-bench-abs}{14.2}
\setval{8b.mt-grpo.minus.gdpo-noaux.writing}{+1.5}
\setval{8b.mt-grpo.minus.gdpo-noaux.writing-abs}{1.5}
\setval{8b.mt-grpo.minus.gdpo-noaux.writingbench}{$-$3.3}
\setval{8b.mt-grpo.minus.gdpo-noaux.writingbench-abs}{3.3}
\setval{8b.mt-grpo.minus.gdpo-w0.5.aime24}{$-$4.8}
\setval{8b.mt-grpo.minus.gdpo-w0.5.aime24-abs}{4.8}
\setval{8b.mt-grpo.minus.gdpo-w0.5.aime25}{+0.3}
\setval{8b.mt-grpo.minus.gdpo-w0.5.aime25-abs}{0.3}
\setval{8b.mt-grpo.minus.gdpo-w0.5.amc23}{+2.2}
\setval{8b.mt-grpo.minus.gdpo-w0.5.amc23-abs}{2.2}
\setval{8b.mt-grpo.minus.gdpo-w0.5.avg}{$-$1.1}
\setval{8b.mt-grpo.minus.gdpo-w0.5.avg-abs}{1.1}
\setval{8b.mt-grpo.minus.gdpo-w0.5.code}{+1.5}
\setval{8b.mt-grpo.minus.gdpo-w0.5.code-abs}{1.5}
\setval{8b.mt-grpo.minus.gdpo-w0.5.creative-writing-v3}{$-$1.2}
\setval{8b.mt-grpo.minus.gdpo-w0.5.creative-writing-v3-abs}{1.2}
\setval{8b.mt-grpo.minus.gdpo-w0.5.guidebench}{$-$0.9}
\setval{8b.mt-grpo.minus.gdpo-w0.5.guidebench-abs}{0.9}
\setval{8b.mt-grpo.minus.gdpo-w0.5.humaneval-plus}{+1.3}
\setval{8b.mt-grpo.minus.gdpo-w0.5.humaneval-plus-abs}{1.3}
\setval{8b.mt-grpo.minus.gdpo-w0.5.if}{$-$0.7}
\setval{8b.mt-grpo.minus.gdpo-w0.5.if-abs}{0.7}
\setval{8b.mt-grpo.minus.gdpo-w0.5.ifbench}{$-$0.3}
\setval{8b.mt-grpo.minus.gdpo-w0.5.ifbench-abs}{0.3}
\setval{8b.mt-grpo.minus.gdpo-w0.5.ifeval}{$-$0.8}
\setval{8b.mt-grpo.minus.gdpo-w0.5.ifeval-abs}{0.8}
\setval{8b.mt-grpo.minus.gdpo-w0.5.judge2.avg}{$-$0.6}
\setval{8b.mt-grpo.minus.gdpo-w0.5.judge2.avg-abs}{0.6}
\setval{8b.mt-grpo.minus.gdpo-w0.5.judge2.creative-writing-v3}{+1.2}
\setval{8b.mt-grpo.minus.gdpo-w0.5.judge2.creative-writing-v3-abs}{1.2}
\setval{8b.mt-grpo.minus.gdpo-w0.5.judge2.longbench-write}{$-$1.2}
\setval{8b.mt-grpo.minus.gdpo-w0.5.judge2.longbench-write-abs}{1.2}
\setval{8b.mt-grpo.minus.gdpo-w0.5.judge2.writing}{$-$0.2}
\setval{8b.mt-grpo.minus.gdpo-w0.5.judge2.writing-abs}{0.2}
\setval{8b.mt-grpo.minus.gdpo-w0.5.judge2.writingbench}{$-$0.7}
\setval{8b.mt-grpo.minus.gdpo-w0.5.judge2.writingbench-abs}{0.7}
\setval{8b.mt-grpo.minus.gdpo-w0.5.livecodebench}{+0.9}
\setval{8b.mt-grpo.minus.gdpo-w0.5.livecodebench-abs}{0.9}
\setval{8b.mt-grpo.minus.gdpo-w0.5.longbench-write}{$-$3.7}
\setval{8b.mt-grpo.minus.gdpo-w0.5.longbench-write-abs}{3.7}
\setval{8b.mt-grpo.minus.gdpo-w0.5.math}{$-$1.9}
\setval{8b.mt-grpo.minus.gdpo-w0.5.math-abs}{1.9}
\setval{8b.mt-grpo.minus.gdpo-w0.5.math500}{$-$3.5}
\setval{8b.mt-grpo.minus.gdpo-w0.5.math500-abs}{3.5}
\setval{8b.mt-grpo.minus.gdpo-w0.5.mbpp-plus}{+2.4}
\setval{8b.mt-grpo.minus.gdpo-w0.5.mbpp-plus-abs}{2.4}
\setval{8b.mt-grpo.minus.gdpo-w0.5.minerva}{$-$0.2}
\setval{8b.mt-grpo.minus.gdpo-w0.5.minerva-abs}{0.2}
\setval{8b.mt-grpo.minus.gdpo-w0.5.olympiad-bench}{$-$5.4}
\setval{8b.mt-grpo.minus.gdpo-w0.5.olympiad-bench-abs}{5.4}
\setval{8b.mt-grpo.minus.gdpo-w0.5.writing}{$-$2.8}
\setval{8b.mt-grpo.minus.gdpo-w0.5.writing-abs}{2.8}
\setval{8b.mt-grpo.minus.gdpo-w0.5.writingbench}{$-$3.3}
\setval{8b.mt-grpo.minus.gdpo-w0.5.writingbench-abs}{3.3}
\setval{8b.mt-grpo.minus.gdpo.aime24}{$-$6.9}
\setval{8b.mt-grpo.minus.gdpo.aime24-abs}{6.9}
\setval{8b.mt-grpo.minus.gdpo.aime25}{+1.4}
\setval{8b.mt-grpo.minus.gdpo.aime25-abs}{1.4}
\setval{8b.mt-grpo.minus.gdpo.amc23}{+3.4}
\setval{8b.mt-grpo.minus.gdpo.amc23-abs}{3.4}
\setval{8b.mt-grpo.minus.gdpo.avg}{$-$1.1}
\setval{8b.mt-grpo.minus.gdpo.avg-abs}{1.1}
\setval{8b.mt-grpo.minus.gdpo.code}{+1.0}
\setval{8b.mt-grpo.minus.gdpo.code-abs}{1.0}
\setval{8b.mt-grpo.minus.gdpo.creative-writing-v3}{$-$2.6}
\setval{8b.mt-grpo.minus.gdpo.creative-writing-v3-abs}{2.6}
\setval{8b.mt-grpo.minus.gdpo.guidebench}{$-$1.4}
\setval{8b.mt-grpo.minus.gdpo.guidebench-abs}{1.4}
\setval{8b.mt-grpo.minus.gdpo.humaneval-plus}{+1.0}
\setval{8b.mt-grpo.minus.gdpo.humaneval-plus-abs}{1.0}
\setval{8b.mt-grpo.minus.gdpo.if}{$-$1.5}
\setval{8b.mt-grpo.minus.gdpo.if-abs}{1.5}
\setval{8b.mt-grpo.minus.gdpo.ifbench}{$-$2.0}
\setval{8b.mt-grpo.minus.gdpo.ifbench-abs}{2.0}
\setval{8b.mt-grpo.minus.gdpo.ifeval}{$-$1.2}
\setval{8b.mt-grpo.minus.gdpo.ifeval-abs}{1.2}
\setval{8b.mt-grpo.minus.gdpo.judge2.avg}{$-$0.8}
\setval{8b.mt-grpo.minus.gdpo.judge2.avg-abs}{0.8}
\setval{8b.mt-grpo.minus.gdpo.judge2.creative-writing-v3}{+0.5}
\setval{8b.mt-grpo.minus.gdpo.judge2.creative-writing-v3-abs}{0.5}
\setval{8b.mt-grpo.minus.gdpo.judge2.longbench-write}{$-$1.7}
\setval{8b.mt-grpo.minus.gdpo.judge2.longbench-write-abs}{1.7}
\setval{8b.mt-grpo.minus.gdpo.judge2.writing}{$-$1.4}
\setval{8b.mt-grpo.minus.gdpo.judge2.writing-abs}{1.4}
\setval{8b.mt-grpo.minus.gdpo.judge2.writingbench}{$-$2.9}
\setval{8b.mt-grpo.minus.gdpo.judge2.writingbench-abs}{2.9}
\setval{8b.mt-grpo.minus.gdpo.livecodebench}{+1.4}
\setval{8b.mt-grpo.minus.gdpo.livecodebench-abs}{1.4}
\setval{8b.mt-grpo.minus.gdpo.longbench-write}{$-$2.1}
\setval{8b.mt-grpo.minus.gdpo.longbench-write-abs}{2.1}
\setval{8b.mt-grpo.minus.gdpo.math}{$-$1.1}
\setval{8b.mt-grpo.minus.gdpo.math-abs}{1.1}
\setval{8b.mt-grpo.minus.gdpo.math500}{$-$2.4}
\setval{8b.mt-grpo.minus.gdpo.math500-abs}{2.4}
\setval{8b.mt-grpo.minus.gdpo.mbpp-plus}{+0.7}
\setval{8b.mt-grpo.minus.gdpo.mbpp-plus-abs}{0.7}
\setval{8b.mt-grpo.minus.gdpo.minerva}{+1.4}
\setval{8b.mt-grpo.minus.gdpo.minerva-abs}{1.4}
\setval{8b.mt-grpo.minus.gdpo.olympiad-bench}{$-$3.6}
\setval{8b.mt-grpo.minus.gdpo.olympiad-bench-abs}{3.6}
\setval{8b.mt-grpo.minus.gdpo.writing}{$-$3.0}
\setval{8b.mt-grpo.minus.gdpo.writing-abs}{3.0}
\setval{8b.mt-grpo.minus.gdpo.writingbench}{$-$4.4}
\setval{8b.mt-grpo.minus.gdpo.writingbench-abs}{4.4}
\setval{8b.mt-grpo.minus.geoma.aime24}{+0.4}
\setval{8b.mt-grpo.minus.geoma.aime24-abs}{0.4}
\setval{8b.mt-grpo.minus.geoma.aime25}{+3.2}
\setval{8b.mt-grpo.minus.geoma.aime25-abs}{3.2}
\setval{8b.mt-grpo.minus.geoma.amc23}{+2.1}
\setval{8b.mt-grpo.minus.geoma.amc23-abs}{2.1}
\setval{8b.mt-grpo.minus.geoma.avg}{+0.5}
\setval{8b.mt-grpo.minus.geoma.avg-abs}{0.5}
\setval{8b.mt-grpo.minus.geoma.code}{+1.3}
\setval{8b.mt-grpo.minus.geoma.code-abs}{1.3}
\setval{8b.mt-grpo.minus.geoma.creative-writing-v3}{$-$0.1}
\setval{8b.mt-grpo.minus.geoma.creative-writing-v3-abs}{0.1}
\setval{8b.mt-grpo.minus.geoma.guidebench}{$-$1.2}
\setval{8b.mt-grpo.minus.geoma.guidebench-abs}{1.2}
\setval{8b.mt-grpo.minus.geoma.humaneval-plus}{+2.3}
\setval{8b.mt-grpo.minus.geoma.humaneval-plus-abs}{2.3}
\setval{8b.mt-grpo.minus.geoma.if}{+0.6}
\setval{8b.mt-grpo.minus.geoma.if-abs}{0.6}
\setval{8b.mt-grpo.minus.geoma.ifbench}{+3.3}
\setval{8b.mt-grpo.minus.geoma.ifbench-abs}{3.3}
\setval{8b.mt-grpo.minus.geoma.ifeval}{$-$0.4}
\setval{8b.mt-grpo.minus.geoma.ifeval-abs}{0.4}
\setval{8b.mt-grpo.minus.geoma.judge2.avg}{+0.8}
\setval{8b.mt-grpo.minus.geoma.judge2.avg-abs}{0.8}
\setval{8b.mt-grpo.minus.geoma.judge2.creative-writing-v3}{+4.0}
\setval{8b.mt-grpo.minus.geoma.judge2.creative-writing-v3-abs}{4.0}
\setval{8b.mt-grpo.minus.geoma.judge2.longbench-write}{$-$2.4}
\setval{8b.mt-grpo.minus.geoma.judge2.longbench-write-abs}{2.4}
\setval{8b.mt-grpo.minus.geoma.judge2.writing}{$-$0.4}
\setval{8b.mt-grpo.minus.geoma.judge2.writing-abs}{0.4}
\setval{8b.mt-grpo.minus.geoma.judge2.writingbench}{$-$2.8}
\setval{8b.mt-grpo.minus.geoma.judge2.writingbench-abs}{2.8}
\setval{8b.mt-grpo.minus.geoma.livecodebench}{+1.9}
\setval{8b.mt-grpo.minus.geoma.livecodebench-abs}{1.9}
\setval{8b.mt-grpo.minus.geoma.longbench-write}{$-$1.5}
\setval{8b.mt-grpo.minus.geoma.longbench-write-abs}{1.5}
\setval{8b.mt-grpo.minus.geoma.math}{+1.3}
\setval{8b.mt-grpo.minus.geoma.math-abs}{1.3}
\setval{8b.mt-grpo.minus.geoma.math500}{+1.0}
\setval{8b.mt-grpo.minus.geoma.math500-abs}{1.0}
\setval{8b.mt-grpo.minus.geoma.mbpp-plus}{$-$0.3}
\setval{8b.mt-grpo.minus.geoma.mbpp-plus-abs}{0.3}
\setval{8b.mt-grpo.minus.geoma.minerva}{+1.3}
\setval{8b.mt-grpo.minus.geoma.minerva-abs}{1.3}
\setval{8b.mt-grpo.minus.geoma.olympiad-bench}{$-$0.2}
\setval{8b.mt-grpo.minus.geoma.olympiad-bench-abs}{0.2}
\setval{8b.mt-grpo.minus.geoma.writing}{$-$1.9}
\setval{8b.mt-grpo.minus.geoma.writing-abs}{1.9}
\setval{8b.mt-grpo.minus.geoma.writingbench}{$-$4.2}
\setval{8b.mt-grpo.minus.geoma.writingbench-abs}{4.2}
\setval{8b.mt-grpo.minus.grpo-nodiv.aime24}{$-$6.1}
\setval{8b.mt-grpo.minus.grpo-nodiv.aime24-abs}{6.1}
\setval{8b.mt-grpo.minus.grpo-nodiv.aime25}{+0.9}
\setval{8b.mt-grpo.minus.grpo-nodiv.aime25-abs}{0.9}
\setval{8b.mt-grpo.minus.grpo-nodiv.amc23}{+2.0}
\setval{8b.mt-grpo.minus.grpo-nodiv.amc23-abs}{2.0}
\setval{8b.mt-grpo.minus.grpo-nodiv.avg}{$-$0.9}
\setval{8b.mt-grpo.minus.grpo-nodiv.avg-abs}{0.9}
\setval{8b.mt-grpo.minus.grpo-nodiv.code}{+0.2}
\setval{8b.mt-grpo.minus.grpo-nodiv.code-abs}{0.2}
\setval{8b.mt-grpo.minus.grpo-nodiv.creative-writing-v3}{+3.7}
\setval{8b.mt-grpo.minus.grpo-nodiv.creative-writing-v3-abs}{3.7}
\setval{8b.mt-grpo.minus.grpo-nodiv.guidebench}{$-$1.7}
\setval{8b.mt-grpo.minus.grpo-nodiv.guidebench-abs}{1.7}
\setval{8b.mt-grpo.minus.grpo-nodiv.humaneval-plus}{$-$1.3}
\setval{8b.mt-grpo.minus.grpo-nodiv.humaneval-plus-abs}{1.3}
\setval{8b.mt-grpo.minus.grpo-nodiv.if}{$-$1.1}
\setval{8b.mt-grpo.minus.grpo-nodiv.if-abs}{1.1}
\setval{8b.mt-grpo.minus.grpo-nodiv.ifbench}{+1.0}
\setval{8b.mt-grpo.minus.grpo-nodiv.ifbench-abs}{1.0}
\setval{8b.mt-grpo.minus.grpo-nodiv.ifeval}{$-$2.5}
\setval{8b.mt-grpo.minus.grpo-nodiv.ifeval-abs}{2.5}
\setval{8b.mt-grpo.minus.grpo-nodiv.judge2.avg}{$-$0.9}
\setval{8b.mt-grpo.minus.grpo-nodiv.judge2.avg-abs}{0.9}
\setval{8b.mt-grpo.minus.grpo-nodiv.judge2.creative-writing-v3}{+4.0}
\setval{8b.mt-grpo.minus.grpo-nodiv.judge2.creative-writing-v3-abs}{4.0}
\setval{8b.mt-grpo.minus.grpo-nodiv.judge2.longbench-write}{$-$4.1}
\setval{8b.mt-grpo.minus.grpo-nodiv.judge2.longbench-write-abs}{4.1}
\setval{8b.mt-grpo.minus.grpo-nodiv.judge2.writing}{+0.5}
\setval{8b.mt-grpo.minus.grpo-nodiv.judge2.writing-abs}{0.5}
\setval{8b.mt-grpo.minus.grpo-nodiv.judge2.writingbench}{+1.6}
\setval{8b.mt-grpo.minus.grpo-nodiv.judge2.writingbench-abs}{1.6}
\setval{8b.mt-grpo.minus.grpo-nodiv.livecodebench}{+2.7}
\setval{8b.mt-grpo.minus.grpo-nodiv.livecodebench-abs}{2.7}
\setval{8b.mt-grpo.minus.grpo-nodiv.longbench-write}{$-$3.6}
\setval{8b.mt-grpo.minus.grpo-nodiv.longbench-write-abs}{3.6}
\setval{8b.mt-grpo.minus.grpo-nodiv.math}{$-$2.0}
\setval{8b.mt-grpo.minus.grpo-nodiv.math-abs}{2.0}
\setval{8b.mt-grpo.minus.grpo-nodiv.math500}{$-$3.5}
\setval{8b.mt-grpo.minus.grpo-nodiv.math500-abs}{3.5}
\setval{8b.mt-grpo.minus.grpo-nodiv.mbpp-plus}{$-$0.8}
\setval{8b.mt-grpo.minus.grpo-nodiv.mbpp-plus-abs}{0.8}
\setval{8b.mt-grpo.minus.grpo-nodiv.minerva}{+0.2}
\setval{8b.mt-grpo.minus.grpo-nodiv.minerva-abs}{0.2}
\setval{8b.mt-grpo.minus.grpo-nodiv.olympiad-bench}{$-$5.7}
\setval{8b.mt-grpo.minus.grpo-nodiv.olympiad-bench-abs}{5.7}
\setval{8b.mt-grpo.minus.grpo-nodiv.writing}{+0.5}
\setval{8b.mt-grpo.minus.grpo-nodiv.writing-abs}{0.5}
\setval{8b.mt-grpo.minus.grpo-nodiv.writingbench}{+1.4}
\setval{8b.mt-grpo.minus.grpo-nodiv.writingbench-abs}{1.4}
\setval{8b.mt-grpo.minus.grpo.aime24}{$-$1.4}
\setval{8b.mt-grpo.minus.grpo.aime24-abs}{1.4}
\setval{8b.mt-grpo.minus.grpo.aime25}{+0.2}
\setval{8b.mt-grpo.minus.grpo.aime25-abs}{0.2}
\setval{8b.mt-grpo.minus.grpo.amc23}{+1.2}
\setval{8b.mt-grpo.minus.grpo.amc23-abs}{1.2}
\setval{8b.mt-grpo.minus.grpo.avg}{$-$0.7}
\setval{8b.mt-grpo.minus.grpo.avg-abs}{0.7}
\setval{8b.mt-grpo.minus.grpo.code}{+1.0}
\setval{8b.mt-grpo.minus.grpo.code-abs}{1.0}
\setval{8b.mt-grpo.minus.grpo.creative-writing-v3}{$-$2.1}
\setval{8b.mt-grpo.minus.grpo.creative-writing-v3-abs}{2.1}
\setval{8b.mt-grpo.minus.grpo.guidebench}{+1.2}
\setval{8b.mt-grpo.minus.grpo.guidebench-abs}{1.2}
\setval{8b.mt-grpo.minus.grpo.humaneval-plus}{+1.3}
\setval{8b.mt-grpo.minus.grpo.humaneval-plus-abs}{1.3}
\setval{8b.mt-grpo.minus.grpo.if}{$-$1.1}
\setval{8b.mt-grpo.minus.grpo.if-abs}{1.1}
\setval{8b.mt-grpo.minus.grpo.ifbench}{$-$4.7}
\setval{8b.mt-grpo.minus.grpo.ifbench-abs}{4.7}
\setval{8b.mt-grpo.minus.grpo.ifeval}{+0.1}
\setval{8b.mt-grpo.minus.grpo.ifeval-abs}{0.1}
\setval{8b.mt-grpo.minus.grpo.judge2.avg}{+0.0}
\setval{8b.mt-grpo.minus.grpo.judge2.avg-abs}{0.0}
\setval{8b.mt-grpo.minus.grpo.judge2.creative-writing-v3}{+2.3}
\setval{8b.mt-grpo.minus.grpo.judge2.creative-writing-v3-abs}{2.3}
\setval{8b.mt-grpo.minus.grpo.judge2.longbench-write}{+0.0}
\setval{8b.mt-grpo.minus.grpo.judge2.longbench-write-abs}{0.0}
\setval{8b.mt-grpo.minus.grpo.judge2.writing}{+0.3}
\setval{8b.mt-grpo.minus.grpo.judge2.writing-abs}{0.3}
\setval{8b.mt-grpo.minus.grpo.judge2.writingbench}{$-$1.4}
\setval{8b.mt-grpo.minus.grpo.judge2.writingbench-abs}{1.4}
\setval{8b.mt-grpo.minus.grpo.livecodebench}{+1.1}
\setval{8b.mt-grpo.minus.grpo.livecodebench-abs}{1.1}
\setval{8b.mt-grpo.minus.grpo.longbench-write}{$-$2.3}
\setval{8b.mt-grpo.minus.grpo.longbench-write-abs}{2.3}
\setval{8b.mt-grpo.minus.grpo.math}{$-$0.2}
\setval{8b.mt-grpo.minus.grpo.math-abs}{0.2}
\setval{8b.mt-grpo.minus.grpo.math500}{$-$0.1}
\setval{8b.mt-grpo.minus.grpo.math500-abs}{0.1}
\setval{8b.mt-grpo.minus.grpo.mbpp-plus}{+0.6}
\setval{8b.mt-grpo.minus.grpo.mbpp-plus-abs}{0.6}
\setval{8b.mt-grpo.minus.grpo.minerva}{+1.2}
\setval{8b.mt-grpo.minus.grpo.minerva-abs}{1.2}
\setval{8b.mt-grpo.minus.grpo.olympiad-bench}{$-$2.2}
\setval{8b.mt-grpo.minus.grpo.olympiad-bench-abs}{2.2}
\setval{8b.mt-grpo.minus.grpo.writing}{$-$3.1}
\setval{8b.mt-grpo.minus.grpo.writing-abs}{3.1}
\setval{8b.mt-grpo.minus.grpo.writingbench}{$-$4.8}
\setval{8b.mt-grpo.minus.grpo.writingbench-abs}{4.8}
\setval{8b.mt-grpo.minus.papo.aime24}{+2.3}
\setval{8b.mt-grpo.minus.papo.aime24-abs}{2.3}
\setval{8b.mt-grpo.minus.papo.aime25}{+5.6}
\setval{8b.mt-grpo.minus.papo.aime25-abs}{5.6}
\setval{8b.mt-grpo.minus.papo.amc23}{+3.6}
\setval{8b.mt-grpo.minus.papo.amc23-abs}{3.6}
\setval{8b.mt-grpo.minus.papo.avg}{+1.4}
\setval{8b.mt-grpo.minus.papo.avg-abs}{1.4}
\setval{8b.mt-grpo.minus.papo.code}{+1.8}
\setval{8b.mt-grpo.minus.papo.code-abs}{1.8}
\setval{8b.mt-grpo.minus.papo.creative-writing-v3}{$-$0.8}
\setval{8b.mt-grpo.minus.papo.creative-writing-v3-abs}{0.8}
\setval{8b.mt-grpo.minus.papo.guidebench}{$-$1.2}
\setval{8b.mt-grpo.minus.papo.guidebench-abs}{1.2}
\setval{8b.mt-grpo.minus.papo.humaneval-plus}{+2.6}
\setval{8b.mt-grpo.minus.papo.humaneval-plus-abs}{2.6}
\setval{8b.mt-grpo.minus.papo.if}{$-$0.2}
\setval{8b.mt-grpo.minus.papo.if-abs}{0.2}
\setval{8b.mt-grpo.minus.papo.ifbench}{+0.0}
\setval{8b.mt-grpo.minus.papo.ifbench-abs}{0.0}
\setval{8b.mt-grpo.minus.papo.ifeval}{+0.6}
\setval{8b.mt-grpo.minus.papo.ifeval-abs}{0.6}
\setval{8b.mt-grpo.minus.papo.judge2.avg}{+1.5}
\setval{8b.mt-grpo.minus.papo.judge2.avg-abs}{1.5}
\setval{8b.mt-grpo.minus.papo.judge2.creative-writing-v3}{$-$0.3}
\setval{8b.mt-grpo.minus.papo.judge2.creative-writing-v3-abs}{0.3}
\setval{8b.mt-grpo.minus.papo.judge2.longbench-write}{+2.9}
\setval{8b.mt-grpo.minus.papo.judge2.longbench-write-abs}{2.9}
\setval{8b.mt-grpo.minus.papo.judge2.writing}{+0.6}
\setval{8b.mt-grpo.minus.papo.judge2.writing-abs}{0.6}
\setval{8b.mt-grpo.minus.papo.judge2.writingbench}{$-$0.8}
\setval{8b.mt-grpo.minus.papo.judge2.writingbench-abs}{0.8}
\setval{8b.mt-grpo.minus.papo.livecodebench}{+1.0}
\setval{8b.mt-grpo.minus.papo.livecodebench-abs}{1.0}
\setval{8b.mt-grpo.minus.papo.longbench-write}{+1.9}
\setval{8b.mt-grpo.minus.papo.longbench-write-abs}{1.9}
\setval{8b.mt-grpo.minus.papo.math}{+2.7}
\setval{8b.mt-grpo.minus.papo.math-abs}{2.7}
\setval{8b.mt-grpo.minus.papo.math500}{+2.1}
\setval{8b.mt-grpo.minus.papo.math500-abs}{2.1}
\setval{8b.mt-grpo.minus.papo.mbpp-plus}{+1.9}
\setval{8b.mt-grpo.minus.papo.mbpp-plus-abs}{1.9}
\setval{8b.mt-grpo.minus.papo.minerva}{+1.7}
\setval{8b.mt-grpo.minus.papo.minerva-abs}{1.7}
\setval{8b.mt-grpo.minus.papo.olympiad-bench}{+0.9}
\setval{8b.mt-grpo.minus.papo.olympiad-bench-abs}{0.9}
\setval{8b.mt-grpo.minus.papo.writing}{$-$0.1}
\setval{8b.mt-grpo.minus.papo.writing-abs}{0.1}
\setval{8b.mt-grpo.minus.papo.writingbench}{$-$1.5}
\setval{8b.mt-grpo.minus.papo.writingbench-abs}{1.5}
\setval{8b.mt-grpo.minus.saw.aime24}{$-$0.8}
\setval{8b.mt-grpo.minus.saw.aime24-abs}{0.8}
\setval{8b.mt-grpo.minus.saw.aime25}{+0.2}
\setval{8b.mt-grpo.minus.saw.aime25-abs}{0.2}
\setval{8b.mt-grpo.minus.saw.amc23}{+3.4}
\setval{8b.mt-grpo.minus.saw.amc23-abs}{3.4}
\setval{8b.mt-grpo.minus.saw.avg}{$-$0.8}
\setval{8b.mt-grpo.minus.saw.avg-abs}{0.8}
\setval{8b.mt-grpo.minus.saw.code}{+0.4}
\setval{8b.mt-grpo.minus.saw.code-abs}{0.4}
\setval{8b.mt-grpo.minus.saw.creative-writing-v3}{+0.7}
\setval{8b.mt-grpo.minus.saw.creative-writing-v3-abs}{0.7}
\setval{8b.mt-grpo.minus.saw.guidebench}{$-$1.2}
\setval{8b.mt-grpo.minus.saw.guidebench-abs}{1.2}
\setval{8b.mt-grpo.minus.saw.humaneval-plus}{$-$0.8}
\setval{8b.mt-grpo.minus.saw.humaneval-plus-abs}{0.8}
\setval{8b.mt-grpo.minus.saw.if}{$-$1.5}
\setval{8b.mt-grpo.minus.saw.if-abs}{1.5}
\setval{8b.mt-grpo.minus.saw.ifbench}{$-$2.7}
\setval{8b.mt-grpo.minus.saw.ifbench-abs}{2.7}
\setval{8b.mt-grpo.minus.saw.ifeval}{$-$0.7}
\setval{8b.mt-grpo.minus.saw.ifeval-abs}{0.7}
\setval{8b.mt-grpo.minus.saw.judge2.avg}{$-$0.3}
\setval{8b.mt-grpo.minus.saw.judge2.avg-abs}{0.3}
\setval{8b.mt-grpo.minus.saw.judge2.creative-writing-v3}{+5.0}
\setval{8b.mt-grpo.minus.saw.judge2.creative-writing-v3-abs}{5.0}
\setval{8b.mt-grpo.minus.saw.judge2.longbench-write}{$-$2.3}
\setval{8b.mt-grpo.minus.saw.judge2.longbench-write-abs}{2.3}
\setval{8b.mt-grpo.minus.saw.judge2.writing}{+0.4}
\setval{8b.mt-grpo.minus.saw.judge2.writing-abs}{0.4}
\setval{8b.mt-grpo.minus.saw.judge2.writingbench}{$-$1.4}
\setval{8b.mt-grpo.minus.saw.judge2.writingbench-abs}{1.4}
\setval{8b.mt-grpo.minus.saw.livecodebench}{+1.0}
\setval{8b.mt-grpo.minus.saw.livecodebench-abs}{1.0}
\setval{8b.mt-grpo.minus.saw.longbench-write}{$-$3.6}
\setval{8b.mt-grpo.minus.saw.longbench-write-abs}{3.6}
\setval{8b.mt-grpo.minus.saw.math}{$-$0.3}
\setval{8b.mt-grpo.minus.saw.math-abs}{0.3}
\setval{8b.mt-grpo.minus.saw.math500}{$-$1.3}
\setval{8b.mt-grpo.minus.saw.math500-abs}{1.3}
\setval{8b.mt-grpo.minus.saw.mbpp-plus}{+0.9}
\setval{8b.mt-grpo.minus.saw.mbpp-plus-abs}{0.9}
\setval{8b.mt-grpo.minus.saw.minerva}{$-$0.7}
\setval{8b.mt-grpo.minus.saw.minerva-abs}{0.7}
\setval{8b.mt-grpo.minus.saw.olympiad-bench}{$-$2.4}
\setval{8b.mt-grpo.minus.saw.olympiad-bench-abs}{2.4}
\setval{8b.mt-grpo.minus.saw.writing}{$-$2.2}
\setval{8b.mt-grpo.minus.saw.writing-abs}{2.2}
\setval{8b.mt-grpo.minus.saw.writingbench}{$-$3.7}
\setval{8b.mt-grpo.minus.saw.writingbench-abs}{3.7}
\setval{8b.mt-grpo.passing-at-ceiling.code.early}{0.67}
\setval{8b.mt-grpo.passing-at-ceiling.code.late}{0.47}
\setval{8b.mt-grpo.passing-at-ceiling.code.run}{0.53}
\setval{8b.mt-grpo.passing-at-ceiling.if.early}{0.30}
\setval{8b.mt-grpo.passing-at-ceiling.if.late}{0.44}
\setval{8b.mt-grpo.passing-at-ceiling.if.run}{0.41}
\setval{8b.mt-grpo.passing-at-ceiling.math.early}{1.00}
\setval{8b.mt-grpo.passing-at-ceiling.math.late}{1.00}
\setval{8b.mt-grpo.passing-at-ceiling.math.run}{1.00}
\setval{8b.mt-grpo.passing-at-ceiling.writing.early}{0.00}
\setval{8b.mt-grpo.passing-at-ceiling.writing.late}{0.00}
\setval{8b.mt-grpo.passing-at-ceiling.writing.run}{0.00}
\setval{8b.mt-grpo.primary-level.code.early}{0.53}
\setval{8b.mt-grpo.primary-level.code.late}{0.63}
\setval{8b.mt-grpo.primary-level.code.run}{0.60}
\setval{8b.mt-grpo.primary-level.if.early}{0.36}
\setval{8b.mt-grpo.primary-level.if.late}{0.72}
\setval{8b.mt-grpo.primary-level.if.run}{0.64}
\setval{8b.mt-grpo.primary-level.math.early}{0.57}
\setval{8b.mt-grpo.primary-level.math.late}{0.69}
\setval{8b.mt-grpo.primary-level.math.run}{0.64}
\setval{8b.mt-grpo.primary-level.writing.early}{0.51}
\setval{8b.mt-grpo.primary-level.writing.late}{0.75}
\setval{8b.mt-grpo.primary-level.writing.run}{0.70}
\setval{8b.papo.aux-change.code}{+0.19}
\setval{8b.papo.aux-change.if}{+0.18}
\setval{8b.papo.aux-change.math}{+0.15}
\setval{8b.papo.aux-change.writing}{+0.15}
\setval{8b.papo.aux-change.writing-rm}{+0.33}
\setval{8b.papo.aux-full.code.early}{0.80}
\setval{8b.papo.aux-full.code.late}{0.99}
\setval{8b.papo.aux-full.code.run}{0.97}
\setval{8b.papo.aux-full.if.early}{0.74}
\setval{8b.papo.aux-full.if.late}{0.95}
\setval{8b.papo.aux-full.if.run}{0.91}
\setval{8b.papo.aux-full.math.early}{0.63}
\setval{8b.papo.aux-full.math.late}{0.98}
\setval{8b.papo.aux-full.math.run}{0.93}
\setval{8b.papo.aux-full.writing-rm.early}{0.06}
\setval{8b.papo.aux-full.writing-rm.late}{0.67}
\setval{8b.papo.aux-full.writing-rm.run}{0.54}
\setval{8b.papo.aux-full.writing.early}{0.71}
\setval{8b.papo.aux-full.writing.late}{0.86}
\setval{8b.papo.aux-full.writing.run}{0.79}
\setval{8b.papo.aux-level.code.early}{0.95}
\setval{8b.papo.aux-level.code.late}{1.00}
\setval{8b.papo.aux-level.code.run}{0.99}
\setval{8b.papo.aux-level.if.early}{0.89}
\setval{8b.papo.aux-level.if.late}{0.97}
\setval{8b.papo.aux-level.if.run}{0.95}
\setval{8b.papo.aux-level.math.early}{0.92}
\setval{8b.papo.aux-level.math.late}{1.00}
\setval{8b.papo.aux-level.math.run}{0.98}
\setval{8b.papo.aux-level.writing-rm.early}{0.77}
\setval{8b.papo.aux-level.writing-rm.late}{0.97}
\setval{8b.papo.aux-level.writing-rm.run}{0.94}
\setval{8b.papo.aux-level.writing.early}{0.93}
\setval{8b.papo.aux-level.writing.late}{0.99}
\setval{8b.papo.aux-level.writing.run}{0.97}
\setval{8b.papo.bench.aime24}{12.2}
\setval{8b.papo.bench.aime25}{10.1}
\setval{8b.papo.bench.amc23}{62.0}
\setval{8b.papo.bench.creative-writing-v3}{51.1}
\setval{8b.papo.bench.guidebench}{84.4}
\setval{8b.papo.bench.humaneval-plus}{82.5}
\setval{8b.papo.bench.ifbench}{37.3}
\setval{8b.papo.bench.ifeval}{86.3}
\setval{8b.papo.bench.livecodebench}{26.6}
\setval{8b.papo.bench.longbench-write}{65.5}
\setval{8b.papo.bench.math500}{78.5}
\setval{8b.papo.bench.mbpp-plus}{73.4}
\setval{8b.papo.bench.minerva}{49.3}
\setval{8b.papo.bench.olympiad-bench}{44.8}
\setval{8b.papo.bench.writingbench}{73.2}
\setval{8b.papo.eval.avg}{55.8}
\setval{8b.papo.eval.code}{60.8}
\setval{8b.papo.eval.if}{69.3}
\setval{8b.papo.eval.math}{42.8}
\setval{8b.papo.eval.writing}{63.3}
\setval{8b.papo.gate.code.against.early}{0.03}
\setval{8b.papo.gate.code.against.late}{0.01}
\setval{8b.papo.gate.code.against.run}{0.01}
\setval{8b.papo.gate.code.agreement}{+0.003}
\setval{8b.papo.gate.code.ceiling.early}{0.07}
\setval{8b.papo.gate.code.ceiling.late}{0.02}
\setval{8b.papo.gate.code.ceiling.run}{0.03}
\setval{8b.papo.gate.code.cosine.early}{0.86}
\setval{8b.papo.gate.code.cosine.late}{0.81}
\setval{8b.papo.gate.code.cosine.run}{0.84}
\setval{8b.papo.gate.if.against.early}{0.04}
\setval{8b.papo.gate.if.against.late}{0.03}
\setval{8b.papo.gate.if.against.run}{0.03}
\setval{8b.papo.gate.if.agreement}{+0.012}
\setval{8b.papo.gate.if.ceiling.early}{0.04}
\setval{8b.papo.gate.if.ceiling.late}{0.11}
\setval{8b.papo.gate.if.ceiling.run}{0.09}
\setval{8b.papo.gate.if.cosine.early}{0.78}
\setval{8b.papo.gate.if.cosine.late}{0.54}
\setval{8b.papo.gate.if.cosine.run}{0.63}
\setval{8b.papo.gate.math.against.early}{0.00}
\setval{8b.papo.gate.math.against.late}{0.00}
\setval{8b.papo.gate.math.against.run}{0.00}
\setval{8b.papo.gate.math.ceiling.early}{0.20}
\setval{8b.papo.gate.math.ceiling.late}{0.03}
\setval{8b.papo.gate.math.ceiling.run}{0.08}
\setval{8b.papo.gate.math.cosine.early}{0.87}
\setval{8b.papo.gate.math.cosine.late}{0.98}
\setval{8b.papo.gate.math.cosine.run}{0.95}
\setval{8b.papo.gate.writing-rm.against.early}{0.19}
\setval{8b.papo.gate.writing-rm.against.late}{0.50}
\setval{8b.papo.gate.writing-rm.against.run}{0.46}
\setval{8b.papo.gate.writing-rm.agreement}{+0.047}
\setval{8b.papo.gate.writing-rm.ceiling.early}{0.00}
\setval{8b.papo.gate.writing-rm.ceiling.late}{0.00}
\setval{8b.papo.gate.writing-rm.ceiling.run}{0.00}
\setval{8b.papo.gate.writing-rm.cosine.early}{0.71}
\setval{8b.papo.gate.writing-rm.cosine.late}{0.14}
\setval{8b.papo.gate.writing-rm.cosine.run}{0.23}
\setval{8b.papo.gate.writing.against.early}{0.23}
\setval{8b.papo.gate.writing.against.late}{0.49}
\setval{8b.papo.gate.writing.against.run}{0.49}
\setval{8b.papo.gate.writing.agreement}{$-$0.029}
\setval{8b.papo.gate.writing.ceiling.early}{0.00}
\setval{8b.papo.gate.writing.ceiling.late}{0.00}
\setval{8b.papo.gate.writing.ceiling.run}{0.00}
\setval{8b.papo.gate.writing.cosine.early}{0.57}
\setval{8b.papo.gate.writing.cosine.late}{0.10}
\setval{8b.papo.gate.writing.cosine.run}{0.14}
\setval{8b.papo.judge2.avg}{53.7}
\setval{8b.papo.judge2.creative-writing-v3}{41.9}
\setval{8b.papo.judge2.longbench-write}{59.5}
\setval{8b.papo.judge2.writing}{52.6}
\setval{8b.papo.judge2.writingbench}{56.3}
\setval{8b.papo.last-step}{300}
\setval{8b.papo.minus.base.aime24}{+0.8}
\setval{8b.papo.minus.base.aime24-abs}{0.8}
\setval{8b.papo.minus.base.aime25}{+1.6}
\setval{8b.papo.minus.base.aime25-abs}{1.6}
\setval{8b.papo.minus.base.amc23}{+11.8}
\setval{8b.papo.minus.base.amc23-abs}{11.8}
\setval{8b.papo.minus.base.avg}{+14.4}
\setval{8b.papo.minus.base.avg-abs}{14.4}
\setval{8b.papo.minus.base.code}{+4.8}
\setval{8b.papo.minus.base.code-abs}{4.8}
\setval{8b.papo.minus.base.creative-writing-v3}{+24.4}
\setval{8b.papo.minus.base.creative-writing-v3-abs}{24.4}
\setval{8b.papo.minus.base.guidebench}{+38.8}
\setval{8b.papo.minus.base.guidebench-abs}{38.8}
\setval{8b.papo.minus.base.humaneval-plus}{+2.4}
\setval{8b.papo.minus.base.humaneval-plus-abs}{2.4}
\setval{8b.papo.minus.base.if}{+30.7}
\setval{8b.papo.minus.base.if-abs}{30.7}
\setval{8b.papo.minus.base.ifbench}{+20.0}
\setval{8b.papo.minus.base.ifbench-abs}{20.0}
\setval{8b.papo.minus.base.ifeval}{+33.3}
\setval{8b.papo.minus.base.ifeval-abs}{33.3}
\setval{8b.papo.minus.base.judge2.avg}{+12.6}
\setval{8b.papo.minus.base.judge2.avg-abs}{12.6}
\setval{8b.papo.minus.base.judge2.creative-writing-v3}{+15.4}
\setval{8b.papo.minus.base.judge2.creative-writing-v3-abs}{15.4}
\setval{8b.papo.minus.base.judge2.longbench-write}{+11.5}
\setval{8b.papo.minus.base.judge2.longbench-write-abs}{11.5}
\setval{8b.papo.minus.base.judge2.writing}{+16.4}
\setval{8b.papo.minus.base.judge2.writing-abs}{16.4}
\setval{8b.papo.minus.base.judge2.writingbench}{+22.3}
\setval{8b.papo.minus.base.judge2.writingbench-abs}{22.3}
\setval{8b.papo.minus.base.livecodebench}{+5.7}
\setval{8b.papo.minus.base.livecodebench-abs}{5.7}
\setval{8b.papo.minus.base.longbench-write}{+14.1}
\setval{8b.papo.minus.base.longbench-write-abs}{14.1}
\setval{8b.papo.minus.base.math}{+5.7}
\setval{8b.papo.minus.base.math-abs}{5.7}
\setval{8b.papo.minus.base.math500}{+6.0}
\setval{8b.papo.minus.base.math500-abs}{6.0}
\setval{8b.papo.minus.base.mbpp-plus}{+6.1}
\setval{8b.papo.minus.base.mbpp-plus-abs}{6.1}
\setval{8b.papo.minus.base.minerva}{+8.4}
\setval{8b.papo.minus.base.minerva-abs}{8.4}
\setval{8b.papo.minus.base.olympiad-bench}{+5.4}
\setval{8b.papo.minus.base.olympiad-bench-abs}{5.4}
\setval{8b.papo.minus.base.writing}{+25.0}
\setval{8b.papo.minus.base.writing-abs}{25.0}
\setval{8b.papo.minus.base.writingbench}{+36.5}
\setval{8b.papo.minus.base.writingbench-abs}{36.5}
\setval{8b.papo.minus.concur-free.aime24}{$-$7.5}
\setval{8b.papo.minus.concur-free.aime24-abs}{7.5}
\setval{8b.papo.minus.concur-free.aime25}{$-$6.5}
\setval{8b.papo.minus.concur-free.aime25-abs}{6.5}
\setval{8b.papo.minus.concur-free.amc23}{$-$2.7}
\setval{8b.papo.minus.concur-free.amc23-abs}{2.7}
\setval{8b.papo.minus.concur-free.avg}{$-$4.4}
\setval{8b.papo.minus.concur-free.avg-abs}{4.4}
\setval{8b.papo.minus.concur-free.code}{$-$1.7}
\setval{8b.papo.minus.concur-free.code-abs}{1.7}
\setval{8b.papo.minus.concur-free.creative-writing-v3}{$-$0.5}
\setval{8b.papo.minus.concur-free.creative-writing-v3-abs}{0.5}
\setval{8b.papo.minus.concur-free.guidebench}{+1.4}
\setval{8b.papo.minus.concur-free.guidebench-abs}{1.4}
\setval{8b.papo.minus.concur-free.humaneval-plus}{$-$2.8}
\setval{8b.papo.minus.concur-free.humaneval-plus-abs}{2.8}
\setval{8b.papo.minus.concur-free.if}{$-$0.9}
\setval{8b.papo.minus.concur-free.if-abs}{0.9}
\setval{8b.papo.minus.concur-free.ifbench}{$-$1.3}
\setval{8b.papo.minus.concur-free.ifbench-abs}{1.3}
\setval{8b.papo.minus.concur-free.ifeval}{$-$2.9}
\setval{8b.papo.minus.concur-free.ifeval-abs}{2.9}
\setval{8b.papo.minus.concur-free.judge2.avg}{$-$3.6}
\setval{8b.papo.minus.concur-free.judge2.avg-abs}{3.6}
\setval{8b.papo.minus.concur-free.judge2.creative-writing-v3}{$-$0.6}
\setval{8b.papo.minus.concur-free.judge2.creative-writing-v3-abs}{0.6}
\setval{8b.papo.minus.concur-free.judge2.longbench-write}{$-$14.6}
\setval{8b.papo.minus.concur-free.judge2.longbench-write-abs}{14.6}
\setval{8b.papo.minus.concur-free.judge2.writing}{$-$6.5}
\setval{8b.papo.minus.concur-free.judge2.writing-abs}{6.5}
\setval{8b.papo.minus.concur-free.judge2.writingbench}{$-$4.4}
\setval{8b.papo.minus.concur-free.judge2.writingbench-abs}{4.4}
\setval{8b.papo.minus.concur-free.livecodebench}{$-$0.5}
\setval{8b.papo.minus.concur-free.livecodebench-abs}{0.5}
\setval{8b.papo.minus.concur-free.longbench-write}{$-$17.5}
\setval{8b.papo.minus.concur-free.longbench-write-abs}{17.5}
\setval{8b.papo.minus.concur-free.math}{$-$4.5}
\setval{8b.papo.minus.concur-free.math-abs}{4.5}
\setval{8b.papo.minus.concur-free.math500}{$-$5.0}
\setval{8b.papo.minus.concur-free.math500-abs}{5.0}
\setval{8b.papo.minus.concur-free.mbpp-plus}{$-$1.7}
\setval{8b.papo.minus.concur-free.mbpp-plus-abs}{1.7}
\setval{8b.papo.minus.concur-free.minerva}{$-$0.4}
\setval{8b.papo.minus.concur-free.minerva-abs}{0.4}
\setval{8b.papo.minus.concur-free.olympiad-bench}{$-$5.2}
\setval{8b.papo.minus.concur-free.olympiad-bench-abs}{5.2}
\setval{8b.papo.minus.concur-free.writing}{$-$10.5}
\setval{8b.papo.minus.concur-free.writing-abs}{10.5}
\setval{8b.papo.minus.concur-free.writingbench}{$-$13.6}
\setval{8b.papo.minus.concur-free.writingbench-abs}{13.6}
\setval{8b.papo.minus.concur-tied.aime24}{$-$19.1}
\setval{8b.papo.minus.concur-tied.aime24-abs}{19.1}
\setval{8b.papo.minus.concur-tied.aime25}{$-$13.9}
\setval{8b.papo.minus.concur-tied.aime25-abs}{13.9}
\setval{8b.papo.minus.concur-tied.amc23}{$-$16.5}
\setval{8b.papo.minus.concur-tied.amc23-abs}{16.5}
\setval{8b.papo.minus.concur-tied.avg}{$-$8.2}
\setval{8b.papo.minus.concur-tied.avg-abs}{8.2}
\setval{8b.papo.minus.concur-tied.code}{$-$2.3}
\setval{8b.papo.minus.concur-tied.code-abs}{2.3}
\setval{8b.papo.minus.concur-tied.creative-writing-v3}{$-$1.3}
\setval{8b.papo.minus.concur-tied.creative-writing-v3-abs}{1.3}
\setval{8b.papo.minus.concur-tied.guidebench}{+1.0}
\setval{8b.papo.minus.concur-tied.guidebench-abs}{1.0}
\setval{8b.papo.minus.concur-tied.humaneval-plus}{$-$4.9}
\setval{8b.papo.minus.concur-tied.humaneval-plus-abs}{4.9}
\setval{8b.papo.minus.concur-tied.if}{$-$1.2}
\setval{8b.papo.minus.concur-tied.if-abs}{1.2}
\setval{8b.papo.minus.concur-tied.ifbench}{$-$2.0}
\setval{8b.papo.minus.concur-tied.ifbench-abs}{2.0}
\setval{8b.papo.minus.concur-tied.ifeval}{$-$2.5}
\setval{8b.papo.minus.concur-tied.ifeval-abs}{2.5}
\setval{8b.papo.minus.concur-tied.judge2.avg}{$-$7.6}
\setval{8b.papo.minus.concur-tied.judge2.avg-abs}{7.6}
\setval{8b.papo.minus.concur-tied.judge2.creative-writing-v3}{+0.1}
\setval{8b.papo.minus.concur-tied.judge2.creative-writing-v3-abs}{0.1}
\setval{8b.papo.minus.concur-tied.judge2.longbench-write}{$-$15.2}
\setval{8b.papo.minus.concur-tied.judge2.longbench-write-abs}{15.2}
\setval{8b.papo.minus.concur-tied.judge2.writing}{$-$7.5}
\setval{8b.papo.minus.concur-tied.judge2.writing-abs}{7.5}
\setval{8b.papo.minus.concur-tied.judge2.writingbench}{$-$7.5}
\setval{8b.papo.minus.concur-tied.judge2.writingbench-abs}{7.5}
\setval{8b.papo.minus.concur-tied.livecodebench}{$-$0.6}
\setval{8b.papo.minus.concur-tied.livecodebench-abs}{0.6}
\setval{8b.papo.minus.concur-tied.longbench-write}{$-$15.7}
\setval{8b.papo.minus.concur-tied.longbench-write-abs}{15.7}
\setval{8b.papo.minus.concur-tied.math}{$-$13.6}
\setval{8b.papo.minus.concur-tied.math-abs}{13.6}
\setval{8b.papo.minus.concur-tied.math500}{$-$11.1}
\setval{8b.papo.minus.concur-tied.math500-abs}{11.1}
\setval{8b.papo.minus.concur-tied.mbpp-plus}{$-$1.5}
\setval{8b.papo.minus.concur-tied.mbpp-plus-abs}{1.5}
\setval{8b.papo.minus.concur-tied.minerva}{$-$6.1}
\setval{8b.papo.minus.concur-tied.minerva-abs}{6.1}
\setval{8b.papo.minus.concur-tied.olympiad-bench}{$-$14.7}
\setval{8b.papo.minus.concur-tied.olympiad-bench-abs}{14.7}
\setval{8b.papo.minus.concur-tied.writing}{$-$10.3}
\setval{8b.papo.minus.concur-tied.writing-abs}{10.3}
\setval{8b.papo.minus.concur-tied.writingbench}{$-$14.0}
\setval{8b.papo.minus.concur-tied.writingbench-abs}{14.0}
\setval{8b.papo.minus.concur.aime24}{$-$16.6}
\setval{8b.papo.minus.concur.aime24-abs}{16.6}
\setval{8b.papo.minus.concur.aime25}{$-$13.0}
\setval{8b.papo.minus.concur.aime25-abs}{13.0}
\setval{8b.papo.minus.concur.amc23}{$-$16.9}
\setval{8b.papo.minus.concur.amc23-abs}{16.9}
\setval{8b.papo.minus.concur.avg}{$-$7.3}
\setval{8b.papo.minus.concur.avg-abs}{7.3}
\setval{8b.papo.minus.concur.code}{+0.1}
\setval{8b.papo.minus.concur.code-abs}{0.1}
\setval{8b.papo.minus.concur.creative-writing-v3}{+1.2}
\setval{8b.papo.minus.concur.creative-writing-v3-abs}{1.2}
\setval{8b.papo.minus.concur.guidebench}{+1.0}
\setval{8b.papo.minus.concur.guidebench-abs}{1.0}
\setval{8b.papo.minus.concur.humaneval-plus}{$-$0.2}
\setval{8b.papo.minus.concur.humaneval-plus-abs}{0.2}
\setval{8b.papo.minus.concur.if}{$-$0.7}
\setval{8b.papo.minus.concur.if-abs}{0.7}
\setval{8b.papo.minus.concur.ifbench}{$-$1.3}
\setval{8b.papo.minus.concur.ifbench-abs}{1.3}
\setval{8b.papo.minus.concur.ifeval}{$-$1.8}
\setval{8b.papo.minus.concur.ifeval-abs}{1.8}
\setval{8b.papo.minus.concur.judge2.avg}{$-$6.7}
\setval{8b.papo.minus.concur.judge2.avg-abs}{6.7}
\setval{8b.papo.minus.concur.judge2.creative-writing-v3}{+4.1}
\setval{8b.papo.minus.concur.judge2.creative-writing-v3-abs}{4.1}
\setval{8b.papo.minus.concur.judge2.longbench-write}{$-$15.9}
\setval{8b.papo.minus.concur.judge2.longbench-write-abs}{15.9}
\setval{8b.papo.minus.concur.judge2.writing}{$-$6.5}
\setval{8b.papo.minus.concur.judge2.writing-abs}{6.5}
\setval{8b.papo.minus.concur.judge2.writingbench}{$-$7.6}
\setval{8b.papo.minus.concur.judge2.writingbench-abs}{7.6}
\setval{8b.papo.minus.concur.livecodebench}{+0.3}
\setval{8b.papo.minus.concur.livecodebench-abs}{0.3}
\setval{8b.papo.minus.concur.longbench-write}{$-$16.1}
\setval{8b.papo.minus.concur.longbench-write-abs}{16.1}
\setval{8b.papo.minus.concur.math}{$-$13.1}
\setval{8b.papo.minus.concur.math-abs}{13.1}
\setval{8b.papo.minus.concur.math500}{$-$11.7}
\setval{8b.papo.minus.concur.math500-abs}{11.7}
\setval{8b.papo.minus.concur.mbpp-plus}{+0.2}
\setval{8b.papo.minus.concur.mbpp-plus-abs}{0.2}
\setval{8b.papo.minus.concur.minerva}{$-$4.9}
\setval{8b.papo.minus.concur.minerva-abs}{4.9}
\setval{8b.papo.minus.concur.olympiad-bench}{$-$15.6}
\setval{8b.papo.minus.concur.olympiad-bench-abs}{15.6}
\setval{8b.papo.minus.concur.writing}{$-$9.9}
\setval{8b.papo.minus.concur.writing-abs}{9.9}
\setval{8b.papo.minus.concur.writingbench}{$-$14.9}
\setval{8b.papo.minus.concur.writingbench-abs}{14.9}
\setval{8b.papo.minus.gd2po.aime24}{$-$4.1}
\setval{8b.papo.minus.gd2po.aime24-abs}{4.1}
\setval{8b.papo.minus.gd2po.aime25}{$-$2.1}
\setval{8b.papo.minus.gd2po.aime25-abs}{2.1}
\setval{8b.papo.minus.gd2po.amc23}{+2.4}
\setval{8b.papo.minus.gd2po.amc23-abs}{2.4}
\setval{8b.papo.minus.gd2po.avg}{$-$1.6}
\setval{8b.papo.minus.gd2po.avg-abs}{1.6}
\setval{8b.papo.minus.gd2po.code}{$-$0.9}
\setval{8b.papo.minus.gd2po.code-abs}{0.9}
\setval{8b.papo.minus.gd2po.creative-writing-v3}{+3.0}
\setval{8b.papo.minus.gd2po.creative-writing-v3-abs}{3.0}
\setval{8b.papo.minus.gd2po.guidebench}{+0.7}
\setval{8b.papo.minus.gd2po.guidebench-abs}{0.7}
\setval{8b.papo.minus.gd2po.humaneval-plus}{$-$1.5}
\setval{8b.papo.minus.gd2po.humaneval-plus-abs}{1.5}
\setval{8b.papo.minus.gd2po.if}{$-$2.7}
\setval{8b.papo.minus.gd2po.if-abs}{2.7}
\setval{8b.papo.minus.gd2po.ifbench}{$-$5.7}
\setval{8b.papo.minus.gd2po.ifbench-abs}{5.7}
\setval{8b.papo.minus.gd2po.ifeval}{$-$3.0}
\setval{8b.papo.minus.gd2po.ifeval-abs}{3.0}
\setval{8b.papo.minus.gd2po.judge2.avg}{$-$1.2}
\setval{8b.papo.minus.gd2po.judge2.avg-abs}{1.2}
\setval{8b.papo.minus.gd2po.judge2.creative-writing-v3}{+5.1}
\setval{8b.papo.minus.gd2po.judge2.creative-writing-v3-abs}{5.1}
\setval{8b.papo.minus.gd2po.judge2.longbench-write}{$-$2.6}
\setval{8b.papo.minus.gd2po.judge2.longbench-write-abs}{2.6}
\setval{8b.papo.minus.gd2po.judge2.writing}{+1.2}
\setval{8b.papo.minus.gd2po.judge2.writing-abs}{1.2}
\setval{8b.papo.minus.gd2po.judge2.writingbench}{+1.1}
\setval{8b.papo.minus.gd2po.judge2.writingbench-abs}{1.1}
\setval{8b.papo.minus.gd2po.livecodebench}{+0.3}
\setval{8b.papo.minus.gd2po.livecodebench-abs}{0.3}
\setval{8b.papo.minus.gd2po.longbench-write}{$-$4.4}
\setval{8b.papo.minus.gd2po.longbench-write-abs}{4.4}
\setval{8b.papo.minus.gd2po.math}{$-$1.8}
\setval{8b.papo.minus.gd2po.math-abs}{1.8}
\setval{8b.papo.minus.gd2po.math500}{$-$2.7}
\setval{8b.papo.minus.gd2po.math500-abs}{2.7}
\setval{8b.papo.minus.gd2po.mbpp-plus}{$-$1.7}
\setval{8b.papo.minus.gd2po.mbpp-plus-abs}{1.7}
\setval{8b.papo.minus.gd2po.minerva}{$-$0.6}
\setval{8b.papo.minus.gd2po.minerva-abs}{0.6}
\setval{8b.papo.minus.gd2po.olympiad-bench}{$-$3.8}
\setval{8b.papo.minus.gd2po.olympiad-bench-abs}{3.8}
\setval{8b.papo.minus.gd2po.writing}{$-$0.7}
\setval{8b.papo.minus.gd2po.writing-abs}{0.7}
\setval{8b.papo.minus.gd2po.writingbench}{$-$0.7}
\setval{8b.papo.minus.gd2po.writingbench-abs}{0.7}
\setval{8b.papo.minus.gdpo-noaux.aime24}{$-$15.7}
\setval{8b.papo.minus.gdpo-noaux.aime24-abs}{15.7}
\setval{8b.papo.minus.gdpo-noaux.aime25}{$-$13.9}
\setval{8b.papo.minus.gdpo-noaux.aime25-abs}{13.9}
\setval{8b.papo.minus.gdpo-noaux.amc23}{$-$15.1}
\setval{8b.papo.minus.gdpo-noaux.amc23-abs}{15.1}
\setval{8b.papo.minus.gdpo-noaux.avg}{$-$4.7}
\setval{8b.papo.minus.gdpo-noaux.avg-abs}{4.7}
\setval{8b.papo.minus.gdpo-noaux.code}{$-$1.9}
\setval{8b.papo.minus.gdpo-noaux.code-abs}{1.9}
\setval{8b.papo.minus.gdpo-noaux.creative-writing-v3}{+7.3}
\setval{8b.papo.minus.gdpo-noaux.creative-writing-v3-abs}{7.3}
\setval{8b.papo.minus.gdpo-noaux.guidebench}{+3.0}
\setval{8b.papo.minus.gdpo-noaux.guidebench-abs}{3.0}
\setval{8b.papo.minus.gdpo-noaux.humaneval-plus}{$-$3.5}
\setval{8b.papo.minus.gdpo-noaux.humaneval-plus-abs}{3.5}
\setval{8b.papo.minus.gdpo-noaux.if}{+2.1}
\setval{8b.papo.minus.gdpo-noaux.if-abs}{2.1}
\setval{8b.papo.minus.gdpo-noaux.ifbench}{+3.7}
\setval{8b.papo.minus.gdpo-noaux.ifbench-abs}{3.7}
\setval{8b.papo.minus.gdpo-noaux.ifeval}{$-$0.2}
\setval{8b.papo.minus.gdpo-noaux.ifeval-abs}{0.2}
\setval{8b.papo.minus.gdpo-noaux.judge2.avg}{$-$3.4}
\setval{8b.papo.minus.gdpo-noaux.judge2.avg-abs}{3.4}
\setval{8b.papo.minus.gdpo-noaux.judge2.creative-writing-v3}{+13.0}
\setval{8b.papo.minus.gdpo-noaux.judge2.creative-writing-v3-abs}{13.0}
\setval{8b.papo.minus.gdpo-noaux.judge2.longbench-write}{+5.1}
\setval{8b.papo.minus.gdpo-noaux.judge2.longbench-write-abs}{5.1}
\setval{8b.papo.minus.gdpo-noaux.judge2.writing}{+8.1}
\setval{8b.papo.minus.gdpo-noaux.judge2.writing-abs}{8.1}
\setval{8b.papo.minus.gdpo-noaux.judge2.writingbench}{+6.3}
\setval{8b.papo.minus.gdpo-noaux.judge2.writingbench-abs}{6.3}
\setval{8b.papo.minus.gdpo-noaux.livecodebench}{$-$0.7}
\setval{8b.papo.minus.gdpo-noaux.livecodebench-abs}{0.7}
\setval{8b.papo.minus.gdpo-noaux.longbench-write}{$-$0.5}
\setval{8b.papo.minus.gdpo-noaux.longbench-write-abs}{0.5}
\setval{8b.papo.minus.gdpo-noaux.math}{$-$12.7}
\setval{8b.papo.minus.gdpo-noaux.math-abs}{12.7}
\setval{8b.papo.minus.gdpo-noaux.math500}{$-$11.8}
\setval{8b.papo.minus.gdpo-noaux.math500-abs}{11.8}
\setval{8b.papo.minus.gdpo-noaux.mbpp-plus}{$-$1.5}
\setval{8b.papo.minus.gdpo-noaux.mbpp-plus-abs}{1.5}
\setval{8b.papo.minus.gdpo-noaux.minerva}{$-$4.6}
\setval{8b.papo.minus.gdpo-noaux.minerva-abs}{4.6}
\setval{8b.papo.minus.gdpo-noaux.olympiad-bench}{$-$15.1}
\setval{8b.papo.minus.gdpo-noaux.olympiad-bench-abs}{15.1}
\setval{8b.papo.minus.gdpo-noaux.writing}{+1.7}
\setval{8b.papo.minus.gdpo-noaux.writing-abs}{1.7}
\setval{8b.papo.minus.gdpo-noaux.writingbench}{$-$1.8}
\setval{8b.papo.minus.gdpo-noaux.writingbench-abs}{1.8}
\setval{8b.papo.minus.gdpo-w0.5.aime24}{$-$7.1}
\setval{8b.papo.minus.gdpo-w0.5.aime24-abs}{7.1}
\setval{8b.papo.minus.gdpo-w0.5.aime25}{$-$5.3}
\setval{8b.papo.minus.gdpo-w0.5.aime25-abs}{5.3}
\setval{8b.papo.minus.gdpo-w0.5.amc23}{$-$1.4}
\setval{8b.papo.minus.gdpo-w0.5.amc23-abs}{1.4}
\setval{8b.papo.minus.gdpo-w0.5.avg}{$-$2.5}
\setval{8b.papo.minus.gdpo-w0.5.avg-abs}{2.5}
\setval{8b.papo.minus.gdpo-w0.5.code}{$-$0.3}
\setval{8b.papo.minus.gdpo-w0.5.code-abs}{0.3}
\setval{8b.papo.minus.gdpo-w0.5.creative-writing-v3}{$-$0.4}
\setval{8b.papo.minus.gdpo-w0.5.creative-writing-v3-abs}{0.4}
\setval{8b.papo.minus.gdpo-w0.5.guidebench}{+0.4}
\setval{8b.papo.minus.gdpo-w0.5.guidebench-abs}{0.4}
\setval{8b.papo.minus.gdpo-w0.5.humaneval-plus}{$-$1.3}
\setval{8b.papo.minus.gdpo-w0.5.humaneval-plus-abs}{1.3}
\setval{8b.papo.minus.gdpo-w0.5.if}{$-$0.5}
\setval{8b.papo.minus.gdpo-w0.5.if-abs}{0.5}
\setval{8b.papo.minus.gdpo-w0.5.ifbench}{$-$0.3}
\setval{8b.papo.minus.gdpo-w0.5.ifbench-abs}{0.3}
\setval{8b.papo.minus.gdpo-w0.5.ifeval}{$-$1.4}
\setval{8b.papo.minus.gdpo-w0.5.ifeval-abs}{1.4}
\setval{8b.papo.minus.gdpo-w0.5.judge2.avg}{$-$2.2}
\setval{8b.papo.minus.gdpo-w0.5.judge2.avg-abs}{2.2}
\setval{8b.papo.minus.gdpo-w0.5.judge2.creative-writing-v3}{+1.5}
\setval{8b.papo.minus.gdpo-w0.5.judge2.creative-writing-v3-abs}{1.5}
\setval{8b.papo.minus.gdpo-w0.5.judge2.longbench-write}{$-$4.1}
\setval{8b.papo.minus.gdpo-w0.5.judge2.longbench-write-abs}{4.1}
\setval{8b.papo.minus.gdpo-w0.5.judge2.writing}{$-$0.8}
\setval{8b.papo.minus.gdpo-w0.5.judge2.writing-abs}{0.8}
\setval{8b.papo.minus.gdpo-w0.5.judge2.writingbench}{+0.1}
\setval{8b.papo.minus.gdpo-w0.5.judge2.writingbench-abs}{0.1}
\setval{8b.papo.minus.gdpo-w0.5.livecodebench}{$-$0.2}
\setval{8b.papo.minus.gdpo-w0.5.livecodebench-abs}{0.2}
\setval{8b.papo.minus.gdpo-w0.5.longbench-write}{$-$5.7}
\setval{8b.papo.minus.gdpo-w0.5.longbench-write-abs}{5.7}
\setval{8b.papo.minus.gdpo-w0.5.math}{$-$4.6}
\setval{8b.papo.minus.gdpo-w0.5.math-abs}{4.6}
\setval{8b.papo.minus.gdpo-w0.5.math500}{$-$5.6}
\setval{8b.papo.minus.gdpo-w0.5.math500-abs}{5.6}
\setval{8b.papo.minus.gdpo-w0.5.mbpp-plus}{+0.6}
\setval{8b.papo.minus.gdpo-w0.5.mbpp-plus-abs}{0.6}
\setval{8b.papo.minus.gdpo-w0.5.minerva}{$-$1.9}
\setval{8b.papo.minus.gdpo-w0.5.minerva-abs}{1.9}
\setval{8b.papo.minus.gdpo-w0.5.olympiad-bench}{$-$6.3}
\setval{8b.papo.minus.gdpo-w0.5.olympiad-bench-abs}{6.3}
\setval{8b.papo.minus.gdpo-w0.5.writing}{$-$2.6}
\setval{8b.papo.minus.gdpo-w0.5.writing-abs}{2.6}
\setval{8b.papo.minus.gdpo-w0.5.writingbench}{$-$1.8}
\setval{8b.papo.minus.gdpo-w0.5.writingbench-abs}{1.8}
\setval{8b.papo.minus.gdpo.aime24}{$-$9.2}
\setval{8b.papo.minus.gdpo.aime24-abs}{9.2}
\setval{8b.papo.minus.gdpo.aime25}{$-$4.3}
\setval{8b.papo.minus.gdpo.aime25-abs}{4.3}
\setval{8b.papo.minus.gdpo.amc23}{$-$0.2}
\setval{8b.papo.minus.gdpo.amc23-abs}{0.2}
\setval{8b.papo.minus.gdpo.avg}{$-$2.5}
\setval{8b.papo.minus.gdpo.avg-abs}{2.5}
\setval{8b.papo.minus.gdpo.code}{$-$0.8}
\setval{8b.papo.minus.gdpo.code-abs}{0.8}
\setval{8b.papo.minus.gdpo.creative-writing-v3}{$-$1.8}
\setval{8b.papo.minus.gdpo.creative-writing-v3-abs}{1.8}
\setval{8b.papo.minus.gdpo.guidebench}{$-$0.2}
\setval{8b.papo.minus.gdpo.guidebench-abs}{0.2}
\setval{8b.papo.minus.gdpo.humaneval-plus}{$-$1.6}
\setval{8b.papo.minus.gdpo.humaneval-plus-abs}{1.6}
\setval{8b.papo.minus.gdpo.if}{$-$1.3}
\setval{8b.papo.minus.gdpo.if-abs}{1.3}
\setval{8b.papo.minus.gdpo.ifbench}{$-$2.0}
\setval{8b.papo.minus.gdpo.ifbench-abs}{2.0}
\setval{8b.papo.minus.gdpo.ifeval}{$-$1.8}
\setval{8b.papo.minus.gdpo.ifeval-abs}{1.8}
\setval{8b.papo.minus.gdpo.judge2.avg}{$-$2.3}
\setval{8b.papo.minus.gdpo.judge2.avg-abs}{2.3}
\setval{8b.papo.minus.gdpo.judge2.creative-writing-v3}{+0.8}
\setval{8b.papo.minus.gdpo.judge2.creative-writing-v3-abs}{0.8}
\setval{8b.papo.minus.gdpo.judge2.longbench-write}{$-$4.6}
\setval{8b.papo.minus.gdpo.judge2.longbench-write-abs}{4.6}
\setval{8b.papo.minus.gdpo.judge2.writing}{$-$2.0}
\setval{8b.papo.minus.gdpo.judge2.writing-abs}{2.0}
\setval{8b.papo.minus.gdpo.judge2.writingbench}{$-$2.1}
\setval{8b.papo.minus.gdpo.judge2.writingbench-abs}{2.1}
\setval{8b.papo.minus.gdpo.livecodebench}{+0.4}
\setval{8b.papo.minus.gdpo.livecodebench-abs}{0.4}
\setval{8b.papo.minus.gdpo.longbench-write}{$-$4.1}
\setval{8b.papo.minus.gdpo.longbench-write-abs}{4.1}
\setval{8b.papo.minus.gdpo.math}{$-$3.8}
\setval{8b.papo.minus.gdpo.math-abs}{3.8}
\setval{8b.papo.minus.gdpo.math500}{$-$4.5}
\setval{8b.papo.minus.gdpo.math500-abs}{4.5}
\setval{8b.papo.minus.gdpo.mbpp-plus}{$-$1.1}
\setval{8b.papo.minus.gdpo.mbpp-plus-abs}{1.1}
\setval{8b.papo.minus.gdpo.minerva}{$-$0.3}
\setval{8b.papo.minus.gdpo.minerva-abs}{0.3}
\setval{8b.papo.minus.gdpo.olympiad-bench}{$-$4.5}
\setval{8b.papo.minus.gdpo.olympiad-bench-abs}{4.5}
\setval{8b.papo.minus.gdpo.writing}{$-$2.9}
\setval{8b.papo.minus.gdpo.writing-abs}{2.9}
\setval{8b.papo.minus.gdpo.writingbench}{$-$2.9}
\setval{8b.papo.minus.gdpo.writingbench-abs}{2.9}
\setval{8b.papo.minus.geoma.aime24}{$-$1.9}
\setval{8b.papo.minus.geoma.aime24-abs}{1.9}
\setval{8b.papo.minus.geoma.aime25}{$-$2.4}
\setval{8b.papo.minus.geoma.aime25-abs}{2.4}
\setval{8b.papo.minus.geoma.amc23}{$-$1.5}
\setval{8b.papo.minus.geoma.amc23-abs}{1.5}
\setval{8b.papo.minus.geoma.avg}{$-$0.9}
\setval{8b.papo.minus.geoma.avg-abs}{0.9}
\setval{8b.papo.minus.geoma.code}{$-$0.5}
\setval{8b.papo.minus.geoma.code-abs}{0.5}
\setval{8b.papo.minus.geoma.creative-writing-v3}{+0.8}
\setval{8b.papo.minus.geoma.creative-writing-v3-abs}{0.8}
\setval{8b.papo.minus.geoma.guidebench}{+0.0}
\setval{8b.papo.minus.geoma.guidebench-abs}{0.0}
\setval{8b.papo.minus.geoma.humaneval-plus}{$-$0.2}
\setval{8b.papo.minus.geoma.humaneval-plus-abs}{0.2}
\setval{8b.papo.minus.geoma.if}{+0.8}
\setval{8b.papo.minus.geoma.if-abs}{0.8}
\setval{8b.papo.minus.geoma.ifbench}{+3.3}
\setval{8b.papo.minus.geoma.ifbench-abs}{3.3}
\setval{8b.papo.minus.geoma.ifeval}{$-$1.0}
\setval{8b.papo.minus.geoma.ifeval-abs}{1.0}
\setval{8b.papo.minus.geoma.judge2.avg}{$-$0.7}
\setval{8b.papo.minus.geoma.judge2.avg-abs}{0.7}
\setval{8b.papo.minus.geoma.judge2.creative-writing-v3}{+4.3}
\setval{8b.papo.minus.geoma.judge2.creative-writing-v3-abs}{4.3}
\setval{8b.papo.minus.geoma.judge2.longbench-write}{$-$5.3}
\setval{8b.papo.minus.geoma.judge2.longbench-write-abs}{5.3}
\setval{8b.papo.minus.geoma.judge2.writing}{$-$1.0}
\setval{8b.papo.minus.geoma.judge2.writing-abs}{1.0}
\setval{8b.papo.minus.geoma.judge2.writingbench}{$-$2.0}
\setval{8b.papo.minus.geoma.judge2.writingbench-abs}{2.0}
\setval{8b.papo.minus.geoma.livecodebench}{+0.9}
\setval{8b.papo.minus.geoma.livecodebench-abs}{0.9}
\setval{8b.papo.minus.geoma.longbench-write}{$-$3.5}
\setval{8b.papo.minus.geoma.longbench-write-abs}{3.5}
\setval{8b.papo.minus.geoma.math}{$-$1.4}
\setval{8b.papo.minus.geoma.math-abs}{1.4}
\setval{8b.papo.minus.geoma.math500}{$-$1.2}
\setval{8b.papo.minus.geoma.math500-abs}{1.2}
\setval{8b.papo.minus.geoma.mbpp-plus}{$-$2.2}
\setval{8b.papo.minus.geoma.mbpp-plus-abs}{2.2}
\setval{8b.papo.minus.geoma.minerva}{$-$0.4}
\setval{8b.papo.minus.geoma.minerva-abs}{0.4}
\setval{8b.papo.minus.geoma.olympiad-bench}{$-$1.2}
\setval{8b.papo.minus.geoma.olympiad-bench-abs}{1.2}
\setval{8b.papo.minus.geoma.writing}{$-$1.8}
\setval{8b.papo.minus.geoma.writing-abs}{1.8}
\setval{8b.papo.minus.geoma.writingbench}{$-$2.7}
\setval{8b.papo.minus.geoma.writingbench-abs}{2.7}
\setval{8b.papo.minus.grpo-nodiv.aime24}{$-$8.4}
\setval{8b.papo.minus.grpo-nodiv.aime24-abs}{8.4}
\setval{8b.papo.minus.grpo-nodiv.aime25}{$-$4.7}
\setval{8b.papo.minus.grpo-nodiv.aime25-abs}{4.7}
\setval{8b.papo.minus.grpo-nodiv.amc23}{$-$1.6}
\setval{8b.papo.minus.grpo-nodiv.amc23-abs}{1.6}
\setval{8b.papo.minus.grpo-nodiv.avg}{$-$2.3}
\setval{8b.papo.minus.grpo-nodiv.avg-abs}{2.3}
\setval{8b.papo.minus.grpo-nodiv.code}{$-$1.6}
\setval{8b.papo.minus.grpo-nodiv.code-abs}{1.6}
\setval{8b.papo.minus.grpo-nodiv.creative-writing-v3}{+4.5}
\setval{8b.papo.minus.grpo-nodiv.creative-writing-v3-abs}{4.5}
\setval{8b.papo.minus.grpo-nodiv.guidebench}{$-$0.5}
\setval{8b.papo.minus.grpo-nodiv.guidebench-abs}{0.5}
\setval{8b.papo.minus.grpo-nodiv.humaneval-plus}{$-$3.8}
\setval{8b.papo.minus.grpo-nodiv.humaneval-plus-abs}{3.8}
\setval{8b.papo.minus.grpo-nodiv.if}{$-$0.9}
\setval{8b.papo.minus.grpo-nodiv.if-abs}{0.9}
\setval{8b.papo.minus.grpo-nodiv.ifbench}{+1.0}
\setval{8b.papo.minus.grpo-nodiv.ifbench-abs}{1.0}
\setval{8b.papo.minus.grpo-nodiv.ifeval}{$-$3.1}
\setval{8b.papo.minus.grpo-nodiv.ifeval-abs}{3.1}
\setval{8b.papo.minus.grpo-nodiv.judge2.avg}{$-$2.4}
\setval{8b.papo.minus.grpo-nodiv.judge2.avg-abs}{2.4}
\setval{8b.papo.minus.grpo-nodiv.judge2.creative-writing-v3}{+4.4}
\setval{8b.papo.minus.grpo-nodiv.judge2.creative-writing-v3-abs}{4.4}
\setval{8b.papo.minus.grpo-nodiv.judge2.longbench-write}{$-$7.0}
\setval{8b.papo.minus.grpo-nodiv.judge2.longbench-write-abs}{7.0}
\setval{8b.papo.minus.grpo-nodiv.judge2.writing}{$-$0.1}
\setval{8b.papo.minus.grpo-nodiv.judge2.writing-abs}{0.1}
\setval{8b.papo.minus.grpo-nodiv.judge2.writingbench}{+2.4}
\setval{8b.papo.minus.grpo-nodiv.judge2.writingbench-abs}{2.4}
\setval{8b.papo.minus.grpo-nodiv.livecodebench}{+1.7}
\setval{8b.papo.minus.grpo-nodiv.livecodebench-abs}{1.7}
\setval{8b.papo.minus.grpo-nodiv.longbench-write}{$-$5.6}
\setval{8b.papo.minus.grpo-nodiv.longbench-write-abs}{5.6}
\setval{8b.papo.minus.grpo-nodiv.math}{$-$4.7}
\setval{8b.papo.minus.grpo-nodiv.math-abs}{4.7}
\setval{8b.papo.minus.grpo-nodiv.math500}{$-$5.7}
\setval{8b.papo.minus.grpo-nodiv.math500-abs}{5.7}
\setval{8b.papo.minus.grpo-nodiv.mbpp-plus}{$-$2.6}
\setval{8b.papo.minus.grpo-nodiv.mbpp-plus-abs}{2.6}
\setval{8b.papo.minus.grpo-nodiv.minerva}{$-$1.5}
\setval{8b.papo.minus.grpo-nodiv.minerva-abs}{1.5}
\setval{8b.papo.minus.grpo-nodiv.olympiad-bench}{$-$6.7}
\setval{8b.papo.minus.grpo-nodiv.olympiad-bench-abs}{6.7}
\setval{8b.papo.minus.grpo-nodiv.writing}{+0.6}
\setval{8b.papo.minus.grpo-nodiv.writing-abs}{0.6}
\setval{8b.papo.minus.grpo-nodiv.writingbench}{+2.9}
\setval{8b.papo.minus.grpo-nodiv.writingbench-abs}{2.9}
\setval{8b.papo.minus.grpo.aime24}{$-$3.6}
\setval{8b.papo.minus.grpo.aime24-abs}{3.6}
\setval{8b.papo.minus.grpo.aime25}{$-$5.4}
\setval{8b.papo.minus.grpo.aime25-abs}{5.4}
\setval{8b.papo.minus.grpo.amc23}{$-$2.3}
\setval{8b.papo.minus.grpo.amc23-abs}{2.3}
\setval{8b.papo.minus.grpo.avg}{$-$2.1}
\setval{8b.papo.minus.grpo.avg-abs}{2.1}
\setval{8b.papo.minus.grpo.code}{$-$0.8}
\setval{8b.papo.minus.grpo.code-abs}{0.8}
\setval{8b.papo.minus.grpo.creative-writing-v3}{$-$1.3}
\setval{8b.papo.minus.grpo.creative-writing-v3-abs}{1.3}
\setval{8b.papo.minus.grpo.guidebench}{+2.5}
\setval{8b.papo.minus.grpo.guidebench-abs}{2.5}
\setval{8b.papo.minus.grpo.humaneval-plus}{$-$1.2}
\setval{8b.papo.minus.grpo.humaneval-plus-abs}{1.2}
\setval{8b.papo.minus.grpo.if}{$-$0.9}
\setval{8b.papo.minus.grpo.if-abs}{0.9}
\setval{8b.papo.minus.grpo.ifbench}{$-$4.7}
\setval{8b.papo.minus.grpo.ifbench-abs}{4.7}
\setval{8b.papo.minus.grpo.ifeval}{$-$0.5}
\setval{8b.papo.minus.grpo.ifeval-abs}{0.5}
\setval{8b.papo.minus.grpo.judge2.avg}{$-$1.5}
\setval{8b.papo.minus.grpo.judge2.avg-abs}{1.5}
\setval{8b.papo.minus.grpo.judge2.creative-writing-v3}{+2.6}
\setval{8b.papo.minus.grpo.judge2.creative-writing-v3-abs}{2.6}
\setval{8b.papo.minus.grpo.judge2.longbench-write}{$-$2.9}
\setval{8b.papo.minus.grpo.judge2.longbench-write-abs}{2.9}
\setval{8b.papo.minus.grpo.judge2.writing}{$-$0.3}
\setval{8b.papo.minus.grpo.judge2.writing-abs}{0.3}
\setval{8b.papo.minus.grpo.judge2.writingbench}{$-$0.6}
\setval{8b.papo.minus.grpo.judge2.writingbench-abs}{0.6}
\setval{8b.papo.minus.grpo.livecodebench}{+0.1}
\setval{8b.papo.minus.grpo.livecodebench-abs}{0.1}
\setval{8b.papo.minus.grpo.longbench-write}{$-$4.3}
\setval{8b.papo.minus.grpo.longbench-write-abs}{4.3}
\setval{8b.papo.minus.grpo.math}{$-$2.9}
\setval{8b.papo.minus.grpo.math-abs}{2.9}
\setval{8b.papo.minus.grpo.math500}{$-$2.2}
\setval{8b.papo.minus.grpo.math500-abs}{2.2}
\setval{8b.papo.minus.grpo.mbpp-plus}{$-$1.2}
\setval{8b.papo.minus.grpo.mbpp-plus-abs}{1.2}
\setval{8b.papo.minus.grpo.minerva}{$-$0.5}
\setval{8b.papo.minus.grpo.minerva-abs}{0.5}
\setval{8b.papo.minus.grpo.olympiad-bench}{$-$3.1}
\setval{8b.papo.minus.grpo.olympiad-bench-abs}{3.1}
\setval{8b.papo.minus.grpo.writing}{$-$3.0}
\setval{8b.papo.minus.grpo.writing-abs}{3.0}
\setval{8b.papo.minus.grpo.writingbench}{$-$3.3}
\setval{8b.papo.minus.grpo.writingbench-abs}{3.3}
\setval{8b.papo.minus.mt-grpo.aime24}{$-$2.3}
\setval{8b.papo.minus.mt-grpo.aime24-abs}{2.3}
\setval{8b.papo.minus.mt-grpo.aime25}{$-$5.6}
\setval{8b.papo.minus.mt-grpo.aime25-abs}{5.6}
\setval{8b.papo.minus.mt-grpo.amc23}{$-$3.6}
\setval{8b.papo.minus.mt-grpo.amc23-abs}{3.6}
\setval{8b.papo.minus.mt-grpo.avg}{$-$1.4}
\setval{8b.papo.minus.mt-grpo.avg-abs}{1.4}
\setval{8b.papo.minus.mt-grpo.code}{$-$1.8}
\setval{8b.papo.minus.mt-grpo.code-abs}{1.8}
\setval{8b.papo.minus.mt-grpo.creative-writing-v3}{+0.8}
\setval{8b.papo.minus.mt-grpo.creative-writing-v3-abs}{0.8}
\setval{8b.papo.minus.mt-grpo.guidebench}{+1.2}
\setval{8b.papo.minus.mt-grpo.guidebench-abs}{1.2}
\setval{8b.papo.minus.mt-grpo.humaneval-plus}{$-$2.6}
\setval{8b.papo.minus.mt-grpo.humaneval-plus-abs}{2.6}
\setval{8b.papo.minus.mt-grpo.if}{+0.2}
\setval{8b.papo.minus.mt-grpo.if-abs}{0.2}
\setval{8b.papo.minus.mt-grpo.ifbench}{+0.0}
\setval{8b.papo.minus.mt-grpo.ifbench-abs}{0.0}
\setval{8b.papo.minus.mt-grpo.ifeval}{$-$0.6}
\setval{8b.papo.minus.mt-grpo.ifeval-abs}{0.6}
\setval{8b.papo.minus.mt-grpo.judge2.avg}{$-$1.5}
\setval{8b.papo.minus.mt-grpo.judge2.avg-abs}{1.5}
\setval{8b.papo.minus.mt-grpo.judge2.creative-writing-v3}{+0.3}
\setval{8b.papo.minus.mt-grpo.judge2.creative-writing-v3-abs}{0.3}
\setval{8b.papo.minus.mt-grpo.judge2.longbench-write}{$-$2.9}
\setval{8b.papo.minus.mt-grpo.judge2.longbench-write-abs}{2.9}
\setval{8b.papo.minus.mt-grpo.judge2.writing}{$-$0.6}
\setval{8b.papo.minus.mt-grpo.judge2.writing-abs}{0.6}
\setval{8b.papo.minus.mt-grpo.judge2.writingbench}{+0.8}
\setval{8b.papo.minus.mt-grpo.judge2.writingbench-abs}{0.8}
\setval{8b.papo.minus.mt-grpo.livecodebench}{$-$1.0}
\setval{8b.papo.minus.mt-grpo.livecodebench-abs}{1.0}
\setval{8b.papo.minus.mt-grpo.longbench-write}{$-$1.9}
\setval{8b.papo.minus.mt-grpo.longbench-write-abs}{1.9}
\setval{8b.papo.minus.mt-grpo.math}{$-$2.7}
\setval{8b.papo.minus.mt-grpo.math-abs}{2.7}
\setval{8b.papo.minus.mt-grpo.math500}{$-$2.1}
\setval{8b.papo.minus.mt-grpo.math500-abs}{2.1}
\setval{8b.papo.minus.mt-grpo.mbpp-plus}{$-$1.9}
\setval{8b.papo.minus.mt-grpo.mbpp-plus-abs}{1.9}
\setval{8b.papo.minus.mt-grpo.minerva}{$-$1.7}
\setval{8b.papo.minus.mt-grpo.minerva-abs}{1.7}
\setval{8b.papo.minus.mt-grpo.olympiad-bench}{$-$0.9}
\setval{8b.papo.minus.mt-grpo.olympiad-bench-abs}{0.9}
\setval{8b.papo.minus.mt-grpo.writing}{+0.1}
\setval{8b.papo.minus.mt-grpo.writing-abs}{0.1}
\setval{8b.papo.minus.mt-grpo.writingbench}{+1.5}
\setval{8b.papo.minus.mt-grpo.writingbench-abs}{1.5}
\setval{8b.papo.minus.saw.aime24}{$-$3.1}
\setval{8b.papo.minus.saw.aime24-abs}{3.1}
\setval{8b.papo.minus.saw.aime25}{$-$5.4}
\setval{8b.papo.minus.saw.aime25-abs}{5.4}
\setval{8b.papo.minus.saw.amc23}{$-$0.2}
\setval{8b.papo.minus.saw.amc23-abs}{0.2}
\setval{8b.papo.minus.saw.avg}{$-$2.2}
\setval{8b.papo.minus.saw.avg-abs}{2.2}
\setval{8b.papo.minus.saw.code}{$-$1.4}
\setval{8b.papo.minus.saw.code-abs}{1.4}
\setval{8b.papo.minus.saw.creative-writing-v3}{+1.5}
\setval{8b.papo.minus.saw.creative-writing-v3-abs}{1.5}
\setval{8b.papo.minus.saw.guidebench}{+0.1}
\setval{8b.papo.minus.saw.guidebench-abs}{0.1}
\setval{8b.papo.minus.saw.humaneval-plus}{$-$3.4}
\setval{8b.papo.minus.saw.humaneval-plus-abs}{3.4}
\setval{8b.papo.minus.saw.if}{$-$1.3}
\setval{8b.papo.minus.saw.if-abs}{1.3}
\setval{8b.papo.minus.saw.ifbench}{$-$2.7}
\setval{8b.papo.minus.saw.ifbench-abs}{2.7}
\setval{8b.papo.minus.saw.ifeval}{$-$1.3}
\setval{8b.papo.minus.saw.ifeval-abs}{1.3}
\setval{8b.papo.minus.saw.judge2.avg}{$-$1.8}
\setval{8b.papo.minus.saw.judge2.avg-abs}{1.8}
\setval{8b.papo.minus.saw.judge2.creative-writing-v3}{+5.3}
\setval{8b.papo.minus.saw.judge2.creative-writing-v3-abs}{5.3}
\setval{8b.papo.minus.saw.judge2.longbench-write}{$-$5.3}
\setval{8b.papo.minus.saw.judge2.longbench-write-abs}{5.3}
\setval{8b.papo.minus.saw.judge2.writing}{$-$0.2}
\setval{8b.papo.minus.saw.judge2.writing-abs}{0.2}
\setval{8b.papo.minus.saw.judge2.writingbench}{$-$0.6}
\setval{8b.papo.minus.saw.judge2.writingbench-abs}{0.6}
\setval{8b.papo.minus.saw.livecodebench}{$-$0.1}
\setval{8b.papo.minus.saw.livecodebench-abs}{0.1}
\setval{8b.papo.minus.saw.longbench-write}{$-$5.5}
\setval{8b.papo.minus.saw.longbench-write-abs}{5.5}
\setval{8b.papo.minus.saw.math}{$-$3.0}
\setval{8b.papo.minus.saw.math-abs}{3.0}
\setval{8b.papo.minus.saw.math500}{$-$3.4}
\setval{8b.papo.minus.saw.math500-abs}{3.4}
\setval{8b.papo.minus.saw.mbpp-plus}{$-$0.9}
\setval{8b.papo.minus.saw.mbpp-plus-abs}{0.9}
\setval{8b.papo.minus.saw.minerva}{$-$2.4}
\setval{8b.papo.minus.saw.minerva-abs}{2.4}
\setval{8b.papo.minus.saw.olympiad-bench}{$-$3.3}
\setval{8b.papo.minus.saw.olympiad-bench-abs}{3.3}
\setval{8b.papo.minus.saw.writing}{$-$2.1}
\setval{8b.papo.minus.saw.writing-abs}{2.1}
\setval{8b.papo.minus.saw.writingbench}{$-$2.2}
\setval{8b.papo.minus.saw.writingbench-abs}{2.2}
\setval{8b.papo.passing-at-ceiling.code.early}{0.71}
\setval{8b.papo.passing-at-ceiling.code.late}{0.76}
\setval{8b.papo.passing-at-ceiling.code.run}{0.75}
\setval{8b.papo.passing-at-ceiling.if.early}{0.33}
\setval{8b.papo.passing-at-ceiling.if.late}{0.57}
\setval{8b.papo.passing-at-ceiling.if.run}{0.49}
\setval{8b.papo.passing-at-ceiling.math.early}{1.00}
\setval{8b.papo.passing-at-ceiling.math.late}{1.00}
\setval{8b.papo.passing-at-ceiling.math.run}{1.00}
\setval{8b.papo.passing-at-ceiling.writing.early}{0.00}
\setval{8b.papo.passing-at-ceiling.writing.late}{0.00}
\setval{8b.papo.passing-at-ceiling.writing.run}{0.00}
\setval{8b.papo.primary-cosine.code.early}{0.90}
\setval{8b.papo.primary-cosine.code.late}{0.98}
\setval{8b.papo.primary-cosine.code.run}{0.97}
\setval{8b.papo.primary-cosine.if.early}{0.70}
\setval{8b.papo.primary-cosine.if.late}{0.75}
\setval{8b.papo.primary-cosine.if.run}{0.77}
\setval{8b.papo.primary-cosine.math.early}{0.84}
\setval{8b.papo.primary-cosine.math.late}{0.97}
\setval{8b.papo.primary-cosine.math.run}{0.93}
\setval{8b.papo.primary-cosine.writing.early}{0.76}
\setval{8b.papo.primary-cosine.writing.late}{0.79}
\setval{8b.papo.primary-cosine.writing.run}{0.76}
\setval{8b.papo.primary-level.code.early}{0.54}
\setval{8b.papo.primary-level.code.late}{0.76}
\setval{8b.papo.primary-level.code.run}{0.71}
\setval{8b.papo.primary-level.if.early}{0.36}
\setval{8b.papo.primary-level.if.late}{0.75}
\setval{8b.papo.primary-level.if.run}{0.65}
\setval{8b.papo.primary-level.math.early}{0.48}
\setval{8b.papo.primary-level.math.late}{0.65}
\setval{8b.papo.primary-level.math.run}{0.60}
\setval{8b.papo.primary-level.writing.early}{0.53}
\setval{8b.papo.primary-level.writing.late}{0.76}
\setval{8b.papo.primary-level.writing.run}{0.71}
\setval{8b.papo.solved.code.early}{0.04}
\setval{8b.papo.solved.code.late}{0.01}
\setval{8b.papo.solved.code.run}{0.02}
\setval{8b.papo.solved.if.early}{0.03}
\setval{8b.papo.solved.if.late}{0.16}
\setval{8b.papo.solved.if.run}{0.11}
\setval{8b.papo.solved.math.early}{0.13}
\setval{8b.papo.solved.math.late}{0.03}
\setval{8b.papo.solved.math.run}{0.06}
\setval{8b.papo.solved.writing.early}{0.00}
\setval{8b.papo.solved.writing.late}{0.00}
\setval{8b.papo.solved.writing.run}{0.00}
\setval{8b.papo.update-share.code.early}{0.18}
\setval{8b.papo.update-share.code.late}{0.15}
\setval{8b.papo.update-share.code.run}{0.15}
\setval{8b.papo.update-share.if.early}{0.28}
\setval{8b.papo.update-share.if.late}{0.20}
\setval{8b.papo.update-share.if.run}{0.22}
\setval{8b.papo.update-share.math.early}{0.17}
\setval{8b.papo.update-share.math.late}{0.15}
\setval{8b.papo.update-share.math.run}{0.15}
\setval{8b.papo.update-share.writing.early}{0.36}
\setval{8b.papo.update-share.writing.late}{0.51}
\setval{8b.papo.update-share.writing.run}{0.48}
\setval{8b.saw.aux-change.code}{+0.18}
\setval{8b.saw.aux-change.if}{+0.19}
\setval{8b.saw.aux-change.math}{+0.13}
\setval{8b.saw.aux-change.writing}{+0.14}
\setval{8b.saw.aux-change.writing-rm}{+0.33}
\setval{8b.saw.aux-full.code.early}{0.71}
\setval{8b.saw.aux-full.code.late}{0.97}
\setval{8b.saw.aux-full.code.run}{0.95}
\setval{8b.saw.aux-full.if.early}{0.73}
\setval{8b.saw.aux-full.if.late}{0.94}
\setval{8b.saw.aux-full.if.run}{0.91}
\setval{8b.saw.aux-full.math.early}{0.57}
\setval{8b.saw.aux-full.math.late}{0.89}
\setval{8b.saw.aux-full.math.run}{0.84}
\setval{8b.saw.aux-full.writing-rm.early}{0.07}
\setval{8b.saw.aux-full.writing-rm.late}{0.65}
\setval{8b.saw.aux-full.writing-rm.run}{0.51}
\setval{8b.saw.aux-full.writing.early}{0.65}
\setval{8b.saw.aux-full.writing.late}{0.81}
\setval{8b.saw.aux-full.writing.run}{0.75}
\setval{8b.saw.aux-level.code.early}{0.93}
\setval{8b.saw.aux-level.code.late}{0.99}
\setval{8b.saw.aux-level.code.run}{0.99}
\setval{8b.saw.aux-level.if.early}{0.88}
\setval{8b.saw.aux-level.if.late}{0.97}
\setval{8b.saw.aux-level.if.run}{0.95}
\setval{8b.saw.aux-level.math.early}{0.91}
\setval{8b.saw.aux-level.math.late}{0.98}
\setval{8b.saw.aux-level.math.run}{0.97}
\setval{8b.saw.aux-level.writing-rm.early}{0.77}
\setval{8b.saw.aux-level.writing-rm.late}{0.97}
\setval{8b.saw.aux-level.writing-rm.run}{0.94}
\setval{8b.saw.aux-level.writing.early}{0.91}
\setval{8b.saw.aux-level.writing.late}{0.98}
\setval{8b.saw.aux-level.writing.run}{0.96}
\setval{8b.saw.bench.aime24}{15.3}
\setval{8b.saw.bench.aime25}{15.5}
\setval{8b.saw.bench.amc23}{62.3}
\setval{8b.saw.bench.creative-writing-v3}{49.7}
\setval{8b.saw.bench.guidebench}{84.3}
\setval{8b.saw.bench.humaneval-plus}{85.9}
\setval{8b.saw.bench.ifbench}{40.0}
\setval{8b.saw.bench.ifeval}{87.6}
\setval{8b.saw.bench.livecodebench}{26.7}
\setval{8b.saw.bench.longbench-write}{71.0}
\setval{8b.saw.bench.math500}{81.9}
\setval{8b.saw.bench.mbpp-plus}{74.3}
\setval{8b.saw.bench.minerva}{51.8}
\setval{8b.saw.bench.olympiad-bench}{48.1}
\setval{8b.saw.bench.writingbench}{75.4}
\setval{8b.saw.eval.avg}{58.0}
\setval{8b.saw.eval.code}{62.3}
\setval{8b.saw.eval.if}{70.6}
\setval{8b.saw.eval.math}{45.8}
\setval{8b.saw.eval.writing}{65.4}
\setval{8b.saw.gate.code.against.early}{0.04}
\setval{8b.saw.gate.code.against.late}{0.02}
\setval{8b.saw.gate.code.against.run}{0.02}
\setval{8b.saw.gate.code.agreement}{+0.005}
\setval{8b.saw.gate.code.ceiling.early}{0.11}
\setval{8b.saw.gate.code.ceiling.late}{0.03}
\setval{8b.saw.gate.code.ceiling.run}{0.04}
\setval{8b.saw.gate.code.cosine.early}{0.83}
\setval{8b.saw.gate.code.cosine.late}{0.77}
\setval{8b.saw.gate.code.cosine.run}{0.81}
\setval{8b.saw.gate.if.against.early}{0.04}
\setval{8b.saw.gate.if.against.late}{0.06}
\setval{8b.saw.gate.if.against.run}{0.04}
\setval{8b.saw.gate.if.agreement}{+0.011}
\setval{8b.saw.gate.if.ceiling.early}{0.05}
\setval{8b.saw.gate.if.ceiling.late}{0.14}
\setval{8b.saw.gate.if.ceiling.run}{0.10}
\setval{8b.saw.gate.if.cosine.early}{0.76}
\setval{8b.saw.gate.if.cosine.late}{0.50}
\setval{8b.saw.gate.if.cosine.run}{0.60}
\setval{8b.saw.gate.math.against.early}{0.00}
\setval{8b.saw.gate.math.against.late}{0.00}
\setval{8b.saw.gate.math.against.run}{0.00}
\setval{8b.saw.gate.math.ceiling.early}{0.22}
\setval{8b.saw.gate.math.ceiling.late}{0.20}
\setval{8b.saw.gate.math.ceiling.run}{0.19}
\setval{8b.saw.gate.math.cosine.early}{0.86}
\setval{8b.saw.gate.math.cosine.late}{0.88}
\setval{8b.saw.gate.math.cosine.run}{0.89}
\setval{8b.saw.gate.writing-rm.against.early}{0.20}
\setval{8b.saw.gate.writing-rm.against.late}{0.51}
\setval{8b.saw.gate.writing-rm.against.run}{0.48}
\setval{8b.saw.gate.writing-rm.agreement}{+0.050}
\setval{8b.saw.gate.writing-rm.ceiling.early}{0.00}
\setval{8b.saw.gate.writing-rm.ceiling.late}{0.00}
\setval{8b.saw.gate.writing-rm.ceiling.run}{0.00}
\setval{8b.saw.gate.writing-rm.cosine.early}{0.69}
\setval{8b.saw.gate.writing-rm.cosine.late}{0.13}
\setval{8b.saw.gate.writing-rm.cosine.run}{0.22}
\setval{8b.saw.gate.writing.against.early}{0.27}
\setval{8b.saw.gate.writing.against.late}{0.59}
\setval{8b.saw.gate.writing.against.run}{0.53}
\setval{8b.saw.gate.writing.agreement}{$-$0.032}
\setval{8b.saw.gate.writing.ceiling.early}{0.00}
\setval{8b.saw.gate.writing.ceiling.late}{0.00}
\setval{8b.saw.gate.writing.ceiling.run}{0.00}
\setval{8b.saw.gate.writing.cosine.early}{0.53}
\setval{8b.saw.gate.writing.cosine.late}{0.04}
\setval{8b.saw.gate.writing.cosine.run}{0.12}
\setval{8b.saw.judge2.avg}{55.5}
\setval{8b.saw.judge2.creative-writing-v3}{36.6}
\setval{8b.saw.judge2.longbench-write}{64.8}
\setval{8b.saw.judge2.writing}{52.8}
\setval{8b.saw.judge2.writingbench}{56.9}
\setval{8b.saw.last-step}{300}
\setval{8b.saw.minus.base.aime24}{+4.0}
\setval{8b.saw.minus.base.aime24-abs}{4.0}
\setval{8b.saw.minus.base.aime25}{+7.0}
\setval{8b.saw.minus.base.aime25-abs}{7.0}
\setval{8b.saw.minus.base.amc23}{+12.0}
\setval{8b.saw.minus.base.amc23-abs}{12.0}
\setval{8b.saw.minus.base.avg}{+16.5}
\setval{8b.saw.minus.base.avg-abs}{16.5}
\setval{8b.saw.minus.base.code}{+6.2}
\setval{8b.saw.minus.base.code-abs}{6.2}
\setval{8b.saw.minus.base.creative-writing-v3}{+23.0}
\setval{8b.saw.minus.base.creative-writing-v3-abs}{23.0}
\setval{8b.saw.minus.base.guidebench}{+38.7}
\setval{8b.saw.minus.base.guidebench-abs}{38.7}
\setval{8b.saw.minus.base.humaneval-plus}{+5.8}
\setval{8b.saw.minus.base.humaneval-plus-abs}{5.8}
\setval{8b.saw.minus.base.if}{+32.0}
\setval{8b.saw.minus.base.if-abs}{32.0}
\setval{8b.saw.minus.base.ifbench}{+22.7}
\setval{8b.saw.minus.base.ifbench-abs}{22.7}
\setval{8b.saw.minus.base.ifeval}{+34.7}
\setval{8b.saw.minus.base.ifeval-abs}{34.7}
\setval{8b.saw.minus.base.judge2.avg}{+14.4}
\setval{8b.saw.minus.base.judge2.avg-abs}{14.4}
\setval{8b.saw.minus.base.judge2.creative-writing-v3}{+10.1}
\setval{8b.saw.minus.base.judge2.creative-writing-v3-abs}{10.1}
\setval{8b.saw.minus.base.judge2.longbench-write}{+16.8}
\setval{8b.saw.minus.base.judge2.longbench-write-abs}{16.8}
\setval{8b.saw.minus.base.judge2.writing}{+16.6}
\setval{8b.saw.minus.base.judge2.writing-abs}{16.6}
\setval{8b.saw.minus.base.judge2.writingbench}{+22.9}
\setval{8b.saw.minus.base.judge2.writingbench-abs}{22.9}
\setval{8b.saw.minus.base.livecodebench}{+5.8}
\setval{8b.saw.minus.base.livecodebench-abs}{5.8}
\setval{8b.saw.minus.base.longbench-write}{+19.6}
\setval{8b.saw.minus.base.longbench-write-abs}{19.6}
\setval{8b.saw.minus.base.math}{+8.7}
\setval{8b.saw.minus.base.math-abs}{8.7}
\setval{8b.saw.minus.base.math500}{+9.5}
\setval{8b.saw.minus.base.math500-abs}{9.5}
\setval{8b.saw.minus.base.mbpp-plus}{+7.1}
\setval{8b.saw.minus.base.mbpp-plus-abs}{7.1}
\setval{8b.saw.minus.base.minerva}{+10.9}
\setval{8b.saw.minus.base.minerva-abs}{10.9}
\setval{8b.saw.minus.base.olympiad-bench}{+8.7}
\setval{8b.saw.minus.base.olympiad-bench-abs}{8.7}
\setval{8b.saw.minus.base.writing}{+27.1}
\setval{8b.saw.minus.base.writing-abs}{27.1}
\setval{8b.saw.minus.base.writingbench}{+38.6}
\setval{8b.saw.minus.base.writingbench-abs}{38.6}
\setval{8b.saw.minus.concur-free.aime24}{$-$4.4}
\setval{8b.saw.minus.concur-free.aime24-abs}{4.4}
\setval{8b.saw.minus.concur-free.aime25}{$-$1.0}
\setval{8b.saw.minus.concur-free.aime25-abs}{1.0}
\setval{8b.saw.minus.concur-free.amc23}{$-$2.4}
\setval{8b.saw.minus.concur-free.amc23-abs}{2.4}
\setval{8b.saw.minus.concur-free.avg}{$-$2.3}
\setval{8b.saw.minus.concur-free.avg-abs}{2.3}
\setval{8b.saw.minus.concur-free.code}{$-$0.2}
\setval{8b.saw.minus.concur-free.code-abs}{0.2}
\setval{8b.saw.minus.concur-free.creative-writing-v3}{$-$2.0}
\setval{8b.saw.minus.concur-free.creative-writing-v3-abs}{2.0}
\setval{8b.saw.minus.concur-free.guidebench}{+1.3}
\setval{8b.saw.minus.concur-free.guidebench-abs}{1.3}
\setval{8b.saw.minus.concur-free.humaneval-plus}{+0.5}
\setval{8b.saw.minus.concur-free.humaneval-plus-abs}{0.5}
\setval{8b.saw.minus.concur-free.if}{+0.4}
\setval{8b.saw.minus.concur-free.if-abs}{0.4}
\setval{8b.saw.minus.concur-free.ifbench}{+1.3}
\setval{8b.saw.minus.concur-free.ifbench-abs}{1.3}
\setval{8b.saw.minus.concur-free.ifeval}{$-$1.6}
\setval{8b.saw.minus.concur-free.ifeval-abs}{1.6}
\setval{8b.saw.minus.concur-free.judge2.avg}{$-$1.9}
\setval{8b.saw.minus.concur-free.judge2.avg-abs}{1.9}
\setval{8b.saw.minus.concur-free.judge2.creative-writing-v3}{$-$5.9}
\setval{8b.saw.minus.concur-free.judge2.creative-writing-v3-abs}{5.9}
\setval{8b.saw.minus.concur-free.judge2.longbench-write}{$-$9.3}
\setval{8b.saw.minus.concur-free.judge2.longbench-write-abs}{9.3}
\setval{8b.saw.minus.concur-free.judge2.writing}{$-$6.3}
\setval{8b.saw.minus.concur-free.judge2.writing-abs}{6.3}
\setval{8b.saw.minus.concur-free.judge2.writingbench}{$-$3.7}
\setval{8b.saw.minus.concur-free.judge2.writingbench-abs}{3.7}
\setval{8b.saw.minus.concur-free.livecodebench}{$-$0.4}
\setval{8b.saw.minus.concur-free.livecodebench-abs}{0.4}
\setval{8b.saw.minus.concur-free.longbench-write}{$-$11.9}
\setval{8b.saw.minus.concur-free.longbench-write-abs}{11.9}
\setval{8b.saw.minus.concur-free.math}{$-$1.5}
\setval{8b.saw.minus.concur-free.math-abs}{1.5}
\setval{8b.saw.minus.concur-free.math500}{$-$1.5}
\setval{8b.saw.minus.concur-free.math500-abs}{1.5}
\setval{8b.saw.minus.concur-free.mbpp-plus}{$-$0.8}
\setval{8b.saw.minus.concur-free.mbpp-plus-abs}{0.8}
\setval{8b.saw.minus.concur-free.minerva}{+2.0}
\setval{8b.saw.minus.concur-free.minerva-abs}{2.0}
\setval{8b.saw.minus.concur-free.olympiad-bench}{$-$1.9}
\setval{8b.saw.minus.concur-free.olympiad-bench-abs}{1.9}
\setval{8b.saw.minus.concur-free.writing}{$-$8.4}
\setval{8b.saw.minus.concur-free.writing-abs}{8.4}
\setval{8b.saw.minus.concur-free.writingbench}{$-$11.4}
\setval{8b.saw.minus.concur-free.writingbench-abs}{11.4}
\setval{8b.saw.minus.concur-tied.aime24}{$-$15.9}
\setval{8b.saw.minus.concur-tied.aime24-abs}{15.9}
\setval{8b.saw.minus.concur-tied.aime25}{$-$8.4}
\setval{8b.saw.minus.concur-tied.aime25-abs}{8.4}
\setval{8b.saw.minus.concur-tied.amc23}{$-$16.2}
\setval{8b.saw.minus.concur-tied.amc23-abs}{16.2}
\setval{8b.saw.minus.concur-tied.avg}{$-$6.0}
\setval{8b.saw.minus.concur-tied.avg-abs}{6.0}
\setval{8b.saw.minus.concur-tied.code}{$-$0.9}
\setval{8b.saw.minus.concur-tied.code-abs}{0.9}
\setval{8b.saw.minus.concur-tied.creative-writing-v3}{$-$2.8}
\setval{8b.saw.minus.concur-tied.creative-writing-v3-abs}{2.8}
\setval{8b.saw.minus.concur-tied.guidebench}{+0.9}
\setval{8b.saw.minus.concur-tied.guidebench-abs}{0.9}
\setval{8b.saw.minus.concur-tied.humaneval-plus}{$-$1.5}
\setval{8b.saw.minus.concur-tied.humaneval-plus-abs}{1.5}
\setval{8b.saw.minus.concur-tied.if}{+0.1}
\setval{8b.saw.minus.concur-tied.if-abs}{0.1}
\setval{8b.saw.minus.concur-tied.ifbench}{+0.7}
\setval{8b.saw.minus.concur-tied.ifbench-abs}{0.7}
\setval{8b.saw.minus.concur-tied.ifeval}{$-$1.2}
\setval{8b.saw.minus.concur-tied.ifeval-abs}{1.2}
\setval{8b.saw.minus.concur-tied.judge2.avg}{$-$5.8}
\setval{8b.saw.minus.concur-tied.judge2.avg-abs}{5.8}
\setval{8b.saw.minus.concur-tied.judge2.creative-writing-v3}{$-$5.2}
\setval{8b.saw.minus.concur-tied.judge2.creative-writing-v3-abs}{5.2}
\setval{8b.saw.minus.concur-tied.judge2.longbench-write}{$-$10.0}
\setval{8b.saw.minus.concur-tied.judge2.longbench-write-abs}{10.0}
\setval{8b.saw.minus.concur-tied.judge2.writing}{$-$7.3}
\setval{8b.saw.minus.concur-tied.judge2.writing-abs}{7.3}
\setval{8b.saw.minus.concur-tied.judge2.writingbench}{$-$6.8}
\setval{8b.saw.minus.concur-tied.judge2.writingbench-abs}{6.8}
\setval{8b.saw.minus.concur-tied.livecodebench}{$-$0.5}
\setval{8b.saw.minus.concur-tied.livecodebench-abs}{0.5}
\setval{8b.saw.minus.concur-tied.longbench-write}{$-$10.2}
\setval{8b.saw.minus.concur-tied.longbench-write-abs}{10.2}
\setval{8b.saw.minus.concur-tied.math}{$-$10.6}
\setval{8b.saw.minus.concur-tied.math-abs}{10.6}
\setval{8b.saw.minus.concur-tied.math500}{$-$7.7}
\setval{8b.saw.minus.concur-tied.math500-abs}{7.7}
\setval{8b.saw.minus.concur-tied.mbpp-plus}{$-$0.5}
\setval{8b.saw.minus.concur-tied.mbpp-plus-abs}{0.5}
\setval{8b.saw.minus.concur-tied.minerva}{$-$3.6}
\setval{8b.saw.minus.concur-tied.minerva-abs}{3.6}
\setval{8b.saw.minus.concur-tied.olympiad-bench}{$-$11.4}
\setval{8b.saw.minus.concur-tied.olympiad-bench-abs}{11.4}
\setval{8b.saw.minus.concur-tied.writing}{$-$8.2}
\setval{8b.saw.minus.concur-tied.writing-abs}{8.2}
\setval{8b.saw.minus.concur-tied.writingbench}{$-$11.8}
\setval{8b.saw.minus.concur-tied.writingbench-abs}{11.8}
\setval{8b.saw.minus.concur.aime24}{$-$13.4}
\setval{8b.saw.minus.concur.aime24-abs}{13.4}
\setval{8b.saw.minus.concur.aime25}{$-$7.6}
\setval{8b.saw.minus.concur.aime25-abs}{7.6}
\setval{8b.saw.minus.concur.amc23}{$-$16.6}
\setval{8b.saw.minus.concur.amc23-abs}{16.6}
\setval{8b.saw.minus.concur.avg}{$-$5.2}
\setval{8b.saw.minus.concur.avg-abs}{5.2}
\setval{8b.saw.minus.concur.code}{+1.5}
\setval{8b.saw.minus.concur.code-abs}{1.5}
\setval{8b.saw.minus.concur.creative-writing-v3}{$-$0.2}
\setval{8b.saw.minus.concur.creative-writing-v3-abs}{0.2}
\setval{8b.saw.minus.concur.guidebench}{+0.9}
\setval{8b.saw.minus.concur.guidebench-abs}{0.9}
\setval{8b.saw.minus.concur.humaneval-plus}{+3.1}
\setval{8b.saw.minus.concur.humaneval-plus-abs}{3.1}
\setval{8b.saw.minus.concur.if}{+0.6}
\setval{8b.saw.minus.concur.if-abs}{0.6}
\setval{8b.saw.minus.concur.ifbench}{+1.3}
\setval{8b.saw.minus.concur.ifbench-abs}{1.3}
\setval{8b.saw.minus.concur.ifeval}{$-$0.5}
\setval{8b.saw.minus.concur.ifeval-abs}{0.5}
\setval{8b.saw.minus.concur.judge2.avg}{$-$4.9}
\setval{8b.saw.minus.concur.judge2.avg-abs}{4.9}
\setval{8b.saw.minus.concur.judge2.creative-writing-v3}{$-$1.2}
\setval{8b.saw.minus.concur.judge2.creative-writing-v3-abs}{1.2}
\setval{8b.saw.minus.concur.judge2.longbench-write}{$-$10.7}
\setval{8b.saw.minus.concur.judge2.longbench-write-abs}{10.7}
\setval{8b.saw.minus.concur.judge2.writing}{$-$6.3}
\setval{8b.saw.minus.concur.judge2.writing-abs}{6.3}
\setval{8b.saw.minus.concur.judge2.writingbench}{$-$7.0}
\setval{8b.saw.minus.concur.judge2.writingbench-abs}{7.0}
\setval{8b.saw.minus.concur.livecodebench}{+0.3}
\setval{8b.saw.minus.concur.livecodebench-abs}{0.3}
\setval{8b.saw.minus.concur.longbench-write}{$-$10.6}
\setval{8b.saw.minus.concur.longbench-write-abs}{10.6}
\setval{8b.saw.minus.concur.math}{$-$10.1}
\setval{8b.saw.minus.concur.math-abs}{10.1}
\setval{8b.saw.minus.concur.math500}{$-$8.2}
\setval{8b.saw.minus.concur.math500-abs}{8.2}
\setval{8b.saw.minus.concur.mbpp-plus}{+1.1}
\setval{8b.saw.minus.concur.mbpp-plus-abs}{1.1}
\setval{8b.saw.minus.concur.minerva}{$-$2.4}
\setval{8b.saw.minus.concur.minerva-abs}{2.4}
\setval{8b.saw.minus.concur.olympiad-bench}{$-$12.2}
\setval{8b.saw.minus.concur.olympiad-bench-abs}{12.2}
\setval{8b.saw.minus.concur.writing}{$-$7.8}
\setval{8b.saw.minus.concur.writing-abs}{7.8}
\setval{8b.saw.minus.concur.writingbench}{$-$12.7}
\setval{8b.saw.minus.concur.writingbench-abs}{12.7}
\setval{8b.saw.minus.gd2po.aime24}{$-$0.9}
\setval{8b.saw.minus.gd2po.aime24-abs}{0.9}
\setval{8b.saw.minus.gd2po.aime25}{+3.3}
\setval{8b.saw.minus.gd2po.aime25-abs}{3.3}
\setval{8b.saw.minus.gd2po.amc23}{+2.7}
\setval{8b.saw.minus.gd2po.amc23-abs}{2.7}
\setval{8b.saw.minus.gd2po.avg}{+0.6}
\setval{8b.saw.minus.gd2po.avg-abs}{0.6}
\setval{8b.saw.minus.gd2po.code}{+0.5}
\setval{8b.saw.minus.gd2po.code-abs}{0.5}
\setval{8b.saw.minus.gd2po.creative-writing-v3}{+1.5}
\setval{8b.saw.minus.gd2po.creative-writing-v3-abs}{1.5}
\setval{8b.saw.minus.gd2po.guidebench}{+0.6}
\setval{8b.saw.minus.gd2po.guidebench-abs}{0.6}
\setval{8b.saw.minus.gd2po.humaneval-plus}{+1.9}
\setval{8b.saw.minus.gd2po.humaneval-plus-abs}{1.9}
\setval{8b.saw.minus.gd2po.if}{$-$1.4}
\setval{8b.saw.minus.gd2po.if-abs}{1.4}
\setval{8b.saw.minus.gd2po.ifbench}{$-$3.0}
\setval{8b.saw.minus.gd2po.ifbench-abs}{3.0}
\setval{8b.saw.minus.gd2po.ifeval}{$-$1.7}
\setval{8b.saw.minus.gd2po.ifeval-abs}{1.7}
\setval{8b.saw.minus.gd2po.judge2.avg}{+0.6}
\setval{8b.saw.minus.gd2po.judge2.avg-abs}{0.6}
\setval{8b.saw.minus.gd2po.judge2.creative-writing-v3}{$-$0.2}
\setval{8b.saw.minus.gd2po.judge2.creative-writing-v3-abs}{0.2}
\setval{8b.saw.minus.gd2po.judge2.longbench-write}{+2.6}
\setval{8b.saw.minus.gd2po.judge2.longbench-write-abs}{2.6}
\setval{8b.saw.minus.gd2po.judge2.writing}{+1.4}
\setval{8b.saw.minus.gd2po.judge2.writing-abs}{1.4}
\setval{8b.saw.minus.gd2po.judge2.writingbench}{+1.7}
\setval{8b.saw.minus.gd2po.judge2.writingbench-abs}{1.7}
\setval{8b.saw.minus.gd2po.livecodebench}{+0.4}
\setval{8b.saw.minus.gd2po.livecodebench-abs}{0.4}
\setval{8b.saw.minus.gd2po.longbench-write}{+1.1}
\setval{8b.saw.minus.gd2po.longbench-write-abs}{1.1}
\setval{8b.saw.minus.gd2po.math}{+1.2}
\setval{8b.saw.minus.gd2po.math-abs}{1.2}
\setval{8b.saw.minus.gd2po.math500}{+0.7}
\setval{8b.saw.minus.gd2po.math500-abs}{0.7}
\setval{8b.saw.minus.gd2po.mbpp-plus}{$-$0.7}
\setval{8b.saw.minus.gd2po.mbpp-plus-abs}{0.7}
\setval{8b.saw.minus.gd2po.minerva}{+1.8}
\setval{8b.saw.minus.gd2po.minerva-abs}{1.8}
\setval{8b.saw.minus.gd2po.olympiad-bench}{$-$0.4}
\setval{8b.saw.minus.gd2po.olympiad-bench-abs}{0.4}
\setval{8b.saw.minus.gd2po.writing}{+1.4}
\setval{8b.saw.minus.gd2po.writing-abs}{1.4}
\setval{8b.saw.minus.gd2po.writingbench}{+1.5}
\setval{8b.saw.minus.gd2po.writingbench-abs}{1.5}
\setval{8b.saw.minus.gdpo-noaux.aime24}{$-$12.6}
\setval{8b.saw.minus.gdpo-noaux.aime24-abs}{12.6}
\setval{8b.saw.minus.gdpo-noaux.aime25}{$-$8.4}
\setval{8b.saw.minus.gdpo-noaux.aime25-abs}{8.4}
\setval{8b.saw.minus.gdpo-noaux.amc23}{$-$14.8}
\setval{8b.saw.minus.gdpo-noaux.amc23-abs}{14.8}
\setval{8b.saw.minus.gdpo-noaux.avg}{$-$2.5}
\setval{8b.saw.minus.gdpo-noaux.avg-abs}{2.5}
\setval{8b.saw.minus.gdpo-noaux.code}{$-$0.4}
\setval{8b.saw.minus.gdpo-noaux.code-abs}{0.4}
\setval{8b.saw.minus.gdpo-noaux.creative-writing-v3}{+5.8}
\setval{8b.saw.minus.gdpo-noaux.creative-writing-v3-abs}{5.8}
\setval{8b.saw.minus.gdpo-noaux.guidebench}{+2.9}
\setval{8b.saw.minus.gdpo-noaux.guidebench-abs}{2.9}
\setval{8b.saw.minus.gdpo-noaux.humaneval-plus}{$-$0.1}
\setval{8b.saw.minus.gdpo-noaux.humaneval-plus-abs}{0.1}
\setval{8b.saw.minus.gdpo-noaux.if}{+3.4}
\setval{8b.saw.minus.gdpo-noaux.if-abs}{3.4}
\setval{8b.saw.minus.gdpo-noaux.ifbench}{+6.3}
\setval{8b.saw.minus.gdpo-noaux.ifbench-abs}{6.3}
\setval{8b.saw.minus.gdpo-noaux.ifeval}{+1.1}
\setval{8b.saw.minus.gdpo-noaux.ifeval-abs}{1.1}
\setval{8b.saw.minus.gdpo-noaux.judge2.avg}{$-$1.6}
\setval{8b.saw.minus.gdpo-noaux.judge2.avg-abs}{1.6}
\setval{8b.saw.minus.gdpo-noaux.judge2.creative-writing-v3}{+7.7}
\setval{8b.saw.minus.gdpo-noaux.judge2.creative-writing-v3-abs}{7.7}
\setval{8b.saw.minus.gdpo-noaux.judge2.longbench-write}{+10.3}
\setval{8b.saw.minus.gdpo-noaux.judge2.longbench-write-abs}{10.3}
\setval{8b.saw.minus.gdpo-noaux.judge2.writing}{+8.3}
\setval{8b.saw.minus.gdpo-noaux.judge2.writing-abs}{8.3}
\setval{8b.saw.minus.gdpo-noaux.judge2.writingbench}{+6.9}
\setval{8b.saw.minus.gdpo-noaux.judge2.writingbench-abs}{6.9}
\setval{8b.saw.minus.gdpo-noaux.livecodebench}{$-$0.7}
\setval{8b.saw.minus.gdpo-noaux.livecodebench-abs}{0.7}
\setval{8b.saw.minus.gdpo-noaux.longbench-write}{+5.1}
\setval{8b.saw.minus.gdpo-noaux.longbench-write-abs}{5.1}
\setval{8b.saw.minus.gdpo-noaux.math}{$-$9.7}
\setval{8b.saw.minus.gdpo-noaux.math-abs}{9.7}
\setval{8b.saw.minus.gdpo-noaux.math500}{$-$8.3}
\setval{8b.saw.minus.gdpo-noaux.math500-abs}{8.3}
\setval{8b.saw.minus.gdpo-noaux.mbpp-plus}{$-$0.5}
\setval{8b.saw.minus.gdpo-noaux.mbpp-plus-abs}{0.5}
\setval{8b.saw.minus.gdpo-noaux.minerva}{$-$2.2}
\setval{8b.saw.minus.gdpo-noaux.minerva-abs}{2.2}
\setval{8b.saw.minus.gdpo-noaux.olympiad-bench}{$-$11.8}
\setval{8b.saw.minus.gdpo-noaux.olympiad-bench-abs}{11.8}
\setval{8b.saw.minus.gdpo-noaux.writing}{+3.8}
\setval{8b.saw.minus.gdpo-noaux.writing-abs}{3.8}
\setval{8b.saw.minus.gdpo-noaux.writingbench}{+0.4}
\setval{8b.saw.minus.gdpo-noaux.writingbench-abs}{0.4}
\setval{8b.saw.minus.gdpo-w0.5.aime24}{$-$4.0}
\setval{8b.saw.minus.gdpo-w0.5.aime24-abs}{4.0}
\setval{8b.saw.minus.gdpo-w0.5.aime25}{+0.1}
\setval{8b.saw.minus.gdpo-w0.5.aime25-abs}{0.1}
\setval{8b.saw.minus.gdpo-w0.5.amc23}{$-$1.2}
\setval{8b.saw.minus.gdpo-w0.5.amc23-abs}{1.2}
\setval{8b.saw.minus.gdpo-w0.5.avg}{$-$0.4}
\setval{8b.saw.minus.gdpo-w0.5.avg-abs}{0.4}
\setval{8b.saw.minus.gdpo-w0.5.code}{+1.2}
\setval{8b.saw.minus.gdpo-w0.5.code-abs}{1.2}
\setval{8b.saw.minus.gdpo-w0.5.creative-writing-v3}{$-$1.9}
\setval{8b.saw.minus.gdpo-w0.5.creative-writing-v3-abs}{1.9}
\setval{8b.saw.minus.gdpo-w0.5.guidebench}{+0.3}
\setval{8b.saw.minus.gdpo-w0.5.guidebench-abs}{0.3}
\setval{8b.saw.minus.gdpo-w0.5.humaneval-plus}{+2.1}
\setval{8b.saw.minus.gdpo-w0.5.humaneval-plus-abs}{2.1}
\setval{8b.saw.minus.gdpo-w0.5.if}{+0.8}
\setval{8b.saw.minus.gdpo-w0.5.if-abs}{0.8}
\setval{8b.saw.minus.gdpo-w0.5.ifbench}{+2.3}
\setval{8b.saw.minus.gdpo-w0.5.ifbench-abs}{2.3}
\setval{8b.saw.minus.gdpo-w0.5.ifeval}{$-$0.1}
\setval{8b.saw.minus.gdpo-w0.5.ifeval-abs}{0.1}
\setval{8b.saw.minus.gdpo-w0.5.judge2.avg}{$-$0.4}
\setval{8b.saw.minus.gdpo-w0.5.judge2.avg-abs}{0.4}
\setval{8b.saw.minus.gdpo-w0.5.judge2.creative-writing-v3}{$-$3.8}
\setval{8b.saw.minus.gdpo-w0.5.judge2.creative-writing-v3-abs}{3.8}
\setval{8b.saw.minus.gdpo-w0.5.judge2.longbench-write}{+1.1}
\setval{8b.saw.minus.gdpo-w0.5.judge2.longbench-write-abs}{1.1}
\setval{8b.saw.minus.gdpo-w0.5.judge2.writing}{$-$0.6}
\setval{8b.saw.minus.gdpo-w0.5.judge2.writing-abs}{0.6}
\setval{8b.saw.minus.gdpo-w0.5.judge2.writingbench}{+0.7}
\setval{8b.saw.minus.gdpo-w0.5.judge2.writingbench-abs}{0.7}
\setval{8b.saw.minus.gdpo-w0.5.livecodebench}{$-$0.1}
\setval{8b.saw.minus.gdpo-w0.5.livecodebench-abs}{0.1}
\setval{8b.saw.minus.gdpo-w0.5.longbench-write}{$-$0.2}
\setval{8b.saw.minus.gdpo-w0.5.longbench-write-abs}{0.2}
\setval{8b.saw.minus.gdpo-w0.5.math}{$-$1.6}
\setval{8b.saw.minus.gdpo-w0.5.math-abs}{1.6}
\setval{8b.saw.minus.gdpo-w0.5.math500}{$-$2.1}
\setval{8b.saw.minus.gdpo-w0.5.math500-abs}{2.1}
\setval{8b.saw.minus.gdpo-w0.5.mbpp-plus}{+1.5}
\setval{8b.saw.minus.gdpo-w0.5.mbpp-plus-abs}{1.5}
\setval{8b.saw.minus.gdpo-w0.5.minerva}{+0.6}
\setval{8b.saw.minus.gdpo-w0.5.minerva-abs}{0.6}
\setval{8b.saw.minus.gdpo-w0.5.olympiad-bench}{$-$3.0}
\setval{8b.saw.minus.gdpo-w0.5.olympiad-bench-abs}{3.0}
\setval{8b.saw.minus.gdpo-w0.5.writing}{$-$0.6}
\setval{8b.saw.minus.gdpo-w0.5.writing-abs}{0.6}
\setval{8b.saw.minus.gdpo-w0.5.writingbench}{+0.4}
\setval{8b.saw.minus.gdpo-w0.5.writingbench-abs}{0.4}
\setval{8b.saw.minus.gdpo.aime24}{$-$6.0}
\setval{8b.saw.minus.gdpo.aime24-abs}{6.0}
\setval{8b.saw.minus.gdpo.aime25}{+1.1}
\setval{8b.saw.minus.gdpo.aime25-abs}{1.1}
\setval{8b.saw.minus.gdpo.amc23}{+0.1}
\setval{8b.saw.minus.gdpo.amc23-abs}{0.1}
\setval{8b.saw.minus.gdpo.avg}{$-$0.4}
\setval{8b.saw.minus.gdpo.avg-abs}{0.4}
\setval{8b.saw.minus.gdpo.code}{+0.7}
\setval{8b.saw.minus.gdpo.code-abs}{0.7}
\setval{8b.saw.minus.gdpo.creative-writing-v3}{$-$3.2}
\setval{8b.saw.minus.gdpo.creative-writing-v3-abs}{3.2}
\setval{8b.saw.minus.gdpo.guidebench}{$-$0.3}
\setval{8b.saw.minus.gdpo.guidebench-abs}{0.3}
\setval{8b.saw.minus.gdpo.humaneval-plus}{+1.8}
\setval{8b.saw.minus.gdpo.humaneval-plus-abs}{1.8}
\setval{8b.saw.minus.gdpo.if}{+0.0}
\setval{8b.saw.minus.gdpo.if-abs}{0.0}
\setval{8b.saw.minus.gdpo.ifbench}{+0.7}
\setval{8b.saw.minus.gdpo.ifbench-abs}{0.7}
\setval{8b.saw.minus.gdpo.ifeval}{$-$0.5}
\setval{8b.saw.minus.gdpo.ifeval-abs}{0.5}
\setval{8b.saw.minus.gdpo.judge2.avg}{$-$0.6}
\setval{8b.saw.minus.gdpo.judge2.avg-abs}{0.6}
\setval{8b.saw.minus.gdpo.judge2.creative-writing-v3}{$-$4.5}
\setval{8b.saw.minus.gdpo.judge2.creative-writing-v3-abs}{4.5}
\setval{8b.saw.minus.gdpo.judge2.longbench-write}{+0.6}
\setval{8b.saw.minus.gdpo.judge2.longbench-write-abs}{0.6}
\setval{8b.saw.minus.gdpo.judge2.writing}{$-$1.8}
\setval{8b.saw.minus.gdpo.judge2.writing-abs}{1.8}
\setval{8b.saw.minus.gdpo.judge2.writingbench}{$-$1.4}
\setval{8b.saw.minus.gdpo.judge2.writingbench-abs}{1.4}
\setval{8b.saw.minus.gdpo.livecodebench}{+0.5}
\setval{8b.saw.minus.gdpo.livecodebench-abs}{0.5}
\setval{8b.saw.minus.gdpo.longbench-write}{+1.5}
\setval{8b.saw.minus.gdpo.longbench-write-abs}{1.5}
\setval{8b.saw.minus.gdpo.math}{$-$0.8}
\setval{8b.saw.minus.gdpo.math-abs}{0.8}
\setval{8b.saw.minus.gdpo.math500}{$-$1.1}
\setval{8b.saw.minus.gdpo.math500-abs}{1.1}
\setval{8b.saw.minus.gdpo.mbpp-plus}{$-$0.2}
\setval{8b.saw.minus.gdpo.mbpp-plus-abs}{0.2}
\setval{8b.saw.minus.gdpo.minerva}{+2.2}
\setval{8b.saw.minus.gdpo.minerva-abs}{2.2}
\setval{8b.saw.minus.gdpo.olympiad-bench}{$-$1.2}
\setval{8b.saw.minus.gdpo.olympiad-bench-abs}{1.2}
\setval{8b.saw.minus.gdpo.writing}{$-$0.8}
\setval{8b.saw.minus.gdpo.writing-abs}{0.8}
\setval{8b.saw.minus.gdpo.writingbench}{$-$0.7}
\setval{8b.saw.minus.gdpo.writingbench-abs}{0.7}
\setval{8b.saw.minus.geoma.aime24}{+1.3}
\setval{8b.saw.minus.geoma.aime24-abs}{1.3}
\setval{8b.saw.minus.geoma.aime25}{+3.0}
\setval{8b.saw.minus.geoma.aime25-abs}{3.0}
\setval{8b.saw.minus.geoma.amc23}{$-$1.2}
\setval{8b.saw.minus.geoma.amc23-abs}{1.2}
\setval{8b.saw.minus.geoma.avg}{+1.3}
\setval{8b.saw.minus.geoma.avg-abs}{1.3}
\setval{8b.saw.minus.geoma.code}{+0.9}
\setval{8b.saw.minus.geoma.code-abs}{0.9}
\setval{8b.saw.minus.geoma.creative-writing-v3}{$-$0.7}
\setval{8b.saw.minus.geoma.creative-writing-v3-abs}{0.7}
\setval{8b.saw.minus.geoma.guidebench}{$-$0.1}
\setval{8b.saw.minus.geoma.guidebench-abs}{0.1}
\setval{8b.saw.minus.geoma.humaneval-plus}{+3.1}
\setval{8b.saw.minus.geoma.humaneval-plus-abs}{3.1}
\setval{8b.saw.minus.geoma.if}{+2.1}
\setval{8b.saw.minus.geoma.if-abs}{2.1}
\setval{8b.saw.minus.geoma.ifbench}{+6.0}
\setval{8b.saw.minus.geoma.ifbench-abs}{6.0}
\setval{8b.saw.minus.geoma.ifeval}{+0.4}
\setval{8b.saw.minus.geoma.ifeval-abs}{0.4}
\setval{8b.saw.minus.geoma.judge2.avg}{+1.1}
\setval{8b.saw.minus.geoma.judge2.avg-abs}{1.1}
\setval{8b.saw.minus.geoma.judge2.creative-writing-v3}{$-$1.0}
\setval{8b.saw.minus.geoma.judge2.creative-writing-v3-abs}{1.0}
\setval{8b.saw.minus.geoma.judge2.longbench-write}{$-$0.1}
\setval{8b.saw.minus.geoma.judge2.longbench-write-abs}{0.1}
\setval{8b.saw.minus.geoma.judge2.writing}{$-$0.8}
\setval{8b.saw.minus.geoma.judge2.writing-abs}{0.8}
\setval{8b.saw.minus.geoma.judge2.writingbench}{$-$1.4}
\setval{8b.saw.minus.geoma.judge2.writingbench-abs}{1.4}
\setval{8b.saw.minus.geoma.livecodebench}{+1.0}
\setval{8b.saw.minus.geoma.livecodebench-abs}{1.0}
\setval{8b.saw.minus.geoma.longbench-write}{+2.1}
\setval{8b.saw.minus.geoma.longbench-write-abs}{2.1}
\setval{8b.saw.minus.geoma.math}{+1.6}
\setval{8b.saw.minus.geoma.math-abs}{1.6}
\setval{8b.saw.minus.geoma.math500}{+2.3}
\setval{8b.saw.minus.geoma.math500-abs}{2.3}
\setval{8b.saw.minus.geoma.mbpp-plus}{$-$1.3}
\setval{8b.saw.minus.geoma.mbpp-plus-abs}{1.3}
\setval{8b.saw.minus.geoma.minerva}{+2.0}
\setval{8b.saw.minus.geoma.minerva-abs}{2.0}
\setval{8b.saw.minus.geoma.olympiad-bench}{+2.1}
\setval{8b.saw.minus.geoma.olympiad-bench-abs}{2.1}
\setval{8b.saw.minus.geoma.writing}{+0.3}
\setval{8b.saw.minus.geoma.writing-abs}{0.3}
\setval{8b.saw.minus.geoma.writingbench}{$-$0.5}
\setval{8b.saw.minus.geoma.writingbench-abs}{0.5}
\setval{8b.saw.minus.grpo-nodiv.aime24}{$-$5.3}
\setval{8b.saw.minus.grpo-nodiv.aime24-abs}{5.3}
\setval{8b.saw.minus.grpo-nodiv.aime25}{+0.7}
\setval{8b.saw.minus.grpo-nodiv.aime25-abs}{0.7}
\setval{8b.saw.minus.grpo-nodiv.amc23}{$-$1.3}
\setval{8b.saw.minus.grpo-nodiv.amc23-abs}{1.3}
\setval{8b.saw.minus.grpo-nodiv.avg}{$-$0.1}
\setval{8b.saw.minus.grpo-nodiv.avg-abs}{0.1}
\setval{8b.saw.minus.grpo-nodiv.code}{$-$0.2}
\setval{8b.saw.minus.grpo-nodiv.code-abs}{0.2}
\setval{8b.saw.minus.grpo-nodiv.creative-writing-v3}{+3.0}
\setval{8b.saw.minus.grpo-nodiv.creative-writing-v3-abs}{3.0}
\setval{8b.saw.minus.grpo-nodiv.guidebench}{$-$0.6}
\setval{8b.saw.minus.grpo-nodiv.guidebench-abs}{0.6}
\setval{8b.saw.minus.grpo-nodiv.humaneval-plus}{$-$0.5}
\setval{8b.saw.minus.grpo-nodiv.humaneval-plus-abs}{0.5}
\setval{8b.saw.minus.grpo-nodiv.if}{+0.4}
\setval{8b.saw.minus.grpo-nodiv.if-abs}{0.4}
\setval{8b.saw.minus.grpo-nodiv.ifbench}{+3.7}
\setval{8b.saw.minus.grpo-nodiv.ifbench-abs}{3.7}
\setval{8b.saw.minus.grpo-nodiv.ifeval}{$-$1.8}
\setval{8b.saw.minus.grpo-nodiv.ifeval-abs}{1.8}
\setval{8b.saw.minus.grpo-nodiv.judge2.avg}{$-$0.6}
\setval{8b.saw.minus.grpo-nodiv.judge2.avg-abs}{0.6}
\setval{8b.saw.minus.grpo-nodiv.judge2.creative-writing-v3}{$-$0.9}
\setval{8b.saw.minus.grpo-nodiv.judge2.creative-writing-v3-abs}{0.9}
\setval{8b.saw.minus.grpo-nodiv.judge2.longbench-write}{$-$1.7}
\setval{8b.saw.minus.grpo-nodiv.judge2.longbench-write-abs}{1.7}
\setval{8b.saw.minus.grpo-nodiv.judge2.writing}{+0.1}
\setval{8b.saw.minus.grpo-nodiv.judge2.writing-abs}{0.1}
\setval{8b.saw.minus.grpo-nodiv.judge2.writingbench}{+3.0}
\setval{8b.saw.minus.grpo-nodiv.judge2.writingbench-abs}{3.0}
\setval{8b.saw.minus.grpo-nodiv.livecodebench}{+1.7}
\setval{8b.saw.minus.grpo-nodiv.livecodebench-abs}{1.7}
\setval{8b.saw.minus.grpo-nodiv.longbench-write}{+0.0}
\setval{8b.saw.minus.grpo-nodiv.longbench-write-abs}{0.0}
\setval{8b.saw.minus.grpo-nodiv.math}{$-$1.8}
\setval{8b.saw.minus.grpo-nodiv.math-abs}{1.8}
\setval{8b.saw.minus.grpo-nodiv.math500}{$-$2.2}
\setval{8b.saw.minus.grpo-nodiv.math500-abs}{2.2}
\setval{8b.saw.minus.grpo-nodiv.mbpp-plus}{$-$1.7}
\setval{8b.saw.minus.grpo-nodiv.mbpp-plus-abs}{1.7}
\setval{8b.saw.minus.grpo-nodiv.minerva}{+1.0}
\setval{8b.saw.minus.grpo-nodiv.minerva-abs}{1.0}
\setval{8b.saw.minus.grpo-nodiv.olympiad-bench}{$-$3.4}
\setval{8b.saw.minus.grpo-nodiv.olympiad-bench-abs}{3.4}
\setval{8b.saw.minus.grpo-nodiv.writing}{+2.7}
\setval{8b.saw.minus.grpo-nodiv.writing-abs}{2.7}
\setval{8b.saw.minus.grpo-nodiv.writingbench}{+5.1}
\setval{8b.saw.minus.grpo-nodiv.writingbench-abs}{5.1}
\setval{8b.saw.minus.grpo.aime24}{$-$0.5}
\setval{8b.saw.minus.grpo.aime24-abs}{0.5}
\setval{8b.saw.minus.grpo.aime25}{+0.0}
\setval{8b.saw.minus.grpo.aime25-abs}{0.0}
\setval{8b.saw.minus.grpo.amc23}{$-$2.1}
\setval{8b.saw.minus.grpo.amc23-abs}{2.1}
\setval{8b.saw.minus.grpo.avg}{+0.1}
\setval{8b.saw.minus.grpo.avg-abs}{0.1}
\setval{8b.saw.minus.grpo.code}{+0.7}
\setval{8b.saw.minus.grpo.code-abs}{0.7}
\setval{8b.saw.minus.grpo.creative-writing-v3}{$-$2.8}
\setval{8b.saw.minus.grpo.creative-writing-v3-abs}{2.8}
\setval{8b.saw.minus.grpo.guidebench}{+2.4}
\setval{8b.saw.minus.grpo.guidebench-abs}{2.4}
\setval{8b.saw.minus.grpo.humaneval-plus}{+2.1}
\setval{8b.saw.minus.grpo.humaneval-plus-abs}{2.1}
\setval{8b.saw.minus.grpo.if}{+0.4}
\setval{8b.saw.minus.grpo.if-abs}{0.4}
\setval{8b.saw.minus.grpo.ifbench}{$-$2.0}
\setval{8b.saw.minus.grpo.ifbench-abs}{2.0}
\setval{8b.saw.minus.grpo.ifeval}{+0.8}
\setval{8b.saw.minus.grpo.ifeval-abs}{0.8}
\setval{8b.saw.minus.grpo.judge2.avg}{+0.2}
\setval{8b.saw.minus.grpo.judge2.avg-abs}{0.2}
\setval{8b.saw.minus.grpo.judge2.creative-writing-v3}{$-$2.7}
\setval{8b.saw.minus.grpo.judge2.creative-writing-v3-abs}{2.7}
\setval{8b.saw.minus.grpo.judge2.longbench-write}{+2.4}
\setval{8b.saw.minus.grpo.judge2.longbench-write-abs}{2.4}
\setval{8b.saw.minus.grpo.judge2.writing}{$-$0.1}
\setval{8b.saw.minus.grpo.judge2.writing-abs}{0.1}
\setval{8b.saw.minus.grpo.judge2.writingbench}{+0.0}
\setval{8b.saw.minus.grpo.judge2.writingbench-abs}{0.0}
\setval{8b.saw.minus.grpo.livecodebench}{+0.2}
\setval{8b.saw.minus.grpo.livecodebench-abs}{0.2}
\setval{8b.saw.minus.grpo.longbench-write}{+1.2}
\setval{8b.saw.minus.grpo.longbench-write-abs}{1.2}
\setval{8b.saw.minus.grpo.math}{+0.1}
\setval{8b.saw.minus.grpo.math-abs}{0.1}
\setval{8b.saw.minus.grpo.math500}{+1.3}
\setval{8b.saw.minus.grpo.math500-abs}{1.3}
\setval{8b.saw.minus.grpo.mbpp-plus}{$-$0.3}
\setval{8b.saw.minus.grpo.mbpp-plus-abs}{0.3}
\setval{8b.saw.minus.grpo.minerva}{+1.9}
\setval{8b.saw.minus.grpo.minerva-abs}{1.9}
\setval{8b.saw.minus.grpo.olympiad-bench}{+0.2}
\setval{8b.saw.minus.grpo.olympiad-bench-abs}{0.2}
\setval{8b.saw.minus.grpo.writing}{$-$0.9}
\setval{8b.saw.minus.grpo.writing-abs}{0.9}
\setval{8b.saw.minus.grpo.writingbench}{$-$1.1}
\setval{8b.saw.minus.grpo.writingbench-abs}{1.1}
\setval{8b.saw.minus.mt-grpo.aime24}{+0.8}
\setval{8b.saw.minus.mt-grpo.aime24-abs}{0.8}
\setval{8b.saw.minus.mt-grpo.aime25}{$-$0.2}
\setval{8b.saw.minus.mt-grpo.aime25-abs}{0.2}
\setval{8b.saw.minus.mt-grpo.amc23}{$-$3.4}
\setval{8b.saw.minus.mt-grpo.amc23-abs}{3.4}
\setval{8b.saw.minus.mt-grpo.avg}{+0.8}
\setval{8b.saw.minus.mt-grpo.avg-abs}{0.8}
\setval{8b.saw.minus.mt-grpo.code}{$-$0.4}
\setval{8b.saw.minus.mt-grpo.code-abs}{0.4}
\setval{8b.saw.minus.mt-grpo.creative-writing-v3}{$-$0.7}
\setval{8b.saw.minus.mt-grpo.creative-writing-v3-abs}{0.7}
\setval{8b.saw.minus.mt-grpo.guidebench}{+1.2}
\setval{8b.saw.minus.mt-grpo.guidebench-abs}{1.2}
\setval{8b.saw.minus.mt-grpo.humaneval-plus}{+0.8}
\setval{8b.saw.minus.mt-grpo.humaneval-plus-abs}{0.8}
\setval{8b.saw.minus.mt-grpo.if}{+1.5}
\setval{8b.saw.minus.mt-grpo.if-abs}{1.5}
\setval{8b.saw.minus.mt-grpo.ifbench}{+2.7}
\setval{8b.saw.minus.mt-grpo.ifbench-abs}{2.7}
\setval{8b.saw.minus.mt-grpo.ifeval}{+0.7}
\setval{8b.saw.minus.mt-grpo.ifeval-abs}{0.7}
\setval{8b.saw.minus.mt-grpo.judge2.avg}{+0.3}
\setval{8b.saw.minus.mt-grpo.judge2.avg-abs}{0.3}
\setval{8b.saw.minus.mt-grpo.judge2.creative-writing-v3}{$-$5.0}
\setval{8b.saw.minus.mt-grpo.judge2.creative-writing-v3-abs}{5.0}
\setval{8b.saw.minus.mt-grpo.judge2.longbench-write}{+2.3}
\setval{8b.saw.minus.mt-grpo.judge2.longbench-write-abs}{2.3}
\setval{8b.saw.minus.mt-grpo.judge2.writing}{$-$0.4}
\setval{8b.saw.minus.mt-grpo.judge2.writing-abs}{0.4}
\setval{8b.saw.minus.mt-grpo.judge2.writingbench}{+1.4}
\setval{8b.saw.minus.mt-grpo.judge2.writingbench-abs}{1.4}
\setval{8b.saw.minus.mt-grpo.livecodebench}{$-$1.0}
\setval{8b.saw.minus.mt-grpo.livecodebench-abs}{1.0}
\setval{8b.saw.minus.mt-grpo.longbench-write}{+3.6}
\setval{8b.saw.minus.mt-grpo.longbench-write-abs}{3.6}
\setval{8b.saw.minus.mt-grpo.math}{+0.3}
\setval{8b.saw.minus.mt-grpo.math-abs}{0.3}
\setval{8b.saw.minus.mt-grpo.math500}{+1.3}
\setval{8b.saw.minus.mt-grpo.math500-abs}{1.3}
\setval{8b.saw.minus.mt-grpo.mbpp-plus}{$-$0.9}
\setval{8b.saw.minus.mt-grpo.mbpp-plus-abs}{0.9}
\setval{8b.saw.minus.mt-grpo.minerva}{+0.7}
\setval{8b.saw.minus.mt-grpo.minerva-abs}{0.7}
\setval{8b.saw.minus.mt-grpo.olympiad-bench}{+2.4}
\setval{8b.saw.minus.mt-grpo.olympiad-bench-abs}{2.4}
\setval{8b.saw.minus.mt-grpo.writing}{+2.2}
\setval{8b.saw.minus.mt-grpo.writing-abs}{2.2}
\setval{8b.saw.minus.mt-grpo.writingbench}{+3.7}
\setval{8b.saw.minus.mt-grpo.writingbench-abs}{3.7}
\setval{8b.saw.minus.papo.aime24}{+3.1}
\setval{8b.saw.minus.papo.aime24-abs}{3.1}
\setval{8b.saw.minus.papo.aime25}{+5.4}
\setval{8b.saw.minus.papo.aime25-abs}{5.4}
\setval{8b.saw.minus.papo.amc23}{+0.2}
\setval{8b.saw.minus.papo.amc23-abs}{0.2}
\setval{8b.saw.minus.papo.avg}{+2.2}
\setval{8b.saw.minus.papo.avg-abs}{2.2}
\setval{8b.saw.minus.papo.code}{+1.4}
\setval{8b.saw.minus.papo.code-abs}{1.4}
\setval{8b.saw.minus.papo.creative-writing-v3}{$-$1.5}
\setval{8b.saw.minus.papo.creative-writing-v3-abs}{1.5}
\setval{8b.saw.minus.papo.guidebench}{$-$0.1}
\setval{8b.saw.minus.papo.guidebench-abs}{0.1}
\setval{8b.saw.minus.papo.humaneval-plus}{+3.4}
\setval{8b.saw.minus.papo.humaneval-plus-abs}{3.4}
\setval{8b.saw.minus.papo.if}{+1.3}
\setval{8b.saw.minus.papo.if-abs}{1.3}
\setval{8b.saw.minus.papo.ifbench}{+2.7}
\setval{8b.saw.minus.papo.ifbench-abs}{2.7}
\setval{8b.saw.minus.papo.ifeval}{+1.3}
\setval{8b.saw.minus.papo.ifeval-abs}{1.3}
\setval{8b.saw.minus.papo.judge2.avg}{+1.8}
\setval{8b.saw.minus.papo.judge2.avg-abs}{1.8}
\setval{8b.saw.minus.papo.judge2.creative-writing-v3}{$-$5.3}
\setval{8b.saw.minus.papo.judge2.creative-writing-v3-abs}{5.3}
\setval{8b.saw.minus.papo.judge2.longbench-write}{+5.3}
\setval{8b.saw.minus.papo.judge2.longbench-write-abs}{5.3}
\setval{8b.saw.minus.papo.judge2.writing}{+0.2}
\setval{8b.saw.minus.papo.judge2.writing-abs}{0.2}
\setval{8b.saw.minus.papo.judge2.writingbench}{+0.6}
\setval{8b.saw.minus.papo.judge2.writingbench-abs}{0.6}
\setval{8b.saw.minus.papo.livecodebench}{+0.1}
\setval{8b.saw.minus.papo.livecodebench-abs}{0.1}
\setval{8b.saw.minus.papo.longbench-write}{+5.5}
\setval{8b.saw.minus.papo.longbench-write-abs}{5.5}
\setval{8b.saw.minus.papo.math}{+3.0}
\setval{8b.saw.minus.papo.math-abs}{3.0}
\setval{8b.saw.minus.papo.math500}{+3.4}
\setval{8b.saw.minus.papo.math500-abs}{3.4}
\setval{8b.saw.minus.papo.mbpp-plus}{+0.9}
\setval{8b.saw.minus.papo.mbpp-plus-abs}{0.9}
\setval{8b.saw.minus.papo.minerva}{+2.4}
\setval{8b.saw.minus.papo.minerva-abs}{2.4}
\setval{8b.saw.minus.papo.olympiad-bench}{+3.3}
\setval{8b.saw.minus.papo.olympiad-bench-abs}{3.3}
\setval{8b.saw.minus.papo.writing}{+2.1}
\setval{8b.saw.minus.papo.writing-abs}{2.1}
\setval{8b.saw.minus.papo.writingbench}{+2.2}
\setval{8b.saw.minus.papo.writingbench-abs}{2.2}
\setval{8b.saw.passing-at-ceiling.code.early}{0.71}
\setval{8b.saw.passing-at-ceiling.code.late}{0.75}
\setval{8b.saw.passing-at-ceiling.code.run}{0.74}
\setval{8b.saw.passing-at-ceiling.if.early}{0.33}
\setval{8b.saw.passing-at-ceiling.if.late}{0.58}
\setval{8b.saw.passing-at-ceiling.if.run}{0.50}
\setval{8b.saw.passing-at-ceiling.math.early}{1.00}
\setval{8b.saw.passing-at-ceiling.math.late}{1.00}
\setval{8b.saw.passing-at-ceiling.math.run}{1.00}
\setval{8b.saw.passing-at-ceiling.writing.early}{0.00}
\setval{8b.saw.passing-at-ceiling.writing.late}{0.01}
\setval{8b.saw.passing-at-ceiling.writing.run}{0.00}
\setval{8b.saw.primary-cosine.code.early}{0.95}
\setval{8b.saw.primary-cosine.code.late}{0.95}
\setval{8b.saw.primary-cosine.code.run}{0.96}
\setval{8b.saw.primary-cosine.if.early}{0.88}
\setval{8b.saw.primary-cosine.if.late}{0.76}
\setval{8b.saw.primary-cosine.if.run}{0.82}
\setval{8b.saw.primary-cosine.math.early}{0.96}
\setval{8b.saw.primary-cosine.math.late}{0.97}
\setval{8b.saw.primary-cosine.math.run}{0.97}
\setval{8b.saw.primary-cosine.writing.early}{0.83}
\setval{8b.saw.primary-cosine.writing.late}{0.78}
\setval{8b.saw.primary-cosine.writing.run}{0.75}
\setval{8b.saw.primary-level.code.early}{0.54}
\setval{8b.saw.primary-level.code.late}{0.77}
\setval{8b.saw.primary-level.code.run}{0.72}
\setval{8b.saw.primary-level.if.early}{0.36}
\setval{8b.saw.primary-level.if.late}{0.77}
\setval{8b.saw.primary-level.if.run}{0.67}
\setval{8b.saw.primary-level.math.early}{0.49}
\setval{8b.saw.primary-level.math.late}{0.71}
\setval{8b.saw.primary-level.math.run}{0.64}
\setval{8b.saw.primary-level.writing.early}{0.54}
\setval{8b.saw.primary-level.writing.late}{0.76}
\setval{8b.saw.primary-level.writing.run}{0.72}
\setval{8b.saw.solved.code.early}{0.07}
\setval{8b.saw.solved.code.late}{0.02}
\setval{8b.saw.solved.code.run}{0.03}
\setval{8b.saw.solved.if.early}{0.03}
\setval{8b.saw.solved.if.late}{0.20}
\setval{8b.saw.solved.if.run}{0.14}
\setval{8b.saw.solved.math.early}{0.15}
\setval{8b.saw.solved.math.late}{0.17}
\setval{8b.saw.solved.math.run}{0.15}
\setval{8b.saw.solved.writing.early}{0.00}
\setval{8b.saw.solved.writing.late}{0.00}
\setval{8b.saw.solved.writing.run}{0.00}
\setval{8b.saw.update-share.code.early}{0.19}
\setval{8b.saw.update-share.code.late}{0.22}
\setval{8b.saw.update-share.code.run}{0.22}
\setval{8b.saw.update-share.if.early}{0.23}
\setval{8b.saw.update-share.if.late}{0.16}
\setval{8b.saw.update-share.if.run}{0.20}
\setval{8b.saw.update-share.math.early}{0.18}
\setval{8b.saw.update-share.math.late}{0.27}
\setval{8b.saw.update-share.math.run}{0.25}
\setval{8b.saw.update-share.writing.early}{0.40}
\setval{8b.saw.update-share.writing.late}{0.35}
\setval{8b.saw.update-share.writing.run}{0.33}
\setval{8b.start.aux-level.code}{0.81}
\setval{8b.start.aux-level.if}{0.79}
\setval{8b.start.aux-level.math}{0.84}
\setval{8b.start.aux-level.writing}{0.84}
\setval{8b.start.aux-level.writing-rm}{0.64}
\setval{8b.train.step}{300}
\setval{eval.count}{15}
\setval{eval.count.code}{3}
\setval{eval.count.if}{3}
\setval{eval.count.math}{6}
\setval{eval.count.writing}{3}
\setval{eval.judge.creative-writing-v3.max-new-tokens}{4,096}
\setval{eval.judge.creative-writing-v3.temperature}{0}
\setval{eval.judge.longbench-write.max-new-tokens}{1,024}
\setval{eval.judge.longbench-write.temperature}{0.5}
\setval{eval.judge.writingbench.max-new-tokens}{2,048}
\setval{eval.judge.writingbench.temperature}{1.0}
\setval{eval.judge.writingbench.top-p}{0.95}
\setval{eval.problems.max}{1,042}
\setval{eval.problems.min}{30}
\setval{olmo.base.bench.aime24}{8.5}
\setval{olmo.base.bench.aime25}{7.4}
\setval{olmo.base.bench.amc23}{44.5}
\setval{olmo.base.bench.creative-writing-v3}{37.9}
\setval{olmo.base.bench.guidebench}{61.4}
\setval{olmo.base.bench.humaneval-plus}{69.2}
\setval{olmo.base.bench.ifbench}{26.7}
\setval{olmo.base.bench.ifeval}{84.4}
\setval{olmo.base.bench.livecodebench}{16.9}
\setval{olmo.base.bench.longbench-write}{57.5}
\setval{olmo.base.bench.math500}{70.1}
\setval{olmo.base.bench.mbpp-plus}{61.2}
\setval{olmo.base.bench.minerva}{35.3}
\setval{olmo.base.bench.olympiad-bench}{36.3}
\setval{olmo.base.bench.writingbench}{44.7}
\setval{olmo.base.eval.avg}{44.1}
\setval{olmo.base.eval.code}{49.1}
\setval{olmo.base.eval.if}{57.5}
\setval{olmo.base.eval.math}{33.7}
\setval{olmo.base.eval.writing}{46.7}
\setval{olmo.base.judge2.avg}{43.0}
\setval{olmo.base.judge2.creative-writing-v3}{31.7}
\setval{olmo.base.judge2.longbench-write}{53.2}
\setval{olmo.base.judge2.writing}{40.8}
\setval{olmo.base.judge2.writingbench}{37.7}
\setval{olmo.base.minus.concur-free.aime24}{$-$18.2}
\setval{olmo.base.minus.concur-free.aime24-abs}{18.2}
\setval{olmo.base.minus.concur-free.aime25}{$-$14.7}
\setval{olmo.base.minus.concur-free.aime25-abs}{14.7}
\setval{olmo.base.minus.concur-free.amc23}{$-$23.0}
\setval{olmo.base.minus.concur-free.amc23-abs}{23.0}
\setval{olmo.base.minus.concur-free.avg}{$-$11.3}
\setval{olmo.base.minus.concur-free.avg-abs}{11.3}
\setval{olmo.base.minus.concur-free.code}{$-$6.5}
\setval{olmo.base.minus.concur-free.code-abs}{6.5}
\setval{olmo.base.minus.concur-free.creative-writing-v3}{$-$11.9}
\setval{olmo.base.minus.concur-free.creative-writing-v3-abs}{11.9}
\setval{olmo.base.minus.concur-free.guidebench}{+1.2}
\setval{olmo.base.minus.concur-free.guidebench-abs}{1.2}
\setval{olmo.base.minus.concur-free.humaneval-plus}{$-$7.4}
\setval{olmo.base.minus.concur-free.humaneval-plus-abs}{7.4}
\setval{olmo.base.minus.concur-free.if}{$-$2.2}
\setval{olmo.base.minus.concur-free.if-abs}{2.2}
\setval{olmo.base.minus.concur-free.ifbench}{$-$5.0}
\setval{olmo.base.minus.concur-free.ifbench-abs}{5.0}
\setval{olmo.base.minus.concur-free.ifeval}{$-$2.9}
\setval{olmo.base.minus.concur-free.ifeval-abs}{2.9}
\setval{olmo.base.minus.concur-free.judge2.avg}{$-$9.7}
\setval{olmo.base.minus.concur-free.judge2.avg-abs}{9.7}
\setval{olmo.base.minus.concur-free.judge2.creative-writing-v3}{$-$5.9}
\setval{olmo.base.minus.concur-free.judge2.creative-writing-v3-abs}{5.9}
\setval{olmo.base.minus.concur-free.judge2.longbench-write}{$-$10.7}
\setval{olmo.base.minus.concur-free.judge2.longbench-write-abs}{10.7}
\setval{olmo.base.minus.concur-free.judge2.writing}{$-$9.0}
\setval{olmo.base.minus.concur-free.judge2.writing-abs}{9.0}
\setval{olmo.base.minus.concur-free.judge2.writingbench}{$-$10.4}
\setval{olmo.base.minus.concur-free.judge2.writingbench-abs}{10.4}
\setval{olmo.base.minus.concur-free.livecodebench}{$-$8.8}
\setval{olmo.base.minus.concur-free.livecodebench-abs}{8.8}
\setval{olmo.base.minus.concur-free.longbench-write}{$-$14.6}
\setval{olmo.base.minus.concur-free.longbench-write-abs}{14.6}
\setval{olmo.base.minus.concur-free.math}{$-$15.3}
\setval{olmo.base.minus.concur-free.math-abs}{15.3}
\setval{olmo.base.minus.concur-free.math500}{$-$14.2}
\setval{olmo.base.minus.concur-free.math500-abs}{14.2}
\setval{olmo.base.minus.concur-free.mbpp-plus}{$-$3.3}
\setval{olmo.base.minus.concur-free.mbpp-plus-abs}{3.3}
\setval{olmo.base.minus.concur-free.minerva}{$-$4.6}
\setval{olmo.base.minus.concur-free.minerva-abs}{4.6}
\setval{olmo.base.minus.concur-free.olympiad-bench}{$-$17.0}
\setval{olmo.base.minus.concur-free.olympiad-bench-abs}{17.0}
\setval{olmo.base.minus.concur-free.writing}{$-$17.3}
\setval{olmo.base.minus.concur-free.writing-abs}{17.3}
\setval{olmo.base.minus.concur-free.writingbench}{$-$25.5}
\setval{olmo.base.minus.concur-free.writingbench-abs}{25.5}
\setval{olmo.base.minus.concur-tied.aime24}{$-$25.9}
\setval{olmo.base.minus.concur-tied.aime24-abs}{25.9}
\setval{olmo.base.minus.concur-tied.aime25}{$-$22.5}
\setval{olmo.base.minus.concur-tied.aime25-abs}{22.5}
\setval{olmo.base.minus.concur-tied.amc23}{$-$35.3}
\setval{olmo.base.minus.concur-tied.amc23-abs}{35.3}
\setval{olmo.base.minus.concur-tied.avg}{$-$15.0}
\setval{olmo.base.minus.concur-tied.avg-abs}{15.0}
\setval{olmo.base.minus.concur-tied.code}{$-$12.0}
\setval{olmo.base.minus.concur-tied.code-abs}{12.0}
\setval{olmo.base.minus.concur-tied.creative-writing-v3}{$-$11.1}
\setval{olmo.base.minus.concur-tied.creative-writing-v3-abs}{11.1}
\setval{olmo.base.minus.concur-tied.guidebench}{+2.9}
\setval{olmo.base.minus.concur-tied.guidebench-abs}{2.9}
\setval{olmo.base.minus.concur-tied.humaneval-plus}{$-$14.1}
\setval{olmo.base.minus.concur-tied.humaneval-plus-abs}{14.1}
\setval{olmo.base.minus.concur-tied.if}{$-$2.5}
\setval{olmo.base.minus.concur-tied.if-abs}{2.5}
\setval{olmo.base.minus.concur-tied.ifbench}{$-$7.0}
\setval{olmo.base.minus.concur-tied.ifbench-abs}{7.0}
\setval{olmo.base.minus.concur-tied.ifeval}{$-$3.2}
\setval{olmo.base.minus.concur-tied.ifeval-abs}{3.2}
\setval{olmo.base.minus.concur-tied.judge2.avg}{$-$13.4}
\setval{olmo.base.minus.concur-tied.judge2.avg-abs}{13.4}
\setval{olmo.base.minus.concur-tied.judge2.creative-writing-v3}{$-$5.5}
\setval{olmo.base.minus.concur-tied.judge2.creative-writing-v3-abs}{5.5}
\setval{olmo.base.minus.concur-tied.judge2.longbench-write}{$-$7.3}
\setval{olmo.base.minus.concur-tied.judge2.longbench-write-abs}{7.3}
\setval{olmo.base.minus.concur-tied.judge2.writing}{$-$6.9}
\setval{olmo.base.minus.concur-tied.judge2.writing-abs}{6.9}
\setval{olmo.base.minus.concur-tied.judge2.writingbench}{$-$7.9}
\setval{olmo.base.minus.concur-tied.judge2.writingbench-abs}{7.9}
\setval{olmo.base.minus.concur-tied.livecodebench}{$-$12.2}
\setval{olmo.base.minus.concur-tied.livecodebench-abs}{12.2}
\setval{olmo.base.minus.concur-tied.longbench-write}{$-$11.8}
\setval{olmo.base.minus.concur-tied.longbench-write-abs}{11.8}
\setval{olmo.base.minus.concur-tied.math}{$-$22.8}
\setval{olmo.base.minus.concur-tied.math-abs}{22.8}
\setval{olmo.base.minus.concur-tied.math500}{$-$19.3}
\setval{olmo.base.minus.concur-tied.math500-abs}{19.3}
\setval{olmo.base.minus.concur-tied.mbpp-plus}{$-$9.8}
\setval{olmo.base.minus.concur-tied.mbpp-plus-abs}{9.8}
\setval{olmo.base.minus.concur-tied.minerva}{$-$8.8}
\setval{olmo.base.minus.concur-tied.minerva-abs}{8.8}
\setval{olmo.base.minus.concur-tied.olympiad-bench}{$-$25.3}
\setval{olmo.base.minus.concur-tied.olympiad-bench-abs}{25.3}
\setval{olmo.base.minus.concur-tied.writing}{$-$14.8}
\setval{olmo.base.minus.concur-tied.writing-abs}{14.8}
\setval{olmo.base.minus.concur-tied.writingbench}{$-$21.4}
\setval{olmo.base.minus.concur-tied.writingbench-abs}{21.4}
\setval{olmo.base.minus.concur.aime24}{$-$27.8}
\setval{olmo.base.minus.concur.aime24-abs}{27.8}
\setval{olmo.base.minus.concur.aime25}{$-$24.8}
\setval{olmo.base.minus.concur.aime25-abs}{24.8}
\setval{olmo.base.minus.concur.amc23}{$-$39.0}
\setval{olmo.base.minus.concur.amc23-abs}{39.0}
\setval{olmo.base.minus.concur.avg}{$-$15.2}
\setval{olmo.base.minus.concur.avg-abs}{15.2}
\setval{olmo.base.minus.concur.code}{$-$11.1}
\setval{olmo.base.minus.concur.code-abs}{11.1}
\setval{olmo.base.minus.concur.creative-writing-v3}{$-$11.3}
\setval{olmo.base.minus.concur.creative-writing-v3-abs}{11.3}
\setval{olmo.base.minus.concur.guidebench}{+4.5}
\setval{olmo.base.minus.concur.guidebench-abs}{4.5}
\setval{olmo.base.minus.concur.humaneval-plus}{$-$13.5}
\setval{olmo.base.minus.concur.humaneval-plus-abs}{13.5}
\setval{olmo.base.minus.concur.if}{$-$1.4}
\setval{olmo.base.minus.concur.if-abs}{1.4}
\setval{olmo.base.minus.concur.ifbench}{$-$5.3}
\setval{olmo.base.minus.concur.ifbench-abs}{5.3}
\setval{olmo.base.minus.concur.ifeval}{$-$3.4}
\setval{olmo.base.minus.concur.ifeval-abs}{3.4}
\setval{olmo.base.minus.concur.judge2.avg}{$-$13.6}
\setval{olmo.base.minus.concur.judge2.avg-abs}{13.6}
\setval{olmo.base.minus.concur.judge2.creative-writing-v3}{$-$6.8}
\setval{olmo.base.minus.concur.judge2.creative-writing-v3-abs}{6.8}
\setval{olmo.base.minus.concur.judge2.longbench-write}{$-$5.2}
\setval{olmo.base.minus.concur.judge2.longbench-write-abs}{5.2}
\setval{olmo.base.minus.concur.judge2.writing}{$-$6.4}
\setval{olmo.base.minus.concur.judge2.writing-abs}{6.4}
\setval{olmo.base.minus.concur.judge2.writingbench}{$-$7.2}
\setval{olmo.base.minus.concur.judge2.writingbench-abs}{7.2}
\setval{olmo.base.minus.concur.livecodebench}{$-$13.5}
\setval{olmo.base.minus.concur.livecodebench-abs}{13.5}
\setval{olmo.base.minus.concur.longbench-write}{$-$11.0}
\setval{olmo.base.minus.concur.longbench-write-abs}{11.0}
\setval{olmo.base.minus.concur.math}{$-$24.6}
\setval{olmo.base.minus.concur.math-abs}{24.6}
\setval{olmo.base.minus.concur.math500}{$-$20.4}
\setval{olmo.base.minus.concur.math500-abs}{20.4}
\setval{olmo.base.minus.concur.mbpp-plus}{$-$6.3}
\setval{olmo.base.minus.concur.mbpp-plus-abs}{6.3}
\setval{olmo.base.minus.concur.minerva}{$-$9.5}
\setval{olmo.base.minus.concur.minerva-abs}{9.5}
\setval{olmo.base.minus.concur.olympiad-bench}{$-$26.1}
\setval{olmo.base.minus.concur.olympiad-bench-abs}{26.1}
\setval{olmo.base.minus.concur.writing}{$-$14.3}
\setval{olmo.base.minus.concur.writing-abs}{14.3}
\setval{olmo.base.minus.concur.writingbench}{$-$20.7}
\setval{olmo.base.minus.concur.writingbench-abs}{20.7}
\setval{olmo.base.minus.gd2po.aime24}{$-$19.8}
\setval{olmo.base.minus.gd2po.aime24-abs}{19.8}
\setval{olmo.base.minus.gd2po.aime25}{$-$16.9}
\setval{olmo.base.minus.gd2po.aime25-abs}{16.9}
\setval{olmo.base.minus.gd2po.amc23}{$-$27.7}
\setval{olmo.base.minus.gd2po.amc23-abs}{27.7}
\setval{olmo.base.minus.gd2po.avg}{$-$11.5}
\setval{olmo.base.minus.gd2po.avg-abs}{11.5}
\setval{olmo.base.minus.gd2po.code}{$-$5.8}
\setval{olmo.base.minus.gd2po.code-abs}{5.8}
\setval{olmo.base.minus.gd2po.creative-writing-v3}{$-$10.4}
\setval{olmo.base.minus.gd2po.creative-writing-v3-abs}{10.4}
\setval{olmo.base.minus.gd2po.guidebench}{+0.3}
\setval{olmo.base.minus.gd2po.guidebench-abs}{0.3}
\setval{olmo.base.minus.gd2po.humaneval-plus}{$-$7.6}
\setval{olmo.base.minus.gd2po.humaneval-plus-abs}{7.6}
\setval{olmo.base.minus.gd2po.if}{$-$2.9}
\setval{olmo.base.minus.gd2po.if-abs}{2.9}
\setval{olmo.base.minus.gd2po.ifbench}{$-$5.7}
\setval{olmo.base.minus.gd2po.ifbench-abs}{5.7}
\setval{olmo.base.minus.gd2po.ifeval}{$-$3.2}
\setval{olmo.base.minus.gd2po.ifeval-abs}{3.2}
\setval{olmo.base.minus.gd2po.judge2.avg}{$-$10.3}
\setval{olmo.base.minus.gd2po.judge2.avg-abs}{10.3}
\setval{olmo.base.minus.gd2po.judge2.creative-writing-v3}{$-$6.2}
\setval{olmo.base.minus.gd2po.judge2.creative-writing-v3-abs}{6.2}
\setval{olmo.base.minus.gd2po.judge2.longbench-write}{$-$6.7}
\setval{olmo.base.minus.gd2po.judge2.longbench-write-abs}{6.7}
\setval{olmo.base.minus.gd2po.judge2.writing}{$-$7.9}
\setval{olmo.base.minus.gd2po.judge2.writing-abs}{7.9}
\setval{olmo.base.minus.gd2po.judge2.writingbench}{$-$10.7}
\setval{olmo.base.minus.gd2po.judge2.writingbench-abs}{10.7}
\setval{olmo.base.minus.gd2po.livecodebench}{$-$6.3}
\setval{olmo.base.minus.gd2po.livecodebench-abs}{6.3}
\setval{olmo.base.minus.gd2po.longbench-write}{$-$10.7}
\setval{olmo.base.minus.gd2po.longbench-write-abs}{10.7}
\setval{olmo.base.minus.gd2po.math}{$-$17.4}
\setval{olmo.base.minus.gd2po.math-abs}{17.4}
\setval{olmo.base.minus.gd2po.math500}{$-$15.6}
\setval{olmo.base.minus.gd2po.math500-abs}{15.6}
\setval{olmo.base.minus.gd2po.mbpp-plus}{$-$3.5}
\setval{olmo.base.minus.gd2po.mbpp-plus-abs}{3.5}
\setval{olmo.base.minus.gd2po.minerva}{$-$4.8}
\setval{olmo.base.minus.gd2po.minerva-abs}{4.8}
\setval{olmo.base.minus.gd2po.olympiad-bench}{$-$19.8}
\setval{olmo.base.minus.gd2po.olympiad-bench-abs}{19.8}
\setval{olmo.base.minus.gd2po.writing}{$-$14.0}
\setval{olmo.base.minus.gd2po.writing-abs}{14.0}
\setval{olmo.base.minus.gd2po.writingbench}{$-$21.0}
\setval{olmo.base.minus.gd2po.writingbench-abs}{21.0}
\setval{olmo.base.minus.gdpo-noaux.aime24}{$-$29.8}
\setval{olmo.base.minus.gdpo-noaux.aime24-abs}{29.8}
\setval{olmo.base.minus.gdpo-noaux.aime25}{$-$26.3}
\setval{olmo.base.minus.gdpo-noaux.aime25-abs}{26.3}
\setval{olmo.base.minus.gdpo-noaux.amc23}{$-$39.4}
\setval{olmo.base.minus.gdpo-noaux.amc23-abs}{39.4}
\setval{olmo.base.minus.gdpo-noaux.avg}{$-$13.7}
\setval{olmo.base.minus.gdpo-noaux.avg-abs}{13.7}
\setval{olmo.base.minus.gdpo-noaux.code}{$-$13.3}
\setval{olmo.base.minus.gdpo-noaux.code-abs}{13.3}
\setval{olmo.base.minus.gdpo-noaux.creative-writing-v3}{$-$0.3}
\setval{olmo.base.minus.gdpo-noaux.creative-writing-v3-abs}{0.3}
\setval{olmo.base.minus.gdpo-noaux.guidebench}{+5.7}
\setval{olmo.base.minus.gdpo-noaux.guidebench-abs}{5.7}
\setval{olmo.base.minus.gdpo-noaux.humaneval-plus}{$-$16.0}
\setval{olmo.base.minus.gdpo-noaux.humaneval-plus-abs}{16.0}
\setval{olmo.base.minus.gdpo-noaux.if}{$-$0.5}
\setval{olmo.base.minus.gdpo-noaux.if-abs}{0.5}
\setval{olmo.base.minus.gdpo-noaux.ifbench}{$-$6.7}
\setval{olmo.base.minus.gdpo-noaux.ifbench-abs}{6.7}
\setval{olmo.base.minus.gdpo-noaux.ifeval}{$-$0.4}
\setval{olmo.base.minus.gdpo-noaux.ifeval-abs}{0.4}
\setval{olmo.base.minus.gdpo-noaux.judge2.avg}{$-$12.7}
\setval{olmo.base.minus.gdpo-noaux.judge2.avg-abs}{12.7}
\setval{olmo.base.minus.gdpo-noaux.judge2.creative-writing-v3}{+0.7}
\setval{olmo.base.minus.gdpo-noaux.judge2.creative-writing-v3-abs}{0.7}
\setval{olmo.base.minus.gdpo-noaux.judge2.longbench-write}{+6.5}
\setval{olmo.base.minus.gdpo-noaux.judge2.longbench-write-abs}{6.5}
\setval{olmo.base.minus.gdpo-noaux.judge2.writing}{+1.7}
\setval{olmo.base.minus.gdpo-noaux.judge2.writing-abs}{1.7}
\setval{olmo.base.minus.gdpo-noaux.judge2.writingbench}{$-$2.1}
\setval{olmo.base.minus.gdpo-noaux.judge2.writingbench-abs}{2.1}
\setval{olmo.base.minus.gdpo-noaux.livecodebench}{$-$14.7}
\setval{olmo.base.minus.gdpo-noaux.livecodebench-abs}{14.7}
\setval{olmo.base.minus.gdpo-noaux.longbench-write}{+1.5}
\setval{olmo.base.minus.gdpo-noaux.longbench-write-abs}{1.5}
\setval{olmo.base.minus.gdpo-noaux.math}{$-$25.8}
\setval{olmo.base.minus.gdpo-noaux.math-abs}{25.8}
\setval{olmo.base.minus.gdpo-noaux.math500}{$-$20.8}
\setval{olmo.base.minus.gdpo-noaux.math500-abs}{20.8}
\setval{olmo.base.minus.gdpo-noaux.mbpp-plus}{$-$9.3}
\setval{olmo.base.minus.gdpo-noaux.mbpp-plus-abs}{9.3}
\setval{olmo.base.minus.gdpo-noaux.minerva}{$-$10.9}
\setval{olmo.base.minus.gdpo-noaux.minerva-abs}{10.9}
\setval{olmo.base.minus.gdpo-noaux.olympiad-bench}{$-$27.5}
\setval{olmo.base.minus.gdpo-noaux.olympiad-bench-abs}{27.5}
\setval{olmo.base.minus.gdpo-noaux.writing}{$-$3.2}
\setval{olmo.base.minus.gdpo-noaux.writing-abs}{3.2}
\setval{olmo.base.minus.gdpo-noaux.writingbench}{$-$10.7}
\setval{olmo.base.minus.gdpo-noaux.writingbench-abs}{10.7}
\setval{olmo.base.minus.gdpo-w0.5.aime24}{$-$21.0}
\setval{olmo.base.minus.gdpo-w0.5.aime24-abs}{21.0}
\setval{olmo.base.minus.gdpo-w0.5.aime25}{$-$19.9}
\setval{olmo.base.minus.gdpo-w0.5.aime25-abs}{19.9}
\setval{olmo.base.minus.gdpo-w0.5.amc23}{$-$29.5}
\setval{olmo.base.minus.gdpo-w0.5.amc23-abs}{29.5}
\setval{olmo.base.minus.gdpo-w0.5.avg}{$-$11.5}
\setval{olmo.base.minus.gdpo-w0.5.avg-abs}{11.5}
\setval{olmo.base.minus.gdpo-w0.5.code}{$-$6.7}
\setval{olmo.base.minus.gdpo-w0.5.code-abs}{6.7}
\setval{olmo.base.minus.gdpo-w0.5.creative-writing-v3}{$-$10.2}
\setval{olmo.base.minus.gdpo-w0.5.creative-writing-v3-abs}{10.2}
\setval{olmo.base.minus.gdpo-w0.5.guidebench}{+9.5}
\setval{olmo.base.minus.gdpo-w0.5.guidebench-abs}{9.5}
\setval{olmo.base.minus.gdpo-w0.5.humaneval-plus}{$-$8.5}
\setval{olmo.base.minus.gdpo-w0.5.humaneval-plus-abs}{8.5}
\setval{olmo.base.minus.gdpo-w0.5.if}{+0.1}
\setval{olmo.base.minus.gdpo-w0.5.if-abs}{0.1}
\setval{olmo.base.minus.gdpo-w0.5.ifbench}{$-$6.0}
\setval{olmo.base.minus.gdpo-w0.5.ifbench-abs}{6.0}
\setval{olmo.base.minus.gdpo-w0.5.ifeval}{$-$3.1}
\setval{olmo.base.minus.gdpo-w0.5.ifeval-abs}{3.1}
\setval{olmo.base.minus.gdpo-w0.5.judge2.avg}{$-$10.3}
\setval{olmo.base.minus.gdpo-w0.5.judge2.avg-abs}{10.3}
\setval{olmo.base.minus.gdpo-w0.5.judge2.creative-writing-v3}{$-$7.8}
\setval{olmo.base.minus.gdpo-w0.5.judge2.creative-writing-v3-abs}{7.8}
\setval{olmo.base.minus.gdpo-w0.5.judge2.longbench-write}{$-$3.4}
\setval{olmo.base.minus.gdpo-w0.5.judge2.longbench-write-abs}{3.4}
\setval{olmo.base.minus.gdpo-w0.5.judge2.writing}{$-$6.7}
\setval{olmo.base.minus.gdpo-w0.5.judge2.writing-abs}{6.7}
\setval{olmo.base.minus.gdpo-w0.5.judge2.writingbench}{$-$9.0}
\setval{olmo.base.minus.gdpo-w0.5.judge2.writingbench-abs}{9.0}
\setval{olmo.base.minus.gdpo-w0.5.livecodebench}{$-$7.9}
\setval{olmo.base.minus.gdpo-w0.5.livecodebench-abs}{7.9}
\setval{olmo.base.minus.gdpo-w0.5.longbench-write}{$-$6.3}
\setval{olmo.base.minus.gdpo-w0.5.longbench-write-abs}{6.3}
\setval{olmo.base.minus.gdpo-w0.5.math}{$-$19.1}
\setval{olmo.base.minus.gdpo-w0.5.math-abs}{19.1}
\setval{olmo.base.minus.gdpo-w0.5.math500}{$-$16.9}
\setval{olmo.base.minus.gdpo-w0.5.math500-abs}{16.9}
\setval{olmo.base.minus.gdpo-w0.5.mbpp-plus}{$-$3.7}
\setval{olmo.base.minus.gdpo-w0.5.mbpp-plus-abs}{3.7}
\setval{olmo.base.minus.gdpo-w0.5.minerva}{$-$6.2}
\setval{olmo.base.minus.gdpo-w0.5.minerva-abs}{6.2}
\setval{olmo.base.minus.gdpo-w0.5.olympiad-bench}{$-$21.2}
\setval{olmo.base.minus.gdpo-w0.5.olympiad-bench-abs}{21.2}
\setval{olmo.base.minus.gdpo-w0.5.writing}{$-$12.7}
\setval{olmo.base.minus.gdpo-w0.5.writing-abs}{12.7}
\setval{olmo.base.minus.gdpo-w0.5.writingbench}{$-$21.8}
\setval{olmo.base.minus.gdpo-w0.5.writingbench-abs}{21.8}
\setval{olmo.base.minus.gdpo.aime24}{$-$19.3}
\setval{olmo.base.minus.gdpo.aime24-abs}{19.3}
\setval{olmo.base.minus.gdpo.aime25}{$-$15.2}
\setval{olmo.base.minus.gdpo.aime25-abs}{15.2}
\setval{olmo.base.minus.gdpo.amc23}{$-$25.0}
\setval{olmo.base.minus.gdpo.amc23-abs}{25.0}
\setval{olmo.base.minus.gdpo.avg}{$-$9.8}
\setval{olmo.base.minus.gdpo.avg-abs}{9.8}
\setval{olmo.base.minus.gdpo.code}{$-$5.6}
\setval{olmo.base.minus.gdpo.code-abs}{5.6}
\setval{olmo.base.minus.gdpo.creative-writing-v3}{$-$9.2}
\setval{olmo.base.minus.gdpo.creative-writing-v3-abs}{9.2}
\setval{olmo.base.minus.gdpo.guidebench}{+10.6}
\setval{olmo.base.minus.gdpo.guidebench-abs}{10.6}
\setval{olmo.base.minus.gdpo.humaneval-plus}{$-$7.1}
\setval{olmo.base.minus.gdpo.humaneval-plus-abs}{7.1}
\setval{olmo.base.minus.gdpo.if}{+1.8}
\setval{olmo.base.minus.gdpo.if-abs}{1.8}
\setval{olmo.base.minus.gdpo.ifbench}{$-$3.0}
\setval{olmo.base.minus.gdpo.ifbench-abs}{3.0}
\setval{olmo.base.minus.gdpo.ifeval}{$-$2.0}
\setval{olmo.base.minus.gdpo.ifeval-abs}{2.0}
\setval{olmo.base.minus.gdpo.judge2.avg}{$-$8.6}
\setval{olmo.base.minus.gdpo.judge2.avg-abs}{8.6}
\setval{olmo.base.minus.gdpo.judge2.creative-writing-v3}{$-$6.2}
\setval{olmo.base.minus.gdpo.judge2.creative-writing-v3-abs}{6.2}
\setval{olmo.base.minus.gdpo.judge2.longbench-write}{$-$2.5}
\setval{olmo.base.minus.gdpo.judge2.longbench-write-abs}{2.5}
\setval{olmo.base.minus.gdpo.judge2.writing}{$-$6.0}
\setval{olmo.base.minus.gdpo.judge2.writing-abs}{6.0}
\setval{olmo.base.minus.gdpo.judge2.writingbench}{$-$9.4}
\setval{olmo.base.minus.gdpo.judge2.writingbench-abs}{9.4}
\setval{olmo.base.minus.gdpo.livecodebench}{$-$6.3}
\setval{olmo.base.minus.gdpo.livecodebench-abs}{6.3}
\setval{olmo.base.minus.gdpo.longbench-write}{$-$5.8}
\setval{olmo.base.minus.gdpo.longbench-write-abs}{5.8}
\setval{olmo.base.minus.gdpo.math}{$-$16.5}
\setval{olmo.base.minus.gdpo.math-abs}{16.5}
\setval{olmo.base.minus.gdpo.math500}{$-$15.1}
\setval{olmo.base.minus.gdpo.math500-abs}{15.1}
\setval{olmo.base.minus.gdpo.mbpp-plus}{$-$3.5}
\setval{olmo.base.minus.gdpo.mbpp-plus-abs}{3.5}
\setval{olmo.base.minus.gdpo.minerva}{$-$5.1}
\setval{olmo.base.minus.gdpo.minerva-abs}{5.1}
\setval{olmo.base.minus.gdpo.olympiad-bench}{$-$19.6}
\setval{olmo.base.minus.gdpo.olympiad-bench-abs}{19.6}
\setval{olmo.base.minus.gdpo.writing}{$-$11.9}
\setval{olmo.base.minus.gdpo.writing-abs}{11.9}
\setval{olmo.base.minus.gdpo.writingbench}{$-$20.6}
\setval{olmo.base.minus.gdpo.writingbench-abs}{20.6}
\setval{olmo.base.minus.geoma.aime24}{$-$15.2}
\setval{olmo.base.minus.geoma.aime24-abs}{15.2}
\setval{olmo.base.minus.geoma.aime25}{$-$14.6}
\setval{olmo.base.minus.geoma.aime25-abs}{14.6}
\setval{olmo.base.minus.geoma.amc23}{$-$19.4}
\setval{olmo.base.minus.geoma.amc23-abs}{19.4}
\setval{olmo.base.minus.geoma.avg}{$-$8.8}
\setval{olmo.base.minus.geoma.avg-abs}{8.8}
\setval{olmo.base.minus.geoma.code}{$-$4.8}
\setval{olmo.base.minus.geoma.code-abs}{4.8}
\setval{olmo.base.minus.geoma.creative-writing-v3}{$-$11.7}
\setval{olmo.base.minus.geoma.creative-writing-v3-abs}{11.7}
\setval{olmo.base.minus.geoma.guidebench}{+14.8}
\setval{olmo.base.minus.geoma.guidebench-abs}{14.8}
\setval{olmo.base.minus.geoma.humaneval-plus}{$-$5.4}
\setval{olmo.base.minus.geoma.humaneval-plus-abs}{5.4}
\setval{olmo.base.minus.geoma.if}{+2.5}
\setval{olmo.base.minus.geoma.if-abs}{2.5}
\setval{olmo.base.minus.geoma.ifbench}{$-$3.7}
\setval{olmo.base.minus.geoma.ifbench-abs}{3.7}
\setval{olmo.base.minus.geoma.ifeval}{$-$3.6}
\setval{olmo.base.minus.geoma.ifeval-abs}{3.6}
\setval{olmo.base.minus.geoma.judge2.avg}{$-$7.8}
\setval{olmo.base.minus.geoma.judge2.avg-abs}{7.8}
\setval{olmo.base.minus.geoma.judge2.creative-writing-v3}{$-$10.1}
\setval{olmo.base.minus.geoma.judge2.creative-writing-v3-abs}{10.1}
\setval{olmo.base.minus.geoma.judge2.longbench-write}{$-$5.0}
\setval{olmo.base.minus.geoma.judge2.longbench-write-abs}{5.0}
\setval{olmo.base.minus.geoma.judge2.writing}{$-$8.7}
\setval{olmo.base.minus.geoma.judge2.writing-abs}{8.7}
\setval{olmo.base.minus.geoma.judge2.writingbench}{$-$11.0}
\setval{olmo.base.minus.geoma.judge2.writingbench-abs}{11.0}
\setval{olmo.base.minus.geoma.livecodebench}{$-$5.5}
\setval{olmo.base.minus.geoma.livecodebench-abs}{5.5}
\setval{olmo.base.minus.geoma.longbench-write}{$-$7.9}
\setval{olmo.base.minus.geoma.longbench-write-abs}{7.9}
\setval{olmo.base.minus.geoma.math}{$-$13.9}
\setval{olmo.base.minus.geoma.math-abs}{13.9}
\setval{olmo.base.minus.geoma.math500}{$-$13.1}
\setval{olmo.base.minus.geoma.math500-abs}{13.1}
\setval{olmo.base.minus.geoma.mbpp-plus}{$-$3.5}
\setval{olmo.base.minus.geoma.mbpp-plus-abs}{3.5}
\setval{olmo.base.minus.geoma.minerva}{$-$4.7}
\setval{olmo.base.minus.geoma.minerva-abs}{4.7}
\setval{olmo.base.minus.geoma.olympiad-bench}{$-$16.5}
\setval{olmo.base.minus.geoma.olympiad-bench-abs}{16.5}
\setval{olmo.base.minus.geoma.writing}{$-$13.9}
\setval{olmo.base.minus.geoma.writing-abs}{13.9}
\setval{olmo.base.minus.geoma.writingbench}{$-$22.2}
\setval{olmo.base.minus.geoma.writingbench-abs}{22.2}
\setval{olmo.base.minus.grpo-nodiv.aime24}{$-$23.2}
\setval{olmo.base.minus.grpo-nodiv.aime24-abs}{23.2}
\setval{olmo.base.minus.grpo-nodiv.aime25}{$-$20.2}
\setval{olmo.base.minus.grpo-nodiv.aime25-abs}{20.2}
\setval{olmo.base.minus.grpo-nodiv.amc23}{$-$28.5}
\setval{olmo.base.minus.grpo-nodiv.amc23-abs}{28.5}
\setval{olmo.base.minus.grpo-nodiv.avg}{$-$11.8}
\setval{olmo.base.minus.grpo-nodiv.avg-abs}{11.8}
\setval{olmo.base.minus.grpo-nodiv.code}{$-$6.3}
\setval{olmo.base.minus.grpo-nodiv.code-abs}{6.3}
\setval{olmo.base.minus.grpo-nodiv.creative-writing-v3}{$-$9.1}
\setval{olmo.base.minus.grpo-nodiv.creative-writing-v3-abs}{9.1}
\setval{olmo.base.minus.grpo-nodiv.guidebench}{+6.0}
\setval{olmo.base.minus.grpo-nodiv.guidebench-abs}{6.0}
\setval{olmo.base.minus.grpo-nodiv.humaneval-plus}{$-$8.2}
\setval{olmo.base.minus.grpo-nodiv.humaneval-plus-abs}{8.2}
\setval{olmo.base.minus.grpo-nodiv.if}{$-$0.7}
\setval{olmo.base.minus.grpo-nodiv.if-abs}{0.7}
\setval{olmo.base.minus.grpo-nodiv.ifbench}{$-$4.7}
\setval{olmo.base.minus.grpo-nodiv.ifbench-abs}{4.7}
\setval{olmo.base.minus.grpo-nodiv.ifeval}{$-$3.4}
\setval{olmo.base.minus.grpo-nodiv.ifeval-abs}{3.4}
\setval{olmo.base.minus.grpo-nodiv.judge2.avg}{$-$10.9}
\setval{olmo.base.minus.grpo-nodiv.judge2.avg-abs}{10.9}
\setval{olmo.base.minus.grpo-nodiv.judge2.creative-writing-v3}{$-$7.2}
\setval{olmo.base.minus.grpo-nodiv.judge2.creative-writing-v3-abs}{7.2}
\setval{olmo.base.minus.grpo-nodiv.judge2.longbench-write}{$-$7.9}
\setval{olmo.base.minus.grpo-nodiv.judge2.longbench-write-abs}{7.9}
\setval{olmo.base.minus.grpo-nodiv.judge2.writing}{$-$8.1}
\setval{olmo.base.minus.grpo-nodiv.judge2.writing-abs}{8.1}
\setval{olmo.base.minus.grpo-nodiv.judge2.writingbench}{$-$9.3}
\setval{olmo.base.minus.grpo-nodiv.judge2.writingbench-abs}{9.3}
\setval{olmo.base.minus.grpo-nodiv.livecodebench}{$-$6.7}
\setval{olmo.base.minus.grpo-nodiv.livecodebench-abs}{6.7}
\setval{olmo.base.minus.grpo-nodiv.longbench-write}{$-$10.1}
\setval{olmo.base.minus.grpo-nodiv.longbench-write-abs}{10.1}
\setval{olmo.base.minus.grpo-nodiv.math}{$-$19.7}
\setval{olmo.base.minus.grpo-nodiv.math-abs}{19.7}
\setval{olmo.base.minus.grpo-nodiv.math500}{$-$17.4}
\setval{olmo.base.minus.grpo-nodiv.math500-abs}{17.4}
\setval{olmo.base.minus.grpo-nodiv.mbpp-plus}{$-$4.1}
\setval{olmo.base.minus.grpo-nodiv.mbpp-plus-abs}{4.1}
\setval{olmo.base.minus.grpo-nodiv.minerva}{$-$6.5}
\setval{olmo.base.minus.grpo-nodiv.minerva-abs}{6.5}
\setval{olmo.base.minus.grpo-nodiv.olympiad-bench}{$-$22.2}
\setval{olmo.base.minus.grpo-nodiv.olympiad-bench-abs}{22.2}
\setval{olmo.base.minus.grpo-nodiv.writing}{$-$12.5}
\setval{olmo.base.minus.grpo-nodiv.writing-abs}{12.5}
\setval{olmo.base.minus.grpo-nodiv.writingbench}{$-$18.4}
\setval{olmo.base.minus.grpo-nodiv.writingbench-abs}{18.4}
\setval{olmo.base.minus.grpo.aime24}{$-$20.9}
\setval{olmo.base.minus.grpo.aime24-abs}{20.9}
\setval{olmo.base.minus.grpo.aime25}{$-$17.3}
\setval{olmo.base.minus.grpo.aime25-abs}{17.3}
\setval{olmo.base.minus.grpo.amc23}{$-$26.3}
\setval{olmo.base.minus.grpo.amc23-abs}{26.3}
\setval{olmo.base.minus.grpo.avg}{$-$11.1}
\setval{olmo.base.minus.grpo.avg-abs}{11.1}
\setval{olmo.base.minus.grpo.code}{$-$6.0}
\setval{olmo.base.minus.grpo.code-abs}{6.0}
\setval{olmo.base.minus.grpo.creative-writing-v3}{$-$11.5}
\setval{olmo.base.minus.grpo.creative-writing-v3-abs}{11.5}
\setval{olmo.base.minus.grpo.guidebench}{+5.0}
\setval{olmo.base.minus.grpo.guidebench-abs}{5.0}
\setval{olmo.base.minus.grpo.humaneval-plus}{$-$8.1}
\setval{olmo.base.minus.grpo.humaneval-plus-abs}{8.1}
\setval{olmo.base.minus.grpo.if}{$-$1.1}
\setval{olmo.base.minus.grpo.if-abs}{1.1}
\setval{olmo.base.minus.grpo.ifbench}{$-$5.3}
\setval{olmo.base.minus.grpo.ifbench-abs}{5.3}
\setval{olmo.base.minus.grpo.ifeval}{$-$3.0}
\setval{olmo.base.minus.grpo.ifeval-abs}{3.0}
\setval{olmo.base.minus.grpo.judge2.avg}{$-$9.8}
\setval{olmo.base.minus.grpo.judge2.avg-abs}{9.8}
\setval{olmo.base.minus.grpo.judge2.creative-writing-v3}{$-$7.0}
\setval{olmo.base.minus.grpo.judge2.creative-writing-v3-abs}{7.0}
\setval{olmo.base.minus.grpo.judge2.longbench-write}{$-$3.9}
\setval{olmo.base.minus.grpo.judge2.longbench-write-abs}{3.9}
\setval{olmo.base.minus.grpo.judge2.writing}{$-$7.2}
\setval{olmo.base.minus.grpo.judge2.writing-abs}{7.2}
\setval{olmo.base.minus.grpo.judge2.writingbench}{$-$10.8}
\setval{olmo.base.minus.grpo.judge2.writingbench-abs}{10.8}
\setval{olmo.base.minus.grpo.livecodebench}{$-$5.5}
\setval{olmo.base.minus.grpo.livecodebench-abs}{5.5}
\setval{olmo.base.minus.grpo.longbench-write}{$-$6.7}
\setval{olmo.base.minus.grpo.longbench-write-abs}{6.7}
\setval{olmo.base.minus.grpo.math}{$-$17.2}
\setval{olmo.base.minus.grpo.math-abs}{17.2}
\setval{olmo.base.minus.grpo.math500}{$-$14.8}
\setval{olmo.base.minus.grpo.math500-abs}{14.8}
\setval{olmo.base.minus.grpo.mbpp-plus}{$-$4.6}
\setval{olmo.base.minus.grpo.mbpp-plus-abs}{4.6}
\setval{olmo.base.minus.grpo.minerva}{$-$4.6}
\setval{olmo.base.minus.grpo.minerva-abs}{4.6}
\setval{olmo.base.minus.grpo.olympiad-bench}{$-$19.4}
\setval{olmo.base.minus.grpo.olympiad-bench-abs}{19.4}
\setval{olmo.base.minus.grpo.writing}{$-$14.0}
\setval{olmo.base.minus.grpo.writing-abs}{14.0}
\setval{olmo.base.minus.grpo.writingbench}{$-$23.7}
\setval{olmo.base.minus.grpo.writingbench-abs}{23.7}
\setval{olmo.base.minus.mt-grpo.aime24}{$-$28.0}
\setval{olmo.base.minus.mt-grpo.aime24-abs}{28.0}
\setval{olmo.base.minus.mt-grpo.aime25}{$-$20.7}
\setval{olmo.base.minus.mt-grpo.aime25-abs}{20.7}
\setval{olmo.base.minus.mt-grpo.amc23}{$-$32.9}
\setval{olmo.base.minus.mt-grpo.amc23-abs}{32.9}
\setval{olmo.base.minus.mt-grpo.avg}{$-$12.6}
\setval{olmo.base.minus.mt-grpo.avg-abs}{12.6}
\setval{olmo.base.minus.mt-grpo.code}{$-$5.3}
\setval{olmo.base.minus.mt-grpo.code-abs}{5.3}
\setval{olmo.base.minus.mt-grpo.creative-writing-v3}{$-$12.8}
\setval{olmo.base.minus.mt-grpo.creative-writing-v3-abs}{12.8}
\setval{olmo.base.minus.mt-grpo.guidebench}{+6.9}
\setval{olmo.base.minus.mt-grpo.guidebench-abs}{6.9}
\setval{olmo.base.minus.mt-grpo.humaneval-plus}{$-$6.5}
\setval{olmo.base.minus.mt-grpo.humaneval-plus-abs}{6.5}
\setval{olmo.base.minus.mt-grpo.if}{$-$0.4}
\setval{olmo.base.minus.mt-grpo.if-abs}{0.4}
\setval{olmo.base.minus.mt-grpo.ifbench}{$-$5.3}
\setval{olmo.base.minus.mt-grpo.ifbench-abs}{5.3}
\setval{olmo.base.minus.mt-grpo.ifeval}{$-$2.6}
\setval{olmo.base.minus.mt-grpo.ifeval-abs}{2.6}
\setval{olmo.base.minus.mt-grpo.judge2.avg}{$-$11.3}
\setval{olmo.base.minus.mt-grpo.judge2.avg-abs}{11.3}
\setval{olmo.base.minus.mt-grpo.judge2.creative-writing-v3}{$-$8.2}
\setval{olmo.base.minus.mt-grpo.judge2.creative-writing-v3-abs}{8.2}
\setval{olmo.base.minus.mt-grpo.judge2.longbench-write}{$-$7.2}
\setval{olmo.base.minus.mt-grpo.judge2.longbench-write-abs}{7.2}
\setval{olmo.base.minus.mt-grpo.judge2.writing}{$-$8.4}
\setval{olmo.base.minus.mt-grpo.judge2.writing-abs}{8.4}
\setval{olmo.base.minus.mt-grpo.judge2.writingbench}{$-$9.8}
\setval{olmo.base.minus.mt-grpo.judge2.writingbench-abs}{9.8}
\setval{olmo.base.minus.mt-grpo.livecodebench}{$-$5.8}
\setval{olmo.base.minus.mt-grpo.livecodebench-abs}{5.8}
\setval{olmo.base.minus.mt-grpo.longbench-write}{$-$8.9}
\setval{olmo.base.minus.mt-grpo.longbench-write-abs}{8.9}
\setval{olmo.base.minus.mt-grpo.math}{$-$21.1}
\setval{olmo.base.minus.mt-grpo.math-abs}{21.1}
\setval{olmo.base.minus.mt-grpo.math500}{$-$16.8}
\setval{olmo.base.minus.mt-grpo.math500-abs}{16.8}
\setval{olmo.base.minus.mt-grpo.mbpp-plus}{$-$3.7}
\setval{olmo.base.minus.mt-grpo.mbpp-plus-abs}{3.7}
\setval{olmo.base.minus.mt-grpo.minerva}{$-$6.7}
\setval{olmo.base.minus.mt-grpo.minerva-abs}{6.7}
\setval{olmo.base.minus.mt-grpo.olympiad-bench}{$-$21.8}
\setval{olmo.base.minus.mt-grpo.olympiad-bench-abs}{21.8}
\setval{olmo.base.minus.mt-grpo.writing}{$-$14.9}
\setval{olmo.base.minus.mt-grpo.writing-abs}{14.9}
\setval{olmo.base.minus.mt-grpo.writingbench}{$-$23.0}
\setval{olmo.base.minus.mt-grpo.writingbench-abs}{23.0}
\setval{olmo.base.minus.papo.aime24}{$-$13.6}
\setval{olmo.base.minus.papo.aime24-abs}{13.6}
\setval{olmo.base.minus.papo.aime25}{$-$12.9}
\setval{olmo.base.minus.papo.aime25-abs}{12.9}
\setval{olmo.base.minus.papo.amc23}{$-$14.8}
\setval{olmo.base.minus.papo.amc23-abs}{14.8}
\setval{olmo.base.minus.papo.avg}{$-$8.2}
\setval{olmo.base.minus.papo.avg-abs}{8.2}
\setval{olmo.base.minus.papo.code}{$-$4.8}
\setval{olmo.base.minus.papo.code-abs}{4.8}
\setval{olmo.base.minus.papo.creative-writing-v3}{$-$9.1}
\setval{olmo.base.minus.papo.creative-writing-v3-abs}{9.1}
\setval{olmo.base.minus.papo.guidebench}{+2.8}
\setval{olmo.base.minus.papo.guidebench-abs}{2.8}
\setval{olmo.base.minus.papo.humaneval-plus}{$-$5.5}
\setval{olmo.base.minus.papo.humaneval-plus-abs}{5.5}
\setval{olmo.base.minus.papo.if}{$-$1.0}
\setval{olmo.base.minus.papo.if-abs}{1.0}
\setval{olmo.base.minus.papo.ifbench}{$-$3.0}
\setval{olmo.base.minus.papo.ifbench-abs}{3.0}
\setval{olmo.base.minus.papo.ifeval}{$-$2.8}
\setval{olmo.base.minus.papo.ifeval-abs}{2.8}
\setval{olmo.base.minus.papo.judge2.avg}{$-$7.0}
\setval{olmo.base.minus.papo.judge2.avg-abs}{7.0}
\setval{olmo.base.minus.papo.judge2.creative-writing-v3}{$-$6.3}
\setval{olmo.base.minus.papo.judge2.creative-writing-v3-abs}{6.3}
\setval{olmo.base.minus.papo.judge2.longbench-write}{$-$2.4}
\setval{olmo.base.minus.papo.judge2.longbench-write-abs}{2.4}
\setval{olmo.base.minus.papo.judge2.writing}{$-$6.2}
\setval{olmo.base.minus.papo.judge2.writing-abs}{6.2}
\setval{olmo.base.minus.papo.judge2.writingbench}{$-$9.8}
\setval{olmo.base.minus.papo.judge2.writingbench-abs}{9.8}
\setval{olmo.base.minus.papo.livecodebench}{$-$5.0}
\setval{olmo.base.minus.papo.livecodebench-abs}{5.0}
\setval{olmo.base.minus.papo.longbench-write}{$-$7.3}
\setval{olmo.base.minus.papo.longbench-write-abs}{7.3}
\setval{olmo.base.minus.papo.math}{$-$11.5}
\setval{olmo.base.minus.papo.math-abs}{11.5}
\setval{olmo.base.minus.papo.math500}{$-$10.8}
\setval{olmo.base.minus.papo.math500-abs}{10.8}
\setval{olmo.base.minus.papo.mbpp-plus}{$-$3.8}
\setval{olmo.base.minus.papo.mbpp-plus-abs}{3.8}
\setval{olmo.base.minus.papo.minerva}{$-$3.3}
\setval{olmo.base.minus.papo.minerva-abs}{3.3}
\setval{olmo.base.minus.papo.olympiad-bench}{$-$13.5}
\setval{olmo.base.minus.papo.olympiad-bench-abs}{13.5}
\setval{olmo.base.minus.papo.writing}{$-$12.3}
\setval{olmo.base.minus.papo.writing-abs}{12.3}
\setval{olmo.base.minus.papo.writingbench}{$-$20.6}
\setval{olmo.base.minus.papo.writingbench-abs}{20.6}
\setval{olmo.base.minus.saw.aime24}{$-$17.6}
\setval{olmo.base.minus.saw.aime24-abs}{17.6}
\setval{olmo.base.minus.saw.aime25}{$-$17.8}
\setval{olmo.base.minus.saw.aime25-abs}{17.8}
\setval{olmo.base.minus.saw.amc23}{$-$22.3}
\setval{olmo.base.minus.saw.amc23-abs}{22.3}
\setval{olmo.base.minus.saw.avg}{$-$10.3}
\setval{olmo.base.minus.saw.avg-abs}{10.3}
\setval{olmo.base.minus.saw.code}{$-$5.4}
\setval{olmo.base.minus.saw.code-abs}{5.4}
\setval{olmo.base.minus.saw.creative-writing-v3}{$-$10.0}
\setval{olmo.base.minus.saw.creative-writing-v3-abs}{10.0}
\setval{olmo.base.minus.saw.guidebench}{+6.7}
\setval{olmo.base.minus.saw.guidebench-abs}{6.7}
\setval{olmo.base.minus.saw.humaneval-plus}{$-$7.1}
\setval{olmo.base.minus.saw.humaneval-plus-abs}{7.1}
\setval{olmo.base.minus.saw.if}{$-$1.2}
\setval{olmo.base.minus.saw.if-abs}{1.2}
\setval{olmo.base.minus.saw.ifbench}{$-$6.0}
\setval{olmo.base.minus.saw.ifbench-abs}{6.0}
\setval{olmo.base.minus.saw.ifeval}{$-$4.2}
\setval{olmo.base.minus.saw.ifeval-abs}{4.2}
\setval{olmo.base.minus.saw.judge2.avg}{$-$9.3}
\setval{olmo.base.minus.saw.judge2.avg-abs}{9.3}
\setval{olmo.base.minus.saw.judge2.creative-writing-v3}{$-$9.3}
\setval{olmo.base.minus.saw.judge2.creative-writing-v3-abs}{9.3}
\setval{olmo.base.minus.saw.judge2.longbench-write}{$-$4.2}
\setval{olmo.base.minus.saw.judge2.longbench-write-abs}{4.2}
\setval{olmo.base.minus.saw.judge2.writing}{$-$8.1}
\setval{olmo.base.minus.saw.judge2.writing-abs}{8.1}
\setval{olmo.base.minus.saw.judge2.writingbench}{$-$10.7}
\setval{olmo.base.minus.saw.judge2.writingbench-abs}{10.7}
\setval{olmo.base.minus.saw.livecodebench}{$-$5.0}
\setval{olmo.base.minus.saw.livecodebench-abs}{5.0}
\setval{olmo.base.minus.saw.longbench-write}{$-$7.5}
\setval{olmo.base.minus.saw.longbench-write-abs}{7.5}
\setval{olmo.base.minus.saw.math}{$-$15.8}
\setval{olmo.base.minus.saw.math-abs}{15.8}
\setval{olmo.base.minus.saw.math500}{$-$14.4}
\setval{olmo.base.minus.saw.math500-abs}{14.4}
\setval{olmo.base.minus.saw.mbpp-plus}{$-$4.2}
\setval{olmo.base.minus.saw.mbpp-plus-abs}{4.2}
\setval{olmo.base.minus.saw.minerva}{$-$5.0}
\setval{olmo.base.minus.saw.minerva-abs}{5.0}
\setval{olmo.base.minus.saw.olympiad-bench}{$-$17.5}
\setval{olmo.base.minus.saw.olympiad-bench-abs}{17.5}
\setval{olmo.base.minus.saw.writing}{$-$13.4}
\setval{olmo.base.minus.saw.writing-abs}{13.4}
\setval{olmo.base.minus.saw.writingbench}{$-$22.6}
\setval{olmo.base.minus.saw.writingbench-abs}{22.6}
\setval{olmo.concur-free.aux-change.code}{+0.09}
\setval{olmo.concur-free.aux-change.if}{+0.11}
\setval{olmo.concur-free.aux-change.math}{$-$0.14}
\setval{olmo.concur-free.aux-change.writing}{$-$0.69}
\setval{olmo.concur-free.aux-change.writing-rm}{+0.23}
\setval{olmo.concur-free.aux-full.code.early}{0.73}
\setval{olmo.concur-free.aux-full.code.late}{0.82}
\setval{olmo.concur-free.aux-full.code.run}{0.82}
\setval{olmo.concur-free.aux-full.if.early}{0.57}
\setval{olmo.concur-free.aux-full.if.late}{0.92}
\setval{olmo.concur-free.aux-full.if.run}{0.81}
\setval{olmo.concur-free.aux-full.math.early}{0.53}
\setval{olmo.concur-free.aux-full.math.late}{0.52}
\setval{olmo.concur-free.aux-full.math.run}{0.43}
\setval{olmo.concur-free.aux-full.writing-rm.early}{0.02}
\setval{olmo.concur-free.aux-full.writing-rm.late}{0.07}
\setval{olmo.concur-free.aux-full.writing-rm.run}{0.04}
\setval{olmo.concur-free.aux-full.writing.early}{0.49}
\setval{olmo.concur-free.aux-full.writing.late}{0.08}
\setval{olmo.concur-free.aux-full.writing.run}{0.13}
\setval{olmo.concur-free.aux-level.code.early}{0.89}
\setval{olmo.concur-free.aux-level.code.late}{0.90}
\setval{olmo.concur-free.aux-level.code.run}{0.91}
\setval{olmo.concur-free.aux-level.if.early}{0.89}
\setval{olmo.concur-free.aux-level.if.late}{0.98}
\setval{olmo.concur-free.aux-level.if.run}{0.96}
\setval{olmo.concur-free.aux-level.math.early}{0.90}
\setval{olmo.concur-free.aux-level.math.late}{0.76}
\setval{olmo.concur-free.aux-level.math.run}{0.74}
\setval{olmo.concur-free.aux-level.writing-rm.early}{0.60}
\setval{olmo.concur-free.aux-level.writing-rm.late}{0.80}
\setval{olmo.concur-free.aux-level.writing-rm.run}{0.73}
\setval{olmo.concur-free.aux-level.writing.early}{0.93}
\setval{olmo.concur-free.aux-level.writing.late}{0.25}
\setval{olmo.concur-free.aux-level.writing.run}{0.49}
\setval{olmo.concur-free.bench.aime24}{26.8}
\setval{olmo.concur-free.bench.aime25}{22.1}
\setval{olmo.concur-free.bench.amc23}{67.5}
\setval{olmo.concur-free.bench.creative-writing-v3}{49.8}
\setval{olmo.concur-free.bench.guidebench}{60.2}
\setval{olmo.concur-free.bench.humaneval-plus}{76.6}
\setval{olmo.concur-free.bench.ifbench}{31.7}
\setval{olmo.concur-free.bench.ifeval}{87.3}
\setval{olmo.concur-free.bench.livecodebench}{25.7}
\setval{olmo.concur-free.bench.longbench-write}{72.0}
\setval{olmo.concur-free.bench.math500}{84.4}
\setval{olmo.concur-free.bench.mbpp-plus}{64.5}
\setval{olmo.concur-free.bench.minerva}{39.9}
\setval{olmo.concur-free.bench.olympiad-bench}{53.3}
\setval{olmo.concur-free.bench.writingbench}{70.2}
\setval{olmo.concur-free.eval.avg}{55.5}
\setval{olmo.concur-free.eval.code}{55.6}
\setval{olmo.concur-free.eval.if}{59.7}
\setval{olmo.concur-free.eval.math}{49.0}
\setval{olmo.concur-free.eval.writing}{64.0}
\setval{olmo.concur-free.gate.code.against.early}{0.06}
\setval{olmo.concur-free.gate.code.against.late}{0.17}
\setval{olmo.concur-free.gate.code.against.run}{0.13}
\setval{olmo.concur-free.gate.code.agreement}{$-$0.042}
\setval{olmo.concur-free.gate.code.ceiling.early}{0.24}
\setval{olmo.concur-free.gate.code.ceiling.late}{0.31}
\setval{olmo.concur-free.gate.code.ceiling.run}{0.27}
\setval{olmo.concur-free.gate.code.cosine.early}{0.73}
\setval{olmo.concur-free.gate.code.cosine.late}{0.45}
\setval{olmo.concur-free.gate.code.cosine.run}{0.55}
\setval{olmo.concur-free.gate.if.against.early}{0.10}
\setval{olmo.concur-free.gate.if.against.late}{0.05}
\setval{olmo.concur-free.gate.if.against.run}{0.08}
\setval{olmo.concur-free.gate.if.agreement}{+0.053}
\setval{olmo.concur-free.gate.if.ceiling.early}{0.22}
\setval{olmo.concur-free.gate.if.ceiling.late}{0.18}
\setval{olmo.concur-free.gate.if.ceiling.run}{0.21}
\setval{olmo.concur-free.gate.if.cosine.early}{0.52}
\setval{olmo.concur-free.gate.if.cosine.late}{0.53}
\setval{olmo.concur-free.gate.if.cosine.run}{0.52}
\setval{olmo.concur-free.gate.math.against.early}{0.00}
\setval{olmo.concur-free.gate.math.against.late}{0.00}
\setval{olmo.concur-free.gate.math.against.run}{0.00}
\setval{olmo.concur-free.gate.math.ceiling.early}{0.14}
\setval{olmo.concur-free.gate.math.ceiling.late}{0.48}
\setval{olmo.concur-free.gate.math.ceiling.run}{0.43}
\setval{olmo.concur-free.gate.math.cosine.early}{0.90}
\setval{olmo.concur-free.gate.math.cosine.late}{0.63}
\setval{olmo.concur-free.gate.math.cosine.run}{0.66}
\setval{olmo.concur-free.gate.writing-rm.against.early}{0.22}
\setval{olmo.concur-free.gate.writing-rm.against.late}{0.54}
\setval{olmo.concur-free.gate.writing-rm.against.run}{0.47}
\setval{olmo.concur-free.gate.writing-rm.agreement}{+0.091}
\setval{olmo.concur-free.gate.writing-rm.ceiling.early}{0.00}
\setval{olmo.concur-free.gate.writing-rm.ceiling.late}{0.00}
\setval{olmo.concur-free.gate.writing-rm.ceiling.run}{0.00}
\setval{olmo.concur-free.gate.writing-rm.cosine.early}{0.69}
\setval{olmo.concur-free.gate.writing-rm.cosine.late}{0.20}
\setval{olmo.concur-free.gate.writing-rm.cosine.run}{0.33}
\setval{olmo.concur-free.gate.writing.against.early}{0.27}
\setval{olmo.concur-free.gate.writing.against.late}{0.67}
\setval{olmo.concur-free.gate.writing.against.run}{0.62}
\setval{olmo.concur-free.gate.writing.agreement}{$-$0.066}
\setval{olmo.concur-free.gate.writing.ceiling.early}{0.00}
\setval{olmo.concur-free.gate.writing.ceiling.late}{0.00}
\setval{olmo.concur-free.gate.writing.ceiling.run}{0.00}
\setval{olmo.concur-free.gate.writing.cosine.early}{0.58}
\setval{olmo.concur-free.gate.writing.cosine.late}{-0.01}
\setval{olmo.concur-free.gate.writing.cosine.run}{0.08}
\setval{olmo.concur-free.judge2.avg}{52.6}
\setval{olmo.concur-free.judge2.creative-writing-v3}{37.6}
\setval{olmo.concur-free.judge2.longbench-write}{63.9}
\setval{olmo.concur-free.judge2.writing}{49.8}
\setval{olmo.concur-free.judge2.writingbench}{48.0}
\setval{olmo.concur-free.last-step}{300}
\setval{olmo.concur-free.minus.base.aime24}{+18.2}
\setval{olmo.concur-free.minus.base.aime24-abs}{18.2}
\setval{olmo.concur-free.minus.base.aime25}{+14.7}
\setval{olmo.concur-free.minus.base.aime25-abs}{14.7}
\setval{olmo.concur-free.minus.base.amc23}{+23.0}
\setval{olmo.concur-free.minus.base.amc23-abs}{23.0}
\setval{olmo.concur-free.minus.base.avg}{+11.3}
\setval{olmo.concur-free.minus.base.avg-abs}{11.3}
\setval{olmo.concur-free.minus.base.code}{+6.5}
\setval{olmo.concur-free.minus.base.code-abs}{6.5}
\setval{olmo.concur-free.minus.base.creative-writing-v3}{+11.9}
\setval{olmo.concur-free.minus.base.creative-writing-v3-abs}{11.9}
\setval{olmo.concur-free.minus.base.guidebench}{$-$1.2}
\setval{olmo.concur-free.minus.base.guidebench-abs}{1.2}
\setval{olmo.concur-free.minus.base.humaneval-plus}{+7.4}
\setval{olmo.concur-free.minus.base.humaneval-plus-abs}{7.4}
\setval{olmo.concur-free.minus.base.if}{+2.2}
\setval{olmo.concur-free.minus.base.if-abs}{2.2}
\setval{olmo.concur-free.minus.base.ifbench}{+5.0}
\setval{olmo.concur-free.minus.base.ifbench-abs}{5.0}
\setval{olmo.concur-free.minus.base.ifeval}{+2.9}
\setval{olmo.concur-free.minus.base.ifeval-abs}{2.9}
\setval{olmo.concur-free.minus.base.judge2.avg}{+9.7}
\setval{olmo.concur-free.minus.base.judge2.avg-abs}{9.7}
\setval{olmo.concur-free.minus.base.judge2.creative-writing-v3}{+5.9}
\setval{olmo.concur-free.minus.base.judge2.creative-writing-v3-abs}{5.9}
\setval{olmo.concur-free.minus.base.judge2.longbench-write}{+10.7}
\setval{olmo.concur-free.minus.base.judge2.longbench-write-abs}{10.7}
\setval{olmo.concur-free.minus.base.judge2.writing}{+9.0}
\setval{olmo.concur-free.minus.base.judge2.writing-abs}{9.0}
\setval{olmo.concur-free.minus.base.judge2.writingbench}{+10.4}
\setval{olmo.concur-free.minus.base.judge2.writingbench-abs}{10.4}
\setval{olmo.concur-free.minus.base.livecodebench}{+8.8}
\setval{olmo.concur-free.minus.base.livecodebench-abs}{8.8}
\setval{olmo.concur-free.minus.base.longbench-write}{+14.6}
\setval{olmo.concur-free.minus.base.longbench-write-abs}{14.6}
\setval{olmo.concur-free.minus.base.math}{+15.3}
\setval{olmo.concur-free.minus.base.math-abs}{15.3}
\setval{olmo.concur-free.minus.base.math500}{+14.2}
\setval{olmo.concur-free.minus.base.math500-abs}{14.2}
\setval{olmo.concur-free.minus.base.mbpp-plus}{+3.3}
\setval{olmo.concur-free.minus.base.mbpp-plus-abs}{3.3}
\setval{olmo.concur-free.minus.base.minerva}{+4.6}
\setval{olmo.concur-free.minus.base.minerva-abs}{4.6}
\setval{olmo.concur-free.minus.base.olympiad-bench}{+17.0}
\setval{olmo.concur-free.minus.base.olympiad-bench-abs}{17.0}
\setval{olmo.concur-free.minus.base.writing}{+17.3}
\setval{olmo.concur-free.minus.base.writing-abs}{17.3}
\setval{olmo.concur-free.minus.base.writingbench}{+25.5}
\setval{olmo.concur-free.minus.base.writingbench-abs}{25.5}
\setval{olmo.concur-free.minus.concur-tied.aime24}{$-$7.7}
\setval{olmo.concur-free.minus.concur-tied.aime24-abs}{7.7}
\setval{olmo.concur-free.minus.concur-tied.aime25}{$-$7.8}
\setval{olmo.concur-free.minus.concur-tied.aime25-abs}{7.8}
\setval{olmo.concur-free.minus.concur-tied.amc23}{$-$12.3}
\setval{olmo.concur-free.minus.concur-tied.amc23-abs}{12.3}
\setval{olmo.concur-free.minus.concur-tied.avg}{$-$3.7}
\setval{olmo.concur-free.minus.concur-tied.avg-abs}{3.7}
\setval{olmo.concur-free.minus.concur-tied.code}{$-$5.5}
\setval{olmo.concur-free.minus.concur-tied.code-abs}{5.5}
\setval{olmo.concur-free.minus.concur-tied.creative-writing-v3}{+0.7}
\setval{olmo.concur-free.minus.concur-tied.creative-writing-v3-abs}{0.7}
\setval{olmo.concur-free.minus.concur-tied.guidebench}{+1.6}
\setval{olmo.concur-free.minus.concur-tied.guidebench-abs}{1.6}
\setval{olmo.concur-free.minus.concur-tied.humaneval-plus}{$-$6.6}
\setval{olmo.concur-free.minus.concur-tied.humaneval-plus-abs}{6.6}
\setval{olmo.concur-free.minus.concur-tied.if}{$-$0.2}
\setval{olmo.concur-free.minus.concur-tied.if-abs}{0.2}
\setval{olmo.concur-free.minus.concur-tied.ifbench}{$-$2.0}
\setval{olmo.concur-free.minus.concur-tied.ifbench-abs}{2.0}
\setval{olmo.concur-free.minus.concur-tied.ifeval}{$-$0.4}
\setval{olmo.concur-free.minus.concur-tied.ifeval-abs}{0.4}
\setval{olmo.concur-free.minus.concur-tied.judge2.avg}{$-$3.8}
\setval{olmo.concur-free.minus.concur-tied.judge2.avg-abs}{3.8}
\setval{olmo.concur-free.minus.concur-tied.judge2.creative-writing-v3}{+0.4}
\setval{olmo.concur-free.minus.concur-tied.judge2.creative-writing-v3-abs}{0.4}
\setval{olmo.concur-free.minus.concur-tied.judge2.longbench-write}{+3.4}
\setval{olmo.concur-free.minus.concur-tied.judge2.longbench-write-abs}{3.4}
\setval{olmo.concur-free.minus.concur-tied.judge2.writing}{+2.1}
\setval{olmo.concur-free.minus.concur-tied.judge2.writing-abs}{2.1}
\setval{olmo.concur-free.minus.concur-tied.judge2.writingbench}{+2.4}
\setval{olmo.concur-free.minus.concur-tied.judge2.writingbench-abs}{2.4}
\setval{olmo.concur-free.minus.concur-tied.livecodebench}{$-$3.4}
\setval{olmo.concur-free.minus.concur-tied.livecodebench-abs}{3.4}
\setval{olmo.concur-free.minus.concur-tied.longbench-write}{+2.8}
\setval{olmo.concur-free.minus.concur-tied.longbench-write-abs}{2.8}
\setval{olmo.concur-free.minus.concur-tied.math}{$-$7.6}
\setval{olmo.concur-free.minus.concur-tied.math-abs}{7.6}
\setval{olmo.concur-free.minus.concur-tied.math500}{$-$5.0}
\setval{olmo.concur-free.minus.concur-tied.math500-abs}{5.0}
\setval{olmo.concur-free.minus.concur-tied.mbpp-plus}{$-$6.6}
\setval{olmo.concur-free.minus.concur-tied.mbpp-plus-abs}{6.6}
\setval{olmo.concur-free.minus.concur-tied.minerva}{$-$4.1}
\setval{olmo.concur-free.minus.concur-tied.minerva-abs}{4.1}
\setval{olmo.concur-free.minus.concur-tied.olympiad-bench}{$-$8.3}
\setval{olmo.concur-free.minus.concur-tied.olympiad-bench-abs}{8.3}
\setval{olmo.concur-free.minus.concur-tied.writing}{+2.5}
\setval{olmo.concur-free.minus.concur-tied.writing-abs}{2.5}
\setval{olmo.concur-free.minus.concur-tied.writingbench}{+4.1}
\setval{olmo.concur-free.minus.concur-tied.writingbench-abs}{4.1}
\setval{olmo.concur-free.minus.concur.aime24}{$-$9.6}
\setval{olmo.concur-free.minus.concur.aime24-abs}{9.6}
\setval{olmo.concur-free.minus.concur.aime25}{$-$10.1}
\setval{olmo.concur-free.minus.concur.aime25-abs}{10.1}
\setval{olmo.concur-free.minus.concur.amc23}{$-$16.0}
\setval{olmo.concur-free.minus.concur.amc23-abs}{16.0}
\setval{olmo.concur-free.minus.concur.avg}{$-$3.9}
\setval{olmo.concur-free.minus.concur.avg-abs}{3.9}
\setval{olmo.concur-free.minus.concur.code}{$-$4.6}
\setval{olmo.concur-free.minus.concur.code-abs}{4.6}
\setval{olmo.concur-free.minus.concur.creative-writing-v3}{+0.6}
\setval{olmo.concur-free.minus.concur.creative-writing-v3-abs}{0.6}
\setval{olmo.concur-free.minus.concur.guidebench}{+3.3}
\setval{olmo.concur-free.minus.concur.guidebench-abs}{3.3}
\setval{olmo.concur-free.minus.concur.humaneval-plus}{$-$6.1}
\setval{olmo.concur-free.minus.concur.humaneval-plus-abs}{6.1}
\setval{olmo.concur-free.minus.concur.if}{+0.8}
\setval{olmo.concur-free.minus.concur.if-abs}{0.8}
\setval{olmo.concur-free.minus.concur.ifbench}{$-$0.3}
\setval{olmo.concur-free.minus.concur.ifbench-abs}{0.3}
\setval{olmo.concur-free.minus.concur.ifeval}{$-$0.5}
\setval{olmo.concur-free.minus.concur.ifeval-abs}{0.5}
\setval{olmo.concur-free.minus.concur.judge2.avg}{$-$4.0}
\setval{olmo.concur-free.minus.concur.judge2.avg-abs}{4.0}
\setval{olmo.concur-free.minus.concur.judge2.creative-writing-v3}{$-$0.9}
\setval{olmo.concur-free.minus.concur.judge2.creative-writing-v3-abs}{0.9}
\setval{olmo.concur-free.minus.concur.judge2.longbench-write}{+5.5}
\setval{olmo.concur-free.minus.concur.judge2.longbench-write-abs}{5.5}
\setval{olmo.concur-free.minus.concur.judge2.writing}{+2.6}
\setval{olmo.concur-free.minus.concur.judge2.writing-abs}{2.6}
\setval{olmo.concur-free.minus.concur.judge2.writingbench}{+3.1}
\setval{olmo.concur-free.minus.concur.judge2.writingbench-abs}{3.1}
\setval{olmo.concur-free.minus.concur.livecodebench}{$-$4.7}
\setval{olmo.concur-free.minus.concur.livecodebench-abs}{4.7}
\setval{olmo.concur-free.minus.concur.longbench-write}{+3.6}
\setval{olmo.concur-free.minus.concur.longbench-write-abs}{3.6}
\setval{olmo.concur-free.minus.concur.math}{$-$9.3}
\setval{olmo.concur-free.minus.concur.math-abs}{9.3}
\setval{olmo.concur-free.minus.concur.math500}{$-$6.2}
\setval{olmo.concur-free.minus.concur.math500-abs}{6.2}
\setval{olmo.concur-free.minus.concur.mbpp-plus}{$-$3.0}
\setval{olmo.concur-free.minus.concur.mbpp-plus-abs}{3.0}
\setval{olmo.concur-free.minus.concur.minerva}{$-$4.9}
\setval{olmo.concur-free.minus.concur.minerva-abs}{4.9}
\setval{olmo.concur-free.minus.concur.olympiad-bench}{$-$9.1}
\setval{olmo.concur-free.minus.concur.olympiad-bench-abs}{9.1}
\setval{olmo.concur-free.minus.concur.writing}{+3.0}
\setval{olmo.concur-free.minus.concur.writing-abs}{3.0}
\setval{olmo.concur-free.minus.concur.writingbench}{+4.8}
\setval{olmo.concur-free.minus.concur.writingbench-abs}{4.8}
\setval{olmo.concur-free.minus.gd2po.aime24}{$-$1.6}
\setval{olmo.concur-free.minus.gd2po.aime24-abs}{1.6}
\setval{olmo.concur-free.minus.gd2po.aime25}{$-$2.2}
\setval{olmo.concur-free.minus.gd2po.aime25-abs}{2.2}
\setval{olmo.concur-free.minus.gd2po.amc23}{$-$4.7}
\setval{olmo.concur-free.minus.gd2po.amc23-abs}{4.7}
\setval{olmo.concur-free.minus.gd2po.avg}{$-$0.2}
\setval{olmo.concur-free.minus.gd2po.avg-abs}{0.2}
\setval{olmo.concur-free.minus.gd2po.code}{+0.7}
\setval{olmo.concur-free.minus.gd2po.code-abs}{0.7}
\setval{olmo.concur-free.minus.gd2po.creative-writing-v3}{+1.5}
\setval{olmo.concur-free.minus.gd2po.creative-writing-v3-abs}{1.5}
\setval{olmo.concur-free.minus.gd2po.guidebench}{$-$1.0}
\setval{olmo.concur-free.minus.gd2po.guidebench-abs}{1.0}
\setval{olmo.concur-free.minus.gd2po.humaneval-plus}{$-$0.1}
\setval{olmo.concur-free.minus.gd2po.humaneval-plus-abs}{0.1}
\setval{olmo.concur-free.minus.gd2po.if}{$-$0.7}
\setval{olmo.concur-free.minus.gd2po.if-abs}{0.7}
\setval{olmo.concur-free.minus.gd2po.ifbench}{$-$0.7}
\setval{olmo.concur-free.minus.gd2po.ifbench-abs}{0.7}
\setval{olmo.concur-free.minus.gd2po.ifeval}{$-$0.4}
\setval{olmo.concur-free.minus.gd2po.ifeval-abs}{0.4}
\setval{olmo.concur-free.minus.gd2po.judge2.avg}{$-$0.6}
\setval{olmo.concur-free.minus.gd2po.judge2.avg-abs}{0.6}
\setval{olmo.concur-free.minus.gd2po.judge2.creative-writing-v3}{$-$0.3}
\setval{olmo.concur-free.minus.gd2po.judge2.creative-writing-v3-abs}{0.3}
\setval{olmo.concur-free.minus.gd2po.judge2.longbench-write}{+4.0}
\setval{olmo.concur-free.minus.gd2po.judge2.longbench-write-abs}{4.0}
\setval{olmo.concur-free.minus.gd2po.judge2.writing}{+1.1}
\setval{olmo.concur-free.minus.gd2po.judge2.writing-abs}{1.1}
\setval{olmo.concur-free.minus.gd2po.judge2.writingbench}{$-$0.3}
\setval{olmo.concur-free.minus.gd2po.judge2.writingbench-abs}{0.3}
\setval{olmo.concur-free.minus.gd2po.livecodebench}{+2.5}
\setval{olmo.concur-free.minus.gd2po.livecodebench-abs}{2.5}
\setval{olmo.concur-free.minus.gd2po.longbench-write}{+3.8}
\setval{olmo.concur-free.minus.gd2po.longbench-write-abs}{3.8}
\setval{olmo.concur-free.minus.gd2po.math}{$-$2.1}
\setval{olmo.concur-free.minus.gd2po.math-abs}{2.1}
\setval{olmo.concur-free.minus.gd2po.math500}{$-$1.4}
\setval{olmo.concur-free.minus.gd2po.math500-abs}{1.4}
\setval{olmo.concur-free.minus.gd2po.mbpp-plus}{$-$0.3}
\setval{olmo.concur-free.minus.gd2po.mbpp-plus-abs}{0.3}
\setval{olmo.concur-free.minus.gd2po.minerva}{$-$0.2}
\setval{olmo.concur-free.minus.gd2po.minerva-abs}{0.2}
\setval{olmo.concur-free.minus.gd2po.olympiad-bench}{$-$2.8}
\setval{olmo.concur-free.minus.gd2po.olympiad-bench-abs}{2.8}
\setval{olmo.concur-free.minus.gd2po.writing}{+3.3}
\setval{olmo.concur-free.minus.gd2po.writing-abs}{3.3}
\setval{olmo.concur-free.minus.gd2po.writingbench}{+4.5}
\setval{olmo.concur-free.minus.gd2po.writingbench-abs}{4.5}
\setval{olmo.concur-free.minus.gdpo-noaux.aime24}{$-$11.6}
\setval{olmo.concur-free.minus.gdpo-noaux.aime24-abs}{11.6}
\setval{olmo.concur-free.minus.gdpo-noaux.aime25}{$-$11.6}
\setval{olmo.concur-free.minus.gdpo-noaux.aime25-abs}{11.6}
\setval{olmo.concur-free.minus.gdpo-noaux.amc23}{$-$16.4}
\setval{olmo.concur-free.minus.gdpo-noaux.amc23-abs}{16.4}
\setval{olmo.concur-free.minus.gdpo-noaux.avg}{$-$2.4}
\setval{olmo.concur-free.minus.gdpo-noaux.avg-abs}{2.4}
\setval{olmo.concur-free.minus.gdpo-noaux.code}{$-$6.8}
\setval{olmo.concur-free.minus.gdpo-noaux.code-abs}{6.8}
\setval{olmo.concur-free.minus.gdpo-noaux.creative-writing-v3}{+11.6}
\setval{olmo.concur-free.minus.gdpo-noaux.creative-writing-v3-abs}{11.6}
\setval{olmo.concur-free.minus.gdpo-noaux.guidebench}{+4.4}
\setval{olmo.concur-free.minus.gdpo-noaux.guidebench-abs}{4.4}
\setval{olmo.concur-free.minus.gdpo-noaux.humaneval-plus}{$-$8.5}
\setval{olmo.concur-free.minus.gdpo-noaux.humaneval-plus-abs}{8.5}
\setval{olmo.concur-free.minus.gdpo-noaux.if}{+1.8}
\setval{olmo.concur-free.minus.gdpo-noaux.if-abs}{1.8}
\setval{olmo.concur-free.minus.gdpo-noaux.ifbench}{$-$1.7}
\setval{olmo.concur-free.minus.gdpo-noaux.ifbench-abs}{1.7}
\setval{olmo.concur-free.minus.gdpo-noaux.ifeval}{+2.5}
\setval{olmo.concur-free.minus.gdpo-noaux.ifeval-abs}{2.5}
\setval{olmo.concur-free.minus.gdpo-noaux.judge2.avg}{$-$3.0}
\setval{olmo.concur-free.minus.gdpo-noaux.judge2.avg-abs}{3.0}
\setval{olmo.concur-free.minus.gdpo-noaux.judge2.creative-writing-v3}{+6.6}
\setval{olmo.concur-free.minus.gdpo-noaux.judge2.creative-writing-v3-abs}{6.6}
\setval{olmo.concur-free.minus.gdpo-noaux.judge2.longbench-write}{+17.3}
\setval{olmo.concur-free.minus.gdpo-noaux.judge2.longbench-write-abs}{17.3}
\setval{olmo.concur-free.minus.gdpo-noaux.judge2.writing}{+10.7}
\setval{olmo.concur-free.minus.gdpo-noaux.judge2.writing-abs}{10.7}
\setval{olmo.concur-free.minus.gdpo-noaux.judge2.writingbench}{+8.3}
\setval{olmo.concur-free.minus.gdpo-noaux.judge2.writingbench-abs}{8.3}
\setval{olmo.concur-free.minus.gdpo-noaux.livecodebench}{$-$5.9}
\setval{olmo.concur-free.minus.gdpo-noaux.livecodebench-abs}{5.9}
\setval{olmo.concur-free.minus.gdpo-noaux.longbench-write}{+16.1}
\setval{olmo.concur-free.minus.gdpo-noaux.longbench-write-abs}{16.1}
\setval{olmo.concur-free.minus.gdpo-noaux.math}{$-$10.5}
\setval{olmo.concur-free.minus.gdpo-noaux.math-abs}{10.5}
\setval{olmo.concur-free.minus.gdpo-noaux.math500}{$-$6.5}
\setval{olmo.concur-free.minus.gdpo-noaux.math500-abs}{6.5}
\setval{olmo.concur-free.minus.gdpo-noaux.mbpp-plus}{$-$6.0}
\setval{olmo.concur-free.minus.gdpo-noaux.mbpp-plus-abs}{6.0}
\setval{olmo.concur-free.minus.gdpo-noaux.minerva}{$-$6.2}
\setval{olmo.concur-free.minus.gdpo-noaux.minerva-abs}{6.2}
\setval{olmo.concur-free.minus.gdpo-noaux.olympiad-bench}{$-$10.5}
\setval{olmo.concur-free.minus.gdpo-noaux.olympiad-bench-abs}{10.5}
\setval{olmo.concur-free.minus.gdpo-noaux.writing}{+14.1}
\setval{olmo.concur-free.minus.gdpo-noaux.writing-abs}{14.1}
\setval{olmo.concur-free.minus.gdpo-noaux.writingbench}{+14.7}
\setval{olmo.concur-free.minus.gdpo-noaux.writingbench-abs}{14.7}
\setval{olmo.concur-free.minus.gdpo-w0.5.aime24}{$-$2.8}
\setval{olmo.concur-free.minus.gdpo-w0.5.aime24-abs}{2.8}
\setval{olmo.concur-free.minus.gdpo-w0.5.aime25}{$-$5.2}
\setval{olmo.concur-free.minus.gdpo-w0.5.aime25-abs}{5.2}
\setval{olmo.concur-free.minus.gdpo-w0.5.amc23}{$-$6.5}
\setval{olmo.concur-free.minus.gdpo-w0.5.amc23-abs}{6.5}
\setval{olmo.concur-free.minus.gdpo-w0.5.avg}{$-$0.2}
\setval{olmo.concur-free.minus.gdpo-w0.5.avg-abs}{0.2}
\setval{olmo.concur-free.minus.gdpo-w0.5.code}{$-$0.2}
\setval{olmo.concur-free.minus.gdpo-w0.5.code-abs}{0.2}
\setval{olmo.concur-free.minus.gdpo-w0.5.creative-writing-v3}{+1.7}
\setval{olmo.concur-free.minus.gdpo-w0.5.creative-writing-v3-abs}{1.7}
\setval{olmo.concur-free.minus.gdpo-w0.5.guidebench}{+8.3}
\setval{olmo.concur-free.minus.gdpo-w0.5.guidebench-abs}{8.3}
\setval{olmo.concur-free.minus.gdpo-w0.5.humaneval-plus}{$-$1.1}
\setval{olmo.concur-free.minus.gdpo-w0.5.humaneval-plus-abs}{1.1}
\setval{olmo.concur-free.minus.gdpo-w0.5.if}{+2.3}
\setval{olmo.concur-free.minus.gdpo-w0.5.if-abs}{2.3}
\setval{olmo.concur-free.minus.gdpo-w0.5.ifbench}{$-$1.0}
\setval{olmo.concur-free.minus.gdpo-w0.5.ifbench-abs}{1.0}
\setval{olmo.concur-free.minus.gdpo-w0.5.ifeval}{$-$0.2}
\setval{olmo.concur-free.minus.gdpo-w0.5.ifeval-abs}{0.2}
\setval{olmo.concur-free.minus.gdpo-w0.5.judge2.avg}{$-$0.7}
\setval{olmo.concur-free.minus.gdpo-w0.5.judge2.avg-abs}{0.7}
\setval{olmo.concur-free.minus.gdpo-w0.5.judge2.creative-writing-v3}{$-$1.9}
\setval{olmo.concur-free.minus.gdpo-w0.5.judge2.creative-writing-v3-abs}{1.9}
\setval{olmo.concur-free.minus.gdpo-w0.5.judge2.longbench-write}{+7.3}
\setval{olmo.concur-free.minus.gdpo-w0.5.judge2.longbench-write-abs}{7.3}
\setval{olmo.concur-free.minus.gdpo-w0.5.judge2.writing}{+2.3}
\setval{olmo.concur-free.minus.gdpo-w0.5.judge2.writing-abs}{2.3}
\setval{olmo.concur-free.minus.gdpo-w0.5.judge2.writingbench}{+1.3}
\setval{olmo.concur-free.minus.gdpo-w0.5.judge2.writingbench-abs}{1.3}
\setval{olmo.concur-free.minus.gdpo-w0.5.livecodebench}{+0.9}
\setval{olmo.concur-free.minus.gdpo-w0.5.livecodebench-abs}{0.9}
\setval{olmo.concur-free.minus.gdpo-w0.5.longbench-write}{+8.3}
\setval{olmo.concur-free.minus.gdpo-w0.5.longbench-write-abs}{8.3}
\setval{olmo.concur-free.minus.gdpo-w0.5.math}{$-$3.8}
\setval{olmo.concur-free.minus.gdpo-w0.5.math-abs}{3.8}
\setval{olmo.concur-free.minus.gdpo-w0.5.math500}{$-$2.7}
\setval{olmo.concur-free.minus.gdpo-w0.5.math500-abs}{2.7}
\setval{olmo.concur-free.minus.gdpo-w0.5.mbpp-plus}{$-$0.4}
\setval{olmo.concur-free.minus.gdpo-w0.5.mbpp-plus-abs}{0.4}
\setval{olmo.concur-free.minus.gdpo-w0.5.minerva}{$-$1.6}
\setval{olmo.concur-free.minus.gdpo-w0.5.minerva-abs}{1.6}
\setval{olmo.concur-free.minus.gdpo-w0.5.olympiad-bench}{$-$4.2}
\setval{olmo.concur-free.minus.gdpo-w0.5.olympiad-bench-abs}{4.2}
\setval{olmo.concur-free.minus.gdpo-w0.5.writing}{+4.6}
\setval{olmo.concur-free.minus.gdpo-w0.5.writing-abs}{4.6}
\setval{olmo.concur-free.minus.gdpo-w0.5.writingbench}{+3.7}
\setval{olmo.concur-free.minus.gdpo-w0.5.writingbench-abs}{3.7}
\setval{olmo.concur-free.minus.gdpo.aime24}{$-$1.0}
\setval{olmo.concur-free.minus.gdpo.aime24-abs}{1.0}
\setval{olmo.concur-free.minus.gdpo.aime25}{$-$0.5}
\setval{olmo.concur-free.minus.gdpo.aime25-abs}{0.5}
\setval{olmo.concur-free.minus.gdpo.amc23}{$-$2.0}
\setval{olmo.concur-free.minus.gdpo.amc23-abs}{2.0}
\setval{olmo.concur-free.minus.gdpo.avg}{+1.6}
\setval{olmo.concur-free.minus.gdpo.avg-abs}{1.6}
\setval{olmo.concur-free.minus.gdpo.code}{+0.9}
\setval{olmo.concur-free.minus.gdpo.code-abs}{0.9}
\setval{olmo.concur-free.minus.gdpo.creative-writing-v3}{+2.6}
\setval{olmo.concur-free.minus.gdpo.creative-writing-v3-abs}{2.6}
\setval{olmo.concur-free.minus.gdpo.guidebench}{+9.3}
\setval{olmo.concur-free.minus.gdpo.guidebench-abs}{9.3}
\setval{olmo.concur-free.minus.gdpo.humaneval-plus}{+0.3}
\setval{olmo.concur-free.minus.gdpo.humaneval-plus-abs}{0.3}
\setval{olmo.concur-free.minus.gdpo.if}{+4.0}
\setval{olmo.concur-free.minus.gdpo.if-abs}{4.0}
\setval{olmo.concur-free.minus.gdpo.ifbench}{+2.0}
\setval{olmo.concur-free.minus.gdpo.ifbench-abs}{2.0}
\setval{olmo.concur-free.minus.gdpo.ifeval}{+0.8}
\setval{olmo.concur-free.minus.gdpo.ifeval-abs}{0.8}
\setval{olmo.concur-free.minus.gdpo.judge2.avg}{+1.1}
\setval{olmo.concur-free.minus.gdpo.judge2.avg-abs}{1.1}
\setval{olmo.concur-free.minus.gdpo.judge2.creative-writing-v3}{$-$0.3}
\setval{olmo.concur-free.minus.gdpo.judge2.creative-writing-v3-abs}{0.3}
\setval{olmo.concur-free.minus.gdpo.judge2.longbench-write}{+8.3}
\setval{olmo.concur-free.minus.gdpo.judge2.longbench-write-abs}{8.3}
\setval{olmo.concur-free.minus.gdpo.judge2.writing}{+3.0}
\setval{olmo.concur-free.minus.gdpo.judge2.writing-abs}{3.0}
\setval{olmo.concur-free.minus.gdpo.judge2.writingbench}{+1.0}
\setval{olmo.concur-free.minus.gdpo.judge2.writingbench-abs}{1.0}
\setval{olmo.concur-free.minus.gdpo.livecodebench}{+2.5}
\setval{olmo.concur-free.minus.gdpo.livecodebench-abs}{2.5}
\setval{olmo.concur-free.minus.gdpo.longbench-write}{+8.8}
\setval{olmo.concur-free.minus.gdpo.longbench-write-abs}{8.8}
\setval{olmo.concur-free.minus.gdpo.math}{$-$1.3}
\setval{olmo.concur-free.minus.gdpo.math-abs}{1.3}
\setval{olmo.concur-free.minus.gdpo.math500}{$-$0.9}
\setval{olmo.concur-free.minus.gdpo.math500-abs}{0.9}
\setval{olmo.concur-free.minus.gdpo.mbpp-plus}{$-$0.3}
\setval{olmo.concur-free.minus.gdpo.mbpp-plus-abs}{0.3}
\setval{olmo.concur-free.minus.gdpo.minerva}{$-$0.4}
\setval{olmo.concur-free.minus.gdpo.minerva-abs}{0.4}
\setval{olmo.concur-free.minus.gdpo.olympiad-bench}{$-$2.6}
\setval{olmo.concur-free.minus.gdpo.olympiad-bench-abs}{2.6}
\setval{olmo.concur-free.minus.gdpo.writing}{+5.4}
\setval{olmo.concur-free.minus.gdpo.writing-abs}{5.4}
\setval{olmo.concur-free.minus.gdpo.writingbench}{+4.9}
\setval{olmo.concur-free.minus.gdpo.writingbench-abs}{4.9}
\setval{olmo.concur-free.minus.geoma.aime24}{+3.0}
\setval{olmo.concur-free.minus.geoma.aime24-abs}{3.0}
\setval{olmo.concur-free.minus.geoma.aime25}{+0.1}
\setval{olmo.concur-free.minus.geoma.aime25-abs}{0.1}
\setval{olmo.concur-free.minus.geoma.amc23}{+3.6}
\setval{olmo.concur-free.minus.geoma.amc23-abs}{3.6}
\setval{olmo.concur-free.minus.geoma.avg}{+2.5}
\setval{olmo.concur-free.minus.geoma.avg-abs}{2.5}
\setval{olmo.concur-free.minus.geoma.code}{+1.7}
\setval{olmo.concur-free.minus.geoma.code-abs}{1.7}
\setval{olmo.concur-free.minus.geoma.creative-writing-v3}{+0.2}
\setval{olmo.concur-free.minus.geoma.creative-writing-v3-abs}{0.2}
\setval{olmo.concur-free.minus.geoma.guidebench}{+13.5}
\setval{olmo.concur-free.minus.geoma.guidebench-abs}{13.5}
\setval{olmo.concur-free.minus.geoma.humaneval-plus}{+2.0}
\setval{olmo.concur-free.minus.geoma.humaneval-plus-abs}{2.0}
\setval{olmo.concur-free.minus.geoma.if}{+4.7}
\setval{olmo.concur-free.minus.geoma.if-abs}{4.7}
\setval{olmo.concur-free.minus.geoma.ifbench}{+1.3}
\setval{olmo.concur-free.minus.geoma.ifbench-abs}{1.3}
\setval{olmo.concur-free.minus.geoma.ifeval}{$-$0.7}
\setval{olmo.concur-free.minus.geoma.ifeval-abs}{0.7}
\setval{olmo.concur-free.minus.geoma.judge2.avg}{+1.9}
\setval{olmo.concur-free.minus.geoma.judge2.avg-abs}{1.9}
\setval{olmo.concur-free.minus.geoma.judge2.creative-writing-v3}{$-$4.2}
\setval{olmo.concur-free.minus.geoma.judge2.creative-writing-v3-abs}{4.2}
\setval{olmo.concur-free.minus.geoma.judge2.longbench-write}{+5.7}
\setval{olmo.concur-free.minus.geoma.judge2.longbench-write-abs}{5.7}
\setval{olmo.concur-free.minus.geoma.judge2.writing}{+0.3}
\setval{olmo.concur-free.minus.geoma.judge2.writing-abs}{0.3}
\setval{olmo.concur-free.minus.geoma.judge2.writingbench}{$-$0.6}
\setval{olmo.concur-free.minus.geoma.judge2.writingbench-abs}{0.6}
\setval{olmo.concur-free.minus.geoma.livecodebench}{+3.3}
\setval{olmo.concur-free.minus.geoma.livecodebench-abs}{3.3}
\setval{olmo.concur-free.minus.geoma.longbench-write}{+6.7}
\setval{olmo.concur-free.minus.geoma.longbench-write-abs}{6.7}
\setval{olmo.concur-free.minus.geoma.math}{+1.4}
\setval{olmo.concur-free.minus.geoma.math-abs}{1.4}
\setval{olmo.concur-free.minus.geoma.math500}{+1.2}
\setval{olmo.concur-free.minus.geoma.math500-abs}{1.2}
\setval{olmo.concur-free.minus.geoma.mbpp-plus}{$-$0.2}
\setval{olmo.concur-free.minus.geoma.mbpp-plus-abs}{0.2}
\setval{olmo.concur-free.minus.geoma.minerva}{+0.0}
\setval{olmo.concur-free.minus.geoma.minerva-abs}{0.0}
\setval{olmo.concur-free.minus.geoma.olympiad-bench}{+0.5}
\setval{olmo.concur-free.minus.geoma.olympiad-bench-abs}{0.5}
\setval{olmo.concur-free.minus.geoma.writing}{+3.4}
\setval{olmo.concur-free.minus.geoma.writing-abs}{3.4}
\setval{olmo.concur-free.minus.geoma.writingbench}{+3.2}
\setval{olmo.concur-free.minus.geoma.writingbench-abs}{3.2}
\setval{olmo.concur-free.minus.grpo-nodiv.aime24}{$-$5.0}
\setval{olmo.concur-free.minus.grpo-nodiv.aime24-abs}{5.0}
\setval{olmo.concur-free.minus.grpo-nodiv.aime25}{$-$5.5}
\setval{olmo.concur-free.minus.grpo-nodiv.aime25-abs}{5.5}
\setval{olmo.concur-free.minus.grpo-nodiv.amc23}{$-$5.5}
\setval{olmo.concur-free.minus.grpo-nodiv.amc23-abs}{5.5}
\setval{olmo.concur-free.minus.grpo-nodiv.avg}{$-$0.5}
\setval{olmo.concur-free.minus.grpo-nodiv.avg-abs}{0.5}
\setval{olmo.concur-free.minus.grpo-nodiv.code}{+0.2}
\setval{olmo.concur-free.minus.grpo-nodiv.code-abs}{0.2}
\setval{olmo.concur-free.minus.grpo-nodiv.creative-writing-v3}{+2.8}
\setval{olmo.concur-free.minus.grpo-nodiv.creative-writing-v3-abs}{2.8}
\setval{olmo.concur-free.minus.grpo-nodiv.guidebench}{+4.7}
\setval{olmo.concur-free.minus.grpo-nodiv.guidebench-abs}{4.7}
\setval{olmo.concur-free.minus.grpo-nodiv.humaneval-plus}{$-$0.7}
\setval{olmo.concur-free.minus.grpo-nodiv.humaneval-plus-abs}{0.7}
\setval{olmo.concur-free.minus.grpo-nodiv.if}{+1.5}
\setval{olmo.concur-free.minus.grpo-nodiv.if-abs}{1.5}
\setval{olmo.concur-free.minus.grpo-nodiv.ifbench}{+0.3}
\setval{olmo.concur-free.minus.grpo-nodiv.ifbench-abs}{0.3}
\setval{olmo.concur-free.minus.grpo-nodiv.ifeval}{$-$0.5}
\setval{olmo.concur-free.minus.grpo-nodiv.ifeval-abs}{0.5}
\setval{olmo.concur-free.minus.grpo-nodiv.judge2.avg}{$-$1.2}
\setval{olmo.concur-free.minus.grpo-nodiv.judge2.avg-abs}{1.2}
\setval{olmo.concur-free.minus.grpo-nodiv.judge2.creative-writing-v3}{$-$1.3}
\setval{olmo.concur-free.minus.grpo-nodiv.judge2.creative-writing-v3-abs}{1.3}
\setval{olmo.concur-free.minus.grpo-nodiv.judge2.longbench-write}{+2.8}
\setval{olmo.concur-free.minus.grpo-nodiv.judge2.longbench-write-abs}{2.8}
\setval{olmo.concur-free.minus.grpo-nodiv.judge2.writing}{+0.9}
\setval{olmo.concur-free.minus.grpo-nodiv.judge2.writing-abs}{0.9}
\setval{olmo.concur-free.minus.grpo-nodiv.judge2.writingbench}{+1.1}
\setval{olmo.concur-free.minus.grpo-nodiv.judge2.writingbench-abs}{1.1}
\setval{olmo.concur-free.minus.grpo-nodiv.livecodebench}{+2.1}
\setval{olmo.concur-free.minus.grpo-nodiv.livecodebench-abs}{2.1}
\setval{olmo.concur-free.minus.grpo-nodiv.longbench-write}{+4.5}
\setval{olmo.concur-free.minus.grpo-nodiv.longbench-write-abs}{4.5}
\setval{olmo.concur-free.minus.grpo-nodiv.math}{$-$4.4}
\setval{olmo.concur-free.minus.grpo-nodiv.math-abs}{4.4}
\setval{olmo.concur-free.minus.grpo-nodiv.math500}{$-$3.1}
\setval{olmo.concur-free.minus.grpo-nodiv.math500-abs}{3.1}
\setval{olmo.concur-free.minus.grpo-nodiv.mbpp-plus}{$-$0.8}
\setval{olmo.concur-free.minus.grpo-nodiv.mbpp-plus-abs}{0.8}
\setval{olmo.concur-free.minus.grpo-nodiv.minerva}{$-$1.9}
\setval{olmo.concur-free.minus.grpo-nodiv.minerva-abs}{1.9}
\setval{olmo.concur-free.minus.grpo-nodiv.olympiad-bench}{$-$5.2}
\setval{olmo.concur-free.minus.grpo-nodiv.olympiad-bench-abs}{5.2}
\setval{olmo.concur-free.minus.grpo-nodiv.writing}{+4.8}
\setval{olmo.concur-free.minus.grpo-nodiv.writing-abs}{4.8}
\setval{olmo.concur-free.minus.grpo-nodiv.writingbench}{+7.0}
\setval{olmo.concur-free.minus.grpo-nodiv.writingbench-abs}{7.0}
\setval{olmo.concur-free.minus.grpo.aime24}{$-$2.7}
\setval{olmo.concur-free.minus.grpo.aime24-abs}{2.7}
\setval{olmo.concur-free.minus.grpo.aime25}{$-$2.6}
\setval{olmo.concur-free.minus.grpo.aime25-abs}{2.6}
\setval{olmo.concur-free.minus.grpo.amc23}{$-$3.4}
\setval{olmo.concur-free.minus.grpo.amc23-abs}{3.4}
\setval{olmo.concur-free.minus.grpo.avg}{+0.2}
\setval{olmo.concur-free.minus.grpo.avg-abs}{0.2}
\setval{olmo.concur-free.minus.grpo.code}{+0.4}
\setval{olmo.concur-free.minus.grpo.code-abs}{0.4}
\setval{olmo.concur-free.minus.grpo.creative-writing-v3}{+0.4}
\setval{olmo.concur-free.minus.grpo.creative-writing-v3-abs}{0.4}
\setval{olmo.concur-free.minus.grpo.guidebench}{+3.7}
\setval{olmo.concur-free.minus.grpo.guidebench-abs}{3.7}
\setval{olmo.concur-free.minus.grpo.humaneval-plus}{$-$0.7}
\setval{olmo.concur-free.minus.grpo.humaneval-plus-abs}{0.7}
\setval{olmo.concur-free.minus.grpo.if}{+1.1}
\setval{olmo.concur-free.minus.grpo.if-abs}{1.1}
\setval{olmo.concur-free.minus.grpo.ifbench}{$-$0.3}
\setval{olmo.concur-free.minus.grpo.ifbench-abs}{0.3}
\setval{olmo.concur-free.minus.grpo.ifeval}{$-$0.1}
\setval{olmo.concur-free.minus.grpo.ifeval-abs}{0.1}
\setval{olmo.concur-free.minus.grpo.judge2.avg}{$-$0.1}
\setval{olmo.concur-free.minus.grpo.judge2.avg-abs}{0.1}
\setval{olmo.concur-free.minus.grpo.judge2.creative-writing-v3}{$-$1.1}
\setval{olmo.concur-free.minus.grpo.judge2.creative-writing-v3-abs}{1.1}
\setval{olmo.concur-free.minus.grpo.judge2.longbench-write}{+6.8}
\setval{olmo.concur-free.minus.grpo.judge2.longbench-write-abs}{6.8}
\setval{olmo.concur-free.minus.grpo.judge2.writing}{+1.8}
\setval{olmo.concur-free.minus.grpo.judge2.writing-abs}{1.8}
\setval{olmo.concur-free.minus.grpo.judge2.writingbench}{$-$0.5}
\setval{olmo.concur-free.minus.grpo.judge2.writingbench-abs}{0.5}
\setval{olmo.concur-free.minus.grpo.livecodebench}{+3.3}
\setval{olmo.concur-free.minus.grpo.livecodebench-abs}{3.3}
\setval{olmo.concur-free.minus.grpo.longbench-write}{+7.8}
\setval{olmo.concur-free.minus.grpo.longbench-write-abs}{7.8}
\setval{olmo.concur-free.minus.grpo.math}{$-$1.9}
\setval{olmo.concur-free.minus.grpo.math-abs}{1.9}
\setval{olmo.concur-free.minus.grpo.math500}{$-$0.6}
\setval{olmo.concur-free.minus.grpo.math500-abs}{0.6}
\setval{olmo.concur-free.minus.grpo.mbpp-plus}{$-$1.3}
\setval{olmo.concur-free.minus.grpo.mbpp-plus-abs}{1.3}
\setval{olmo.concur-free.minus.grpo.minerva}{+0.0}
\setval{olmo.concur-free.minus.grpo.minerva-abs}{0.0}
\setval{olmo.concur-free.minus.grpo.olympiad-bench}{$-$2.4}
\setval{olmo.concur-free.minus.grpo.olympiad-bench-abs}{2.4}
\setval{olmo.concur-free.minus.grpo.writing}{+3.3}
\setval{olmo.concur-free.minus.grpo.writing-abs}{3.3}
\setval{olmo.concur-free.minus.grpo.writingbench}{+1.7}
\setval{olmo.concur-free.minus.grpo.writingbench-abs}{1.7}
\setval{olmo.concur-free.minus.mt-grpo.aime24}{$-$9.8}
\setval{olmo.concur-free.minus.mt-grpo.aime24-abs}{9.8}
\setval{olmo.concur-free.minus.mt-grpo.aime25}{$-$6.0}
\setval{olmo.concur-free.minus.mt-grpo.aime25-abs}{6.0}
\setval{olmo.concur-free.minus.mt-grpo.amc23}{$-$9.9}
\setval{olmo.concur-free.minus.mt-grpo.amc23-abs}{9.9}
\setval{olmo.concur-free.minus.mt-grpo.avg}{$-$1.3}
\setval{olmo.concur-free.minus.mt-grpo.avg-abs}{1.3}
\setval{olmo.concur-free.minus.mt-grpo.code}{+1.2}
\setval{olmo.concur-free.minus.mt-grpo.code-abs}{1.2}
\setval{olmo.concur-free.minus.mt-grpo.creative-writing-v3}{$-$0.9}
\setval{olmo.concur-free.minus.mt-grpo.creative-writing-v3-abs}{0.9}
\setval{olmo.concur-free.minus.mt-grpo.guidebench}{+5.7}
\setval{olmo.concur-free.minus.mt-grpo.guidebench-abs}{5.7}
\setval{olmo.concur-free.minus.mt-grpo.humaneval-plus}{+0.9}
\setval{olmo.concur-free.minus.mt-grpo.humaneval-plus-abs}{0.9}
\setval{olmo.concur-free.minus.mt-grpo.if}{+1.9}
\setval{olmo.concur-free.minus.mt-grpo.if-abs}{1.9}
\setval{olmo.concur-free.minus.mt-grpo.ifbench}{$-$0.3}
\setval{olmo.concur-free.minus.mt-grpo.ifbench-abs}{0.3}
\setval{olmo.concur-free.minus.mt-grpo.ifeval}{+0.2}
\setval{olmo.concur-free.minus.mt-grpo.ifeval-abs}{0.2}
\setval{olmo.concur-free.minus.mt-grpo.judge2.avg}{$-$1.6}
\setval{olmo.concur-free.minus.mt-grpo.judge2.avg-abs}{1.6}
\setval{olmo.concur-free.minus.mt-grpo.judge2.creative-writing-v3}{$-$2.3}
\setval{olmo.concur-free.minus.mt-grpo.judge2.creative-writing-v3-abs}{2.3}
\setval{olmo.concur-free.minus.mt-grpo.judge2.longbench-write}{+3.5}
\setval{olmo.concur-free.minus.mt-grpo.judge2.longbench-write-abs}{3.5}
\setval{olmo.concur-free.minus.mt-grpo.judge2.writing}{+0.6}
\setval{olmo.concur-free.minus.mt-grpo.judge2.writing-abs}{0.6}
\setval{olmo.concur-free.minus.mt-grpo.judge2.writingbench}{+0.5}
\setval{olmo.concur-free.minus.mt-grpo.judge2.writingbench-abs}{0.5}
\setval{olmo.concur-free.minus.mt-grpo.livecodebench}{+3.0}
\setval{olmo.concur-free.minus.mt-grpo.livecodebench-abs}{3.0}
\setval{olmo.concur-free.minus.mt-grpo.longbench-write}{+5.7}
\setval{olmo.concur-free.minus.mt-grpo.longbench-write-abs}{5.7}
\setval{olmo.concur-free.minus.mt-grpo.math}{$-$5.8}
\setval{olmo.concur-free.minus.mt-grpo.math-abs}{5.8}
\setval{olmo.concur-free.minus.mt-grpo.math500}{$-$2.6}
\setval{olmo.concur-free.minus.mt-grpo.math500-abs}{2.6}
\setval{olmo.concur-free.minus.mt-grpo.mbpp-plus}{$-$0.5}
\setval{olmo.concur-free.minus.mt-grpo.mbpp-plus-abs}{0.5}
\setval{olmo.concur-free.minus.mt-grpo.minerva}{$-$2.0}
\setval{olmo.concur-free.minus.mt-grpo.minerva-abs}{2.0}
\setval{olmo.concur-free.minus.mt-grpo.olympiad-bench}{$-$4.8}
\setval{olmo.concur-free.minus.mt-grpo.olympiad-bench-abs}{4.8}
\setval{olmo.concur-free.minus.mt-grpo.writing}{+2.4}
\setval{olmo.concur-free.minus.mt-grpo.writing-abs}{2.4}
\setval{olmo.concur-free.minus.mt-grpo.writingbench}{+2.5}
\setval{olmo.concur-free.minus.mt-grpo.writingbench-abs}{2.5}
\setval{olmo.concur-free.minus.papo.aime24}{+4.6}
\setval{olmo.concur-free.minus.papo.aime24-abs}{4.6}
\setval{olmo.concur-free.minus.papo.aime25}{+1.8}
\setval{olmo.concur-free.minus.papo.aime25-abs}{1.8}
\setval{olmo.concur-free.minus.papo.amc23}{+8.1}
\setval{olmo.concur-free.minus.papo.amc23-abs}{8.1}
\setval{olmo.concur-free.minus.papo.avg}{+3.1}
\setval{olmo.concur-free.minus.papo.avg-abs}{3.1}
\setval{olmo.concur-free.minus.papo.code}{+1.7}
\setval{olmo.concur-free.minus.papo.code-abs}{1.7}
\setval{olmo.concur-free.minus.papo.creative-writing-v3}{+2.8}
\setval{olmo.concur-free.minus.papo.creative-writing-v3-abs}{2.8}
\setval{olmo.concur-free.minus.papo.guidebench}{+1.5}
\setval{olmo.concur-free.minus.papo.guidebench-abs}{1.5}
\setval{olmo.concur-free.minus.papo.humaneval-plus}{+1.9}
\setval{olmo.concur-free.minus.papo.humaneval-plus-abs}{1.9}
\setval{olmo.concur-free.minus.papo.if}{+1.2}
\setval{olmo.concur-free.minus.papo.if-abs}{1.2}
\setval{olmo.concur-free.minus.papo.ifbench}{+2.0}
\setval{olmo.concur-free.minus.papo.ifbench-abs}{2.0}
\setval{olmo.concur-free.minus.papo.ifeval}{+0.1}
\setval{olmo.concur-free.minus.papo.ifeval-abs}{0.1}
\setval{olmo.concur-free.minus.papo.judge2.avg}{+2.7}
\setval{olmo.concur-free.minus.papo.judge2.avg-abs}{2.7}
\setval{olmo.concur-free.minus.papo.judge2.creative-writing-v3}{$-$0.4}
\setval{olmo.concur-free.minus.papo.judge2.creative-writing-v3-abs}{0.4}
\setval{olmo.concur-free.minus.papo.judge2.longbench-write}{+8.3}
\setval{olmo.concur-free.minus.papo.judge2.longbench-write-abs}{8.3}
\setval{olmo.concur-free.minus.papo.judge2.writing}{+2.8}
\setval{olmo.concur-free.minus.papo.judge2.writing-abs}{2.8}
\setval{olmo.concur-free.minus.papo.judge2.writingbench}{+0.5}
\setval{olmo.concur-free.minus.papo.judge2.writingbench-abs}{0.5}
\setval{olmo.concur-free.minus.papo.livecodebench}{+3.8}
\setval{olmo.concur-free.minus.papo.livecodebench-abs}{3.8}
\setval{olmo.concur-free.minus.papo.longbench-write}{+7.3}
\setval{olmo.concur-free.minus.papo.longbench-write-abs}{7.3}
\setval{olmo.concur-free.minus.papo.math}{+3.8}
\setval{olmo.concur-free.minus.papo.math-abs}{3.8}
\setval{olmo.concur-free.minus.papo.math500}{+3.5}
\setval{olmo.concur-free.minus.papo.math500-abs}{3.5}
\setval{olmo.concur-free.minus.papo.mbpp-plus}{$-$0.6}
\setval{olmo.concur-free.minus.papo.mbpp-plus-abs}{0.6}
\setval{olmo.concur-free.minus.papo.minerva}{+1.4}
\setval{olmo.concur-free.minus.papo.minerva-abs}{1.4}
\setval{olmo.concur-free.minus.papo.olympiad-bench}{+3.5}
\setval{olmo.concur-free.minus.papo.olympiad-bench-abs}{3.5}
\setval{olmo.concur-free.minus.papo.writing}{+5.0}
\setval{olmo.concur-free.minus.papo.writing-abs}{5.0}
\setval{olmo.concur-free.minus.papo.writingbench}{+4.9}
\setval{olmo.concur-free.minus.papo.writingbench-abs}{4.9}
\setval{olmo.concur-free.minus.saw.aime24}{+0.6}
\setval{olmo.concur-free.minus.saw.aime24-abs}{0.6}
\setval{olmo.concur-free.minus.saw.aime25}{$-$3.1}
\setval{olmo.concur-free.minus.saw.aime25-abs}{3.1}
\setval{olmo.concur-free.minus.saw.amc23}{+0.6}
\setval{olmo.concur-free.minus.saw.amc23-abs}{0.6}
\setval{olmo.concur-free.minus.saw.avg}{+1.0}
\setval{olmo.concur-free.minus.saw.avg-abs}{1.0}
\setval{olmo.concur-free.minus.saw.code}{+1.1}
\setval{olmo.concur-free.minus.saw.code-abs}{1.1}
\setval{olmo.concur-free.minus.saw.creative-writing-v3}{+1.9}
\setval{olmo.concur-free.minus.saw.creative-writing-v3-abs}{1.9}
\setval{olmo.concur-free.minus.saw.guidebench}{+5.5}
\setval{olmo.concur-free.minus.saw.guidebench-abs}{5.5}
\setval{olmo.concur-free.minus.saw.humaneval-plus}{+0.3}
\setval{olmo.concur-free.minus.saw.humaneval-plus-abs}{0.3}
\setval{olmo.concur-free.minus.saw.if}{+1.1}
\setval{olmo.concur-free.minus.saw.if-abs}{1.1}
\setval{olmo.concur-free.minus.saw.ifbench}{$-$1.0}
\setval{olmo.concur-free.minus.saw.ifbench-abs}{1.0}
\setval{olmo.concur-free.minus.saw.ifeval}{$-$1.3}
\setval{olmo.concur-free.minus.saw.ifeval-abs}{1.3}
\setval{olmo.concur-free.minus.saw.judge2.avg}{+0.4}
\setval{olmo.concur-free.minus.saw.judge2.avg-abs}{0.4}
\setval{olmo.concur-free.minus.saw.judge2.creative-writing-v3}{$-$3.4}
\setval{olmo.concur-free.minus.saw.judge2.creative-writing-v3-abs}{3.4}
\setval{olmo.concur-free.minus.saw.judge2.longbench-write}{+6.5}
\setval{olmo.concur-free.minus.saw.judge2.longbench-write-abs}{6.5}
\setval{olmo.concur-free.minus.saw.judge2.writing}{+0.9}
\setval{olmo.concur-free.minus.saw.judge2.writing-abs}{0.9}
\setval{olmo.concur-free.minus.saw.judge2.writingbench}{$-$0.4}
\setval{olmo.concur-free.minus.saw.judge2.writingbench-abs}{0.4}
\setval{olmo.concur-free.minus.saw.livecodebench}{+3.8}
\setval{olmo.concur-free.minus.saw.livecodebench-abs}{3.8}
\setval{olmo.concur-free.minus.saw.longbench-write}{+7.1}
\setval{olmo.concur-free.minus.saw.longbench-write-abs}{7.1}
\setval{olmo.concur-free.minus.saw.math}{$-$0.5}
\setval{olmo.concur-free.minus.saw.math-abs}{0.5}
\setval{olmo.concur-free.minus.saw.math500}{$-$0.1}
\setval{olmo.concur-free.minus.saw.math500-abs}{0.1}
\setval{olmo.concur-free.minus.saw.mbpp-plus}{$-$0.9}
\setval{olmo.concur-free.minus.saw.mbpp-plus-abs}{0.9}
\setval{olmo.concur-free.minus.saw.minerva}{$-$0.4}
\setval{olmo.concur-free.minus.saw.minerva-abs}{0.4}
\setval{olmo.concur-free.minus.saw.olympiad-bench}{$-$0.5}
\setval{olmo.concur-free.minus.saw.olympiad-bench-abs}{0.5}
\setval{olmo.concur-free.minus.saw.writing}{+3.9}
\setval{olmo.concur-free.minus.saw.writing-abs}{3.9}
\setval{olmo.concur-free.minus.saw.writingbench}{+2.8}
\setval{olmo.concur-free.minus.saw.writingbench-abs}{2.8}
\setval{olmo.concur-free.passing-at-ceiling.code.early}{0.84}
\setval{olmo.concur-free.passing-at-ceiling.code.late}{0.85}
\setval{olmo.concur-free.passing-at-ceiling.code.run}{0.84}
\setval{olmo.concur-free.passing-at-ceiling.if.early}{0.48}
\setval{olmo.concur-free.passing-at-ceiling.if.late}{0.56}
\setval{olmo.concur-free.passing-at-ceiling.if.run}{0.52}
\setval{olmo.concur-free.passing-at-ceiling.math.early}{1.00}
\setval{olmo.concur-free.passing-at-ceiling.math.late}{1.00}
\setval{olmo.concur-free.passing-at-ceiling.math.run}{1.00}
\setval{olmo.concur-free.passing-at-ceiling.writing.early}{0.00}
\setval{olmo.concur-free.passing-at-ceiling.writing.late}{0.00}
\setval{olmo.concur-free.passing-at-ceiling.writing.run}{0.00}
\setval{olmo.concur-free.primary-cosine.code.early}{0.95}
\setval{olmo.concur-free.primary-cosine.code.late}{0.91}
\setval{olmo.concur-free.primary-cosine.code.run}{0.92}
\setval{olmo.concur-free.primary-cosine.if.early}{0.82}
\setval{olmo.concur-free.primary-cosine.if.late}{0.80}
\setval{olmo.concur-free.primary-cosine.if.run}{0.81}
\setval{olmo.concur-free.primary-cosine.math.early}{0.98}
\setval{olmo.concur-free.primary-cosine.math.late}{0.90}
\setval{olmo.concur-free.primary-cosine.math.run}{0.91}
\setval{olmo.concur-free.primary-cosine.writing.early}{0.96}
\setval{olmo.concur-free.primary-cosine.writing.late}{0.93}
\setval{olmo.concur-free.primary-cosine.writing.run}{0.93}
\setval{olmo.concur-free.primary-level.code.early}{0.58}
\setval{olmo.concur-free.primary-level.code.late}{0.83}
\setval{olmo.concur-free.primary-level.code.run}{0.76}
\setval{olmo.concur-free.primary-level.if.early}{0.63}
\setval{olmo.concur-free.primary-level.if.late}{0.73}
\setval{olmo.concur-free.primary-level.if.run}{0.69}
\setval{olmo.concur-free.primary-level.math.early}{0.34}
\setval{olmo.concur-free.primary-level.math.late}{0.64}
\setval{olmo.concur-free.primary-level.math.run}{0.56}
\setval{olmo.concur-free.primary-level.writing.early}{0.50}
\setval{olmo.concur-free.primary-level.writing.late}{0.74}
\setval{olmo.concur-free.primary-level.writing.run}{0.67}
\setval{olmo.concur-free.solved.code.early}{0.14}
\setval{olmo.concur-free.solved.code.late}{0.17}
\setval{olmo.concur-free.solved.code.run}{0.16}
\setval{olmo.concur-free.solved.if.early}{0.17}
\setval{olmo.concur-free.solved.if.late}{0.18}
\setval{olmo.concur-free.solved.if.run}{0.18}
\setval{olmo.concur-free.solved.math.early}{0.07}
\setval{olmo.concur-free.solved.math.late}{0.23}
\setval{olmo.concur-free.solved.math.run}{0.18}
\setval{olmo.concur-free.solved.writing.early}{0.00}
\setval{olmo.concur-free.solved.writing.late}{0.00}
\setval{olmo.concur-free.solved.writing.run}{0.00}
\setval{olmo.concur-free.trust.code.agreement}{$-$0.042}
\setval{olmo.concur-free.trust.code.errors}{$-$9.8}
\setval{olmo.concur-free.trust.code.first-free}{--}
\setval{olmo.concur-free.trust.code.free-share}{0.00}
\setval{olmo.concur-free.trust.if.agreement}{+0.053}
\setval{olmo.concur-free.trust.if.errors}{+17.1}
\setval{olmo.concur-free.trust.if.first-free}{2}
\setval{olmo.concur-free.trust.if.free-share}{1.00}
\setval{olmo.concur-free.trust.math.first-free}{--}
\setval{olmo.concur-free.trust.math.free-share}{0.00}
\setval{olmo.concur-free.trust.writing-rm.agreement}{+0.091}
\setval{olmo.concur-free.trust.writing-rm.errors}{+44.8}
\setval{olmo.concur-free.trust.writing-rm.first-free}{2}
\setval{olmo.concur-free.trust.writing-rm.free-share}{1.00}
\setval{olmo.concur-free.trust.writing.agreement}{$-$0.066}
\setval{olmo.concur-free.trust.writing.errors}{$-$39.1}
\setval{olmo.concur-free.trust.writing.first-free}{--}
\setval{olmo.concur-free.trust.writing.free-share}{0.00}
\setval{olmo.concur-free.update-share.code.early}{0.17}
\setval{olmo.concur-free.update-share.code.late}{0.18}
\setval{olmo.concur-free.update-share.code.run}{0.17}
\setval{olmo.concur-free.update-share.if.early}{0.19}
\setval{olmo.concur-free.update-share.if.late}{0.22}
\setval{olmo.concur-free.update-share.if.run}{0.22}
\setval{olmo.concur-free.update-share.math.early}{0.19}
\setval{olmo.concur-free.update-share.math.late}{0.25}
\setval{olmo.concur-free.update-share.math.run}{0.23}
\setval{olmo.concur-free.update-share.writing.early}{0.44}
\setval{olmo.concur-free.update-share.writing.late}{0.36}
\setval{olmo.concur-free.update-share.writing.run}{0.38}
\setval{olmo.concur-tied.aux-change.code}{$-$0.21}
\setval{olmo.concur-tied.aux-change.if}{+0.10}
\setval{olmo.concur-tied.aux-change.math}{$-$0.73}
\setval{olmo.concur-tied.aux-change.writing}{$-$0.69}
\setval{olmo.concur-tied.aux-change.writing-rm}{+0.23}
\setval{olmo.concur-tied.aux-full.code.early}{0.67}
\setval{olmo.concur-tied.aux-full.code.late}{0.44}
\setval{olmo.concur-tied.aux-full.code.run}{0.53}
\setval{olmo.concur-tied.aux-full.if.early}{0.56}
\setval{olmo.concur-tied.aux-full.if.late}{0.87}
\setval{olmo.concur-tied.aux-full.if.run}{0.72}
\setval{olmo.concur-tied.aux-full.math.early}{0.51}
\setval{olmo.concur-tied.aux-full.math.late}{0.03}
\setval{olmo.concur-tied.aux-full.math.run}{0.12}
\setval{olmo.concur-tied.aux-full.writing-rm.early}{0.02}
\setval{olmo.concur-tied.aux-full.writing-rm.late}{0.06}
\setval{olmo.concur-tied.aux-full.writing-rm.run}{0.04}
\setval{olmo.concur-tied.aux-full.writing.early}{0.47}
\setval{olmo.concur-tied.aux-full.writing.late}{0.09}
\setval{olmo.concur-tied.aux-full.writing.run}{0.13}
\setval{olmo.concur-tied.aux-level.code.early}{0.86}
\setval{olmo.concur-tied.aux-level.code.late}{0.60}
\setval{olmo.concur-tied.aux-level.code.run}{0.70}
\setval{olmo.concur-tied.aux-level.if.early}{0.89}
\setval{olmo.concur-tied.aux-level.if.late}{0.97}
\setval{olmo.concur-tied.aux-level.if.run}{0.94}
\setval{olmo.concur-tied.aux-level.math.early}{0.90}
\setval{olmo.concur-tied.aux-level.math.late}{0.17}
\setval{olmo.concur-tied.aux-level.math.run}{0.39}
\setval{olmo.concur-tied.aux-level.writing-rm.early}{0.60}
\setval{olmo.concur-tied.aux-level.writing-rm.late}{0.79}
\setval{olmo.concur-tied.aux-level.writing-rm.run}{0.73}
\setval{olmo.concur-tied.aux-level.writing.early}{0.93}
\setval{olmo.concur-tied.aux-level.writing.late}{0.26}
\setval{olmo.concur-tied.aux-level.writing.run}{0.46}
\setval{olmo.concur-tied.bench.aime24}{34.5}
\setval{olmo.concur-tied.bench.aime25}{29.9}
\setval{olmo.concur-tied.bench.amc23}{79.8}
\setval{olmo.concur-tied.bench.creative-writing-v3}{49.0}
\setval{olmo.concur-tied.bench.guidebench}{58.5}
\setval{olmo.concur-tied.bench.humaneval-plus}{83.3}
\setval{olmo.concur-tied.bench.ifbench}{33.7}
\setval{olmo.concur-tied.bench.ifeval}{87.6}
\setval{olmo.concur-tied.bench.livecodebench}{29.0}
\setval{olmo.concur-tied.bench.longbench-write}{69.2}
\setval{olmo.concur-tied.bench.math500}{89.4}
\setval{olmo.concur-tied.bench.mbpp-plus}{71.1}
\setval{olmo.concur-tied.bench.minerva}{44.0}
\setval{olmo.concur-tied.bench.olympiad-bench}{61.6}
\setval{olmo.concur-tied.bench.writingbench}{66.1}
\setval{olmo.concur-tied.best-count.all}{10}
\setval{olmo.concur-tied.best-count.code}{3}
\setval{olmo.concur-tied.best-count.if}{1}
\setval{olmo.concur-tied.best-count.math}{5}
\setval{olmo.concur-tied.best-count.writing}{1}
\setval{olmo.concur-tied.eval.avg}{59.1}
\setval{olmo.concur-tied.eval.code}{61.1}
\setval{olmo.concur-tied.eval.if}{60.0}
\setval{olmo.concur-tied.eval.math}{56.5}
\setval{olmo.concur-tied.eval.writing}{61.5}
\setval{olmo.concur-tied.gate.code.against.early}{0.06}
\setval{olmo.concur-tied.gate.code.against.late}{0.10}
\setval{olmo.concur-tied.gate.code.against.run}{0.11}
\setval{olmo.concur-tied.gate.code.agreement}{$-$0.010}
\setval{olmo.concur-tied.gate.code.ceiling.early}{0.28}
\setval{olmo.concur-tied.gate.code.ceiling.late}{0.72}
\setval{olmo.concur-tied.gate.code.ceiling.run}{0.57}
\setval{olmo.concur-tied.gate.code.cosine.early}{0.70}
\setval{olmo.concur-tied.gate.code.cosine.late}{0.15}
\setval{olmo.concur-tied.gate.code.cosine.run}{0.33}
\setval{olmo.concur-tied.gate.if.against.early}{0.10}
\setval{olmo.concur-tied.gate.if.against.late}{0.07}
\setval{olmo.concur-tied.gate.if.against.run}{0.09}
\setval{olmo.concur-tied.gate.if.agreement}{+0.064}
\setval{olmo.concur-tied.gate.if.ceiling.early}{0.23}
\setval{olmo.concur-tied.gate.if.ceiling.late}{0.27}
\setval{olmo.concur-tied.gate.if.ceiling.run}{0.26}
\setval{olmo.concur-tied.gate.if.cosine.early}{0.51}
\setval{olmo.concur-tied.gate.if.cosine.late}{0.48}
\setval{olmo.concur-tied.gate.if.cosine.run}{0.48}
\setval{olmo.concur-tied.gate.math.against.early}{0.00}
\setval{olmo.concur-tied.gate.math.against.late}{0.00}
\setval{olmo.concur-tied.gate.math.against.run}{0.00}
\setval{olmo.concur-tied.gate.math.ceiling.early}{0.15}
\setval{olmo.concur-tied.gate.math.ceiling.late}{0.88}
\setval{olmo.concur-tied.gate.math.ceiling.run}{0.70}
\setval{olmo.concur-tied.gate.math.cosine.early}{0.90}
\setval{olmo.concur-tied.gate.math.cosine.late}{0.16}
\setval{olmo.concur-tied.gate.math.cosine.run}{0.37}
\setval{olmo.concur-tied.gate.writing-rm.against.early}{0.22}
\setval{olmo.concur-tied.gate.writing-rm.against.late}{0.55}
\setval{olmo.concur-tied.gate.writing-rm.against.run}{0.47}
\setval{olmo.concur-tied.gate.writing-rm.agreement}{+0.093}
\setval{olmo.concur-tied.gate.writing-rm.ceiling.early}{0.00}
\setval{olmo.concur-tied.gate.writing-rm.ceiling.late}{0.00}
\setval{olmo.concur-tied.gate.writing-rm.ceiling.run}{0.00}
\setval{olmo.concur-tied.gate.writing-rm.cosine.early}{0.69}
\setval{olmo.concur-tied.gate.writing-rm.cosine.late}{0.20}
\setval{olmo.concur-tied.gate.writing-rm.cosine.run}{0.33}
\setval{olmo.concur-tied.gate.writing.against.early}{0.27}
\setval{olmo.concur-tied.gate.writing.against.late}{0.67}
\setval{olmo.concur-tied.gate.writing.against.run}{0.63}
\setval{olmo.concur-tied.gate.writing.agreement}{$-$0.063}
\setval{olmo.concur-tied.gate.writing.ceiling.early}{0.00}
\setval{olmo.concur-tied.gate.writing.ceiling.late}{0.00}
\setval{olmo.concur-tied.gate.writing.ceiling.run}{0.00}
\setval{olmo.concur-tied.gate.writing.cosine.early}{0.57}
\setval{olmo.concur-tied.gate.writing.cosine.late}{0.01}
\setval{olmo.concur-tied.gate.writing.cosine.run}{0.07}
\setval{olmo.concur-tied.judge2.avg}{56.4}
\setval{olmo.concur-tied.judge2.creative-writing-v3}{37.2}
\setval{olmo.concur-tied.judge2.longbench-write}{60.5}
\setval{olmo.concur-tied.judge2.writing}{47.8}
\setval{olmo.concur-tied.judge2.writingbench}{45.6}
\setval{olmo.concur-tied.last-step}{299}
\setval{olmo.concur-tied.minus.base.aime24}{+25.9}
\setval{olmo.concur-tied.minus.base.aime24-abs}{25.9}
\setval{olmo.concur-tied.minus.base.aime25}{+22.5}
\setval{olmo.concur-tied.minus.base.aime25-abs}{22.5}
\setval{olmo.concur-tied.minus.base.amc23}{+35.3}
\setval{olmo.concur-tied.minus.base.amc23-abs}{35.3}
\setval{olmo.concur-tied.minus.base.avg}{+15.0}
\setval{olmo.concur-tied.minus.base.avg-abs}{15.0}
\setval{olmo.concur-tied.minus.base.code}{+12.0}
\setval{olmo.concur-tied.minus.base.code-abs}{12.0}
\setval{olmo.concur-tied.minus.base.creative-writing-v3}{+11.1}
\setval{olmo.concur-tied.minus.base.creative-writing-v3-abs}{11.1}
\setval{olmo.concur-tied.minus.base.guidebench}{$-$2.9}
\setval{olmo.concur-tied.minus.base.guidebench-abs}{2.9}
\setval{olmo.concur-tied.minus.base.humaneval-plus}{+14.1}
\setval{olmo.concur-tied.minus.base.humaneval-plus-abs}{14.1}
\setval{olmo.concur-tied.minus.base.if}{+2.5}
\setval{olmo.concur-tied.minus.base.if-abs}{2.5}
\setval{olmo.concur-tied.minus.base.ifbench}{+7.0}
\setval{olmo.concur-tied.minus.base.ifbench-abs}{7.0}
\setval{olmo.concur-tied.minus.base.ifeval}{+3.2}
\setval{olmo.concur-tied.minus.base.ifeval-abs}{3.2}
\setval{olmo.concur-tied.minus.base.judge2.avg}{+13.4}
\setval{olmo.concur-tied.minus.base.judge2.avg-abs}{13.4}
\setval{olmo.concur-tied.minus.base.judge2.creative-writing-v3}{+5.5}
\setval{olmo.concur-tied.minus.base.judge2.creative-writing-v3-abs}{5.5}
\setval{olmo.concur-tied.minus.base.judge2.longbench-write}{+7.3}
\setval{olmo.concur-tied.minus.base.judge2.longbench-write-abs}{7.3}
\setval{olmo.concur-tied.minus.base.judge2.writing}{+6.9}
\setval{olmo.concur-tied.minus.base.judge2.writing-abs}{6.9}
\setval{olmo.concur-tied.minus.base.judge2.writingbench}{+7.9}
\setval{olmo.concur-tied.minus.base.judge2.writingbench-abs}{7.9}
\setval{olmo.concur-tied.minus.base.livecodebench}{+12.2}
\setval{olmo.concur-tied.minus.base.livecodebench-abs}{12.2}
\setval{olmo.concur-tied.minus.base.longbench-write}{+11.8}
\setval{olmo.concur-tied.minus.base.longbench-write-abs}{11.8}
\setval{olmo.concur-tied.minus.base.math}{+22.8}
\setval{olmo.concur-tied.minus.base.math-abs}{22.8}
\setval{olmo.concur-tied.minus.base.math500}{+19.3}
\setval{olmo.concur-tied.minus.base.math500-abs}{19.3}
\setval{olmo.concur-tied.minus.base.mbpp-plus}{+9.8}
\setval{olmo.concur-tied.minus.base.mbpp-plus-abs}{9.8}
\setval{olmo.concur-tied.minus.base.minerva}{+8.8}
\setval{olmo.concur-tied.minus.base.minerva-abs}{8.8}
\setval{olmo.concur-tied.minus.base.olympiad-bench}{+25.3}
\setval{olmo.concur-tied.minus.base.olympiad-bench-abs}{25.3}
\setval{olmo.concur-tied.minus.base.writing}{+14.8}
\setval{olmo.concur-tied.minus.base.writing-abs}{14.8}
\setval{olmo.concur-tied.minus.base.writingbench}{+21.4}
\setval{olmo.concur-tied.minus.base.writingbench-abs}{21.4}
\setval{olmo.concur-tied.minus.best.bench.aime24}{$-$2.1}
\setval{olmo.concur-tied.minus.best.bench.aime25}{+1.8}
\setval{olmo.concur-tied.minus.best.bench.amc23}{+2.4}
\setval{olmo.concur-tied.minus.best.bench.creative-writing-v3}{$-$1.7}
\setval{olmo.concur-tied.minus.best.bench.guidebench}{$-$2.6}
\setval{olmo.concur-tied.minus.best.bench.humaneval-plus}{+6.0}
\setval{olmo.concur-tied.minus.best.bench.ifbench}{+1.0}
\setval{olmo.concur-tied.minus.best.bench.ifeval}{$-$1.0}
\setval{olmo.concur-tied.minus.best.bench.livecodebench}{+5.8}
\setval{olmo.concur-tied.minus.best.bench.longbench-write}{+1.0}
\setval{olmo.concur-tied.minus.best.bench.math500}{+2.5}
\setval{olmo.concur-tied.minus.best.bench.mbpp-plus}{+5.3}
\setval{olmo.concur-tied.minus.best.bench.minerva}{+2.1}
\setval{olmo.concur-tied.minus.best.bench.olympiad-bench}{+3.5}
\setval{olmo.concur-tied.minus.best.bench.writingbench}{$-$2.3}
\setval{olmo.concur-tied.minus.best.eval.avg}{+2.4}
\setval{olmo.concur-tied.minus.best.eval.code}{+6.0}
\setval{olmo.concur-tied.minus.best.eval.if}{$-$0.4}
\setval{olmo.concur-tied.minus.best.eval.math}{+1.7}
\setval{olmo.concur-tied.minus.best.eval.writing}{$-$0.1}
\setval{olmo.concur-tied.minus.concur-free.aime24}{+7.7}
\setval{olmo.concur-tied.minus.concur-free.aime24-abs}{7.7}
\setval{olmo.concur-tied.minus.concur-free.aime25}{+7.8}
\setval{olmo.concur-tied.minus.concur-free.aime25-abs}{7.8}
\setval{olmo.concur-tied.minus.concur-free.amc23}{+12.3}
\setval{olmo.concur-tied.minus.concur-free.amc23-abs}{12.3}
\setval{olmo.concur-tied.minus.concur-free.avg}{+3.7}
\setval{olmo.concur-tied.minus.concur-free.avg-abs}{3.7}
\setval{olmo.concur-tied.minus.concur-free.code}{+5.5}
\setval{olmo.concur-tied.minus.concur-free.code-abs}{5.5}
\setval{olmo.concur-tied.minus.concur-free.creative-writing-v3}{$-$0.7}
\setval{olmo.concur-tied.minus.concur-free.creative-writing-v3-abs}{0.7}
\setval{olmo.concur-tied.minus.concur-free.guidebench}{$-$1.6}
\setval{olmo.concur-tied.minus.concur-free.guidebench-abs}{1.6}
\setval{olmo.concur-tied.minus.concur-free.humaneval-plus}{+6.6}
\setval{olmo.concur-tied.minus.concur-free.humaneval-plus-abs}{6.6}
\setval{olmo.concur-tied.minus.concur-free.if}{+0.2}
\setval{olmo.concur-tied.minus.concur-free.if-abs}{0.2}
\setval{olmo.concur-tied.minus.concur-free.ifbench}{+2.0}
\setval{olmo.concur-tied.minus.concur-free.ifbench-abs}{2.0}
\setval{olmo.concur-tied.minus.concur-free.ifeval}{+0.4}
\setval{olmo.concur-tied.minus.concur-free.ifeval-abs}{0.4}
\setval{olmo.concur-tied.minus.concur-free.judge2.avg}{+3.8}
\setval{olmo.concur-tied.minus.concur-free.judge2.avg-abs}{3.8}
\setval{olmo.concur-tied.minus.concur-free.judge2.creative-writing-v3}{$-$0.4}
\setval{olmo.concur-tied.minus.concur-free.judge2.creative-writing-v3-abs}{0.4}
\setval{olmo.concur-tied.minus.concur-free.judge2.longbench-write}{$-$3.4}
\setval{olmo.concur-tied.minus.concur-free.judge2.longbench-write-abs}{3.4}
\setval{olmo.concur-tied.minus.concur-free.judge2.writing}{$-$2.1}
\setval{olmo.concur-tied.minus.concur-free.judge2.writing-abs}{2.1}
\setval{olmo.concur-tied.minus.concur-free.judge2.writingbench}{$-$2.4}
\setval{olmo.concur-tied.minus.concur-free.judge2.writingbench-abs}{2.4}
\setval{olmo.concur-tied.minus.concur-free.livecodebench}{+3.4}
\setval{olmo.concur-tied.minus.concur-free.livecodebench-abs}{3.4}
\setval{olmo.concur-tied.minus.concur-free.longbench-write}{$-$2.8}
\setval{olmo.concur-tied.minus.concur-free.longbench-write-abs}{2.8}
\setval{olmo.concur-tied.minus.concur-free.math}{+7.6}
\setval{olmo.concur-tied.minus.concur-free.math-abs}{7.6}
\setval{olmo.concur-tied.minus.concur-free.math500}{+5.0}
\setval{olmo.concur-tied.minus.concur-free.math500-abs}{5.0}
\setval{olmo.concur-tied.minus.concur-free.mbpp-plus}{+6.6}
\setval{olmo.concur-tied.minus.concur-free.mbpp-plus-abs}{6.6}
\setval{olmo.concur-tied.minus.concur-free.minerva}{+4.1}
\setval{olmo.concur-tied.minus.concur-free.minerva-abs}{4.1}
\setval{olmo.concur-tied.minus.concur-free.olympiad-bench}{+8.3}
\setval{olmo.concur-tied.minus.concur-free.olympiad-bench-abs}{8.3}
\setval{olmo.concur-tied.minus.concur-free.writing}{$-$2.5}
\setval{olmo.concur-tied.minus.concur-free.writing-abs}{2.5}
\setval{olmo.concur-tied.minus.concur-free.writingbench}{$-$4.1}
\setval{olmo.concur-tied.minus.concur-free.writingbench-abs}{4.1}
\setval{olmo.concur-tied.minus.concur.aime24}{$-$1.9}
\setval{olmo.concur-tied.minus.concur.aime24-abs}{1.9}
\setval{olmo.concur-tied.minus.concur.aime25}{$-$2.3}
\setval{olmo.concur-tied.minus.concur.aime25-abs}{2.3}
\setval{olmo.concur-tied.minus.concur.amc23}{$-$3.7}
\setval{olmo.concur-tied.minus.concur.amc23-abs}{3.7}
\setval{olmo.concur-tied.minus.concur.avg}{$-$0.2}
\setval{olmo.concur-tied.minus.concur.avg-abs}{0.2}
\setval{olmo.concur-tied.minus.concur.code}{+0.9}
\setval{olmo.concur-tied.minus.concur.code-abs}{0.9}
\setval{olmo.concur-tied.minus.concur.creative-writing-v3}{$-$0.2}
\setval{olmo.concur-tied.minus.concur.creative-writing-v3-abs}{0.2}
\setval{olmo.concur-tied.minus.concur.guidebench}{+1.6}
\setval{olmo.concur-tied.minus.concur.guidebench-abs}{1.6}
\setval{olmo.concur-tied.minus.concur.humaneval-plus}{+0.5}
\setval{olmo.concur-tied.minus.concur.humaneval-plus-abs}{0.5}
\setval{olmo.concur-tied.minus.concur.if}{+1.1}
\setval{olmo.concur-tied.minus.concur.if-abs}{1.1}
\setval{olmo.concur-tied.minus.concur.ifbench}{+1.7}
\setval{olmo.concur-tied.minus.concur.ifbench-abs}{1.7}
\setval{olmo.concur-tied.minus.concur.ifeval}{$-$0.1}
\setval{olmo.concur-tied.minus.concur.ifeval-abs}{0.1}
\setval{olmo.concur-tied.minus.concur.judge2.avg}{$-$0.2}
\setval{olmo.concur-tied.minus.concur.judge2.avg-abs}{0.2}
\setval{olmo.concur-tied.minus.concur.judge2.creative-writing-v3}{$-$1.3}
\setval{olmo.concur-tied.minus.concur.judge2.creative-writing-v3-abs}{1.3}
\setval{olmo.concur-tied.minus.concur.judge2.longbench-write}{+2.1}
\setval{olmo.concur-tied.minus.concur.judge2.longbench-write-abs}{2.1}
\setval{olmo.concur-tied.minus.concur.judge2.writing}{+0.5}
\setval{olmo.concur-tied.minus.concur.judge2.writing-abs}{0.5}
\setval{olmo.concur-tied.minus.concur.judge2.writingbench}{+0.7}
\setval{olmo.concur-tied.minus.concur.judge2.writingbench-abs}{0.7}
\setval{olmo.concur-tied.minus.concur.livecodebench}{$-$1.3}
\setval{olmo.concur-tied.minus.concur.livecodebench-abs}{1.3}
\setval{olmo.concur-tied.minus.concur.longbench-write}{+0.7}
\setval{olmo.concur-tied.minus.concur.longbench-write-abs}{0.7}
\setval{olmo.concur-tied.minus.concur.math}{$-$1.8}
\setval{olmo.concur-tied.minus.concur.math-abs}{1.8}
\setval{olmo.concur-tied.minus.concur.math500}{$-$1.1}
\setval{olmo.concur-tied.minus.concur.math500-abs}{1.1}
\setval{olmo.concur-tied.minus.concur.mbpp-plus}{+3.5}
\setval{olmo.concur-tied.minus.concur.mbpp-plus-abs}{3.5}
\setval{olmo.concur-tied.minus.concur.minerva}{$-$0.8}
\setval{olmo.concur-tied.minus.concur.minerva-abs}{0.8}
\setval{olmo.concur-tied.minus.concur.olympiad-bench}{$-$0.8}
\setval{olmo.concur-tied.minus.concur.olympiad-bench-abs}{0.8}
\setval{olmo.concur-tied.minus.concur.writing}{+0.4}
\setval{olmo.concur-tied.minus.concur.writing-abs}{0.4}
\setval{olmo.concur-tied.minus.concur.writingbench}{+0.7}
\setval{olmo.concur-tied.minus.concur.writingbench-abs}{0.7}
\setval{olmo.concur-tied.minus.gd2po.aime24}{+6.1}
\setval{olmo.concur-tied.minus.gd2po.aime24-abs}{6.1}
\setval{olmo.concur-tied.minus.gd2po.aime25}{+5.6}
\setval{olmo.concur-tied.minus.gd2po.aime25-abs}{5.6}
\setval{olmo.concur-tied.minus.gd2po.amc23}{+7.7}
\setval{olmo.concur-tied.minus.gd2po.amc23-abs}{7.7}
\setval{olmo.concur-tied.minus.gd2po.avg}{+3.5}
\setval{olmo.concur-tied.minus.gd2po.avg-abs}{3.5}
\setval{olmo.concur-tied.minus.gd2po.code}{+6.2}
\setval{olmo.concur-tied.minus.gd2po.code-abs}{6.2}
\setval{olmo.concur-tied.minus.gd2po.creative-writing-v3}{+0.7}
\setval{olmo.concur-tied.minus.gd2po.creative-writing-v3-abs}{0.7}
\setval{olmo.concur-tied.minus.gd2po.guidebench}{$-$2.6}
\setval{olmo.concur-tied.minus.gd2po.guidebench-abs}{2.6}
\setval{olmo.concur-tied.minus.gd2po.humaneval-plus}{+6.5}
\setval{olmo.concur-tied.minus.gd2po.humaneval-plus-abs}{6.5}
\setval{olmo.concur-tied.minus.gd2po.if}{$-$0.4}
\setval{olmo.concur-tied.minus.gd2po.if-abs}{0.4}
\setval{olmo.concur-tied.minus.gd2po.ifbench}{+1.3}
\setval{olmo.concur-tied.minus.gd2po.ifbench-abs}{1.3}
\setval{olmo.concur-tied.minus.gd2po.ifeval}{+0.0}
\setval{olmo.concur-tied.minus.gd2po.ifeval-abs}{0.0}
\setval{olmo.concur-tied.minus.gd2po.judge2.avg}{+3.1}
\setval{olmo.concur-tied.minus.gd2po.judge2.avg-abs}{3.1}
\setval{olmo.concur-tied.minus.gd2po.judge2.creative-writing-v3}{$-$0.7}
\setval{olmo.concur-tied.minus.gd2po.judge2.creative-writing-v3-abs}{0.7}
\setval{olmo.concur-tied.minus.gd2po.judge2.longbench-write}{+0.6}
\setval{olmo.concur-tied.minus.gd2po.judge2.longbench-write-abs}{0.6}
\setval{olmo.concur-tied.minus.gd2po.judge2.writing}{$-$1.0}
\setval{olmo.concur-tied.minus.gd2po.judge2.writing-abs}{1.0}
\setval{olmo.concur-tied.minus.gd2po.judge2.writingbench}{$-$2.8}
\setval{olmo.concur-tied.minus.gd2po.judge2.writingbench-abs}{2.8}
\setval{olmo.concur-tied.minus.gd2po.livecodebench}{+5.8}
\setval{olmo.concur-tied.minus.gd2po.livecodebench-abs}{5.8}
\setval{olmo.concur-tied.minus.gd2po.longbench-write}{+1.0}
\setval{olmo.concur-tied.minus.gd2po.longbench-write-abs}{1.0}
\setval{olmo.concur-tied.minus.gd2po.math}{+5.4}
\setval{olmo.concur-tied.minus.gd2po.math-abs}{5.4}
\setval{olmo.concur-tied.minus.gd2po.math500}{+3.7}
\setval{olmo.concur-tied.minus.gd2po.math500-abs}{3.7}
\setval{olmo.concur-tied.minus.gd2po.mbpp-plus}{+6.3}
\setval{olmo.concur-tied.minus.gd2po.mbpp-plus-abs}{6.3}
\setval{olmo.concur-tied.minus.gd2po.minerva}{+3.9}
\setval{olmo.concur-tied.minus.gd2po.minerva-abs}{3.9}
\setval{olmo.concur-tied.minus.gd2po.olympiad-bench}{+5.5}
\setval{olmo.concur-tied.minus.gd2po.olympiad-bench-abs}{5.5}
\setval{olmo.concur-tied.minus.gd2po.writing}{+0.7}
\setval{olmo.concur-tied.minus.gd2po.writing-abs}{0.7}
\setval{olmo.concur-tied.minus.gd2po.writingbench}{+0.4}
\setval{olmo.concur-tied.minus.gd2po.writingbench-abs}{0.4}
\setval{olmo.concur-tied.minus.gdpo-noaux.aime24}{$-$3.9}
\setval{olmo.concur-tied.minus.gdpo-noaux.aime24-abs}{3.9}
\setval{olmo.concur-tied.minus.gdpo-noaux.aime25}{$-$3.8}
\setval{olmo.concur-tied.minus.gdpo-noaux.aime25-abs}{3.8}
\setval{olmo.concur-tied.minus.gdpo-noaux.amc23}{$-$4.1}
\setval{olmo.concur-tied.minus.gdpo-noaux.amc23-abs}{4.1}
\setval{olmo.concur-tied.minus.gdpo-noaux.avg}{+1.3}
\setval{olmo.concur-tied.minus.gdpo-noaux.avg-abs}{1.3}
\setval{olmo.concur-tied.minus.gdpo-noaux.code}{$-$1.3}
\setval{olmo.concur-tied.minus.gdpo-noaux.code-abs}{1.3}
\setval{olmo.concur-tied.minus.gdpo-noaux.creative-writing-v3}{+10.8}
\setval{olmo.concur-tied.minus.gdpo-noaux.creative-writing-v3-abs}{10.8}
\setval{olmo.concur-tied.minus.gdpo-noaux.guidebench}{+2.8}
\setval{olmo.concur-tied.minus.gdpo-noaux.guidebench-abs}{2.8}
\setval{olmo.concur-tied.minus.gdpo-noaux.humaneval-plus}{$-$1.9}
\setval{olmo.concur-tied.minus.gdpo-noaux.humaneval-plus-abs}{1.9}
\setval{olmo.concur-tied.minus.gdpo-noaux.if}{+2.0}
\setval{olmo.concur-tied.minus.gdpo-noaux.if-abs}{2.0}
\setval{olmo.concur-tied.minus.gdpo-noaux.ifbench}{+0.3}
\setval{olmo.concur-tied.minus.gdpo-noaux.ifbench-abs}{0.3}
\setval{olmo.concur-tied.minus.gdpo-noaux.ifeval}{+2.9}
\setval{olmo.concur-tied.minus.gdpo-noaux.ifeval-abs}{2.9}
\setval{olmo.concur-tied.minus.gdpo-noaux.judge2.avg}{+0.7}
\setval{olmo.concur-tied.minus.gdpo-noaux.judge2.avg-abs}{0.7}
\setval{olmo.concur-tied.minus.gdpo-noaux.judge2.creative-writing-v3}{+6.2}
\setval{olmo.concur-tied.minus.gdpo-noaux.judge2.creative-writing-v3-abs}{6.2}
\setval{olmo.concur-tied.minus.gdpo-noaux.judge2.longbench-write}{+13.9}
\setval{olmo.concur-tied.minus.gdpo-noaux.judge2.longbench-write-abs}{13.9}
\setval{olmo.concur-tied.minus.gdpo-noaux.judge2.writing}{+8.7}
\setval{olmo.concur-tied.minus.gdpo-noaux.judge2.writing-abs}{8.7}
\setval{olmo.concur-tied.minus.gdpo-noaux.judge2.writingbench}{+5.8}
\setval{olmo.concur-tied.minus.gdpo-noaux.judge2.writingbench-abs}{5.8}
\setval{olmo.concur-tied.minus.gdpo-noaux.livecodebench}{$-$2.5}
\setval{olmo.concur-tied.minus.gdpo-noaux.livecodebench-abs}{2.5}
\setval{olmo.concur-tied.minus.gdpo-noaux.longbench-write}{+13.3}
\setval{olmo.concur-tied.minus.gdpo-noaux.longbench-write-abs}{13.3}
\setval{olmo.concur-tied.minus.gdpo-noaux.math}{$-$2.9}
\setval{olmo.concur-tied.minus.gdpo-noaux.math-abs}{2.9}
\setval{olmo.concur-tied.minus.gdpo-noaux.math500}{$-$1.5}
\setval{olmo.concur-tied.minus.gdpo-noaux.math500-abs}{1.5}
\setval{olmo.concur-tied.minus.gdpo-noaux.mbpp-plus}{+0.6}
\setval{olmo.concur-tied.minus.gdpo-noaux.mbpp-plus-abs}{0.6}
\setval{olmo.concur-tied.minus.gdpo-noaux.minerva}{$-$2.1}
\setval{olmo.concur-tied.minus.gdpo-noaux.minerva-abs}{2.1}
\setval{olmo.concur-tied.minus.gdpo-noaux.olympiad-bench}{$-$2.2}
\setval{olmo.concur-tied.minus.gdpo-noaux.olympiad-bench-abs}{2.2}
\setval{olmo.concur-tied.minus.gdpo-noaux.writing}{+11.6}
\setval{olmo.concur-tied.minus.gdpo-noaux.writing-abs}{11.6}
\setval{olmo.concur-tied.minus.gdpo-noaux.writingbench}{+10.7}
\setval{olmo.concur-tied.minus.gdpo-noaux.writingbench-abs}{10.7}
\setval{olmo.concur-tied.minus.gdpo-w0.5.aime24}{+4.9}
\setval{olmo.concur-tied.minus.gdpo-w0.5.aime24-abs}{4.9}
\setval{olmo.concur-tied.minus.gdpo-w0.5.aime25}{+2.6}
\setval{olmo.concur-tied.minus.gdpo-w0.5.aime25-abs}{2.6}
\setval{olmo.concur-tied.minus.gdpo-w0.5.amc23}{+5.9}
\setval{olmo.concur-tied.minus.gdpo-w0.5.amc23-abs}{5.9}
\setval{olmo.concur-tied.minus.gdpo-w0.5.avg}{+3.5}
\setval{olmo.concur-tied.minus.gdpo-w0.5.avg-abs}{3.5}
\setval{olmo.concur-tied.minus.gdpo-w0.5.code}{+5.3}
\setval{olmo.concur-tied.minus.gdpo-w0.5.code-abs}{5.3}
\setval{olmo.concur-tied.minus.gdpo-w0.5.creative-writing-v3}{+0.9}
\setval{olmo.concur-tied.minus.gdpo-w0.5.creative-writing-v3-abs}{0.9}
\setval{olmo.concur-tied.minus.gdpo-w0.5.guidebench}{+6.6}
\setval{olmo.concur-tied.minus.gdpo-w0.5.guidebench-abs}{6.6}
\setval{olmo.concur-tied.minus.gdpo-w0.5.humaneval-plus}{+5.5}
\setval{olmo.concur-tied.minus.gdpo-w0.5.humaneval-plus-abs}{5.5}
\setval{olmo.concur-tied.minus.gdpo-w0.5.if}{+2.6}
\setval{olmo.concur-tied.minus.gdpo-w0.5.if-abs}{2.6}
\setval{olmo.concur-tied.minus.gdpo-w0.5.ifbench}{+1.0}
\setval{olmo.concur-tied.minus.gdpo-w0.5.ifbench-abs}{1.0}
\setval{olmo.concur-tied.minus.gdpo-w0.5.ifeval}{+0.1}
\setval{olmo.concur-tied.minus.gdpo-w0.5.ifeval-abs}{0.1}
\setval{olmo.concur-tied.minus.gdpo-w0.5.judge2.avg}{+3.1}
\setval{olmo.concur-tied.minus.gdpo-w0.5.judge2.avg-abs}{3.1}
\setval{olmo.concur-tied.minus.gdpo-w0.5.judge2.creative-writing-v3}{$-$2.2}
\setval{olmo.concur-tied.minus.gdpo-w0.5.judge2.creative-writing-v3-abs}{2.2}
\setval{olmo.concur-tied.minus.gdpo-w0.5.judge2.longbench-write}{+3.9}
\setval{olmo.concur-tied.minus.gdpo-w0.5.judge2.longbench-write-abs}{3.9}
\setval{olmo.concur-tied.minus.gdpo-w0.5.judge2.writing}{+0.2}
\setval{olmo.concur-tied.minus.gdpo-w0.5.judge2.writing-abs}{0.2}
\setval{olmo.concur-tied.minus.gdpo-w0.5.judge2.writingbench}{$-$1.1}
\setval{olmo.concur-tied.minus.gdpo-w0.5.judge2.writingbench-abs}{1.1}
\setval{olmo.concur-tied.minus.gdpo-w0.5.livecodebench}{+4.2}
\setval{olmo.concur-tied.minus.gdpo-w0.5.livecodebench-abs}{4.2}
\setval{olmo.concur-tied.minus.gdpo-w0.5.longbench-write}{+5.5}
\setval{olmo.concur-tied.minus.gdpo-w0.5.longbench-write-abs}{5.5}
\setval{olmo.concur-tied.minus.gdpo-w0.5.math}{+3.7}
\setval{olmo.concur-tied.minus.gdpo-w0.5.math-abs}{3.7}
\setval{olmo.concur-tied.minus.gdpo-w0.5.math500}{+2.4}
\setval{olmo.concur-tied.minus.gdpo-w0.5.math500-abs}{2.4}
\setval{olmo.concur-tied.minus.gdpo-w0.5.mbpp-plus}{+6.1}
\setval{olmo.concur-tied.minus.gdpo-w0.5.mbpp-plus-abs}{6.1}
\setval{olmo.concur-tied.minus.gdpo-w0.5.minerva}{+2.5}
\setval{olmo.concur-tied.minus.gdpo-w0.5.minerva-abs}{2.5}
\setval{olmo.concur-tied.minus.gdpo-w0.5.olympiad-bench}{+4.1}
\setval{olmo.concur-tied.minus.gdpo-w0.5.olympiad-bench-abs}{4.1}
\setval{olmo.concur-tied.minus.gdpo-w0.5.writing}{+2.0}
\setval{olmo.concur-tied.minus.gdpo-w0.5.writing-abs}{2.0}
\setval{olmo.concur-tied.minus.gdpo-w0.5.writingbench}{$-$0.4}
\setval{olmo.concur-tied.minus.gdpo-w0.5.writingbench-abs}{0.4}
\setval{olmo.concur-tied.minus.gdpo.aime24}{+6.7}
\setval{olmo.concur-tied.minus.gdpo.aime24-abs}{6.7}
\setval{olmo.concur-tied.minus.gdpo.aime25}{+7.3}
\setval{olmo.concur-tied.minus.gdpo.aime25-abs}{7.3}
\setval{olmo.concur-tied.minus.gdpo.amc23}{+10.3}
\setval{olmo.concur-tied.minus.gdpo.amc23-abs}{10.3}
\setval{olmo.concur-tied.minus.gdpo.avg}{+5.2}
\setval{olmo.concur-tied.minus.gdpo.avg-abs}{5.2}
\setval{olmo.concur-tied.minus.gdpo.code}{+6.4}
\setval{olmo.concur-tied.minus.gdpo.code-abs}{6.4}
\setval{olmo.concur-tied.minus.gdpo.creative-writing-v3}{+1.9}
\setval{olmo.concur-tied.minus.gdpo.creative-writing-v3-abs}{1.9}
\setval{olmo.concur-tied.minus.gdpo.guidebench}{+7.7}
\setval{olmo.concur-tied.minus.gdpo.guidebench-abs}{7.7}
\setval{olmo.concur-tied.minus.gdpo.humaneval-plus}{+7.0}
\setval{olmo.concur-tied.minus.gdpo.humaneval-plus-abs}{7.0}
\setval{olmo.concur-tied.minus.gdpo.if}{+4.3}
\setval{olmo.concur-tied.minus.gdpo.if-abs}{4.3}
\setval{olmo.concur-tied.minus.gdpo.ifbench}{+4.0}
\setval{olmo.concur-tied.minus.gdpo.ifbench-abs}{4.0}
\setval{olmo.concur-tied.minus.gdpo.ifeval}{+1.2}
\setval{olmo.concur-tied.minus.gdpo.ifeval-abs}{1.2}
\setval{olmo.concur-tied.minus.gdpo.judge2.avg}{+4.8}
\setval{olmo.concur-tied.minus.gdpo.judge2.avg-abs}{4.8}
\setval{olmo.concur-tied.minus.gdpo.judge2.creative-writing-v3}{$-$0.6}
\setval{olmo.concur-tied.minus.gdpo.judge2.creative-writing-v3-abs}{0.6}
\setval{olmo.concur-tied.minus.gdpo.judge2.longbench-write}{+4.9}
\setval{olmo.concur-tied.minus.gdpo.judge2.longbench-write-abs}{4.9}
\setval{olmo.concur-tied.minus.gdpo.judge2.writing}{+0.9}
\setval{olmo.concur-tied.minus.gdpo.judge2.writing-abs}{0.9}
\setval{olmo.concur-tied.minus.gdpo.judge2.writingbench}{$-$1.5}
\setval{olmo.concur-tied.minus.gdpo.judge2.writingbench-abs}{1.5}
\setval{olmo.concur-tied.minus.gdpo.livecodebench}{+5.9}
\setval{olmo.concur-tied.minus.gdpo.livecodebench-abs}{5.9}
\setval{olmo.concur-tied.minus.gdpo.longbench-write}{+6.0}
\setval{olmo.concur-tied.minus.gdpo.longbench-write-abs}{6.0}
\setval{olmo.concur-tied.minus.gdpo.math}{+6.3}
\setval{olmo.concur-tied.minus.gdpo.math-abs}{6.3}
\setval{olmo.concur-tied.minus.gdpo.math500}{+4.2}
\setval{olmo.concur-tied.minus.gdpo.math500-abs}{4.2}
\setval{olmo.concur-tied.minus.gdpo.mbpp-plus}{+6.3}
\setval{olmo.concur-tied.minus.gdpo.mbpp-plus-abs}{6.3}
\setval{olmo.concur-tied.minus.gdpo.minerva}{+3.7}
\setval{olmo.concur-tied.minus.gdpo.minerva-abs}{3.7}
\setval{olmo.concur-tied.minus.gdpo.olympiad-bench}{+5.7}
\setval{olmo.concur-tied.minus.gdpo.olympiad-bench-abs}{5.7}
\setval{olmo.concur-tied.minus.gdpo.writing}{+2.9}
\setval{olmo.concur-tied.minus.gdpo.writing-abs}{2.9}
\setval{olmo.concur-tied.minus.gdpo.writingbench}{+0.8}
\setval{olmo.concur-tied.minus.gdpo.writingbench-abs}{0.8}
\setval{olmo.concur-tied.minus.geoma.aime24}{+10.7}
\setval{olmo.concur-tied.minus.geoma.aime24-abs}{10.7}
\setval{olmo.concur-tied.minus.geoma.aime25}{+7.9}
\setval{olmo.concur-tied.minus.geoma.aime25-abs}{7.9}
\setval{olmo.concur-tied.minus.geoma.amc23}{+15.9}
\setval{olmo.concur-tied.minus.geoma.amc23-abs}{15.9}
\setval{olmo.concur-tied.minus.geoma.avg}{+6.2}
\setval{olmo.concur-tied.minus.geoma.avg-abs}{6.2}
\setval{olmo.concur-tied.minus.geoma.code}{+7.2}
\setval{olmo.concur-tied.minus.geoma.code-abs}{7.2}
\setval{olmo.concur-tied.minus.geoma.creative-writing-v3}{$-$0.6}
\setval{olmo.concur-tied.minus.geoma.creative-writing-v3-abs}{0.6}
\setval{olmo.concur-tied.minus.geoma.guidebench}{+11.9}
\setval{olmo.concur-tied.minus.geoma.guidebench-abs}{11.9}
\setval{olmo.concur-tied.minus.geoma.humaneval-plus}{+8.7}
\setval{olmo.concur-tied.minus.geoma.humaneval-plus-abs}{8.7}
\setval{olmo.concur-tied.minus.geoma.if}{+5.0}
\setval{olmo.concur-tied.minus.geoma.if-abs}{5.0}
\setval{olmo.concur-tied.minus.geoma.ifbench}{+3.3}
\setval{olmo.concur-tied.minus.geoma.ifbench-abs}{3.3}
\setval{olmo.concur-tied.minus.geoma.ifeval}{$-$0.4}
\setval{olmo.concur-tied.minus.geoma.ifeval-abs}{0.4}
\setval{olmo.concur-tied.minus.geoma.judge2.avg}{+5.7}
\setval{olmo.concur-tied.minus.geoma.judge2.avg-abs}{5.7}
\setval{olmo.concur-tied.minus.geoma.judge2.creative-writing-v3}{$-$4.6}
\setval{olmo.concur-tied.minus.geoma.judge2.creative-writing-v3-abs}{4.6}
\setval{olmo.concur-tied.minus.geoma.judge2.longbench-write}{+2.3}
\setval{olmo.concur-tied.minus.geoma.judge2.longbench-write-abs}{2.3}
\setval{olmo.concur-tied.minus.geoma.judge2.writing}{$-$1.8}
\setval{olmo.concur-tied.minus.geoma.judge2.writing-abs}{1.8}
\setval{olmo.concur-tied.minus.geoma.judge2.writingbench}{$-$3.0}
\setval{olmo.concur-tied.minus.geoma.judge2.writingbench-abs}{3.0}
\setval{olmo.concur-tied.minus.geoma.livecodebench}{+6.7}
\setval{olmo.concur-tied.minus.geoma.livecodebench-abs}{6.7}
\setval{olmo.concur-tied.minus.geoma.longbench-write}{+3.9}
\setval{olmo.concur-tied.minus.geoma.longbench-write-abs}{3.9}
\setval{olmo.concur-tied.minus.geoma.math}{+9.0}
\setval{olmo.concur-tied.minus.geoma.math-abs}{9.0}
\setval{olmo.concur-tied.minus.geoma.math500}{+6.2}
\setval{olmo.concur-tied.minus.geoma.math500-abs}{6.2}
\setval{olmo.concur-tied.minus.geoma.mbpp-plus}{+6.3}
\setval{olmo.concur-tied.minus.geoma.mbpp-plus-abs}{6.3}
\setval{olmo.concur-tied.minus.geoma.minerva}{+4.1}
\setval{olmo.concur-tied.minus.geoma.minerva-abs}{4.1}
\setval{olmo.concur-tied.minus.geoma.olympiad-bench}{+8.8}
\setval{olmo.concur-tied.minus.geoma.olympiad-bench-abs}{8.8}
\setval{olmo.concur-tied.minus.geoma.writing}{+0.8}
\setval{olmo.concur-tied.minus.geoma.writing-abs}{0.8}
\setval{olmo.concur-tied.minus.geoma.writingbench}{$-$0.8}
\setval{olmo.concur-tied.minus.geoma.writingbench-abs}{0.8}
\setval{olmo.concur-tied.minus.grpo-nodiv.aime24}{+2.7}
\setval{olmo.concur-tied.minus.grpo-nodiv.aime24-abs}{2.7}
\setval{olmo.concur-tied.minus.grpo-nodiv.aime25}{+2.3}
\setval{olmo.concur-tied.minus.grpo-nodiv.aime25-abs}{2.3}
\setval{olmo.concur-tied.minus.grpo-nodiv.amc23}{+6.8}
\setval{olmo.concur-tied.minus.grpo-nodiv.amc23-abs}{6.8}
\setval{olmo.concur-tied.minus.grpo-nodiv.avg}{+3.2}
\setval{olmo.concur-tied.minus.grpo-nodiv.avg-abs}{3.2}
\setval{olmo.concur-tied.minus.grpo-nodiv.code}{+5.7}
\setval{olmo.concur-tied.minus.grpo-nodiv.code-abs}{5.7}
\setval{olmo.concur-tied.minus.grpo-nodiv.creative-writing-v3}{+2.1}
\setval{olmo.concur-tied.minus.grpo-nodiv.creative-writing-v3-abs}{2.1}
\setval{olmo.concur-tied.minus.grpo-nodiv.guidebench}{+3.1}
\setval{olmo.concur-tied.minus.grpo-nodiv.guidebench-abs}{3.1}
\setval{olmo.concur-tied.minus.grpo-nodiv.humaneval-plus}{+5.9}
\setval{olmo.concur-tied.minus.grpo-nodiv.humaneval-plus-abs}{5.9}
\setval{olmo.concur-tied.minus.grpo-nodiv.if}{+1.8}
\setval{olmo.concur-tied.minus.grpo-nodiv.if-abs}{1.8}
\setval{olmo.concur-tied.minus.grpo-nodiv.ifbench}{+2.3}
\setval{olmo.concur-tied.minus.grpo-nodiv.ifbench-abs}{2.3}
\setval{olmo.concur-tied.minus.grpo-nodiv.ifeval}{$-$0.1}
\setval{olmo.concur-tied.minus.grpo-nodiv.ifeval-abs}{0.1}
\setval{olmo.concur-tied.minus.grpo-nodiv.judge2.avg}{+2.5}
\setval{olmo.concur-tied.minus.grpo-nodiv.judge2.avg-abs}{2.5}
\setval{olmo.concur-tied.minus.grpo-nodiv.judge2.creative-writing-v3}{$-$1.6}
\setval{olmo.concur-tied.minus.grpo-nodiv.judge2.creative-writing-v3-abs}{1.6}
\setval{olmo.concur-tied.minus.grpo-nodiv.judge2.longbench-write}{$-$0.6}
\setval{olmo.concur-tied.minus.grpo-nodiv.judge2.longbench-write-abs}{0.6}
\setval{olmo.concur-tied.minus.grpo-nodiv.judge2.writing}{$-$1.2}
\setval{olmo.concur-tied.minus.grpo-nodiv.judge2.writing-abs}{1.2}
\setval{olmo.concur-tied.minus.grpo-nodiv.judge2.writingbench}{$-$1.4}
\setval{olmo.concur-tied.minus.grpo-nodiv.judge2.writingbench-abs}{1.4}
\setval{olmo.concur-tied.minus.grpo-nodiv.livecodebench}{+5.4}
\setval{olmo.concur-tied.minus.grpo-nodiv.livecodebench-abs}{5.4}
\setval{olmo.concur-tied.minus.grpo-nodiv.longbench-write}{+1.7}
\setval{olmo.concur-tied.minus.grpo-nodiv.longbench-write-abs}{1.7}
\setval{olmo.concur-tied.minus.grpo-nodiv.math}{+3.2}
\setval{olmo.concur-tied.minus.grpo-nodiv.math-abs}{3.2}
\setval{olmo.concur-tied.minus.grpo-nodiv.math500}{+1.9}
\setval{olmo.concur-tied.minus.grpo-nodiv.math500-abs}{1.9}
\setval{olmo.concur-tied.minus.grpo-nodiv.mbpp-plus}{+5.7}
\setval{olmo.concur-tied.minus.grpo-nodiv.mbpp-plus-abs}{5.7}
\setval{olmo.concur-tied.minus.grpo-nodiv.minerva}{+2.2}
\setval{olmo.concur-tied.minus.grpo-nodiv.minerva-abs}{2.2}
\setval{olmo.concur-tied.minus.grpo-nodiv.olympiad-bench}{+3.1}
\setval{olmo.concur-tied.minus.grpo-nodiv.olympiad-bench-abs}{3.1}
\setval{olmo.concur-tied.minus.grpo-nodiv.writing}{+2.2}
\setval{olmo.concur-tied.minus.grpo-nodiv.writing-abs}{2.2}
\setval{olmo.concur-tied.minus.grpo-nodiv.writingbench}{+3.0}
\setval{olmo.concur-tied.minus.grpo-nodiv.writingbench-abs}{3.0}
\setval{olmo.concur-tied.minus.grpo.aime24}{+5.0}
\setval{olmo.concur-tied.minus.grpo.aime24-abs}{5.0}
\setval{olmo.concur-tied.minus.grpo.aime25}{+5.2}
\setval{olmo.concur-tied.minus.grpo.aime25-abs}{5.2}
\setval{olmo.concur-tied.minus.grpo.amc23}{+9.0}
\setval{olmo.concur-tied.minus.grpo.amc23-abs}{9.0}
\setval{olmo.concur-tied.minus.grpo.avg}{+3.9}
\setval{olmo.concur-tied.minus.grpo.avg-abs}{3.9}
\setval{olmo.concur-tied.minus.grpo.code}{+6.0}
\setval{olmo.concur-tied.minus.grpo.code-abs}{6.0}
\setval{olmo.concur-tied.minus.grpo.creative-writing-v3}{$-$0.3}
\setval{olmo.concur-tied.minus.grpo.creative-writing-v3-abs}{0.3}
\setval{olmo.concur-tied.minus.grpo.guidebench}{+2.1}
\setval{olmo.concur-tied.minus.grpo.guidebench-abs}{2.1}
\setval{olmo.concur-tied.minus.grpo.humaneval-plus}{+6.0}
\setval{olmo.concur-tied.minus.grpo.humaneval-plus-abs}{6.0}
\setval{olmo.concur-tied.minus.grpo.if}{+1.3}
\setval{olmo.concur-tied.minus.grpo.if-abs}{1.3}
\setval{olmo.concur-tied.minus.grpo.ifbench}{+1.7}
\setval{olmo.concur-tied.minus.grpo.ifbench-abs}{1.7}
\setval{olmo.concur-tied.minus.grpo.ifeval}{+0.2}
\setval{olmo.concur-tied.minus.grpo.ifeval-abs}{0.2}
\setval{olmo.concur-tied.minus.grpo.judge2.avg}{+3.6}
\setval{olmo.concur-tied.minus.grpo.judge2.avg-abs}{3.6}
\setval{olmo.concur-tied.minus.grpo.judge2.creative-writing-v3}{$-$1.4}
\setval{olmo.concur-tied.minus.grpo.judge2.creative-writing-v3-abs}{1.4}
\setval{olmo.concur-tied.minus.grpo.judge2.longbench-write}{+3.4}
\setval{olmo.concur-tied.minus.grpo.judge2.longbench-write-abs}{3.4}
\setval{olmo.concur-tied.minus.grpo.judge2.writing}{$-$0.3}
\setval{olmo.concur-tied.minus.grpo.judge2.writing-abs}{0.3}
\setval{olmo.concur-tied.minus.grpo.judge2.writingbench}{$-$2.9}
\setval{olmo.concur-tied.minus.grpo.judge2.writingbench-abs}{2.9}
\setval{olmo.concur-tied.minus.grpo.livecodebench}{+6.7}
\setval{olmo.concur-tied.minus.grpo.livecodebench-abs}{6.7}
\setval{olmo.concur-tied.minus.grpo.longbench-write}{+5.0}
\setval{olmo.concur-tied.minus.grpo.longbench-write-abs}{5.0}
\setval{olmo.concur-tied.minus.grpo.math}{+5.6}
\setval{olmo.concur-tied.minus.grpo.math-abs}{5.6}
\setval{olmo.concur-tied.minus.grpo.math500}{+4.5}
\setval{olmo.concur-tied.minus.grpo.math500-abs}{4.5}
\setval{olmo.concur-tied.minus.grpo.mbpp-plus}{+5.3}
\setval{olmo.concur-tied.minus.grpo.mbpp-plus-abs}{5.3}
\setval{olmo.concur-tied.minus.grpo.minerva}{+4.1}
\setval{olmo.concur-tied.minus.grpo.minerva-abs}{4.1}
\setval{olmo.concur-tied.minus.grpo.olympiad-bench}{+5.9}
\setval{olmo.concur-tied.minus.grpo.olympiad-bench-abs}{5.9}
\setval{olmo.concur-tied.minus.grpo.writing}{+0.8}
\setval{olmo.concur-tied.minus.grpo.writing-abs}{0.8}
\setval{olmo.concur-tied.minus.grpo.writingbench}{$-$2.3}
\setval{olmo.concur-tied.minus.grpo.writingbench-abs}{2.3}
\setval{olmo.concur-tied.minus.mt-grpo.aime24}{$-$2.1}
\setval{olmo.concur-tied.minus.mt-grpo.aime24-abs}{2.1}
\setval{olmo.concur-tied.minus.mt-grpo.aime25}{+1.8}
\setval{olmo.concur-tied.minus.mt-grpo.aime25-abs}{1.8}
\setval{olmo.concur-tied.minus.mt-grpo.amc23}{+2.4}
\setval{olmo.concur-tied.minus.mt-grpo.amc23-abs}{2.4}
\setval{olmo.concur-tied.minus.mt-grpo.avg}{+2.4}
\setval{olmo.concur-tied.minus.mt-grpo.avg-abs}{2.4}
\setval{olmo.concur-tied.minus.mt-grpo.code}{+6.7}
\setval{olmo.concur-tied.minus.mt-grpo.code-abs}{6.7}
\setval{olmo.concur-tied.minus.mt-grpo.creative-writing-v3}{$-$1.7}
\setval{olmo.concur-tied.minus.mt-grpo.creative-writing-v3-abs}{1.7}
\setval{olmo.concur-tied.minus.mt-grpo.guidebench}{+4.0}
\setval{olmo.concur-tied.minus.mt-grpo.guidebench-abs}{4.0}
\setval{olmo.concur-tied.minus.mt-grpo.humaneval-plus}{+7.6}
\setval{olmo.concur-tied.minus.mt-grpo.humaneval-plus-abs}{7.6}
\setval{olmo.concur-tied.minus.mt-grpo.if}{+2.1}
\setval{olmo.concur-tied.minus.mt-grpo.if-abs}{2.1}
\setval{olmo.concur-tied.minus.mt-grpo.ifbench}{+1.7}
\setval{olmo.concur-tied.minus.mt-grpo.ifbench-abs}{1.7}
\setval{olmo.concur-tied.minus.mt-grpo.ifeval}{+0.6}
\setval{olmo.concur-tied.minus.mt-grpo.ifeval-abs}{0.6}
\setval{olmo.concur-tied.minus.mt-grpo.judge2.avg}{+2.1}
\setval{olmo.concur-tied.minus.mt-grpo.judge2.avg-abs}{2.1}
\setval{olmo.concur-tied.minus.mt-grpo.judge2.creative-writing-v3}{$-$2.7}
\setval{olmo.concur-tied.minus.mt-grpo.judge2.creative-writing-v3-abs}{2.7}
\setval{olmo.concur-tied.minus.mt-grpo.judge2.longbench-write}{+0.1}
\setval{olmo.concur-tied.minus.mt-grpo.judge2.longbench-write-abs}{0.1}
\setval{olmo.concur-tied.minus.mt-grpo.judge2.writing}{$-$1.5}
\setval{olmo.concur-tied.minus.mt-grpo.judge2.writing-abs}{1.5}
\setval{olmo.concur-tied.minus.mt-grpo.judge2.writingbench}{$-$1.9}
\setval{olmo.concur-tied.minus.mt-grpo.judge2.writingbench-abs}{1.9}
\setval{olmo.concur-tied.minus.mt-grpo.livecodebench}{+6.4}
\setval{olmo.concur-tied.minus.mt-grpo.livecodebench-abs}{6.4}
\setval{olmo.concur-tied.minus.mt-grpo.longbench-write}{+2.9}
\setval{olmo.concur-tied.minus.mt-grpo.longbench-write-abs}{2.9}
\setval{olmo.concur-tied.minus.mt-grpo.math}{+1.7}
\setval{olmo.concur-tied.minus.mt-grpo.math-abs}{1.7}
\setval{olmo.concur-tied.minus.mt-grpo.math500}{+2.5}
\setval{olmo.concur-tied.minus.mt-grpo.math500-abs}{2.5}
\setval{olmo.concur-tied.minus.mt-grpo.mbpp-plus}{+6.1}
\setval{olmo.concur-tied.minus.mt-grpo.mbpp-plus-abs}{6.1}
\setval{olmo.concur-tied.minus.mt-grpo.minerva}{+2.1}
\setval{olmo.concur-tied.minus.mt-grpo.minerva-abs}{2.1}
\setval{olmo.concur-tied.minus.mt-grpo.olympiad-bench}{+3.5}
\setval{olmo.concur-tied.minus.mt-grpo.olympiad-bench-abs}{3.5}
\setval{olmo.concur-tied.minus.mt-grpo.writing}{$-$0.1}
\setval{olmo.concur-tied.minus.mt-grpo.writing-abs}{0.1}
\setval{olmo.concur-tied.minus.mt-grpo.writingbench}{$-$1.6}
\setval{olmo.concur-tied.minus.mt-grpo.writingbench-abs}{1.6}
\setval{olmo.concur-tied.minus.papo.aime24}{+12.3}
\setval{olmo.concur-tied.minus.papo.aime24-abs}{12.3}
\setval{olmo.concur-tied.minus.papo.aime25}{+9.6}
\setval{olmo.concur-tied.minus.papo.aime25-abs}{9.6}
\setval{olmo.concur-tied.minus.papo.amc23}{+20.5}
\setval{olmo.concur-tied.minus.papo.amc23-abs}{20.5}
\setval{olmo.concur-tied.minus.papo.avg}{+6.8}
\setval{olmo.concur-tied.minus.papo.avg-abs}{6.8}
\setval{olmo.concur-tied.minus.papo.code}{+7.2}
\setval{olmo.concur-tied.minus.papo.code-abs}{7.2}
\setval{olmo.concur-tied.minus.papo.creative-writing-v3}{+2.1}
\setval{olmo.concur-tied.minus.papo.creative-writing-v3-abs}{2.1}
\setval{olmo.concur-tied.minus.papo.guidebench}{$-$0.1}
\setval{olmo.concur-tied.minus.papo.guidebench-abs}{0.1}
\setval{olmo.concur-tied.minus.papo.humaneval-plus}{+8.5}
\setval{olmo.concur-tied.minus.papo.humaneval-plus-abs}{8.5}
\setval{olmo.concur-tied.minus.papo.if}{+1.5}
\setval{olmo.concur-tied.minus.papo.if-abs}{1.5}
\setval{olmo.concur-tied.minus.papo.ifbench}{+4.0}
\setval{olmo.concur-tied.minus.papo.ifbench-abs}{4.0}
\setval{olmo.concur-tied.minus.papo.ifeval}{+0.5}
\setval{olmo.concur-tied.minus.papo.ifeval-abs}{0.5}
\setval{olmo.concur-tied.minus.papo.judge2.avg}{+6.4}
\setval{olmo.concur-tied.minus.papo.judge2.avg-abs}{6.4}
\setval{olmo.concur-tied.minus.papo.judge2.creative-writing-v3}{$-$0.8}
\setval{olmo.concur-tied.minus.papo.judge2.creative-writing-v3-abs}{0.8}
\setval{olmo.concur-tied.minus.papo.judge2.longbench-write}{+4.9}
\setval{olmo.concur-tied.minus.papo.judge2.longbench-write-abs}{4.9}
\setval{olmo.concur-tied.minus.papo.judge2.writing}{+0.7}
\setval{olmo.concur-tied.minus.papo.judge2.writing-abs}{0.7}
\setval{olmo.concur-tied.minus.papo.judge2.writingbench}{$-$1.9}
\setval{olmo.concur-tied.minus.papo.judge2.writingbench-abs}{1.9}
\setval{olmo.concur-tied.minus.papo.livecodebench}{+7.1}
\setval{olmo.concur-tied.minus.papo.livecodebench-abs}{7.1}
\setval{olmo.concur-tied.minus.papo.longbench-write}{+4.5}
\setval{olmo.concur-tied.minus.papo.longbench-write-abs}{4.5}
\setval{olmo.concur-tied.minus.papo.math}{+11.4}
\setval{olmo.concur-tied.minus.papo.math-abs}{11.4}
\setval{olmo.concur-tied.minus.papo.math500}{+8.5}
\setval{olmo.concur-tied.minus.papo.math500-abs}{8.5}
\setval{olmo.concur-tied.minus.papo.mbpp-plus}{+6.0}
\setval{olmo.concur-tied.minus.papo.mbpp-plus-abs}{6.0}
\setval{olmo.concur-tied.minus.papo.minerva}{+5.5}
\setval{olmo.concur-tied.minus.papo.minerva-abs}{5.5}
\setval{olmo.concur-tied.minus.papo.olympiad-bench}{+11.8}
\setval{olmo.concur-tied.minus.papo.olympiad-bench-abs}{11.8}
\setval{olmo.concur-tied.minus.papo.writing}{+2.4}
\setval{olmo.concur-tied.minus.papo.writing-abs}{2.4}
\setval{olmo.concur-tied.minus.papo.writingbench}{+0.8}
\setval{olmo.concur-tied.minus.papo.writingbench-abs}{0.8}
\setval{olmo.concur-tied.minus.saw.aime24}{+8.3}
\setval{olmo.concur-tied.minus.saw.aime24-abs}{8.3}
\setval{olmo.concur-tied.minus.saw.aime25}{+4.7}
\setval{olmo.concur-tied.minus.saw.aime25-abs}{4.7}
\setval{olmo.concur-tied.minus.saw.amc23}{+13.0}
\setval{olmo.concur-tied.minus.saw.amc23-abs}{13.0}
\setval{olmo.concur-tied.minus.saw.avg}{+4.7}
\setval{olmo.concur-tied.minus.saw.avg-abs}{4.7}
\setval{olmo.concur-tied.minus.saw.code}{+6.6}
\setval{olmo.concur-tied.minus.saw.code-abs}{6.6}
\setval{olmo.concur-tied.minus.saw.creative-writing-v3}{+1.1}
\setval{olmo.concur-tied.minus.saw.creative-writing-v3-abs}{1.1}
\setval{olmo.concur-tied.minus.saw.guidebench}{+3.8}
\setval{olmo.concur-tied.minus.saw.guidebench-abs}{3.8}
\setval{olmo.concur-tied.minus.saw.humaneval-plus}{+7.0}
\setval{olmo.concur-tied.minus.saw.humaneval-plus-abs}{7.0}
\setval{olmo.concur-tied.minus.saw.if}{+1.3}
\setval{olmo.concur-tied.minus.saw.if-abs}{1.3}
\setval{olmo.concur-tied.minus.saw.ifbench}{+1.0}
\setval{olmo.concur-tied.minus.saw.ifbench-abs}{1.0}
\setval{olmo.concur-tied.minus.saw.ifeval}{$-$1.0}
\setval{olmo.concur-tied.minus.saw.ifeval-abs}{1.0}
\setval{olmo.concur-tied.minus.saw.judge2.avg}{+4.2}
\setval{olmo.concur-tied.minus.saw.judge2.avg-abs}{4.2}
\setval{olmo.concur-tied.minus.saw.judge2.creative-writing-v3}{$-$3.8}
\setval{olmo.concur-tied.minus.saw.judge2.creative-writing-v3-abs}{3.8}
\setval{olmo.concur-tied.minus.saw.judge2.longbench-write}{+3.1}
\setval{olmo.concur-tied.minus.saw.judge2.longbench-write-abs}{3.1}
\setval{olmo.concur-tied.minus.saw.judge2.writing}{$-$1.2}
\setval{olmo.concur-tied.minus.saw.judge2.writing-abs}{1.2}
\setval{olmo.concur-tied.minus.saw.judge2.writingbench}{$-$2.8}
\setval{olmo.concur-tied.minus.saw.judge2.writingbench-abs}{2.8}
\setval{olmo.concur-tied.minus.saw.livecodebench}{+7.1}
\setval{olmo.concur-tied.minus.saw.livecodebench-abs}{7.1}
\setval{olmo.concur-tied.minus.saw.longbench-write}{+4.3}
\setval{olmo.concur-tied.minus.saw.longbench-write-abs}{4.3}
\setval{olmo.concur-tied.minus.saw.math}{+7.1}
\setval{olmo.concur-tied.minus.saw.math-abs}{7.1}
\setval{olmo.concur-tied.minus.saw.math500}{+4.9}
\setval{olmo.concur-tied.minus.saw.math500-abs}{4.9}
\setval{olmo.concur-tied.minus.saw.mbpp-plus}{+5.6}
\setval{olmo.concur-tied.minus.saw.mbpp-plus-abs}{5.6}
\setval{olmo.concur-tied.minus.saw.minerva}{+3.7}
\setval{olmo.concur-tied.minus.saw.minerva-abs}{3.7}
\setval{olmo.concur-tied.minus.saw.olympiad-bench}{+7.8}
\setval{olmo.concur-tied.minus.saw.olympiad-bench-abs}{7.8}
\setval{olmo.concur-tied.minus.saw.writing}{+1.4}
\setval{olmo.concur-tied.minus.saw.writing-abs}{1.4}
\setval{olmo.concur-tied.minus.saw.writingbench}{$-$1.2}
\setval{olmo.concur-tied.minus.saw.writingbench-abs}{1.2}
\setval{olmo.concur-tied.passing-at-ceiling.code.early}{0.84}
\setval{olmo.concur-tied.passing-at-ceiling.code.late}{0.88}
\setval{olmo.concur-tied.passing-at-ceiling.code.run}{0.86}
\setval{olmo.concur-tied.passing-at-ceiling.if.early}{0.49}
\setval{olmo.concur-tied.passing-at-ceiling.if.late}{0.56}
\setval{olmo.concur-tied.passing-at-ceiling.if.run}{0.51}
\setval{olmo.concur-tied.passing-at-ceiling.math.early}{1.00}
\setval{olmo.concur-tied.passing-at-ceiling.math.late}{1.00}
\setval{olmo.concur-tied.passing-at-ceiling.math.run}{1.00}
\setval{olmo.concur-tied.passing-at-ceiling.writing.early}{0.00}
\setval{olmo.concur-tied.passing-at-ceiling.writing.late}{0.01}
\setval{olmo.concur-tied.passing-at-ceiling.writing.run}{0.00}
\setval{olmo.concur-tied.primary-cosine.code.early}{0.98}
\setval{olmo.concur-tied.primary-cosine.code.late}{0.95}
\setval{olmo.concur-tied.primary-cosine.code.run}{0.96}
\setval{olmo.concur-tied.primary-cosine.if.early}{0.84}
\setval{olmo.concur-tied.primary-cosine.if.late}{0.82}
\setval{olmo.concur-tied.primary-cosine.if.run}{0.83}
\setval{olmo.concur-tied.primary-cosine.math.early}{0.99}
\setval{olmo.concur-tied.primary-cosine.math.late}{0.97}
\setval{olmo.concur-tied.primary-cosine.math.run}{0.97}
\setval{olmo.concur-tied.primary-cosine.writing.early}{0.96}
\setval{olmo.concur-tied.primary-cosine.writing.late}{0.93}
\setval{olmo.concur-tied.primary-cosine.writing.run}{0.94}
\setval{olmo.concur-tied.primary-level.code.early}{0.58}
\setval{olmo.concur-tied.primary-level.code.late}{0.85}
\setval{olmo.concur-tied.primary-level.code.run}{0.78}
\setval{olmo.concur-tied.primary-level.if.early}{0.63}
\setval{olmo.concur-tied.primary-level.if.late}{0.72}
\setval{olmo.concur-tied.primary-level.if.run}{0.68}
\setval{olmo.concur-tied.primary-level.math.early}{0.34}
\setval{olmo.concur-tied.primary-level.math.late}{0.72}
\setval{olmo.concur-tied.primary-level.math.run}{0.62}
\setval{olmo.concur-tied.primary-level.writing.early}{0.50}
\setval{olmo.concur-tied.primary-level.writing.late}{0.75}
\setval{olmo.concur-tied.primary-level.writing.run}{0.68}
\setval{olmo.concur-tied.solved.code.early}{0.17}
\setval{olmo.concur-tied.solved.code.late}{0.44}
\setval{olmo.concur-tied.solved.code.run}{0.34}
\setval{olmo.concur-tied.solved.if.early}{0.18}
\setval{olmo.concur-tied.solved.if.late}{0.22}
\setval{olmo.concur-tied.solved.if.run}{0.21}
\setval{olmo.concur-tied.solved.math.early}{0.08}
\setval{olmo.concur-tied.solved.math.late}{0.36}
\setval{olmo.concur-tied.solved.math.run}{0.28}
\setval{olmo.concur-tied.solved.writing.early}{0.00}
\setval{olmo.concur-tied.solved.writing.late}{0.00}
\setval{olmo.concur-tied.solved.writing.run}{0.00}
\setval{olmo.concur-tied.trust.code.agreement}{$-$0.010}
\setval{olmo.concur-tied.trust.code.errors}{$-$2.2}
\setval{olmo.concur-tied.trust.code.first-free}{5}
\setval{olmo.concur-tied.trust.code.free-share}{0.02}
\setval{olmo.concur-tied.trust.if.agreement}{+0.064}
\setval{olmo.concur-tied.trust.if.errors}{+18.0}
\setval{olmo.concur-tied.trust.if.first-free}{2}
\setval{olmo.concur-tied.trust.if.free-share}{0.99}
\setval{olmo.concur-tied.trust.math.first-free}{--}
\setval{olmo.concur-tied.trust.math.free-share}{0.00}
\setval{olmo.concur-tied.trust.writing-rm.agreement}{+0.093}
\setval{olmo.concur-tied.trust.writing-rm.errors}{+45.9}
\setval{olmo.concur-tied.trust.writing-rm.first-free}{2}
\setval{olmo.concur-tied.trust.writing-rm.free-share}{0.99}
\setval{olmo.concur-tied.trust.writing.agreement}{$-$0.063}
\setval{olmo.concur-tied.trust.writing.errors}{$-$38.0}
\setval{olmo.concur-tied.trust.writing.first-free}{--}
\setval{olmo.concur-tied.trust.writing.free-share}{0.00}
\setval{olmo.concur-tied.update-share.code.early}{0.17}
\setval{olmo.concur-tied.update-share.code.late}{0.12}
\setval{olmo.concur-tied.update-share.code.run}{0.15}
\setval{olmo.concur-tied.update-share.if.early}{0.19}
\setval{olmo.concur-tied.update-share.if.late}{0.29}
\setval{olmo.concur-tied.update-share.if.run}{0.26}
\setval{olmo.concur-tied.update-share.math.early}{0.18}
\setval{olmo.concur-tied.update-share.math.late}{0.10}
\setval{olmo.concur-tied.update-share.math.run}{0.13}
\setval{olmo.concur-tied.update-share.writing.early}{0.46}
\setval{olmo.concur-tied.update-share.writing.late}{0.49}
\setval{olmo.concur-tied.update-share.writing.run}{0.46}
\setval{olmo.concur.aux-change.code}{$-$0.25}
\setval{olmo.concur.aux-change.if}{+0.11}
\setval{olmo.concur.aux-change.math}{$-$0.72}
\setval{olmo.concur.aux-change.writing}{$-$0.76}
\setval{olmo.concur.aux-change.writing-rm}{+0.22}
\setval{olmo.concur.aux-full.code.early}{0.66}
\setval{olmo.concur.aux-full.code.late}{0.42}
\setval{olmo.concur.aux-full.code.run}{0.54}
\setval{olmo.concur.aux-full.if.early}{0.57}
\setval{olmo.concur.aux-full.if.late}{0.91}
\setval{olmo.concur.aux-full.if.run}{0.80}
\setval{olmo.concur.aux-full.math.early}{0.51}
\setval{olmo.concur.aux-full.math.late}{0.03}
\setval{olmo.concur.aux-full.math.run}{0.12}
\setval{olmo.concur.aux-full.writing-rm.early}{0.02}
\setval{olmo.concur.aux-full.writing-rm.late}{0.05}
\setval{olmo.concur.aux-full.writing-rm.run}{0.04}
\setval{olmo.concur.aux-full.writing.early}{0.49}
\setval{olmo.concur.aux-full.writing.late}{0.05}
\setval{olmo.concur.aux-full.writing.run}{0.12}
\setval{olmo.concur.aux-level.code.early}{0.85}
\setval{olmo.concur.aux-level.code.late}{0.56}
\setval{olmo.concur.aux-level.code.run}{0.69}
\setval{olmo.concur.aux-level.if.early}{0.90}
\setval{olmo.concur.aux-level.if.late}{0.98}
\setval{olmo.concur.aux-level.if.run}{0.96}
\setval{olmo.concur.aux-level.math.early}{0.89}
\setval{olmo.concur.aux-level.math.late}{0.18}
\setval{olmo.concur.aux-level.math.run}{0.36}
\setval{olmo.concur.aux-level.writing-rm.early}{0.60}
\setval{olmo.concur.aux-level.writing-rm.late}{0.79}
\setval{olmo.concur.aux-level.writing-rm.run}{0.72}
\setval{olmo.concur.aux-level.writing.early}{0.94}
\setval{olmo.concur.aux-level.writing.late}{0.18}
\setval{olmo.concur.aux-level.writing.run}{0.45}
\setval{olmo.concur.bench.aime24}{36.4}
\setval{olmo.concur.bench.aime25}{32.2}
\setval{olmo.concur.bench.amc23}{83.5}
\setval{olmo.concur.bench.creative-writing-v3}{49.2}
\setval{olmo.concur.bench.guidebench}{56.9}
\setval{olmo.concur.bench.humaneval-plus}{82.7}
\setval{olmo.concur.bench.ifbench}{32.0}
\setval{olmo.concur.bench.ifeval}{87.8}
\setval{olmo.concur.bench.livecodebench}{30.3}
\setval{olmo.concur.bench.longbench-write}{68.5}
\setval{olmo.concur.bench.math500}{90.5}
\setval{olmo.concur.bench.mbpp-plus}{67.5}
\setval{olmo.concur.bench.minerva}{44.8}
\setval{olmo.concur.bench.olympiad-bench}{62.4}
\setval{olmo.concur.bench.writingbench}{65.4}
\setval{olmo.concur.best-count.all}{9}
\setval{olmo.concur.best-count.code}{3}
\setval{olmo.concur.best-count.if}{0}
\setval{olmo.concur.best-count.math}{5}
\setval{olmo.concur.best-count.writing}{1}
\setval{olmo.concur.eval.avg}{59.4}
\setval{olmo.concur.eval.code}{60.2}
\setval{olmo.concur.eval.if}{58.9}
\setval{olmo.concur.eval.math}{58.3}
\setval{olmo.concur.eval.writing}{61.0}
\setval{olmo.concur.gate.code.against.early}{0.08}
\setval{olmo.concur.gate.code.against.late}{0.10}
\setval{olmo.concur.gate.code.against.run}{0.12}
\setval{olmo.concur.gate.code.agreement}{$-$0.011}
\setval{olmo.concur.gate.code.ceiling.early}{0.29}
\setval{olmo.concur.gate.code.ceiling.late}{0.75}
\setval{olmo.concur.gate.code.ceiling.run}{0.56}
\setval{olmo.concur.gate.code.cosine.early}{0.68}
\setval{olmo.concur.gate.code.cosine.late}{0.14}
\setval{olmo.concur.gate.code.cosine.run}{0.33}
\setval{olmo.concur.gate.if.against.early}{0.11}
\setval{olmo.concur.gate.if.against.late}{0.06}
\setval{olmo.concur.gate.if.against.run}{0.07}
\setval{olmo.concur.gate.if.agreement}{+0.053}
\setval{olmo.concur.gate.if.ceiling.early}{0.22}
\setval{olmo.concur.gate.if.ceiling.late}{0.19}
\setval{olmo.concur.gate.if.ceiling.run}{0.22}
\setval{olmo.concur.gate.if.cosine.early}{0.50}
\setval{olmo.concur.gate.if.cosine.late}{0.52}
\setval{olmo.concur.gate.if.cosine.run}{0.52}
\setval{olmo.concur.gate.math.against.early}{0.00}
\setval{olmo.concur.gate.math.against.late}{0.00}
\setval{olmo.concur.gate.math.against.run}{0.00}
\setval{olmo.concur.gate.math.ceiling.early}{0.16}
\setval{olmo.concur.gate.math.ceiling.late}{0.89}
\setval{olmo.concur.gate.math.ceiling.run}{0.72}
\setval{olmo.concur.gate.math.cosine.early}{0.90}
\setval{olmo.concur.gate.math.cosine.late}{0.15}
\setval{olmo.concur.gate.math.cosine.run}{0.34}
\setval{olmo.concur.gate.writing-rm.against.early}{0.22}
\setval{olmo.concur.gate.writing-rm.against.late}{0.56}
\setval{olmo.concur.gate.writing-rm.against.run}{0.46}
\setval{olmo.concur.gate.writing-rm.agreement}{+0.093}
\setval{olmo.concur.gate.writing-rm.ceiling.early}{0.00}
\setval{olmo.concur.gate.writing-rm.ceiling.late}{0.00}
\setval{olmo.concur.gate.writing-rm.ceiling.run}{0.00}
\setval{olmo.concur.gate.writing-rm.cosine.early}{0.69}
\setval{olmo.concur.gate.writing-rm.cosine.late}{0.18}
\setval{olmo.concur.gate.writing-rm.cosine.run}{0.33}
\setval{olmo.concur.gate.writing.against.early}{0.27}
\setval{olmo.concur.gate.writing.against.late}{0.69}
\setval{olmo.concur.gate.writing.against.run}{0.63}
\setval{olmo.concur.gate.writing.agreement}{$-$0.068}
\setval{olmo.concur.gate.writing.ceiling.early}{0.00}
\setval{olmo.concur.gate.writing.ceiling.late}{0.00}
\setval{olmo.concur.gate.writing.ceiling.run}{0.00}
\setval{olmo.concur.gate.writing.cosine.early}{0.57}
\setval{olmo.concur.gate.writing.cosine.late}{-0.01}
\setval{olmo.concur.gate.writing.cosine.run}{0.07}
\setval{olmo.concur.judge2.avg}{56.6}
\setval{olmo.concur.judge2.creative-writing-v3}{38.5}
\setval{olmo.concur.judge2.longbench-write}{58.4}
\setval{olmo.concur.judge2.writing}{47.2}
\setval{olmo.concur.judge2.writingbench}{44.9}
\setval{olmo.concur.last-step}{300}
\setval{olmo.concur.minus.base.aime24}{+27.8}
\setval{olmo.concur.minus.base.aime24-abs}{27.8}
\setval{olmo.concur.minus.base.aime25}{+24.8}
\setval{olmo.concur.minus.base.aime25-abs}{24.8}
\setval{olmo.concur.minus.base.amc23}{+39.0}
\setval{olmo.concur.minus.base.amc23-abs}{39.0}
\setval{olmo.concur.minus.base.avg}{+15.2}
\setval{olmo.concur.minus.base.avg-abs}{15.2}
\setval{olmo.concur.minus.base.code}{+11.1}
\setval{olmo.concur.minus.base.code-abs}{11.1}
\setval{olmo.concur.minus.base.creative-writing-v3}{+11.3}
\setval{olmo.concur.minus.base.creative-writing-v3-abs}{11.3}
\setval{olmo.concur.minus.base.guidebench}{$-$4.5}
\setval{olmo.concur.minus.base.guidebench-abs}{4.5}
\setval{olmo.concur.minus.base.humaneval-plus}{+13.5}
\setval{olmo.concur.minus.base.humaneval-plus-abs}{13.5}
\setval{olmo.concur.minus.base.if}{+1.4}
\setval{olmo.concur.minus.base.if-abs}{1.4}
\setval{olmo.concur.minus.base.ifbench}{+5.3}
\setval{olmo.concur.minus.base.ifbench-abs}{5.3}
\setval{olmo.concur.minus.base.ifeval}{+3.4}
\setval{olmo.concur.minus.base.ifeval-abs}{3.4}
\setval{olmo.concur.minus.base.judge2.avg}{+13.6}
\setval{olmo.concur.minus.base.judge2.avg-abs}{13.6}
\setval{olmo.concur.minus.base.judge2.creative-writing-v3}{+6.8}
\setval{olmo.concur.minus.base.judge2.creative-writing-v3-abs}{6.8}
\setval{olmo.concur.minus.base.judge2.longbench-write}{+5.2}
\setval{olmo.concur.minus.base.judge2.longbench-write-abs}{5.2}
\setval{olmo.concur.minus.base.judge2.writing}{+6.4}
\setval{olmo.concur.minus.base.judge2.writing-abs}{6.4}
\setval{olmo.concur.minus.base.judge2.writingbench}{+7.2}
\setval{olmo.concur.minus.base.judge2.writingbench-abs}{7.2}
\setval{olmo.concur.minus.base.livecodebench}{+13.5}
\setval{olmo.concur.minus.base.livecodebench-abs}{13.5}
\setval{olmo.concur.minus.base.longbench-write}{+11.0}
\setval{olmo.concur.minus.base.longbench-write-abs}{11.0}
\setval{olmo.concur.minus.base.math}{+24.6}
\setval{olmo.concur.minus.base.math-abs}{24.6}
\setval{olmo.concur.minus.base.math500}{+20.4}
\setval{olmo.concur.minus.base.math500-abs}{20.4}
\setval{olmo.concur.minus.base.mbpp-plus}{+6.3}
\setval{olmo.concur.minus.base.mbpp-plus-abs}{6.3}
\setval{olmo.concur.minus.base.minerva}{+9.5}
\setval{olmo.concur.minus.base.minerva-abs}{9.5}
\setval{olmo.concur.minus.base.olympiad-bench}{+26.1}
\setval{olmo.concur.minus.base.olympiad-bench-abs}{26.1}
\setval{olmo.concur.minus.base.writing}{+14.3}
\setval{olmo.concur.minus.base.writing-abs}{14.3}
\setval{olmo.concur.minus.base.writingbench}{+20.7}
\setval{olmo.concur.minus.base.writingbench-abs}{20.7}
\setval{olmo.concur.minus.best.bench.aime24}{$-$0.2}
\setval{olmo.concur.minus.best.bench.aime25}{+4.1}
\setval{olmo.concur.minus.best.bench.amc23}{+6.1}
\setval{olmo.concur.minus.best.bench.creative-writing-v3}{$-$1.5}
\setval{olmo.concur.minus.best.bench.guidebench}{$-$4.2}
\setval{olmo.concur.minus.best.bench.humaneval-plus}{+5.4}
\setval{olmo.concur.minus.best.bench.ifbench}{$-$0.7}
\setval{olmo.concur.minus.best.bench.ifeval}{$-$0.8}
\setval{olmo.concur.minus.best.bench.livecodebench}{+7.1}
\setval{olmo.concur.minus.best.bench.longbench-write}{+0.3}
\setval{olmo.concur.minus.best.bench.math500}{+3.6}
\setval{olmo.concur.minus.best.bench.mbpp-plus}{+1.7}
\setval{olmo.concur.minus.best.bench.minerva}{+2.8}
\setval{olmo.concur.minus.best.bench.olympiad-bench}{+4.4}
\setval{olmo.concur.minus.best.bench.writingbench}{$-$3.0}
\setval{olmo.concur.minus.best.eval.avg}{+2.6}
\setval{olmo.concur.minus.best.eval.code}{+5.1}
\setval{olmo.concur.minus.best.eval.if}{$-$1.5}
\setval{olmo.concur.minus.best.eval.math}{+3.5}
\setval{olmo.concur.minus.best.eval.writing}{$-$0.5}
\setval{olmo.concur.minus.concur-free.aime24}{+9.6}
\setval{olmo.concur.minus.concur-free.aime24-abs}{9.6}
\setval{olmo.concur.minus.concur-free.aime25}{+10.1}
\setval{olmo.concur.minus.concur-free.aime25-abs}{10.1}
\setval{olmo.concur.minus.concur-free.amc23}{+16.0}
\setval{olmo.concur.minus.concur-free.amc23-abs}{16.0}
\setval{olmo.concur.minus.concur-free.avg}{+3.9}
\setval{olmo.concur.minus.concur-free.avg-abs}{3.9}
\setval{olmo.concur.minus.concur-free.code}{+4.6}
\setval{olmo.concur.minus.concur-free.code-abs}{4.6}
\setval{olmo.concur.minus.concur-free.creative-writing-v3}{$-$0.6}
\setval{olmo.concur.minus.concur-free.creative-writing-v3-abs}{0.6}
\setval{olmo.concur.minus.concur-free.guidebench}{$-$3.3}
\setval{olmo.concur.minus.concur-free.guidebench-abs}{3.3}
\setval{olmo.concur.minus.concur-free.humaneval-plus}{+6.1}
\setval{olmo.concur.minus.concur-free.humaneval-plus-abs}{6.1}
\setval{olmo.concur.minus.concur-free.if}{$-$0.8}
\setval{olmo.concur.minus.concur-free.if-abs}{0.8}
\setval{olmo.concur.minus.concur-free.ifbench}{+0.3}
\setval{olmo.concur.minus.concur-free.ifbench-abs}{0.3}
\setval{olmo.concur.minus.concur-free.ifeval}{+0.5}
\setval{olmo.concur.minus.concur-free.ifeval-abs}{0.5}
\setval{olmo.concur.minus.concur-free.judge2.avg}{+4.0}
\setval{olmo.concur.minus.concur-free.judge2.avg-abs}{4.0}
\setval{olmo.concur.minus.concur-free.judge2.creative-writing-v3}{+0.9}
\setval{olmo.concur.minus.concur-free.judge2.creative-writing-v3-abs}{0.9}
\setval{olmo.concur.minus.concur-free.judge2.longbench-write}{$-$5.5}
\setval{olmo.concur.minus.concur-free.judge2.longbench-write-abs}{5.5}
\setval{olmo.concur.minus.concur-free.judge2.writing}{$-$2.6}
\setval{olmo.concur.minus.concur-free.judge2.writing-abs}{2.6}
\setval{olmo.concur.minus.concur-free.judge2.writingbench}{$-$3.1}
\setval{olmo.concur.minus.concur-free.judge2.writingbench-abs}{3.1}
\setval{olmo.concur.minus.concur-free.livecodebench}{+4.7}
\setval{olmo.concur.minus.concur-free.livecodebench-abs}{4.7}
\setval{olmo.concur.minus.concur-free.longbench-write}{$-$3.6}
\setval{olmo.concur.minus.concur-free.longbench-write-abs}{3.6}
\setval{olmo.concur.minus.concur-free.math}{+9.3}
\setval{olmo.concur.minus.concur-free.math-abs}{9.3}
\setval{olmo.concur.minus.concur-free.math500}{+6.2}
\setval{olmo.concur.minus.concur-free.math500-abs}{6.2}
\setval{olmo.concur.minus.concur-free.mbpp-plus}{+3.0}
\setval{olmo.concur.minus.concur-free.mbpp-plus-abs}{3.0}
\setval{olmo.concur.minus.concur-free.minerva}{+4.9}
\setval{olmo.concur.minus.concur-free.minerva-abs}{4.9}
\setval{olmo.concur.minus.concur-free.olympiad-bench}{+9.1}
\setval{olmo.concur.minus.concur-free.olympiad-bench-abs}{9.1}
\setval{olmo.concur.minus.concur-free.writing}{$-$3.0}
\setval{olmo.concur.minus.concur-free.writing-abs}{3.0}
\setval{olmo.concur.minus.concur-free.writingbench}{$-$4.8}
\setval{olmo.concur.minus.concur-free.writingbench-abs}{4.8}
\setval{olmo.concur.minus.concur-tied.aime24}{+1.9}
\setval{olmo.concur.minus.concur-tied.aime24-abs}{1.9}
\setval{olmo.concur.minus.concur-tied.aime25}{+2.3}
\setval{olmo.concur.minus.concur-tied.aime25-abs}{2.3}
\setval{olmo.concur.minus.concur-tied.amc23}{+3.7}
\setval{olmo.concur.minus.concur-tied.amc23-abs}{3.7}
\setval{olmo.concur.minus.concur-tied.avg}{+0.2}
\setval{olmo.concur.minus.concur-tied.avg-abs}{0.2}
\setval{olmo.concur.minus.concur-tied.code}{$-$0.9}
\setval{olmo.concur.minus.concur-tied.code-abs}{0.9}
\setval{olmo.concur.minus.concur-tied.creative-writing-v3}{+0.2}
\setval{olmo.concur.minus.concur-tied.creative-writing-v3-abs}{0.2}
\setval{olmo.concur.minus.concur-tied.guidebench}{$-$1.6}
\setval{olmo.concur.minus.concur-tied.guidebench-abs}{1.6}
\setval{olmo.concur.minus.concur-tied.humaneval-plus}{$-$0.5}
\setval{olmo.concur.minus.concur-tied.humaneval-plus-abs}{0.5}
\setval{olmo.concur.minus.concur-tied.if}{$-$1.1}
\setval{olmo.concur.minus.concur-tied.if-abs}{1.1}
\setval{olmo.concur.minus.concur-tied.ifbench}{$-$1.7}
\setval{olmo.concur.minus.concur-tied.ifbench-abs}{1.7}
\setval{olmo.concur.minus.concur-tied.ifeval}{+0.1}
\setval{olmo.concur.minus.concur-tied.ifeval-abs}{0.1}
\setval{olmo.concur.minus.concur-tied.judge2.avg}{+0.2}
\setval{olmo.concur.minus.concur-tied.judge2.avg-abs}{0.2}
\setval{olmo.concur.minus.concur-tied.judge2.creative-writing-v3}{+1.3}
\setval{olmo.concur.minus.concur-tied.judge2.creative-writing-v3-abs}{1.3}
\setval{olmo.concur.minus.concur-tied.judge2.longbench-write}{$-$2.1}
\setval{olmo.concur.minus.concur-tied.judge2.longbench-write-abs}{2.1}
\setval{olmo.concur.minus.concur-tied.judge2.writing}{$-$0.5}
\setval{olmo.concur.minus.concur-tied.judge2.writing-abs}{0.5}
\setval{olmo.concur.minus.concur-tied.judge2.writingbench}{$-$0.7}
\setval{olmo.concur.minus.concur-tied.judge2.writingbench-abs}{0.7}
\setval{olmo.concur.minus.concur-tied.livecodebench}{+1.3}
\setval{olmo.concur.minus.concur-tied.livecodebench-abs}{1.3}
\setval{olmo.concur.minus.concur-tied.longbench-write}{$-$0.7}
\setval{olmo.concur.minus.concur-tied.longbench-write-abs}{0.7}
\setval{olmo.concur.minus.concur-tied.math}{+1.8}
\setval{olmo.concur.minus.concur-tied.math-abs}{1.8}
\setval{olmo.concur.minus.concur-tied.math500}{+1.1}
\setval{olmo.concur.minus.concur-tied.math500-abs}{1.1}
\setval{olmo.concur.minus.concur-tied.mbpp-plus}{$-$3.5}
\setval{olmo.concur.minus.concur-tied.mbpp-plus-abs}{3.5}
\setval{olmo.concur.minus.concur-tied.minerva}{+0.8}
\setval{olmo.concur.minus.concur-tied.minerva-abs}{0.8}
\setval{olmo.concur.minus.concur-tied.olympiad-bench}{+0.8}
\setval{olmo.concur.minus.concur-tied.olympiad-bench-abs}{0.8}
\setval{olmo.concur.minus.concur-tied.writing}{$-$0.4}
\setval{olmo.concur.minus.concur-tied.writing-abs}{0.4}
\setval{olmo.concur.minus.concur-tied.writingbench}{$-$0.7}
\setval{olmo.concur.minus.concur-tied.writingbench-abs}{0.7}
\setval{olmo.concur.minus.gd2po.aime24}{+8.0}
\setval{olmo.concur.minus.gd2po.aime24-abs}{8.0}
\setval{olmo.concur.minus.gd2po.aime25}{+7.9}
\setval{olmo.concur.minus.gd2po.aime25-abs}{7.9}
\setval{olmo.concur.minus.gd2po.amc23}{+11.3}
\setval{olmo.concur.minus.gd2po.amc23-abs}{11.3}
\setval{olmo.concur.minus.gd2po.avg}{+3.7}
\setval{olmo.concur.minus.gd2po.avg-abs}{3.7}
\setval{olmo.concur.minus.gd2po.code}{+5.3}
\setval{olmo.concur.minus.gd2po.code-abs}{5.3}
\setval{olmo.concur.minus.gd2po.creative-writing-v3}{+0.9}
\setval{olmo.concur.minus.gd2po.creative-writing-v3-abs}{0.9}
\setval{olmo.concur.minus.gd2po.guidebench}{$-$4.2}
\setval{olmo.concur.minus.gd2po.guidebench-abs}{4.2}
\setval{olmo.concur.minus.gd2po.humaneval-plus}{+6.0}
\setval{olmo.concur.minus.gd2po.humaneval-plus-abs}{6.0}
\setval{olmo.concur.minus.gd2po.if}{$-$1.5}
\setval{olmo.concur.minus.gd2po.if-abs}{1.5}
\setval{olmo.concur.minus.gd2po.ifbench}{$-$0.3}
\setval{olmo.concur.minus.gd2po.ifbench-abs}{0.3}
\setval{olmo.concur.minus.gd2po.ifeval}{+0.1}
\setval{olmo.concur.minus.gd2po.ifeval-abs}{0.1}
\setval{olmo.concur.minus.gd2po.judge2.avg}{+3.3}
\setval{olmo.concur.minus.gd2po.judge2.avg-abs}{3.3}
\setval{olmo.concur.minus.gd2po.judge2.creative-writing-v3}{+0.6}
\setval{olmo.concur.minus.gd2po.judge2.creative-writing-v3-abs}{0.6}
\setval{olmo.concur.minus.gd2po.judge2.longbench-write}{$-$1.5}
\setval{olmo.concur.minus.gd2po.judge2.longbench-write-abs}{1.5}
\setval{olmo.concur.minus.gd2po.judge2.writing}{$-$1.5}
\setval{olmo.concur.minus.gd2po.judge2.writing-abs}{1.5}
\setval{olmo.concur.minus.gd2po.judge2.writingbench}{$-$3.5}
\setval{olmo.concur.minus.gd2po.judge2.writingbench-abs}{3.5}
\setval{olmo.concur.minus.gd2po.livecodebench}{+7.1}
\setval{olmo.concur.minus.gd2po.livecodebench-abs}{7.1}
\setval{olmo.concur.minus.gd2po.longbench-write}{+0.3}
\setval{olmo.concur.minus.gd2po.longbench-write-abs}{0.3}
\setval{olmo.concur.minus.gd2po.math}{+7.2}
\setval{olmo.concur.minus.gd2po.math-abs}{7.2}
\setval{olmo.concur.minus.gd2po.math500}{+4.8}
\setval{olmo.concur.minus.gd2po.math500-abs}{4.8}
\setval{olmo.concur.minus.gd2po.mbpp-plus}{+2.8}
\setval{olmo.concur.minus.gd2po.mbpp-plus-abs}{2.8}
\setval{olmo.concur.minus.gd2po.minerva}{+4.7}
\setval{olmo.concur.minus.gd2po.minerva-abs}{4.7}
\setval{olmo.concur.minus.gd2po.olympiad-bench}{+6.3}
\setval{olmo.concur.minus.gd2po.olympiad-bench-abs}{6.3}
\setval{olmo.concur.minus.gd2po.writing}{+0.3}
\setval{olmo.concur.minus.gd2po.writing-abs}{0.3}
\setval{olmo.concur.minus.gd2po.writingbench}{$-$0.3}
\setval{olmo.concur.minus.gd2po.writingbench-abs}{0.3}
\setval{olmo.concur.minus.gdpo-noaux.aime24}{$-$2.0}
\setval{olmo.concur.minus.gdpo-noaux.aime24-abs}{2.0}
\setval{olmo.concur.minus.gdpo-noaux.aime25}{$-$1.5}
\setval{olmo.concur.minus.gdpo-noaux.aime25-abs}{1.5}
\setval{olmo.concur.minus.gdpo-noaux.amc23}{$-$0.4}
\setval{olmo.concur.minus.gdpo-noaux.amc23-abs}{0.4}
\setval{olmo.concur.minus.gdpo-noaux.avg}{+1.5}
\setval{olmo.concur.minus.gdpo-noaux.avg-abs}{1.5}
\setval{olmo.concur.minus.gdpo-noaux.code}{$-$2.2}
\setval{olmo.concur.minus.gdpo-noaux.code-abs}{2.2}
\setval{olmo.concur.minus.gdpo-noaux.creative-writing-v3}{+11.0}
\setval{olmo.concur.minus.gdpo-noaux.creative-writing-v3-abs}{11.0}
\setval{olmo.concur.minus.gdpo-noaux.guidebench}{+1.2}
\setval{olmo.concur.minus.gdpo-noaux.guidebench-abs}{1.2}
\setval{olmo.concur.minus.gdpo-noaux.humaneval-plus}{$-$2.4}
\setval{olmo.concur.minus.gdpo-noaux.humaneval-plus-abs}{2.4}
\setval{olmo.concur.minus.gdpo-noaux.if}{+0.9}
\setval{olmo.concur.minus.gdpo-noaux.if-abs}{0.9}
\setval{olmo.concur.minus.gdpo-noaux.ifbench}{$-$1.3}
\setval{olmo.concur.minus.gdpo-noaux.ifbench-abs}{1.3}
\setval{olmo.concur.minus.gdpo-noaux.ifeval}{+3.0}
\setval{olmo.concur.minus.gdpo-noaux.ifeval-abs}{3.0}
\setval{olmo.concur.minus.gdpo-noaux.judge2.avg}{+0.9}
\setval{olmo.concur.minus.gdpo-noaux.judge2.avg-abs}{0.9}
\setval{olmo.concur.minus.gdpo-noaux.judge2.creative-writing-v3}{+7.5}
\setval{olmo.concur.minus.gdpo-noaux.judge2.creative-writing-v3-abs}{7.5}
\setval{olmo.concur.minus.gdpo-noaux.judge2.longbench-write}{+11.7}
\setval{olmo.concur.minus.gdpo-noaux.judge2.longbench-write-abs}{11.7}
\setval{olmo.concur.minus.gdpo-noaux.judge2.writing}{+8.1}
\setval{olmo.concur.minus.gdpo-noaux.judge2.writing-abs}{8.1}
\setval{olmo.concur.minus.gdpo-noaux.judge2.writingbench}{+5.1}
\setval{olmo.concur.minus.gdpo-noaux.judge2.writingbench-abs}{5.1}
\setval{olmo.concur.minus.gdpo-noaux.livecodebench}{$-$1.2}
\setval{olmo.concur.minus.gdpo-noaux.livecodebench-abs}{1.2}
\setval{olmo.concur.minus.gdpo-noaux.longbench-write}{+12.5}
\setval{olmo.concur.minus.gdpo-noaux.longbench-write-abs}{12.5}
\setval{olmo.concur.minus.gdpo-noaux.math}{$-$1.1}
\setval{olmo.concur.minus.gdpo-noaux.math-abs}{1.1}
\setval{olmo.concur.minus.gdpo-noaux.math500}{$-$0.3}
\setval{olmo.concur.minus.gdpo-noaux.math500-abs}{0.3}
\setval{olmo.concur.minus.gdpo-noaux.mbpp-plus}{$-$3.0}
\setval{olmo.concur.minus.gdpo-noaux.mbpp-plus-abs}{3.0}
\setval{olmo.concur.minus.gdpo-noaux.minerva}{$-$1.3}
\setval{olmo.concur.minus.gdpo-noaux.minerva-abs}{1.3}
\setval{olmo.concur.minus.gdpo-noaux.olympiad-bench}{$-$1.3}
\setval{olmo.concur.minus.gdpo-noaux.olympiad-bench-abs}{1.3}
\setval{olmo.concur.minus.gdpo-noaux.writing}{+11.2}
\setval{olmo.concur.minus.gdpo-noaux.writing-abs}{11.2}
\setval{olmo.concur.minus.gdpo-noaux.writingbench}{+10.0}
\setval{olmo.concur.minus.gdpo-noaux.writingbench-abs}{10.0}
\setval{olmo.concur.minus.gdpo-w0.5.aime24}{+6.8}
\setval{olmo.concur.minus.gdpo-w0.5.aime24-abs}{6.8}
\setval{olmo.concur.minus.gdpo-w0.5.aime25}{+4.9}
\setval{olmo.concur.minus.gdpo-w0.5.aime25-abs}{4.9}
\setval{olmo.concur.minus.gdpo-w0.5.amc23}{+9.5}
\setval{olmo.concur.minus.gdpo-w0.5.amc23-abs}{9.5}
\setval{olmo.concur.minus.gdpo-w0.5.avg}{+3.7}
\setval{olmo.concur.minus.gdpo-w0.5.avg-abs}{3.7}
\setval{olmo.concur.minus.gdpo-w0.5.code}{+4.4}
\setval{olmo.concur.minus.gdpo-w0.5.code-abs}{4.4}
\setval{olmo.concur.minus.gdpo-w0.5.creative-writing-v3}{+1.1}
\setval{olmo.concur.minus.gdpo-w0.5.creative-writing-v3-abs}{1.1}
\setval{olmo.concur.minus.gdpo-w0.5.guidebench}{+5.0}
\setval{olmo.concur.minus.gdpo-w0.5.guidebench-abs}{5.0}
\setval{olmo.concur.minus.gdpo-w0.5.humaneval-plus}{+5.0}
\setval{olmo.concur.minus.gdpo-w0.5.humaneval-plus-abs}{5.0}
\setval{olmo.concur.minus.gdpo-w0.5.if}{+1.5}
\setval{olmo.concur.minus.gdpo-w0.5.if-abs}{1.5}
\setval{olmo.concur.minus.gdpo-w0.5.ifbench}{$-$0.7}
\setval{olmo.concur.minus.gdpo-w0.5.ifbench-abs}{0.7}
\setval{olmo.concur.minus.gdpo-w0.5.ifeval}{+0.2}
\setval{olmo.concur.minus.gdpo-w0.5.ifeval-abs}{0.2}
\setval{olmo.concur.minus.gdpo-w0.5.judge2.avg}{+3.3}
\setval{olmo.concur.minus.gdpo-w0.5.judge2.avg-abs}{3.3}
\setval{olmo.concur.minus.gdpo-w0.5.judge2.creative-writing-v3}{$-$1.0}
\setval{olmo.concur.minus.gdpo-w0.5.judge2.creative-writing-v3-abs}{1.0}
\setval{olmo.concur.minus.gdpo-w0.5.judge2.longbench-write}{+1.8}
\setval{olmo.concur.minus.gdpo-w0.5.judge2.longbench-write-abs}{1.8}
\setval{olmo.concur.minus.gdpo-w0.5.judge2.writing}{$-$0.3}
\setval{olmo.concur.minus.gdpo-w0.5.judge2.writing-abs}{0.3}
\setval{olmo.concur.minus.gdpo-w0.5.judge2.writingbench}{$-$1.8}
\setval{olmo.concur.minus.gdpo-w0.5.judge2.writingbench-abs}{1.8}
\setval{olmo.concur.minus.gdpo-w0.5.livecodebench}{+5.5}
\setval{olmo.concur.minus.gdpo-w0.5.livecodebench-abs}{5.5}
\setval{olmo.concur.minus.gdpo-w0.5.longbench-write}{+4.8}
\setval{olmo.concur.minus.gdpo-w0.5.longbench-write-abs}{4.8}
\setval{olmo.concur.minus.gdpo-w0.5.math}{+5.5}
\setval{olmo.concur.minus.gdpo-w0.5.math-abs}{5.5}
\setval{olmo.concur.minus.gdpo-w0.5.math500}{+3.5}
\setval{olmo.concur.minus.gdpo-w0.5.math500-abs}{3.5}
\setval{olmo.concur.minus.gdpo-w0.5.mbpp-plus}{+2.6}
\setval{olmo.concur.minus.gdpo-w0.5.mbpp-plus-abs}{2.6}
\setval{olmo.concur.minus.gdpo-w0.5.minerva}{+3.3}
\setval{olmo.concur.minus.gdpo-w0.5.minerva-abs}{3.3}
\setval{olmo.concur.minus.gdpo-w0.5.olympiad-bench}{+4.9}
\setval{olmo.concur.minus.gdpo-w0.5.olympiad-bench-abs}{4.9}
\setval{olmo.concur.minus.gdpo-w0.5.writing}{+1.6}
\setval{olmo.concur.minus.gdpo-w0.5.writing-abs}{1.6}
\setval{olmo.concur.minus.gdpo-w0.5.writingbench}{$-$1.1}
\setval{olmo.concur.minus.gdpo-w0.5.writingbench-abs}{1.1}
\setval{olmo.concur.minus.gdpo.aime24}{+8.5}
\setval{olmo.concur.minus.gdpo.aime24-abs}{8.5}
\setval{olmo.concur.minus.gdpo.aime25}{+9.6}
\setval{olmo.concur.minus.gdpo.aime25-abs}{9.6}
\setval{olmo.concur.minus.gdpo.amc23}{+14.0}
\setval{olmo.concur.minus.gdpo.amc23-abs}{14.0}
\setval{olmo.concur.minus.gdpo.avg}{+5.5}
\setval{olmo.concur.minus.gdpo.avg-abs}{5.5}
\setval{olmo.concur.minus.gdpo.code}{+5.5}
\setval{olmo.concur.minus.gdpo.code-abs}{5.5}
\setval{olmo.concur.minus.gdpo.creative-writing-v3}{+2.1}
\setval{olmo.concur.minus.gdpo.creative-writing-v3-abs}{2.1}
\setval{olmo.concur.minus.gdpo.guidebench}{+6.0}
\setval{olmo.concur.minus.gdpo.guidebench-abs}{6.0}
\setval{olmo.concur.minus.gdpo.humaneval-plus}{+6.4}
\setval{olmo.concur.minus.gdpo.humaneval-plus-abs}{6.4}
\setval{olmo.concur.minus.gdpo.if}{+3.2}
\setval{olmo.concur.minus.gdpo.if-abs}{3.2}
\setval{olmo.concur.minus.gdpo.ifbench}{+2.3}
\setval{olmo.concur.minus.gdpo.ifbench-abs}{2.3}
\setval{olmo.concur.minus.gdpo.ifeval}{+1.3}
\setval{olmo.concur.minus.gdpo.ifeval-abs}{1.3}
\setval{olmo.concur.minus.gdpo.judge2.avg}{+5.0}
\setval{olmo.concur.minus.gdpo.judge2.avg-abs}{5.0}
\setval{olmo.concur.minus.gdpo.judge2.creative-writing-v3}{+0.6}
\setval{olmo.concur.minus.gdpo.judge2.creative-writing-v3-abs}{0.6}
\setval{olmo.concur.minus.gdpo.judge2.longbench-write}{+2.7}
\setval{olmo.concur.minus.gdpo.judge2.longbench-write-abs}{2.7}
\setval{olmo.concur.minus.gdpo.judge2.writing}{+0.4}
\setval{olmo.concur.minus.gdpo.judge2.writing-abs}{0.4}
\setval{olmo.concur.minus.gdpo.judge2.writingbench}{$-$2.2}
\setval{olmo.concur.minus.gdpo.judge2.writingbench-abs}{2.2}
\setval{olmo.concur.minus.gdpo.livecodebench}{+7.2}
\setval{olmo.concur.minus.gdpo.livecodebench-abs}{7.2}
\setval{olmo.concur.minus.gdpo.longbench-write}{+5.2}
\setval{olmo.concur.minus.gdpo.longbench-write-abs}{5.2}
\setval{olmo.concur.minus.gdpo.math}{+8.1}
\setval{olmo.concur.minus.gdpo.math-abs}{8.1}
\setval{olmo.concur.minus.gdpo.math500}{+5.3}
\setval{olmo.concur.minus.gdpo.math500-abs}{5.3}
\setval{olmo.concur.minus.gdpo.mbpp-plus}{+2.8}
\setval{olmo.concur.minus.gdpo.mbpp-plus-abs}{2.8}
\setval{olmo.concur.minus.gdpo.minerva}{+4.4}
\setval{olmo.concur.minus.gdpo.minerva-abs}{4.4}
\setval{olmo.concur.minus.gdpo.olympiad-bench}{+6.5}
\setval{olmo.concur.minus.gdpo.olympiad-bench-abs}{6.5}
\setval{olmo.concur.minus.gdpo.writing}{+2.5}
\setval{olmo.concur.minus.gdpo.writing-abs}{2.5}
\setval{olmo.concur.minus.gdpo.writingbench}{+0.1}
\setval{olmo.concur.minus.gdpo.writingbench-abs}{0.1}
\setval{olmo.concur.minus.geoma.aime24}{+12.6}
\setval{olmo.concur.minus.geoma.aime24-abs}{12.6}
\setval{olmo.concur.minus.geoma.aime25}{+10.2}
\setval{olmo.concur.minus.geoma.aime25-abs}{10.2}
\setval{olmo.concur.minus.geoma.amc23}{+19.6}
\setval{olmo.concur.minus.geoma.amc23-abs}{19.6}
\setval{olmo.concur.minus.geoma.avg}{+6.4}
\setval{olmo.concur.minus.geoma.avg-abs}{6.4}
\setval{olmo.concur.minus.geoma.code}{+6.3}
\setval{olmo.concur.minus.geoma.code-abs}{6.3}
\setval{olmo.concur.minus.geoma.creative-writing-v3}{$-$0.4}
\setval{olmo.concur.minus.geoma.creative-writing-v3-abs}{0.4}
\setval{olmo.concur.minus.geoma.guidebench}{+10.3}
\setval{olmo.concur.minus.geoma.guidebench-abs}{10.3}
\setval{olmo.concur.minus.geoma.humaneval-plus}{+8.1}
\setval{olmo.concur.minus.geoma.humaneval-plus-abs}{8.1}
\setval{olmo.concur.minus.geoma.if}{+3.9}
\setval{olmo.concur.minus.geoma.if-abs}{3.9}
\setval{olmo.concur.minus.geoma.ifbench}{+1.7}
\setval{olmo.concur.minus.geoma.ifbench-abs}{1.7}
\setval{olmo.concur.minus.geoma.ifeval}{$-$0.2}
\setval{olmo.concur.minus.geoma.ifeval-abs}{0.2}
\setval{olmo.concur.minus.geoma.judge2.avg}{+5.9}
\setval{olmo.concur.minus.geoma.judge2.avg-abs}{5.9}
\setval{olmo.concur.minus.geoma.judge2.creative-writing-v3}{$-$3.3}
\setval{olmo.concur.minus.geoma.judge2.creative-writing-v3-abs}{3.3}
\setval{olmo.concur.minus.geoma.judge2.longbench-write}{+0.2}
\setval{olmo.concur.minus.geoma.judge2.longbench-write-abs}{0.2}
\setval{olmo.concur.minus.geoma.judge2.writing}{$-$2.3}
\setval{olmo.concur.minus.geoma.judge2.writing-abs}{2.3}
\setval{olmo.concur.minus.geoma.judge2.writingbench}{$-$3.7}
\setval{olmo.concur.minus.geoma.judge2.writingbench-abs}{3.7}
\setval{olmo.concur.minus.geoma.livecodebench}{+8.0}
\setval{olmo.concur.minus.geoma.livecodebench-abs}{8.0}
\setval{olmo.concur.minus.geoma.longbench-write}{+3.2}
\setval{olmo.concur.minus.geoma.longbench-write-abs}{3.2}
\setval{olmo.concur.minus.geoma.math}{+10.7}
\setval{olmo.concur.minus.geoma.math-abs}{10.7}
\setval{olmo.concur.minus.geoma.math500}{+7.4}
\setval{olmo.concur.minus.geoma.math500-abs}{7.4}
\setval{olmo.concur.minus.geoma.mbpp-plus}{+2.8}
\setval{olmo.concur.minus.geoma.mbpp-plus-abs}{2.8}
\setval{olmo.concur.minus.geoma.minerva}{+4.8}
\setval{olmo.concur.minus.geoma.minerva-abs}{4.8}
\setval{olmo.concur.minus.geoma.olympiad-bench}{+9.7}
\setval{olmo.concur.minus.geoma.olympiad-bench-abs}{9.7}
\setval{olmo.concur.minus.geoma.writing}{+0.4}
\setval{olmo.concur.minus.geoma.writing-abs}{0.4}
\setval{olmo.concur.minus.geoma.writingbench}{$-$1.5}
\setval{olmo.concur.minus.geoma.writingbench-abs}{1.5}
\setval{olmo.concur.minus.grpo-nodiv.aime24}{+4.6}
\setval{olmo.concur.minus.grpo-nodiv.aime24-abs}{4.6}
\setval{olmo.concur.minus.grpo-nodiv.aime25}{+4.6}
\setval{olmo.concur.minus.grpo-nodiv.aime25-abs}{4.6}
\setval{olmo.concur.minus.grpo-nodiv.amc23}{+10.5}
\setval{olmo.concur.minus.grpo-nodiv.amc23-abs}{10.5}
\setval{olmo.concur.minus.grpo-nodiv.avg}{+3.4}
\setval{olmo.concur.minus.grpo-nodiv.avg-abs}{3.4}
\setval{olmo.concur.minus.grpo-nodiv.code}{+4.8}
\setval{olmo.concur.minus.grpo-nodiv.code-abs}{4.8}
\setval{olmo.concur.minus.grpo-nodiv.creative-writing-v3}{+2.2}
\setval{olmo.concur.minus.grpo-nodiv.creative-writing-v3-abs}{2.2}
\setval{olmo.concur.minus.grpo-nodiv.guidebench}{+1.4}
\setval{olmo.concur.minus.grpo-nodiv.guidebench-abs}{1.4}
\setval{olmo.concur.minus.grpo-nodiv.humaneval-plus}{+5.4}
\setval{olmo.concur.minus.grpo-nodiv.humaneval-plus-abs}{5.4}
\setval{olmo.concur.minus.grpo-nodiv.if}{+0.7}
\setval{olmo.concur.minus.grpo-nodiv.if-abs}{0.7}
\setval{olmo.concur.minus.grpo-nodiv.ifbench}{+0.7}
\setval{olmo.concur.minus.grpo-nodiv.ifbench-abs}{0.7}
\setval{olmo.concur.minus.grpo-nodiv.ifeval}{+0.0}
\setval{olmo.concur.minus.grpo-nodiv.ifeval-abs}{0.0}
\setval{olmo.concur.minus.grpo-nodiv.judge2.avg}{+2.7}
\setval{olmo.concur.minus.grpo-nodiv.judge2.avg-abs}{2.7}
\setval{olmo.concur.minus.grpo-nodiv.judge2.creative-writing-v3}{$-$0.4}
\setval{olmo.concur.minus.grpo-nodiv.judge2.creative-writing-v3-abs}{0.4}
\setval{olmo.concur.minus.grpo-nodiv.judge2.longbench-write}{$-$2.8}
\setval{olmo.concur.minus.grpo-nodiv.judge2.longbench-write-abs}{2.8}
\setval{olmo.concur.minus.grpo-nodiv.judge2.writing}{$-$1.7}
\setval{olmo.concur.minus.grpo-nodiv.judge2.writing-abs}{1.7}
\setval{olmo.concur.minus.grpo-nodiv.judge2.writingbench}{$-$2.1}
\setval{olmo.concur.minus.grpo-nodiv.judge2.writingbench-abs}{2.1}
\setval{olmo.concur.minus.grpo-nodiv.livecodebench}{+6.7}
\setval{olmo.concur.minus.grpo-nodiv.livecodebench-abs}{6.7}
\setval{olmo.concur.minus.grpo-nodiv.longbench-write}{+1.0}
\setval{olmo.concur.minus.grpo-nodiv.longbench-write-abs}{1.0}
\setval{olmo.concur.minus.grpo-nodiv.math}{+4.9}
\setval{olmo.concur.minus.grpo-nodiv.math-abs}{4.9}
\setval{olmo.concur.minus.grpo-nodiv.math500}{+3.0}
\setval{olmo.concur.minus.grpo-nodiv.math500-abs}{3.0}
\setval{olmo.concur.minus.grpo-nodiv.mbpp-plus}{+2.2}
\setval{olmo.concur.minus.grpo-nodiv.mbpp-plus-abs}{2.2}
\setval{olmo.concur.minus.grpo-nodiv.minerva}{+3.0}
\setval{olmo.concur.minus.grpo-nodiv.minerva-abs}{3.0}
\setval{olmo.concur.minus.grpo-nodiv.olympiad-bench}{+4.0}
\setval{olmo.concur.minus.grpo-nodiv.olympiad-bench-abs}{4.0}
\setval{olmo.concur.minus.grpo-nodiv.writing}{+1.8}
\setval{olmo.concur.minus.grpo-nodiv.writing-abs}{1.8}
\setval{olmo.concur.minus.grpo-nodiv.writingbench}{+2.3}
\setval{olmo.concur.minus.grpo-nodiv.writingbench-abs}{2.3}
\setval{olmo.concur.minus.grpo.aime24}{+6.9}
\setval{olmo.concur.minus.grpo.aime24-abs}{6.9}
\setval{olmo.concur.minus.grpo.aime25}{+7.5}
\setval{olmo.concur.minus.grpo.aime25-abs}{7.5}
\setval{olmo.concur.minus.grpo.amc23}{+12.7}
\setval{olmo.concur.minus.grpo.amc23-abs}{12.7}
\setval{olmo.concur.minus.grpo.avg}{+4.1}
\setval{olmo.concur.minus.grpo.avg-abs}{4.1}
\setval{olmo.concur.minus.grpo.code}{+5.1}
\setval{olmo.concur.minus.grpo.code-abs}{5.1}
\setval{olmo.concur.minus.grpo.creative-writing-v3}{$-$0.2}
\setval{olmo.concur.minus.grpo.creative-writing-v3-abs}{0.2}
\setval{olmo.concur.minus.grpo.guidebench}{+0.5}
\setval{olmo.concur.minus.grpo.guidebench-abs}{0.5}
\setval{olmo.concur.minus.grpo.humaneval-plus}{+5.4}
\setval{olmo.concur.minus.grpo.humaneval-plus-abs}{5.4}
\setval{olmo.concur.minus.grpo.if}{+0.3}
\setval{olmo.concur.minus.grpo.if-abs}{0.3}
\setval{olmo.concur.minus.grpo.ifbench}{+0.0}
\setval{olmo.concur.minus.grpo.ifbench-abs}{0.0}
\setval{olmo.concur.minus.grpo.ifeval}{+0.4}
\setval{olmo.concur.minus.grpo.ifeval-abs}{0.4}
\setval{olmo.concur.minus.grpo.judge2.avg}{+3.9}
\setval{olmo.concur.minus.grpo.judge2.avg-abs}{3.9}
\setval{olmo.concur.minus.grpo.judge2.creative-writing-v3}{$-$0.2}
\setval{olmo.concur.minus.grpo.judge2.creative-writing-v3-abs}{0.2}
\setval{olmo.concur.minus.grpo.judge2.longbench-write}{+1.3}
\setval{olmo.concur.minus.grpo.judge2.longbench-write-abs}{1.3}
\setval{olmo.concur.minus.grpo.judge2.writing}{$-$0.8}
\setval{olmo.concur.minus.grpo.judge2.writing-abs}{0.8}
\setval{olmo.concur.minus.grpo.judge2.writingbench}{$-$3.6}
\setval{olmo.concur.minus.grpo.judge2.writingbench-abs}{3.6}
\setval{olmo.concur.minus.grpo.livecodebench}{+8.0}
\setval{olmo.concur.minus.grpo.livecodebench-abs}{8.0}
\setval{olmo.concur.minus.grpo.longbench-write}{+4.3}
\setval{olmo.concur.minus.grpo.longbench-write-abs}{4.3}
\setval{olmo.concur.minus.grpo.math}{+7.4}
\setval{olmo.concur.minus.grpo.math-abs}{7.4}
\setval{olmo.concur.minus.grpo.math500}{+5.6}
\setval{olmo.concur.minus.grpo.math500-abs}{5.6}
\setval{olmo.concur.minus.grpo.mbpp-plus}{+1.7}
\setval{olmo.concur.minus.grpo.mbpp-plus-abs}{1.7}
\setval{olmo.concur.minus.grpo.minerva}{+4.9}
\setval{olmo.concur.minus.grpo.minerva-abs}{4.9}
\setval{olmo.concur.minus.grpo.olympiad-bench}{+6.7}
\setval{olmo.concur.minus.grpo.olympiad-bench-abs}{6.7}
\setval{olmo.concur.minus.grpo.writing}{+0.4}
\setval{olmo.concur.minus.grpo.writing-abs}{0.4}
\setval{olmo.concur.minus.grpo.writingbench}{$-$3.0}
\setval{olmo.concur.minus.grpo.writingbench-abs}{3.0}
\setval{olmo.concur.minus.mt-grpo.aime24}{$-$0.2}
\setval{olmo.concur.minus.mt-grpo.aime24-abs}{0.2}
\setval{olmo.concur.minus.mt-grpo.aime25}{+4.1}
\setval{olmo.concur.minus.mt-grpo.aime25-abs}{4.1}
\setval{olmo.concur.minus.mt-grpo.amc23}{+6.1}
\setval{olmo.concur.minus.mt-grpo.amc23-abs}{6.1}
\setval{olmo.concur.minus.mt-grpo.avg}{+2.6}
\setval{olmo.concur.minus.mt-grpo.avg-abs}{2.6}
\setval{olmo.concur.minus.mt-grpo.code}{+5.8}
\setval{olmo.concur.minus.mt-grpo.code-abs}{5.8}
\setval{olmo.concur.minus.mt-grpo.creative-writing-v3}{$-$1.5}
\setval{olmo.concur.minus.mt-grpo.creative-writing-v3-abs}{1.5}
\setval{olmo.concur.minus.mt-grpo.guidebench}{+2.4}
\setval{olmo.concur.minus.mt-grpo.guidebench-abs}{2.4}
\setval{olmo.concur.minus.mt-grpo.humaneval-plus}{+7.0}
\setval{olmo.concur.minus.mt-grpo.humaneval-plus-abs}{7.0}
\setval{olmo.concur.minus.mt-grpo.if}{+1.0}
\setval{olmo.concur.minus.mt-grpo.if-abs}{1.0}
\setval{olmo.concur.minus.mt-grpo.ifbench}{+0.0}
\setval{olmo.concur.minus.mt-grpo.ifbench-abs}{0.0}
\setval{olmo.concur.minus.mt-grpo.ifeval}{+0.7}
\setval{olmo.concur.minus.mt-grpo.ifeval-abs}{0.7}
\setval{olmo.concur.minus.mt-grpo.judge2.avg}{+2.3}
\setval{olmo.concur.minus.mt-grpo.judge2.avg-abs}{2.3}
\setval{olmo.concur.minus.mt-grpo.judge2.creative-writing-v3}{$-$1.4}
\setval{olmo.concur.minus.mt-grpo.judge2.creative-writing-v3-abs}{1.4}
\setval{olmo.concur.minus.mt-grpo.judge2.longbench-write}{$-$2.1}
\setval{olmo.concur.minus.mt-grpo.judge2.longbench-write-abs}{2.1}
\setval{olmo.concur.minus.mt-grpo.judge2.writing}{$-$2.0}
\setval{olmo.concur.minus.mt-grpo.judge2.writing-abs}{2.0}
\setval{olmo.concur.minus.mt-grpo.judge2.writingbench}{$-$2.6}
\setval{olmo.concur.minus.mt-grpo.judge2.writingbench-abs}{2.6}
\setval{olmo.concur.minus.mt-grpo.livecodebench}{+7.7}
\setval{olmo.concur.minus.mt-grpo.livecodebench-abs}{7.7}
\setval{olmo.concur.minus.mt-grpo.longbench-write}{+2.1}
\setval{olmo.concur.minus.mt-grpo.longbench-write-abs}{2.1}
\setval{olmo.concur.minus.mt-grpo.math}{+3.5}
\setval{olmo.concur.minus.mt-grpo.math-abs}{3.5}
\setval{olmo.concur.minus.mt-grpo.math500}{+3.6}
\setval{olmo.concur.minus.mt-grpo.math500-abs}{3.6}
\setval{olmo.concur.minus.mt-grpo.mbpp-plus}{+2.6}
\setval{olmo.concur.minus.mt-grpo.mbpp-plus-abs}{2.6}
\setval{olmo.concur.minus.mt-grpo.minerva}{+2.8}
\setval{olmo.concur.minus.mt-grpo.minerva-abs}{2.8}
\setval{olmo.concur.minus.mt-grpo.olympiad-bench}{+4.4}
\setval{olmo.concur.minus.mt-grpo.olympiad-bench-abs}{4.4}
\setval{olmo.concur.minus.mt-grpo.writing}{$-$0.5}
\setval{olmo.concur.minus.mt-grpo.writing-abs}{0.5}
\setval{olmo.concur.minus.mt-grpo.writingbench}{$-$2.2}
\setval{olmo.concur.minus.mt-grpo.writingbench-abs}{2.2}
\setval{olmo.concur.minus.papo.aime24}{+14.2}
\setval{olmo.concur.minus.papo.aime24-abs}{14.2}
\setval{olmo.concur.minus.papo.aime25}{+11.9}
\setval{olmo.concur.minus.papo.aime25-abs}{11.9}
\setval{olmo.concur.minus.papo.amc23}{+24.1}
\setval{olmo.concur.minus.papo.amc23-abs}{24.1}
\setval{olmo.concur.minus.papo.avg}{+7.0}
\setval{olmo.concur.minus.papo.avg-abs}{7.0}
\setval{olmo.concur.minus.papo.code}{+6.3}
\setval{olmo.concur.minus.papo.code-abs}{6.3}
\setval{olmo.concur.minus.papo.creative-writing-v3}{+2.2}
\setval{olmo.concur.minus.papo.creative-writing-v3-abs}{2.2}
\setval{olmo.concur.minus.papo.guidebench}{$-$1.7}
\setval{olmo.concur.minus.papo.guidebench-abs}{1.7}
\setval{olmo.concur.minus.papo.humaneval-plus}{+8.0}
\setval{olmo.concur.minus.papo.humaneval-plus-abs}{8.0}
\setval{olmo.concur.minus.papo.if}{+0.4}
\setval{olmo.concur.minus.papo.if-abs}{0.4}
\setval{olmo.concur.minus.papo.ifbench}{+2.3}
\setval{olmo.concur.minus.papo.ifbench-abs}{2.3}
\setval{olmo.concur.minus.papo.ifeval}{+0.6}
\setval{olmo.concur.minus.papo.ifeval-abs}{0.6}
\setval{olmo.concur.minus.papo.judge2.avg}{+6.6}
\setval{olmo.concur.minus.papo.judge2.avg-abs}{6.6}
\setval{olmo.concur.minus.papo.judge2.creative-writing-v3}{+0.5}
\setval{olmo.concur.minus.papo.judge2.creative-writing-v3-abs}{0.5}
\setval{olmo.concur.minus.papo.judge2.longbench-write}{+2.7}
\setval{olmo.concur.minus.papo.judge2.longbench-write-abs}{2.7}
\setval{olmo.concur.minus.papo.judge2.writing}{+0.2}
\setval{olmo.concur.minus.papo.judge2.writing-abs}{0.2}
\setval{olmo.concur.minus.papo.judge2.writingbench}{$-$2.6}
\setval{olmo.concur.minus.papo.judge2.writingbench-abs}{2.6}
\setval{olmo.concur.minus.papo.livecodebench}{+8.5}
\setval{olmo.concur.minus.papo.livecodebench-abs}{8.5}
\setval{olmo.concur.minus.papo.longbench-write}{+3.8}
\setval{olmo.concur.minus.papo.longbench-write-abs}{3.8}
\setval{olmo.concur.minus.papo.math}{+13.1}
\setval{olmo.concur.minus.papo.math-abs}{13.1}
\setval{olmo.concur.minus.papo.math500}{+9.7}
\setval{olmo.concur.minus.papo.math500-abs}{9.7}
\setval{olmo.concur.minus.papo.mbpp-plus}{+2.5}
\setval{olmo.concur.minus.papo.mbpp-plus-abs}{2.5}
\setval{olmo.concur.minus.papo.minerva}{+6.2}
\setval{olmo.concur.minus.papo.minerva-abs}{6.2}
\setval{olmo.concur.minus.papo.olympiad-bench}{+12.7}
\setval{olmo.concur.minus.papo.olympiad-bench-abs}{12.7}
\setval{olmo.concur.minus.papo.writing}{+2.0}
\setval{olmo.concur.minus.papo.writing-abs}{2.0}
\setval{olmo.concur.minus.papo.writingbench}{+0.1}
\setval{olmo.concur.minus.papo.writingbench-abs}{0.1}
\setval{olmo.concur.minus.saw.aime24}{+10.2}
\setval{olmo.concur.minus.saw.aime24-abs}{10.2}
\setval{olmo.concur.minus.saw.aime25}{+7.0}
\setval{olmo.concur.minus.saw.aime25-abs}{7.0}
\setval{olmo.concur.minus.saw.amc23}{+16.6}
\setval{olmo.concur.minus.saw.amc23-abs}{16.6}
\setval{olmo.concur.minus.saw.avg}{+4.9}
\setval{olmo.concur.minus.saw.avg-abs}{4.9}
\setval{olmo.concur.minus.saw.code}{+5.7}
\setval{olmo.concur.minus.saw.code-abs}{5.7}
\setval{olmo.concur.minus.saw.creative-writing-v3}{+1.3}
\setval{olmo.concur.minus.saw.creative-writing-v3-abs}{1.3}
\setval{olmo.concur.minus.saw.guidebench}{+2.2}
\setval{olmo.concur.minus.saw.guidebench-abs}{2.2}
\setval{olmo.concur.minus.saw.humaneval-plus}{+6.4}
\setval{olmo.concur.minus.saw.humaneval-plus-abs}{6.4}
\setval{olmo.concur.minus.saw.if}{+0.2}
\setval{olmo.concur.minus.saw.if-abs}{0.2}
\setval{olmo.concur.minus.saw.ifbench}{$-$0.7}
\setval{olmo.concur.minus.saw.ifbench-abs}{0.7}
\setval{olmo.concur.minus.saw.ifeval}{$-$0.8}
\setval{olmo.concur.minus.saw.ifeval-abs}{0.8}
\setval{olmo.concur.minus.saw.judge2.avg}{+4.4}
\setval{olmo.concur.minus.saw.judge2.avg-abs}{4.4}
\setval{olmo.concur.minus.saw.judge2.creative-writing-v3}{$-$2.5}
\setval{olmo.concur.minus.saw.judge2.creative-writing-v3-abs}{2.5}
\setval{olmo.concur.minus.saw.judge2.longbench-write}{+1.0}
\setval{olmo.concur.minus.saw.judge2.longbench-write-abs}{1.0}
\setval{olmo.concur.minus.saw.judge2.writing}{$-$1.7}
\setval{olmo.concur.minus.saw.judge2.writing-abs}{1.7}
\setval{olmo.concur.minus.saw.judge2.writingbench}{$-$3.5}
\setval{olmo.concur.minus.saw.judge2.writingbench-abs}{3.5}
\setval{olmo.concur.minus.saw.livecodebench}{+8.5}
\setval{olmo.concur.minus.saw.livecodebench-abs}{8.5}
\setval{olmo.concur.minus.saw.longbench-write}{+3.6}
\setval{olmo.concur.minus.saw.longbench-write-abs}{3.6}
\setval{olmo.concur.minus.saw.math}{+8.8}
\setval{olmo.concur.minus.saw.math-abs}{8.8}
\setval{olmo.concur.minus.saw.math500}{+6.0}
\setval{olmo.concur.minus.saw.math500-abs}{6.0}
\setval{olmo.concur.minus.saw.mbpp-plus}{+2.1}
\setval{olmo.concur.minus.saw.mbpp-plus-abs}{2.1}
\setval{olmo.concur.minus.saw.minerva}{+4.5}
\setval{olmo.concur.minus.saw.minerva-abs}{4.5}
\setval{olmo.concur.minus.saw.olympiad-bench}{+8.6}
\setval{olmo.concur.minus.saw.olympiad-bench-abs}{8.6}
\setval{olmo.concur.minus.saw.writing}{+1.0}
\setval{olmo.concur.minus.saw.writing-abs}{1.0}
\setval{olmo.concur.minus.saw.writingbench}{$-$1.9}
\setval{olmo.concur.minus.saw.writingbench-abs}{1.9}
\setval{olmo.concur.passing-at-ceiling.code.early}{0.84}
\setval{olmo.concur.passing-at-ceiling.code.late}{0.88}
\setval{olmo.concur.passing-at-ceiling.code.run}{0.86}
\setval{olmo.concur.passing-at-ceiling.if.early}{0.48}
\setval{olmo.concur.passing-at-ceiling.if.late}{0.56}
\setval{olmo.concur.passing-at-ceiling.if.run}{0.52}
\setval{olmo.concur.passing-at-ceiling.math.early}{1.00}
\setval{olmo.concur.passing-at-ceiling.math.late}{1.00}
\setval{olmo.concur.passing-at-ceiling.math.run}{1.00}
\setval{olmo.concur.passing-at-ceiling.writing.early}{0.00}
\setval{olmo.concur.passing-at-ceiling.writing.late}{0.01}
\setval{olmo.concur.passing-at-ceiling.writing.run}{0.00}
\setval{olmo.concur.primary-cosine.code.early}{0.98}
\setval{olmo.concur.primary-cosine.code.late}{0.95}
\setval{olmo.concur.primary-cosine.code.run}{0.96}
\setval{olmo.concur.primary-cosine.if.early}{0.82}
\setval{olmo.concur.primary-cosine.if.late}{0.80}
\setval{olmo.concur.primary-cosine.if.run}{0.81}
\setval{olmo.concur.primary-cosine.math.early}{0.99}
\setval{olmo.concur.primary-cosine.math.late}{0.97}
\setval{olmo.concur.primary-cosine.math.run}{0.97}
\setval{olmo.concur.primary-cosine.writing.early}{0.95}
\setval{olmo.concur.primary-cosine.writing.late}{0.93}
\setval{olmo.concur.primary-cosine.writing.run}{0.94}
\setval{olmo.concur.primary-level.code.early}{0.58}
\setval{olmo.concur.primary-level.code.late}{0.86}
\setval{olmo.concur.primary-level.code.run}{0.78}
\setval{olmo.concur.primary-level.if.early}{0.63}
\setval{olmo.concur.primary-level.if.late}{0.73}
\setval{olmo.concur.primary-level.if.run}{0.68}
\setval{olmo.concur.primary-level.math.early}{0.35}
\setval{olmo.concur.primary-level.math.late}{0.73}
\setval{olmo.concur.primary-level.math.run}{0.63}
\setval{olmo.concur.primary-level.writing.early}{0.50}
\setval{olmo.concur.primary-level.writing.late}{0.75}
\setval{olmo.concur.primary-level.writing.run}{0.67}
\setval{olmo.concur.solved.code.early}{0.17}
\setval{olmo.concur.solved.code.late}{0.45}
\setval{olmo.concur.solved.code.run}{0.32}
\setval{olmo.concur.solved.if.early}{0.17}
\setval{olmo.concur.solved.if.late}{0.18}
\setval{olmo.concur.solved.if.run}{0.19}
\setval{olmo.concur.solved.math.early}{0.06}
\setval{olmo.concur.solved.math.late}{0.38}
\setval{olmo.concur.solved.math.run}{0.28}
\setval{olmo.concur.solved.writing.early}{0.00}
\setval{olmo.concur.solved.writing.late}{0.00}
\setval{olmo.concur.solved.writing.run}{0.00}
\setval{olmo.concur.trust.code.agreement}{$-$0.011}
\setval{olmo.concur.trust.code.errors}{$-$2.5}
\setval{olmo.concur.trust.code.first-free}{--}
\setval{olmo.concur.trust.code.free-share}{0.00}
\setval{olmo.concur.trust.if.agreement}{+0.053}
\setval{olmo.concur.trust.if.errors}{+16.9}
\setval{olmo.concur.trust.if.first-free}{4}
\setval{olmo.concur.trust.if.free-share}{0.99}
\setval{olmo.concur.trust.math.first-free}{--}
\setval{olmo.concur.trust.math.free-share}{0.00}
\setval{olmo.concur.trust.writing-rm.agreement}{+0.093}
\setval{olmo.concur.trust.writing-rm.errors}{+46.2}
\setval{olmo.concur.trust.writing-rm.first-free}{3}
\setval{olmo.concur.trust.writing-rm.free-share}{0.99}
\setval{olmo.concur.trust.writing.agreement}{$-$0.069}
\setval{olmo.concur.trust.writing.errors}{$-$41.2}
\setval{olmo.concur.trust.writing.first-free}{--}
\setval{olmo.concur.trust.writing.free-share}{0.00}
\setval{olmo.concur.update-share.code.early}{0.16}
\setval{olmo.concur.update-share.code.late}{0.12}
\setval{olmo.concur.update-share.code.run}{0.14}
\setval{olmo.concur.update-share.if.early}{0.20}
\setval{olmo.concur.update-share.if.late}{0.31}
\setval{olmo.concur.update-share.if.run}{0.27}
\setval{olmo.concur.update-share.math.early}{0.19}
\setval{olmo.concur.update-share.math.late}{0.10}
\setval{olmo.concur.update-share.math.run}{0.13}
\setval{olmo.concur.update-share.writing.early}{0.45}
\setval{olmo.concur.update-share.writing.late}{0.47}
\setval{olmo.concur.update-share.writing.run}{0.46}
\setval{olmo.estimators.aux-full.code.run-max}{0.94}
\setval{olmo.estimators.aux-full.code.run-min}{0.79}
\setval{olmo.estimators.aux-full.if.run-max}{0.91}
\setval{olmo.estimators.aux-full.if.run-min}{0.76}
\setval{olmo.estimators.aux-full.math.run-max}{0.73}
\setval{olmo.estimators.aux-full.math.run-min}{0.46}
\setval{olmo.estimators.aux-full.writing-rm.run-max}{0.17}
\setval{olmo.estimators.aux-full.writing-rm.run-min}{0.06}
\setval{olmo.estimators.aux-full.writing.run-max}{0.78}
\setval{olmo.estimators.aux-full.writing.run-min}{0.48}
\setval{olmo.estimators.avg-spread}{3.8}
\setval{olmo.estimators.bench.aime24.max}{36.6}
\setval{olmo.estimators.bench.aime24.min}{22.2}
\setval{olmo.estimators.bench.aime25.max}{28.1}
\setval{olmo.estimators.bench.aime25.min}{20.3}
\setval{olmo.estimators.bench.amc23.max}{77.4}
\setval{olmo.estimators.bench.amc23.min}{59.4}
\setval{olmo.estimators.bench.creative-writing-v3.max}{50.7}
\setval{olmo.estimators.bench.creative-writing-v3.min}{47.0}
\setval{olmo.estimators.bench.guidebench.max}{61.1}
\setval{olmo.estimators.bench.guidebench.min}{46.6}
\setval{olmo.estimators.bench.humaneval-plus.max}{77.3}
\setval{olmo.estimators.bench.humaneval-plus.min}{74.6}
\setval{olmo.estimators.bench.ifbench.max}{32.7}
\setval{olmo.estimators.bench.ifbench.min}{29.7}
\setval{olmo.estimators.bench.ifeval.max}{88.6}
\setval{olmo.estimators.bench.ifeval.min}{86.5}
\setval{olmo.estimators.bench.livecodebench.max}{23.2}
\setval{olmo.estimators.bench.livecodebench.min}{21.9}
\setval{olmo.estimators.bench.longbench-write.max}{68.2}
\setval{olmo.estimators.bench.longbench-write.min}{63.3}
\setval{olmo.estimators.bench.math500.max}{86.9}
\setval{olmo.estimators.bench.math500.min}{80.9}
\setval{olmo.estimators.bench.mbpp-plus.max}{65.8}
\setval{olmo.estimators.bench.mbpp-plus.min}{64.7}
\setval{olmo.estimators.bench.minerva.max}{42.0}
\setval{olmo.estimators.bench.minerva.min}{38.6}
\setval{olmo.estimators.bench.olympiad-bench.max}{58.0}
\setval{olmo.estimators.bench.olympiad-bench.min}{49.7}
\setval{olmo.estimators.bench.writingbench.max}{68.5}
\setval{olmo.estimators.bench.writingbench.min}{65.3}
\setval{olmo.estimators.eval.avg.max}{56.7}
\setval{olmo.estimators.eval.avg.min}{52.4}
\setval{olmo.estimators.eval.code.max}{55.2}
\setval{olmo.estimators.eval.code.min}{53.9}
\setval{olmo.estimators.eval.if.max}{60.4}
\setval{olmo.estimators.eval.if.min}{55.0}
\setval{olmo.estimators.eval.math.max}{54.8}
\setval{olmo.estimators.eval.math.min}{45.2}
\setval{olmo.estimators.eval.writing.max}{61.6}
\setval{olmo.estimators.eval.writing.min}{58.6}
\setval{olmo.estimators.minus.gdpo-noaux.avg-max}{$-$1.1}
\setval{olmo.estimators.minus.gdpo-noaux.avg-min}{$-$5.5}
\setval{olmo.gd2po.aux-change.code}{+0.15}
\setval{olmo.gd2po.aux-change.if}{+0.11}
\setval{olmo.gd2po.aux-change.math}{$-$0.10}
\setval{olmo.gd2po.aux-change.writing}{$-$0.02}
\setval{olmo.gd2po.aux-change.writing-rm}{+0.24}
\setval{olmo.gd2po.aux-full.code.early}{0.74}
\setval{olmo.gd2po.aux-full.code.late}{0.90}
\setval{olmo.gd2po.aux-full.code.run}{0.88}
\setval{olmo.gd2po.aux-full.if.early}{0.58}
\setval{olmo.gd2po.aux-full.if.late}{0.94}
\setval{olmo.gd2po.aux-full.if.run}{0.84}
\setval{olmo.gd2po.aux-full.math.early}{0.52}
\setval{olmo.gd2po.aux-full.math.late}{0.61}
\setval{olmo.gd2po.aux-full.math.run}{0.47}
\setval{olmo.gd2po.aux-full.writing-rm.early}{0.02}
\setval{olmo.gd2po.aux-full.writing-rm.late}{0.12}
\setval{olmo.gd2po.aux-full.writing-rm.run}{0.06}
\setval{olmo.gd2po.aux-full.writing.early}{0.55}
\setval{olmo.gd2po.aux-full.writing.late}{0.58}
\setval{olmo.gd2po.aux-full.writing.run}{0.48}
\setval{olmo.gd2po.aux-level.code.early}{0.89}
\setval{olmo.gd2po.aux-level.code.late}{0.96}
\setval{olmo.gd2po.aux-level.code.run}{0.96}
\setval{olmo.gd2po.aux-level.if.early}{0.90}
\setval{olmo.gd2po.aux-level.if.late}{0.99}
\setval{olmo.gd2po.aux-level.if.run}{0.97}
\setval{olmo.gd2po.aux-level.math.early}{0.90}
\setval{olmo.gd2po.aux-level.math.late}{0.80}
\setval{olmo.gd2po.aux-level.math.run}{0.77}
\setval{olmo.gd2po.aux-level.writing-rm.early}{0.60}
\setval{olmo.gd2po.aux-level.writing-rm.late}{0.81}
\setval{olmo.gd2po.aux-level.writing-rm.run}{0.73}
\setval{olmo.gd2po.aux-level.writing.early}{0.95}
\setval{olmo.gd2po.aux-level.writing.late}{0.93}
\setval{olmo.gd2po.aux-level.writing.run}{0.92}
\setval{olmo.gd2po.bench.aime24}{28.3}
\setval{olmo.gd2po.bench.aime25}{24.3}
\setval{olmo.gd2po.bench.amc23}{72.2}
\setval{olmo.gd2po.bench.creative-writing-v3}{48.3}
\setval{olmo.gd2po.bench.guidebench}{61.1}
\setval{olmo.gd2po.bench.humaneval-plus}{76.8}
\setval{olmo.gd2po.bench.ifbench}{32.3}
\setval{olmo.gd2po.bench.ifeval}{87.6}
\setval{olmo.gd2po.bench.livecodebench}{23.2}
\setval{olmo.gd2po.bench.longbench-write}{68.2}
\setval{olmo.gd2po.bench.math500}{85.7}
\setval{olmo.gd2po.bench.mbpp-plus}{64.8}
\setval{olmo.gd2po.bench.minerva}{40.1}
\setval{olmo.gd2po.bench.olympiad-bench}{56.1}
\setval{olmo.gd2po.bench.writingbench}{65.7}
\setval{olmo.gd2po.eval.avg}{55.7}
\setval{olmo.gd2po.eval.code}{54.9}
\setval{olmo.gd2po.eval.if}{60.4}
\setval{olmo.gd2po.eval.math}{51.1}
\setval{olmo.gd2po.eval.writing}{60.8}
\setval{olmo.gd2po.gate.code.against.early}{0.06}
\setval{olmo.gd2po.gate.code.against.late}{0.10}
\setval{olmo.gd2po.gate.code.against.run}{0.09}
\setval{olmo.gd2po.gate.code.agreement}{$-$0.017}
\setval{olmo.gd2po.gate.code.ceiling.early}{0.23}
\setval{olmo.gd2po.gate.code.ceiling.late}{0.23}
\setval{olmo.gd2po.gate.code.ceiling.run}{0.21}
\setval{olmo.gd2po.gate.code.cosine.early}{0.74}
\setval{olmo.gd2po.gate.code.cosine.late}{0.59}
\setval{olmo.gd2po.gate.code.cosine.run}{0.66}
\setval{olmo.gd2po.gate.if.against.early}{0.10}
\setval{olmo.gd2po.gate.if.against.late}{0.04}
\setval{olmo.gd2po.gate.if.against.run}{0.07}
\setval{olmo.gd2po.gate.if.agreement}{+0.040}
\setval{olmo.gd2po.gate.if.ceiling.early}{0.22}
\setval{olmo.gd2po.gate.if.ceiling.late}{0.16}
\setval{olmo.gd2po.gate.if.ceiling.run}{0.19}
\setval{olmo.gd2po.gate.if.cosine.early}{0.52}
\setval{olmo.gd2po.gate.if.cosine.late}{0.54}
\setval{olmo.gd2po.gate.if.cosine.run}{0.54}
\setval{olmo.gd2po.gate.math.against.early}{0.00}
\setval{olmo.gd2po.gate.math.against.late}{0.00}
\setval{olmo.gd2po.gate.math.against.run}{0.00}
\setval{olmo.gd2po.gate.math.ceiling.early}{0.15}
\setval{olmo.gd2po.gate.math.ceiling.late}{0.47}
\setval{olmo.gd2po.gate.math.ceiling.run}{0.42}
\setval{olmo.gd2po.gate.math.cosine.early}{0.90}
\setval{olmo.gd2po.gate.math.cosine.late}{0.65}
\setval{olmo.gd2po.gate.math.cosine.run}{0.68}
\setval{olmo.gd2po.gate.writing-rm.against.early}{0.21}
\setval{olmo.gd2po.gate.writing-rm.against.late}{0.49}
\setval{olmo.gd2po.gate.writing-rm.against.run}{0.43}
\setval{olmo.gd2po.gate.writing-rm.agreement}{+0.099}
\setval{olmo.gd2po.gate.writing-rm.ceiling.early}{0.00}
\setval{olmo.gd2po.gate.writing-rm.ceiling.late}{0.00}
\setval{olmo.gd2po.gate.writing-rm.ceiling.run}{0.00}
\setval{olmo.gd2po.gate.writing-rm.cosine.early}{0.70}
\setval{olmo.gd2po.gate.writing-rm.cosine.late}{0.27}
\setval{olmo.gd2po.gate.writing-rm.cosine.run}{0.38}
\setval{olmo.gd2po.gate.writing.against.early}{0.24}
\setval{olmo.gd2po.gate.writing.against.late}{0.58}
\setval{olmo.gd2po.gate.writing.against.run}{0.52}
\setval{olmo.gd2po.gate.writing.agreement}{$-$0.080}
\setval{olmo.gd2po.gate.writing.ceiling.early}{0.00}
\setval{olmo.gd2po.gate.writing.ceiling.late}{0.00}
\setval{olmo.gd2po.gate.writing.ceiling.run}{0.00}
\setval{olmo.gd2po.gate.writing.cosine.early}{0.61}
\setval{olmo.gd2po.gate.writing.cosine.late}{0.10}
\setval{olmo.gd2po.gate.writing.cosine.run}{0.21}
\setval{olmo.gd2po.judge2.avg}{53.2}
\setval{olmo.gd2po.judge2.creative-writing-v3}{37.9}
\setval{olmo.gd2po.judge2.longbench-write}{59.9}
\setval{olmo.gd2po.judge2.writing}{48.7}
\setval{olmo.gd2po.judge2.writingbench}{48.4}
\setval{olmo.gd2po.last-step}{300}
\setval{olmo.gd2po.minus.base.aime24}{+19.8}
\setval{olmo.gd2po.minus.base.aime24-abs}{19.8}
\setval{olmo.gd2po.minus.base.aime25}{+16.9}
\setval{olmo.gd2po.minus.base.aime25-abs}{16.9}
\setval{olmo.gd2po.minus.base.amc23}{+27.7}
\setval{olmo.gd2po.minus.base.amc23-abs}{27.7}
\setval{olmo.gd2po.minus.base.avg}{+11.5}
\setval{olmo.gd2po.minus.base.avg-abs}{11.5}
\setval{olmo.gd2po.minus.base.code}{+5.8}
\setval{olmo.gd2po.minus.base.code-abs}{5.8}
\setval{olmo.gd2po.minus.base.creative-writing-v3}{+10.4}
\setval{olmo.gd2po.minus.base.creative-writing-v3-abs}{10.4}
\setval{olmo.gd2po.minus.base.guidebench}{$-$0.3}
\setval{olmo.gd2po.minus.base.guidebench-abs}{0.3}
\setval{olmo.gd2po.minus.base.humaneval-plus}{+7.6}
\setval{olmo.gd2po.minus.base.humaneval-plus-abs}{7.6}
\setval{olmo.gd2po.minus.base.if}{+2.9}
\setval{olmo.gd2po.minus.base.if-abs}{2.9}
\setval{olmo.gd2po.minus.base.ifbench}{+5.7}
\setval{olmo.gd2po.minus.base.ifbench-abs}{5.7}
\setval{olmo.gd2po.minus.base.ifeval}{+3.2}
\setval{olmo.gd2po.minus.base.ifeval-abs}{3.2}
\setval{olmo.gd2po.minus.base.judge2.avg}{+10.3}
\setval{olmo.gd2po.minus.base.judge2.avg-abs}{10.3}
\setval{olmo.gd2po.minus.base.judge2.creative-writing-v3}{+6.2}
\setval{olmo.gd2po.minus.base.judge2.creative-writing-v3-abs}{6.2}
\setval{olmo.gd2po.minus.base.judge2.longbench-write}{+6.7}
\setval{olmo.gd2po.minus.base.judge2.longbench-write-abs}{6.7}
\setval{olmo.gd2po.minus.base.judge2.writing}{+7.9}
\setval{olmo.gd2po.minus.base.judge2.writing-abs}{7.9}
\setval{olmo.gd2po.minus.base.judge2.writingbench}{+10.7}
\setval{olmo.gd2po.minus.base.judge2.writingbench-abs}{10.7}
\setval{olmo.gd2po.minus.base.livecodebench}{+6.3}
\setval{olmo.gd2po.minus.base.livecodebench-abs}{6.3}
\setval{olmo.gd2po.minus.base.longbench-write}{+10.7}
\setval{olmo.gd2po.minus.base.longbench-write-abs}{10.7}
\setval{olmo.gd2po.minus.base.math}{+17.4}
\setval{olmo.gd2po.minus.base.math-abs}{17.4}
\setval{olmo.gd2po.minus.base.math500}{+15.6}
\setval{olmo.gd2po.minus.base.math500-abs}{15.6}
\setval{olmo.gd2po.minus.base.mbpp-plus}{+3.5}
\setval{olmo.gd2po.minus.base.mbpp-plus-abs}{3.5}
\setval{olmo.gd2po.minus.base.minerva}{+4.8}
\setval{olmo.gd2po.minus.base.minerva-abs}{4.8}
\setval{olmo.gd2po.minus.base.olympiad-bench}{+19.8}
\setval{olmo.gd2po.minus.base.olympiad-bench-abs}{19.8}
\setval{olmo.gd2po.minus.base.writing}{+14.0}
\setval{olmo.gd2po.minus.base.writing-abs}{14.0}
\setval{olmo.gd2po.minus.base.writingbench}{+21.0}
\setval{olmo.gd2po.minus.base.writingbench-abs}{21.0}
\setval{olmo.gd2po.minus.concur-free.aime24}{+1.6}
\setval{olmo.gd2po.minus.concur-free.aime24-abs}{1.6}
\setval{olmo.gd2po.minus.concur-free.aime25}{+2.2}
\setval{olmo.gd2po.minus.concur-free.aime25-abs}{2.2}
\setval{olmo.gd2po.minus.concur-free.amc23}{+4.7}
\setval{olmo.gd2po.minus.concur-free.amc23-abs}{4.7}
\setval{olmo.gd2po.minus.concur-free.avg}{+0.2}
\setval{olmo.gd2po.minus.concur-free.avg-abs}{0.2}
\setval{olmo.gd2po.minus.concur-free.code}{$-$0.7}
\setval{olmo.gd2po.minus.concur-free.code-abs}{0.7}
\setval{olmo.gd2po.minus.concur-free.creative-writing-v3}{$-$1.5}
\setval{olmo.gd2po.minus.concur-free.creative-writing-v3-abs}{1.5}
\setval{olmo.gd2po.minus.concur-free.guidebench}{+1.0}
\setval{olmo.gd2po.minus.concur-free.guidebench-abs}{1.0}
\setval{olmo.gd2po.minus.concur-free.humaneval-plus}{+0.1}
\setval{olmo.gd2po.minus.concur-free.humaneval-plus-abs}{0.1}
\setval{olmo.gd2po.minus.concur-free.if}{+0.7}
\setval{olmo.gd2po.minus.concur-free.if-abs}{0.7}
\setval{olmo.gd2po.minus.concur-free.ifbench}{+0.7}
\setval{olmo.gd2po.minus.concur-free.ifbench-abs}{0.7}
\setval{olmo.gd2po.minus.concur-free.ifeval}{+0.4}
\setval{olmo.gd2po.minus.concur-free.ifeval-abs}{0.4}
\setval{olmo.gd2po.minus.concur-free.judge2.avg}{+0.6}
\setval{olmo.gd2po.minus.concur-free.judge2.avg-abs}{0.6}
\setval{olmo.gd2po.minus.concur-free.judge2.creative-writing-v3}{+0.3}
\setval{olmo.gd2po.minus.concur-free.judge2.creative-writing-v3-abs}{0.3}
\setval{olmo.gd2po.minus.concur-free.judge2.longbench-write}{$-$4.0}
\setval{olmo.gd2po.minus.concur-free.judge2.longbench-write-abs}{4.0}
\setval{olmo.gd2po.minus.concur-free.judge2.writing}{$-$1.1}
\setval{olmo.gd2po.minus.concur-free.judge2.writing-abs}{1.1}
\setval{olmo.gd2po.minus.concur-free.judge2.writingbench}{+0.3}
\setval{olmo.gd2po.minus.concur-free.judge2.writingbench-abs}{0.3}
\setval{olmo.gd2po.minus.concur-free.livecodebench}{$-$2.5}
\setval{olmo.gd2po.minus.concur-free.livecodebench-abs}{2.5}
\setval{olmo.gd2po.minus.concur-free.longbench-write}{$-$3.8}
\setval{olmo.gd2po.minus.concur-free.longbench-write-abs}{3.8}
\setval{olmo.gd2po.minus.concur-free.math}{+2.1}
\setval{olmo.gd2po.minus.concur-free.math-abs}{2.1}
\setval{olmo.gd2po.minus.concur-free.math500}{+1.4}
\setval{olmo.gd2po.minus.concur-free.math500-abs}{1.4}
\setval{olmo.gd2po.minus.concur-free.mbpp-plus}{+0.3}
\setval{olmo.gd2po.minus.concur-free.mbpp-plus-abs}{0.3}
\setval{olmo.gd2po.minus.concur-free.minerva}{+0.2}
\setval{olmo.gd2po.minus.concur-free.minerva-abs}{0.2}
\setval{olmo.gd2po.minus.concur-free.olympiad-bench}{+2.8}
\setval{olmo.gd2po.minus.concur-free.olympiad-bench-abs}{2.8}
\setval{olmo.gd2po.minus.concur-free.writing}{$-$3.3}
\setval{olmo.gd2po.minus.concur-free.writing-abs}{3.3}
\setval{olmo.gd2po.minus.concur-free.writingbench}{$-$4.5}
\setval{olmo.gd2po.minus.concur-free.writingbench-abs}{4.5}
\setval{olmo.gd2po.minus.concur-tied.aime24}{$-$6.1}
\setval{olmo.gd2po.minus.concur-tied.aime24-abs}{6.1}
\setval{olmo.gd2po.minus.concur-tied.aime25}{$-$5.6}
\setval{olmo.gd2po.minus.concur-tied.aime25-abs}{5.6}
\setval{olmo.gd2po.minus.concur-tied.amc23}{$-$7.7}
\setval{olmo.gd2po.minus.concur-tied.amc23-abs}{7.7}
\setval{olmo.gd2po.minus.concur-tied.avg}{$-$3.5}
\setval{olmo.gd2po.minus.concur-tied.avg-abs}{3.5}
\setval{olmo.gd2po.minus.concur-tied.code}{$-$6.2}
\setval{olmo.gd2po.minus.concur-tied.code-abs}{6.2}
\setval{olmo.gd2po.minus.concur-tied.creative-writing-v3}{$-$0.7}
\setval{olmo.gd2po.minus.concur-tied.creative-writing-v3-abs}{0.7}
\setval{olmo.gd2po.minus.concur-tied.guidebench}{+2.6}
\setval{olmo.gd2po.minus.concur-tied.guidebench-abs}{2.6}
\setval{olmo.gd2po.minus.concur-tied.humaneval-plus}{$-$6.5}
\setval{olmo.gd2po.minus.concur-tied.humaneval-plus-abs}{6.5}
\setval{olmo.gd2po.minus.concur-tied.if}{+0.4}
\setval{olmo.gd2po.minus.concur-tied.if-abs}{0.4}
\setval{olmo.gd2po.minus.concur-tied.ifbench}{$-$1.3}
\setval{olmo.gd2po.minus.concur-tied.ifbench-abs}{1.3}
\setval{olmo.gd2po.minus.concur-tied.ifeval}{+0.0}
\setval{olmo.gd2po.minus.concur-tied.ifeval-abs}{0.0}
\setval{olmo.gd2po.minus.concur-tied.judge2.avg}{$-$3.1}
\setval{olmo.gd2po.minus.concur-tied.judge2.avg-abs}{3.1}
\setval{olmo.gd2po.minus.concur-tied.judge2.creative-writing-v3}{+0.7}
\setval{olmo.gd2po.minus.concur-tied.judge2.creative-writing-v3-abs}{0.7}
\setval{olmo.gd2po.minus.concur-tied.judge2.longbench-write}{$-$0.6}
\setval{olmo.gd2po.minus.concur-tied.judge2.longbench-write-abs}{0.6}
\setval{olmo.gd2po.minus.concur-tied.judge2.writing}{+1.0}
\setval{olmo.gd2po.minus.concur-tied.judge2.writing-abs}{1.0}
\setval{olmo.gd2po.minus.concur-tied.judge2.writingbench}{+2.8}
\setval{olmo.gd2po.minus.concur-tied.judge2.writingbench-abs}{2.8}
\setval{olmo.gd2po.minus.concur-tied.livecodebench}{$-$5.8}
\setval{olmo.gd2po.minus.concur-tied.livecodebench-abs}{5.8}
\setval{olmo.gd2po.minus.concur-tied.longbench-write}{$-$1.0}
\setval{olmo.gd2po.minus.concur-tied.longbench-write-abs}{1.0}
\setval{olmo.gd2po.minus.concur-tied.math}{$-$5.4}
\setval{olmo.gd2po.minus.concur-tied.math-abs}{5.4}
\setval{olmo.gd2po.minus.concur-tied.math500}{$-$3.7}
\setval{olmo.gd2po.minus.concur-tied.math500-abs}{3.7}
\setval{olmo.gd2po.minus.concur-tied.mbpp-plus}{$-$6.3}
\setval{olmo.gd2po.minus.concur-tied.mbpp-plus-abs}{6.3}
\setval{olmo.gd2po.minus.concur-tied.minerva}{$-$3.9}
\setval{olmo.gd2po.minus.concur-tied.minerva-abs}{3.9}
\setval{olmo.gd2po.minus.concur-tied.olympiad-bench}{$-$5.5}
\setval{olmo.gd2po.minus.concur-tied.olympiad-bench-abs}{5.5}
\setval{olmo.gd2po.minus.concur-tied.writing}{$-$0.7}
\setval{olmo.gd2po.minus.concur-tied.writing-abs}{0.7}
\setval{olmo.gd2po.minus.concur-tied.writingbench}{$-$0.4}
\setval{olmo.gd2po.minus.concur-tied.writingbench-abs}{0.4}
\setval{olmo.gd2po.minus.concur.aime24}{$-$8.0}
\setval{olmo.gd2po.minus.concur.aime24-abs}{8.0}
\setval{olmo.gd2po.minus.concur.aime25}{$-$7.9}
\setval{olmo.gd2po.minus.concur.aime25-abs}{7.9}
\setval{olmo.gd2po.minus.concur.amc23}{$-$11.3}
\setval{olmo.gd2po.minus.concur.amc23-abs}{11.3}
\setval{olmo.gd2po.minus.concur.avg}{$-$3.7}
\setval{olmo.gd2po.minus.concur.avg-abs}{3.7}
\setval{olmo.gd2po.minus.concur.code}{$-$5.3}
\setval{olmo.gd2po.minus.concur.code-abs}{5.3}
\setval{olmo.gd2po.minus.concur.creative-writing-v3}{$-$0.9}
\setval{olmo.gd2po.minus.concur.creative-writing-v3-abs}{0.9}
\setval{olmo.gd2po.minus.concur.guidebench}{+4.2}
\setval{olmo.gd2po.minus.concur.guidebench-abs}{4.2}
\setval{olmo.gd2po.minus.concur.humaneval-plus}{$-$6.0}
\setval{olmo.gd2po.minus.concur.humaneval-plus-abs}{6.0}
\setval{olmo.gd2po.minus.concur.if}{+1.5}
\setval{olmo.gd2po.minus.concur.if-abs}{1.5}
\setval{olmo.gd2po.minus.concur.ifbench}{+0.3}
\setval{olmo.gd2po.minus.concur.ifbench-abs}{0.3}
\setval{olmo.gd2po.minus.concur.ifeval}{$-$0.1}
\setval{olmo.gd2po.minus.concur.ifeval-abs}{0.1}
\setval{olmo.gd2po.minus.concur.judge2.avg}{$-$3.3}
\setval{olmo.gd2po.minus.concur.judge2.avg-abs}{3.3}
\setval{olmo.gd2po.minus.concur.judge2.creative-writing-v3}{$-$0.6}
\setval{olmo.gd2po.minus.concur.judge2.creative-writing-v3-abs}{0.6}
\setval{olmo.gd2po.minus.concur.judge2.longbench-write}{+1.5}
\setval{olmo.gd2po.minus.concur.judge2.longbench-write-abs}{1.5}
\setval{olmo.gd2po.minus.concur.judge2.writing}{+1.5}
\setval{olmo.gd2po.minus.concur.judge2.writing-abs}{1.5}
\setval{olmo.gd2po.minus.concur.judge2.writingbench}{+3.5}
\setval{olmo.gd2po.minus.concur.judge2.writingbench-abs}{3.5}
\setval{olmo.gd2po.minus.concur.livecodebench}{$-$7.1}
\setval{olmo.gd2po.minus.concur.livecodebench-abs}{7.1}
\setval{olmo.gd2po.minus.concur.longbench-write}{$-$0.3}
\setval{olmo.gd2po.minus.concur.longbench-write-abs}{0.3}
\setval{olmo.gd2po.minus.concur.math}{$-$7.2}
\setval{olmo.gd2po.minus.concur.math-abs}{7.2}
\setval{olmo.gd2po.minus.concur.math500}{$-$4.8}
\setval{olmo.gd2po.minus.concur.math500-abs}{4.8}
\setval{olmo.gd2po.minus.concur.mbpp-plus}{$-$2.8}
\setval{olmo.gd2po.minus.concur.mbpp-plus-abs}{2.8}
\setval{olmo.gd2po.minus.concur.minerva}{$-$4.7}
\setval{olmo.gd2po.minus.concur.minerva-abs}{4.7}
\setval{olmo.gd2po.minus.concur.olympiad-bench}{$-$6.3}
\setval{olmo.gd2po.minus.concur.olympiad-bench-abs}{6.3}
\setval{olmo.gd2po.minus.concur.writing}{$-$0.3}
\setval{olmo.gd2po.minus.concur.writing-abs}{0.3}
\setval{olmo.gd2po.minus.concur.writingbench}{+0.3}
\setval{olmo.gd2po.minus.concur.writingbench-abs}{0.3}
\setval{olmo.gd2po.minus.gdpo-noaux.aime24}{$-$10.0}
\setval{olmo.gd2po.minus.gdpo-noaux.aime24-abs}{10.0}
\setval{olmo.gd2po.minus.gdpo-noaux.aime25}{$-$9.4}
\setval{olmo.gd2po.minus.gdpo-noaux.aime25-abs}{9.4}
\setval{olmo.gd2po.minus.gdpo-noaux.amc23}{$-$11.7}
\setval{olmo.gd2po.minus.gdpo-noaux.amc23-abs}{11.7}
\setval{olmo.gd2po.minus.gdpo-noaux.avg}{$-$2.2}
\setval{olmo.gd2po.minus.gdpo-noaux.avg-abs}{2.2}
\setval{olmo.gd2po.minus.gdpo-noaux.code}{$-$7.5}
\setval{olmo.gd2po.minus.gdpo-noaux.code-abs}{7.5}
\setval{olmo.gd2po.minus.gdpo-noaux.creative-writing-v3}{+10.1}
\setval{olmo.gd2po.minus.gdpo-noaux.creative-writing-v3-abs}{10.1}
\setval{olmo.gd2po.minus.gdpo-noaux.guidebench}{+5.4}
\setval{olmo.gd2po.minus.gdpo-noaux.guidebench-abs}{5.4}
\setval{olmo.gd2po.minus.gdpo-noaux.humaneval-plus}{$-$8.4}
\setval{olmo.gd2po.minus.gdpo-noaux.humaneval-plus-abs}{8.4}
\setval{olmo.gd2po.minus.gdpo-noaux.if}{+2.4}
\setval{olmo.gd2po.minus.gdpo-noaux.if-abs}{2.4}
\setval{olmo.gd2po.minus.gdpo-noaux.ifbench}{$-$1.0}
\setval{olmo.gd2po.minus.gdpo-noaux.ifbench-abs}{1.0}
\setval{olmo.gd2po.minus.gdpo-noaux.ifeval}{+2.9}
\setval{olmo.gd2po.minus.gdpo-noaux.ifeval-abs}{2.9}
\setval{olmo.gd2po.minus.gdpo-noaux.judge2.avg}{$-$2.4}
\setval{olmo.gd2po.minus.gdpo-noaux.judge2.avg-abs}{2.4}
\setval{olmo.gd2po.minus.gdpo-noaux.judge2.creative-writing-v3}{+6.9}
\setval{olmo.gd2po.minus.gdpo-noaux.judge2.creative-writing-v3-abs}{6.9}
\setval{olmo.gd2po.minus.gdpo-noaux.judge2.longbench-write}{+13.3}
\setval{olmo.gd2po.minus.gdpo-noaux.judge2.longbench-write-abs}{13.3}
\setval{olmo.gd2po.minus.gdpo-noaux.judge2.writing}{+9.6}
\setval{olmo.gd2po.minus.gdpo-noaux.judge2.writing-abs}{9.6}
\setval{olmo.gd2po.minus.gdpo-noaux.judge2.writingbench}{+8.6}
\setval{olmo.gd2po.minus.gdpo-noaux.judge2.writingbench-abs}{8.6}
\setval{olmo.gd2po.minus.gdpo-noaux.livecodebench}{$-$8.3}
\setval{olmo.gd2po.minus.gdpo-noaux.livecodebench-abs}{8.3}
\setval{olmo.gd2po.minus.gdpo-noaux.longbench-write}{+12.2}
\setval{olmo.gd2po.minus.gdpo-noaux.longbench-write-abs}{12.2}
\setval{olmo.gd2po.minus.gdpo-noaux.math}{$-$8.3}
\setval{olmo.gd2po.minus.gdpo-noaux.math-abs}{8.3}
\setval{olmo.gd2po.minus.gdpo-noaux.math500}{$-$5.2}
\setval{olmo.gd2po.minus.gdpo-noaux.math500-abs}{5.2}
\setval{olmo.gd2po.minus.gdpo-noaux.mbpp-plus}{$-$5.7}
\setval{olmo.gd2po.minus.gdpo-noaux.mbpp-plus-abs}{5.7}
\setval{olmo.gd2po.minus.gdpo-noaux.minerva}{$-$6.0}
\setval{olmo.gd2po.minus.gdpo-noaux.minerva-abs}{6.0}
\setval{olmo.gd2po.minus.gdpo-noaux.olympiad-bench}{$-$7.7}
\setval{olmo.gd2po.minus.gdpo-noaux.olympiad-bench-abs}{7.7}
\setval{olmo.gd2po.minus.gdpo-noaux.writing}{+10.9}
\setval{olmo.gd2po.minus.gdpo-noaux.writing-abs}{10.9}
\setval{olmo.gd2po.minus.gdpo-noaux.writingbench}{+10.3}
\setval{olmo.gd2po.minus.gdpo-noaux.writingbench-abs}{10.3}
\setval{olmo.gd2po.minus.gdpo-w0.5.aime24}{$-$1.2}
\setval{olmo.gd2po.minus.gdpo-w0.5.aime24-abs}{1.2}
\setval{olmo.gd2po.minus.gdpo-w0.5.aime25}{$-$3.0}
\setval{olmo.gd2po.minus.gdpo-w0.5.aime25-abs}{3.0}
\setval{olmo.gd2po.minus.gdpo-w0.5.amc23}{$-$1.8}
\setval{olmo.gd2po.minus.gdpo-w0.5.amc23-abs}{1.8}
\setval{olmo.gd2po.minus.gdpo-w0.5.avg}{+0.0}
\setval{olmo.gd2po.minus.gdpo-w0.5.avg-abs}{0.0}
\setval{olmo.gd2po.minus.gdpo-w0.5.code}{$-$0.9}
\setval{olmo.gd2po.minus.gdpo-w0.5.code-abs}{0.9}
\setval{olmo.gd2po.minus.gdpo-w0.5.creative-writing-v3}{+0.2}
\setval{olmo.gd2po.minus.gdpo-w0.5.creative-writing-v3-abs}{0.2}
\setval{olmo.gd2po.minus.gdpo-w0.5.guidebench}{+9.2}
\setval{olmo.gd2po.minus.gdpo-w0.5.guidebench-abs}{9.2}
\setval{olmo.gd2po.minus.gdpo-w0.5.humaneval-plus}{$-$1.0}
\setval{olmo.gd2po.minus.gdpo-w0.5.humaneval-plus-abs}{1.0}
\setval{olmo.gd2po.minus.gdpo-w0.5.if}{+3.0}
\setval{olmo.gd2po.minus.gdpo-w0.5.if-abs}{3.0}
\setval{olmo.gd2po.minus.gdpo-w0.5.ifbench}{$-$0.3}
\setval{olmo.gd2po.minus.gdpo-w0.5.ifbench-abs}{0.3}
\setval{olmo.gd2po.minus.gdpo-w0.5.ifeval}{+0.1}
\setval{olmo.gd2po.minus.gdpo-w0.5.ifeval-abs}{0.1}
\setval{olmo.gd2po.minus.gdpo-w0.5.judge2.avg}{+0.0}
\setval{olmo.gd2po.minus.gdpo-w0.5.judge2.avg-abs}{0.0}
\setval{olmo.gd2po.minus.gdpo-w0.5.judge2.creative-writing-v3}{$-$1.5}
\setval{olmo.gd2po.minus.gdpo-w0.5.judge2.creative-writing-v3-abs}{1.5}
\setval{olmo.gd2po.minus.gdpo-w0.5.judge2.longbench-write}{+3.3}
\setval{olmo.gd2po.minus.gdpo-w0.5.judge2.longbench-write-abs}{3.3}
\setval{olmo.gd2po.minus.gdpo-w0.5.judge2.writing}{+1.1}
\setval{olmo.gd2po.minus.gdpo-w0.5.judge2.writing-abs}{1.1}
\setval{olmo.gd2po.minus.gdpo-w0.5.judge2.writingbench}{+1.7}
\setval{olmo.gd2po.minus.gdpo-w0.5.judge2.writingbench-abs}{1.7}
\setval{olmo.gd2po.minus.gdpo-w0.5.livecodebench}{$-$1.6}
\setval{olmo.gd2po.minus.gdpo-w0.5.livecodebench-abs}{1.6}
\setval{olmo.gd2po.minus.gdpo-w0.5.longbench-write}{+4.5}
\setval{olmo.gd2po.minus.gdpo-w0.5.longbench-write-abs}{4.5}
\setval{olmo.gd2po.minus.gdpo-w0.5.math}{$-$1.7}
\setval{olmo.gd2po.minus.gdpo-w0.5.math-abs}{1.7}
\setval{olmo.gd2po.minus.gdpo-w0.5.math500}{$-$1.3}
\setval{olmo.gd2po.minus.gdpo-w0.5.math500-abs}{1.3}
\setval{olmo.gd2po.minus.gdpo-w0.5.mbpp-plus}{$-$0.1}
\setval{olmo.gd2po.minus.gdpo-w0.5.mbpp-plus-abs}{0.1}
\setval{olmo.gd2po.minus.gdpo-w0.5.minerva}{$-$1.4}
\setval{olmo.gd2po.minus.gdpo-w0.5.minerva-abs}{1.4}
\setval{olmo.gd2po.minus.gdpo-w0.5.olympiad-bench}{$-$1.4}
\setval{olmo.gd2po.minus.gdpo-w0.5.olympiad-bench-abs}{1.4}
\setval{olmo.gd2po.minus.gdpo-w0.5.writing}{+1.3}
\setval{olmo.gd2po.minus.gdpo-w0.5.writing-abs}{1.3}
\setval{olmo.gd2po.minus.gdpo-w0.5.writingbench}{$-$0.8}
\setval{olmo.gd2po.minus.gdpo-w0.5.writingbench-abs}{0.8}
\setval{olmo.gd2po.minus.gdpo.aime24}{+0.5}
\setval{olmo.gd2po.minus.gdpo.aime24-abs}{0.5}
\setval{olmo.gd2po.minus.gdpo.aime25}{+1.7}
\setval{olmo.gd2po.minus.gdpo.aime25-abs}{1.7}
\setval{olmo.gd2po.minus.gdpo.amc23}{+2.7}
\setval{olmo.gd2po.minus.gdpo.amc23-abs}{2.7}
\setval{olmo.gd2po.minus.gdpo.avg}{+1.8}
\setval{olmo.gd2po.minus.gdpo.avg-abs}{1.8}
\setval{olmo.gd2po.minus.gdpo.code}{+0.2}
\setval{olmo.gd2po.minus.gdpo.code-abs}{0.2}
\setval{olmo.gd2po.minus.gdpo.creative-writing-v3}{+1.2}
\setval{olmo.gd2po.minus.gdpo.creative-writing-v3-abs}{1.2}
\setval{olmo.gd2po.minus.gdpo.guidebench}{+10.3}
\setval{olmo.gd2po.minus.gdpo.guidebench-abs}{10.3}
\setval{olmo.gd2po.minus.gdpo.humaneval-plus}{+0.4}
\setval{olmo.gd2po.minus.gdpo.humaneval-plus-abs}{0.4}
\setval{olmo.gd2po.minus.gdpo.if}{+4.7}
\setval{olmo.gd2po.minus.gdpo.if-abs}{4.7}
\setval{olmo.gd2po.minus.gdpo.ifbench}{+2.7}
\setval{olmo.gd2po.minus.gdpo.ifbench-abs}{2.7}
\setval{olmo.gd2po.minus.gdpo.ifeval}{+1.2}
\setval{olmo.gd2po.minus.gdpo.ifeval-abs}{1.2}
\setval{olmo.gd2po.minus.gdpo.judge2.avg}{+1.7}
\setval{olmo.gd2po.minus.gdpo.judge2.avg-abs}{1.7}
\setval{olmo.gd2po.minus.gdpo.judge2.creative-writing-v3}{+0.1}
\setval{olmo.gd2po.minus.gdpo.judge2.creative-writing-v3-abs}{0.1}
\setval{olmo.gd2po.minus.gdpo.judge2.longbench-write}{+4.3}
\setval{olmo.gd2po.minus.gdpo.judge2.longbench-write-abs}{4.3}
\setval{olmo.gd2po.minus.gdpo.judge2.writing}{+1.9}
\setval{olmo.gd2po.minus.gdpo.judge2.writing-abs}{1.9}
\setval{olmo.gd2po.minus.gdpo.judge2.writingbench}{+1.3}
\setval{olmo.gd2po.minus.gdpo.judge2.writingbench-abs}{1.3}
\setval{olmo.gd2po.minus.gdpo.livecodebench}{+0.1}
\setval{olmo.gd2po.minus.gdpo.livecodebench-abs}{0.1}
\setval{olmo.gd2po.minus.gdpo.longbench-write}{+4.9}
\setval{olmo.gd2po.minus.gdpo.longbench-write-abs}{4.9}
\setval{olmo.gd2po.minus.gdpo.math}{+0.9}
\setval{olmo.gd2po.minus.gdpo.math-abs}{0.9}
\setval{olmo.gd2po.minus.gdpo.math500}{+0.5}
\setval{olmo.gd2po.minus.gdpo.math500-abs}{0.5}
\setval{olmo.gd2po.minus.gdpo.mbpp-plus}{+0.0}
\setval{olmo.gd2po.minus.gdpo.mbpp-plus-abs}{0.0}
\setval{olmo.gd2po.minus.gdpo.minerva}{$-$0.2}
\setval{olmo.gd2po.minus.gdpo.minerva-abs}{0.2}
\setval{olmo.gd2po.minus.gdpo.olympiad-bench}{+0.2}
\setval{olmo.gd2po.minus.gdpo.olympiad-bench-abs}{0.2}
\setval{olmo.gd2po.minus.gdpo.writing}{+2.2}
\setval{olmo.gd2po.minus.gdpo.writing-abs}{2.2}
\setval{olmo.gd2po.minus.gdpo.writingbench}{+0.4}
\setval{olmo.gd2po.minus.gdpo.writingbench-abs}{0.4}
\setval{olmo.gd2po.minus.geoma.aime24}{+4.6}
\setval{olmo.gd2po.minus.geoma.aime24-abs}{4.6}
\setval{olmo.gd2po.minus.geoma.aime25}{+2.3}
\setval{olmo.gd2po.minus.geoma.aime25-abs}{2.3}
\setval{olmo.gd2po.minus.geoma.amc23}{+8.3}
\setval{olmo.gd2po.minus.geoma.amc23-abs}{8.3}
\setval{olmo.gd2po.minus.geoma.avg}{+2.7}
\setval{olmo.gd2po.minus.geoma.avg-abs}{2.7}
\setval{olmo.gd2po.minus.geoma.code}{+1.0}
\setval{olmo.gd2po.minus.geoma.code-abs}{1.0}
\setval{olmo.gd2po.minus.geoma.creative-writing-v3}{$-$1.3}
\setval{olmo.gd2po.minus.geoma.creative-writing-v3-abs}{1.3}
\setval{olmo.gd2po.minus.geoma.guidebench}{+14.5}
\setval{olmo.gd2po.minus.geoma.guidebench-abs}{14.5}
\setval{olmo.gd2po.minus.geoma.humaneval-plus}{+2.1}
\setval{olmo.gd2po.minus.geoma.humaneval-plus-abs}{2.1}
\setval{olmo.gd2po.minus.geoma.if}{+5.4}
\setval{olmo.gd2po.minus.geoma.if-abs}{5.4}
\setval{olmo.gd2po.minus.geoma.ifbench}{+2.0}
\setval{olmo.gd2po.minus.geoma.ifbench-abs}{2.0}
\setval{olmo.gd2po.minus.geoma.ifeval}{$-$0.4}
\setval{olmo.gd2po.minus.geoma.ifeval-abs}{0.4}
\setval{olmo.gd2po.minus.geoma.judge2.avg}{+2.5}
\setval{olmo.gd2po.minus.geoma.judge2.avg-abs}{2.5}
\setval{olmo.gd2po.minus.geoma.judge2.creative-writing-v3}{$-$3.9}
\setval{olmo.gd2po.minus.geoma.judge2.creative-writing-v3-abs}{3.9}
\setval{olmo.gd2po.minus.geoma.judge2.longbench-write}{+1.7}
\setval{olmo.gd2po.minus.geoma.judge2.longbench-write-abs}{1.7}
\setval{olmo.gd2po.minus.geoma.judge2.writing}{$-$0.8}
\setval{olmo.gd2po.minus.geoma.judge2.writing-abs}{0.8}
\setval{olmo.gd2po.minus.geoma.judge2.writingbench}{$-$0.3}
\setval{olmo.gd2po.minus.geoma.judge2.writingbench-abs}{0.3}
\setval{olmo.gd2po.minus.geoma.livecodebench}{+0.9}
\setval{olmo.gd2po.minus.geoma.livecodebench-abs}{0.9}
\setval{olmo.gd2po.minus.geoma.longbench-write}{+2.9}
\setval{olmo.gd2po.minus.geoma.longbench-write-abs}{2.9}
\setval{olmo.gd2po.minus.geoma.math}{+3.5}
\setval{olmo.gd2po.minus.geoma.math-abs}{3.5}
\setval{olmo.gd2po.minus.geoma.math500}{+2.5}
\setval{olmo.gd2po.minus.geoma.math500-abs}{2.5}
\setval{olmo.gd2po.minus.geoma.mbpp-plus}{+0.1}
\setval{olmo.gd2po.minus.geoma.mbpp-plus-abs}{0.1}
\setval{olmo.gd2po.minus.geoma.minerva}{+0.1}
\setval{olmo.gd2po.minus.geoma.minerva-abs}{0.1}
\setval{olmo.gd2po.minus.geoma.olympiad-bench}{+3.3}
\setval{olmo.gd2po.minus.geoma.olympiad-bench-abs}{3.3}
\setval{olmo.gd2po.minus.geoma.writing}{+0.1}
\setval{olmo.gd2po.minus.geoma.writing-abs}{0.1}
\setval{olmo.gd2po.minus.geoma.writingbench}{$-$1.2}
\setval{olmo.gd2po.minus.geoma.writingbench-abs}{1.2}
\setval{olmo.gd2po.minus.grpo-nodiv.aime24}{$-$3.4}
\setval{olmo.gd2po.minus.grpo-nodiv.aime24-abs}{3.4}
\setval{olmo.gd2po.minus.grpo-nodiv.aime25}{$-$3.3}
\setval{olmo.gd2po.minus.grpo-nodiv.aime25-abs}{3.3}
\setval{olmo.gd2po.minus.grpo-nodiv.amc23}{$-$0.9}
\setval{olmo.gd2po.minus.grpo-nodiv.amc23-abs}{0.9}
\setval{olmo.gd2po.minus.grpo-nodiv.avg}{$-$0.3}
\setval{olmo.gd2po.minus.grpo-nodiv.avg-abs}{0.3}
\setval{olmo.gd2po.minus.grpo-nodiv.code}{$-$0.5}
\setval{olmo.gd2po.minus.grpo-nodiv.code-abs}{0.5}
\setval{olmo.gd2po.minus.grpo-nodiv.creative-writing-v3}{+1.3}
\setval{olmo.gd2po.minus.grpo-nodiv.creative-writing-v3-abs}{1.3}
\setval{olmo.gd2po.minus.grpo-nodiv.guidebench}{+5.7}
\setval{olmo.gd2po.minus.grpo-nodiv.guidebench-abs}{5.7}
\setval{olmo.gd2po.minus.grpo-nodiv.humaneval-plus}{$-$0.6}
\setval{olmo.gd2po.minus.grpo-nodiv.humaneval-plus-abs}{0.6}
\setval{olmo.gd2po.minus.grpo-nodiv.if}{+2.2}
\setval{olmo.gd2po.minus.grpo-nodiv.if-abs}{2.2}
\setval{olmo.gd2po.minus.grpo-nodiv.ifbench}{+1.0}
\setval{olmo.gd2po.minus.grpo-nodiv.ifbench-abs}{1.0}
\setval{olmo.gd2po.minus.grpo-nodiv.ifeval}{$-$0.1}
\setval{olmo.gd2po.minus.grpo-nodiv.ifeval-abs}{0.1}
\setval{olmo.gd2po.minus.grpo-nodiv.judge2.avg}{$-$0.6}
\setval{olmo.gd2po.minus.grpo-nodiv.judge2.avg-abs}{0.6}
\setval{olmo.gd2po.minus.grpo-nodiv.judge2.creative-writing-v3}{$-$0.9}
\setval{olmo.gd2po.minus.grpo-nodiv.judge2.creative-writing-v3-abs}{0.9}
\setval{olmo.gd2po.minus.grpo-nodiv.judge2.longbench-write}{$-$1.2}
\setval{olmo.gd2po.minus.grpo-nodiv.judge2.longbench-write-abs}{1.2}
\setval{olmo.gd2po.minus.grpo-nodiv.judge2.writing}{$-$0.3}
\setval{olmo.gd2po.minus.grpo-nodiv.judge2.writing-abs}{0.3}
\setval{olmo.gd2po.minus.grpo-nodiv.judge2.writingbench}{+1.4}
\setval{olmo.gd2po.minus.grpo-nodiv.judge2.writingbench-abs}{1.4}
\setval{olmo.gd2po.minus.grpo-nodiv.livecodebench}{$-$0.4}
\setval{olmo.gd2po.minus.grpo-nodiv.livecodebench-abs}{0.4}
\setval{olmo.gd2po.minus.grpo-nodiv.longbench-write}{+0.7}
\setval{olmo.gd2po.minus.grpo-nodiv.longbench-write-abs}{0.7}
\setval{olmo.gd2po.minus.grpo-nodiv.math}{$-$2.2}
\setval{olmo.gd2po.minus.grpo-nodiv.math-abs}{2.2}
\setval{olmo.gd2po.minus.grpo-nodiv.math500}{$-$1.8}
\setval{olmo.gd2po.minus.grpo-nodiv.math500-abs}{1.8}
\setval{olmo.gd2po.minus.grpo-nodiv.mbpp-plus}{$-$0.6}
\setval{olmo.gd2po.minus.grpo-nodiv.mbpp-plus-abs}{0.6}
\setval{olmo.gd2po.minus.grpo-nodiv.minerva}{$-$1.7}
\setval{olmo.gd2po.minus.grpo-nodiv.minerva-abs}{1.7}
\setval{olmo.gd2po.minus.grpo-nodiv.olympiad-bench}{$-$2.4}
\setval{olmo.gd2po.minus.grpo-nodiv.olympiad-bench-abs}{2.4}
\setval{olmo.gd2po.minus.grpo-nodiv.writing}{+1.5}
\setval{olmo.gd2po.minus.grpo-nodiv.writing-abs}{1.5}
\setval{olmo.gd2po.minus.grpo-nodiv.writingbench}{+2.6}
\setval{olmo.gd2po.minus.grpo-nodiv.writingbench-abs}{2.6}
\setval{olmo.gd2po.minus.grpo.aime24}{$-$1.1}
\setval{olmo.gd2po.minus.grpo.aime24-abs}{1.1}
\setval{olmo.gd2po.minus.grpo.aime25}{$-$0.4}
\setval{olmo.gd2po.minus.grpo.aime25-abs}{0.4}
\setval{olmo.gd2po.minus.grpo.amc23}{+1.3}
\setval{olmo.gd2po.minus.grpo.amc23-abs}{1.3}
\setval{olmo.gd2po.minus.grpo.avg}{+0.4}
\setval{olmo.gd2po.minus.grpo.avg-abs}{0.4}
\setval{olmo.gd2po.minus.grpo.code}{$-$0.2}
\setval{olmo.gd2po.minus.grpo.code-abs}{0.2}
\setval{olmo.gd2po.minus.grpo.creative-writing-v3}{$-$1.1}
\setval{olmo.gd2po.minus.grpo.creative-writing-v3-abs}{1.1}
\setval{olmo.gd2po.minus.grpo.guidebench}{+4.7}
\setval{olmo.gd2po.minus.grpo.guidebench-abs}{4.7}
\setval{olmo.gd2po.minus.grpo.humaneval-plus}{$-$0.5}
\setval{olmo.gd2po.minus.grpo.humaneval-plus-abs}{0.5}
\setval{olmo.gd2po.minus.grpo.if}{+1.8}
\setval{olmo.gd2po.minus.grpo.if-abs}{1.8}
\setval{olmo.gd2po.minus.grpo.ifbench}{+0.3}
\setval{olmo.gd2po.minus.grpo.ifbench-abs}{0.3}
\setval{olmo.gd2po.minus.grpo.ifeval}{+0.2}
\setval{olmo.gd2po.minus.grpo.ifeval-abs}{0.2}
\setval{olmo.gd2po.minus.grpo.judge2.avg}{+0.5}
\setval{olmo.gd2po.minus.grpo.judge2.avg-abs}{0.5}
\setval{olmo.gd2po.minus.grpo.judge2.creative-writing-v3}{$-$0.7}
\setval{olmo.gd2po.minus.grpo.judge2.creative-writing-v3-abs}{0.7}
\setval{olmo.gd2po.minus.grpo.judge2.longbench-write}{+2.8}
\setval{olmo.gd2po.minus.grpo.judge2.longbench-write-abs}{2.8}
\setval{olmo.gd2po.minus.grpo.judge2.writing}{+0.7}
\setval{olmo.gd2po.minus.grpo.judge2.writing-abs}{0.7}
\setval{olmo.gd2po.minus.grpo.judge2.writingbench}{$-$0.1}
\setval{olmo.gd2po.minus.grpo.judge2.writingbench-abs}{0.1}
\setval{olmo.gd2po.minus.grpo.livecodebench}{+0.9}
\setval{olmo.gd2po.minus.grpo.livecodebench-abs}{0.9}
\setval{olmo.gd2po.minus.grpo.longbench-write}{+4.0}
\setval{olmo.gd2po.minus.grpo.longbench-write-abs}{4.0}
\setval{olmo.gd2po.minus.grpo.math}{+0.2}
\setval{olmo.gd2po.minus.grpo.math-abs}{0.2}
\setval{olmo.gd2po.minus.grpo.math500}{+0.8}
\setval{olmo.gd2po.minus.grpo.math500-abs}{0.8}
\setval{olmo.gd2po.minus.grpo.mbpp-plus}{$-$1.0}
\setval{olmo.gd2po.minus.grpo.mbpp-plus-abs}{1.0}
\setval{olmo.gd2po.minus.grpo.minerva}{+0.2}
\setval{olmo.gd2po.minus.grpo.minerva-abs}{0.2}
\setval{olmo.gd2po.minus.grpo.olympiad-bench}{+0.4}
\setval{olmo.gd2po.minus.grpo.olympiad-bench-abs}{0.4}
\setval{olmo.gd2po.minus.grpo.writing}{+0.1}
\setval{olmo.gd2po.minus.grpo.writing-abs}{0.1}
\setval{olmo.gd2po.minus.grpo.writingbench}{$-$2.7}
\setval{olmo.gd2po.minus.grpo.writingbench-abs}{2.7}
\setval{olmo.gd2po.minus.mt-grpo.aime24}{$-$8.2}
\setval{olmo.gd2po.minus.mt-grpo.aime24-abs}{8.2}
\setval{olmo.gd2po.minus.mt-grpo.aime25}{$-$3.9}
\setval{olmo.gd2po.minus.mt-grpo.aime25-abs}{3.9}
\setval{olmo.gd2po.minus.mt-grpo.amc23}{$-$5.2}
\setval{olmo.gd2po.minus.mt-grpo.amc23-abs}{5.2}
\setval{olmo.gd2po.minus.mt-grpo.avg}{$-$1.1}
\setval{olmo.gd2po.minus.mt-grpo.avg-abs}{1.1}
\setval{olmo.gd2po.minus.mt-grpo.code}{+0.5}
\setval{olmo.gd2po.minus.mt-grpo.code-abs}{0.5}
\setval{olmo.gd2po.minus.mt-grpo.creative-writing-v3}{$-$2.4}
\setval{olmo.gd2po.minus.mt-grpo.creative-writing-v3-abs}{2.4}
\setval{olmo.gd2po.minus.mt-grpo.guidebench}{+6.6}
\setval{olmo.gd2po.minus.mt-grpo.guidebench-abs}{6.6}
\setval{olmo.gd2po.minus.mt-grpo.humaneval-plus}{+1.0}
\setval{olmo.gd2po.minus.mt-grpo.humaneval-plus-abs}{1.0}
\setval{olmo.gd2po.minus.mt-grpo.if}{+2.5}
\setval{olmo.gd2po.minus.mt-grpo.if-abs}{2.5}
\setval{olmo.gd2po.minus.mt-grpo.ifbench}{+0.3}
\setval{olmo.gd2po.minus.mt-grpo.ifbench-abs}{0.3}
\setval{olmo.gd2po.minus.mt-grpo.ifeval}{+0.6}
\setval{olmo.gd2po.minus.mt-grpo.ifeval-abs}{0.6}
\setval{olmo.gd2po.minus.mt-grpo.judge2.avg}{$-$1.0}
\setval{olmo.gd2po.minus.mt-grpo.judge2.avg-abs}{1.0}
\setval{olmo.gd2po.minus.mt-grpo.judge2.creative-writing-v3}{$-$2.0}
\setval{olmo.gd2po.minus.mt-grpo.judge2.creative-writing-v3-abs}{2.0}
\setval{olmo.gd2po.minus.mt-grpo.judge2.longbench-write}{$-$0.5}
\setval{olmo.gd2po.minus.mt-grpo.judge2.longbench-write-abs}{0.5}
\setval{olmo.gd2po.minus.mt-grpo.judge2.writing}{$-$0.6}
\setval{olmo.gd2po.minus.mt-grpo.judge2.writing-abs}{0.6}
\setval{olmo.gd2po.minus.mt-grpo.judge2.writingbench}{+0.8}
\setval{olmo.gd2po.minus.mt-grpo.judge2.writingbench-abs}{0.8}
\setval{olmo.gd2po.minus.mt-grpo.livecodebench}{+0.6}
\setval{olmo.gd2po.minus.mt-grpo.livecodebench-abs}{0.6}
\setval{olmo.gd2po.minus.mt-grpo.longbench-write}{+1.8}
\setval{olmo.gd2po.minus.mt-grpo.longbench-write-abs}{1.8}
\setval{olmo.gd2po.minus.mt-grpo.math}{$-$3.7}
\setval{olmo.gd2po.minus.mt-grpo.math-abs}{3.7}
\setval{olmo.gd2po.minus.mt-grpo.math500}{$-$1.2}
\setval{olmo.gd2po.minus.mt-grpo.math500-abs}{1.2}
\setval{olmo.gd2po.minus.mt-grpo.mbpp-plus}{$-$0.2}
\setval{olmo.gd2po.minus.mt-grpo.mbpp-plus-abs}{0.2}
\setval{olmo.gd2po.minus.mt-grpo.minerva}{$-$1.8}
\setval{olmo.gd2po.minus.mt-grpo.minerva-abs}{1.8}
\setval{olmo.gd2po.minus.mt-grpo.olympiad-bench}{$-$2.0}
\setval{olmo.gd2po.minus.mt-grpo.olympiad-bench-abs}{2.0}
\setval{olmo.gd2po.minus.mt-grpo.writing}{$-$0.8}
\setval{olmo.gd2po.minus.mt-grpo.writing-abs}{0.8}
\setval{olmo.gd2po.minus.mt-grpo.writingbench}{$-$2.0}
\setval{olmo.gd2po.minus.mt-grpo.writingbench-abs}{2.0}
\setval{olmo.gd2po.minus.papo.aime24}{+6.1}
\setval{olmo.gd2po.minus.papo.aime24-abs}{6.1}
\setval{olmo.gd2po.minus.papo.aime25}{+4.0}
\setval{olmo.gd2po.minus.papo.aime25-abs}{4.0}
\setval{olmo.gd2po.minus.papo.amc23}{+12.8}
\setval{olmo.gd2po.minus.papo.amc23-abs}{12.8}
\setval{olmo.gd2po.minus.papo.avg}{+3.3}
\setval{olmo.gd2po.minus.papo.avg-abs}{3.3}
\setval{olmo.gd2po.minus.papo.code}{+1.0}
\setval{olmo.gd2po.minus.papo.code-abs}{1.0}
\setval{olmo.gd2po.minus.papo.creative-writing-v3}{+1.3}
\setval{olmo.gd2po.minus.papo.creative-writing-v3-abs}{1.3}
\setval{olmo.gd2po.minus.papo.guidebench}{+2.5}
\setval{olmo.gd2po.minus.papo.guidebench-abs}{2.5}
\setval{olmo.gd2po.minus.papo.humaneval-plus}{+2.0}
\setval{olmo.gd2po.minus.papo.humaneval-plus-abs}{2.0}
\setval{olmo.gd2po.minus.papo.if}{+1.9}
\setval{olmo.gd2po.minus.papo.if-abs}{1.9}
\setval{olmo.gd2po.minus.papo.ifbench}{+2.7}
\setval{olmo.gd2po.minus.papo.ifbench-abs}{2.7}
\setval{olmo.gd2po.minus.papo.ifeval}{+0.5}
\setval{olmo.gd2po.minus.papo.ifeval-abs}{0.5}
\setval{olmo.gd2po.minus.papo.judge2.avg}{+3.3}
\setval{olmo.gd2po.minus.papo.judge2.avg-abs}{3.3}
\setval{olmo.gd2po.minus.papo.judge2.creative-writing-v3}{$-$0.1}
\setval{olmo.gd2po.minus.papo.judge2.creative-writing-v3-abs}{0.1}
\setval{olmo.gd2po.minus.papo.judge2.longbench-write}{+4.3}
\setval{olmo.gd2po.minus.papo.judge2.longbench-write-abs}{4.3}
\setval{olmo.gd2po.minus.papo.judge2.writing}{+1.7}
\setval{olmo.gd2po.minus.papo.judge2.writing-abs}{1.7}
\setval{olmo.gd2po.minus.papo.judge2.writingbench}{+0.9}
\setval{olmo.gd2po.minus.papo.judge2.writingbench-abs}{0.9}
\setval{olmo.gd2po.minus.papo.livecodebench}{+1.3}
\setval{olmo.gd2po.minus.papo.livecodebench-abs}{1.3}
\setval{olmo.gd2po.minus.papo.longbench-write}{+3.5}
\setval{olmo.gd2po.minus.papo.longbench-write-abs}{3.5}
\setval{olmo.gd2po.minus.papo.math}{+5.9}
\setval{olmo.gd2po.minus.papo.math-abs}{5.9}
\setval{olmo.gd2po.minus.papo.math500}{+4.8}
\setval{olmo.gd2po.minus.papo.math500-abs}{4.8}
\setval{olmo.gd2po.minus.papo.mbpp-plus}{$-$0.3}
\setval{olmo.gd2po.minus.papo.mbpp-plus-abs}{0.3}
\setval{olmo.gd2po.minus.papo.minerva}{+1.6}
\setval{olmo.gd2po.minus.papo.minerva-abs}{1.6}
\setval{olmo.gd2po.minus.papo.olympiad-bench}{+6.3}
\setval{olmo.gd2po.minus.papo.olympiad-bench-abs}{6.3}
\setval{olmo.gd2po.minus.papo.writing}{+1.7}
\setval{olmo.gd2po.minus.papo.writing-abs}{1.7}
\setval{olmo.gd2po.minus.papo.writingbench}{+0.4}
\setval{olmo.gd2po.minus.papo.writingbench-abs}{0.4}
\setval{olmo.gd2po.minus.saw.aime24}{+2.2}
\setval{olmo.gd2po.minus.saw.aime24-abs}{2.2}
\setval{olmo.gd2po.minus.saw.aime25}{$-$0.9}
\setval{olmo.gd2po.minus.saw.aime25-abs}{0.9}
\setval{olmo.gd2po.minus.saw.amc23}{+5.3}
\setval{olmo.gd2po.minus.saw.amc23-abs}{5.3}
\setval{olmo.gd2po.minus.saw.avg}{+1.2}
\setval{olmo.gd2po.minus.saw.avg-abs}{1.2}
\setval{olmo.gd2po.minus.saw.code}{+0.4}
\setval{olmo.gd2po.minus.saw.code-abs}{0.4}
\setval{olmo.gd2po.minus.saw.creative-writing-v3}{+0.4}
\setval{olmo.gd2po.minus.saw.creative-writing-v3-abs}{0.4}
\setval{olmo.gd2po.minus.saw.guidebench}{+6.4}
\setval{olmo.gd2po.minus.saw.guidebench-abs}{6.4}
\setval{olmo.gd2po.minus.saw.humaneval-plus}{+0.4}
\setval{olmo.gd2po.minus.saw.humaneval-plus-abs}{0.4}
\setval{olmo.gd2po.minus.saw.if}{+1.7}
\setval{olmo.gd2po.minus.saw.if-abs}{1.7}
\setval{olmo.gd2po.minus.saw.ifbench}{$-$0.3}
\setval{olmo.gd2po.minus.saw.ifbench-abs}{0.3}
\setval{olmo.gd2po.minus.saw.ifeval}{$-$1.0}
\setval{olmo.gd2po.minus.saw.ifeval-abs}{1.0}
\setval{olmo.gd2po.minus.saw.judge2.avg}{+1.0}
\setval{olmo.gd2po.minus.saw.judge2.avg-abs}{1.0}
\setval{olmo.gd2po.minus.saw.judge2.creative-writing-v3}{$-$3.1}
\setval{olmo.gd2po.minus.saw.judge2.creative-writing-v3-abs}{3.1}
\setval{olmo.gd2po.minus.saw.judge2.longbench-write}{+2.5}
\setval{olmo.gd2po.minus.saw.judge2.longbench-write-abs}{2.5}
\setval{olmo.gd2po.minus.saw.judge2.writing}{$-$0.2}
\setval{olmo.gd2po.minus.saw.judge2.writing-abs}{0.2}
\setval{olmo.gd2po.minus.saw.judge2.writingbench}{$-$0.1}
\setval{olmo.gd2po.minus.saw.judge2.writingbench-abs}{0.1}
\setval{olmo.gd2po.minus.saw.livecodebench}{+1.3}
\setval{olmo.gd2po.minus.saw.livecodebench-abs}{1.3}
\setval{olmo.gd2po.minus.saw.longbench-write}{+3.3}
\setval{olmo.gd2po.minus.saw.longbench-write-abs}{3.3}
\setval{olmo.gd2po.minus.saw.math}{+1.6}
\setval{olmo.gd2po.minus.saw.math-abs}{1.6}
\setval{olmo.gd2po.minus.saw.math500}{+1.2}
\setval{olmo.gd2po.minus.saw.math500-abs}{1.2}
\setval{olmo.gd2po.minus.saw.mbpp-plus}{$-$0.6}
\setval{olmo.gd2po.minus.saw.mbpp-plus-abs}{0.6}
\setval{olmo.gd2po.minus.saw.minerva}{$-$0.2}
\setval{olmo.gd2po.minus.saw.minerva-abs}{0.2}
\setval{olmo.gd2po.minus.saw.olympiad-bench}{+2.3}
\setval{olmo.gd2po.minus.saw.olympiad-bench-abs}{2.3}
\setval{olmo.gd2po.minus.saw.writing}{+0.7}
\setval{olmo.gd2po.minus.saw.writing-abs}{0.7}
\setval{olmo.gd2po.minus.saw.writingbench}{$-$1.6}
\setval{olmo.gd2po.minus.saw.writingbench-abs}{1.6}
\setval{olmo.gd2po.passing-at-ceiling.code.early}{0.84}
\setval{olmo.gd2po.passing-at-ceiling.code.late}{0.85}
\setval{olmo.gd2po.passing-at-ceiling.code.run}{0.84}
\setval{olmo.gd2po.passing-at-ceiling.if.early}{0.48}
\setval{olmo.gd2po.passing-at-ceiling.if.late}{0.56}
\setval{olmo.gd2po.passing-at-ceiling.if.run}{0.52}
\setval{olmo.gd2po.passing-at-ceiling.math.early}{1.00}
\setval{olmo.gd2po.passing-at-ceiling.math.late}{1.00}
\setval{olmo.gd2po.passing-at-ceiling.math.run}{1.00}
\setval{olmo.gd2po.passing-at-ceiling.writing.early}{0.00}
\setval{olmo.gd2po.passing-at-ceiling.writing.late}{0.00}
\setval{olmo.gd2po.passing-at-ceiling.writing.run}{0.00}
\setval{olmo.gd2po.primary-cosine.code.early}{0.91}
\setval{olmo.gd2po.primary-cosine.code.late}{0.86}
\setval{olmo.gd2po.primary-cosine.code.run}{0.89}
\setval{olmo.gd2po.primary-cosine.if.early}{0.63}
\setval{olmo.gd2po.primary-cosine.if.late}{0.71}
\setval{olmo.gd2po.primary-cosine.if.run}{0.68}
\setval{olmo.gd2po.primary-cosine.math.early}{0.97}
\setval{olmo.gd2po.primary-cosine.math.late}{0.88}
\setval{olmo.gd2po.primary-cosine.math.run}{0.89}
\setval{olmo.gd2po.primary-cosine.writing.early}{0.81}
\setval{olmo.gd2po.primary-cosine.writing.late}{0.58}
\setval{olmo.gd2po.primary-cosine.writing.run}{0.62}
\setval{olmo.gd2po.primary-level.code.early}{0.58}
\setval{olmo.gd2po.primary-level.code.late}{0.84}
\setval{olmo.gd2po.primary-level.code.run}{0.76}
\setval{olmo.gd2po.primary-level.if.early}{0.63}
\setval{olmo.gd2po.primary-level.if.late}{0.74}
\setval{olmo.gd2po.primary-level.if.run}{0.69}
\setval{olmo.gd2po.primary-level.math.early}{0.34}
\setval{olmo.gd2po.primary-level.math.late}{0.67}
\setval{olmo.gd2po.primary-level.math.run}{0.57}
\setval{olmo.gd2po.primary-level.writing.early}{0.49}
\setval{olmo.gd2po.primary-level.writing.late}{0.68}
\setval{olmo.gd2po.primary-level.writing.run}{0.63}
\setval{olmo.gd2po.solved.code.early}{0.13}
\setval{olmo.gd2po.solved.code.late}{0.15}
\setval{olmo.gd2po.solved.code.run}{0.14}
\setval{olmo.gd2po.solved.if.early}{0.17}
\setval{olmo.gd2po.solved.if.late}{0.17}
\setval{olmo.gd2po.solved.if.run}{0.17}
\setval{olmo.gd2po.solved.math.early}{0.07}
\setval{olmo.gd2po.solved.math.late}{0.24}
\setval{olmo.gd2po.solved.math.run}{0.19}
\setval{olmo.gd2po.solved.writing.early}{0.00}
\setval{olmo.gd2po.solved.writing.late}{0.00}
\setval{olmo.gd2po.solved.writing.run}{0.00}
\setval{olmo.gd2po.update-share.code.early}{0.22}
\setval{olmo.gd2po.update-share.code.late}{0.21}
\setval{olmo.gd2po.update-share.code.run}{0.24}
\setval{olmo.gd2po.update-share.if.early}{0.13}
\setval{olmo.gd2po.update-share.if.late}{0.20}
\setval{olmo.gd2po.update-share.if.run}{0.18}
\setval{olmo.gd2po.update-share.math.early}{0.27}
\setval{olmo.gd2po.update-share.math.late}{0.34}
\setval{olmo.gd2po.update-share.math.run}{0.31}
\setval{olmo.gd2po.update-share.writing.early}{0.38}
\setval{olmo.gd2po.update-share.writing.late}{0.25}
\setval{olmo.gd2po.update-share.writing.run}{0.27}
\setval{olmo.gdpo-noaux.aux-change.code}{$-$0.77}
\setval{olmo.gdpo-noaux.aux-change.if}{$-$0.11}
\setval{olmo.gdpo-noaux.aux-change.math}{$-$0.82}
\setval{olmo.gdpo-noaux.aux-change.writing}{$-$0.87}
\setval{olmo.gdpo-noaux.aux-change.writing-rm}{+0.15}
\setval{olmo.gdpo-noaux.aux-full.code.early}{0.57}
\setval{olmo.gdpo-noaux.aux-full.code.late}{0.01}
\setval{olmo.gdpo-noaux.aux-full.code.run}{0.12}
\setval{olmo.gdpo-noaux.aux-full.if.early}{0.54}
\setval{olmo.gdpo-noaux.aux-full.if.late}{0.36}
\setval{olmo.gdpo-noaux.aux-full.if.run}{0.35}
\setval{olmo.gdpo-noaux.aux-full.math.early}{0.48}
\setval{olmo.gdpo-noaux.aux-full.math.late}{0.01}
\setval{olmo.gdpo-noaux.aux-full.math.run}{0.08}
\setval{olmo.gdpo-noaux.aux-full.writing-rm.early}{0.02}
\setval{olmo.gdpo-noaux.aux-full.writing-rm.late}{0.02}
\setval{olmo.gdpo-noaux.aux-full.writing-rm.run}{0.02}
\setval{olmo.gdpo-noaux.aux-full.writing.early}{0.49}
\setval{olmo.gdpo-noaux.aux-full.writing.late}{0.02}
\setval{olmo.gdpo-noaux.aux-full.writing.run}{0.07}
\setval{olmo.gdpo-noaux.aux-level.code.early}{0.77}
\setval{olmo.gdpo-noaux.aux-level.code.late}{0.04}
\setval{olmo.gdpo-noaux.aux-level.code.run}{0.21}
\setval{olmo.gdpo-noaux.aux-level.if.early}{0.88}
\setval{olmo.gdpo-noaux.aux-level.if.late}{0.77}
\setval{olmo.gdpo-noaux.aux-level.if.run}{0.76}
\setval{olmo.gdpo-noaux.aux-level.math.early}{0.85}
\setval{olmo.gdpo-noaux.aux-level.math.late}{0.08}
\setval{olmo.gdpo-noaux.aux-level.math.run}{0.26}
\setval{olmo.gdpo-noaux.aux-level.writing-rm.early}{0.60}
\setval{olmo.gdpo-noaux.aux-level.writing-rm.late}{0.72}
\setval{olmo.gdpo-noaux.aux-level.writing-rm.run}{0.69}
\setval{olmo.gdpo-noaux.aux-level.writing.early}{0.94}
\setval{olmo.gdpo-noaux.aux-level.writing.late}{0.08}
\setval{olmo.gdpo-noaux.aux-level.writing.run}{0.26}
\setval{olmo.gdpo-noaux.bench.aime24}{38.3}
\setval{olmo.gdpo-noaux.bench.aime25}{33.6}
\setval{olmo.gdpo-noaux.bench.amc23}{83.9}
\setval{olmo.gdpo-noaux.bench.creative-writing-v3}{38.2}
\setval{olmo.gdpo-noaux.bench.guidebench}{55.8}
\setval{olmo.gdpo-noaux.bench.humaneval-plus}{85.2}
\setval{olmo.gdpo-noaux.bench.ifbench}{33.3}
\setval{olmo.gdpo-noaux.bench.ifeval}{84.8}
\setval{olmo.gdpo-noaux.bench.livecodebench}{31.5}
\setval{olmo.gdpo-noaux.bench.longbench-write}{56.0}
\setval{olmo.gdpo-noaux.bench.math500}{90.9}
\setval{olmo.gdpo-noaux.bench.mbpp-plus}{70.5}
\setval{olmo.gdpo-noaux.bench.minerva}{46.2}
\setval{olmo.gdpo-noaux.bench.olympiad-bench}{63.7}
\setval{olmo.gdpo-noaux.bench.writingbench}{55.5}
\setval{olmo.gdpo-noaux.eval.avg}{57.8}
\setval{olmo.gdpo-noaux.eval.code}{62.4}
\setval{olmo.gdpo-noaux.eval.if}{58.0}
\setval{olmo.gdpo-noaux.eval.math}{59.4}
\setval{olmo.gdpo-noaux.eval.writing}{49.9}
\setval{olmo.gdpo-noaux.gate.code.against.early}{0.09}
\setval{olmo.gdpo-noaux.gate.code.against.late}{0.09}
\setval{olmo.gdpo-noaux.gate.code.against.run}{0.09}
\setval{olmo.gdpo-noaux.gate.code.agreement}{+0.005}
\setval{olmo.gdpo-noaux.gate.code.ceiling.early}{0.35}
\setval{olmo.gdpo-noaux.gate.code.ceiling.late}{0.68}
\setval{olmo.gdpo-noaux.gate.code.ceiling.run}{0.67}
\setval{olmo.gdpo-noaux.gate.code.cosine.early}{0.60}
\setval{olmo.gdpo-noaux.gate.code.cosine.late}{0.10}
\setval{olmo.gdpo-noaux.gate.code.cosine.run}{0.18}
\setval{olmo.gdpo-noaux.gate.if.against.early}{0.10}
\setval{olmo.gdpo-noaux.gate.if.against.late}{0.11}
\setval{olmo.gdpo-noaux.gate.if.against.run}{0.11}
\setval{olmo.gdpo-noaux.gate.if.agreement}{+0.130}
\setval{olmo.gdpo-noaux.gate.if.ceiling.early}{0.22}
\setval{olmo.gdpo-noaux.gate.if.ceiling.late}{0.33}
\setval{olmo.gdpo-noaux.gate.if.ceiling.run}{0.28}
\setval{olmo.gdpo-noaux.gate.if.cosine.early}{0.50}
\setval{olmo.gdpo-noaux.gate.if.cosine.late}{0.33}
\setval{olmo.gdpo-noaux.gate.if.cosine.run}{0.38}
\setval{olmo.gdpo-noaux.gate.math.against.early}{0.00}
\setval{olmo.gdpo-noaux.gate.math.against.late}{0.00}
\setval{olmo.gdpo-noaux.gate.math.against.run}{0.00}
\setval{olmo.gdpo-noaux.gate.math.ceiling.early}{0.19}
\setval{olmo.gdpo-noaux.gate.math.ceiling.late}{0.90}
\setval{olmo.gdpo-noaux.gate.math.ceiling.run}{0.78}
\setval{olmo.gdpo-noaux.gate.math.cosine.early}{0.86}
\setval{olmo.gdpo-noaux.gate.math.cosine.late}{0.11}
\setval{olmo.gdpo-noaux.gate.math.cosine.run}{0.27}
\setval{olmo.gdpo-noaux.gate.writing-rm.against.early}{0.21}
\setval{olmo.gdpo-noaux.gate.writing-rm.against.late}{0.54}
\setval{olmo.gdpo-noaux.gate.writing-rm.against.run}{0.45}
\setval{olmo.gdpo-noaux.gate.writing-rm.agreement}{+0.101}
\setval{olmo.gdpo-noaux.gate.writing-rm.ceiling.early}{0.00}
\setval{olmo.gdpo-noaux.gate.writing-rm.ceiling.late}{0.00}
\setval{olmo.gdpo-noaux.gate.writing-rm.ceiling.run}{0.00}
\setval{olmo.gdpo-noaux.gate.writing-rm.cosine.early}{0.71}
\setval{olmo.gdpo-noaux.gate.writing-rm.cosine.late}{0.22}
\setval{olmo.gdpo-noaux.gate.writing-rm.cosine.run}{0.35}
\setval{olmo.gdpo-noaux.gate.writing.against.early}{0.25}
\setval{olmo.gdpo-noaux.gate.writing.against.late}{0.65}
\setval{olmo.gdpo-noaux.gate.writing.against.run}{0.62}
\setval{olmo.gdpo-noaux.gate.writing.agreement}{$-$0.036}
\setval{olmo.gdpo-noaux.gate.writing.ceiling.early}{0.00}
\setval{olmo.gdpo-noaux.gate.writing.ceiling.late}{0.00}
\setval{olmo.gdpo-noaux.gate.writing.ceiling.run}{0.00}
\setval{olmo.gdpo-noaux.gate.writing.cosine.early}{0.60}
\setval{olmo.gdpo-noaux.gate.writing.cosine.late}{0.01}
\setval{olmo.gdpo-noaux.gate.writing.cosine.run}{0.07}
\setval{olmo.gdpo-noaux.judge2.avg}{55.7}
\setval{olmo.gdpo-noaux.judge2.creative-writing-v3}{31.0}
\setval{olmo.gdpo-noaux.judge2.longbench-write}{46.6}
\setval{olmo.gdpo-noaux.judge2.writing}{39.1}
\setval{olmo.gdpo-noaux.judge2.writingbench}{39.8}
\setval{olmo.gdpo-noaux.last-step}{299}
\setval{olmo.gdpo-noaux.minus.base.aime24}{+29.8}
\setval{olmo.gdpo-noaux.minus.base.aime24-abs}{29.8}
\setval{olmo.gdpo-noaux.minus.base.aime25}{+26.3}
\setval{olmo.gdpo-noaux.minus.base.aime25-abs}{26.3}
\setval{olmo.gdpo-noaux.minus.base.amc23}{+39.4}
\setval{olmo.gdpo-noaux.minus.base.amc23-abs}{39.4}
\setval{olmo.gdpo-noaux.minus.base.avg}{+13.7}
\setval{olmo.gdpo-noaux.minus.base.avg-abs}{13.7}
\setval{olmo.gdpo-noaux.minus.base.code}{+13.3}
\setval{olmo.gdpo-noaux.minus.base.code-abs}{13.3}
\setval{olmo.gdpo-noaux.minus.base.creative-writing-v3}{+0.3}
\setval{olmo.gdpo-noaux.minus.base.creative-writing-v3-abs}{0.3}
\setval{olmo.gdpo-noaux.minus.base.guidebench}{$-$5.7}
\setval{olmo.gdpo-noaux.minus.base.guidebench-abs}{5.7}
\setval{olmo.gdpo-noaux.minus.base.humaneval-plus}{+16.0}
\setval{olmo.gdpo-noaux.minus.base.humaneval-plus-abs}{16.0}
\setval{olmo.gdpo-noaux.minus.base.if}{+0.5}
\setval{olmo.gdpo-noaux.minus.base.if-abs}{0.5}
\setval{olmo.gdpo-noaux.minus.base.ifbench}{+6.7}
\setval{olmo.gdpo-noaux.minus.base.ifbench-abs}{6.7}
\setval{olmo.gdpo-noaux.minus.base.ifeval}{+0.4}
\setval{olmo.gdpo-noaux.minus.base.ifeval-abs}{0.4}
\setval{olmo.gdpo-noaux.minus.base.judge2.avg}{+12.7}
\setval{olmo.gdpo-noaux.minus.base.judge2.avg-abs}{12.7}
\setval{olmo.gdpo-noaux.minus.base.judge2.creative-writing-v3}{$-$0.7}
\setval{olmo.gdpo-noaux.minus.base.judge2.creative-writing-v3-abs}{0.7}
\setval{olmo.gdpo-noaux.minus.base.judge2.longbench-write}{$-$6.5}
\setval{olmo.gdpo-noaux.minus.base.judge2.longbench-write-abs}{6.5}
\setval{olmo.gdpo-noaux.minus.base.judge2.writing}{$-$1.7}
\setval{olmo.gdpo-noaux.minus.base.judge2.writing-abs}{1.7}
\setval{olmo.gdpo-noaux.minus.base.judge2.writingbench}{+2.1}
\setval{olmo.gdpo-noaux.minus.base.judge2.writingbench-abs}{2.1}
\setval{olmo.gdpo-noaux.minus.base.livecodebench}{+14.7}
\setval{olmo.gdpo-noaux.minus.base.livecodebench-abs}{14.7}
\setval{olmo.gdpo-noaux.minus.base.longbench-write}{$-$1.5}
\setval{olmo.gdpo-noaux.minus.base.longbench-write-abs}{1.5}
\setval{olmo.gdpo-noaux.minus.base.math}{+25.8}
\setval{olmo.gdpo-noaux.minus.base.math-abs}{25.8}
\setval{olmo.gdpo-noaux.minus.base.math500}{+20.8}
\setval{olmo.gdpo-noaux.minus.base.math500-abs}{20.8}
\setval{olmo.gdpo-noaux.minus.base.mbpp-plus}{+9.3}
\setval{olmo.gdpo-noaux.minus.base.mbpp-plus-abs}{9.3}
\setval{olmo.gdpo-noaux.minus.base.minerva}{+10.9}
\setval{olmo.gdpo-noaux.minus.base.minerva-abs}{10.9}
\setval{olmo.gdpo-noaux.minus.base.olympiad-bench}{+27.5}
\setval{olmo.gdpo-noaux.minus.base.olympiad-bench-abs}{27.5}
\setval{olmo.gdpo-noaux.minus.base.writing}{+3.2}
\setval{olmo.gdpo-noaux.minus.base.writing-abs}{3.2}
\setval{olmo.gdpo-noaux.minus.base.writingbench}{+10.7}
\setval{olmo.gdpo-noaux.minus.base.writingbench-abs}{10.7}
\setval{olmo.gdpo-noaux.minus.concur-free.aime24}{+11.6}
\setval{olmo.gdpo-noaux.minus.concur-free.aime24-abs}{11.6}
\setval{olmo.gdpo-noaux.minus.concur-free.aime25}{+11.6}
\setval{olmo.gdpo-noaux.minus.concur-free.aime25-abs}{11.6}
\setval{olmo.gdpo-noaux.minus.concur-free.amc23}{+16.4}
\setval{olmo.gdpo-noaux.minus.concur-free.amc23-abs}{16.4}
\setval{olmo.gdpo-noaux.minus.concur-free.avg}{+2.4}
\setval{olmo.gdpo-noaux.minus.concur-free.avg-abs}{2.4}
\setval{olmo.gdpo-noaux.minus.concur-free.code}{+6.8}
\setval{olmo.gdpo-noaux.minus.concur-free.code-abs}{6.8}
\setval{olmo.gdpo-noaux.minus.concur-free.creative-writing-v3}{$-$11.6}
\setval{olmo.gdpo-noaux.minus.concur-free.creative-writing-v3-abs}{11.6}
\setval{olmo.gdpo-noaux.minus.concur-free.guidebench}{$-$4.4}
\setval{olmo.gdpo-noaux.minus.concur-free.guidebench-abs}{4.4}
\setval{olmo.gdpo-noaux.minus.concur-free.humaneval-plus}{+8.5}
\setval{olmo.gdpo-noaux.minus.concur-free.humaneval-plus-abs}{8.5}
\setval{olmo.gdpo-noaux.minus.concur-free.if}{$-$1.8}
\setval{olmo.gdpo-noaux.minus.concur-free.if-abs}{1.8}
\setval{olmo.gdpo-noaux.minus.concur-free.ifbench}{+1.7}
\setval{olmo.gdpo-noaux.minus.concur-free.ifbench-abs}{1.7}
\setval{olmo.gdpo-noaux.minus.concur-free.ifeval}{$-$2.5}
\setval{olmo.gdpo-noaux.minus.concur-free.ifeval-abs}{2.5}
\setval{olmo.gdpo-noaux.minus.concur-free.judge2.avg}{+3.0}
\setval{olmo.gdpo-noaux.minus.concur-free.judge2.avg-abs}{3.0}
\setval{olmo.gdpo-noaux.minus.concur-free.judge2.creative-writing-v3}{$-$6.6}
\setval{olmo.gdpo-noaux.minus.concur-free.judge2.creative-writing-v3-abs}{6.6}
\setval{olmo.gdpo-noaux.minus.concur-free.judge2.longbench-write}{$-$17.3}
\setval{olmo.gdpo-noaux.minus.concur-free.judge2.longbench-write-abs}{17.3}
\setval{olmo.gdpo-noaux.minus.concur-free.judge2.writing}{$-$10.7}
\setval{olmo.gdpo-noaux.minus.concur-free.judge2.writing-abs}{10.7}
\setval{olmo.gdpo-noaux.minus.concur-free.judge2.writingbench}{$-$8.3}
\setval{olmo.gdpo-noaux.minus.concur-free.judge2.writingbench-abs}{8.3}
\setval{olmo.gdpo-noaux.minus.concur-free.livecodebench}{+5.9}
\setval{olmo.gdpo-noaux.minus.concur-free.livecodebench-abs}{5.9}
\setval{olmo.gdpo-noaux.minus.concur-free.longbench-write}{$-$16.1}
\setval{olmo.gdpo-noaux.minus.concur-free.longbench-write-abs}{16.1}
\setval{olmo.gdpo-noaux.minus.concur-free.math}{+10.5}
\setval{olmo.gdpo-noaux.minus.concur-free.math-abs}{10.5}
\setval{olmo.gdpo-noaux.minus.concur-free.math500}{+6.5}
\setval{olmo.gdpo-noaux.minus.concur-free.math500-abs}{6.5}
\setval{olmo.gdpo-noaux.minus.concur-free.mbpp-plus}{+6.0}
\setval{olmo.gdpo-noaux.minus.concur-free.mbpp-plus-abs}{6.0}
\setval{olmo.gdpo-noaux.minus.concur-free.minerva}{+6.2}
\setval{olmo.gdpo-noaux.minus.concur-free.minerva-abs}{6.2}
\setval{olmo.gdpo-noaux.minus.concur-free.olympiad-bench}{+10.5}
\setval{olmo.gdpo-noaux.minus.concur-free.olympiad-bench-abs}{10.5}
\setval{olmo.gdpo-noaux.minus.concur-free.writing}{$-$14.1}
\setval{olmo.gdpo-noaux.minus.concur-free.writing-abs}{14.1}
\setval{olmo.gdpo-noaux.minus.concur-free.writingbench}{$-$14.7}
\setval{olmo.gdpo-noaux.minus.concur-free.writingbench-abs}{14.7}
\setval{olmo.gdpo-noaux.minus.concur-tied.aime24}{+3.9}
\setval{olmo.gdpo-noaux.minus.concur-tied.aime24-abs}{3.9}
\setval{olmo.gdpo-noaux.minus.concur-tied.aime25}{+3.8}
\setval{olmo.gdpo-noaux.minus.concur-tied.aime25-abs}{3.8}
\setval{olmo.gdpo-noaux.minus.concur-tied.amc23}{+4.1}
\setval{olmo.gdpo-noaux.minus.concur-tied.amc23-abs}{4.1}
\setval{olmo.gdpo-noaux.minus.concur-tied.avg}{$-$1.3}
\setval{olmo.gdpo-noaux.minus.concur-tied.avg-abs}{1.3}
\setval{olmo.gdpo-noaux.minus.concur-tied.code}{+1.3}
\setval{olmo.gdpo-noaux.minus.concur-tied.code-abs}{1.3}
\setval{olmo.gdpo-noaux.minus.concur-tied.creative-writing-v3}{$-$10.8}
\setval{olmo.gdpo-noaux.minus.concur-tied.creative-writing-v3-abs}{10.8}
\setval{olmo.gdpo-noaux.minus.concur-tied.guidebench}{$-$2.8}
\setval{olmo.gdpo-noaux.minus.concur-tied.guidebench-abs}{2.8}
\setval{olmo.gdpo-noaux.minus.concur-tied.humaneval-plus}{+1.9}
\setval{olmo.gdpo-noaux.minus.concur-tied.humaneval-plus-abs}{1.9}
\setval{olmo.gdpo-noaux.minus.concur-tied.if}{$-$2.0}
\setval{olmo.gdpo-noaux.minus.concur-tied.if-abs}{2.0}
\setval{olmo.gdpo-noaux.minus.concur-tied.ifbench}{$-$0.3}
\setval{olmo.gdpo-noaux.minus.concur-tied.ifbench-abs}{0.3}
\setval{olmo.gdpo-noaux.minus.concur-tied.ifeval}{$-$2.9}
\setval{olmo.gdpo-noaux.minus.concur-tied.ifeval-abs}{2.9}
\setval{olmo.gdpo-noaux.minus.concur-tied.judge2.avg}{$-$0.7}
\setval{olmo.gdpo-noaux.minus.concur-tied.judge2.avg-abs}{0.7}
\setval{olmo.gdpo-noaux.minus.concur-tied.judge2.creative-writing-v3}{$-$6.2}
\setval{olmo.gdpo-noaux.minus.concur-tied.judge2.creative-writing-v3-abs}{6.2}
\setval{olmo.gdpo-noaux.minus.concur-tied.judge2.longbench-write}{$-$13.9}
\setval{olmo.gdpo-noaux.minus.concur-tied.judge2.longbench-write-abs}{13.9}
\setval{olmo.gdpo-noaux.minus.concur-tied.judge2.writing}{$-$8.7}
\setval{olmo.gdpo-noaux.minus.concur-tied.judge2.writing-abs}{8.7}
\setval{olmo.gdpo-noaux.minus.concur-tied.judge2.writingbench}{$-$5.8}
\setval{olmo.gdpo-noaux.minus.concur-tied.judge2.writingbench-abs}{5.8}
\setval{olmo.gdpo-noaux.minus.concur-tied.livecodebench}{+2.5}
\setval{olmo.gdpo-noaux.minus.concur-tied.livecodebench-abs}{2.5}
\setval{olmo.gdpo-noaux.minus.concur-tied.longbench-write}{$-$13.3}
\setval{olmo.gdpo-noaux.minus.concur-tied.longbench-write-abs}{13.3}
\setval{olmo.gdpo-noaux.minus.concur-tied.math}{+2.9}
\setval{olmo.gdpo-noaux.minus.concur-tied.math-abs}{2.9}
\setval{olmo.gdpo-noaux.minus.concur-tied.math500}{+1.5}
\setval{olmo.gdpo-noaux.minus.concur-tied.math500-abs}{1.5}
\setval{olmo.gdpo-noaux.minus.concur-tied.mbpp-plus}{$-$0.6}
\setval{olmo.gdpo-noaux.minus.concur-tied.mbpp-plus-abs}{0.6}
\setval{olmo.gdpo-noaux.minus.concur-tied.minerva}{+2.1}
\setval{olmo.gdpo-noaux.minus.concur-tied.minerva-abs}{2.1}
\setval{olmo.gdpo-noaux.minus.concur-tied.olympiad-bench}{+2.2}
\setval{olmo.gdpo-noaux.minus.concur-tied.olympiad-bench-abs}{2.2}
\setval{olmo.gdpo-noaux.minus.concur-tied.writing}{$-$11.6}
\setval{olmo.gdpo-noaux.minus.concur-tied.writing-abs}{11.6}
\setval{olmo.gdpo-noaux.minus.concur-tied.writingbench}{$-$10.7}
\setval{olmo.gdpo-noaux.minus.concur-tied.writingbench-abs}{10.7}
\setval{olmo.gdpo-noaux.minus.concur.aime24}{+2.0}
\setval{olmo.gdpo-noaux.minus.concur.aime24-abs}{2.0}
\setval{olmo.gdpo-noaux.minus.concur.aime25}{+1.5}
\setval{olmo.gdpo-noaux.minus.concur.aime25-abs}{1.5}
\setval{olmo.gdpo-noaux.minus.concur.amc23}{+0.4}
\setval{olmo.gdpo-noaux.minus.concur.amc23-abs}{0.4}
\setval{olmo.gdpo-noaux.minus.concur.avg}{$-$1.5}
\setval{olmo.gdpo-noaux.minus.concur.avg-abs}{1.5}
\setval{olmo.gdpo-noaux.minus.concur.code}{+2.2}
\setval{olmo.gdpo-noaux.minus.concur.code-abs}{2.2}
\setval{olmo.gdpo-noaux.minus.concur.creative-writing-v3}{$-$11.0}
\setval{olmo.gdpo-noaux.minus.concur.creative-writing-v3-abs}{11.0}
\setval{olmo.gdpo-noaux.minus.concur.guidebench}{$-$1.2}
\setval{olmo.gdpo-noaux.minus.concur.guidebench-abs}{1.2}
\setval{olmo.gdpo-noaux.minus.concur.humaneval-plus}{+2.4}
\setval{olmo.gdpo-noaux.minus.concur.humaneval-plus-abs}{2.4}
\setval{olmo.gdpo-noaux.minus.concur.if}{$-$0.9}
\setval{olmo.gdpo-noaux.minus.concur.if-abs}{0.9}
\setval{olmo.gdpo-noaux.minus.concur.ifbench}{+1.3}
\setval{olmo.gdpo-noaux.minus.concur.ifbench-abs}{1.3}
\setval{olmo.gdpo-noaux.minus.concur.ifeval}{$-$3.0}
\setval{olmo.gdpo-noaux.minus.concur.ifeval-abs}{3.0}
\setval{olmo.gdpo-noaux.minus.concur.judge2.avg}{$-$0.9}
\setval{olmo.gdpo-noaux.minus.concur.judge2.avg-abs}{0.9}
\setval{olmo.gdpo-noaux.minus.concur.judge2.creative-writing-v3}{$-$7.5}
\setval{olmo.gdpo-noaux.minus.concur.judge2.creative-writing-v3-abs}{7.5}
\setval{olmo.gdpo-noaux.minus.concur.judge2.longbench-write}{$-$11.7}
\setval{olmo.gdpo-noaux.minus.concur.judge2.longbench-write-abs}{11.7}
\setval{olmo.gdpo-noaux.minus.concur.judge2.writing}{$-$8.1}
\setval{olmo.gdpo-noaux.minus.concur.judge2.writing-abs}{8.1}
\setval{olmo.gdpo-noaux.minus.concur.judge2.writingbench}{$-$5.1}
\setval{olmo.gdpo-noaux.minus.concur.judge2.writingbench-abs}{5.1}
\setval{olmo.gdpo-noaux.minus.concur.livecodebench}{+1.2}
\setval{olmo.gdpo-noaux.minus.concur.livecodebench-abs}{1.2}
\setval{olmo.gdpo-noaux.minus.concur.longbench-write}{$-$12.5}
\setval{olmo.gdpo-noaux.minus.concur.longbench-write-abs}{12.5}
\setval{olmo.gdpo-noaux.minus.concur.math}{+1.1}
\setval{olmo.gdpo-noaux.minus.concur.math-abs}{1.1}
\setval{olmo.gdpo-noaux.minus.concur.math500}{+0.3}
\setval{olmo.gdpo-noaux.minus.concur.math500-abs}{0.3}
\setval{olmo.gdpo-noaux.minus.concur.mbpp-plus}{+3.0}
\setval{olmo.gdpo-noaux.minus.concur.mbpp-plus-abs}{3.0}
\setval{olmo.gdpo-noaux.minus.concur.minerva}{+1.3}
\setval{olmo.gdpo-noaux.minus.concur.minerva-abs}{1.3}
\setval{olmo.gdpo-noaux.minus.concur.olympiad-bench}{+1.3}
\setval{olmo.gdpo-noaux.minus.concur.olympiad-bench-abs}{1.3}
\setval{olmo.gdpo-noaux.minus.concur.writing}{$-$11.2}
\setval{olmo.gdpo-noaux.minus.concur.writing-abs}{11.2}
\setval{olmo.gdpo-noaux.minus.concur.writingbench}{$-$10.0}
\setval{olmo.gdpo-noaux.minus.concur.writingbench-abs}{10.0}
\setval{olmo.gdpo-noaux.minus.gd2po.aime24}{+10.0}
\setval{olmo.gdpo-noaux.minus.gd2po.aime24-abs}{10.0}
\setval{olmo.gdpo-noaux.minus.gd2po.aime25}{+9.4}
\setval{olmo.gdpo-noaux.minus.gd2po.aime25-abs}{9.4}
\setval{olmo.gdpo-noaux.minus.gd2po.amc23}{+11.7}
\setval{olmo.gdpo-noaux.minus.gd2po.amc23-abs}{11.7}
\setval{olmo.gdpo-noaux.minus.gd2po.avg}{+2.2}
\setval{olmo.gdpo-noaux.minus.gd2po.avg-abs}{2.2}
\setval{olmo.gdpo-noaux.minus.gd2po.code}{+7.5}
\setval{olmo.gdpo-noaux.minus.gd2po.code-abs}{7.5}
\setval{olmo.gdpo-noaux.minus.gd2po.creative-writing-v3}{$-$10.1}
\setval{olmo.gdpo-noaux.minus.gd2po.creative-writing-v3-abs}{10.1}
\setval{olmo.gdpo-noaux.minus.gd2po.guidebench}{$-$5.4}
\setval{olmo.gdpo-noaux.minus.gd2po.guidebench-abs}{5.4}
\setval{olmo.gdpo-noaux.minus.gd2po.humaneval-plus}{+8.4}
\setval{olmo.gdpo-noaux.minus.gd2po.humaneval-plus-abs}{8.4}
\setval{olmo.gdpo-noaux.minus.gd2po.if}{$-$2.4}
\setval{olmo.gdpo-noaux.minus.gd2po.if-abs}{2.4}
\setval{olmo.gdpo-noaux.minus.gd2po.ifbench}{+1.0}
\setval{olmo.gdpo-noaux.minus.gd2po.ifbench-abs}{1.0}
\setval{olmo.gdpo-noaux.minus.gd2po.ifeval}{$-$2.9}
\setval{olmo.gdpo-noaux.minus.gd2po.ifeval-abs}{2.9}
\setval{olmo.gdpo-noaux.minus.gd2po.judge2.avg}{+2.4}
\setval{olmo.gdpo-noaux.minus.gd2po.judge2.avg-abs}{2.4}
\setval{olmo.gdpo-noaux.minus.gd2po.judge2.creative-writing-v3}{$-$6.9}
\setval{olmo.gdpo-noaux.minus.gd2po.judge2.creative-writing-v3-abs}{6.9}
\setval{olmo.gdpo-noaux.minus.gd2po.judge2.longbench-write}{$-$13.3}
\setval{olmo.gdpo-noaux.minus.gd2po.judge2.longbench-write-abs}{13.3}
\setval{olmo.gdpo-noaux.minus.gd2po.judge2.writing}{$-$9.6}
\setval{olmo.gdpo-noaux.minus.gd2po.judge2.writing-abs}{9.6}
\setval{olmo.gdpo-noaux.minus.gd2po.judge2.writingbench}{$-$8.6}
\setval{olmo.gdpo-noaux.minus.gd2po.judge2.writingbench-abs}{8.6}
\setval{olmo.gdpo-noaux.minus.gd2po.livecodebench}{+8.3}
\setval{olmo.gdpo-noaux.minus.gd2po.livecodebench-abs}{8.3}
\setval{olmo.gdpo-noaux.minus.gd2po.longbench-write}{$-$12.2}
\setval{olmo.gdpo-noaux.minus.gd2po.longbench-write-abs}{12.2}
\setval{olmo.gdpo-noaux.minus.gd2po.math}{+8.3}
\setval{olmo.gdpo-noaux.minus.gd2po.math-abs}{8.3}
\setval{olmo.gdpo-noaux.minus.gd2po.math500}{+5.2}
\setval{olmo.gdpo-noaux.minus.gd2po.math500-abs}{5.2}
\setval{olmo.gdpo-noaux.minus.gd2po.mbpp-plus}{+5.7}
\setval{olmo.gdpo-noaux.minus.gd2po.mbpp-plus-abs}{5.7}
\setval{olmo.gdpo-noaux.minus.gd2po.minerva}{+6.0}
\setval{olmo.gdpo-noaux.minus.gd2po.minerva-abs}{6.0}
\setval{olmo.gdpo-noaux.minus.gd2po.olympiad-bench}{+7.7}
\setval{olmo.gdpo-noaux.minus.gd2po.olympiad-bench-abs}{7.7}
\setval{olmo.gdpo-noaux.minus.gd2po.writing}{$-$10.9}
\setval{olmo.gdpo-noaux.minus.gd2po.writing-abs}{10.9}
\setval{olmo.gdpo-noaux.minus.gd2po.writingbench}{$-$10.3}
\setval{olmo.gdpo-noaux.minus.gd2po.writingbench-abs}{10.3}
\setval{olmo.gdpo-noaux.minus.gdpo-w0.5.aime24}{+8.8}
\setval{olmo.gdpo-noaux.minus.gdpo-w0.5.aime24-abs}{8.8}
\setval{olmo.gdpo-noaux.minus.gdpo-w0.5.aime25}{+6.4}
\setval{olmo.gdpo-noaux.minus.gdpo-w0.5.aime25-abs}{6.4}
\setval{olmo.gdpo-noaux.minus.gdpo-w0.5.amc23}{+9.9}
\setval{olmo.gdpo-noaux.minus.gdpo-w0.5.amc23-abs}{9.9}
\setval{olmo.gdpo-noaux.minus.gdpo-w0.5.avg}{+2.2}
\setval{olmo.gdpo-noaux.minus.gdpo-w0.5.avg-abs}{2.2}
\setval{olmo.gdpo-noaux.minus.gdpo-w0.5.code}{+6.6}
\setval{olmo.gdpo-noaux.minus.gdpo-w0.5.code-abs}{6.6}
\setval{olmo.gdpo-noaux.minus.gdpo-w0.5.creative-writing-v3}{$-$9.9}
\setval{olmo.gdpo-noaux.minus.gdpo-w0.5.creative-writing-v3-abs}{9.9}
\setval{olmo.gdpo-noaux.minus.gdpo-w0.5.guidebench}{+3.8}
\setval{olmo.gdpo-noaux.minus.gdpo-w0.5.guidebench-abs}{3.8}
\setval{olmo.gdpo-noaux.minus.gdpo-w0.5.humaneval-plus}{+7.4}
\setval{olmo.gdpo-noaux.minus.gdpo-w0.5.humaneval-plus-abs}{7.4}
\setval{olmo.gdpo-noaux.minus.gdpo-w0.5.if}{+0.6}
\setval{olmo.gdpo-noaux.minus.gdpo-w0.5.if-abs}{0.6}
\setval{olmo.gdpo-noaux.minus.gdpo-w0.5.ifbench}{+0.7}
\setval{olmo.gdpo-noaux.minus.gdpo-w0.5.ifbench-abs}{0.7}
\setval{olmo.gdpo-noaux.minus.gdpo-w0.5.ifeval}{$-$2.8}
\setval{olmo.gdpo-noaux.minus.gdpo-w0.5.ifeval-abs}{2.8}
\setval{olmo.gdpo-noaux.minus.gdpo-w0.5.judge2.avg}{+2.4}
\setval{olmo.gdpo-noaux.minus.gdpo-w0.5.judge2.avg-abs}{2.4}
\setval{olmo.gdpo-noaux.minus.gdpo-w0.5.judge2.creative-writing-v3}{$-$8.5}
\setval{olmo.gdpo-noaux.minus.gdpo-w0.5.judge2.creative-writing-v3-abs}{8.5}
\setval{olmo.gdpo-noaux.minus.gdpo-w0.5.judge2.longbench-write}{$-$10.0}
\setval{olmo.gdpo-noaux.minus.gdpo-w0.5.judge2.longbench-write-abs}{10.0}
\setval{olmo.gdpo-noaux.minus.gdpo-w0.5.judge2.writing}{$-$8.5}
\setval{olmo.gdpo-noaux.minus.gdpo-w0.5.judge2.writing-abs}{8.5}
\setval{olmo.gdpo-noaux.minus.gdpo-w0.5.judge2.writingbench}{$-$6.9}
\setval{olmo.gdpo-noaux.minus.gdpo-w0.5.judge2.writingbench-abs}{6.9}
\setval{olmo.gdpo-noaux.minus.gdpo-w0.5.livecodebench}{+6.7}
\setval{olmo.gdpo-noaux.minus.gdpo-w0.5.livecodebench-abs}{6.7}
\setval{olmo.gdpo-noaux.minus.gdpo-w0.5.longbench-write}{$-$7.7}
\setval{olmo.gdpo-noaux.minus.gdpo-w0.5.longbench-write-abs}{7.7}
\setval{olmo.gdpo-noaux.minus.gdpo-w0.5.math}{+6.6}
\setval{olmo.gdpo-noaux.minus.gdpo-w0.5.math-abs}{6.6}
\setval{olmo.gdpo-noaux.minus.gdpo-w0.5.math500}{+3.8}
\setval{olmo.gdpo-noaux.minus.gdpo-w0.5.math500-abs}{3.8}
\setval{olmo.gdpo-noaux.minus.gdpo-w0.5.mbpp-plus}{+5.6}
\setval{olmo.gdpo-noaux.minus.gdpo-w0.5.mbpp-plus-abs}{5.6}
\setval{olmo.gdpo-noaux.minus.gdpo-w0.5.minerva}{+4.6}
\setval{olmo.gdpo-noaux.minus.gdpo-w0.5.minerva-abs}{4.6}
\setval{olmo.gdpo-noaux.minus.gdpo-w0.5.olympiad-bench}{+6.2}
\setval{olmo.gdpo-noaux.minus.gdpo-w0.5.olympiad-bench-abs}{6.2}
\setval{olmo.gdpo-noaux.minus.gdpo-w0.5.writing}{$-$9.6}
\setval{olmo.gdpo-noaux.minus.gdpo-w0.5.writing-abs}{9.6}
\setval{olmo.gdpo-noaux.minus.gdpo-w0.5.writingbench}{$-$11.0}
\setval{olmo.gdpo-noaux.minus.gdpo-w0.5.writingbench-abs}{11.0}
\setval{olmo.gdpo-noaux.minus.gdpo.aime24}{+10.5}
\setval{olmo.gdpo-noaux.minus.gdpo.aime24-abs}{10.5}
\setval{olmo.gdpo-noaux.minus.gdpo.aime25}{+11.0}
\setval{olmo.gdpo-noaux.minus.gdpo.aime25-abs}{11.0}
\setval{olmo.gdpo-noaux.minus.gdpo.amc23}{+14.4}
\setval{olmo.gdpo-noaux.minus.gdpo.amc23-abs}{14.4}
\setval{olmo.gdpo-noaux.minus.gdpo.avg}{+3.9}
\setval{olmo.gdpo-noaux.minus.gdpo.avg-abs}{3.9}
\setval{olmo.gdpo-noaux.minus.gdpo.code}{+7.7}
\setval{olmo.gdpo-noaux.minus.gdpo.code-abs}{7.7}
\setval{olmo.gdpo-noaux.minus.gdpo.creative-writing-v3}{$-$8.9}
\setval{olmo.gdpo-noaux.minus.gdpo.creative-writing-v3-abs}{8.9}
\setval{olmo.gdpo-noaux.minus.gdpo.guidebench}{+4.9}
\setval{olmo.gdpo-noaux.minus.gdpo.guidebench-abs}{4.9}
\setval{olmo.gdpo-noaux.minus.gdpo.humaneval-plus}{+8.8}
\setval{olmo.gdpo-noaux.minus.gdpo.humaneval-plus-abs}{8.8}
\setval{olmo.gdpo-noaux.minus.gdpo.if}{+2.3}
\setval{olmo.gdpo-noaux.minus.gdpo.if-abs}{2.3}
\setval{olmo.gdpo-noaux.minus.gdpo.ifbench}{+3.7}
\setval{olmo.gdpo-noaux.minus.gdpo.ifbench-abs}{3.7}
\setval{olmo.gdpo-noaux.minus.gdpo.ifeval}{$-$1.7}
\setval{olmo.gdpo-noaux.minus.gdpo.ifeval-abs}{1.7}
\setval{olmo.gdpo-noaux.minus.gdpo.judge2.avg}{+4.1}
\setval{olmo.gdpo-noaux.minus.gdpo.judge2.avg-abs}{4.1}
\setval{olmo.gdpo-noaux.minus.gdpo.judge2.creative-writing-v3}{$-$6.9}
\setval{olmo.gdpo-noaux.minus.gdpo.judge2.creative-writing-v3-abs}{6.9}
\setval{olmo.gdpo-noaux.minus.gdpo.judge2.longbench-write}{$-$9.0}
\setval{olmo.gdpo-noaux.minus.gdpo.judge2.longbench-write-abs}{9.0}
\setval{olmo.gdpo-noaux.minus.gdpo.judge2.writing}{$-$7.7}
\setval{olmo.gdpo-noaux.minus.gdpo.judge2.writing-abs}{7.7}
\setval{olmo.gdpo-noaux.minus.gdpo.judge2.writingbench}{$-$7.3}
\setval{olmo.gdpo-noaux.minus.gdpo.judge2.writingbench-abs}{7.3}
\setval{olmo.gdpo-noaux.minus.gdpo.livecodebench}{+8.4}
\setval{olmo.gdpo-noaux.minus.gdpo.livecodebench-abs}{8.4}
\setval{olmo.gdpo-noaux.minus.gdpo.longbench-write}{$-$7.3}
\setval{olmo.gdpo-noaux.minus.gdpo.longbench-write-abs}{7.3}
\setval{olmo.gdpo-noaux.minus.gdpo.math}{+9.2}
\setval{olmo.gdpo-noaux.minus.gdpo.math-abs}{9.2}
\setval{olmo.gdpo-noaux.minus.gdpo.math500}{+5.6}
\setval{olmo.gdpo-noaux.minus.gdpo.math500-abs}{5.6}
\setval{olmo.gdpo-noaux.minus.gdpo.mbpp-plus}{+5.7}
\setval{olmo.gdpo-noaux.minus.gdpo.mbpp-plus-abs}{5.7}
\setval{olmo.gdpo-noaux.minus.gdpo.minerva}{+5.8}
\setval{olmo.gdpo-noaux.minus.gdpo.minerva-abs}{5.8}
\setval{olmo.gdpo-noaux.minus.gdpo.olympiad-bench}{+7.8}
\setval{olmo.gdpo-noaux.minus.gdpo.olympiad-bench-abs}{7.8}
\setval{olmo.gdpo-noaux.minus.gdpo.writing}{$-$8.7}
\setval{olmo.gdpo-noaux.minus.gdpo.writing-abs}{8.7}
\setval{olmo.gdpo-noaux.minus.gdpo.writingbench}{$-$9.8}
\setval{olmo.gdpo-noaux.minus.gdpo.writingbench-abs}{9.8}
\setval{olmo.gdpo-noaux.minus.geoma.aime24}{+14.6}
\setval{olmo.gdpo-noaux.minus.geoma.aime24-abs}{14.6}
\setval{olmo.gdpo-noaux.minus.geoma.aime25}{+11.7}
\setval{olmo.gdpo-noaux.minus.geoma.aime25-abs}{11.7}
\setval{olmo.gdpo-noaux.minus.geoma.amc23}{+20.0}
\setval{olmo.gdpo-noaux.minus.geoma.amc23-abs}{20.0}
\setval{olmo.gdpo-noaux.minus.geoma.avg}{+4.9}
\setval{olmo.gdpo-noaux.minus.geoma.avg-abs}{4.9}
\setval{olmo.gdpo-noaux.minus.geoma.code}{+8.5}
\setval{olmo.gdpo-noaux.minus.geoma.code-abs}{8.5}
\setval{olmo.gdpo-noaux.minus.geoma.creative-writing-v3}{$-$11.4}
\setval{olmo.gdpo-noaux.minus.geoma.creative-writing-v3-abs}{11.4}
\setval{olmo.gdpo-noaux.minus.geoma.guidebench}{+9.1}
\setval{olmo.gdpo-noaux.minus.geoma.guidebench-abs}{9.1}
\setval{olmo.gdpo-noaux.minus.geoma.humaneval-plus}{+10.5}
\setval{olmo.gdpo-noaux.minus.geoma.humaneval-plus-abs}{10.5}
\setval{olmo.gdpo-noaux.minus.geoma.if}{+3.0}
\setval{olmo.gdpo-noaux.minus.geoma.if-abs}{3.0}
\setval{olmo.gdpo-noaux.minus.geoma.ifbench}{+3.0}
\setval{olmo.gdpo-noaux.minus.geoma.ifbench-abs}{3.0}
\setval{olmo.gdpo-noaux.minus.geoma.ifeval}{$-$3.2}
\setval{olmo.gdpo-noaux.minus.geoma.ifeval-abs}{3.2}
\setval{olmo.gdpo-noaux.minus.geoma.judge2.avg}{+5.0}
\setval{olmo.gdpo-noaux.minus.geoma.judge2.avg-abs}{5.0}
\setval{olmo.gdpo-noaux.minus.geoma.judge2.creative-writing-v3}{$-$10.8}
\setval{olmo.gdpo-noaux.minus.geoma.judge2.creative-writing-v3-abs}{10.8}
\setval{olmo.gdpo-noaux.minus.geoma.judge2.longbench-write}{$-$11.6}
\setval{olmo.gdpo-noaux.minus.geoma.judge2.longbench-write-abs}{11.6}
\setval{olmo.gdpo-noaux.minus.geoma.judge2.writing}{$-$10.4}
\setval{olmo.gdpo-noaux.minus.geoma.judge2.writing-abs}{10.4}
\setval{olmo.gdpo-noaux.minus.geoma.judge2.writingbench}{$-$8.9}
\setval{olmo.gdpo-noaux.minus.geoma.judge2.writingbench-abs}{8.9}
\setval{olmo.gdpo-noaux.minus.geoma.livecodebench}{+9.2}
\setval{olmo.gdpo-noaux.minus.geoma.livecodebench-abs}{9.2}
\setval{olmo.gdpo-noaux.minus.geoma.longbench-write}{$-$9.3}
\setval{olmo.gdpo-noaux.minus.geoma.longbench-write-abs}{9.3}
\setval{olmo.gdpo-noaux.minus.geoma.math}{+11.9}
\setval{olmo.gdpo-noaux.minus.geoma.math-abs}{11.9}
\setval{olmo.gdpo-noaux.minus.geoma.math500}{+7.7}
\setval{olmo.gdpo-noaux.minus.geoma.math500-abs}{7.7}
\setval{olmo.gdpo-noaux.minus.geoma.mbpp-plus}{+5.8}
\setval{olmo.gdpo-noaux.minus.geoma.mbpp-plus-abs}{5.8}
\setval{olmo.gdpo-noaux.minus.geoma.minerva}{+6.2}
\setval{olmo.gdpo-noaux.minus.geoma.minerva-abs}{6.2}
\setval{olmo.gdpo-noaux.minus.geoma.olympiad-bench}{+11.0}
\setval{olmo.gdpo-noaux.minus.geoma.olympiad-bench-abs}{11.0}
\setval{olmo.gdpo-noaux.minus.geoma.writing}{$-$10.7}
\setval{olmo.gdpo-noaux.minus.geoma.writing-abs}{10.7}
\setval{olmo.gdpo-noaux.minus.geoma.writingbench}{$-$11.5}
\setval{olmo.gdpo-noaux.minus.geoma.writingbench-abs}{11.5}
\setval{olmo.gdpo-noaux.minus.grpo-nodiv.aime24}{+6.6}
\setval{olmo.gdpo-noaux.minus.grpo-nodiv.aime24-abs}{6.6}
\setval{olmo.gdpo-noaux.minus.grpo-nodiv.aime25}{+6.0}
\setval{olmo.gdpo-noaux.minus.grpo-nodiv.aime25-abs}{6.0}
\setval{olmo.gdpo-noaux.minus.grpo-nodiv.amc23}{+10.9}
\setval{olmo.gdpo-noaux.minus.grpo-nodiv.amc23-abs}{10.9}
\setval{olmo.gdpo-noaux.minus.grpo-nodiv.avg}{+1.9}
\setval{olmo.gdpo-noaux.minus.grpo-nodiv.avg-abs}{1.9}
\setval{olmo.gdpo-noaux.minus.grpo-nodiv.code}{+7.0}
\setval{olmo.gdpo-noaux.minus.grpo-nodiv.code-abs}{7.0}
\setval{olmo.gdpo-noaux.minus.grpo-nodiv.creative-writing-v3}{$-$8.8}
\setval{olmo.gdpo-noaux.minus.grpo-nodiv.creative-writing-v3-abs}{8.8}
\setval{olmo.gdpo-noaux.minus.grpo-nodiv.guidebench}{+0.3}
\setval{olmo.gdpo-noaux.minus.grpo-nodiv.guidebench-abs}{0.3}
\setval{olmo.gdpo-noaux.minus.grpo-nodiv.humaneval-plus}{+7.8}
\setval{olmo.gdpo-noaux.minus.grpo-nodiv.humaneval-plus-abs}{7.8}
\setval{olmo.gdpo-noaux.minus.grpo-nodiv.if}{$-$0.2}
\setval{olmo.gdpo-noaux.minus.grpo-nodiv.if-abs}{0.2}
\setval{olmo.gdpo-noaux.minus.grpo-nodiv.ifbench}{+2.0}
\setval{olmo.gdpo-noaux.minus.grpo-nodiv.ifbench-abs}{2.0}
\setval{olmo.gdpo-noaux.minus.grpo-nodiv.ifeval}{$-$3.0}
\setval{olmo.gdpo-noaux.minus.grpo-nodiv.ifeval-abs}{3.0}
\setval{olmo.gdpo-noaux.minus.grpo-nodiv.judge2.avg}{+1.8}
\setval{olmo.gdpo-noaux.minus.grpo-nodiv.judge2.avg-abs}{1.8}
\setval{olmo.gdpo-noaux.minus.grpo-nodiv.judge2.creative-writing-v3}{$-$7.9}
\setval{olmo.gdpo-noaux.minus.grpo-nodiv.judge2.creative-writing-v3-abs}{7.9}
\setval{olmo.gdpo-noaux.minus.grpo-nodiv.judge2.longbench-write}{$-$14.5}
\setval{olmo.gdpo-noaux.minus.grpo-nodiv.judge2.longbench-write-abs}{14.5}
\setval{olmo.gdpo-noaux.minus.grpo-nodiv.judge2.writing}{$-$9.9}
\setval{olmo.gdpo-noaux.minus.grpo-nodiv.judge2.writing-abs}{9.9}
\setval{olmo.gdpo-noaux.minus.grpo-nodiv.judge2.writingbench}{$-$7.2}
\setval{olmo.gdpo-noaux.minus.grpo-nodiv.judge2.writingbench-abs}{7.2}
\setval{olmo.gdpo-noaux.minus.grpo-nodiv.livecodebench}{+7.9}
\setval{olmo.gdpo-noaux.minus.grpo-nodiv.livecodebench-abs}{7.9}
\setval{olmo.gdpo-noaux.minus.grpo-nodiv.longbench-write}{$-$11.6}
\setval{olmo.gdpo-noaux.minus.grpo-nodiv.longbench-write-abs}{11.6}
\setval{olmo.gdpo-noaux.minus.grpo-nodiv.math}{+6.1}
\setval{olmo.gdpo-noaux.minus.grpo-nodiv.math-abs}{6.1}
\setval{olmo.gdpo-noaux.minus.grpo-nodiv.math500}{+3.4}
\setval{olmo.gdpo-noaux.minus.grpo-nodiv.math500-abs}{3.4}
\setval{olmo.gdpo-noaux.minus.grpo-nodiv.mbpp-plus}{+5.2}
\setval{olmo.gdpo-noaux.minus.grpo-nodiv.mbpp-plus-abs}{5.2}
\setval{olmo.gdpo-noaux.minus.grpo-nodiv.minerva}{+4.3}
\setval{olmo.gdpo-noaux.minus.grpo-nodiv.minerva-abs}{4.3}
\setval{olmo.gdpo-noaux.minus.grpo-nodiv.olympiad-bench}{+5.3}
\setval{olmo.gdpo-noaux.minus.grpo-nodiv.olympiad-bench-abs}{5.3}
\setval{olmo.gdpo-noaux.minus.grpo-nodiv.writing}{$-$9.3}
\setval{olmo.gdpo-noaux.minus.grpo-nodiv.writing-abs}{9.3}
\setval{olmo.gdpo-noaux.minus.grpo-nodiv.writingbench}{$-$7.7}
\setval{olmo.gdpo-noaux.minus.grpo-nodiv.writingbench-abs}{7.7}
\setval{olmo.gdpo-noaux.minus.grpo.aime24}{+8.9}
\setval{olmo.gdpo-noaux.minus.grpo.aime24-abs}{8.9}
\setval{olmo.gdpo-noaux.minus.grpo.aime25}{+9.0}
\setval{olmo.gdpo-noaux.minus.grpo.aime25-abs}{9.0}
\setval{olmo.gdpo-noaux.minus.grpo.amc23}{+13.0}
\setval{olmo.gdpo-noaux.minus.grpo.amc23-abs}{13.0}
\setval{olmo.gdpo-noaux.minus.grpo.avg}{+2.6}
\setval{olmo.gdpo-noaux.minus.grpo.avg-abs}{2.6}
\setval{olmo.gdpo-noaux.minus.grpo.code}{+7.3}
\setval{olmo.gdpo-noaux.minus.grpo.code-abs}{7.3}
\setval{olmo.gdpo-noaux.minus.grpo.creative-writing-v3}{$-$11.2}
\setval{olmo.gdpo-noaux.minus.grpo.creative-writing-v3-abs}{11.2}
\setval{olmo.gdpo-noaux.minus.grpo.guidebench}{$-$0.7}
\setval{olmo.gdpo-noaux.minus.grpo.guidebench-abs}{0.7}
\setval{olmo.gdpo-noaux.minus.grpo.humaneval-plus}{+7.9}
\setval{olmo.gdpo-noaux.minus.grpo.humaneval-plus-abs}{7.9}
\setval{olmo.gdpo-noaux.minus.grpo.if}{$-$0.7}
\setval{olmo.gdpo-noaux.minus.grpo.if-abs}{0.7}
\setval{olmo.gdpo-noaux.minus.grpo.ifbench}{+1.3}
\setval{olmo.gdpo-noaux.minus.grpo.ifbench-abs}{1.3}
\setval{olmo.gdpo-noaux.minus.grpo.ifeval}{$-$2.6}
\setval{olmo.gdpo-noaux.minus.grpo.ifeval-abs}{2.6}
\setval{olmo.gdpo-noaux.minus.grpo.judge2.avg}{+2.9}
\setval{olmo.gdpo-noaux.minus.grpo.judge2.avg-abs}{2.9}
\setval{olmo.gdpo-noaux.minus.grpo.judge2.creative-writing-v3}{$-$7.7}
\setval{olmo.gdpo-noaux.minus.grpo.judge2.creative-writing-v3-abs}{7.7}
\setval{olmo.gdpo-noaux.minus.grpo.judge2.longbench-write}{$-$10.5}
\setval{olmo.gdpo-noaux.minus.grpo.judge2.longbench-write-abs}{10.5}
\setval{olmo.gdpo-noaux.minus.grpo.judge2.writing}{$-$8.9}
\setval{olmo.gdpo-noaux.minus.grpo.judge2.writing-abs}{8.9}
\setval{olmo.gdpo-noaux.minus.grpo.judge2.writingbench}{$-$8.7}
\setval{olmo.gdpo-noaux.minus.grpo.judge2.writingbench-abs}{8.7}
\setval{olmo.gdpo-noaux.minus.grpo.livecodebench}{+9.2}
\setval{olmo.gdpo-noaux.minus.grpo.livecodebench-abs}{9.2}
\setval{olmo.gdpo-noaux.minus.grpo.longbench-write}{$-$8.2}
\setval{olmo.gdpo-noaux.minus.grpo.longbench-write-abs}{8.2}
\setval{olmo.gdpo-noaux.minus.grpo.math}{+8.5}
\setval{olmo.gdpo-noaux.minus.grpo.math-abs}{8.5}
\setval{olmo.gdpo-noaux.minus.grpo.math500}{+5.9}
\setval{olmo.gdpo-noaux.minus.grpo.math500-abs}{5.9}
\setval{olmo.gdpo-noaux.minus.grpo.mbpp-plus}{+4.7}
\setval{olmo.gdpo-noaux.minus.grpo.mbpp-plus-abs}{4.7}
\setval{olmo.gdpo-noaux.minus.grpo.minerva}{+6.2}
\setval{olmo.gdpo-noaux.minus.grpo.minerva-abs}{6.2}
\setval{olmo.gdpo-noaux.minus.grpo.olympiad-bench}{+8.1}
\setval{olmo.gdpo-noaux.minus.grpo.olympiad-bench-abs}{8.1}
\setval{olmo.gdpo-noaux.minus.grpo.writing}{$-$10.8}
\setval{olmo.gdpo-noaux.minus.grpo.writing-abs}{10.8}
\setval{olmo.gdpo-noaux.minus.grpo.writingbench}{$-$13.0}
\setval{olmo.gdpo-noaux.minus.grpo.writingbench-abs}{13.0}
\setval{olmo.gdpo-noaux.minus.mt-grpo.aime24}{+1.8}
\setval{olmo.gdpo-noaux.minus.mt-grpo.aime24-abs}{1.8}
\setval{olmo.gdpo-noaux.minus.mt-grpo.aime25}{+5.5}
\setval{olmo.gdpo-noaux.minus.mt-grpo.aime25-abs}{5.5}
\setval{olmo.gdpo-noaux.minus.mt-grpo.amc23}{+6.5}
\setval{olmo.gdpo-noaux.minus.mt-grpo.amc23-abs}{6.5}
\setval{olmo.gdpo-noaux.minus.mt-grpo.avg}{+1.1}
\setval{olmo.gdpo-noaux.minus.mt-grpo.avg-abs}{1.1}
\setval{olmo.gdpo-noaux.minus.mt-grpo.code}{+8.0}
\setval{olmo.gdpo-noaux.minus.mt-grpo.code-abs}{8.0}
\setval{olmo.gdpo-noaux.minus.mt-grpo.creative-writing-v3}{$-$12.5}
\setval{olmo.gdpo-noaux.minus.mt-grpo.creative-writing-v3-abs}{12.5}
\setval{olmo.gdpo-noaux.minus.mt-grpo.guidebench}{+1.2}
\setval{olmo.gdpo-noaux.minus.mt-grpo.guidebench-abs}{1.2}
\setval{olmo.gdpo-noaux.minus.mt-grpo.humaneval-plus}{+9.5}
\setval{olmo.gdpo-noaux.minus.mt-grpo.humaneval-plus-abs}{9.5}
\setval{olmo.gdpo-noaux.minus.mt-grpo.if}{+0.1}
\setval{olmo.gdpo-noaux.minus.mt-grpo.if-abs}{0.1}
\setval{olmo.gdpo-noaux.minus.mt-grpo.ifbench}{+1.3}
\setval{olmo.gdpo-noaux.minus.mt-grpo.ifbench-abs}{1.3}
\setval{olmo.gdpo-noaux.minus.mt-grpo.ifeval}{$-$2.3}
\setval{olmo.gdpo-noaux.minus.mt-grpo.ifeval-abs}{2.3}
\setval{olmo.gdpo-noaux.minus.mt-grpo.judge2.avg}{+1.4}
\setval{olmo.gdpo-noaux.minus.mt-grpo.judge2.avg-abs}{1.4}
\setval{olmo.gdpo-noaux.minus.mt-grpo.judge2.creative-writing-v3}{$-$8.9}
\setval{olmo.gdpo-noaux.minus.mt-grpo.judge2.creative-writing-v3-abs}{8.9}
\setval{olmo.gdpo-noaux.minus.mt-grpo.judge2.longbench-write}{$-$13.8}
\setval{olmo.gdpo-noaux.minus.mt-grpo.judge2.longbench-write-abs}{13.8}
\setval{olmo.gdpo-noaux.minus.mt-grpo.judge2.writing}{$-$10.2}
\setval{olmo.gdpo-noaux.minus.mt-grpo.judge2.writing-abs}{10.2}
\setval{olmo.gdpo-noaux.minus.mt-grpo.judge2.writingbench}{$-$7.8}
\setval{olmo.gdpo-noaux.minus.mt-grpo.judge2.writingbench-abs}{7.8}
\setval{olmo.gdpo-noaux.minus.mt-grpo.livecodebench}{+8.9}
\setval{olmo.gdpo-noaux.minus.mt-grpo.livecodebench-abs}{8.9}
\setval{olmo.gdpo-noaux.minus.mt-grpo.longbench-write}{$-$10.4}
\setval{olmo.gdpo-noaux.minus.mt-grpo.longbench-write-abs}{10.4}
\setval{olmo.gdpo-noaux.minus.mt-grpo.math}{+4.6}
\setval{olmo.gdpo-noaux.minus.mt-grpo.math-abs}{4.6}
\setval{olmo.gdpo-noaux.minus.mt-grpo.math500}{+4.0}
\setval{olmo.gdpo-noaux.minus.mt-grpo.math500-abs}{4.0}
\setval{olmo.gdpo-noaux.minus.mt-grpo.mbpp-plus}{+5.5}
\setval{olmo.gdpo-noaux.minus.mt-grpo.mbpp-plus-abs}{5.5}
\setval{olmo.gdpo-noaux.minus.mt-grpo.minerva}{+4.2}
\setval{olmo.gdpo-noaux.minus.mt-grpo.minerva-abs}{4.2}
\setval{olmo.gdpo-noaux.minus.mt-grpo.olympiad-bench}{+5.7}
\setval{olmo.gdpo-noaux.minus.mt-grpo.olympiad-bench-abs}{5.7}
\setval{olmo.gdpo-noaux.minus.mt-grpo.writing}{$-$11.7}
\setval{olmo.gdpo-noaux.minus.mt-grpo.writing-abs}{11.7}
\setval{olmo.gdpo-noaux.minus.mt-grpo.writingbench}{$-$12.2}
\setval{olmo.gdpo-noaux.minus.mt-grpo.writingbench-abs}{12.2}
\setval{olmo.gdpo-noaux.minus.papo.aime24}{+16.1}
\setval{olmo.gdpo-noaux.minus.papo.aime24-abs}{16.1}
\setval{olmo.gdpo-noaux.minus.papo.aime25}{+13.3}
\setval{olmo.gdpo-noaux.minus.papo.aime25-abs}{13.3}
\setval{olmo.gdpo-noaux.minus.papo.amc23}{+24.5}
\setval{olmo.gdpo-noaux.minus.papo.amc23-abs}{24.5}
\setval{olmo.gdpo-noaux.minus.papo.avg}{+5.5}
\setval{olmo.gdpo-noaux.minus.papo.avg-abs}{5.5}
\setval{olmo.gdpo-noaux.minus.papo.code}{+8.5}
\setval{olmo.gdpo-noaux.minus.papo.code-abs}{8.5}
\setval{olmo.gdpo-noaux.minus.papo.creative-writing-v3}{$-$8.8}
\setval{olmo.gdpo-noaux.minus.papo.creative-writing-v3-abs}{8.8}
\setval{olmo.gdpo-noaux.minus.papo.guidebench}{$-$2.9}
\setval{olmo.gdpo-noaux.minus.papo.guidebench-abs}{2.9}
\setval{olmo.gdpo-noaux.minus.papo.humaneval-plus}{+10.4}
\setval{olmo.gdpo-noaux.minus.papo.humaneval-plus-abs}{10.4}
\setval{olmo.gdpo-noaux.minus.papo.if}{$-$0.5}
\setval{olmo.gdpo-noaux.minus.papo.if-abs}{0.5}
\setval{olmo.gdpo-noaux.minus.papo.ifbench}{+3.7}
\setval{olmo.gdpo-noaux.minus.papo.ifbench-abs}{3.7}
\setval{olmo.gdpo-noaux.minus.papo.ifeval}{$-$2.4}
\setval{olmo.gdpo-noaux.minus.papo.ifeval-abs}{2.4}
\setval{olmo.gdpo-noaux.minus.papo.judge2.avg}{+5.7}
\setval{olmo.gdpo-noaux.minus.papo.judge2.avg-abs}{5.7}
\setval{olmo.gdpo-noaux.minus.papo.judge2.creative-writing-v3}{$-$7.0}
\setval{olmo.gdpo-noaux.minus.papo.judge2.creative-writing-v3-abs}{7.0}
\setval{olmo.gdpo-noaux.minus.papo.judge2.longbench-write}{$-$9.0}
\setval{olmo.gdpo-noaux.minus.papo.judge2.longbench-write-abs}{9.0}
\setval{olmo.gdpo-noaux.minus.papo.judge2.writing}{$-$7.9}
\setval{olmo.gdpo-noaux.minus.papo.judge2.writing-abs}{7.9}
\setval{olmo.gdpo-noaux.minus.papo.judge2.writingbench}{$-$7.7}
\setval{olmo.gdpo-noaux.minus.papo.judge2.writingbench-abs}{7.7}
\setval{olmo.gdpo-noaux.minus.papo.livecodebench}{+9.7}
\setval{olmo.gdpo-noaux.minus.papo.livecodebench-abs}{9.7}
\setval{olmo.gdpo-noaux.minus.papo.longbench-write}{$-$8.7}
\setval{olmo.gdpo-noaux.minus.papo.longbench-write-abs}{8.7}
\setval{olmo.gdpo-noaux.minus.papo.math}{+14.3}
\setval{olmo.gdpo-noaux.minus.papo.math-abs}{14.3}
\setval{olmo.gdpo-noaux.minus.papo.math500}{+10.0}
\setval{olmo.gdpo-noaux.minus.papo.math500-abs}{10.0}
\setval{olmo.gdpo-noaux.minus.papo.mbpp-plus}{+5.4}
\setval{olmo.gdpo-noaux.minus.papo.mbpp-plus-abs}{5.4}
\setval{olmo.gdpo-noaux.minus.papo.minerva}{+7.6}
\setval{olmo.gdpo-noaux.minus.papo.minerva-abs}{7.6}
\setval{olmo.gdpo-noaux.minus.papo.olympiad-bench}{+14.0}
\setval{olmo.gdpo-noaux.minus.papo.olympiad-bench-abs}{14.0}
\setval{olmo.gdpo-noaux.minus.papo.writing}{$-$9.1}
\setval{olmo.gdpo-noaux.minus.papo.writing-abs}{9.1}
\setval{olmo.gdpo-noaux.minus.papo.writingbench}{$-$9.9}
\setval{olmo.gdpo-noaux.minus.papo.writingbench-abs}{9.9}
\setval{olmo.gdpo-noaux.minus.saw.aime24}{+12.2}
\setval{olmo.gdpo-noaux.minus.saw.aime24-abs}{12.2}
\setval{olmo.gdpo-noaux.minus.saw.aime25}{+8.4}
\setval{olmo.gdpo-noaux.minus.saw.aime25-abs}{8.4}
\setval{olmo.gdpo-noaux.minus.saw.amc23}{+17.0}
\setval{olmo.gdpo-noaux.minus.saw.amc23-abs}{17.0}
\setval{olmo.gdpo-noaux.minus.saw.avg}{+3.4}
\setval{olmo.gdpo-noaux.minus.saw.avg-abs}{3.4}
\setval{olmo.gdpo-noaux.minus.saw.code}{+7.9}
\setval{olmo.gdpo-noaux.minus.saw.code-abs}{7.9}
\setval{olmo.gdpo-noaux.minus.saw.creative-writing-v3}{$-$9.7}
\setval{olmo.gdpo-noaux.minus.saw.creative-writing-v3-abs}{9.7}
\setval{olmo.gdpo-noaux.minus.saw.guidebench}{+1.1}
\setval{olmo.gdpo-noaux.minus.saw.guidebench-abs}{1.1}
\setval{olmo.gdpo-noaux.minus.saw.humaneval-plus}{+8.8}
\setval{olmo.gdpo-noaux.minus.saw.humaneval-plus-abs}{8.8}
\setval{olmo.gdpo-noaux.minus.saw.if}{$-$0.7}
\setval{olmo.gdpo-noaux.minus.saw.if-abs}{0.7}
\setval{olmo.gdpo-noaux.minus.saw.ifbench}{+0.7}
\setval{olmo.gdpo-noaux.minus.saw.ifbench-abs}{0.7}
\setval{olmo.gdpo-noaux.minus.saw.ifeval}{$-$3.8}
\setval{olmo.gdpo-noaux.minus.saw.ifeval-abs}{3.8}
\setval{olmo.gdpo-noaux.minus.saw.judge2.avg}{+3.5}
\setval{olmo.gdpo-noaux.minus.saw.judge2.avg-abs}{3.5}
\setval{olmo.gdpo-noaux.minus.saw.judge2.creative-writing-v3}{$-$10.0}
\setval{olmo.gdpo-noaux.minus.saw.judge2.creative-writing-v3-abs}{10.0}
\setval{olmo.gdpo-noaux.minus.saw.judge2.longbench-write}{$-$10.8}
\setval{olmo.gdpo-noaux.minus.saw.judge2.longbench-write-abs}{10.8}
\setval{olmo.gdpo-noaux.minus.saw.judge2.writing}{$-$9.8}
\setval{olmo.gdpo-noaux.minus.saw.judge2.writing-abs}{9.8}
\setval{olmo.gdpo-noaux.minus.saw.judge2.writingbench}{$-$8.7}
\setval{olmo.gdpo-noaux.minus.saw.judge2.writingbench-abs}{8.7}
\setval{olmo.gdpo-noaux.minus.saw.livecodebench}{+9.7}
\setval{olmo.gdpo-noaux.minus.saw.livecodebench-abs}{9.7}
\setval{olmo.gdpo-noaux.minus.saw.longbench-write}{$-$8.9}
\setval{olmo.gdpo-noaux.minus.saw.longbench-write-abs}{8.9}
\setval{olmo.gdpo-noaux.minus.saw.math}{+10.0}
\setval{olmo.gdpo-noaux.minus.saw.math-abs}{10.0}
\setval{olmo.gdpo-noaux.minus.saw.math500}{+6.4}
\setval{olmo.gdpo-noaux.minus.saw.math500-abs}{6.4}
\setval{olmo.gdpo-noaux.minus.saw.mbpp-plus}{+5.1}
\setval{olmo.gdpo-noaux.minus.saw.mbpp-plus-abs}{5.1}
\setval{olmo.gdpo-noaux.minus.saw.minerva}{+5.8}
\setval{olmo.gdpo-noaux.minus.saw.minerva-abs}{5.8}
\setval{olmo.gdpo-noaux.minus.saw.olympiad-bench}{+9.9}
\setval{olmo.gdpo-noaux.minus.saw.olympiad-bench-abs}{9.9}
\setval{olmo.gdpo-noaux.minus.saw.writing}{$-$10.2}
\setval{olmo.gdpo-noaux.minus.saw.writing-abs}{10.2}
\setval{olmo.gdpo-noaux.minus.saw.writingbench}{$-$11.9}
\setval{olmo.gdpo-noaux.minus.saw.writingbench-abs}{11.9}
\setval{olmo.gdpo-noaux.passing-at-ceiling.code.early}{0.84}
\setval{olmo.gdpo-noaux.passing-at-ceiling.code.late}{0.89}
\setval{olmo.gdpo-noaux.passing-at-ceiling.code.run}{0.88}
\setval{olmo.gdpo-noaux.passing-at-ceiling.if.early}{0.48}
\setval{olmo.gdpo-noaux.passing-at-ceiling.if.late}{0.51}
\setval{olmo.gdpo-noaux.passing-at-ceiling.if.run}{0.48}
\setval{olmo.gdpo-noaux.passing-at-ceiling.math.early}{1.00}
\setval{olmo.gdpo-noaux.passing-at-ceiling.math.late}{1.00}
\setval{olmo.gdpo-noaux.passing-at-ceiling.math.run}{1.00}
\setval{olmo.gdpo-noaux.passing-at-ceiling.writing.early}{0.00}
\setval{olmo.gdpo-noaux.passing-at-ceiling.writing.late}{0.01}
\setval{olmo.gdpo-noaux.passing-at-ceiling.writing.run}{0.00}
\setval{olmo.gdpo-noaux.primary-cosine.code.early}{1.00}
\setval{olmo.gdpo-noaux.primary-cosine.code.late}{1.00}
\setval{olmo.gdpo-noaux.primary-cosine.code.run}{1.00}
\setval{olmo.gdpo-noaux.primary-cosine.if.early}{0.76}
\setval{olmo.gdpo-noaux.primary-cosine.if.late}{0.81}
\setval{olmo.gdpo-noaux.primary-cosine.if.run}{0.79}
\setval{olmo.gdpo-noaux.primary-cosine.math.early}{1.00}
\setval{olmo.gdpo-noaux.primary-cosine.math.late}{1.00}
\setval{olmo.gdpo-noaux.primary-cosine.math.run}{1.00}
\setval{olmo.gdpo-noaux.primary-cosine.writing.early}{1.00}
\setval{olmo.gdpo-noaux.primary-cosine.writing.late}{1.00}
\setval{olmo.gdpo-noaux.primary-cosine.writing.run}{1.00}
\setval{olmo.gdpo-noaux.primary-level.code.early}{0.59}
\setval{olmo.gdpo-noaux.primary-level.code.late}{0.87}
\setval{olmo.gdpo-noaux.primary-level.code.run}{0.80}
\setval{olmo.gdpo-noaux.primary-level.if.early}{0.63}
\setval{olmo.gdpo-noaux.primary-level.if.late}{0.68}
\setval{olmo.gdpo-noaux.primary-level.if.run}{0.65}
\setval{olmo.gdpo-noaux.primary-level.math.early}{0.35}
\setval{olmo.gdpo-noaux.primary-level.math.late}{0.74}
\setval{olmo.gdpo-noaux.primary-level.math.run}{0.65}
\setval{olmo.gdpo-noaux.primary-level.writing.early}{0.49}
\setval{olmo.gdpo-noaux.primary-level.writing.late}{0.74}
\setval{olmo.gdpo-noaux.primary-level.writing.run}{0.68}
\setval{olmo.gdpo-noaux.solved.code.early}{0.18}
\setval{olmo.gdpo-noaux.solved.code.late}{0.12}
\setval{olmo.gdpo-noaux.solved.code.run}{0.23}
\setval{olmo.gdpo-noaux.solved.if.early}{0.17}
\setval{olmo.gdpo-noaux.solved.if.late}{0.25}
\setval{olmo.gdpo-noaux.solved.if.run}{0.21}
\setval{olmo.gdpo-noaux.solved.math.early}{0.06}
\setval{olmo.gdpo-noaux.solved.math.late}{0.29}
\setval{olmo.gdpo-noaux.solved.math.run}{0.26}
\setval{olmo.gdpo-noaux.solved.writing.early}{0.00}
\setval{olmo.gdpo-noaux.solved.writing.late}{0.00}
\setval{olmo.gdpo-noaux.solved.writing.run}{0.00}
\setval{olmo.gdpo-noaux.update-share.code.early}{0.20}
\setval{olmo.gdpo-noaux.update-share.code.late}{0.15}
\setval{olmo.gdpo-noaux.update-share.code.run}{0.17}
\setval{olmo.gdpo-noaux.update-share.if.early}{0.33}
\setval{olmo.gdpo-noaux.update-share.if.late}{0.32}
\setval{olmo.gdpo-noaux.update-share.if.run}{0.31}
\setval{olmo.gdpo-noaux.update-share.math.early}{0.20}
\setval{olmo.gdpo-noaux.update-share.math.late}{0.16}
\setval{olmo.gdpo-noaux.update-share.math.run}{0.18}
\setval{olmo.gdpo-noaux.update-share.writing.early}{0.27}
\setval{olmo.gdpo-noaux.update-share.writing.late}{0.37}
\setval{olmo.gdpo-noaux.update-share.writing.run}{0.34}
\setval{olmo.gdpo-w0.5.aux-change.code}{+0.09}
\setval{olmo.gdpo-w0.5.aux-change.if}{+0.11}
\setval{olmo.gdpo-w0.5.aux-change.math}{$-$0.20}
\setval{olmo.gdpo-w0.5.aux-change.writing}{+0.00}
\setval{olmo.gdpo-w0.5.aux-change.writing-rm}{+0.26}
\setval{olmo.gdpo-w0.5.aux-full.code.early}{0.73}
\setval{olmo.gdpo-w0.5.aux-full.code.late}{0.82}
\setval{olmo.gdpo-w0.5.aux-full.code.run}{0.81}
\setval{olmo.gdpo-w0.5.aux-full.if.early}{0.59}
\setval{olmo.gdpo-w0.5.aux-full.if.late}{0.93}
\setval{olmo.gdpo-w0.5.aux-full.if.run}{0.82}
\setval{olmo.gdpo-w0.5.aux-full.math.early}{0.53}
\setval{olmo.gdpo-w0.5.aux-full.math.late}{0.45}
\setval{olmo.gdpo-w0.5.aux-full.math.run}{0.36}
\setval{olmo.gdpo-w0.5.aux-full.writing-rm.early}{0.02}
\setval{olmo.gdpo-w0.5.aux-full.writing-rm.late}{0.17}
\setval{olmo.gdpo-w0.5.aux-full.writing-rm.run}{0.07}
\setval{olmo.gdpo-w0.5.aux-full.writing.early}{0.55}
\setval{olmo.gdpo-w0.5.aux-full.writing.late}{0.66}
\setval{olmo.gdpo-w0.5.aux-full.writing.run}{0.49}
\setval{olmo.gdpo-w0.5.aux-level.code.early}{0.89}
\setval{olmo.gdpo-w0.5.aux-level.code.late}{0.90}
\setval{olmo.gdpo-w0.5.aux-level.code.run}{0.91}
\setval{olmo.gdpo-w0.5.aux-level.if.early}{0.90}
\setval{olmo.gdpo-w0.5.aux-level.if.late}{0.99}
\setval{olmo.gdpo-w0.5.aux-level.if.run}{0.96}
\setval{olmo.gdpo-w0.5.aux-level.math.early}{0.91}
\setval{olmo.gdpo-w0.5.aux-level.math.late}{0.70}
\setval{olmo.gdpo-w0.5.aux-level.math.run}{0.68}
\setval{olmo.gdpo-w0.5.aux-level.writing-rm.early}{0.60}
\setval{olmo.gdpo-w0.5.aux-level.writing-rm.late}{0.83}
\setval{olmo.gdpo-w0.5.aux-level.writing-rm.run}{0.74}
\setval{olmo.gdpo-w0.5.aux-level.writing.early}{0.95}
\setval{olmo.gdpo-w0.5.aux-level.writing.late}{0.95}
\setval{olmo.gdpo-w0.5.aux-level.writing.run}{0.91}
\setval{olmo.gdpo-w0.5.bench.aime24}{29.6}
\setval{olmo.gdpo-w0.5.bench.aime25}{27.3}
\setval{olmo.gdpo-w0.5.bench.amc23}{74.0}
\setval{olmo.gdpo-w0.5.bench.creative-writing-v3}{48.1}
\setval{olmo.gdpo-w0.5.bench.guidebench}{51.9}
\setval{olmo.gdpo-w0.5.bench.humaneval-plus}{77.7}
\setval{olmo.gdpo-w0.5.bench.ifbench}{32.7}
\setval{olmo.gdpo-w0.5.bench.ifeval}{87.5}
\setval{olmo.gdpo-w0.5.bench.livecodebench}{24.8}
\setval{olmo.gdpo-w0.5.bench.longbench-write}{63.7}
\setval{olmo.gdpo-w0.5.bench.math500}{87.0}
\setval{olmo.gdpo-w0.5.bench.mbpp-plus}{64.9}
\setval{olmo.gdpo-w0.5.bench.minerva}{41.5}
\setval{olmo.gdpo-w0.5.bench.olympiad-bench}{57.5}
\setval{olmo.gdpo-w0.5.bench.writingbench}{66.5}
\setval{olmo.gdpo-w0.5.eval.avg}{55.7}
\setval{olmo.gdpo-w0.5.eval.code}{55.8}
\setval{olmo.gdpo-w0.5.eval.if}{57.4}
\setval{olmo.gdpo-w0.5.eval.math}{52.8}
\setval{olmo.gdpo-w0.5.eval.writing}{59.4}
\setval{olmo.gdpo-w0.5.gate.code.against.early}{0.06}
\setval{olmo.gdpo-w0.5.gate.code.against.late}{0.15}
\setval{olmo.gdpo-w0.5.gate.code.against.run}{0.13}
\setval{olmo.gdpo-w0.5.gate.code.agreement}{$-$0.033}
\setval{olmo.gdpo-w0.5.gate.code.ceiling.early}{0.25}
\setval{olmo.gdpo-w0.5.gate.code.ceiling.late}{0.35}
\setval{olmo.gdpo-w0.5.gate.code.ceiling.run}{0.29}
\setval{olmo.gdpo-w0.5.gate.code.cosine.early}{0.72}
\setval{olmo.gdpo-w0.5.gate.code.cosine.late}{0.44}
\setval{olmo.gdpo-w0.5.gate.code.cosine.run}{0.55}
\setval{olmo.gdpo-w0.5.gate.if.against.early}{0.10}
\setval{olmo.gdpo-w0.5.gate.if.against.late}{0.05}
\setval{olmo.gdpo-w0.5.gate.if.against.run}{0.07}
\setval{olmo.gdpo-w0.5.gate.if.agreement}{+0.051}
\setval{olmo.gdpo-w0.5.gate.if.ceiling.early}{0.22}
\setval{olmo.gdpo-w0.5.gate.if.ceiling.late}{0.17}
\setval{olmo.gdpo-w0.5.gate.if.ceiling.run}{0.21}
\setval{olmo.gdpo-w0.5.gate.if.cosine.early}{0.51}
\setval{olmo.gdpo-w0.5.gate.if.cosine.late}{0.52}
\setval{olmo.gdpo-w0.5.gate.if.cosine.run}{0.52}
\setval{olmo.gdpo-w0.5.gate.math.against.early}{0.00}
\setval{olmo.gdpo-w0.5.gate.math.against.late}{0.00}
\setval{olmo.gdpo-w0.5.gate.math.against.run}{0.00}
\setval{olmo.gdpo-w0.5.gate.math.ceiling.early}{0.14}
\setval{olmo.gdpo-w0.5.gate.math.ceiling.late}{0.60}
\setval{olmo.gdpo-w0.5.gate.math.ceiling.run}{0.52}
\setval{olmo.gdpo-w0.5.gate.math.cosine.early}{0.91}
\setval{olmo.gdpo-w0.5.gate.math.cosine.late}{0.52}
\setval{olmo.gdpo-w0.5.gate.math.cosine.run}{0.58}
\setval{olmo.gdpo-w0.5.gate.writing-rm.against.early}{0.22}
\setval{olmo.gdpo-w0.5.gate.writing-rm.against.late}{0.51}
\setval{olmo.gdpo-w0.5.gate.writing-rm.against.run}{0.44}
\setval{olmo.gdpo-w0.5.gate.writing-rm.agreement}{+0.094}
\setval{olmo.gdpo-w0.5.gate.writing-rm.ceiling.early}{0.00}
\setval{olmo.gdpo-w0.5.gate.writing-rm.ceiling.late}{0.00}
\setval{olmo.gdpo-w0.5.gate.writing-rm.ceiling.run}{0.00}
\setval{olmo.gdpo-w0.5.gate.writing-rm.cosine.early}{0.69}
\setval{olmo.gdpo-w0.5.gate.writing-rm.cosine.late}{0.24}
\setval{olmo.gdpo-w0.5.gate.writing-rm.cosine.run}{0.36}
\setval{olmo.gdpo-w0.5.gate.writing.against.early}{0.25}
\setval{olmo.gdpo-w0.5.gate.writing.against.late}{0.58}
\setval{olmo.gdpo-w0.5.gate.writing.against.run}{0.54}
\setval{olmo.gdpo-w0.5.gate.writing.agreement}{$-$0.076}
\setval{olmo.gdpo-w0.5.gate.writing.ceiling.early}{0.00}
\setval{olmo.gdpo-w0.5.gate.writing.ceiling.late}{0.00}
\setval{olmo.gdpo-w0.5.gate.writing.ceiling.run}{0.00}
\setval{olmo.gdpo-w0.5.gate.writing.cosine.early}{0.59}
\setval{olmo.gdpo-w0.5.gate.writing.cosine.late}{0.11}
\setval{olmo.gdpo-w0.5.gate.writing.cosine.run}{0.19}
\setval{olmo.gdpo-w0.5.judge2.avg}{53.3}
\setval{olmo.gdpo-w0.5.judge2.creative-writing-v3}{39.4}
\setval{olmo.gdpo-w0.5.judge2.longbench-write}{56.6}
\setval{olmo.gdpo-w0.5.judge2.writing}{47.6}
\setval{olmo.gdpo-w0.5.judge2.writingbench}{46.7}
\setval{olmo.gdpo-w0.5.last-step}{300}
\setval{olmo.gdpo-w0.5.minus.base.aime24}{+21.0}
\setval{olmo.gdpo-w0.5.minus.base.aime24-abs}{21.0}
\setval{olmo.gdpo-w0.5.minus.base.aime25}{+19.9}
\setval{olmo.gdpo-w0.5.minus.base.aime25-abs}{19.9}
\setval{olmo.gdpo-w0.5.minus.base.amc23}{+29.5}
\setval{olmo.gdpo-w0.5.minus.base.amc23-abs}{29.5}
\setval{olmo.gdpo-w0.5.minus.base.avg}{+11.5}
\setval{olmo.gdpo-w0.5.minus.base.avg-abs}{11.5}
\setval{olmo.gdpo-w0.5.minus.base.code}{+6.7}
\setval{olmo.gdpo-w0.5.minus.base.code-abs}{6.7}
\setval{olmo.gdpo-w0.5.minus.base.creative-writing-v3}{+10.2}
\setval{olmo.gdpo-w0.5.minus.base.creative-writing-v3-abs}{10.2}
\setval{olmo.gdpo-w0.5.minus.base.guidebench}{$-$9.5}
\setval{olmo.gdpo-w0.5.minus.base.guidebench-abs}{9.5}
\setval{olmo.gdpo-w0.5.minus.base.humaneval-plus}{+8.5}
\setval{olmo.gdpo-w0.5.minus.base.humaneval-plus-abs}{8.5}
\setval{olmo.gdpo-w0.5.minus.base.if}{$-$0.1}
\setval{olmo.gdpo-w0.5.minus.base.if-abs}{0.1}
\setval{olmo.gdpo-w0.5.minus.base.ifbench}{+6.0}
\setval{olmo.gdpo-w0.5.minus.base.ifbench-abs}{6.0}
\setval{olmo.gdpo-w0.5.minus.base.ifeval}{+3.1}
\setval{olmo.gdpo-w0.5.minus.base.ifeval-abs}{3.1}
\setval{olmo.gdpo-w0.5.minus.base.judge2.avg}{+10.3}
\setval{olmo.gdpo-w0.5.minus.base.judge2.avg-abs}{10.3}
\setval{olmo.gdpo-w0.5.minus.base.judge2.creative-writing-v3}{+7.8}
\setval{olmo.gdpo-w0.5.minus.base.judge2.creative-writing-v3-abs}{7.8}
\setval{olmo.gdpo-w0.5.minus.base.judge2.longbench-write}{+3.4}
\setval{olmo.gdpo-w0.5.minus.base.judge2.longbench-write-abs}{3.4}
\setval{olmo.gdpo-w0.5.minus.base.judge2.writing}{+6.7}
\setval{olmo.gdpo-w0.5.minus.base.judge2.writing-abs}{6.7}
\setval{olmo.gdpo-w0.5.minus.base.judge2.writingbench}{+9.0}
\setval{olmo.gdpo-w0.5.minus.base.judge2.writingbench-abs}{9.0}
\setval{olmo.gdpo-w0.5.minus.base.livecodebench}{+7.9}
\setval{olmo.gdpo-w0.5.minus.base.livecodebench-abs}{7.9}
\setval{olmo.gdpo-w0.5.minus.base.longbench-write}{+6.3}
\setval{olmo.gdpo-w0.5.minus.base.longbench-write-abs}{6.3}
\setval{olmo.gdpo-w0.5.minus.base.math}{+19.1}
\setval{olmo.gdpo-w0.5.minus.base.math-abs}{19.1}
\setval{olmo.gdpo-w0.5.minus.base.math500}{+16.9}
\setval{olmo.gdpo-w0.5.minus.base.math500-abs}{16.9}
\setval{olmo.gdpo-w0.5.minus.base.mbpp-plus}{+3.7}
\setval{olmo.gdpo-w0.5.minus.base.mbpp-plus-abs}{3.7}
\setval{olmo.gdpo-w0.5.minus.base.minerva}{+6.2}
\setval{olmo.gdpo-w0.5.minus.base.minerva-abs}{6.2}
\setval{olmo.gdpo-w0.5.minus.base.olympiad-bench}{+21.2}
\setval{olmo.gdpo-w0.5.minus.base.olympiad-bench-abs}{21.2}
\setval{olmo.gdpo-w0.5.minus.base.writing}{+12.7}
\setval{olmo.gdpo-w0.5.minus.base.writing-abs}{12.7}
\setval{olmo.gdpo-w0.5.minus.base.writingbench}{+21.8}
\setval{olmo.gdpo-w0.5.minus.base.writingbench-abs}{21.8}
\setval{olmo.gdpo-w0.5.minus.concur-free.aime24}{+2.8}
\setval{olmo.gdpo-w0.5.minus.concur-free.aime24-abs}{2.8}
\setval{olmo.gdpo-w0.5.minus.concur-free.aime25}{+5.2}
\setval{olmo.gdpo-w0.5.minus.concur-free.aime25-abs}{5.2}
\setval{olmo.gdpo-w0.5.minus.concur-free.amc23}{+6.5}
\setval{olmo.gdpo-w0.5.minus.concur-free.amc23-abs}{6.5}
\setval{olmo.gdpo-w0.5.minus.concur-free.avg}{+0.2}
\setval{olmo.gdpo-w0.5.minus.concur-free.avg-abs}{0.2}
\setval{olmo.gdpo-w0.5.minus.concur-free.code}{+0.2}
\setval{olmo.gdpo-w0.5.minus.concur-free.code-abs}{0.2}
\setval{olmo.gdpo-w0.5.minus.concur-free.creative-writing-v3}{$-$1.7}
\setval{olmo.gdpo-w0.5.minus.concur-free.creative-writing-v3-abs}{1.7}
\setval{olmo.gdpo-w0.5.minus.concur-free.guidebench}{$-$8.3}
\setval{olmo.gdpo-w0.5.minus.concur-free.guidebench-abs}{8.3}
\setval{olmo.gdpo-w0.5.minus.concur-free.humaneval-plus}{+1.1}
\setval{olmo.gdpo-w0.5.minus.concur-free.humaneval-plus-abs}{1.1}
\setval{olmo.gdpo-w0.5.minus.concur-free.if}{$-$2.3}
\setval{olmo.gdpo-w0.5.minus.concur-free.if-abs}{2.3}
\setval{olmo.gdpo-w0.5.minus.concur-free.ifbench}{+1.0}
\setval{olmo.gdpo-w0.5.minus.concur-free.ifbench-abs}{1.0}
\setval{olmo.gdpo-w0.5.minus.concur-free.ifeval}{+0.2}
\setval{olmo.gdpo-w0.5.minus.concur-free.ifeval-abs}{0.2}
\setval{olmo.gdpo-w0.5.minus.concur-free.judge2.avg}{+0.7}
\setval{olmo.gdpo-w0.5.minus.concur-free.judge2.avg-abs}{0.7}
\setval{olmo.gdpo-w0.5.minus.concur-free.judge2.creative-writing-v3}{+1.9}
\setval{olmo.gdpo-w0.5.minus.concur-free.judge2.creative-writing-v3-abs}{1.9}
\setval{olmo.gdpo-w0.5.minus.concur-free.judge2.longbench-write}{$-$7.3}
\setval{olmo.gdpo-w0.5.minus.concur-free.judge2.longbench-write-abs}{7.3}
\setval{olmo.gdpo-w0.5.minus.concur-free.judge2.writing}{$-$2.3}
\setval{olmo.gdpo-w0.5.minus.concur-free.judge2.writing-abs}{2.3}
\setval{olmo.gdpo-w0.5.minus.concur-free.judge2.writingbench}{$-$1.3}
\setval{olmo.gdpo-w0.5.minus.concur-free.judge2.writingbench-abs}{1.3}
\setval{olmo.gdpo-w0.5.minus.concur-free.livecodebench}{$-$0.9}
\setval{olmo.gdpo-w0.5.minus.concur-free.livecodebench-abs}{0.9}
\setval{olmo.gdpo-w0.5.minus.concur-free.longbench-write}{$-$8.3}
\setval{olmo.gdpo-w0.5.minus.concur-free.longbench-write-abs}{8.3}
\setval{olmo.gdpo-w0.5.minus.concur-free.math}{+3.8}
\setval{olmo.gdpo-w0.5.minus.concur-free.math-abs}{3.8}
\setval{olmo.gdpo-w0.5.minus.concur-free.math500}{+2.7}
\setval{olmo.gdpo-w0.5.minus.concur-free.math500-abs}{2.7}
\setval{olmo.gdpo-w0.5.minus.concur-free.mbpp-plus}{+0.4}
\setval{olmo.gdpo-w0.5.minus.concur-free.mbpp-plus-abs}{0.4}
\setval{olmo.gdpo-w0.5.minus.concur-free.minerva}{+1.6}
\setval{olmo.gdpo-w0.5.minus.concur-free.minerva-abs}{1.6}
\setval{olmo.gdpo-w0.5.minus.concur-free.olympiad-bench}{+4.2}
\setval{olmo.gdpo-w0.5.minus.concur-free.olympiad-bench-abs}{4.2}
\setval{olmo.gdpo-w0.5.minus.concur-free.writing}{$-$4.6}
\setval{olmo.gdpo-w0.5.minus.concur-free.writing-abs}{4.6}
\setval{olmo.gdpo-w0.5.minus.concur-free.writingbench}{$-$3.7}
\setval{olmo.gdpo-w0.5.minus.concur-free.writingbench-abs}{3.7}
\setval{olmo.gdpo-w0.5.minus.concur-tied.aime24}{$-$4.9}
\setval{olmo.gdpo-w0.5.minus.concur-tied.aime24-abs}{4.9}
\setval{olmo.gdpo-w0.5.minus.concur-tied.aime25}{$-$2.6}
\setval{olmo.gdpo-w0.5.minus.concur-tied.aime25-abs}{2.6}
\setval{olmo.gdpo-w0.5.minus.concur-tied.amc23}{$-$5.9}
\setval{olmo.gdpo-w0.5.minus.concur-tied.amc23-abs}{5.9}
\setval{olmo.gdpo-w0.5.minus.concur-tied.avg}{$-$3.5}
\setval{olmo.gdpo-w0.5.minus.concur-tied.avg-abs}{3.5}
\setval{olmo.gdpo-w0.5.minus.concur-tied.code}{$-$5.3}
\setval{olmo.gdpo-w0.5.minus.concur-tied.code-abs}{5.3}
\setval{olmo.gdpo-w0.5.minus.concur-tied.creative-writing-v3}{$-$0.9}
\setval{olmo.gdpo-w0.5.minus.concur-tied.creative-writing-v3-abs}{0.9}
\setval{olmo.gdpo-w0.5.minus.concur-tied.guidebench}{$-$6.6}
\setval{olmo.gdpo-w0.5.minus.concur-tied.guidebench-abs}{6.6}
\setval{olmo.gdpo-w0.5.minus.concur-tied.humaneval-plus}{$-$5.5}
\setval{olmo.gdpo-w0.5.minus.concur-tied.humaneval-plus-abs}{5.5}
\setval{olmo.gdpo-w0.5.minus.concur-tied.if}{$-$2.6}
\setval{olmo.gdpo-w0.5.minus.concur-tied.if-abs}{2.6}
\setval{olmo.gdpo-w0.5.minus.concur-tied.ifbench}{$-$1.0}
\setval{olmo.gdpo-w0.5.minus.concur-tied.ifbench-abs}{1.0}
\setval{olmo.gdpo-w0.5.minus.concur-tied.ifeval}{$-$0.1}
\setval{olmo.gdpo-w0.5.minus.concur-tied.ifeval-abs}{0.1}
\setval{olmo.gdpo-w0.5.minus.concur-tied.judge2.avg}{$-$3.1}
\setval{olmo.gdpo-w0.5.minus.concur-tied.judge2.avg-abs}{3.1}
\setval{olmo.gdpo-w0.5.minus.concur-tied.judge2.creative-writing-v3}{+2.2}
\setval{olmo.gdpo-w0.5.minus.concur-tied.judge2.creative-writing-v3-abs}{2.2}
\setval{olmo.gdpo-w0.5.minus.concur-tied.judge2.longbench-write}{$-$3.9}
\setval{olmo.gdpo-w0.5.minus.concur-tied.judge2.longbench-write-abs}{3.9}
\setval{olmo.gdpo-w0.5.minus.concur-tied.judge2.writing}{$-$0.2}
\setval{olmo.gdpo-w0.5.minus.concur-tied.judge2.writing-abs}{0.2}
\setval{olmo.gdpo-w0.5.minus.concur-tied.judge2.writingbench}{+1.1}
\setval{olmo.gdpo-w0.5.minus.concur-tied.judge2.writingbench-abs}{1.1}
\setval{olmo.gdpo-w0.5.minus.concur-tied.livecodebench}{$-$4.2}
\setval{olmo.gdpo-w0.5.minus.concur-tied.livecodebench-abs}{4.2}
\setval{olmo.gdpo-w0.5.minus.concur-tied.longbench-write}{$-$5.5}
\setval{olmo.gdpo-w0.5.minus.concur-tied.longbench-write-abs}{5.5}
\setval{olmo.gdpo-w0.5.minus.concur-tied.math}{$-$3.7}
\setval{olmo.gdpo-w0.5.minus.concur-tied.math-abs}{3.7}
\setval{olmo.gdpo-w0.5.minus.concur-tied.math500}{$-$2.4}
\setval{olmo.gdpo-w0.5.minus.concur-tied.math500-abs}{2.4}
\setval{olmo.gdpo-w0.5.minus.concur-tied.mbpp-plus}{$-$6.1}
\setval{olmo.gdpo-w0.5.minus.concur-tied.mbpp-plus-abs}{6.1}
\setval{olmo.gdpo-w0.5.minus.concur-tied.minerva}{$-$2.5}
\setval{olmo.gdpo-w0.5.minus.concur-tied.minerva-abs}{2.5}
\setval{olmo.gdpo-w0.5.minus.concur-tied.olympiad-bench}{$-$4.1}
\setval{olmo.gdpo-w0.5.minus.concur-tied.olympiad-bench-abs}{4.1}
\setval{olmo.gdpo-w0.5.minus.concur-tied.writing}{$-$2.0}
\setval{olmo.gdpo-w0.5.minus.concur-tied.writing-abs}{2.0}
\setval{olmo.gdpo-w0.5.minus.concur-tied.writingbench}{+0.4}
\setval{olmo.gdpo-w0.5.minus.concur-tied.writingbench-abs}{0.4}
\setval{olmo.gdpo-w0.5.minus.concur.aime24}{$-$6.8}
\setval{olmo.gdpo-w0.5.minus.concur.aime24-abs}{6.8}
\setval{olmo.gdpo-w0.5.minus.concur.aime25}{$-$4.9}
\setval{olmo.gdpo-w0.5.minus.concur.aime25-abs}{4.9}
\setval{olmo.gdpo-w0.5.minus.concur.amc23}{$-$9.5}
\setval{olmo.gdpo-w0.5.minus.concur.amc23-abs}{9.5}
\setval{olmo.gdpo-w0.5.minus.concur.avg}{$-$3.7}
\setval{olmo.gdpo-w0.5.minus.concur.avg-abs}{3.7}
\setval{olmo.gdpo-w0.5.minus.concur.code}{$-$4.4}
\setval{olmo.gdpo-w0.5.minus.concur.code-abs}{4.4}
\setval{olmo.gdpo-w0.5.minus.concur.creative-writing-v3}{$-$1.1}
\setval{olmo.gdpo-w0.5.minus.concur.creative-writing-v3-abs}{1.1}
\setval{olmo.gdpo-w0.5.minus.concur.guidebench}{$-$5.0}
\setval{olmo.gdpo-w0.5.minus.concur.guidebench-abs}{5.0}
\setval{olmo.gdpo-w0.5.minus.concur.humaneval-plus}{$-$5.0}
\setval{olmo.gdpo-w0.5.minus.concur.humaneval-plus-abs}{5.0}
\setval{olmo.gdpo-w0.5.minus.concur.if}{$-$1.5}
\setval{olmo.gdpo-w0.5.minus.concur.if-abs}{1.5}
\setval{olmo.gdpo-w0.5.minus.concur.ifbench}{+0.7}
\setval{olmo.gdpo-w0.5.minus.concur.ifbench-abs}{0.7}
\setval{olmo.gdpo-w0.5.minus.concur.ifeval}{$-$0.2}
\setval{olmo.gdpo-w0.5.minus.concur.ifeval-abs}{0.2}
\setval{olmo.gdpo-w0.5.minus.concur.judge2.avg}{$-$3.3}
\setval{olmo.gdpo-w0.5.minus.concur.judge2.avg-abs}{3.3}
\setval{olmo.gdpo-w0.5.minus.concur.judge2.creative-writing-v3}{+1.0}
\setval{olmo.gdpo-w0.5.minus.concur.judge2.creative-writing-v3-abs}{1.0}
\setval{olmo.gdpo-w0.5.minus.concur.judge2.longbench-write}{$-$1.8}
\setval{olmo.gdpo-w0.5.minus.concur.judge2.longbench-write-abs}{1.8}
\setval{olmo.gdpo-w0.5.minus.concur.judge2.writing}{+0.3}
\setval{olmo.gdpo-w0.5.minus.concur.judge2.writing-abs}{0.3}
\setval{olmo.gdpo-w0.5.minus.concur.judge2.writingbench}{+1.8}
\setval{olmo.gdpo-w0.5.minus.concur.judge2.writingbench-abs}{1.8}
\setval{olmo.gdpo-w0.5.minus.concur.livecodebench}{$-$5.5}
\setval{olmo.gdpo-w0.5.minus.concur.livecodebench-abs}{5.5}
\setval{olmo.gdpo-w0.5.minus.concur.longbench-write}{$-$4.8}
\setval{olmo.gdpo-w0.5.minus.concur.longbench-write-abs}{4.8}
\setval{olmo.gdpo-w0.5.minus.concur.math}{$-$5.5}
\setval{olmo.gdpo-w0.5.minus.concur.math-abs}{5.5}
\setval{olmo.gdpo-w0.5.minus.concur.math500}{$-$3.5}
\setval{olmo.gdpo-w0.5.minus.concur.math500-abs}{3.5}
\setval{olmo.gdpo-w0.5.minus.concur.mbpp-plus}{$-$2.6}
\setval{olmo.gdpo-w0.5.minus.concur.mbpp-plus-abs}{2.6}
\setval{olmo.gdpo-w0.5.minus.concur.minerva}{$-$3.3}
\setval{olmo.gdpo-w0.5.minus.concur.minerva-abs}{3.3}
\setval{olmo.gdpo-w0.5.minus.concur.olympiad-bench}{$-$4.9}
\setval{olmo.gdpo-w0.5.minus.concur.olympiad-bench-abs}{4.9}
\setval{olmo.gdpo-w0.5.minus.concur.writing}{$-$1.6}
\setval{olmo.gdpo-w0.5.minus.concur.writing-abs}{1.6}
\setval{olmo.gdpo-w0.5.minus.concur.writingbench}{+1.1}
\setval{olmo.gdpo-w0.5.minus.concur.writingbench-abs}{1.1}
\setval{olmo.gdpo-w0.5.minus.gd2po.aime24}{+1.2}
\setval{olmo.gdpo-w0.5.minus.gd2po.aime24-abs}{1.2}
\setval{olmo.gdpo-w0.5.minus.gd2po.aime25}{+3.0}
\setval{olmo.gdpo-w0.5.minus.gd2po.aime25-abs}{3.0}
\setval{olmo.gdpo-w0.5.minus.gd2po.amc23}{+1.8}
\setval{olmo.gdpo-w0.5.minus.gd2po.amc23-abs}{1.8}
\setval{olmo.gdpo-w0.5.minus.gd2po.avg}{+0.0}
\setval{olmo.gdpo-w0.5.minus.gd2po.avg-abs}{0.0}
\setval{olmo.gdpo-w0.5.minus.gd2po.code}{+0.9}
\setval{olmo.gdpo-w0.5.minus.gd2po.code-abs}{0.9}
\setval{olmo.gdpo-w0.5.minus.gd2po.creative-writing-v3}{$-$0.2}
\setval{olmo.gdpo-w0.5.minus.gd2po.creative-writing-v3-abs}{0.2}
\setval{olmo.gdpo-w0.5.minus.gd2po.guidebench}{$-$9.2}
\setval{olmo.gdpo-w0.5.minus.gd2po.guidebench-abs}{9.2}
\setval{olmo.gdpo-w0.5.minus.gd2po.humaneval-plus}{+1.0}
\setval{olmo.gdpo-w0.5.minus.gd2po.humaneval-plus-abs}{1.0}
\setval{olmo.gdpo-w0.5.minus.gd2po.if}{$-$3.0}
\setval{olmo.gdpo-w0.5.minus.gd2po.if-abs}{3.0}
\setval{olmo.gdpo-w0.5.minus.gd2po.ifbench}{+0.3}
\setval{olmo.gdpo-w0.5.minus.gd2po.ifbench-abs}{0.3}
\setval{olmo.gdpo-w0.5.minus.gd2po.ifeval}{$-$0.1}
\setval{olmo.gdpo-w0.5.minus.gd2po.ifeval-abs}{0.1}
\setval{olmo.gdpo-w0.5.minus.gd2po.judge2.avg}{+0.0}
\setval{olmo.gdpo-w0.5.minus.gd2po.judge2.avg-abs}{0.0}
\setval{olmo.gdpo-w0.5.minus.gd2po.judge2.creative-writing-v3}{+1.5}
\setval{olmo.gdpo-w0.5.minus.gd2po.judge2.creative-writing-v3-abs}{1.5}
\setval{olmo.gdpo-w0.5.minus.gd2po.judge2.longbench-write}{$-$3.3}
\setval{olmo.gdpo-w0.5.minus.gd2po.judge2.longbench-write-abs}{3.3}
\setval{olmo.gdpo-w0.5.minus.gd2po.judge2.writing}{$-$1.1}
\setval{olmo.gdpo-w0.5.minus.gd2po.judge2.writing-abs}{1.1}
\setval{olmo.gdpo-w0.5.minus.gd2po.judge2.writingbench}{$-$1.7}
\setval{olmo.gdpo-w0.5.minus.gd2po.judge2.writingbench-abs}{1.7}
\setval{olmo.gdpo-w0.5.minus.gd2po.livecodebench}{+1.6}
\setval{olmo.gdpo-w0.5.minus.gd2po.livecodebench-abs}{1.6}
\setval{olmo.gdpo-w0.5.minus.gd2po.longbench-write}{$-$4.5}
\setval{olmo.gdpo-w0.5.minus.gd2po.longbench-write-abs}{4.5}
\setval{olmo.gdpo-w0.5.minus.gd2po.math}{+1.7}
\setval{olmo.gdpo-w0.5.minus.gd2po.math-abs}{1.7}
\setval{olmo.gdpo-w0.5.minus.gd2po.math500}{+1.3}
\setval{olmo.gdpo-w0.5.minus.gd2po.math500-abs}{1.3}
\setval{olmo.gdpo-w0.5.minus.gd2po.mbpp-plus}{+0.1}
\setval{olmo.gdpo-w0.5.minus.gd2po.mbpp-plus-abs}{0.1}
\setval{olmo.gdpo-w0.5.minus.gd2po.minerva}{+1.4}
\setval{olmo.gdpo-w0.5.minus.gd2po.minerva-abs}{1.4}
\setval{olmo.gdpo-w0.5.minus.gd2po.olympiad-bench}{+1.4}
\setval{olmo.gdpo-w0.5.minus.gd2po.olympiad-bench-abs}{1.4}
\setval{olmo.gdpo-w0.5.minus.gd2po.writing}{$-$1.3}
\setval{olmo.gdpo-w0.5.minus.gd2po.writing-abs}{1.3}
\setval{olmo.gdpo-w0.5.minus.gd2po.writingbench}{+0.8}
\setval{olmo.gdpo-w0.5.minus.gd2po.writingbench-abs}{0.8}
\setval{olmo.gdpo-w0.5.minus.gdpo-noaux.aime24}{$-$8.8}
\setval{olmo.gdpo-w0.5.minus.gdpo-noaux.aime24-abs}{8.8}
\setval{olmo.gdpo-w0.5.minus.gdpo-noaux.aime25}{$-$6.4}
\setval{olmo.gdpo-w0.5.minus.gdpo-noaux.aime25-abs}{6.4}
\setval{olmo.gdpo-w0.5.minus.gdpo-noaux.amc23}{$-$9.9}
\setval{olmo.gdpo-w0.5.minus.gdpo-noaux.amc23-abs}{9.9}
\setval{olmo.gdpo-w0.5.minus.gdpo-noaux.avg}{$-$2.2}
\setval{olmo.gdpo-w0.5.minus.gdpo-noaux.avg-abs}{2.2}
\setval{olmo.gdpo-w0.5.minus.gdpo-noaux.code}{$-$6.6}
\setval{olmo.gdpo-w0.5.minus.gdpo-noaux.code-abs}{6.6}
\setval{olmo.gdpo-w0.5.minus.gdpo-noaux.creative-writing-v3}{+9.9}
\setval{olmo.gdpo-w0.5.minus.gdpo-noaux.creative-writing-v3-abs}{9.9}
\setval{olmo.gdpo-w0.5.minus.gdpo-noaux.guidebench}{$-$3.8}
\setval{olmo.gdpo-w0.5.minus.gdpo-noaux.guidebench-abs}{3.8}
\setval{olmo.gdpo-w0.5.minus.gdpo-noaux.humaneval-plus}{$-$7.4}
\setval{olmo.gdpo-w0.5.minus.gdpo-noaux.humaneval-plus-abs}{7.4}
\setval{olmo.gdpo-w0.5.minus.gdpo-noaux.if}{$-$0.6}
\setval{olmo.gdpo-w0.5.minus.gdpo-noaux.if-abs}{0.6}
\setval{olmo.gdpo-w0.5.minus.gdpo-noaux.ifbench}{$-$0.7}
\setval{olmo.gdpo-w0.5.minus.gdpo-noaux.ifbench-abs}{0.7}
\setval{olmo.gdpo-w0.5.minus.gdpo-noaux.ifeval}{+2.8}
\setval{olmo.gdpo-w0.5.minus.gdpo-noaux.ifeval-abs}{2.8}
\setval{olmo.gdpo-w0.5.minus.gdpo-noaux.judge2.avg}{$-$2.4}
\setval{olmo.gdpo-w0.5.minus.gdpo-noaux.judge2.avg-abs}{2.4}
\setval{olmo.gdpo-w0.5.minus.gdpo-noaux.judge2.creative-writing-v3}{+8.5}
\setval{olmo.gdpo-w0.5.minus.gdpo-noaux.judge2.creative-writing-v3-abs}{8.5}
\setval{olmo.gdpo-w0.5.minus.gdpo-noaux.judge2.longbench-write}{+10.0}
\setval{olmo.gdpo-w0.5.minus.gdpo-noaux.judge2.longbench-write-abs}{10.0}
\setval{olmo.gdpo-w0.5.minus.gdpo-noaux.judge2.writing}{+8.5}
\setval{olmo.gdpo-w0.5.minus.gdpo-noaux.judge2.writing-abs}{8.5}
\setval{olmo.gdpo-w0.5.minus.gdpo-noaux.judge2.writingbench}{+6.9}
\setval{olmo.gdpo-w0.5.minus.gdpo-noaux.judge2.writingbench-abs}{6.9}
\setval{olmo.gdpo-w0.5.minus.gdpo-noaux.livecodebench}{$-$6.7}
\setval{olmo.gdpo-w0.5.minus.gdpo-noaux.livecodebench-abs}{6.7}
\setval{olmo.gdpo-w0.5.minus.gdpo-noaux.longbench-write}{+7.7}
\setval{olmo.gdpo-w0.5.minus.gdpo-noaux.longbench-write-abs}{7.7}
\setval{olmo.gdpo-w0.5.minus.gdpo-noaux.math}{$-$6.6}
\setval{olmo.gdpo-w0.5.minus.gdpo-noaux.math-abs}{6.6}
\setval{olmo.gdpo-w0.5.minus.gdpo-noaux.math500}{$-$3.8}
\setval{olmo.gdpo-w0.5.minus.gdpo-noaux.math500-abs}{3.8}
\setval{olmo.gdpo-w0.5.minus.gdpo-noaux.mbpp-plus}{$-$5.6}
\setval{olmo.gdpo-w0.5.minus.gdpo-noaux.mbpp-plus-abs}{5.6}
\setval{olmo.gdpo-w0.5.minus.gdpo-noaux.minerva}{$-$4.6}
\setval{olmo.gdpo-w0.5.minus.gdpo-noaux.minerva-abs}{4.6}
\setval{olmo.gdpo-w0.5.minus.gdpo-noaux.olympiad-bench}{$-$6.2}
\setval{olmo.gdpo-w0.5.minus.gdpo-noaux.olympiad-bench-abs}{6.2}
\setval{olmo.gdpo-w0.5.minus.gdpo-noaux.writing}{+9.6}
\setval{olmo.gdpo-w0.5.minus.gdpo-noaux.writing-abs}{9.6}
\setval{olmo.gdpo-w0.5.minus.gdpo-noaux.writingbench}{+11.0}
\setval{olmo.gdpo-w0.5.minus.gdpo-noaux.writingbench-abs}{11.0}
\setval{olmo.gdpo-w0.5.minus.gdpo.aime24}{+1.8}
\setval{olmo.gdpo-w0.5.minus.gdpo.aime24-abs}{1.8}
\setval{olmo.gdpo-w0.5.minus.gdpo.aime25}{+4.7}
\setval{olmo.gdpo-w0.5.minus.gdpo.aime25-abs}{4.7}
\setval{olmo.gdpo-w0.5.minus.gdpo.amc23}{+4.5}
\setval{olmo.gdpo-w0.5.minus.gdpo.amc23-abs}{4.5}
\setval{olmo.gdpo-w0.5.minus.gdpo.avg}{+1.8}
\setval{olmo.gdpo-w0.5.minus.gdpo.avg-abs}{1.8}
\setval{olmo.gdpo-w0.5.minus.gdpo.code}{+1.1}
\setval{olmo.gdpo-w0.5.minus.gdpo.code-abs}{1.1}
\setval{olmo.gdpo-w0.5.minus.gdpo.creative-writing-v3}{+1.0}
\setval{olmo.gdpo-w0.5.minus.gdpo.creative-writing-v3-abs}{1.0}
\setval{olmo.gdpo-w0.5.minus.gdpo.guidebench}{+1.1}
\setval{olmo.gdpo-w0.5.minus.gdpo.guidebench-abs}{1.1}
\setval{olmo.gdpo-w0.5.minus.gdpo.humaneval-plus}{+1.4}
\setval{olmo.gdpo-w0.5.minus.gdpo.humaneval-plus-abs}{1.4}
\setval{olmo.gdpo-w0.5.minus.gdpo.if}{+1.7}
\setval{olmo.gdpo-w0.5.minus.gdpo.if-abs}{1.7}
\setval{olmo.gdpo-w0.5.minus.gdpo.ifbench}{+3.0}
\setval{olmo.gdpo-w0.5.minus.gdpo.ifbench-abs}{3.0}
\setval{olmo.gdpo-w0.5.minus.gdpo.ifeval}{+1.1}
\setval{olmo.gdpo-w0.5.minus.gdpo.ifeval-abs}{1.1}
\setval{olmo.gdpo-w0.5.minus.gdpo.judge2.avg}{+1.7}
\setval{olmo.gdpo-w0.5.minus.gdpo.judge2.avg-abs}{1.7}
\setval{olmo.gdpo-w0.5.minus.gdpo.judge2.creative-writing-v3}{+1.6}
\setval{olmo.gdpo-w0.5.minus.gdpo.judge2.creative-writing-v3-abs}{1.6}
\setval{olmo.gdpo-w0.5.minus.gdpo.judge2.longbench-write}{+0.9}
\setval{olmo.gdpo-w0.5.minus.gdpo.judge2.longbench-write-abs}{0.9}
\setval{olmo.gdpo-w0.5.minus.gdpo.judge2.writing}{+0.7}
\setval{olmo.gdpo-w0.5.minus.gdpo.judge2.writing-abs}{0.7}
\setval{olmo.gdpo-w0.5.minus.gdpo.judge2.writingbench}{$-$0.4}
\setval{olmo.gdpo-w0.5.minus.gdpo.judge2.writingbench-abs}{0.4}
\setval{olmo.gdpo-w0.5.minus.gdpo.livecodebench}{+1.7}
\setval{olmo.gdpo-w0.5.minus.gdpo.livecodebench-abs}{1.7}
\setval{olmo.gdpo-w0.5.minus.gdpo.longbench-write}{+0.5}
\setval{olmo.gdpo-w0.5.minus.gdpo.longbench-write-abs}{0.5}
\setval{olmo.gdpo-w0.5.minus.gdpo.math}{+2.6}
\setval{olmo.gdpo-w0.5.minus.gdpo.math-abs}{2.6}
\setval{olmo.gdpo-w0.5.minus.gdpo.math500}{+1.8}
\setval{olmo.gdpo-w0.5.minus.gdpo.math500-abs}{1.8}
\setval{olmo.gdpo-w0.5.minus.gdpo.mbpp-plus}{+0.2}
\setval{olmo.gdpo-w0.5.minus.gdpo.mbpp-plus-abs}{0.2}
\setval{olmo.gdpo-w0.5.minus.gdpo.minerva}{+1.2}
\setval{olmo.gdpo-w0.5.minus.gdpo.minerva-abs}{1.2}
\setval{olmo.gdpo-w0.5.minus.gdpo.olympiad-bench}{+1.6}
\setval{olmo.gdpo-w0.5.minus.gdpo.olympiad-bench-abs}{1.6}
\setval{olmo.gdpo-w0.5.minus.gdpo.writing}{+0.9}
\setval{olmo.gdpo-w0.5.minus.gdpo.writing-abs}{0.9}
\setval{olmo.gdpo-w0.5.minus.gdpo.writingbench}{+1.2}
\setval{olmo.gdpo-w0.5.minus.gdpo.writingbench-abs}{1.2}
\setval{olmo.gdpo-w0.5.minus.geoma.aime24}{+5.8}
\setval{olmo.gdpo-w0.5.minus.geoma.aime24-abs}{5.8}
\setval{olmo.gdpo-w0.5.minus.geoma.aime25}{+5.3}
\setval{olmo.gdpo-w0.5.minus.geoma.aime25-abs}{5.3}
\setval{olmo.gdpo-w0.5.minus.geoma.amc23}{+10.1}
\setval{olmo.gdpo-w0.5.minus.geoma.amc23-abs}{10.1}
\setval{olmo.gdpo-w0.5.minus.geoma.avg}{+2.7}
\setval{olmo.gdpo-w0.5.minus.geoma.avg-abs}{2.7}
\setval{olmo.gdpo-w0.5.minus.geoma.code}{+1.9}
\setval{olmo.gdpo-w0.5.minus.geoma.code-abs}{1.9}
\setval{olmo.gdpo-w0.5.minus.geoma.creative-writing-v3}{$-$1.5}
\setval{olmo.gdpo-w0.5.minus.geoma.creative-writing-v3-abs}{1.5}
\setval{olmo.gdpo-w0.5.minus.geoma.guidebench}{+5.3}
\setval{olmo.gdpo-w0.5.minus.geoma.guidebench-abs}{5.3}
\setval{olmo.gdpo-w0.5.minus.geoma.humaneval-plus}{+3.1}
\setval{olmo.gdpo-w0.5.minus.geoma.humaneval-plus-abs}{3.1}
\setval{olmo.gdpo-w0.5.minus.geoma.if}{+2.4}
\setval{olmo.gdpo-w0.5.minus.geoma.if-abs}{2.4}
\setval{olmo.gdpo-w0.5.minus.geoma.ifbench}{+2.3}
\setval{olmo.gdpo-w0.5.minus.geoma.ifbench-abs}{2.3}
\setval{olmo.gdpo-w0.5.minus.geoma.ifeval}{$-$0.5}
\setval{olmo.gdpo-w0.5.minus.geoma.ifeval-abs}{0.5}
\setval{olmo.gdpo-w0.5.minus.geoma.judge2.avg}{+2.6}
\setval{olmo.gdpo-w0.5.minus.geoma.judge2.avg-abs}{2.6}
\setval{olmo.gdpo-w0.5.minus.geoma.judge2.creative-writing-v3}{$-$2.4}
\setval{olmo.gdpo-w0.5.minus.geoma.judge2.creative-writing-v3-abs}{2.4}
\setval{olmo.gdpo-w0.5.minus.geoma.judge2.longbench-write}{$-$1.6}
\setval{olmo.gdpo-w0.5.minus.geoma.judge2.longbench-write-abs}{1.6}
\setval{olmo.gdpo-w0.5.minus.geoma.judge2.writing}{$-$2.0}
\setval{olmo.gdpo-w0.5.minus.geoma.judge2.writing-abs}{2.0}
\setval{olmo.gdpo-w0.5.minus.geoma.judge2.writingbench}{$-$1.9}
\setval{olmo.gdpo-w0.5.minus.geoma.judge2.writingbench-abs}{1.9}
\setval{olmo.gdpo-w0.5.minus.geoma.livecodebench}{+2.5}
\setval{olmo.gdpo-w0.5.minus.geoma.livecodebench-abs}{2.5}
\setval{olmo.gdpo-w0.5.minus.geoma.longbench-write}{$-$1.6}
\setval{olmo.gdpo-w0.5.minus.geoma.longbench-write-abs}{1.6}
\setval{olmo.gdpo-w0.5.minus.geoma.math}{+5.2}
\setval{olmo.gdpo-w0.5.minus.geoma.math-abs}{5.2}
\setval{olmo.gdpo-w0.5.minus.geoma.math500}{+3.9}
\setval{olmo.gdpo-w0.5.minus.geoma.math500-abs}{3.9}
\setval{olmo.gdpo-w0.5.minus.geoma.mbpp-plus}{+0.2}
\setval{olmo.gdpo-w0.5.minus.geoma.mbpp-plus-abs}{0.2}
\setval{olmo.gdpo-w0.5.minus.geoma.minerva}{+1.6}
\setval{olmo.gdpo-w0.5.minus.geoma.minerva-abs}{1.6}
\setval{olmo.gdpo-w0.5.minus.geoma.olympiad-bench}{+4.8}
\setval{olmo.gdpo-w0.5.minus.geoma.olympiad-bench-abs}{4.8}
\setval{olmo.gdpo-w0.5.minus.geoma.writing}{$-$1.2}
\setval{olmo.gdpo-w0.5.minus.geoma.writing-abs}{1.2}
\setval{olmo.gdpo-w0.5.minus.geoma.writingbench}{$-$0.5}
\setval{olmo.gdpo-w0.5.minus.geoma.writingbench-abs}{0.5}
\setval{olmo.gdpo-w0.5.minus.grpo-nodiv.aime24}{$-$2.2}
\setval{olmo.gdpo-w0.5.minus.grpo-nodiv.aime24-abs}{2.2}
\setval{olmo.gdpo-w0.5.minus.grpo-nodiv.aime25}{$-$0.3}
\setval{olmo.gdpo-w0.5.minus.grpo-nodiv.aime25-abs}{0.3}
\setval{olmo.gdpo-w0.5.minus.grpo-nodiv.amc23}{+0.9}
\setval{olmo.gdpo-w0.5.minus.grpo-nodiv.amc23-abs}{0.9}
\setval{olmo.gdpo-w0.5.minus.grpo-nodiv.avg}{$-$0.3}
\setval{olmo.gdpo-w0.5.minus.grpo-nodiv.avg-abs}{0.3}
\setval{olmo.gdpo-w0.5.minus.grpo-nodiv.code}{+0.4}
\setval{olmo.gdpo-w0.5.minus.grpo-nodiv.code-abs}{0.4}
\setval{olmo.gdpo-w0.5.minus.grpo-nodiv.creative-writing-v3}{+1.1}
\setval{olmo.gdpo-w0.5.minus.grpo-nodiv.creative-writing-v3-abs}{1.1}
\setval{olmo.gdpo-w0.5.minus.grpo-nodiv.guidebench}{$-$3.6}
\setval{olmo.gdpo-w0.5.minus.grpo-nodiv.guidebench-abs}{3.6}
\setval{olmo.gdpo-w0.5.minus.grpo-nodiv.humaneval-plus}{+0.4}
\setval{olmo.gdpo-w0.5.minus.grpo-nodiv.humaneval-plus-abs}{0.4}
\setval{olmo.gdpo-w0.5.minus.grpo-nodiv.if}{$-$0.8}
\setval{olmo.gdpo-w0.5.minus.grpo-nodiv.if-abs}{0.8}
\setval{olmo.gdpo-w0.5.minus.grpo-nodiv.ifbench}{+1.3}
\setval{olmo.gdpo-w0.5.minus.grpo-nodiv.ifbench-abs}{1.3}
\setval{olmo.gdpo-w0.5.minus.grpo-nodiv.ifeval}{$-$0.2}
\setval{olmo.gdpo-w0.5.minus.grpo-nodiv.ifeval-abs}{0.2}
\setval{olmo.gdpo-w0.5.minus.grpo-nodiv.judge2.avg}{$-$0.6}
\setval{olmo.gdpo-w0.5.minus.grpo-nodiv.judge2.avg-abs}{0.6}
\setval{olmo.gdpo-w0.5.minus.grpo-nodiv.judge2.creative-writing-v3}{+0.6}
\setval{olmo.gdpo-w0.5.minus.grpo-nodiv.judge2.creative-writing-v3-abs}{0.6}
\setval{olmo.gdpo-w0.5.minus.grpo-nodiv.judge2.longbench-write}{$-$4.5}
\setval{olmo.gdpo-w0.5.minus.grpo-nodiv.judge2.longbench-write-abs}{4.5}
\setval{olmo.gdpo-w0.5.minus.grpo-nodiv.judge2.writing}{$-$1.4}
\setval{olmo.gdpo-w0.5.minus.grpo-nodiv.judge2.writing-abs}{1.4}
\setval{olmo.gdpo-w0.5.minus.grpo-nodiv.judge2.writingbench}{$-$0.3}
\setval{olmo.gdpo-w0.5.minus.grpo-nodiv.judge2.writingbench-abs}{0.3}
\setval{olmo.gdpo-w0.5.minus.grpo-nodiv.livecodebench}{+1.2}
\setval{olmo.gdpo-w0.5.minus.grpo-nodiv.livecodebench-abs}{1.2}
\setval{olmo.gdpo-w0.5.minus.grpo-nodiv.longbench-write}{$-$3.8}
\setval{olmo.gdpo-w0.5.minus.grpo-nodiv.longbench-write-abs}{3.8}
\setval{olmo.gdpo-w0.5.minus.grpo-nodiv.math}{$-$0.5}
\setval{olmo.gdpo-w0.5.minus.grpo-nodiv.math-abs}{0.5}
\setval{olmo.gdpo-w0.5.minus.grpo-nodiv.math500}{$-$0.5}
\setval{olmo.gdpo-w0.5.minus.grpo-nodiv.math500-abs}{0.5}
\setval{olmo.gdpo-w0.5.minus.grpo-nodiv.mbpp-plus}{$-$0.4}
\setval{olmo.gdpo-w0.5.minus.grpo-nodiv.mbpp-plus-abs}{0.4}
\setval{olmo.gdpo-w0.5.minus.grpo-nodiv.minerva}{$-$0.3}
\setval{olmo.gdpo-w0.5.minus.grpo-nodiv.minerva-abs}{0.3}
\setval{olmo.gdpo-w0.5.minus.grpo-nodiv.olympiad-bench}{$-$0.9}
\setval{olmo.gdpo-w0.5.minus.grpo-nodiv.olympiad-bench-abs}{0.9}
\setval{olmo.gdpo-w0.5.minus.grpo-nodiv.writing}{+0.2}
\setval{olmo.gdpo-w0.5.minus.grpo-nodiv.writing-abs}{0.2}
\setval{olmo.gdpo-w0.5.minus.grpo-nodiv.writingbench}{+3.3}
\setval{olmo.gdpo-w0.5.minus.grpo-nodiv.writingbench-abs}{3.3}
\setval{olmo.gdpo-w0.5.minus.grpo.aime24}{+0.1}
\setval{olmo.gdpo-w0.5.minus.grpo.aime24-abs}{0.1}
\setval{olmo.gdpo-w0.5.minus.grpo.aime25}{+2.6}
\setval{olmo.gdpo-w0.5.minus.grpo.aime25-abs}{2.6}
\setval{olmo.gdpo-w0.5.minus.grpo.amc23}{+3.1}
\setval{olmo.gdpo-w0.5.minus.grpo.amc23-abs}{3.1}
\setval{olmo.gdpo-w0.5.minus.grpo.avg}{+0.4}
\setval{olmo.gdpo-w0.5.minus.grpo.avg-abs}{0.4}
\setval{olmo.gdpo-w0.5.minus.grpo.code}{+0.7}
\setval{olmo.gdpo-w0.5.minus.grpo.code-abs}{0.7}
\setval{olmo.gdpo-w0.5.minus.grpo.creative-writing-v3}{$-$1.3}
\setval{olmo.gdpo-w0.5.minus.grpo.creative-writing-v3-abs}{1.3}
\setval{olmo.gdpo-w0.5.minus.grpo.guidebench}{$-$4.5}
\setval{olmo.gdpo-w0.5.minus.grpo.guidebench-abs}{4.5}
\setval{olmo.gdpo-w0.5.minus.grpo.humaneval-plus}{+0.4}
\setval{olmo.gdpo-w0.5.minus.grpo.humaneval-plus-abs}{0.4}
\setval{olmo.gdpo-w0.5.minus.grpo.if}{$-$1.2}
\setval{olmo.gdpo-w0.5.minus.grpo.if-abs}{1.2}
\setval{olmo.gdpo-w0.5.minus.grpo.ifbench}{+0.7}
\setval{olmo.gdpo-w0.5.minus.grpo.ifbench-abs}{0.7}
\setval{olmo.gdpo-w0.5.minus.grpo.ifeval}{+0.1}
\setval{olmo.gdpo-w0.5.minus.grpo.ifeval-abs}{0.1}
\setval{olmo.gdpo-w0.5.minus.grpo.judge2.avg}{+0.5}
\setval{olmo.gdpo-w0.5.minus.grpo.judge2.avg-abs}{0.5}
\setval{olmo.gdpo-w0.5.minus.grpo.judge2.creative-writing-v3}{+0.8}
\setval{olmo.gdpo-w0.5.minus.grpo.judge2.creative-writing-v3-abs}{0.8}
\setval{olmo.gdpo-w0.5.minus.grpo.judge2.longbench-write}{$-$0.5}
\setval{olmo.gdpo-w0.5.minus.grpo.judge2.longbench-write-abs}{0.5}
\setval{olmo.gdpo-w0.5.minus.grpo.judge2.writing}{$-$0.5}
\setval{olmo.gdpo-w0.5.minus.grpo.judge2.writing-abs}{0.5}
\setval{olmo.gdpo-w0.5.minus.grpo.judge2.writingbench}{$-$1.8}
\setval{olmo.gdpo-w0.5.minus.grpo.judge2.writingbench-abs}{1.8}
\setval{olmo.gdpo-w0.5.minus.grpo.livecodebench}{+2.5}
\setval{olmo.gdpo-w0.5.minus.grpo.livecodebench-abs}{2.5}
\setval{olmo.gdpo-w0.5.minus.grpo.longbench-write}{$-$0.5}
\setval{olmo.gdpo-w0.5.minus.grpo.longbench-write-abs}{0.5}
\setval{olmo.gdpo-w0.5.minus.grpo.math}{+1.9}
\setval{olmo.gdpo-w0.5.minus.grpo.math-abs}{1.9}
\setval{olmo.gdpo-w0.5.minus.grpo.math500}{+2.1}
\setval{olmo.gdpo-w0.5.minus.grpo.math500-abs}{2.1}
\setval{olmo.gdpo-w0.5.minus.grpo.mbpp-plus}{$-$0.9}
\setval{olmo.gdpo-w0.5.minus.grpo.mbpp-plus-abs}{0.9}
\setval{olmo.gdpo-w0.5.minus.grpo.minerva}{+1.6}
\setval{olmo.gdpo-w0.5.minus.grpo.minerva-abs}{1.6}
\setval{olmo.gdpo-w0.5.minus.grpo.olympiad-bench}{+1.8}
\setval{olmo.gdpo-w0.5.minus.grpo.olympiad-bench-abs}{1.8}
\setval{olmo.gdpo-w0.5.minus.grpo.writing}{$-$1.2}
\setval{olmo.gdpo-w0.5.minus.grpo.writing-abs}{1.2}
\setval{olmo.gdpo-w0.5.minus.grpo.writingbench}{$-$2.0}
\setval{olmo.gdpo-w0.5.minus.grpo.writingbench-abs}{2.0}
\setval{olmo.gdpo-w0.5.minus.mt-grpo.aime24}{$-$7.0}
\setval{olmo.gdpo-w0.5.minus.mt-grpo.aime24-abs}{7.0}
\setval{olmo.gdpo-w0.5.minus.mt-grpo.aime25}{$-$0.8}
\setval{olmo.gdpo-w0.5.minus.mt-grpo.aime25-abs}{0.8}
\setval{olmo.gdpo-w0.5.minus.mt-grpo.amc23}{$-$3.4}
\setval{olmo.gdpo-w0.5.minus.mt-grpo.amc23-abs}{3.4}
\setval{olmo.gdpo-w0.5.minus.mt-grpo.avg}{$-$1.1}
\setval{olmo.gdpo-w0.5.minus.mt-grpo.avg-abs}{1.1}
\setval{olmo.gdpo-w0.5.minus.mt-grpo.code}{+1.4}
\setval{olmo.gdpo-w0.5.minus.mt-grpo.code-abs}{1.4}
\setval{olmo.gdpo-w0.5.minus.mt-grpo.creative-writing-v3}{$-$2.6}
\setval{olmo.gdpo-w0.5.minus.mt-grpo.creative-writing-v3-abs}{2.6}
\setval{olmo.gdpo-w0.5.minus.mt-grpo.guidebench}{$-$2.6}
\setval{olmo.gdpo-w0.5.minus.mt-grpo.guidebench-abs}{2.6}
\setval{olmo.gdpo-w0.5.minus.mt-grpo.humaneval-plus}{+2.0}
\setval{olmo.gdpo-w0.5.minus.mt-grpo.humaneval-plus-abs}{2.0}
\setval{olmo.gdpo-w0.5.minus.mt-grpo.if}{$-$0.5}
\setval{olmo.gdpo-w0.5.minus.mt-grpo.if-abs}{0.5}
\setval{olmo.gdpo-w0.5.minus.mt-grpo.ifbench}{+0.7}
\setval{olmo.gdpo-w0.5.minus.mt-grpo.ifbench-abs}{0.7}
\setval{olmo.gdpo-w0.5.minus.mt-grpo.ifeval}{+0.5}
\setval{olmo.gdpo-w0.5.minus.mt-grpo.ifeval-abs}{0.5}
\setval{olmo.gdpo-w0.5.minus.mt-grpo.judge2.avg}{$-$1.0}
\setval{olmo.gdpo-w0.5.minus.mt-grpo.judge2.avg-abs}{1.0}
\setval{olmo.gdpo-w0.5.minus.mt-grpo.judge2.creative-writing-v3}{$-$0.5}
\setval{olmo.gdpo-w0.5.minus.mt-grpo.judge2.creative-writing-v3-abs}{0.5}
\setval{olmo.gdpo-w0.5.minus.mt-grpo.judge2.longbench-write}{$-$3.8}
\setval{olmo.gdpo-w0.5.minus.mt-grpo.judge2.longbench-write-abs}{3.8}
\setval{olmo.gdpo-w0.5.minus.mt-grpo.judge2.writing}{$-$1.7}
\setval{olmo.gdpo-w0.5.minus.mt-grpo.judge2.writing-abs}{1.7}
\setval{olmo.gdpo-w0.5.minus.mt-grpo.judge2.writingbench}{$-$0.8}
\setval{olmo.gdpo-w0.5.minus.mt-grpo.judge2.writingbench-abs}{0.8}
\setval{olmo.gdpo-w0.5.minus.mt-grpo.livecodebench}{+2.2}
\setval{olmo.gdpo-w0.5.minus.mt-grpo.livecodebench-abs}{2.2}
\setval{olmo.gdpo-w0.5.minus.mt-grpo.longbench-write}{$-$2.6}
\setval{olmo.gdpo-w0.5.minus.mt-grpo.longbench-write-abs}{2.6}
\setval{olmo.gdpo-w0.5.minus.mt-grpo.math}{$-$2.0}
\setval{olmo.gdpo-w0.5.minus.mt-grpo.math-abs}{2.0}
\setval{olmo.gdpo-w0.5.minus.mt-grpo.math500}{+0.1}
\setval{olmo.gdpo-w0.5.minus.mt-grpo.math500-abs}{0.1}
\setval{olmo.gdpo-w0.5.minus.mt-grpo.mbpp-plus}{$-$0.1}
\setval{olmo.gdpo-w0.5.minus.mt-grpo.mbpp-plus-abs}{0.1}
\setval{olmo.gdpo-w0.5.minus.mt-grpo.minerva}{$-$0.4}
\setval{olmo.gdpo-w0.5.minus.mt-grpo.minerva-abs}{0.4}
\setval{olmo.gdpo-w0.5.minus.mt-grpo.olympiad-bench}{$-$0.5}
\setval{olmo.gdpo-w0.5.minus.mt-grpo.olympiad-bench-abs}{0.5}
\setval{olmo.gdpo-w0.5.minus.mt-grpo.writing}{$-$2.1}
\setval{olmo.gdpo-w0.5.minus.mt-grpo.writing-abs}{2.1}
\setval{olmo.gdpo-w0.5.minus.mt-grpo.writingbench}{$-$1.2}
\setval{olmo.gdpo-w0.5.minus.mt-grpo.writingbench-abs}{1.2}
\setval{olmo.gdpo-w0.5.minus.papo.aime24}{+7.4}
\setval{olmo.gdpo-w0.5.minus.papo.aime24-abs}{7.4}
\setval{olmo.gdpo-w0.5.minus.papo.aime25}{+7.0}
\setval{olmo.gdpo-w0.5.minus.papo.aime25-abs}{7.0}
\setval{olmo.gdpo-w0.5.minus.papo.amc23}{+14.6}
\setval{olmo.gdpo-w0.5.minus.papo.amc23-abs}{14.6}
\setval{olmo.gdpo-w0.5.minus.papo.avg}{+3.3}
\setval{olmo.gdpo-w0.5.minus.papo.avg-abs}{3.3}
\setval{olmo.gdpo-w0.5.minus.papo.code}{+1.9}
\setval{olmo.gdpo-w0.5.minus.papo.code-abs}{1.9}
\setval{olmo.gdpo-w0.5.minus.papo.creative-writing-v3}{+1.1}
\setval{olmo.gdpo-w0.5.minus.papo.creative-writing-v3-abs}{1.1}
\setval{olmo.gdpo-w0.5.minus.papo.guidebench}{$-$6.7}
\setval{olmo.gdpo-w0.5.minus.papo.guidebench-abs}{6.7}
\setval{olmo.gdpo-w0.5.minus.papo.humaneval-plus}{+3.0}
\setval{olmo.gdpo-w0.5.minus.papo.humaneval-plus-abs}{3.0}
\setval{olmo.gdpo-w0.5.minus.papo.if}{$-$1.1}
\setval{olmo.gdpo-w0.5.minus.papo.if-abs}{1.1}
\setval{olmo.gdpo-w0.5.minus.papo.ifbench}{+3.0}
\setval{olmo.gdpo-w0.5.minus.papo.ifbench-abs}{3.0}
\setval{olmo.gdpo-w0.5.minus.papo.ifeval}{+0.4}
\setval{olmo.gdpo-w0.5.minus.papo.ifeval-abs}{0.4}
\setval{olmo.gdpo-w0.5.minus.papo.judge2.avg}{+3.3}
\setval{olmo.gdpo-w0.5.minus.papo.judge2.avg-abs}{3.3}
\setval{olmo.gdpo-w0.5.minus.papo.judge2.creative-writing-v3}{+1.4}
\setval{olmo.gdpo-w0.5.minus.papo.judge2.creative-writing-v3-abs}{1.4}
\setval{olmo.gdpo-w0.5.minus.papo.judge2.longbench-write}{+1.0}
\setval{olmo.gdpo-w0.5.minus.papo.judge2.longbench-write-abs}{1.0}
\setval{olmo.gdpo-w0.5.minus.papo.judge2.writing}{+0.5}
\setval{olmo.gdpo-w0.5.minus.papo.judge2.writing-abs}{0.5}
\setval{olmo.gdpo-w0.5.minus.papo.judge2.writingbench}{$-$0.8}
\setval{olmo.gdpo-w0.5.minus.papo.judge2.writingbench-abs}{0.8}
\setval{olmo.gdpo-w0.5.minus.papo.livecodebench}{+2.9}
\setval{olmo.gdpo-w0.5.minus.papo.livecodebench-abs}{2.9}
\setval{olmo.gdpo-w0.5.minus.papo.longbench-write}{$-$1.0}
\setval{olmo.gdpo-w0.5.minus.papo.longbench-write-abs}{1.0}
\setval{olmo.gdpo-w0.5.minus.papo.math}{+7.6}
\setval{olmo.gdpo-w0.5.minus.papo.math-abs}{7.6}
\setval{olmo.gdpo-w0.5.minus.papo.math500}{+6.2}
\setval{olmo.gdpo-w0.5.minus.papo.math500-abs}{6.2}
\setval{olmo.gdpo-w0.5.minus.papo.mbpp-plus}{$-$0.2}
\setval{olmo.gdpo-w0.5.minus.papo.mbpp-plus-abs}{0.2}
\setval{olmo.gdpo-w0.5.minus.papo.minerva}{+3.0}
\setval{olmo.gdpo-w0.5.minus.papo.minerva-abs}{3.0}
\setval{olmo.gdpo-w0.5.minus.papo.olympiad-bench}{+7.8}
\setval{olmo.gdpo-w0.5.minus.papo.olympiad-bench-abs}{7.8}
\setval{olmo.gdpo-w0.5.minus.papo.writing}{+0.4}
\setval{olmo.gdpo-w0.5.minus.papo.writing-abs}{0.4}
\setval{olmo.gdpo-w0.5.minus.papo.writingbench}{+1.1}
\setval{olmo.gdpo-w0.5.minus.papo.writingbench-abs}{1.1}
\setval{olmo.gdpo-w0.5.minus.saw.aime24}{+3.4}
\setval{olmo.gdpo-w0.5.minus.saw.aime24-abs}{3.4}
\setval{olmo.gdpo-w0.5.minus.saw.aime25}{+2.1}
\setval{olmo.gdpo-w0.5.minus.saw.aime25-abs}{2.1}
\setval{olmo.gdpo-w0.5.minus.saw.amc23}{+7.1}
\setval{olmo.gdpo-w0.5.minus.saw.amc23-abs}{7.1}
\setval{olmo.gdpo-w0.5.minus.saw.avg}{+1.2}
\setval{olmo.gdpo-w0.5.minus.saw.avg-abs}{1.2}
\setval{olmo.gdpo-w0.5.minus.saw.code}{+1.3}
\setval{olmo.gdpo-w0.5.minus.saw.code-abs}{1.3}
\setval{olmo.gdpo-w0.5.minus.saw.creative-writing-v3}{+0.2}
\setval{olmo.gdpo-w0.5.minus.saw.creative-writing-v3-abs}{0.2}
\setval{olmo.gdpo-w0.5.minus.saw.guidebench}{$-$2.8}
\setval{olmo.gdpo-w0.5.minus.saw.guidebench-abs}{2.8}
\setval{olmo.gdpo-w0.5.minus.saw.humaneval-plus}{+1.4}
\setval{olmo.gdpo-w0.5.minus.saw.humaneval-plus-abs}{1.4}
\setval{olmo.gdpo-w0.5.minus.saw.if}{$-$1.3}
\setval{olmo.gdpo-w0.5.minus.saw.if-abs}{1.3}
\setval{olmo.gdpo-w0.5.minus.saw.ifbench}{+0.0}
\setval{olmo.gdpo-w0.5.minus.saw.ifbench-abs}{0.0}
\setval{olmo.gdpo-w0.5.minus.saw.ifeval}{$-$1.1}
\setval{olmo.gdpo-w0.5.minus.saw.ifeval-abs}{1.1}
\setval{olmo.gdpo-w0.5.minus.saw.judge2.avg}{+1.1}
\setval{olmo.gdpo-w0.5.minus.saw.judge2.avg-abs}{1.1}
\setval{olmo.gdpo-w0.5.minus.saw.judge2.creative-writing-v3}{$-$1.6}
\setval{olmo.gdpo-w0.5.minus.saw.judge2.creative-writing-v3-abs}{1.6}
\setval{olmo.gdpo-w0.5.minus.saw.judge2.longbench-write}{$-$0.8}
\setval{olmo.gdpo-w0.5.minus.saw.judge2.longbench-write-abs}{0.8}
\setval{olmo.gdpo-w0.5.minus.saw.judge2.writing}{$-$1.4}
\setval{olmo.gdpo-w0.5.minus.saw.judge2.writing-abs}{1.4}
\setval{olmo.gdpo-w0.5.minus.saw.judge2.writingbench}{$-$1.7}
\setval{olmo.gdpo-w0.5.minus.saw.judge2.writingbench-abs}{1.7}
\setval{olmo.gdpo-w0.5.minus.saw.livecodebench}{+2.9}
\setval{olmo.gdpo-w0.5.minus.saw.livecodebench-abs}{2.9}
\setval{olmo.gdpo-w0.5.minus.saw.longbench-write}{$-$1.2}
\setval{olmo.gdpo-w0.5.minus.saw.longbench-write-abs}{1.2}
\setval{olmo.gdpo-w0.5.minus.saw.math}{+3.3}
\setval{olmo.gdpo-w0.5.minus.saw.math-abs}{3.3}
\setval{olmo.gdpo-w0.5.minus.saw.math500}{+2.6}
\setval{olmo.gdpo-w0.5.minus.saw.math500-abs}{2.6}
\setval{olmo.gdpo-w0.5.minus.saw.mbpp-plus}{$-$0.5}
\setval{olmo.gdpo-w0.5.minus.saw.mbpp-plus-abs}{0.5}
\setval{olmo.gdpo-w0.5.minus.saw.minerva}{+1.2}
\setval{olmo.gdpo-w0.5.minus.saw.minerva-abs}{1.2}
\setval{olmo.gdpo-w0.5.minus.saw.olympiad-bench}{+3.7}
\setval{olmo.gdpo-w0.5.minus.saw.olympiad-bench-abs}{3.7}
\setval{olmo.gdpo-w0.5.minus.saw.writing}{$-$0.6}
\setval{olmo.gdpo-w0.5.minus.saw.writing-abs}{0.6}
\setval{olmo.gdpo-w0.5.minus.saw.writingbench}{$-$0.9}
\setval{olmo.gdpo-w0.5.minus.saw.writingbench-abs}{0.9}
\setval{olmo.gdpo-w0.5.passing-at-ceiling.code.early}{0.84}
\setval{olmo.gdpo-w0.5.passing-at-ceiling.code.late}{0.86}
\setval{olmo.gdpo-w0.5.passing-at-ceiling.code.run}{0.84}
\setval{olmo.gdpo-w0.5.passing-at-ceiling.if.early}{0.48}
\setval{olmo.gdpo-w0.5.passing-at-ceiling.if.late}{0.57}
\setval{olmo.gdpo-w0.5.passing-at-ceiling.if.run}{0.52}
\setval{olmo.gdpo-w0.5.passing-at-ceiling.math.early}{1.00}
\setval{olmo.gdpo-w0.5.passing-at-ceiling.math.late}{1.00}
\setval{olmo.gdpo-w0.5.passing-at-ceiling.math.run}{1.00}
\setval{olmo.gdpo-w0.5.passing-at-ceiling.writing.early}{0.00}
\setval{olmo.gdpo-w0.5.passing-at-ceiling.writing.late}{0.00}
\setval{olmo.gdpo-w0.5.passing-at-ceiling.writing.run}{0.00}
\setval{olmo.gdpo-w0.5.primary-cosine.code.early}{0.97}
\setval{olmo.gdpo-w0.5.primary-cosine.code.late}{0.94}
\setval{olmo.gdpo-w0.5.primary-cosine.code.run}{0.95}
\setval{olmo.gdpo-w0.5.primary-cosine.if.early}{0.80}
\setval{olmo.gdpo-w0.5.primary-cosine.if.late}{0.79}
\setval{olmo.gdpo-w0.5.primary-cosine.if.run}{0.80}
\setval{olmo.gdpo-w0.5.primary-cosine.math.early}{0.99}
\setval{olmo.gdpo-w0.5.primary-cosine.math.late}{0.94}
\setval{olmo.gdpo-w0.5.primary-cosine.math.run}{0.95}
\setval{olmo.gdpo-w0.5.primary-cosine.writing.early}{0.94}
\setval{olmo.gdpo-w0.5.primary-cosine.writing.late}{0.87}
\setval{olmo.gdpo-w0.5.primary-cosine.writing.run}{0.88}
\setval{olmo.gdpo-w0.5.primary-level.code.early}{0.59}
\setval{olmo.gdpo-w0.5.primary-level.code.late}{0.84}
\setval{olmo.gdpo-w0.5.primary-level.code.run}{0.77}
\setval{olmo.gdpo-w0.5.primary-level.if.early}{0.63}
\setval{olmo.gdpo-w0.5.primary-level.if.late}{0.74}
\setval{olmo.gdpo-w0.5.primary-level.if.run}{0.69}
\setval{olmo.gdpo-w0.5.primary-level.math.early}{0.34}
\setval{olmo.gdpo-w0.5.primary-level.math.late}{0.69}
\setval{olmo.gdpo-w0.5.primary-level.math.run}{0.60}
\setval{olmo.gdpo-w0.5.primary-level.writing.early}{0.50}
\setval{olmo.gdpo-w0.5.primary-level.writing.late}{0.70}
\setval{olmo.gdpo-w0.5.primary-level.writing.run}{0.65}
\setval{olmo.gdpo-w0.5.solved.code.early}{0.14}
\setval{olmo.gdpo-w0.5.solved.code.late}{0.20}
\setval{olmo.gdpo-w0.5.solved.code.run}{0.17}
\setval{olmo.gdpo-w0.5.solved.if.early}{0.17}
\setval{olmo.gdpo-w0.5.solved.if.late}{0.18}
\setval{olmo.gdpo-w0.5.solved.if.run}{0.18}
\setval{olmo.gdpo-w0.5.solved.math.early}{0.07}
\setval{olmo.gdpo-w0.5.solved.math.late}{0.32}
\setval{olmo.gdpo-w0.5.solved.math.run}{0.23}
\setval{olmo.gdpo-w0.5.solved.writing.early}{0.00}
\setval{olmo.gdpo-w0.5.solved.writing.late}{0.00}
\setval{olmo.gdpo-w0.5.solved.writing.run}{0.00}
\setval{olmo.gdpo-w0.5.update-share.code.early}{0.19}
\setval{olmo.gdpo-w0.5.update-share.code.late}{0.17}
\setval{olmo.gdpo-w0.5.update-share.code.run}{0.18}
\setval{olmo.gdpo-w0.5.update-share.if.early}{0.23}
\setval{olmo.gdpo-w0.5.update-share.if.late}{0.23}
\setval{olmo.gdpo-w0.5.update-share.if.run}{0.24}
\setval{olmo.gdpo-w0.5.update-share.math.early}{0.20}
\setval{olmo.gdpo-w0.5.update-share.math.late}{0.22}
\setval{olmo.gdpo-w0.5.update-share.math.run}{0.21}
\setval{olmo.gdpo-w0.5.update-share.writing.early}{0.39}
\setval{olmo.gdpo-w0.5.update-share.writing.late}{0.38}
\setval{olmo.gdpo-w0.5.update-share.writing.run}{0.37}
\setval{olmo.gdpo.aux-change.code}{+0.16}
\setval{olmo.gdpo.aux-change.if}{+0.12}
\setval{olmo.gdpo.aux-change.math}{$-$0.13}
\setval{olmo.gdpo.aux-change.writing}{+0.02}
\setval{olmo.gdpo.aux-change.writing-rm}{+0.29}
\setval{olmo.gdpo.aux-full.code.early}{0.74}
\setval{olmo.gdpo.aux-full.code.late}{0.91}
\setval{olmo.gdpo.aux-full.code.run}{0.89}
\setval{olmo.gdpo.aux-full.if.early}{0.60}
\setval{olmo.gdpo.aux-full.if.late}{0.95}
\setval{olmo.gdpo.aux-full.if.run}{0.87}
\setval{olmo.gdpo.aux-full.math.early}{0.53}
\setval{olmo.gdpo.aux-full.math.late}{0.54}
\setval{olmo.gdpo.aux-full.math.run}{0.46}
\setval{olmo.gdpo.aux-full.writing-rm.early}{0.02}
\setval{olmo.gdpo.aux-full.writing-rm.late}{0.25}
\setval{olmo.gdpo.aux-full.writing-rm.run}{0.11}
\setval{olmo.gdpo.aux-full.writing.early}{0.57}
\setval{olmo.gdpo.aux-full.writing.late}{0.75}
\setval{olmo.gdpo.aux-full.writing.run}{0.64}
\setval{olmo.gdpo.aux-level.code.early}{0.90}
\setval{olmo.gdpo.aux-level.code.late}{0.96}
\setval{olmo.gdpo.aux-level.code.run}{0.97}
\setval{olmo.gdpo.aux-level.if.early}{0.90}
\setval{olmo.gdpo.aux-level.if.late}{0.99}
\setval{olmo.gdpo.aux-level.if.run}{0.97}
\setval{olmo.gdpo.aux-level.math.early}{0.90}
\setval{olmo.gdpo.aux-level.math.late}{0.77}
\setval{olmo.gdpo.aux-level.math.run}{0.76}
\setval{olmo.gdpo.aux-level.writing-rm.early}{0.60}
\setval{olmo.gdpo.aux-level.writing-rm.late}{0.86}
\setval{olmo.gdpo.aux-level.writing-rm.run}{0.76}
\setval{olmo.gdpo.aux-level.writing.early}{0.95}
\setval{olmo.gdpo.aux-level.writing.late}{0.97}
\setval{olmo.gdpo.aux-level.writing.run}{0.95}
\setval{olmo.gdpo.bench.aime24}{27.8}
\setval{olmo.gdpo.bench.aime25}{22.6}
\setval{olmo.gdpo.bench.amc23}{69.5}
\setval{olmo.gdpo.bench.creative-writing-v3}{47.1}
\setval{olmo.gdpo.bench.guidebench}{50.9}
\setval{olmo.gdpo.bench.humaneval-plus}{76.3}
\setval{olmo.gdpo.bench.ifbench}{29.7}
\setval{olmo.gdpo.bench.ifeval}{86.5}
\setval{olmo.gdpo.bench.livecodebench}{23.1}
\setval{olmo.gdpo.bench.longbench-write}{63.3}
\setval{olmo.gdpo.bench.math500}{85.2}
\setval{olmo.gdpo.bench.mbpp-plus}{64.8}
\setval{olmo.gdpo.bench.minerva}{40.4}
\setval{olmo.gdpo.bench.olympiad-bench}{55.9}
\setval{olmo.gdpo.bench.writingbench}{65.3}
\setval{olmo.gdpo.eval.avg}{53.9}
\setval{olmo.gdpo.eval.code}{54.7}
\setval{olmo.gdpo.eval.if}{55.7}
\setval{olmo.gdpo.eval.math}{50.2}
\setval{olmo.gdpo.eval.writing}{58.6}
\setval{olmo.gdpo.gate.code.against.early}{0.06}
\setval{olmo.gdpo.gate.code.against.late}{0.10}
\setval{olmo.gdpo.gate.code.against.run}{0.07}
\setval{olmo.gdpo.gate.code.agreement}{$-$0.014}
\setval{olmo.gdpo.gate.code.ceiling.early}{0.21}
\setval{olmo.gdpo.gate.code.ceiling.late}{0.21}
\setval{olmo.gdpo.gate.code.ceiling.run}{0.18}
\setval{olmo.gdpo.gate.code.cosine.early}{0.74}
\setval{olmo.gdpo.gate.code.cosine.late}{0.60}
\setval{olmo.gdpo.gate.code.cosine.run}{0.71}
\setval{olmo.gdpo.gate.if.against.early}{0.10}
\setval{olmo.gdpo.gate.if.against.late}{0.04}
\setval{olmo.gdpo.gate.if.against.run}{0.06}
\setval{olmo.gdpo.gate.if.agreement}{+0.042}
\setval{olmo.gdpo.gate.if.ceiling.early}{0.22}
\setval{olmo.gdpo.gate.if.ceiling.late}{0.13}
\setval{olmo.gdpo.gate.if.ceiling.run}{0.18}
\setval{olmo.gdpo.gate.if.cosine.early}{0.51}
\setval{olmo.gdpo.gate.if.cosine.late}{0.53}
\setval{olmo.gdpo.gate.if.cosine.run}{0.55}
\setval{olmo.gdpo.gate.math.against.early}{0.00}
\setval{olmo.gdpo.gate.math.against.late}{0.00}
\setval{olmo.gdpo.gate.math.against.run}{0.00}
\setval{olmo.gdpo.gate.math.ceiling.early}{0.14}
\setval{olmo.gdpo.gate.math.ceiling.late}{0.51}
\setval{olmo.gdpo.gate.math.ceiling.run}{0.42}
\setval{olmo.gdpo.gate.math.cosine.early}{0.90}
\setval{olmo.gdpo.gate.math.cosine.late}{0.61}
\setval{olmo.gdpo.gate.math.cosine.run}{0.67}
\setval{olmo.gdpo.gate.writing-rm.against.early}{0.22}
\setval{olmo.gdpo.gate.writing-rm.against.late}{0.50}
\setval{olmo.gdpo.gate.writing-rm.against.run}{0.43}
\setval{olmo.gdpo.gate.writing-rm.agreement}{+0.094}
\setval{olmo.gdpo.gate.writing-rm.ceiling.early}{0.00}
\setval{olmo.gdpo.gate.writing-rm.ceiling.late}{0.00}
\setval{olmo.gdpo.gate.writing-rm.ceiling.run}{0.00}
\setval{olmo.gdpo.gate.writing-rm.cosine.early}{0.69}
\setval{olmo.gdpo.gate.writing-rm.cosine.late}{0.24}
\setval{olmo.gdpo.gate.writing-rm.cosine.run}{0.37}
\setval{olmo.gdpo.gate.writing.against.early}{0.24}
\setval{olmo.gdpo.gate.writing.against.late}{0.52}
\setval{olmo.gdpo.gate.writing.against.run}{0.47}
\setval{olmo.gdpo.gate.writing.agreement}{$-$0.060}
\setval{olmo.gdpo.gate.writing.ceiling.early}{0.00}
\setval{olmo.gdpo.gate.writing.ceiling.late}{0.00}
\setval{olmo.gdpo.gate.writing.ceiling.run}{0.00}
\setval{olmo.gdpo.gate.writing.cosine.early}{0.60}
\setval{olmo.gdpo.gate.writing.cosine.late}{0.15}
\setval{olmo.gdpo.gate.writing.cosine.run}{0.24}
\setval{olmo.gdpo.judge2.avg}{51.5}
\setval{olmo.gdpo.judge2.creative-writing-v3}{37.8}
\setval{olmo.gdpo.judge2.longbench-write}{55.6}
\setval{olmo.gdpo.judge2.writing}{46.8}
\setval{olmo.gdpo.judge2.writingbench}{47.1}
\setval{olmo.gdpo.last-step}{300}
\setval{olmo.gdpo.minus.base.aime24}{+19.3}
\setval{olmo.gdpo.minus.base.aime24-abs}{19.3}
\setval{olmo.gdpo.minus.base.aime25}{+15.2}
\setval{olmo.gdpo.minus.base.aime25-abs}{15.2}
\setval{olmo.gdpo.minus.base.amc23}{+25.0}
\setval{olmo.gdpo.minus.base.amc23-abs}{25.0}
\setval{olmo.gdpo.minus.base.avg}{+9.8}
\setval{olmo.gdpo.minus.base.avg-abs}{9.8}
\setval{olmo.gdpo.minus.base.code}{+5.6}
\setval{olmo.gdpo.minus.base.code-abs}{5.6}
\setval{olmo.gdpo.minus.base.creative-writing-v3}{+9.2}
\setval{olmo.gdpo.minus.base.creative-writing-v3-abs}{9.2}
\setval{olmo.gdpo.minus.base.guidebench}{$-$10.6}
\setval{olmo.gdpo.minus.base.guidebench-abs}{10.6}
\setval{olmo.gdpo.minus.base.humaneval-plus}{+7.1}
\setval{olmo.gdpo.minus.base.humaneval-plus-abs}{7.1}
\setval{olmo.gdpo.minus.base.if}{$-$1.8}
\setval{olmo.gdpo.minus.base.if-abs}{1.8}
\setval{olmo.gdpo.minus.base.ifbench}{+3.0}
\setval{olmo.gdpo.minus.base.ifbench-abs}{3.0}
\setval{olmo.gdpo.minus.base.ifeval}{+2.0}
\setval{olmo.gdpo.minus.base.ifeval-abs}{2.0}
\setval{olmo.gdpo.minus.base.judge2.avg}{+8.6}
\setval{olmo.gdpo.minus.base.judge2.avg-abs}{8.6}
\setval{olmo.gdpo.minus.base.judge2.creative-writing-v3}{+6.2}
\setval{olmo.gdpo.minus.base.judge2.creative-writing-v3-abs}{6.2}
\setval{olmo.gdpo.minus.base.judge2.longbench-write}{+2.5}
\setval{olmo.gdpo.minus.base.judge2.longbench-write-abs}{2.5}
\setval{olmo.gdpo.minus.base.judge2.writing}{+6.0}
\setval{olmo.gdpo.minus.base.judge2.writing-abs}{6.0}
\setval{olmo.gdpo.minus.base.judge2.writingbench}{+9.4}
\setval{olmo.gdpo.minus.base.judge2.writingbench-abs}{9.4}
\setval{olmo.gdpo.minus.base.livecodebench}{+6.3}
\setval{olmo.gdpo.minus.base.livecodebench-abs}{6.3}
\setval{olmo.gdpo.minus.base.longbench-write}{+5.8}
\setval{olmo.gdpo.minus.base.longbench-write-abs}{5.8}
\setval{olmo.gdpo.minus.base.math}{+16.5}
\setval{olmo.gdpo.minus.base.math-abs}{16.5}
\setval{olmo.gdpo.minus.base.math500}{+15.1}
\setval{olmo.gdpo.minus.base.math500-abs}{15.1}
\setval{olmo.gdpo.minus.base.mbpp-plus}{+3.5}
\setval{olmo.gdpo.minus.base.mbpp-plus-abs}{3.5}
\setval{olmo.gdpo.minus.base.minerva}{+5.1}
\setval{olmo.gdpo.minus.base.minerva-abs}{5.1}
\setval{olmo.gdpo.minus.base.olympiad-bench}{+19.6}
\setval{olmo.gdpo.minus.base.olympiad-bench-abs}{19.6}
\setval{olmo.gdpo.minus.base.writing}{+11.9}
\setval{olmo.gdpo.minus.base.writing-abs}{11.9}
\setval{olmo.gdpo.minus.base.writingbench}{+20.6}
\setval{olmo.gdpo.minus.base.writingbench-abs}{20.6}
\setval{olmo.gdpo.minus.concur-free.aime24}{+1.0}
\setval{olmo.gdpo.minus.concur-free.aime24-abs}{1.0}
\setval{olmo.gdpo.minus.concur-free.aime25}{+0.5}
\setval{olmo.gdpo.minus.concur-free.aime25-abs}{0.5}
\setval{olmo.gdpo.minus.concur-free.amc23}{+2.0}
\setval{olmo.gdpo.minus.concur-free.amc23-abs}{2.0}
\setval{olmo.gdpo.minus.concur-free.avg}{$-$1.6}
\setval{olmo.gdpo.minus.concur-free.avg-abs}{1.6}
\setval{olmo.gdpo.minus.concur-free.code}{$-$0.9}
\setval{olmo.gdpo.minus.concur-free.code-abs}{0.9}
\setval{olmo.gdpo.minus.concur-free.creative-writing-v3}{$-$2.6}
\setval{olmo.gdpo.minus.concur-free.creative-writing-v3-abs}{2.6}
\setval{olmo.gdpo.minus.concur-free.guidebench}{$-$9.3}
\setval{olmo.gdpo.minus.concur-free.guidebench-abs}{9.3}
\setval{olmo.gdpo.minus.concur-free.humaneval-plus}{$-$0.3}
\setval{olmo.gdpo.minus.concur-free.humaneval-plus-abs}{0.3}
\setval{olmo.gdpo.minus.concur-free.if}{$-$4.0}
\setval{olmo.gdpo.minus.concur-free.if-abs}{4.0}
\setval{olmo.gdpo.minus.concur-free.ifbench}{$-$2.0}
\setval{olmo.gdpo.minus.concur-free.ifbench-abs}{2.0}
\setval{olmo.gdpo.minus.concur-free.ifeval}{$-$0.8}
\setval{olmo.gdpo.minus.concur-free.ifeval-abs}{0.8}
\setval{olmo.gdpo.minus.concur-free.judge2.avg}{$-$1.1}
\setval{olmo.gdpo.minus.concur-free.judge2.avg-abs}{1.1}
\setval{olmo.gdpo.minus.concur-free.judge2.creative-writing-v3}{+0.3}
\setval{olmo.gdpo.minus.concur-free.judge2.creative-writing-v3-abs}{0.3}
\setval{olmo.gdpo.minus.concur-free.judge2.longbench-write}{$-$8.3}
\setval{olmo.gdpo.minus.concur-free.judge2.longbench-write-abs}{8.3}
\setval{olmo.gdpo.minus.concur-free.judge2.writing}{$-$3.0}
\setval{olmo.gdpo.minus.concur-free.judge2.writing-abs}{3.0}
\setval{olmo.gdpo.minus.concur-free.judge2.writingbench}{$-$1.0}
\setval{olmo.gdpo.minus.concur-free.judge2.writingbench-abs}{1.0}
\setval{olmo.gdpo.minus.concur-free.livecodebench}{$-$2.5}
\setval{olmo.gdpo.minus.concur-free.livecodebench-abs}{2.5}
\setval{olmo.gdpo.minus.concur-free.longbench-write}{$-$8.8}
\setval{olmo.gdpo.minus.concur-free.longbench-write-abs}{8.8}
\setval{olmo.gdpo.minus.concur-free.math}{+1.3}
\setval{olmo.gdpo.minus.concur-free.math-abs}{1.3}
\setval{olmo.gdpo.minus.concur-free.math500}{+0.9}
\setval{olmo.gdpo.minus.concur-free.math500-abs}{0.9}
\setval{olmo.gdpo.minus.concur-free.mbpp-plus}{+0.3}
\setval{olmo.gdpo.minus.concur-free.mbpp-plus-abs}{0.3}
\setval{olmo.gdpo.minus.concur-free.minerva}{+0.4}
\setval{olmo.gdpo.minus.concur-free.minerva-abs}{0.4}
\setval{olmo.gdpo.minus.concur-free.olympiad-bench}{+2.6}
\setval{olmo.gdpo.minus.concur-free.olympiad-bench-abs}{2.6}
\setval{olmo.gdpo.minus.concur-free.writing}{$-$5.4}
\setval{olmo.gdpo.minus.concur-free.writing-abs}{5.4}
\setval{olmo.gdpo.minus.concur-free.writingbench}{$-$4.9}
\setval{olmo.gdpo.minus.concur-free.writingbench-abs}{4.9}
\setval{olmo.gdpo.minus.concur-tied.aime24}{$-$6.7}
\setval{olmo.gdpo.minus.concur-tied.aime24-abs}{6.7}
\setval{olmo.gdpo.minus.concur-tied.aime25}{$-$7.3}
\setval{olmo.gdpo.minus.concur-tied.aime25-abs}{7.3}
\setval{olmo.gdpo.minus.concur-tied.amc23}{$-$10.3}
\setval{olmo.gdpo.minus.concur-tied.amc23-abs}{10.3}
\setval{olmo.gdpo.minus.concur-tied.avg}{$-$5.2}
\setval{olmo.gdpo.minus.concur-tied.avg-abs}{5.2}
\setval{olmo.gdpo.minus.concur-tied.code}{$-$6.4}
\setval{olmo.gdpo.minus.concur-tied.code-abs}{6.4}
\setval{olmo.gdpo.minus.concur-tied.creative-writing-v3}{$-$1.9}
\setval{olmo.gdpo.minus.concur-tied.creative-writing-v3-abs}{1.9}
\setval{olmo.gdpo.minus.concur-tied.guidebench}{$-$7.7}
\setval{olmo.gdpo.minus.concur-tied.guidebench-abs}{7.7}
\setval{olmo.gdpo.minus.concur-tied.humaneval-plus}{$-$7.0}
\setval{olmo.gdpo.minus.concur-tied.humaneval-plus-abs}{7.0}
\setval{olmo.gdpo.minus.concur-tied.if}{$-$4.3}
\setval{olmo.gdpo.minus.concur-tied.if-abs}{4.3}
\setval{olmo.gdpo.minus.concur-tied.ifbench}{$-$4.0}
\setval{olmo.gdpo.minus.concur-tied.ifbench-abs}{4.0}
\setval{olmo.gdpo.minus.concur-tied.ifeval}{$-$1.2}
\setval{olmo.gdpo.minus.concur-tied.ifeval-abs}{1.2}
\setval{olmo.gdpo.minus.concur-tied.judge2.avg}{$-$4.8}
\setval{olmo.gdpo.minus.concur-tied.judge2.avg-abs}{4.8}
\setval{olmo.gdpo.minus.concur-tied.judge2.creative-writing-v3}{+0.6}
\setval{olmo.gdpo.minus.concur-tied.judge2.creative-writing-v3-abs}{0.6}
\setval{olmo.gdpo.minus.concur-tied.judge2.longbench-write}{$-$4.9}
\setval{olmo.gdpo.minus.concur-tied.judge2.longbench-write-abs}{4.9}
\setval{olmo.gdpo.minus.concur-tied.judge2.writing}{$-$0.9}
\setval{olmo.gdpo.minus.concur-tied.judge2.writing-abs}{0.9}
\setval{olmo.gdpo.minus.concur-tied.judge2.writingbench}{+1.5}
\setval{olmo.gdpo.minus.concur-tied.judge2.writingbench-abs}{1.5}
\setval{olmo.gdpo.minus.concur-tied.livecodebench}{$-$5.9}
\setval{olmo.gdpo.minus.concur-tied.livecodebench-abs}{5.9}
\setval{olmo.gdpo.minus.concur-tied.longbench-write}{$-$6.0}
\setval{olmo.gdpo.minus.concur-tied.longbench-write-abs}{6.0}
\setval{olmo.gdpo.minus.concur-tied.math}{$-$6.3}
\setval{olmo.gdpo.minus.concur-tied.math-abs}{6.3}
\setval{olmo.gdpo.minus.concur-tied.math500}{$-$4.2}
\setval{olmo.gdpo.minus.concur-tied.math500-abs}{4.2}
\setval{olmo.gdpo.minus.concur-tied.mbpp-plus}{$-$6.3}
\setval{olmo.gdpo.minus.concur-tied.mbpp-plus-abs}{6.3}
\setval{olmo.gdpo.minus.concur-tied.minerva}{$-$3.7}
\setval{olmo.gdpo.minus.concur-tied.minerva-abs}{3.7}
\setval{olmo.gdpo.minus.concur-tied.olympiad-bench}{$-$5.7}
\setval{olmo.gdpo.minus.concur-tied.olympiad-bench-abs}{5.7}
\setval{olmo.gdpo.minus.concur-tied.writing}{$-$2.9}
\setval{olmo.gdpo.minus.concur-tied.writing-abs}{2.9}
\setval{olmo.gdpo.minus.concur-tied.writingbench}{$-$0.8}
\setval{olmo.gdpo.minus.concur-tied.writingbench-abs}{0.8}
\setval{olmo.gdpo.minus.concur.aime24}{$-$8.5}
\setval{olmo.gdpo.minus.concur.aime24-abs}{8.5}
\setval{olmo.gdpo.minus.concur.aime25}{$-$9.6}
\setval{olmo.gdpo.minus.concur.aime25-abs}{9.6}
\setval{olmo.gdpo.minus.concur.amc23}{$-$14.0}
\setval{olmo.gdpo.minus.concur.amc23-abs}{14.0}
\setval{olmo.gdpo.minus.concur.avg}{$-$5.5}
\setval{olmo.gdpo.minus.concur.avg-abs}{5.5}
\setval{olmo.gdpo.minus.concur.code}{$-$5.5}
\setval{olmo.gdpo.minus.concur.code-abs}{5.5}
\setval{olmo.gdpo.minus.concur.creative-writing-v3}{$-$2.1}
\setval{olmo.gdpo.minus.concur.creative-writing-v3-abs}{2.1}
\setval{olmo.gdpo.minus.concur.guidebench}{$-$6.0}
\setval{olmo.gdpo.minus.concur.guidebench-abs}{6.0}
\setval{olmo.gdpo.minus.concur.humaneval-plus}{$-$6.4}
\setval{olmo.gdpo.minus.concur.humaneval-plus-abs}{6.4}
\setval{olmo.gdpo.minus.concur.if}{$-$3.2}
\setval{olmo.gdpo.minus.concur.if-abs}{3.2}
\setval{olmo.gdpo.minus.concur.ifbench}{$-$2.3}
\setval{olmo.gdpo.minus.concur.ifbench-abs}{2.3}
\setval{olmo.gdpo.minus.concur.ifeval}{$-$1.3}
\setval{olmo.gdpo.minus.concur.ifeval-abs}{1.3}
\setval{olmo.gdpo.minus.concur.judge2.avg}{$-$5.0}
\setval{olmo.gdpo.minus.concur.judge2.avg-abs}{5.0}
\setval{olmo.gdpo.minus.concur.judge2.creative-writing-v3}{$-$0.6}
\setval{olmo.gdpo.minus.concur.judge2.creative-writing-v3-abs}{0.6}
\setval{olmo.gdpo.minus.concur.judge2.longbench-write}{$-$2.7}
\setval{olmo.gdpo.minus.concur.judge2.longbench-write-abs}{2.7}
\setval{olmo.gdpo.minus.concur.judge2.writing}{$-$0.4}
\setval{olmo.gdpo.minus.concur.judge2.writing-abs}{0.4}
\setval{olmo.gdpo.minus.concur.judge2.writingbench}{+2.2}
\setval{olmo.gdpo.minus.concur.judge2.writingbench-abs}{2.2}
\setval{olmo.gdpo.minus.concur.livecodebench}{$-$7.2}
\setval{olmo.gdpo.minus.concur.livecodebench-abs}{7.2}
\setval{olmo.gdpo.minus.concur.longbench-write}{$-$5.2}
\setval{olmo.gdpo.minus.concur.longbench-write-abs}{5.2}
\setval{olmo.gdpo.minus.concur.math}{$-$8.1}
\setval{olmo.gdpo.minus.concur.math-abs}{8.1}
\setval{olmo.gdpo.minus.concur.math500}{$-$5.3}
\setval{olmo.gdpo.minus.concur.math500-abs}{5.3}
\setval{olmo.gdpo.minus.concur.mbpp-plus}{$-$2.8}
\setval{olmo.gdpo.minus.concur.mbpp-plus-abs}{2.8}
\setval{olmo.gdpo.minus.concur.minerva}{$-$4.4}
\setval{olmo.gdpo.minus.concur.minerva-abs}{4.4}
\setval{olmo.gdpo.minus.concur.olympiad-bench}{$-$6.5}
\setval{olmo.gdpo.minus.concur.olympiad-bench-abs}{6.5}
\setval{olmo.gdpo.minus.concur.writing}{$-$2.5}
\setval{olmo.gdpo.minus.concur.writing-abs}{2.5}
\setval{olmo.gdpo.minus.concur.writingbench}{$-$0.1}
\setval{olmo.gdpo.minus.concur.writingbench-abs}{0.1}
\setval{olmo.gdpo.minus.gd2po.aime24}{$-$0.5}
\setval{olmo.gdpo.minus.gd2po.aime24-abs}{0.5}
\setval{olmo.gdpo.minus.gd2po.aime25}{$-$1.7}
\setval{olmo.gdpo.minus.gd2po.aime25-abs}{1.7}
\setval{olmo.gdpo.minus.gd2po.amc23}{$-$2.7}
\setval{olmo.gdpo.minus.gd2po.amc23-abs}{2.7}
\setval{olmo.gdpo.minus.gd2po.avg}{$-$1.8}
\setval{olmo.gdpo.minus.gd2po.avg-abs}{1.8}
\setval{olmo.gdpo.minus.gd2po.code}{$-$0.2}
\setval{olmo.gdpo.minus.gd2po.code-abs}{0.2}
\setval{olmo.gdpo.minus.gd2po.creative-writing-v3}{$-$1.2}
\setval{olmo.gdpo.minus.gd2po.creative-writing-v3-abs}{1.2}
\setval{olmo.gdpo.minus.gd2po.guidebench}{$-$10.3}
\setval{olmo.gdpo.minus.gd2po.guidebench-abs}{10.3}
\setval{olmo.gdpo.minus.gd2po.humaneval-plus}{$-$0.4}
\setval{olmo.gdpo.minus.gd2po.humaneval-plus-abs}{0.4}
\setval{olmo.gdpo.minus.gd2po.if}{$-$4.7}
\setval{olmo.gdpo.minus.gd2po.if-abs}{4.7}
\setval{olmo.gdpo.minus.gd2po.ifbench}{$-$2.7}
\setval{olmo.gdpo.minus.gd2po.ifbench-abs}{2.7}
\setval{olmo.gdpo.minus.gd2po.ifeval}{$-$1.2}
\setval{olmo.gdpo.minus.gd2po.ifeval-abs}{1.2}
\setval{olmo.gdpo.minus.gd2po.judge2.avg}{$-$1.7}
\setval{olmo.gdpo.minus.gd2po.judge2.avg-abs}{1.7}
\setval{olmo.gdpo.minus.gd2po.judge2.creative-writing-v3}{$-$0.1}
\setval{olmo.gdpo.minus.gd2po.judge2.creative-writing-v3-abs}{0.1}
\setval{olmo.gdpo.minus.gd2po.judge2.longbench-write}{$-$4.3}
\setval{olmo.gdpo.minus.gd2po.judge2.longbench-write-abs}{4.3}
\setval{olmo.gdpo.minus.gd2po.judge2.writing}{$-$1.9}
\setval{olmo.gdpo.minus.gd2po.judge2.writing-abs}{1.9}
\setval{olmo.gdpo.minus.gd2po.judge2.writingbench}{$-$1.3}
\setval{olmo.gdpo.minus.gd2po.judge2.writingbench-abs}{1.3}
\setval{olmo.gdpo.minus.gd2po.livecodebench}{$-$0.1}
\setval{olmo.gdpo.minus.gd2po.livecodebench-abs}{0.1}
\setval{olmo.gdpo.minus.gd2po.longbench-write}{$-$4.9}
\setval{olmo.gdpo.minus.gd2po.longbench-write-abs}{4.9}
\setval{olmo.gdpo.minus.gd2po.math}{$-$0.9}
\setval{olmo.gdpo.minus.gd2po.math-abs}{0.9}
\setval{olmo.gdpo.minus.gd2po.math500}{$-$0.5}
\setval{olmo.gdpo.minus.gd2po.math500-abs}{0.5}
\setval{olmo.gdpo.minus.gd2po.mbpp-plus}{+0.0}
\setval{olmo.gdpo.minus.gd2po.mbpp-plus-abs}{0.0}
\setval{olmo.gdpo.minus.gd2po.minerva}{+0.2}
\setval{olmo.gdpo.minus.gd2po.minerva-abs}{0.2}
\setval{olmo.gdpo.minus.gd2po.olympiad-bench}{$-$0.2}
\setval{olmo.gdpo.minus.gd2po.olympiad-bench-abs}{0.2}
\setval{olmo.gdpo.minus.gd2po.writing}{$-$2.2}
\setval{olmo.gdpo.minus.gd2po.writing-abs}{2.2}
\setval{olmo.gdpo.minus.gd2po.writingbench}{$-$0.4}
\setval{olmo.gdpo.minus.gd2po.writingbench-abs}{0.4}
\setval{olmo.gdpo.minus.gdpo-noaux.aime24}{$-$10.5}
\setval{olmo.gdpo.minus.gdpo-noaux.aime24-abs}{10.5}
\setval{olmo.gdpo.minus.gdpo-noaux.aime25}{$-$11.0}
\setval{olmo.gdpo.minus.gdpo-noaux.aime25-abs}{11.0}
\setval{olmo.gdpo.minus.gdpo-noaux.amc23}{$-$14.4}
\setval{olmo.gdpo.minus.gdpo-noaux.amc23-abs}{14.4}
\setval{olmo.gdpo.minus.gdpo-noaux.avg}{$-$3.9}
\setval{olmo.gdpo.minus.gdpo-noaux.avg-abs}{3.9}
\setval{olmo.gdpo.minus.gdpo-noaux.code}{$-$7.7}
\setval{olmo.gdpo.minus.gdpo-noaux.code-abs}{7.7}
\setval{olmo.gdpo.minus.gdpo-noaux.creative-writing-v3}{+8.9}
\setval{olmo.gdpo.minus.gdpo-noaux.creative-writing-v3-abs}{8.9}
\setval{olmo.gdpo.minus.gdpo-noaux.guidebench}{$-$4.9}
\setval{olmo.gdpo.minus.gdpo-noaux.guidebench-abs}{4.9}
\setval{olmo.gdpo.minus.gdpo-noaux.humaneval-plus}{$-$8.8}
\setval{olmo.gdpo.minus.gdpo-noaux.humaneval-plus-abs}{8.8}
\setval{olmo.gdpo.minus.gdpo-noaux.if}{$-$2.3}
\setval{olmo.gdpo.minus.gdpo-noaux.if-abs}{2.3}
\setval{olmo.gdpo.minus.gdpo-noaux.ifbench}{$-$3.7}
\setval{olmo.gdpo.minus.gdpo-noaux.ifbench-abs}{3.7}
\setval{olmo.gdpo.minus.gdpo-noaux.ifeval}{+1.7}
\setval{olmo.gdpo.minus.gdpo-noaux.ifeval-abs}{1.7}
\setval{olmo.gdpo.minus.gdpo-noaux.judge2.avg}{$-$4.1}
\setval{olmo.gdpo.minus.gdpo-noaux.judge2.avg-abs}{4.1}
\setval{olmo.gdpo.minus.gdpo-noaux.judge2.creative-writing-v3}{+6.9}
\setval{olmo.gdpo.minus.gdpo-noaux.judge2.creative-writing-v3-abs}{6.9}
\setval{olmo.gdpo.minus.gdpo-noaux.judge2.longbench-write}{+9.0}
\setval{olmo.gdpo.minus.gdpo-noaux.judge2.longbench-write-abs}{9.0}
\setval{olmo.gdpo.minus.gdpo-noaux.judge2.writing}{+7.7}
\setval{olmo.gdpo.minus.gdpo-noaux.judge2.writing-abs}{7.7}
\setval{olmo.gdpo.minus.gdpo-noaux.judge2.writingbench}{+7.3}
\setval{olmo.gdpo.minus.gdpo-noaux.judge2.writingbench-abs}{7.3}
\setval{olmo.gdpo.minus.gdpo-noaux.livecodebench}{$-$8.4}
\setval{olmo.gdpo.minus.gdpo-noaux.livecodebench-abs}{8.4}
\setval{olmo.gdpo.minus.gdpo-noaux.longbench-write}{+7.3}
\setval{olmo.gdpo.minus.gdpo-noaux.longbench-write-abs}{7.3}
\setval{olmo.gdpo.minus.gdpo-noaux.math}{$-$9.2}
\setval{olmo.gdpo.minus.gdpo-noaux.math-abs}{9.2}
\setval{olmo.gdpo.minus.gdpo-noaux.math500}{$-$5.6}
\setval{olmo.gdpo.minus.gdpo-noaux.math500-abs}{5.6}
\setval{olmo.gdpo.minus.gdpo-noaux.mbpp-plus}{$-$5.7}
\setval{olmo.gdpo.minus.gdpo-noaux.mbpp-plus-abs}{5.7}
\setval{olmo.gdpo.minus.gdpo-noaux.minerva}{$-$5.8}
\setval{olmo.gdpo.minus.gdpo-noaux.minerva-abs}{5.8}
\setval{olmo.gdpo.minus.gdpo-noaux.olympiad-bench}{$-$7.8}
\setval{olmo.gdpo.minus.gdpo-noaux.olympiad-bench-abs}{7.8}
\setval{olmo.gdpo.minus.gdpo-noaux.writing}{+8.7}
\setval{olmo.gdpo.minus.gdpo-noaux.writing-abs}{8.7}
\setval{olmo.gdpo.minus.gdpo-noaux.writingbench}{+9.8}
\setval{olmo.gdpo.minus.gdpo-noaux.writingbench-abs}{9.8}
\setval{olmo.gdpo.minus.gdpo-w0.5.aime24}{$-$1.8}
\setval{olmo.gdpo.minus.gdpo-w0.5.aime24-abs}{1.8}
\setval{olmo.gdpo.minus.gdpo-w0.5.aime25}{$-$4.7}
\setval{olmo.gdpo.minus.gdpo-w0.5.aime25-abs}{4.7}
\setval{olmo.gdpo.minus.gdpo-w0.5.amc23}{$-$4.5}
\setval{olmo.gdpo.minus.gdpo-w0.5.amc23-abs}{4.5}
\setval{olmo.gdpo.minus.gdpo-w0.5.avg}{$-$1.8}
\setval{olmo.gdpo.minus.gdpo-w0.5.avg-abs}{1.8}
\setval{olmo.gdpo.minus.gdpo-w0.5.code}{$-$1.1}
\setval{olmo.gdpo.minus.gdpo-w0.5.code-abs}{1.1}
\setval{olmo.gdpo.minus.gdpo-w0.5.creative-writing-v3}{$-$1.0}
\setval{olmo.gdpo.minus.gdpo-w0.5.creative-writing-v3-abs}{1.0}
\setval{olmo.gdpo.minus.gdpo-w0.5.guidebench}{$-$1.1}
\setval{olmo.gdpo.minus.gdpo-w0.5.guidebench-abs}{1.1}
\setval{olmo.gdpo.minus.gdpo-w0.5.humaneval-plus}{$-$1.4}
\setval{olmo.gdpo.minus.gdpo-w0.5.humaneval-plus-abs}{1.4}
\setval{olmo.gdpo.minus.gdpo-w0.5.if}{$-$1.7}
\setval{olmo.gdpo.minus.gdpo-w0.5.if-abs}{1.7}
\setval{olmo.gdpo.minus.gdpo-w0.5.ifbench}{$-$3.0}
\setval{olmo.gdpo.minus.gdpo-w0.5.ifbench-abs}{3.0}
\setval{olmo.gdpo.minus.gdpo-w0.5.ifeval}{$-$1.1}
\setval{olmo.gdpo.minus.gdpo-w0.5.ifeval-abs}{1.1}
\setval{olmo.gdpo.minus.gdpo-w0.5.judge2.avg}{$-$1.7}
\setval{olmo.gdpo.minus.gdpo-w0.5.judge2.avg-abs}{1.7}
\setval{olmo.gdpo.minus.gdpo-w0.5.judge2.creative-writing-v3}{$-$1.6}
\setval{olmo.gdpo.minus.gdpo-w0.5.judge2.creative-writing-v3-abs}{1.6}
\setval{olmo.gdpo.minus.gdpo-w0.5.judge2.longbench-write}{$-$0.9}
\setval{olmo.gdpo.minus.gdpo-w0.5.judge2.longbench-write-abs}{0.9}
\setval{olmo.gdpo.minus.gdpo-w0.5.judge2.writing}{$-$0.7}
\setval{olmo.gdpo.minus.gdpo-w0.5.judge2.writing-abs}{0.7}
\setval{olmo.gdpo.minus.gdpo-w0.5.judge2.writingbench}{+0.4}
\setval{olmo.gdpo.minus.gdpo-w0.5.judge2.writingbench-abs}{0.4}
\setval{olmo.gdpo.minus.gdpo-w0.5.livecodebench}{$-$1.7}
\setval{olmo.gdpo.minus.gdpo-w0.5.livecodebench-abs}{1.7}
\setval{olmo.gdpo.minus.gdpo-w0.5.longbench-write}{$-$0.5}
\setval{olmo.gdpo.minus.gdpo-w0.5.longbench-write-abs}{0.5}
\setval{olmo.gdpo.minus.gdpo-w0.5.math}{$-$2.6}
\setval{olmo.gdpo.minus.gdpo-w0.5.math-abs}{2.6}
\setval{olmo.gdpo.minus.gdpo-w0.5.math500}{$-$1.8}
\setval{olmo.gdpo.minus.gdpo-w0.5.math500-abs}{1.8}
\setval{olmo.gdpo.minus.gdpo-w0.5.mbpp-plus}{$-$0.2}
\setval{olmo.gdpo.minus.gdpo-w0.5.mbpp-plus-abs}{0.2}
\setval{olmo.gdpo.minus.gdpo-w0.5.minerva}{$-$1.2}
\setval{olmo.gdpo.minus.gdpo-w0.5.minerva-abs}{1.2}
\setval{olmo.gdpo.minus.gdpo-w0.5.olympiad-bench}{$-$1.6}
\setval{olmo.gdpo.minus.gdpo-w0.5.olympiad-bench-abs}{1.6}
\setval{olmo.gdpo.minus.gdpo-w0.5.writing}{$-$0.9}
\setval{olmo.gdpo.minus.gdpo-w0.5.writing-abs}{0.9}
\setval{olmo.gdpo.minus.gdpo-w0.5.writingbench}{$-$1.2}
\setval{olmo.gdpo.minus.gdpo-w0.5.writingbench-abs}{1.2}
\setval{olmo.gdpo.minus.geoma.aime24}{+4.1}
\setval{olmo.gdpo.minus.geoma.aime24-abs}{4.1}
\setval{olmo.gdpo.minus.geoma.aime25}{+0.6}
\setval{olmo.gdpo.minus.geoma.aime25-abs}{0.6}
\setval{olmo.gdpo.minus.geoma.amc23}{+5.6}
\setval{olmo.gdpo.minus.geoma.amc23-abs}{5.6}
\setval{olmo.gdpo.minus.geoma.avg}{+1.0}
\setval{olmo.gdpo.minus.geoma.avg-abs}{1.0}
\setval{olmo.gdpo.minus.geoma.code}{+0.8}
\setval{olmo.gdpo.minus.geoma.code-abs}{0.8}
\setval{olmo.gdpo.minus.geoma.creative-writing-v3}{$-$2.5}
\setval{olmo.gdpo.minus.geoma.creative-writing-v3-abs}{2.5}
\setval{olmo.gdpo.minus.geoma.guidebench}{+4.2}
\setval{olmo.gdpo.minus.geoma.guidebench-abs}{4.2}
\setval{olmo.gdpo.minus.geoma.humaneval-plus}{+1.7}
\setval{olmo.gdpo.minus.geoma.humaneval-plus-abs}{1.7}
\setval{olmo.gdpo.minus.geoma.if}{+0.7}
\setval{olmo.gdpo.minus.geoma.if-abs}{0.7}
\setval{olmo.gdpo.minus.geoma.ifbench}{$-$0.7}
\setval{olmo.gdpo.minus.geoma.ifbench-abs}{0.7}
\setval{olmo.gdpo.minus.geoma.ifeval}{$-$1.6}
\setval{olmo.gdpo.minus.geoma.ifeval-abs}{1.6}
\setval{olmo.gdpo.minus.geoma.judge2.avg}{+0.8}
\setval{olmo.gdpo.minus.geoma.judge2.avg-abs}{0.8}
\setval{olmo.gdpo.minus.geoma.judge2.creative-writing-v3}{$-$3.9}
\setval{olmo.gdpo.minus.geoma.judge2.creative-writing-v3-abs}{3.9}
\setval{olmo.gdpo.minus.geoma.judge2.longbench-write}{$-$2.5}
\setval{olmo.gdpo.minus.geoma.judge2.longbench-write-abs}{2.5}
\setval{olmo.gdpo.minus.geoma.judge2.writing}{$-$2.7}
\setval{olmo.gdpo.minus.geoma.judge2.writing-abs}{2.7}
\setval{olmo.gdpo.minus.geoma.judge2.writingbench}{$-$1.6}
\setval{olmo.gdpo.minus.geoma.judge2.writingbench-abs}{1.6}
\setval{olmo.gdpo.minus.geoma.livecodebench}{+0.8}
\setval{olmo.gdpo.minus.geoma.livecodebench-abs}{0.8}
\setval{olmo.gdpo.minus.geoma.longbench-write}{$-$2.1}
\setval{olmo.gdpo.minus.geoma.longbench-write-abs}{2.1}
\setval{olmo.gdpo.minus.geoma.math}{+2.7}
\setval{olmo.gdpo.minus.geoma.math-abs}{2.7}
\setval{olmo.gdpo.minus.geoma.math500}{+2.1}
\setval{olmo.gdpo.minus.geoma.math500-abs}{2.1}
\setval{olmo.gdpo.minus.geoma.mbpp-plus}{+0.0}
\setval{olmo.gdpo.minus.geoma.mbpp-plus-abs}{0.0}
\setval{olmo.gdpo.minus.geoma.minerva}{+0.4}
\setval{olmo.gdpo.minus.geoma.minerva-abs}{0.4}
\setval{olmo.gdpo.minus.geoma.olympiad-bench}{+3.1}
\setval{olmo.gdpo.minus.geoma.olympiad-bench-abs}{3.1}
\setval{olmo.gdpo.minus.geoma.writing}{$-$2.1}
\setval{olmo.gdpo.minus.geoma.writing-abs}{2.1}
\setval{olmo.gdpo.minus.geoma.writingbench}{$-$1.7}
\setval{olmo.gdpo.minus.geoma.writingbench-abs}{1.7}
\setval{olmo.gdpo.minus.grpo-nodiv.aime24}{$-$4.0}
\setval{olmo.gdpo.minus.grpo-nodiv.aime24-abs}{4.0}
\setval{olmo.gdpo.minus.grpo-nodiv.aime25}{$-$5.0}
\setval{olmo.gdpo.minus.grpo-nodiv.aime25-abs}{5.0}
\setval{olmo.gdpo.minus.grpo-nodiv.amc23}{$-$3.5}
\setval{olmo.gdpo.minus.grpo-nodiv.amc23-abs}{3.5}
\setval{olmo.gdpo.minus.grpo-nodiv.avg}{$-$2.0}
\setval{olmo.gdpo.minus.grpo-nodiv.avg-abs}{2.0}
\setval{olmo.gdpo.minus.grpo-nodiv.code}{$-$0.7}
\setval{olmo.gdpo.minus.grpo-nodiv.code-abs}{0.7}
\setval{olmo.gdpo.minus.grpo-nodiv.creative-writing-v3}{+0.2}
\setval{olmo.gdpo.minus.grpo-nodiv.creative-writing-v3-abs}{0.2}
\setval{olmo.gdpo.minus.grpo-nodiv.guidebench}{$-$4.6}
\setval{olmo.gdpo.minus.grpo-nodiv.guidebench-abs}{4.6}
\setval{olmo.gdpo.minus.grpo-nodiv.humaneval-plus}{$-$1.0}
\setval{olmo.gdpo.minus.grpo-nodiv.humaneval-plus-abs}{1.0}
\setval{olmo.gdpo.minus.grpo-nodiv.if}{$-$2.5}
\setval{olmo.gdpo.minus.grpo-nodiv.if-abs}{2.5}
\setval{olmo.gdpo.minus.grpo-nodiv.ifbench}{$-$1.7}
\setval{olmo.gdpo.minus.grpo-nodiv.ifbench-abs}{1.7}
\setval{olmo.gdpo.minus.grpo-nodiv.ifeval}{$-$1.3}
\setval{olmo.gdpo.minus.grpo-nodiv.ifeval-abs}{1.3}
\setval{olmo.gdpo.minus.grpo-nodiv.judge2.avg}{$-$2.3}
\setval{olmo.gdpo.minus.grpo-nodiv.judge2.avg-abs}{2.3}
\setval{olmo.gdpo.minus.grpo-nodiv.judge2.creative-writing-v3}{$-$1.0}
\setval{olmo.gdpo.minus.grpo-nodiv.judge2.creative-writing-v3-abs}{1.0}
\setval{olmo.gdpo.minus.grpo-nodiv.judge2.longbench-write}{$-$5.5}
\setval{olmo.gdpo.minus.grpo-nodiv.judge2.longbench-write-abs}{5.5}
\setval{olmo.gdpo.minus.grpo-nodiv.judge2.writing}{$-$2.1}
\setval{olmo.gdpo.minus.grpo-nodiv.judge2.writing-abs}{2.1}
\setval{olmo.gdpo.minus.grpo-nodiv.judge2.writingbench}{+0.1}
\setval{olmo.gdpo.minus.grpo-nodiv.judge2.writingbench-abs}{0.1}
\setval{olmo.gdpo.minus.grpo-nodiv.livecodebench}{$-$0.5}
\setval{olmo.gdpo.minus.grpo-nodiv.livecodebench-abs}{0.5}
\setval{olmo.gdpo.minus.grpo-nodiv.longbench-write}{$-$4.3}
\setval{olmo.gdpo.minus.grpo-nodiv.longbench-write-abs}{4.3}
\setval{olmo.gdpo.minus.grpo-nodiv.math}{$-$3.1}
\setval{olmo.gdpo.minus.grpo-nodiv.math-abs}{3.1}
\setval{olmo.gdpo.minus.grpo-nodiv.math500}{$-$2.3}
\setval{olmo.gdpo.minus.grpo-nodiv.math500-abs}{2.3}
\setval{olmo.gdpo.minus.grpo-nodiv.mbpp-plus}{$-$0.6}
\setval{olmo.gdpo.minus.grpo-nodiv.mbpp-plus-abs}{0.6}
\setval{olmo.gdpo.minus.grpo-nodiv.minerva}{$-$1.4}
\setval{olmo.gdpo.minus.grpo-nodiv.minerva-abs}{1.4}
\setval{olmo.gdpo.minus.grpo-nodiv.olympiad-bench}{$-$2.6}
\setval{olmo.gdpo.minus.grpo-nodiv.olympiad-bench-abs}{2.6}
\setval{olmo.gdpo.minus.grpo-nodiv.writing}{$-$0.7}
\setval{olmo.gdpo.minus.grpo-nodiv.writing-abs}{0.7}
\setval{olmo.gdpo.minus.grpo-nodiv.writingbench}{+2.1}
\setval{olmo.gdpo.minus.grpo-nodiv.writingbench-abs}{2.1}
\setval{olmo.gdpo.minus.grpo.aime24}{$-$1.7}
\setval{olmo.gdpo.minus.grpo.aime24-abs}{1.7}
\setval{olmo.gdpo.minus.grpo.aime25}{$-$2.1}
\setval{olmo.gdpo.minus.grpo.aime25-abs}{2.1}
\setval{olmo.gdpo.minus.grpo.amc23}{$-$1.3}
\setval{olmo.gdpo.minus.grpo.amc23-abs}{1.3}
\setval{olmo.gdpo.minus.grpo.avg}{$-$1.4}
\setval{olmo.gdpo.minus.grpo.avg-abs}{1.4}
\setval{olmo.gdpo.minus.grpo.code}{$-$0.4}
\setval{olmo.gdpo.minus.grpo.code-abs}{0.4}
\setval{olmo.gdpo.minus.grpo.creative-writing-v3}{$-$2.2}
\setval{olmo.gdpo.minus.grpo.creative-writing-v3-abs}{2.2}
\setval{olmo.gdpo.minus.grpo.guidebench}{$-$5.6}
\setval{olmo.gdpo.minus.grpo.guidebench-abs}{5.6}
\setval{olmo.gdpo.minus.grpo.humaneval-plus}{$-$1.0}
\setval{olmo.gdpo.minus.grpo.humaneval-plus-abs}{1.0}
\setval{olmo.gdpo.minus.grpo.if}{$-$3.0}
\setval{olmo.gdpo.minus.grpo.if-abs}{3.0}
\setval{olmo.gdpo.minus.grpo.ifbench}{$-$2.3}
\setval{olmo.gdpo.minus.grpo.ifbench-abs}{2.3}
\setval{olmo.gdpo.minus.grpo.ifeval}{$-$1.0}
\setval{olmo.gdpo.minus.grpo.ifeval-abs}{1.0}
\setval{olmo.gdpo.minus.grpo.judge2.avg}{$-$1.2}
\setval{olmo.gdpo.minus.grpo.judge2.avg-abs}{1.2}
\setval{olmo.gdpo.minus.grpo.judge2.creative-writing-v3}{$-$0.8}
\setval{olmo.gdpo.minus.grpo.judge2.creative-writing-v3-abs}{0.8}
\setval{olmo.gdpo.minus.grpo.judge2.longbench-write}{$-$1.4}
\setval{olmo.gdpo.minus.grpo.judge2.longbench-write-abs}{1.4}
\setval{olmo.gdpo.minus.grpo.judge2.writing}{$-$1.2}
\setval{olmo.gdpo.minus.grpo.judge2.writing-abs}{1.2}
\setval{olmo.gdpo.minus.grpo.judge2.writingbench}{$-$1.4}
\setval{olmo.gdpo.minus.grpo.judge2.writingbench-abs}{1.4}
\setval{olmo.gdpo.minus.grpo.livecodebench}{+0.8}
\setval{olmo.gdpo.minus.grpo.livecodebench-abs}{0.8}
\setval{olmo.gdpo.minus.grpo.longbench-write}{$-$0.9}
\setval{olmo.gdpo.minus.grpo.longbench-write-abs}{0.9}
\setval{olmo.gdpo.minus.grpo.math}{$-$0.7}
\setval{olmo.gdpo.minus.grpo.math-abs}{0.7}
\setval{olmo.gdpo.minus.grpo.math500}{+0.3}
\setval{olmo.gdpo.minus.grpo.math500-abs}{0.3}
\setval{olmo.gdpo.minus.grpo.mbpp-plus}{$-$1.0}
\setval{olmo.gdpo.minus.grpo.mbpp-plus-abs}{1.0}
\setval{olmo.gdpo.minus.grpo.minerva}{+0.4}
\setval{olmo.gdpo.minus.grpo.minerva-abs}{0.4}
\setval{olmo.gdpo.minus.grpo.olympiad-bench}{+0.2}
\setval{olmo.gdpo.minus.grpo.olympiad-bench-abs}{0.2}
\setval{olmo.gdpo.minus.grpo.writing}{$-$2.1}
\setval{olmo.gdpo.minus.grpo.writing-abs}{2.1}
\setval{olmo.gdpo.minus.grpo.writingbench}{$-$3.2}
\setval{olmo.gdpo.minus.grpo.writingbench-abs}{3.2}
\setval{olmo.gdpo.minus.mt-grpo.aime24}{$-$8.8}
\setval{olmo.gdpo.minus.mt-grpo.aime24-abs}{8.8}
\setval{olmo.gdpo.minus.mt-grpo.aime25}{$-$5.5}
\setval{olmo.gdpo.minus.mt-grpo.aime25-abs}{5.5}
\setval{olmo.gdpo.minus.mt-grpo.amc23}{$-$7.9}
\setval{olmo.gdpo.minus.mt-grpo.amc23-abs}{7.9}
\setval{olmo.gdpo.minus.mt-grpo.avg}{$-$2.8}
\setval{olmo.gdpo.minus.mt-grpo.avg-abs}{2.8}
\setval{olmo.gdpo.minus.mt-grpo.code}{+0.3}
\setval{olmo.gdpo.minus.mt-grpo.code-abs}{0.3}
\setval{olmo.gdpo.minus.mt-grpo.creative-writing-v3}{$-$3.6}
\setval{olmo.gdpo.minus.mt-grpo.creative-writing-v3-abs}{3.6}
\setval{olmo.gdpo.minus.mt-grpo.guidebench}{$-$3.6}
\setval{olmo.gdpo.minus.mt-grpo.guidebench-abs}{3.6}
\setval{olmo.gdpo.minus.mt-grpo.humaneval-plus}{+0.6}
\setval{olmo.gdpo.minus.mt-grpo.humaneval-plus-abs}{0.6}
\setval{olmo.gdpo.minus.mt-grpo.if}{$-$2.2}
\setval{olmo.gdpo.minus.mt-grpo.if-abs}{2.2}
\setval{olmo.gdpo.minus.mt-grpo.ifbench}{$-$2.3}
\setval{olmo.gdpo.minus.mt-grpo.ifbench-abs}{2.3}
\setval{olmo.gdpo.minus.mt-grpo.ifeval}{$-$0.6}
\setval{olmo.gdpo.minus.mt-grpo.ifeval-abs}{0.6}
\setval{olmo.gdpo.minus.mt-grpo.judge2.avg}{$-$2.7}
\setval{olmo.gdpo.minus.mt-grpo.judge2.avg-abs}{2.7}
\setval{olmo.gdpo.minus.mt-grpo.judge2.creative-writing-v3}{$-$2.0}
\setval{olmo.gdpo.minus.mt-grpo.judge2.creative-writing-v3-abs}{2.0}
\setval{olmo.gdpo.minus.mt-grpo.judge2.longbench-write}{$-$4.8}
\setval{olmo.gdpo.minus.mt-grpo.judge2.longbench-write-abs}{4.8}
\setval{olmo.gdpo.minus.mt-grpo.judge2.writing}{$-$2.4}
\setval{olmo.gdpo.minus.mt-grpo.judge2.writing-abs}{2.4}
\setval{olmo.gdpo.minus.mt-grpo.judge2.writingbench}{$-$0.5}
\setval{olmo.gdpo.minus.mt-grpo.judge2.writingbench-abs}{0.5}
\setval{olmo.gdpo.minus.mt-grpo.livecodebench}{+0.5}
\setval{olmo.gdpo.minus.mt-grpo.livecodebench-abs}{0.5}
\setval{olmo.gdpo.minus.mt-grpo.longbench-write}{$-$3.1}
\setval{olmo.gdpo.minus.mt-grpo.longbench-write-abs}{3.1}
\setval{olmo.gdpo.minus.mt-grpo.math}{$-$4.6}
\setval{olmo.gdpo.minus.mt-grpo.math-abs}{4.6}
\setval{olmo.gdpo.minus.mt-grpo.math500}{$-$1.7}
\setval{olmo.gdpo.minus.mt-grpo.math500-abs}{1.7}
\setval{olmo.gdpo.minus.mt-grpo.mbpp-plus}{$-$0.2}
\setval{olmo.gdpo.minus.mt-grpo.mbpp-plus-abs}{0.2}
\setval{olmo.gdpo.minus.mt-grpo.minerva}{$-$1.6}
\setval{olmo.gdpo.minus.mt-grpo.minerva-abs}{1.6}
\setval{olmo.gdpo.minus.mt-grpo.olympiad-bench}{$-$2.1}
\setval{olmo.gdpo.minus.mt-grpo.olympiad-bench-abs}{2.1}
\setval{olmo.gdpo.minus.mt-grpo.writing}{$-$3.0}
\setval{olmo.gdpo.minus.mt-grpo.writing-abs}{3.0}
\setval{olmo.gdpo.minus.mt-grpo.writingbench}{$-$2.4}
\setval{olmo.gdpo.minus.mt-grpo.writingbench-abs}{2.4}
\setval{olmo.gdpo.minus.papo.aime24}{+5.6}
\setval{olmo.gdpo.minus.papo.aime24-abs}{5.6}
\setval{olmo.gdpo.minus.papo.aime25}{+2.3}
\setval{olmo.gdpo.minus.papo.aime25-abs}{2.3}
\setval{olmo.gdpo.minus.papo.amc23}{+10.2}
\setval{olmo.gdpo.minus.papo.amc23-abs}{10.2}
\setval{olmo.gdpo.minus.papo.avg}{+1.5}
\setval{olmo.gdpo.minus.papo.avg-abs}{1.5}
\setval{olmo.gdpo.minus.papo.code}{+0.8}
\setval{olmo.gdpo.minus.papo.code-abs}{0.8}
\setval{olmo.gdpo.minus.papo.creative-writing-v3}{+0.2}
\setval{olmo.gdpo.minus.papo.creative-writing-v3-abs}{0.2}
\setval{olmo.gdpo.minus.papo.guidebench}{$-$7.8}
\setval{olmo.gdpo.minus.papo.guidebench-abs}{7.8}
\setval{olmo.gdpo.minus.papo.humaneval-plus}{+1.6}
\setval{olmo.gdpo.minus.papo.humaneval-plus-abs}{1.6}
\setval{olmo.gdpo.minus.papo.if}{$-$2.8}
\setval{olmo.gdpo.minus.papo.if-abs}{2.8}
\setval{olmo.gdpo.minus.papo.ifbench}{+0.0}
\setval{olmo.gdpo.minus.papo.ifbench-abs}{0.0}
\setval{olmo.gdpo.minus.papo.ifeval}{$-$0.7}
\setval{olmo.gdpo.minus.papo.ifeval-abs}{0.7}
\setval{olmo.gdpo.minus.papo.judge2.avg}{+1.6}
\setval{olmo.gdpo.minus.papo.judge2.avg-abs}{1.6}
\setval{olmo.gdpo.minus.papo.judge2.creative-writing-v3}{$-$0.1}
\setval{olmo.gdpo.minus.papo.judge2.creative-writing-v3-abs}{0.1}
\setval{olmo.gdpo.minus.papo.judge2.longbench-write}{+0.0}
\setval{olmo.gdpo.minus.papo.judge2.longbench-write-abs}{0.0}
\setval{olmo.gdpo.minus.papo.judge2.writing}{$-$0.2}
\setval{olmo.gdpo.minus.papo.judge2.writing-abs}{0.2}
\setval{olmo.gdpo.minus.papo.judge2.writingbench}{$-$0.4}
\setval{olmo.gdpo.minus.papo.judge2.writingbench-abs}{0.4}
\setval{olmo.gdpo.minus.papo.livecodebench}{+1.3}
\setval{olmo.gdpo.minus.papo.livecodebench-abs}{1.3}
\setval{olmo.gdpo.minus.papo.longbench-write}{$-$1.5}
\setval{olmo.gdpo.minus.papo.longbench-write-abs}{1.5}
\setval{olmo.gdpo.minus.papo.math}{+5.1}
\setval{olmo.gdpo.minus.papo.math-abs}{5.1}
\setval{olmo.gdpo.minus.papo.math500}{+4.4}
\setval{olmo.gdpo.minus.papo.math500-abs}{4.4}
\setval{olmo.gdpo.minus.papo.mbpp-plus}{$-$0.3}
\setval{olmo.gdpo.minus.papo.mbpp-plus-abs}{0.3}
\setval{olmo.gdpo.minus.papo.minerva}{+1.8}
\setval{olmo.gdpo.minus.papo.minerva-abs}{1.8}
\setval{olmo.gdpo.minus.papo.olympiad-bench}{+6.2}
\setval{olmo.gdpo.minus.papo.olympiad-bench-abs}{6.2}
\setval{olmo.gdpo.minus.papo.writing}{$-$0.4}
\setval{olmo.gdpo.minus.papo.writing-abs}{0.4}
\setval{olmo.gdpo.minus.papo.writingbench}{+0.0}
\setval{olmo.gdpo.minus.papo.writingbench-abs}{0.0}
\setval{olmo.gdpo.minus.saw.aime24}{+1.7}
\setval{olmo.gdpo.minus.saw.aime24-abs}{1.7}
\setval{olmo.gdpo.minus.saw.aime25}{$-$2.6}
\setval{olmo.gdpo.minus.saw.aime25-abs}{2.6}
\setval{olmo.gdpo.minus.saw.amc23}{+2.7}
\setval{olmo.gdpo.minus.saw.amc23-abs}{2.7}
\setval{olmo.gdpo.minus.saw.avg}{$-$0.6}
\setval{olmo.gdpo.minus.saw.avg-abs}{0.6}
\setval{olmo.gdpo.minus.saw.code}{+0.2}
\setval{olmo.gdpo.minus.saw.code-abs}{0.2}
\setval{olmo.gdpo.minus.saw.creative-writing-v3}{$-$0.8}
\setval{olmo.gdpo.minus.saw.creative-writing-v3-abs}{0.8}
\setval{olmo.gdpo.minus.saw.guidebench}{$-$3.8}
\setval{olmo.gdpo.minus.saw.guidebench-abs}{3.8}
\setval{olmo.gdpo.minus.saw.humaneval-plus}{+0.0}
\setval{olmo.gdpo.minus.saw.humaneval-plus-abs}{0.0}
\setval{olmo.gdpo.minus.saw.if}{$-$3.0}
\setval{olmo.gdpo.minus.saw.if-abs}{3.0}
\setval{olmo.gdpo.minus.saw.ifbench}{$-$3.0}
\setval{olmo.gdpo.minus.saw.ifbench-abs}{3.0}
\setval{olmo.gdpo.minus.saw.ifeval}{$-$2.2}
\setval{olmo.gdpo.minus.saw.ifeval-abs}{2.2}
\setval{olmo.gdpo.minus.saw.judge2.avg}{$-$0.7}
\setval{olmo.gdpo.minus.saw.judge2.avg-abs}{0.7}
\setval{olmo.gdpo.minus.saw.judge2.creative-writing-v3}{$-$3.2}
\setval{olmo.gdpo.minus.saw.judge2.creative-writing-v3-abs}{3.2}
\setval{olmo.gdpo.minus.saw.judge2.longbench-write}{$-$1.7}
\setval{olmo.gdpo.minus.saw.judge2.longbench-write-abs}{1.7}
\setval{olmo.gdpo.minus.saw.judge2.writing}{$-$2.1}
\setval{olmo.gdpo.minus.saw.judge2.writing-abs}{2.1}
\setval{olmo.gdpo.minus.saw.judge2.writingbench}{$-$1.4}
\setval{olmo.gdpo.minus.saw.judge2.writingbench-abs}{1.4}
\setval{olmo.gdpo.minus.saw.livecodebench}{+1.3}
\setval{olmo.gdpo.minus.saw.livecodebench-abs}{1.3}
\setval{olmo.gdpo.minus.saw.longbench-write}{$-$1.7}
\setval{olmo.gdpo.minus.saw.longbench-write-abs}{1.7}
\setval{olmo.gdpo.minus.saw.math}{+0.8}
\setval{olmo.gdpo.minus.saw.math-abs}{0.8}
\setval{olmo.gdpo.minus.saw.math500}{+0.8}
\setval{olmo.gdpo.minus.saw.math500-abs}{0.8}
\setval{olmo.gdpo.minus.saw.mbpp-plus}{$-$0.7}
\setval{olmo.gdpo.minus.saw.mbpp-plus-abs}{0.7}
\setval{olmo.gdpo.minus.saw.minerva}{+0.0}
\setval{olmo.gdpo.minus.saw.minerva-abs}{0.0}
\setval{olmo.gdpo.minus.saw.olympiad-bench}{+2.1}
\setval{olmo.gdpo.minus.saw.olympiad-bench-abs}{2.1}
\setval{olmo.gdpo.minus.saw.writing}{$-$1.5}
\setval{olmo.gdpo.minus.saw.writing-abs}{1.5}
\setval{olmo.gdpo.minus.saw.writingbench}{$-$2.0}
\setval{olmo.gdpo.minus.saw.writingbench-abs}{2.0}
\setval{olmo.gdpo.passing-at-ceiling.code.early}{0.84}
\setval{olmo.gdpo.passing-at-ceiling.code.late}{0.84}
\setval{olmo.gdpo.passing-at-ceiling.code.run}{0.84}
\setval{olmo.gdpo.passing-at-ceiling.if.early}{0.48}
\setval{olmo.gdpo.passing-at-ceiling.if.late}{0.57}
\setval{olmo.gdpo.passing-at-ceiling.if.run}{0.53}
\setval{olmo.gdpo.passing-at-ceiling.math.early}{1.00}
\setval{olmo.gdpo.passing-at-ceiling.math.late}{1.00}
\setval{olmo.gdpo.passing-at-ceiling.math.run}{1.00}
\setval{olmo.gdpo.passing-at-ceiling.writing.early}{0.00}
\setval{olmo.gdpo.passing-at-ceiling.writing.late}{0.00}
\setval{olmo.gdpo.passing-at-ceiling.writing.run}{0.00}
\setval{olmo.gdpo.primary-cosine.code.early}{0.93}
\setval{olmo.gdpo.primary-cosine.code.late}{0.89}
\setval{olmo.gdpo.primary-cosine.code.run}{0.92}
\setval{olmo.gdpo.primary-cosine.if.early}{0.78}
\setval{olmo.gdpo.primary-cosine.if.late}{0.79}
\setval{olmo.gdpo.primary-cosine.if.run}{0.79}
\setval{olmo.gdpo.primary-cosine.math.early}{0.97}
\setval{olmo.gdpo.primary-cosine.math.late}{0.89}
\setval{olmo.gdpo.primary-cosine.math.run}{0.91}
\setval{olmo.gdpo.primary-cosine.writing.early}{0.88}
\setval{olmo.gdpo.primary-cosine.writing.late}{0.72}
\setval{olmo.gdpo.primary-cosine.writing.run}{0.76}
\setval{olmo.gdpo.primary-level.code.early}{0.58}
\setval{olmo.gdpo.primary-level.code.late}{0.83}
\setval{olmo.gdpo.primary-level.code.run}{0.75}
\setval{olmo.gdpo.primary-level.if.early}{0.63}
\setval{olmo.gdpo.primary-level.if.late}{0.75}
\setval{olmo.gdpo.primary-level.if.run}{0.70}
\setval{olmo.gdpo.primary-level.math.early}{0.34}
\setval{olmo.gdpo.primary-level.math.late}{0.67}
\setval{olmo.gdpo.primary-level.math.run}{0.57}
\setval{olmo.gdpo.primary-level.writing.early}{0.50}
\setval{olmo.gdpo.primary-level.writing.late}{0.69}
\setval{olmo.gdpo.primary-level.writing.run}{0.64}
\setval{olmo.gdpo.solved.code.early}{0.12}
\setval{olmo.gdpo.solved.code.late}{0.14}
\setval{olmo.gdpo.solved.code.run}{0.12}
\setval{olmo.gdpo.solved.if.early}{0.18}
\setval{olmo.gdpo.solved.if.late}{0.17}
\setval{olmo.gdpo.solved.if.run}{0.17}
\setval{olmo.gdpo.solved.math.early}{0.07}
\setval{olmo.gdpo.solved.math.late}{0.26}
\setval{olmo.gdpo.solved.math.run}{0.19}
\setval{olmo.gdpo.solved.writing.early}{0.00}
\setval{olmo.gdpo.solved.writing.late}{0.00}
\setval{olmo.gdpo.solved.writing.run}{0.00}
\setval{olmo.gdpo.update-share.code.early}{0.18}
\setval{olmo.gdpo.update-share.code.late}{0.17}
\setval{olmo.gdpo.update-share.code.run}{0.18}
\setval{olmo.gdpo.update-share.if.early}{0.19}
\setval{olmo.gdpo.update-share.if.late}{0.18}
\setval{olmo.gdpo.update-share.if.run}{0.19}
\setval{olmo.gdpo.update-share.math.early}{0.19}
\setval{olmo.gdpo.update-share.math.late}{0.24}
\setval{olmo.gdpo.update-share.math.run}{0.22}
\setval{olmo.gdpo.update-share.writing.early}{0.44}
\setval{olmo.gdpo.update-share.writing.late}{0.40}
\setval{olmo.gdpo.update-share.writing.run}{0.41}
\setval{olmo.geoma.aux-change.code}{+0.18}
\setval{olmo.geoma.aux-change.if}{+0.12}
\setval{olmo.geoma.aux-change.math}{$-$0.04}
\setval{olmo.geoma.aux-change.writing}{+0.02}
\setval{olmo.geoma.aux-change.writing-rm}{+0.30}
\setval{olmo.geoma.aux-full.code.early}{0.77}
\setval{olmo.geoma.aux-full.code.late}{0.95}
\setval{olmo.geoma.aux-full.code.run}{0.93}
\setval{olmo.geoma.aux-full.if.early}{0.62}
\setval{olmo.geoma.aux-full.if.late}{0.95}
\setval{olmo.geoma.aux-full.if.run}{0.90}
\setval{olmo.geoma.aux-full.math.early}{0.55}
\setval{olmo.geoma.aux-full.math.late}{0.67}
\setval{olmo.geoma.aux-full.math.run}{0.59}
\setval{olmo.geoma.aux-full.writing-rm.early}{0.02}
\setval{olmo.geoma.aux-full.writing-rm.late}{0.30}
\setval{olmo.geoma.aux-full.writing-rm.run}{0.13}
\setval{olmo.geoma.aux-full.writing.early}{0.58}
\setval{olmo.geoma.aux-full.writing.late}{0.76}
\setval{olmo.geoma.aux-full.writing.run}{0.68}
\setval{olmo.geoma.aux-level.code.early}{0.92}
\setval{olmo.geoma.aux-level.code.late}{0.99}
\setval{olmo.geoma.aux-level.code.run}{0.98}
\setval{olmo.geoma.aux-level.if.early}{0.91}
\setval{olmo.geoma.aux-level.if.late}{0.99}
\setval{olmo.geoma.aux-level.if.run}{0.98}
\setval{olmo.geoma.aux-level.math.early}{0.91}
\setval{olmo.geoma.aux-level.math.late}{0.86}
\setval{olmo.geoma.aux-level.math.run}{0.84}
\setval{olmo.geoma.aux-level.writing-rm.early}{0.60}
\setval{olmo.geoma.aux-level.writing-rm.late}{0.87}
\setval{olmo.geoma.aux-level.writing-rm.run}{0.77}
\setval{olmo.geoma.aux-level.writing.early}{0.95}
\setval{olmo.geoma.aux-level.writing.late}{0.97}
\setval{olmo.geoma.aux-level.writing.run}{0.96}
\setval{olmo.geoma.bench.aime24}{23.8}
\setval{olmo.geoma.bench.aime25}{22.0}
\setval{olmo.geoma.bench.amc23}{63.9}
\setval{olmo.geoma.bench.creative-writing-v3}{49.6}
\setval{olmo.geoma.bench.guidebench}{46.6}
\setval{olmo.geoma.bench.humaneval-plus}{74.6}
\setval{olmo.geoma.bench.ifbench}{30.3}
\setval{olmo.geoma.bench.ifeval}{88.0}
\setval{olmo.geoma.bench.livecodebench}{22.3}
\setval{olmo.geoma.bench.longbench-write}{65.3}
\setval{olmo.geoma.bench.math500}{83.2}
\setval{olmo.geoma.bench.mbpp-plus}{64.7}
\setval{olmo.geoma.bench.minerva}{40.0}
\setval{olmo.geoma.bench.olympiad-bench}{52.7}
\setval{olmo.geoma.bench.writingbench}{67.0}
\setval{olmo.geoma.eval.avg}{52.9}
\setval{olmo.geoma.eval.code}{53.9}
\setval{olmo.geoma.eval.if}{55.0}
\setval{olmo.geoma.eval.math}{47.6}
\setval{olmo.geoma.eval.writing}{60.6}
\setval{olmo.geoma.gate.code.against.early}{0.06}
\setval{olmo.geoma.gate.code.against.late}{0.05}
\setval{olmo.geoma.gate.code.against.run}{0.04}
\setval{olmo.geoma.gate.code.agreement}{$-$0.001}
\setval{olmo.geoma.gate.code.ceiling.early}{0.19}
\setval{olmo.geoma.gate.code.ceiling.late}{0.12}
\setval{olmo.geoma.gate.code.ceiling.run}{0.12}
\setval{olmo.geoma.gate.code.cosine.early}{0.76}
\setval{olmo.geoma.gate.code.cosine.late}{0.74}
\setval{olmo.geoma.gate.code.cosine.run}{0.78}
\setval{olmo.geoma.gate.if.against.early}{0.09}
\setval{olmo.geoma.gate.if.against.late}{0.04}
\setval{olmo.geoma.gate.if.against.run}{0.06}
\setval{olmo.geoma.gate.if.agreement}{+0.033}
\setval{olmo.geoma.gate.if.ceiling.early}{0.23}
\setval{olmo.geoma.gate.if.ceiling.late}{0.15}
\setval{olmo.geoma.gate.if.ceiling.run}{0.16}
\setval{olmo.geoma.gate.if.cosine.early}{0.52}
\setval{olmo.geoma.gate.if.cosine.late}{0.53}
\setval{olmo.geoma.gate.if.cosine.run}{0.57}
\setval{olmo.geoma.gate.math.against.early}{0.00}
\setval{olmo.geoma.gate.math.against.late}{0.00}
\setval{olmo.geoma.gate.math.against.run}{0.00}
\setval{olmo.geoma.gate.math.ceiling.early}{0.13}
\setval{olmo.geoma.gate.math.ceiling.late}{0.34}
\setval{olmo.geoma.gate.math.ceiling.run}{0.31}
\setval{olmo.geoma.gate.math.cosine.early}{0.91}
\setval{olmo.geoma.gate.math.cosine.late}{0.75}
\setval{olmo.geoma.gate.math.cosine.run}{0.77}
\setval{olmo.geoma.gate.writing-rm.against.early}{0.22}
\setval{olmo.geoma.gate.writing-rm.against.late}{0.48}
\setval{olmo.geoma.gate.writing-rm.against.run}{0.42}
\setval{olmo.geoma.gate.writing-rm.agreement}{+0.091}
\setval{olmo.geoma.gate.writing-rm.ceiling.early}{0.00}
\setval{olmo.geoma.gate.writing-rm.ceiling.late}{0.00}
\setval{olmo.geoma.gate.writing-rm.ceiling.run}{0.00}
\setval{olmo.geoma.gate.writing-rm.cosine.early}{0.69}
\setval{olmo.geoma.gate.writing-rm.cosine.late}{0.26}
\setval{olmo.geoma.gate.writing-rm.cosine.run}{0.38}
\setval{olmo.geoma.gate.writing.against.early}{0.25}
\setval{olmo.geoma.gate.writing.against.late}{0.49}
\setval{olmo.geoma.gate.writing.against.run}{0.44}
\setval{olmo.geoma.gate.writing.agreement}{$-$0.054}
\setval{olmo.geoma.gate.writing.ceiling.early}{0.00}
\setval{olmo.geoma.gate.writing.ceiling.late}{0.00}
\setval{olmo.geoma.gate.writing.ceiling.run}{0.00}
\setval{olmo.geoma.gate.writing.cosine.early}{0.59}
\setval{olmo.geoma.gate.writing.cosine.late}{0.18}
\setval{olmo.geoma.gate.writing.cosine.run}{0.28}
\setval{olmo.geoma.judge2.avg}{50.7}
\setval{olmo.geoma.judge2.creative-writing-v3}{41.8}
\setval{olmo.geoma.judge2.longbench-write}{58.2}
\setval{olmo.geoma.judge2.writing}{49.5}
\setval{olmo.geoma.judge2.writingbench}{48.6}
\setval{olmo.geoma.last-step}{300}
\setval{olmo.geoma.minus.base.aime24}{+15.2}
\setval{olmo.geoma.minus.base.aime24-abs}{15.2}
\setval{olmo.geoma.minus.base.aime25}{+14.6}
\setval{olmo.geoma.minus.base.aime25-abs}{14.6}
\setval{olmo.geoma.minus.base.amc23}{+19.4}
\setval{olmo.geoma.minus.base.amc23-abs}{19.4}
\setval{olmo.geoma.minus.base.avg}{+8.8}
\setval{olmo.geoma.minus.base.avg-abs}{8.8}
\setval{olmo.geoma.minus.base.code}{+4.8}
\setval{olmo.geoma.minus.base.code-abs}{4.8}
\setval{olmo.geoma.minus.base.creative-writing-v3}{+11.7}
\setval{olmo.geoma.minus.base.creative-writing-v3-abs}{11.7}
\setval{olmo.geoma.minus.base.guidebench}{$-$14.8}
\setval{olmo.geoma.minus.base.guidebench-abs}{14.8}
\setval{olmo.geoma.minus.base.humaneval-plus}{+5.4}
\setval{olmo.geoma.minus.base.humaneval-plus-abs}{5.4}
\setval{olmo.geoma.minus.base.if}{$-$2.5}
\setval{olmo.geoma.minus.base.if-abs}{2.5}
\setval{olmo.geoma.minus.base.ifbench}{+3.7}
\setval{olmo.geoma.minus.base.ifbench-abs}{3.7}
\setval{olmo.geoma.minus.base.ifeval}{+3.6}
\setval{olmo.geoma.minus.base.ifeval-abs}{3.6}
\setval{olmo.geoma.minus.base.judge2.avg}{+7.8}
\setval{olmo.geoma.minus.base.judge2.avg-abs}{7.8}
\setval{olmo.geoma.minus.base.judge2.creative-writing-v3}{+10.1}
\setval{olmo.geoma.minus.base.judge2.creative-writing-v3-abs}{10.1}
\setval{olmo.geoma.minus.base.judge2.longbench-write}{+5.0}
\setval{olmo.geoma.minus.base.judge2.longbench-write-abs}{5.0}
\setval{olmo.geoma.minus.base.judge2.writing}{+8.7}
\setval{olmo.geoma.minus.base.judge2.writing-abs}{8.7}
\setval{olmo.geoma.minus.base.judge2.writingbench}{+11.0}
\setval{olmo.geoma.minus.base.judge2.writingbench-abs}{11.0}
\setval{olmo.geoma.minus.base.livecodebench}{+5.5}
\setval{olmo.geoma.minus.base.livecodebench-abs}{5.5}
\setval{olmo.geoma.minus.base.longbench-write}{+7.9}
\setval{olmo.geoma.minus.base.longbench-write-abs}{7.9}
\setval{olmo.geoma.minus.base.math}{+13.9}
\setval{olmo.geoma.minus.base.math-abs}{13.9}
\setval{olmo.geoma.minus.base.math500}{+13.1}
\setval{olmo.geoma.minus.base.math500-abs}{13.1}
\setval{olmo.geoma.minus.base.mbpp-plus}{+3.5}
\setval{olmo.geoma.minus.base.mbpp-plus-abs}{3.5}
\setval{olmo.geoma.minus.base.minerva}{+4.7}
\setval{olmo.geoma.minus.base.minerva-abs}{4.7}
\setval{olmo.geoma.minus.base.olympiad-bench}{+16.5}
\setval{olmo.geoma.minus.base.olympiad-bench-abs}{16.5}
\setval{olmo.geoma.minus.base.writing}{+13.9}
\setval{olmo.geoma.minus.base.writing-abs}{13.9}
\setval{olmo.geoma.minus.base.writingbench}{+22.2}
\setval{olmo.geoma.minus.base.writingbench-abs}{22.2}
\setval{olmo.geoma.minus.concur-free.aime24}{$-$3.0}
\setval{olmo.geoma.minus.concur-free.aime24-abs}{3.0}
\setval{olmo.geoma.minus.concur-free.aime25}{$-$0.1}
\setval{olmo.geoma.minus.concur-free.aime25-abs}{0.1}
\setval{olmo.geoma.minus.concur-free.amc23}{$-$3.6}
\setval{olmo.geoma.minus.concur-free.amc23-abs}{3.6}
\setval{olmo.geoma.minus.concur-free.avg}{$-$2.5}
\setval{olmo.geoma.minus.concur-free.avg-abs}{2.5}
\setval{olmo.geoma.minus.concur-free.code}{$-$1.7}
\setval{olmo.geoma.minus.concur-free.code-abs}{1.7}
\setval{olmo.geoma.minus.concur-free.creative-writing-v3}{$-$0.2}
\setval{olmo.geoma.minus.concur-free.creative-writing-v3-abs}{0.2}
\setval{olmo.geoma.minus.concur-free.guidebench}{$-$13.5}
\setval{olmo.geoma.minus.concur-free.guidebench-abs}{13.5}
\setval{olmo.geoma.minus.concur-free.humaneval-plus}{$-$2.0}
\setval{olmo.geoma.minus.concur-free.humaneval-plus-abs}{2.0}
\setval{olmo.geoma.minus.concur-free.if}{$-$4.7}
\setval{olmo.geoma.minus.concur-free.if-abs}{4.7}
\setval{olmo.geoma.minus.concur-free.ifbench}{$-$1.3}
\setval{olmo.geoma.minus.concur-free.ifbench-abs}{1.3}
\setval{olmo.geoma.minus.concur-free.ifeval}{+0.7}
\setval{olmo.geoma.minus.concur-free.ifeval-abs}{0.7}
\setval{olmo.geoma.minus.concur-free.judge2.avg}{$-$1.9}
\setval{olmo.geoma.minus.concur-free.judge2.avg-abs}{1.9}
\setval{olmo.geoma.minus.concur-free.judge2.creative-writing-v3}{+4.2}
\setval{olmo.geoma.minus.concur-free.judge2.creative-writing-v3-abs}{4.2}
\setval{olmo.geoma.minus.concur-free.judge2.longbench-write}{$-$5.7}
\setval{olmo.geoma.minus.concur-free.judge2.longbench-write-abs}{5.7}
\setval{olmo.geoma.minus.concur-free.judge2.writing}{$-$0.3}
\setval{olmo.geoma.minus.concur-free.judge2.writing-abs}{0.3}
\setval{olmo.geoma.minus.concur-free.judge2.writingbench}{+0.6}
\setval{olmo.geoma.minus.concur-free.judge2.writingbench-abs}{0.6}
\setval{olmo.geoma.minus.concur-free.livecodebench}{$-$3.3}
\setval{olmo.geoma.minus.concur-free.livecodebench-abs}{3.3}
\setval{olmo.geoma.minus.concur-free.longbench-write}{$-$6.7}
\setval{olmo.geoma.minus.concur-free.longbench-write-abs}{6.7}
\setval{olmo.geoma.minus.concur-free.math}{$-$1.4}
\setval{olmo.geoma.minus.concur-free.math-abs}{1.4}
\setval{olmo.geoma.minus.concur-free.math500}{$-$1.2}
\setval{olmo.geoma.minus.concur-free.math500-abs}{1.2}
\setval{olmo.geoma.minus.concur-free.mbpp-plus}{+0.2}
\setval{olmo.geoma.minus.concur-free.mbpp-plus-abs}{0.2}
\setval{olmo.geoma.minus.concur-free.minerva}{+0.0}
\setval{olmo.geoma.minus.concur-free.minerva-abs}{0.0}
\setval{olmo.geoma.minus.concur-free.olympiad-bench}{$-$0.5}
\setval{olmo.geoma.minus.concur-free.olympiad-bench-abs}{0.5}
\setval{olmo.geoma.minus.concur-free.writing}{$-$3.4}
\setval{olmo.geoma.minus.concur-free.writing-abs}{3.4}
\setval{olmo.geoma.minus.concur-free.writingbench}{$-$3.2}
\setval{olmo.geoma.minus.concur-free.writingbench-abs}{3.2}
\setval{olmo.geoma.minus.concur-tied.aime24}{$-$10.7}
\setval{olmo.geoma.minus.concur-tied.aime24-abs}{10.7}
\setval{olmo.geoma.minus.concur-tied.aime25}{$-$7.9}
\setval{olmo.geoma.minus.concur-tied.aime25-abs}{7.9}
\setval{olmo.geoma.minus.concur-tied.amc23}{$-$15.9}
\setval{olmo.geoma.minus.concur-tied.amc23-abs}{15.9}
\setval{olmo.geoma.minus.concur-tied.avg}{$-$6.2}
\setval{olmo.geoma.minus.concur-tied.avg-abs}{6.2}
\setval{olmo.geoma.minus.concur-tied.code}{$-$7.2}
\setval{olmo.geoma.minus.concur-tied.code-abs}{7.2}
\setval{olmo.geoma.minus.concur-tied.creative-writing-v3}{+0.6}
\setval{olmo.geoma.minus.concur-tied.creative-writing-v3-abs}{0.6}
\setval{olmo.geoma.minus.concur-tied.guidebench}{$-$11.9}
\setval{olmo.geoma.minus.concur-tied.guidebench-abs}{11.9}
\setval{olmo.geoma.minus.concur-tied.humaneval-plus}{$-$8.7}
\setval{olmo.geoma.minus.concur-tied.humaneval-plus-abs}{8.7}
\setval{olmo.geoma.minus.concur-tied.if}{$-$5.0}
\setval{olmo.geoma.minus.concur-tied.if-abs}{5.0}
\setval{olmo.geoma.minus.concur-tied.ifbench}{$-$3.3}
\setval{olmo.geoma.minus.concur-tied.ifbench-abs}{3.3}
\setval{olmo.geoma.minus.concur-tied.ifeval}{+0.4}
\setval{olmo.geoma.minus.concur-tied.ifeval-abs}{0.4}
\setval{olmo.geoma.minus.concur-tied.judge2.avg}{$-$5.7}
\setval{olmo.geoma.minus.concur-tied.judge2.avg-abs}{5.7}
\setval{olmo.geoma.minus.concur-tied.judge2.creative-writing-v3}{+4.6}
\setval{olmo.geoma.minus.concur-tied.judge2.creative-writing-v3-abs}{4.6}
\setval{olmo.geoma.minus.concur-tied.judge2.longbench-write}{$-$2.3}
\setval{olmo.geoma.minus.concur-tied.judge2.longbench-write-abs}{2.3}
\setval{olmo.geoma.minus.concur-tied.judge2.writing}{+1.8}
\setval{olmo.geoma.minus.concur-tied.judge2.writing-abs}{1.8}
\setval{olmo.geoma.minus.concur-tied.judge2.writingbench}{+3.0}
\setval{olmo.geoma.minus.concur-tied.judge2.writingbench-abs}{3.0}
\setval{olmo.geoma.minus.concur-tied.livecodebench}{$-$6.7}
\setval{olmo.geoma.minus.concur-tied.livecodebench-abs}{6.7}
\setval{olmo.geoma.minus.concur-tied.longbench-write}{$-$3.9}
\setval{olmo.geoma.minus.concur-tied.longbench-write-abs}{3.9}
\setval{olmo.geoma.minus.concur-tied.math}{$-$9.0}
\setval{olmo.geoma.minus.concur-tied.math-abs}{9.0}
\setval{olmo.geoma.minus.concur-tied.math500}{$-$6.2}
\setval{olmo.geoma.minus.concur-tied.math500-abs}{6.2}
\setval{olmo.geoma.minus.concur-tied.mbpp-plus}{$-$6.3}
\setval{olmo.geoma.minus.concur-tied.mbpp-plus-abs}{6.3}
\setval{olmo.geoma.minus.concur-tied.minerva}{$-$4.1}
\setval{olmo.geoma.minus.concur-tied.minerva-abs}{4.1}
\setval{olmo.geoma.minus.concur-tied.olympiad-bench}{$-$8.8}
\setval{olmo.geoma.minus.concur-tied.olympiad-bench-abs}{8.8}
\setval{olmo.geoma.minus.concur-tied.writing}{$-$0.8}
\setval{olmo.geoma.minus.concur-tied.writing-abs}{0.8}
\setval{olmo.geoma.minus.concur-tied.writingbench}{+0.8}
\setval{olmo.geoma.minus.concur-tied.writingbench-abs}{0.8}
\setval{olmo.geoma.minus.concur.aime24}{$-$12.6}
\setval{olmo.geoma.minus.concur.aime24-abs}{12.6}
\setval{olmo.geoma.minus.concur.aime25}{$-$10.2}
\setval{olmo.geoma.minus.concur.aime25-abs}{10.2}
\setval{olmo.geoma.minus.concur.amc23}{$-$19.6}
\setval{olmo.geoma.minus.concur.amc23-abs}{19.6}
\setval{olmo.geoma.minus.concur.avg}{$-$6.4}
\setval{olmo.geoma.minus.concur.avg-abs}{6.4}
\setval{olmo.geoma.minus.concur.code}{$-$6.3}
\setval{olmo.geoma.minus.concur.code-abs}{6.3}
\setval{olmo.geoma.minus.concur.creative-writing-v3}{+0.4}
\setval{olmo.geoma.minus.concur.creative-writing-v3-abs}{0.4}
\setval{olmo.geoma.minus.concur.guidebench}{$-$10.3}
\setval{olmo.geoma.minus.concur.guidebench-abs}{10.3}
\setval{olmo.geoma.minus.concur.humaneval-plus}{$-$8.1}
\setval{olmo.geoma.minus.concur.humaneval-plus-abs}{8.1}
\setval{olmo.geoma.minus.concur.if}{$-$3.9}
\setval{olmo.geoma.minus.concur.if-abs}{3.9}
\setval{olmo.geoma.minus.concur.ifbench}{$-$1.7}
\setval{olmo.geoma.minus.concur.ifbench-abs}{1.7}
\setval{olmo.geoma.minus.concur.ifeval}{+0.2}
\setval{olmo.geoma.minus.concur.ifeval-abs}{0.2}
\setval{olmo.geoma.minus.concur.judge2.avg}{$-$5.9}
\setval{olmo.geoma.minus.concur.judge2.avg-abs}{5.9}
\setval{olmo.geoma.minus.concur.judge2.creative-writing-v3}{+3.3}
\setval{olmo.geoma.minus.concur.judge2.creative-writing-v3-abs}{3.3}
\setval{olmo.geoma.minus.concur.judge2.longbench-write}{$-$0.2}
\setval{olmo.geoma.minus.concur.judge2.longbench-write-abs}{0.2}
\setval{olmo.geoma.minus.concur.judge2.writing}{+2.3}
\setval{olmo.geoma.minus.concur.judge2.writing-abs}{2.3}
\setval{olmo.geoma.minus.concur.judge2.writingbench}{+3.7}
\setval{olmo.geoma.minus.concur.judge2.writingbench-abs}{3.7}
\setval{olmo.geoma.minus.concur.livecodebench}{$-$8.0}
\setval{olmo.geoma.minus.concur.livecodebench-abs}{8.0}
\setval{olmo.geoma.minus.concur.longbench-write}{$-$3.2}
\setval{olmo.geoma.minus.concur.longbench-write-abs}{3.2}
\setval{olmo.geoma.minus.concur.math}{$-$10.7}
\setval{olmo.geoma.minus.concur.math-abs}{10.7}
\setval{olmo.geoma.minus.concur.math500}{$-$7.4}
\setval{olmo.geoma.minus.concur.math500-abs}{7.4}
\setval{olmo.geoma.minus.concur.mbpp-plus}{$-$2.8}
\setval{olmo.geoma.minus.concur.mbpp-plus-abs}{2.8}
\setval{olmo.geoma.minus.concur.minerva}{$-$4.8}
\setval{olmo.geoma.minus.concur.minerva-abs}{4.8}
\setval{olmo.geoma.minus.concur.olympiad-bench}{$-$9.7}
\setval{olmo.geoma.minus.concur.olympiad-bench-abs}{9.7}
\setval{olmo.geoma.minus.concur.writing}{$-$0.4}
\setval{olmo.geoma.minus.concur.writing-abs}{0.4}
\setval{olmo.geoma.minus.concur.writingbench}{+1.5}
\setval{olmo.geoma.minus.concur.writingbench-abs}{1.5}
\setval{olmo.geoma.minus.gd2po.aime24}{$-$4.6}
\setval{olmo.geoma.minus.gd2po.aime24-abs}{4.6}
\setval{olmo.geoma.minus.gd2po.aime25}{$-$2.3}
\setval{olmo.geoma.minus.gd2po.aime25-abs}{2.3}
\setval{olmo.geoma.minus.gd2po.amc23}{$-$8.3}
\setval{olmo.geoma.minus.gd2po.amc23-abs}{8.3}
\setval{olmo.geoma.minus.gd2po.avg}{$-$2.7}
\setval{olmo.geoma.minus.gd2po.avg-abs}{2.7}
\setval{olmo.geoma.minus.gd2po.code}{$-$1.0}
\setval{olmo.geoma.minus.gd2po.code-abs}{1.0}
\setval{olmo.geoma.minus.gd2po.creative-writing-v3}{+1.3}
\setval{olmo.geoma.minus.gd2po.creative-writing-v3-abs}{1.3}
\setval{olmo.geoma.minus.gd2po.guidebench}{$-$14.5}
\setval{olmo.geoma.minus.gd2po.guidebench-abs}{14.5}
\setval{olmo.geoma.minus.gd2po.humaneval-plus}{$-$2.1}
\setval{olmo.geoma.minus.gd2po.humaneval-plus-abs}{2.1}
\setval{olmo.geoma.minus.gd2po.if}{$-$5.4}
\setval{olmo.geoma.minus.gd2po.if-abs}{5.4}
\setval{olmo.geoma.minus.gd2po.ifbench}{$-$2.0}
\setval{olmo.geoma.minus.gd2po.ifbench-abs}{2.0}
\setval{olmo.geoma.minus.gd2po.ifeval}{+0.4}
\setval{olmo.geoma.minus.gd2po.ifeval-abs}{0.4}
\setval{olmo.geoma.minus.gd2po.judge2.avg}{$-$2.5}
\setval{olmo.geoma.minus.gd2po.judge2.avg-abs}{2.5}
\setval{olmo.geoma.minus.gd2po.judge2.creative-writing-v3}{+3.9}
\setval{olmo.geoma.minus.gd2po.judge2.creative-writing-v3-abs}{3.9}
\setval{olmo.geoma.minus.gd2po.judge2.longbench-write}{$-$1.7}
\setval{olmo.geoma.minus.gd2po.judge2.longbench-write-abs}{1.7}
\setval{olmo.geoma.minus.gd2po.judge2.writing}{+0.8}
\setval{olmo.geoma.minus.gd2po.judge2.writing-abs}{0.8}
\setval{olmo.geoma.minus.gd2po.judge2.writingbench}{+0.3}
\setval{olmo.geoma.minus.gd2po.judge2.writingbench-abs}{0.3}
\setval{olmo.geoma.minus.gd2po.livecodebench}{$-$0.9}
\setval{olmo.geoma.minus.gd2po.livecodebench-abs}{0.9}
\setval{olmo.geoma.minus.gd2po.longbench-write}{$-$2.9}
\setval{olmo.geoma.minus.gd2po.longbench-write-abs}{2.9}
\setval{olmo.geoma.minus.gd2po.math}{$-$3.5}
\setval{olmo.geoma.minus.gd2po.math-abs}{3.5}
\setval{olmo.geoma.minus.gd2po.math500}{$-$2.5}
\setval{olmo.geoma.minus.gd2po.math500-abs}{2.5}
\setval{olmo.geoma.minus.gd2po.mbpp-plus}{$-$0.1}
\setval{olmo.geoma.minus.gd2po.mbpp-plus-abs}{0.1}
\setval{olmo.geoma.minus.gd2po.minerva}{$-$0.1}
\setval{olmo.geoma.minus.gd2po.minerva-abs}{0.1}
\setval{olmo.geoma.minus.gd2po.olympiad-bench}{$-$3.3}
\setval{olmo.geoma.minus.gd2po.olympiad-bench-abs}{3.3}
\setval{olmo.geoma.minus.gd2po.writing}{$-$0.1}
\setval{olmo.geoma.minus.gd2po.writing-abs}{0.1}
\setval{olmo.geoma.minus.gd2po.writingbench}{+1.2}
\setval{olmo.geoma.minus.gd2po.writingbench-abs}{1.2}
\setval{olmo.geoma.minus.gdpo-noaux.aime24}{$-$14.6}
\setval{olmo.geoma.minus.gdpo-noaux.aime24-abs}{14.6}
\setval{olmo.geoma.minus.gdpo-noaux.aime25}{$-$11.7}
\setval{olmo.geoma.minus.gdpo-noaux.aime25-abs}{11.7}
\setval{olmo.geoma.minus.gdpo-noaux.amc23}{$-$20.0}
\setval{olmo.geoma.minus.gdpo-noaux.amc23-abs}{20.0}
\setval{olmo.geoma.minus.gdpo-noaux.avg}{$-$4.9}
\setval{olmo.geoma.minus.gdpo-noaux.avg-abs}{4.9}
\setval{olmo.geoma.minus.gdpo-noaux.code}{$-$8.5}
\setval{olmo.geoma.minus.gdpo-noaux.code-abs}{8.5}
\setval{olmo.geoma.minus.gdpo-noaux.creative-writing-v3}{+11.4}
\setval{olmo.geoma.minus.gdpo-noaux.creative-writing-v3-abs}{11.4}
\setval{olmo.geoma.minus.gdpo-noaux.guidebench}{$-$9.1}
\setval{olmo.geoma.minus.gdpo-noaux.guidebench-abs}{9.1}
\setval{olmo.geoma.minus.gdpo-noaux.humaneval-plus}{$-$10.5}
\setval{olmo.geoma.minus.gdpo-noaux.humaneval-plus-abs}{10.5}
\setval{olmo.geoma.minus.gdpo-noaux.if}{$-$3.0}
\setval{olmo.geoma.minus.gdpo-noaux.if-abs}{3.0}
\setval{olmo.geoma.minus.gdpo-noaux.ifbench}{$-$3.0}
\setval{olmo.geoma.minus.gdpo-noaux.ifbench-abs}{3.0}
\setval{olmo.geoma.minus.gdpo-noaux.ifeval}{+3.2}
\setval{olmo.geoma.minus.gdpo-noaux.ifeval-abs}{3.2}
\setval{olmo.geoma.minus.gdpo-noaux.judge2.avg}{$-$5.0}
\setval{olmo.geoma.minus.gdpo-noaux.judge2.avg-abs}{5.0}
\setval{olmo.geoma.minus.gdpo-noaux.judge2.creative-writing-v3}{+10.8}
\setval{olmo.geoma.minus.gdpo-noaux.judge2.creative-writing-v3-abs}{10.8}
\setval{olmo.geoma.minus.gdpo-noaux.judge2.longbench-write}{+11.6}
\setval{olmo.geoma.minus.gdpo-noaux.judge2.longbench-write-abs}{11.6}
\setval{olmo.geoma.minus.gdpo-noaux.judge2.writing}{+10.4}
\setval{olmo.geoma.minus.gdpo-noaux.judge2.writing-abs}{10.4}
\setval{olmo.geoma.minus.gdpo-noaux.judge2.writingbench}{+8.9}
\setval{olmo.geoma.minus.gdpo-noaux.judge2.writingbench-abs}{8.9}
\setval{olmo.geoma.minus.gdpo-noaux.livecodebench}{$-$9.2}
\setval{olmo.geoma.minus.gdpo-noaux.livecodebench-abs}{9.2}
\setval{olmo.geoma.minus.gdpo-noaux.longbench-write}{+9.3}
\setval{olmo.geoma.minus.gdpo-noaux.longbench-write-abs}{9.3}
\setval{olmo.geoma.minus.gdpo-noaux.math}{$-$11.9}
\setval{olmo.geoma.minus.gdpo-noaux.math-abs}{11.9}
\setval{olmo.geoma.minus.gdpo-noaux.math500}{$-$7.7}
\setval{olmo.geoma.minus.gdpo-noaux.math500-abs}{7.7}
\setval{olmo.geoma.minus.gdpo-noaux.mbpp-plus}{$-$5.8}
\setval{olmo.geoma.minus.gdpo-noaux.mbpp-plus-abs}{5.8}
\setval{olmo.geoma.minus.gdpo-noaux.minerva}{$-$6.2}
\setval{olmo.geoma.minus.gdpo-noaux.minerva-abs}{6.2}
\setval{olmo.geoma.minus.gdpo-noaux.olympiad-bench}{$-$11.0}
\setval{olmo.geoma.minus.gdpo-noaux.olympiad-bench-abs}{11.0}
\setval{olmo.geoma.minus.gdpo-noaux.writing}{+10.7}
\setval{olmo.geoma.minus.gdpo-noaux.writing-abs}{10.7}
\setval{olmo.geoma.minus.gdpo-noaux.writingbench}{+11.5}
\setval{olmo.geoma.minus.gdpo-noaux.writingbench-abs}{11.5}
\setval{olmo.geoma.minus.gdpo-w0.5.aime24}{$-$5.8}
\setval{olmo.geoma.minus.gdpo-w0.5.aime24-abs}{5.8}
\setval{olmo.geoma.minus.gdpo-w0.5.aime25}{$-$5.3}
\setval{olmo.geoma.minus.gdpo-w0.5.aime25-abs}{5.3}
\setval{olmo.geoma.minus.gdpo-w0.5.amc23}{$-$10.1}
\setval{olmo.geoma.minus.gdpo-w0.5.amc23-abs}{10.1}
\setval{olmo.geoma.minus.gdpo-w0.5.avg}{$-$2.7}
\setval{olmo.geoma.minus.gdpo-w0.5.avg-abs}{2.7}
\setval{olmo.geoma.minus.gdpo-w0.5.code}{$-$1.9}
\setval{olmo.geoma.minus.gdpo-w0.5.code-abs}{1.9}
\setval{olmo.geoma.minus.gdpo-w0.5.creative-writing-v3}{+1.5}
\setval{olmo.geoma.minus.gdpo-w0.5.creative-writing-v3-abs}{1.5}
\setval{olmo.geoma.minus.gdpo-w0.5.guidebench}{$-$5.3}
\setval{olmo.geoma.minus.gdpo-w0.5.guidebench-abs}{5.3}
\setval{olmo.geoma.minus.gdpo-w0.5.humaneval-plus}{$-$3.1}
\setval{olmo.geoma.minus.gdpo-w0.5.humaneval-plus-abs}{3.1}
\setval{olmo.geoma.minus.gdpo-w0.5.if}{$-$2.4}
\setval{olmo.geoma.minus.gdpo-w0.5.if-abs}{2.4}
\setval{olmo.geoma.minus.gdpo-w0.5.ifbench}{$-$2.3}
\setval{olmo.geoma.minus.gdpo-w0.5.ifbench-abs}{2.3}
\setval{olmo.geoma.minus.gdpo-w0.5.ifeval}{+0.5}
\setval{olmo.geoma.minus.gdpo-w0.5.ifeval-abs}{0.5}
\setval{olmo.geoma.minus.gdpo-w0.5.judge2.avg}{$-$2.6}
\setval{olmo.geoma.minus.gdpo-w0.5.judge2.avg-abs}{2.6}
\setval{olmo.geoma.minus.gdpo-w0.5.judge2.creative-writing-v3}{+2.4}
\setval{olmo.geoma.minus.gdpo-w0.5.judge2.creative-writing-v3-abs}{2.4}
\setval{olmo.geoma.minus.gdpo-w0.5.judge2.longbench-write}{+1.6}
\setval{olmo.geoma.minus.gdpo-w0.5.judge2.longbench-write-abs}{1.6}
\setval{olmo.geoma.minus.gdpo-w0.5.judge2.writing}{+2.0}
\setval{olmo.geoma.minus.gdpo-w0.5.judge2.writing-abs}{2.0}
\setval{olmo.geoma.minus.gdpo-w0.5.judge2.writingbench}{+1.9}
\setval{olmo.geoma.minus.gdpo-w0.5.judge2.writingbench-abs}{1.9}
\setval{olmo.geoma.minus.gdpo-w0.5.livecodebench}{$-$2.5}
\setval{olmo.geoma.minus.gdpo-w0.5.livecodebench-abs}{2.5}
\setval{olmo.geoma.minus.gdpo-w0.5.longbench-write}{+1.6}
\setval{olmo.geoma.minus.gdpo-w0.5.longbench-write-abs}{1.6}
\setval{olmo.geoma.minus.gdpo-w0.5.math}{$-$5.2}
\setval{olmo.geoma.minus.gdpo-w0.5.math-abs}{5.2}
\setval{olmo.geoma.minus.gdpo-w0.5.math500}{$-$3.9}
\setval{olmo.geoma.minus.gdpo-w0.5.math500-abs}{3.9}
\setval{olmo.geoma.minus.gdpo-w0.5.mbpp-plus}{$-$0.2}
\setval{olmo.geoma.minus.gdpo-w0.5.mbpp-plus-abs}{0.2}
\setval{olmo.geoma.minus.gdpo-w0.5.minerva}{$-$1.6}
\setval{olmo.geoma.minus.gdpo-w0.5.minerva-abs}{1.6}
\setval{olmo.geoma.minus.gdpo-w0.5.olympiad-bench}{$-$4.8}
\setval{olmo.geoma.minus.gdpo-w0.5.olympiad-bench-abs}{4.8}
\setval{olmo.geoma.minus.gdpo-w0.5.writing}{+1.2}
\setval{olmo.geoma.minus.gdpo-w0.5.writing-abs}{1.2}
\setval{olmo.geoma.minus.gdpo-w0.5.writingbench}{+0.5}
\setval{olmo.geoma.minus.gdpo-w0.5.writingbench-abs}{0.5}
\setval{olmo.geoma.minus.gdpo.aime24}{$-$4.1}
\setval{olmo.geoma.minus.gdpo.aime24-abs}{4.1}
\setval{olmo.geoma.minus.gdpo.aime25}{$-$0.6}
\setval{olmo.geoma.minus.gdpo.aime25-abs}{0.6}
\setval{olmo.geoma.minus.gdpo.amc23}{$-$5.6}
\setval{olmo.geoma.minus.gdpo.amc23-abs}{5.6}
\setval{olmo.geoma.minus.gdpo.avg}{$-$1.0}
\setval{olmo.geoma.minus.gdpo.avg-abs}{1.0}
\setval{olmo.geoma.minus.gdpo.code}{$-$0.8}
\setval{olmo.geoma.minus.gdpo.code-abs}{0.8}
\setval{olmo.geoma.minus.gdpo.creative-writing-v3}{+2.5}
\setval{olmo.geoma.minus.gdpo.creative-writing-v3-abs}{2.5}
\setval{olmo.geoma.minus.gdpo.guidebench}{$-$4.2}
\setval{olmo.geoma.minus.gdpo.guidebench-abs}{4.2}
\setval{olmo.geoma.minus.gdpo.humaneval-plus}{$-$1.7}
\setval{olmo.geoma.minus.gdpo.humaneval-plus-abs}{1.7}
\setval{olmo.geoma.minus.gdpo.if}{$-$0.7}
\setval{olmo.geoma.minus.gdpo.if-abs}{0.7}
\setval{olmo.geoma.minus.gdpo.ifbench}{+0.7}
\setval{olmo.geoma.minus.gdpo.ifbench-abs}{0.7}
\setval{olmo.geoma.minus.gdpo.ifeval}{+1.6}
\setval{olmo.geoma.minus.gdpo.ifeval-abs}{1.6}
\setval{olmo.geoma.minus.gdpo.judge2.avg}{$-$0.8}
\setval{olmo.geoma.minus.gdpo.judge2.avg-abs}{0.8}
\setval{olmo.geoma.minus.gdpo.judge2.creative-writing-v3}{+3.9}
\setval{olmo.geoma.minus.gdpo.judge2.creative-writing-v3-abs}{3.9}
\setval{olmo.geoma.minus.gdpo.judge2.longbench-write}{+2.5}
\setval{olmo.geoma.minus.gdpo.judge2.longbench-write-abs}{2.5}
\setval{olmo.geoma.minus.gdpo.judge2.writing}{+2.7}
\setval{olmo.geoma.minus.gdpo.judge2.writing-abs}{2.7}
\setval{olmo.geoma.minus.gdpo.judge2.writingbench}{+1.6}
\setval{olmo.geoma.minus.gdpo.judge2.writingbench-abs}{1.6}
\setval{olmo.geoma.minus.gdpo.livecodebench}{$-$0.8}
\setval{olmo.geoma.minus.gdpo.livecodebench-abs}{0.8}
\setval{olmo.geoma.minus.gdpo.longbench-write}{+2.1}
\setval{olmo.geoma.minus.gdpo.longbench-write-abs}{2.1}
\setval{olmo.geoma.minus.gdpo.math}{$-$2.7}
\setval{olmo.geoma.minus.gdpo.math-abs}{2.7}
\setval{olmo.geoma.minus.gdpo.math500}{$-$2.1}
\setval{olmo.geoma.minus.gdpo.math500-abs}{2.1}
\setval{olmo.geoma.minus.gdpo.mbpp-plus}{+0.0}
\setval{olmo.geoma.minus.gdpo.mbpp-plus-abs}{0.0}
\setval{olmo.geoma.minus.gdpo.minerva}{$-$0.4}
\setval{olmo.geoma.minus.gdpo.minerva-abs}{0.4}
\setval{olmo.geoma.minus.gdpo.olympiad-bench}{$-$3.1}
\setval{olmo.geoma.minus.gdpo.olympiad-bench-abs}{3.1}
\setval{olmo.geoma.minus.gdpo.writing}{+2.1}
\setval{olmo.geoma.minus.gdpo.writing-abs}{2.1}
\setval{olmo.geoma.minus.gdpo.writingbench}{+1.7}
\setval{olmo.geoma.minus.gdpo.writingbench-abs}{1.7}
\setval{olmo.geoma.minus.grpo-nodiv.aime24}{$-$8.0}
\setval{olmo.geoma.minus.grpo-nodiv.aime24-abs}{8.0}
\setval{olmo.geoma.minus.grpo-nodiv.aime25}{$-$5.6}
\setval{olmo.geoma.minus.grpo-nodiv.aime25-abs}{5.6}
\setval{olmo.geoma.minus.grpo-nodiv.amc23}{$-$9.1}
\setval{olmo.geoma.minus.grpo-nodiv.amc23-abs}{9.1}
\setval{olmo.geoma.minus.grpo-nodiv.avg}{$-$3.0}
\setval{olmo.geoma.minus.grpo-nodiv.avg-abs}{3.0}
\setval{olmo.geoma.minus.grpo-nodiv.code}{$-$1.5}
\setval{olmo.geoma.minus.grpo-nodiv.code-abs}{1.5}
\setval{olmo.geoma.minus.grpo-nodiv.creative-writing-v3}{+2.6}
\setval{olmo.geoma.minus.grpo-nodiv.creative-writing-v3-abs}{2.6}
\setval{olmo.geoma.minus.grpo-nodiv.guidebench}{$-$8.8}
\setval{olmo.geoma.minus.grpo-nodiv.guidebench-abs}{8.8}
\setval{olmo.geoma.minus.grpo-nodiv.humaneval-plus}{$-$2.7}
\setval{olmo.geoma.minus.grpo-nodiv.humaneval-plus-abs}{2.7}
\setval{olmo.geoma.minus.grpo-nodiv.if}{$-$3.2}
\setval{olmo.geoma.minus.grpo-nodiv.if-abs}{3.2}
\setval{olmo.geoma.minus.grpo-nodiv.ifbench}{$-$1.0}
\setval{olmo.geoma.minus.grpo-nodiv.ifbench-abs}{1.0}
\setval{olmo.geoma.minus.grpo-nodiv.ifeval}{+0.2}
\setval{olmo.geoma.minus.grpo-nodiv.ifeval-abs}{0.2}
\setval{olmo.geoma.minus.grpo-nodiv.judge2.avg}{$-$3.1}
\setval{olmo.geoma.minus.grpo-nodiv.judge2.avg-abs}{3.1}
\setval{olmo.geoma.minus.grpo-nodiv.judge2.creative-writing-v3}{+2.9}
\setval{olmo.geoma.minus.grpo-nodiv.judge2.creative-writing-v3-abs}{2.9}
\setval{olmo.geoma.minus.grpo-nodiv.judge2.longbench-write}{$-$2.9}
\setval{olmo.geoma.minus.grpo-nodiv.judge2.longbench-write-abs}{2.9}
\setval{olmo.geoma.minus.grpo-nodiv.judge2.writing}{+0.6}
\setval{olmo.geoma.minus.grpo-nodiv.judge2.writing-abs}{0.6}
\setval{olmo.geoma.minus.grpo-nodiv.judge2.writingbench}{+1.7}
\setval{olmo.geoma.minus.grpo-nodiv.judge2.writingbench-abs}{1.7}
\setval{olmo.geoma.minus.grpo-nodiv.livecodebench}{$-$1.3}
\setval{olmo.geoma.minus.grpo-nodiv.livecodebench-abs}{1.3}
\setval{olmo.geoma.minus.grpo-nodiv.longbench-write}{$-$2.2}
\setval{olmo.geoma.minus.grpo-nodiv.longbench-write-abs}{2.2}
\setval{olmo.geoma.minus.grpo-nodiv.math}{$-$5.8}
\setval{olmo.geoma.minus.grpo-nodiv.math-abs}{5.8}
\setval{olmo.geoma.minus.grpo-nodiv.math500}{$-$4.3}
\setval{olmo.geoma.minus.grpo-nodiv.math500-abs}{4.3}
\setval{olmo.geoma.minus.grpo-nodiv.mbpp-plus}{$-$0.6}
\setval{olmo.geoma.minus.grpo-nodiv.mbpp-plus-abs}{0.6}
\setval{olmo.geoma.minus.grpo-nodiv.minerva}{$-$1.8}
\setval{olmo.geoma.minus.grpo-nodiv.minerva-abs}{1.8}
\setval{olmo.geoma.minus.grpo-nodiv.olympiad-bench}{$-$5.7}
\setval{olmo.geoma.minus.grpo-nodiv.olympiad-bench-abs}{5.7}
\setval{olmo.geoma.minus.grpo-nodiv.writing}{+1.4}
\setval{olmo.geoma.minus.grpo-nodiv.writing-abs}{1.4}
\setval{olmo.geoma.minus.grpo-nodiv.writingbench}{+3.8}
\setval{olmo.geoma.minus.grpo-nodiv.writingbench-abs}{3.8}
\setval{olmo.geoma.minus.grpo.aime24}{$-$5.7}
\setval{olmo.geoma.minus.grpo.aime24-abs}{5.7}
\setval{olmo.geoma.minus.grpo.aime25}{$-$2.7}
\setval{olmo.geoma.minus.grpo.aime25-abs}{2.7}
\setval{olmo.geoma.minus.grpo.amc23}{$-$7.0}
\setval{olmo.geoma.minus.grpo.amc23-abs}{7.0}
\setval{olmo.geoma.minus.grpo.avg}{$-$2.3}
\setval{olmo.geoma.minus.grpo.avg-abs}{2.3}
\setval{olmo.geoma.minus.grpo.code}{$-$1.2}
\setval{olmo.geoma.minus.grpo.code-abs}{1.2}
\setval{olmo.geoma.minus.grpo.creative-writing-v3}{+0.2}
\setval{olmo.geoma.minus.grpo.creative-writing-v3-abs}{0.2}
\setval{olmo.geoma.minus.grpo.guidebench}{$-$9.8}
\setval{olmo.geoma.minus.grpo.guidebench-abs}{9.8}
\setval{olmo.geoma.minus.grpo.humaneval-plus}{$-$2.7}
\setval{olmo.geoma.minus.grpo.humaneval-plus-abs}{2.7}
\setval{olmo.geoma.minus.grpo.if}{$-$3.6}
\setval{olmo.geoma.minus.grpo.if-abs}{3.6}
\setval{olmo.geoma.minus.grpo.ifbench}{$-$1.7}
\setval{olmo.geoma.minus.grpo.ifbench-abs}{1.7}
\setval{olmo.geoma.minus.grpo.ifeval}{+0.6}
\setval{olmo.geoma.minus.grpo.ifeval-abs}{0.6}
\setval{olmo.geoma.minus.grpo.judge2.avg}{$-$2.0}
\setval{olmo.geoma.minus.grpo.judge2.avg-abs}{2.0}
\setval{olmo.geoma.minus.grpo.judge2.creative-writing-v3}{+3.2}
\setval{olmo.geoma.minus.grpo.judge2.creative-writing-v3-abs}{3.2}
\setval{olmo.geoma.minus.grpo.judge2.longbench-write}{+1.1}
\setval{olmo.geoma.minus.grpo.judge2.longbench-write-abs}{1.1}
\setval{olmo.geoma.minus.grpo.judge2.writing}{+1.5}
\setval{olmo.geoma.minus.grpo.judge2.writing-abs}{1.5}
\setval{olmo.geoma.minus.grpo.judge2.writingbench}{+0.1}
\setval{olmo.geoma.minus.grpo.judge2.writingbench-abs}{0.1}
\setval{olmo.geoma.minus.grpo.livecodebench}{+0.0}
\setval{olmo.geoma.minus.grpo.livecodebench-abs}{0.0}
\setval{olmo.geoma.minus.grpo.longbench-write}{+1.1}
\setval{olmo.geoma.minus.grpo.longbench-write-abs}{1.1}
\setval{olmo.geoma.minus.grpo.math}{$-$3.3}
\setval{olmo.geoma.minus.grpo.math-abs}{3.3}
\setval{olmo.geoma.minus.grpo.math500}{$-$1.8}
\setval{olmo.geoma.minus.grpo.math500-abs}{1.8}
\setval{olmo.geoma.minus.grpo.mbpp-plus}{$-$1.1}
\setval{olmo.geoma.minus.grpo.mbpp-plus-abs}{1.1}
\setval{olmo.geoma.minus.grpo.minerva}{+0.0}
\setval{olmo.geoma.minus.grpo.minerva-abs}{0.0}
\setval{olmo.geoma.minus.grpo.olympiad-bench}{$-$2.9}
\setval{olmo.geoma.minus.grpo.olympiad-bench-abs}{2.9}
\setval{olmo.geoma.minus.grpo.writing}{$-$0.1}
\setval{olmo.geoma.minus.grpo.writing-abs}{0.1}
\setval{olmo.geoma.minus.grpo.writingbench}{$-$1.5}
\setval{olmo.geoma.minus.grpo.writingbench-abs}{1.5}
\setval{olmo.geoma.minus.mt-grpo.aime24}{$-$12.8}
\setval{olmo.geoma.minus.mt-grpo.aime24-abs}{12.8}
\setval{olmo.geoma.minus.mt-grpo.aime25}{$-$6.1}
\setval{olmo.geoma.minus.mt-grpo.aime25-abs}{6.1}
\setval{olmo.geoma.minus.mt-grpo.amc23}{$-$13.5}
\setval{olmo.geoma.minus.mt-grpo.amc23-abs}{13.5}
\setval{olmo.geoma.minus.mt-grpo.avg}{$-$3.8}
\setval{olmo.geoma.minus.mt-grpo.avg-abs}{3.8}
\setval{olmo.geoma.minus.mt-grpo.code}{$-$0.5}
\setval{olmo.geoma.minus.mt-grpo.code-abs}{0.5}
\setval{olmo.geoma.minus.mt-grpo.creative-writing-v3}{$-$1.1}
\setval{olmo.geoma.minus.mt-grpo.creative-writing-v3-abs}{1.1}
\setval{olmo.geoma.minus.mt-grpo.guidebench}{$-$7.9}
\setval{olmo.geoma.minus.mt-grpo.guidebench-abs}{7.9}
\setval{olmo.geoma.minus.mt-grpo.humaneval-plus}{$-$1.1}
\setval{olmo.geoma.minus.mt-grpo.humaneval-plus-abs}{1.1}
\setval{olmo.geoma.minus.mt-grpo.if}{$-$2.9}
\setval{olmo.geoma.minus.mt-grpo.if-abs}{2.9}
\setval{olmo.geoma.minus.mt-grpo.ifbench}{$-$1.7}
\setval{olmo.geoma.minus.mt-grpo.ifbench-abs}{1.7}
\setval{olmo.geoma.minus.mt-grpo.ifeval}{+1.0}
\setval{olmo.geoma.minus.mt-grpo.ifeval-abs}{1.0}
\setval{olmo.geoma.minus.mt-grpo.judge2.avg}{$-$3.5}
\setval{olmo.geoma.minus.mt-grpo.judge2.avg-abs}{3.5}
\setval{olmo.geoma.minus.mt-grpo.judge2.creative-writing-v3}{+1.9}
\setval{olmo.geoma.minus.mt-grpo.judge2.creative-writing-v3-abs}{1.9}
\setval{olmo.geoma.minus.mt-grpo.judge2.longbench-write}{$-$2.2}
\setval{olmo.geoma.minus.mt-grpo.judge2.longbench-write-abs}{2.2}
\setval{olmo.geoma.minus.mt-grpo.judge2.writing}{+0.3}
\setval{olmo.geoma.minus.mt-grpo.judge2.writing-abs}{0.3}
\setval{olmo.geoma.minus.mt-grpo.judge2.writingbench}{+1.1}
\setval{olmo.geoma.minus.mt-grpo.judge2.writingbench-abs}{1.1}
\setval{olmo.geoma.minus.mt-grpo.livecodebench}{$-$0.3}
\setval{olmo.geoma.minus.mt-grpo.livecodebench-abs}{0.3}
\setval{olmo.geoma.minus.mt-grpo.longbench-write}{$-$1.0}
\setval{olmo.geoma.minus.mt-grpo.longbench-write-abs}{1.0}
\setval{olmo.geoma.minus.mt-grpo.math}{$-$7.2}
\setval{olmo.geoma.minus.mt-grpo.math-abs}{7.2}
\setval{olmo.geoma.minus.mt-grpo.math500}{$-$3.7}
\setval{olmo.geoma.minus.mt-grpo.math500-abs}{3.7}
\setval{olmo.geoma.minus.mt-grpo.mbpp-plus}{$-$0.2}
\setval{olmo.geoma.minus.mt-grpo.mbpp-plus-abs}{0.2}
\setval{olmo.geoma.minus.mt-grpo.minerva}{$-$2.0}
\setval{olmo.geoma.minus.mt-grpo.minerva-abs}{2.0}
\setval{olmo.geoma.minus.mt-grpo.olympiad-bench}{$-$5.3}
\setval{olmo.geoma.minus.mt-grpo.olympiad-bench-abs}{5.3}
\setval{olmo.geoma.minus.mt-grpo.writing}{$-$1.0}
\setval{olmo.geoma.minus.mt-grpo.writing-abs}{1.0}
\setval{olmo.geoma.minus.mt-grpo.writingbench}{$-$0.7}
\setval{olmo.geoma.minus.mt-grpo.writingbench-abs}{0.7}
\setval{olmo.geoma.minus.papo.aime24}{+1.6}
\setval{olmo.geoma.minus.papo.aime24-abs}{1.6}
\setval{olmo.geoma.minus.papo.aime25}{+1.7}
\setval{olmo.geoma.minus.papo.aime25-abs}{1.7}
\setval{olmo.geoma.minus.papo.amc23}{+4.5}
\setval{olmo.geoma.minus.papo.amc23-abs}{4.5}
\setval{olmo.geoma.minus.papo.avg}{+0.6}
\setval{olmo.geoma.minus.papo.avg-abs}{0.6}
\setval{olmo.geoma.minus.papo.code}{+0.0}
\setval{olmo.geoma.minus.papo.code-abs}{0.0}
\setval{olmo.geoma.minus.papo.creative-writing-v3}{+2.6}
\setval{olmo.geoma.minus.papo.creative-writing-v3-abs}{2.6}
\setval{olmo.geoma.minus.papo.guidebench}{$-$12.0}
\setval{olmo.geoma.minus.papo.guidebench-abs}{12.0}
\setval{olmo.geoma.minus.papo.humaneval-plus}{$-$0.1}
\setval{olmo.geoma.minus.papo.humaneval-plus-abs}{0.1}
\setval{olmo.geoma.minus.papo.if}{$-$3.5}
\setval{olmo.geoma.minus.papo.if-abs}{3.5}
\setval{olmo.geoma.minus.papo.ifbench}{+0.7}
\setval{olmo.geoma.minus.papo.ifbench-abs}{0.7}
\setval{olmo.geoma.minus.papo.ifeval}{+0.8}
\setval{olmo.geoma.minus.papo.ifeval-abs}{0.8}
\setval{olmo.geoma.minus.papo.judge2.avg}{+0.8}
\setval{olmo.geoma.minus.papo.judge2.avg-abs}{0.8}
\setval{olmo.geoma.minus.papo.judge2.creative-writing-v3}{+3.8}
\setval{olmo.geoma.minus.papo.judge2.creative-writing-v3-abs}{3.8}
\setval{olmo.geoma.minus.papo.judge2.longbench-write}{+2.6}
\setval{olmo.geoma.minus.papo.judge2.longbench-write-abs}{2.6}
\setval{olmo.geoma.minus.papo.judge2.writing}{+2.5}
\setval{olmo.geoma.minus.papo.judge2.writing-abs}{2.5}
\setval{olmo.geoma.minus.papo.judge2.writingbench}{+1.1}
\setval{olmo.geoma.minus.papo.judge2.writingbench-abs}{1.1}
\setval{olmo.geoma.minus.papo.livecodebench}{+0.5}
\setval{olmo.geoma.minus.papo.livecodebench-abs}{0.5}
\setval{olmo.geoma.minus.papo.longbench-write}{+0.6}
\setval{olmo.geoma.minus.papo.longbench-write-abs}{0.6}
\setval{olmo.geoma.minus.papo.math}{+2.4}
\setval{olmo.geoma.minus.papo.math-abs}{2.4}
\setval{olmo.geoma.minus.papo.math500}{+2.3}
\setval{olmo.geoma.minus.papo.math500-abs}{2.3}
\setval{olmo.geoma.minus.papo.mbpp-plus}{$-$0.3}
\setval{olmo.geoma.minus.papo.mbpp-plus-abs}{0.3}
\setval{olmo.geoma.minus.papo.minerva}{+1.4}
\setval{olmo.geoma.minus.papo.minerva-abs}{1.4}
\setval{olmo.geoma.minus.papo.olympiad-bench}{+3.0}
\setval{olmo.geoma.minus.papo.olympiad-bench-abs}{3.0}
\setval{olmo.geoma.minus.papo.writing}{+1.6}
\setval{olmo.geoma.minus.papo.writing-abs}{1.6}
\setval{olmo.geoma.minus.papo.writingbench}{+1.6}
\setval{olmo.geoma.minus.papo.writingbench-abs}{1.6}
\setval{olmo.geoma.minus.saw.aime24}{$-$2.4}
\setval{olmo.geoma.minus.saw.aime24-abs}{2.4}
\setval{olmo.geoma.minus.saw.aime25}{$-$3.2}
\setval{olmo.geoma.minus.saw.aime25-abs}{3.2}
\setval{olmo.geoma.minus.saw.amc23}{$-$3.0}
\setval{olmo.geoma.minus.saw.amc23-abs}{3.0}
\setval{olmo.geoma.minus.saw.avg}{$-$1.5}
\setval{olmo.geoma.minus.saw.avg-abs}{1.5}
\setval{olmo.geoma.minus.saw.code}{$-$0.6}
\setval{olmo.geoma.minus.saw.code-abs}{0.6}
\setval{olmo.geoma.minus.saw.creative-writing-v3}{+1.7}
\setval{olmo.geoma.minus.saw.creative-writing-v3-abs}{1.7}
\setval{olmo.geoma.minus.saw.guidebench}{$-$8.1}
\setval{olmo.geoma.minus.saw.guidebench-abs}{8.1}
\setval{olmo.geoma.minus.saw.humaneval-plus}{$-$1.7}
\setval{olmo.geoma.minus.saw.humaneval-plus-abs}{1.7}
\setval{olmo.geoma.minus.saw.if}{$-$3.7}
\setval{olmo.geoma.minus.saw.if-abs}{3.7}
\setval{olmo.geoma.minus.saw.ifbench}{$-$2.3}
\setval{olmo.geoma.minus.saw.ifbench-abs}{2.3}
\setval{olmo.geoma.minus.saw.ifeval}{$-$0.6}
\setval{olmo.geoma.minus.saw.ifeval-abs}{0.6}
\setval{olmo.geoma.minus.saw.judge2.avg}{$-$1.5}
\setval{olmo.geoma.minus.saw.judge2.avg-abs}{1.5}
\setval{olmo.geoma.minus.saw.judge2.creative-writing-v3}{+0.8}
\setval{olmo.geoma.minus.saw.judge2.creative-writing-v3-abs}{0.8}
\setval{olmo.geoma.minus.saw.judge2.longbench-write}{+0.8}
\setval{olmo.geoma.minus.saw.judge2.longbench-write-abs}{0.8}
\setval{olmo.geoma.minus.saw.judge2.writing}{+0.6}
\setval{olmo.geoma.minus.saw.judge2.writing-abs}{0.6}
\setval{olmo.geoma.minus.saw.judge2.writingbench}{+0.2}
\setval{olmo.geoma.minus.saw.judge2.writingbench-abs}{0.2}
\setval{olmo.geoma.minus.saw.livecodebench}{+0.5}
\setval{olmo.geoma.minus.saw.livecodebench-abs}{0.5}
\setval{olmo.geoma.minus.saw.longbench-write}{+0.4}
\setval{olmo.geoma.minus.saw.longbench-write-abs}{0.4}
\setval{olmo.geoma.minus.saw.math}{$-$1.9}
\setval{olmo.geoma.minus.saw.math-abs}{1.9}
\setval{olmo.geoma.minus.saw.math500}{$-$1.3}
\setval{olmo.geoma.minus.saw.math500-abs}{1.3}
\setval{olmo.geoma.minus.saw.mbpp-plus}{$-$0.7}
\setval{olmo.geoma.minus.saw.mbpp-plus-abs}{0.7}
\setval{olmo.geoma.minus.saw.minerva}{$-$0.4}
\setval{olmo.geoma.minus.saw.minerva-abs}{0.4}
\setval{olmo.geoma.minus.saw.olympiad-bench}{$-$1.1}
\setval{olmo.geoma.minus.saw.olympiad-bench-abs}{1.1}
\setval{olmo.geoma.minus.saw.writing}{+0.6}
\setval{olmo.geoma.minus.saw.writing-abs}{0.6}
\setval{olmo.geoma.minus.saw.writingbench}{$-$0.4}
\setval{olmo.geoma.minus.saw.writingbench-abs}{0.4}
\setval{olmo.geoma.passing-at-ceiling.code.early}{0.83}
\setval{olmo.geoma.passing-at-ceiling.code.late}{0.85}
\setval{olmo.geoma.passing-at-ceiling.code.run}{0.83}
\setval{olmo.geoma.passing-at-ceiling.if.early}{0.49}
\setval{olmo.geoma.passing-at-ceiling.if.late}{0.57}
\setval{olmo.geoma.passing-at-ceiling.if.run}{0.54}
\setval{olmo.geoma.passing-at-ceiling.math.early}{1.00}
\setval{olmo.geoma.passing-at-ceiling.math.late}{1.00}
\setval{olmo.geoma.passing-at-ceiling.math.run}{1.00}
\setval{olmo.geoma.passing-at-ceiling.writing.early}{0.00}
\setval{olmo.geoma.passing-at-ceiling.writing.late}{0.00}
\setval{olmo.geoma.passing-at-ceiling.writing.run}{0.00}
\setval{olmo.geoma.primary-cosine.code.early}{0.87}
\setval{olmo.geoma.primary-cosine.code.late}{0.90}
\setval{olmo.geoma.primary-cosine.code.run}{0.91}
\setval{olmo.geoma.primary-cosine.if.early}{0.63}
\setval{olmo.geoma.primary-cosine.if.late}{0.70}
\setval{olmo.geoma.primary-cosine.if.run}{0.70}
\setval{olmo.geoma.primary-cosine.math.early}{0.93}
\setval{olmo.geoma.primary-cosine.math.late}{0.87}
\setval{olmo.geoma.primary-cosine.math.run}{0.88}
\setval{olmo.geoma.primary-cosine.writing.early}{0.83}
\setval{olmo.geoma.primary-cosine.writing.late}{0.71}
\setval{olmo.geoma.primary-cosine.writing.run}{0.73}
\setval{olmo.geoma.primary-level.code.early}{0.59}
\setval{olmo.geoma.primary-level.code.late}{0.82}
\setval{olmo.geoma.primary-level.code.run}{0.74}
\setval{olmo.geoma.primary-level.if.early}{0.63}
\setval{olmo.geoma.primary-level.if.late}{0.75}
\setval{olmo.geoma.primary-level.if.run}{0.70}
\setval{olmo.geoma.primary-level.math.early}{0.34}
\setval{olmo.geoma.primary-level.math.late}{0.63}
\setval{olmo.geoma.primary-level.math.run}{0.54}
\setval{olmo.geoma.primary-level.writing.early}{0.50}
\setval{olmo.geoma.primary-level.writing.late}{0.69}
\setval{olmo.geoma.primary-level.writing.run}{0.63}
\setval{olmo.geoma.solved.code.early}{0.12}
\setval{olmo.geoma.solved.code.late}{0.09}
\setval{olmo.geoma.solved.code.run}{0.09}
\setval{olmo.geoma.solved.if.early}{0.17}
\setval{olmo.geoma.solved.if.late}{0.18}
\setval{olmo.geoma.solved.if.run}{0.16}
\setval{olmo.geoma.solved.math.early}{0.06}
\setval{olmo.geoma.solved.math.late}{0.17}
\setval{olmo.geoma.solved.math.run}{0.13}
\setval{olmo.geoma.solved.writing.early}{0.00}
\setval{olmo.geoma.solved.writing.late}{0.00}
\setval{olmo.geoma.solved.writing.run}{0.00}
\setval{olmo.geoma.update-share.code.early}{0.23}
\setval{olmo.geoma.update-share.code.late}{0.16}
\setval{olmo.geoma.update-share.code.run}{0.18}
\setval{olmo.geoma.update-share.if.early}{0.27}
\setval{olmo.geoma.update-share.if.late}{0.23}
\setval{olmo.geoma.update-share.if.run}{0.24}
\setval{olmo.geoma.update-share.math.early}{0.22}
\setval{olmo.geoma.update-share.math.late}{0.26}
\setval{olmo.geoma.update-share.math.run}{0.25}
\setval{olmo.geoma.update-share.writing.early}{0.28}
\setval{olmo.geoma.update-share.writing.late}{0.35}
\setval{olmo.geoma.update-share.writing.run}{0.32}
\setval{olmo.grpo-nodiv.aux-change.code}{+0.16}
\setval{olmo.grpo-nodiv.aux-change.if}{+0.11}
\setval{olmo.grpo-nodiv.aux-change.math}{$-$0.15}
\setval{olmo.grpo-nodiv.aux-change.writing}{+0.01}
\setval{olmo.grpo-nodiv.aux-change.writing-rm}{+0.21}
\setval{olmo.grpo-nodiv.aux-full.code.early}{0.76}
\setval{olmo.grpo-nodiv.aux-full.code.late}{0.88}
\setval{olmo.grpo-nodiv.aux-full.code.run}{0.86}
\setval{olmo.grpo-nodiv.aux-full.if.early}{0.59}
\setval{olmo.grpo-nodiv.aux-full.if.late}{0.89}
\setval{olmo.grpo-nodiv.aux-full.if.run}{0.79}
\setval{olmo.grpo-nodiv.aux-full.math.early}{0.53}
\setval{olmo.grpo-nodiv.aux-full.math.late}{0.53}
\setval{olmo.grpo-nodiv.aux-full.math.run}{0.40}
\setval{olmo.grpo-nodiv.aux-full.writing-rm.early}{0.02}
\setval{olmo.grpo-nodiv.aux-full.writing-rm.late}{0.09}
\setval{olmo.grpo-nodiv.aux-full.writing-rm.run}{0.05}
\setval{olmo.grpo-nodiv.aux-full.writing.early}{0.56}
\setval{olmo.grpo-nodiv.aux-full.writing.late}{0.59}
\setval{olmo.grpo-nodiv.aux-full.writing.run}{0.55}
\setval{olmo.grpo-nodiv.aux-level.code.early}{0.92}
\setval{olmo.grpo-nodiv.aux-level.code.late}{0.97}
\setval{olmo.grpo-nodiv.aux-level.code.run}{0.96}
\setval{olmo.grpo-nodiv.aux-level.if.early}{0.90}
\setval{olmo.grpo-nodiv.aux-level.if.late}{0.98}
\setval{olmo.grpo-nodiv.aux-level.if.run}{0.96}
\setval{olmo.grpo-nodiv.aux-level.math.early}{0.91}
\setval{olmo.grpo-nodiv.aux-level.math.late}{0.75}
\setval{olmo.grpo-nodiv.aux-level.math.run}{0.72}
\setval{olmo.grpo-nodiv.aux-level.writing-rm.early}{0.60}
\setval{olmo.grpo-nodiv.aux-level.writing-rm.late}{0.78}
\setval{olmo.grpo-nodiv.aux-level.writing-rm.run}{0.72}
\setval{olmo.grpo-nodiv.aux-level.writing.early}{0.95}
\setval{olmo.grpo-nodiv.aux-level.writing.late}{0.95}
\setval{olmo.grpo-nodiv.aux-level.writing.run}{0.95}
\setval{olmo.grpo-nodiv.bench.aime24}{31.8}
\setval{olmo.grpo-nodiv.bench.aime25}{27.6}
\setval{olmo.grpo-nodiv.bench.amc23}{73.0}
\setval{olmo.grpo-nodiv.bench.creative-writing-v3}{47.0}
\setval{olmo.grpo-nodiv.bench.guidebench}{55.5}
\setval{olmo.grpo-nodiv.bench.humaneval-plus}{77.4}
\setval{olmo.grpo-nodiv.bench.ifbench}{31.3}
\setval{olmo.grpo-nodiv.bench.ifeval}{87.8}
\setval{olmo.grpo-nodiv.bench.livecodebench}{23.6}
\setval{olmo.grpo-nodiv.bench.longbench-write}{67.5}
\setval{olmo.grpo-nodiv.bench.math500}{87.5}
\setval{olmo.grpo-nodiv.bench.mbpp-plus}{65.3}
\setval{olmo.grpo-nodiv.bench.minerva}{41.8}
\setval{olmo.grpo-nodiv.bench.olympiad-bench}{58.4}
\setval{olmo.grpo-nodiv.bench.writingbench}{63.2}
\setval{olmo.grpo-nodiv.eval.avg}{55.9}
\setval{olmo.grpo-nodiv.eval.code}{55.4}
\setval{olmo.grpo-nodiv.eval.if}{58.2}
\setval{olmo.grpo-nodiv.eval.math}{53.4}
\setval{olmo.grpo-nodiv.eval.writing}{59.2}
\setval{olmo.grpo-nodiv.judge2.avg}{53.9}
\setval{olmo.grpo-nodiv.judge2.creative-writing-v3}{38.8}
\setval{olmo.grpo-nodiv.judge2.longbench-write}{61.1}
\setval{olmo.grpo-nodiv.judge2.writing}{49.0}
\setval{olmo.grpo-nodiv.judge2.writingbench}{47.0}
\setval{olmo.grpo-nodiv.last-step}{300}
\setval{olmo.grpo-nodiv.minus.base.aime24}{+23.2}
\setval{olmo.grpo-nodiv.minus.base.aime24-abs}{23.2}
\setval{olmo.grpo-nodiv.minus.base.aime25}{+20.2}
\setval{olmo.grpo-nodiv.minus.base.aime25-abs}{20.2}
\setval{olmo.grpo-nodiv.minus.base.amc23}{+28.5}
\setval{olmo.grpo-nodiv.minus.base.amc23-abs}{28.5}
\setval{olmo.grpo-nodiv.minus.base.avg}{+11.8}
\setval{olmo.grpo-nodiv.minus.base.avg-abs}{11.8}
\setval{olmo.grpo-nodiv.minus.base.code}{+6.3}
\setval{olmo.grpo-nodiv.minus.base.code-abs}{6.3}
\setval{olmo.grpo-nodiv.minus.base.creative-writing-v3}{+9.1}
\setval{olmo.grpo-nodiv.minus.base.creative-writing-v3-abs}{9.1}
\setval{olmo.grpo-nodiv.minus.base.guidebench}{$-$6.0}
\setval{olmo.grpo-nodiv.minus.base.guidebench-abs}{6.0}
\setval{olmo.grpo-nodiv.minus.base.humaneval-plus}{+8.2}
\setval{olmo.grpo-nodiv.minus.base.humaneval-plus-abs}{8.2}
\setval{olmo.grpo-nodiv.minus.base.if}{+0.7}
\setval{olmo.grpo-nodiv.minus.base.if-abs}{0.7}
\setval{olmo.grpo-nodiv.minus.base.ifbench}{+4.7}
\setval{olmo.grpo-nodiv.minus.base.ifbench-abs}{4.7}
\setval{olmo.grpo-nodiv.minus.base.ifeval}{+3.4}
\setval{olmo.grpo-nodiv.minus.base.ifeval-abs}{3.4}
\setval{olmo.grpo-nodiv.minus.base.judge2.avg}{+10.9}
\setval{olmo.grpo-nodiv.minus.base.judge2.avg-abs}{10.9}
\setval{olmo.grpo-nodiv.minus.base.judge2.creative-writing-v3}{+7.2}
\setval{olmo.grpo-nodiv.minus.base.judge2.creative-writing-v3-abs}{7.2}
\setval{olmo.grpo-nodiv.minus.base.judge2.longbench-write}{+7.9}
\setval{olmo.grpo-nodiv.minus.base.judge2.longbench-write-abs}{7.9}
\setval{olmo.grpo-nodiv.minus.base.judge2.writing}{+8.1}
\setval{olmo.grpo-nodiv.minus.base.judge2.writing-abs}{8.1}
\setval{olmo.grpo-nodiv.minus.base.judge2.writingbench}{+9.3}
\setval{olmo.grpo-nodiv.minus.base.judge2.writingbench-abs}{9.3}
\setval{olmo.grpo-nodiv.minus.base.livecodebench}{+6.7}
\setval{olmo.grpo-nodiv.minus.base.livecodebench-abs}{6.7}
\setval{olmo.grpo-nodiv.minus.base.longbench-write}{+10.1}
\setval{olmo.grpo-nodiv.minus.base.longbench-write-abs}{10.1}
\setval{olmo.grpo-nodiv.minus.base.math}{+19.7}
\setval{olmo.grpo-nodiv.minus.base.math-abs}{19.7}
\setval{olmo.grpo-nodiv.minus.base.math500}{+17.4}
\setval{olmo.grpo-nodiv.minus.base.math500-abs}{17.4}
\setval{olmo.grpo-nodiv.minus.base.mbpp-plus}{+4.1}
\setval{olmo.grpo-nodiv.minus.base.mbpp-plus-abs}{4.1}
\setval{olmo.grpo-nodiv.minus.base.minerva}{+6.5}
\setval{olmo.grpo-nodiv.minus.base.minerva-abs}{6.5}
\setval{olmo.grpo-nodiv.minus.base.olympiad-bench}{+22.2}
\setval{olmo.grpo-nodiv.minus.base.olympiad-bench-abs}{22.2}
\setval{olmo.grpo-nodiv.minus.base.writing}{+12.5}
\setval{olmo.grpo-nodiv.minus.base.writing-abs}{12.5}
\setval{olmo.grpo-nodiv.minus.base.writingbench}{+18.4}
\setval{olmo.grpo-nodiv.minus.base.writingbench-abs}{18.4}
\setval{olmo.grpo-nodiv.minus.concur-free.aime24}{+5.0}
\setval{olmo.grpo-nodiv.minus.concur-free.aime24-abs}{5.0}
\setval{olmo.grpo-nodiv.minus.concur-free.aime25}{+5.5}
\setval{olmo.grpo-nodiv.minus.concur-free.aime25-abs}{5.5}
\setval{olmo.grpo-nodiv.minus.concur-free.amc23}{+5.5}
\setval{olmo.grpo-nodiv.minus.concur-free.amc23-abs}{5.5}
\setval{olmo.grpo-nodiv.minus.concur-free.avg}{+0.5}
\setval{olmo.grpo-nodiv.minus.concur-free.avg-abs}{0.5}
\setval{olmo.grpo-nodiv.minus.concur-free.code}{$-$0.2}
\setval{olmo.grpo-nodiv.minus.concur-free.code-abs}{0.2}
\setval{olmo.grpo-nodiv.minus.concur-free.creative-writing-v3}{$-$2.8}
\setval{olmo.grpo-nodiv.minus.concur-free.creative-writing-v3-abs}{2.8}
\setval{olmo.grpo-nodiv.minus.concur-free.guidebench}{$-$4.7}
\setval{olmo.grpo-nodiv.minus.concur-free.guidebench-abs}{4.7}
\setval{olmo.grpo-nodiv.minus.concur-free.humaneval-plus}{+0.7}
\setval{olmo.grpo-nodiv.minus.concur-free.humaneval-plus-abs}{0.7}
\setval{olmo.grpo-nodiv.minus.concur-free.if}{$-$1.5}
\setval{olmo.grpo-nodiv.minus.concur-free.if-abs}{1.5}
\setval{olmo.grpo-nodiv.minus.concur-free.ifbench}{$-$0.3}
\setval{olmo.grpo-nodiv.minus.concur-free.ifbench-abs}{0.3}
\setval{olmo.grpo-nodiv.minus.concur-free.ifeval}{+0.5}
\setval{olmo.grpo-nodiv.minus.concur-free.ifeval-abs}{0.5}
\setval{olmo.grpo-nodiv.minus.concur-free.judge2.avg}{+1.2}
\setval{olmo.grpo-nodiv.minus.concur-free.judge2.avg-abs}{1.2}
\setval{olmo.grpo-nodiv.minus.concur-free.judge2.creative-writing-v3}{+1.3}
\setval{olmo.grpo-nodiv.minus.concur-free.judge2.creative-writing-v3-abs}{1.3}
\setval{olmo.grpo-nodiv.minus.concur-free.judge2.longbench-write}{$-$2.8}
\setval{olmo.grpo-nodiv.minus.concur-free.judge2.longbench-write-abs}{2.8}
\setval{olmo.grpo-nodiv.minus.concur-free.judge2.writing}{$-$0.9}
\setval{olmo.grpo-nodiv.minus.concur-free.judge2.writing-abs}{0.9}
\setval{olmo.grpo-nodiv.minus.concur-free.judge2.writingbench}{$-$1.1}
\setval{olmo.grpo-nodiv.minus.concur-free.judge2.writingbench-abs}{1.1}
\setval{olmo.grpo-nodiv.minus.concur-free.livecodebench}{$-$2.1}
\setval{olmo.grpo-nodiv.minus.concur-free.livecodebench-abs}{2.1}
\setval{olmo.grpo-nodiv.minus.concur-free.longbench-write}{$-$4.5}
\setval{olmo.grpo-nodiv.minus.concur-free.longbench-write-abs}{4.5}
\setval{olmo.grpo-nodiv.minus.concur-free.math}{+4.4}
\setval{olmo.grpo-nodiv.minus.concur-free.math-abs}{4.4}
\setval{olmo.grpo-nodiv.minus.concur-free.math500}{+3.1}
\setval{olmo.grpo-nodiv.minus.concur-free.math500-abs}{3.1}
\setval{olmo.grpo-nodiv.minus.concur-free.mbpp-plus}{+0.8}
\setval{olmo.grpo-nodiv.minus.concur-free.mbpp-plus-abs}{0.8}
\setval{olmo.grpo-nodiv.minus.concur-free.minerva}{+1.9}
\setval{olmo.grpo-nodiv.minus.concur-free.minerva-abs}{1.9}
\setval{olmo.grpo-nodiv.minus.concur-free.olympiad-bench}{+5.2}
\setval{olmo.grpo-nodiv.minus.concur-free.olympiad-bench-abs}{5.2}
\setval{olmo.grpo-nodiv.minus.concur-free.writing}{$-$4.8}
\setval{olmo.grpo-nodiv.minus.concur-free.writing-abs}{4.8}
\setval{olmo.grpo-nodiv.minus.concur-free.writingbench}{$-$7.0}
\setval{olmo.grpo-nodiv.minus.concur-free.writingbench-abs}{7.0}
\setval{olmo.grpo-nodiv.minus.concur-tied.aime24}{$-$2.7}
\setval{olmo.grpo-nodiv.minus.concur-tied.aime24-abs}{2.7}
\setval{olmo.grpo-nodiv.minus.concur-tied.aime25}{$-$2.3}
\setval{olmo.grpo-nodiv.minus.concur-tied.aime25-abs}{2.3}
\setval{olmo.grpo-nodiv.minus.concur-tied.amc23}{$-$6.8}
\setval{olmo.grpo-nodiv.minus.concur-tied.amc23-abs}{6.8}
\setval{olmo.grpo-nodiv.minus.concur-tied.avg}{$-$3.2}
\setval{olmo.grpo-nodiv.minus.concur-tied.avg-abs}{3.2}
\setval{olmo.grpo-nodiv.minus.concur-tied.code}{$-$5.7}
\setval{olmo.grpo-nodiv.minus.concur-tied.code-abs}{5.7}
\setval{olmo.grpo-nodiv.minus.concur-tied.creative-writing-v3}{$-$2.1}
\setval{olmo.grpo-nodiv.minus.concur-tied.creative-writing-v3-abs}{2.1}
\setval{olmo.grpo-nodiv.minus.concur-tied.guidebench}{$-$3.1}
\setval{olmo.grpo-nodiv.minus.concur-tied.guidebench-abs}{3.1}
\setval{olmo.grpo-nodiv.minus.concur-tied.humaneval-plus}{$-$5.9}
\setval{olmo.grpo-nodiv.minus.concur-tied.humaneval-plus-abs}{5.9}
\setval{olmo.grpo-nodiv.minus.concur-tied.if}{$-$1.8}
\setval{olmo.grpo-nodiv.minus.concur-tied.if-abs}{1.8}
\setval{olmo.grpo-nodiv.minus.concur-tied.ifbench}{$-$2.3}
\setval{olmo.grpo-nodiv.minus.concur-tied.ifbench-abs}{2.3}
\setval{olmo.grpo-nodiv.minus.concur-tied.ifeval}{+0.1}
\setval{olmo.grpo-nodiv.minus.concur-tied.ifeval-abs}{0.1}
\setval{olmo.grpo-nodiv.minus.concur-tied.judge2.avg}{$-$2.5}
\setval{olmo.grpo-nodiv.minus.concur-tied.judge2.avg-abs}{2.5}
\setval{olmo.grpo-nodiv.minus.concur-tied.judge2.creative-writing-v3}{+1.6}
\setval{olmo.grpo-nodiv.minus.concur-tied.judge2.creative-writing-v3-abs}{1.6}
\setval{olmo.grpo-nodiv.minus.concur-tied.judge2.longbench-write}{+0.6}
\setval{olmo.grpo-nodiv.minus.concur-tied.judge2.longbench-write-abs}{0.6}
\setval{olmo.grpo-nodiv.minus.concur-tied.judge2.writing}{+1.2}
\setval{olmo.grpo-nodiv.minus.concur-tied.judge2.writing-abs}{1.2}
\setval{olmo.grpo-nodiv.minus.concur-tied.judge2.writingbench}{+1.4}
\setval{olmo.grpo-nodiv.minus.concur-tied.judge2.writingbench-abs}{1.4}
\setval{olmo.grpo-nodiv.minus.concur-tied.livecodebench}{$-$5.4}
\setval{olmo.grpo-nodiv.minus.concur-tied.livecodebench-abs}{5.4}
\setval{olmo.grpo-nodiv.minus.concur-tied.longbench-write}{$-$1.7}
\setval{olmo.grpo-nodiv.minus.concur-tied.longbench-write-abs}{1.7}
\setval{olmo.grpo-nodiv.minus.concur-tied.math}{$-$3.2}
\setval{olmo.grpo-nodiv.minus.concur-tied.math-abs}{3.2}
\setval{olmo.grpo-nodiv.minus.concur-tied.math500}{$-$1.9}
\setval{olmo.grpo-nodiv.minus.concur-tied.math500-abs}{1.9}
\setval{olmo.grpo-nodiv.minus.concur-tied.mbpp-plus}{$-$5.7}
\setval{olmo.grpo-nodiv.minus.concur-tied.mbpp-plus-abs}{5.7}
\setval{olmo.grpo-nodiv.minus.concur-tied.minerva}{$-$2.2}
\setval{olmo.grpo-nodiv.minus.concur-tied.minerva-abs}{2.2}
\setval{olmo.grpo-nodiv.minus.concur-tied.olympiad-bench}{$-$3.1}
\setval{olmo.grpo-nodiv.minus.concur-tied.olympiad-bench-abs}{3.1}
\setval{olmo.grpo-nodiv.minus.concur-tied.writing}{$-$2.2}
\setval{olmo.grpo-nodiv.minus.concur-tied.writing-abs}{2.2}
\setval{olmo.grpo-nodiv.minus.concur-tied.writingbench}{$-$3.0}
\setval{olmo.grpo-nodiv.minus.concur-tied.writingbench-abs}{3.0}
\setval{olmo.grpo-nodiv.minus.concur.aime24}{$-$4.6}
\setval{olmo.grpo-nodiv.minus.concur.aime24-abs}{4.6}
\setval{olmo.grpo-nodiv.minus.concur.aime25}{$-$4.6}
\setval{olmo.grpo-nodiv.minus.concur.aime25-abs}{4.6}
\setval{olmo.grpo-nodiv.minus.concur.amc23}{$-$10.5}
\setval{olmo.grpo-nodiv.minus.concur.amc23-abs}{10.5}
\setval{olmo.grpo-nodiv.minus.concur.avg}{$-$3.4}
\setval{olmo.grpo-nodiv.minus.concur.avg-abs}{3.4}
\setval{olmo.grpo-nodiv.minus.concur.code}{$-$4.8}
\setval{olmo.grpo-nodiv.minus.concur.code-abs}{4.8}
\setval{olmo.grpo-nodiv.minus.concur.creative-writing-v3}{$-$2.2}
\setval{olmo.grpo-nodiv.minus.concur.creative-writing-v3-abs}{2.2}
\setval{olmo.grpo-nodiv.minus.concur.guidebench}{$-$1.4}
\setval{olmo.grpo-nodiv.minus.concur.guidebench-abs}{1.4}
\setval{olmo.grpo-nodiv.minus.concur.humaneval-plus}{$-$5.4}
\setval{olmo.grpo-nodiv.minus.concur.humaneval-plus-abs}{5.4}
\setval{olmo.grpo-nodiv.minus.concur.if}{$-$0.7}
\setval{olmo.grpo-nodiv.minus.concur.if-abs}{0.7}
\setval{olmo.grpo-nodiv.minus.concur.ifbench}{$-$0.7}
\setval{olmo.grpo-nodiv.minus.concur.ifbench-abs}{0.7}
\setval{olmo.grpo-nodiv.minus.concur.ifeval}{+0.0}
\setval{olmo.grpo-nodiv.minus.concur.ifeval-abs}{0.0}
\setval{olmo.grpo-nodiv.minus.concur.judge2.avg}{$-$2.7}
\setval{olmo.grpo-nodiv.minus.concur.judge2.avg-abs}{2.7}
\setval{olmo.grpo-nodiv.minus.concur.judge2.creative-writing-v3}{+0.4}
\setval{olmo.grpo-nodiv.minus.concur.judge2.creative-writing-v3-abs}{0.4}
\setval{olmo.grpo-nodiv.minus.concur.judge2.longbench-write}{+2.8}
\setval{olmo.grpo-nodiv.minus.concur.judge2.longbench-write-abs}{2.8}
\setval{olmo.grpo-nodiv.minus.concur.judge2.writing}{+1.7}
\setval{olmo.grpo-nodiv.minus.concur.judge2.writing-abs}{1.7}
\setval{olmo.grpo-nodiv.minus.concur.judge2.writingbench}{+2.1}
\setval{olmo.grpo-nodiv.minus.concur.judge2.writingbench-abs}{2.1}
\setval{olmo.grpo-nodiv.minus.concur.livecodebench}{$-$6.7}
\setval{olmo.grpo-nodiv.minus.concur.livecodebench-abs}{6.7}
\setval{olmo.grpo-nodiv.minus.concur.longbench-write}{$-$1.0}
\setval{olmo.grpo-nodiv.minus.concur.longbench-write-abs}{1.0}
\setval{olmo.grpo-nodiv.minus.concur.math}{$-$4.9}
\setval{olmo.grpo-nodiv.minus.concur.math-abs}{4.9}
\setval{olmo.grpo-nodiv.minus.concur.math500}{$-$3.0}
\setval{olmo.grpo-nodiv.minus.concur.math500-abs}{3.0}
\setval{olmo.grpo-nodiv.minus.concur.mbpp-plus}{$-$2.2}
\setval{olmo.grpo-nodiv.minus.concur.mbpp-plus-abs}{2.2}
\setval{olmo.grpo-nodiv.minus.concur.minerva}{$-$3.0}
\setval{olmo.grpo-nodiv.minus.concur.minerva-abs}{3.0}
\setval{olmo.grpo-nodiv.minus.concur.olympiad-bench}{$-$4.0}
\setval{olmo.grpo-nodiv.minus.concur.olympiad-bench-abs}{4.0}
\setval{olmo.grpo-nodiv.minus.concur.writing}{$-$1.8}
\setval{olmo.grpo-nodiv.minus.concur.writing-abs}{1.8}
\setval{olmo.grpo-nodiv.minus.concur.writingbench}{$-$2.3}
\setval{olmo.grpo-nodiv.minus.concur.writingbench-abs}{2.3}
\setval{olmo.grpo-nodiv.minus.gd2po.aime24}{+3.4}
\setval{olmo.grpo-nodiv.minus.gd2po.aime24-abs}{3.4}
\setval{olmo.grpo-nodiv.minus.gd2po.aime25}{+3.3}
\setval{olmo.grpo-nodiv.minus.gd2po.aime25-abs}{3.3}
\setval{olmo.grpo-nodiv.minus.gd2po.amc23}{+0.9}
\setval{olmo.grpo-nodiv.minus.gd2po.amc23-abs}{0.9}
\setval{olmo.grpo-nodiv.minus.gd2po.avg}{+0.3}
\setval{olmo.grpo-nodiv.minus.gd2po.avg-abs}{0.3}
\setval{olmo.grpo-nodiv.minus.gd2po.code}{+0.5}
\setval{olmo.grpo-nodiv.minus.gd2po.code-abs}{0.5}
\setval{olmo.grpo-nodiv.minus.gd2po.creative-writing-v3}{$-$1.3}
\setval{olmo.grpo-nodiv.minus.gd2po.creative-writing-v3-abs}{1.3}
\setval{olmo.grpo-nodiv.minus.gd2po.guidebench}{$-$5.7}
\setval{olmo.grpo-nodiv.minus.gd2po.guidebench-abs}{5.7}
\setval{olmo.grpo-nodiv.minus.gd2po.humaneval-plus}{+0.6}
\setval{olmo.grpo-nodiv.minus.gd2po.humaneval-plus-abs}{0.6}
\setval{olmo.grpo-nodiv.minus.gd2po.if}{$-$2.2}
\setval{olmo.grpo-nodiv.minus.gd2po.if-abs}{2.2}
\setval{olmo.grpo-nodiv.minus.gd2po.ifbench}{$-$1.0}
\setval{olmo.grpo-nodiv.minus.gd2po.ifbench-abs}{1.0}
\setval{olmo.grpo-nodiv.minus.gd2po.ifeval}{+0.1}
\setval{olmo.grpo-nodiv.minus.gd2po.ifeval-abs}{0.1}
\setval{olmo.grpo-nodiv.minus.gd2po.judge2.avg}{+0.6}
\setval{olmo.grpo-nodiv.minus.gd2po.judge2.avg-abs}{0.6}
\setval{olmo.grpo-nodiv.minus.gd2po.judge2.creative-writing-v3}{+0.9}
\setval{olmo.grpo-nodiv.minus.gd2po.judge2.creative-writing-v3-abs}{0.9}
\setval{olmo.grpo-nodiv.minus.gd2po.judge2.longbench-write}{+1.2}
\setval{olmo.grpo-nodiv.minus.gd2po.judge2.longbench-write-abs}{1.2}
\setval{olmo.grpo-nodiv.minus.gd2po.judge2.writing}{+0.3}
\setval{olmo.grpo-nodiv.minus.gd2po.judge2.writing-abs}{0.3}
\setval{olmo.grpo-nodiv.minus.gd2po.judge2.writingbench}{$-$1.4}
\setval{olmo.grpo-nodiv.minus.gd2po.judge2.writingbench-abs}{1.4}
\setval{olmo.grpo-nodiv.minus.gd2po.livecodebench}{+0.4}
\setval{olmo.grpo-nodiv.minus.gd2po.livecodebench-abs}{0.4}
\setval{olmo.grpo-nodiv.minus.gd2po.longbench-write}{$-$0.7}
\setval{olmo.grpo-nodiv.minus.gd2po.longbench-write-abs}{0.7}
\setval{olmo.grpo-nodiv.minus.gd2po.math}{+2.2}
\setval{olmo.grpo-nodiv.minus.gd2po.math-abs}{2.2}
\setval{olmo.grpo-nodiv.minus.gd2po.math500}{+1.8}
\setval{olmo.grpo-nodiv.minus.gd2po.math500-abs}{1.8}
\setval{olmo.grpo-nodiv.minus.gd2po.mbpp-plus}{+0.6}
\setval{olmo.grpo-nodiv.minus.gd2po.mbpp-plus-abs}{0.6}
\setval{olmo.grpo-nodiv.minus.gd2po.minerva}{+1.7}
\setval{olmo.grpo-nodiv.minus.gd2po.minerva-abs}{1.7}
\setval{olmo.grpo-nodiv.minus.gd2po.olympiad-bench}{+2.4}
\setval{olmo.grpo-nodiv.minus.gd2po.olympiad-bench-abs}{2.4}
\setval{olmo.grpo-nodiv.minus.gd2po.writing}{$-$1.5}
\setval{olmo.grpo-nodiv.minus.gd2po.writing-abs}{1.5}
\setval{olmo.grpo-nodiv.minus.gd2po.writingbench}{$-$2.6}
\setval{olmo.grpo-nodiv.minus.gd2po.writingbench-abs}{2.6}
\setval{olmo.grpo-nodiv.minus.gdpo-noaux.aime24}{$-$6.6}
\setval{olmo.grpo-nodiv.minus.gdpo-noaux.aime24-abs}{6.6}
\setval{olmo.grpo-nodiv.minus.gdpo-noaux.aime25}{$-$6.0}
\setval{olmo.grpo-nodiv.minus.gdpo-noaux.aime25-abs}{6.0}
\setval{olmo.grpo-nodiv.minus.gdpo-noaux.amc23}{$-$10.9}
\setval{olmo.grpo-nodiv.minus.gdpo-noaux.amc23-abs}{10.9}
\setval{olmo.grpo-nodiv.minus.gdpo-noaux.avg}{$-$1.9}
\setval{olmo.grpo-nodiv.minus.gdpo-noaux.avg-abs}{1.9}
\setval{olmo.grpo-nodiv.minus.gdpo-noaux.code}{$-$7.0}
\setval{olmo.grpo-nodiv.minus.gdpo-noaux.code-abs}{7.0}
\setval{olmo.grpo-nodiv.minus.gdpo-noaux.creative-writing-v3}{+8.8}
\setval{olmo.grpo-nodiv.minus.gdpo-noaux.creative-writing-v3-abs}{8.8}
\setval{olmo.grpo-nodiv.minus.gdpo-noaux.guidebench}{$-$0.3}
\setval{olmo.grpo-nodiv.minus.gdpo-noaux.guidebench-abs}{0.3}
\setval{olmo.grpo-nodiv.minus.gdpo-noaux.humaneval-plus}{$-$7.8}
\setval{olmo.grpo-nodiv.minus.gdpo-noaux.humaneval-plus-abs}{7.8}
\setval{olmo.grpo-nodiv.minus.gdpo-noaux.if}{+0.2}
\setval{olmo.grpo-nodiv.minus.gdpo-noaux.if-abs}{0.2}
\setval{olmo.grpo-nodiv.minus.gdpo-noaux.ifbench}{$-$2.0}
\setval{olmo.grpo-nodiv.minus.gdpo-noaux.ifbench-abs}{2.0}
\setval{olmo.grpo-nodiv.minus.gdpo-noaux.ifeval}{+3.0}
\setval{olmo.grpo-nodiv.minus.gdpo-noaux.ifeval-abs}{3.0}
\setval{olmo.grpo-nodiv.minus.gdpo-noaux.judge2.avg}{$-$1.8}
\setval{olmo.grpo-nodiv.minus.gdpo-noaux.judge2.avg-abs}{1.8}
\setval{olmo.grpo-nodiv.minus.gdpo-noaux.judge2.creative-writing-v3}{+7.9}
\setval{olmo.grpo-nodiv.minus.gdpo-noaux.judge2.creative-writing-v3-abs}{7.9}
\setval{olmo.grpo-nodiv.minus.gdpo-noaux.judge2.longbench-write}{+14.5}
\setval{olmo.grpo-nodiv.minus.gdpo-noaux.judge2.longbench-write-abs}{14.5}
\setval{olmo.grpo-nodiv.minus.gdpo-noaux.judge2.writing}{+9.9}
\setval{olmo.grpo-nodiv.minus.gdpo-noaux.judge2.writing-abs}{9.9}
\setval{olmo.grpo-nodiv.minus.gdpo-noaux.judge2.writingbench}{+7.2}
\setval{olmo.grpo-nodiv.minus.gdpo-noaux.judge2.writingbench-abs}{7.2}
\setval{olmo.grpo-nodiv.minus.gdpo-noaux.livecodebench}{$-$7.9}
\setval{olmo.grpo-nodiv.minus.gdpo-noaux.livecodebench-abs}{7.9}
\setval{olmo.grpo-nodiv.minus.gdpo-noaux.longbench-write}{+11.6}
\setval{olmo.grpo-nodiv.minus.gdpo-noaux.longbench-write-abs}{11.6}
\setval{olmo.grpo-nodiv.minus.gdpo-noaux.math}{$-$6.1}
\setval{olmo.grpo-nodiv.minus.gdpo-noaux.math-abs}{6.1}
\setval{olmo.grpo-nodiv.minus.gdpo-noaux.math500}{$-$3.4}
\setval{olmo.grpo-nodiv.minus.gdpo-noaux.math500-abs}{3.4}
\setval{olmo.grpo-nodiv.minus.gdpo-noaux.mbpp-plus}{$-$5.2}
\setval{olmo.grpo-nodiv.minus.gdpo-noaux.mbpp-plus-abs}{5.2}
\setval{olmo.grpo-nodiv.minus.gdpo-noaux.minerva}{$-$4.3}
\setval{olmo.grpo-nodiv.minus.gdpo-noaux.minerva-abs}{4.3}
\setval{olmo.grpo-nodiv.minus.gdpo-noaux.olympiad-bench}{$-$5.3}
\setval{olmo.grpo-nodiv.minus.gdpo-noaux.olympiad-bench-abs}{5.3}
\setval{olmo.grpo-nodiv.minus.gdpo-noaux.writing}{+9.3}
\setval{olmo.grpo-nodiv.minus.gdpo-noaux.writing-abs}{9.3}
\setval{olmo.grpo-nodiv.minus.gdpo-noaux.writingbench}{+7.7}
\setval{olmo.grpo-nodiv.minus.gdpo-noaux.writingbench-abs}{7.7}
\setval{olmo.grpo-nodiv.minus.gdpo-w0.5.aime24}{+2.2}
\setval{olmo.grpo-nodiv.minus.gdpo-w0.5.aime24-abs}{2.2}
\setval{olmo.grpo-nodiv.minus.gdpo-w0.5.aime25}{+0.3}
\setval{olmo.grpo-nodiv.minus.gdpo-w0.5.aime25-abs}{0.3}
\setval{olmo.grpo-nodiv.minus.gdpo-w0.5.amc23}{$-$0.9}
\setval{olmo.grpo-nodiv.minus.gdpo-w0.5.amc23-abs}{0.9}
\setval{olmo.grpo-nodiv.minus.gdpo-w0.5.avg}{+0.3}
\setval{olmo.grpo-nodiv.minus.gdpo-w0.5.avg-abs}{0.3}
\setval{olmo.grpo-nodiv.minus.gdpo-w0.5.code}{$-$0.4}
\setval{olmo.grpo-nodiv.minus.gdpo-w0.5.code-abs}{0.4}
\setval{olmo.grpo-nodiv.minus.gdpo-w0.5.creative-writing-v3}{$-$1.1}
\setval{olmo.grpo-nodiv.minus.gdpo-w0.5.creative-writing-v3-abs}{1.1}
\setval{olmo.grpo-nodiv.minus.gdpo-w0.5.guidebench}{+3.6}
\setval{olmo.grpo-nodiv.minus.gdpo-w0.5.guidebench-abs}{3.6}
\setval{olmo.grpo-nodiv.minus.gdpo-w0.5.humaneval-plus}{$-$0.4}
\setval{olmo.grpo-nodiv.minus.gdpo-w0.5.humaneval-plus-abs}{0.4}
\setval{olmo.grpo-nodiv.minus.gdpo-w0.5.if}{+0.8}
\setval{olmo.grpo-nodiv.minus.gdpo-w0.5.if-abs}{0.8}
\setval{olmo.grpo-nodiv.minus.gdpo-w0.5.ifbench}{$-$1.3}
\setval{olmo.grpo-nodiv.minus.gdpo-w0.5.ifbench-abs}{1.3}
\setval{olmo.grpo-nodiv.minus.gdpo-w0.5.ifeval}{+0.2}
\setval{olmo.grpo-nodiv.minus.gdpo-w0.5.ifeval-abs}{0.2}
\setval{olmo.grpo-nodiv.minus.gdpo-w0.5.judge2.avg}{+0.6}
\setval{olmo.grpo-nodiv.minus.gdpo-w0.5.judge2.avg-abs}{0.6}
\setval{olmo.grpo-nodiv.minus.gdpo-w0.5.judge2.creative-writing-v3}{$-$0.6}
\setval{olmo.grpo-nodiv.minus.gdpo-w0.5.judge2.creative-writing-v3-abs}{0.6}
\setval{olmo.grpo-nodiv.minus.gdpo-w0.5.judge2.longbench-write}{+4.5}
\setval{olmo.grpo-nodiv.minus.gdpo-w0.5.judge2.longbench-write-abs}{4.5}
\setval{olmo.grpo-nodiv.minus.gdpo-w0.5.judge2.writing}{+1.4}
\setval{olmo.grpo-nodiv.minus.gdpo-w0.5.judge2.writing-abs}{1.4}
\setval{olmo.grpo-nodiv.minus.gdpo-w0.5.judge2.writingbench}{+0.3}
\setval{olmo.grpo-nodiv.minus.gdpo-w0.5.judge2.writingbench-abs}{0.3}
\setval{olmo.grpo-nodiv.minus.gdpo-w0.5.livecodebench}{$-$1.2}
\setval{olmo.grpo-nodiv.minus.gdpo-w0.5.livecodebench-abs}{1.2}
\setval{olmo.grpo-nodiv.minus.gdpo-w0.5.longbench-write}{+3.8}
\setval{olmo.grpo-nodiv.minus.gdpo-w0.5.longbench-write-abs}{3.8}
\setval{olmo.grpo-nodiv.minus.gdpo-w0.5.math}{+0.5}
\setval{olmo.grpo-nodiv.minus.gdpo-w0.5.math-abs}{0.5}
\setval{olmo.grpo-nodiv.minus.gdpo-w0.5.math500}{+0.5}
\setval{olmo.grpo-nodiv.minus.gdpo-w0.5.math500-abs}{0.5}
\setval{olmo.grpo-nodiv.minus.gdpo-w0.5.mbpp-plus}{+0.4}
\setval{olmo.grpo-nodiv.minus.gdpo-w0.5.mbpp-plus-abs}{0.4}
\setval{olmo.grpo-nodiv.minus.gdpo-w0.5.minerva}{+0.3}
\setval{olmo.grpo-nodiv.minus.gdpo-w0.5.minerva-abs}{0.3}
\setval{olmo.grpo-nodiv.minus.gdpo-w0.5.olympiad-bench}{+0.9}
\setval{olmo.grpo-nodiv.minus.gdpo-w0.5.olympiad-bench-abs}{0.9}
\setval{olmo.grpo-nodiv.minus.gdpo-w0.5.writing}{$-$0.2}
\setval{olmo.grpo-nodiv.minus.gdpo-w0.5.writing-abs}{0.2}
\setval{olmo.grpo-nodiv.minus.gdpo-w0.5.writingbench}{$-$3.3}
\setval{olmo.grpo-nodiv.minus.gdpo-w0.5.writingbench-abs}{3.3}
\setval{olmo.grpo-nodiv.minus.gdpo.aime24}{+4.0}
\setval{olmo.grpo-nodiv.minus.gdpo.aime24-abs}{4.0}
\setval{olmo.grpo-nodiv.minus.gdpo.aime25}{+5.0}
\setval{olmo.grpo-nodiv.minus.gdpo.aime25-abs}{5.0}
\setval{olmo.grpo-nodiv.minus.gdpo.amc23}{+3.5}
\setval{olmo.grpo-nodiv.minus.gdpo.amc23-abs}{3.5}
\setval{olmo.grpo-nodiv.minus.gdpo.avg}{+2.0}
\setval{olmo.grpo-nodiv.minus.gdpo.avg-abs}{2.0}
\setval{olmo.grpo-nodiv.minus.gdpo.code}{+0.7}
\setval{olmo.grpo-nodiv.minus.gdpo.code-abs}{0.7}
\setval{olmo.grpo-nodiv.minus.gdpo.creative-writing-v3}{$-$0.2}
\setval{olmo.grpo-nodiv.minus.gdpo.creative-writing-v3-abs}{0.2}
\setval{olmo.grpo-nodiv.minus.gdpo.guidebench}{+4.6}
\setval{olmo.grpo-nodiv.minus.gdpo.guidebench-abs}{4.6}
\setval{olmo.grpo-nodiv.minus.gdpo.humaneval-plus}{+1.0}
\setval{olmo.grpo-nodiv.minus.gdpo.humaneval-plus-abs}{1.0}
\setval{olmo.grpo-nodiv.minus.gdpo.if}{+2.5}
\setval{olmo.grpo-nodiv.minus.gdpo.if-abs}{2.5}
\setval{olmo.grpo-nodiv.minus.gdpo.ifbench}{+1.7}
\setval{olmo.grpo-nodiv.minus.gdpo.ifbench-abs}{1.7}
\setval{olmo.grpo-nodiv.minus.gdpo.ifeval}{+1.3}
\setval{olmo.grpo-nodiv.minus.gdpo.ifeval-abs}{1.3}
\setval{olmo.grpo-nodiv.minus.gdpo.judge2.avg}{+2.3}
\setval{olmo.grpo-nodiv.minus.gdpo.judge2.avg-abs}{2.3}
\setval{olmo.grpo-nodiv.minus.gdpo.judge2.creative-writing-v3}{+1.0}
\setval{olmo.grpo-nodiv.minus.gdpo.judge2.creative-writing-v3-abs}{1.0}
\setval{olmo.grpo-nodiv.minus.gdpo.judge2.longbench-write}{+5.5}
\setval{olmo.grpo-nodiv.minus.gdpo.judge2.longbench-write-abs}{5.5}
\setval{olmo.grpo-nodiv.minus.gdpo.judge2.writing}{+2.1}
\setval{olmo.grpo-nodiv.minus.gdpo.judge2.writing-abs}{2.1}
\setval{olmo.grpo-nodiv.minus.gdpo.judge2.writingbench}{$-$0.1}
\setval{olmo.grpo-nodiv.minus.gdpo.judge2.writingbench-abs}{0.1}
\setval{olmo.grpo-nodiv.minus.gdpo.livecodebench}{+0.5}
\setval{olmo.grpo-nodiv.minus.gdpo.livecodebench-abs}{0.5}
\setval{olmo.grpo-nodiv.minus.gdpo.longbench-write}{+4.3}
\setval{olmo.grpo-nodiv.minus.gdpo.longbench-write-abs}{4.3}
\setval{olmo.grpo-nodiv.minus.gdpo.math}{+3.1}
\setval{olmo.grpo-nodiv.minus.gdpo.math-abs}{3.1}
\setval{olmo.grpo-nodiv.minus.gdpo.math500}{+2.3}
\setval{olmo.grpo-nodiv.minus.gdpo.math500-abs}{2.3}
\setval{olmo.grpo-nodiv.minus.gdpo.mbpp-plus}{+0.6}
\setval{olmo.grpo-nodiv.minus.gdpo.mbpp-plus-abs}{0.6}
\setval{olmo.grpo-nodiv.minus.gdpo.minerva}{+1.4}
\setval{olmo.grpo-nodiv.minus.gdpo.minerva-abs}{1.4}
\setval{olmo.grpo-nodiv.minus.gdpo.olympiad-bench}{+2.6}
\setval{olmo.grpo-nodiv.minus.gdpo.olympiad-bench-abs}{2.6}
\setval{olmo.grpo-nodiv.minus.gdpo.writing}{+0.7}
\setval{olmo.grpo-nodiv.minus.gdpo.writing-abs}{0.7}
\setval{olmo.grpo-nodiv.minus.gdpo.writingbench}{$-$2.1}
\setval{olmo.grpo-nodiv.minus.gdpo.writingbench-abs}{2.1}
\setval{olmo.grpo-nodiv.minus.geoma.aime24}{+8.0}
\setval{olmo.grpo-nodiv.minus.geoma.aime24-abs}{8.0}
\setval{olmo.grpo-nodiv.minus.geoma.aime25}{+5.6}
\setval{olmo.grpo-nodiv.minus.geoma.aime25-abs}{5.6}
\setval{olmo.grpo-nodiv.minus.geoma.amc23}{+9.1}
\setval{olmo.grpo-nodiv.minus.geoma.amc23-abs}{9.1}
\setval{olmo.grpo-nodiv.minus.geoma.avg}{+3.0}
\setval{olmo.grpo-nodiv.minus.geoma.avg-abs}{3.0}
\setval{olmo.grpo-nodiv.minus.geoma.code}{+1.5}
\setval{olmo.grpo-nodiv.minus.geoma.code-abs}{1.5}
\setval{olmo.grpo-nodiv.minus.geoma.creative-writing-v3}{$-$2.6}
\setval{olmo.grpo-nodiv.minus.geoma.creative-writing-v3-abs}{2.6}
\setval{olmo.grpo-nodiv.minus.geoma.guidebench}{+8.8}
\setval{olmo.grpo-nodiv.minus.geoma.guidebench-abs}{8.8}
\setval{olmo.grpo-nodiv.minus.geoma.humaneval-plus}{+2.7}
\setval{olmo.grpo-nodiv.minus.geoma.humaneval-plus-abs}{2.7}
\setval{olmo.grpo-nodiv.minus.geoma.if}{+3.2}
\setval{olmo.grpo-nodiv.minus.geoma.if-abs}{3.2}
\setval{olmo.grpo-nodiv.minus.geoma.ifbench}{+1.0}
\setval{olmo.grpo-nodiv.minus.geoma.ifbench-abs}{1.0}
\setval{olmo.grpo-nodiv.minus.geoma.ifeval}{$-$0.2}
\setval{olmo.grpo-nodiv.minus.geoma.ifeval-abs}{0.2}
\setval{olmo.grpo-nodiv.minus.geoma.judge2.avg}{+3.1}
\setval{olmo.grpo-nodiv.minus.geoma.judge2.avg-abs}{3.1}
\setval{olmo.grpo-nodiv.minus.geoma.judge2.creative-writing-v3}{$-$2.9}
\setval{olmo.grpo-nodiv.minus.geoma.judge2.creative-writing-v3-abs}{2.9}
\setval{olmo.grpo-nodiv.minus.geoma.judge2.longbench-write}{+2.9}
\setval{olmo.grpo-nodiv.minus.geoma.judge2.longbench-write-abs}{2.9}
\setval{olmo.grpo-nodiv.minus.geoma.judge2.writing}{$-$0.6}
\setval{olmo.grpo-nodiv.minus.geoma.judge2.writing-abs}{0.6}
\setval{olmo.grpo-nodiv.minus.geoma.judge2.writingbench}{$-$1.7}
\setval{olmo.grpo-nodiv.minus.geoma.judge2.writingbench-abs}{1.7}
\setval{olmo.grpo-nodiv.minus.geoma.livecodebench}{+1.3}
\setval{olmo.grpo-nodiv.minus.geoma.livecodebench-abs}{1.3}
\setval{olmo.grpo-nodiv.minus.geoma.longbench-write}{+2.2}
\setval{olmo.grpo-nodiv.minus.geoma.longbench-write-abs}{2.2}
\setval{olmo.grpo-nodiv.minus.geoma.math}{+5.8}
\setval{olmo.grpo-nodiv.minus.geoma.math-abs}{5.8}
\setval{olmo.grpo-nodiv.minus.geoma.math500}{+4.3}
\setval{olmo.grpo-nodiv.minus.geoma.math500-abs}{4.3}
\setval{olmo.grpo-nodiv.minus.geoma.mbpp-plus}{+0.6}
\setval{olmo.grpo-nodiv.minus.geoma.mbpp-plus-abs}{0.6}
\setval{olmo.grpo-nodiv.minus.geoma.minerva}{+1.8}
\setval{olmo.grpo-nodiv.minus.geoma.minerva-abs}{1.8}
\setval{olmo.grpo-nodiv.minus.geoma.olympiad-bench}{+5.7}
\setval{olmo.grpo-nodiv.minus.geoma.olympiad-bench-abs}{5.7}
\setval{olmo.grpo-nodiv.minus.geoma.writing}{$-$1.4}
\setval{olmo.grpo-nodiv.minus.geoma.writing-abs}{1.4}
\setval{olmo.grpo-nodiv.minus.geoma.writingbench}{$-$3.8}
\setval{olmo.grpo-nodiv.minus.geoma.writingbench-abs}{3.8}
\setval{olmo.grpo-nodiv.minus.grpo.aime24}{+2.3}
\setval{olmo.grpo-nodiv.minus.grpo.aime24-abs}{2.3}
\setval{olmo.grpo-nodiv.minus.grpo.aime25}{+2.9}
\setval{olmo.grpo-nodiv.minus.grpo.aime25-abs}{2.9}
\setval{olmo.grpo-nodiv.minus.grpo.amc23}{+2.2}
\setval{olmo.grpo-nodiv.minus.grpo.amc23-abs}{2.2}
\setval{olmo.grpo-nodiv.minus.grpo.avg}{+0.7}
\setval{olmo.grpo-nodiv.minus.grpo.avg-abs}{0.7}
\setval{olmo.grpo-nodiv.minus.grpo.code}{+0.3}
\setval{olmo.grpo-nodiv.minus.grpo.code-abs}{0.3}
\setval{olmo.grpo-nodiv.minus.grpo.creative-writing-v3}{$-$2.4}
\setval{olmo.grpo-nodiv.minus.grpo.creative-writing-v3-abs}{2.4}
\setval{olmo.grpo-nodiv.minus.grpo.guidebench}{$-$1.0}
\setval{olmo.grpo-nodiv.minus.grpo.guidebench-abs}{1.0}
\setval{olmo.grpo-nodiv.minus.grpo.humaneval-plus}{+0.1}
\setval{olmo.grpo-nodiv.minus.grpo.humaneval-plus-abs}{0.1}
\setval{olmo.grpo-nodiv.minus.grpo.if}{$-$0.4}
\setval{olmo.grpo-nodiv.minus.grpo.if-abs}{0.4}
\setval{olmo.grpo-nodiv.minus.grpo.ifbench}{$-$0.7}
\setval{olmo.grpo-nodiv.minus.grpo.ifbench-abs}{0.7}
\setval{olmo.grpo-nodiv.minus.grpo.ifeval}{+0.4}
\setval{olmo.grpo-nodiv.minus.grpo.ifeval-abs}{0.4}
\setval{olmo.grpo-nodiv.minus.grpo.judge2.avg}{+1.1}
\setval{olmo.grpo-nodiv.minus.grpo.judge2.avg-abs}{1.1}
\setval{olmo.grpo-nodiv.minus.grpo.judge2.creative-writing-v3}{+0.2}
\setval{olmo.grpo-nodiv.minus.grpo.judge2.creative-writing-v3-abs}{0.2}
\setval{olmo.grpo-nodiv.minus.grpo.judge2.longbench-write}{+4.0}
\setval{olmo.grpo-nodiv.minus.grpo.judge2.longbench-write-abs}{4.0}
\setval{olmo.grpo-nodiv.minus.grpo.judge2.writing}{+0.9}
\setval{olmo.grpo-nodiv.minus.grpo.judge2.writing-abs}{0.9}
\setval{olmo.grpo-nodiv.minus.grpo.judge2.writingbench}{$-$1.5}
\setval{olmo.grpo-nodiv.minus.grpo.judge2.writingbench-abs}{1.5}
\setval{olmo.grpo-nodiv.minus.grpo.livecodebench}{+1.3}
\setval{olmo.grpo-nodiv.minus.grpo.livecodebench-abs}{1.3}
\setval{olmo.grpo-nodiv.minus.grpo.longbench-write}{+3.3}
\setval{olmo.grpo-nodiv.minus.grpo.longbench-write-abs}{3.3}
\setval{olmo.grpo-nodiv.minus.grpo.math}{+2.4}
\setval{olmo.grpo-nodiv.minus.grpo.math-abs}{2.4}
\setval{olmo.grpo-nodiv.minus.grpo.math500}{+2.6}
\setval{olmo.grpo-nodiv.minus.grpo.math500-abs}{2.6}
\setval{olmo.grpo-nodiv.minus.grpo.mbpp-plus}{$-$0.4}
\setval{olmo.grpo-nodiv.minus.grpo.mbpp-plus-abs}{0.4}
\setval{olmo.grpo-nodiv.minus.grpo.minerva}{+1.9}
\setval{olmo.grpo-nodiv.minus.grpo.minerva-abs}{1.9}
\setval{olmo.grpo-nodiv.minus.grpo.olympiad-bench}{+2.8}
\setval{olmo.grpo-nodiv.minus.grpo.olympiad-bench-abs}{2.8}
\setval{olmo.grpo-nodiv.minus.grpo.writing}{$-$1.5}
\setval{olmo.grpo-nodiv.minus.grpo.writing-abs}{1.5}
\setval{olmo.grpo-nodiv.minus.grpo.writingbench}{$-$5.3}
\setval{olmo.grpo-nodiv.minus.grpo.writingbench-abs}{5.3}
\setval{olmo.grpo-nodiv.minus.mt-grpo.aime24}{$-$4.8}
\setval{olmo.grpo-nodiv.minus.mt-grpo.aime24-abs}{4.8}
\setval{olmo.grpo-nodiv.minus.mt-grpo.aime25}{$-$0.5}
\setval{olmo.grpo-nodiv.minus.mt-grpo.aime25-abs}{0.5}
\setval{olmo.grpo-nodiv.minus.mt-grpo.amc23}{$-$4.4}
\setval{olmo.grpo-nodiv.minus.mt-grpo.amc23-abs}{4.4}
\setval{olmo.grpo-nodiv.minus.mt-grpo.avg}{$-$0.8}
\setval{olmo.grpo-nodiv.minus.mt-grpo.avg-abs}{0.8}
\setval{olmo.grpo-nodiv.minus.mt-grpo.code}{+1.0}
\setval{olmo.grpo-nodiv.minus.mt-grpo.code-abs}{1.0}
\setval{olmo.grpo-nodiv.minus.mt-grpo.creative-writing-v3}{$-$3.7}
\setval{olmo.grpo-nodiv.minus.mt-grpo.creative-writing-v3-abs}{3.7}
\setval{olmo.grpo-nodiv.minus.mt-grpo.guidebench}{+1.0}
\setval{olmo.grpo-nodiv.minus.mt-grpo.guidebench-abs}{1.0}
\setval{olmo.grpo-nodiv.minus.mt-grpo.humaneval-plus}{+1.6}
\setval{olmo.grpo-nodiv.minus.mt-grpo.humaneval-plus-abs}{1.6}
\setval{olmo.grpo-nodiv.minus.mt-grpo.if}{+0.3}
\setval{olmo.grpo-nodiv.minus.mt-grpo.if-abs}{0.3}
\setval{olmo.grpo-nodiv.minus.mt-grpo.ifbench}{$-$0.7}
\setval{olmo.grpo-nodiv.minus.mt-grpo.ifbench-abs}{0.7}
\setval{olmo.grpo-nodiv.minus.mt-grpo.ifeval}{+0.7}
\setval{olmo.grpo-nodiv.minus.mt-grpo.ifeval-abs}{0.7}
\setval{olmo.grpo-nodiv.minus.mt-grpo.judge2.avg}{$-$0.4}
\setval{olmo.grpo-nodiv.minus.mt-grpo.judge2.avg-abs}{0.4}
\setval{olmo.grpo-nodiv.minus.mt-grpo.judge2.creative-writing-v3}{$-$1.0}
\setval{olmo.grpo-nodiv.minus.mt-grpo.judge2.creative-writing-v3-abs}{1.0}
\setval{olmo.grpo-nodiv.minus.mt-grpo.judge2.longbench-write}{+0.7}
\setval{olmo.grpo-nodiv.minus.mt-grpo.judge2.longbench-write-abs}{0.7}
\setval{olmo.grpo-nodiv.minus.mt-grpo.judge2.writing}{$-$0.3}
\setval{olmo.grpo-nodiv.minus.mt-grpo.judge2.writing-abs}{0.3}
\setval{olmo.grpo-nodiv.minus.mt-grpo.judge2.writingbench}{$-$0.6}
\setval{olmo.grpo-nodiv.minus.mt-grpo.judge2.writingbench-abs}{0.6}
\setval{olmo.grpo-nodiv.minus.mt-grpo.livecodebench}{+1.0}
\setval{olmo.grpo-nodiv.minus.mt-grpo.livecodebench-abs}{1.0}
\setval{olmo.grpo-nodiv.minus.mt-grpo.longbench-write}{+1.2}
\setval{olmo.grpo-nodiv.minus.mt-grpo.longbench-write-abs}{1.2}
\setval{olmo.grpo-nodiv.minus.mt-grpo.math}{$-$1.5}
\setval{olmo.grpo-nodiv.minus.mt-grpo.math-abs}{1.5}
\setval{olmo.grpo-nodiv.minus.mt-grpo.math500}{+0.6}
\setval{olmo.grpo-nodiv.minus.mt-grpo.math500-abs}{0.6}
\setval{olmo.grpo-nodiv.minus.mt-grpo.mbpp-plus}{+0.4}
\setval{olmo.grpo-nodiv.minus.mt-grpo.mbpp-plus-abs}{0.4}
\setval{olmo.grpo-nodiv.minus.mt-grpo.minerva}{$-$0.1}
\setval{olmo.grpo-nodiv.minus.mt-grpo.minerva-abs}{0.1}
\setval{olmo.grpo-nodiv.minus.mt-grpo.olympiad-bench}{+0.4}
\setval{olmo.grpo-nodiv.minus.mt-grpo.olympiad-bench-abs}{0.4}
\setval{olmo.grpo-nodiv.minus.mt-grpo.writing}{$-$2.4}
\setval{olmo.grpo-nodiv.minus.mt-grpo.writing-abs}{2.4}
\setval{olmo.grpo-nodiv.minus.mt-grpo.writingbench}{$-$4.5}
\setval{olmo.grpo-nodiv.minus.mt-grpo.writingbench-abs}{4.5}
\setval{olmo.grpo-nodiv.minus.papo.aime24}{+9.6}
\setval{olmo.grpo-nodiv.minus.papo.aime24-abs}{9.6}
\setval{olmo.grpo-nodiv.minus.papo.aime25}{+7.3}
\setval{olmo.grpo-nodiv.minus.papo.aime25-abs}{7.3}
\setval{olmo.grpo-nodiv.minus.papo.amc23}{+13.7}
\setval{olmo.grpo-nodiv.minus.papo.amc23-abs}{13.7}
\setval{olmo.grpo-nodiv.minus.papo.avg}{+3.6}
\setval{olmo.grpo-nodiv.minus.papo.avg-abs}{3.6}
\setval{olmo.grpo-nodiv.minus.papo.code}{+1.5}
\setval{olmo.grpo-nodiv.minus.papo.code-abs}{1.5}
\setval{olmo.grpo-nodiv.minus.papo.creative-writing-v3}{+0.0}
\setval{olmo.grpo-nodiv.minus.papo.creative-writing-v3-abs}{0.0}
\setval{olmo.grpo-nodiv.minus.papo.guidebench}{$-$3.2}
\setval{olmo.grpo-nodiv.minus.papo.guidebench-abs}{3.2}
\setval{olmo.grpo-nodiv.minus.papo.humaneval-plus}{+2.6}
\setval{olmo.grpo-nodiv.minus.papo.humaneval-plus-abs}{2.6}
\setval{olmo.grpo-nodiv.minus.papo.if}{$-$0.3}
\setval{olmo.grpo-nodiv.minus.papo.if-abs}{0.3}
\setval{olmo.grpo-nodiv.minus.papo.ifbench}{+1.7}
\setval{olmo.grpo-nodiv.minus.papo.ifbench-abs}{1.7}
\setval{olmo.grpo-nodiv.minus.papo.ifeval}{+0.6}
\setval{olmo.grpo-nodiv.minus.papo.ifeval-abs}{0.6}
\setval{olmo.grpo-nodiv.minus.papo.judge2.avg}{+3.9}
\setval{olmo.grpo-nodiv.minus.papo.judge2.avg-abs}{3.9}
\setval{olmo.grpo-nodiv.minus.papo.judge2.creative-writing-v3}{+0.9}
\setval{olmo.grpo-nodiv.minus.papo.judge2.creative-writing-v3-abs}{0.9}
\setval{olmo.grpo-nodiv.minus.papo.judge2.longbench-write}{+5.5}
\setval{olmo.grpo-nodiv.minus.papo.judge2.longbench-write-abs}{5.5}
\setval{olmo.grpo-nodiv.minus.papo.judge2.writing}{+1.9}
\setval{olmo.grpo-nodiv.minus.papo.judge2.writing-abs}{1.9}
\setval{olmo.grpo-nodiv.minus.papo.judge2.writingbench}{$-$0.5}
\setval{olmo.grpo-nodiv.minus.papo.judge2.writingbench-abs}{0.5}
\setval{olmo.grpo-nodiv.minus.papo.livecodebench}{+1.7}
\setval{olmo.grpo-nodiv.minus.papo.livecodebench-abs}{1.7}
\setval{olmo.grpo-nodiv.minus.papo.longbench-write}{+2.8}
\setval{olmo.grpo-nodiv.minus.papo.longbench-write-abs}{2.8}
\setval{olmo.grpo-nodiv.minus.papo.math}{+8.2}
\setval{olmo.grpo-nodiv.minus.papo.math-abs}{8.2}
\setval{olmo.grpo-nodiv.minus.papo.math500}{+6.6}
\setval{olmo.grpo-nodiv.minus.papo.math500-abs}{6.6}
\setval{olmo.grpo-nodiv.minus.papo.mbpp-plus}{+0.3}
\setval{olmo.grpo-nodiv.minus.papo.mbpp-plus-abs}{0.3}
\setval{olmo.grpo-nodiv.minus.papo.minerva}{+3.3}
\setval{olmo.grpo-nodiv.minus.papo.minerva-abs}{3.3}
\setval{olmo.grpo-nodiv.minus.papo.olympiad-bench}{+8.7}
\setval{olmo.grpo-nodiv.minus.papo.olympiad-bench-abs}{8.7}
\setval{olmo.grpo-nodiv.minus.papo.writing}{+0.2}
\setval{olmo.grpo-nodiv.minus.papo.writing-abs}{0.2}
\setval{olmo.grpo-nodiv.minus.papo.writingbench}{$-$2.2}
\setval{olmo.grpo-nodiv.minus.papo.writingbench-abs}{2.2}
\setval{olmo.grpo-nodiv.minus.saw.aime24}{+5.6}
\setval{olmo.grpo-nodiv.minus.saw.aime24-abs}{5.6}
\setval{olmo.grpo-nodiv.minus.saw.aime25}{+2.4}
\setval{olmo.grpo-nodiv.minus.saw.aime25-abs}{2.4}
\setval{olmo.grpo-nodiv.minus.saw.amc23}{+6.2}
\setval{olmo.grpo-nodiv.minus.saw.amc23-abs}{6.2}
\setval{olmo.grpo-nodiv.minus.saw.avg}{+1.5}
\setval{olmo.grpo-nodiv.minus.saw.avg-abs}{1.5}
\setval{olmo.grpo-nodiv.minus.saw.code}{+0.9}
\setval{olmo.grpo-nodiv.minus.saw.code-abs}{0.9}
\setval{olmo.grpo-nodiv.minus.saw.creative-writing-v3}{$-$0.9}
\setval{olmo.grpo-nodiv.minus.saw.creative-writing-v3-abs}{0.9}
\setval{olmo.grpo-nodiv.minus.saw.guidebench}{+0.8}
\setval{olmo.grpo-nodiv.minus.saw.guidebench-abs}{0.8}
\setval{olmo.grpo-nodiv.minus.saw.humaneval-plus}{+1.0}
\setval{olmo.grpo-nodiv.minus.saw.humaneval-plus-abs}{1.0}
\setval{olmo.grpo-nodiv.minus.saw.if}{$-$0.5}
\setval{olmo.grpo-nodiv.minus.saw.if-abs}{0.5}
\setval{olmo.grpo-nodiv.minus.saw.ifbench}{$-$1.3}
\setval{olmo.grpo-nodiv.minus.saw.ifbench-abs}{1.3}
\setval{olmo.grpo-nodiv.minus.saw.ifeval}{$-$0.8}
\setval{olmo.grpo-nodiv.minus.saw.ifeval-abs}{0.8}
\setval{olmo.grpo-nodiv.minus.saw.judge2.avg}{+1.6}
\setval{olmo.grpo-nodiv.minus.saw.judge2.avg-abs}{1.6}
\setval{olmo.grpo-nodiv.minus.saw.judge2.creative-writing-v3}{$-$2.2}
\setval{olmo.grpo-nodiv.minus.saw.judge2.creative-writing-v3-abs}{2.2}
\setval{olmo.grpo-nodiv.minus.saw.judge2.longbench-write}{+3.7}
\setval{olmo.grpo-nodiv.minus.saw.judge2.longbench-write-abs}{3.7}
\setval{olmo.grpo-nodiv.minus.saw.judge2.writing}{+0.0}
\setval{olmo.grpo-nodiv.minus.saw.judge2.writing-abs}{0.0}
\setval{olmo.grpo-nodiv.minus.saw.judge2.writingbench}{$-$1.5}
\setval{olmo.grpo-nodiv.minus.saw.judge2.writingbench-abs}{1.5}
\setval{olmo.grpo-nodiv.minus.saw.livecodebench}{+1.7}
\setval{olmo.grpo-nodiv.minus.saw.livecodebench-abs}{1.7}
\setval{olmo.grpo-nodiv.minus.saw.longbench-write}{+2.6}
\setval{olmo.grpo-nodiv.minus.saw.longbench-write-abs}{2.6}
\setval{olmo.grpo-nodiv.minus.saw.math}{+3.9}
\setval{olmo.grpo-nodiv.minus.saw.math-abs}{3.9}
\setval{olmo.grpo-nodiv.minus.saw.math500}{+3.0}
\setval{olmo.grpo-nodiv.minus.saw.math500-abs}{3.0}
\setval{olmo.grpo-nodiv.minus.saw.mbpp-plus}{$-$0.1}
\setval{olmo.grpo-nodiv.minus.saw.mbpp-plus-abs}{0.1}
\setval{olmo.grpo-nodiv.minus.saw.minerva}{+1.5}
\setval{olmo.grpo-nodiv.minus.saw.minerva-abs}{1.5}
\setval{olmo.grpo-nodiv.minus.saw.olympiad-bench}{+4.6}
\setval{olmo.grpo-nodiv.minus.saw.olympiad-bench-abs}{4.6}
\setval{olmo.grpo-nodiv.minus.saw.writing}{$-$0.8}
\setval{olmo.grpo-nodiv.minus.saw.writing-abs}{0.8}
\setval{olmo.grpo-nodiv.minus.saw.writingbench}{$-$4.2}
\setval{olmo.grpo-nodiv.minus.saw.writingbench-abs}{4.2}
\setval{olmo.grpo-nodiv.passing-at-ceiling.code.early}{0.84}
\setval{olmo.grpo-nodiv.passing-at-ceiling.code.late}{0.85}
\setval{olmo.grpo-nodiv.passing-at-ceiling.code.run}{0.84}
\setval{olmo.grpo-nodiv.passing-at-ceiling.if.early}{0.48}
\setval{olmo.grpo-nodiv.passing-at-ceiling.if.late}{0.56}
\setval{olmo.grpo-nodiv.passing-at-ceiling.if.run}{0.52}
\setval{olmo.grpo-nodiv.passing-at-ceiling.math.early}{1.00}
\setval{olmo.grpo-nodiv.passing-at-ceiling.math.late}{1.00}
\setval{olmo.grpo-nodiv.passing-at-ceiling.math.run}{1.00}
\setval{olmo.grpo-nodiv.passing-at-ceiling.writing.early}{0.00}
\setval{olmo.grpo-nodiv.passing-at-ceiling.writing.late}{0.00}
\setval{olmo.grpo-nodiv.passing-at-ceiling.writing.run}{0.00}
\setval{olmo.grpo-nodiv.primary-level.code.early}{0.60}
\setval{olmo.grpo-nodiv.primary-level.code.late}{0.84}
\setval{olmo.grpo-nodiv.primary-level.code.run}{0.77}
\setval{olmo.grpo-nodiv.primary-level.if.early}{0.63}
\setval{olmo.grpo-nodiv.primary-level.if.late}{0.74}
\setval{olmo.grpo-nodiv.primary-level.if.run}{0.69}
\setval{olmo.grpo-nodiv.primary-level.math.early}{0.35}
\setval{olmo.grpo-nodiv.primary-level.math.late}{0.71}
\setval{olmo.grpo-nodiv.primary-level.math.run}{0.60}
\setval{olmo.grpo-nodiv.primary-level.writing.early}{0.49}
\setval{olmo.grpo-nodiv.primary-level.writing.late}{0.67}
\setval{olmo.grpo-nodiv.primary-level.writing.run}{0.62}
\setval{olmo.grpo.aux-change.code}{+0.17}
\setval{olmo.grpo.aux-change.if}{+0.11}
\setval{olmo.grpo.aux-change.math}{$-$0.08}
\setval{olmo.grpo.aux-change.writing}{+0.02}
\setval{olmo.grpo.aux-change.writing-rm}{+0.30}
\setval{olmo.grpo.aux-full.code.early}{0.76}
\setval{olmo.grpo.aux-full.code.late}{0.94}
\setval{olmo.grpo.aux-full.code.run}{0.91}
\setval{olmo.grpo.aux-full.if.early}{0.60}
\setval{olmo.grpo.aux-full.if.late}{0.94}
\setval{olmo.grpo.aux-full.if.run}{0.87}
\setval{olmo.grpo.aux-full.math.early}{0.55}
\setval{olmo.grpo.aux-full.math.late}{0.66}
\setval{olmo.grpo.aux-full.math.run}{0.53}
\setval{olmo.grpo.aux-full.writing-rm.early}{0.02}
\setval{olmo.grpo.aux-full.writing-rm.late}{0.30}
\setval{olmo.grpo.aux-full.writing-rm.run}{0.13}
\setval{olmo.grpo.aux-full.writing.early}{0.57}
\setval{olmo.grpo.aux-full.writing.late}{0.69}
\setval{olmo.grpo.aux-full.writing.run}{0.61}
\setval{olmo.grpo.aux-level.code.early}{0.91}
\setval{olmo.grpo.aux-level.code.late}{0.98}
\setval{olmo.grpo.aux-level.code.run}{0.97}
\setval{olmo.grpo.aux-level.if.early}{0.91}
\setval{olmo.grpo.aux-level.if.late}{0.99}
\setval{olmo.grpo.aux-level.if.run}{0.97}
\setval{olmo.grpo.aux-level.math.early}{0.91}
\setval{olmo.grpo.aux-level.math.late}{0.82}
\setval{olmo.grpo.aux-level.math.run}{0.79}
\setval{olmo.grpo.aux-level.writing-rm.early}{0.60}
\setval{olmo.grpo.aux-level.writing-rm.late}{0.86}
\setval{olmo.grpo.aux-level.writing-rm.run}{0.77}
\setval{olmo.grpo.aux-level.writing.early}{0.95}
\setval{olmo.grpo.aux-level.writing.late}{0.97}
\setval{olmo.grpo.aux-level.writing.run}{0.96}
\setval{olmo.grpo.bench.aime24}{29.5}
\setval{olmo.grpo.bench.aime25}{24.7}
\setval{olmo.grpo.bench.amc23}{70.9}
\setval{olmo.grpo.bench.creative-writing-v3}{49.4}
\setval{olmo.grpo.bench.guidebench}{56.4}
\setval{olmo.grpo.bench.humaneval-plus}{77.3}
\setval{olmo.grpo.bench.ifbench}{32.0}
\setval{olmo.grpo.bench.ifeval}{87.4}
\setval{olmo.grpo.bench.livecodebench}{22.3}
\setval{olmo.grpo.bench.longbench-write}{64.2}
\setval{olmo.grpo.bench.math500}{85.0}
\setval{olmo.grpo.bench.mbpp-plus}{65.8}
\setval{olmo.grpo.bench.minerva}{39.9}
\setval{olmo.grpo.bench.olympiad-bench}{55.7}
\setval{olmo.grpo.bench.writingbench}{68.5}
\setval{olmo.grpo.eval.avg}{55.3}
\setval{olmo.grpo.eval.code}{55.2}
\setval{olmo.grpo.eval.if}{58.6}
\setval{olmo.grpo.eval.math}{50.9}
\setval{olmo.grpo.eval.writing}{60.7}
\setval{olmo.grpo.judge2.avg}{52.7}
\setval{olmo.grpo.judge2.creative-writing-v3}{38.6}
\setval{olmo.grpo.judge2.longbench-write}{57.1}
\setval{olmo.grpo.judge2.writing}{48.1}
\setval{olmo.grpo.judge2.writingbench}{48.5}
\setval{olmo.grpo.last-step}{300}
\setval{olmo.grpo.minus.base.aime24}{+20.9}
\setval{olmo.grpo.minus.base.aime24-abs}{20.9}
\setval{olmo.grpo.minus.base.aime25}{+17.3}
\setval{olmo.grpo.minus.base.aime25-abs}{17.3}
\setval{olmo.grpo.minus.base.amc23}{+26.3}
\setval{olmo.grpo.minus.base.amc23-abs}{26.3}
\setval{olmo.grpo.minus.base.avg}{+11.1}
\setval{olmo.grpo.minus.base.avg-abs}{11.1}
\setval{olmo.grpo.minus.base.code}{+6.0}
\setval{olmo.grpo.minus.base.code-abs}{6.0}
\setval{olmo.grpo.minus.base.creative-writing-v3}{+11.5}
\setval{olmo.grpo.minus.base.creative-writing-v3-abs}{11.5}
\setval{olmo.grpo.minus.base.guidebench}{$-$5.0}
\setval{olmo.grpo.minus.base.guidebench-abs}{5.0}
\setval{olmo.grpo.minus.base.humaneval-plus}{+8.1}
\setval{olmo.grpo.minus.base.humaneval-plus-abs}{8.1}
\setval{olmo.grpo.minus.base.if}{+1.1}
\setval{olmo.grpo.minus.base.if-abs}{1.1}
\setval{olmo.grpo.minus.base.ifbench}{+5.3}
\setval{olmo.grpo.minus.base.ifbench-abs}{5.3}
\setval{olmo.grpo.minus.base.ifeval}{+3.0}
\setval{olmo.grpo.minus.base.ifeval-abs}{3.0}
\setval{olmo.grpo.minus.base.judge2.avg}{+9.8}
\setval{olmo.grpo.minus.base.judge2.avg-abs}{9.8}
\setval{olmo.grpo.minus.base.judge2.creative-writing-v3}{+7.0}
\setval{olmo.grpo.minus.base.judge2.creative-writing-v3-abs}{7.0}
\setval{olmo.grpo.minus.base.judge2.longbench-write}{+3.9}
\setval{olmo.grpo.minus.base.judge2.longbench-write-abs}{3.9}
\setval{olmo.grpo.minus.base.judge2.writing}{+7.2}
\setval{olmo.grpo.minus.base.judge2.writing-abs}{7.2}
\setval{olmo.grpo.minus.base.judge2.writingbench}{+10.8}
\setval{olmo.grpo.minus.base.judge2.writingbench-abs}{10.8}
\setval{olmo.grpo.minus.base.livecodebench}{+5.5}
\setval{olmo.grpo.minus.base.livecodebench-abs}{5.5}
\setval{olmo.grpo.minus.base.longbench-write}{+6.7}
\setval{olmo.grpo.minus.base.longbench-write-abs}{6.7}
\setval{olmo.grpo.minus.base.math}{+17.2}
\setval{olmo.grpo.minus.base.math-abs}{17.2}
\setval{olmo.grpo.minus.base.math500}{+14.8}
\setval{olmo.grpo.minus.base.math500-abs}{14.8}
\setval{olmo.grpo.minus.base.mbpp-plus}{+4.6}
\setval{olmo.grpo.minus.base.mbpp-plus-abs}{4.6}
\setval{olmo.grpo.minus.base.minerva}{+4.6}
\setval{olmo.grpo.minus.base.minerva-abs}{4.6}
\setval{olmo.grpo.minus.base.olympiad-bench}{+19.4}
\setval{olmo.grpo.minus.base.olympiad-bench-abs}{19.4}
\setval{olmo.grpo.minus.base.writing}{+14.0}
\setval{olmo.grpo.minus.base.writing-abs}{14.0}
\setval{olmo.grpo.minus.base.writingbench}{+23.7}
\setval{olmo.grpo.minus.base.writingbench-abs}{23.7}
\setval{olmo.grpo.minus.concur-free.aime24}{+2.7}
\setval{olmo.grpo.minus.concur-free.aime24-abs}{2.7}
\setval{olmo.grpo.minus.concur-free.aime25}{+2.6}
\setval{olmo.grpo.minus.concur-free.aime25-abs}{2.6}
\setval{olmo.grpo.minus.concur-free.amc23}{+3.4}
\setval{olmo.grpo.minus.concur-free.amc23-abs}{3.4}
\setval{olmo.grpo.minus.concur-free.avg}{$-$0.2}
\setval{olmo.grpo.minus.concur-free.avg-abs}{0.2}
\setval{olmo.grpo.minus.concur-free.code}{$-$0.4}
\setval{olmo.grpo.minus.concur-free.code-abs}{0.4}
\setval{olmo.grpo.minus.concur-free.creative-writing-v3}{$-$0.4}
\setval{olmo.grpo.minus.concur-free.creative-writing-v3-abs}{0.4}
\setval{olmo.grpo.minus.concur-free.guidebench}{$-$3.7}
\setval{olmo.grpo.minus.concur-free.guidebench-abs}{3.7}
\setval{olmo.grpo.minus.concur-free.humaneval-plus}{+0.7}
\setval{olmo.grpo.minus.concur-free.humaneval-plus-abs}{0.7}
\setval{olmo.grpo.minus.concur-free.if}{$-$1.1}
\setval{olmo.grpo.minus.concur-free.if-abs}{1.1}
\setval{olmo.grpo.minus.concur-free.ifbench}{+0.3}
\setval{olmo.grpo.minus.concur-free.ifbench-abs}{0.3}
\setval{olmo.grpo.minus.concur-free.ifeval}{+0.1}
\setval{olmo.grpo.minus.concur-free.ifeval-abs}{0.1}
\setval{olmo.grpo.minus.concur-free.judge2.avg}{+0.1}
\setval{olmo.grpo.minus.concur-free.judge2.avg-abs}{0.1}
\setval{olmo.grpo.minus.concur-free.judge2.creative-writing-v3}{+1.1}
\setval{olmo.grpo.minus.concur-free.judge2.creative-writing-v3-abs}{1.1}
\setval{olmo.grpo.minus.concur-free.judge2.longbench-write}{$-$6.8}
\setval{olmo.grpo.minus.concur-free.judge2.longbench-write-abs}{6.8}
\setval{olmo.grpo.minus.concur-free.judge2.writing}{$-$1.8}
\setval{olmo.grpo.minus.concur-free.judge2.writing-abs}{1.8}
\setval{olmo.grpo.minus.concur-free.judge2.writingbench}{+0.5}
\setval{olmo.grpo.minus.concur-free.judge2.writingbench-abs}{0.5}
\setval{olmo.grpo.minus.concur-free.livecodebench}{$-$3.3}
\setval{olmo.grpo.minus.concur-free.livecodebench-abs}{3.3}
\setval{olmo.grpo.minus.concur-free.longbench-write}{$-$7.8}
\setval{olmo.grpo.minus.concur-free.longbench-write-abs}{7.8}
\setval{olmo.grpo.minus.concur-free.math}{+1.9}
\setval{olmo.grpo.minus.concur-free.math-abs}{1.9}
\setval{olmo.grpo.minus.concur-free.math500}{+0.6}
\setval{olmo.grpo.minus.concur-free.math500-abs}{0.6}
\setval{olmo.grpo.minus.concur-free.mbpp-plus}{+1.3}
\setval{olmo.grpo.minus.concur-free.mbpp-plus-abs}{1.3}
\setval{olmo.grpo.minus.concur-free.minerva}{+0.0}
\setval{olmo.grpo.minus.concur-free.minerva-abs}{0.0}
\setval{olmo.grpo.minus.concur-free.olympiad-bench}{+2.4}
\setval{olmo.grpo.minus.concur-free.olympiad-bench-abs}{2.4}
\setval{olmo.grpo.minus.concur-free.writing}{$-$3.3}
\setval{olmo.grpo.minus.concur-free.writing-abs}{3.3}
\setval{olmo.grpo.minus.concur-free.writingbench}{$-$1.7}
\setval{olmo.grpo.minus.concur-free.writingbench-abs}{1.7}
\setval{olmo.grpo.minus.concur-tied.aime24}{$-$5.0}
\setval{olmo.grpo.minus.concur-tied.aime24-abs}{5.0}
\setval{olmo.grpo.minus.concur-tied.aime25}{$-$5.2}
\setval{olmo.grpo.minus.concur-tied.aime25-abs}{5.2}
\setval{olmo.grpo.minus.concur-tied.amc23}{$-$9.0}
\setval{olmo.grpo.minus.concur-tied.amc23-abs}{9.0}
\setval{olmo.grpo.minus.concur-tied.avg}{$-$3.9}
\setval{olmo.grpo.minus.concur-tied.avg-abs}{3.9}
\setval{olmo.grpo.minus.concur-tied.code}{$-$6.0}
\setval{olmo.grpo.minus.concur-tied.code-abs}{6.0}
\setval{olmo.grpo.minus.concur-tied.creative-writing-v3}{+0.3}
\setval{olmo.grpo.minus.concur-tied.creative-writing-v3-abs}{0.3}
\setval{olmo.grpo.minus.concur-tied.guidebench}{$-$2.1}
\setval{olmo.grpo.minus.concur-tied.guidebench-abs}{2.1}
\setval{olmo.grpo.minus.concur-tied.humaneval-plus}{$-$6.0}
\setval{olmo.grpo.minus.concur-tied.humaneval-plus-abs}{6.0}
\setval{olmo.grpo.minus.concur-tied.if}{$-$1.3}
\setval{olmo.grpo.minus.concur-tied.if-abs}{1.3}
\setval{olmo.grpo.minus.concur-tied.ifbench}{$-$1.7}
\setval{olmo.grpo.minus.concur-tied.ifbench-abs}{1.7}
\setval{olmo.grpo.minus.concur-tied.ifeval}{$-$0.2}
\setval{olmo.grpo.minus.concur-tied.ifeval-abs}{0.2}
\setval{olmo.grpo.minus.concur-tied.judge2.avg}{$-$3.6}
\setval{olmo.grpo.minus.concur-tied.judge2.avg-abs}{3.6}
\setval{olmo.grpo.minus.concur-tied.judge2.creative-writing-v3}{+1.4}
\setval{olmo.grpo.minus.concur-tied.judge2.creative-writing-v3-abs}{1.4}
\setval{olmo.grpo.minus.concur-tied.judge2.longbench-write}{$-$3.4}
\setval{olmo.grpo.minus.concur-tied.judge2.longbench-write-abs}{3.4}
\setval{olmo.grpo.minus.concur-tied.judge2.writing}{+0.3}
\setval{olmo.grpo.minus.concur-tied.judge2.writing-abs}{0.3}
\setval{olmo.grpo.minus.concur-tied.judge2.writingbench}{+2.9}
\setval{olmo.grpo.minus.concur-tied.judge2.writingbench-abs}{2.9}
\setval{olmo.grpo.minus.concur-tied.livecodebench}{$-$6.7}
\setval{olmo.grpo.minus.concur-tied.livecodebench-abs}{6.7}
\setval{olmo.grpo.minus.concur-tied.longbench-write}{$-$5.0}
\setval{olmo.grpo.minus.concur-tied.longbench-write-abs}{5.0}
\setval{olmo.grpo.minus.concur-tied.math}{$-$5.6}
\setval{olmo.grpo.minus.concur-tied.math-abs}{5.6}
\setval{olmo.grpo.minus.concur-tied.math500}{$-$4.5}
\setval{olmo.grpo.minus.concur-tied.math500-abs}{4.5}
\setval{olmo.grpo.minus.concur-tied.mbpp-plus}{$-$5.3}
\setval{olmo.grpo.minus.concur-tied.mbpp-plus-abs}{5.3}
\setval{olmo.grpo.minus.concur-tied.minerva}{$-$4.1}
\setval{olmo.grpo.minus.concur-tied.minerva-abs}{4.1}
\setval{olmo.grpo.minus.concur-tied.olympiad-bench}{$-$5.9}
\setval{olmo.grpo.minus.concur-tied.olympiad-bench-abs}{5.9}
\setval{olmo.grpo.minus.concur-tied.writing}{$-$0.8}
\setval{olmo.grpo.minus.concur-tied.writing-abs}{0.8}
\setval{olmo.grpo.minus.concur-tied.writingbench}{+2.3}
\setval{olmo.grpo.minus.concur-tied.writingbench-abs}{2.3}
\setval{olmo.grpo.minus.concur.aime24}{$-$6.9}
\setval{olmo.grpo.minus.concur.aime24-abs}{6.9}
\setval{olmo.grpo.minus.concur.aime25}{$-$7.5}
\setval{olmo.grpo.minus.concur.aime25-abs}{7.5}
\setval{olmo.grpo.minus.concur.amc23}{$-$12.7}
\setval{olmo.grpo.minus.concur.amc23-abs}{12.7}
\setval{olmo.grpo.minus.concur.avg}{$-$4.1}
\setval{olmo.grpo.minus.concur.avg-abs}{4.1}
\setval{olmo.grpo.minus.concur.code}{$-$5.1}
\setval{olmo.grpo.minus.concur.code-abs}{5.1}
\setval{olmo.grpo.minus.concur.creative-writing-v3}{+0.2}
\setval{olmo.grpo.minus.concur.creative-writing-v3-abs}{0.2}
\setval{olmo.grpo.minus.concur.guidebench}{$-$0.5}
\setval{olmo.grpo.minus.concur.guidebench-abs}{0.5}
\setval{olmo.grpo.minus.concur.humaneval-plus}{$-$5.4}
\setval{olmo.grpo.minus.concur.humaneval-plus-abs}{5.4}
\setval{olmo.grpo.minus.concur.if}{$-$0.3}
\setval{olmo.grpo.minus.concur.if-abs}{0.3}
\setval{olmo.grpo.minus.concur.ifbench}{+0.0}
\setval{olmo.grpo.minus.concur.ifbench-abs}{0.0}
\setval{olmo.grpo.minus.concur.ifeval}{$-$0.4}
\setval{olmo.grpo.minus.concur.ifeval-abs}{0.4}
\setval{olmo.grpo.minus.concur.judge2.avg}{$-$3.9}
\setval{olmo.grpo.minus.concur.judge2.avg-abs}{3.9}
\setval{olmo.grpo.minus.concur.judge2.creative-writing-v3}{+0.2}
\setval{olmo.grpo.minus.concur.judge2.creative-writing-v3-abs}{0.2}
\setval{olmo.grpo.minus.concur.judge2.longbench-write}{$-$1.3}
\setval{olmo.grpo.minus.concur.judge2.longbench-write-abs}{1.3}
\setval{olmo.grpo.minus.concur.judge2.writing}{+0.8}
\setval{olmo.grpo.minus.concur.judge2.writing-abs}{0.8}
\setval{olmo.grpo.minus.concur.judge2.writingbench}{+3.6}
\setval{olmo.grpo.minus.concur.judge2.writingbench-abs}{3.6}
\setval{olmo.grpo.minus.concur.livecodebench}{$-$8.0}
\setval{olmo.grpo.minus.concur.livecodebench-abs}{8.0}
\setval{olmo.grpo.minus.concur.longbench-write}{$-$4.3}
\setval{olmo.grpo.minus.concur.longbench-write-abs}{4.3}
\setval{olmo.grpo.minus.concur.math}{$-$7.4}
\setval{olmo.grpo.minus.concur.math-abs}{7.4}
\setval{olmo.grpo.minus.concur.math500}{$-$5.6}
\setval{olmo.grpo.minus.concur.math500-abs}{5.6}
\setval{olmo.grpo.minus.concur.mbpp-plus}{$-$1.7}
\setval{olmo.grpo.minus.concur.mbpp-plus-abs}{1.7}
\setval{olmo.grpo.minus.concur.minerva}{$-$4.9}
\setval{olmo.grpo.minus.concur.minerva-abs}{4.9}
\setval{olmo.grpo.minus.concur.olympiad-bench}{$-$6.7}
\setval{olmo.grpo.minus.concur.olympiad-bench-abs}{6.7}
\setval{olmo.grpo.minus.concur.writing}{$-$0.4}
\setval{olmo.grpo.minus.concur.writing-abs}{0.4}
\setval{olmo.grpo.minus.concur.writingbench}{+3.0}
\setval{olmo.grpo.minus.concur.writingbench-abs}{3.0}
\setval{olmo.grpo.minus.gd2po.aime24}{+1.1}
\setval{olmo.grpo.minus.gd2po.aime24-abs}{1.1}
\setval{olmo.grpo.minus.gd2po.aime25}{+0.4}
\setval{olmo.grpo.minus.gd2po.aime25-abs}{0.4}
\setval{olmo.grpo.minus.gd2po.amc23}{$-$1.3}
\setval{olmo.grpo.minus.gd2po.amc23-abs}{1.3}
\setval{olmo.grpo.minus.gd2po.avg}{$-$0.4}
\setval{olmo.grpo.minus.gd2po.avg-abs}{0.4}
\setval{olmo.grpo.minus.gd2po.code}{+0.2}
\setval{olmo.grpo.minus.gd2po.code-abs}{0.2}
\setval{olmo.grpo.minus.gd2po.creative-writing-v3}{+1.1}
\setval{olmo.grpo.minus.gd2po.creative-writing-v3-abs}{1.1}
\setval{olmo.grpo.minus.gd2po.guidebench}{$-$4.7}
\setval{olmo.grpo.minus.gd2po.guidebench-abs}{4.7}
\setval{olmo.grpo.minus.gd2po.humaneval-plus}{+0.5}
\setval{olmo.grpo.minus.gd2po.humaneval-plus-abs}{0.5}
\setval{olmo.grpo.minus.gd2po.if}{$-$1.8}
\setval{olmo.grpo.minus.gd2po.if-abs}{1.8}
\setval{olmo.grpo.minus.gd2po.ifbench}{$-$0.3}
\setval{olmo.grpo.minus.gd2po.ifbench-abs}{0.3}
\setval{olmo.grpo.minus.gd2po.ifeval}{$-$0.2}
\setval{olmo.grpo.minus.gd2po.ifeval-abs}{0.2}
\setval{olmo.grpo.minus.gd2po.judge2.avg}{$-$0.5}
\setval{olmo.grpo.minus.gd2po.judge2.avg-abs}{0.5}
\setval{olmo.grpo.minus.gd2po.judge2.creative-writing-v3}{+0.7}
\setval{olmo.grpo.minus.gd2po.judge2.creative-writing-v3-abs}{0.7}
\setval{olmo.grpo.minus.gd2po.judge2.longbench-write}{$-$2.8}
\setval{olmo.grpo.minus.gd2po.judge2.longbench-write-abs}{2.8}
\setval{olmo.grpo.minus.gd2po.judge2.writing}{$-$0.7}
\setval{olmo.grpo.minus.gd2po.judge2.writing-abs}{0.7}
\setval{olmo.grpo.minus.gd2po.judge2.writingbench}{+0.1}
\setval{olmo.grpo.minus.gd2po.judge2.writingbench-abs}{0.1}
\setval{olmo.grpo.minus.gd2po.livecodebench}{$-$0.9}
\setval{olmo.grpo.minus.gd2po.livecodebench-abs}{0.9}
\setval{olmo.grpo.minus.gd2po.longbench-write}{$-$4.0}
\setval{olmo.grpo.minus.gd2po.longbench-write-abs}{4.0}
\setval{olmo.grpo.minus.gd2po.math}{$-$0.2}
\setval{olmo.grpo.minus.gd2po.math-abs}{0.2}
\setval{olmo.grpo.minus.gd2po.math500}{$-$0.8}
\setval{olmo.grpo.minus.gd2po.math500-abs}{0.8}
\setval{olmo.grpo.minus.gd2po.mbpp-plus}{+1.0}
\setval{olmo.grpo.minus.gd2po.mbpp-plus-abs}{1.0}
\setval{olmo.grpo.minus.gd2po.minerva}{$-$0.2}
\setval{olmo.grpo.minus.gd2po.minerva-abs}{0.2}
\setval{olmo.grpo.minus.gd2po.olympiad-bench}{$-$0.4}
\setval{olmo.grpo.minus.gd2po.olympiad-bench-abs}{0.4}
\setval{olmo.grpo.minus.gd2po.writing}{$-$0.1}
\setval{olmo.grpo.minus.gd2po.writing-abs}{0.1}
\setval{olmo.grpo.minus.gd2po.writingbench}{+2.7}
\setval{olmo.grpo.minus.gd2po.writingbench-abs}{2.7}
\setval{olmo.grpo.minus.gdpo-noaux.aime24}{$-$8.9}
\setval{olmo.grpo.minus.gdpo-noaux.aime24-abs}{8.9}
\setval{olmo.grpo.minus.gdpo-noaux.aime25}{$-$9.0}
\setval{olmo.grpo.minus.gdpo-noaux.aime25-abs}{9.0}
\setval{olmo.grpo.minus.gdpo-noaux.amc23}{$-$13.0}
\setval{olmo.grpo.minus.gdpo-noaux.amc23-abs}{13.0}
\setval{olmo.grpo.minus.gdpo-noaux.avg}{$-$2.6}
\setval{olmo.grpo.minus.gdpo-noaux.avg-abs}{2.6}
\setval{olmo.grpo.minus.gdpo-noaux.code}{$-$7.3}
\setval{olmo.grpo.minus.gdpo-noaux.code-abs}{7.3}
\setval{olmo.grpo.minus.gdpo-noaux.creative-writing-v3}{+11.2}
\setval{olmo.grpo.minus.gdpo-noaux.creative-writing-v3-abs}{11.2}
\setval{olmo.grpo.minus.gdpo-noaux.guidebench}{+0.7}
\setval{olmo.grpo.minus.gdpo-noaux.guidebench-abs}{0.7}
\setval{olmo.grpo.minus.gdpo-noaux.humaneval-plus}{$-$7.9}
\setval{olmo.grpo.minus.gdpo-noaux.humaneval-plus-abs}{7.9}
\setval{olmo.grpo.minus.gdpo-noaux.if}{+0.7}
\setval{olmo.grpo.minus.gdpo-noaux.if-abs}{0.7}
\setval{olmo.grpo.minus.gdpo-noaux.ifbench}{$-$1.3}
\setval{olmo.grpo.minus.gdpo-noaux.ifbench-abs}{1.3}
\setval{olmo.grpo.minus.gdpo-noaux.ifeval}{+2.6}
\setval{olmo.grpo.minus.gdpo-noaux.ifeval-abs}{2.6}
\setval{olmo.grpo.minus.gdpo-noaux.judge2.avg}{$-$2.9}
\setval{olmo.grpo.minus.gdpo-noaux.judge2.avg-abs}{2.9}
\setval{olmo.grpo.minus.gdpo-noaux.judge2.creative-writing-v3}{+7.7}
\setval{olmo.grpo.minus.gdpo-noaux.judge2.creative-writing-v3-abs}{7.7}
\setval{olmo.grpo.minus.gdpo-noaux.judge2.longbench-write}{+10.5}
\setval{olmo.grpo.minus.gdpo-noaux.judge2.longbench-write-abs}{10.5}
\setval{olmo.grpo.minus.gdpo-noaux.judge2.writing}{+8.9}
\setval{olmo.grpo.minus.gdpo-noaux.judge2.writing-abs}{8.9}
\setval{olmo.grpo.minus.gdpo-noaux.judge2.writingbench}{+8.7}
\setval{olmo.grpo.minus.gdpo-noaux.judge2.writingbench-abs}{8.7}
\setval{olmo.grpo.minus.gdpo-noaux.livecodebench}{$-$9.2}
\setval{olmo.grpo.minus.gdpo-noaux.livecodebench-abs}{9.2}
\setval{olmo.grpo.minus.gdpo-noaux.longbench-write}{+8.2}
\setval{olmo.grpo.minus.gdpo-noaux.longbench-write-abs}{8.2}
\setval{olmo.grpo.minus.gdpo-noaux.math}{$-$8.5}
\setval{olmo.grpo.minus.gdpo-noaux.math-abs}{8.5}
\setval{olmo.grpo.minus.gdpo-noaux.math500}{$-$5.9}
\setval{olmo.grpo.minus.gdpo-noaux.math500-abs}{5.9}
\setval{olmo.grpo.minus.gdpo-noaux.mbpp-plus}{$-$4.7}
\setval{olmo.grpo.minus.gdpo-noaux.mbpp-plus-abs}{4.7}
\setval{olmo.grpo.minus.gdpo-noaux.minerva}{$-$6.2}
\setval{olmo.grpo.minus.gdpo-noaux.minerva-abs}{6.2}
\setval{olmo.grpo.minus.gdpo-noaux.olympiad-bench}{$-$8.1}
\setval{olmo.grpo.minus.gdpo-noaux.olympiad-bench-abs}{8.1}
\setval{olmo.grpo.minus.gdpo-noaux.writing}{+10.8}
\setval{olmo.grpo.minus.gdpo-noaux.writing-abs}{10.8}
\setval{olmo.grpo.minus.gdpo-noaux.writingbench}{+13.0}
\setval{olmo.grpo.minus.gdpo-noaux.writingbench-abs}{13.0}
\setval{olmo.grpo.minus.gdpo-w0.5.aime24}{$-$0.1}
\setval{olmo.grpo.minus.gdpo-w0.5.aime24-abs}{0.1}
\setval{olmo.grpo.minus.gdpo-w0.5.aime25}{$-$2.6}
\setval{olmo.grpo.minus.gdpo-w0.5.aime25-abs}{2.6}
\setval{olmo.grpo.minus.gdpo-w0.5.amc23}{$-$3.1}
\setval{olmo.grpo.minus.gdpo-w0.5.amc23-abs}{3.1}
\setval{olmo.grpo.minus.gdpo-w0.5.avg}{$-$0.4}
\setval{olmo.grpo.minus.gdpo-w0.5.avg-abs}{0.4}
\setval{olmo.grpo.minus.gdpo-w0.5.code}{$-$0.7}
\setval{olmo.grpo.minus.gdpo-w0.5.code-abs}{0.7}
\setval{olmo.grpo.minus.gdpo-w0.5.creative-writing-v3}{+1.3}
\setval{olmo.grpo.minus.gdpo-w0.5.creative-writing-v3-abs}{1.3}
\setval{olmo.grpo.minus.gdpo-w0.5.guidebench}{+4.5}
\setval{olmo.grpo.minus.gdpo-w0.5.guidebench-abs}{4.5}
\setval{olmo.grpo.minus.gdpo-w0.5.humaneval-plus}{$-$0.4}
\setval{olmo.grpo.minus.gdpo-w0.5.humaneval-plus-abs}{0.4}
\setval{olmo.grpo.minus.gdpo-w0.5.if}{+1.2}
\setval{olmo.grpo.minus.gdpo-w0.5.if-abs}{1.2}
\setval{olmo.grpo.minus.gdpo-w0.5.ifbench}{$-$0.7}
\setval{olmo.grpo.minus.gdpo-w0.5.ifbench-abs}{0.7}
\setval{olmo.grpo.minus.gdpo-w0.5.ifeval}{$-$0.1}
\setval{olmo.grpo.minus.gdpo-w0.5.ifeval-abs}{0.1}
\setval{olmo.grpo.minus.gdpo-w0.5.judge2.avg}{$-$0.5}
\setval{olmo.grpo.minus.gdpo-w0.5.judge2.avg-abs}{0.5}
\setval{olmo.grpo.minus.gdpo-w0.5.judge2.creative-writing-v3}{$-$0.8}
\setval{olmo.grpo.minus.gdpo-w0.5.judge2.creative-writing-v3-abs}{0.8}
\setval{olmo.grpo.minus.gdpo-w0.5.judge2.longbench-write}{+0.5}
\setval{olmo.grpo.minus.gdpo-w0.5.judge2.longbench-write-abs}{0.5}
\setval{olmo.grpo.minus.gdpo-w0.5.judge2.writing}{+0.5}
\setval{olmo.grpo.minus.gdpo-w0.5.judge2.writing-abs}{0.5}
\setval{olmo.grpo.minus.gdpo-w0.5.judge2.writingbench}{+1.8}
\setval{olmo.grpo.minus.gdpo-w0.5.judge2.writingbench-abs}{1.8}
\setval{olmo.grpo.minus.gdpo-w0.5.livecodebench}{$-$2.5}
\setval{olmo.grpo.minus.gdpo-w0.5.livecodebench-abs}{2.5}
\setval{olmo.grpo.minus.gdpo-w0.5.longbench-write}{+0.5}
\setval{olmo.grpo.minus.gdpo-w0.5.longbench-write-abs}{0.5}
\setval{olmo.grpo.minus.gdpo-w0.5.math}{$-$1.9}
\setval{olmo.grpo.minus.gdpo-w0.5.math-abs}{1.9}
\setval{olmo.grpo.minus.gdpo-w0.5.math500}{$-$2.1}
\setval{olmo.grpo.minus.gdpo-w0.5.math500-abs}{2.1}
\setval{olmo.grpo.minus.gdpo-w0.5.mbpp-plus}{+0.9}
\setval{olmo.grpo.minus.gdpo-w0.5.mbpp-plus-abs}{0.9}
\setval{olmo.grpo.minus.gdpo-w0.5.minerva}{$-$1.6}
\setval{olmo.grpo.minus.gdpo-w0.5.minerva-abs}{1.6}
\setval{olmo.grpo.minus.gdpo-w0.5.olympiad-bench}{$-$1.8}
\setval{olmo.grpo.minus.gdpo-w0.5.olympiad-bench-abs}{1.8}
\setval{olmo.grpo.minus.gdpo-w0.5.writing}{+1.2}
\setval{olmo.grpo.minus.gdpo-w0.5.writing-abs}{1.2}
\setval{olmo.grpo.minus.gdpo-w0.5.writingbench}{+2.0}
\setval{olmo.grpo.minus.gdpo-w0.5.writingbench-abs}{2.0}
\setval{olmo.grpo.minus.gdpo.aime24}{+1.7}
\setval{olmo.grpo.minus.gdpo.aime24-abs}{1.7}
\setval{olmo.grpo.minus.gdpo.aime25}{+2.1}
\setval{olmo.grpo.minus.gdpo.aime25-abs}{2.1}
\setval{olmo.grpo.minus.gdpo.amc23}{+1.3}
\setval{olmo.grpo.minus.gdpo.amc23-abs}{1.3}
\setval{olmo.grpo.minus.gdpo.avg}{+1.4}
\setval{olmo.grpo.minus.gdpo.avg-abs}{1.4}
\setval{olmo.grpo.minus.gdpo.code}{+0.4}
\setval{olmo.grpo.minus.gdpo.code-abs}{0.4}
\setval{olmo.grpo.minus.gdpo.creative-writing-v3}{+2.2}
\setval{olmo.grpo.minus.gdpo.creative-writing-v3-abs}{2.2}
\setval{olmo.grpo.minus.gdpo.guidebench}{+5.6}
\setval{olmo.grpo.minus.gdpo.guidebench-abs}{5.6}
\setval{olmo.grpo.minus.gdpo.humaneval-plus}{+1.0}
\setval{olmo.grpo.minus.gdpo.humaneval-plus-abs}{1.0}
\setval{olmo.grpo.minus.gdpo.if}{+3.0}
\setval{olmo.grpo.minus.gdpo.if-abs}{3.0}
\setval{olmo.grpo.minus.gdpo.ifbench}{+2.3}
\setval{olmo.grpo.minus.gdpo.ifbench-abs}{2.3}
\setval{olmo.grpo.minus.gdpo.ifeval}{+1.0}
\setval{olmo.grpo.minus.gdpo.ifeval-abs}{1.0}
\setval{olmo.grpo.minus.gdpo.judge2.avg}{+1.2}
\setval{olmo.grpo.minus.gdpo.judge2.avg-abs}{1.2}
\setval{olmo.grpo.minus.gdpo.judge2.creative-writing-v3}{+0.8}
\setval{olmo.grpo.minus.gdpo.judge2.creative-writing-v3-abs}{0.8}
\setval{olmo.grpo.minus.gdpo.judge2.longbench-write}{+1.4}
\setval{olmo.grpo.minus.gdpo.judge2.longbench-write-abs}{1.4}
\setval{olmo.grpo.minus.gdpo.judge2.writing}{+1.2}
\setval{olmo.grpo.minus.gdpo.judge2.writing-abs}{1.2}
\setval{olmo.grpo.minus.gdpo.judge2.writingbench}{+1.4}
\setval{olmo.grpo.minus.gdpo.judge2.writingbench-abs}{1.4}
\setval{olmo.grpo.minus.gdpo.livecodebench}{$-$0.8}
\setval{olmo.grpo.minus.gdpo.livecodebench-abs}{0.8}
\setval{olmo.grpo.minus.gdpo.longbench-write}{+0.9}
\setval{olmo.grpo.minus.gdpo.longbench-write-abs}{0.9}
\setval{olmo.grpo.minus.gdpo.math}{+0.7}
\setval{olmo.grpo.minus.gdpo.math-abs}{0.7}
\setval{olmo.grpo.minus.gdpo.math500}{$-$0.3}
\setval{olmo.grpo.minus.gdpo.math500-abs}{0.3}
\setval{olmo.grpo.minus.gdpo.mbpp-plus}{+1.0}
\setval{olmo.grpo.minus.gdpo.mbpp-plus-abs}{1.0}
\setval{olmo.grpo.minus.gdpo.minerva}{$-$0.4}
\setval{olmo.grpo.minus.gdpo.minerva-abs}{0.4}
\setval{olmo.grpo.minus.gdpo.olympiad-bench}{$-$0.2}
\setval{olmo.grpo.minus.gdpo.olympiad-bench-abs}{0.2}
\setval{olmo.grpo.minus.gdpo.writing}{+2.1}
\setval{olmo.grpo.minus.gdpo.writing-abs}{2.1}
\setval{olmo.grpo.minus.gdpo.writingbench}{+3.2}
\setval{olmo.grpo.minus.gdpo.writingbench-abs}{3.2}
\setval{olmo.grpo.minus.geoma.aime24}{+5.7}
\setval{olmo.grpo.minus.geoma.aime24-abs}{5.7}
\setval{olmo.grpo.minus.geoma.aime25}{+2.7}
\setval{olmo.grpo.minus.geoma.aime25-abs}{2.7}
\setval{olmo.grpo.minus.geoma.amc23}{+7.0}
\setval{olmo.grpo.minus.geoma.amc23-abs}{7.0}
\setval{olmo.grpo.minus.geoma.avg}{+2.3}
\setval{olmo.grpo.minus.geoma.avg-abs}{2.3}
\setval{olmo.grpo.minus.geoma.code}{+1.2}
\setval{olmo.grpo.minus.geoma.code-abs}{1.2}
\setval{olmo.grpo.minus.geoma.creative-writing-v3}{$-$0.2}
\setval{olmo.grpo.minus.geoma.creative-writing-v3-abs}{0.2}
\setval{olmo.grpo.minus.geoma.guidebench}{+9.8}
\setval{olmo.grpo.minus.geoma.guidebench-abs}{9.8}
\setval{olmo.grpo.minus.geoma.humaneval-plus}{+2.7}
\setval{olmo.grpo.minus.geoma.humaneval-plus-abs}{2.7}
\setval{olmo.grpo.minus.geoma.if}{+3.6}
\setval{olmo.grpo.minus.geoma.if-abs}{3.6}
\setval{olmo.grpo.minus.geoma.ifbench}{+1.7}
\setval{olmo.grpo.minus.geoma.ifbench-abs}{1.7}
\setval{olmo.grpo.minus.geoma.ifeval}{$-$0.6}
\setval{olmo.grpo.minus.geoma.ifeval-abs}{0.6}
\setval{olmo.grpo.minus.geoma.judge2.avg}{+2.0}
\setval{olmo.grpo.minus.geoma.judge2.avg-abs}{2.0}
\setval{olmo.grpo.minus.geoma.judge2.creative-writing-v3}{$-$3.2}
\setval{olmo.grpo.minus.geoma.judge2.creative-writing-v3-abs}{3.2}
\setval{olmo.grpo.minus.geoma.judge2.longbench-write}{$-$1.1}
\setval{olmo.grpo.minus.geoma.judge2.longbench-write-abs}{1.1}
\setval{olmo.grpo.minus.geoma.judge2.writing}{$-$1.5}
\setval{olmo.grpo.minus.geoma.judge2.writing-abs}{1.5}
\setval{olmo.grpo.minus.geoma.judge2.writingbench}{$-$0.1}
\setval{olmo.grpo.minus.geoma.judge2.writingbench-abs}{0.1}
\setval{olmo.grpo.minus.geoma.livecodebench}{+0.0}
\setval{olmo.grpo.minus.geoma.livecodebench-abs}{0.0}
\setval{olmo.grpo.minus.geoma.longbench-write}{$-$1.1}
\setval{olmo.grpo.minus.geoma.longbench-write-abs}{1.1}
\setval{olmo.grpo.minus.geoma.math}{+3.3}
\setval{olmo.grpo.minus.geoma.math-abs}{3.3}
\setval{olmo.grpo.minus.geoma.math500}{+1.8}
\setval{olmo.grpo.minus.geoma.math500-abs}{1.8}
\setval{olmo.grpo.minus.geoma.mbpp-plus}{+1.1}
\setval{olmo.grpo.minus.geoma.mbpp-plus-abs}{1.1}
\setval{olmo.grpo.minus.geoma.minerva}{+0.0}
\setval{olmo.grpo.minus.geoma.minerva-abs}{0.0}
\setval{olmo.grpo.minus.geoma.olympiad-bench}{+2.9}
\setval{olmo.grpo.minus.geoma.olympiad-bench-abs}{2.9}
\setval{olmo.grpo.minus.geoma.writing}{+0.1}
\setval{olmo.grpo.minus.geoma.writing-abs}{0.1}
\setval{olmo.grpo.minus.geoma.writingbench}{+1.5}
\setval{olmo.grpo.minus.geoma.writingbench-abs}{1.5}
\setval{olmo.grpo.minus.grpo-nodiv.aime24}{$-$2.3}
\setval{olmo.grpo.minus.grpo-nodiv.aime24-abs}{2.3}
\setval{olmo.grpo.minus.grpo-nodiv.aime25}{$-$2.9}
\setval{olmo.grpo.minus.grpo-nodiv.aime25-abs}{2.9}
\setval{olmo.grpo.minus.grpo-nodiv.amc23}{$-$2.2}
\setval{olmo.grpo.minus.grpo-nodiv.amc23-abs}{2.2}
\setval{olmo.grpo.minus.grpo-nodiv.avg}{$-$0.7}
\setval{olmo.grpo.minus.grpo-nodiv.avg-abs}{0.7}
\setval{olmo.grpo.minus.grpo-nodiv.code}{$-$0.3}
\setval{olmo.grpo.minus.grpo-nodiv.code-abs}{0.3}
\setval{olmo.grpo.minus.grpo-nodiv.creative-writing-v3}{+2.4}
\setval{olmo.grpo.minus.grpo-nodiv.creative-writing-v3-abs}{2.4}
\setval{olmo.grpo.minus.grpo-nodiv.guidebench}{+1.0}
\setval{olmo.grpo.minus.grpo-nodiv.guidebench-abs}{1.0}
\setval{olmo.grpo.minus.grpo-nodiv.humaneval-plus}{$-$0.1}
\setval{olmo.grpo.minus.grpo-nodiv.humaneval-plus-abs}{0.1}
\setval{olmo.grpo.minus.grpo-nodiv.if}{+0.4}
\setval{olmo.grpo.minus.grpo-nodiv.if-abs}{0.4}
\setval{olmo.grpo.minus.grpo-nodiv.ifbench}{+0.7}
\setval{olmo.grpo.minus.grpo-nodiv.ifbench-abs}{0.7}
\setval{olmo.grpo.minus.grpo-nodiv.ifeval}{$-$0.4}
\setval{olmo.grpo.minus.grpo-nodiv.ifeval-abs}{0.4}
\setval{olmo.grpo.minus.grpo-nodiv.judge2.avg}{$-$1.1}
\setval{olmo.grpo.minus.grpo-nodiv.judge2.avg-abs}{1.1}
\setval{olmo.grpo.minus.grpo-nodiv.judge2.creative-writing-v3}{$-$0.2}
\setval{olmo.grpo.minus.grpo-nodiv.judge2.creative-writing-v3-abs}{0.2}
\setval{olmo.grpo.minus.grpo-nodiv.judge2.longbench-write}{$-$4.0}
\setval{olmo.grpo.minus.grpo-nodiv.judge2.longbench-write-abs}{4.0}
\setval{olmo.grpo.minus.grpo-nodiv.judge2.writing}{$-$0.9}
\setval{olmo.grpo.minus.grpo-nodiv.judge2.writing-abs}{0.9}
\setval{olmo.grpo.minus.grpo-nodiv.judge2.writingbench}{+1.5}
\setval{olmo.grpo.minus.grpo-nodiv.judge2.writingbench-abs}{1.5}
\setval{olmo.grpo.minus.grpo-nodiv.livecodebench}{$-$1.3}
\setval{olmo.grpo.minus.grpo-nodiv.livecodebench-abs}{1.3}
\setval{olmo.grpo.minus.grpo-nodiv.longbench-write}{$-$3.3}
\setval{olmo.grpo.minus.grpo-nodiv.longbench-write-abs}{3.3}
\setval{olmo.grpo.minus.grpo-nodiv.math}{$-$2.4}
\setval{olmo.grpo.minus.grpo-nodiv.math-abs}{2.4}
\setval{olmo.grpo.minus.grpo-nodiv.math500}{$-$2.6}
\setval{olmo.grpo.minus.grpo-nodiv.math500-abs}{2.6}
\setval{olmo.grpo.minus.grpo-nodiv.mbpp-plus}{+0.4}
\setval{olmo.grpo.minus.grpo-nodiv.mbpp-plus-abs}{0.4}
\setval{olmo.grpo.minus.grpo-nodiv.minerva}{$-$1.9}
\setval{olmo.grpo.minus.grpo-nodiv.minerva-abs}{1.9}
\setval{olmo.grpo.minus.grpo-nodiv.olympiad-bench}{$-$2.8}
\setval{olmo.grpo.minus.grpo-nodiv.olympiad-bench-abs}{2.8}
\setval{olmo.grpo.minus.grpo-nodiv.writing}{+1.5}
\setval{olmo.grpo.minus.grpo-nodiv.writing-abs}{1.5}
\setval{olmo.grpo.minus.grpo-nodiv.writingbench}{+5.3}
\setval{olmo.grpo.minus.grpo-nodiv.writingbench-abs}{5.3}
\setval{olmo.grpo.minus.mt-grpo.aime24}{$-$7.1}
\setval{olmo.grpo.minus.mt-grpo.aime24-abs}{7.1}
\setval{olmo.grpo.minus.mt-grpo.aime25}{$-$3.4}
\setval{olmo.grpo.minus.mt-grpo.aime25-abs}{3.4}
\setval{olmo.grpo.minus.mt-grpo.amc23}{$-$6.6}
\setval{olmo.grpo.minus.mt-grpo.amc23-abs}{6.6}
\setval{olmo.grpo.minus.mt-grpo.avg}{$-$1.4}
\setval{olmo.grpo.minus.mt-grpo.avg-abs}{1.4}
\setval{olmo.grpo.minus.mt-grpo.code}{+0.7}
\setval{olmo.grpo.minus.mt-grpo.code-abs}{0.7}
\setval{olmo.grpo.minus.mt-grpo.creative-writing-v3}{$-$1.3}
\setval{olmo.grpo.minus.mt-grpo.creative-writing-v3-abs}{1.3}
\setval{olmo.grpo.minus.mt-grpo.guidebench}{+1.9}
\setval{olmo.grpo.minus.mt-grpo.guidebench-abs}{1.9}
\setval{olmo.grpo.minus.mt-grpo.humaneval-plus}{+1.6}
\setval{olmo.grpo.minus.mt-grpo.humaneval-plus-abs}{1.6}
\setval{olmo.grpo.minus.mt-grpo.if}{+0.8}
\setval{olmo.grpo.minus.mt-grpo.if-abs}{0.8}
\setval{olmo.grpo.minus.mt-grpo.ifbench}{+0.0}
\setval{olmo.grpo.minus.mt-grpo.ifbench-abs}{0.0}
\setval{olmo.grpo.minus.mt-grpo.ifeval}{+0.4}
\setval{olmo.grpo.minus.mt-grpo.ifeval-abs}{0.4}
\setval{olmo.grpo.minus.mt-grpo.judge2.avg}{$-$1.5}
\setval{olmo.grpo.minus.mt-grpo.judge2.avg-abs}{1.5}
\setval{olmo.grpo.minus.mt-grpo.judge2.creative-writing-v3}{$-$1.3}
\setval{olmo.grpo.minus.mt-grpo.judge2.creative-writing-v3-abs}{1.3}
\setval{olmo.grpo.minus.mt-grpo.judge2.longbench-write}{$-$3.3}
\setval{olmo.grpo.minus.mt-grpo.judge2.longbench-write-abs}{3.3}
\setval{olmo.grpo.minus.mt-grpo.judge2.writing}{$-$1.2}
\setval{olmo.grpo.minus.mt-grpo.judge2.writing-abs}{1.2}
\setval{olmo.grpo.minus.mt-grpo.judge2.writingbench}{+1.0}
\setval{olmo.grpo.minus.mt-grpo.judge2.writingbench-abs}{1.0}
\setval{olmo.grpo.minus.mt-grpo.livecodebench}{$-$0.3}
\setval{olmo.grpo.minus.mt-grpo.livecodebench-abs}{0.3}
\setval{olmo.grpo.minus.mt-grpo.longbench-write}{$-$2.2}
\setval{olmo.grpo.minus.mt-grpo.longbench-write-abs}{2.2}
\setval{olmo.grpo.minus.mt-grpo.math}{$-$3.9}
\setval{olmo.grpo.minus.mt-grpo.math-abs}{3.9}
\setval{olmo.grpo.minus.mt-grpo.math500}{$-$2.0}
\setval{olmo.grpo.minus.mt-grpo.math500-abs}{2.0}
\setval{olmo.grpo.minus.mt-grpo.mbpp-plus}{+0.8}
\setval{olmo.grpo.minus.mt-grpo.mbpp-plus-abs}{0.8}
\setval{olmo.grpo.minus.mt-grpo.minerva}{$-$2.0}
\setval{olmo.grpo.minus.mt-grpo.minerva-abs}{2.0}
\setval{olmo.grpo.minus.mt-grpo.olympiad-bench}{$-$2.4}
\setval{olmo.grpo.minus.mt-grpo.olympiad-bench-abs}{2.4}
\setval{olmo.grpo.minus.mt-grpo.writing}{$-$0.9}
\setval{olmo.grpo.minus.mt-grpo.writing-abs}{0.9}
\setval{olmo.grpo.minus.mt-grpo.writingbench}{+0.8}
\setval{olmo.grpo.minus.mt-grpo.writingbench-abs}{0.8}
\setval{olmo.grpo.minus.papo.aime24}{+7.3}
\setval{olmo.grpo.minus.papo.aime24-abs}{7.3}
\setval{olmo.grpo.minus.papo.aime25}{+4.4}
\setval{olmo.grpo.minus.papo.aime25-abs}{4.4}
\setval{olmo.grpo.minus.papo.amc23}{+11.5}
\setval{olmo.grpo.minus.papo.amc23-abs}{11.5}
\setval{olmo.grpo.minus.papo.avg}{+2.9}
\setval{olmo.grpo.minus.papo.avg-abs}{2.9}
\setval{olmo.grpo.minus.papo.code}{+1.2}
\setval{olmo.grpo.minus.papo.code-abs}{1.2}
\setval{olmo.grpo.minus.papo.creative-writing-v3}{+2.4}
\setval{olmo.grpo.minus.papo.creative-writing-v3-abs}{2.4}
\setval{olmo.grpo.minus.papo.guidebench}{$-$2.2}
\setval{olmo.grpo.minus.papo.guidebench-abs}{2.2}
\setval{olmo.grpo.minus.papo.humaneval-plus}{+2.6}
\setval{olmo.grpo.minus.papo.humaneval-plus-abs}{2.6}
\setval{olmo.grpo.minus.papo.if}{+0.1}
\setval{olmo.grpo.minus.papo.if-abs}{0.1}
\setval{olmo.grpo.minus.papo.ifbench}{+2.3}
\setval{olmo.grpo.minus.papo.ifbench-abs}{2.3}
\setval{olmo.grpo.minus.papo.ifeval}{+0.2}
\setval{olmo.grpo.minus.papo.ifeval-abs}{0.2}
\setval{olmo.grpo.minus.papo.judge2.avg}{+2.8}
\setval{olmo.grpo.minus.papo.judge2.avg-abs}{2.8}
\setval{olmo.grpo.minus.papo.judge2.creative-writing-v3}{+0.6}
\setval{olmo.grpo.minus.papo.judge2.creative-writing-v3-abs}{0.6}
\setval{olmo.grpo.minus.papo.judge2.longbench-write}{+1.5}
\setval{olmo.grpo.minus.papo.judge2.longbench-write-abs}{1.5}
\setval{olmo.grpo.minus.papo.judge2.writing}{+1.0}
\setval{olmo.grpo.minus.papo.judge2.writing-abs}{1.0}
\setval{olmo.grpo.minus.papo.judge2.writingbench}{+1.0}
\setval{olmo.grpo.minus.papo.judge2.writingbench-abs}{1.0}
\setval{olmo.grpo.minus.papo.livecodebench}{+0.5}
\setval{olmo.grpo.minus.papo.livecodebench-abs}{0.5}
\setval{olmo.grpo.minus.papo.longbench-write}{$-$0.5}
\setval{olmo.grpo.minus.papo.longbench-write-abs}{0.5}
\setval{olmo.grpo.minus.papo.math}{+5.8}
\setval{olmo.grpo.minus.papo.math-abs}{5.8}
\setval{olmo.grpo.minus.papo.math500}{+4.1}
\setval{olmo.grpo.minus.papo.math500-abs}{4.1}
\setval{olmo.grpo.minus.papo.mbpp-plus}{+0.7}
\setval{olmo.grpo.minus.papo.mbpp-plus-abs}{0.7}
\setval{olmo.grpo.minus.papo.minerva}{+1.4}
\setval{olmo.grpo.minus.papo.minerva-abs}{1.4}
\setval{olmo.grpo.minus.papo.olympiad-bench}{+5.9}
\setval{olmo.grpo.minus.papo.olympiad-bench-abs}{5.9}
\setval{olmo.grpo.minus.papo.writing}{+1.7}
\setval{olmo.grpo.minus.papo.writing-abs}{1.7}
\setval{olmo.grpo.minus.papo.writingbench}{+3.1}
\setval{olmo.grpo.minus.papo.writingbench-abs}{3.1}
\setval{olmo.grpo.minus.saw.aime24}{+3.3}
\setval{olmo.grpo.minus.saw.aime24-abs}{3.3}
\setval{olmo.grpo.minus.saw.aime25}{$-$0.5}
\setval{olmo.grpo.minus.saw.aime25-abs}{0.5}
\setval{olmo.grpo.minus.saw.amc23}{+4.0}
\setval{olmo.grpo.minus.saw.amc23-abs}{4.0}
\setval{olmo.grpo.minus.saw.avg}{+0.8}
\setval{olmo.grpo.minus.saw.avg-abs}{0.8}
\setval{olmo.grpo.minus.saw.code}{+0.6}
\setval{olmo.grpo.minus.saw.code-abs}{0.6}
\setval{olmo.grpo.minus.saw.creative-writing-v3}{+1.5}
\setval{olmo.grpo.minus.saw.creative-writing-v3-abs}{1.5}
\setval{olmo.grpo.minus.saw.guidebench}{+1.7}
\setval{olmo.grpo.minus.saw.guidebench-abs}{1.7}
\setval{olmo.grpo.minus.saw.humaneval-plus}{+1.0}
\setval{olmo.grpo.minus.saw.humaneval-plus-abs}{1.0}
\setval{olmo.grpo.minus.saw.if}{+0.0}
\setval{olmo.grpo.minus.saw.if-abs}{0.0}
\setval{olmo.grpo.minus.saw.ifbench}{$-$0.7}
\setval{olmo.grpo.minus.saw.ifbench-abs}{0.7}
\setval{olmo.grpo.minus.saw.ifeval}{$-$1.2}
\setval{olmo.grpo.minus.saw.ifeval-abs}{1.2}
\setval{olmo.grpo.minus.saw.judge2.avg}{+0.5}
\setval{olmo.grpo.minus.saw.judge2.avg-abs}{0.5}
\setval{olmo.grpo.minus.saw.judge2.creative-writing-v3}{$-$2.4}
\setval{olmo.grpo.minus.saw.judge2.creative-writing-v3-abs}{2.4}
\setval{olmo.grpo.minus.saw.judge2.longbench-write}{$-$0.3}
\setval{olmo.grpo.minus.saw.judge2.longbench-write-abs}{0.3}
\setval{olmo.grpo.minus.saw.judge2.writing}{$-$0.9}
\setval{olmo.grpo.minus.saw.judge2.writing-abs}{0.9}
\setval{olmo.grpo.minus.saw.judge2.writingbench}{+0.1}
\setval{olmo.grpo.minus.saw.judge2.writingbench-abs}{0.1}
\setval{olmo.grpo.minus.saw.livecodebench}{+0.5}
\setval{olmo.grpo.minus.saw.livecodebench-abs}{0.5}
\setval{olmo.grpo.minus.saw.longbench-write}{$-$0.7}
\setval{olmo.grpo.minus.saw.longbench-write-abs}{0.7}
\setval{olmo.grpo.minus.saw.math}{+1.5}
\setval{olmo.grpo.minus.saw.math-abs}{1.5}
\setval{olmo.grpo.minus.saw.math500}{+0.5}
\setval{olmo.grpo.minus.saw.math500-abs}{0.5}
\setval{olmo.grpo.minus.saw.mbpp-plus}{+0.4}
\setval{olmo.grpo.minus.saw.mbpp-plus-abs}{0.4}
\setval{olmo.grpo.minus.saw.minerva}{$-$0.4}
\setval{olmo.grpo.minus.saw.minerva-abs}{0.4}
\setval{olmo.grpo.minus.saw.olympiad-bench}{+1.8}
\setval{olmo.grpo.minus.saw.olympiad-bench-abs}{1.8}
\setval{olmo.grpo.minus.saw.writing}{+0.6}
\setval{olmo.grpo.minus.saw.writing-abs}{0.6}
\setval{olmo.grpo.minus.saw.writingbench}{+1.1}
\setval{olmo.grpo.minus.saw.writingbench-abs}{1.1}
\setval{olmo.grpo.passing-at-ceiling.code.early}{0.84}
\setval{olmo.grpo.passing-at-ceiling.code.late}{0.84}
\setval{olmo.grpo.passing-at-ceiling.code.run}{0.83}
\setval{olmo.grpo.passing-at-ceiling.if.early}{0.49}
\setval{olmo.grpo.passing-at-ceiling.if.late}{0.57}
\setval{olmo.grpo.passing-at-ceiling.if.run}{0.53}
\setval{olmo.grpo.passing-at-ceiling.math.early}{1.00}
\setval{olmo.grpo.passing-at-ceiling.math.late}{1.00}
\setval{olmo.grpo.passing-at-ceiling.math.run}{1.00}
\setval{olmo.grpo.passing-at-ceiling.writing.early}{0.00}
\setval{olmo.grpo.passing-at-ceiling.writing.late}{0.00}
\setval{olmo.grpo.passing-at-ceiling.writing.run}{0.00}
\setval{olmo.grpo.primary-level.code.early}{0.59}
\setval{olmo.grpo.primary-level.code.late}{0.82}
\setval{olmo.grpo.primary-level.code.run}{0.75}
\setval{olmo.grpo.primary-level.if.early}{0.63}
\setval{olmo.grpo.primary-level.if.late}{0.74}
\setval{olmo.grpo.primary-level.if.run}{0.69}
\setval{olmo.grpo.primary-level.math.early}{0.34}
\setval{olmo.grpo.primary-level.math.late}{0.66}
\setval{olmo.grpo.primary-level.math.run}{0.57}
\setval{olmo.grpo.primary-level.writing.early}{0.49}
\setval{olmo.grpo.primary-level.writing.late}{0.70}
\setval{olmo.grpo.primary-level.writing.run}{0.64}
\setval{olmo.mt-grpo.aux-change.code}{+0.12}
\setval{olmo.mt-grpo.aux-change.if}{+0.11}
\setval{olmo.mt-grpo.aux-change.math}{$-$0.11}
\setval{olmo.mt-grpo.aux-change.writing}{+0.02}
\setval{olmo.mt-grpo.aux-change.writing-rm}{+0.28}
\setval{olmo.mt-grpo.aux-full.code.early}{0.72}
\setval{olmo.mt-grpo.aux-full.code.late}{0.82}
\setval{olmo.mt-grpo.aux-full.code.run}{0.79}
\setval{olmo.mt-grpo.aux-full.if.early}{0.57}
\setval{olmo.mt-grpo.aux-full.if.late}{0.88}
\setval{olmo.mt-grpo.aux-full.if.run}{0.76}
\setval{olmo.mt-grpo.aux-full.math.early}{0.47}
\setval{olmo.mt-grpo.aux-full.math.late}{0.58}
\setval{olmo.mt-grpo.aux-full.math.run}{0.50}
\setval{olmo.mt-grpo.aux-full.writing-rm.early}{0.01}
\setval{olmo.mt-grpo.aux-full.writing-rm.late}{0.25}
\setval{olmo.mt-grpo.aux-full.writing-rm.run}{0.09}
\setval{olmo.mt-grpo.aux-full.writing.early}{0.59}
\setval{olmo.mt-grpo.aux-full.writing.late}{0.65}
\setval{olmo.mt-grpo.aux-full.writing.run}{0.60}
\setval{olmo.mt-grpo.aux-level.code.early}{0.90}
\setval{olmo.mt-grpo.aux-level.code.late}{0.93}
\setval{olmo.mt-grpo.aux-level.code.run}{0.92}
\setval{olmo.mt-grpo.aux-level.if.early}{0.90}
\setval{olmo.mt-grpo.aux-level.if.late}{0.98}
\setval{olmo.mt-grpo.aux-level.if.run}{0.95}
\setval{olmo.mt-grpo.aux-level.math.early}{0.86}
\setval{olmo.mt-grpo.aux-level.math.late}{0.79}
\setval{olmo.mt-grpo.aux-level.math.run}{0.75}
\setval{olmo.mt-grpo.aux-level.writing-rm.early}{0.58}
\setval{olmo.mt-grpo.aux-level.writing-rm.late}{0.85}
\setval{olmo.mt-grpo.aux-level.writing-rm.run}{0.73}
\setval{olmo.mt-grpo.aux-level.writing.early}{0.95}
\setval{olmo.mt-grpo.aux-level.writing.late}{0.97}
\setval{olmo.mt-grpo.aux-level.writing.run}{0.96}
\setval{olmo.mt-grpo.bench.aime24}{36.6}
\setval{olmo.mt-grpo.bench.aime25}{28.1}
\setval{olmo.mt-grpo.bench.amc23}{77.4}
\setval{olmo.mt-grpo.bench.creative-writing-v3}{50.7}
\setval{olmo.mt-grpo.bench.guidebench}{54.5}
\setval{olmo.mt-grpo.bench.humaneval-plus}{75.7}
\setval{olmo.mt-grpo.bench.ifbench}{32.0}
\setval{olmo.mt-grpo.bench.ifeval}{87.1}
\setval{olmo.mt-grpo.bench.livecodebench}{22.6}
\setval{olmo.mt-grpo.bench.longbench-write}{66.4}
\setval{olmo.mt-grpo.bench.math500}{86.9}
\setval{olmo.mt-grpo.bench.mbpp-plus}{65.0}
\setval{olmo.mt-grpo.bench.minerva}{42.0}
\setval{olmo.mt-grpo.bench.olympiad-bench}{58.0}
\setval{olmo.mt-grpo.bench.writingbench}{67.7}
\setval{olmo.mt-grpo.eval.avg}{56.7}
\setval{olmo.mt-grpo.eval.code}{54.4}
\setval{olmo.mt-grpo.eval.if}{57.9}
\setval{olmo.mt-grpo.eval.math}{54.8}
\setval{olmo.mt-grpo.eval.writing}{61.6}
\setval{olmo.mt-grpo.judge2.avg}{54.2}
\setval{olmo.mt-grpo.judge2.creative-writing-v3}{39.9}
\setval{olmo.mt-grpo.judge2.longbench-write}{60.4}
\setval{olmo.mt-grpo.judge2.writing}{49.3}
\setval{olmo.mt-grpo.judge2.writingbench}{47.5}
\setval{olmo.mt-grpo.last-step}{300}
\setval{olmo.mt-grpo.minus.base.aime24}{+28.0}
\setval{olmo.mt-grpo.minus.base.aime24-abs}{28.0}
\setval{olmo.mt-grpo.minus.base.aime25}{+20.7}
\setval{olmo.mt-grpo.minus.base.aime25-abs}{20.7}
\setval{olmo.mt-grpo.minus.base.amc23}{+32.9}
\setval{olmo.mt-grpo.minus.base.amc23-abs}{32.9}
\setval{olmo.mt-grpo.minus.base.avg}{+12.6}
\setval{olmo.mt-grpo.minus.base.avg-abs}{12.6}
\setval{olmo.mt-grpo.minus.base.code}{+5.3}
\setval{olmo.mt-grpo.minus.base.code-abs}{5.3}
\setval{olmo.mt-grpo.minus.base.creative-writing-v3}{+12.8}
\setval{olmo.mt-grpo.minus.base.creative-writing-v3-abs}{12.8}
\setval{olmo.mt-grpo.minus.base.guidebench}{$-$6.9}
\setval{olmo.mt-grpo.minus.base.guidebench-abs}{6.9}
\setval{olmo.mt-grpo.minus.base.humaneval-plus}{+6.5}
\setval{olmo.mt-grpo.minus.base.humaneval-plus-abs}{6.5}
\setval{olmo.mt-grpo.minus.base.if}{+0.4}
\setval{olmo.mt-grpo.minus.base.if-abs}{0.4}
\setval{olmo.mt-grpo.minus.base.ifbench}{+5.3}
\setval{olmo.mt-grpo.minus.base.ifbench-abs}{5.3}
\setval{olmo.mt-grpo.minus.base.ifeval}{+2.6}
\setval{olmo.mt-grpo.minus.base.ifeval-abs}{2.6}
\setval{olmo.mt-grpo.minus.base.judge2.avg}{+11.3}
\setval{olmo.mt-grpo.minus.base.judge2.avg-abs}{11.3}
\setval{olmo.mt-grpo.minus.base.judge2.creative-writing-v3}{+8.2}
\setval{olmo.mt-grpo.minus.base.judge2.creative-writing-v3-abs}{8.2}
\setval{olmo.mt-grpo.minus.base.judge2.longbench-write}{+7.2}
\setval{olmo.mt-grpo.minus.base.judge2.longbench-write-abs}{7.2}
\setval{olmo.mt-grpo.minus.base.judge2.writing}{+8.4}
\setval{olmo.mt-grpo.minus.base.judge2.writing-abs}{8.4}
\setval{olmo.mt-grpo.minus.base.judge2.writingbench}{+9.8}
\setval{olmo.mt-grpo.minus.base.judge2.writingbench-abs}{9.8}
\setval{olmo.mt-grpo.minus.base.livecodebench}{+5.8}
\setval{olmo.mt-grpo.minus.base.livecodebench-abs}{5.8}
\setval{olmo.mt-grpo.minus.base.longbench-write}{+8.9}
\setval{olmo.mt-grpo.minus.base.longbench-write-abs}{8.9}
\setval{olmo.mt-grpo.minus.base.math}{+21.1}
\setval{olmo.mt-grpo.minus.base.math-abs}{21.1}
\setval{olmo.mt-grpo.minus.base.math500}{+16.8}
\setval{olmo.mt-grpo.minus.base.math500-abs}{16.8}
\setval{olmo.mt-grpo.minus.base.mbpp-plus}{+3.7}
\setval{olmo.mt-grpo.minus.base.mbpp-plus-abs}{3.7}
\setval{olmo.mt-grpo.minus.base.minerva}{+6.7}
\setval{olmo.mt-grpo.minus.base.minerva-abs}{6.7}
\setval{olmo.mt-grpo.minus.base.olympiad-bench}{+21.8}
\setval{olmo.mt-grpo.minus.base.olympiad-bench-abs}{21.8}
\setval{olmo.mt-grpo.minus.base.writing}{+14.9}
\setval{olmo.mt-grpo.minus.base.writing-abs}{14.9}
\setval{olmo.mt-grpo.minus.base.writingbench}{+23.0}
\setval{olmo.mt-grpo.minus.base.writingbench-abs}{23.0}
\setval{olmo.mt-grpo.minus.concur-free.aime24}{+9.8}
\setval{olmo.mt-grpo.minus.concur-free.aime24-abs}{9.8}
\setval{olmo.mt-grpo.minus.concur-free.aime25}{+6.0}
\setval{olmo.mt-grpo.minus.concur-free.aime25-abs}{6.0}
\setval{olmo.mt-grpo.minus.concur-free.amc23}{+9.9}
\setval{olmo.mt-grpo.minus.concur-free.amc23-abs}{9.9}
\setval{olmo.mt-grpo.minus.concur-free.avg}{+1.3}
\setval{olmo.mt-grpo.minus.concur-free.avg-abs}{1.3}
\setval{olmo.mt-grpo.minus.concur-free.code}{$-$1.2}
\setval{olmo.mt-grpo.minus.concur-free.code-abs}{1.2}
\setval{olmo.mt-grpo.minus.concur-free.creative-writing-v3}{+0.9}
\setval{olmo.mt-grpo.minus.concur-free.creative-writing-v3-abs}{0.9}
\setval{olmo.mt-grpo.minus.concur-free.guidebench}{$-$5.7}
\setval{olmo.mt-grpo.minus.concur-free.guidebench-abs}{5.7}
\setval{olmo.mt-grpo.minus.concur-free.humaneval-plus}{$-$0.9}
\setval{olmo.mt-grpo.minus.concur-free.humaneval-plus-abs}{0.9}
\setval{olmo.mt-grpo.minus.concur-free.if}{$-$1.9}
\setval{olmo.mt-grpo.minus.concur-free.if-abs}{1.9}
\setval{olmo.mt-grpo.minus.concur-free.ifbench}{+0.3}
\setval{olmo.mt-grpo.minus.concur-free.ifbench-abs}{0.3}
\setval{olmo.mt-grpo.minus.concur-free.ifeval}{$-$0.2}
\setval{olmo.mt-grpo.minus.concur-free.ifeval-abs}{0.2}
\setval{olmo.mt-grpo.minus.concur-free.judge2.avg}{+1.6}
\setval{olmo.mt-grpo.minus.concur-free.judge2.avg-abs}{1.6}
\setval{olmo.mt-grpo.minus.concur-free.judge2.creative-writing-v3}{+2.3}
\setval{olmo.mt-grpo.minus.concur-free.judge2.creative-writing-v3-abs}{2.3}
\setval{olmo.mt-grpo.minus.concur-free.judge2.longbench-write}{$-$3.5}
\setval{olmo.mt-grpo.minus.concur-free.judge2.longbench-write-abs}{3.5}
\setval{olmo.mt-grpo.minus.concur-free.judge2.writing}{$-$0.6}
\setval{olmo.mt-grpo.minus.concur-free.judge2.writing-abs}{0.6}
\setval{olmo.mt-grpo.minus.concur-free.judge2.writingbench}{$-$0.5}
\setval{olmo.mt-grpo.minus.concur-free.judge2.writingbench-abs}{0.5}
\setval{olmo.mt-grpo.minus.concur-free.livecodebench}{$-$3.0}
\setval{olmo.mt-grpo.minus.concur-free.livecodebench-abs}{3.0}
\setval{olmo.mt-grpo.minus.concur-free.longbench-write}{$-$5.7}
\setval{olmo.mt-grpo.minus.concur-free.longbench-write-abs}{5.7}
\setval{olmo.mt-grpo.minus.concur-free.math}{+5.8}
\setval{olmo.mt-grpo.minus.concur-free.math-abs}{5.8}
\setval{olmo.mt-grpo.minus.concur-free.math500}{+2.6}
\setval{olmo.mt-grpo.minus.concur-free.math500-abs}{2.6}
\setval{olmo.mt-grpo.minus.concur-free.mbpp-plus}{+0.5}
\setval{olmo.mt-grpo.minus.concur-free.mbpp-plus-abs}{0.5}
\setval{olmo.mt-grpo.minus.concur-free.minerva}{+2.0}
\setval{olmo.mt-grpo.minus.concur-free.minerva-abs}{2.0}
\setval{olmo.mt-grpo.minus.concur-free.olympiad-bench}{+4.8}
\setval{olmo.mt-grpo.minus.concur-free.olympiad-bench-abs}{4.8}
\setval{olmo.mt-grpo.minus.concur-free.writing}{$-$2.4}
\setval{olmo.mt-grpo.minus.concur-free.writing-abs}{2.4}
\setval{olmo.mt-grpo.minus.concur-free.writingbench}{$-$2.5}
\setval{olmo.mt-grpo.minus.concur-free.writingbench-abs}{2.5}
\setval{olmo.mt-grpo.minus.concur-tied.aime24}{+2.1}
\setval{olmo.mt-grpo.minus.concur-tied.aime24-abs}{2.1}
\setval{olmo.mt-grpo.minus.concur-tied.aime25}{$-$1.8}
\setval{olmo.mt-grpo.minus.concur-tied.aime25-abs}{1.8}
\setval{olmo.mt-grpo.minus.concur-tied.amc23}{$-$2.4}
\setval{olmo.mt-grpo.minus.concur-tied.amc23-abs}{2.4}
\setval{olmo.mt-grpo.minus.concur-tied.avg}{$-$2.4}
\setval{olmo.mt-grpo.minus.concur-tied.avg-abs}{2.4}
\setval{olmo.mt-grpo.minus.concur-tied.code}{$-$6.7}
\setval{olmo.mt-grpo.minus.concur-tied.code-abs}{6.7}
\setval{olmo.mt-grpo.minus.concur-tied.creative-writing-v3}{+1.7}
\setval{olmo.mt-grpo.minus.concur-tied.creative-writing-v3-abs}{1.7}
\setval{olmo.mt-grpo.minus.concur-tied.guidebench}{$-$4.0}
\setval{olmo.mt-grpo.minus.concur-tied.guidebench-abs}{4.0}
\setval{olmo.mt-grpo.minus.concur-tied.humaneval-plus}{$-$7.6}
\setval{olmo.mt-grpo.minus.concur-tied.humaneval-plus-abs}{7.6}
\setval{olmo.mt-grpo.minus.concur-tied.if}{$-$2.1}
\setval{olmo.mt-grpo.minus.concur-tied.if-abs}{2.1}
\setval{olmo.mt-grpo.minus.concur-tied.ifbench}{$-$1.7}
\setval{olmo.mt-grpo.minus.concur-tied.ifbench-abs}{1.7}
\setval{olmo.mt-grpo.minus.concur-tied.ifeval}{$-$0.6}
\setval{olmo.mt-grpo.minus.concur-tied.ifeval-abs}{0.6}
\setval{olmo.mt-grpo.minus.concur-tied.judge2.avg}{$-$2.1}
\setval{olmo.mt-grpo.minus.concur-tied.judge2.avg-abs}{2.1}
\setval{olmo.mt-grpo.minus.concur-tied.judge2.creative-writing-v3}{+2.7}
\setval{olmo.mt-grpo.minus.concur-tied.judge2.creative-writing-v3-abs}{2.7}
\setval{olmo.mt-grpo.minus.concur-tied.judge2.longbench-write}{$-$0.1}
\setval{olmo.mt-grpo.minus.concur-tied.judge2.longbench-write-abs}{0.1}
\setval{olmo.mt-grpo.minus.concur-tied.judge2.writing}{+1.5}
\setval{olmo.mt-grpo.minus.concur-tied.judge2.writing-abs}{1.5}
\setval{olmo.mt-grpo.minus.concur-tied.judge2.writingbench}{+1.9}
\setval{olmo.mt-grpo.minus.concur-tied.judge2.writingbench-abs}{1.9}
\setval{olmo.mt-grpo.minus.concur-tied.livecodebench}{$-$6.4}
\setval{olmo.mt-grpo.minus.concur-tied.livecodebench-abs}{6.4}
\setval{olmo.mt-grpo.minus.concur-tied.longbench-write}{$-$2.9}
\setval{olmo.mt-grpo.minus.concur-tied.longbench-write-abs}{2.9}
\setval{olmo.mt-grpo.minus.concur-tied.math}{$-$1.7}
\setval{olmo.mt-grpo.minus.concur-tied.math-abs}{1.7}
\setval{olmo.mt-grpo.minus.concur-tied.math500}{$-$2.5}
\setval{olmo.mt-grpo.minus.concur-tied.math500-abs}{2.5}
\setval{olmo.mt-grpo.minus.concur-tied.mbpp-plus}{$-$6.1}
\setval{olmo.mt-grpo.minus.concur-tied.mbpp-plus-abs}{6.1}
\setval{olmo.mt-grpo.minus.concur-tied.minerva}{$-$2.1}
\setval{olmo.mt-grpo.minus.concur-tied.minerva-abs}{2.1}
\setval{olmo.mt-grpo.minus.concur-tied.olympiad-bench}{$-$3.5}
\setval{olmo.mt-grpo.minus.concur-tied.olympiad-bench-abs}{3.5}
\setval{olmo.mt-grpo.minus.concur-tied.writing}{+0.1}
\setval{olmo.mt-grpo.minus.concur-tied.writing-abs}{0.1}
\setval{olmo.mt-grpo.minus.concur-tied.writingbench}{+1.6}
\setval{olmo.mt-grpo.minus.concur-tied.writingbench-abs}{1.6}
\setval{olmo.mt-grpo.minus.concur.aime24}{+0.2}
\setval{olmo.mt-grpo.minus.concur.aime24-abs}{0.2}
\setval{olmo.mt-grpo.minus.concur.aime25}{$-$4.1}
\setval{olmo.mt-grpo.minus.concur.aime25-abs}{4.1}
\setval{olmo.mt-grpo.minus.concur.amc23}{$-$6.1}
\setval{olmo.mt-grpo.minus.concur.amc23-abs}{6.1}
\setval{olmo.mt-grpo.minus.concur.avg}{$-$2.6}
\setval{olmo.mt-grpo.minus.concur.avg-abs}{2.6}
\setval{olmo.mt-grpo.minus.concur.code}{$-$5.8}
\setval{olmo.mt-grpo.minus.concur.code-abs}{5.8}
\setval{olmo.mt-grpo.minus.concur.creative-writing-v3}{+1.5}
\setval{olmo.mt-grpo.minus.concur.creative-writing-v3-abs}{1.5}
\setval{olmo.mt-grpo.minus.concur.guidebench}{$-$2.4}
\setval{olmo.mt-grpo.minus.concur.guidebench-abs}{2.4}
\setval{olmo.mt-grpo.minus.concur.humaneval-plus}{$-$7.0}
\setval{olmo.mt-grpo.minus.concur.humaneval-plus-abs}{7.0}
\setval{olmo.mt-grpo.minus.concur.if}{$-$1.0}
\setval{olmo.mt-grpo.minus.concur.if-abs}{1.0}
\setval{olmo.mt-grpo.minus.concur.ifbench}{+0.0}
\setval{olmo.mt-grpo.minus.concur.ifbench-abs}{0.0}
\setval{olmo.mt-grpo.minus.concur.ifeval}{$-$0.7}
\setval{olmo.mt-grpo.minus.concur.ifeval-abs}{0.7}
\setval{olmo.mt-grpo.minus.concur.judge2.avg}{$-$2.3}
\setval{olmo.mt-grpo.minus.concur.judge2.avg-abs}{2.3}
\setval{olmo.mt-grpo.minus.concur.judge2.creative-writing-v3}{+1.4}
\setval{olmo.mt-grpo.minus.concur.judge2.creative-writing-v3-abs}{1.4}
\setval{olmo.mt-grpo.minus.concur.judge2.longbench-write}{+2.1}
\setval{olmo.mt-grpo.minus.concur.judge2.longbench-write-abs}{2.1}
\setval{olmo.mt-grpo.minus.concur.judge2.writing}{+2.0}
\setval{olmo.mt-grpo.minus.concur.judge2.writing-abs}{2.0}
\setval{olmo.mt-grpo.minus.concur.judge2.writingbench}{+2.6}
\setval{olmo.mt-grpo.minus.concur.judge2.writingbench-abs}{2.6}
\setval{olmo.mt-grpo.minus.concur.livecodebench}{$-$7.7}
\setval{olmo.mt-grpo.minus.concur.livecodebench-abs}{7.7}
\setval{olmo.mt-grpo.minus.concur.longbench-write}{$-$2.1}
\setval{olmo.mt-grpo.minus.concur.longbench-write-abs}{2.1}
\setval{olmo.mt-grpo.minus.concur.math}{$-$3.5}
\setval{olmo.mt-grpo.minus.concur.math-abs}{3.5}
\setval{olmo.mt-grpo.minus.concur.math500}{$-$3.6}
\setval{olmo.mt-grpo.minus.concur.math500-abs}{3.6}
\setval{olmo.mt-grpo.minus.concur.mbpp-plus}{$-$2.6}
\setval{olmo.mt-grpo.minus.concur.mbpp-plus-abs}{2.6}
\setval{olmo.mt-grpo.minus.concur.minerva}{$-$2.8}
\setval{olmo.mt-grpo.minus.concur.minerva-abs}{2.8}
\setval{olmo.mt-grpo.minus.concur.olympiad-bench}{$-$4.4}
\setval{olmo.mt-grpo.minus.concur.olympiad-bench-abs}{4.4}
\setval{olmo.mt-grpo.minus.concur.writing}{+0.5}
\setval{olmo.mt-grpo.minus.concur.writing-abs}{0.5}
\setval{olmo.mt-grpo.minus.concur.writingbench}{+2.2}
\setval{olmo.mt-grpo.minus.concur.writingbench-abs}{2.2}
\setval{olmo.mt-grpo.minus.gd2po.aime24}{+8.2}
\setval{olmo.mt-grpo.minus.gd2po.aime24-abs}{8.2}
\setval{olmo.mt-grpo.minus.gd2po.aime25}{+3.9}
\setval{olmo.mt-grpo.minus.gd2po.aime25-abs}{3.9}
\setval{olmo.mt-grpo.minus.gd2po.amc23}{+5.2}
\setval{olmo.mt-grpo.minus.gd2po.amc23-abs}{5.2}
\setval{olmo.mt-grpo.minus.gd2po.avg}{+1.1}
\setval{olmo.mt-grpo.minus.gd2po.avg-abs}{1.1}
\setval{olmo.mt-grpo.minus.gd2po.code}{$-$0.5}
\setval{olmo.mt-grpo.minus.gd2po.code-abs}{0.5}
\setval{olmo.mt-grpo.minus.gd2po.creative-writing-v3}{+2.4}
\setval{olmo.mt-grpo.minus.gd2po.creative-writing-v3-abs}{2.4}
\setval{olmo.mt-grpo.minus.gd2po.guidebench}{$-$6.6}
\setval{olmo.mt-grpo.minus.gd2po.guidebench-abs}{6.6}
\setval{olmo.mt-grpo.minus.gd2po.humaneval-plus}{$-$1.0}
\setval{olmo.mt-grpo.minus.gd2po.humaneval-plus-abs}{1.0}
\setval{olmo.mt-grpo.minus.gd2po.if}{$-$2.5}
\setval{olmo.mt-grpo.minus.gd2po.if-abs}{2.5}
\setval{olmo.mt-grpo.minus.gd2po.ifbench}{$-$0.3}
\setval{olmo.mt-grpo.minus.gd2po.ifbench-abs}{0.3}
\setval{olmo.mt-grpo.minus.gd2po.ifeval}{$-$0.6}
\setval{olmo.mt-grpo.minus.gd2po.ifeval-abs}{0.6}
\setval{olmo.mt-grpo.minus.gd2po.judge2.avg}{+1.0}
\setval{olmo.mt-grpo.minus.gd2po.judge2.avg-abs}{1.0}
\setval{olmo.mt-grpo.minus.gd2po.judge2.creative-writing-v3}{+2.0}
\setval{olmo.mt-grpo.minus.gd2po.judge2.creative-writing-v3-abs}{2.0}
\setval{olmo.mt-grpo.minus.gd2po.judge2.longbench-write}{+0.5}
\setval{olmo.mt-grpo.minus.gd2po.judge2.longbench-write-abs}{0.5}
\setval{olmo.mt-grpo.minus.gd2po.judge2.writing}{+0.6}
\setval{olmo.mt-grpo.minus.gd2po.judge2.writing-abs}{0.6}
\setval{olmo.mt-grpo.minus.gd2po.judge2.writingbench}{$-$0.8}
\setval{olmo.mt-grpo.minus.gd2po.judge2.writingbench-abs}{0.8}
\setval{olmo.mt-grpo.minus.gd2po.livecodebench}{$-$0.6}
\setval{olmo.mt-grpo.minus.gd2po.livecodebench-abs}{0.6}
\setval{olmo.mt-grpo.minus.gd2po.longbench-write}{$-$1.8}
\setval{olmo.mt-grpo.minus.gd2po.longbench-write-abs}{1.8}
\setval{olmo.mt-grpo.minus.gd2po.math}{+3.7}
\setval{olmo.mt-grpo.minus.gd2po.math-abs}{3.7}
\setval{olmo.mt-grpo.minus.gd2po.math500}{+1.2}
\setval{olmo.mt-grpo.minus.gd2po.math500-abs}{1.2}
\setval{olmo.mt-grpo.minus.gd2po.mbpp-plus}{+0.2}
\setval{olmo.mt-grpo.minus.gd2po.mbpp-plus-abs}{0.2}
\setval{olmo.mt-grpo.minus.gd2po.minerva}{+1.8}
\setval{olmo.mt-grpo.minus.gd2po.minerva-abs}{1.8}
\setval{olmo.mt-grpo.minus.gd2po.olympiad-bench}{+2.0}
\setval{olmo.mt-grpo.minus.gd2po.olympiad-bench-abs}{2.0}
\setval{olmo.mt-grpo.minus.gd2po.writing}{+0.8}
\setval{olmo.mt-grpo.minus.gd2po.writing-abs}{0.8}
\setval{olmo.mt-grpo.minus.gd2po.writingbench}{+2.0}
\setval{olmo.mt-grpo.minus.gd2po.writingbench-abs}{2.0}
\setval{olmo.mt-grpo.minus.gdpo-noaux.aime24}{$-$1.8}
\setval{olmo.mt-grpo.minus.gdpo-noaux.aime24-abs}{1.8}
\setval{olmo.mt-grpo.minus.gdpo-noaux.aime25}{$-$5.5}
\setval{olmo.mt-grpo.minus.gdpo-noaux.aime25-abs}{5.5}
\setval{olmo.mt-grpo.minus.gdpo-noaux.amc23}{$-$6.5}
\setval{olmo.mt-grpo.minus.gdpo-noaux.amc23-abs}{6.5}
\setval{olmo.mt-grpo.minus.gdpo-noaux.avg}{$-$1.1}
\setval{olmo.mt-grpo.minus.gdpo-noaux.avg-abs}{1.1}
\setval{olmo.mt-grpo.minus.gdpo-noaux.code}{$-$8.0}
\setval{olmo.mt-grpo.minus.gdpo-noaux.code-abs}{8.0}
\setval{olmo.mt-grpo.minus.gdpo-noaux.creative-writing-v3}{+12.5}
\setval{olmo.mt-grpo.minus.gdpo-noaux.creative-writing-v3-abs}{12.5}
\setval{olmo.mt-grpo.minus.gdpo-noaux.guidebench}{$-$1.2}
\setval{olmo.mt-grpo.minus.gdpo-noaux.guidebench-abs}{1.2}
\setval{olmo.mt-grpo.minus.gdpo-noaux.humaneval-plus}{$-$9.5}
\setval{olmo.mt-grpo.minus.gdpo-noaux.humaneval-plus-abs}{9.5}
\setval{olmo.mt-grpo.minus.gdpo-noaux.if}{$-$0.1}
\setval{olmo.mt-grpo.minus.gdpo-noaux.if-abs}{0.1}
\setval{olmo.mt-grpo.minus.gdpo-noaux.ifbench}{$-$1.3}
\setval{olmo.mt-grpo.minus.gdpo-noaux.ifbench-abs}{1.3}
\setval{olmo.mt-grpo.minus.gdpo-noaux.ifeval}{+2.3}
\setval{olmo.mt-grpo.minus.gdpo-noaux.ifeval-abs}{2.3}
\setval{olmo.mt-grpo.minus.gdpo-noaux.judge2.avg}{$-$1.4}
\setval{olmo.mt-grpo.minus.gdpo-noaux.judge2.avg-abs}{1.4}
\setval{olmo.mt-grpo.minus.gdpo-noaux.judge2.creative-writing-v3}{+8.9}
\setval{olmo.mt-grpo.minus.gdpo-noaux.judge2.creative-writing-v3-abs}{8.9}
\setval{olmo.mt-grpo.minus.gdpo-noaux.judge2.longbench-write}{+13.8}
\setval{olmo.mt-grpo.minus.gdpo-noaux.judge2.longbench-write-abs}{13.8}
\setval{olmo.mt-grpo.minus.gdpo-noaux.judge2.writing}{+10.2}
\setval{olmo.mt-grpo.minus.gdpo-noaux.judge2.writing-abs}{10.2}
\setval{olmo.mt-grpo.minus.gdpo-noaux.judge2.writingbench}{+7.8}
\setval{olmo.mt-grpo.minus.gdpo-noaux.judge2.writingbench-abs}{7.8}
\setval{olmo.mt-grpo.minus.gdpo-noaux.livecodebench}{$-$8.9}
\setval{olmo.mt-grpo.minus.gdpo-noaux.livecodebench-abs}{8.9}
\setval{olmo.mt-grpo.minus.gdpo-noaux.longbench-write}{+10.4}
\setval{olmo.mt-grpo.minus.gdpo-noaux.longbench-write-abs}{10.4}
\setval{olmo.mt-grpo.minus.gdpo-noaux.math}{$-$4.6}
\setval{olmo.mt-grpo.minus.gdpo-noaux.math-abs}{4.6}
\setval{olmo.mt-grpo.minus.gdpo-noaux.math500}{$-$4.0}
\setval{olmo.mt-grpo.minus.gdpo-noaux.math500-abs}{4.0}
\setval{olmo.mt-grpo.minus.gdpo-noaux.mbpp-plus}{$-$5.5}
\setval{olmo.mt-grpo.minus.gdpo-noaux.mbpp-plus-abs}{5.5}
\setval{olmo.mt-grpo.minus.gdpo-noaux.minerva}{$-$4.2}
\setval{olmo.mt-grpo.minus.gdpo-noaux.minerva-abs}{4.2}
\setval{olmo.mt-grpo.minus.gdpo-noaux.olympiad-bench}{$-$5.7}
\setval{olmo.mt-grpo.minus.gdpo-noaux.olympiad-bench-abs}{5.7}
\setval{olmo.mt-grpo.minus.gdpo-noaux.writing}{+11.7}
\setval{olmo.mt-grpo.minus.gdpo-noaux.writing-abs}{11.7}
\setval{olmo.mt-grpo.minus.gdpo-noaux.writingbench}{+12.2}
\setval{olmo.mt-grpo.minus.gdpo-noaux.writingbench-abs}{12.2}
\setval{olmo.mt-grpo.minus.gdpo-w0.5.aime24}{+7.0}
\setval{olmo.mt-grpo.minus.gdpo-w0.5.aime24-abs}{7.0}
\setval{olmo.mt-grpo.minus.gdpo-w0.5.aime25}{+0.8}
\setval{olmo.mt-grpo.minus.gdpo-w0.5.aime25-abs}{0.8}
\setval{olmo.mt-grpo.minus.gdpo-w0.5.amc23}{+3.4}
\setval{olmo.mt-grpo.minus.gdpo-w0.5.amc23-abs}{3.4}
\setval{olmo.mt-grpo.minus.gdpo-w0.5.avg}{+1.1}
\setval{olmo.mt-grpo.minus.gdpo-w0.5.avg-abs}{1.1}
\setval{olmo.mt-grpo.minus.gdpo-w0.5.code}{$-$1.4}
\setval{olmo.mt-grpo.minus.gdpo-w0.5.code-abs}{1.4}
\setval{olmo.mt-grpo.minus.gdpo-w0.5.creative-writing-v3}{+2.6}
\setval{olmo.mt-grpo.minus.gdpo-w0.5.creative-writing-v3-abs}{2.6}
\setval{olmo.mt-grpo.minus.gdpo-w0.5.guidebench}{+2.6}
\setval{olmo.mt-grpo.minus.gdpo-w0.5.guidebench-abs}{2.6}
\setval{olmo.mt-grpo.minus.gdpo-w0.5.humaneval-plus}{$-$2.0}
\setval{olmo.mt-grpo.minus.gdpo-w0.5.humaneval-plus-abs}{2.0}
\setval{olmo.mt-grpo.minus.gdpo-w0.5.if}{+0.5}
\setval{olmo.mt-grpo.minus.gdpo-w0.5.if-abs}{0.5}
\setval{olmo.mt-grpo.minus.gdpo-w0.5.ifbench}{$-$0.7}
\setval{olmo.mt-grpo.minus.gdpo-w0.5.ifbench-abs}{0.7}
\setval{olmo.mt-grpo.minus.gdpo-w0.5.ifeval}{$-$0.5}
\setval{olmo.mt-grpo.minus.gdpo-w0.5.ifeval-abs}{0.5}
\setval{olmo.mt-grpo.minus.gdpo-w0.5.judge2.avg}{+1.0}
\setval{olmo.mt-grpo.minus.gdpo-w0.5.judge2.avg-abs}{1.0}
\setval{olmo.mt-grpo.minus.gdpo-w0.5.judge2.creative-writing-v3}{+0.5}
\setval{olmo.mt-grpo.minus.gdpo-w0.5.judge2.creative-writing-v3-abs}{0.5}
\setval{olmo.mt-grpo.minus.gdpo-w0.5.judge2.longbench-write}{+3.8}
\setval{olmo.mt-grpo.minus.gdpo-w0.5.judge2.longbench-write-abs}{3.8}
\setval{olmo.mt-grpo.minus.gdpo-w0.5.judge2.writing}{+1.7}
\setval{olmo.mt-grpo.minus.gdpo-w0.5.judge2.writing-abs}{1.7}
\setval{olmo.mt-grpo.minus.gdpo-w0.5.judge2.writingbench}{+0.8}
\setval{olmo.mt-grpo.minus.gdpo-w0.5.judge2.writingbench-abs}{0.8}
\setval{olmo.mt-grpo.minus.gdpo-w0.5.livecodebench}{$-$2.2}
\setval{olmo.mt-grpo.minus.gdpo-w0.5.livecodebench-abs}{2.2}
\setval{olmo.mt-grpo.minus.gdpo-w0.5.longbench-write}{+2.6}
\setval{olmo.mt-grpo.minus.gdpo-w0.5.longbench-write-abs}{2.6}
\setval{olmo.mt-grpo.minus.gdpo-w0.5.math}{+2.0}
\setval{olmo.mt-grpo.minus.gdpo-w0.5.math-abs}{2.0}
\setval{olmo.mt-grpo.minus.gdpo-w0.5.math500}{$-$0.1}
\setval{olmo.mt-grpo.minus.gdpo-w0.5.math500-abs}{0.1}
\setval{olmo.mt-grpo.minus.gdpo-w0.5.mbpp-plus}{+0.1}
\setval{olmo.mt-grpo.minus.gdpo-w0.5.mbpp-plus-abs}{0.1}
\setval{olmo.mt-grpo.minus.gdpo-w0.5.minerva}{+0.4}
\setval{olmo.mt-grpo.minus.gdpo-w0.5.minerva-abs}{0.4}
\setval{olmo.mt-grpo.minus.gdpo-w0.5.olympiad-bench}{+0.5}
\setval{olmo.mt-grpo.minus.gdpo-w0.5.olympiad-bench-abs}{0.5}
\setval{olmo.mt-grpo.minus.gdpo-w0.5.writing}{+2.1}
\setval{olmo.mt-grpo.minus.gdpo-w0.5.writing-abs}{2.1}
\setval{olmo.mt-grpo.minus.gdpo-w0.5.writingbench}{+1.2}
\setval{olmo.mt-grpo.minus.gdpo-w0.5.writingbench-abs}{1.2}
\setval{olmo.mt-grpo.minus.gdpo.aime24}{+8.8}
\setval{olmo.mt-grpo.minus.gdpo.aime24-abs}{8.8}
\setval{olmo.mt-grpo.minus.gdpo.aime25}{+5.5}
\setval{olmo.mt-grpo.minus.gdpo.aime25-abs}{5.5}
\setval{olmo.mt-grpo.minus.gdpo.amc23}{+7.9}
\setval{olmo.mt-grpo.minus.gdpo.amc23-abs}{7.9}
\setval{olmo.mt-grpo.minus.gdpo.avg}{+2.8}
\setval{olmo.mt-grpo.minus.gdpo.avg-abs}{2.8}
\setval{olmo.mt-grpo.minus.gdpo.code}{$-$0.3}
\setval{olmo.mt-grpo.minus.gdpo.code-abs}{0.3}
\setval{olmo.mt-grpo.minus.gdpo.creative-writing-v3}{+3.6}
\setval{olmo.mt-grpo.minus.gdpo.creative-writing-v3-abs}{3.6}
\setval{olmo.mt-grpo.minus.gdpo.guidebench}{+3.6}
\setval{olmo.mt-grpo.minus.gdpo.guidebench-abs}{3.6}
\setval{olmo.mt-grpo.minus.gdpo.humaneval-plus}{$-$0.6}
\setval{olmo.mt-grpo.minus.gdpo.humaneval-plus-abs}{0.6}
\setval{olmo.mt-grpo.minus.gdpo.if}{+2.2}
\setval{olmo.mt-grpo.minus.gdpo.if-abs}{2.2}
\setval{olmo.mt-grpo.minus.gdpo.ifbench}{+2.3}
\setval{olmo.mt-grpo.minus.gdpo.ifbench-abs}{2.3}
\setval{olmo.mt-grpo.minus.gdpo.ifeval}{+0.6}
\setval{olmo.mt-grpo.minus.gdpo.ifeval-abs}{0.6}
\setval{olmo.mt-grpo.minus.gdpo.judge2.avg}{+2.7}
\setval{olmo.mt-grpo.minus.gdpo.judge2.avg-abs}{2.7}
\setval{olmo.mt-grpo.minus.gdpo.judge2.creative-writing-v3}{+2.0}
\setval{olmo.mt-grpo.minus.gdpo.judge2.creative-writing-v3-abs}{2.0}
\setval{olmo.mt-grpo.minus.gdpo.judge2.longbench-write}{+4.8}
\setval{olmo.mt-grpo.minus.gdpo.judge2.longbench-write-abs}{4.8}
\setval{olmo.mt-grpo.minus.gdpo.judge2.writing}{+2.4}
\setval{olmo.mt-grpo.minus.gdpo.judge2.writing-abs}{2.4}
\setval{olmo.mt-grpo.minus.gdpo.judge2.writingbench}{+0.5}
\setval{olmo.mt-grpo.minus.gdpo.judge2.writingbench-abs}{0.5}
\setval{olmo.mt-grpo.minus.gdpo.livecodebench}{$-$0.5}
\setval{olmo.mt-grpo.minus.gdpo.livecodebench-abs}{0.5}
\setval{olmo.mt-grpo.minus.gdpo.longbench-write}{+3.1}
\setval{olmo.mt-grpo.minus.gdpo.longbench-write-abs}{3.1}
\setval{olmo.mt-grpo.minus.gdpo.math}{+4.6}
\setval{olmo.mt-grpo.minus.gdpo.math-abs}{4.6}
\setval{olmo.mt-grpo.minus.gdpo.math500}{+1.7}
\setval{olmo.mt-grpo.minus.gdpo.math500-abs}{1.7}
\setval{olmo.mt-grpo.minus.gdpo.mbpp-plus}{+0.2}
\setval{olmo.mt-grpo.minus.gdpo.mbpp-plus-abs}{0.2}
\setval{olmo.mt-grpo.minus.gdpo.minerva}{+1.6}
\setval{olmo.mt-grpo.minus.gdpo.minerva-abs}{1.6}
\setval{olmo.mt-grpo.minus.gdpo.olympiad-bench}{+2.1}
\setval{olmo.mt-grpo.minus.gdpo.olympiad-bench-abs}{2.1}
\setval{olmo.mt-grpo.minus.gdpo.writing}{+3.0}
\setval{olmo.mt-grpo.minus.gdpo.writing-abs}{3.0}
\setval{olmo.mt-grpo.minus.gdpo.writingbench}{+2.4}
\setval{olmo.mt-grpo.minus.gdpo.writingbench-abs}{2.4}
\setval{olmo.mt-grpo.minus.geoma.aime24}{+12.8}
\setval{olmo.mt-grpo.minus.geoma.aime24-abs}{12.8}
\setval{olmo.mt-grpo.minus.geoma.aime25}{+6.1}
\setval{olmo.mt-grpo.minus.geoma.aime25-abs}{6.1}
\setval{olmo.mt-grpo.minus.geoma.amc23}{+13.5}
\setval{olmo.mt-grpo.minus.geoma.amc23-abs}{13.5}
\setval{olmo.mt-grpo.minus.geoma.avg}{+3.8}
\setval{olmo.mt-grpo.minus.geoma.avg-abs}{3.8}
\setval{olmo.mt-grpo.minus.geoma.code}{+0.5}
\setval{olmo.mt-grpo.minus.geoma.code-abs}{0.5}
\setval{olmo.mt-grpo.minus.geoma.creative-writing-v3}{+1.1}
\setval{olmo.mt-grpo.minus.geoma.creative-writing-v3-abs}{1.1}
\setval{olmo.mt-grpo.minus.geoma.guidebench}{+7.9}
\setval{olmo.mt-grpo.minus.geoma.guidebench-abs}{7.9}
\setval{olmo.mt-grpo.minus.geoma.humaneval-plus}{+1.1}
\setval{olmo.mt-grpo.minus.geoma.humaneval-plus-abs}{1.1}
\setval{olmo.mt-grpo.minus.geoma.if}{+2.9}
\setval{olmo.mt-grpo.minus.geoma.if-abs}{2.9}
\setval{olmo.mt-grpo.minus.geoma.ifbench}{+1.7}
\setval{olmo.mt-grpo.minus.geoma.ifbench-abs}{1.7}
\setval{olmo.mt-grpo.minus.geoma.ifeval}{$-$1.0}
\setval{olmo.mt-grpo.minus.geoma.ifeval-abs}{1.0}
\setval{olmo.mt-grpo.minus.geoma.judge2.avg}{+3.5}
\setval{olmo.mt-grpo.minus.geoma.judge2.avg-abs}{3.5}
\setval{olmo.mt-grpo.minus.geoma.judge2.creative-writing-v3}{$-$1.9}
\setval{olmo.mt-grpo.minus.geoma.judge2.creative-writing-v3-abs}{1.9}
\setval{olmo.mt-grpo.minus.geoma.judge2.longbench-write}{+2.2}
\setval{olmo.mt-grpo.minus.geoma.judge2.longbench-write-abs}{2.2}
\setval{olmo.mt-grpo.minus.geoma.judge2.writing}{$-$0.3}
\setval{olmo.mt-grpo.minus.geoma.judge2.writing-abs}{0.3}
\setval{olmo.mt-grpo.minus.geoma.judge2.writingbench}{$-$1.1}
\setval{olmo.mt-grpo.minus.geoma.judge2.writingbench-abs}{1.1}
\setval{olmo.mt-grpo.minus.geoma.livecodebench}{+0.3}
\setval{olmo.mt-grpo.minus.geoma.livecodebench-abs}{0.3}
\setval{olmo.mt-grpo.minus.geoma.longbench-write}{+1.0}
\setval{olmo.mt-grpo.minus.geoma.longbench-write-abs}{1.0}
\setval{olmo.mt-grpo.minus.geoma.math}{+7.2}
\setval{olmo.mt-grpo.minus.geoma.math-abs}{7.2}
\setval{olmo.mt-grpo.minus.geoma.math500}{+3.7}
\setval{olmo.mt-grpo.minus.geoma.math500-abs}{3.7}
\setval{olmo.mt-grpo.minus.geoma.mbpp-plus}{+0.2}
\setval{olmo.mt-grpo.minus.geoma.mbpp-plus-abs}{0.2}
\setval{olmo.mt-grpo.minus.geoma.minerva}{+2.0}
\setval{olmo.mt-grpo.minus.geoma.minerva-abs}{2.0}
\setval{olmo.mt-grpo.minus.geoma.olympiad-bench}{+5.3}
\setval{olmo.mt-grpo.minus.geoma.olympiad-bench-abs}{5.3}
\setval{olmo.mt-grpo.minus.geoma.writing}{+1.0}
\setval{olmo.mt-grpo.minus.geoma.writing-abs}{1.0}
\setval{olmo.mt-grpo.minus.geoma.writingbench}{+0.7}
\setval{olmo.mt-grpo.minus.geoma.writingbench-abs}{0.7}
\setval{olmo.mt-grpo.minus.grpo-nodiv.aime24}{+4.8}
\setval{olmo.mt-grpo.minus.grpo-nodiv.aime24-abs}{4.8}
\setval{olmo.mt-grpo.minus.grpo-nodiv.aime25}{+0.5}
\setval{olmo.mt-grpo.minus.grpo-nodiv.aime25-abs}{0.5}
\setval{olmo.mt-grpo.minus.grpo-nodiv.amc23}{+4.4}
\setval{olmo.mt-grpo.minus.grpo-nodiv.amc23-abs}{4.4}
\setval{olmo.mt-grpo.minus.grpo-nodiv.avg}{+0.8}
\setval{olmo.mt-grpo.minus.grpo-nodiv.avg-abs}{0.8}
\setval{olmo.mt-grpo.minus.grpo-nodiv.code}{$-$1.0}
\setval{olmo.mt-grpo.minus.grpo-nodiv.code-abs}{1.0}
\setval{olmo.mt-grpo.minus.grpo-nodiv.creative-writing-v3}{+3.7}
\setval{olmo.mt-grpo.minus.grpo-nodiv.creative-writing-v3-abs}{3.7}
\setval{olmo.mt-grpo.minus.grpo-nodiv.guidebench}{$-$1.0}
\setval{olmo.mt-grpo.minus.grpo-nodiv.guidebench-abs}{1.0}
\setval{olmo.mt-grpo.minus.grpo-nodiv.humaneval-plus}{$-$1.6}
\setval{olmo.mt-grpo.minus.grpo-nodiv.humaneval-plus-abs}{1.6}
\setval{olmo.mt-grpo.minus.grpo-nodiv.if}{$-$0.3}
\setval{olmo.mt-grpo.minus.grpo-nodiv.if-abs}{0.3}
\setval{olmo.mt-grpo.minus.grpo-nodiv.ifbench}{+0.7}
\setval{olmo.mt-grpo.minus.grpo-nodiv.ifbench-abs}{0.7}
\setval{olmo.mt-grpo.minus.grpo-nodiv.ifeval}{$-$0.7}
\setval{olmo.mt-grpo.minus.grpo-nodiv.ifeval-abs}{0.7}
\setval{olmo.mt-grpo.minus.grpo-nodiv.judge2.avg}{+0.4}
\setval{olmo.mt-grpo.minus.grpo-nodiv.judge2.avg-abs}{0.4}
\setval{olmo.mt-grpo.minus.grpo-nodiv.judge2.creative-writing-v3}{+1.0}
\setval{olmo.mt-grpo.minus.grpo-nodiv.judge2.creative-writing-v3-abs}{1.0}
\setval{olmo.mt-grpo.minus.grpo-nodiv.judge2.longbench-write}{$-$0.7}
\setval{olmo.mt-grpo.minus.grpo-nodiv.judge2.longbench-write-abs}{0.7}
\setval{olmo.mt-grpo.minus.grpo-nodiv.judge2.writing}{+0.3}
\setval{olmo.mt-grpo.minus.grpo-nodiv.judge2.writing-abs}{0.3}
\setval{olmo.mt-grpo.minus.grpo-nodiv.judge2.writingbench}{+0.6}
\setval{olmo.mt-grpo.minus.grpo-nodiv.judge2.writingbench-abs}{0.6}
\setval{olmo.mt-grpo.minus.grpo-nodiv.livecodebench}{$-$1.0}
\setval{olmo.mt-grpo.minus.grpo-nodiv.livecodebench-abs}{1.0}
\setval{olmo.mt-grpo.minus.grpo-nodiv.longbench-write}{$-$1.2}
\setval{olmo.mt-grpo.minus.grpo-nodiv.longbench-write-abs}{1.2}
\setval{olmo.mt-grpo.minus.grpo-nodiv.math}{+1.5}
\setval{olmo.mt-grpo.minus.grpo-nodiv.math-abs}{1.5}
\setval{olmo.mt-grpo.minus.grpo-nodiv.math500}{$-$0.6}
\setval{olmo.mt-grpo.minus.grpo-nodiv.math500-abs}{0.6}
\setval{olmo.mt-grpo.minus.grpo-nodiv.mbpp-plus}{$-$0.4}
\setval{olmo.mt-grpo.minus.grpo-nodiv.mbpp-plus-abs}{0.4}
\setval{olmo.mt-grpo.minus.grpo-nodiv.minerva}{+0.1}
\setval{olmo.mt-grpo.minus.grpo-nodiv.minerva-abs}{0.1}
\setval{olmo.mt-grpo.minus.grpo-nodiv.olympiad-bench}{$-$0.4}
\setval{olmo.mt-grpo.minus.grpo-nodiv.olympiad-bench-abs}{0.4}
\setval{olmo.mt-grpo.minus.grpo-nodiv.writing}{+2.4}
\setval{olmo.mt-grpo.minus.grpo-nodiv.writing-abs}{2.4}
\setval{olmo.mt-grpo.minus.grpo-nodiv.writingbench}{+4.5}
\setval{olmo.mt-grpo.minus.grpo-nodiv.writingbench-abs}{4.5}
\setval{olmo.mt-grpo.minus.grpo.aime24}{+7.1}
\setval{olmo.mt-grpo.minus.grpo.aime24-abs}{7.1}
\setval{olmo.mt-grpo.minus.grpo.aime25}{+3.4}
\setval{olmo.mt-grpo.minus.grpo.aime25-abs}{3.4}
\setval{olmo.mt-grpo.minus.grpo.amc23}{+6.6}
\setval{olmo.mt-grpo.minus.grpo.amc23-abs}{6.6}
\setval{olmo.mt-grpo.minus.grpo.avg}{+1.4}
\setval{olmo.mt-grpo.minus.grpo.avg-abs}{1.4}
\setval{olmo.mt-grpo.minus.grpo.code}{$-$0.7}
\setval{olmo.mt-grpo.minus.grpo.code-abs}{0.7}
\setval{olmo.mt-grpo.minus.grpo.creative-writing-v3}{+1.3}
\setval{olmo.mt-grpo.minus.grpo.creative-writing-v3-abs}{1.3}
\setval{olmo.mt-grpo.minus.grpo.guidebench}{$-$1.9}
\setval{olmo.mt-grpo.minus.grpo.guidebench-abs}{1.9}
\setval{olmo.mt-grpo.minus.grpo.humaneval-plus}{$-$1.6}
\setval{olmo.mt-grpo.minus.grpo.humaneval-plus-abs}{1.6}
\setval{olmo.mt-grpo.minus.grpo.if}{$-$0.8}
\setval{olmo.mt-grpo.minus.grpo.if-abs}{0.8}
\setval{olmo.mt-grpo.minus.grpo.ifbench}{+0.0}
\setval{olmo.mt-grpo.minus.grpo.ifbench-abs}{0.0}
\setval{olmo.mt-grpo.minus.grpo.ifeval}{$-$0.4}
\setval{olmo.mt-grpo.minus.grpo.ifeval-abs}{0.4}
\setval{olmo.mt-grpo.minus.grpo.judge2.avg}{+1.5}
\setval{olmo.mt-grpo.minus.grpo.judge2.avg-abs}{1.5}
\setval{olmo.mt-grpo.minus.grpo.judge2.creative-writing-v3}{+1.3}
\setval{olmo.mt-grpo.minus.grpo.judge2.creative-writing-v3-abs}{1.3}
\setval{olmo.mt-grpo.minus.grpo.judge2.longbench-write}{+3.3}
\setval{olmo.mt-grpo.minus.grpo.judge2.longbench-write-abs}{3.3}
\setval{olmo.mt-grpo.minus.grpo.judge2.writing}{+1.2}
\setval{olmo.mt-grpo.minus.grpo.judge2.writing-abs}{1.2}
\setval{olmo.mt-grpo.minus.grpo.judge2.writingbench}{$-$1.0}
\setval{olmo.mt-grpo.minus.grpo.judge2.writingbench-abs}{1.0}
\setval{olmo.mt-grpo.minus.grpo.livecodebench}{+0.3}
\setval{olmo.mt-grpo.minus.grpo.livecodebench-abs}{0.3}
\setval{olmo.mt-grpo.minus.grpo.longbench-write}{+2.2}
\setval{olmo.mt-grpo.minus.grpo.longbench-write-abs}{2.2}
\setval{olmo.mt-grpo.minus.grpo.math}{+3.9}
\setval{olmo.mt-grpo.minus.grpo.math-abs}{3.9}
\setval{olmo.mt-grpo.minus.grpo.math500}{+2.0}
\setval{olmo.mt-grpo.minus.grpo.math500-abs}{2.0}
\setval{olmo.mt-grpo.minus.grpo.mbpp-plus}{$-$0.8}
\setval{olmo.mt-grpo.minus.grpo.mbpp-plus-abs}{0.8}
\setval{olmo.mt-grpo.minus.grpo.minerva}{+2.0}
\setval{olmo.mt-grpo.minus.grpo.minerva-abs}{2.0}
\setval{olmo.mt-grpo.minus.grpo.olympiad-bench}{+2.4}
\setval{olmo.mt-grpo.minus.grpo.olympiad-bench-abs}{2.4}
\setval{olmo.mt-grpo.minus.grpo.writing}{+0.9}
\setval{olmo.mt-grpo.minus.grpo.writing-abs}{0.9}
\setval{olmo.mt-grpo.minus.grpo.writingbench}{$-$0.8}
\setval{olmo.mt-grpo.minus.grpo.writingbench-abs}{0.8}
\setval{olmo.mt-grpo.minus.papo.aime24}{+14.4}
\setval{olmo.mt-grpo.minus.papo.aime24-abs}{14.4}
\setval{olmo.mt-grpo.minus.papo.aime25}{+7.8}
\setval{olmo.mt-grpo.minus.papo.aime25-abs}{7.8}
\setval{olmo.mt-grpo.minus.papo.amc23}{+18.0}
\setval{olmo.mt-grpo.minus.papo.amc23-abs}{18.0}
\setval{olmo.mt-grpo.minus.papo.avg}{+4.4}
\setval{olmo.mt-grpo.minus.papo.avg-abs}{4.4}
\setval{olmo.mt-grpo.minus.papo.code}{+0.5}
\setval{olmo.mt-grpo.minus.papo.code-abs}{0.5}
\setval{olmo.mt-grpo.minus.papo.creative-writing-v3}{+3.7}
\setval{olmo.mt-grpo.minus.papo.creative-writing-v3-abs}{3.7}
\setval{olmo.mt-grpo.minus.papo.guidebench}{$-$4.1}
\setval{olmo.mt-grpo.minus.papo.guidebench-abs}{4.1}
\setval{olmo.mt-grpo.minus.papo.humaneval-plus}{+1.0}
\setval{olmo.mt-grpo.minus.papo.humaneval-plus-abs}{1.0}
\setval{olmo.mt-grpo.minus.papo.if}{$-$0.6}
\setval{olmo.mt-grpo.minus.papo.if-abs}{0.6}
\setval{olmo.mt-grpo.minus.papo.ifbench}{+2.3}
\setval{olmo.mt-grpo.minus.papo.ifbench-abs}{2.3}
\setval{olmo.mt-grpo.minus.papo.ifeval}{$-$0.1}
\setval{olmo.mt-grpo.minus.papo.ifeval-abs}{0.1}
\setval{olmo.mt-grpo.minus.papo.judge2.avg}{+4.3}
\setval{olmo.mt-grpo.minus.papo.judge2.avg-abs}{4.3}
\setval{olmo.mt-grpo.minus.papo.judge2.creative-writing-v3}{+1.9}
\setval{olmo.mt-grpo.minus.papo.judge2.creative-writing-v3-abs}{1.9}
\setval{olmo.mt-grpo.minus.papo.judge2.longbench-write}{+4.8}
\setval{olmo.mt-grpo.minus.papo.judge2.longbench-write-abs}{4.8}
\setval{olmo.mt-grpo.minus.papo.judge2.writing}{+2.2}
\setval{olmo.mt-grpo.minus.papo.judge2.writing-abs}{2.2}
\setval{olmo.mt-grpo.minus.papo.judge2.writingbench}{+0.0}
\setval{olmo.mt-grpo.minus.papo.judge2.writingbench-abs}{0.0}
\setval{olmo.mt-grpo.minus.papo.livecodebench}{+0.7}
\setval{olmo.mt-grpo.minus.papo.livecodebench-abs}{0.7}
\setval{olmo.mt-grpo.minus.papo.longbench-write}{+1.6}
\setval{olmo.mt-grpo.minus.papo.longbench-write-abs}{1.6}
\setval{olmo.mt-grpo.minus.papo.math}{+9.7}
\setval{olmo.mt-grpo.minus.papo.math-abs}{9.7}
\setval{olmo.mt-grpo.minus.papo.math500}{+6.0}
\setval{olmo.mt-grpo.minus.papo.math500-abs}{6.0}
\setval{olmo.mt-grpo.minus.papo.mbpp-plus}{$-$0.1}
\setval{olmo.mt-grpo.minus.papo.mbpp-plus-abs}{0.1}
\setval{olmo.mt-grpo.minus.papo.minerva}{+3.4}
\setval{olmo.mt-grpo.minus.papo.minerva-abs}{3.4}
\setval{olmo.mt-grpo.minus.papo.olympiad-bench}{+8.3}
\setval{olmo.mt-grpo.minus.papo.olympiad-bench-abs}{8.3}
\setval{olmo.mt-grpo.minus.papo.writing}{+2.6}
\setval{olmo.mt-grpo.minus.papo.writing-abs}{2.6}
\setval{olmo.mt-grpo.minus.papo.writingbench}{+2.3}
\setval{olmo.mt-grpo.minus.papo.writingbench-abs}{2.3}
\setval{olmo.mt-grpo.minus.saw.aime24}{+10.4}
\setval{olmo.mt-grpo.minus.saw.aime24-abs}{10.4}
\setval{olmo.mt-grpo.minus.saw.aime25}{+2.9}
\setval{olmo.mt-grpo.minus.saw.aime25-abs}{2.9}
\setval{olmo.mt-grpo.minus.saw.amc23}{+10.5}
\setval{olmo.mt-grpo.minus.saw.amc23-abs}{10.5}
\setval{olmo.mt-grpo.minus.saw.avg}{+2.3}
\setval{olmo.mt-grpo.minus.saw.avg-abs}{2.3}
\setval{olmo.mt-grpo.minus.saw.code}{$-$0.1}
\setval{olmo.mt-grpo.minus.saw.code-abs}{0.1}
\setval{olmo.mt-grpo.minus.saw.creative-writing-v3}{+2.8}
\setval{olmo.mt-grpo.minus.saw.creative-writing-v3-abs}{2.8}
\setval{olmo.mt-grpo.minus.saw.guidebench}{$-$0.2}
\setval{olmo.mt-grpo.minus.saw.guidebench-abs}{0.2}
\setval{olmo.mt-grpo.minus.saw.humaneval-plus}{$-$0.6}
\setval{olmo.mt-grpo.minus.saw.humaneval-plus-abs}{0.6}
\setval{olmo.mt-grpo.minus.saw.if}{$-$0.8}
\setval{olmo.mt-grpo.minus.saw.if-abs}{0.8}
\setval{olmo.mt-grpo.minus.saw.ifbench}{$-$0.7}
\setval{olmo.mt-grpo.minus.saw.ifbench-abs}{0.7}
\setval{olmo.mt-grpo.minus.saw.ifeval}{$-$1.6}
\setval{olmo.mt-grpo.minus.saw.ifeval-abs}{1.6}
\setval{olmo.mt-grpo.minus.saw.judge2.avg}{+2.0}
\setval{olmo.mt-grpo.minus.saw.judge2.avg-abs}{2.0}
\setval{olmo.mt-grpo.minus.saw.judge2.creative-writing-v3}{$-$1.1}
\setval{olmo.mt-grpo.minus.saw.judge2.creative-writing-v3-abs}{1.1}
\setval{olmo.mt-grpo.minus.saw.judge2.longbench-write}{+3.0}
\setval{olmo.mt-grpo.minus.saw.judge2.longbench-write-abs}{3.0}
\setval{olmo.mt-grpo.minus.saw.judge2.writing}{+0.3}
\setval{olmo.mt-grpo.minus.saw.judge2.writing-abs}{0.3}
\setval{olmo.mt-grpo.minus.saw.judge2.writingbench}{$-$0.9}
\setval{olmo.mt-grpo.minus.saw.judge2.writingbench-abs}{0.9}
\setval{olmo.mt-grpo.minus.saw.livecodebench}{+0.7}
\setval{olmo.mt-grpo.minus.saw.livecodebench-abs}{0.7}
\setval{olmo.mt-grpo.minus.saw.longbench-write}{+1.4}
\setval{olmo.mt-grpo.minus.saw.longbench-write-abs}{1.4}
\setval{olmo.mt-grpo.minus.saw.math}{+5.4}
\setval{olmo.mt-grpo.minus.saw.math-abs}{5.4}
\setval{olmo.mt-grpo.minus.saw.math500}{+2.4}
\setval{olmo.mt-grpo.minus.saw.math500-abs}{2.4}
\setval{olmo.mt-grpo.minus.saw.mbpp-plus}{$-$0.4}
\setval{olmo.mt-grpo.minus.saw.mbpp-plus-abs}{0.4}
\setval{olmo.mt-grpo.minus.saw.minerva}{+1.6}
\setval{olmo.mt-grpo.minus.saw.minerva-abs}{1.6}
\setval{olmo.mt-grpo.minus.saw.olympiad-bench}{+4.2}
\setval{olmo.mt-grpo.minus.saw.olympiad-bench-abs}{4.2}
\setval{olmo.mt-grpo.minus.saw.writing}{+1.5}
\setval{olmo.mt-grpo.minus.saw.writing-abs}{1.5}
\setval{olmo.mt-grpo.minus.saw.writingbench}{+0.3}
\setval{olmo.mt-grpo.minus.saw.writingbench-abs}{0.3}
\setval{olmo.mt-grpo.passing-at-ceiling.code.early}{0.80}
\setval{olmo.mt-grpo.passing-at-ceiling.code.late}{0.70}
\setval{olmo.mt-grpo.passing-at-ceiling.code.run}{0.71}
\setval{olmo.mt-grpo.passing-at-ceiling.if.early}{0.49}
\setval{olmo.mt-grpo.passing-at-ceiling.if.late}{0.45}
\setval{olmo.mt-grpo.passing-at-ceiling.if.run}{0.47}
\setval{olmo.mt-grpo.passing-at-ceiling.math.early}{1.00}
\setval{olmo.mt-grpo.passing-at-ceiling.math.late}{1.00}
\setval{olmo.mt-grpo.passing-at-ceiling.math.run}{1.00}
\setval{olmo.mt-grpo.passing-at-ceiling.writing.early}{0.00}
\setval{olmo.mt-grpo.passing-at-ceiling.writing.late}{0.00}
\setval{olmo.mt-grpo.passing-at-ceiling.writing.run}{0.00}
\setval{olmo.mt-grpo.primary-level.code.early}{0.60}
\setval{olmo.mt-grpo.primary-level.code.late}{0.70}
\setval{olmo.mt-grpo.primary-level.code.run}{0.66}
\setval{olmo.mt-grpo.primary-level.if.early}{0.63}
\setval{olmo.mt-grpo.primary-level.if.late}{0.68}
\setval{olmo.mt-grpo.primary-level.if.run}{0.66}
\setval{olmo.mt-grpo.primary-level.math.early}{0.48}
\setval{olmo.mt-grpo.primary-level.math.late}{0.70}
\setval{olmo.mt-grpo.primary-level.math.run}{0.66}
\setval{olmo.mt-grpo.primary-level.writing.early}{0.48}
\setval{olmo.mt-grpo.primary-level.writing.late}{0.69}
\setval{olmo.mt-grpo.primary-level.writing.run}{0.61}
\setval{olmo.papo.aux-change.code}{+0.18}
\setval{olmo.papo.aux-change.if}{+0.12}
\setval{olmo.papo.aux-change.math}{$-$0.01}
\setval{olmo.papo.aux-change.writing}{+0.03}
\setval{olmo.papo.aux-change.writing-rm}{+0.32}
\setval{olmo.papo.aux-full.code.early}{0.79}
\setval{olmo.papo.aux-full.code.late}{0.96}
\setval{olmo.papo.aux-full.code.run}{0.94}
\setval{olmo.papo.aux-full.if.early}{0.65}
\setval{olmo.papo.aux-full.if.late}{0.96}
\setval{olmo.papo.aux-full.if.run}{0.91}
\setval{olmo.papo.aux-full.math.early}{0.56}
\setval{olmo.papo.aux-full.math.late}{0.75}
\setval{olmo.papo.aux-full.math.run}{0.73}
\setval{olmo.papo.aux-full.writing-rm.early}{0.02}
\setval{olmo.papo.aux-full.writing-rm.late}{0.36}
\setval{olmo.papo.aux-full.writing-rm.run}{0.17}
\setval{olmo.papo.aux-full.writing.early}{0.65}
\setval{olmo.papo.aux-full.writing.late}{0.82}
\setval{olmo.papo.aux-full.writing.run}{0.78}
\setval{olmo.papo.aux-level.code.early}{0.93}
\setval{olmo.papo.aux-level.code.late}{0.99}
\setval{olmo.papo.aux-level.code.run}{0.99}
\setval{olmo.papo.aux-level.if.early}{0.92}
\setval{olmo.papo.aux-level.if.late}{0.99}
\setval{olmo.papo.aux-level.if.run}{0.98}
\setval{olmo.papo.aux-level.math.early}{0.92}
\setval{olmo.papo.aux-level.math.late}{0.89}
\setval{olmo.papo.aux-level.math.run}{0.92}
\setval{olmo.papo.aux-level.writing-rm.early}{0.61}
\setval{olmo.papo.aux-level.writing-rm.late}{0.88}
\setval{olmo.papo.aux-level.writing-rm.run}{0.79}
\setval{olmo.papo.aux-level.writing.early}{0.96}
\setval{olmo.papo.aux-level.writing.late}{0.98}
\setval{olmo.papo.aux-level.writing.run}{0.97}
\setval{olmo.papo.bench.aime24}{22.2}
\setval{olmo.papo.bench.aime25}{20.3}
\setval{olmo.papo.bench.amc23}{59.4}
\setval{olmo.papo.bench.creative-writing-v3}{47.0}
\setval{olmo.papo.bench.guidebench}{58.6}
\setval{olmo.papo.bench.humaneval-plus}{74.8}
\setval{olmo.papo.bench.ifbench}{29.7}
\setval{olmo.papo.bench.ifeval}{87.2}
\setval{olmo.papo.bench.livecodebench}{21.9}
\setval{olmo.papo.bench.longbench-write}{64.7}
\setval{olmo.papo.bench.math500}{80.9}
\setval{olmo.papo.bench.mbpp-plus}{65.1}
\setval{olmo.papo.bench.minerva}{38.6}
\setval{olmo.papo.bench.olympiad-bench}{49.7}
\setval{olmo.papo.bench.writingbench}{65.4}
\setval{olmo.papo.eval.avg}{52.4}
\setval{olmo.papo.eval.code}{53.9}
\setval{olmo.papo.eval.if}{58.5}
\setval{olmo.papo.eval.math}{45.2}
\setval{olmo.papo.eval.writing}{59.0}
\setval{olmo.papo.gate.code.against.early}{0.05}
\setval{olmo.papo.gate.code.against.late}{0.04}
\setval{olmo.papo.gate.code.against.run}{0.03}
\setval{olmo.papo.gate.code.agreement}{$-$0.004}
\setval{olmo.papo.gate.code.ceiling.early}{0.18}
\setval{olmo.papo.gate.code.ceiling.late}{0.08}
\setval{olmo.papo.gate.code.ceiling.run}{0.10}
\setval{olmo.papo.gate.code.cosine.early}{0.79}
\setval{olmo.papo.gate.code.cosine.late}{0.79}
\setval{olmo.papo.gate.code.cosine.run}{0.81}
\setval{olmo.papo.gate.if.against.early}{0.08}
\setval{olmo.papo.gate.if.against.late}{0.04}
\setval{olmo.papo.gate.if.against.run}{0.05}
\setval{olmo.papo.gate.if.agreement}{+0.028}
\setval{olmo.papo.gate.if.ceiling.early}{0.22}
\setval{olmo.papo.gate.if.ceiling.late}{0.11}
\setval{olmo.papo.gate.if.ceiling.run}{0.13}
\setval{olmo.papo.gate.if.cosine.early}{0.53}
\setval{olmo.papo.gate.if.cosine.late}{0.56}
\setval{olmo.papo.gate.if.cosine.run}{0.58}
\setval{olmo.papo.gate.math.against.early}{0.00}
\setval{olmo.papo.gate.math.against.late}{0.00}
\setval{olmo.papo.gate.math.against.run}{0.00}
\setval{olmo.papo.gate.math.ceiling.early}{0.13}
\setval{olmo.papo.gate.math.ceiling.late}{0.24}
\setval{olmo.papo.gate.math.ceiling.run}{0.16}
\setval{olmo.papo.gate.math.cosine.early}{0.91}
\setval{olmo.papo.gate.math.cosine.late}{0.82}
\setval{olmo.papo.gate.math.cosine.run}{0.88}
\setval{olmo.papo.gate.writing-rm.against.early}{0.21}
\setval{olmo.papo.gate.writing-rm.against.late}{0.47}
\setval{olmo.papo.gate.writing-rm.against.run}{0.41}
\setval{olmo.papo.gate.writing-rm.agreement}{+0.085}
\setval{olmo.papo.gate.writing-rm.ceiling.early}{0.00}
\setval{olmo.papo.gate.writing-rm.ceiling.late}{0.00}
\setval{olmo.papo.gate.writing-rm.ceiling.run}{0.00}
\setval{olmo.papo.gate.writing-rm.cosine.early}{0.70}
\setval{olmo.papo.gate.writing-rm.cosine.late}{0.26}
\setval{olmo.papo.gate.writing-rm.cosine.run}{0.39}
\setval{olmo.papo.gate.writing.against.early}{0.21}
\setval{olmo.papo.gate.writing.against.late}{0.44}
\setval{olmo.papo.gate.writing.against.run}{0.38}
\setval{olmo.papo.gate.writing.agreement}{$-$0.039}
\setval{olmo.papo.gate.writing.ceiling.early}{0.00}
\setval{olmo.papo.gate.writing.ceiling.late}{0.00}
\setval{olmo.papo.gate.writing.ceiling.run}{0.00}
\setval{olmo.papo.gate.writing.cosine.early}{0.64}
\setval{olmo.papo.gate.writing.cosine.late}{0.21}
\setval{olmo.papo.gate.writing.cosine.run}{0.32}
\setval{olmo.papo.judge2.avg}{50.0}
\setval{olmo.papo.judge2.creative-writing-v3}{38.0}
\setval{olmo.papo.judge2.longbench-write}{55.6}
\setval{olmo.papo.judge2.writing}{47.0}
\setval{olmo.papo.judge2.writingbench}{47.5}
\setval{olmo.papo.last-step}{300}
\setval{olmo.papo.minus.base.aime24}{+13.6}
\setval{olmo.papo.minus.base.aime24-abs}{13.6}
\setval{olmo.papo.minus.base.aime25}{+12.9}
\setval{olmo.papo.minus.base.aime25-abs}{12.9}
\setval{olmo.papo.minus.base.amc23}{+14.8}
\setval{olmo.papo.minus.base.amc23-abs}{14.8}
\setval{olmo.papo.minus.base.avg}{+8.2}
\setval{olmo.papo.minus.base.avg-abs}{8.2}
\setval{olmo.papo.minus.base.code}{+4.8}
\setval{olmo.papo.minus.base.code-abs}{4.8}
\setval{olmo.papo.minus.base.creative-writing-v3}{+9.1}
\setval{olmo.papo.minus.base.creative-writing-v3-abs}{9.1}
\setval{olmo.papo.minus.base.guidebench}{$-$2.8}
\setval{olmo.papo.minus.base.guidebench-abs}{2.8}
\setval{olmo.papo.minus.base.humaneval-plus}{+5.5}
\setval{olmo.papo.minus.base.humaneval-plus-abs}{5.5}
\setval{olmo.papo.minus.base.if}{+1.0}
\setval{olmo.papo.minus.base.if-abs}{1.0}
\setval{olmo.papo.minus.base.ifbench}{+3.0}
\setval{olmo.papo.minus.base.ifbench-abs}{3.0}
\setval{olmo.papo.minus.base.ifeval}{+2.8}
\setval{olmo.papo.minus.base.ifeval-abs}{2.8}
\setval{olmo.papo.minus.base.judge2.avg}{+7.0}
\setval{olmo.papo.minus.base.judge2.avg-abs}{7.0}
\setval{olmo.papo.minus.base.judge2.creative-writing-v3}{+6.3}
\setval{olmo.papo.minus.base.judge2.creative-writing-v3-abs}{6.3}
\setval{olmo.papo.minus.base.judge2.longbench-write}{+2.4}
\setval{olmo.papo.minus.base.judge2.longbench-write-abs}{2.4}
\setval{olmo.papo.minus.base.judge2.writing}{+6.2}
\setval{olmo.papo.minus.base.judge2.writing-abs}{6.2}
\setval{olmo.papo.minus.base.judge2.writingbench}{+9.8}
\setval{olmo.papo.minus.base.judge2.writingbench-abs}{9.8}
\setval{olmo.papo.minus.base.livecodebench}{+5.0}
\setval{olmo.papo.minus.base.livecodebench-abs}{5.0}
\setval{olmo.papo.minus.base.longbench-write}{+7.3}
\setval{olmo.papo.minus.base.longbench-write-abs}{7.3}
\setval{olmo.papo.minus.base.math}{+11.5}
\setval{olmo.papo.minus.base.math-abs}{11.5}
\setval{olmo.papo.minus.base.math500}{+10.8}
\setval{olmo.papo.minus.base.math500-abs}{10.8}
\setval{olmo.papo.minus.base.mbpp-plus}{+3.8}
\setval{olmo.papo.minus.base.mbpp-plus-abs}{3.8}
\setval{olmo.papo.minus.base.minerva}{+3.3}
\setval{olmo.papo.minus.base.minerva-abs}{3.3}
\setval{olmo.papo.minus.base.olympiad-bench}{+13.5}
\setval{olmo.papo.minus.base.olympiad-bench-abs}{13.5}
\setval{olmo.papo.minus.base.writing}{+12.3}
\setval{olmo.papo.minus.base.writing-abs}{12.3}
\setval{olmo.papo.minus.base.writingbench}{+20.6}
\setval{olmo.papo.minus.base.writingbench-abs}{20.6}
\setval{olmo.papo.minus.concur-free.aime24}{$-$4.6}
\setval{olmo.papo.minus.concur-free.aime24-abs}{4.6}
\setval{olmo.papo.minus.concur-free.aime25}{$-$1.8}
\setval{olmo.papo.minus.concur-free.aime25-abs}{1.8}
\setval{olmo.papo.minus.concur-free.amc23}{$-$8.1}
\setval{olmo.papo.minus.concur-free.amc23-abs}{8.1}
\setval{olmo.papo.minus.concur-free.avg}{$-$3.1}
\setval{olmo.papo.minus.concur-free.avg-abs}{3.1}
\setval{olmo.papo.minus.concur-free.code}{$-$1.7}
\setval{olmo.papo.minus.concur-free.code-abs}{1.7}
\setval{olmo.papo.minus.concur-free.creative-writing-v3}{$-$2.8}
\setval{olmo.papo.minus.concur-free.creative-writing-v3-abs}{2.8}
\setval{olmo.papo.minus.concur-free.guidebench}{$-$1.5}
\setval{olmo.papo.minus.concur-free.guidebench-abs}{1.5}
\setval{olmo.papo.minus.concur-free.humaneval-plus}{$-$1.9}
\setval{olmo.papo.minus.concur-free.humaneval-plus-abs}{1.9}
\setval{olmo.papo.minus.concur-free.if}{$-$1.2}
\setval{olmo.papo.minus.concur-free.if-abs}{1.2}
\setval{olmo.papo.minus.concur-free.ifbench}{$-$2.0}
\setval{olmo.papo.minus.concur-free.ifbench-abs}{2.0}
\setval{olmo.papo.minus.concur-free.ifeval}{$-$0.1}
\setval{olmo.papo.minus.concur-free.ifeval-abs}{0.1}
\setval{olmo.papo.minus.concur-free.judge2.avg}{$-$2.7}
\setval{olmo.papo.minus.concur-free.judge2.avg-abs}{2.7}
\setval{olmo.papo.minus.concur-free.judge2.creative-writing-v3}{+0.4}
\setval{olmo.papo.minus.concur-free.judge2.creative-writing-v3-abs}{0.4}
\setval{olmo.papo.minus.concur-free.judge2.longbench-write}{$-$8.3}
\setval{olmo.papo.minus.concur-free.judge2.longbench-write-abs}{8.3}
\setval{olmo.papo.minus.concur-free.judge2.writing}{$-$2.8}
\setval{olmo.papo.minus.concur-free.judge2.writing-abs}{2.8}
\setval{olmo.papo.minus.concur-free.judge2.writingbench}{$-$0.5}
\setval{olmo.papo.minus.concur-free.judge2.writingbench-abs}{0.5}
\setval{olmo.papo.minus.concur-free.livecodebench}{$-$3.8}
\setval{olmo.papo.minus.concur-free.livecodebench-abs}{3.8}
\setval{olmo.papo.minus.concur-free.longbench-write}{$-$7.3}
\setval{olmo.papo.minus.concur-free.longbench-write-abs}{7.3}
\setval{olmo.papo.minus.concur-free.math}{$-$3.8}
\setval{olmo.papo.minus.concur-free.math-abs}{3.8}
\setval{olmo.papo.minus.concur-free.math500}{$-$3.5}
\setval{olmo.papo.minus.concur-free.math500-abs}{3.5}
\setval{olmo.papo.minus.concur-free.mbpp-plus}{+0.6}
\setval{olmo.papo.minus.concur-free.mbpp-plus-abs}{0.6}
\setval{olmo.papo.minus.concur-free.minerva}{$-$1.4}
\setval{olmo.papo.minus.concur-free.minerva-abs}{1.4}
\setval{olmo.papo.minus.concur-free.olympiad-bench}{$-$3.5}
\setval{olmo.papo.minus.concur-free.olympiad-bench-abs}{3.5}
\setval{olmo.papo.minus.concur-free.writing}{$-$5.0}
\setval{olmo.papo.minus.concur-free.writing-abs}{5.0}
\setval{olmo.papo.minus.concur-free.writingbench}{$-$4.9}
\setval{olmo.papo.minus.concur-free.writingbench-abs}{4.9}
\setval{olmo.papo.minus.concur-tied.aime24}{$-$12.3}
\setval{olmo.papo.minus.concur-tied.aime24-abs}{12.3}
\setval{olmo.papo.minus.concur-tied.aime25}{$-$9.6}
\setval{olmo.papo.minus.concur-tied.aime25-abs}{9.6}
\setval{olmo.papo.minus.concur-tied.amc23}{$-$20.5}
\setval{olmo.papo.minus.concur-tied.amc23-abs}{20.5}
\setval{olmo.papo.minus.concur-tied.avg}{$-$6.8}
\setval{olmo.papo.minus.concur-tied.avg-abs}{6.8}
\setval{olmo.papo.minus.concur-tied.code}{$-$7.2}
\setval{olmo.papo.minus.concur-tied.code-abs}{7.2}
\setval{olmo.papo.minus.concur-tied.creative-writing-v3}{$-$2.1}
\setval{olmo.papo.minus.concur-tied.creative-writing-v3-abs}{2.1}
\setval{olmo.papo.minus.concur-tied.guidebench}{+0.1}
\setval{olmo.papo.minus.concur-tied.guidebench-abs}{0.1}
\setval{olmo.papo.minus.concur-tied.humaneval-plus}{$-$8.5}
\setval{olmo.papo.minus.concur-tied.humaneval-plus-abs}{8.5}
\setval{olmo.papo.minus.concur-tied.if}{$-$1.5}
\setval{olmo.papo.minus.concur-tied.if-abs}{1.5}
\setval{olmo.papo.minus.concur-tied.ifbench}{$-$4.0}
\setval{olmo.papo.minus.concur-tied.ifbench-abs}{4.0}
\setval{olmo.papo.minus.concur-tied.ifeval}{$-$0.5}
\setval{olmo.papo.minus.concur-tied.ifeval-abs}{0.5}
\setval{olmo.papo.minus.concur-tied.judge2.avg}{$-$6.4}
\setval{olmo.papo.minus.concur-tied.judge2.avg-abs}{6.4}
\setval{olmo.papo.minus.concur-tied.judge2.creative-writing-v3}{+0.8}
\setval{olmo.papo.minus.concur-tied.judge2.creative-writing-v3-abs}{0.8}
\setval{olmo.papo.minus.concur-tied.judge2.longbench-write}{$-$4.9}
\setval{olmo.papo.minus.concur-tied.judge2.longbench-write-abs}{4.9}
\setval{olmo.papo.minus.concur-tied.judge2.writing}{$-$0.7}
\setval{olmo.papo.minus.concur-tied.judge2.writing-abs}{0.7}
\setval{olmo.papo.minus.concur-tied.judge2.writingbench}{+1.9}
\setval{olmo.papo.minus.concur-tied.judge2.writingbench-abs}{1.9}
\setval{olmo.papo.minus.concur-tied.livecodebench}{$-$7.1}
\setval{olmo.papo.minus.concur-tied.livecodebench-abs}{7.1}
\setval{olmo.papo.minus.concur-tied.longbench-write}{$-$4.5}
\setval{olmo.papo.minus.concur-tied.longbench-write-abs}{4.5}
\setval{olmo.papo.minus.concur-tied.math}{$-$11.4}
\setval{olmo.papo.minus.concur-tied.math-abs}{11.4}
\setval{olmo.papo.minus.concur-tied.math500}{$-$8.5}
\setval{olmo.papo.minus.concur-tied.math500-abs}{8.5}
\setval{olmo.papo.minus.concur-tied.mbpp-plus}{$-$6.0}
\setval{olmo.papo.minus.concur-tied.mbpp-plus-abs}{6.0}
\setval{olmo.papo.minus.concur-tied.minerva}{$-$5.5}
\setval{olmo.papo.minus.concur-tied.minerva-abs}{5.5}
\setval{olmo.papo.minus.concur-tied.olympiad-bench}{$-$11.8}
\setval{olmo.papo.minus.concur-tied.olympiad-bench-abs}{11.8}
\setval{olmo.papo.minus.concur-tied.writing}{$-$2.4}
\setval{olmo.papo.minus.concur-tied.writing-abs}{2.4}
\setval{olmo.papo.minus.concur-tied.writingbench}{$-$0.8}
\setval{olmo.papo.minus.concur-tied.writingbench-abs}{0.8}
\setval{olmo.papo.minus.concur.aime24}{$-$14.2}
\setval{olmo.papo.minus.concur.aime24-abs}{14.2}
\setval{olmo.papo.minus.concur.aime25}{$-$11.9}
\setval{olmo.papo.minus.concur.aime25-abs}{11.9}
\setval{olmo.papo.minus.concur.amc23}{$-$24.1}
\setval{olmo.papo.minus.concur.amc23-abs}{24.1}
\setval{olmo.papo.minus.concur.avg}{$-$7.0}
\setval{olmo.papo.minus.concur.avg-abs}{7.0}
\setval{olmo.papo.minus.concur.code}{$-$6.3}
\setval{olmo.papo.minus.concur.code-abs}{6.3}
\setval{olmo.papo.minus.concur.creative-writing-v3}{$-$2.2}
\setval{olmo.papo.minus.concur.creative-writing-v3-abs}{2.2}
\setval{olmo.papo.minus.concur.guidebench}{+1.7}
\setval{olmo.papo.minus.concur.guidebench-abs}{1.7}
\setval{olmo.papo.minus.concur.humaneval-plus}{$-$8.0}
\setval{olmo.papo.minus.concur.humaneval-plus-abs}{8.0}
\setval{olmo.papo.minus.concur.if}{$-$0.4}
\setval{olmo.papo.minus.concur.if-abs}{0.4}
\setval{olmo.papo.minus.concur.ifbench}{$-$2.3}
\setval{olmo.papo.minus.concur.ifbench-abs}{2.3}
\setval{olmo.papo.minus.concur.ifeval}{$-$0.6}
\setval{olmo.papo.minus.concur.ifeval-abs}{0.6}
\setval{olmo.papo.minus.concur.judge2.avg}{$-$6.6}
\setval{olmo.papo.minus.concur.judge2.avg-abs}{6.6}
\setval{olmo.papo.minus.concur.judge2.creative-writing-v3}{$-$0.5}
\setval{olmo.papo.minus.concur.judge2.creative-writing-v3-abs}{0.5}
\setval{olmo.papo.minus.concur.judge2.longbench-write}{$-$2.7}
\setval{olmo.papo.minus.concur.judge2.longbench-write-abs}{2.7}
\setval{olmo.papo.minus.concur.judge2.writing}{$-$0.2}
\setval{olmo.papo.minus.concur.judge2.writing-abs}{0.2}
\setval{olmo.papo.minus.concur.judge2.writingbench}{+2.6}
\setval{olmo.papo.minus.concur.judge2.writingbench-abs}{2.6}
\setval{olmo.papo.minus.concur.livecodebench}{$-$8.5}
\setval{olmo.papo.minus.concur.livecodebench-abs}{8.5}
\setval{olmo.papo.minus.concur.longbench-write}{$-$3.8}
\setval{olmo.papo.minus.concur.longbench-write-abs}{3.8}
\setval{olmo.papo.minus.concur.math}{$-$13.1}
\setval{olmo.papo.minus.concur.math-abs}{13.1}
\setval{olmo.papo.minus.concur.math500}{$-$9.7}
\setval{olmo.papo.minus.concur.math500-abs}{9.7}
\setval{olmo.papo.minus.concur.mbpp-plus}{$-$2.5}
\setval{olmo.papo.minus.concur.mbpp-plus-abs}{2.5}
\setval{olmo.papo.minus.concur.minerva}{$-$6.2}
\setval{olmo.papo.minus.concur.minerva-abs}{6.2}
\setval{olmo.papo.minus.concur.olympiad-bench}{$-$12.7}
\setval{olmo.papo.minus.concur.olympiad-bench-abs}{12.7}
\setval{olmo.papo.minus.concur.writing}{$-$2.0}
\setval{olmo.papo.minus.concur.writing-abs}{2.0}
\setval{olmo.papo.minus.concur.writingbench}{$-$0.1}
\setval{olmo.papo.minus.concur.writingbench-abs}{0.1}
\setval{olmo.papo.minus.gd2po.aime24}{$-$6.1}
\setval{olmo.papo.minus.gd2po.aime24-abs}{6.1}
\setval{olmo.papo.minus.gd2po.aime25}{$-$4.0}
\setval{olmo.papo.minus.gd2po.aime25-abs}{4.0}
\setval{olmo.papo.minus.gd2po.amc23}{$-$12.8}
\setval{olmo.papo.minus.gd2po.amc23-abs}{12.8}
\setval{olmo.papo.minus.gd2po.avg}{$-$3.3}
\setval{olmo.papo.minus.gd2po.avg-abs}{3.3}
\setval{olmo.papo.minus.gd2po.code}{$-$1.0}
\setval{olmo.papo.minus.gd2po.code-abs}{1.0}
\setval{olmo.papo.minus.gd2po.creative-writing-v3}{$-$1.3}
\setval{olmo.papo.minus.gd2po.creative-writing-v3-abs}{1.3}
\setval{olmo.papo.minus.gd2po.guidebench}{$-$2.5}
\setval{olmo.papo.minus.gd2po.guidebench-abs}{2.5}
\setval{olmo.papo.minus.gd2po.humaneval-plus}{$-$2.0}
\setval{olmo.papo.minus.gd2po.humaneval-plus-abs}{2.0}
\setval{olmo.papo.minus.gd2po.if}{$-$1.9}
\setval{olmo.papo.minus.gd2po.if-abs}{1.9}
\setval{olmo.papo.minus.gd2po.ifbench}{$-$2.7}
\setval{olmo.papo.minus.gd2po.ifbench-abs}{2.7}
\setval{olmo.papo.minus.gd2po.ifeval}{$-$0.5}
\setval{olmo.papo.minus.gd2po.ifeval-abs}{0.5}
\setval{olmo.papo.minus.gd2po.judge2.avg}{$-$3.3}
\setval{olmo.papo.minus.gd2po.judge2.avg-abs}{3.3}
\setval{olmo.papo.minus.gd2po.judge2.creative-writing-v3}{+0.1}
\setval{olmo.papo.minus.gd2po.judge2.creative-writing-v3-abs}{0.1}
\setval{olmo.papo.minus.gd2po.judge2.longbench-write}{$-$4.3}
\setval{olmo.papo.minus.gd2po.judge2.longbench-write-abs}{4.3}
\setval{olmo.papo.minus.gd2po.judge2.writing}{$-$1.7}
\setval{olmo.papo.minus.gd2po.judge2.writing-abs}{1.7}
\setval{olmo.papo.minus.gd2po.judge2.writingbench}{$-$0.9}
\setval{olmo.papo.minus.gd2po.judge2.writingbench-abs}{0.9}
\setval{olmo.papo.minus.gd2po.livecodebench}{$-$1.3}
\setval{olmo.papo.minus.gd2po.livecodebench-abs}{1.3}
\setval{olmo.papo.minus.gd2po.longbench-write}{$-$3.5}
\setval{olmo.papo.minus.gd2po.longbench-write-abs}{3.5}
\setval{olmo.papo.minus.gd2po.math}{$-$5.9}
\setval{olmo.papo.minus.gd2po.math-abs}{5.9}
\setval{olmo.papo.minus.gd2po.math500}{$-$4.8}
\setval{olmo.papo.minus.gd2po.math500-abs}{4.8}
\setval{olmo.papo.minus.gd2po.mbpp-plus}{+0.3}
\setval{olmo.papo.minus.gd2po.mbpp-plus-abs}{0.3}
\setval{olmo.papo.minus.gd2po.minerva}{$-$1.6}
\setval{olmo.papo.minus.gd2po.minerva-abs}{1.6}
\setval{olmo.papo.minus.gd2po.olympiad-bench}{$-$6.3}
\setval{olmo.papo.minus.gd2po.olympiad-bench-abs}{6.3}
\setval{olmo.papo.minus.gd2po.writing}{$-$1.7}
\setval{olmo.papo.minus.gd2po.writing-abs}{1.7}
\setval{olmo.papo.minus.gd2po.writingbench}{$-$0.4}
\setval{olmo.papo.minus.gd2po.writingbench-abs}{0.4}
\setval{olmo.papo.minus.gdpo-noaux.aime24}{$-$16.1}
\setval{olmo.papo.minus.gdpo-noaux.aime24-abs}{16.1}
\setval{olmo.papo.minus.gdpo-noaux.aime25}{$-$13.3}
\setval{olmo.papo.minus.gdpo-noaux.aime25-abs}{13.3}
\setval{olmo.papo.minus.gdpo-noaux.amc23}{$-$24.5}
\setval{olmo.papo.minus.gdpo-noaux.amc23-abs}{24.5}
\setval{olmo.papo.minus.gdpo-noaux.avg}{$-$5.5}
\setval{olmo.papo.minus.gdpo-noaux.avg-abs}{5.5}
\setval{olmo.papo.minus.gdpo-noaux.code}{$-$8.5}
\setval{olmo.papo.minus.gdpo-noaux.code-abs}{8.5}
\setval{olmo.papo.minus.gdpo-noaux.creative-writing-v3}{+8.8}
\setval{olmo.papo.minus.gdpo-noaux.creative-writing-v3-abs}{8.8}
\setval{olmo.papo.minus.gdpo-noaux.guidebench}{+2.9}
\setval{olmo.papo.minus.gdpo-noaux.guidebench-abs}{2.9}
\setval{olmo.papo.minus.gdpo-noaux.humaneval-plus}{$-$10.4}
\setval{olmo.papo.minus.gdpo-noaux.humaneval-plus-abs}{10.4}
\setval{olmo.papo.minus.gdpo-noaux.if}{+0.5}
\setval{olmo.papo.minus.gdpo-noaux.if-abs}{0.5}
\setval{olmo.papo.minus.gdpo-noaux.ifbench}{$-$3.7}
\setval{olmo.papo.minus.gdpo-noaux.ifbench-abs}{3.7}
\setval{olmo.papo.minus.gdpo-noaux.ifeval}{+2.4}
\setval{olmo.papo.minus.gdpo-noaux.ifeval-abs}{2.4}
\setval{olmo.papo.minus.gdpo-noaux.judge2.avg}{$-$5.7}
\setval{olmo.papo.minus.gdpo-noaux.judge2.avg-abs}{5.7}
\setval{olmo.papo.minus.gdpo-noaux.judge2.creative-writing-v3}{+7.0}
\setval{olmo.papo.minus.gdpo-noaux.judge2.creative-writing-v3-abs}{7.0}
\setval{olmo.papo.minus.gdpo-noaux.judge2.longbench-write}{+9.0}
\setval{olmo.papo.minus.gdpo-noaux.judge2.longbench-write-abs}{9.0}
\setval{olmo.papo.minus.gdpo-noaux.judge2.writing}{+7.9}
\setval{olmo.papo.minus.gdpo-noaux.judge2.writing-abs}{7.9}
\setval{olmo.papo.minus.gdpo-noaux.judge2.writingbench}{+7.7}
\setval{olmo.papo.minus.gdpo-noaux.judge2.writingbench-abs}{7.7}
\setval{olmo.papo.minus.gdpo-noaux.livecodebench}{$-$9.7}
\setval{olmo.papo.minus.gdpo-noaux.livecodebench-abs}{9.7}
\setval{olmo.papo.minus.gdpo-noaux.longbench-write}{+8.7}
\setval{olmo.papo.minus.gdpo-noaux.longbench-write-abs}{8.7}
\setval{olmo.papo.minus.gdpo-noaux.math}{$-$14.3}
\setval{olmo.papo.minus.gdpo-noaux.math-abs}{14.3}
\setval{olmo.papo.minus.gdpo-noaux.math500}{$-$10.0}
\setval{olmo.papo.minus.gdpo-noaux.math500-abs}{10.0}
\setval{olmo.papo.minus.gdpo-noaux.mbpp-plus}{$-$5.4}
\setval{olmo.papo.minus.gdpo-noaux.mbpp-plus-abs}{5.4}
\setval{olmo.papo.minus.gdpo-noaux.minerva}{$-$7.6}
\setval{olmo.papo.minus.gdpo-noaux.minerva-abs}{7.6}
\setval{olmo.papo.minus.gdpo-noaux.olympiad-bench}{$-$14.0}
\setval{olmo.papo.minus.gdpo-noaux.olympiad-bench-abs}{14.0}
\setval{olmo.papo.minus.gdpo-noaux.writing}{+9.1}
\setval{olmo.papo.minus.gdpo-noaux.writing-abs}{9.1}
\setval{olmo.papo.minus.gdpo-noaux.writingbench}{+9.9}
\setval{olmo.papo.minus.gdpo-noaux.writingbench-abs}{9.9}
\setval{olmo.papo.minus.gdpo-w0.5.aime24}{$-$7.4}
\setval{olmo.papo.minus.gdpo-w0.5.aime24-abs}{7.4}
\setval{olmo.papo.minus.gdpo-w0.5.aime25}{$-$7.0}
\setval{olmo.papo.minus.gdpo-w0.5.aime25-abs}{7.0}
\setval{olmo.papo.minus.gdpo-w0.5.amc23}{$-$14.6}
\setval{olmo.papo.minus.gdpo-w0.5.amc23-abs}{14.6}
\setval{olmo.papo.minus.gdpo-w0.5.avg}{$-$3.3}
\setval{olmo.papo.minus.gdpo-w0.5.avg-abs}{3.3}
\setval{olmo.papo.minus.gdpo-w0.5.code}{$-$1.9}
\setval{olmo.papo.minus.gdpo-w0.5.code-abs}{1.9}
\setval{olmo.papo.minus.gdpo-w0.5.creative-writing-v3}{$-$1.1}
\setval{olmo.papo.minus.gdpo-w0.5.creative-writing-v3-abs}{1.1}
\setval{olmo.papo.minus.gdpo-w0.5.guidebench}{+6.7}
\setval{olmo.papo.minus.gdpo-w0.5.guidebench-abs}{6.7}
\setval{olmo.papo.minus.gdpo-w0.5.humaneval-plus}{$-$3.0}
\setval{olmo.papo.minus.gdpo-w0.5.humaneval-plus-abs}{3.0}
\setval{olmo.papo.minus.gdpo-w0.5.if}{+1.1}
\setval{olmo.papo.minus.gdpo-w0.5.if-abs}{1.1}
\setval{olmo.papo.minus.gdpo-w0.5.ifbench}{$-$3.0}
\setval{olmo.papo.minus.gdpo-w0.5.ifbench-abs}{3.0}
\setval{olmo.papo.minus.gdpo-w0.5.ifeval}{$-$0.4}
\setval{olmo.papo.minus.gdpo-w0.5.ifeval-abs}{0.4}
\setval{olmo.papo.minus.gdpo-w0.5.judge2.avg}{$-$3.3}
\setval{olmo.papo.minus.gdpo-w0.5.judge2.avg-abs}{3.3}
\setval{olmo.papo.minus.gdpo-w0.5.judge2.creative-writing-v3}{$-$1.4}
\setval{olmo.papo.minus.gdpo-w0.5.judge2.creative-writing-v3-abs}{1.4}
\setval{olmo.papo.minus.gdpo-w0.5.judge2.longbench-write}{$-$1.0}
\setval{olmo.papo.minus.gdpo-w0.5.judge2.longbench-write-abs}{1.0}
\setval{olmo.papo.minus.gdpo-w0.5.judge2.writing}{$-$0.5}
\setval{olmo.papo.minus.gdpo-w0.5.judge2.writing-abs}{0.5}
\setval{olmo.papo.minus.gdpo-w0.5.judge2.writingbench}{+0.8}
\setval{olmo.papo.minus.gdpo-w0.5.judge2.writingbench-abs}{0.8}
\setval{olmo.papo.minus.gdpo-w0.5.livecodebench}{$-$2.9}
\setval{olmo.papo.minus.gdpo-w0.5.livecodebench-abs}{2.9}
\setval{olmo.papo.minus.gdpo-w0.5.longbench-write}{+1.0}
\setval{olmo.papo.minus.gdpo-w0.5.longbench-write-abs}{1.0}
\setval{olmo.papo.minus.gdpo-w0.5.math}{$-$7.6}
\setval{olmo.papo.minus.gdpo-w0.5.math-abs}{7.6}
\setval{olmo.papo.minus.gdpo-w0.5.math500}{$-$6.2}
\setval{olmo.papo.minus.gdpo-w0.5.math500-abs}{6.2}
\setval{olmo.papo.minus.gdpo-w0.5.mbpp-plus}{+0.2}
\setval{olmo.papo.minus.gdpo-w0.5.mbpp-plus-abs}{0.2}
\setval{olmo.papo.minus.gdpo-w0.5.minerva}{$-$3.0}
\setval{olmo.papo.minus.gdpo-w0.5.minerva-abs}{3.0}
\setval{olmo.papo.minus.gdpo-w0.5.olympiad-bench}{$-$7.8}
\setval{olmo.papo.minus.gdpo-w0.5.olympiad-bench-abs}{7.8}
\setval{olmo.papo.minus.gdpo-w0.5.writing}{$-$0.4}
\setval{olmo.papo.minus.gdpo-w0.5.writing-abs}{0.4}
\setval{olmo.papo.minus.gdpo-w0.5.writingbench}{$-$1.1}
\setval{olmo.papo.minus.gdpo-w0.5.writingbench-abs}{1.1}
\setval{olmo.papo.minus.gdpo.aime24}{$-$5.6}
\setval{olmo.papo.minus.gdpo.aime24-abs}{5.6}
\setval{olmo.papo.minus.gdpo.aime25}{$-$2.3}
\setval{olmo.papo.minus.gdpo.aime25-abs}{2.3}
\setval{olmo.papo.minus.gdpo.amc23}{$-$10.2}
\setval{olmo.papo.minus.gdpo.amc23-abs}{10.2}
\setval{olmo.papo.minus.gdpo.avg}{$-$1.5}
\setval{olmo.papo.minus.gdpo.avg-abs}{1.5}
\setval{olmo.papo.minus.gdpo.code}{$-$0.8}
\setval{olmo.papo.minus.gdpo.code-abs}{0.8}
\setval{olmo.papo.minus.gdpo.creative-writing-v3}{$-$0.2}
\setval{olmo.papo.minus.gdpo.creative-writing-v3-abs}{0.2}
\setval{olmo.papo.minus.gdpo.guidebench}{+7.8}
\setval{olmo.papo.minus.gdpo.guidebench-abs}{7.8}
\setval{olmo.papo.minus.gdpo.humaneval-plus}{$-$1.6}
\setval{olmo.papo.minus.gdpo.humaneval-plus-abs}{1.6}
\setval{olmo.papo.minus.gdpo.if}{+2.8}
\setval{olmo.papo.minus.gdpo.if-abs}{2.8}
\setval{olmo.papo.minus.gdpo.ifbench}{+0.0}
\setval{olmo.papo.minus.gdpo.ifbench-abs}{0.0}
\setval{olmo.papo.minus.gdpo.ifeval}{+0.7}
\setval{olmo.papo.minus.gdpo.ifeval-abs}{0.7}
\setval{olmo.papo.minus.gdpo.judge2.avg}{$-$1.6}
\setval{olmo.papo.minus.gdpo.judge2.avg-abs}{1.6}
\setval{olmo.papo.minus.gdpo.judge2.creative-writing-v3}{+0.1}
\setval{olmo.papo.minus.gdpo.judge2.creative-writing-v3-abs}{0.1}
\setval{olmo.papo.minus.gdpo.judge2.longbench-write}{+0.0}
\setval{olmo.papo.minus.gdpo.judge2.longbench-write-abs}{0.0}
\setval{olmo.papo.minus.gdpo.judge2.writing}{+0.2}
\setval{olmo.papo.minus.gdpo.judge2.writing-abs}{0.2}
\setval{olmo.papo.minus.gdpo.judge2.writingbench}{+0.4}
\setval{olmo.papo.minus.gdpo.judge2.writingbench-abs}{0.4}
\setval{olmo.papo.minus.gdpo.livecodebench}{$-$1.3}
\setval{olmo.papo.minus.gdpo.livecodebench-abs}{1.3}
\setval{olmo.papo.minus.gdpo.longbench-write}{+1.5}
\setval{olmo.papo.minus.gdpo.longbench-write-abs}{1.5}
\setval{olmo.papo.minus.gdpo.math}{$-$5.1}
\setval{olmo.papo.minus.gdpo.math-abs}{5.1}
\setval{olmo.papo.minus.gdpo.math500}{$-$4.4}
\setval{olmo.papo.minus.gdpo.math500-abs}{4.4}
\setval{olmo.papo.minus.gdpo.mbpp-plus}{+0.3}
\setval{olmo.papo.minus.gdpo.mbpp-plus-abs}{0.3}
\setval{olmo.papo.minus.gdpo.minerva}{$-$1.8}
\setval{olmo.papo.minus.gdpo.minerva-abs}{1.8}
\setval{olmo.papo.minus.gdpo.olympiad-bench}{$-$6.2}
\setval{olmo.papo.minus.gdpo.olympiad-bench-abs}{6.2}
\setval{olmo.papo.minus.gdpo.writing}{+0.4}
\setval{olmo.papo.minus.gdpo.writing-abs}{0.4}
\setval{olmo.papo.minus.gdpo.writingbench}{+0.0}
\setval{olmo.papo.minus.gdpo.writingbench-abs}{0.0}
\setval{olmo.papo.minus.geoma.aime24}{$-$1.6}
\setval{olmo.papo.minus.geoma.aime24-abs}{1.6}
\setval{olmo.papo.minus.geoma.aime25}{$-$1.7}
\setval{olmo.papo.minus.geoma.aime25-abs}{1.7}
\setval{olmo.papo.minus.geoma.amc23}{$-$4.5}
\setval{olmo.papo.minus.geoma.amc23-abs}{4.5}
\setval{olmo.papo.minus.geoma.avg}{$-$0.6}
\setval{olmo.papo.minus.geoma.avg-abs}{0.6}
\setval{olmo.papo.minus.geoma.code}{+0.0}
\setval{olmo.papo.minus.geoma.code-abs}{0.0}
\setval{olmo.papo.minus.geoma.creative-writing-v3}{$-$2.6}
\setval{olmo.papo.minus.geoma.creative-writing-v3-abs}{2.6}
\setval{olmo.papo.minus.geoma.guidebench}{+12.0}
\setval{olmo.papo.minus.geoma.guidebench-abs}{12.0}
\setval{olmo.papo.minus.geoma.humaneval-plus}{+0.1}
\setval{olmo.papo.minus.geoma.humaneval-plus-abs}{0.1}
\setval{olmo.papo.minus.geoma.if}{+3.5}
\setval{olmo.papo.minus.geoma.if-abs}{3.5}
\setval{olmo.papo.minus.geoma.ifbench}{$-$0.7}
\setval{olmo.papo.minus.geoma.ifbench-abs}{0.7}
\setval{olmo.papo.minus.geoma.ifeval}{$-$0.8}
\setval{olmo.papo.minus.geoma.ifeval-abs}{0.8}
\setval{olmo.papo.minus.geoma.judge2.avg}{$-$0.8}
\setval{olmo.papo.minus.geoma.judge2.avg-abs}{0.8}
\setval{olmo.papo.minus.geoma.judge2.creative-writing-v3}{$-$3.8}
\setval{olmo.papo.minus.geoma.judge2.creative-writing-v3-abs}{3.8}
\setval{olmo.papo.minus.geoma.judge2.longbench-write}{$-$2.6}
\setval{olmo.papo.minus.geoma.judge2.longbench-write-abs}{2.6}
\setval{olmo.papo.minus.geoma.judge2.writing}{$-$2.5}
\setval{olmo.papo.minus.geoma.judge2.writing-abs}{2.5}
\setval{olmo.papo.minus.geoma.judge2.writingbench}{$-$1.1}
\setval{olmo.papo.minus.geoma.judge2.writingbench-abs}{1.1}
\setval{olmo.papo.minus.geoma.livecodebench}{$-$0.5}
\setval{olmo.papo.minus.geoma.livecodebench-abs}{0.5}
\setval{olmo.papo.minus.geoma.longbench-write}{$-$0.6}
\setval{olmo.papo.minus.geoma.longbench-write-abs}{0.6}
\setval{olmo.papo.minus.geoma.math}{$-$2.4}
\setval{olmo.papo.minus.geoma.math-abs}{2.4}
\setval{olmo.papo.minus.geoma.math500}{$-$2.3}
\setval{olmo.papo.minus.geoma.math500-abs}{2.3}
\setval{olmo.papo.minus.geoma.mbpp-plus}{+0.3}
\setval{olmo.papo.minus.geoma.mbpp-plus-abs}{0.3}
\setval{olmo.papo.minus.geoma.minerva}{$-$1.4}
\setval{olmo.papo.minus.geoma.minerva-abs}{1.4}
\setval{olmo.papo.minus.geoma.olympiad-bench}{$-$3.0}
\setval{olmo.papo.minus.geoma.olympiad-bench-abs}{3.0}
\setval{olmo.papo.minus.geoma.writing}{$-$1.6}
\setval{olmo.papo.minus.geoma.writing-abs}{1.6}
\setval{olmo.papo.minus.geoma.writingbench}{$-$1.6}
\setval{olmo.papo.minus.geoma.writingbench-abs}{1.6}
\setval{olmo.papo.minus.grpo-nodiv.aime24}{$-$9.6}
\setval{olmo.papo.minus.grpo-nodiv.aime24-abs}{9.6}
\setval{olmo.papo.minus.grpo-nodiv.aime25}{$-$7.3}
\setval{olmo.papo.minus.grpo-nodiv.aime25-abs}{7.3}
\setval{olmo.papo.minus.grpo-nodiv.amc23}{$-$13.7}
\setval{olmo.papo.minus.grpo-nodiv.amc23-abs}{13.7}
\setval{olmo.papo.minus.grpo-nodiv.avg}{$-$3.6}
\setval{olmo.papo.minus.grpo-nodiv.avg-abs}{3.6}
\setval{olmo.papo.minus.grpo-nodiv.code}{$-$1.5}
\setval{olmo.papo.minus.grpo-nodiv.code-abs}{1.5}
\setval{olmo.papo.minus.grpo-nodiv.creative-writing-v3}{+0.0}
\setval{olmo.papo.minus.grpo-nodiv.creative-writing-v3-abs}{0.0}
\setval{olmo.papo.minus.grpo-nodiv.guidebench}{+3.2}
\setval{olmo.papo.minus.grpo-nodiv.guidebench-abs}{3.2}
\setval{olmo.papo.minus.grpo-nodiv.humaneval-plus}{$-$2.6}
\setval{olmo.papo.minus.grpo-nodiv.humaneval-plus-abs}{2.6}
\setval{olmo.papo.minus.grpo-nodiv.if}{+0.3}
\setval{olmo.papo.minus.grpo-nodiv.if-abs}{0.3}
\setval{olmo.papo.minus.grpo-nodiv.ifbench}{$-$1.7}
\setval{olmo.papo.minus.grpo-nodiv.ifbench-abs}{1.7}
\setval{olmo.papo.minus.grpo-nodiv.ifeval}{$-$0.6}
\setval{olmo.papo.minus.grpo-nodiv.ifeval-abs}{0.6}
\setval{olmo.papo.minus.grpo-nodiv.judge2.avg}{$-$3.9}
\setval{olmo.papo.minus.grpo-nodiv.judge2.avg-abs}{3.9}
\setval{olmo.papo.minus.grpo-nodiv.judge2.creative-writing-v3}{$-$0.9}
\setval{olmo.papo.minus.grpo-nodiv.judge2.creative-writing-v3-abs}{0.9}
\setval{olmo.papo.minus.grpo-nodiv.judge2.longbench-write}{$-$5.5}
\setval{olmo.papo.minus.grpo-nodiv.judge2.longbench-write-abs}{5.5}
\setval{olmo.papo.minus.grpo-nodiv.judge2.writing}{$-$1.9}
\setval{olmo.papo.minus.grpo-nodiv.judge2.writing-abs}{1.9}
\setval{olmo.papo.minus.grpo-nodiv.judge2.writingbench}{+0.5}
\setval{olmo.papo.minus.grpo-nodiv.judge2.writingbench-abs}{0.5}
\setval{olmo.papo.minus.grpo-nodiv.livecodebench}{$-$1.7}
\setval{olmo.papo.minus.grpo-nodiv.livecodebench-abs}{1.7}
\setval{olmo.papo.minus.grpo-nodiv.longbench-write}{$-$2.8}
\setval{olmo.papo.minus.grpo-nodiv.longbench-write-abs}{2.8}
\setval{olmo.papo.minus.grpo-nodiv.math}{$-$8.2}
\setval{olmo.papo.minus.grpo-nodiv.math-abs}{8.2}
\setval{olmo.papo.minus.grpo-nodiv.math500}{$-$6.6}
\setval{olmo.papo.minus.grpo-nodiv.math500-abs}{6.6}
\setval{olmo.papo.minus.grpo-nodiv.mbpp-plus}{$-$0.3}
\setval{olmo.papo.minus.grpo-nodiv.mbpp-plus-abs}{0.3}
\setval{olmo.papo.minus.grpo-nodiv.minerva}{$-$3.3}
\setval{olmo.papo.minus.grpo-nodiv.minerva-abs}{3.3}
\setval{olmo.papo.minus.grpo-nodiv.olympiad-bench}{$-$8.7}
\setval{olmo.papo.minus.grpo-nodiv.olympiad-bench-abs}{8.7}
\setval{olmo.papo.minus.grpo-nodiv.writing}{$-$0.2}
\setval{olmo.papo.minus.grpo-nodiv.writing-abs}{0.2}
\setval{olmo.papo.minus.grpo-nodiv.writingbench}{+2.2}
\setval{olmo.papo.minus.grpo-nodiv.writingbench-abs}{2.2}
\setval{olmo.papo.minus.grpo.aime24}{$-$7.3}
\setval{olmo.papo.minus.grpo.aime24-abs}{7.3}
\setval{olmo.papo.minus.grpo.aime25}{$-$4.4}
\setval{olmo.papo.minus.grpo.aime25-abs}{4.4}
\setval{olmo.papo.minus.grpo.amc23}{$-$11.5}
\setval{olmo.papo.minus.grpo.amc23-abs}{11.5}
\setval{olmo.papo.minus.grpo.avg}{$-$2.9}
\setval{olmo.papo.minus.grpo.avg-abs}{2.9}
\setval{olmo.papo.minus.grpo.code}{$-$1.2}
\setval{olmo.papo.minus.grpo.code-abs}{1.2}
\setval{olmo.papo.minus.grpo.creative-writing-v3}{$-$2.4}
\setval{olmo.papo.minus.grpo.creative-writing-v3-abs}{2.4}
\setval{olmo.papo.minus.grpo.guidebench}{+2.2}
\setval{olmo.papo.minus.grpo.guidebench-abs}{2.2}
\setval{olmo.papo.minus.grpo.humaneval-plus}{$-$2.6}
\setval{olmo.papo.minus.grpo.humaneval-plus-abs}{2.6}
\setval{olmo.papo.minus.grpo.if}{$-$0.1}
\setval{olmo.papo.minus.grpo.if-abs}{0.1}
\setval{olmo.papo.minus.grpo.ifbench}{$-$2.3}
\setval{olmo.papo.minus.grpo.ifbench-abs}{2.3}
\setval{olmo.papo.minus.grpo.ifeval}{$-$0.2}
\setval{olmo.papo.minus.grpo.ifeval-abs}{0.2}
\setval{olmo.papo.minus.grpo.judge2.avg}{$-$2.8}
\setval{olmo.papo.minus.grpo.judge2.avg-abs}{2.8}
\setval{olmo.papo.minus.grpo.judge2.creative-writing-v3}{$-$0.6}
\setval{olmo.papo.minus.grpo.judge2.creative-writing-v3-abs}{0.6}
\setval{olmo.papo.minus.grpo.judge2.longbench-write}{$-$1.5}
\setval{olmo.papo.minus.grpo.judge2.longbench-write-abs}{1.5}
\setval{olmo.papo.minus.grpo.judge2.writing}{$-$1.0}
\setval{olmo.papo.minus.grpo.judge2.writing-abs}{1.0}
\setval{olmo.papo.minus.grpo.judge2.writingbench}{$-$1.0}
\setval{olmo.papo.minus.grpo.judge2.writingbench-abs}{1.0}
\setval{olmo.papo.minus.grpo.livecodebench}{$-$0.5}
\setval{olmo.papo.minus.grpo.livecodebench-abs}{0.5}
\setval{olmo.papo.minus.grpo.longbench-write}{+0.5}
\setval{olmo.papo.minus.grpo.longbench-write-abs}{0.5}
\setval{olmo.papo.minus.grpo.math}{$-$5.8}
\setval{olmo.papo.minus.grpo.math-abs}{5.8}
\setval{olmo.papo.minus.grpo.math500}{$-$4.1}
\setval{olmo.papo.minus.grpo.math500-abs}{4.1}
\setval{olmo.papo.minus.grpo.mbpp-plus}{$-$0.7}
\setval{olmo.papo.minus.grpo.mbpp-plus-abs}{0.7}
\setval{olmo.papo.minus.grpo.minerva}{$-$1.4}
\setval{olmo.papo.minus.grpo.minerva-abs}{1.4}
\setval{olmo.papo.minus.grpo.olympiad-bench}{$-$5.9}
\setval{olmo.papo.minus.grpo.olympiad-bench-abs}{5.9}
\setval{olmo.papo.minus.grpo.writing}{$-$1.7}
\setval{olmo.papo.minus.grpo.writing-abs}{1.7}
\setval{olmo.papo.minus.grpo.writingbench}{$-$3.1}
\setval{olmo.papo.minus.grpo.writingbench-abs}{3.1}
\setval{olmo.papo.minus.mt-grpo.aime24}{$-$14.4}
\setval{olmo.papo.minus.mt-grpo.aime24-abs}{14.4}
\setval{olmo.papo.minus.mt-grpo.aime25}{$-$7.8}
\setval{olmo.papo.minus.mt-grpo.aime25-abs}{7.8}
\setval{olmo.papo.minus.mt-grpo.amc23}{$-$18.0}
\setval{olmo.papo.minus.mt-grpo.amc23-abs}{18.0}
\setval{olmo.papo.minus.mt-grpo.avg}{$-$4.4}
\setval{olmo.papo.minus.mt-grpo.avg-abs}{4.4}
\setval{olmo.papo.minus.mt-grpo.code}{$-$0.5}
\setval{olmo.papo.minus.mt-grpo.code-abs}{0.5}
\setval{olmo.papo.minus.mt-grpo.creative-writing-v3}{$-$3.7}
\setval{olmo.papo.minus.mt-grpo.creative-writing-v3-abs}{3.7}
\setval{olmo.papo.minus.mt-grpo.guidebench}{+4.1}
\setval{olmo.papo.minus.mt-grpo.guidebench-abs}{4.1}
\setval{olmo.papo.minus.mt-grpo.humaneval-plus}{$-$1.0}
\setval{olmo.papo.minus.mt-grpo.humaneval-plus-abs}{1.0}
\setval{olmo.papo.minus.mt-grpo.if}{+0.6}
\setval{olmo.papo.minus.mt-grpo.if-abs}{0.6}
\setval{olmo.papo.minus.mt-grpo.ifbench}{$-$2.3}
\setval{olmo.papo.minus.mt-grpo.ifbench-abs}{2.3}
\setval{olmo.papo.minus.mt-grpo.ifeval}{+0.1}
\setval{olmo.papo.minus.mt-grpo.ifeval-abs}{0.1}
\setval{olmo.papo.minus.mt-grpo.judge2.avg}{$-$4.3}
\setval{olmo.papo.minus.mt-grpo.judge2.avg-abs}{4.3}
\setval{olmo.papo.minus.mt-grpo.judge2.creative-writing-v3}{$-$1.9}
\setval{olmo.papo.minus.mt-grpo.judge2.creative-writing-v3-abs}{1.9}
\setval{olmo.papo.minus.mt-grpo.judge2.longbench-write}{$-$4.8}
\setval{olmo.papo.minus.mt-grpo.judge2.longbench-write-abs}{4.8}
\setval{olmo.papo.minus.mt-grpo.judge2.writing}{$-$2.2}
\setval{olmo.papo.minus.mt-grpo.judge2.writing-abs}{2.2}
\setval{olmo.papo.minus.mt-grpo.judge2.writingbench}{+0.0}
\setval{olmo.papo.minus.mt-grpo.judge2.writingbench-abs}{0.0}
\setval{olmo.papo.minus.mt-grpo.livecodebench}{$-$0.7}
\setval{olmo.papo.minus.mt-grpo.livecodebench-abs}{0.7}
\setval{olmo.papo.minus.mt-grpo.longbench-write}{$-$1.6}
\setval{olmo.papo.minus.mt-grpo.longbench-write-abs}{1.6}
\setval{olmo.papo.minus.mt-grpo.math}{$-$9.7}
\setval{olmo.papo.minus.mt-grpo.math-abs}{9.7}
\setval{olmo.papo.minus.mt-grpo.math500}{$-$6.0}
\setval{olmo.papo.minus.mt-grpo.math500-abs}{6.0}
\setval{olmo.papo.minus.mt-grpo.mbpp-plus}{+0.1}
\setval{olmo.papo.minus.mt-grpo.mbpp-plus-abs}{0.1}
\setval{olmo.papo.minus.mt-grpo.minerva}{$-$3.4}
\setval{olmo.papo.minus.mt-grpo.minerva-abs}{3.4}
\setval{olmo.papo.minus.mt-grpo.olympiad-bench}{$-$8.3}
\setval{olmo.papo.minus.mt-grpo.olympiad-bench-abs}{8.3}
\setval{olmo.papo.minus.mt-grpo.writing}{$-$2.6}
\setval{olmo.papo.minus.mt-grpo.writing-abs}{2.6}
\setval{olmo.papo.minus.mt-grpo.writingbench}{$-$2.3}
\setval{olmo.papo.minus.mt-grpo.writingbench-abs}{2.3}
\setval{olmo.papo.minus.saw.aime24}{$-$4.0}
\setval{olmo.papo.minus.saw.aime24-abs}{4.0}
\setval{olmo.papo.minus.saw.aime25}{$-$4.9}
\setval{olmo.papo.minus.saw.aime25-abs}{4.9}
\setval{olmo.papo.minus.saw.amc23}{$-$7.5}
\setval{olmo.papo.minus.saw.amc23-abs}{7.5}
\setval{olmo.papo.minus.saw.avg}{$-$2.1}
\setval{olmo.papo.minus.saw.avg-abs}{2.1}
\setval{olmo.papo.minus.saw.code}{$-$0.6}
\setval{olmo.papo.minus.saw.code-abs}{0.6}
\setval{olmo.papo.minus.saw.creative-writing-v3}{$-$0.9}
\setval{olmo.papo.minus.saw.creative-writing-v3-abs}{0.9}
\setval{olmo.papo.minus.saw.guidebench}{+3.9}
\setval{olmo.papo.minus.saw.guidebench-abs}{3.9}
\setval{olmo.papo.minus.saw.humaneval-plus}{$-$1.6}
\setval{olmo.papo.minus.saw.humaneval-plus-abs}{1.6}
\setval{olmo.papo.minus.saw.if}{$-$0.2}
\setval{olmo.papo.minus.saw.if-abs}{0.2}
\setval{olmo.papo.minus.saw.ifbench}{$-$3.0}
\setval{olmo.papo.minus.saw.ifbench-abs}{3.0}
\setval{olmo.papo.minus.saw.ifeval}{$-$1.4}
\setval{olmo.papo.minus.saw.ifeval-abs}{1.4}
\setval{olmo.papo.minus.saw.judge2.avg}{$-$2.3}
\setval{olmo.papo.minus.saw.judge2.avg-abs}{2.3}
\setval{olmo.papo.minus.saw.judge2.creative-writing-v3}{$-$3.0}
\setval{olmo.papo.minus.saw.judge2.creative-writing-v3-abs}{3.0}
\setval{olmo.papo.minus.saw.judge2.longbench-write}{$-$1.8}
\setval{olmo.papo.minus.saw.judge2.longbench-write-abs}{1.8}
\setval{olmo.papo.minus.saw.judge2.writing}{$-$1.9}
\setval{olmo.papo.minus.saw.judge2.writing-abs}{1.9}
\setval{olmo.papo.minus.saw.judge2.writingbench}{$-$0.9}
\setval{olmo.papo.minus.saw.judge2.writingbench-abs}{0.9}
\setval{olmo.papo.minus.saw.livecodebench}{+0.0}
\setval{olmo.papo.minus.saw.livecodebench-abs}{0.0}
\setval{olmo.papo.minus.saw.longbench-write}{$-$0.2}
\setval{olmo.papo.minus.saw.longbench-write-abs}{0.2}
\setval{olmo.papo.minus.saw.math}{$-$4.3}
\setval{olmo.papo.minus.saw.math-abs}{4.3}
\setval{olmo.papo.minus.saw.math500}{$-$3.6}
\setval{olmo.papo.minus.saw.math500-abs}{3.6}
\setval{olmo.papo.minus.saw.mbpp-plus}{$-$0.3}
\setval{olmo.papo.minus.saw.mbpp-plus-abs}{0.3}
\setval{olmo.papo.minus.saw.minerva}{$-$1.8}
\setval{olmo.papo.minus.saw.minerva-abs}{1.8}
\setval{olmo.papo.minus.saw.olympiad-bench}{$-$4.1}
\setval{olmo.papo.minus.saw.olympiad-bench-abs}{4.1}
\setval{olmo.papo.minus.saw.writing}{$-$1.0}
\setval{olmo.papo.minus.saw.writing-abs}{1.0}
\setval{olmo.papo.minus.saw.writingbench}{$-$2.0}
\setval{olmo.papo.minus.saw.writingbench-abs}{2.0}
\setval{olmo.papo.passing-at-ceiling.code.early}{0.84}
\setval{olmo.papo.passing-at-ceiling.code.late}{0.84}
\setval{olmo.papo.passing-at-ceiling.code.run}{0.83}
\setval{olmo.papo.passing-at-ceiling.if.early}{0.49}
\setval{olmo.papo.passing-at-ceiling.if.late}{0.58}
\setval{olmo.papo.passing-at-ceiling.if.run}{0.54}
\setval{olmo.papo.passing-at-ceiling.math.early}{1.00}
\setval{olmo.papo.passing-at-ceiling.math.late}{1.00}
\setval{olmo.papo.passing-at-ceiling.math.run}{1.00}
\setval{olmo.papo.passing-at-ceiling.writing.early}{0.00}
\setval{olmo.papo.passing-at-ceiling.writing.late}{0.00}
\setval{olmo.papo.passing-at-ceiling.writing.run}{0.00}
\setval{olmo.papo.primary-cosine.code.early}{0.85}
\setval{olmo.papo.primary-cosine.code.late}{0.93}
\setval{olmo.papo.primary-cosine.code.run}{0.92}
\setval{olmo.papo.primary-cosine.if.early}{0.64}
\setval{olmo.papo.primary-cosine.if.late}{0.73}
\setval{olmo.papo.primary-cosine.if.run}{0.71}
\setval{olmo.papo.primary-cosine.math.early}{0.87}
\setval{olmo.papo.primary-cosine.math.late}{0.85}
\setval{olmo.papo.primary-cosine.math.run}{0.87}
\setval{olmo.papo.primary-cosine.writing.early}{0.73}
\setval{olmo.papo.primary-cosine.writing.late}{0.73}
\setval{olmo.papo.primary-cosine.writing.run}{0.72}
\setval{olmo.papo.primary-level.code.early}{0.58}
\setval{olmo.papo.primary-level.code.late}{0.80}
\setval{olmo.papo.primary-level.code.run}{0.73}
\setval{olmo.papo.primary-level.if.early}{0.64}
\setval{olmo.papo.primary-level.if.late}{0.74}
\setval{olmo.papo.primary-level.if.run}{0.70}
\setval{olmo.papo.primary-level.math.early}{0.34}
\setval{olmo.papo.primary-level.math.late}{0.57}
\setval{olmo.papo.primary-level.math.run}{0.46}
\setval{olmo.papo.primary-level.writing.early}{0.49}
\setval{olmo.papo.primary-level.writing.late}{0.69}
\setval{olmo.papo.primary-level.writing.run}{0.63}
\setval{olmo.papo.solved.code.early}{0.11}
\setval{olmo.papo.solved.code.late}{0.06}
\setval{olmo.papo.solved.code.run}{0.08}
\setval{olmo.papo.solved.if.early}{0.17}
\setval{olmo.papo.solved.if.late}{0.17}
\setval{olmo.papo.solved.if.run}{0.16}
\setval{olmo.papo.solved.math.early}{0.07}
\setval{olmo.papo.solved.math.late}{0.10}
\setval{olmo.papo.solved.math.run}{0.08}
\setval{olmo.papo.solved.writing.early}{0.00}
\setval{olmo.papo.solved.writing.late}{0.00}
\setval{olmo.papo.solved.writing.run}{0.00}
\setval{olmo.papo.update-share.code.early}{0.17}
\setval{olmo.papo.update-share.code.late}{0.12}
\setval{olmo.papo.update-share.code.run}{0.13}
\setval{olmo.papo.update-share.if.early}{0.32}
\setval{olmo.papo.update-share.if.late}{0.20}
\setval{olmo.papo.update-share.if.run}{0.23}
\setval{olmo.papo.update-share.math.early}{0.17}
\setval{olmo.papo.update-share.math.late}{0.21}
\setval{olmo.papo.update-share.math.run}{0.19}
\setval{olmo.papo.update-share.writing.early}{0.35}
\setval{olmo.papo.update-share.writing.late}{0.47}
\setval{olmo.papo.update-share.writing.run}{0.45}
\setval{olmo.saw.aux-change.code}{+0.16}
\setval{olmo.saw.aux-change.if}{+0.12}
\setval{olmo.saw.aux-change.math}{$-$0.08}
\setval{olmo.saw.aux-change.writing}{+0.02}
\setval{olmo.saw.aux-change.writing-rm}{+0.29}
\setval{olmo.saw.aux-full.code.early}{0.75}
\setval{olmo.saw.aux-full.code.late}{0.92}
\setval{olmo.saw.aux-full.code.run}{0.90}
\setval{olmo.saw.aux-full.if.early}{0.60}
\setval{olmo.saw.aux-full.if.late}{0.95}
\setval{olmo.saw.aux-full.if.run}{0.87}
\setval{olmo.saw.aux-full.math.early}{0.53}
\setval{olmo.saw.aux-full.math.late}{0.62}
\setval{olmo.saw.aux-full.math.run}{0.52}
\setval{olmo.saw.aux-full.writing-rm.early}{0.02}
\setval{olmo.saw.aux-full.writing-rm.late}{0.27}
\setval{olmo.saw.aux-full.writing-rm.run}{0.11}
\setval{olmo.saw.aux-full.writing.early}{0.56}
\setval{olmo.saw.aux-full.writing.late}{0.76}
\setval{olmo.saw.aux-full.writing.run}{0.68}
\setval{olmo.saw.aux-level.code.early}{0.90}
\setval{olmo.saw.aux-level.code.late}{0.97}
\setval{olmo.saw.aux-level.code.run}{0.97}
\setval{olmo.saw.aux-level.if.early}{0.90}
\setval{olmo.saw.aux-level.if.late}{0.99}
\setval{olmo.saw.aux-level.if.run}{0.97}
\setval{olmo.saw.aux-level.math.early}{0.90}
\setval{olmo.saw.aux-level.math.late}{0.83}
\setval{olmo.saw.aux-level.math.run}{0.79}
\setval{olmo.saw.aux-level.writing-rm.early}{0.60}
\setval{olmo.saw.aux-level.writing-rm.late}{0.86}
\setval{olmo.saw.aux-level.writing-rm.run}{0.76}
\setval{olmo.saw.aux-level.writing.early}{0.95}
\setval{olmo.saw.aux-level.writing.late}{0.97}
\setval{olmo.saw.aux-level.writing.run}{0.96}
\setval{olmo.saw.bench.aime24}{26.1}
\setval{olmo.saw.bench.aime25}{25.2}
\setval{olmo.saw.bench.amc23}{66.9}
\setval{olmo.saw.bench.creative-writing-v3}{47.9}
\setval{olmo.saw.bench.guidebench}{54.7}
\setval{olmo.saw.bench.humaneval-plus}{76.3}
\setval{olmo.saw.bench.ifbench}{32.7}
\setval{olmo.saw.bench.ifeval}{88.6}
\setval{olmo.saw.bench.livecodebench}{21.9}
\setval{olmo.saw.bench.longbench-write}{64.9}
\setval{olmo.saw.bench.math500}{84.5}
\setval{olmo.saw.bench.mbpp-plus}{65.4}
\setval{olmo.saw.bench.minerva}{40.3}
\setval{olmo.saw.bench.olympiad-bench}{53.8}
\setval{olmo.saw.bench.writingbench}{67.4}
\setval{olmo.saw.eval.avg}{54.4}
\setval{olmo.saw.eval.code}{54.6}
\setval{olmo.saw.eval.if}{58.7}
\setval{olmo.saw.eval.math}{49.5}
\setval{olmo.saw.eval.writing}{60.1}
\setval{olmo.saw.gate.code.against.early}{0.06}
\setval{olmo.saw.gate.code.against.late}{0.08}
\setval{olmo.saw.gate.code.against.run}{0.07}
\setval{olmo.saw.gate.code.agreement}{$-$0.018}
\setval{olmo.saw.gate.code.ceiling.early}{0.23}
\setval{olmo.saw.gate.code.ceiling.late}{0.21}
\setval{olmo.saw.gate.code.ceiling.run}{0.18}
\setval{olmo.saw.gate.code.cosine.early}{0.74}
\setval{olmo.saw.gate.code.cosine.late}{0.64}
\setval{olmo.saw.gate.code.cosine.run}{0.71}
\setval{olmo.saw.gate.if.against.early}{0.10}
\setval{olmo.saw.gate.if.against.late}{0.05}
\setval{olmo.saw.gate.if.against.run}{0.06}
\setval{olmo.saw.gate.if.agreement}{+0.039}
\setval{olmo.saw.gate.if.ceiling.early}{0.23}
\setval{olmo.saw.gate.if.ceiling.late}{0.14}
\setval{olmo.saw.gate.if.ceiling.run}{0.18}
\setval{olmo.saw.gate.if.cosine.early}{0.51}
\setval{olmo.saw.gate.if.cosine.late}{0.55}
\setval{olmo.saw.gate.if.cosine.run}{0.55}
\setval{olmo.saw.gate.math.against.early}{0.00}
\setval{olmo.saw.gate.math.against.late}{0.00}
\setval{olmo.saw.gate.math.against.run}{0.00}
\setval{olmo.saw.gate.math.ceiling.early}{0.15}
\setval{olmo.saw.gate.math.ceiling.late}{0.40}
\setval{olmo.saw.gate.math.ceiling.run}{0.37}
\setval{olmo.saw.gate.math.cosine.early}{0.90}
\setval{olmo.saw.gate.math.cosine.late}{0.70}
\setval{olmo.saw.gate.math.cosine.run}{0.72}
\setval{olmo.saw.gate.writing-rm.against.early}{0.21}
\setval{olmo.saw.gate.writing-rm.against.late}{0.49}
\setval{olmo.saw.gate.writing-rm.against.run}{0.43}
\setval{olmo.saw.gate.writing-rm.agreement}{+0.091}
\setval{olmo.saw.gate.writing-rm.ceiling.early}{0.00}
\setval{olmo.saw.gate.writing-rm.ceiling.late}{0.00}
\setval{olmo.saw.gate.writing-rm.ceiling.run}{0.00}
\setval{olmo.saw.gate.writing-rm.cosine.early}{0.70}
\setval{olmo.saw.gate.writing-rm.cosine.late}{0.25}
\setval{olmo.saw.gate.writing-rm.cosine.run}{0.38}
\setval{olmo.saw.gate.writing.against.early}{0.24}
\setval{olmo.saw.gate.writing.against.late}{0.51}
\setval{olmo.saw.gate.writing.against.run}{0.45}
\setval{olmo.saw.gate.writing.agreement}{$-$0.054}
\setval{olmo.saw.gate.writing.ceiling.early}{0.00}
\setval{olmo.saw.gate.writing.ceiling.late}{0.00}
\setval{olmo.saw.gate.writing.ceiling.run}{0.00}
\setval{olmo.saw.gate.writing.cosine.early}{0.60}
\setval{olmo.saw.gate.writing.cosine.late}{0.16}
\setval{olmo.saw.gate.writing.cosine.run}{0.27}
\setval{olmo.saw.judge2.avg}{52.2}
\setval{olmo.saw.judge2.creative-writing-v3}{41.0}
\setval{olmo.saw.judge2.longbench-write}{57.4}
\setval{olmo.saw.judge2.writing}{48.9}
\setval{olmo.saw.judge2.writingbench}{48.4}
\setval{olmo.saw.last-step}{300}
\setval{olmo.saw.minus.base.aime24}{+17.6}
\setval{olmo.saw.minus.base.aime24-abs}{17.6}
\setval{olmo.saw.minus.base.aime25}{+17.8}
\setval{olmo.saw.minus.base.aime25-abs}{17.8}
\setval{olmo.saw.minus.base.amc23}{+22.3}
\setval{olmo.saw.minus.base.amc23-abs}{22.3}
\setval{olmo.saw.minus.base.avg}{+10.3}
\setval{olmo.saw.minus.base.avg-abs}{10.3}
\setval{olmo.saw.minus.base.code}{+5.4}
\setval{olmo.saw.minus.base.code-abs}{5.4}
\setval{olmo.saw.minus.base.creative-writing-v3}{+10.0}
\setval{olmo.saw.minus.base.creative-writing-v3-abs}{10.0}
\setval{olmo.saw.minus.base.guidebench}{$-$6.7}
\setval{olmo.saw.minus.base.guidebench-abs}{6.7}
\setval{olmo.saw.minus.base.humaneval-plus}{+7.1}
\setval{olmo.saw.minus.base.humaneval-plus-abs}{7.1}
\setval{olmo.saw.minus.base.if}{+1.2}
\setval{olmo.saw.minus.base.if-abs}{1.2}
\setval{olmo.saw.minus.base.ifbench}{+6.0}
\setval{olmo.saw.minus.base.ifbench-abs}{6.0}
\setval{olmo.saw.minus.base.ifeval}{+4.2}
\setval{olmo.saw.minus.base.ifeval-abs}{4.2}
\setval{olmo.saw.minus.base.judge2.avg}{+9.3}
\setval{olmo.saw.minus.base.judge2.avg-abs}{9.3}
\setval{olmo.saw.minus.base.judge2.creative-writing-v3}{+9.3}
\setval{olmo.saw.minus.base.judge2.creative-writing-v3-abs}{9.3}
\setval{olmo.saw.minus.base.judge2.longbench-write}{+4.2}
\setval{olmo.saw.minus.base.judge2.longbench-write-abs}{4.2}
\setval{olmo.saw.minus.base.judge2.writing}{+8.1}
\setval{olmo.saw.minus.base.judge2.writing-abs}{8.1}
\setval{olmo.saw.minus.base.judge2.writingbench}{+10.7}
\setval{olmo.saw.minus.base.judge2.writingbench-abs}{10.7}
\setval{olmo.saw.minus.base.livecodebench}{+5.0}
\setval{olmo.saw.minus.base.livecodebench-abs}{5.0}
\setval{olmo.saw.minus.base.longbench-write}{+7.5}
\setval{olmo.saw.minus.base.longbench-write-abs}{7.5}
\setval{olmo.saw.minus.base.math}{+15.8}
\setval{olmo.saw.minus.base.math-abs}{15.8}
\setval{olmo.saw.minus.base.math500}{+14.4}
\setval{olmo.saw.minus.base.math500-abs}{14.4}
\setval{olmo.saw.minus.base.mbpp-plus}{+4.2}
\setval{olmo.saw.minus.base.mbpp-plus-abs}{4.2}
\setval{olmo.saw.minus.base.minerva}{+5.0}
\setval{olmo.saw.minus.base.minerva-abs}{5.0}
\setval{olmo.saw.minus.base.olympiad-bench}{+17.5}
\setval{olmo.saw.minus.base.olympiad-bench-abs}{17.5}
\setval{olmo.saw.minus.base.writing}{+13.4}
\setval{olmo.saw.minus.base.writing-abs}{13.4}
\setval{olmo.saw.minus.base.writingbench}{+22.6}
\setval{olmo.saw.minus.base.writingbench-abs}{22.6}
\setval{olmo.saw.minus.concur-free.aime24}{$-$0.6}
\setval{olmo.saw.minus.concur-free.aime24-abs}{0.6}
\setval{olmo.saw.minus.concur-free.aime25}{+3.1}
\setval{olmo.saw.minus.concur-free.aime25-abs}{3.1}
\setval{olmo.saw.minus.concur-free.amc23}{$-$0.6}
\setval{olmo.saw.minus.concur-free.amc23-abs}{0.6}
\setval{olmo.saw.minus.concur-free.avg}{$-$1.0}
\setval{olmo.saw.minus.concur-free.avg-abs}{1.0}
\setval{olmo.saw.minus.concur-free.code}{$-$1.1}
\setval{olmo.saw.minus.concur-free.code-abs}{1.1}
\setval{olmo.saw.minus.concur-free.creative-writing-v3}{$-$1.9}
\setval{olmo.saw.minus.concur-free.creative-writing-v3-abs}{1.9}
\setval{olmo.saw.minus.concur-free.guidebench}{$-$5.5}
\setval{olmo.saw.minus.concur-free.guidebench-abs}{5.5}
\setval{olmo.saw.minus.concur-free.humaneval-plus}{$-$0.3}
\setval{olmo.saw.minus.concur-free.humaneval-plus-abs}{0.3}
\setval{olmo.saw.minus.concur-free.if}{$-$1.1}
\setval{olmo.saw.minus.concur-free.if-abs}{1.1}
\setval{olmo.saw.minus.concur-free.ifbench}{+1.0}
\setval{olmo.saw.minus.concur-free.ifbench-abs}{1.0}
\setval{olmo.saw.minus.concur-free.ifeval}{+1.3}
\setval{olmo.saw.minus.concur-free.ifeval-abs}{1.3}
\setval{olmo.saw.minus.concur-free.judge2.avg}{$-$0.4}
\setval{olmo.saw.minus.concur-free.judge2.avg-abs}{0.4}
\setval{olmo.saw.minus.concur-free.judge2.creative-writing-v3}{+3.4}
\setval{olmo.saw.minus.concur-free.judge2.creative-writing-v3-abs}{3.4}
\setval{olmo.saw.minus.concur-free.judge2.longbench-write}{$-$6.5}
\setval{olmo.saw.minus.concur-free.judge2.longbench-write-abs}{6.5}
\setval{olmo.saw.minus.concur-free.judge2.writing}{$-$0.9}
\setval{olmo.saw.minus.concur-free.judge2.writing-abs}{0.9}
\setval{olmo.saw.minus.concur-free.judge2.writingbench}{+0.4}
\setval{olmo.saw.minus.concur-free.judge2.writingbench-abs}{0.4}
\setval{olmo.saw.minus.concur-free.livecodebench}{$-$3.8}
\setval{olmo.saw.minus.concur-free.livecodebench-abs}{3.8}
\setval{olmo.saw.minus.concur-free.longbench-write}{$-$7.1}
\setval{olmo.saw.minus.concur-free.longbench-write-abs}{7.1}
\setval{olmo.saw.minus.concur-free.math}{+0.5}
\setval{olmo.saw.minus.concur-free.math-abs}{0.5}
\setval{olmo.saw.minus.concur-free.math500}{+0.1}
\setval{olmo.saw.minus.concur-free.math500-abs}{0.1}
\setval{olmo.saw.minus.concur-free.mbpp-plus}{+0.9}
\setval{olmo.saw.minus.concur-free.mbpp-plus-abs}{0.9}
\setval{olmo.saw.minus.concur-free.minerva}{+0.4}
\setval{olmo.saw.minus.concur-free.minerva-abs}{0.4}
\setval{olmo.saw.minus.concur-free.olympiad-bench}{+0.5}
\setval{olmo.saw.minus.concur-free.olympiad-bench-abs}{0.5}
\setval{olmo.saw.minus.concur-free.writing}{$-$3.9}
\setval{olmo.saw.minus.concur-free.writing-abs}{3.9}
\setval{olmo.saw.minus.concur-free.writingbench}{$-$2.8}
\setval{olmo.saw.minus.concur-free.writingbench-abs}{2.8}
\setval{olmo.saw.minus.concur-tied.aime24}{$-$8.3}
\setval{olmo.saw.minus.concur-tied.aime24-abs}{8.3}
\setval{olmo.saw.minus.concur-tied.aime25}{$-$4.7}
\setval{olmo.saw.minus.concur-tied.aime25-abs}{4.7}
\setval{olmo.saw.minus.concur-tied.amc23}{$-$13.0}
\setval{olmo.saw.minus.concur-tied.amc23-abs}{13.0}
\setval{olmo.saw.minus.concur-tied.avg}{$-$4.7}
\setval{olmo.saw.minus.concur-tied.avg-abs}{4.7}
\setval{olmo.saw.minus.concur-tied.code}{$-$6.6}
\setval{olmo.saw.minus.concur-tied.code-abs}{6.6}
\setval{olmo.saw.minus.concur-tied.creative-writing-v3}{$-$1.1}
\setval{olmo.saw.minus.concur-tied.creative-writing-v3-abs}{1.1}
\setval{olmo.saw.minus.concur-tied.guidebench}{$-$3.8}
\setval{olmo.saw.minus.concur-tied.guidebench-abs}{3.8}
\setval{olmo.saw.minus.concur-tied.humaneval-plus}{$-$7.0}
\setval{olmo.saw.minus.concur-tied.humaneval-plus-abs}{7.0}
\setval{olmo.saw.minus.concur-tied.if}{$-$1.3}
\setval{olmo.saw.minus.concur-tied.if-abs}{1.3}
\setval{olmo.saw.minus.concur-tied.ifbench}{$-$1.0}
\setval{olmo.saw.minus.concur-tied.ifbench-abs}{1.0}
\setval{olmo.saw.minus.concur-tied.ifeval}{+1.0}
\setval{olmo.saw.minus.concur-tied.ifeval-abs}{1.0}
\setval{olmo.saw.minus.concur-tied.judge2.avg}{$-$4.2}
\setval{olmo.saw.minus.concur-tied.judge2.avg-abs}{4.2}
\setval{olmo.saw.minus.concur-tied.judge2.creative-writing-v3}{+3.8}
\setval{olmo.saw.minus.concur-tied.judge2.creative-writing-v3-abs}{3.8}
\setval{olmo.saw.minus.concur-tied.judge2.longbench-write}{$-$3.1}
\setval{olmo.saw.minus.concur-tied.judge2.longbench-write-abs}{3.1}
\setval{olmo.saw.minus.concur-tied.judge2.writing}{+1.2}
\setval{olmo.saw.minus.concur-tied.judge2.writing-abs}{1.2}
\setval{olmo.saw.minus.concur-tied.judge2.writingbench}{+2.8}
\setval{olmo.saw.minus.concur-tied.judge2.writingbench-abs}{2.8}
\setval{olmo.saw.minus.concur-tied.livecodebench}{$-$7.1}
\setval{olmo.saw.minus.concur-tied.livecodebench-abs}{7.1}
\setval{olmo.saw.minus.concur-tied.longbench-write}{$-$4.3}
\setval{olmo.saw.minus.concur-tied.longbench-write-abs}{4.3}
\setval{olmo.saw.minus.concur-tied.math}{$-$7.1}
\setval{olmo.saw.minus.concur-tied.math-abs}{7.1}
\setval{olmo.saw.minus.concur-tied.math500}{$-$4.9}
\setval{olmo.saw.minus.concur-tied.math500-abs}{4.9}
\setval{olmo.saw.minus.concur-tied.mbpp-plus}{$-$5.6}
\setval{olmo.saw.minus.concur-tied.mbpp-plus-abs}{5.6}
\setval{olmo.saw.minus.concur-tied.minerva}{$-$3.7}
\setval{olmo.saw.minus.concur-tied.minerva-abs}{3.7}
\setval{olmo.saw.minus.concur-tied.olympiad-bench}{$-$7.8}
\setval{olmo.saw.minus.concur-tied.olympiad-bench-abs}{7.8}
\setval{olmo.saw.minus.concur-tied.writing}{$-$1.4}
\setval{olmo.saw.minus.concur-tied.writing-abs}{1.4}
\setval{olmo.saw.minus.concur-tied.writingbench}{+1.2}
\setval{olmo.saw.minus.concur-tied.writingbench-abs}{1.2}
\setval{olmo.saw.minus.concur.aime24}{$-$10.2}
\setval{olmo.saw.minus.concur.aime24-abs}{10.2}
\setval{olmo.saw.minus.concur.aime25}{$-$7.0}
\setval{olmo.saw.minus.concur.aime25-abs}{7.0}
\setval{olmo.saw.minus.concur.amc23}{$-$16.6}
\setval{olmo.saw.minus.concur.amc23-abs}{16.6}
\setval{olmo.saw.minus.concur.avg}{$-$4.9}
\setval{olmo.saw.minus.concur.avg-abs}{4.9}
\setval{olmo.saw.minus.concur.code}{$-$5.7}
\setval{olmo.saw.minus.concur.code-abs}{5.7}
\setval{olmo.saw.minus.concur.creative-writing-v3}{$-$1.3}
\setval{olmo.saw.minus.concur.creative-writing-v3-abs}{1.3}
\setval{olmo.saw.minus.concur.guidebench}{$-$2.2}
\setval{olmo.saw.minus.concur.guidebench-abs}{2.2}
\setval{olmo.saw.minus.concur.humaneval-plus}{$-$6.4}
\setval{olmo.saw.minus.concur.humaneval-plus-abs}{6.4}
\setval{olmo.saw.minus.concur.if}{$-$0.2}
\setval{olmo.saw.minus.concur.if-abs}{0.2}
\setval{olmo.saw.minus.concur.ifbench}{+0.7}
\setval{olmo.saw.minus.concur.ifbench-abs}{0.7}
\setval{olmo.saw.minus.concur.ifeval}{+0.8}
\setval{olmo.saw.minus.concur.ifeval-abs}{0.8}
\setval{olmo.saw.minus.concur.judge2.avg}{$-$4.4}
\setval{olmo.saw.minus.concur.judge2.avg-abs}{4.4}
\setval{olmo.saw.minus.concur.judge2.creative-writing-v3}{+2.5}
\setval{olmo.saw.minus.concur.judge2.creative-writing-v3-abs}{2.5}
\setval{olmo.saw.minus.concur.judge2.longbench-write}{$-$1.0}
\setval{olmo.saw.minus.concur.judge2.longbench-write-abs}{1.0}
\setval{olmo.saw.minus.concur.judge2.writing}{+1.7}
\setval{olmo.saw.minus.concur.judge2.writing-abs}{1.7}
\setval{olmo.saw.minus.concur.judge2.writingbench}{+3.5}
\setval{olmo.saw.minus.concur.judge2.writingbench-abs}{3.5}
\setval{olmo.saw.minus.concur.livecodebench}{$-$8.5}
\setval{olmo.saw.minus.concur.livecodebench-abs}{8.5}
\setval{olmo.saw.minus.concur.longbench-write}{$-$3.6}
\setval{olmo.saw.minus.concur.longbench-write-abs}{3.6}
\setval{olmo.saw.minus.concur.math}{$-$8.8}
\setval{olmo.saw.minus.concur.math-abs}{8.8}
\setval{olmo.saw.minus.concur.math500}{$-$6.0}
\setval{olmo.saw.minus.concur.math500-abs}{6.0}
\setval{olmo.saw.minus.concur.mbpp-plus}{$-$2.1}
\setval{olmo.saw.minus.concur.mbpp-plus-abs}{2.1}
\setval{olmo.saw.minus.concur.minerva}{$-$4.5}
\setval{olmo.saw.minus.concur.minerva-abs}{4.5}
\setval{olmo.saw.minus.concur.olympiad-bench}{$-$8.6}
\setval{olmo.saw.minus.concur.olympiad-bench-abs}{8.6}
\setval{olmo.saw.minus.concur.writing}{$-$1.0}
\setval{olmo.saw.minus.concur.writing-abs}{1.0}
\setval{olmo.saw.minus.concur.writingbench}{+1.9}
\setval{olmo.saw.minus.concur.writingbench-abs}{1.9}
\setval{olmo.saw.minus.gd2po.aime24}{$-$2.2}
\setval{olmo.saw.minus.gd2po.aime24-abs}{2.2}
\setval{olmo.saw.minus.gd2po.aime25}{+0.9}
\setval{olmo.saw.minus.gd2po.aime25-abs}{0.9}
\setval{olmo.saw.minus.gd2po.amc23}{$-$5.3}
\setval{olmo.saw.minus.gd2po.amc23-abs}{5.3}
\setval{olmo.saw.minus.gd2po.avg}{$-$1.2}
\setval{olmo.saw.minus.gd2po.avg-abs}{1.2}
\setval{olmo.saw.minus.gd2po.code}{$-$0.4}
\setval{olmo.saw.minus.gd2po.code-abs}{0.4}
\setval{olmo.saw.minus.gd2po.creative-writing-v3}{$-$0.4}
\setval{olmo.saw.minus.gd2po.creative-writing-v3-abs}{0.4}
\setval{olmo.saw.minus.gd2po.guidebench}{$-$6.4}
\setval{olmo.saw.minus.gd2po.guidebench-abs}{6.4}
\setval{olmo.saw.minus.gd2po.humaneval-plus}{$-$0.4}
\setval{olmo.saw.minus.gd2po.humaneval-plus-abs}{0.4}
\setval{olmo.saw.minus.gd2po.if}{$-$1.7}
\setval{olmo.saw.minus.gd2po.if-abs}{1.7}
\setval{olmo.saw.minus.gd2po.ifbench}{+0.3}
\setval{olmo.saw.minus.gd2po.ifbench-abs}{0.3}
\setval{olmo.saw.minus.gd2po.ifeval}{+1.0}
\setval{olmo.saw.minus.gd2po.ifeval-abs}{1.0}
\setval{olmo.saw.minus.gd2po.judge2.avg}{$-$1.0}
\setval{olmo.saw.minus.gd2po.judge2.avg-abs}{1.0}
\setval{olmo.saw.minus.gd2po.judge2.creative-writing-v3}{+3.1}
\setval{olmo.saw.minus.gd2po.judge2.creative-writing-v3-abs}{3.1}
\setval{olmo.saw.minus.gd2po.judge2.longbench-write}{$-$2.5}
\setval{olmo.saw.minus.gd2po.judge2.longbench-write-abs}{2.5}
\setval{olmo.saw.minus.gd2po.judge2.writing}{+0.2}
\setval{olmo.saw.minus.gd2po.judge2.writing-abs}{0.2}
\setval{olmo.saw.minus.gd2po.judge2.writingbench}{+0.1}
\setval{olmo.saw.minus.gd2po.judge2.writingbench-abs}{0.1}
\setval{olmo.saw.minus.gd2po.livecodebench}{$-$1.3}
\setval{olmo.saw.minus.gd2po.livecodebench-abs}{1.3}
\setval{olmo.saw.minus.gd2po.longbench-write}{$-$3.3}
\setval{olmo.saw.minus.gd2po.longbench-write-abs}{3.3}
\setval{olmo.saw.minus.gd2po.math}{$-$1.6}
\setval{olmo.saw.minus.gd2po.math-abs}{1.6}
\setval{olmo.saw.minus.gd2po.math500}{$-$1.2}
\setval{olmo.saw.minus.gd2po.math500-abs}{1.2}
\setval{olmo.saw.minus.gd2po.mbpp-plus}{+0.6}
\setval{olmo.saw.minus.gd2po.mbpp-plus-abs}{0.6}
\setval{olmo.saw.minus.gd2po.minerva}{+0.2}
\setval{olmo.saw.minus.gd2po.minerva-abs}{0.2}
\setval{olmo.saw.minus.gd2po.olympiad-bench}{$-$2.3}
\setval{olmo.saw.minus.gd2po.olympiad-bench-abs}{2.3}
\setval{olmo.saw.minus.gd2po.writing}{$-$0.7}
\setval{olmo.saw.minus.gd2po.writing-abs}{0.7}
\setval{olmo.saw.minus.gd2po.writingbench}{+1.6}
\setval{olmo.saw.minus.gd2po.writingbench-abs}{1.6}
\setval{olmo.saw.minus.gdpo-noaux.aime24}{$-$12.2}
\setval{olmo.saw.minus.gdpo-noaux.aime24-abs}{12.2}
\setval{olmo.saw.minus.gdpo-noaux.aime25}{$-$8.4}
\setval{olmo.saw.minus.gdpo-noaux.aime25-abs}{8.4}
\setval{olmo.saw.minus.gdpo-noaux.amc23}{$-$17.0}
\setval{olmo.saw.minus.gdpo-noaux.amc23-abs}{17.0}
\setval{olmo.saw.minus.gdpo-noaux.avg}{$-$3.4}
\setval{olmo.saw.minus.gdpo-noaux.avg-abs}{3.4}
\setval{olmo.saw.minus.gdpo-noaux.code}{$-$7.9}
\setval{olmo.saw.minus.gdpo-noaux.code-abs}{7.9}
\setval{olmo.saw.minus.gdpo-noaux.creative-writing-v3}{+9.7}
\setval{olmo.saw.minus.gdpo-noaux.creative-writing-v3-abs}{9.7}
\setval{olmo.saw.minus.gdpo-noaux.guidebench}{$-$1.1}
\setval{olmo.saw.minus.gdpo-noaux.guidebench-abs}{1.1}
\setval{olmo.saw.minus.gdpo-noaux.humaneval-plus}{$-$8.8}
\setval{olmo.saw.minus.gdpo-noaux.humaneval-plus-abs}{8.8}
\setval{olmo.saw.minus.gdpo-noaux.if}{+0.7}
\setval{olmo.saw.minus.gdpo-noaux.if-abs}{0.7}
\setval{olmo.saw.minus.gdpo-noaux.ifbench}{$-$0.7}
\setval{olmo.saw.minus.gdpo-noaux.ifbench-abs}{0.7}
\setval{olmo.saw.minus.gdpo-noaux.ifeval}{+3.8}
\setval{olmo.saw.minus.gdpo-noaux.ifeval-abs}{3.8}
\setval{olmo.saw.minus.gdpo-noaux.judge2.avg}{$-$3.5}
\setval{olmo.saw.minus.gdpo-noaux.judge2.avg-abs}{3.5}
\setval{olmo.saw.minus.gdpo-noaux.judge2.creative-writing-v3}{+10.0}
\setval{olmo.saw.minus.gdpo-noaux.judge2.creative-writing-v3-abs}{10.0}
\setval{olmo.saw.minus.gdpo-noaux.judge2.longbench-write}{+10.8}
\setval{olmo.saw.minus.gdpo-noaux.judge2.longbench-write-abs}{10.8}
\setval{olmo.saw.minus.gdpo-noaux.judge2.writing}{+9.8}
\setval{olmo.saw.minus.gdpo-noaux.judge2.writing-abs}{9.8}
\setval{olmo.saw.minus.gdpo-noaux.judge2.writingbench}{+8.7}
\setval{olmo.saw.minus.gdpo-noaux.judge2.writingbench-abs}{8.7}
\setval{olmo.saw.minus.gdpo-noaux.livecodebench}{$-$9.7}
\setval{olmo.saw.minus.gdpo-noaux.livecodebench-abs}{9.7}
\setval{olmo.saw.minus.gdpo-noaux.longbench-write}{+8.9}
\setval{olmo.saw.minus.gdpo-noaux.longbench-write-abs}{8.9}
\setval{olmo.saw.minus.gdpo-noaux.math}{$-$10.0}
\setval{olmo.saw.minus.gdpo-noaux.math-abs}{10.0}
\setval{olmo.saw.minus.gdpo-noaux.math500}{$-$6.4}
\setval{olmo.saw.minus.gdpo-noaux.math500-abs}{6.4}
\setval{olmo.saw.minus.gdpo-noaux.mbpp-plus}{$-$5.1}
\setval{olmo.saw.minus.gdpo-noaux.mbpp-plus-abs}{5.1}
\setval{olmo.saw.minus.gdpo-noaux.minerva}{$-$5.8}
\setval{olmo.saw.minus.gdpo-noaux.minerva-abs}{5.8}
\setval{olmo.saw.minus.gdpo-noaux.olympiad-bench}{$-$9.9}
\setval{olmo.saw.minus.gdpo-noaux.olympiad-bench-abs}{9.9}
\setval{olmo.saw.minus.gdpo-noaux.writing}{+10.2}
\setval{olmo.saw.minus.gdpo-noaux.writing-abs}{10.2}
\setval{olmo.saw.minus.gdpo-noaux.writingbench}{+11.9}
\setval{olmo.saw.minus.gdpo-noaux.writingbench-abs}{11.9}
\setval{olmo.saw.minus.gdpo-w0.5.aime24}{$-$3.4}
\setval{olmo.saw.minus.gdpo-w0.5.aime24-abs}{3.4}
\setval{olmo.saw.minus.gdpo-w0.5.aime25}{$-$2.1}
\setval{olmo.saw.minus.gdpo-w0.5.aime25-abs}{2.1}
\setval{olmo.saw.minus.gdpo-w0.5.amc23}{$-$7.1}
\setval{olmo.saw.minus.gdpo-w0.5.amc23-abs}{7.1}
\setval{olmo.saw.minus.gdpo-w0.5.avg}{$-$1.2}
\setval{olmo.saw.minus.gdpo-w0.5.avg-abs}{1.2}
\setval{olmo.saw.minus.gdpo-w0.5.code}{$-$1.3}
\setval{olmo.saw.minus.gdpo-w0.5.code-abs}{1.3}
\setval{olmo.saw.minus.gdpo-w0.5.creative-writing-v3}{$-$0.2}
\setval{olmo.saw.minus.gdpo-w0.5.creative-writing-v3-abs}{0.2}
\setval{olmo.saw.minus.gdpo-w0.5.guidebench}{+2.8}
\setval{olmo.saw.minus.gdpo-w0.5.guidebench-abs}{2.8}
\setval{olmo.saw.minus.gdpo-w0.5.humaneval-plus}{$-$1.4}
\setval{olmo.saw.minus.gdpo-w0.5.humaneval-plus-abs}{1.4}
\setval{olmo.saw.minus.gdpo-w0.5.if}{+1.3}
\setval{olmo.saw.minus.gdpo-w0.5.if-abs}{1.3}
\setval{olmo.saw.minus.gdpo-w0.5.ifbench}{+0.0}
\setval{olmo.saw.minus.gdpo-w0.5.ifbench-abs}{0.0}
\setval{olmo.saw.minus.gdpo-w0.5.ifeval}{+1.1}
\setval{olmo.saw.minus.gdpo-w0.5.ifeval-abs}{1.1}
\setval{olmo.saw.minus.gdpo-w0.5.judge2.avg}{$-$1.1}
\setval{olmo.saw.minus.gdpo-w0.5.judge2.avg-abs}{1.1}
\setval{olmo.saw.minus.gdpo-w0.5.judge2.creative-writing-v3}{+1.6}
\setval{olmo.saw.minus.gdpo-w0.5.judge2.creative-writing-v3-abs}{1.6}
\setval{olmo.saw.minus.gdpo-w0.5.judge2.longbench-write}{+0.8}
\setval{olmo.saw.minus.gdpo-w0.5.judge2.longbench-write-abs}{0.8}
\setval{olmo.saw.minus.gdpo-w0.5.judge2.writing}{+1.4}
\setval{olmo.saw.minus.gdpo-w0.5.judge2.writing-abs}{1.4}
\setval{olmo.saw.minus.gdpo-w0.5.judge2.writingbench}{+1.7}
\setval{olmo.saw.minus.gdpo-w0.5.judge2.writingbench-abs}{1.7}
\setval{olmo.saw.minus.gdpo-w0.5.livecodebench}{$-$2.9}
\setval{olmo.saw.minus.gdpo-w0.5.livecodebench-abs}{2.9}
\setval{olmo.saw.minus.gdpo-w0.5.longbench-write}{+1.2}
\setval{olmo.saw.minus.gdpo-w0.5.longbench-write-abs}{1.2}
\setval{olmo.saw.minus.gdpo-w0.5.math}{$-$3.3}
\setval{olmo.saw.minus.gdpo-w0.5.math-abs}{3.3}
\setval{olmo.saw.minus.gdpo-w0.5.math500}{$-$2.6}
\setval{olmo.saw.minus.gdpo-w0.5.math500-abs}{2.6}
\setval{olmo.saw.minus.gdpo-w0.5.mbpp-plus}{+0.5}
\setval{olmo.saw.minus.gdpo-w0.5.mbpp-plus-abs}{0.5}
\setval{olmo.saw.minus.gdpo-w0.5.minerva}{$-$1.2}
\setval{olmo.saw.minus.gdpo-w0.5.minerva-abs}{1.2}
\setval{olmo.saw.minus.gdpo-w0.5.olympiad-bench}{$-$3.7}
\setval{olmo.saw.minus.gdpo-w0.5.olympiad-bench-abs}{3.7}
\setval{olmo.saw.minus.gdpo-w0.5.writing}{+0.6}
\setval{olmo.saw.minus.gdpo-w0.5.writing-abs}{0.6}
\setval{olmo.saw.minus.gdpo-w0.5.writingbench}{+0.9}
\setval{olmo.saw.minus.gdpo-w0.5.writingbench-abs}{0.9}
\setval{olmo.saw.minus.gdpo.aime24}{$-$1.7}
\setval{olmo.saw.minus.gdpo.aime24-abs}{1.7}
\setval{olmo.saw.minus.gdpo.aime25}{+2.6}
\setval{olmo.saw.minus.gdpo.aime25-abs}{2.6}
\setval{olmo.saw.minus.gdpo.amc23}{$-$2.7}
\setval{olmo.saw.minus.gdpo.amc23-abs}{2.7}
\setval{olmo.saw.minus.gdpo.avg}{+0.6}
\setval{olmo.saw.minus.gdpo.avg-abs}{0.6}
\setval{olmo.saw.minus.gdpo.code}{$-$0.2}
\setval{olmo.saw.minus.gdpo.code-abs}{0.2}
\setval{olmo.saw.minus.gdpo.creative-writing-v3}{+0.8}
\setval{olmo.saw.minus.gdpo.creative-writing-v3-abs}{0.8}
\setval{olmo.saw.minus.gdpo.guidebench}{+3.8}
\setval{olmo.saw.minus.gdpo.guidebench-abs}{3.8}
\setval{olmo.saw.minus.gdpo.humaneval-plus}{+0.0}
\setval{olmo.saw.minus.gdpo.humaneval-plus-abs}{0.0}
\setval{olmo.saw.minus.gdpo.if}{+3.0}
\setval{olmo.saw.minus.gdpo.if-abs}{3.0}
\setval{olmo.saw.minus.gdpo.ifbench}{+3.0}
\setval{olmo.saw.minus.gdpo.ifbench-abs}{3.0}
\setval{olmo.saw.minus.gdpo.ifeval}{+2.2}
\setval{olmo.saw.minus.gdpo.ifeval-abs}{2.2}
\setval{olmo.saw.minus.gdpo.judge2.avg}{+0.7}
\setval{olmo.saw.minus.gdpo.judge2.avg-abs}{0.7}
\setval{olmo.saw.minus.gdpo.judge2.creative-writing-v3}{+3.2}
\setval{olmo.saw.minus.gdpo.judge2.creative-writing-v3-abs}{3.2}
\setval{olmo.saw.minus.gdpo.judge2.longbench-write}{+1.7}
\setval{olmo.saw.minus.gdpo.judge2.longbench-write-abs}{1.7}
\setval{olmo.saw.minus.gdpo.judge2.writing}{+2.1}
\setval{olmo.saw.minus.gdpo.judge2.writing-abs}{2.1}
\setval{olmo.saw.minus.gdpo.judge2.writingbench}{+1.4}
\setval{olmo.saw.minus.gdpo.judge2.writingbench-abs}{1.4}
\setval{olmo.saw.minus.gdpo.livecodebench}{$-$1.3}
\setval{olmo.saw.minus.gdpo.livecodebench-abs}{1.3}
\setval{olmo.saw.minus.gdpo.longbench-write}{+1.7}
\setval{olmo.saw.minus.gdpo.longbench-write-abs}{1.7}
\setval{olmo.saw.minus.gdpo.math}{$-$0.8}
\setval{olmo.saw.minus.gdpo.math-abs}{0.8}
\setval{olmo.saw.minus.gdpo.math500}{$-$0.8}
\setval{olmo.saw.minus.gdpo.math500-abs}{0.8}
\setval{olmo.saw.minus.gdpo.mbpp-plus}{+0.7}
\setval{olmo.saw.minus.gdpo.mbpp-plus-abs}{0.7}
\setval{olmo.saw.minus.gdpo.minerva}{+0.0}
\setval{olmo.saw.minus.gdpo.minerva-abs}{0.0}
\setval{olmo.saw.minus.gdpo.olympiad-bench}{$-$2.1}
\setval{olmo.saw.minus.gdpo.olympiad-bench-abs}{2.1}
\setval{olmo.saw.minus.gdpo.writing}{+1.5}
\setval{olmo.saw.minus.gdpo.writing-abs}{1.5}
\setval{olmo.saw.minus.gdpo.writingbench}{+2.0}
\setval{olmo.saw.minus.gdpo.writingbench-abs}{2.0}
\setval{olmo.saw.minus.geoma.aime24}{+2.4}
\setval{olmo.saw.minus.geoma.aime24-abs}{2.4}
\setval{olmo.saw.minus.geoma.aime25}{+3.2}
\setval{olmo.saw.minus.geoma.aime25-abs}{3.2}
\setval{olmo.saw.minus.geoma.amc23}{+3.0}
\setval{olmo.saw.minus.geoma.amc23-abs}{3.0}
\setval{olmo.saw.minus.geoma.avg}{+1.5}
\setval{olmo.saw.minus.geoma.avg-abs}{1.5}
\setval{olmo.saw.minus.geoma.code}{+0.6}
\setval{olmo.saw.minus.geoma.code-abs}{0.6}
\setval{olmo.saw.minus.geoma.creative-writing-v3}{$-$1.7}
\setval{olmo.saw.minus.geoma.creative-writing-v3-abs}{1.7}
\setval{olmo.saw.minus.geoma.guidebench}{+8.1}
\setval{olmo.saw.minus.geoma.guidebench-abs}{8.1}
\setval{olmo.saw.minus.geoma.humaneval-plus}{+1.7}
\setval{olmo.saw.minus.geoma.humaneval-plus-abs}{1.7}
\setval{olmo.saw.minus.geoma.if}{+3.7}
\setval{olmo.saw.minus.geoma.if-abs}{3.7}
\setval{olmo.saw.minus.geoma.ifbench}{+2.3}
\setval{olmo.saw.minus.geoma.ifbench-abs}{2.3}
\setval{olmo.saw.minus.geoma.ifeval}{+0.6}
\setval{olmo.saw.minus.geoma.ifeval-abs}{0.6}
\setval{olmo.saw.minus.geoma.judge2.avg}{+1.5}
\setval{olmo.saw.minus.geoma.judge2.avg-abs}{1.5}
\setval{olmo.saw.minus.geoma.judge2.creative-writing-v3}{$-$0.8}
\setval{olmo.saw.minus.geoma.judge2.creative-writing-v3-abs}{0.8}
\setval{olmo.saw.minus.geoma.judge2.longbench-write}{$-$0.8}
\setval{olmo.saw.minus.geoma.judge2.longbench-write-abs}{0.8}
\setval{olmo.saw.minus.geoma.judge2.writing}{$-$0.6}
\setval{olmo.saw.minus.geoma.judge2.writing-abs}{0.6}
\setval{olmo.saw.minus.geoma.judge2.writingbench}{$-$0.2}
\setval{olmo.saw.minus.geoma.judge2.writingbench-abs}{0.2}
\setval{olmo.saw.minus.geoma.livecodebench}{$-$0.5}
\setval{olmo.saw.minus.geoma.livecodebench-abs}{0.5}
\setval{olmo.saw.minus.geoma.longbench-write}{$-$0.4}
\setval{olmo.saw.minus.geoma.longbench-write-abs}{0.4}
\setval{olmo.saw.minus.geoma.math}{+1.9}
\setval{olmo.saw.minus.geoma.math-abs}{1.9}
\setval{olmo.saw.minus.geoma.math500}{+1.3}
\setval{olmo.saw.minus.geoma.math500-abs}{1.3}
\setval{olmo.saw.minus.geoma.mbpp-plus}{+0.7}
\setval{olmo.saw.minus.geoma.mbpp-plus-abs}{0.7}
\setval{olmo.saw.minus.geoma.minerva}{+0.4}
\setval{olmo.saw.minus.geoma.minerva-abs}{0.4}
\setval{olmo.saw.minus.geoma.olympiad-bench}{+1.1}
\setval{olmo.saw.minus.geoma.olympiad-bench-abs}{1.1}
\setval{olmo.saw.minus.geoma.writing}{$-$0.6}
\setval{olmo.saw.minus.geoma.writing-abs}{0.6}
\setval{olmo.saw.minus.geoma.writingbench}{+0.4}
\setval{olmo.saw.minus.geoma.writingbench-abs}{0.4}
\setval{olmo.saw.minus.grpo-nodiv.aime24}{$-$5.6}
\setval{olmo.saw.minus.grpo-nodiv.aime24-abs}{5.6}
\setval{olmo.saw.minus.grpo-nodiv.aime25}{$-$2.4}
\setval{olmo.saw.minus.grpo-nodiv.aime25-abs}{2.4}
\setval{olmo.saw.minus.grpo-nodiv.amc23}{$-$6.2}
\setval{olmo.saw.minus.grpo-nodiv.amc23-abs}{6.2}
\setval{olmo.saw.minus.grpo-nodiv.avg}{$-$1.5}
\setval{olmo.saw.minus.grpo-nodiv.avg-abs}{1.5}
\setval{olmo.saw.minus.grpo-nodiv.code}{$-$0.9}
\setval{olmo.saw.minus.grpo-nodiv.code-abs}{0.9}
\setval{olmo.saw.minus.grpo-nodiv.creative-writing-v3}{+0.9}
\setval{olmo.saw.minus.grpo-nodiv.creative-writing-v3-abs}{0.9}
\setval{olmo.saw.minus.grpo-nodiv.guidebench}{$-$0.8}
\setval{olmo.saw.minus.grpo-nodiv.guidebench-abs}{0.8}
\setval{olmo.saw.minus.grpo-nodiv.humaneval-plus}{$-$1.0}
\setval{olmo.saw.minus.grpo-nodiv.humaneval-plus-abs}{1.0}
\setval{olmo.saw.minus.grpo-nodiv.if}{+0.5}
\setval{olmo.saw.minus.grpo-nodiv.if-abs}{0.5}
\setval{olmo.saw.minus.grpo-nodiv.ifbench}{+1.3}
\setval{olmo.saw.minus.grpo-nodiv.ifbench-abs}{1.3}
\setval{olmo.saw.minus.grpo-nodiv.ifeval}{+0.8}
\setval{olmo.saw.minus.grpo-nodiv.ifeval-abs}{0.8}
\setval{olmo.saw.minus.grpo-nodiv.judge2.avg}{$-$1.6}
\setval{olmo.saw.minus.grpo-nodiv.judge2.avg-abs}{1.6}
\setval{olmo.saw.minus.grpo-nodiv.judge2.creative-writing-v3}{+2.2}
\setval{olmo.saw.minus.grpo-nodiv.judge2.creative-writing-v3-abs}{2.2}
\setval{olmo.saw.minus.grpo-nodiv.judge2.longbench-write}{$-$3.7}
\setval{olmo.saw.minus.grpo-nodiv.judge2.longbench-write-abs}{3.7}
\setval{olmo.saw.minus.grpo-nodiv.judge2.writing}{+0.0}
\setval{olmo.saw.minus.grpo-nodiv.judge2.writing-abs}{0.0}
\setval{olmo.saw.minus.grpo-nodiv.judge2.writingbench}{+1.5}
\setval{olmo.saw.minus.grpo-nodiv.judge2.writingbench-abs}{1.5}
\setval{olmo.saw.minus.grpo-nodiv.livecodebench}{$-$1.7}
\setval{olmo.saw.minus.grpo-nodiv.livecodebench-abs}{1.7}
\setval{olmo.saw.minus.grpo-nodiv.longbench-write}{$-$2.6}
\setval{olmo.saw.minus.grpo-nodiv.longbench-write-abs}{2.6}
\setval{olmo.saw.minus.grpo-nodiv.math}{$-$3.9}
\setval{olmo.saw.minus.grpo-nodiv.math-abs}{3.9}
\setval{olmo.saw.minus.grpo-nodiv.math500}{$-$3.0}
\setval{olmo.saw.minus.grpo-nodiv.math500-abs}{3.0}
\setval{olmo.saw.minus.grpo-nodiv.mbpp-plus}{+0.1}
\setval{olmo.saw.minus.grpo-nodiv.mbpp-plus-abs}{0.1}
\setval{olmo.saw.minus.grpo-nodiv.minerva}{$-$1.5}
\setval{olmo.saw.minus.grpo-nodiv.minerva-abs}{1.5}
\setval{olmo.saw.minus.grpo-nodiv.olympiad-bench}{$-$4.6}
\setval{olmo.saw.minus.grpo-nodiv.olympiad-bench-abs}{4.6}
\setval{olmo.saw.minus.grpo-nodiv.writing}{+0.8}
\setval{olmo.saw.minus.grpo-nodiv.writing-abs}{0.8}
\setval{olmo.saw.minus.grpo-nodiv.writingbench}{+4.2}
\setval{olmo.saw.minus.grpo-nodiv.writingbench-abs}{4.2}
\setval{olmo.saw.minus.grpo.aime24}{$-$3.3}
\setval{olmo.saw.minus.grpo.aime24-abs}{3.3}
\setval{olmo.saw.minus.grpo.aime25}{+0.5}
\setval{olmo.saw.minus.grpo.aime25-abs}{0.5}
\setval{olmo.saw.minus.grpo.amc23}{$-$4.0}
\setval{olmo.saw.minus.grpo.amc23-abs}{4.0}
\setval{olmo.saw.minus.grpo.avg}{$-$0.8}
\setval{olmo.saw.minus.grpo.avg-abs}{0.8}
\setval{olmo.saw.minus.grpo.code}{$-$0.6}
\setval{olmo.saw.minus.grpo.code-abs}{0.6}
\setval{olmo.saw.minus.grpo.creative-writing-v3}{$-$1.5}
\setval{olmo.saw.minus.grpo.creative-writing-v3-abs}{1.5}
\setval{olmo.saw.minus.grpo.guidebench}{$-$1.7}
\setval{olmo.saw.minus.grpo.guidebench-abs}{1.7}
\setval{olmo.saw.minus.grpo.humaneval-plus}{$-$1.0}
\setval{olmo.saw.minus.grpo.humaneval-plus-abs}{1.0}
\setval{olmo.saw.minus.grpo.if}{+0.0}
\setval{olmo.saw.minus.grpo.if-abs}{0.0}
\setval{olmo.saw.minus.grpo.ifbench}{+0.7}
\setval{olmo.saw.minus.grpo.ifbench-abs}{0.7}
\setval{olmo.saw.minus.grpo.ifeval}{+1.2}
\setval{olmo.saw.minus.grpo.ifeval-abs}{1.2}
\setval{olmo.saw.minus.grpo.judge2.avg}{$-$0.5}
\setval{olmo.saw.minus.grpo.judge2.avg-abs}{0.5}
\setval{olmo.saw.minus.grpo.judge2.creative-writing-v3}{+2.4}
\setval{olmo.saw.minus.grpo.judge2.creative-writing-v3-abs}{2.4}
\setval{olmo.saw.minus.grpo.judge2.longbench-write}{+0.3}
\setval{olmo.saw.minus.grpo.judge2.longbench-write-abs}{0.3}
\setval{olmo.saw.minus.grpo.judge2.writing}{+0.9}
\setval{olmo.saw.minus.grpo.judge2.writing-abs}{0.9}
\setval{olmo.saw.minus.grpo.judge2.writingbench}{$-$0.1}
\setval{olmo.saw.minus.grpo.judge2.writingbench-abs}{0.1}
\setval{olmo.saw.minus.grpo.livecodebench}{$-$0.5}
\setval{olmo.saw.minus.grpo.livecodebench-abs}{0.5}
\setval{olmo.saw.minus.grpo.longbench-write}{+0.7}
\setval{olmo.saw.minus.grpo.longbench-write-abs}{0.7}
\setval{olmo.saw.minus.grpo.math}{$-$1.5}
\setval{olmo.saw.minus.grpo.math-abs}{1.5}
\setval{olmo.saw.minus.grpo.math500}{$-$0.5}
\setval{olmo.saw.minus.grpo.math500-abs}{0.5}
\setval{olmo.saw.minus.grpo.mbpp-plus}{$-$0.4}
\setval{olmo.saw.minus.grpo.mbpp-plus-abs}{0.4}
\setval{olmo.saw.minus.grpo.minerva}{+0.4}
\setval{olmo.saw.minus.grpo.minerva-abs}{0.4}
\setval{olmo.saw.minus.grpo.olympiad-bench}{$-$1.8}
\setval{olmo.saw.minus.grpo.olympiad-bench-abs}{1.8}
\setval{olmo.saw.minus.grpo.writing}{$-$0.6}
\setval{olmo.saw.minus.grpo.writing-abs}{0.6}
\setval{olmo.saw.minus.grpo.writingbench}{$-$1.1}
\setval{olmo.saw.minus.grpo.writingbench-abs}{1.1}
\setval{olmo.saw.minus.mt-grpo.aime24}{$-$10.4}
\setval{olmo.saw.minus.mt-grpo.aime24-abs}{10.4}
\setval{olmo.saw.minus.mt-grpo.aime25}{$-$2.9}
\setval{olmo.saw.minus.mt-grpo.aime25-abs}{2.9}
\setval{olmo.saw.minus.mt-grpo.amc23}{$-$10.5}
\setval{olmo.saw.minus.mt-grpo.amc23-abs}{10.5}
\setval{olmo.saw.minus.mt-grpo.avg}{$-$2.3}
\setval{olmo.saw.minus.mt-grpo.avg-abs}{2.3}
\setval{olmo.saw.minus.mt-grpo.code}{+0.1}
\setval{olmo.saw.minus.mt-grpo.code-abs}{0.1}
\setval{olmo.saw.minus.mt-grpo.creative-writing-v3}{$-$2.8}
\setval{olmo.saw.minus.mt-grpo.creative-writing-v3-abs}{2.8}
\setval{olmo.saw.minus.mt-grpo.guidebench}{+0.2}
\setval{olmo.saw.minus.mt-grpo.guidebench-abs}{0.2}
\setval{olmo.saw.minus.mt-grpo.humaneval-plus}{+0.6}
\setval{olmo.saw.minus.mt-grpo.humaneval-plus-abs}{0.6}
\setval{olmo.saw.minus.mt-grpo.if}{+0.8}
\setval{olmo.saw.minus.mt-grpo.if-abs}{0.8}
\setval{olmo.saw.minus.mt-grpo.ifbench}{+0.7}
\setval{olmo.saw.minus.mt-grpo.ifbench-abs}{0.7}
\setval{olmo.saw.minus.mt-grpo.ifeval}{+1.6}
\setval{olmo.saw.minus.mt-grpo.ifeval-abs}{1.6}
\setval{olmo.saw.minus.mt-grpo.judge2.avg}{$-$2.0}
\setval{olmo.saw.minus.mt-grpo.judge2.avg-abs}{2.0}
\setval{olmo.saw.minus.mt-grpo.judge2.creative-writing-v3}{+1.1}
\setval{olmo.saw.minus.mt-grpo.judge2.creative-writing-v3-abs}{1.1}
\setval{olmo.saw.minus.mt-grpo.judge2.longbench-write}{$-$3.0}
\setval{olmo.saw.minus.mt-grpo.judge2.longbench-write-abs}{3.0}
\setval{olmo.saw.minus.mt-grpo.judge2.writing}{$-$0.3}
\setval{olmo.saw.minus.mt-grpo.judge2.writing-abs}{0.3}
\setval{olmo.saw.minus.mt-grpo.judge2.writingbench}{+0.9}
\setval{olmo.saw.minus.mt-grpo.judge2.writingbench-abs}{0.9}
\setval{olmo.saw.minus.mt-grpo.livecodebench}{$-$0.7}
\setval{olmo.saw.minus.mt-grpo.livecodebench-abs}{0.7}
\setval{olmo.saw.minus.mt-grpo.longbench-write}{$-$1.4}
\setval{olmo.saw.minus.mt-grpo.longbench-write-abs}{1.4}
\setval{olmo.saw.minus.mt-grpo.math}{$-$5.4}
\setval{olmo.saw.minus.mt-grpo.math-abs}{5.4}
\setval{olmo.saw.minus.mt-grpo.math500}{$-$2.4}
\setval{olmo.saw.minus.mt-grpo.math500-abs}{2.4}
\setval{olmo.saw.minus.mt-grpo.mbpp-plus}{+0.4}
\setval{olmo.saw.minus.mt-grpo.mbpp-plus-abs}{0.4}
\setval{olmo.saw.minus.mt-grpo.minerva}{$-$1.6}
\setval{olmo.saw.minus.mt-grpo.minerva-abs}{1.6}
\setval{olmo.saw.minus.mt-grpo.olympiad-bench}{$-$4.2}
\setval{olmo.saw.minus.mt-grpo.olympiad-bench-abs}{4.2}
\setval{olmo.saw.minus.mt-grpo.writing}{$-$1.5}
\setval{olmo.saw.minus.mt-grpo.writing-abs}{1.5}
\setval{olmo.saw.minus.mt-grpo.writingbench}{$-$0.3}
\setval{olmo.saw.minus.mt-grpo.writingbench-abs}{0.3}
\setval{olmo.saw.minus.papo.aime24}{+4.0}
\setval{olmo.saw.minus.papo.aime24-abs}{4.0}
\setval{olmo.saw.minus.papo.aime25}{+4.9}
\setval{olmo.saw.minus.papo.aime25-abs}{4.9}
\setval{olmo.saw.minus.papo.amc23}{+7.5}
\setval{olmo.saw.minus.papo.amc23-abs}{7.5}
\setval{olmo.saw.minus.papo.avg}{+2.1}
\setval{olmo.saw.minus.papo.avg-abs}{2.1}
\setval{olmo.saw.minus.papo.code}{+0.6}
\setval{olmo.saw.minus.papo.code-abs}{0.6}
\setval{olmo.saw.minus.papo.creative-writing-v3}{+0.9}
\setval{olmo.saw.minus.papo.creative-writing-v3-abs}{0.9}
\setval{olmo.saw.minus.papo.guidebench}{$-$3.9}
\setval{olmo.saw.minus.papo.guidebench-abs}{3.9}
\setval{olmo.saw.minus.papo.humaneval-plus}{+1.6}
\setval{olmo.saw.minus.papo.humaneval-plus-abs}{1.6}
\setval{olmo.saw.minus.papo.if}{+0.2}
\setval{olmo.saw.minus.papo.if-abs}{0.2}
\setval{olmo.saw.minus.papo.ifbench}{+3.0}
\setval{olmo.saw.minus.papo.ifbench-abs}{3.0}
\setval{olmo.saw.minus.papo.ifeval}{+1.4}
\setval{olmo.saw.minus.papo.ifeval-abs}{1.4}
\setval{olmo.saw.minus.papo.judge2.avg}{+2.3}
\setval{olmo.saw.minus.papo.judge2.avg-abs}{2.3}
\setval{olmo.saw.minus.papo.judge2.creative-writing-v3}{+3.0}
\setval{olmo.saw.minus.papo.judge2.creative-writing-v3-abs}{3.0}
\setval{olmo.saw.minus.papo.judge2.longbench-write}{+1.8}
\setval{olmo.saw.minus.papo.judge2.longbench-write-abs}{1.8}
\setval{olmo.saw.minus.papo.judge2.writing}{+1.9}
\setval{olmo.saw.minus.papo.judge2.writing-abs}{1.9}
\setval{olmo.saw.minus.papo.judge2.writingbench}{+0.9}
\setval{olmo.saw.minus.papo.judge2.writingbench-abs}{0.9}
\setval{olmo.saw.minus.papo.livecodebench}{+0.0}
\setval{olmo.saw.minus.papo.livecodebench-abs}{0.0}
\setval{olmo.saw.minus.papo.longbench-write}{+0.2}
\setval{olmo.saw.minus.papo.longbench-write-abs}{0.2}
\setval{olmo.saw.minus.papo.math}{+4.3}
\setval{olmo.saw.minus.papo.math-abs}{4.3}
\setval{olmo.saw.minus.papo.math500}{+3.6}
\setval{olmo.saw.minus.papo.math500-abs}{3.6}
\setval{olmo.saw.minus.papo.mbpp-plus}{+0.3}
\setval{olmo.saw.minus.papo.mbpp-plus-abs}{0.3}
\setval{olmo.saw.minus.papo.minerva}{+1.8}
\setval{olmo.saw.minus.papo.minerva-abs}{1.8}
\setval{olmo.saw.minus.papo.olympiad-bench}{+4.1}
\setval{olmo.saw.minus.papo.olympiad-bench-abs}{4.1}
\setval{olmo.saw.minus.papo.writing}{+1.0}
\setval{olmo.saw.minus.papo.writing-abs}{1.0}
\setval{olmo.saw.minus.papo.writingbench}{+2.0}
\setval{olmo.saw.minus.papo.writingbench-abs}{2.0}
\setval{olmo.saw.passing-at-ceiling.code.early}{0.84}
\setval{olmo.saw.passing-at-ceiling.code.late}{0.85}
\setval{olmo.saw.passing-at-ceiling.code.run}{0.84}
\setval{olmo.saw.passing-at-ceiling.if.early}{0.48}
\setval{olmo.saw.passing-at-ceiling.if.late}{0.57}
\setval{olmo.saw.passing-at-ceiling.if.run}{0.53}
\setval{olmo.saw.passing-at-ceiling.math.early}{1.00}
\setval{olmo.saw.passing-at-ceiling.math.late}{1.00}
\setval{olmo.saw.passing-at-ceiling.math.run}{1.00}
\setval{olmo.saw.passing-at-ceiling.writing.early}{0.00}
\setval{olmo.saw.passing-at-ceiling.writing.late}{0.00}
\setval{olmo.saw.passing-at-ceiling.writing.run}{0.00}
\setval{olmo.saw.primary-cosine.code.early}{0.92}
\setval{olmo.saw.primary-cosine.code.late}{0.90}
\setval{olmo.saw.primary-cosine.code.run}{0.92}
\setval{olmo.saw.primary-cosine.if.early}{0.75}
\setval{olmo.saw.primary-cosine.if.late}{0.76}
\setval{olmo.saw.primary-cosine.if.run}{0.76}
\setval{olmo.saw.primary-cosine.math.early}{0.97}
\setval{olmo.saw.primary-cosine.math.late}{0.90}
\setval{olmo.saw.primary-cosine.math.run}{0.91}
\setval{olmo.saw.primary-cosine.writing.early}{0.86}
\setval{olmo.saw.primary-cosine.writing.late}{0.68}
\setval{olmo.saw.primary-cosine.writing.run}{0.72}
\setval{olmo.saw.primary-level.code.early}{0.58}
\setval{olmo.saw.primary-level.code.late}{0.82}
\setval{olmo.saw.primary-level.code.run}{0.75}
\setval{olmo.saw.primary-level.if.early}{0.63}
\setval{olmo.saw.primary-level.if.late}{0.74}
\setval{olmo.saw.primary-level.if.run}{0.69}
\setval{olmo.saw.primary-level.math.early}{0.34}
\setval{olmo.saw.primary-level.math.late}{0.64}
\setval{olmo.saw.primary-level.math.run}{0.55}
\setval{olmo.saw.primary-level.writing.early}{0.49}
\setval{olmo.saw.primary-level.writing.late}{0.69}
\setval{olmo.saw.primary-level.writing.run}{0.64}
\setval{olmo.saw.solved.code.early}{0.14}
\setval{olmo.saw.solved.code.late}{0.15}
\setval{olmo.saw.solved.code.run}{0.13}
\setval{olmo.saw.solved.if.early}{0.18}
\setval{olmo.saw.solved.if.late}{0.17}
\setval{olmo.saw.solved.if.run}{0.17}
\setval{olmo.saw.solved.math.early}{0.07}
\setval{olmo.saw.solved.math.late}{0.19}
\setval{olmo.saw.solved.math.run}{0.16}
\setval{olmo.saw.solved.writing.early}{0.00}
\setval{olmo.saw.solved.writing.late}{0.00}
\setval{olmo.saw.solved.writing.run}{0.00}
\setval{olmo.saw.update-share.code.early}{0.18}
\setval{olmo.saw.update-share.code.late}{0.17}
\setval{olmo.saw.update-share.code.run}{0.17}
\setval{olmo.saw.update-share.if.early}{0.19}
\setval{olmo.saw.update-share.if.late}{0.18}
\setval{olmo.saw.update-share.if.run}{0.19}
\setval{olmo.saw.update-share.math.early}{0.19}
\setval{olmo.saw.update-share.math.late}{0.26}
\setval{olmo.saw.update-share.math.run}{0.23}
\setval{olmo.saw.update-share.writing.early}{0.45}
\setval{olmo.saw.update-share.writing.late}{0.39}
\setval{olmo.saw.update-share.writing.run}{0.41}
\setval{olmo.start.aux-level.code}{0.81}
\setval{olmo.start.aux-level.if}{0.87}
\setval{olmo.start.aux-level.math}{0.90}
\setval{olmo.start.aux-level.writing}{0.95}
\setval{olmo.start.aux-level.writing-rm}{0.57}
\setval{olmo.train.step}{299}

\title{CONCUR: Concordant Auxiliary Rewards for Multi-Task, Multi-Reward Reinforcement Learning}

\author{
\makebox[\textwidth][c]{
\textbf{Jixuan Leng}$^{1}$\thanks{Work done while interning at Google.}
\qquad
\textbf{Si Si}$^{2}$
\qquad
\textbf{Fei Wang}$^{3}$
\qquad
\textbf{Cho-Jui Hsieh}$^{4}$
}
\\[0.5em]
\makebox[\textwidth][c]{
\textbf{Jiong Zhang}$^{3}$
\qquad
\textbf{Hsiang-Fu Yu}$^{3}$
\qquad
\textbf{Inderjit S. Dhillon}$^{3}$
}
\\[1em]
$^{1}$CMU \quad
$^{2}$Google DeepMind \quad
$^{3}$Google \quad
$^{4}$UCLA
}

\begin{document}
\maketitle
\begin{abstract}
When a Large Language Model (LLM) is trained with Reinforcement Learning (RL) on multiple tasks, a task's primary reward can \emph{gate} auxiliary rewards such as a length reward: on a failing response, an auxiliary cannot exceed the primary.
However, group-relative advantage estimators treat a gated auxiliary like any other component, so it can reverse the primary's order, and at the primary's ceiling only auxiliaries order responses.
Keeping gated auxiliaries may hurt some tasks, while dropping them may hurt others.
We propose \concur, an advantage estimator that replaces the standardized scores of each gated auxiliary with the closest scores that satisfy two conditions: responses with higher primary rewards cannot receive lower scores, and responses at the primary's ceiling receive equal scores.
\concur-T applies both conditions throughout training, and \concur-A lifts the second for an auxiliary that agrees with the primary's order more often than it reverses it among passing responses.
On three models from two families and across math, code, instruction following, and writing, both variants score higher on average than all baselines and than when all the gated auxiliaries are dropped.
\end{abstract}

\section{Introduction}\label{sec:intro}
\vspace{-0.2em}

Reinforcement Learning (RL) is now a standard way to improve the capabilities of Large Language Models (LLMs)~\citep{guo2025deepseekr1,lambert2025tulu3}.
One setup trains a single model on several tasks in one run~\citep{pardinas2026apriel,ramesh2026mtgrpo}, where each task can score a \emph{response} with several \emph{reward components}.
We study this multi-task, multi-reward setup, in which one component of each task is the \emph{primary} that measures the task's own success, such as the correctness of a math answer.
Some of the others are auxiliary components that the primary \emph{gates}.
We follow the reward design of \citet{bai2026rdpo}, in which a response \emph{passes} if its primary is at least one half, and a \emph{gated auxiliary} keeps its own score on a response that passes and cannot exceed the primary on one that fails.
Group-relative advantage estimators treat a gated auxiliary like any other component.
For example, GDPO~\citep{liu2026gdpo} standardizes each component within a \emph{group} of responses to the same prompt and sums the standardized scores at the \emph{declared weights} specified by the reward design.

\begin{figure*}[t]
\centering
\includegraphics[width=\textwidth]{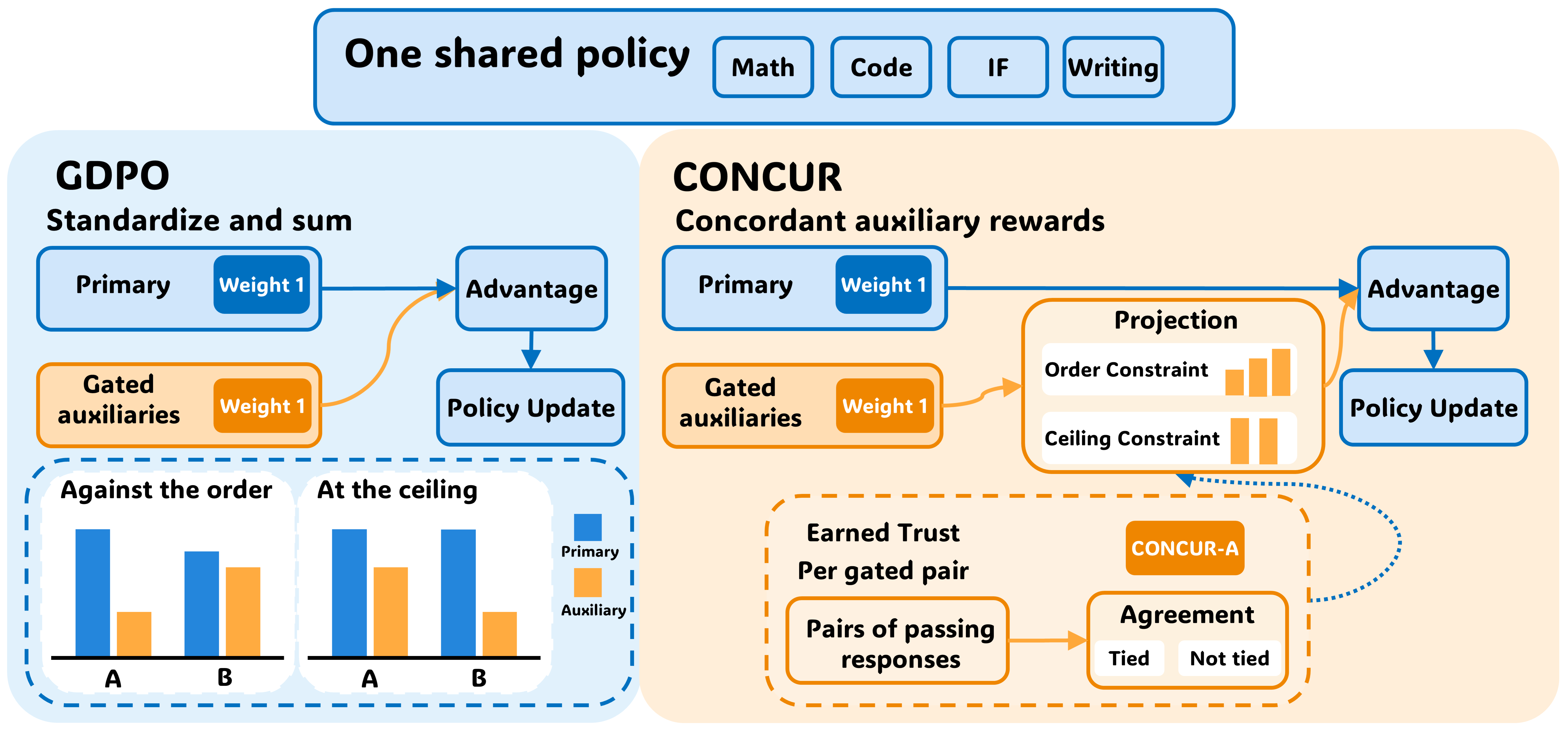}\vspace{-0.3em}
\caption{Left: GDPO standardizes reward components in a group and sums them at the declared weights, and the insets show where an auxiliary departs from its primary.
Right: \concur projects each gated auxiliary's scores onto the closest ones that concur with the primary, and in \concur-A earned trust decides whether a ceiling is tied.}
\label{fig:framework}
\vspace{-0.8em}
\end{figure*}

What should an estimator do with a gated auxiliary?
We train three LLMs with GDPO on a mixture of math, code, instruction following (IF), and writing, in which every task's primary gates a length reward and writing's primary also gates a reward-model score (RM).
We call a task together with one of its gated auxiliaries a \emph{gated pair}, and the mixture has five.
Neither keeping nor dropping the gated auxiliaries is satisfactory.
As a \emph{reference}, we run GDPO with every gated auxiliary at weight $0$.
GDPO at the declared weights scores well below the reference on math and below it on code, but above it on writing.
The left of \fig{fig:framework} illustrates the two places where a gated auxiliary departs from its primary during training: against the primary's order, where a response with a lower primary scores higher, and \emph{at the ceiling}, where the primary equals its maximum and cannot separate responses.
We measure each as a share of the sum of an auxiliary's squared standardized scores.
As \fig{fig:gate} shows, the share against the order grows to more than half on writing, and the share at the ceiling is largest on math, whose binary primary scores correct responses equally.
At the ceiling the auxiliary acts alone. In a group whose responses all reach it, the primary's standardized scores are zero, while standardization scales the auxiliary's raw differences, however small, to unit variance.

The gate gives the primary priority over the auxiliaries that it gates, but it constrains only the responses that fail.
We propose \concur, an advantage estimator that extends this priority to every pair of responses in a group, as the right of \fig{fig:framework} illustrates.
\concur does this with a projection: within each group, it replaces each gated auxiliary's standardized scores with the closest scores that \emph{concur} with the primary.
Scores concur if they satisfy two constraints. The \emph{order constraint} requires that no response with a higher primary scores lower, and the \emph{ceiling constraint}, where it applies, requires that responses at the ceiling score alike.
When the ceiling constraint applies to a gated pair, we say that the pair's ceiling is \emph{tied}.
\concur-T ties every gated pair's ceiling throughout training, has no setting and no state, and is our default.
\concur-A is the adaptive form, and we call its rule \emph{earned trust}. It lifts a gated pair's tie while the auxiliary agrees with the primary's order more often than it reverses it among the passing responses that the primary separates.
\concur uses no gradient and keeps every reward component at its declared weight.

Our contributions are summarized as follows:
\begin{itemize}[leftmargin=*,nosep]
\item We analyze how a gated auxiliary enters group-relative advantages, show that in training it departs from its primary both against the primary's order and at its ceiling, and find that neither keeping nor dropping gated auxiliaries is satisfactory.
\item We propose \concur, an advantage estimator that projects each gated auxiliary's standardized scores onto the closest scores that concur with the primary, in a default and an adaptive variant.
\item We prove that the projection never reverses the primary's order within a group and that, with a binary primary, it keeps only the auxiliary's difference between correct and wrong responses.
\item We show that, on three models from two families and across math, code, instruction-following, and writing benchmarks, both variants achieve higher scores on average than seven existing estimators and than when the gated auxiliaries are dropped.
\end{itemize}

\section{Related Work}\label{sec:related}
\vspace{-0.2em}

\textbf{Multi-reward RL.}
Group-relative estimators differ in where they normalize: GRPO standardizes the weighted sum within a group, GDPO and MO-GRPO standardize each component before the sum, and RDPO maps each component to quantile scores~\citep{shao2024deepseekmath,liu2026gdpo,ichihara2025mogrpo,bai2026rdpo}.
Others account for the correlation between components~\citep{anonymous2026geoma,hu2026corrgrpo}, change the weights of losses, tasks, or reward components during training~\citep{kendall2018multi,chen2018gradnorm,lu2026dynamic,ramesh2026mtgrpo,he2026saw}, or filter out responses whose per-component advantages disagree in sign~\citep{liu2026gd2po}.
Two recent works drop the division by the standard deviation of the rewards in a group~\citep{kassenaar2026learning,wang2026darkroom}, as Dr.\ GRPO already does~\citep{liu2025understanding}.
All these methods treat every reward component alike, whereas \concur changes only the gated auxiliaries' scores in each response group.

\textbf{Priority among rewards.}
A common reward design, called conditioning or gating, lets one reward count in full only when another one reaches a threshold~\citep{kimi2025k15,arora2025training,aggarwal2025l1,song2025walk,liu2026gdpo,bai2026rdpo}.
\concur takes such a gate as given and makes its priority hold for any two responses of a group.
Other designs let a second reward order the correct responses only~\citep{tan2026papo,peng2026ddca,li2026drpo,fang2026orpg}.
On a binary primary, they make the opposite choice from \concur. They keep the auxiliary's differences among correct responses, which \concur's ceiling constraint removes, and drop its difference between correct and wrong responses, which \concur keeps.
Appendix~\ref{app:related} discusses other ways to enforce a priority, such as constrained RL, lexicographic RL, and methods that act in gradient space.

\textbf{Order-preserving projections.}
Isotonic regression finds the values that respect a given order and are closest to the input values, and the pool-adjacent-violators algorithm computes it along a chain~\citep{ayer1955empirical}.
The most related work is POISE, which applies this projection to the score labels in the fixed training corpus of a robot reward model, so that they follow the order of pairwise preference labels~\citep{wang2026trustroboreward}.
\concur projects at every training step onto the order of the primary, which is another reward component of the same responses, and adds the ceiling constraint.

\section{Preliminaries}\label{sec:prelim}
\vspace{-0.2em}

\begin{figure*}[t]
\centering
\includegraphics[width=\textwidth]{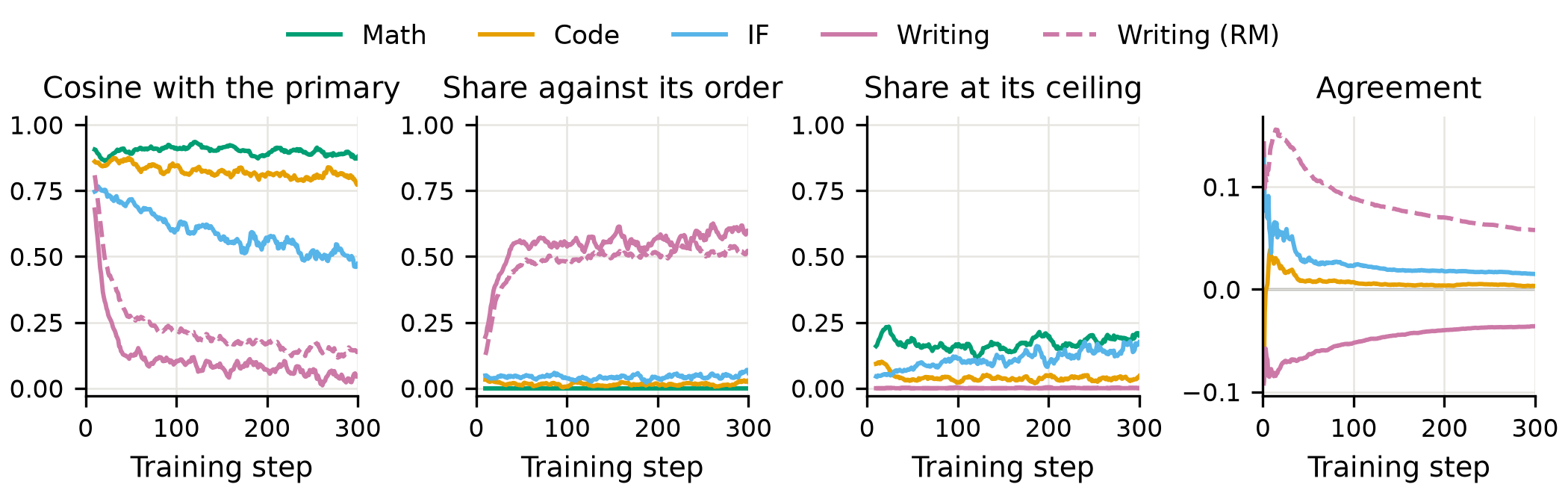}\vspace{-0.3em}
\caption{GDPO at the declared weights on \qwenfour{}, one line per gated pair.
The panels show the auxiliary's cosine with its primary, its shares of energy against the order and at the ceiling from \eqn{eq:shares}, as 10-step running means, and the cumulative mean of the agreement of \eqn{eq:agreement}, which is not defined for Math's binary primary.}
\label{fig:gate}
\vspace{-0.8em}
\end{figure*}

\subsection{Multi-Task RL with Gated Rewards}\label{sec:setting}
\vspace{-0.2em}
The policy $\pi_\theta$ samples a \emph{group} of $G$ responses to each prompt $x$ of task $t\in\mathcal T$.
Each \emph{reward component} $k\in\mathcal K_t$ gives response $i$ a score $r_{ik}\in[0,1]$ and has a \emph{declared weight} $w_k$.
One component of each task, indexed by $j$, is its \emph{primary}, which measures the task's own success.
We write $p_i=r_{ij}$ for the primary score of response $i$, and the primary \emph{separates} two responses $i$ and $i'$ if $p_i\ne p_{i'}$.
A response \emph{passes} if $p_i\ge\frac12$ and \emph{fails} otherwise, and it is \emph{at the ceiling} if $p_i=1$, which is the primary's maximum.
The primary \emph{gates} a set $\mathcal A_t\subset\mathcal K_t\setminus\{j\}$ of \emph{gated auxiliaries}: $r_{ik}=\min\{s_{ik},p_i\}$ if $p_i<\frac12$ and $r_{ik}=s_{ik}$ otherwise, where $s_{ik}\in[0,1]$ is the auxiliary's own score.
The components, their gates, and the declared weights form the \emph{reward design}.

We mix equal shares of math, code, IF, and writing prompts, with every declared weight $1$.
The primary is binary correctness for math, the unit-test pass rate for code, and a weighted share of the criteria that a response meets for IF and writing~\citep{li2026rubrichub}.
A program checks IF's criteria.
Writing's criteria form a rubric, and a judge model decides which of them a response meets.
Every task's primary gates a \emph{length reward}, which is full for a response no longer than a per-prompt reference length and falls quadratically to zero at twice that length, and writing's primary also gates a \emph{reward-model score} (RM)~\citep{liu2026skywork}, rescaled to $[0,1]$.
Each gated auxiliary and its task together form a \emph{gated pair}, of which the mixture has five: Math, Code, IF, and Writing are each a task with its own length reward, and Writing (RM) is writing with its RM.
IF also has one component that is not gated, which is the judge's score on one further criterion.
Appendix~\ref{app:roles} shows every component's role, and Appendix~\ref{app:data} gives the data and rewards.

\textbf{Group-relative advantages.}\label{sec:gdpo}
GDPO extends GRPO by standardizing each reward component within a group before taking the weighted sum:
\begin{align}
z_{ik}&=\frac{r_{ik}-\bar r_k}{d_k},\qquad u_i=\sum_{k\in\mathcal K_t}w_kz_{ik},\label{eq:gdpo}\\
A_i&=\W(u)_i=(u_i-\bar u)/\hat\sigma_u,\label{eq:whiten}
\end{align}
where $\bar r_k$ and $d_k$ are the group's mean and sample standard deviation, $z_{ik}=0$ if $d_k=0$, and the \emph{batch whitening} $\W$, which preserves the order of $u$, computes $\bar u$ and $\hat\sigma_u$ over all tasks.
We call $z_{ik}$ a \emph{standardized score} and $u_i$ a \emph{summed score}, and we write $z_k=(z_{ik})_i\in\R^G$ for the group as a whole.

\subsection{A Motivating Experiment}\label{sec:diagnosis}
\vspace{-0.2em}
We train \qwenfour{}~\citep{yang2025qwen3} with GDPO at the declared weights and, on every training batch, compare each gated auxiliary's standardized scores $z$ with its primary's.
\fig{fig:gate} plots the comparison, and \tab{tab:gate} in the appendix gives its values.
We also train GDPO with every gated auxiliary at weight $0$, which is a different reward design.
We call this run the \emph{reference} and label it GDPO (no aux.).
An auxiliary's \emph{energy} in a group is $\lVert z\rVert^2$.
Let $\hat z^{\le}$ be the scores closest to $z$ in which no response with a higher primary scores lower, and let $\hat z^{=}$ be the closest scores that are also equal on the responses at the ceiling.
The energy of every nonzero $z$ then splits into exactly three nonnegative shares that sum to one (more in Section~\ref{sec:theory}):
\begin{equation}\label{eq:shares}
\begin{aligned}
\text{against the order}&=1-\lVert\hat z^{\le}\rVert^2/\lVert z\rVert^2,\\
\text{at the ceiling}&=\bigl(\lVert\hat z^{\le}\rVert^2-\lVert\hat z^{=}\rVert^2\bigr)/\lVert z\rVert^2,\\
\text{kept}&=\lVert\hat z^{=}\rVert^2/\lVert z\rVert^2.
\end{aligned}
\end{equation}
We also measure whether an auxiliary orders passing responses as its primary does.
Take a group and a gated pair of task $t$ with auxiliary $k$, and let $P=\{i:p_i\ge\frac12\}$ be the group's passing responses.
If the primary separates at least two of them, the group gives a \emph{reading} of the pair's \emph{agreement},
\begin{equation}\label{eq:agreement}
a_{x,k}=\frac{\sum_{(i,i')\in P^2}\operatorname{sign}(p_i-p_{i'})\operatorname{sign}(r_{ik}-r_{i'k})}{\lvert\{(i,i')\in P^2:p_i\ne p_{i'}\}\rvert}.
\end{equation}
Among the pairs in $P$ that the primary separates, it is the share that the auxiliary orders as the primary does minus the share that it orders the other way.

\textbf{(O1) At the start of training, a gated auxiliary mostly repeats its primary on math, code, and IF.}
In the first panel, the cosine between an auxiliary's standardized scores and its primary's is at least \val{4b.gdpo.gate.if.cosine.early} on Math, Code, and IF over the first 20 training steps.
Even a high cosine leaves room for a large effect on training. On math, where this cosine is highest, the reference scores well above GDPO.

\textbf{(O2) An auxiliary departs from its primary in two places: against its order and at its ceiling.}
On Math, the share against the order is zero by construction, because the gate sets a wrong response's auxiliary to $0$.
It stays small on Code and IF, and on the two writing pairs it grows during training to more than half.
The share at the ceiling is largest on Math.
Over the last 20 training steps, a share \val{4b.gdpo.solved.math.late} of the math groups in the update have every response at the ceiling.
There the primary is constant, and the other components alone order the group.

\textbf{(O3) Dropping the gated auxiliaries raises the score on math and lowers it on writing.}
On our evaluation benchmarks, the reference scores \val{4b.gdpo-noaux.minus.gdpo.math-abs} points above GDPO on math, above it on code as well, and \val{4b.gdpo-noaux.minus.gdpo.writing-abs} points below it on writing (more in Section~\ref{sec:reference}).
Whether an auxiliary agrees with its primary varies by pair. In \fig{fig:gate}, the agreement ends positive for writing's RM and negative for writing's length reward, each by a few hundredths.

These observations motivate the two constraints of \concur: its order constraint removes the share against the order, and its ceiling constraint, wherever it applies, removes the share at the ceiling.

\section{\texorpdfstring{\underline{Conc}ordant A\underline{u}xiliary \underline{R}ewards}{Concordant Auxiliary Rewards}}\label{sec:method}
\vspace{-0.2em}
Within each group, \concur replaces each gated auxiliary's standardized scores with the closest scores that \emph{concur} with the primary.
Scores concur if they satisfy the order constraint and, at the steps where the pair's ceiling is \emph{tied}, the ceiling constraint as well.
The \emph{ceiling policy} decides when a pair's ceiling is tied, and the two variants of \concur differ only in this policy.
\concur-T uses the constant policy that keeps all ceilings tied, and we recommend it as the default variant.
The adaptive form \concur-A lifts a pair's tie while the pair's auxiliary has \emph{earned trust}.
The other components keep their standardized scores, and the sum of all scores at the declared weights is then whitened over the whole batch.
\concur needs nothing beyond the reward design. It uses the gate as it is declared, without having to know what a given gated auxiliary rewards.
Algorithm~\ref{alg:concur} in the appendix states the full computation for one batch.

\subsection{The Projection}\label{sec:project}
\vspace{-0.2em}
Take one group, with primaries $p\in[0,1]^G$.
The vectors of scores $v\in\R^G$ that satisfy the order constraint form the \emph{order cone} $C_{\le}(p)=\{v\in\R^G:\ p_i>p_{i'}\;\Rightarrow\;v_i\ge v_{i'}\}$.
The order constraint has no margin: any difference in the primary, however small, outranks the auxiliary.
At the ceiling, equal primaries are not an artifact of sampling, since a prompt that the policy solves puts its whole group at the ceiling, where the components other than the primary determine the entire order and each has unit variance.
The ceiling therefore gets its own constraint, and $C_{=}(p)=\{v\in C_{\le}(p):\ p_i=p_{i'}=1\;\Rightarrow\;v_i=v_{i'}\}$ is the cone of those vectors that satisfy both the order constraint and this one.

\concur moves an auxiliary's standardized scores as little as the constraints allow: for each gated auxiliary $k\in\mathcal A_t$ and for each group, it takes the Euclidean projection of the group's scores $z_k$,
\begin{equation}\label{eq:project}
\hat z_k=\Pi_{C}(z_k)=\argmin_{v\in C}\lVert v-z_k\rVert^2,
\end{equation}
where $C=C_{=}(p)$ if the pair's ceiling is tied at that training step and $C=C_{\le}(p)$ if it is not, and the summed score $u_i$ then uses $\hat z_k$ in place of $z_k$:
\begin{equation}\label{eq:concur}
\begin{aligned}
u_i&=\sum_{k\in\mathcal K_t\setminus\mathcal A_t}w_kz_{ik}+\sum_{k\in\mathcal A_t}w_k\hat z_{ik},\\
A_i&=\W(u)_i.
\end{aligned}
\end{equation}
The projection exists, is unique, and leaves $z_k$ unchanged exactly when $z_k$ concurs (more in Section~\ref{sec:theory}).
Its exact computation has two steps, which Appendix~\ref{app:projection} details.
First, if the ceiling is tied, the scores of the responses at the ceiling are replaced by their mean.
Then the pool-adjacent-violators algorithm of isotonic regression pools neighboring responses whose scores are out of order along the primary, giving each pool its mean.
The projected scores are not standardized again. That would restore unit variance to what the projection leaves.

\subsection{Earned Trust}\label{sec:trust}
\vspace{-0.2em}
Tying a ceiling has a cost. It removes all the credit that the auxiliary assigns among the responses at the ceiling, which can be most of a task's passing responses.
\concur-A restores this credit for a gated pair whose auxiliary orders passing responses as the primary does, which it measures in every group where the primary separates two of them.

For a gated pair $(t,k)$, the \emph{pooled agreement} $\bar a_{t,k}$ is the mean of all readings $a_{x,k}$ of its agreement since the start of training, including those of the current batch, and $\mathrm{se}_{t,k}$ is its standard error.
A pair's ceiling is tied by default, and it is not tied at a step if $\bar a_{t,k}>2\,\mathrm{se}_{t,k}$.
While its ceiling is not tied, we say that the pair's auxiliary \emph{has earned trust}.
A binary primary scores every passing response alike, so it never gives a reading. Its ceiling stays tied throughout training, and the two variants compute the same summed scores on such a task.
With a graded primary, \concur-A suits a reward design in which an auxiliary that agrees with the primary should keep ordering the responses at the ceiling, and \concur-T suits one in which it should not.
Appendix~\ref{app:trust} details the rule and its assumption.

\section{Analysis}\label{sec:theory}\label{sec:theory-ceiling}\label{sec:theory-project}
\vspace{-0.2em}
Appendix~\ref{app:proofs} provides the derivations and the complete proofs of the two statements in this section.

\textbf{The ceiling.}
Take a group whose responses are all at the ceiling.
The primary is constant there, so $z_{ij}=0$ for every response $i$, while every component $k$ with $d_k>0$ has standardized scores of unit sample variance, whatever the size of its raw differences.
For example, suppose $G=8$ and a binary primary gates one auxiliary, both at weight $1$.
If all eight responses are correct and exactly one of them has a lower auxiliary score than the others, whether by $0.01$ or by $0.5$, that response has $u_i=-7/\sqrt8\approx-2.47$.
If four responses are correct with a full auxiliary score and four are wrong, each wrong response has $u_i=-\sqrt{7/2}\approx-1.87$.
A difference of $0.01$ in the auxiliary thus costs a correct response more summed score in the first group than a wrong answer costs in the second.
Under \concur with the ceiling tied, the gated auxiliary contributes nothing to that first group.

\begin{proposition}[The projection]\label{prop:project}
For primaries $p$, take any vector $z\in\R^G$ and $C\in\{C_{\le}(p),C_{=}(p)\}$.
\begin{enumerate}[label=(\roman*),leftmargin=*,nosep]
\item \emph{Existence.} $C$ is a closed convex cone that contains the constant vectors, so the projection $\hat z=\Pi_C(z)$ of $z$ exists and it is unique.
\item \emph{Computation.} Averaging the scores at a tied ceiling and then pooling the adjacent violators along the primary computes $\hat z$ exactly.
\item \emph{Mean.} $\sum_i\hat z_i=\sum_iz_i$.
\item \emph{Energy.} $\lVert z\rVert^2=\lVert\hat z\rVert^2+\lVert z-\hat z\rVert^2$, and $\hat z=z$ exactly when $z\in C$.
With $\hat z^{\le}=\Pi_{C_{\le}(p)}(z)$ and $\hat z^{=}=\Pi_{C_{=}(p)}(z)$, we have $\lVert\hat z^{=}\rVert\le\lVert\hat z^{\le}\rVert\le\lVert z\rVert$, so for $z\ne0$ the shares of \eqn{eq:shares} are nonnegative and sum to one.
\item \emph{Order.} $(p_i-p_{i'})(\hat z_i-\hat z_{i'})\ge0$ for every pair of responses, so the group covariance of $\hat z$ with any nondecreasing function of the primary is nonnegative, and in particular $\langle\hat z,z_j\rangle\ge0$ for the primary's own scores $z_j$.
\end{enumerate}
\end{proposition}

By (iv), a projected auxiliary has no more energy than under GDPO.
By (v), on a task with positive declared weights whose primary gates every other component, \concur's summed score always orders any two responses of a group with different primaries as the primary does, whereas GDPO's need not.
A component that is not gated, such as IF's judged criterion, can still reverse this order.

\begin{corollary}[A binary primary]\label{cor:binary}
Let the primary of a given task be binary and gate its auxiliary $k$.
\begin{enumerate}[label=(\roman*),leftmargin=*,nosep]
\item A failing response has $r_{ik}=0$, so $z_k\in C_{\le}(p)$: the order constraint changes nothing, and with the ceiling not tied \concur's sum $u$ is GDPO's on any group of the task.
\item With the ceiling tied, in a group with both passes and failures, $\hat z_k=c\,z_j$ with $c=\langle z_k,z_j\rangle/\lVert z_j\rVert^2\in[0,1]$, which is the cosine between $z_k$ and $z_j$ whenever $z_k$ is nonzero.
\item With the ceiling tied, in a group of passes only, $\hat z_k=0$, whatever the scores in $z_k$ are.
\end{enumerate}
\end{corollary}

On a task such as math, whose only components are a binary primary and one gated auxiliary at weight $1$ each, \concur's summed score with the ceiling tied is $u=(1+c)\,z_j$ in a group with passes and failures and $u=0$ in a group of passes only.
These are the summed scores of GDPO without that auxiliary, multiplied by a per-group factor $1+c\in[1,2]$.
The gated auxiliary therefore keeps its difference between passes and failures, which adds to the primary's, and what the projection removes is only its order among the passes within a group.

\section{Experiments}\label{sec:exp}\label{sec:setup}
\vspace{-0.2em}
In this section, we run experiments with \qwenfour{}, \qweneight{}~\citep{yang2025qwen3}, and \olmo{}~\citep{olmo2025olmo3} to address four questions:
(1) Does \concur improve on the estimators at the declared weights?
(2) Does it improve on dropping the gated auxiliaries?
(3) What does each part of \concur add?
(4) Is \concur in effect a change of weight?

\begin{table*}[t]
\centering
\footnotesize
\setlength{\tabcolsep}{2pt}
\renewcommand{\arraystretch}{1.2}
\resizebox{\ifdim\width>\linewidth\linewidth\else\width\fi}{!}{%
\begin{tabular}{lccccccc!{\color{tableblue}\vrule width 1.4pt}cccc!{\color{tableblue}\vrule width 1.4pt}cccc!{\color{tableblue}\vrule width 1.4pt}cccc!{\color{tableblue}\vrule width 1.4pt}c}
\toprule
 & \multicolumn{7}{c!{\hspace{1.4pt}}}{Math} & \multicolumn{4}{c!{\hspace{1.4pt}}}{Code} & \multicolumn{4}{c!{\hspace{1.4pt}}}{IF} & \multicolumn{4}{c!{\hspace{1.4pt}}}{Writing} &  \\
\cmidrule(lr{\dimexpr\cmidrulekern+1.4pt\relax}){2-8} \cmidrule(lr{\dimexpr\cmidrulekern+1.4pt\relax}){9-12} \cmidrule(lr{\dimexpr\cmidrulekern+1.4pt\relax}){13-16} \cmidrule(lr{\dimexpr\cmidrulekern+1.4pt\relax}){17-20}
 & AIME24 & AIME25 & MATH & OBench & Minerva & AMC23 & \textit{Mean} & HEval+ & MBPP+ & LCB-v6 & \textit{Mean} & IFEval & IFBench & GBench & \textit{Mean} & WBench & CW-v3 & LBWrite & \textit{Mean} & Avg \\
\midrule
\addlinespace[2pt]
\multicolumn{21}{>{\columncolor{tablepink}}l}{\textbf{Qwen3-4B-Base}} \\
\addlinespace[2pt]
\quad \textcolor{tablegray}{\textit{Base}} & \textcolor{tablegray}{\textit{\phantom{0}7.81}} & \textcolor{tablegray}{\textit{\phantom{0}7.08}} & \textcolor{tablegray}{\textit{67.26}} & \textcolor{tablegray}{\textit{35.62}} & \textcolor{tablegray}{\textit{31.61}} & \textcolor{tablegray}{\textit{43.91}} & \textcolor{tablegray}{\textit{32.21}} & \textcolor{tablegray}{\textit{65.06}} & \textcolor{tablegray}{\textit{55.58}} & \textcolor{tablegray}{\textit{13.20}} & \textcolor{tablegray}{\textit{44.61}} & \textcolor{tablegray}{\textit{47.36}} & \textcolor{tablegray}{\textit{17.00}} & \textcolor{tablegray}{\textit{24.86}} & \textcolor{tablegray}{\textit{29.74}} & \textcolor{tablegray}{\textit{32.42}} & \textcolor{tablegray}{\textit{24.76}} & \textcolor{tablegray}{\textit{44.44}} & \textcolor{tablegray}{\textit{33.87}} & \textcolor{tablegray}{\textit{34.53}} \\
\quad GRPO & 11.98 & 12.60 & 80.25 & 44.63 & 45.28 & 54.69 & 41.57 & 82.74 & 71.16 & 24.00 & 59.30 & 86.45 & \underline{37.67} & 77.93 & 67.35 & 74.25 & \underline{47.32} & 69.21 & 63.59 & 54.68 \\
\quad GDPO & \phantom{0}9.06 & \phantom{0}9.90 & 79.29 & 44.66 & 45.48 & 59.61 & 41.33 & 81.83 & 70.87 & 24.40 & 59.03 & 85.73 & 36.33 & 80.61 & 67.56 & 71.94 & 45.97 & 69.01 & 62.31 & 54.31 \\
\quad GD$^2$PO & 11.88 & \phantom{0}6.77 & 77.67 & 43.55 & 47.19 & 57.73 & 40.80 & \underline{83.41} & \underline{71.46} & 23.49 & \underline{59.45} & 87.65 & \underline{37.67} & 79.08 & 68.13 & 70.16 & 44.68 & 68.94 & 61.26 & 54.09 \\
\quad GEOMA & 11.56 & 10.63 & 78.85 & 43.84 & 46.94 & 55.39 & 41.20 & 80.24 & 70.21 & \textbf{25.66} & 58.70 & \underline{87.77} & 36.00 & 78.50 & 67.42 & 69.53 & 45.86 & 65.83 & 60.41 & 53.79 \\
\quad SAW & 12.50 & 11.15 & 77.84 & 43.86 & 48.14 & 56.64 & 41.69 & 80.73 & 70.11 & 22.80 & 57.88 & 86.93 & \textbf{40.33} & \textbf{81.00} & \textbf{69.42} & 69.88 & 45.06 & 66.21 & 60.38 & 54.21 \\
\quad MT-GRPO & 10.73 & \phantom{0}6.15 & 76.99 & 42.51 & 45.98 & 59.77 & 40.35 & 79.09 & 70.05 & 22.63 & 57.26 & 86.69 & 34.33 & 79.65 & 66.89 & 70.14 & 44.51 & 66.81 & 60.49 & 53.07 \\
\quad PAPO & \phantom{0}9.27 & \phantom{0}7.50 & 76.53 & 41.48 & 46.53 & 53.52 & 39.14 & 83.05 & 70.71 & 23.54 & 59.10 & 85.85 & 33.33 & \underline{80.71} & 66.63 & 66.50 & 46.53 & 63.97 & 59.00 & 52.60 \\
\rowcolor{tablelightblue}\quad CONCUR-T & \underline{19.48} & \underline{16.35} & \textbf{85.51} & \underline{52.85} & \underline{49.12} & \underline{69.22} & \underline{48.76} & \textbf{84.02} & 68.60 & 24.74 & 59.12 & \underline{87.77} & \underline{37.67} & 79.17 & 68.20 & \textbf{80.19} & \textbf{47.84} & \textbf{77.74} & \textbf{68.59} & \underline{58.69} \\
\rowcolor{tablelightblue}\quad CONCUR-A & \textbf{20.52} & \textbf{18.75} & \underline{85.28} & \textbf{53.58} & \textbf{49.22} & \textbf{69.38} & \textbf{49.45} & 82.44 & \textbf{71.51} & \underline{25.14} & \textbf{59.70} & \textbf{88.13} & \textbf{40.33} & 77.35 & \underline{68.60} & \underline{79.43} & 44.71 & \underline{76.31} & \underline{66.82} & \textbf{58.80} \\
\addlinespace[2pt]
\multicolumn{21}{>{\columncolor{tablepink}}l}{\textbf{Qwen3-8B-Base}} \\
\addlinespace[2pt]
\quad \textcolor{tablegray}{\textit{Base}} & \textcolor{tablegray}{\textit{11.35}} & \textcolor{tablegray}{\textit{\phantom{0}8.54}} & \textcolor{tablegray}{\textit{72.46}} & \textcolor{tablegray}{\textit{39.44}} & \textcolor{tablegray}{\textit{40.91}} & \textcolor{tablegray}{\textit{50.23}} & \textcolor{tablegray}{\textit{37.16}} & \textcolor{tablegray}{\textit{80.06}} & \textcolor{tablegray}{\textit{67.28}} & \textcolor{tablegray}{\textit{20.91}} & \textcolor{tablegray}{\textit{56.08}} & \textcolor{tablegray}{\textit{53.00}} & \textcolor{tablegray}{\textit{17.33}} & \textcolor{tablegray}{\textit{45.59}} & \textcolor{tablegray}{\textit{38.64}} & \textcolor{tablegray}{\textit{36.75}} & \textcolor{tablegray}{\textit{26.72}} & \textcolor{tablegray}{\textit{51.42}} & \textcolor{tablegray}{\textit{38.30}} & \textcolor{tablegray}{\textit{41.47}} \\
\quad GRPO & 15.83 & 15.52 & 80.64 & 47.94 & 49.86 & 64.38 & 45.70 & 83.72 & 74.66 & 26.51 & 61.63 & 86.81 & \underline{42.00} & 81.86 & 70.22 & 76.48 & \underline{52.48} & 69.78 & 66.25 & 57.90 \\
\quad GDPO & 21.35 & 14.38 & 82.99 & 49.32 & 49.61 & 62.19 & 46.64 & 84.09 & 74.52 & 26.23 & 61.61 & 88.13 & 39.33 & \textbf{84.55} & \underline{70.67} & 76.09 & \textbf{52.90} & 69.57 & 66.19 & 58.35 \\
\quad GD$^2$PO & 16.25 & 12.19 & 81.21 & 48.56 & 49.92 & 59.61 & 44.62 & 83.96 & 75.08 & 26.29 & 61.78 & \textbf{89.33} & \textbf{43.00} & 83.69 & \textbf{72.00} & 73.89 & 48.16 & 69.88 & 63.98 & 57.40 \\
\quad GEOMA & 14.06 & 12.50 & 79.63 & 45.99 & 49.77 & 63.52 & 44.24 & 82.74 & \textbf{75.61} & 25.71 & 61.36 & 87.29 & 34.00 & \underline{84.36} & 68.55 & 75.94 & 50.40 & 68.97 & 65.10 & 56.70 \\
\quad SAW & 15.31 & 15.52 & 81.91 & 48.13 & 51.77 & 62.27 & 45.82 & \underline{85.85} & 74.34 & 26.69 & 62.29 & 87.65 & 40.00 & 84.26 & 70.64 & 75.40 & 49.67 & 71.03 & 65.37 & 57.99 \\
\quad MT-GRPO & 14.48 & 15.73 & 80.58 & 45.75 & 51.05 & 65.63 & 45.53 & 85.06 & \underline{75.26} & \textbf{27.66} & \underline{62.66} & 86.93 & 37.33 & 83.11 & 69.12 & 71.70 & 50.33 & 67.45 & 63.16 & 57.20 \\
\quad PAPO & 12.19 & 10.10 & 78.48 & 44.81 & 49.33 & 62.03 & 42.82 & 82.50 & 73.41 & 26.63 & 60.85 & 86.33 & 37.33 & \underline{84.36} & 69.34 & 73.20 & 51.15 & 65.50 & 63.28 & 55.82 \\
\rowcolor{tablelightblue}\quad CONCUR-T & \textbf{31.25} & \textbf{23.96} & \underline{89.61} & \underline{59.55} & \textbf{55.40} & \underline{78.52} & \textbf{56.38} & \textbf{87.38} & 74.87 & \underline{27.20} & \textbf{63.15} & \underline{88.85} & 39.33 & 83.40 & 70.53 & \underline{87.19} & 52.42 & \underline{81.22} & \textbf{73.61} & \textbf{64.01} \\
\rowcolor{tablelightblue}\quad CONCUR-A & \underline{28.75} & \underline{23.13} & \textbf{90.13} & \textbf{60.37} & \underline{54.19} & \textbf{78.91} & \underline{55.91} & 82.74 & 73.20 & 26.34 & 60.76 & 88.13 & 38.67 & 83.40 & 70.06 & \textbf{88.06} & 49.92 & \textbf{81.60} & \underline{73.19} & \underline{63.17} \\
\addlinespace[2pt]
\multicolumn{21}{>{\columncolor{tablepink}}l}{\textbf{OLMo-3-7B-Instruct-SFT}} \\
\addlinespace[2pt]
\quad \textcolor{tablegray}{\textit{Base}} & \textcolor{tablegray}{\textit{\phantom{0}8.54}} & \textcolor{tablegray}{\textit{\phantom{0}7.40}} & \textcolor{tablegray}{\textit{70.12}} & \textcolor{tablegray}{\textit{36.27}} & \textcolor{tablegray}{\textit{35.29}} & \textcolor{tablegray}{\textit{44.53}} & \textcolor{tablegray}{\textit{33.69}} & \textcolor{tablegray}{\textit{69.21}} & \textcolor{tablegray}{\textit{61.24}} & \textcolor{tablegray}{\textit{16.86}} & \textcolor{tablegray}{\textit{49.10}} & \textcolor{tablegray}{\textit{84.41}} & \textcolor{tablegray}{\textit{26.67}} & \textcolor{tablegray}{\textit{61.42}} & \textcolor{tablegray}{\textit{57.50}} & \textcolor{tablegray}{\textit{44.74}} & \textcolor{tablegray}{\textit{37.91}} & \textcolor{tablegray}{\textit{57.47}} & \textcolor{tablegray}{\textit{46.71}} & \textcolor{tablegray}{\textit{44.14}} \\
\quad GRPO & 29.48 & 24.69 & 84.95 & 55.67 & 39.94 & 70.86 & 50.93 & 77.32 & 65.79 & 22.34 & 55.15 & 87.41 & 32.00 & 56.43 & 58.61 & \textbf{68.47} & 49.37 & 64.21 & 60.68 & 55.26 \\
\quad GDPO & 27.81 & 22.60 & 85.24 & 55.89 & 40.36 & 69.53 & 50.24 & 76.34 & 64.76 & 23.14 & 54.75 & 86.45 & 29.67 & 50.86 & 55.66 & 65.31 & 47.14 & 63.27 & 58.57 & 53.89 \\
\quad GD$^2$PO & 28.33 & 24.27 & 85.72 & 56.07 & 40.12 & 72.19 & 51.12 & 76.77 & 64.79 & 23.20 & 54.92 & 87.65 & 32.33 & \textbf{61.13} & \textbf{60.37} & 65.74 & 48.31 & 68.21 & 60.75 & 55.66 \\
\quad GEOMA & 23.75 & 21.98 & 83.18 & 52.74 & 39.97 & 63.91 & 47.59 & 74.63 & 64.74 & 22.34 & 53.90 & \underline{88.01} & 30.33 & 46.64 & 54.99 & 66.99 & \underline{49.59} & 65.32 & 60.63 & 52.94 \\
\quad SAW & 26.15 & 25.21 & 84.49 & 53.82 & 40.34 & 66.88 & 49.48 & 76.34 & 65.42 & 21.89 & 54.55 & \textbf{88.61} & \underline{32.67} & 54.70 & 58.66 & 67.36 & 47.91 & 64.93 & 60.07 & 54.45 \\
\quad MT-GRPO & \textbf{36.56} & 28.13 & 86.91 & 58.04 & 41.96 & 77.42 & 54.84 & 75.73 & 64.97 & 22.63 & 54.44 & 87.05 & 32.00 & 54.51 & 57.85 & \underline{67.69} & \textbf{50.70} & 66.37 & \textbf{61.59} & 56.71 \\
\quad PAPO & 22.19 & 20.31 & 80.88 & 49.74 & 38.56 & 59.38 & 45.17 & 74.76 & 65.08 & 21.89 & 53.91 & 87.17 & 29.67 & \underline{58.64} & 58.49 & 65.35 & 46.98 & 64.73 & 59.02 & 52.35 \\
\rowcolor{tablelightblue}\quad CONCUR-T & 34.48 & \underline{29.90} & \underline{89.41} & \underline{61.57} & \underline{44.05} & \underline{79.84} & \underline{56.54} & \textbf{83.29} & \textbf{71.06} & \underline{29.03} & \textbf{61.13} & 87.65 & \textbf{33.67} & 58.54 & \underline{59.95} & 66.14 & 49.04 & \textbf{69.23} & \underline{61.47} & \underline{59.13} \\
\rowcolor{tablelightblue}\quad CONCUR-A & \underline{36.35} & \textbf{32.19} & \textbf{90.53} & \textbf{62.41} & \textbf{44.81} & \textbf{83.52} & \textbf{58.30} & \underline{82.74} & \underline{67.54} & \textbf{30.34} & \underline{60.21} & 87.77 & 32.00 & 56.91 & 58.89 & 65.45 & 49.21 & \underline{68.49} & 61.05 & \textbf{59.35} \\
\bottomrule
\end{tabular}%
}
\vspace{-0.3em}
\caption{Evaluation scores of the seven estimators and the two variants of \concur. Each model's best score is in bold, its second best is underlined, and the base policy is in gray italics.
Eight benchmark names are shortened: MATH is MATH-500, OBench is OlympiadBench, HEval+ is HumanEval+, LCB-v6 is LiveCodeBench v6, GBench is GuideBench, WBench is WritingBench, CW-v3 is Creative Writing v3, and LBWrite denotes LongBench-Write.}
\label{tab:main}
\vspace{-0.8em}
\end{table*}
\textbf{Baselines.}
We compare \concur with seven other advantage estimators at the declared weights:
\begin{itemize}[leftmargin=*,nosep]
\item \textbf{GRPO}~\citep{shao2024deepseekmath}. Standardizes the weighted sum of the components within a group.
\item \textbf{GDPO}~\citep{liu2026gdpo}. Standardizes each component within a group before summing them.
\item \textbf{GD$^2$PO}~\citep{liu2026gd2po}. Hard variant: drops responses whose component advantages conflict.
\item \textbf{GEOMA}~\citep{anonymous2026geoma}. Aggregates preconditioned components by Nash social welfare.
\item \textbf{SAW}~\citep{he2026saw}. Rescales each component's weight by its batch coefficient of variation.
\item \textbf{MT-GRPO}~\citep{ramesh2026mtgrpo}. Reweights the tasks toward those that are currently weaker.
\item \textbf{PAPO}~\citep{tan2026papo}. Normalizes each process reward among the passing responses only.
\end{itemize}

\textbf{Training.}
We train on the four-task mixture with $G=8$ responses to each prompt.
Following prior work~\citep{khatri2025scalerl,mistral2025magistral}, we exclude from the update the groups whose responses all receive identical component scores.
Every run takes 300 steps of 512 prompts, at a constant learning rate of $10^{-6}$ (more in Appendix~\ref{app:runs}).

\textbf{Evaluation.}
We evaluate the base policy and each run's final checkpoint on \val{eval.count} benchmarks in four \emph{task groups}: \val{eval.count.math} math~\citep{hendrycks2021math,lightman2024verify,lewkowycz2022minerva,he2024olympiadbench}, \val{eval.count.code} code~\citep{chen2021evaluating,austin2021program,liu2023evalplus,jain2025livecodebench}, \val{eval.count.if} IF~\citep{zhou2023instruction,pyatkin2025ifbench,diao2025guidebench}, and \val{eval.count.writing} writing~\citep{wu2025writingbench,paech2025creativewriting,bai2025longwriter}.
A task group's score is the mean over its benchmarks, and \emph{Avg} is the mean over all benchmarks.
We score the writing benchmarks with the judge model used in training, which we call the \emph{training judge}.
The same responses are also scored by a second judge, which comes from another model family and is not used in training.

\subsection{Q1: \texorpdfstring{\concur}{CONCUR} Has the Highest Avg}\label{sec:results}\label{sec:where}
\vspace{-0.2em}
As shown in \tab{tab:main}, on all three models \concur in either variant has the highest Avg of the runs at the declared weights, more than four points above GDPO, and both variants are also the top two under the second judge.
Their largest gain is on math, at least \val{olmo.concur-tied.minus.gdpo.math-abs} points above GDPO and above every estimator.
Five of the other estimators change how components are normalized, weighted, or filtered but treat a gated auxiliary like any other component, and none of them moves Avg by more than three points from GDPO on any model.
PAPO keeps the auxiliary's order among correct responses, which \concur removes on math, and has the lowest math score of the seven estimators on every model.
Under the second judge, both variants score above GDPO on the writing task group on every model and above all seven estimators on the two Qwen models, but below five of them on \olmo{}, as \tab{tab:second-judge} in the appendix shows.

\subsection{Q2: \texorpdfstring{\concur}{CONCUR} Beats Dropping Auxiliaries}\label{sec:reference}
\vspace{-0.2em}
\begin{figure*}[t]
\centering
\begin{minipage}[c]{0.44\textwidth}\centering
\includegraphics[width=\linewidth]{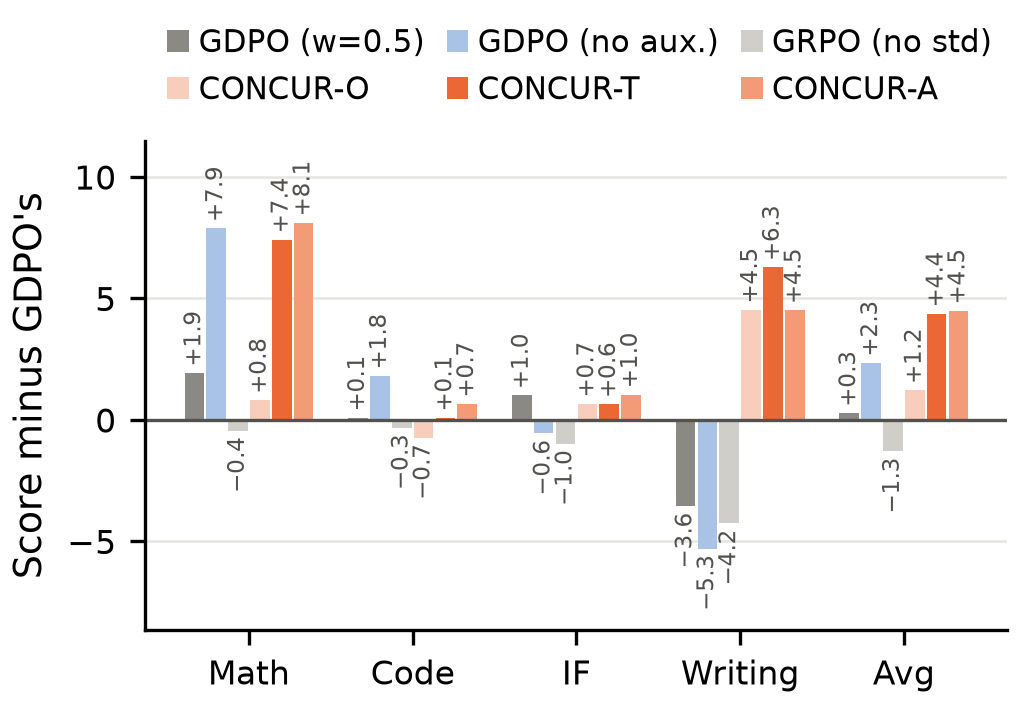}\\[-0.1em]
{\small (a) \qwenfour{}}
\end{minipage}\hspace{0.096\textwidth}
\begin{minipage}[c]{0.44\textwidth}\centering
\includegraphics[width=\linewidth]{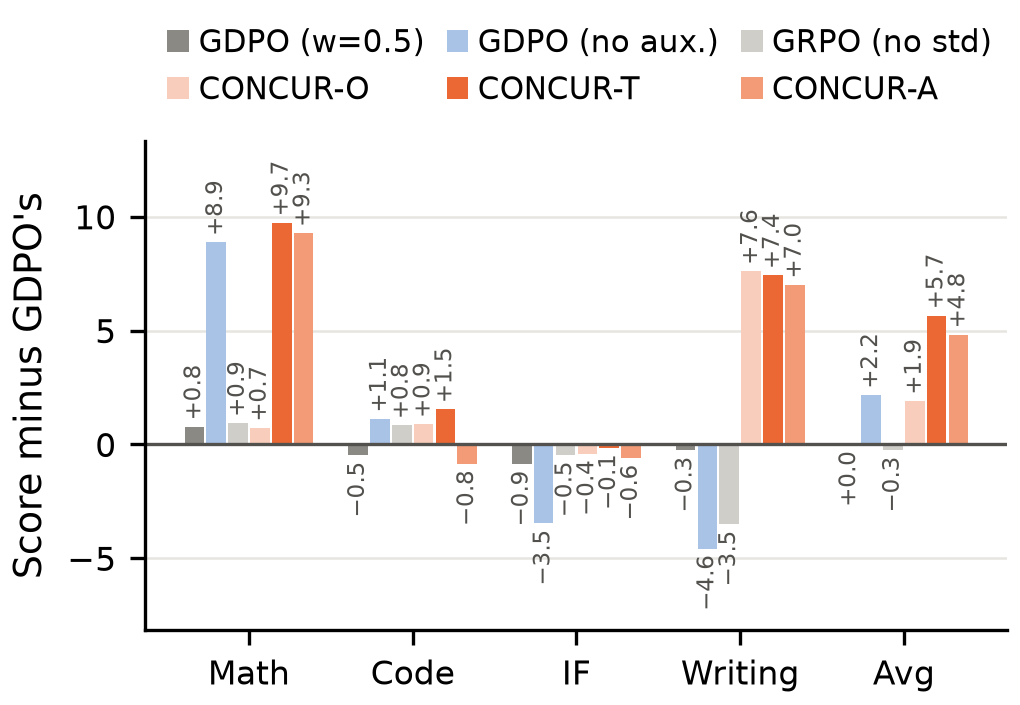}\\[-0.1em]
{\small (b) \qweneight{}}
\end{minipage}\\[0.4em]
\begin{minipage}[c]{0.44\textwidth}\centering
\includegraphics[width=\linewidth]{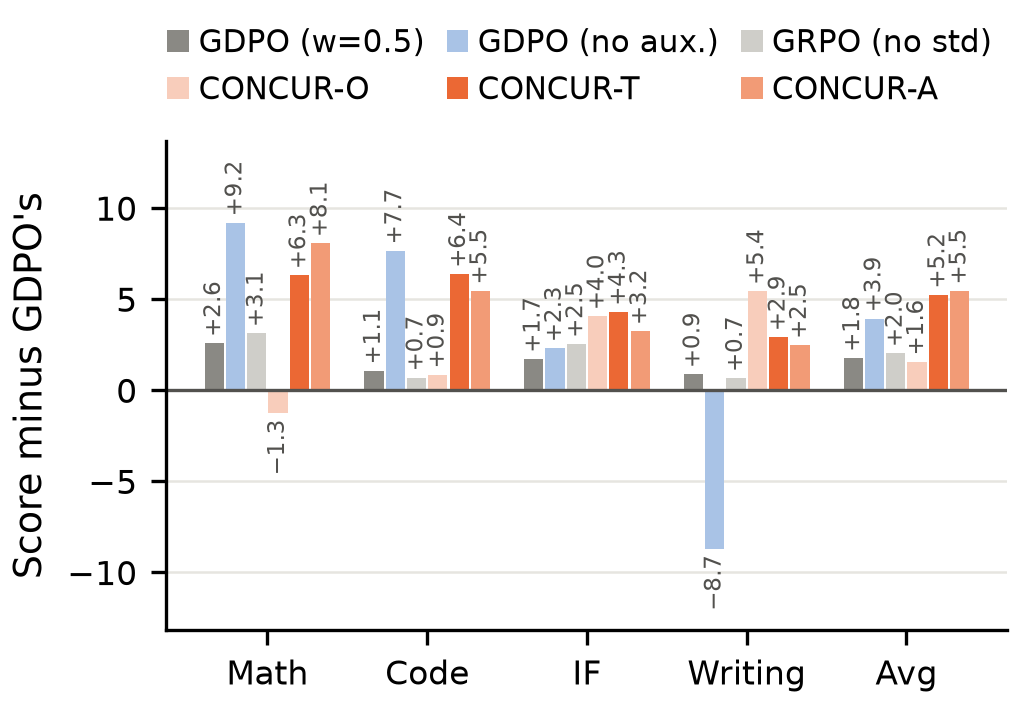}\\[-0.1em]
{\small (c) \olmo{}}
\end{minipage}\hspace{0.03\textwidth}
\begin{minipage}[c]{0.506\textwidth}\centering
\includegraphics[width=\linewidth]{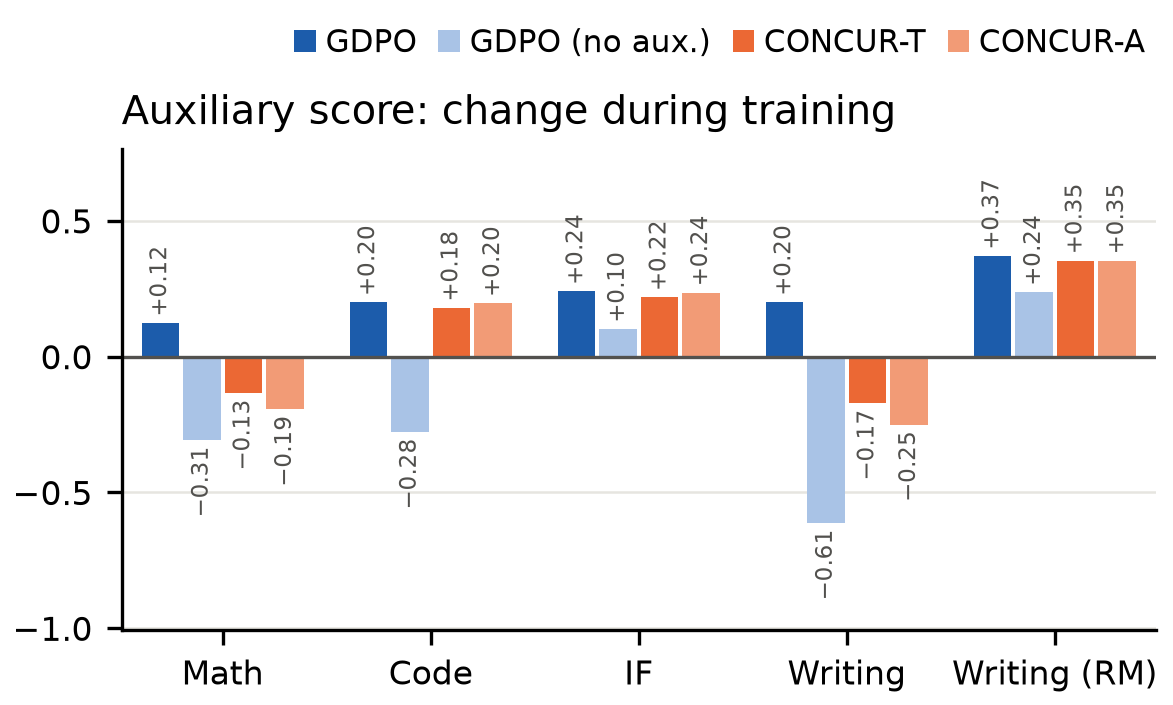}\\[-0.1em]
{\small (d) Auxiliary scores on \qwenfour{}}
\end{minipage}\vspace{-0.3em}
\caption{Panels (a) to (c): each task group's score and Avg minus those of GDPO. Panel (d): each gated auxiliary's mean score over the passing responses, as the mean over the last 20 training steps minus the base policy's level.}
\label{fig:ablation}
\vspace{-0.8em}
\end{figure*}
\fig{fig:ablation}(a) to (c) compare both variants with GDPO and with the reference GDPO (no aux.).
On every model, dropping the gated auxiliaries raises the math score by more than seven points and lowers the writing score by more than four, so neither choice is better on both.
The reference scores below GDPO on all three writing benchmarks and under both judges.
Both variants keep most of the gain on math and score above GDPO on writing, and on Avg they are above both runs.
On \olmo{}, the reference still scores above both variants on math and on code.

\textbf{What happens to the auxiliaries' scores?}
To track what the auxiliaries reward, we look at training on \qwenfour{}.
\fig{fig:ablation}(d) shows how each gated auxiliary's mean score over passing responses changes over training.
Under both variants, this score rises on code, on IF, and for writing's RM about as much as under GDPO, and it falls on math and for writing's length reward.
These two pairs correspond to the two places that the motivating experiment measured: on math, every passing response is at the ceiling, and on writing, more than half of the length reward's energy is against the primary's order.
Writing's RM also has more than half of its energy against the order, but unlike the length reward, it agrees with the primary more often than it reverses it among passing responses.
In the reference, this score ends lower than under either variant on all five pairs, so both variants give up less of what the auxiliaries reward than dropping them does.
The same holds on the other two models, and on \olmo{}, where code's ceiling is tied at every step, this score also falls on code under both variants. Appendix~\ref{app:ablation} shows their panels and every benchmark's scores.

\textbf{What happens to the primary?}
The bottom row of \fig{fig:alignment} tracks the primary's mean score over all responses.
On math and on writing, it rises more under both variants than under GDPO and nearly as much as under the reference, on all three models (more in Appendix~\ref{app:ablation}).
The reference trains these two tasks on the primary alone, so under GDPO the gated auxiliaries hold the primary back, and under \concur they do much less.
On writing, the reference reaches about the same primary as both variants but scores below GDPO on the benchmarks, which also measure what the auxiliaries reward, and \concur keeps more of it.

\subsection{Q3: What Each Part of \texorpdfstring{\concur}{CONCUR} Adds}\label{sec:ablation}
\vspace{-0.2em}
We also run the ablation \concur-O, which applies the order constraint alone. \fig{fig:ablation}(a) to (c) show its scores along with those of the other runs.

\begin{figure*}[t]
\centering
\includegraphics[width=\textwidth]{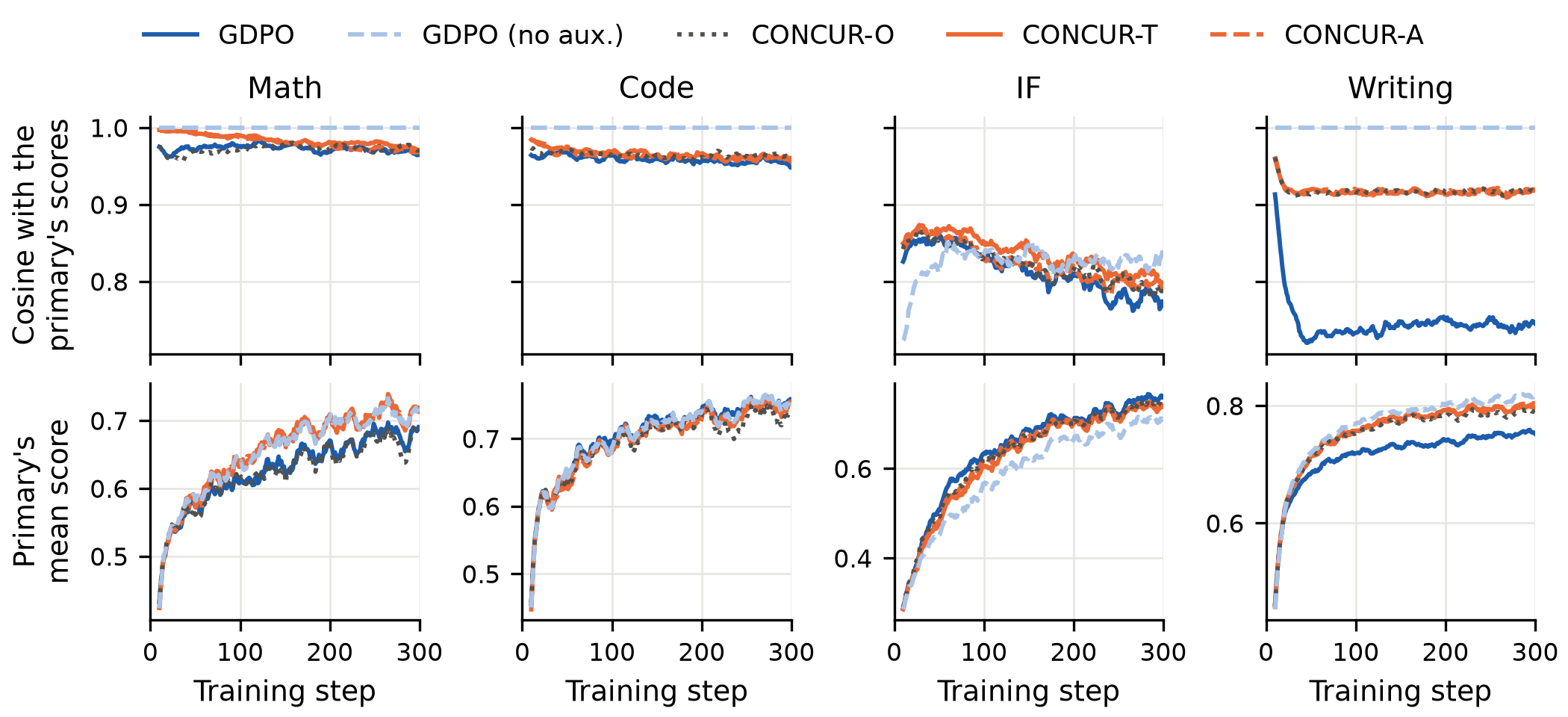}\vspace{-0.3em}
\caption{Cosine between each task's advantages and the standardized scores of its primary (top) and the primary's mean score over all responses (bottom) during training on \qwenfour{}, shown as 10-step running means.}
\label{fig:alignment}
\vspace{-0.8em}
\end{figure*}
\textbf{The order constraint.}
Compared with GDPO, \concur-O raises the writing score on every model, by at least \val{4b.concur-free.minus.gdpo.writing-abs} points, and writing is the task where the share against the order is large.
Under the second judge, \concur-O also scores above GDPO on writing on every model.
Few of writing's passing responses are at the ceiling, so in almost every writing group \concur-T and \concur-A compute what \concur-O computes.
In training, the three runs reach about the same primary on writing, above that of GDPO.
On \qwenfour{}, \fig{fig:alignment} shows that the cosine between writing's advantages and its primary's standardized scores falls under GDPO and stays high under both variants. Over the last 20 steps on all three models, this cosine is at most \val{8b.gdpo.primary-cosine.writing.late} under GDPO and at least \val{4b.concur-tied.primary-cosine.writing.late} under both variants.
On code and IF, where the share against the order stays small, \concur-O scores about as high as GDPO or higher on the benchmarks.
By Corollary~\ref{cor:binary}, \concur-O computes the summed scores of GDPO on every math group, so the order constraint does not act on math.

\textbf{The ceiling constraint.}
\concur-T adds the ceiling constraint on all pairs at all steps, and its math score is at least \val{4b.concur-tied.minus.concur-free.math-abs} points above that of \concur-O.
The gain on math comes from the ceiling constraint. \concur-T, \concur-A, and the reference are the three runs that give math's auxiliary no credit among correct responses, and all three score far above GDPO and \concur-O on math, with a higher math accuracy in training as well.
Yet math takes a smaller part of the update, since the ceiling constraint sets to zero the summed scores of a math group whose responses are all correct, and such a group stays in the batch whitening. Over the last 20 steps, math's share of the batch's summed squared advantages is at least \val{olmo.gdpo.update-share.math.late} under GDPO and at most \val{4b.concur-tied.update-share.math.late} under \concur-T (more in Appendix~\ref{app:mechanism}).
The ceiling constraint changes the other three task groups less, and its largest effect there is on the code score of \olmo{}, which it raises by \val{olmo.concur-tied.minus.concur-free.code-abs} points relative to \concur-O.
Under GDPO, the share at the ceiling on code is \val{olmo.gdpo.gate.code.ceiling.late} on that model and only \val{4b.gdpo.gate.code.ceiling.late} on each of the two Qwen models.

\textbf{The ceiling policy.}\label{sec:trust-results}\label{sec:cross}
The two variants end within one point of each other on Avg.
They differ in what they compute, as \tab{tab:trust} in the appendix details.
\concur-A keeps the tie for math, whose binary primary gives no reading, and for writing's length reward, and lifts it at almost every step for IF and for writing's RM.
On code it lifts the tie at most steps on the two Qwen models and keeps it at every step on \olmo{}, where code's length reward reverses the primary's order more often than it agrees with it. There \concur-A also scores \val{olmo.concur.minus.concur-free.code-abs} points above \concur-O on code.
Lifting a tie changes scores only where passing responses reach the ceiling. IF and code on the two Qwen models have more than half of their passing responses there, while writing's RM has almost none.
On \qweneight{}, code's pooled agreement is barely positive, \concur-A lifts the tie at almost every step, and its code score is \val{8b.concur-tied.minus.concur.code-abs} points below that of \concur-T.
As readings accumulate, the standard error shrinks, so the rule tests only the sign of the pooled agreement. This case is one reason why \concur-T is our default variant.

\subsection{Q4: Is \texorpdfstring{\concur}{CONCUR} a Change of Weight?}\label{sec:mechanism}
\vspace{-0.2em}
A weight scales an auxiliary on every response of every group by one factor, whereas the projection changes an auxiliary's scores within a group, differently in each group, and not at all where they already concur.
On a binary primary that gates one auxiliary, as on math, \concur comes close to a change of weight. By Corollary~\ref{cor:binary}, its summed scores in a group are GDPO's without the auxiliary, scaled by the group's own factor, and the math scores of the two variants are within \val{olmo.concur-tied.minus.gdpo-noaux.math-abs} points of the reference's.
At a smaller positive weight, the auxiliary alone still orders a group whose responses are all at the ceiling.
GDPO ($w=0.5$), which sets every gated auxiliary to weight $0.5$, gains far less on math than the reference does and scores below both variants on Avg on all three models.
The length reward's score on math moves under it about as under GDPO but falls under both variants.
On a graded primary, the projection is not a single factor on the auxiliary. On writing, GDPO at each of the three weights $1$, $0.5$, and $0$ scores below \concur-O and below the two variants on all three models.

\textbf{Dropping the division.}
GRPO (no std), which drops GRPO's division by the group's standard deviation, scores more than six points below the reference on math, and more than three points below both variants on Avg (more in Appendix~\ref{app:ablation}). Without this division, math's auxiliary still gets credit among correct responses, as under GDPO.

\section{Conclusion}\label{sec:conclusion}
\vspace{-0.2em}
\concur extends the gate's priority to every pair of responses in a group: it projects each gated auxiliary's standardized scores onto the closest scores that concur with the primary.
On Avg over \val{eval.count} benchmarks, the default \concur-T and the adaptive \concur-A both score more than four points above GDPO on all three models, and they are also above every other estimator in our comparison and above a reference that drops the gated auxiliaries.
The order constraint alone raises the writing score, the ceiling constraint raises the math score, and the two variants end within one point of each other on Avg.
In training, both variants raise the primary on math and on writing nearly as much as dropping the gated auxiliaries does, and they give up less of what the auxiliaries reward, as every gated auxiliary's mean score over the passing responses ends higher than under the reference on every model.\label{body-end}
\pagebreak

\clearpage
\section*{Limitations}
Our evidence comes from one four-task mixture.
Its gated auxiliaries are four length rewards and one reward-model score.
We test one intermediate weight.
We measure what the auxiliaries reward only on training rollouts, since the benchmarks do not score it.
The order constraint is a design choice, and it is wrong when the designer would let an auxiliary outrank a small difference in the primary.
It constrains how credit is assigned within a group, not the policy's expected primary after an update.
Where the primary is a judge's score, as on writing, \concur makes the gated auxiliaries follow that judge, which may share preferences with writing's reward model~\citep{singhal2024long,dubois2024length}.
The gains over GDPO on the writing task group are smaller under a second judge, as \tab{tab:second-judge} reports.
Earned trust is also a design choice. It assumes that an auxiliary's agreement with the primary where the primary separates passing responses carries over to the ceiling, where it separates none, and our mixture does not test this.

\bibliography{custom}

\appendix
\section{Additional Related Work}\label{app:related}
In this section, we discuss other ways to enforce a priority among rewards and further work on reward models and group-wise objectives.
Constrained RL maximizes one reward subject to a constraint on another~\citep{achiam2017constrained,tessler2019rcpo,dai2024safe,girgis2026constrained,kancheti2026fgrpo}, lexicographic RL maximizes its rewards in a fixed order of priority~\citep{skalse2022lexicographic,tercan2024tlo}, and methods in gradient space weight or project one gradient so that it does not oppose another~\citep{du2018adapting,yu2020gradient,qiu2025lppg,munn2025gcr}.
\concur applies the priority to every pair of responses in a group, directly on their scores, with no gradient and no multiplier.
Reward models and automatic judges can favor longer or more confident responses regardless of quality~\citep{singhal2024long,dubois2024length,leng2025taming}, and in our mixture the RM is an auxiliary that writing's primary gates.
Outside RL, preference optimization goes beyond pairs and contrasts several responses to the same prompt jointly, as a ranked list or as a set~\citep{song2024pro,gupta2024mpo,leng2026groupdpo}.

\section{Method Details}\label{app:method}
In this section, we list the role of every reward component in the mixture, describe how the projection is computed, and specify the rule of earned trust.

\subsection{Gates and Roles}\label{app:roles}
In \tab{tab:roles}, math's primary is binary, and the primaries of code, IF, and writing are graded.
In our implementation, a response is at the ceiling if its primary is within $10^{-6}$ of the maximum value $1$.
\begin{table}[htbp]
\centering
\small
\setlength{\tabcolsep}{4pt}
\resizebox{\ifdim\width>\linewidth\linewidth\else\width\fi}{!}{%
\begin{tabular}{@{}llll@{}}
\toprule
Task & Primary & Gated auxiliaries & Not gated \\
\midrule
Math & correctness (binary) & length reward & -- \\
Code & unit-test pass rate & length reward & -- \\
IF & rule-checked criteria met & length reward & judged criterion \\
Writing & judged criteria met & length reward, RM & -- \\
\bottomrule
\end{tabular}%
}
\caption{Reward components of the mixture: a task's primary gates each auxiliary in its row, all at weight $1$.}
\label{tab:roles}\label{tab:mixed}
\end{table}
\subsection{The Projection}\label{app:projection}
For one group and one gated auxiliary $k$, the projection of the standardized scores $z_k$ in \eqn{eq:project} is computed in two steps, as Algorithm~\ref{alg:concur} states.
\emph{Averaging step.} If the pair's ceiling is tied, the scores of the responses at the ceiling are replaced by their mean.
\emph{Pooling step.} The responses are sorted in ascending order by primary, and those with equal primaries by their score after the averaging step.
The sorted scores are then scanned in order.
Each score starts a block of its own, and while the last block has a lower mean than the one before it, the two merge.
At the end, every response receives the mean of the block that holds it.
Responses that share a primary below the ceiling are not constrained among themselves.
Their equal primaries can be an artifact of sampling, since another response to the same prompt can have a higher primary, and the order constraint then keeps all of their auxiliary scores at or below that response's.
\begin{algorithm}[htbp]
\caption{\concur for one batch. Line~2 is earned trust, which only \concur-A carries out.}\label{alg:concur}
\small
\begin{algorithmic}[1]
\Statex \textbf{Input (\concur-A):} for every gated pair $(t,k)$, the number $n$, the mean $\bar a_{t,k}$, and the standard error $\mathrm{se}_{t,k}$ of the readings $a_{x,k}$ of \eqn{eq:agreement} from all groups so far, including this batch's
\Statex \textbf{For each task $t$ and gated auxiliary $k\in\mathcal A_t$}
\State $\mathrm{tied}(t,k)\gets$ true
\State \concur-A: $\mathrm{tied}(t,k)\gets$ false if $n\ge61$ and $\bar a_{t,k}>2\,\mathrm{se}_{t,k}$
\Statex \textbf{For each group (prompt $x$ of task $t$)}
\State $z_{ik}\gets r_{ik}$ standardized within the group, for every $k\in\mathcal K_t$
\For{$k\in\mathcal A_t$}
\State \textbf{if} $\mathrm{tied}(t,k)$ \textbf{then} replace the $z_{ik}$ of the responses at the ceiling with their mean
\State $\hat z_k\gets$ sort in ascending order by (primary, score), then pool adjacent violators
\EndFor
\State $u_i\gets\sum_{k\in\mathcal K_t\setminus\mathcal A_t}w_kz_{ik}+\sum_{k\in\mathcal A_t}w_k\hat z_{ik}$
\Statex \textbf{For the batch}
\State $A\gets\W(u)$
\end{algorithmic}
\end{algorithm}
\subsection{Earned Trust}\label{app:trust}
Let $a_1,\dots,a_n$ be all readings of the gated pair $(t,k)$ since the start of training, including those of the current step's batch.
Their mean and its standard error are given by
\begin{equation}\label{eq:app-se}
\begin{aligned}
\bar a_{t,k}&=\frac1n\sum_{m=1}^{n}a_m,\\
\mathrm{se}_{t,k}&=\sqrt{\frac{1}{n(n-1)}\sum_{m=1}^{n}\bigl(a_m-\bar a_{t,k}\bigr)^2}.
\end{aligned}
\end{equation}
At a step, the pair's ceiling is not tied if $n\ge61$ and $\bar a_{t,k}>2\,\mathrm{se}_{t,k}$, and it is tied otherwise.
The rule has three settings: the margin of two standard errors, the minimum number of readings, and pooling over all of training.
The minimum of $61$ readings comes from Student's $t$ distribution: with $61$ readings and thus $60$ degrees of freedom, its two-sided $95\%$ critical value is $2.00$ to two decimals.
A pair with a graded primary has $61$ readings within a few training steps, and the standard error shrinks as readings accumulate, so the rule tests only the sign of the pooled agreement.
The readings are not independent draws from one distribution, since the policy changes between steps and every reading has the same weight, so we use the margin as a fixed threshold on the pooled agreement and attach no significance level to it.
A ceiling that is not tied at one step can be tied again at a later one.
No reading compares two responses at the ceiling: they have the same primary, so every pair of responses that \eqn{eq:agreement} counts has a response below the ceiling.
The rule assumes that if an auxiliary agrees with the primary's order on such pairs more often than it reverses that order, it also tracks what the primary measures among the responses at the ceiling.

\section{Analysis and Proofs}\label{app:proofs}
In this section, we show how standardization acts at the ceiling, prove Proposition~\ref{prop:project} and Corollary~\ref{cor:binary}, and add a few remarks on these two statements.

A group has $G\ge2$ responses with primaries $p\in[0,1]^G$, and $H=\{i:p_i=1\}$ is the set of responses at the ceiling.
Vectors lie in $\R^G$, with $\langle a,b\rangle=\sum_ia_ib_i$ and $\lVert a\rVert^2=\langle a,a\rangle$, and $\one$ is the vector of ones.
The standardized scores $z_k$ of \eqn{eq:gdpo} satisfy $\langle z_k,\one\rangle=0$, and since $d_k$ is the sample standard deviation with $d_k^2=\frac{1}{G-1}\sum_i(r_{ik}-\bar r_k)^2$, they also satisfy $\lVert z_k\rVert^2=G-1$ if $d_k>0$.
The two facts below are standard.
Let $C\subset\R^G$ be a closed convex cone and $z\in\R^G$.
\emph{Fact 1.}
The function $v\mapsto\lVert v-z\rVert^2$ has exactly one minimizer on $C$, which we write $\Pi_C(z)$.
\emph{Fact 2.}
A point $\hat z\in C$ equals $\Pi_C(z)$ if and only if it satisfies
\begin{align}
\langle z-\hat z,\,\hat z\rangle&=0\quad\text{and}\label{eq:app-orth}\\
\langle z-\hat z,\,v\rangle&\le0\quad\text{for every }v\in C.\label{eq:app-polar}
\end{align}

\subsection{Standardization at the Ceiling}\label{app:proof-ceiling}
In a group whose responses are all at the ceiling, every response $i$ has $r_{ij}=1$, so $d_j=0$, which gives $z_{ij}=0$.
A component $k$ with $d_k>0$ has $\lVert z_k\rVert^2=G-1$ whatever the size of its raw differences, so its standardized scores have the sample variance $\frac{1}{G-1}\sum_iz_{ik}^2=1$.
By \eqn{eq:gdpo}, $u_i=\sum_{k\ne j}w_kz_{ik}$, so the remaining components alone determine the whole order of such a group.

\textbf{The example.}
Let $G=8$, and let a binary primary gate one auxiliary, both at weight $1$.
In the first group, all responses are correct, seven with the auxiliary score $s$ and one with $s-\delta$, where $\delta>0$.
The deviations from the mean are $\delta/8$ for the seven and $-7\delta/8$ for the one, and the sample variance is $d_k^2=\frac17\bigl(7\cdot\frac{\delta^2}{64}+\frac{49\,\delta^2}{64}\bigr)=\frac{\delta^2}{8}$.
The one response has $z_{ik}=-\frac{7\delta/8}{\delta/\sqrt8}=-\frac{7}{\sqrt8}\approx-2.47$, which does not depend on $\delta$, and since $z_{ij}=0$, this is its $u_i$.
In the second group, four responses are correct with auxiliary score $1$ and four are wrong.
A wrong response has primary $0$, so the gate sets its auxiliary score to $\min\{s_{ik},0\}=0$, and both components score the group $(1,1,1,1,0,0,0,0)$, with mean $\frac12$ and sample variance $\frac17\cdot8\cdot\frac14=\frac27$.
Their standardized scores are $\pm\frac12\big/\sqrt{2/7}=\pm\sqrt{7/8}$, so a wrong response has $u_i=-2\sqrt{7/8}=-\sqrt{7/2}\approx-1.87$.

\subsection{Proof of Proposition~\ref{prop:project}}\label{app:proof-project}
\textbf{Item (i) (existence).}
The set $C_{\le}(p)$ is the intersection of the closed half-spaces $\{v:v_i\ge v_{i'}\}$ over the pairs with $p_i>p_{i'}$, or all of $\R^G$ if there is no such pair.
The boundary of each half-space contains the origin, so each is a closed convex cone, and so is their intersection.
The vectors that are constant on $H$ form a linear subspace $S$, and $C_{=}(p)=C_{\le}(p)\cap S$ is again a closed convex cone.
A constant vector satisfies every inequality and every equality above, so both cones contain the constant vectors, among them $0$ and $\pm\one$.
By Fact 1, $\hat z=\Pi_C(z)$ exists and is unique for every $z\in\R^G$.

\textbf{Item (ii) (computation).}
Let $P_S$ be the map that replaces the entries of a vector on $H$ with their mean and leaves its other entries unchanged.
Let $y=P_Sz$ if the ceiling is tied and $y=z$ if it is not, so that $y$ is the result of the averaging step of Appendix~\ref{app:projection}, and let $w=\Pi_{C_{\le}(p)}(y)$.
\emph{Step (a): $w_m\le w_{m'}$ for any two responses $m$ and $m'$ with the same primary and $y_m\le y_{m'}$.}
Let $w'$ be $w$ with the entries $m$ and $m'$ exchanged.
The two responses have the same primary, so $w'\in C_{\le}(p)$.
Suppose that $w_m>w_{m'}$, so that $w'\ne w$.
Expanding the squares gives $\lVert w'-y\rVert^2-\lVert w-y\rVert^2=2\,(w_m-w_{m'})(y_m-y_{m'})\le0$, because $y_m\le y_{m'}$.
Then $w'\in C_{\le}(p)$ is at least as close to $y$ as $w$ is, which contradicts the uniqueness of $w$.

\emph{Step (b): $w=\hat z$.}
If the ceiling is not tied, this holds by definition.
If it is tied, the vector $z-y$ is zero outside $H$ and sums to zero over $H$, where every $v\in S$ is constant, so $\langle z-y,v\rangle=0$ for $v\in S$.
For $v\in S$ the vector $v-y$ lies in $S$, and therefore $\lVert v-z\rVert^2=\lVert v-y\rVert^2+\lVert y-z\rVert^2$, whose last term does not depend on $v$.
Since $C_{=}(p)\subset S$, the point of $C_{=}(p)$ closest to $z$ is the point of $C_{=}(p)$ closest to $y$, so $\hat z=\Pi_{C_{=}(p)}(y)$.
The vector $y$ is constant on $H$, whose responses share one primary, so step (a) gives $w_m\le w_{m'}$ and $w_{m'}\le w_m$ for all $m,m'\in H$, and $w\in C_{\le}(p)\cap S=C_{=}(p)$.
The point $w$ is closest to $y$ in $C_{\le}(p)$ and lies in the subset $C_{=}(p)$, and therefore $w=\Pi_{C_{=}(p)}(y)=\hat z$.

\emph{Step (c): the pooling step returns $w$.}
Number the responses in the order of the pooling step's sort, so that $p_1\le p_2\le\dots\le p_G$, and $y_m\le y_{m+1}$ whenever $p_m=p_{m+1}$.
Let $K=\{v:v_1\le v_2\le\dots\le v_G\}$, which is a closed convex cone.
First, $K\subset C_{\le}(p)$: if $p_i>p_{i'}$, then $i>i'$ in this numbering, and $v_i\ge v_{i'}$ for $v\in K$.
Second, $w$ lies in $K$: if $p_m<p_{m+1}$, then $w_m\le w_{m+1}$ because $w\in C_{\le}(p)$, and if $p_m=p_{m+1}$, then $w_m\le w_{m+1}$ by step (a).
Third, $w$ is closest to $y$ in $C_{\le}(p)$ and lies in the subset $K$, so $w=\Pi_K(y)$, which is the isotonic regression of the sequence $y_1,\dots,y_G$.
After the sort, the pooling step applies the pool-adjacent-violators algorithm to $y_1,\dots,y_G$, which computes the isotonic regression of a sequence~\citep{ayer1955empirical}, so that the two steps together return $w=\hat z$.

\textbf{Item (iii) (mean).}
The cone $C$ contains $\pm\one$, so \eqn{eq:app-polar} with $v=\one$ and with $v=-\one$ gives $\langle z-\hat z,\one\rangle=0$, and it follows that $\sum_i\hat z_i=\sum_iz_i$.

\textbf{Item (iv) (energy).}
By \eqn{eq:app-orth}, $\lVert z\rVert^2=\lVert\hat z+(z-\hat z)\rVert^2=\lVert\hat z\rVert^2+\lVert z-\hat z\rVert^2$, so $\lVert\hat z\rVert\le\lVert z\rVert$.
If $z\in C$, then $z$ is itself the closest point of $C$, so $\hat z=z$. Conversely, if $\hat z=z$, then $z\in C$ because $\hat z\in C$.
Let $\hat z^{\le}=\Pi_{C_{\le}(p)}(z)$ and $\hat z^{=}=\Pi_{C_{=}(p)}(z)$.
Since $C_{=}(p)\subset C_{\le}(p)$, the closest point of the smaller cone is at least as far from $z$, so $\lVert z-\hat z^{=}\rVert\ge\lVert z-\hat z^{\le}\rVert$.
The energy identity for each cone then gives $\lVert\hat z^{=}\rVert^2=\lVert z\rVert^2-\lVert z-\hat z^{=}\rVert^2\le\lVert z\rVert^2-\lVert z-\hat z^{\le}\rVert^2=\lVert\hat z^{\le}\rVert^2$.
For $z\ne0$, the three shares of \eqn{eq:shares} are therefore nonnegative, and they sum to one, since $\lVert\hat z^{\le}\rVert^2$ and $\lVert\hat z^{=}\rVert^2$ cancel.

\textbf{Item (v) (order).}
Since $\hat z\in C\subset C_{\le}(p)$, $p_i>p_{i'}$ implies $\hat z_i\ge\hat z_{i'}$, so $(p_i-p_{i'})(\hat z_i-\hat z_{i'})\ge0$ for every pair.
Let $f$ be a nondecreasing function and $f_i=f(p_i)$.
Then $(f_i-f_{i'})(\hat z_i-\hat z_{i'})\ge0$ for every pair as well: if $p_i>p_{i'}$ both factors are nonnegative, and if $p_i=p_{i'}$ the first factor is zero.
For any two vectors $a,b$ with means $\bar a,\bar b$,
\[
\sum_{i,i'}(a_i-a_{i'})(b_i-b_{i'})=2G\sum_i(a_i-\bar a)(b_i-\bar b) .
\]
With $a_i=f_i$ and $b=\hat z$, the left side is a sum of nonnegative terms, so the group covariance of $\hat z$ with $f(p)$ is nonnegative.
For $f(p)=p$ and $d_j>0$, the right side is $2G\,d_j\,\langle\hat z,z_j\rangle$, because $p_i-\bar p=d_j\,z_{ij}$ and $\sum_iz_{ij}=0$, so $\langle\hat z,z_j\rangle\ge0$.
If $d_j=0$, then $z_j=0$ and $\langle\hat z,z_j\rangle=0$.
\qed

\subsection{Proof of Corollary~\ref{cor:binary}}\label{app:proof-binary}
A response with $p_i=1$ passes and is at the ceiling, and a response with $p_i=0$ fails, so $H$ is the set of passes, and we write $F$ for the set of failures and $n=\lvert H\rvert$.
Under the gate, a failure has $r_{ik}=\min\{s_{ik},0\}=0$, while a pass has $r_{ik}=s_{ik}\ge0$.

\emph{Part (i).}
The only pairs with $p_i>p_{i'}$ have $i\in H$ and $i'\in F$.
If $d_k=0$, then $z_k=0\in C_{\le}(p)$, and if $d_k>0$, then $z_{ik}-z_{i'k}=(r_{ik}-r_{i'k})/d_k=s_{ik}/d_k\ge0$ for such a pair.
In both cases $z_k\in C_{\le}(p)$, and $\Pi_{C_{\le}(p)}(z_k)=z_k$ by item (iv) of Proposition~\ref{prop:project}.
The share against the order in \eqn{eq:shares} is therefore zero on a binary primary.
If none of the task's gated pairs has a tied ceiling, \eqn{eq:concur} gives its groups the summed scores of \eqn{eq:gdpo}.

\emph{Part (ii).}
Both passes and failures are present, so $1\le n\le G-1$ and $d_j>0$.
If $d_k=0$, then $z_k=0$ and $\hat z_k=0$, and the claim holds with $c=0$.
Let $d_k>0$, and let $\bar s_H=\frac1n\sum_{i\in H}r_{ik}\ge0$ be the mean auxiliary score of the passes.
With $S$ and $P_S$ as in the proof of Proposition~\ref{prop:project}, the vector $y=P_Sz_k$ has the entries $(\bar s_H-\bar r_k)/d_k$ on $H$ and $-\bar r_k/d_k$ on $F$.
It is constant on $H$, and its entries on $H$ exceed those on $F$ by $\bar s_H/d_k\ge0$, so $y\in C_{=}(p)$.
By step (b) of that proof and item (iv), $\hat z_k=\Pi_{C_{=}(p)}(y)=y$.
The vector $\hat z_k$ is constant on $H$ and constant on $F$, and it sums to zero by item (iii).
The same holds for $z_j$, whose entries are $(1-\frac nG)/d_j$ on $H$ and $-\frac nG/d_j$ on $F$, and $z_j\ne0$.
The vectors that are constant on $H$, constant on $F$ and sum to zero form a line, so $\hat z_k=c\,z_j$ with $c=\langle\hat z_k,z_j\rangle/\lVert z_j\rVert^2$.
Since $z_j\in S$ and $z_k-P_Sz_k$ is orthogonal to $S$, $\langle\hat z_k,z_j\rangle=\langle z_k,z_j\rangle$, which gives $c=\langle z_k,z_j\rangle/\lVert z_j\rVert^2$.
The entries on $H$ exceed those on $F$ by $\bar s_H/d_k$ in $\hat z_k$ and by $1/d_j$ in $z_j$, so $c=\bar s_H\,d_j/d_k\ge0$.
Since $\lVert z_k\rVert^2=\lVert z_j\rVert^2=G-1$, $c=\langle z_k,z_j\rangle/(\lVert z_k\rVert\,\lVert z_j\rVert)$ is their cosine, so $c\le1$.

\emph{Part (iii).}
In a group of passes only, every response is in $H$, so $C_{=}(p)$ is the set of constant vectors.
The constant vector closest to $z_k$ is its mean times $\one$, which is zero.
\qed

\subsection{Remarks on the Statements}\label{app:remarks}
The factor $1+c$ that follows Corollary~\ref{cor:binary} is $2$ exactly when the group's correct responses share one positive auxiliary score.
Take a binary primary, a group with both passes and failures, and an auxiliary with $d_k>0$. By the proof of part (ii) of the corollary, the tied ceiling keeps $\hat z_k=P_Sz_k$, so the part $z_k-\hat z_k$ that it removes is zero on the failures and equals $(r_{ik}-\bar s_H)/d_k$ on the passes.
PAPO's standardization in \eqn{eq:papo} divides the same differences $r_{ik}-\bar s_H$ by the standard deviation over the passes and sets the failures to zero. Up to a positive factor, it therefore keeps exactly the part of $z_k$ that the tied ceiling removes, and it drops the part $\hat z_k=c\,z_j$ that the tied ceiling keeps instead.

\section{Experimental Details}\label{app:exp}
In this section, we describe the data and rewards, the training runs and baselines, and the evaluation.
\begin{table*}[t]
\centering
\small
\setlength{\tabcolsep}{4pt}
\resizebox{\ifdim\width>\linewidth\linewidth\else\width\fi}{!}{%
\begin{tabular}{@{}ll rrcccl ll@{}}
\toprule
Task group & Benchmark & Problems & Responses & Response Len & Temperature & Top-$p$ & Other & Metric & Judge \\
\midrule
Math & AIME24 & 30 & 32 & 32,768 & 0.6 & 0.95 & -- & accuracy & -- \\
 & AIME25 & 30 & 32 & 32,768 & 0.6 & 0.95 & -- & accuracy & -- \\
 & MATH-500 & 500 & 32 & 32,768 & 0.6 & 0.95 & -- & accuracy & -- \\
 & OlympiadBench & 674 & 32 & 32,768 & 0.6 & 0.95 & -- & accuracy & -- \\
 & Minerva & 272 & 32 & 32,768 & 0.6 & 0.95 & -- & accuracy & -- \\
 & AMC23 & 40 & 32 & 32,768 & 0.6 & 0.95 & -- & accuracy & -- \\
\midrule
Code & HumanEval+ & 164 & 10 & 32,768 & 0.6 & 0.95 & -- & pass@1 & -- \\
 & MBPP+ & 378 & 10 & 32,768 & 0.6 & 0.95 & -- & pass@1 & -- \\
 & LiveCodeBench v6 & 175 & 10 & 32,768 & 0.6 & 0.95 & -- & pass@1 & -- \\
\midrule
IF & IFEval & 541 & 1 & 32,768 & 0.6 & 0.95 & -- & instruction-level strict accuracy & -- \\
 & IFBench & 300 & 1 & 32,768 & 0.6 & 0.95 & -- & prompt-level loose accuracy & -- \\
 & GuideBench & 1,042 & 1 & 32,768 & 0 & 1.0 & -- & accuracy & -- \\
\midrule
Writing & WritingBench & 1,000 & 1 & 32,768 & 0.7 & 0.8 & top-$k$ 20 & overall score & \texttt{Gemma-4-26B-A4B-IT} \\
 & Creative Writing v3 & 96 & 1 & 32,768 & 0.7 & 1.0 & min-$p$ 0.1 & rubric score & \texttt{Gemma-4-26B-A4B-IT} \\
 & LongBench-Write & 120 & 1 & 32,768 & 0.5 & 1.0 & -- & overall score & \texttt{Gemma-4-26B-A4B-IT} \\
\bottomrule
\end{tabular}%
}

\caption{Settings of the evaluation for each benchmark.
Every metric is reported on a scale from 0 to 100.}
\label{tab:eval-config}
\end{table*}

\subsection{Data and Rewards}\label{app:data}
\textbf{Data.}
Math prompts come from DeepScaleR~\citep{deepscaler2025} and DAPO-Math-17k~\citep{yu2025dapo}, code prompts from Eurus-2-RL~\citep{cui2025prime}, Dolci-RL-Zero-Code~\citep{olmo2025olmo3}, and Open-R1~\citep{openr1}, and IF and writing prompts from RubricHub~\citep{li2026rubrichub}.
We deduplicate prompts across tasks, remove exact and embedded copies of evaluation questions, and keep prompts of at most 2,048 tokens.
For each base policy, we sample eight \emph{base responses} to every candidate prompt at temperature $1$ and top-$p$ $1$, cut them off at the model's response limit in training, and score them with the primaries and, on writing, with the reward model (RM).
On math, code, and writing, we drop the prompts whose eight base responses all end before the limit with a primary of $1$, and those on which no base response passes make up a fixed 10\% of each task's training prompts. On IF, we set no quota and keep the prompts whose base responses all have a primary of $1$, since the judged criterion can still separate their responses.
Each of the four tasks has 9,838 training prompts on \qwenfour{}, 11,705 on \qweneight{}, and 10,850 on \olmo{}.

\textbf{The length reward.}
For response $i$ of $|y_i|$ generated tokens to a prompt $x$ with \emph{reference length} $\ell_x$, the length reward's own score $s_{ik}$ is given by
\begin{equation}\label{eq:length}
s_{ik}=\max\Bigl\{0,\;1-\Bigl(\frac{(|y_i|-\ell_x)_+}{\ell_x}\Bigr)^{2}\Bigr\},
\end{equation}
where $(a)_+=\max\{a,0\}$.
The reference length $\ell_x$ is the mean number of generated tokens over the prompt's \emph{reference set}, which consists of its base responses that end before the limit and have a primary of $1$.
On code, IF, and writing, a prompt with fewer than two such responses instead uses its base responses that end before the limit and pass.
A prompt whose reference set is empty takes a task-level reference length.
Over the task's prompts with a reference set, we regress the logarithm of the reference length on the share of base responses outside the set.
The task-level length is the fit at a share of $1$ if the $95\%$ bootstrap interval of the slope over 2,000 resamples lies above zero, and it is the mean reference length of these prompts otherwise.

\textbf{Judge and reward model.}
The training judge is \texttt{Gemma-4-26B-\allowbreak A4B-\allowbreak IT}~\citep{gemma2026gemma4}, and the RM is \texttt{Skywork-\allowbreak Reward-\allowbreak V2-\allowbreak Qwen3-\allowbreak 8B}~\citep{liu2026skywork}.
The RM's raw score $\rho$ is rescaled to $(\rho-\rho_{\mathrm{lo}})/(\rho_{\mathrm{hi}}-\rho_{\mathrm{lo}})$ and clipped to $[0,1]$, where $\rho_{\mathrm{lo}}$ and $\rho_{\mathrm{hi}}$ are the 1st and 99th percentiles of its raw scores on all base responses of the training writing prompts, and they stay fixed for all runs of a base policy throughout their training.

\begin{table}[htbp]
\centering
\small
\resizebox{\ifdim\width>\linewidth\linewidth\else\width\fi}{!}{%
\begin{tabular}{@{}lc@{}}
\toprule
Hyperparameter & Value \\
\midrule
Backend & megatron \\
Rollout & sglang \\
Optimizer & Adam \\
Learning Rate & 1e-6 \\
LR Scheduler & constant \\
Weight Decay & 0.0 \\
Grad Clip & 1.0 \\
Batch Size & 512 \\
Mini Batch Size & 64 \\
Num Responses & 8 \\
Temperature & 1.0 \\
Clip Ratio Low / High & 0.2 / 0.28 \\
Loss Agg & token-mean \\
Prompt Len & 2048 \\
Response Len & 4096 (\qwenfour{}), 8192 (others) \\
Num Steps & 300 \\
MT-GRPO LR / $\lambda_{\mathrm{bias}}$ & 0.025 / 0.3 \\
\bottomrule
\end{tabular}%
}
\caption{Hyperparameters of the runs.}
\label{tab:hparams}
\end{table}
\subsection{Training, Runs, and Baselines}\label{app:runs}
\textbf{Training.}
\tab{tab:hparams} lists the training hyperparameters.
The estimator's advantages enter a clipped surrogate objective~\citep{schulman2017proximal}, without a KL or entropy term.
Every run except MT-GRPO excludes the groups with identical component scores before applying its estimator.
Each run uses 4 nodes of 8 NVIDIA RTX PRO 6000 GPUs, which also serve the training judge and the RM.

\begin{table*}[t]
\centering
\small
\setlength{\tabcolsep}{4pt}
\resizebox{\ifdim\width>\linewidth\linewidth\else\width\fi}{!}{%
\begin{tabular}{@{}lccccccccc@{}}
\toprule
 & \multicolumn{2}{c}{Cosine} & \multicolumn{2}{c}{Against order} & \multicolumn{2}{c}{At ceiling} & \multicolumn{2}{c}{Passing at ceiling} & Agreement \\
\cmidrule(lr){2-3} \cmidrule(lr){4-5} \cmidrule(lr){6-7} \cmidrule(lr){8-9}
Pair & Early & Late & Early & Late & Early & Late & Early & Late & \\
\midrule
Math & 0.89 & 0.89 & 0.00 & 0.00 & 0.19 & 0.19 & 1.00 & 1.00 & -- \\
Code & 0.85 & 0.79 & 0.03 & 0.03 & 0.09 & 0.04 & 0.70 & 0.74 & +0.003 \\
IF & 0.75 & 0.48 & 0.04 & 0.05 & 0.05 & 0.18 & 0.29 & 0.57 & +0.015 \\
Writing & 0.51 & 0.04 & 0.29 & 0.60 & 0.00 & 0.00 & 0.00 & 0.00 & $-$0.036 \\
Writing (RM) & 0.67 & 0.15 & 0.22 & 0.51 & 0.00 & 0.00 & 0.00 & 0.00 & +0.058 \\
\bottomrule
\end{tabular}%
}

\caption{The gated auxiliaries under GDPO at the declared weights on \qwenfour{}, early (steps 1--20) and late (the last 20).
Passing at ceiling is the share of passing responses at the ceiling, and Agreement is pooled over the run.}
\label{tab:gate}
\end{table*}
\begin{table}[t]
\centering
\small
\setlength{\tabcolsep}{4pt}
\resizebox{\ifdim\width>\linewidth\linewidth\else\width\fi}{!}{%
\begin{tabular}{@{}lcccccc@{}}
\toprule
 & Passing & Pooled & \multicolumn{3}{c}{In standard errors} & Not \\
\cmidrule(lr){4-6}
Pair & at ceiling & agreement & -A & -T & -O & tied \\
\midrule
\multicolumn{7}{@{}l}{\qwenfour{}} \\
\quad Math & 1.00 & -- & -- & -- & -- & 0.00 \\
\quad Code & 0.74 & +0.012 & +4.2 & +1.9 & +1.4 & 0.85 \\
\quad IF & 0.55 & +0.018 & +7.4 & +6.5 & +7.0 & 0.95 \\
\quad Writing & 0.02 & $-$0.054 & $-$36.8 & $-$32.7 & $-$35.1 & 0.00 \\
\quad Writing (RM) & 0.02 & +0.061 & +37.4 & +35.7 & +36.2 & 1.00 \\
\midrule
\multicolumn{7}{@{}l}{\qweneight{}} \\
\quad Math & 1.00 & -- & -- & -- & -- & 0.00 \\
\quad Code & 0.76 & +0.011 & +3.8 & +1.2 & +2.3 & 0.97 \\
\quad IF & 0.56 & +0.018 & +7.3 & +8.4 & +7.2 & 0.95 \\
\quad Writing & 0.02 & $-$0.047 & $-$34.1 & $-$34.0 & $-$33.7 & 0.00 \\
\quad Writing (RM) & 0.02 & +0.051 & +31.6 & +30.5 & +29.8 & 1.00 \\
\midrule
\multicolumn{7}{@{}l}{\olmo{}} \\
\quad Math & 1.00 & -- & -- & -- & -- & 0.00 \\
\quad Code & 0.88 & $-$0.011 & $-$2.5 & $-$2.2 & $-$9.8 & 0.00 \\
\quad IF & 0.56 & +0.053 & +16.9 & +18.0 & +17.1 & 0.99 \\
\quad Writing & 0.01 & $-$0.069 & $-$41.2 & $-$38.0 & $-$39.1 & 0.00 \\
\quad Writing (RM) & 0.01 & +0.093 & +46.2 & +45.9 & +44.8 & 0.99 \\
\bottomrule
\end{tabular}%
}

\caption{What earned trust reads and decides for each gated pair: the share of passing responses at the ceiling over the last 20 training steps, the pooled agreement of \concur-A at its last training step, each run's pooled agreement in standard errors, which \concur-T and \concur-O only log, and the share of all training steps at which \concur-A leaves the pair's ceiling not tied.}
\label{tab:trust}
\end{table}
\textbf{Runs and baselines.}
On each model, we run the seven estimators at the declared weights, the two variants of \concur, \concur-O, the reference, GDPO ($w=0.5$), and GRPO (no std).
All runs on a model, including the reference, share the training prompts, the training recipe, and the rewards with their gates.
Each baseline uses the settings of its paper or public code where reported and the shared recipe otherwise.
MT-GRPO keeps its own filter and its ratio-preserving rejection sampling with the settings of its released script, so it generates more responses per step than every other run.
All runs whiten their summed scores over the batch except the five whose estimators lack this step: GRPO, GRPO (no std), GEOMA, MT-GRPO, and PAPO.

\textbf{PAPO.}
We apply PAPO's advantage rule with a task's primary as its outcome reward, the passing responses $P=\{i:p_i\ge\frac12\}$ as its correct set, and each gated auxiliary $k\in\mathcal A_t$ as a process reward:
\begin{equation}\label{eq:papo}
\begin{aligned}
\tilde z_{ik}&=\begin{cases}(r_{ik}-\bar r^{P}_{k})/d^{P}_{k}&i\in P,\\0&i\notin P,\end{cases}\\
A_i&=\sum_{k\in\mathcal K_t\setminus\mathcal A_t}w_kz_{ik}+\sum_{k\in\mathcal A_t}w_k\tilde z_{ik},
\end{aligned}
\end{equation}
where $\bar r^{P}_{k}$ and $d^{P}_{k}$ are the mean and the sample standard deviation of the auxiliary over $P$, and $\tilde z_{ik}$ is zero on every response of a group in which fewer than two responses pass or $d^{P}_{k}$ is numerically zero.

\subsection{Evaluation}\label{app:eval}
\tab{tab:eval-config} gives each benchmark's settings, metric, and judge, which are the same for every run.
Creative Writing v3 is scored by its rubric alone, and LongBench-Write's score is the mean of a quality score from the judge and a length score.
The training judge scores each writing benchmark by its own criteria, and the second judge is \texttt{Qwen3.6-\allowbreak 35B-\allowbreak A3B}~\citep{qwen36_35b_a3b} with thinking off.
\tab{tab:second-judge} gives the scores of every run under this judge.

\section{Additional Results}\label{app:results}
In this section, we report the training statistics behind the motivating experiment, the decisions of earned trust, the shares of energy and of the update during training, and the results of the two variants, the ablation, and the reference on all three models.

\begin{figure*}[t]
\centering
\includegraphics[width=0.472\textwidth]{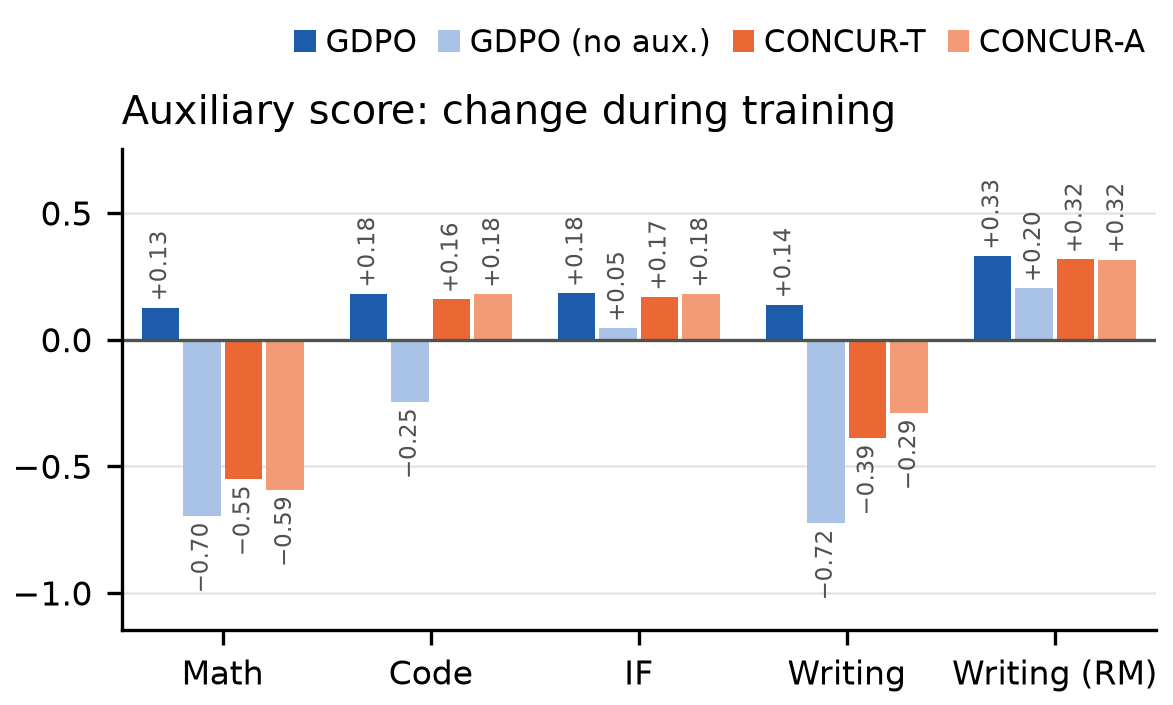}\hfill
\includegraphics[width=0.472\textwidth]{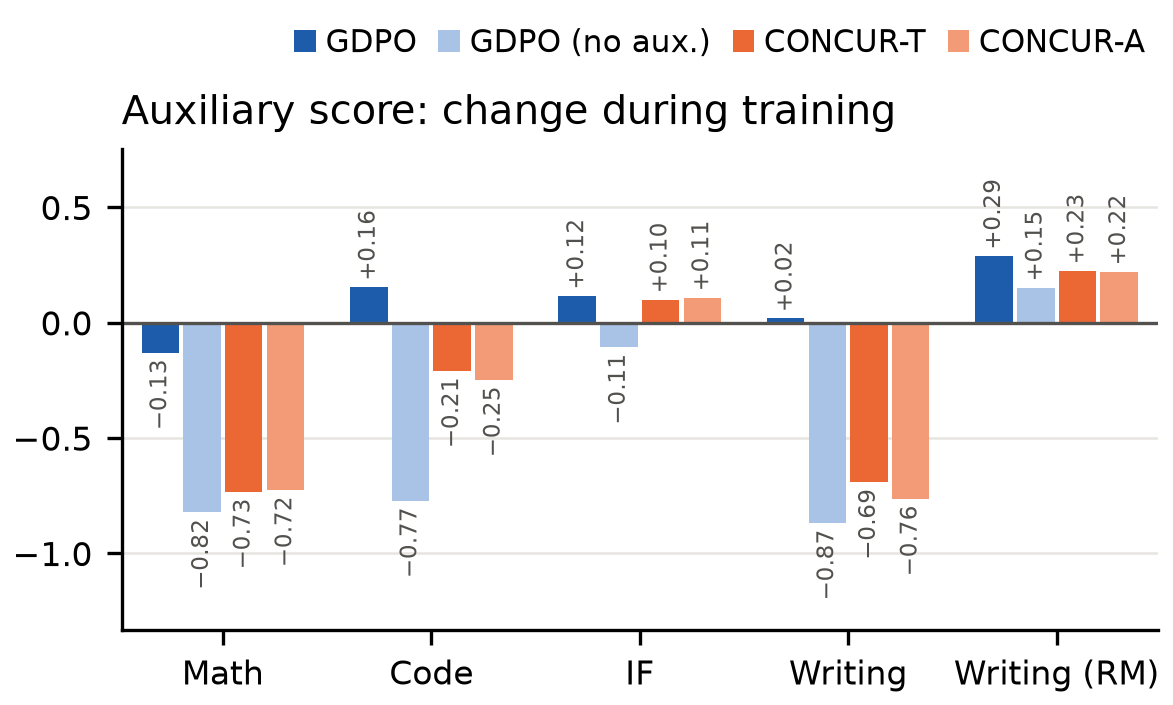}
\caption{Each gated auxiliary's mean score over the passing responses on \qweneight{} (left) and \olmo{} (right): the mean over the last 20 training steps minus the base policy's level of the same model.}
\label{fig:aux-others}
\end{figure*}
\begin{figure*}[t]
\centering
\includegraphics[width=\textwidth]{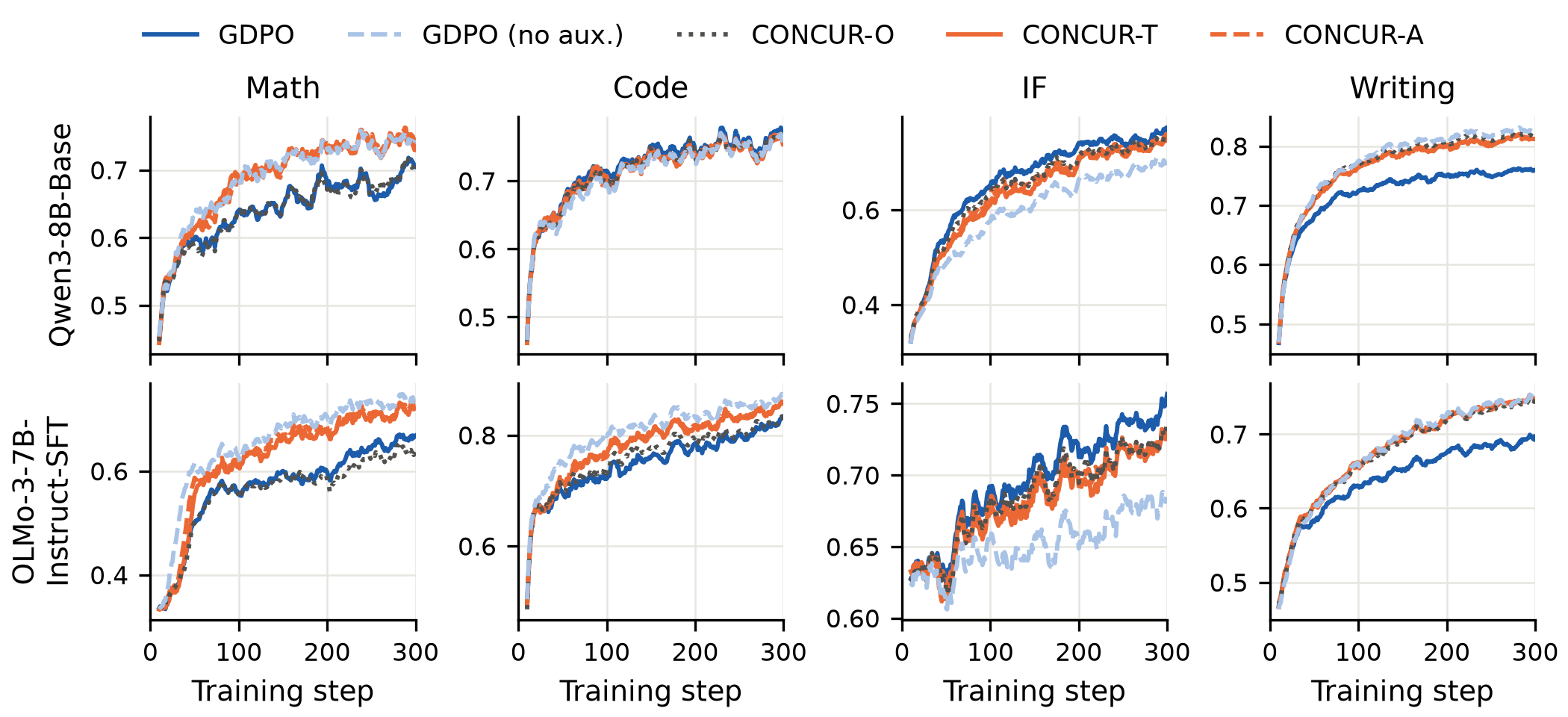}
\caption{The primary's mean score over all responses during training on \qweneight{} (top) and \olmo{} (bottom) under GDPO, the reference, and \concur, shown as 10-step running means.}
\label{fig:primary}
\end{figure*}
\begin{table*}[!t]
\centering
\footnotesize
\setlength{\tabcolsep}{2pt}
\renewcommand{\arraystretch}{1.2}
\resizebox{\ifdim\width>\linewidth\linewidth\else\width\fi}{!}{%
\begin{tabular}{lccccccc!{\color{tableblue}\vrule width 1.4pt}cccc!{\color{tableblue}\vrule width 1.4pt}cccc!{\color{tableblue}\vrule width 1.4pt}cccc!{\color{tableblue}\vrule width 1.4pt}c}
\toprule
 & \multicolumn{7}{c!{\hspace{1.4pt}}}{Math} & \multicolumn{4}{c!{\hspace{1.4pt}}}{Code} & \multicolumn{4}{c!{\hspace{1.4pt}}}{IF} & \multicolumn{4}{c!{\hspace{1.4pt}}}{Writing} &  \\
\cmidrule(lr{\dimexpr\cmidrulekern+1.4pt\relax}){2-8} \cmidrule(lr{\dimexpr\cmidrulekern+1.4pt\relax}){9-12} \cmidrule(lr{\dimexpr\cmidrulekern+1.4pt\relax}){13-16} \cmidrule(lr{\dimexpr\cmidrulekern+1.4pt\relax}){17-20}
 & AIME24 & AIME25 & MATH & OBench & Minerva & AMC23 & \textit{Mean} & HEval+ & MBPP+ & LCB-v6 & \textit{Mean} & IFEval & IFBench & GBench & \textit{Mean} & WBench & CW-v3 & LBWrite & \textit{Mean} & Avg \\
\midrule
\addlinespace[2pt]
\multicolumn{21}{>{\columncolor{tablepink}}l}{\textbf{Qwen3-4B-Base}} \\
\addlinespace[2pt]
\quad GDPO & \phantom{0}9.06 & \phantom{0}9.90 & 79.29 & 44.66 & 45.48 & 59.61 & 41.33 & 81.83 & 70.87 & 24.40 & 59.03 & 85.73 & 36.33 & \textbf{80.61} & 67.56 & 71.94 & \underline{45.97} & 69.01 & 62.31 & 54.31 \\
\quad GDPO ($w=0.5$) & 14.58 & 10.31 & 81.79 & 48.46 & 46.12 & 58.36 & 43.27 & 82.01 & 71.19 & 24.17 & 59.12 & 86.81 & \underline{39.33} & \underline{79.56} & \underline{68.57} & 70.74 & 38.11 & 67.41 & 58.75 & 54.60 \\
\quad GDPO (no aux.) & \textbf{23.44} & \textbf{18.75} & 84.76 & \underline{53.16} & \textbf{49.52} & 65.70 & \underline{49.22} & \textbf{85.49} & \textbf{72.33} & \underline{24.74} & \textbf{60.85} & 84.77 & \textbf{40.33} & 75.91 & 67.01 & 67.35 & 39.33 & 64.36 & 57.01 & 56.66 \\
\quad GRPO (no std) & 12.40 & \phantom{0}8.33 & 79.49 & 44.81 & 45.80 & 54.53 & 40.89 & 80.49 & 71.19 & 24.46 & 58.71 & 87.29 & 33.67 & 78.79 & 66.58 & 65.63 & 42.53 & 66.00 & 58.06 & 53.03 \\
\quad CONCUR-O & 11.88 & 14.58 & 80.32 & 45.73 & 46.25 & 54.14 & 42.15 & 81.10 & 69.66 & 24.17 & 58.31 & 86.69 & 39.00 & 78.98 & 68.22 & 77.31 & 45.42 & \textbf{77.78} & \underline{66.84} & 55.53 \\
\rowcolor{tablelightblue}\quad CONCUR-T & 19.48 & \underline{16.35} & \textbf{85.51} & 52.85 & 49.12 & \underline{69.22} & 48.76 & \underline{84.02} & 68.60 & \underline{24.74} & 59.12 & \underline{87.77} & 37.67 & 79.17 & 68.20 & \textbf{80.19} & \textbf{47.84} & \underline{77.74} & \textbf{68.59} & \underline{58.69} \\
\rowcolor{tablelightblue}\quad CONCUR-A & \underline{20.52} & \textbf{18.75} & \underline{85.28} & \textbf{53.58} & \underline{49.22} & \textbf{69.38} & \textbf{49.45} & 82.44 & \underline{71.51} & \textbf{25.14} & \underline{59.70} & \textbf{88.13} & \textbf{40.33} & 77.35 & \textbf{68.60} & \underline{79.43} & 44.71 & 76.31 & 66.82 & \textbf{58.80} \\
\addlinespace[2pt]
\multicolumn{21}{>{\columncolor{tablepink}}l}{\textbf{Qwen3-8B-Base}} \\
\addlinespace[2pt]
\quad GDPO & 21.35 & 14.38 & 82.99 & 49.32 & 49.61 & 62.19 & 46.64 & 84.09 & 74.52 & 26.23 & 61.61 & 88.13 & \textbf{39.33} & \underline{84.55} & \textbf{70.67} & 76.09 & \textbf{52.90} & 69.57 & 66.19 & 58.35 \\
\quad GDPO ($w=0.5$) & 19.27 & 15.42 & 84.06 & 51.12 & 51.22 & 63.44 & 47.42 & 83.78 & 72.83 & 26.80 & 61.14 & 87.77 & 37.67 & 83.97 & 69.80 & 75.04 & 51.56 & 71.20 & 65.93 & 58.34 \\
\quad GDPO (no aux.) & 27.92 & \textbf{23.96} & \textbf{90.24} & \underline{59.92} & 53.98 & 77.11 & 55.52 & 85.98 & 74.87 & \textbf{27.37} & \underline{62.74} & 86.57 & 33.67 & 81.38 & 67.21 & 75.00 & 43.89 & 65.95 & 61.61 & 60.52 \\
\quad GRPO (no std) & 20.63 & 14.79 & 84.13 & 51.50 & 50.80 & 63.59 & 47.57 & \underline{86.34} & \textbf{76.03} & 24.97 & 62.45 & \textbf{89.45} & 36.33 & \textbf{84.84} & 70.21 & 70.33 & 46.66 & 71.06 & 62.68 & 58.10 \\
\quad CONCUR-O & 19.69 & 16.56 & 83.43 & 50.04 & 49.78 & 64.69 & 47.37 & 85.30 & \underline{75.11} & 27.09 & 62.50 & \underline{89.21} & \underline{38.67} & 82.92 & 70.26 & 86.78 & 51.64 & \textbf{82.97} & \textbf{73.80} & 60.26 \\
\rowcolor{tablelightblue}\quad CONCUR-T & \textbf{31.25} & \textbf{23.96} & 89.61 & 59.55 & \textbf{55.40} & \underline{78.52} & \textbf{56.38} & \textbf{87.38} & 74.87 & \underline{27.20} & \textbf{63.15} & 88.85 & \textbf{39.33} & 83.40 & \underline{70.53} & \underline{87.19} & \underline{52.42} & 81.22 & \underline{73.61} & \textbf{64.01} \\
\rowcolor{tablelightblue}\quad CONCUR-A & \underline{28.75} & \underline{23.13} & \underline{90.13} & \textbf{60.37} & \underline{54.19} & \textbf{78.91} & \underline{55.91} & 82.74 & 73.20 & 26.34 & 60.76 & 88.13 & \underline{38.67} & 83.40 & 70.06 & \textbf{88.06} & 49.92 & \underline{81.60} & 73.19 & \underline{63.17} \\
\addlinespace[2pt]
\multicolumn{21}{>{\columncolor{tablepink}}l}{\textbf{OLMo-3-7B-Instruct-SFT}} \\
\addlinespace[2pt]
\quad GDPO & 27.81 & 22.60 & 85.24 & 55.89 & 40.36 & 69.53 & 50.24 & 76.34 & 64.76 & 23.14 & 54.75 & 86.45 & 29.67 & 50.86 & 55.66 & 65.31 & 47.14 & 63.27 & 58.57 & 53.89 \\
\quad GDPO ($w=0.5$) & 29.58 & 27.29 & 87.04 & 57.51 & 41.53 & 73.98 & 52.82 & 77.74 & 64.92 & 24.80 & 55.82 & 87.53 & 32.67 & 51.92 & 57.37 & \underline{66.50} & 48.10 & 63.73 & 59.44 & 55.66 \\
\quad GDPO (no aux.) & \textbf{38.33} & \textbf{33.65} & \textbf{90.88} & \textbf{63.74} & \textbf{46.15} & \textbf{83.91} & \textbf{59.44} & \textbf{85.18} & \underline{70.50} & \textbf{31.54} & \textbf{62.41} & 84.77 & \underline{33.33} & 55.76 & 57.95 & 55.46 & 38.21 & 55.98 & 49.89 & 57.83 \\
\quad GRPO (no std) & 31.77 & 27.60 & 87.51 & 58.44 & 41.81 & 73.05 & 53.36 & 77.38 & 65.34 & 23.60 & 55.44 & \textbf{87.77} & 31.33 & 55.47 & 58.19 & 63.18 & 46.98 & 67.53 & 59.23 & 55.92 \\
\quad CONCUR-O & 26.77 & 22.08 & 84.36 & 53.28 & 39.92 & 67.50 & 48.99 & 76.65 & 64.50 & 25.66 & 55.60 & 87.29 & 31.67 & \textbf{60.17} & \underline{59.71} & \textbf{70.21} & \textbf{49.78} & \textbf{72.04} & \textbf{64.01} & 55.46 \\
\rowcolor{tablelightblue}\quad CONCUR-T & 34.48 & 29.90 & 89.41 & 61.57 & 44.05 & 79.84 & 56.54 & \underline{83.29} & \textbf{71.06} & 29.03 & \underline{61.13} & \underline{87.65} & \textbf{33.67} & \underline{58.54} & \textbf{59.95} & 66.14 & 49.04 & \underline{69.23} & \underline{61.47} & \underline{59.13} \\
\rowcolor{tablelightblue}\quad CONCUR-A & \underline{36.35} & \underline{32.19} & \underline{90.53} & \underline{62.41} & \underline{44.81} & \underline{83.52} & \underline{58.30} & 82.74 & 67.54 & \underline{30.34} & 60.21 & \textbf{87.77} & 32.00 & 56.91 & 58.89 & 65.45 & \underline{49.21} & 68.49 & 61.05 & \textbf{59.35} \\
\bottomrule
\end{tabular}%
}

\caption{Evaluation scores of GDPO, GDPO~($w=0.5$), the reference GDPO~(no aux.), GRPO~(no std), \concur-O, and the two variants of \concur on every benchmark, with the same benchmark names and formatting as \tab{tab:main}.}
\label{tab:ablation}
\bigskip
\footnotesize
\setlength{\tabcolsep}{2pt}
\renewcommand{\arraystretch}{1.2}
\resizebox{\ifdim\width>\linewidth\linewidth\else\width\fi}{!}{%
\begin{tabular}{lccccc!{\color{tableblue}\vrule width 1.4pt}ccccc!{\color{tableblue}\vrule width 1.4pt}ccccc}
\toprule
 & \multicolumn{5}{c!{\hspace{1.4pt}}}{Qwen3-4B-Base} & \multicolumn{5}{c!{\hspace{1.4pt}}}{Qwen3-8B-Base} & \multicolumn{5}{c}{OLMo-3-7B-Instruct-SFT} \\
\cmidrule(lr{\dimexpr\cmidrulekern+1.4pt\relax}){2-6} \cmidrule(lr{\dimexpr\cmidrulekern+1.4pt\relax}){7-11} \cmidrule(lr){12-16}
 & WBench & CW-v3 & LBWrite & \textit{Mean} & Avg & WBench & CW-v3 & LBWrite & \textit{Mean} & Avg & WBench & CW-v3 & LBWrite & \textit{Mean} & Avg \\
\midrule
\textcolor{tablegray}{\textit{Base}} & \textcolor{tablegray}{\textit{30.48}} & \textcolor{tablegray}{\textit{20.86}} & \textcolor{tablegray}{\textit{40.92}} & \textcolor{tablegray}{\textit{30.75}} & \textcolor{tablegray}{\textit{33.91}} & \textcolor{tablegray}{\textit{34.00}} & \textcolor{tablegray}{\textit{26.43}} & \textcolor{tablegray}{\textit{48.05}} & \textcolor{tablegray}{\textit{36.16}} & \textcolor{tablegray}{\textit{41.04}} & \textcolor{tablegray}{\textit{37.68}} & \textcolor{tablegray}{\textit{31.68}} & \textcolor{tablegray}{\textit{53.17}} & \textcolor{tablegray}{\textit{40.84}} & \textcolor{tablegray}{\textit{42.97}} \\
GRPO & 53.44 & 35.10 & 60.72 & 49.75 & 51.91 & 56.88 & 39.26 & 62.43 & 52.86 & 55.22 & \underline{48.49} & 38.63 & 57.07 & 48.07 & 52.74 \\
GDPO & 54.02 & 34.38 & 62.31 & 50.23 & 51.90 & 58.35 & 41.02 & 64.17 & 54.51 & 56.02 & 47.05 & 37.84 & 55.65 & 46.85 & 51.55 \\
GD$^2$PO & 52.54 & 32.89 & 60.97 & 48.80 & 51.60 & 55.21 & 36.73 & 62.17 & 51.37 & 54.88 & 48.36 & 37.91 & 59.90 & 48.72 & 53.25 \\
GEOMA & 50.68 & 33.12 & 57.67 & 47.16 & 51.14 & 58.29 & 37.53 & 64.87 & 53.56 & 54.39 & \textbf{48.63} & \textbf{41.79} & 58.19 & \underline{49.54} & 50.72 \\
SAW & 53.38 & \textbf{35.59} & 60.04 & 49.67 & 52.07 & 56.91 & 36.56 & 64.81 & 52.76 & 55.47 & 48.42 & \underline{41.00} & 57.39 & 48.94 & 52.22 \\
MT-GRPO & 53.26 & 34.06 & 60.78 & 49.37 & 50.85 & 55.49 & 41.52 & 62.46 & 53.16 & 55.20 & 47.52 & 39.89 & 60.41 & 49.27 & 54.25 \\
PAPO & 52.15 & 34.22 & 57.54 & 47.97 & 50.40 & 56.29 & \underline{41.85} & 59.54 & 52.56 & 53.68 & 47.50 & 37.99 & 55.61 & 47.04 & 49.96 \\
GDPO ($w=0.5$) & 53.60 & 30.06 & 60.53 & 48.07 & 52.46 & 56.21 & 40.35 & 63.66 & 53.41 & 55.84 & 46.69 & 39.44 & 56.59 & 47.57 & 53.28 \\
GDPO (no aux.) & 46.03 & 23.35 & 50.77 & 40.05 & 53.27 & 50.02 & 28.86 & 54.49 & 44.46 & 57.09 & 39.77 & 30.96 & 46.62 & 39.12 & 55.67 \\
GRPO (no std) & 49.56 & 30.51 & 58.61 & 46.23 & 50.66 & 53.87 & 37.48 & 66.55 & 52.63 & 56.09 & 46.97 & 38.84 & \underline{61.11} & 48.97 & 53.87 \\
CONCUR-O & 54.45 & \underline{35.27} & \textbf{69.35} & \textbf{53.02} & 52.77 & 60.65 & \textbf{42.50} & 74.10 & \underline{59.08} & 57.32 & 48.03 & 37.57 & \textbf{63.90} & \textbf{49.83} & 52.62 \\
\rowcolor{tablelightblue}CONCUR-T & \underline{55.33} & 33.98 & \underline{67.97} & \underline{52.43} & \underline{55.45} & \underline{63.76} & 41.77 & \underline{74.78} & \textbf{60.10} & \textbf{61.31} & 45.59 & 37.21 & 60.50 & 47.77 & \underline{56.39} \\
\rowcolor{tablelightblue}CONCUR-A & \textbf{56.05} & 32.36 & 66.64 & 51.68 & \textbf{55.78} & \textbf{63.92} & 37.75 & \textbf{75.49} & 59.05 & \underline{60.34} & 44.90 & 38.48 & 58.35 & 47.25 & \textbf{56.59} \\
\bottomrule
\end{tabular}%
}

\caption{Writing benchmarks and Avg under the second judge, with the names and formatting of \tab{tab:main}.}
\label{tab:second-judge}
\end{table*}
\subsection{Gated Auxiliaries under GDPO}\label{app:diagnosis}
\tab{tab:gate} lists the training statistics behind the three observations of the motivating experiment: the cosine of (O1), the shares of (O2), and the agreement of (O3).
The share at the ceiling grows on IF to nearly its value on Math and stays small on Code.
A training statistic of one step pools a task's groups in that step's batch: in a cosine or a share of \eqn{eq:shares}, each inner product and each squared norm is summed over the groups before the ratio is taken.
The share of passing responses at the ceiling is taken over the task's passing responses in the batch, and early and late values are means over 20 steps.
Among a task's groups that enter the update, the share with all responses at the ceiling is \val{4b.gdpo.solved.math.early} early and \val{4b.gdpo.solved.math.late} late on math, \val{4b.gdpo.solved.code.early} and \val{4b.gdpo.solved.code.late} on code, and \val{4b.gdpo.solved.if.early} and \val{4b.gdpo.solved.if.late} on IF, and it stays \val{4b.gdpo.solved.writing.late} on writing.

\textbf{An auxiliary independent of the primary.}
Let a group's primaries be distinct, and let the auxiliary's scores be independent normal draws that do not depend on the primary.
For a standard normal vector in $\R^G$, the expected squared norm of its projection onto $C_{\le}(p)$ is $H_G=\sum_{m=1}^{G}1/m$, of which the constant vectors account for $1$, and the vector's direction is independent of its norm.
The projection therefore keeps an expected share $(H_G-1)/(G-1)$ of the auxiliary's energy, and the remainder is the expected share against the order, which rounds to $0.75$ at the group size of $G=8$.

\subsection{Trust during Training}\label{app:trust-results}
A decision of earned trust changes a group's scores only if at least two of its responses are at the ceiling.
Over the last 20 training steps of the \concur-A run on \qwenfour{}, all responses are at the ceiling in a share \val{4b.concur.solved.if.late} of the IF groups and \val{4b.concur.solved.code.late} of the code groups that enter the update. In these groups, the auxiliary keeps GDPO's standardized scores at the steps where the pair's tie is lifted.

\subsection{Shares during Training}\label{app:mechanism}
Over the last 20 training steps of the \concur-A run on \qwenfour{}, the share against the order is zero on math by construction, \val{4b.concur.gate.code.against.late} on code, \val{4b.concur.gate.if.against.late} on IF, \val{4b.concur.gate.writing.against.late} for writing's length reward, and \val{4b.concur.gate.writing-rm.against.late} for its RM.
Over the same steps, the share at the ceiling is \val{4b.concur.gate.math.ceiling.late} on math, \val{4b.concur.gate.code.ceiling.late} on code, \val{4b.concur.gate.if.ceiling.late} on IF, and no more than \val{4b.concur.gate.writing.ceiling.late} for writing's two gated auxiliaries.

\textbf{Shares of the update.}
Over the last 20 training steps, math's share of the batch's summed squared advantages is \val{4b.gdpo.update-share.math.late}, \val{8b.gdpo.update-share.math.late}, and \val{olmo.gdpo.update-share.math.late} under GDPO and \val{4b.concur-tied.update-share.math.late}, \val{8b.concur-tied.update-share.math.late}, and \val{olmo.concur-tied.update-share.math.late} under \concur-T on \qwenfour{}, \qweneight{}, and \olmo{}, respectively.

\subsection{Variants, Ablation, and Reference}\label{app:ablation}
\tab{tab:ablation} gives every benchmark for GDPO, GDPO ($w=0.5$), the reference, GRPO (no std), \concur-O, and the two variants of \concur.
\fig{fig:aux-others} shows the auxiliaries' scores on \qweneight{} and \olmo{}, as \fig{fig:ablation}(d) does on \qwenfour{}.
In both figures, the \emph{base policy's level} of a gated pair is the auxiliary's mean score over the task's passing responses at the first step, averaged over the model's runs, and a run's change is its mean score over the last 20 training steps minus this level.
Under GDPO ($w=0.5$), which both figures leave out, the length reward's change on math is \val{4b.gdpo-w0.5.aux-change.math}, \val{8b.gdpo-w0.5.aux-change.math}, and \val{olmo.gdpo-w0.5.aux-change.math} on the three models in that order, compared with \val{4b.gdpo.aux-change.math}, \val{8b.gdpo.aux-change.math}, and \val{olmo.gdpo.aux-change.math} under GDPO.
As the bottom row of \fig{fig:alignment} does on \qwenfour{}, \fig{fig:primary} shows the primary's mean score over all responses on \qweneight{} and \olmo{} for the same runs.
On \olmo{}, where code's share at the ceiling is the largest, code's pass rate in training also ends higher under both variants than under GDPO and \concur-O.

\end{document}